UNIVERSITAT POLITÈCNICA DE CATALUNYA

**MASTER EN MÉTODOS NUMÉRICOS PARA CÁLCULO Y DISEÑO EN INGENIERÍA**

Population-Based Methods:

# PARTICLE SWARM OPTIMIZATION

– Development of a General-Purpose Optimizer and Applications –

PART I

Master's Thesis

Submitted by:
*Mauro Sebastián Innocente*

Supervised by:
Dr. *Johann Sienz*
Civil & Computational Engineering Centre
University of Wales Swansea

# SUMMARY

Optimization is a multi-disciplinary field concerning mathematicians, physicists, economists, biologists and engineers, among others. Some of the problems they have to face are inherently optimization problems (e.g. travelling salesman problem, scheduling, structural optimal design), while some others can be solved as if they were, by minimizing a conveniently defined error function (e.g. systems of equations, training of artificial neural networks).

Although all these problems involve optimizing some pre-defined criterion, they can be very different from one another. For instance, there are a finite number of solutions for discrete and combinatorial problems, whereas there are infinite solutions for continuous problems; the optimum's location, or its value, may be either static or dynamic; the problem may be single-objective or multi-objective, etc. This thesis is concerned with continuous, static, and single-objective optimization problems, subject to inequality constraints only. Nevertheless, some methods to handle other kinds of problems are briefly reviewed in **SECTION I**.

The "particle swarm optimization" paradigm was inspired by previous simulations of the cooperative behaviour observed in social beings. It is a bottom-up, randomly weighted, population-based method whose ability to optimize emerges from local, individual-to-individual interactions. As opposed to traditional methods, it can deal with different kinds of problems with few or no adaptation due to the fact that it does not optimize by profiting from problem-specific features of the problem at issue but by means of a parallel, cooperative exploration of the search-space carried out by a population of individuals.

The main goal of this thesis consists of developing an optimizer that can perform reasonably well on most problems. Hence, the influence of the settings of the algorithm's parameters on the behaviour of the system is studied, some general-purpose settings are sought, and some variations to the canonical version are proposed aiming to turn it into a more general-purpose optimizer. Since no termination condition is included in the canonical version, this thesis is also concerned with the design of some stopping criteria which allow the iterative search to be terminated if further significant improvement is unlikely, or if a certain number of time-steps are reached. In addition to that, some constraint-handling techniques are incorporated into the canonical algorithm in order to handle inequality constraints. Finally, the capabilities of the proposed general-purpose optimizers are illustrated by optimizing a few benchmark problems.

# ACKNOWLEDGEMENTS

I would like to thank my master's thesis' supervisor, Dr. Johann Sienz, for his invaluable guidance, advice, encouragement, facilities' supply, financial aid, and friendly support beyond the academic issues. I would also like to thank the "particle swarm optimization" researchers, James Kennedy, Maurice Clerc, Ioan Cristian Trelea and Russell Eberhart, for replying my e-mails and confirming my suspicions regarding the common misuse of the random weights in the literature, especially when writing the algorithm in vector notation.

Thanks to Dr. Stuart Bates for lending me his genetic algorithm software for the Latin Hyper-cube Sampling (LHS) to be used for the particles' initialization, although a thorough analysis of the use of this technique for this purpose could not be carried out due to time constraints.

I would like to thank Dr. Eugenio Oñate, from Universitat Politècnica de Catalunya (Spain), and Professor Ken Morgan, from University of Wales Swansea, for contacting me with my thesis' supervisor, Dr. Johann Sienz. I would also like to thank Lelia Zielonka, from CIMNE (International Centre for Numerical Methods in Engineering, Spain), for her permanent willingness to deal, and efficiency in dealing, with any kind of administrative issue.

I must express my immense gratitude to my parents and my three sisters for their endless, unconditional support and encouragement in every matter. None of my achievements would have been possible without them. I would also like to acknowledge the support of the remaining members of my "big family", especially that of my grandmother, Olga Baez.

I would like to thank the feedback received from the Applied Mechanics Department of the Universidad Nacional del Nordeste (Argentina) with regards to my questions related to numerical methods. In particular, I would like to express my gratitude to Pablo Alejandro Beneyto for his help and encouragement along my undergraduate and postgraduate studies.

Thanks to Amelia Cazorla for her friendly company and for keeping our shared flat running while I was attending the master's modules in Barcelona. My truthful gratitude to Héctor Ulises Levatti for our endless interesting discussions, for his friendly company, and for his permanent readiness to help me out when needed. I would also like to thank Omar Sued and Norma Nardi for their friendly company and support, and for their genuine willingness to help me in every matter, including sharing their flat with me in Barcelona.

Thanks to María Andrade, Lisa Roberts and Nora de la Quintana for their genuine, friendly company during my stay in Swansea. I would also like to thank Silvia Más Denia, Eva Mantiñán Fraga and Santiago Waessle for their friendship, for their kindness in sharing their flat with me when needed, and for making my stay in Swansea way more pleasant.

Thanks to Davide Deganello for his friendly company, for sharing his flat with me when needed, and for the numerous discussions with regards to the particle swarms and to the artificial neural networks—among other subjects—during so many coffee breaks.

I would like to thank Dr. Lluís Torres Llinàs, from Universitat de Girona, for his flexibility in my timetables during my work there, so that this thesis could be completed. I would also like to express my thanks to Marta Baena Muñoz for her help in keeping my flat running while finishing this thesis and for her friendly company during my stay in Girona.

Finally, making Jaques Riget's words my own [78], I would like to express my truthful gratitude to the "swarm" of friends in Argentina, Spain and Wales, who had something to do with this thesis one way or another.

# CONTENTS

## Chapter 1

## INTRODUCTION

# SECTION I

# BACKGROUND

## Chapter 2

## OPTIMIZATION

Chapter 4

EVOLUTIONARY ALGORITHMS

Chapter 5

## SWARM INTELLIGENCE

SECTION II

# ALGORITHM RESEARCH

Chapter 6

## PRELIMINARY ANALYSIS OF THE PARAMETERS

Chapter 7

# STOPPING CRITERIA

## Chapter 8
## FURTHER ANALYSIS OF THE PARAMETERS

## Chapter 9
## GENERAL-PURPOSE OPTIMIZERS

## Chapter 10

## CONSTRAINED PARTICLE SWARM OPTIMIZATION

# SECTION III

# APPLICATIONS

## Chapter 11

## FUNCTION OPTIMIZATION

Chapter 12

## TRAINING OF A SIMPLE ARTIFICIAL NEURAL NETWORK

Chapter 13

## ENGINEERING PROBLEMS

SECTION IV

# FINAL CONCLUSIONS

Chapter 14

## FINAL CONCLUSIONS

# SECTION V

# APPENDICES

## Appendix 1

## TRADITIONAL TRAINING OF ARTIFICIAL NEURONS

## Appendix 2

## BRIEF REVIEW OF GENETICS

## Appendix 3

## OPTIMIZERS' BENCHMARK TEST FUNCTIONS

# Appendix 4
## DIGITAL CONTENT

A4.1 Experiments carried out along section 6.4

A4.2 Experiments carried out along chapter 7

A4.3 Experiments carried out along chapter 8

A4.4 Experiments carried out along chapter 9

A4.5 Experiments carried out along chapter 10

A4.6 Experiments carried out along chapter 11

A4.7 Animations

A4.8 Images

A4.9 Matlab PSO codes

A4.10 Traditional Artificial Neuron's training (see Chapter 3).xls

# LIST OF FIGURES

# LIST OF TABLES

# NOMENCLATURE

| | | |
|---|---|---|
| $aw$ | : | Acceleration weight |
| ACO | : | Ant Colony Optimization / Ant Colony Optimizer |
| ACM | : | Adaptive Culture Model |
| ACS | : | Ant Colony System |
| AI | : | Artificial Intelligence |
| AL | : | Artificial Life |
| ALife | : | Artificial Life |
| ANN | : | Artificial Neural Network |
| AS | : | Ant System |
| $\mathcal{B}(\hat{\mathbf{x}}, \varepsilon)$ | : | Neighbourhood of $\hat{\mathbf{x}}$ defined by a hyper-sphere of radius $\varepsilon$ |
| b-PSO | : | Binary Particle Swarm Optimizer |
| B-PSO | : | Basic Particle Swarm Optimizer $\mathbf{C}_i^{(t)} = \mathbf{C}\left(\boldsymbol{\sigma}_i^{(t)}, \boldsymbol{\alpha}_i^{(t)}\right)$ |
| BSt-PSO | : | Basic, Standard Particle Swarm Optimizer |
| BStLd-PSO | : | Basic, Standard Particle Swarm Optimizer with Linearly time-decreasing inertia weight |
| BStSd-PSO | : | Basic, Standard Particle Swarm Optimizer with Sigmoidly time-decreasing inertia weight |
| BSw-PSO | : | Basic Particle Swarm Optimizer with time-Swapping learning weights |
| BSwLd-PSO | : | Basic Particle Swarm Optimizer with time-Swapping learning weights and Linearly time-decreasing inertia weight |
| BSwSd-PSO | : | Basic Particle Swarm Optimizer with time-Swapping learning weights and Sigmoidly time-decreasing inertia weight |
| $\mathbf{C}_i^{(t)}$ | : | Inverse of the covariance matrix, which information is fully contained by the vectors of standard deviation and rotation angles, i.e. $\mathbf{C}_i^{(t)} = \mathbf{C}\left(\boldsymbol{\sigma}_i^{(t)}, \boldsymbol{\alpha}_i^{(t)}\right)$, for individual $i$ at generation $t$ |
| C-PSO | : | Constricted Particle Swarm Optimizer |
| CA | : | Cellular Automaton |
| CAs | : | Cellular Automata |
| $\mathbf{cg}$ | : | Centre of gravity of the swarm |
| $cgbest$ | : | Minimum conflict found so far by any particle in the swarm |
| $cgworst$ | : | Maximum conflict found so far by any particle in a specialized sub-swarm |
| $cpbest_s$ | : | Minimum conflict found so far by particle $s$ |
| $c_s$ | : | Present conflict of particle $s$ |
| CPU | : | Central Processing Unit |
| DoL | : | Division of Labour |
| EA | : | Evolutionary Algorithm |
| EC | : | Evolutionary Computation |
| EP | : | Evolutionary Programming |
| ES | : | Evolution Strategy |
| $e^{(k)}$ | : | Absolute error regarding the conflict values at the $k^{\text{th}}$ time-step |
| $e_c^{(k)}$ | : | Absolute error regarding the design variables' values at the $k^{\text{th}}$ time-step |
| $re_i^{(k)}$ | : | Absolute error regarding the conflict values at the $k^{\text{th}}$ time-step, relative to the initial conflict value |
| $re_{ci}^{(k)}$ | : | Absolute error regarding the design variables' values at the $k^{\text{th}}$ time-step, relative to the initial Euclidean norm |
| $re^{(k)}$ | : | Absolute error regarding the conflict values at the $k^{\text{th}}$ time-step, relative to the current conflict value |
| $re_c^{(k)}$ | : | Absolute error regarding the design variables' values at the $k^{\text{th}}$ time-step, relative to the current Euclidean norm |
| $\mathcal{F}$ | : | Feasible part of the search-space |
| FSM | : | Finite State Machine |
| G-PSO | : | General Particle Swarm Optimizer |
| GA | : | Genetic Algorithm |
| $gbest_j^{(t-1)}$ | : | Coordinate $j$ of the best position found by any particle in the swarm up to time-step ($t$-1) |
| GP | : | Genetic Programming |
| GP-PSO | : | General-Purpose Particle Swarm Optimizer |
| IQ | : | Intelligence Quotient |
| $iw$ | : | Individuality weight |
| LHS | : | Latin Hypercube Sampling |
| MA | : | Memetic Algorithm |
| MLP | : | Multi-Layer Perceptron |
| $\mathcal{N}$ | : | Set of positive integer numbers |
| $N_{(0,1)}$ | : | Random number generated from a zero-mean Gaussian normal distribution with standard deviation equal to one, resampled anew each time it is referenced |
| $\overline{N}_{(0,1)}$ | : | Random number generated from a zero-mean Gaussian normal distribution with standard deviation equal to one, resampled anew for each individual and for each generation, but kept constant within each individual |
| $\mathbf{N}_{(0,\boldsymbol{\sigma})}$ | : | Vector whose components are random numbers generated from Gaussian zero-mean normal distributions with standard deviations equal to the corresponding components of the vector $\boldsymbol{\sigma}$, resampled anew each time it is referenced |
| NN | : | Neural Network |
| O-PSO | : | Original Particle Swarm Optimizer |
| $\mathbf{p}_s$ | : | Present position of particle $s$ |
| $\mathbf{pbest}_s$ | : | Best position found so far by particle $s$ |
| $pbest_{ij}^{(t-1)}$ | : | Coordinate $j$ of the best position found by particle $i$ up to time-step $(t-1)$ |
| PE | : | Processing Element |
| PSO | : | Particle Swarm Optimization / Particle Swarm Optimizer |
| PSO-G | : | Global Particle Swarm Optimizer |
| PSO-L | : | Local Particle Swarm Optimizer |
| $\mathcal{R}$ | : | Set of the real numbers |
| $\mathcal{R}^n$ | : | $n$-dimensional vectorial space, whose dimensions are the set of the real numbers |
| $\mathcal{S}$ | : | Search-space |
| S-ACO | : | Simple Ant Colony Optimization |
| SA | : | Simulated Annealing |
| SI | : | Swarm Intelligence |
| SM | : | Simplex Method |
| SO | : | Self-Organization |
| $sw$ | : | Sociality weight |
| TP | : | Transportation Problem |
| TS | : | Tabu Search |
| TSP | : | Travelling Salesman Problem |
| $U_{(0,1)}$ | : | Random number generated from a uniform distribution in the range $[0,1]$, resampled anew each time it is referenced |
| $\vec{\mathbf{v}}$ | : | Unitary basis vector |
| $v_{\max}$ | : | Maximum value that the components of the particles' velocity are allowed to take on |
| $w$ | : | Inertia weight |
| $x$ | : | Scalar |
| $\tilde{x}_{i,j}$ | : | Integer decoded from a bit-string |
| $\mathbf{x}$ | : | Vector |
| $\bar{\mathbf{x}}$ | : | Potential solution vector |
| $\breve{\mathbf{x}}$ | : | Local solution vector |
| $\hat{\mathbf{x}}$ | : | Global solution vector |
| $\mathbf{X}$ | : | Matrix |
| $Z$ | : | Set of integer numbers |
| $\boldsymbol{\alpha}_i^{(t)}$ | : | vector of rotation angles for individual $i$ at generation $t$ |
| $\boldsymbol{\sigma}_i^{(t)}$ | : | vector of standard deviations for individual $i$ at generation $t$ |
| $\chi$ | : | Constriction factor |

# Chapter 1

# INTRODUCTION

It is said that the stupid person does not learn from their own experiences, the intelligent person does learn from their own experiences, while the clever person learns from other people's experiences.

## 1.1 Introduction

Optimization is the process of seeking the combination of variables that leads to the best performance of a model, usually subject to a set of constraints. Thus, different combinations of values of the "variables" allow trying different candidate solutions, the "constraints" limit the valid combinations, and an "optimality criterion" allows telling better from worse. Traditional optimization methods exhibit serious drawbacks such as a number of requirements that either the objective function or the constraints must comply with for the method to be suitable, and their typical inability to escape local optima.

Evolution and learning comprise stochastic systems which rely on randomness to introduce creativity, since logically inferred responses do not innovate. Thus, evolution takes place by means of stochastic operators that alter the genetic information of the members of a population, while the individuals that better cope with the environment are more likely to survive and pass their genes to next generations. In turn, learning relies in making random decisions when there is not enough information to make strictly logical inferences, and then apprehending the results. Both evolution and learning are processes that organisms undergo to adapt to the environment so as to better cope with it. Since the adaptation is carried out by seeking the best responses, they can be viewed as optimization processes. Therefore, several optimization methods have been developed inspired by biological evolution and learning.

Evolution does not make sense for an isolated individual, except for the local exploration of new genetic compositions that can be made by the mutation of a single individual's genetic code. However, mutation seldom occurs within an individual's life-span, and the evolution's true power lies in the rare mutations and in the exchange of genetic information that occur during reproduction between individuals within a population. The aggregation of the genetic material of the different individuals is known as the "genetic pool".

In contrast, learning does make sense within a single individual, although there is a small range of situations that an individual can experience along its life-time. Clearly, it would be a remarkable step forward if this individual could have access to the knowledge gained by others. The experiences of numerous individuals along the current and past generations are stored in the so-called "culture". Thus, an individual can learn from its own experiences, from observing others' successful behaviours, and/or from culture. Notice that the culture can store experiences undergone by individuals that were never in contact with the individual at issue.

Evolutionary algorithms are population-based optimization methods which were inspired by evolution processes that natural organisms undergo to adapt to the environment, where new regions of the search-space are explored by altering the genetic code, which is composed of the object variables themselves or by a mapping from them. The so-called "fitness" associated to the new genetic codes are then evaluated, and a higher probability of survival is assigned to organisms that present themselves as fitter. Thus, the genetic materials which result in better coping with the environment are more likely to survive and to be passed to future generations.

The particle swarm optimization paradigm is a population-based method which was inspired by previous simulations of social processes. The main idea is that the achievements of a population of social beings overcome the sum of their individual achievements. The culture is spread—and updated—by means of individual-to-individual interactions, while individuals tend to seek agreement as they tend to imitate the most successful ones. Thus, cooperation among social beings overcomes competition. The typical example is that of a bird flock searching for food: they always find some without any prior knowledge about its location.

Thus, the evolutionary algorithms and the particle swarm optimization paradigm consist of decentralized and self-organized systems whose bottom-up ability to optimize emerges in a higher level than that of the individuals form local interactions among them. Perhaps the most characteristic feature of these methods is that they do not use the programmers' expertise to optimize the problem at issue. While this makes it difficult to understand how optimization actually occurs, these algorithms show astonishing robustness in dealing with many kinds of complex problems they were not specifically designed for. Hence they seem to be the right choice for the development of a general-purpose optimizer.

## 1.2 Motivation

Optimization is part of every-day life. Finding the shortest path to work or scheduling the activities of the day so as to minimize the time spent are typical examples. Likewise, energy companies try to minimize the energy losses along their networks so as to supply the service to a higher number of people and to maximize profits, structural optimization is concerned with the weight or cost minimization, the design of cars and aircrafts need to solve complex, multi-objective optimization problems to deliver their products, etc. In addition, almost any real-world problem that is not inherently an optimization problem can be turned into one by defining a measure of error which is to be minimized. Thus, problems such as solving systems of equations and training artificial neural networks can be viewed as optimization problems. Therefore, optimization problems arise in different areas such as mathematics, physics, chemistry, engineering, architecture, economics, management, biology, etc.

A waiter does not need to calculate the mechanics and dynamics of the problem to take a tray full of plates from the kitchen to the customers' tables. In fact, when the equilibrium is about to be broken, the appropriate system of forces is introduced to restore the equilibrium, without the need of an exhaustive, deterministic analysis. Likewise, the modern optimization methods such as the particle swarm optimization and the evolutionary algorithms do not use problem-specific details of the problem to solve it, as opposed to traditional methods. All that is needed is a function of the variables that allows differentiating better from worse. This feature makes them especially suitable for general-purpose algorithms, while it makes it possible to handle complex problems which traditional optimization methods are not able to deal with, and problems which no traditional method has been yet developed for. Among the modern methods, the population-based ones—which carry out a parallel search—appear to lead to more accurate solutions. In turn, the particle swarm optimization method is chosen over evolutionary algorithms because it is claimed to be computationally cheaper and remarkably easier to be programmed. Besides, it is also claimed to typically find more accurate solutions.

## 1.3 Objectives

The particle swarm optimization method is a relatively new method whose canonical version is suitable for unconstrained optimization problems only, and is not equipped with any error estimator that allows terminating the search when a certain degree of accuracy is achieved.

This thesis intends to introduce the particle swarm optimization method, to discuss its main strengths and weaknesses, to understand the influence of the settings of the parameters on the behaviour of the system, to find the settings that are appropriate for a general-purpose optimizer, and to incorporate some stopping criteria and some constraint-handling techniques. Finally, all these wrapped together is expected to lead to a general-purpose optimizer, which is to be applied to a number of benchmark problems in order to illustrate its capabilities.

## 1.4 Methodology

The influence of each parameter of the algorithm on the behaviour of the swarm is analyzed in terms of the evolution of the best solution found so far, of the average of the current candidate solutions, of its ability to and speed of clustering, and of its reluctance to getting trapped in sub-optimal solutions. Thus, some tunings taken from the literature are tried and some others are proposed, discussed, and tested on a suite of benchmark functions.

Some measures of the degree of clustering of the particles and of the evolution of the best and average solutions are developed, and the termination conditions are implemented by setting some thresholds for these measures below which the search is to be terminated. The evolution of these measures is studied by running the optimizer without termination conditions on a suite of benchmark functions. Then, the stopping criteria is developed by grouping some of these termination conditions for which the permitted values are set so as to comply with the permitted absolute errors found in the literature for the benchmark functions considered. The influence of the parameters' settings is studied again with the stopping criteria incorporated.

A few constraint-handling techniques are incorporated into some selected unconstrained optimizers, and their suitability is studied on the optimization of a suite of three constrained optimization problems. The application of the resulting optimizer to a number of benchmark problems serves the function of both testing the optimizer and illustrating its capabilities.

## 1.5 Lay out of the thesis

This thesis is divided into five sections:

**SECTION I** comprises five chapters devoted to an extensive review of the optimization and artificial intelligence fields. **Chapter 2** presents the main concepts behind optimization, where some traditional methods are reviewed, some are just mentioned, and the population-based ones are introduced. **Chapter 3** and **Chapter 4** consist of a review of artificial intelligence—where the stress is put on artificial neural networks—and of the evolutionary algorithms, respectively. Finally, **Chapter 5** presents the main concepts underlying swarm intelligence, and introduces the "ant colony optimization" and the "particle swarm optimization" methods.

**SECTION II** comprises five chapters dedicated to specific research on the particle swarm optimization paradigm. A preliminary analysis of the parameters of the basic algorithm is undertaken in **Chapter 6**, where every search is run along a fixed number of time-steps. The stopping criteria are developed along **Chapter 7**. Further analyses of the parameters' settings are carried out along **Chapter 8**, with the stopping criteria already incorporated. **Chapter 9** is devoted to the development of general-purpose unconstrained optimizers, profiting from the experimental results and derived conclusions obtained from chapters 6 to 8. **Chapter 10** presents a few constraint-handling techniques, which are incorporated into one selected general-purpose unconstrained optimizer taken from chapter 9.

**SECTION III** is composed of three chapters that illustrate a few possible applications of the general-purpose optimizers proposed by the end of chapter 10. **Chapter 11** is dedicated to illustrate the ability of four optimizers to deal with a set of six benchmark functions with hyper-cube-like boundary constraints for symmetric and asymmetric initializations, and also with another suite of five constrained benchmark functions and one non-convex unconstrained benchmark function. **Chapter 12** shows the learning of two simple artificial neural networks composed of three neurons each for the logical "xor" problem. Finally, the capabilities of two selected general-purpose optimizers are shown by optimizing three benchmark engineering problems along **Chapter 13**.

**SECTION IV**—which is composed of **Chapter 14** only—summarizes the achievements, concluding remarks and future research avenues proposed.

**SECTION V** contains one digital and three written appendices.

# SECTION I

# B A C K G R O U N D

# Chapter 2

# OPTIMIZATION

This chapter is a brief and necessarily incomplete review of the optimization field. Some of the basic concepts are presented. The dissertation is directed to problem solving, disregarding the other important stages of a complete optimization analysis. Some of the most popular methods are outlined. Only some general concepts regarding the population-based methods are discussed because they will be dealt with in detail from Chapter 4 on.

## 2.1 Introduction

Optimization is the process of seeking the best alternative according to a specific criterion, typically subject to a number of constraints. Thus, for problems in which the qualities of any answer can be quantified in a numerical value, optimization is the process of finding the permitted combination of variables in the problem that either minimizes or maximizes that value. Hence the **variables** allow trying different alternatives, the **constraints** limit the valid alternatives, and the **objective** allows differentiating better from worse.

Although optimization is ideally in quest for the best solution possible, this is often not the case in real-world problems, where successive improvement is already a great success. Besides, the concept of "possible" is remarkably subjective. Frequently, the optimization process is stopped because no more improvement is being achieved or simply because "time is up", despite not even knowing how good the best solution found so far is. Notice that the expression "the best solution" suggests that there is more than one solution and that every one possesses a different degree of "goodness" as opposed to root finding, for instance, where any solution is as good as any other.

Finding the shortest route to work or deciding which products and in which amounts to buy in different supermarkets in order to spend as little as possible are typical every day real-life problems that can be put under the label of "optimization problems" straightaway. Other problems that are inherently optimization problems could be for example finding the shape of a channel so that the energy losses are at the minimum keeping the area of the cross-section constant, finding the shortest route through a number of cities, finding the amounts of a product to ship from each of a number of service points to a number of destinations so that each destination receives a certain amount and the total cost is at the minimum, developing a product with the best quality at the lowest price, designing a plant so as to maximize the production at the lowest cost and without exceeding the pollution permitted, etc. However, almost any real-world problem that is not inherently an optimization problem can be turned into one. Think for instance of finding the value of "*x*" so that $2 \cdot x = 5$. A convenient measure of error such as $e(x) = (5 - 2 \cdot x)^2$ can be defined, so that its minimization would lead to the solution of the original problem. Thus, problems such as solving a system of equations, curve-fitting, and training artificial neural networks (see **Chapter 3**), can be treated as optimization problems. Since optimization problems are typically harder to solve than others, and almost any problem can be turned into an optimization problem, the techniques developed for them serve well for other purposes. *In general, any abstract task to be accomplished can be thought of as solving a problem, which, in turn, can be perceived as a search through a space of potential solutions. Since we are usually after the best solution, we can view this task as an optimization process* [20].

Hence optimization problems arise in different areas such as mathematics, physics, chemistry, engineering, architecture, economics, management, biology, etc. The behaviour of an existing system can be sometimes analyzed directly observing the system, in which case the situation is real and the results are reliable, but the number of situations analyzed is necessarily limited and the structure of the system is fixed. In order to allow the analysis of different alternatives, to allow the analysis when this risks the operating of the real system, or even when the real system is not tangible, the development of a model becomes necessary. The models can be physical (e.g. prototypes) or they can be symbolic (e.g. drawings or mathematical models). In any case, a model is merely an interpretation of a problem rather than the problem itself, which always requires a number of simplifications.

*All models are a simplification of the real world; otherwise they would be as complex and unwieldy as the natural setting itself. The process of problem solving consists of two separate general steps: (1) creating a model of the problem, and (2) using that model to generate a solution… If our model has a high degree of fidelity, we can have more confidence that our solution will be meaningful. In contrast, if the model has too many unfulfilled assumptions and rough approximations, the solution may be meaningless, or worse.* [54]

This work is concerned with the symbolic models, more precisely with the mathematical models. Thus, the plain words in which real problems are posed must be transformed into mathematical language in order to attempt to solve them by means of mathematical techniques. This is called the "formulation of the problem", which introduces a number of simplifications to the real problem.

The question is whether to make a lot of simplifications so that the available methods can solve the model, or develop a model as close as possible to the real problem and approximate the techniques used to solve it. Whatever approach is decided, the next step is to solve the proposed model. Traditional approaches require specific characteristics of the equations involved, so that the model is forced to assume the corresponding simplifications to suite the solving techniques. This approach results in an exact solution of an approximate model (e.g. the analytical solution of "structural analysis" problems by the "strength of materials" approach, linear programming, gradient descent optimization techniques, etc.). At present, the alternative of developing more precise models independent from the solution technique, and then attempting to solve them by means of a set of available toolboxes including modern heuristics and hybrid methods which do not severely limit the model are claimed to lead to better results. This approach results in an approximate solution of a model as exact as possible (e.g. the "finite element" approach of "structural analysis" problems, evolutionary algorithms, particle swarm optimization, etc.). The second approach usually outperforms the first one when dealing with real-world problems.

The accuracy of the model is vital to the success of the optimization process. The formulation of the problem must express precisely what is desired to be solved, for which the definition of a so-called "cost function" that successfully measures the idea of optimality is critical. The setting of the constraints in agreement with the real problem is also critical, since otherwise the solution might be meaningless.

The function to be optimized in a mathematical model is traditionally called "cost function", although other names such as "objective function", "evaluation function" and "fitness function" are frequent in the literature. In this thesis, the **objective function** is a function that relates the real problem to the model, while the **cost function** is the function that is to be optimized, which is not specifically related to the problem but to the optimization method. The problem variables are called from here on **object variables**.

This thesis is focused on the methods to solve the well-posed problem, assuming that its formulation has been successfully performed. In particular, the **particle swarm optimization** approach will be investigated further, from **Chapter 5** forth.

# 2.2 Mathematical optimization

The complete analysis of an optimization problem is performed in four broad stages:

1. Analysis and definition of the problem
2. Formulation of the problem (development of the model)
3. Solution of the model
4. Validation of the model

Only the solution of a well-posed mathematical model is dealt with here, while the other stages are well beyond the scope of this work. Thus, the problem of optimization is typically structured as a function of some object variables, often in the presence of some constraints.

## 2.2.1 Variety of optimization problems

It is important to remark at this point the tremendous differences between the different kinds of optimization problems, which allow being handled by completely different approaches.

*Perhaps the most significant difference lies in the fact that in some problems vectors describe solutions and optimal solutions, whereas in other cases functions are needed to formulate and solve the problem. This … results in a difference between optimization techniques for these two categories of problems. The situation is similar to the case of equations or systems of equations in which we are interested in a vector solution … and differential equations where the unknown is a function. In the first case, we talk about mathematical programming; in the second, about variational problems* [63].

Think about the so-called "transportation problem" (TP), which consists of a certain product that is to be shipped in amounts $o_i$ from $n$ service points to $m$ destinations, where it is to be received in amounts $d_j$. The cost of shipping a unit of product from every origin to every destination is also known, and the problem consists of finding the amount of product to be sent from each origin to each destination so that the total transportation cost is at minimum.

Think now about the problem of designing the profile of the cross-section of a channel, keeping its area fixed, so that the losses of energy are at minimum.

Clearly, the first case is an example of mathematical programming while the second is a variational problem. This work focuses on mathematical programming problems, from here forth referred to simply as “optimization”.

In turn, the optimization problems are typically differentiated taking into account the kind of models that can represent them. Therefore they are sometimes divided into “linear” and “nonlinear” problems. Other times they are classified as “linear”, “differentiable”, “convex”, “integer”, “mixed-integer”, “non-differentiable” [1], etc. This dissertation is more interested in dividing them into “continuous”, “combinatorial” and “binary” optimization problems, because the particle swarm optimization approach was originally developed for continuous search-spaces, no matter whether the problem is linear or nonlinear, differentiable or non-differentiable. Thus, further chapters will be concerned mainly with continuous optimization problems. In fact, only continuous optimization problems are dealt with from **Chapter 6** on.

#### 2.2.1.1 Continuous optimization

A problem of continuous optimization is a problem whose object variables can take real number values. Note that this means that the search-space is continuous, without this implying that the feasible space or the function to be optimized need to be continuous. Of course, since there are infinite real numbers, and even within a certain range, infinite real numbers are contained within it, optimizing by force brute, i.e. by trying every possible solution and keeping the best, is out of question. Hence the problem has no solution when the landscape that represents it is random. However, there is almost always a function that models the landscape, which does not need to be continuous, by means of which information about the systematic relationships between the points in the landscape can be obtained.

The measure to indicate the distance between two different solutions is typically, but not exclusively, the Euclidean norm. Thus, in high dimensional problems, the information about the distance between points and the differences between their cost function values give useful information to guide the search. It seems reasonable to think that if moving in a certain direction improved the fitness in the last step, it would do the same again if moving in the same direction. Moreover, if an algorithm could find a reasonably good solution somewhere, it is very likely that a better solution can be found nearby.

[1] Note that the membership to a group does not exclude the membership to another. For instance, a linear problem is always differentiable.

### 2.2.1.2 Combinatorial optimization

Combinatorial optimization is a completely different matter. Here the aim is at finding the optimal arrangement of the elements so as to optimize a result. Therefore, combinatorial optimization has a finite number of possible solutions although this number becomes typically intractable for real-world problems. One of the most typical combinatorial optimization problems is the travelling salesman problem (TSP), in which the shortest path through a number of cities has to be found so that each city is visited only once, and the travel ends at the same city where it started. Since it is a loop, the starting point does not really matter, so that the lists of cities A, B, C and C, B, A are exactly the same. For a problem consisting of only 50 cities, $\frac{49!}{2}$ trials would be necessary to test all possible solutions. Since this is clearly not feasible, the search must be guided by optimally allocating the trials without trying all possible solutions. There are numberless variations of the TSP such as the cases where some cities cannot be connected to each other. The constraints eliminate some solutions thus making the set of possible solutions smaller, but they also introduce a higher degree of complexity into the problem, since the feasibility of every trial has to be verified.

Traditional methods to deal with combinatorial problems are sequential. For instance, one can start randomly generating a tour for the TSP, and calculating the total length. Then, successive tours are generated step by step, calculating the accumulative distances, so that when the distance is longer than the shortest found so far, all the possible tours from that point on are eliminated. Hence entire areas of the search-space will not be searched.

However, it has been realized that nature does not optimize in a sequential fashion but in a parallel manner. The new methods, which are based on natural metaphors, attempt to mimic that behaviour. Even the artificial intelligence field has been greatly influenced by this view: the “artificial life” paradigm as a route to artificial intelligence is an active research field at present. This is discussed in some detail in **Chapter 3**.

There are many traditional ways of dealing with these kinds of problems that will not be discussed in details within this thesis, which is not particularly focused on combinatorial problems. Nevertheless, a few traditional methods are briefly outlined later in this chapter, and some population-based methods suitable for combinatorial problems such as the “genetic algorithms” and the “ant colony optimization” method will be discussed in some detail in **Chapter 4** and **Chapter 5**, respectively. In fact, a way of handling a TSP by means of the “evolutionary programming” method is briefly explained in section **4.3.3.3.1**. Beware that evolution in nature is a kind of combinatorial problem, consisting of reordering the genetic material to better cope with the environmental challenges, although new genetic material is infrequently introduced by means of mutations.

#### 2.2.1.3 Binary optimization

Binary optimization problems are those whose object variables can take one of two values such us true-false, on-off, negative-positive, cold-hot, etc. Hence they are typically decision problems although the approach is very versatile, since almost anything can be represented to any degree of precision by a binary alphabet. Thus, some continuous and combinatorial problems can also be handled by binary algorithms. Since given two binary values only a discrete number of values are in between, the binary optimization problems are discrete. Hence there are a finite number of possible solutions, although this number is typically huge.

The binary search-space consists of an $n$-dimensional hyper-cube, whose vertices are all the possible solutions to the problem at hand. Therefore, there are $2^n$ points in the discrete binary hyper-space to be searched, one of which is the global best sought. Notice that for $n$ = 1, the hyper-space is a unitary segment, for $n$ = 2 it is a square of unitary edges, for $n$ = 3 it is a cube of unitary edges, whereas for higher dimensions it is a unitary hyper-cube.

A very simple example of how to deal with a continuous optimization problem by means of a binary genetic algorithm is shown in section **4.3.4.1**.

### 2.2.2 Optimization problem

Let $\mathcal{S}$ be the search-space and $\mathcal{F} \subseteq \mathcal{S}$ its feasible part. The latter consists of the aggregation of all the valid solutions to the problem (i.e. all those that do not violate any constraint).

An optimization problem consists of finding the vector $\hat{\mathbf{x}} \in \mathcal{F}$ such that $f(\hat{\mathbf{x}}) \le f(\mathbf{x})$ for minimization problems, or $f(\hat{\mathbf{x}}) \ge f(\mathbf{x})$ for maximization ones. Without loss of generality, optimization will mean minimization from here on, since if $f(\hat{\mathbf{x}}) \le f(\mathbf{x}) \Rightarrow -f(\hat{\mathbf{x}}) \ge -f(\mathbf{x})$. In other words, $\min\{f(\mathbf{x})\} = -\max\{-f(\mathbf{x})\}$. Therefore, an optimization problem can be formulated as:

$$\begin{aligned}&\text{Minimize } f(\mathbf{x})\\&\text{subject to } \mathbf{x} \in \mathcal{F}\end{aligned} \tag{2. 1}$$

More precisely, the problem can be rewritten as shown in equation **(2. 2)**.

$$\begin{aligned}&\text{Minimize } f(\mathbf{x})\\&\text{subject to } \begin{cases} g_j(\mathbf{x}) \ge 0 & ; \quad j = 1, \dots, q \\ g_j(\mathbf{x}) = 0 & ; \quad j = q+1, \dots, m \end{cases}\end{aligned} \tag{2. 2}$$

Where:

- $\mathbf{x} \in \mathcal{S}$ is the vector of object variables
- $f(\cdot) : \mathcal{S} \to \mathcal{E}$ is the function to be optimized
- $g_j(\cdot)$ are the constraint functions
- $\mathcal{S} \subseteq \mathcal{R}^n \;\wedge\; \mathcal{E} \subseteq \mathcal{R}$
- $\mathcal{R}$ is the set of real numbers

The goal is to find the vector $\hat{\mathbf{x}} \in \mathcal{F}$ such that:

$$\forall \mathbf{x} \in \mathcal{F} : f(\mathbf{x}) \ge f(\hat{\mathbf{x}}) \tag{2. 3}$$

Where $f(\hat{\mathbf{x}})$ is a global minimum and $\hat{\mathbf{x}}$ is its location.

For a general constrained optimization problem, $\mathcal{F}$ is typically defined as:

$$\mathcal{F} = \{\mathbf{x} \in \mathcal{S} \mid g_j(\mathbf{x}) \ge 0 \;\; \forall j \in \{1, \dots, q\} \;\wedge\; g_j(\mathbf{x}) = 0 \;\; \forall j \in \{q+1, \dots, m\}\} \tag{2. 4}$$

Equations **(2. 2)** to **(2. 4)** define minimization problems with "greater than or equal to" constraints. This does not lose generality since for problems with "less than or equal to" constraints, the function $k_j(\mathbf{x}) = -g_j(\mathbf{x})$ can be defined so that if $g_j(\mathbf{x}) \ge 0 \;\Rightarrow\; k_j(\mathbf{x}) \le 0$.

Notice that in equation **(2. 2)** each equality constraint could be replaced by two inequality constraints: "greater than or equal to" plus "less than or equal to".

### 2.2.3 The objective and the cost functions

The **objective** of an optimization problem defines the goal that is sought, so that it is given in plain words, being independent from the mathematical model designed to solve it.

Some kinds of so-called multi-objective optimization problems seek to achieve more than one objective which are typically conflicting, such as the goal of producing the "best" and "the cheapest" product at the same time. However, in order to apply optimization techniques, a function that returns a single value is required so that the qualities of the solutions can be compared. Thus, one of the alternatives consists of building up a function that includes all the objectives, which are weighted according to some criteria. Another option consists of defining a threshold allowed for each objective except one, which is then optimized. More complex methods developed to deal with multi-objective optimization problems are well beyond the scope of this thesis, which will only deal with mono-objective optimization problems.

#### 2.2.3.1 The objective function

While the objectives of the optimization problem are given in plain words, its mathematical formulation is called the **objective function**. Therefore the objective function maps the search-space to the space of the objective(s) of the optimization problem, so that it acts as a link between the real problem and the model of it that is to be optimized. Notice that the objective function returns a vector for multi-objective optimization problems and a scalar for mono-objective optimization problems.

#### 2.2.3.2 The cost function

The **cost function** is the actual function to be minimized. It maps the output of the objective function to the real numbers, thus giving a way to decide which solutions are better than others so as to guide the search. The name is due to the fact that the goal of the first optimization problems was at minimizing costs. In evolutionary algorithms, it is known as the **fitness function** instead due to the natural evolution metaphor under which they are based, which seeks the maximization of the organisms' fitness. Since the particle swarm optimizers are based under the metaphor of human social behaviour (see sections **5.4** and **5.6**), the function to be minimized by them will be referred to as the **conflict function**.

The cost function and the objective function can coincide in some cases, they can be proportional, or there can be an arbitrarily complex mapping between them, according to the problem at hand. As previously mentioned, it is assumed in this thesis that the cost function has already been constructed in such a way that the optimal solution serves the objective(s) well. The cost function plot is sometimes called landscape or topography because of its appearance in two-dimensional search-spaces. Thus, decreasing the cost function is equivalent to moving downhill in the landscape, which is plotted by assigning one dimension to each variable and one dimension for the cost values.

## 2.2.4 Constraints

Although the constraints allow concentrating the search into limited areas thus reducing the size of the search-space, each potential solution must be verified not to violate the constraints. Hence the problem usually becomes harder to be solved than its unconstrained counterpart.

There are many different kinds of constraints. Thus, the restrictions to the values that the object variables can take are commonly posed mathematically by equality and inequality constraint functions as shown in equation **(2. 2)**. Other constraints can restrict the values that the objective function can take, or they can restrict some variables when some others take certain values. Any arbitrarily complex constraint can be in principle posed according to the design of the model that represents the problem. Thus, not only does the algorithm have to minimize the cost function, but it also has to comply with all the constraints of the problem.

The majority of the optimization techniques for continuous optimization problems require the differentiability of the cost and of the constraint functions. However, for modern heuristic techniques like population-based methods, this restriction often does not apply.

## 2.2.5 Global and local optimization

If the search is limited to an area around a certain point of the search-space, it is said that this area is the neighbourhood of that point. Depending on the characteristics of the search-space, a neighbourhood can be defined differently. For instance, in binary search-spaces it is said that the neighbourhood of a certain point is comprised by all the points that are separated from the point at issue by a hamming-distance[2] equal to one. For continuous optimization problems whose search-space is the vectorial space $\mathcal{R}^n$, the distance between two points is typically defined as the Euclidean norm. Thus, its neighbourhood can be described as all the points within a hyper-sphere whose centre is the point at issue, as shown in equation **(2. 5)**.

$$\mathcal{B}(\breve{\mathbf{x}}, \varepsilon) = \left\{\mathbf{x} \in \mathcal{R}^n : \left\|\mathbf{x} - \breve{\mathbf{x}}\right\| \leq \varepsilon\right\} \quad \text{where } \varepsilon \in \mathcal{R} \wedge \varepsilon \neq 0 \qquad \textbf{(2. 5)}$$

Usually, the optimization task is complicated by the existence of nonlinear cost functions with multiple local minima. A local minimum is the minimum within a neighbourhood included in the feasible search-space. Thus, it is said that $\breve{\mathbf{x}}$ is the location of a local minimum if:

$$\exists\, \varepsilon \in \mathcal{R} > 0 \mid f(\breve{\mathbf{x}}) \leq f(\mathbf{x}) \;\; \forall \mathbf{x} \in \mathcal{B}(\breve{\mathbf{x}}, \varepsilon) \subseteq \mathcal{F} \qquad \textbf{(2. 6)}$$

If $\mathcal{B} = \mathcal{F} \;\; \Rightarrow \;\; \breve{\mathbf{x}} = \hat{\mathbf{x}}$ is the location of the global minimum.

# 2.3 Optimization methods

It is important to remark at this point the tremendous differences between the different kinds of optimization problems, which must be dealt with by completely different approaches.

The great variety of optimization problems has been briefly described, and it has been claimed that their tremendous differences oblige to deal with each kind by completely different approaches. Thus, the optimization methods could be classified according to the kind of optimization problems that they are capable of handling, so that they could be divided into "linear", "non-linear", "differentiable", "convex", "integer", "mix-integer" and "non-differentiable" optimization methods. Other classifications take into account the features of the algorithms rather than the problems they can handle, so that they could be classified as being "gradient-based" or "gradient-free", "exact" or "approximate", "deterministic" or "probabilistic", "analytical" or "heuristic", "single-based" or "population-based", etc. There can be found numerous different and often conflicting classifications, where every method can belong simultaneously to more than one category. In addition to that, words like

[2] The hamming-distance between two points in a binary space is the number of bits that need to be flipped to move from one to the other.

"heuristic" do not have a precise definition when used in this context, and some derived words like "meta-heuristic" make a precise classification even more difficult.

Gallagher [37] suggests that *heuristic (or approximate) algorithms aim to find a good solution to a problem in a reasonable amount of computation time – but with no guarantee of "goodness" or "efficiency"… Two broad classes of heuristics are "Constructive methods" and "Local search methods"*.

The term derives from the Greek "heuriskein" that means to find or discover. However, despite the etymology of the work, its use is more frequently related to techniques that do not guarantee to find anything. In fact, they usually consist of methods based on common sense, natural metaphors or even more general, methods whose behaviour is not fully and deterministically understood.

Reeves et al. [65] offer: *A heuristic is a technique which seeks good (i.e. near-optimal) solutions at a reasonable computational cost without being able to guarantee either feasibility or optimality, or even in many cases to state how close to optimality a particular feasible solution is*.

Instead, a meta-heuristic is claimed to be *a top-level general strategy which guides other heuristics to search for feasible solutions in domains where the task is hard. …Examples of meta-heuristics are tabu search, simulated annealing, genetic algorithms and memetic algorithms* [23].

Gallagher [37], in turn, suggests that *meta-heuristics are (roughly) high-level strategies that combine lower-level techniques for exploration and exploitation of the search space.*

Instead, Glover (from [10]) defines a meta-heuristic as *a master strategy that guides and modifies other heuristics to produce solutions beyond those that are normally generated in a quest for local optimality*, whereas Batista [10] claims that meta-heuristics, *which are designed to search for global optima, provide a means for approximately solving complex optimization problems. However, they cannot guarantee that the best solution found after termination criteria are satisfied is indeed a global optimal solution to the problem…*

The membership of some paradigms to the group of the heuristics is sometimes doubtful. In addition to that, the bounds between heuristics and meta-heuristics are even fuzzier because any small subroutine within a heuristic method might be claimed to turn it into a meta-heuristic one. To avoid at least this last problem, both heuristics and meta-heuristics are included in this work within the "modern heuristic techniques". Note, however, that if a heuristic classifies as a traditional method, the latter prevails in the following classification.

**OPTIMIZATION METHODS**

**TRADITIONAL METHODS**
**Deterministic**
(Based on precise mathematical theories)

**Analytical**

- Simplex method
- Local search:
*Hill-climber*
*Gradient-based methods*
- Tabu Search

**Construct solutions**

- Divide and conquer
- Branch and bound

**MODERN HEURISTIC TECHNIQUES**
**Probabilistic and approximate**
(Make use of techniques that are not directly related to optimization processes)

**Single solution**

- Random search
- Random walk
- Greedy search
- Stochastic hill-climber
- Simulated annealing

**Population-based**

- Evolutionary Algorithms:
*Evolution Strategies*
*Evolutionary Programming*
*Genetic Algorithms*
*Genetic Programming*
- Swarm-intelligence-based:
Ant Colony Optimization
Particle Swarm Optimization

**Fig. 2. 1**: Classification of the different optimization methods.

Although "tabu search" makes use of techniques that are not directly related to optimization processes such as keeping a list of temporarily banned locations, it is also a deterministic method derived from the local search, hence it has been included within the "traditional methods". The general criterion was to classify all purely deterministic methods as traditional, and the probabilistic methods as modern heuristics. However, this classification intends by no means to be rigid, since there can certainly be many valid arguments that could modify it. Nevertheless, it seems especially appropriate for this thesis, which is particularly concerned with the population-based methods. Some selected methods are outlined in the rest of the chapter, while only a basic introduction to population-based methods is included because they will be discussed in some details in later chapters.

## 2.4 Traditional methods

While the boundaries that separate traditional methods from heuristics are fuzzy, traditional methods are typically characterized as being deterministic, in spite of the fact that they do not necessarily find exact solutions. Numberless traditional methods have been developed to guide the search through the problem space in quest for the optimal solution. Although these methods have proven through years their capabilities of finding either exact solutions or good approximations, they have at least three main disadvantages:

1. They typically move form one point to another in the search-space by using some deterministic rule, what makes them likely to get trapped into local optima. Some methods were developed to avoid this weakness, but then they end up not being very robust or simply being just computationally too expensive to deal with real-world problems.

2. They require mathematically well-defined problems to be operable, whilst real-world problems do not usually lend themselves to that. In addition, not only do they need to evaluate the function to be optimized but they also require additional information such as gradients, which limits the kinds of functions they can handle (they must be at least continuous).

3. They are designed to solve a certain kind of problem, what leads to a great number of problem-targeted algorithms. When a new non-standard problem arises, a new algorithm must be designed to deal with it, provided the problem is mathematically well-defined.

Some well known traditional methods are briefly discussed hereafter with the goal of making what population-based methods can offer clearer in future chapters.

## 2.4.1 Linear programming: The simplex method

The simplex method (SM) is a method that can only handle optimization problems whose cost and constraint functions are linear. A possible formulation is shown in equation **(2. 7)**.

$$\begin{aligned} &\text{Minimize} \quad f(\mathbf{x}) = \mathbf{c}\cdot\mathbf{x} \\ &\text{subject to} \quad \begin{cases} \mathbf{A}\cdot\mathbf{x} \geq \mathbf{a} \\ \mathbf{B}\cdot\mathbf{x} = \mathbf{b} \\ \mathbf{D}\cdot\mathbf{x} \leq \mathbf{d} \end{cases} \end{aligned} \tag{2. 7}$$

Where the vectors **a**, **b**, **c, d** and the matrices **A**, **B** and **D** are data of the problem.

Notice that the constraints are lines in 2-dimensional, planes in 3-dimensional, and hyper-planes in higher-dimensional search-spaces. Thus, a finite feasible region has the form of a polygon, a polyhedron, or a hyper-polyhedron, respectively. Since the cost function is also linear, it can be represented by straight contour lines for 2-dimensional problems (**Fig. 2. 2**).

The SM requires that the problem formulated in equation **(2. 7)** is rewritten in its standard form, as shown in equation **(2. 8)**. A “greater than or equal to” constraint can be turned into a “less than or equal to” constraint by multiplying both sides of the inequality by -1. In order to transform the inequality into equality constraints, a variable is added to each inequality equation, so that a constraint like $\sum_i a_{ij} \cdot x_i \le a_j$ is transformed into $\sum_i a_{ij} \cdot x_i + y_j = a_j$, where $y_j \ge 0$. To ensure that all the variables are greater than or equal to zero, each variable that do not have that restriction in the original problem is replaced by two other variables such that $x_i = z'_i - z''_i$, where $z'_i \ge 0 \;\wedge\; z''_i \ge 0$. Therefore, the formulation **(2. 7)** can be rewritten as:

$$\text{Minimize} \quad f(\mathbf{x}) = \mathbf{c} \cdot \mathbf{x}$$
$$\text{subject to} \begin{cases} \mathbf{A} \cdot \mathbf{x} = \mathbf{a} \\ \mathbf{x} \ge \mathbf{0} \end{cases} \tag{2. 8}$$

Note that the vectors **c**, **x**, **a**, and the matrix **A** in **(2. 8)** are not the same as in **(2. 7)**.

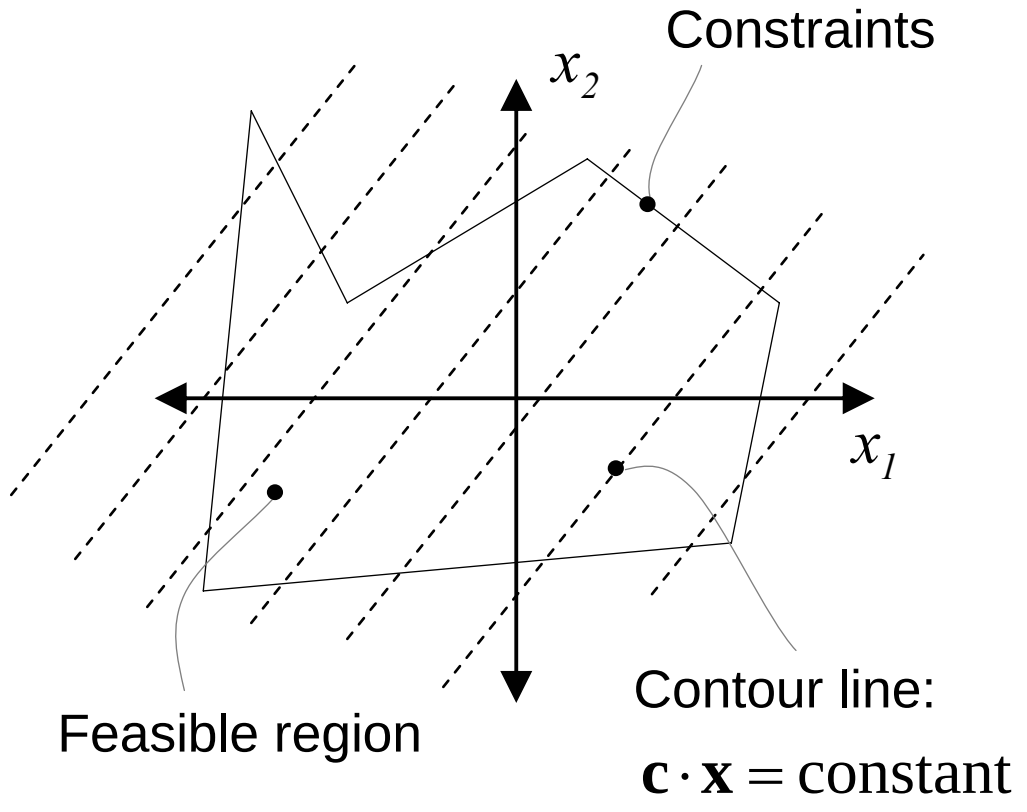

**Fig. 2. 2**: Example of a feasible region and level lines in a 2-dimensional linear problem. Each side of the polygon that represents the feasible region is given by a linear constraint, each one of which divides the plane in one feasible and one infeasible semi-planes. The intersection of the six feasible semi-planes gives shape to the feasible polygon. The dotted lines represent the traces of the hyper-plane $\mathbf{c} \cdot \mathbf{x}$ on planes parallel to $(x_1, x_2)$, placed at different levels.

If the linear system $\mathbf{A} \cdot \mathbf{x} = \mathbf{a}$ is such that for the *m* x *n* matrix **A** it happens that *m* > *n*, then *m* – *n* equations of the system need to be eliminated in order to obtain an equivalent square *n* x *n* system. Hence the solution of the problem, if any, is the solution of this linear system.

Suppose the system is such that there are more variables than equations (i.e. *n* > *m*). Any square matrix of size and rank equal to *m* that can be generated with the columns of **A** is called a “basic matrix” $(\mathbf{A_B})$, and the matrix containing the remaining columns is called a “non-basic matrix” $(\mathbf{A_N})$. The *m* variables corresponding to the columns of $\mathbf{A_B}$ are called “basic variables”, and the ones corresponding to the columns of $\mathbf{A_N}$ are “non-basic variables”.

Thus, the system of equations that represents the constraints can be rewritten as follows:

$$\mathbf{A} \cdot \mathbf{x} = \begin{pmatrix} \mathbf{A_B} & \mathbf{A_N} \end{pmatrix} \cdot \begin{Bmatrix} \mathbf{x_B} \\ \mathbf{x_N} \end{Bmatrix} = \mathbf{a} \tag{2. 9}$$

If $\mathbf{x_N} = \mathbf{0}$, the system $\mathbf{A_B} \cdot \mathbf{x_B} = \mathbf{a}$ has a unique solution $\mathbf{\bar{x}_B} = \mathbf{A_B^{-1}} \cdot \mathbf{a}$. If all its components are "greater than or equal to" zero, the vector $\begin{Bmatrix} \mathbf{\bar{x}_B} \\ \mathbf{0} \end{Bmatrix}$ is a so-called "basic solution".

The cost function corresponding to a "basic solution" is shown in equation **(2. 10)**.

$$\bar{f} = \begin{pmatrix} \mathbf{c_B} & \mathbf{c_N} \end{pmatrix} \cdot \begin{Bmatrix} \mathbf{\bar{x}_B} \\ \mathbf{0} \end{Bmatrix} = \mathbf{c_B} \cdot \mathbf{\bar{x}_B} \qquad \textbf{(2. 10)}$$

The basic solutions are the hyper-vertices of the feasible region (see **Fig. 2. 2**). It is obvious that, since the cost function is linear, the optimal solution must be on the boundaries of the feasible region. Besides, since all those boundaries are linear, the minimum value of the cost function on each side of the hyper-polyhedron will be at one end, while its maximum value will be at the other. Hence the optimum will be located in one of the vertices, thus turning the problem into a discrete one. Therefore the optimum is among the basic solutions!

The "fundamental theorem of the linear programming", whose proof is beyond the scope of this dissertation, asserts that, given a problem formulated as shown in **(2. 8)**:

1. If the problem admits at least one solution, then it admits at least one "basic solution".
2. If the problem admits at least one optimal solution, then there is also at least one optimal solution that is a basic solution.

Notice that for low-dimensional problems all the basic solutions can be evaluated (i.e. all the vertices of the feasible region) so that the one with the minimal cost is the solution to the problem. However, the problem becomes intractable by brute force (i.e. by trying all possible solutions) for a moderately high number of variables. The SM aims to organize the operations so that there is no need to go through all the vertices to find the optimal solution.

The method starts at any particular basic solution and then jumps to an adjacent vertex at least as good as the present vertex. Beware that the values of the original variables place the solutions in the search-space, whereas the values of the auxiliary variables introduced to turn inequality into equality constraints give information about the distance from the point at issue to the corresponding boundary, since the points within the feasible region have these auxiliary components positive. Setting any variable to zero implies searching a border, so that setting several variables to zero searches the intersection of all the corresponding borders. Thus, let $\bar{\mathbf{x}} = \begin{Bmatrix} \bar{\mathbf{x}}_{\mathbf{B}} \\ \mathbf{0} \end{Bmatrix}$, where $\bar{\mathbf{x}}_{\mathbf{B}} \in \mathcal{R}^{m} \wedge \mathbf{0} \in \mathcal{R}^{n-m}$ be a first feasible extremal point (basic solution). The basic iterative step consists of setting one component of $\bar{\mathbf{x}}_{\mathbf{B}}$ to zero (leaving variable) and letting one component of $\mathbf{0}$ become positive (entering variable), provided the cost is lowered and the solution is feasible.

Let $\mathbf{A} \cdot \bar{\mathbf{x}} = \left( \mathbf{A}_{\mathbf{B}} \quad \mathbf{A}_{\mathbf{N}} \right) \cdot \begin{Bmatrix} \bar{\mathbf{x}}_{\mathbf{B}} \\ \mathbf{0} \end{Bmatrix} = \mathbf{a} \quad \Rightarrow \quad \bar{\mathbf{x}}_{\mathbf{B}} = \mathbf{A}_{\mathbf{B}}^{-1} \cdot \mathbf{a}$ **(2. 11)**

Suppose $\bar{\mathbf{x}}_{\mathbf{B}} \geq \mathbf{0}$, so that it is a basic solution. Then, the cost for this basic solution is:

$$\bar{f} = \mathbf{c} \cdot \bar{\mathbf{x}} = \left( \mathbf{c}_{\mathbf{B}} \quad \mathbf{c}_{\mathbf{N}} \right) \cdot \begin{pmatrix} \bar{\mathbf{x}}_{\mathbf{B}} \\ \mathbf{0} \end{pmatrix} = \mathbf{c}_{\mathbf{B}} \cdot \bar{\mathbf{x}}_{\mathbf{B}} = \mathbf{c}_{\mathbf{B}} \cdot \mathbf{A}_{\mathbf{B}}^{-1} \cdot \mathbf{a} \qquad \textbf{(2. 12)}$$

The move from $\bar{\mathbf{x}} = \begin{Bmatrix} \bar{\mathbf{x}}_{\mathbf{B}} \\ \mathbf{0} \end{Bmatrix}$ to $\mathbf{x} = \begin{Bmatrix} \mathbf{x}_{\mathbf{B}} \\ \mathbf{x}_{\mathbf{N}} \end{Bmatrix}$, where $\mathbf{x}_{\mathbf{N}}$ is at the user's disposal, will only take place if the three following conditions are fulfilled:

1. The constraints are satisfied: $\mathbf{A} \cdot \mathbf{x} = \left( \mathbf{A}_{\mathbf{B}} \quad \mathbf{A}_{\mathbf{N}} \right) \cdot \begin{Bmatrix} \mathbf{x}_{\mathbf{B}} \\ \mathbf{x}_{\mathbf{N}} \end{Bmatrix} = \mathbf{a}$
2. The cost is lowered: $f = \mathbf{c} \cdot \mathbf{x} < \bar{f} = \bar{\mathbf{c}} \cdot \bar{\mathbf{x}}$
3. All the components of the new vector are non-negative: $\mathbf{x} \geq \mathbf{0}$

The first condition $\mathbf{A}_{\mathbf{B}} \cdot \mathbf{x}_{\mathbf{B}} + \mathbf{A}_{\mathbf{N}} \cdot \mathbf{x}_{\mathbf{N}} = \mathbf{a}$ forces:

$\mathbf{x}_{\mathbf{B}} = \mathbf{A}_{\mathbf{B}}^{-1} \cdot \left( \mathbf{a} - \mathbf{A}_{\mathbf{N}} \cdot \mathbf{x}_{\mathbf{N}} \right) \quad \Rightarrow \quad \mathbf{x}_{\mathbf{B}} = \bar{\mathbf{x}}_{\mathbf{B}} - \mathbf{A}_{\mathbf{B}}^{-1} \cdot \mathbf{A}_{\mathbf{N}} \cdot \mathbf{x}_{\mathbf{N}}$, so that the new cost is: **(2. 13)**

$$f = \mathbf{c} \cdot \mathbf{x} = \left( \mathbf{c}_{\mathbf{B}} \quad \mathbf{c}_{\mathbf{N}} \right) \cdot \begin{pmatrix} \mathbf{x}_{\mathbf{B}} \\ \mathbf{x}_{\mathbf{N}} \end{pmatrix} = \mathbf{c}_{\mathbf{B}} \cdot \mathbf{x}_{\mathbf{B}} + \mathbf{c}_{\mathbf{N}} \cdot \mathbf{x}_{\mathbf{N}} = \mathbf{c}_{\mathbf{B}} \cdot \left( \bar{\mathbf{x}}_{\mathbf{B}} - \mathbf{A}_{\mathbf{B}}^{-1} \cdot \mathbf{A}_{\mathbf{N}} \cdot \mathbf{x}_{\mathbf{N}} \right) + \mathbf{c}_{\mathbf{N}} \cdot \mathbf{x}_{\mathbf{N}} \qquad \textbf{(2. 14)}$$

Therefore, $f = \mathbf{c}_{\mathbf{B}} \cdot \bar{\mathbf{x}}_{\mathbf{B}} + \left( \mathbf{c}_{\mathbf{N}} - \mathbf{c}_{\mathbf{B}} \cdot \mathbf{A}_{\mathbf{B}}^{-1} \cdot \mathbf{A}_{\mathbf{N}} \right) \cdot \mathbf{x}_{\mathbf{N}} \quad \Rightarrow \quad f = \bar{f} + \mathbf{r} \cdot \mathbf{x}_{\mathbf{N}}$ **(2. 15)**

Since $\mathbf{x} \geq \mathbf{0} \Rightarrow \mathbf{x}_{\mathbf{N}} \geq \mathbf{0}$ (condition 3), the components of the vector $\mathbf{r} = \mathbf{c}_{\mathbf{N}} - \mathbf{c}_{\mathbf{B}} \cdot \mathbf{A}_{\mathbf{B}}^{-1} \cdot \mathbf{A}_{\mathbf{N}}$ will dictate whether condition 2 has been fulfilled. If its components are all non-negative, it is not possible to decrease the cost, and the present solution is the optimal solution to the problem formulated in equation **(2. 8)**. Note that all the transformations performed on the real variables of the original problem in order to suit the formulation to the SM need now to be reversed!

If the vector $\mathbf{r}$ has non-negative components, the cost might still be decreased. In order to decide how to make a jump to another vertex, the SM proposes to modify $\mathbf{x}_{\mathbf{N}}$ according to:

$$\mathbf{x_N} = t \cdot \vec{\mathbf{v}} \qquad \textbf{(2. 16)}$$

Where $t \geq 0$ and $\vec{\mathbf{v}}$ is a unitary basis vector. Thus, $t$ is equal to zero for the initial vertex.

Observing equation **(2. 15)**, it seems clear that the goal is to make $\mathbf{r} \cdot \mathbf{x_N} = \mathbf{r} \cdot t \cdot \vec{\mathbf{v}}$ as negative as possible for the next iteration, so that $\vec{\mathbf{v}}$ must be chosen such that it is the basis vector corresponding to the most negative value of **r**. Thus, the "entering variable" is chosen.

For the selection of the "leaving variable", the SM proposes as a general procedure examining the ratios of the vectors $\overline{\mathbf{x}}_\mathbf{B}$ and $\mathbf{A_B^{-1}} \cdot \mathbf{A_N} \cdot \vec{\mathbf{v}}$ componentwise (see **(2. 13)**). The component of $\overline{\mathbf{x}}_\mathbf{B}$ with the least ratio among the positive ones is selected as the leaving variable. If there is no positive ratio, the problem does not admit an optimal solution.

A more detailed review of this method would considerably enlarge the discussion, which is not the aim of this dissertation.

Beware that the SM can only deal with problems with linear cost and constraint functions, whilst real-world problems are almost never linear. Sometimes problems are "linearized" by force so as to suit the method, with the obvious consequences. In addition to that, many variables need to be added to suit the formulation **(2. 8)**, thus enlarging the search-space!

## 2.4.2 Local search

Local search methods iteratively compare a current solution to a transformation of itself, so that when the transformation is better than the current solution, it becomes the new solution. These methods can only guarantee to find a local optimum.

### 2.4.2.1 Hill-climber

The hill-climber is a local search method developed for discrete search-spaces. It evaluates all the neighbours of a current point, the best one of which is selected to compete with the current solution. If the new point is better than the current, it becomes the new solution.

Different algorithms can be implemented varying the definition of the neighbourhood. For example, the neighbourhood of a current point in a binary search-space could be defined as all those points that are separated from the current position by a hamming-distance equal to one.

The success or failure of the method is directly influenced by the selection of the starting point. Usually, several hill-climbers are started from a large variety of initial points, and the best solution among all of them is the one picked as the "global best", although there is no guarantee to be even close to the real global optimum. Nevertheless, improvement of the solution in real world-problems is already a great success.

It must be remarked that regardless the names of the algorithms, it is assumed here that optimization means minimization. Thus, methods like "hill-climber" and "steepest descent" are assumed to be conveniently adjusted for minimization problems.

### 2.4.2.2 Gradient-based methods

The gradient-based methods are suitable for continuous optimization problems, with the limitation that the function that is to be optimized must be at least continuous. Like the rest of the local search methods, they can only guarantee to find a local optimum. There are many different methods that rely, one way or another, on the gradient information to guide the search. Only their common underlying principles are very briefly discussed here.

The central idea is to find the maximum directional derivative from the current position in the search-space. This direction is given by the gradient $-\nabla f(\mathbf{x})$. Thus, the "steepest descent" method starts with a candidate solution $\mathbf{x}_k$ and iteratively generates new solutions following the update rule in equation **(2. 17)**.

$\mathbf{x}_{k+1} = \mathbf{x}_k - \alpha_k \cdot \nabla(\mathbf{x}_k)$, where $\alpha_k$ is the step size. **(2. 17)**

The bigger the step size the less local, but also the less precise, the search becomes. An interesting alternative is to design an adaptive $\alpha_k$.

Different criteria to select the direction to follow for the next candidate solution give shape to methods like "conjugate directions" and "conjugate gradients" methods.

Incorporating second order derivative information to the update rule gives shape to a number of Newton's methods, which enhance the convergence rate to the detriment of the robustness and at the computationally expensive cost of calculating Hessian matrices and their inverses. Moreover, Newton's methods extend the requirement of continuity to the function derivative.

### 2.4.3 Tabu search

Tabu search (TS) is different from the hill-climber in that it maintains a tabu list of recently visited solutions that are excluded from being the next current solution. Hence, the search is forced to explore new areas and can sometimes escape local optima after reaching them.

Since combinatorial optimization problems are out of the scope of this work, this method will not be discussed in much further details, although a brief explanation is given hereafter.

One of the classical combinatorial optimization problems are the permutation problems, such as the TSP. In these problems, each solution is represented by a list of the elements (say cities) to be permutated.

The neighbourhood of the current solution is typically defined as all the lists that differ from the current in one swap (i.e. in one exchange of positions). For instance, suppose the current solution that is given by the string:

| | | | |
|---|---|---|---|
| 1 | 2 | 3 | 4 |

All its neighbours are:

| | | | |
|---|---|---|---|
| 2 | 1 | 3 | 4 |
| 3 | 2 | 1 | 4 |
| 4 | 2 | 3 | 1 |
| 1 | 3 | 2 | 4 |
| 1 | 2 | 4 | 3 |

Given a current solution, the neighbours' solutions are calculated, choosing the one which results in best improvement. Suppose that this was attained by swapping the elements 2 and 3:

| | | | | | | | | |
|---|---|---|---|---|---|---|---|---|
| 1 | 2 | 3 | 4 | → | 1 | 3 | 2 | 4 |

For the next $n$ steps the swap of this pair is forbidden. A typical data structure for the tabu list is as follows:

| | 2 | 3 | 4 |
|---|---|---|---|
| 1 | | II | I |
| 2 | | III | |
| 3 | | | |

The numbers outside the table are the elements of the list, whereas the numbers inside it are the number of iterations left for the corresponding elements to be allowed to swap again. In this case $n$ = 3, so that the last swap was performed between the elements 2 and 3, while the previous was between elements 1 and 3, and the previous between elements 1 and 4. Therefore, even if one of these swapping leads to the best improvement in the next step, those choices are still forbidden. In addition, if no swapping leads to an improvement, a decision that causes temporal deterioration is compulsory. This might eventually help to escape local optima. Many important details of the method as well as its numerous alternatives were omitted here because they are not of interest within this work.

### 2.4.4 Divide and conquer

Some apparently complex problems lend themselves to being broken up into smaller simple problems. This process can continue until the sub-problems are so simple that they can be solved even by hand. Then, the method assembles an overall solution by combining the solutions of the sub-problems. Of course, most problems do not allow this procedure.

*The divide and conquer principle is intrinsic to many sorting algorithms… Polynomial and matrix multiplications can also be accomplished using this approach* [54].

Although this method is mentioned here for completeness, its precise details are well beyond the scope of this dissertation.

### 2.4.5 Branch and bound

The branch and bound method is suitable for big sized combinatorial problems such as the TSP. It differs from an exhaustive search through all the possible solutions in that it applies some heuristics in order to eliminate some non-promising areas of the search-space.

The first part of the method consists of setting a lower bound so that when a new solution has a cost that is already greater than the present lower bound, there is no need to evaluate how bad the solution is. Thus, entire areas of the search-space are not searched.

The initial lower bound could be calculated by generating a first tour at random, but there are special techniques to find a better initial value. The lower bound is very important, since the better it is, the faster the algorithm due to the elimination of more solutions.

The precise techniques to generate a first lower bound as well as the manner of eliminating parts of the search space will not be discussed here for obvious reasons.

## 2.5 Modern heuristic techniques

*There are many classic algorithms that are designed to search spaces for an optimum solution. In fact, there are so many algorithms that it's natural to wonder why there's such a plethora of choices. The sad answer is that none of these traditional methods is robust. Every time the problem changes you have to change the algorithm* [54].

Modern heuristics are general-purpose optimization methods which will do well on almost any type of problem even though a specifically designed problem-targeted algorithm would be probably more efficient. As a general feature, modern heuristics make use of techniques that are not directly related to optimization processes, heavily relying on stochastic operators.

Their condition of general purpose optimization algorithms together with their capabilities of escaping local optima, ridges and plateaus, are their major advantages. Most modern heuristic techniques imitate life or some other physical processes in nature to develop problem-solving techniques. Schwefel [68] suggests that the *…answer to why we imitate life on computers stems from an observed mismatch between the range of traditional crisp computing methods and the tasks we want to tackle today. That does not mean to abandon traditional methods, but merely to add new ones to our toolboxes for solving problems.*

## 2.5.1 Random search

This method is a simple way of searching binary spaces, though it can easily be adapted to continuous problems. Random allocations are generated and evaluated successively, always keeping the best value in memory. After an arbitrary number of trials, the best solution found is the one taken as the solution of the problem. It is not surprising that this method does not obtain impressive results!

## 2.5.2 Random walk

This method is an alternative to the random search. It consists of a point randomly allocated, whose coordinates in the binary hyper-space[3] are randomly modified by flipping one bit at a time in successive iterations. Each trial is evaluated, keeping in memory the best position found so far. As opposed to the random search, here the initial trial is of great importance because the next position is based on the current point. This algorithm is suitable for exploitation rather than for exploration purposes, focusing on a region of the search-space.

Both the random search and the random walk methods perform extremely poorly for problems with strong interactions among variables because they only modify one coordinate at a time.

[3] It is not difficult to think of means to adapt the method to continuous spaces.

## 2.5.3 Greedy search

The greedy search consists of a slight modification over the random walk. An initial point is positioned and evaluated. One coordinate of the binary search-space is flipped, but now the modification is kept only if the flipping results in a better performance. Thus, temporal deterioration of solution is not allowed. The performance does not increase much with respect to the random search and random walk, and it shows the same poor behaviour when the problem at issue presents strong interactions among the object variables.

## 2.5.4 Stochastic hill-climber

This is an improved variation to the random walk. A random bit-string is generated and evaluated. If nothing is known about the locations of good potential solutions, any initial trial anywhere in the search-space is as good as any other. Typically, initialization near the boundaries is avoided, so as to prevent from the possibility that the best value happen to be placed in a boundary of the search-space that is opposite to that of the initialization.

Then, the pattern (instead of a single bit) is modified. A typical means of implementing this is by setting a probability of flipping a bit (typically around 10%), so that on average, one every ten bits along the bit-string is flipped per iteration[4]. The new pattern is kept if it shows a better performance. Otherwise, the previous pattern is retained.

This way, the problem of not capitalizing the patterns of bits that work together (so-called “building blocks” in genetic algorithms), “may” be overcome. Therefore, this method might work for problems with strong interactions among the variables.

The problem with this method is how to set the probability threshold. On the one hand, if it is too high, it favours exploration, jumping from hill to hill, finding good areas, but failing at finding the “top of the hill”. On the other hand, if the mutation rate is too low, it guarantees to find the “top of the hill” of the area where the “individual” is placed, thus finding only a local optimum. Low mutation rates do not allow escaping local optima. A dynamically adapting mutation rate can be thought of, so that a high value enables the algorithm to search for a good area (exploration), and a later low value (exploitation) enables it to reach the top.

[4] A random number (from a uniform distribution) between 0 and 1 is generated for each bit. If it happens to be less than 0.10, the bit is flipped, keeping its value otherwise.

Notice that if the mutation rate is $\frac{1}{m}$, where $m$ is number of bits in the string, the method is the same as the "greedy search". Instead, if the mutation rate is 0.5, half the bits are flipped per iteration, so that the method becomes the random search. This is because the latter flips, on average, half the bits, since they are all randomly generated again.

## 2.5.5 Simulated annealing

Simulated annealing (SA) is a method suitable to deal with combinatorial optimization problems. It was developed under the natural metaphor of pre-heated molecules cooling into a crystalline pattern. … *In a molten metal the molecules move chaotically, and as the metal cools they begin to find patterns of connectivity with neighbouring molecules, until they cool into a nice orderly pattern – an optimum* [47].

In the same fashion as the stochastic hill-climber, SA generates a bit-string randomly. Then, it randomly flips bits selected according to a probability threshold, replacing the bit-string if the new one performs better. However, if it performs worse, it can still be kept according to another probability threshold, admitting temporary deteriorations which in turn might allow escaping local optima. Hence the algorithm is more likely to find the global optimum, or at least a better local optimum, than the stochastic hill-climber that only allows constant improvement. It is fair to remark that this concept of temporal deterioration has been introduced in many of the modern optimization approaches.

The probability threshold that a solution has of being accepted despite performing worse than the previous one is a function of the system "temperature". Since the system is "cooling", the probabilities of accepting poorer solutions decrease over time. The effect is that the system possesses more exploration abilities in the early iterations, slowly swapping into higher exploitation abilities as the iterations go by.

Notice that this is the first method up to now, which makes use of natural metaphors in order to optimize. The particle swarm optimization method, despite being different in very many important aspects, belongs to this group of natural-metaphor-inspired algorithms.

The name of the algorithm turns out to be unfortunate because it does not intend to simulate the natural process but to solve an optimization problem by a means that resembles pre-heated molecules cooling into a crystalline pattern. Natural metaphors sometimes are just a source of inspiration, or a clearer means of explaining how the method works when it turns out to be more difficult to understand it in an abstract fashion. Simulation is a totally different matter.

## 2.5.6 Population-based methods

Among the different branches of modern heuristic techniques, the so-called population-based methods comprise a set of techniques of most promising applications. They have already proven to be applicable to a variety of optimization problems, as well as data mining, pattern recognition, classification, prediction, machine learning, automatic programming, scheduling, supply-chain management, medical diagnosis, among many other tasks. These outstandingly robust methods outperform traditional methods when the problems at hand are of high complexity, such the case of NP-Hard[5] problems. They successively update a population of candidate solutions replacing the current population by a new better one, usually keeping the size of the population constant for each iteration. Thus, a parallel exploration of many optima can take place simultaneously, instead of the sequential exploration performed by traditional methods, whose point-to-point search is usually not capable of overcoming local pathologies. Therefore, the likelihood of getting trapped into a bad local optimum diminishes dramatically.

In addition to that, population-based methods only require the cost function[6] information to guide the search, without the need of obtaining additional information that is auxiliary to the problem (e.g. gradients, Hessian matrices, etc.).

Most (if not all) population-based methods are based on processes that occur in nature. Thus, as opposed to traditional methods, they make use of probabilistic rather than deterministic transition rules, applying stochastic operators to operations that guide the search, without implying that the search being carried out is random. Even when such intelligent beings as human beings face a new problem that has never been experienced before, they make a random decision, apprehending the consequent result. Random operators are necessary to include creativity and unpredictability into the system, as well as to avoid getting stuck into a systematic pattern. Moreover, they also allow making a decision when there is no clue that allows inferring what the next step should be. Notice that none of these can be accomplished by deterministic rules, no matter how complex they can become.

[5] NP stands for "Nondeterministic Polynomial". *For NP-Hard problems, no known algorithms are able to generate the best answer in an amount of time that grows only as a polynomial function of the elements in the problem* [32].

[6] In this dissertation, the cost function will be called **fitness function** when dealing with evolutionary algorithms, and **conflict function** when dealing with particle swarm optimizers.

Thus, the initial positions of the particles in the search space are typically decided at random, unless there is a "clue" for a better initialization. For the next iteration, some algorithms (e.g. evolutionary programming) try new allocations in the search-space by randomly varying the current positions, adding a **fitness-based** selection scheme so as to avoid the search to become random. By keeping only the best solutions, the new population is guaranteed to be placed in more promising areas. Other algorithms (e.g. particle swarm optimization) make use of **conflict-based** deterministic rules that point the current solutions to more promising areas of the search-space. Random variables weigh the deterministic rule in order to introduce creativity. Notice that creativity does not follow inference rules! Thus, new areas of the search-space are explored along their route to the most promising areas known so far.

Population-based methods basically rely on the behaviour emerging from the interactions among the individuals. Therefore, the behaviour of the whole system cannot be predicted by analyzing individuals in isolation, since each individual is aware neither of the behaviour of the group nor of the goals of the group, but it simply performs its role within the population. Thus, creativity emerges in a level higher than the individual's, due to the randomness embedded in the interactions. The concept is that by means of cooperation, the achievement of the population outperforms the sum of the achievements of all the isolated individuals. In population-based methods under the metaphor of cooperative behaviour, this means that cooperation overcomes competition.

These methods can find approximate solutions only. However, the aim in real-world problems is usually at finding a near-optimal solution rather than an exact one. For instance, Schwefel [68] suggests that the focus in evolutionary algorithms should be on improvement rather than on optimization. To clarify this concept, Kennedy et al. [47] offer their "law of sufficiency": *If a solution is good enough, and it is fast enough, and it is cheap enough, then it is sufficient.*

For example, consider the TSP, in which the shortest path through a number of cities has to be found. Testing all possible solutions for a tour through $n$ cities would require $\frac{(n-1)!}{2}$ trials. Clearly, for increasing numbers of cities, the problem becomes intractable very soon, and testing all possible solutions is out of the question. In contrast, population-based methods are well capable of finding very good solutions in a few hundreds of iterations.

Although the "efficacy" of a problem-solver is typically given by the "convergence theorem", which states that by having unlimited time, the algorithm must be able to find the global optimum with probability one: $\mathrm{P}\left(\lim_{t\to\infty}\mathbf{x}_t = \hat{\mathbf{x}}\right) = 1$, for practical applications it is not the global convergence with probability one what one is most interested in, but rather the capability of the algorithm to find a solution which is better than the best solution known so far. The "efficiency" of the algorithm is then measured by its rate of convergence.

## 2.5.7 Closure

This chapter gave a basic introduction to the field of optimization. Some of the different kinds of optimization problems were addressed putting aside the variational problems, which search for the optimal solution within a space of functions. In addition, the main broad stages of a complete optimization analysis were mentioned, while the attention was guided to the search of a solution to a pre-defined model. Thus, the problem is reduced to find the location of the optimal solution to a given function. That is to say, the solution is a vector of numbers!

Even within this kind of optimization problems, the combinatorial problems are not of interest in this thesis, since the particle swarm optimization method was originally developed for continuous optimization problems, which this dissertation is focused on. Hence little attention was paid to combinatorial problems, although they were not completely put aside because some population-based methods like genetic algorithms are suitable for these kinds of problems, while keeping strong links to the particle swarm optimization method.

A few traditional optimization methods were also outlined. Some of them were discussed in more detail than others, according to their importance in relation to the particle swarm approach. Thus, the SM was discussed in some detail in order to show its complexity in relation to the very simplified linear problems it can handle, and the number of variables that are necessary to add to the search-space just to formulate a limited linear model in such a way that the method can be applied. Likewise, gradient-based methods could have been discussed in more detail due to the fact that they can also be competitors to the particle swarm method, but this would have considerably enlarged this chapter. Nevertheless, their limited scope to continuous functions (or even to continuous function derivative), their lack of robustness and the need of expensive calculus of Hessian matrices and their inverses, were outlined as some of their weaknesses. Other methods more suitable for combinatorial problems were very briefly discussed, while some more attention was paid to TS because it introduces the idea of allowing a temporal deterioration of a solution to escape local optima. Another method that also allows temporal deterioration is SA, which additionally happens to be the first method inspired by natural processes. Some other related methods that are not very usable in practice (random search, random walk, greedy search and stochastic hill-climber) were also outlined because they serve as a guide towards the SA approach. The latter was not discussed in more detail because of space limitations. Nevertheless, the features that relate it to the evolutionary algorithms and to the swarm-intelligence-based methods were mentioned avoiding the details related to the precise thermodynamic metaphor.

Finally, the population-based methods were introduced without discussing their precise techniques, which will be dealt with in further chapters.

The particle swarm optimization approach is based on the intelligence that emerges from social interactions among individuals, so that the method is closely related to the field of artificial intelligence. Likewise, some views of evolution suggest that the latter is strongly linked to intelligence because evolution is a process of adaptation that creates biological beings of increasing intelligence. Thus, it is also related to the field of artificial intelligence.

Besides, both particle swarm optimization and the evolutionary algorithms are population-based methods that perform a parallel search of the problem space by relying on stochastic operators as well as on sharing the individuals' knowledge. In addition, the relatively new field called "artificial life" deals with artificial beings which can display life, so that they can learn, share knowledge, evolve, etc. Therefore, it encompasses the swarm-intelligence-based methods, the evolutionary algorithms, and many others, in a single field. In fact, artificial life is sometimes claimed to be a new route to artificial intelligence.

There are evident links amongst artificial intelligence, evolutionary algorithms, and particle swarm optimization. Therefore, they will all be discussed in some detail in next chapters.

# Chapter 3

# ARTIFICIAL INTELLIGENCE

This chapter gives a brief overview of the field of artificial intelligence, which is argued here to be composed of three main paradigms: the "symbolic paradigm", the "connectionist paradigm", and the "artificial life paradigm". They are briefly reviewed, and their capabilities and limitations are pointed out. The "expert systems" are discussed as the main applications of the "symbolic paradigm", the "artificial neural networks" as the main applications of the "connectionist paradigm", while the "evolutionary algorithms" and the "swarm-intelligence-based methods" are argued to be encompassed by the "artificial life paradigm". Thus, the possibility of handling optimization problems by means of artificial intelligence techniques becomes clear. The artificial neural networks are dealt with in more detail because their inclusion in this chapter is two-fold: on the one hand, they are the main applications of the connectionist paradigm as a means of generating artificial intelligence; on the other hand, they are also viewed as engineering devices that perform universal function approximations. However, they require training, which happens to be a very complex and difficult task that is traditionally performed by gradient-based techniques with their obvious limitations. Thus, such traditional training techniques are discussed towards the end of this chapter, while an alternative training by means of a particle swarm optimizer is performed in Chapter 12.

## 3.1 Introduction

Artificial intelligence (AI) is an important branch of computer science, which deals *with intelligent behaviour, learning, and adaptation in machines. Research in AI is concerned with producing useful machines to automate human tasks requiring intelligent behaviour. Examples include: …handwriting recognition, speech recognition, and face recognition… As such, it has become an engineering discipline, focused on providing solutions to practical problems. ...AI systems are now in routine use in many business, hospitals and military units around the world, as well as being built into many common home computer software applications and video games* [80].

Since its origins in 1956, the AI field has been in quest for understanding, modelling and designing intelligent systems. At the beginning, the aim was at creating a general intelligent agent, but the target was soon realized to be more difficult than it had been expected. Thus, the target turned into creating problem-specific (arguably) intelligent systems. So far, it had been assumed that all knowledge could be reduced to symbols, and the reasoning to a crisp-logic-based manipulation of those symbols. Thus, the "symbolic paradigm[1]" was developed.

Later, the massively parallel structure of the brain, which was realized to lead to astonishing results in handling the tasks which the symbolic approach had difficulty in dealing with, namely perceptual tasks, inspired a new approach called the "connectionist paradigm"[2].

At present, the AI field divides into different "schools of thought", although which those schools are, and their exact boundaries, are not clearly defined. Hence some researchers suggest two mainstreams, where "classical AI" refers to the "symbolic paradigm" and its derivatives (e.g. expert systems), while "computational intelligence" refers to all the other paradigms such as "artificial neural networks" (ANNs), "fuzzy systems", "evolutionary algorithms" (EAs), "particle swarm optimization" (PSO), "ant colony optimization" (ACO), etc. Others include both the "symbolic paradigm" and the "connectionist paradigm" into "classical AI", while "modern AI" comprises all the remaining approaches.

A new field[3] called "artificial life" (AL or ALife) encompasses a great number of paradigms based upon the metaphors of biological organisms' behaviours, such as EAs, PSO, ACO, "cellular automaton" (CA) and "agent-based methods". Notice that some of the paradigms included before in the "computational intelligence" school of thought (or in the "modern AI") are now included in AL. Since AL aims to create living organisms that require intelligence to survive, it can be seen as a route to AI. Thus, the AI field is split in this thesis into three so-called **paradigms[4] of AI**: the **symbolic paradigm**, the **connectionist paradigm**, and **AL**.

*The earliest computer scientists … were motivated in large part by visions of imbuing computer programs with intelligence, with the life-like ability to self-replicate, and with the adaptive capability to learn and to control their environments. These early pioneers of computer science were as much interested in biology and psychology as in electronics, and they looked to natural systems as guiding metaphors for how to achieve their visions. It should be no surprise, then, that from the earliest days*

---

[1] Paradigm: *a set of assumptions, concepts, values, and practices that constitutes a way of viewing reality for the community that shares them, especially in an intellectual discipline* [23]. This is a term widely used in this work.

[2] Although the "connectionist paradigm" is posterior, the first attempts to mimic neural networks were contemporary to the first developments on the "symbolic paradigm".

[3] Although it was officially created in 1987, the first related works date back to 1960s.

[4] Beware that the terms "school of thought", "paradigm" and "approach" might be used indistinctly. Furthermore, there is not such a term as meta-paradigm, so that a paradigm might encompass other paradigms.

*computers were applied not only to calculating missile trajectories and deciphering military codes but also to modelling the brain, mimicking human learning, and simulating biological evolution* [55].

Before dealing with AI, it is tempting to try to define intelligence, but despite long-standing efforts trying to define such concept, no widely accepted definition could be coined. Hence it is not intended here to develop a precise definition but just to present some interesting concepts from the literature, in order to have a "sense" of what intelligence and AI might be.

### 3.1.1 Intelligence

*Intelligence is usually said to involve mental capabilities such as the ability to reason, plan, solve problems, think abstractly, comprehend ideas and language, and learn* [80]. It is also claimed to be *the capacity to acquire and apply knowledge* [23], and *the faculty of thought and reason* [23].

Minsky suggests that *intelligence … means … the ability to solve hard problems* (from [35]), whereas Fogel et al. [35] argue that *all problems are hard until you know how to solve them*.

*The concept of intelligence has always been intended to distinguish "better" from "worse". Intelligence can be simply defined by a set of measures that support the experts' opinion of what comprises a "good" mind* [47]. Some widely agreed definitions have included *qualities such as memory, problem-solving ability, and verbal and mathematical abilities…* [47].

Notice that these definitions consist of an enumeration of qualities, while it seems more appropriate to say that those are just symptoms of intelligence. If attempted to define intelligence by enumerating its symptoms, all of them should be specified!

Furthermore, intelligence has been traditionally used to refer to humans' mental abilities, but there is no reason to limit the definition to humans. A good definition should be applicable to other animals and to computer programs as well.

*The study of intelligence in psychology has been dominated by a focus on testing, ways to measure the trait, and has suffered from a lack of success at defining what it is! …Intelligence has always been considered as a trait of the individual… …we will tend to view intelligence from the viewpoint of the population. …The achievements of outstanding individuals make us all more intelligent…* [47]. This concept is central to this thesis, since the PSO paradigm is based upon the intelligence that emerges from social cooperative interactions.

## 3.1.2 Artificial intelligence

A classical definition asserts that AI *is defined as intelligence exhibited by an artificial (non-natural, manufactured) entity. Such a system is generally assumed to be a computer* [80]. Another definition argues that it is *the ability of a computer or other machine to perform those activities that are normally thought to require intelligence* [23].

Of course, without a clear definition of intelligence, the previous definitions are imprecise. Besides, the assertion that simply the performance of an activity that requires intelligence implies the performer being intelligent is arguable: is a simple calculator intelligent just because performing arithmetic calculations require intelligence?

*Research in AI has typically passed over investigations of the primal causative factors of intelligence to more rapidly obtain the immediate consequences of intelligence… Efficient theorem proving, pattern recognition, and tree searching are symptoms of intelligent behaviour. But systems that can accomplish these feats are not, simply by consequence, intelligent*[5]*… They solve problems, but they do not solve the problem of how to solve problems* [35].

Rich argues that *artificial intelligence is the study of how to make computers do things at which, at the moment, people* (Bäck [6] suggests "living things" instead) *are better* (from [6]).

There has been great disagreement among the computer science community regarding the definition of AI for several decades. A traditional test to decide upon the intelligence of a machine, the "Turing test", states that if when an isolated individual who types questions to the computer and to another individual, cannot tell whether the answers come from the computer or from the individual, the machine is intelligent. The uncomfortable issue is that the most stupid human being can do extremely well in the Turing test, and the simplest computer can do extremely well in a traditional intelligence quotient (IQ) test for humans.

Fogel et al. [35] claim that a proper definition of intelligence should apply to humans and machines equally well, and suggest that *intelligence may be defined as the capability of a system to adapt its behaviour to meet its goals in a range of environments* (from [35]).

[5] Notice the contradiction of this assertion with respect to the definition of AI as *the ability of a computer or other machine to perform those activities that are normally thought to require intelligence* [23]!

## 3.2 Paradigms of artificial intelligence

Due to the lack of success to create general intelligent agents, the problem degenerated into many different specific problems. Namely perception, machine learning, game playing, theorem proving, reasoning methods, forecasting, problem-solving, automatic programming, learning and language processing, pattern recognition, modelling of evolution, etc.

From the very beginning, the field of AI focused on trying to model human's intelligence, either in its "behavioural rules" or in its "neurophysiology", since it was believed that intelligence was an individual characteristic, whilst humans are the most intelligent creatures.

Thus …*two alternative approaches were pursued to model intelligence: on the one hand, there was the symbolic approach which was a mathematically oriented way of abstractly describing processes leading to intelligent behaviour. On the other hand, there was a rather physiologically oriented approach, which favoured the modelling of brain functions in order to reverse-engineer*[6] *intelligence* [39].

However, although both main traditional paradigms were contemporary, findings by Minsky et al. (from [39])[7] led the funding for research on the neural network approach to be cut off, so that all the research in AI and cognitive science was oriented to the symbolic paradigm.

### 3.2.1 Symbolic paradigm

*The earliest approaches to artificial intelligence assumed that human intelligence is a matter of processing symbols. "Symbol processing" means that a problem is embedded in a universe of symbols, which are like algebraic variables; that is, a symbol is a discrete unit of knowledge … that can be manipulated according to some rules of logic* [47]. *The symbolic approach to artificial intelligence assumes the so-called "physical symbol system hypothesis" which asserts that intelligent systems can appropriately be described as symbol systems* [39].

In spite of the fact that the first electronic computers were aimed at performing numerical calculations, it was soon realized that there was not much difference with handling symbols instead, since the machines manipulated the stored bit-strings containing the units of information, without knowing anything about what they stood for. Thus, the symbolic paradigm was heavily influenced by the architecture of the conventional computers, also known as von Neumann machines[8], whose schematic representation is shown in **Fig. 3. 1**.

[6] Reverse-engineer: *to analyze a product to try to figure out its components, construction, and inner workings, often with the intent of creating something similar* [23].

[7] Minsky et al. (1969) (from [39]) proved that *although the early neural network mechanisms could learn anything a neuron could represent, a neuron could represent only very little*.

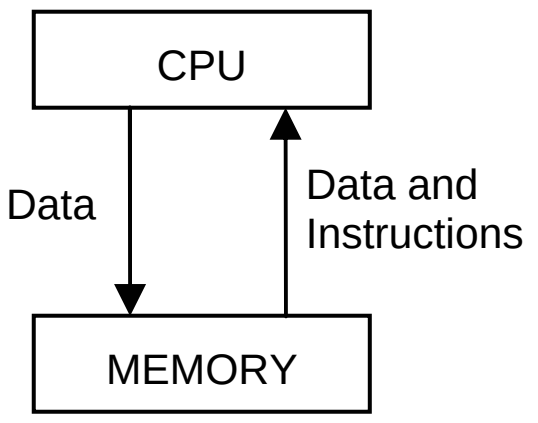

1. Fetch an instruction from memory.
2. Fetch any data that is required by the instruction from memory.
3. Process the data (i.e. execute the instruction).
4. Store the results in memory.
5. Go to step 1.

**Fig. 3. 1**: Representation of a generic von Neumann machine.

There are many problems that can be formalized in terms of an algorithm, which can then be broken down into a set of simpler statements which, in turn, can be reduced to the instructions that the central processing unit (CPU) executes. The statements are expressed in terms of symbols as units of knowledge, to which humans assign some meanings. It was hoped that all knowledge could be reduced to the manipulation of symbols according to certain rules, and that the whole thing could be implemented in a von Neumann machine. However, these machines need to be told in advance and in great detail the series of steps required to perform the algorithm (i.e. a computer program) because conventional computers cannot deal with noisy or incomplete data. Furthermore, not every problem can be formalized in terms of an algorithm, especially the highly complex problems that are found in real life.

Since the origins of AI in 1956, there was the vision of a general intelligent agent, not limited to intellectual tasks but also having human-like sensors to interact with the physical world. The agent is given a baggage of "knowledge" in the form of a set of different methods that would be eventually needed to deal with different situations. Thus, there must also be a component to define which method is to be used to deal with every given situation, and another one to store the specific conditions of the outer world in a suitable data structure, in order to determine the next action to take, and to notice changes in the external conditions.

*A symbol system has a memory capable of storing, retaining, and retrieving symbols and symbol structures. It has a set of information processes that form symbol structures as a function of sensory*

[8] *A von Neumann machine is a model created by John von Neumann for a computing machine that uses a single storage structure to hold both the set of instructions on how to perform the computation and the data required or generated by the computation. Most modern computers use this von Neumann architecture. Computers using this architecture are said to be "von Neumann machines"* [80].

*stimuli. Furthermore, it has a set of information processes which produce symbol structures which cause motor actions and modify symbol structures in memory in a variety of ways. A symbol system interacts with its environment in two different ways:*

- *It receives sensory stimuli from the environment which it converts into internal symbol structures.*
- *It acts upon the environment in ways determined by symbol structures that it produces by internal information processes.*

*Thus, its behaviour can be influenced by both its current environment through its sensory inputs, and by its previous environments through the information it has stored in memory from its experiences* [39].

The symbol systems describe the problem by manipulating symbols as discrete elementary units. For instance, in high-level programming languages, possible types of symbols are “integer”, “real”, “character”, “Boolean”, etc., while the manipulation rules are typically sequential arithmetic and crisp logic operations. However, these methods are far too simple to handle the complex situations that a “general intelligent agent” is expected to deal with. The ambitious goal of creating a general intelligent agent was never quite achieved, and it was put aside in favour of the development of “expert systems”.

A system, generally speaking, can be characterised with different levels of description. For instance, see the traditional levels of a computer system description shown in **Fig. 3. 2**. Notice that different levels of description are at the same time different levels of abstraction.

1. Device Level
2. Circuit Level
3. Logic Level
   a) Logic Gate Sub-Level
   b) Register Transfer Sub-Level
4. Program Level
   a) Machine Code Level
   b) Assembler Program Level
   c) Problem Oriented Programming Language Level

**Fig. 3. 2**: Typical levels of increasing degree of abstraction for describing the operation of a computer system (from [39]).

*A description at the device level reflects the physical properties of the involved devices…* which *…become increasingly invisible at higher levels of system description. The circuit level disregards certain interconnection patterns between certain devices and considers only compound structures of several devices…* which have functionalities that *…are useful for composing even more complex functions. Such more complex functions are the elementary units of the next higher level of description or abstraction… Since the more abstract levels of descriptions are essentially simplifications of the descriptions at less abstract levels, it is clearly possible to describe the operation of a complex computer system at the device level*, though *this is usually overly difficult* [39].

At high levels of abstraction such as the program level, lower level interactions such as the voltages that take place at interconnections between transistors are ignored, without that resulting in limitations to the description of the overall computer system's input-output behaviour. For instance, the arbitrary possible interconnections of transistors, capacitors, and resistors at the device level lead to extremely complex behaviours of the overall system, difficult to handle, to understand, and to describe at that level. Therefore, functional easy-to-use units are built up from the elementary units at one level, which become the units of the next higher level. Thus, the modular[9] design of a system allows the input-output behaviour of the entire system to be described at any given level of abstraction, and any description can be decomposed into its next lower level of abstraction.

There are many other highly important aspects of the symbolic paradigm such as the concepts of the "knowledge level", "rationality", "knowledge representation and reasoning", and "reasoning with uncertainty", which cannot be discussed in this work for obvious reasons. Notice however that the subject matter is of great importance for further work on the development of a "general-purpose robust optimizer".

## 3.2.2 Connectionist paradigm

*It seems clear that humans learn word meanings from contexts… The view of high-dimensional semantic space is a bottom-up view of language, as opposed to the top-down imposition of definitional rules in the usual dictionary reference* [47].

Connectionism *claims that the idea of symbol systems is insufficient for describing exhaustively truly intelligent systems. The connectionist claim is backed up by philosophical analysis that human thought and intelligence rest essentially on "non-symbolic structures"* [39].

Around 1986, the neural network (NN) approach, which had come to be known as the "connectionist paradigm", returned to challenge the dominance of the "symbolic paradigm". Some of the reasons for its renaissance were the failure of the symbolic paradigm in achieving the great goals it had been predicted to, as well as several philosophical objections claiming that the human thought does not use a fixed representation of its environment. Besides, developments in micro-electronics made it possible to build powerful massively parallel computers aimed at simulating intelligence and life.

[9] "Modular design" means here that the functionality of the units at a given level is independent of the interactions among the units in lower levels.

Therefore, the massively parallel and highly interconnected brain-like structure of the new computer systems became increasingly popular, to the extent that some researchers claimed that the AI field should turn into that direction. In any case, the two approaches became the two mainstreams and competitors once again[10].

*Many engineers welcomed the neural network models and the associated learning procedures, as they dealt with continuous values and the mathematical foundations of the techniques were continuous mathematics. This was the type of mathematics they were familiar with from their own discipline. This contrasted … with the learning techniques which had been developed in the symbolic paradigm* [39].

Notice that although the connectionist paradigm is influenced by different fields such as AI, cognitive science, philosophy, psychology and sociology, the present work focuses primarily on the engineering point of view.

Hoffman [39] claims that *…engineers often look at neural networks rather as general function approximators than as a paradigm for the development and understanding of comprehensible intelligent systems*. Despite the somehow disappointing accuracy of Hoffman's assertion, it is claimed here that there is no need of such a restriction, since the development of general-purpose, robust and self-adapting optimizers does not differ much from the quest for the general intelligent system that AI has been pursuing for so long. Thus, although the scope of this thesis is far too limited to quest for such a goal, further work in this regard seems definitely worthy. Notice that the population-based optimization methods are based on "intelligent" techniques to search a huge high-dimensional hyper-space!

Since this thesis is focused on the PSO paradigm, the ANNs will be dealt with mainly as engineering devices that need to be trained. However, the quest for intelligent optimizers is not abandoned, and the intelligent behaviour is approached by means of a massively parallel search of the problem space performed by a population of cooperative particles. Thus, the whole swarm could be claimed to be an intelligent agent!

[10] Notice the parallelism of the "phenotypic approaches" and "genotypic approaches" in EAs (see **Chapter 4**), with the "symbolic approach" (behavioural) and "connectionist approach" (neurophysiologic) in AI!

### 3.2.3 Artificial life

AL is a relatively new field that encompasses all the paradigms based upon the metaphor of biological organisms. As a route to AI, AL does not intend to model biological life but to create new life that would display intelligence, by exploiting principles underlying living organisms[11]. When researchers in AL concern themselves with intelligence at all, they tend *to focus on the "bottom-up" nature of emergent behaviours*[12] [80].

AL includes paradigms such as "evolution strategies" (ESs), "evolutionary programming" (EP), "genetic algorithms" (GAs), "genetic programming" (GP), PSO and ACO. These paradigms will be discussed in some details in further chapters because they can deal with optimization problems, whilst other paradigms more related with artificial living creatures, namely cellular automata (CAs) and biomorphs, are well beyond the scope of this work.

*Von Neumann was one of the first people ever to think of ways to create artificial life. He came up with the ideas behind automata. He felt that, in principle there was nothing different between a carbon-based life form and an automata except for level of complexity. Saw no reason why this couldn't be done. A sufficiently complex finite automaton should be able to carry on any life functions that a living creature could…* [52] (from [60]).

*This quest for creating and improving artificial creatures is with the hopes that one day they will reach a point where they will even start developing an intelligence of their own. This would be the ultimate goal in artificial life - to make something that could actually think the way humans do!* [60][13].

Kennedy et al. [47] wonder: *Where is the boundary between life and inanimate physical objects, really? And how about those scientists who argue that … an insect colony is a superorganism – doesn't that make the so called death of one ant something less than the lost of a life, something more like cutting hair or losing a tooth? … the creation of … life-like beings in computer programs, with goal-seeking behaviours, capable of self-reproduction, learning and reasoning, and even evolution in their*

[11] *…Perhaps the question is not "how do we make computers simulate carbon based life?" as much as it is "how do we make computers who share the properties of carbon based life?"* [60]

[12] *An emergent behaviour or emergent property can appear when a number of simple entities (agents) operate in an environment, forming more complex behaviours as a collective. The property itself is often unpredictable and unprecedented, and represents a new level of the system's evolution. The complex behaviour or properties are not a property of any single such entity, nor can they easily be predicted or deduced from behaviour in the lower-level entities. The shape and behaviour of a flock of birds or school of fish are good examples* [80].

[13] Note the contradiction between "…developing an intelligence of their own" and "…think the way humans do". The aim in AL is precisely to leave behind the traditional reliance on human's knowledge.

*digital environments, blurs as well the division between living and nonliving systems. Creatures in artificial life programs may be able to do all the things that living things do. Who is to say that they are not themselves alive? And why does it matter?*

If attempting to create artificial living creatures, the question of what life is arises. In the same fashion as in AI, no crisp definition is sought, but some interesting ideas are taken from the literature, which give a sense of what life and AL might be.

### 3.2.3.1 Life

A typical definition argues that life is *the property or quality that distinguishes living organisms from dead organisms and inanimate matter, manifested in functions such as metabolism, growth, reproduction, and response to stimuli or adaptation to the environment originating from within the organism* [23]. What the definition does not specify is which those properties must be displayed simultaneously for the thing to be considered alive. For instance, think of ...*the property of reproduction, which most people consider one of the most important qualities: Does a single creature without this ability suddenly cease to live? Is a human who is impotent no longer alive?* [60].

*In biology, a lifeform has traditionally been considered to be a member of a population whose members can exhibit all the following phenomena at least once during their existence: growth, metabolism*[14]*, motion*[15]*, reproduction*[16] *and response to stimuli*[17] [80].

*One interesting idea, discussed by Steven Levy* [52]*, is that life is possibly a continuum rather than the binary values of alive or not alive. Things might actually have relative levels of "aliveness" depending on various qualities such as reproduction abilities, complexity, and many more. ... But even if the idea of a continuum is accepted, the question still remains as to just what those qualities should be* (from [60]).

It was claimed before that a definition of "intelligence" should be applicable to any and not only to carbon-based beings. Likewise, a definition of "life" should be independent of the material the life is made out of.

Given that the definitions of "life" are based on a number of properties that a living being must manifest, and taking into account Levy's interpretation of aliveness as being continuous rather than binary [52], things can be claimed to have degrees of aliveness, according to the properties they display. This concept is valid for both carbon-based and silicon-based beings.

[14] *Metabolism: growing by absorbing and reorganizing mass and excreting waste* [80].

[15] *Motion: either moving itself, or having internal motion* [80].

[16] *Reproduction: ability to create entities that are similar to, yet separate form, itself* [80].

[17] *Response to stimuli: ability to measure properties of its surrounding environment, and act upon certain conditions* [80].

Consider for example the following classification, from less to more alive: inanimate matters, cars[18], EAs' agents[19], viruses[20], fire[21], bacteria, plants, worker bees, mules[22], animals that can reproduce. Of course, what the properties are and the order of importance are arguable, and there will surely be disagreements with this ordering.

### 3.2.3.2 Artificial life

Common definitions from the dictionary argue that *artificial life ... is the study of life through the use of human-made analogs of living systems* [80], or that it consists of *the simulation of biological phenomena through the use of computer models, robotics, or biochemistry* [23]. Another definition claims that AL is …*the study of synthetic systems which behave like natural living systems in some way* [23].

In this thesis, AL is simply thought of as man-made life, which can display different degrees of aliveness and different degrees of intelligence. Perhaps the most distinctive feature of the AL paradigms is that the intelligent behaviour is an emergent property. This means that the complex intelligent behaviour is formed from the lower level individual-to-individual interactions, apparently not related to the higher level resulting property. Thus, the emergent behaviour cannot be easily inferred from the lower level individuals' behaviours, so that it can be neither "intended to" nor "implemented" on purpose in AL.

*It does not appear possible to unambiguously decide whether a phenomenon should be classified as emergent, and even in the cases where classification is agreed upon it rarely helps to explain the phenomena in any deep way. In fact, calling a phenomenon emergent is sometimes used in lieu of any better explanation. One reason why emergent behaviour occurs is that the number of interactions between components of a system increases combinatorially with the number of components, thus*

[18] They display motion, as opposed to inanimate matters.

[19] They display motion, they reproduce, and they adapt to the environment by evolving their structure to improve their performance (see **Chapter 4**). They display some degree of intelligence!

[20] They display a carbon-based genetic code of their own, but do not grow and cannot reproduce outside a host cell.

[21] Notice that according to the previous definitions, fire displays growth, metabolism, motion, reproduction and response to stimuli! However, it is not good at adapting to different environments (thus, it is not very intelligent).

[22] Worker bees and mules are almost fully "alive", but they cannot reproduce. Notice that if the whole colony of bees is considered as a super-organism, it is fully "alive"!

*potentially allowing for many new and subtle types of behaviour to emerge.* However, *...merely having a large number of interactions is not enough by itself to guarantee emergent behaviour; many of the interactions may be negligible or irrelevant, or they may cancel each other out* [80][23].

It has been argued that "emergence", like "randomness", is simply a word to cover up the ignorance about the relationships between causes and effects in these complex systems. An emergent feature of an artificial system means in this work that it was not specifically and intentionally programmed to appear.

#### 3.2.3.3 Game of life

Gardner (from [47]) argued that *...an arbitrary starting point for the story of the modern paradigm known as artificial life is ... mathematician John Conway's "game of life"*. In short, it consists of a grid of binary elements arranged in a torus fashion[24], which are re-arranged in a two dimensional rectangular plane. Each "pixel" in the grid can present one of two states (on or off), which depends on a simple set of rules:

- Every pixel is surrounded by eight other pixels.
- Whenever less than two of the surrounding pixels are on, a living pixel dies of loneliness.
- Whenever more than three surrounding pixels are on, a living pixel dies of overcrowding.
- If the pixel is "off" and is surrounded by three living pixels, it is born (i.e. it turns on).

Thus, the state of a pixel in the next time step depends on the present state, and on the states of the neighbour pixels. When this is implemented into a computer program, a sequence of never repeating patterns appear on the screen, which might last a very long time before all the pixels "die out". This could be considered the origin of AL. Here the patterns are not intentionally programmed but they emerge, unpredicted, from a very simple set of rules that are based on local interactions among adjacent pixels only!

---

[23] *In some cases, a large number of interactions can in fact work against the emergence of interesting behaviour, by creating a lot of "noise" to drown out any emerging "signal"; the emergent behaviour may need to be temporarily isolated from other interactions before it reaches enough critical mass to be self-supporting. Thus it is not just the sheer number of connections between components which encourages emergence; it is also how these connections are organised. ...In some cases, the system has to reach a combined threshold of diversity, organisation, and connectivity before emergent behaviour appears. ...Systems with emergent properties or emergent structures may appear to defy entropic principles and the second law of thermodynamics, because they form and increase order despite the lack of command and central control. This is possible because open systems can extract information and order out of the environment* [80].

[24] This means that when observing the grid in a rectangular shape, opposite borders are in reality adjacent.

### 3.2.3.4 Some other "artificial creatures"

Some other well known artificial creatures are the biomorphs created by Dawkins (from [47]), which are encoded into a chromosome of nine genes, each one of which can take the form of any of ten alleles. The genes encode the rules for the development of the biomorph. EAs are applied so that the biomorph undergoes evolution. Other examples are:

Cellular automaton: A cellular CA is a very simple virtual machine from which complex behaviour emerges. The simplest version is binary and one-dimensional. Thus, a ring is displayed for each iteration in the form of a line segment, where each bit or cell can take the "on" or "off" position. Successive iterations generate different segments, which are placed one below the other, giving shape to a certain pattern. Each cell's next state depends on the present state and on the neighbours' states, according to a pre-defined set of rules that are applied to every cell in the present bit-string. The states in the new bit-string are calculated and printed in the next row. The evolution of the system can lead to a homogeneous state, to periodic structures, to a chaotic pattern, or to complex localized structures. The game of life is a particular case of a CA that leads to this last situation.

*Computer Viruses: This is usually one of the simplest levels of artificial life, on a par with how real viruses compare to animal life. They do, however, exhibit most of the "required" life signs, such as reproduction, integration of parts, unpredictability, and such.*

*Robotics: These are physical manifestations of experiments in artificial life. They usually have the properties of complexity, integration of parts, irritability, movement, and a few others, but the lack the ability to reproduce. Research is, however, being done on creating robots that will create more of themselves* [73] (from [60]).

### 3.2.3.5 Learning and evolution

It is widely accepted that the "two great stochastic systems", "mind" and "evolution", follow parallel paths. They both adapt by heavily relying on probabilistic choices to introduce unpredictability and creativity, thus behaving very much alike.

*Some theorists argue that mental processes are very similar to evolutionary ones, where hypotheses or ideas are proposed, tested, and either accepted or rejected by a population. ...the "memetic" view, proposed by Dawkins ... suggests that ideas and other cultural symbols and patterns, called "memes",*

*act like genes; they evolve through selection, with mutation and recombination just like biological genes, increasing their frequency in the population if they are adaptive, dying out if they're not. …Ideas spread through imitation; for instance, one person expresses an idea in the presence of another, who adopts the idea for his or her own. This person then expresses the idea, probably in its own words, adding the possibility of mutation, and the meme replicates through the population* [47].

Hence these two processes, which are typically used as natural metaphors for artificial engineering problem-solving algorithms, often compete with each other to solve the same kinds of problems. However, learning in nature takes place during each individual's lifetime, whereas evolution does through generations. Without taking into account certain negligible mutations, an individual's genetic code does not change during its lifetime.

The big question is then if these processes are independent or if they interact affecting each other. That is to say, does learning affect evolution? Does something that has to be learned today insert into one's genetic code through evolution, so that one is born already knowing it? Does animals' instinct have anything to do with this?

**The "Baldwin effect"**

*Learning during one's lifetime does not directly affect one's genetic makeup; consequently, things learned during one's individual's lifetime cannot be transmitted directly to its offspring. However, some evolutionary biologists have discussed an indirect effect of learning on evolution… The idea … is that if learning helps survival, then organisms best able to learn will have the most offspring and increase the frequency of the genes responsible for learning. If the environment is stable so that the best things to learn remain constant, then this can lead indirectly to a genetic encoding of a trait that originally had to be learned. In short, the capacity to acquire a desired trait allows the learning organism to survive preferentially and gives genetic variation the possibility of independently discovering the desired trait… In this indirect way, learning can affect evolution, even if what is learned cannot be transmitted genetically* [56].

Levy [52] has implemented a "harsh" artificial world inhabited by four different kinds of creatures, who had to learn from experience, trying to find out which plants in the environment were comestible. Some beings had the ability to learn, some to evolve, some both abilities, whereas some had neither. Beings without the ability to learn died out very fast, with little difference between the ones who could evolve and those who were not given the ability to do so. The ones with both abilities were the ones that lasted longer, though the effect of evolution took very long to differentiate beings with both abilities from the ones with only the ability to learn. After three million time-steps, the beings with both abilities knew the information from birth! In time, only these beings were left in the world.

## 3.3 Expert systems

Once the neurophysiologic approach was put aside around 1969, all the research in cognitive science and AI was conducted in the symbolic paradigm. While the predictions were exciting, the actual achievements were rather disappointing, to the extent that the aspirations to develop a general intelligent system were practically abandoned, and the work was then oriented towards the development of problem-specific intelligent systems.

*…the focus in the development of expert systems was to develop useful systems with a limited scope of intelligent behaviour. This was opposed to the earlier hopes to develop principles … which would lead to systems of general intelligence. Expert Systems were conceived to adopt substantial aspects of human (expert) reasoning as well as to capture the relevant knowledge of the domain of expertise in order to produce sensible advice* [39].

Expert systems were originally conceived to simulate human expert behaviour within a restricted domain of expertise. It was later noticed that it is frequently very difficult to ensure a system’s performance at the expert level, what could be particularly dangerous in systems used, for example, to perform medical diagnosis. In addition to that, human experts know the boundaries of their expertise, and would rather admit their ignorance than give a senseless answer. Furthermore, their boundaries smoothly degrade, in contrast to the strict boundaries of expert systems. Thus, the role of expert systems today is more as assistant advisors of actual human experts than as experts themselves.

The “first generation of expert systems”, also known as “rule-based expert systems”, consists of systems that merely have a rule base which contains the problem-specific part of the system (i.e. the “domain knowledge”) and a rule-inferring mechanism that decides which of the rules are to be applied in a certain situation, and chains rules together so that the desired conclusion can be reached. Notice that the desired conclusion is already implemented within the system. In other words, the system merely selects one of the possible solutions. The “second generation of experts systems” differ in that they separate the knowledge about how to do the task (“task knowledge”) from the “domain knowledge”. The “task knowledge” encompasses all the knowledge regarding how to perform the task, and it is still implemented within the “symbol level” of abstraction. Differently, the “domain knowledge” encompasses all the knowledge about the environment or domain in which the system should function, and introduces a new higher level of abstraction: The “knowledge level”. Recall that in “rule-based expert systems”, all the knowledge is represented in a single level of abstraction.

The “behavioural rules” are defined and programmed in such a way that the “expert system” is expected to respond to stimuli in the same manner as the expert it emulates would do, but the rules are sometimes conflictive, and many experts may provide different rules for the same situation. Furthermore, notice that expert systems take the knowledge from human experts who already have it, so that these systems cannot solve any of the problems for which humans have not developed sufficient expertise. Nowadays, these kinds of systems are not considered to be intelligent, since they only solve problems by means of deterministically programmed rules that rely on humans’ knowledge.

Bäck [7] offers: *Following a commonly accepted characterization, a system is computationally intelligent if it deals only with numerical data, does not use knowledge in the classical expert system sense, and exhibits computational adaptivity, fault tolerance, and speed and error rates approaching human performance*.

A big limitation of the symbolic paradigm consists of its assumption that at a certain point the user assigns meanings to the symbols, or that they can figure them out from a previous chain of logically linked other symbols. The logic rules can infer facts and/or meanings from other facts and/or meanings that were in turn deduced from others, and so on, according to the logic laws that rule the reasoning. At most, this can go back to the beginning of the program, but in the end, a symbolic algorithm depends on some facts that are given to it by the user. Another important limitation of the symbolic paradigm is that it is based on crisp logic, whose black-and-white nature does not reflect real life processes, since in real life something is not true or false but “kind of true” and “kind of false”, in different degrees. For instance, in continuous problems, a variable like *“temperature” might have a range of states: cold, cool, moderate, warm, hot, very hot. Yet the change from one state to another is not precisely defined. At no point can an increase of a tenth of a degree be said to change “this is warm” into “this is hot”…* [18].

# 3.4 Fuzzy systems

Fuzziness refers to non-statistical imprecision and vagueness in information and data. In set theory an element either belongs to a set or not, in crisp logic a statement either is true or false, whereas fuzziness reflects a degree of membership or truth in the range [0,1], where "0" means absolutely false (or absolutely not belonging to) and "1" means absolutely true (or absolutely belonging to). It is an expansion of the classical set theory, and a generalization of the conventional Boolean TRUE – FALSE logic (so-called "crisp logic").

Fuzzy systems are similar to conventional systems, with the difference that the former contain fuzzifiers to convert the inputs of the system into a fuzzy representation, and defuzzifiers to convert the output of the fuzzy process into a numerically precise solution.

In fuzzy systems, all the *rules are implemented in parallel, all at once, and the conclusion is obtained by combining the outputs of all the if-then rules and defuzzifying the combination to produce a crisp output* [47].

## 3.4.1 Fuzzy sets

In order to clarify the concept of fuzzy membership, it is typical to discuss its difference with respect to a probability. Suppose on the one hand the statement that there is a probability of 0.7 that tomorrow the weather will be windy, and on the other hand the statement that tomorrow the weather will have a membership to the set of windy weathers equal to 0.7. By tomorrow, the weather will be either windy or not in the first case (only two possible membership values), whereas in the second the weather will be undoubtedly "quite" windy, although the exact notion of how windy a membership of 0.7 stands for, is subjective.

There are two alternatives to represent a membership function: continuous or discrete. The former consist of a mathematical function where the argument is the element and the output is its membership, while the latter is just a list of discrete points. Each element in a fuzzy set is represented by an ordered pair, where the first component is the element itself and the second is its membership to the fuzzy set. Note that a four-year-old kid might, for example, belong to the set of very young people (say with a membership of 0.99), to the set of young people (say with a membership of 0.87) and to the set of old people (say with a membership of 0.05).

The operations on fuzzy sets are defined in the same fashion as on the classic set theory. Consider, for instance, the two basic set operations: “intersection” and “union”. Let A and B be fuzzy sets on a mutual universe, and let $m\mathrm{A}(x)$ and $m\mathrm{B}(x)$ be the memberships of the elements of the universe to the sets A and B. Let also $\mathrm{C} = \mathrm{A} \cap \mathrm{B}$ and $\mathrm{D} = \mathrm{A} \cup \mathrm{B}$.

Then, $m\mathrm{C}(x) = \min(m\mathrm{A}(x), m\mathrm{B}(x))$, and $m\mathrm{D}(x) = \max(m\mathrm{A}(x), m\mathrm{B}(x))$.

That is to say that the membership of an element to a set defined as the intersection of the fuzzy sets A and B is equal to the minimum of its memberships to those sets, while its membership to a set defined as the union of A and B is equal to its maximum membership. The reason for this is obvious: if an element is 0.2 member of A and 0.8 member of B, it is 0.2 member of both at the same time, and 0.8 member of any of them. All the set operations defined for crisp sets are mathematically defined for fuzzy sets. Refer to [43] for an easy-to-read review of the different set operations in fuzzy sets, as well as their properties.

## 3.4.2 Fuzzy logic

Crisp logic is the basis of the symbols’ manipulation in the symbolic paradigm. However, since applications like “expert systems” are expected to arrive to the conclusions that human experts would, it seems reasonable to attempt to introduce fuzzy logic within their rules in order to mimic the humans’ reasoning more accurately.

In fuzzy logic a proposition may be true, false, or have some intermediate truth values. They can be continuous, but typically they are allowed to take one of a series of discrete values (multi-valued logic). All the typical connectives for crisp logic, namely “not”, “and”, “or”, “if-then” and “if and only if”, are defined for fuzzy logic.

For instance, the table of truth for the fuzzy logic “or” problem is the same as the “union” operation in fuzzy sets. That is to say that if a sentence “a” has a truth value equal to say 0.5, and a sentence “b” has a truth value equal to say 0.3, the truth value for “a or b” is 0.5.

Likewise, the fuzzy logic “and” problem is equivalent to the “intersection operation” in fuzzy sets, so that the truth value for “a and b” in the previous example is 0.3.

It must be noticed, however, that there are some problems with the “if-then” connective, for which very many different solutions have been proposed (refer to [43]).

*Expert systems have been the most obvious recipients of the benefits of fuzzy logic, since their domain is often inherently fuzzy… Fuzzy systems … provide a rich and meaningful addition to standard logic. The mathematics generated by these theories is consistent, and fuzzy logic may be a generalization of classic logic. The applications which may be generated from or adapted to fuzzy logic are wide-ranging, and provide the opportunity for modelling of conditions which are inherently imprecisely defined, despite the concerns of classical logicians. Many systems may be modelled, simulated, and even replicated with the help of fuzzy systems, not the least of which is human reasoning itself* [12].

Fuzzy expert systems are succeeding where conventional expert systems have failed due to rule sets that do not cover all the situations. Fuzzy rule sets typically require far fewer and simpler rules than conventional AI systems. Moreover, fuzzy systems can be evolved by EAs.

## 3.5 Artificial Neural Networks

The "neurophysiologic" branch of AI developed the ANN paradigm by roughly mimicking the highly interconnected parallel structure of the brain. However, it *…should be noted that neural networks are often used for pattern classification tasks and other engineering problems, which is quite different from the connectionist claim that connectionists systems, composed of a large number of neurons, are suitable for building complex intelligent systems* [39].

It is not the objective of this thesis to explore the possibilities of creating general, intelligent systems (although that would certainly solve all the problems dealt with here!), but to discuss the ANN paradigm as an engineering tool. The ANN approach as a route to the creation of general intelligent systems is definitely worthy of further work, even if the research is "only" limited to optimization tasks, since a general-purpose robust intelligent optimizer can be thought of as a simplified general intelligent system with its range of applications limited to optimization tasks (much less ambitious, whilst still following the same line). Therefore, ANNs are thought of as engineering devices in this work, while its great expectations as a means of creating AI are left for future and parallel work.

From a practical engineering point of view, ANNs are basically nonlinear mapping functions that seek to find the relation between some input and output uncorrelated data. Thus, as an engineering tool, ANNs are claimed to possess the ability to perform universal function approximation (from [35]).

Some of the advantageous features of a neural network (NN) are that systems using them typically take shorter to prototype than systems using more conventional approaches. Furthermore, a NN-based system tolerates missing, noisy and incorrect data better than most others. The disadvantage is that it gives no explanation of the behaviour of the system, since it just links the data without theoretical background, so that no information of the process is given. However, the structure of these devices needs to be designed, and the strengths of the connections need to be set by means of a so-called "training technique". Therefore, there are many kinds of ANNs that differ mainly in their style of learning and in their architecture.

Regarding the style of learning, two main groups can be distinguished, namely supervised and unsupervised learning. The former successively corrects possible errors in the performance of the ANN in relation to a training data set provided by the user. In contrast, the latter works with unclassified training data, attempting to find a meaningful structure in the presented data. Only supervised learning is discussed in this work, since the aim is to show the goodness of the PSO paradigm at handling the task, while the latter requires the output of the training data set to calculate the fitness of its particles.

With regards to the architecture[25], two main groups can be distinguished as well, namely feed-forward networks and feed-back (or recurrent) networks. In the former, the output of a neuron becomes the input of other neurons that belong to layers that are closer to the entire network's output(s)[26]. In contrast, recurrent networks can have signals travelling in both directions by introducing loops into the network.

Notice that while only feed-forward networks, composed of a short amount of processing elements (PEs), are dealt with in this thesis, the human brain is believed to rely on extremely complex and massively parallel feed-back networks, composed of thousands of millions of neurons. Since the ANN paradigm is considered here as an engineering tool, it is natural to deal with the already proven efficient and relatively simple multi-layer perceptron (MLP) model[27]. However, for further work in quest for a general purpose intelligent robust optimizer, further research on feed-back networks should be seriously taken into consideration.

---

[25] An ANN is composed of a collection of neurons, commonly organized in layers. Other types of organizations are outside the scope of the present work.

[26] That is to say that the network has signals travelling in one way only, from input to output.

[27] See section **3.5.5.2**.

### 3.5.1 Biological background

A detailed outline of the anatomy of the human brain is of course out of the scope of this work. However, since the ANN approach was developed under that biological metaphor, it seems fair to include a brief and simplistic introduction to the subject matter.

The human brain is estimated to contain around $10^{10}$ to $10^{12}$ neurons[28], each of which has between $10^3$ to $10^4$ connections to other neurons, called synapses[29]. The different types of specialized neurons are arranged forming a huge massively interconnected network, which are coupled with receptors and effectors to interact with the environment. Thus, the receptors provide the input to the network, whereas the effectors control the reaction of the whole human body to the stimulus.

The specialized neurons that directly receive the input from the sensors are called "sensor neurons", whereas the ones responsible for the control of the muscles are called "motor neurons". In between, there are many different types of neurons which receive the inputs form other neurons' synaptic endbulbs, and pass impulses to other neurons' dendrites (see **Fig. 3. 3**).

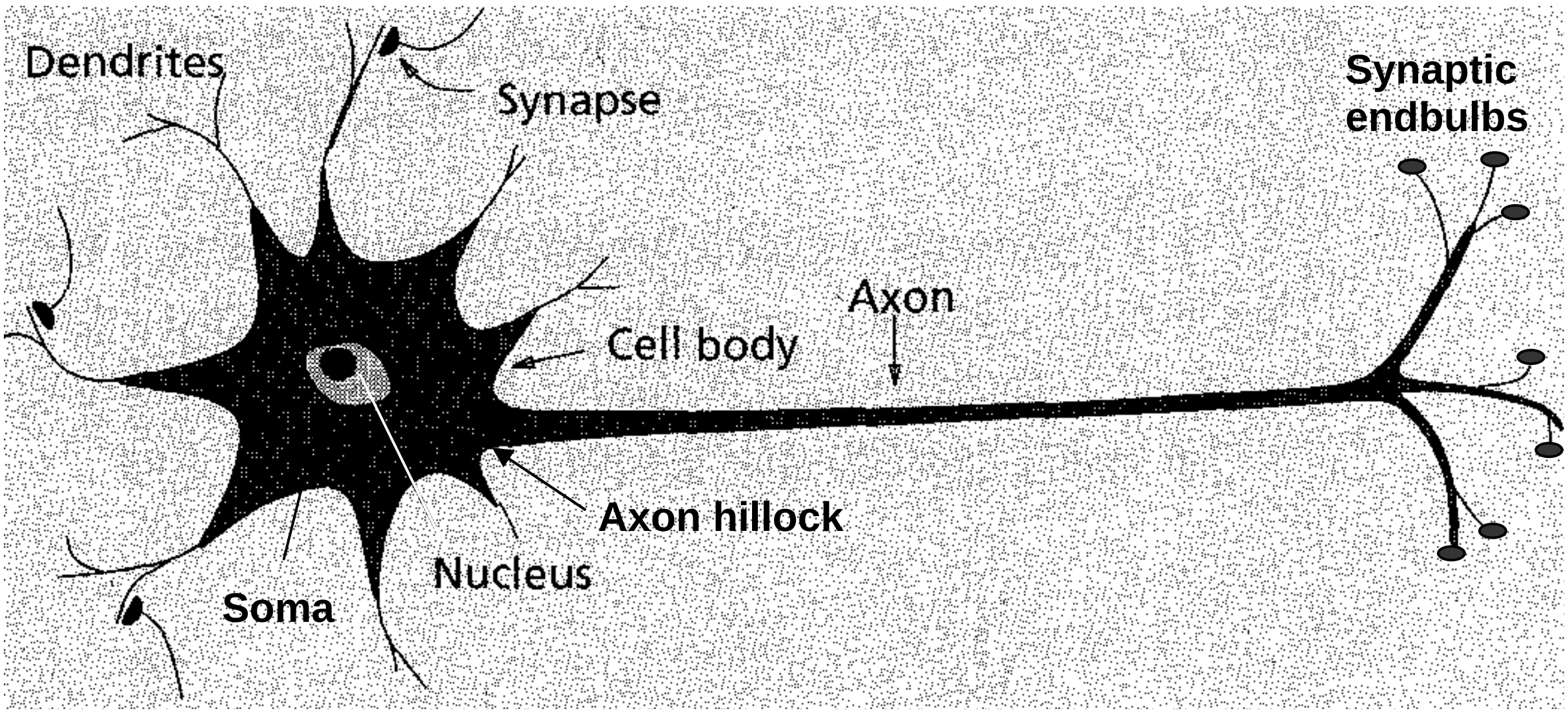

**Fig. 3. 3**: Sketch of a biological neuron. The synaptic endbulbs are in charge of transmitting the signal emitted by the neuron to the neurons it is connected to, through the synapses [42].

[28] Neuron: *Any of the impulse-conducting cells that constitute the brain, spinal column, and nerves, consisting of a nucleated cell body with one or more dendrites and a single axon* [23]. It is also called nerve cell, or brain cell.

[29] Synapse: *Junction across which a nerve impulse passes from an axon terminal to a neuron, muscle cell, or gland cell* [23].

The neurons are connected to each other in an extremely complex way. The neural network cannot be thought of as a simple stimulus-response chain from sensors to effectors. The continuously changing interconnections shape all kinds of internal structures, distinguished by feed-back processes, so that the output depends on both the stimuli and the present state of the network. When an input is received, the new signals are mixed up with the remaining signals from previous experiences that are still travelling through some of the numberless feed-back cycles of the network. Thus, the human brain's network has memory of past experiences. Furthermore, the connections and their strengths are not stationary!

When a neuron fires, a positive or negative charge is sent to the dendrites connected to its synaptic endbulbs[30]. The total strength of the signal received by a certain neuron consists of the spatial summation of all the weak charges received by its dendrites, coming from different neurons. A temporal summation also takes place, through which a rapid series of weak signals is converted into a large one (accumulated in the axon hillock). Then, the aggregate signal is passed to the soma, and only if it is greater than a threshold, the neuron fires sending a signal through the axon to the synaptic endbulbs. The strength of the emitted signal is constant, regardless how much greater than the threshold the aggregate input is, and regardless of the axon's divisions. The strength of the emitted signal is the same in every synaptic endbulb.

The aim of ANNs is to bring together the abilities of human brains and computers by trying to mimic in a very simplified manner some aspects of the information processing in biological brains, thus bringing computers closer to brain's capabilities, while still keeping those desired features they already have.

## 3.5.2 Biological brains versus conventional computers

The brain's network of neurons forms a massively parallel information processing system, as opposed to conventional computers, in which a single processor executes a single series of instructions. A few clear differences between biological brains and conventional computers are summarized in **Table 3. 1**.

[30] The detailed biological process is much more complex, involving processes such as the signal transmission from the synaptic endbulbs to the dendrites occurring by means of transmitter molecules that diffuse across a very small synaptic cleft, or some of the signal transmissions being rather chemical than electrical, etc. As previously mentioned, this is not a thorough analysis of the functioning of the human brain!

In spite of being equipped with such a slow processor, the massively parallel interconnections of very simple PEs gives the biological brains the capability of overcoming some difficulties and outperform conventional computers in several tasks such as recognizing a face among a crowd or from an angle never encountered before, handwriting recognition despite its many particularities never seen before, crossing a street without calculating distances and velocities, or even reading a text which simply contains many mistakes. Biological brains can easily accept noisy data and even partial damage, thanks to their redundant connections.

| | von Neumann machines (conventional computers) | Biological Neural System |
|---|---|---|
| Processor | Complex<br>High Speed: $10^9$ Hz<br>One or a few | Simple<br>Low Speed: 100 Hz<br>A large number |
| Computing | Centralized<br>Sequential<br>Stored programs | Distributed<br>Parallel<br>Self-learning |
| Reliability | Very vulnerable | Robust |
| Fault tolerance | No | Yes |
| Expertise | Numerical and symbolic manipulations | Perceptual problems |
| Operating environment | Well-defined and well-constrained | Poorly-defined and unconstrained |

**Table 3. 1**: Comparison between the main features of biological brains and conventional computers.

## 3.5.3 Artificial neural networks versus biological brains

The complexity of the brain's behaviour is due to the huge number of neurons in addition to the complex structure of their interconnections. The reason why it appears to be so difficult, maybe impossible, to simulate biological brains by means of ANNs is that the design of the architecture of the network is not trivial, being perhaps impossible to define which node should be connected to which, the weights of the connections, and the thresholds' settings.

Furthermore, neither the weights, nor the connections, nor the thresholds are stationary but continuously changing in a biological brain. Notice that the non-stationary nature of the connections between neurons is encompassed in the changing weights, since a weight equal to zero is equivalent to not being connected at all, and the change in a weight's sign implies the swap of the nature of the signal (from excitatory to inhibitory, and the other way round).

However, although far from simulating a biological brain, a brain-like designed ANN yet performs well when dealing with perceptual tasks, very much like biological brains, therefore complementing conventional computers.

### 3.5.4 Artificial neural networks versus conventional computers

The symbolic paradigm consists of indivisible entities (symbols) which can give shape to more complex entities by means of simple rules. Of course once the symbols and the rules are known, this is a very powerful method, but defining them can be extremely difficult and tedious. For instance, to define a symbol, some other symbols and rules might be necessary[31].

In contrast, biological brains work in parallel, being able to recognize a symbol immediately without analyzing sub-symbols (for instance, you recognize a cow without counting its legs). This is because the brain works in parallel, and all the information is summed up.

*An artificial neural network is a computational paradigm that differs substantially from those based on the standard von Neumann architecture. Feature recognition ANNs generally learn from experience, instead of being expressly programmed with rules as in conventional artificial intelligence…* [64].

Conventional computers need to be told in a very precise manner the exact series of steps required to perform an algorithm, assuming it is possible to find an algorithm to describe the problem! Thus, they fetch an instruction and all the required data from memory, to later execute the instruction. They do not accept noisy data or memory damage.

However, perceptual tasks, for instance, are almost impossible to formalize by means of an algorithm. The astonishing wide diversity of associations biological brains can perform is absolutely out of reach for conventional computers.

ANNs were developed to overcome these limitations by attempting to incorporate biological brains' desirable features. Thus, information is stored in a set of weights instead of in a computer program.

Similarly to biological brains and in contrast to conventional computers, ANNs are robust when facing hardware damage or noisy data, mainly due to their redundant structure. *Partial destruction of the network leads to the corresponding degradation of performance. However, some network capabilities may be retained even with major network damage* [59]. Besides, ANNs are expected to deal with unseen patterns by generalizing from the training set.

---

[31] Recall the already explained modular structure of the symbolic systems, which have different levels of abstraction. As mentioned in [59], a cow might be a symbol, but defining a cow is not trivial. Although it surely can be done by using symbols like legs, udders, etc., handling perceptual problems soon becomes tedious or even unapproachable by the symbolic paradigm.

To summarize, ANNs are good at perceptual tasks, which are precisely the kinds of problems which conventional symbolic paradigm fails in dealing with, thus complementing each other.

## 3.5.5 Artificial neural networks' mathematical issues

The neurons in biological neural networks are represented in ANNs by relatively simple PEs called nodes. Sometimes the change of denominations is avoided, and the nodes are simply referred to as neurons, the ANNs as neural networks, and so on. Form here forth, the biological terms and their artificial counterparts are used indistinctly.

Thus, each neuron can be thought of as a node, and a connection linking two neurons as an edge, which has a weight associated representing the strength of the interaction between the two neurons at issue. Further, the sign of the weight represents whether the influence of a neuron over the other is stimulatory or inhibitory, and a weight equal to zero represents the break of the connection.

Therefore, a single biological neuron could be grossly modelled as shown in **Fig. 3. 4**.

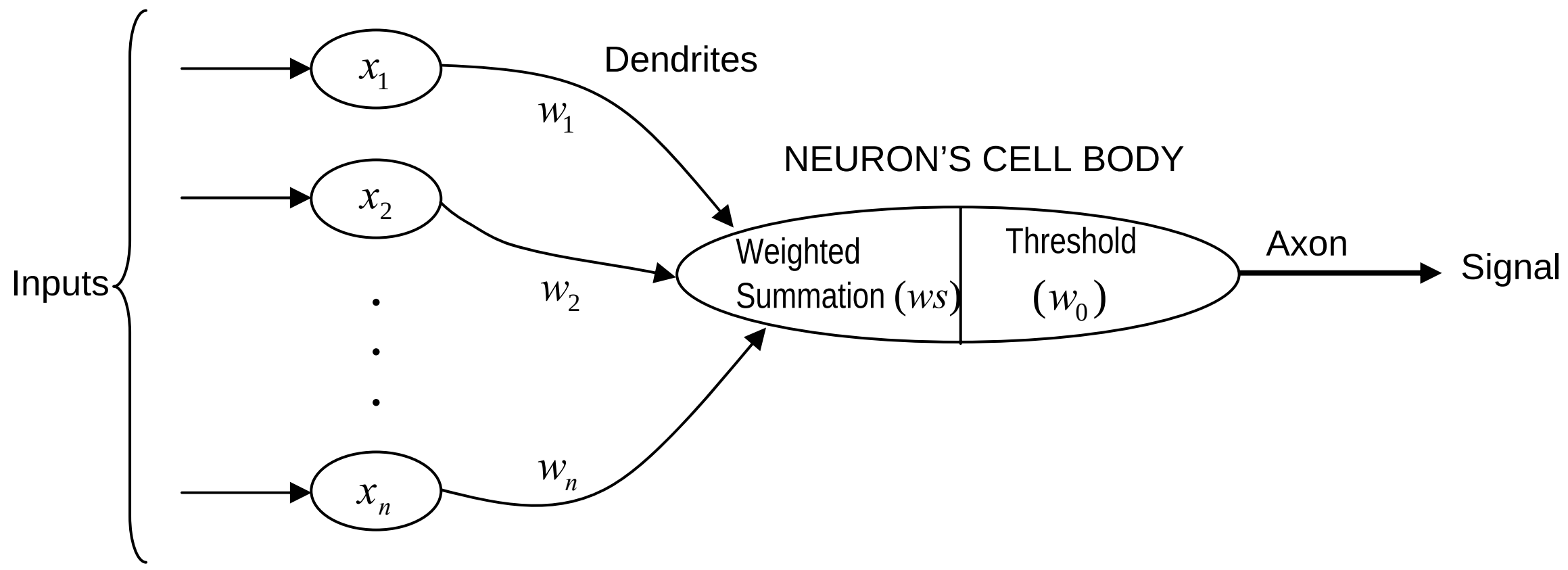

**Fig. 3. 4**: A single neuron model. The neuron's activation function consists of a threshold that has to be overcome by the weighted summation of its inputs in order to fire.

### 3.5.5.1 The artificial neuron

Obviously, the simplest ANN possible is composed of a single artificial neuron. The first artificial neuron was developed based on the model of the biological one (see **Fig. 3. 4**), and was called "perceptron" (see **Fig. 3. 5**).

### 3.5.5.1.1 The perceptron

The $n$ inputs $x_i$ to a perceptron $j$ are $n$ signals coming from $n$ synapses whose strengths are represented by $n$ weights $w_{ji}$. The weighted signals are summed up to produce the aggregate signal that the perceptron receives $ws_j$, as stated in **(3. 1)**. If and only if the summation reaches the threshold $w_{j0}$, the neuron fires (i.e. $y_j = 1$). Note that the weight $w_{j0}$ can be thought of as an actual threshold, or as a weight corresponding to an input $x_0 = -1$.

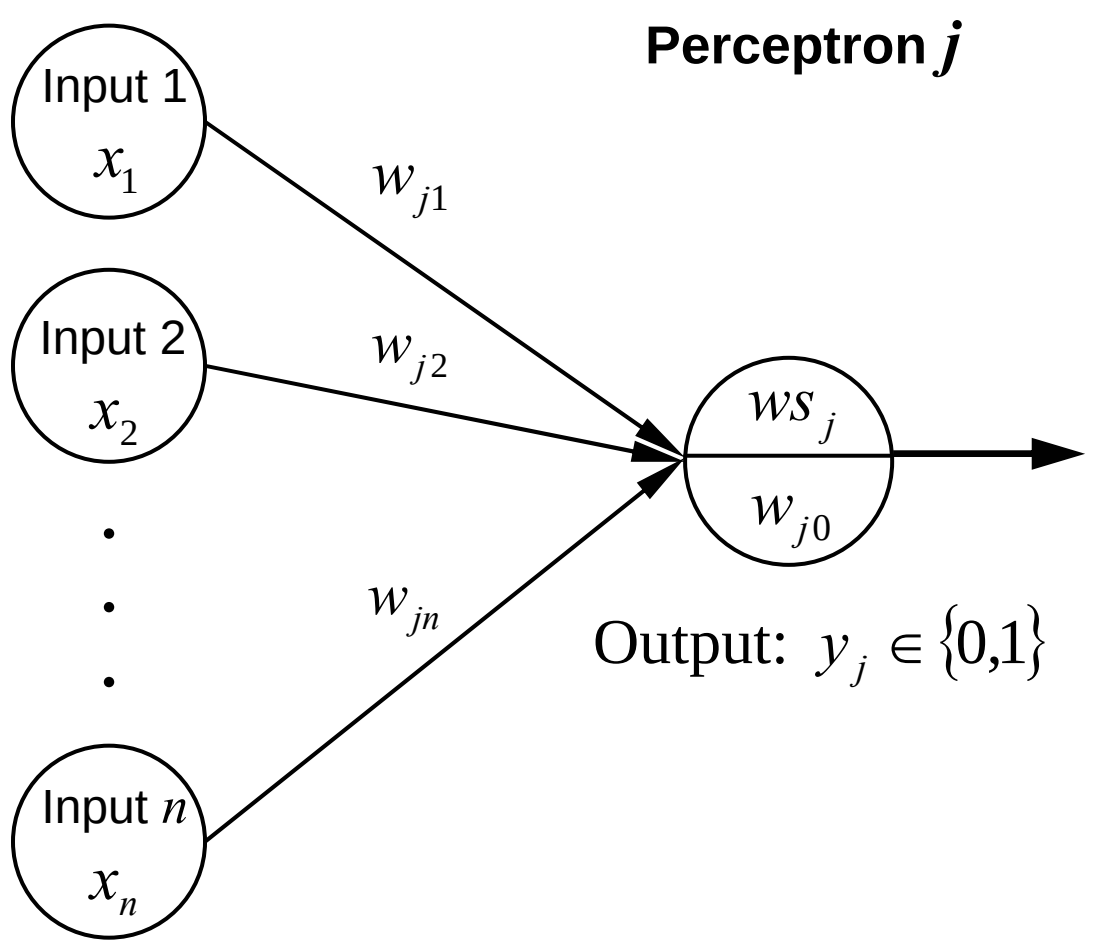

**Fig. 3. 5**: The perceptron.

$$\left. \begin{aligned} ws_j &= \sum_{i=1}^{n} w_{ji} \cdot x_i \\ lc_j &= ws_j - w_{j0} = \sum_{i=0}^{n} w_{ji} \cdot x_i \end{aligned} \right\} \; ; \; \text{if} \; \begin{cases} lc_j \geq 0 \quad \text{or} \quad ws_j \geq w_{j0} \quad \Rightarrow \quad y_j = 1 \\ lc_j < 0 \quad \text{or} \quad ws_j < w_{j0} \quad \Rightarrow \quad y_j = 0 \end{cases} \qquad \textbf{(3. 1)}$$

Where:

- $ws_j$: weighted summation
- $lc_j$: linear combination
- $w_{j0}$: neuron $j$'s threshold
- $x_0 = -1$

Therefore, a perceptron is an artificial neuron composed of an "adder", which generates the aggregate input signal, and by a "linear threshold activation function" (see **Fig. 3. 6**), which is said to be linear because it separates two "classes" by a hyper-plane, whose position within the hyper-space containing all the patterns needs to be learned. Hence a perceptron can be seen as a "binary classifier" that maps its vector of inputs into a scalar, whose sign is used to classify the input pattern into one of the two learned classes.

In order to facilitate the manipulation of the mathematical expressions, from here forth, unless specifically stated otherwise, $w_{j0}$ will be considered a component of the vector of weights rather than a threshold, and $x_0 = -1$ will be included into the vector of inputs. Thus,

$$\mathbf{w}_j = \{w_{j0} \quad w_{j1} \quad w_{j2} \quad \cdots \quad w_{jn}\} \quad \text{and} \quad \mathbf{x} = \begin{Bmatrix} x_0 = -1 \\ x_1 \\ x_2 \\ \vdots \\ x_n \end{Bmatrix} \quad \Rightarrow \quad lc_j = \mathbf{w}_j \cdot \mathbf{x} \tag{3. 2}$$

Although it has been claimed that ANNs are universal function approximators, the first immediate application of a perceptron is to binary classify patterns, so that the discussion is initially focused here on classification problems. Notice that in order to perform such a task, the patterns presented to the perceptron need to be first recognized, for which they must be represented by some kind of description[32]. *The most frequently used type of pattern description is the description by a set of features… Each pattern is described by its respective values for each of the features… …the patterns can be regarded as points in the n-dimensional Euclidean space* [39].

If the features' values (i.e. the variables of the $n$-dimensional Euclidean space) are thought of as a perceptron $j$'s inputs, its threshold activation function divides the features hyper-space in two parts. Thus, the patterns can be classified as belonging to one or the other of the two sub-hyper-spaces separated by a hyper-plane, whose equation is $lc_j = 0$.

For example, for the very simple case of two features, the threshold function states:

$$w_{j1} \cdot x_1 + w_{j2} \cdot x_2 \geq w_{j0} \tag{3. 3}$$

If equation **(3. 3)** is true, the pattern belongs to one class $(y_j = 1)$. Otherwise, it belongs to the other $(y_j = 0)$. In this example, the two classes are separated by a straight line (see **Fig. 3. 6**).

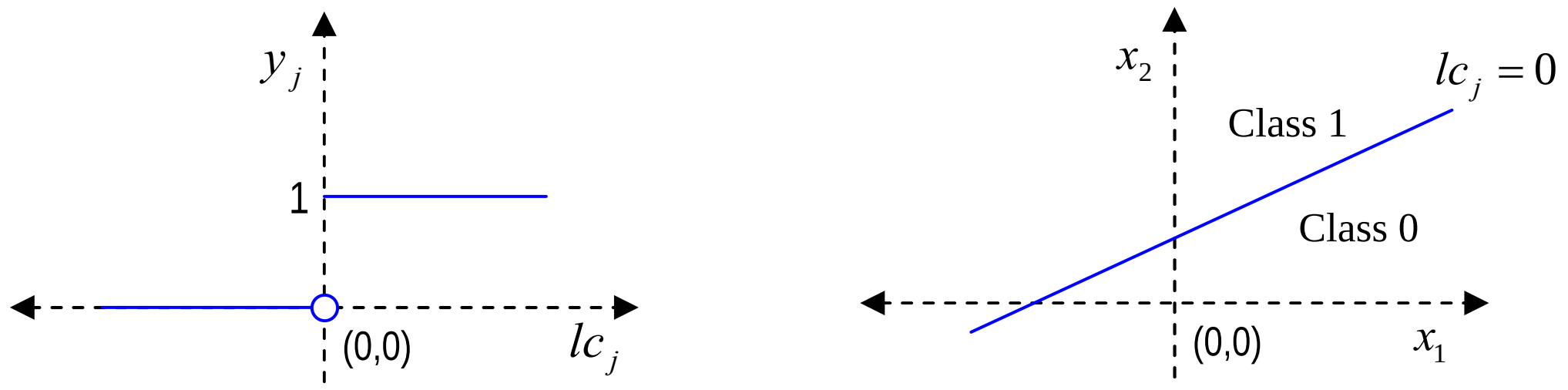

**Fig. 3. 6**: Left: Perceptron's "linear threshold activation function" (or "step transfer function"). If the linear combination of the features is greater than zero, the pattern belongs to one class; otherwise, it belongs to the other.

Right: Two-dimensional features space. The two classes are separated by a straight line, which is said to be the "separation function" or the "decision boundary".

[32] Pattern recognition problems encompass tasks such as interpreting visual images or recognizing voices.

However, the position of the hyper-plane that separates the two classes is defined by the perceptron's weights, which are initially unknown.

In order to train a perceptron $j$, a training data set consisting of patterns whose classification is known in advance is offered. The error of the neuron is given by the difference between the desired output and the one that the perceptron returns. To correct this error, its weights need to be adjusted. Therefore, the perceptron $j$ is offered a set of $s$ input vectors $\left(\mathbf{x}^{(k)}\right)$ and their corresponding desired outputs $\left(t^{(k)} \in \{0,1\}\right)$: $\left\{\left[\mathbf{x}^{(1)},t^{(1)}\right],\left[\mathbf{x}^{(2)},t^{(2)}\right], \ldots ,\left[\mathbf{x}^{(s)},t^{(s)}\right]\right\}$.

The response of the perceptron $j$ when the $k^{\text{th}}$ pattern is offered is shown in equation **(3. 4)**.

$$\begin{aligned} y_j^{(k)} &= 1 \quad \text{if} \quad lc_j^{(k)} = \mathbf{w}_j^{(k)} \cdot \mathbf{x}^{(k)} \geq 0 \\ y_j^{(k)} &= 0 \quad \text{if} \quad lc_j^{(k)} = \mathbf{w}_j^{(k)} \cdot \mathbf{x}^{(k)} < 0 \end{aligned} \tag{3. 4}$$

Then, $\delta_j^{(k)} = \left(t_j^{(k)} - y_j^{(k)}\right) \in \{-1,0,1\}$ is the error of the neuron $j$ when the $k^{\text{th}}$ training pattern is offered, which has to be corrected by adjusting the parameters of the system according to a learning algorithm[33].

The so-called ***δ*-rule** establishes that the correction of each weight $\left(\Delta w_{ji}^{(k)}\right)$ is proportional to the error $\left(\delta_j^{(k)}\right)$, and to the value of the corresponding input $\left(x_i^{(k)}\right)$, as shown in equation **(3. 5)**.

$$\begin{aligned} &\Delta w_{ji}^{(k)} = \eta \cdot \delta_j^{(k)} \cdot x_i^{(k)} = \eta \cdot \left(t_j^{(k)} - y_j^{(k)}\right) \cdot x_i^{(k)} \\ &w_{ji}^{(k+1)} = w_{ji}^{(k)} + \Delta w_{ji}^{(k)} \quad \Rightarrow \quad \mathbf{w}_j^{(k+1)} = \mathbf{w}_j^{(k)} + \mathbf{\Delta w}_j^{(k+1)} \\ &i = 0, 1, 2, \ldots, n \end{aligned} \tag{3. 5}$$

Where $\mathbf{\Delta w}_j^{(k)} \in \left\{-\mathbf{x}^{(k)}, \mathbf{0}, \mathbf{x}^{(k)}\right\}$, and $\eta$ is a constant greater than zero that handles the step sizes of the weights' updates.

The corrections of the weights are made proportional to the input because it is assumed that the bigger the input the greater its influence on the overall error.

The whole training procedure, which is proven to be capable of finding a separating hyper-plane if one exists, is as follows:

1- Initialize all the weights to zero $\left(\mathbf{w}_j^{(k)} = \mathbf{0}\right)$.

[33] Since the perceptron learns from a training data set, it is said that it learns from experience.

2- Select a pattern ($k$) from the training data set and classify it.

3- While the stopping criterion is not attained, update the weights according to **(3. 5)**. Notice that if the classification is correct, $\Delta \mathbf{w}_j^{(k)} = \mathbf{0}$.

4- **If** all the patterns have not been analyzed, **then** select the next pattern, classify it, and go back to step 3.

Some of the simple successful applications of this perceptron are the computation of the logical operations "and", "or" and "not". Consider, for instance, the logical operation "and". It is easy to realize by common sense that one solution could be the one shown in **Fig. 3. 7**, but common sense is not enough for bigger problems, and the perceptron needs to learn its weights. In order to show how the ***δ*-rule** works, the problem is solved in details hereafter, for the case of two inputs only. The training data set and a possible solution are shown in **Fig. 3. 8**. The problem has been solved for $\eta$ = 1, and all the partial results are gathered in **Table 3. 2**.

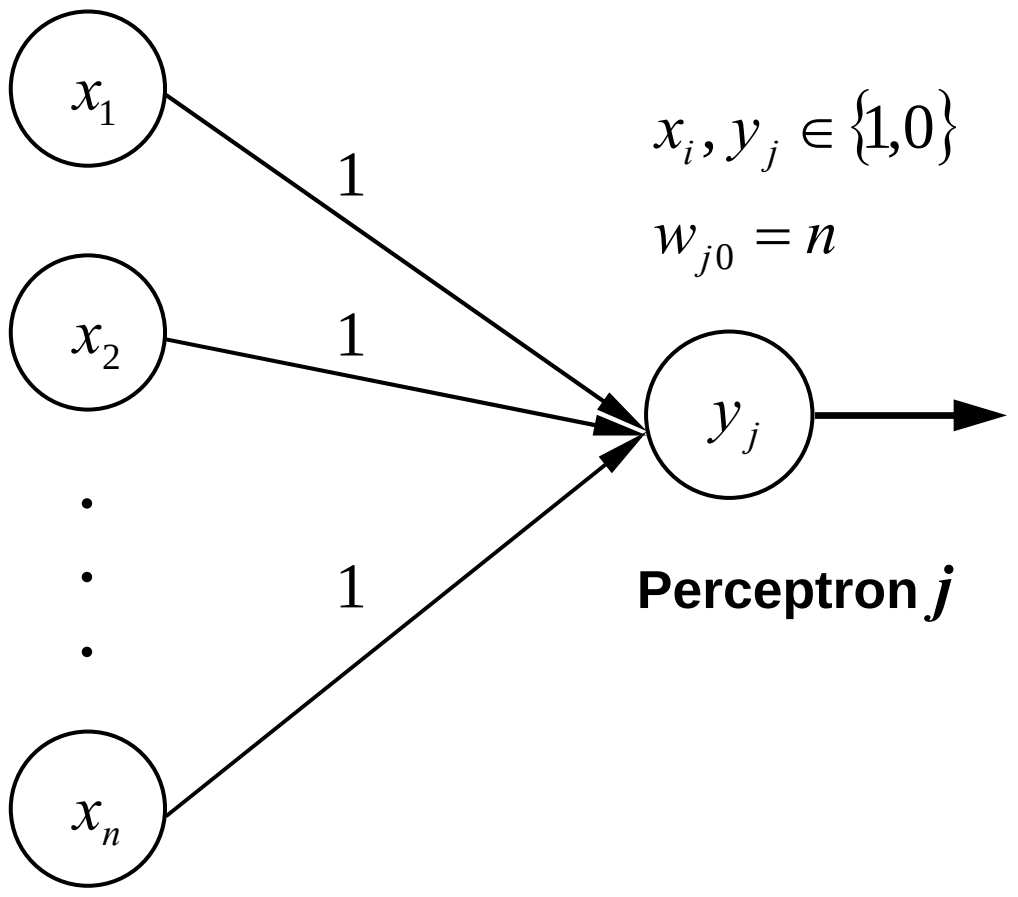

**Fig. 3. 7**: Computation of the logical operation "and" by means of a perceptron $j$.

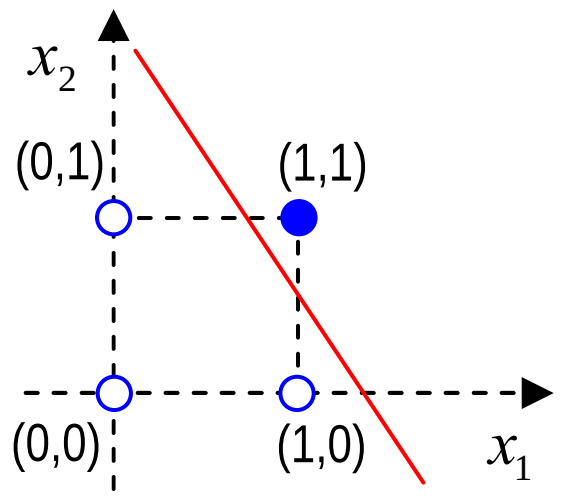

| $x_1$ | $x_2$ | $t$ (target) |
|---|---|---|
| 1 | 0 | 0 |
| 0 | 1 | 0 |
| 1 | 1 | 1 |
| 0 | 0 | 0 |

**Fig. 3. 8**: The logical operation "and". If solved by a perceptron, any line like the red one is a solution.

It is fair to remark that this is a very simple problem whose training patterns do not present noise. In many real life problems the training patterns are noisy and conflicting, so that there is no set of weights that reduces the error to zero. Then, the goal is to minimize it instead. The logical operations "or" (common sense suggests the weights $w_{j0} = w_{j1} = w_{j2} = 1$) and "not" (common sense suggests $w_{j0} = 0$ and $w_{j1} = -1$) can be solved in the same fashion.

| Pattern | $x_0$ | $x_1$ | $x_2$ | $w_{j0}$ | $w_{j1}$ | $w_{j2}$ | $t_j$ | $lc_j$ | $y_j$ | $\delta_j$ | $\Delta w_{j0}$ | $\Delta w_{j1}$ | $\Delta w_{j2}$ |
|---|---|---|---|---|---|---|---|---|---|---|---|---|---|
| 1 | -1 | 1.00 | 0.00 | 0.00 | 0.00 | 0.00 | 0.00 | 0.00 | 1.00 | -1.00 | 1.00 | -1.00 | 0.00 |
| 2 | -1 | 0.00 | 1.00 | 1.00 | -1.00 | 0.00 | 0.00 | -1.00 | 0.00 | 0.00 | 0.00 | 0.00 | 0.00 |
| 3 | -1 | 1.00 | 1.00 | 1.00 | -1.00 | 0.00 | 1.00 | -2.00 | 0.00 | 1.00 | -1.00 | 1.00 | 1.00 |
| 4 | -1 | 0.00 | 0.00 | 0.00 | 0.00 | 1.00 | 0.00 | 0.00 | 1.00 | -1.00 | 1.00 | 0.00 | 0.00 |
| 1 | -1 | 1.00 | 0.00 | 1.00 | 0.00 | 1.00 | 0.00 | -1.00 | 0.00 | 0.00 | 0.00 | 0.00 | 0.00 |
| 2 | -1 | 0.00 | 1.00 | 1.00 | 0.00 | 1.00 | 0.00 | 0.00 | 1.00 | -1.00 | 1.00 | 0.00 | -1.00 |
| 3 | -1 | 1.00 | 1.00 | 2.00 | 0.00 | 0.00 | 1.00 | -2.00 | 0.00 | 1.00 | -1.00 | 1.00 | 1.00 |
| 4 | -1 | 0.00 | 0.00 | 1.00 | 1.00 | 1.00 | 0.00 | -1.00 | 0.00 | 0.00 | 0.00 | 0.00 | 0.00 |
| 1 | -1 | 1.00 | 0.00 | 1.00 | 1.00 | 1.00 | 0.00 | 0.00 | 1.00 | -1.00 | 1.00 | -1.00 | 0.00 |
| 2 | -1 | 0.00 | 1.00 | 2.00 | 0.00 | 1.00 | 0.00 | -1.00 | 0.00 | 0.00 | 0.00 | 0.00 | 0.00 |
| 3 | -1 | 1.00 | 1.00 | 2.00 | 0.00 | 1.00 | 1.00 | -1.00 | 0.00 | 1.00 | -1.00 | 1.00 | 1.00 |
| 4 | -1 | 0.00 | 0.00 | 1.00 | 1.00 | 2.00 | 0.00 | -1.00 | 0.00 | 0.00 | 0.00 | 0.00 | 0.00 |
| 1 | -1 | 1.00 | 0.00 | 1.00 | 1.00 | 2.00 | 0.00 | 0.00 | 1.00 | -1.00 | 1.00 | -1.00 | 0.00 |
| 2 | -1 | 0.00 | 1.00 | 2.00 | 0.00 | 2.00 | 0.00 | 0.00 | 1.00 | -1.00 | 1.00 | 0.00 | -1.00 |
| 3 | -1 | 1.00 | 1.00 | 3.00 | 0.00 | 1.00 | 1.00 | -2.00 | 0.00 | 1.00 | -1.00 | 1.00 | 1.00 |
| 4 | -1 | 0.00 | 0.00 | 2.00 | 1.00 | 2.00 | 0.00 | -2.00 | 0.00 | 0.00 | 0.00 | 0.00 | 0.00 |
| 1 | -1 | 1.00 | 0.00 | 2.00 | 1.00 | 2.00 | 0.00 | -1.00 | 0.00 | 0.00 | 0.00 | 0.00 | 0.00 |
| 2 | -1 | 0.00 | 1.00 | 2.00 | 1.00 | 2.00 | 0.00 | 0.00 | 1.00 | -1.00 | 1.00 | 0.00 | -1.00 |
| 3 | -1 | 1.00 | 1.00 | 3.00 | 1.00 | 1.00 | 1.00 | -1.00 | 0.00 | 1.00 | -1.00 | 1.00 | 1.00 |
| 4 | -1 | 0.00 | 0.00 | 2.00 | 2.00 | 2.00 | 0.00 | -2.00 | 0.00 | 0.00 | 0.00 | 0.00 | 0.00 |
| 1 | -1 | 1.00 | 0.00 | 2.00 | 2.00 | 2.00 | 0.00 | 0.00 | 1.00 | -1.00 | 1.00 | -1.00 | 0.00 |
| 2 | -1 | 0.00 | 1.00 | 3.00 | 1.00 | 2.00 | 0.00 | -1.00 | 0.00 | 0.00 | 0.00 | 0.00 | 0.00 |
| 3 | -1 | 1.00 | 1.00 | 3.00 | 1.00 | 2.00 | 1.00 | 0.00 | 1.00 | 0.00 | 0.00 | 0.00 | 0.00 |
| 4 | -1 | 0.00 | 0.00 | 3.00 | 1.00 | 2.00 | 0.00 | -3.00 | 0.00 | 0.00 | 0.00 | 0.00 | 0.00 |
| 1 | -1 | 1.00 | 0.00 | 3.00 | 1.00 | 2.00 | 0.00 | -2.00 | 0.00 | 0.00 | 0.00 | 0.00 | 0.00 |
| 2 | -1 | 0.00 | 1.00 | 3.00 | 1.00 | 2.00 | 0.00 | -1.00 | 0.00 | 0.00 | 0.00 | 0.00 | 0.00 |
| 3 | -1 | 1.00 | 1.00 | 3.00 | 1.00 | 2.00 | 1.00 | 0.00 | 1.00 | 0.00 | 0.00 | 0.00 | 0.00 |
| 4 | -1 | 0.00 | 0.00 | 3.00 | 1.00 | 2.00 | 0.00 | -3.00 | 0.00 | 0.00 | 0.00 | 0.00 | 0.00 |

**Table 3. 2**: Process of updating the perceptron *j*'s weights to solve the logic operation "and". Notice that the last time the four patterns are presented, there is no error in their classification. Observe that the weights found here are not the same as the ones proposed in **Fig. 3. 7**.

Refer to **Appendix 1** for a slightly more complex case, where the inputs are real-valued.

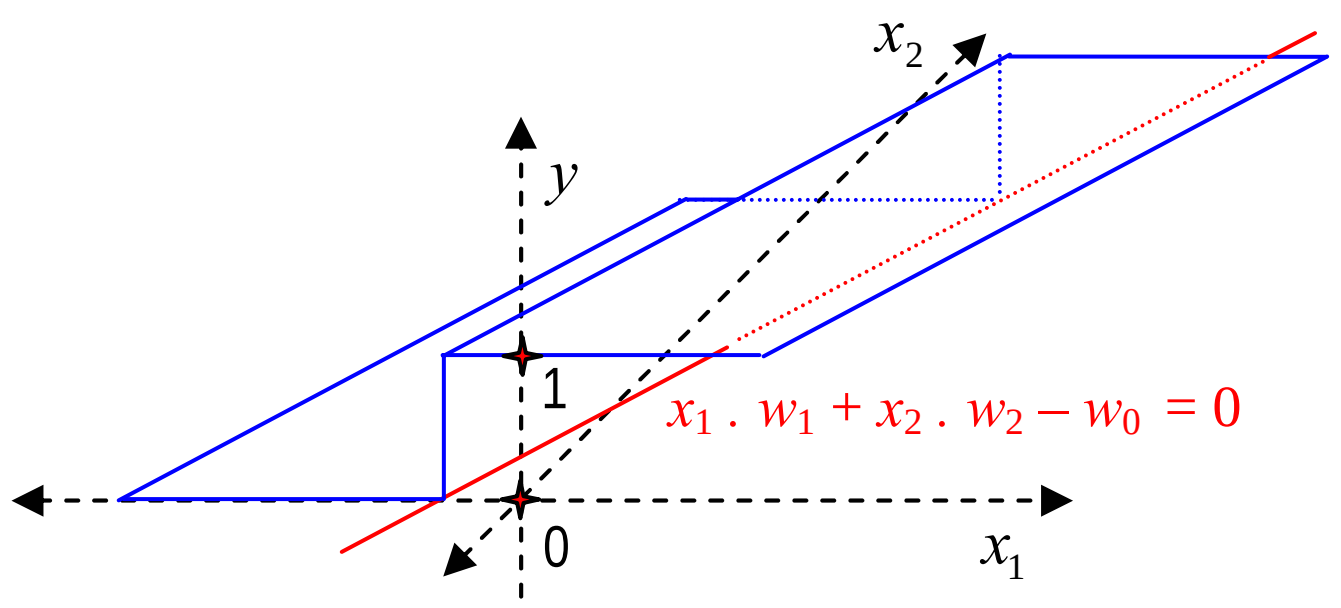

**Fig. 3. 9**: Perceptron's "linear threshold activation function" interpreted as an approximation function. The red line is the linear decision boundary, contained by the feature space (plane "$x_1, x_2$").

Learning on a perceptron is guaranteed. The "perceptron convergence theorem" asserts that if a perceptron is capable of finding a solution to a certain problem, the learning rule will find the solution in a finite number of steps. For proof of the theorem, refer to: **Perceptrons**, by Minsky and Papert (1969).

It has been argued before that ANNs can be thought of as universal function approximators, but only classification tasks have been discussed so far. However, each training pattern is nothing but a set of inputs with its corresponding output, so that this could be seen as a function approximation, as shown in **Fig. 3. 9**. Notice that the approximated function is the doorstep-like surface, and not the linear separation function.

### 3.5.5.1.2 The linear artificial neuron

Although this is the simplest artificial neuron, it is very useful in practice. For instance, it is typically used in the output layers of ANNs that are aimed at function approximation, since their output value does not have restrictions. Like the perceptron, this artificial neuron (see **Fig. 3. 10**) is composed of an "adder" that sums up the $n$ weighted inputs to produce the aggregate signal, and of a linear so-called "transfer function" (see **Fig. 3. 10**, **Fig. 3. 11**) instead of the "linear threshold function". Typically, the linear transfer function consists of multiplying the linear combination of the inputs by one, so that the linear artificial neuron simply maps an input vector into a scalar by means of a linear combination. Thus, the linear artificial neuron can be seen as a "function approximator" (see **Fig. 3. 11**), as opposed to the perceptron, which was viewed as a "classifier" (see **Fig. 3. 6**).

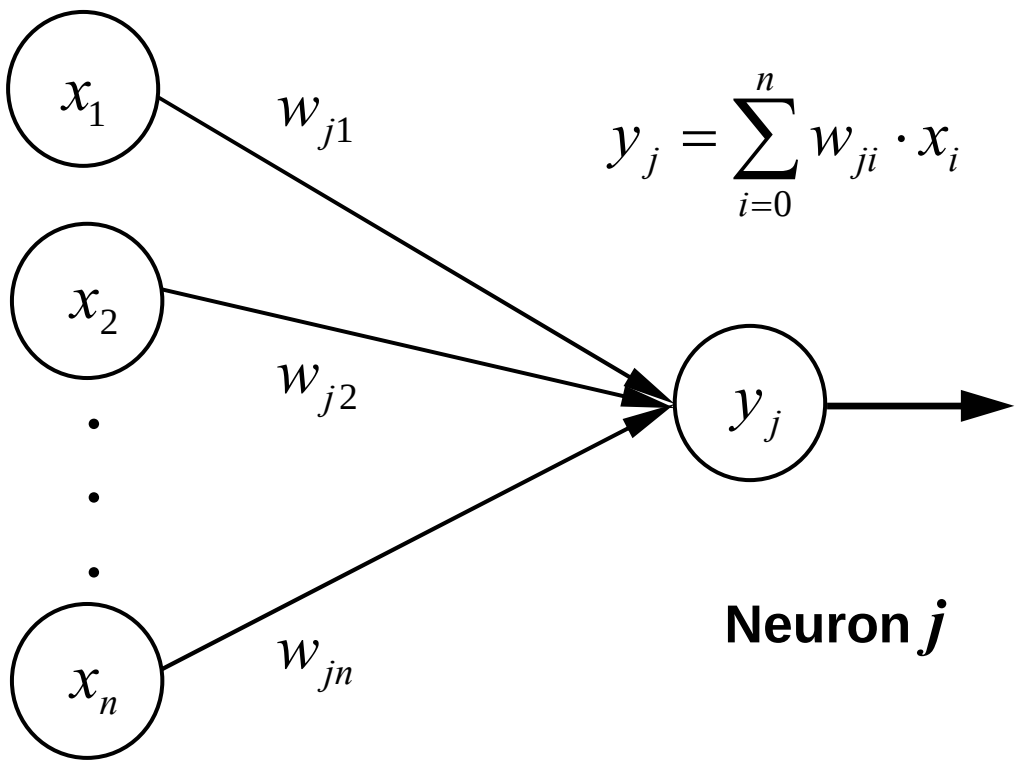

**Fig. 3. 10**: The linear artificial neuron.

Comparing to the view of the perceptron as an approximator (**Fig. 3. 9**), the linear artificial neuron approximates a hyper-plane instead of a "hyper-doorstep" (see **Fig. 3. 11**, right).

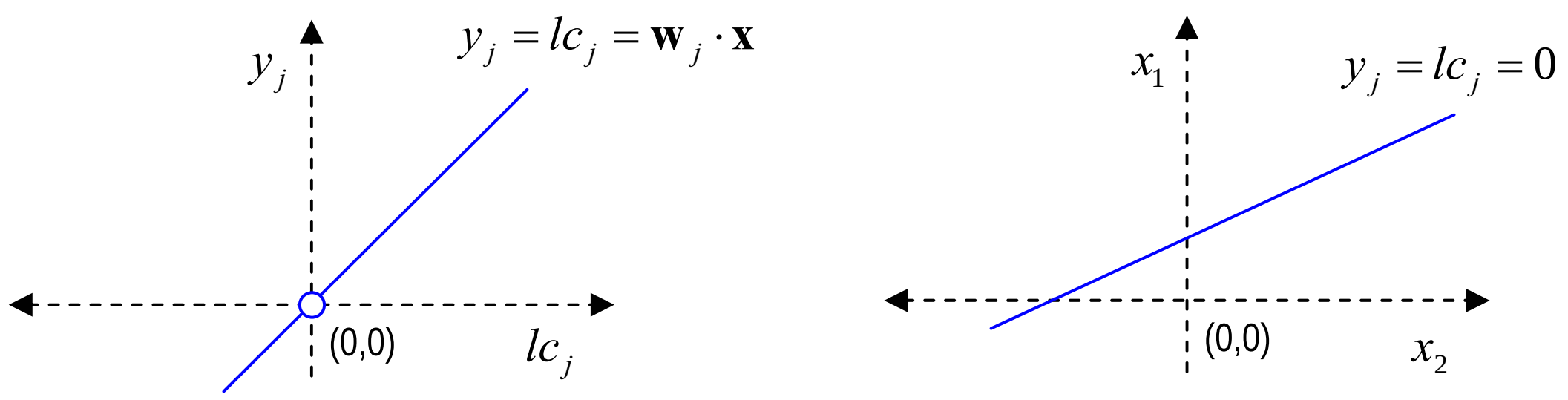

**Fig. 3. 11**: Left: Transfer function of a linear artificial neuron $j$.

Right: Two-dimensional features space. The blue line is not a decision boundary but the trace of the approximation function (plane $y_j = 0$) on the plane "$x_1, x_2$".

The hyper-plane that approximates the patterns has to be positioned in the hyper-space by setting the weights, which are initially unknown. Therefore the neuron needs to be trained. Hence a neuron $j$ is offered a set of $s$ input vectors $\left(\mathbf{x}_j^{(k)} \in \mathcal{R}^n\right)$, and their corresponding desired outputs $\left(t_j^{(k)} \in \mathcal{R}\right)$: $\left\{\left[\mathbf{x}^{(1)}, t^{(1)}\right], \left[\mathbf{x}^{(2)}, t^{(2)}\right], \dots, \left[\mathbf{x}^{(s)}, t^{(s)}\right]\right\}$.

The neuron's weights have to be adjusted to fit the patterns. The error of the neuron $j$ when the $k^{\text{th}}$ pattern is offered is given by the difference between the desired output and the one that the neuron returns $\left(\delta_j^{(k)} = t_j^{(k)} - y_j^{(k)}\right)$. The response of the neuron $j$ when the $k^{\text{th}}$ pattern is offered is given by equation **(3. 6)**.

$$y_j^{(k)} = \mathbf{w}_j^{(k)} \cdot \mathbf{x}^{(k)} \qquad \textbf{(3. 6)}$$

Where $n$ is the number of inputs to the neuron.

For each training set, the error could be deterministically corrected so that the neuron's mistake is overcome in one step (absolute correction rule). Thus, the new weights could be:

$$\breve{\mathbf{w}}_j^{(k)} = \mathbf{w}_j^{(k)} + \frac{\delta_j^{(k)}}{\sum_{i=0}^{n} x_i}, \text{ so that } t_j^{(k)} = \breve{\mathbf{w}}_j^{(k)} \cdot \mathbf{x}^{(k)} \qquad \textbf{(3. 7)}$$

Where $\breve{\mathbf{w}}_j^{(k)}$ is a local solution vector of weights for the neuron $j$ and the pattern ($k$) only.

This would be one of the infinite possible solutions for the pattern ($k$), but when the network is offered the next pattern ($k$+1), it forgets the previous learning. Instead, the ***δ*-rule** proposes to update the weights of neuron $j$ when pattern ($k$) is offered as stated by equation **(3. 8)**:

$$\begin{aligned} &\Delta w_{ji}^{(k)} = \eta \cdot \delta_j^{(k)} \cdot x_i^{(k)} = \eta \cdot \left(t_j^{(k)} - y_j^{(k)}\right) \cdot x_i^{(k)} \\ &w_{ji}^{(k+1)} = w_{ji}^{(k)} + \Delta w_{ji}^{(k)} \quad \Rightarrow \quad \mathbf{w}_j^{(k+1)} = \mathbf{w}_j^{(k)} + \mathbf{\Delta w}_j^{(k)} \\ &i = 0, 1, 2, \dots, n \end{aligned} \qquad \textbf{(3. 8)}$$

Notice that $\delta_j^{(k)}$ and $\Delta w_{ji}^{(k)}$ can now take any value. The parameter $\eta$ determines the learning speed. In this case the mistake is not corrected immediately, but a repeated application of the method, with a small $\eta$, changes the weights slowly until the mistake is either overcome or minimized. However the method converges only to a local optimum. A linear artificial neuron with the correction rule embedded in it is called ADALINE (ADAptive LINear Element).

### 3.5.5.1.3 The nonlinear artificial neuron

Both linear threshold activation functions and linear transfer functions are too limited.

When dealing with classification problems, for instance, the distribution of the patterns does not always allow the separation of the classes by a linear function, either due to noise in the available features' values or simply due to more complex phenomena. A simple means of dealing with this situation consists of using stepwise linear separation functions (**Fig. 3. 12**). Another alternative is to replace the linear by a more appropriate non-linear separation function.

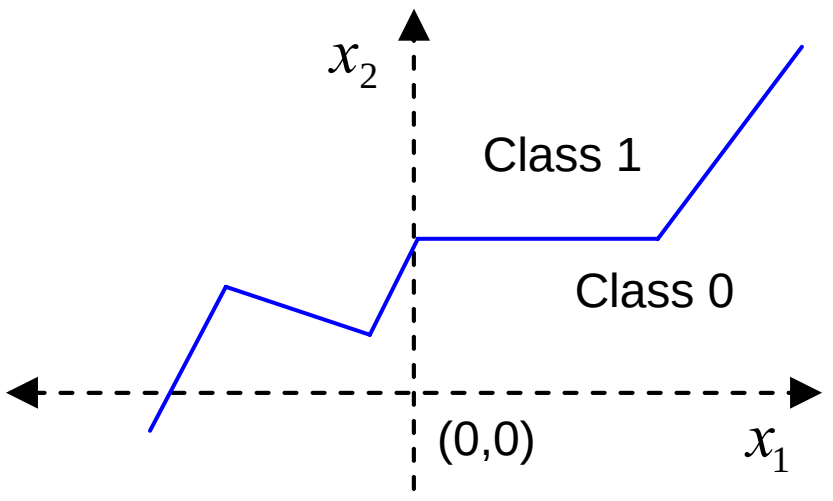

**Fig. 3. 12**: Example of a stepwise linear separation function.

When dealing with approximation tasks, the capabilities of the linear artificial neurons are limited to approximating hyper-planes (**Fig. 3. 11**). To extend the capabilities to approximating more complex functions, the nonlinear artificial neurons are composed of an "adder" that generates the aggregate signal, plus a non-linear transfer function. Very popular and useful nonlinear transfer functions are sigmoid functions (**Fig. 3. 13**), which have the feature of mapping the inputs into the interval (0, 1) or (-1, 1):

$$y = \frac{1}{1+e^{-lc}} \qquad \text{or} \qquad y = \tanh(lc) = \frac{2}{1+e^{-2\cdot lc}} - 1 \tag{3. 9}$$

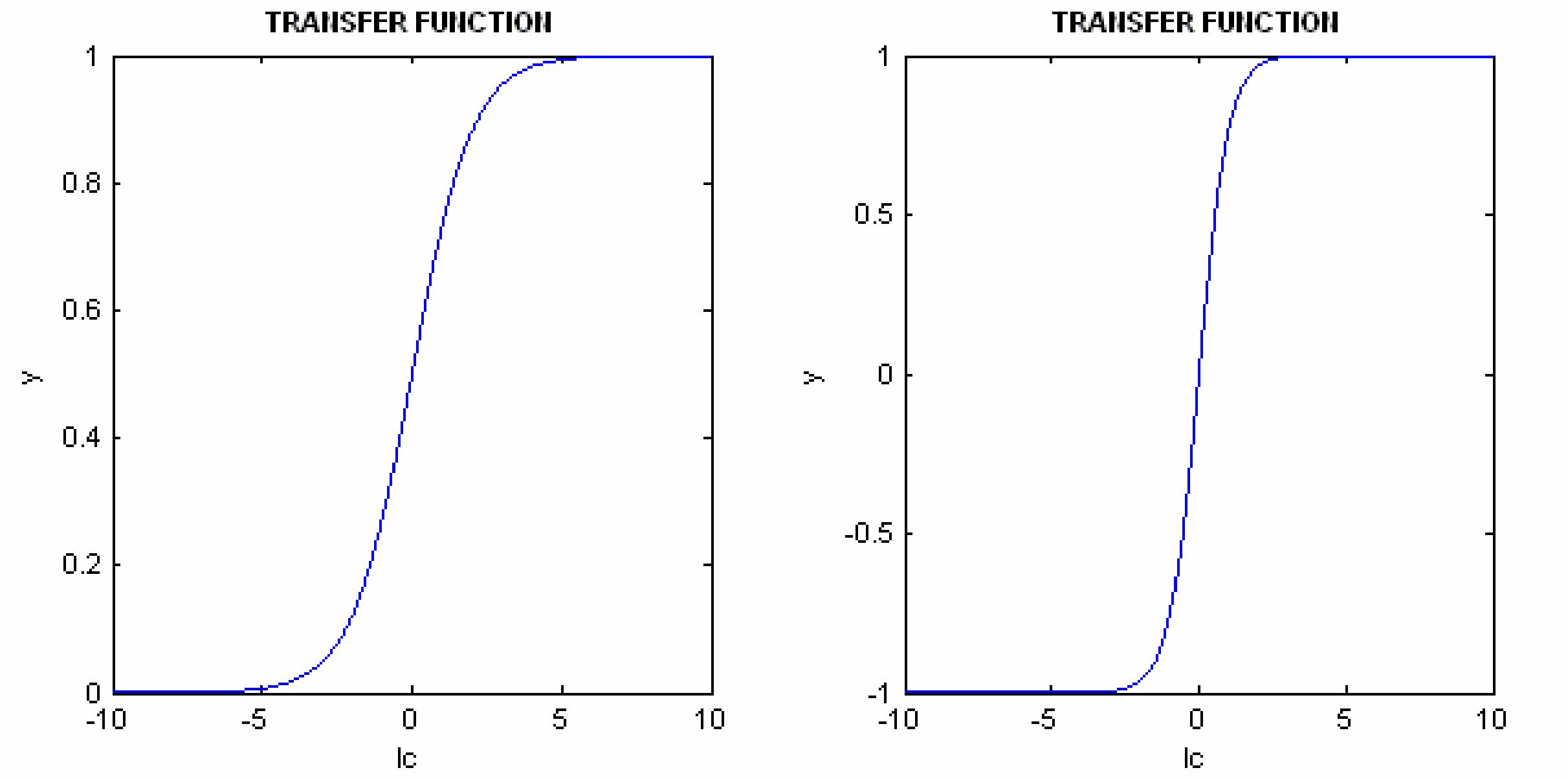

**Fig. 3. 13**: Two commonly used sigmoid functions. The graph on the left corresponds to the first, and the one on the right to the second of the equations **(3. 9)**.

In order to train this neuron, a set of $s$ input vectors $\mathbf{x}^{(k)}$ (patterns) and their corresponding desired outputs $t_j^{(k)}$ are offered to the neuron: $\left\{\left[\mathbf{x}^{(1)},t^{(1)}\right],\left[\mathbf{x}^{(2)},t^{(2)}\right],\ \ldots\ ,\left[\mathbf{x}^{(s)},t^{(s)}\right]\right\}$. The actual response of a neuron $j$ when the $k^{\text{th}}$ pattern is offered is given by equation **(3. 10)**.

$$y_j^{(k)} = f_j\left(\mathbf{w}_j^{(k)} \cdot \mathbf{x}^{(k)}\right) \tag{3. 10}$$

Where $n$ is the number of inputs to the neuron.

The error in the output of a neuron $j$ when the $k^{\text{th}}$ pattern is offered is $\left(t_j^{(k)} - y_j^{(k)}\right)$. The aim is to adjust the weights so as to minimize the error $e_j$ stated in equation **(3. 11)**. Beware that in general, $\delta_j^{(k)} \neq \left(t_j^{(k)} - y_j^{(k)}\right)$ for the nonlinear neuron, as shown in equation **(3. 13)**.

$$e_j = \frac{1}{s}\cdot\sum_{k=1}^{s} e_j^{(k)} \quad ; \quad e_j^{(k)} = \frac{\left(t_j^{(k)} - y_j^{(k)}\right)^2}{2} \tag{3. 11}$$

Where $s$ is the number of patterns.

Note that $e_j^{(k)}$ is a measure of error for the neuron $j$ when receiving the pattern ($k$).

If equation **(3. 10)** is inserted into equation **(3. 11)**, the error is a function of the weights, so that a **gradient-descent** strategy is used to learn the weights that minimize the error function:

$$w_{ji}^{(k+1)} = w_{ji}^{(k)} - \eta\cdot\frac{\partial e_j^{(k)}}{\partial w_{ji}^{(k)}} \tag{3. 12}$$

If $\frac{\partial e_j^{(k)}}{\partial w_{ji}^{(k)}}$ is equal to zero, the weight $w_{ji}^{(k+1)}$ (connecting neuron $j$ with its $i^{\text{th}}$ input) does not need to be adjusted. The sign (-) in equation **(3. 12)** is due to the fact that if $\frac{\partial e_j^{(k)}}{\partial w_{ji}^{(k)}} > 0$, the weight needs to be reduced (the desired value is greater than the one obtained by the neuron). The parameter $\eta$ handles the step sizes of the weights' incremental corrections $\Delta w_{ji}^{(k)}$.

$$\text{Let } \delta_j^{(k)} = -\frac{\partial e_j^{(k)}}{\partial lc_j^{(k)}} = -\frac{\partial e_j^{(k)}}{\partial y_j^{(k)}}\cdot\frac{d\left(y_j^{(k)}\right)}{d\left(lc_j^{(k)}\right)} = \left(t_j^{(k)} - y_j^{(k)}\right)\cdot\frac{d\left(y_j^{(k)}\right)}{d\left(lc_j^{(k)}\right)} \tag{3. 13}$$

Making use of equations **(3. 2)**, **(3. 10)**, **(3. 11)** and **(3. 13)**:

$$\frac{\partial e_j^{(k)}}{\partial w_{ji}^{(k)}}=\frac{\partial e_j^{(k)}}{\partial y_j^{(k)}}\cdot\frac{\partial y_j^{(k)}}{\partial w_{ji}^{(k)}}=\frac{\partial e_j^{(k)}}{\partial y_j^{(k)}}\cdot\frac{d\left(y_j^{(k)}\right)}{d\left(lc_j^{(k)}\right)}\cdot\frac{\partial lc_j^{(k)}}{\partial w_{ji}^{(k)}}=-\delta_j^{(k)}\cdot\frac{\partial lc_j^{(k)}}{\partial w_{ji}^{(k)}}\quad\Rightarrow\quad\frac{\partial e_j^{(k)}}{\partial w_{ji}^{(k)}}=-\delta_j^{(k)}\cdot x_i^{(k)} \tag{3. 14}$$

Thus, by inserting equations **(3. 13)** and **(3. 14)** into equation **(3. 12)**:

$$w_{ji}^{(k+1)}=w_{ji}^{(k)}+\eta\cdot\delta_j^{(k)}\cdot x_i^{(k)}=w_{ji}^{(k)}+\eta\cdot\left[\left(t_j^{(k)}-y_j^{(k)}\right)\cdot\frac{d\left(y_j^{(k)}\right)}{d\left(lc_j^{(k)}\right)}\right]\cdot x_i^{(k)} \tag{3. 15}$$

Notice that for the "linear artificial neuron", $\frac{d\left(y_j^{(k)}\right)}{d\left(lc_j^{(k)}\right)}=1$, so that $\delta_j^{(k)}=\left(t_j^{(k)}-y_j^{(k)}\right)$. Therefore:

$w_{ji}^{(k+1)}=w_{ji}^{(k)}+\eta\cdot\delta_j^{(k)}\cdot x_i^{(k)}=w_{ji}^{(k)}+\eta\cdot\left(t_j^{(k)}-y_j^{(k)}\right)\cdot x_i^{(k)}$ is identical to the ***δ*-rule**!

For the nonlinear artificial neuron, suppose that the nonlinear transfer function is:

$$y_j^{(k)}=\frac{1}{1+\mathrm{e}^{-lc_j^{(k)}}}\quad\Rightarrow\quad\frac{dy_j^{(k)}}{d\left(lc_j^{(k)}\right)}=y_j^{(k)}\cdot\left(1-y_j^{(k)}\right) \tag{3. 16}$$

Hence, inserting equation **(3. 16)** into **(3. 15)**:

$$w_{ji}^{(k+1)}=w_{ji}^{(k)}+\eta\cdot\delta_j^{(k)}\cdot x_i^{(k)}=w_{ji}^{(k)}+\eta\cdot\left[\left(t_j^{(k)}-y_j^{(k)}\right)\cdot y_j^{(k)}\cdot\left(1-y_j^{(k)}\right)\right]\cdot x_i^{(k)}, \tag{3. 17}$$

or in matrix notation:

$$\begin{aligned}\mathbf{w}_j^{(k+1)}&=\mathbf{w}_j^{(k)}+\eta\cdot\delta_j^{(k)}\cdot\mathbf{x}^{(k)}=\mathbf{w}_j^{(k)}+\eta\cdot\left[\left(t_j^{(k)}-y_j^{(k)}\right)\cdot\frac{d\left(y_j^{(k)}\right)}{d\left(lc_j^{(k)}\right)}\right]\cdot\mathbf{x}^{(k)}\\ \mathbf{w}_j^{(k+1)}&=\mathbf{w}_j^{(k)}+\eta\cdot\left[\left(t_j^{(k)}-y_j^{(k)}\right)\cdot y_j^{(k)}\cdot\left(1-y_j^{(k)}\right)\right]\cdot\mathbf{x}^{(k)}\end{aligned} \tag{3. 18}$$

Thus, a simple means of training a nonlinear artificial neuron is defined by equation **(3. 18)**. Note that it is also valid for linear artificial neurons, but not for perceptrons.

Although the nonlinear artificial neuron can solve many problems, Minsky et al. (from [39]) proved that single artificial neurons[34] have severe limitations. A very simple problem that makes the assertion evident is the logical operation "xor". To perform this operation, the neurons need to be cascaded. Although it is not obvious, it is not too difficult to deduce by common sense how to arrange the neurons and how to set the weights of their connections to solve that problem (see **Fig. 3. 14**). Clearly, any desired region could eventually be confined by means of several linear perceptrons, while another perceptron could gather all the partial decisions and make the final one.

[34] Conflicting definitions of "perceptron" are frequent in the literature. The artificial neuron of any kind is sometimes referred to as a "perceptron". Here the term is reserved to neurons with "linear threshold activation functions".

However, it is not possible for bigger problems to deduce by common sense the structure, the weights and the thresholds of the network. For instance, suppose the structure is defined, and the weights and thresholds are randomly chosen. A training data set makes it possible to obtain the error in the network's output, which has to be either eliminated or minimized by adjusting the parameters. However, it is not clear how to distribute the responsibilities throughout the whole network.

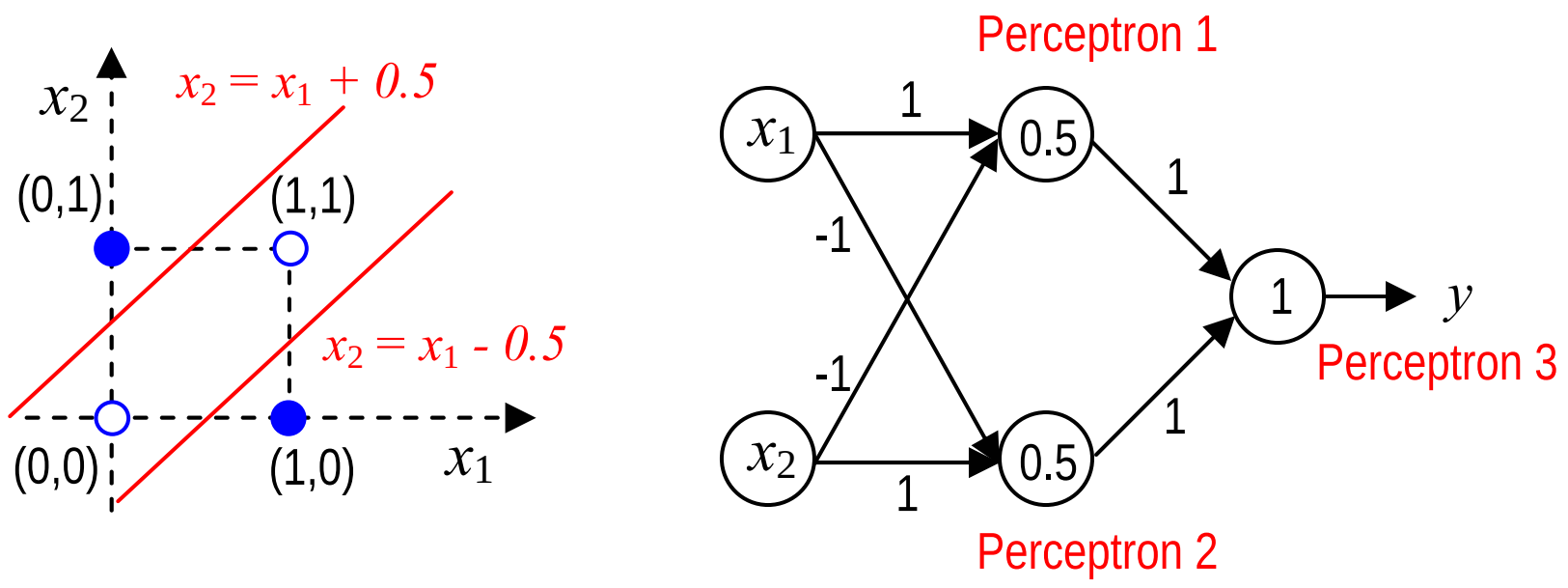

| $x_1$ | $x_2$ | $t$ (target) |
|---|---|---|
| 1 | 0 | 1 |
| 0 | 1 | 1 |
| 1 | 1 | 0 |
| 0 | 0 | 0 |

**Fig. 3. 14**: On the left: The logical operation "xor". The two classes cannot be separated by a single perceptron. A proposed solution is given by the two red straight lines: one class is between the lines, while the other is outside.

In the middle: A proposed ANN that solves the problem. The upper perceptron, whose decision boundary is the upper red line on the left, makes a decision. The lower perceptron, whose decision boundary is the lower red line, makes a decision as well. A third perceptron receives both decisions and makes its own (final) decision. With these proposed weights and thresholds, the third perceptron would fire if and only if only one of the first two perceptrons fires. Notice that the number inside each perceptron is its corresponding threshold. There are infinite possible solutions for this simple problem, with the one proposed here being obtained by common sense.

On the right: Truth table and training data set for the logical operation "xor".

### 3.5.5.2 Multi-layer perceptron

The MLP[35] is probably the ANN most widely used for engineering purposes, especially for function approximation and pattern classification problems. It is a feed-forward network within which each neuron receives inputs from every perceptron in the previous layer, and sends its output to every perceptron in the following layer (see **Fig. 3. 15**).

[35] The name MLP is kept because it is common in the literature, but its components are not limited to perceptrons.

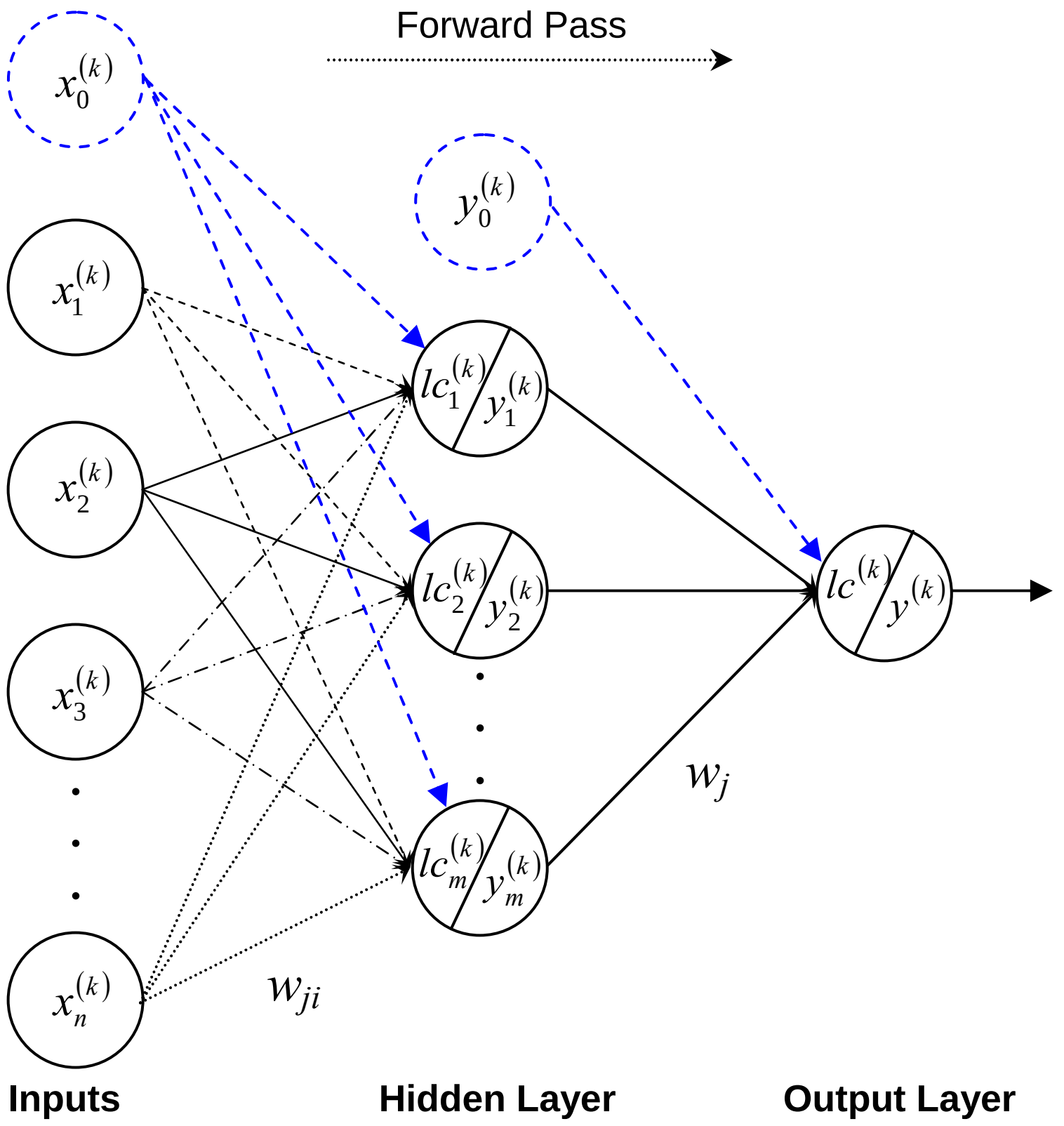

$$lc_j^{(k)} = \sum_{i=0}^{n} w_{ji} \cdot x_i^{(k)}$$
$$y_j^{(k)} = f^h\left(lc_j^{(k)}\right)$$
$$lc^{(k)} = \sum_{j=0}^{m} w_j \cdot y_j^{(k)}$$
$$y^{(k)} = f^o\left(lc^{(k)}\right)$$

Where:

- $lc_j^{(k)}$: linear combination of the inputs to the neuron $j$, corresponding to pattern $(k)$.
- $w_{ji}$: weight of the $i^{th}$ input to neuron $j$ (fixed values once the network is trained).
- $x_0^{(k)} = y_0^{(k)} = -1 \quad \forall(k)$
- $y_j^{(k)}$: neuron $j$'s output for pattern $(k)$.
- $f^h(\cdot)$: transfer function of the neurons in the hidden layer.
- $w_j$: weight of the $j^{th}$ input to the output neuron.
- $f^o(\cdot)$: transfer function of the output neuron.

**Fig. 3. 15**: A typical feed-forward artificial neural network (multi-layer perceptron) with only one hidden layer and one artificial neuron in the output layer.

A critical issue here is the shape of the transfer functions, although they usually fall into one of the three previously explained categories: step-functions, linear functions and sigmoid functions. Furthermore, different types of transfer functions are commonly used for different layers within the same network. For instance, it is frequent to find neurons with sigmoid transfer functions in the hidden layer(s) and neurons with linear transfer functions in the output layer for function approximation, or perceptrons with threshold activation functions (step-functions) for classification tasks.

The output vector from the hidden layer in **Fig. 3. 15** is shown in equation **(3. 19)**.

$$\mathbf{y}^{(k)} = \mathbf{F}^{h}\left(\mathbf{W}^{h} \cdot \mathbf{x}^{(k)}\right) \tag{3. 19}$$

Where:

- $\mathbf{y}^{(k)}$: output vector from the hidden layer, corresponding to pattern (*k*).
- $\mathbf{F}^{h}(\cdot)$: $\mathcal{R}^{m} \rightarrow \mathcal{M}^{m}$, where $\mathcal{M} \subseteq \mathcal{R}$. Typically, $\mathcal{M}:[0,1]$ or $[0,-1]$.
- $\mathbf{y}^{(k)} = \mathbf{F}^{h}\left(\mathbf{lc}^{(k)}\right) = \begin{Bmatrix} f_{1}^{h}\left(lc_{1}^{(k)}\right) \\ \vdots \\ f_{m}^{h}\left(lc_{m}^{(k)}\right) \end{Bmatrix}$, where $f_{j}^{h}(\cdot)$ are typically sigmoid functions.
- $\mathbf{W}^{h}$ : matrix of weights corresponding to the hidden layer. Size: $m$ x $(n+1)$.
- $\mathbf{x}^{(k)}$: pattern (*k*).
- $\mathbf{W}^{h} = \begin{pmatrix} w_{1,0} & w_{1,1} & \cdots & w_{1,n} \\ w_{2,1} & w_{2,2} & \cdots & w_{2,n} \\ \vdots & \vdots & \ddots & \vdots \\ w_{m,1} & w_{m,2} & \cdots & w_{m,n} \end{pmatrix}$ and $\mathbf{x}^{(k)} = \begin{Bmatrix} x_{0}^{(k)} = -1 \\ x_{1}^{(k)} \\ \vdots \\ x_{n}^{(k)} \end{Bmatrix}$.

In MLPs, all the layers typically follow the same equations. However, the output layer is not commonly mapped into a restricted interval. Typically, $\mathcal{M} = \mathcal{R}$ for the output layer.

Equations **(3. 20)** show the equations for the particular case of an output layer consisting of a single neuron, as shown in **Fig. 3. 15**.

$$y^{(k)} = f^{o}\left(\mathbf{w}^{o} \cdot \mathbf{y}^{(k)}\right) \tag{3. 20}$$

Where:

- $y^{(k)}$: output scalar of the whole network, corresponding to pattern (*k*).
- $f^{o}$: $\mathcal{R} \rightarrow \mathcal{M}$, where $\mathcal{M} \subseteq \mathcal{R}$. Typically, $\mathcal{M} = \mathcal{R}$ and $f^{o}$ is a linear function.
- $\mathbf{w}^{o}$: row vector of weights corresponding to the output layer. Size: $(m+1)$.
- $\mathbf{x}^{(k)}$: pattern (*k*).
- $\mathbf{w}^{h} = \{w_{1,0} \quad w_{1,1} \quad \cdots \quad w_{1,m}\}$ and $\mathbf{y}^{(k)} = \begin{Bmatrix} y_{0}^{(k)} = -1 \\ y_{1}^{(k)} \\ \vdots \\ y_{m}^{(k)} \end{Bmatrix}$.

Without loss of generality, this work will be focused from here forth on the type of network shown in **Fig. 3. 15**. Therefore, the equations of interest are **(3. 19)** and **(3. 20)**. Thus, a MLP has been introduced, which is expected to handle the situations where the single neurons succeed, plus those ones where the latter fail.

However, in the same fashion as the single neurons', the MLP's weights need to be learned. When training a perceptron, the problem is how to assign the responsibility of each weight for a given error. Now, the same problem still applies, in addition to the problem of lacking the target output values for the neurons in the hidden layer!

**Learning by the generalized *δ*-rule (back-propagation rule)**

All the ANNs, including the single neuron, must be provided for training purposes with an error detector and with a processor to modify their weights. The former determines the errors in the network's output(s) in relation to a set of desired responses (training patterns) that has to be provided as well, while the latter guides the weights' adjustments to fit the training patterns. Finally, another set of patterns is provided to test the network's learned weights.

The basic principle of the training techniques consists of modifying the weights so as to compensate for the errors that the network makes with respect to the training patterns, the manner of doing which leads to the different learning algorithms. Recall that only the training of a well posed feed-forward ANN is dealt with here. The important issue of designing the network's structure and feed-back ANNs are beyond the scope of this thesis.

The goal of the backpropagation rule is defined to adjust the weights so as to minimize a certain measure of the network's error. Notice that in equation **(3. 11)**, the measures of error were defined for a single neuron $j$ and for the pattern ($k$) as: $e_j^{(k)} = \frac{\left(t_j^{(k)} - y_j^{(k)}\right)^2}{2}$, and the mean error considering all the patterns as: $e_j = \frac{1}{s} \cdot \sum_{k=1}^{s} e_j^{(k)}$. For networks, the measures of error are:

$$
\begin{aligned}
e^{(k)} &= \sum_{j=1}^{w} e_j^{(k)} \\
e &= \frac{1}{s} \cdot \sum_{k=1}^{s} e^{(k)}
\end{aligned}
\qquad \textbf{(3. 21)}
$$

Where $w$ is the number of neurons in the output layer, and $s$ is the number of training patterns.

Hence the aim is to adjust the weights so as to minimize the error $e$ stated in equation **(3. 21)**. The correction of each neuron's weights is made according to equation **(3. 15)**:

$w_{ji}^{(k+1)} = w_{ji}^{(k)} + \eta \cdot \delta_j^{(k)} \cdot x_i^{(k)}$, where $\delta_j^{(k)} = -\frac{\partial e_j^{(k)}}{\partial y_j^{(k)}} \cdot \frac{d\left(y_j^{(k)}\right)}{d\left(lc_j^{(k)}\right)}$, according to equation **(3. 13)**.

Two different cases can be distinguished to deduce $\frac{\partial e_j^{(k)}}{\partial y_j^{(k)}}$ : "Output" and "Hidden" neurons.

In the first case, the weights' correction rule is the same as that of the nonlinear artificial neuron, as shown in equation **(3. 13)** and reproduced in equation **(3. 22)**.

$$\delta_j^{(k)} = \left(t_j^{(k)} - y_j^{(k)}\right) \cdot \frac{d\left(y_j^{(k)}\right)}{d\left(lc_j^{(k)}\right)} \quad \textbf{(3. 22)}$$

Where $j$ is an output neuron.

In the second case, the neuron $j$ belongs to a hidden layer. Each neuron in a hidden layer is in some degree responsible for all the outputs' errors in the network, so that the error to minimize for each pattern ($k$) is not $e_j^{(k)}$ but $e^{(k)}$, which involves all the output units, as stated in equation **(3. 21)**.

$$\delta_j^{(k)} = -\frac{\partial e^{(k)}}{\partial y_j^{(k)}} \cdot \frac{d\left(y_j^{(k)}\right)}{d\left(lc_j^{(k)}\right)} \quad \textbf{(3. 23)}$$

$$\frac{\partial e^{(k)}}{\partial y_j^{(k)}} = \frac{\partial\left(\sum_{q=1}^{w} e_q^{(k)}\right)}{\partial y_j^{(k)}} = \sum_{q=1}^{w} \frac{\partial e_q^{(k)}}{\partial lc_q^{(k)}} \cdot \frac{\partial lc_q^{(k)}}{\partial y_j^{(k)}} = \sum_{q=1}^{w} \frac{\partial e_q^{(k)}}{\partial lc_q^{(k)}} \cdot \frac{\partial}{\partial y_j^{(k)}}\left(\sum_{i=1}^{z} w_{qi}^{(k)} \cdot y_i^{(k)}\right) \quad \textbf{(3. 24)}$$

$$\frac{\partial e^{(k)}}{\partial y_j^{(k)}} = \sum_{q=1}^{w} \frac{\partial e_q^{(k)}}{\partial lc_q^{(k)}} \cdot w_{qj}^{(k)} = \sum_{q=1}^{w} \delta_q^{(k)} \cdot w_{qj}^{(k)} \quad \textbf{(3. 25)}$$

Where $j$ is a hidden neuron.

Notice that $j$ is a hidden neuron, $q$ is an output neuron, $z$ is the number of neurons in the hidden layer, and $w$ is the number of neurons in the output layer. Thus, the network's weights' correction rule is as stated in equations **(3. 26)**, **(3. 27)** and **(3. 28)**.

$$w_{ji}^{(k+1)} = w_{ji}^{(k)} + \eta \cdot \delta_j^{(k)} \cdot x_i^{(k)} \quad \textbf{(3. 26)}$$

Where, for output neurons:

$$\delta_j^{(k)} = \left(t_j^{(k)} - y_j^{(k)}\right) \cdot \frac{d\left(y_j^{(k)}\right)}{d\left(lc_j^{(k)}\right)} \tag{3. 27}$$

while for hidden neurons:

$$\delta_j^{(k)} = \left(\sum_{q=1}^{w} \delta_q^{(k)} \cdot w_{qj}^{(k)}\right) \cdot \frac{d\left(y_j^{(k)}\right)}{d\left(lc_j^{(k)}\right)} \tag{3. 28}$$

For multiple hidden layers, the summation in equation **(3. 28)** runs from one to the number of neurons in the next higher layer.

Thus, once all the weights associated with the highest hidden layer are adjusted, the procedure continues in the same fashion towards the inputs. Then, the next pattern is presented and the process is repeated iteratively until the weights tend to converge.

Theoretically, this method provides a means of training MLPs with any number of layers and any number of neurons per layer (not suitable for feed-back networks). More precisely, the method is suitable for feed-forward networks, even if they are not organized in layers: …*the network does not have to be organized in layers (any pattern of connectivity that permits a partial ordering of the nodes from input to output is allowed)* [75].

At present, traditional methods for training ANNs are based on back-propagation gradient descent techniques. Recently, however, some attempts have been successfully made to train the networks by evolutionary and particle swarm techniques.

## 3.5.6 Further comments on artificial neural networks

### 3.5.6.1 Momentum

The ANN performs a nonlinear input-output mapping. In addition to that, the error function used for training purposes introduces nonlinearity in the output of the ANN. Thus, this double nonlinearity on the variables (weights) of the function to be minimized for the training of the network generates many local optima.

The gradient-based back-propagation algorithm is likely to get trapped in any local minimum. A technique that can help to improve its performance in this regard is the addition of a "momentum" term to equation **(3. 26)**, as shown in **(3. 29)**:

$$w_{ji}^{(k+1)} = w_{ji}^{(k)} + \eta_1 \cdot \delta_j^{(k)} \cdot x_i^{(k)} + \eta_2 \cdot m_{ji}^{(k)} \quad ; \quad 0 \leq \eta_2 \leq 1$$
$$m_{ji}^{(k)} = w_{ji}^{(k)} - w_{ji}^{(k-1)} \qquad \textbf{(3. 29)}$$

Thus, the momentum term simply adds a fraction of the previous weight's update, so that when the gradient continuously changes direction, this technique smoothes out the variations.

### 3.5.6.2 Training, testing and overfitting

The weights of the network are trained by means of an algorithm that minimizes a certain measure of error, as previously explained. The ANN learns its weights form certain training patterns, so as to return an admissible error. However, when handling new data that were not used during the training, the level of error is commonly greater. Therefore, although it is tempting to use as many patterns as possible for the training procedure, the available data must be partitioned in two, keeping one part for testing the learned weights.

The amount of data used for training and for testing is problem dependent, though usually the available data is randomly divided into equal sets. *If several attempts to train the neural network do not result in an acceptable error, it may be useful to add more nodes to the structure of the network. A larger network can approximate more functions, but larger networks are also prone to overfitting the available data. This is the same problem that occurs when trying to fit polynomials to data* [54]. Overfitting the data results in lower errors when the network is offered the training samples, but errors are likely to be high for unseen patterns.

In general, it is not known in advance how many hidden neurons would be needed to produce a good approximation to the data. Typically, if the number of neurons is too small, the approximation function that the network can generate does not fit the data properly. If it is too high, overfitting occurs. Overfitting is likely to become a problem if there are few training patterns available. Conversely, when the number of patterns tends to infinite, overfitting is not a danger because the noise associated with the data ends up being negligible.

A simple way of avoiding overfitting consists of dividing the data into three sets: a training set, a validation set, and a test set. The network is trained using the first set of data, stopping the training every now and then to test with the second set of data. Usually the errors on the validation data set are higher than on the training set. When the updates of the weights keep on diminishing the error on the training data set, but the error in the validation data set stops decreasing (typically it starts increasing slowly), the network starts overfitting. Therefore, the training is stopped at the time when the error in the validation data set is at its minimum. Finally, the third independent data set is used to test the ability of the trained network to generalize. The disadvantage of this method is that it requires a data-rich situation (in which case the risk of overfitting decreases!).

There are basically two “styles” of training: “incremental training”, and “batch training”.

The incremental (or adaptive) training updates the weights each time a pattern is presented, so that the training patterns are sequentially presented to the network. Notice that all the learning algorithms previously discussed were written following this strategy.

Conversely, the batch training updates the weights only after all the training data were presented to the network. Therefore, the error function to minimize encompasses the errors of all the presented patterns. The latter is the one onto which emphasis will be put in this work.

Many different techniques to train ANNs can be found in the literature, depending on the transfer functions and on the architecture of the network. However, not only do these techniques require a specific design according to the kinds of neurons and to the structure of the network, but also they can only find a local optimum which is not even guaranteed to be a reasonably good one. Traditional training methods are gradient-based, so that they do not perform a good exploration of the search-space, and they require differentiable transfer functions. Furthermore, the training results depend to a great extent on the initialization of the weights. Suppose, for instance, the very simple case of a single linear artificial neuron, whose correction rule is $w_{ji}^{(k+1)} = w_{ji}^{(k)} + \eta \cdot \delta_j^{(k)} \cdot x_i^{(k)}$, and suppose for simplicity that all the inputs have the same sign. Then, all the weights’ updates will be either positive or negative!

Differently, the PSO method can be used to train any kind of network, no matter of the transfer function (which can be non-differentiable), and no matter of the architecture of the network. Furthermore, since the ANN performs a nonlinear mapping, whose output is an input to another nonlinear function that computes the error to be minimized, the latter result in a function with many local optima. The PSO algorithm, as opposed to gradient-based methods, is well capable of escaping poor local optima so as to converge into a global optimum or at least a good local one. Therefore, there is no need to develop ANN-specific training algorithms. Similar to the PSO approach, the EAs are alternative methods to train

ANNs. *There has already been a great deal of effort directed toward evolving neural networks… …The evolutionary process cannot only search for the best set of weights for a fixed neural architecture, but also for the best architecture at the same time. All that is required is to encode the manner in which the connections between neurons are represented* [54].

### 3.5.6.3 Feed-back artificial neural networks

Feed-forward networks are very widespread for engineering applications because they are precisely that: "engineering devices". Hence they perform well at classification and function approximation, where the outputs do not affect the inputs. However, even for engineering problems such as forecasting, time-series prediction and any kind of problem where the outputs depend at least on the present state of the system in addition to the inputs, the outputs are fed-back by means of loops within the network. Of course, for cognitive science, feed-back (or recurrent) networks are the most interesting ones, since, for instance, some kinds of memory can be generated by allowing "old" signals to keep on travelling through complex loops inside the network while new signals are being received.

However, recurrent networks are out of the scope of this work, and are only mentioned here for completeness. Notice, though, that the PSO paradigm seems, at first glance, suitable for training this difficult-to-train kind of networks. This subject matter is left for future work.

### 3.5.6.4 Application areas

The main applications of the feed-forward networks are "pattern classification" and "function approximation". Other applications more related to the ANNs as a means of creating AI are "associative memory" and "optimization".

**Pattern classification**

The input layer of the network is fed with a set of patterns, each one of which is composed of a set of features. For each received pattern, the output of the network establishes what class the pattern offered belongs to. Typically, when there are only two possible classes, the output layer is composed of one perceptron only (if the pattern offered makes it fire, the pattern belongs to one class, otherwise to the other). For a greater number of classes, there usually are as many output neurons as there are number of classes.

In order to perform classification tasks, the weights need to be set. Fitting the weights to a given training data set is called “learning”, which is typically performed by a back-propagation gradient descent technique over a suitable error surface (note that there are many techniques following this line). The alternative presented later in this thesis is learning the weights by means of a particle swarm optimizer.

### Function approximation

Given an input vector, the network is expected to obtain the output corresponding to a certain phenomenon for which no analytical expression is known. Therefore, instead of having discrete outputs representing different classes, the output here is a real number. Typical transfer functions are linear or sigmoid functions, as previously explained.

In the same fashion as in pattern classification, the weights need to be learned, and the most common techniques are also based on back-propagation gradient descent techniques.

This approach is useful for function approximation and also for time-series predictions, where the aim is at predicting future values of a series given previous values, as explained when discussing the first applications of the EP paradigm.

*Function approximation is the task of learning or constructing a function that generates approximately the same outputs from input vectors as the process being modelled, based on some available training data… …A single hidden layer neural network is sufficient for a network to be a universal function approximator… …Theoretically, with a sufficient number of nodes in the hidden layer, any nonlinear function can be approximated* [2].

### Associative memory

*In this application there is no input/output distinction. Rather the network learns a set of identity mappings by fitting* $w_{ij}$ *to memory patterns… If an incomplete or partly erroneous memory is presented to the network, it completes the pattern. In a certain sense, feature recognizers are special cases of associative memories* [64].

### Optimization

Although this thesis deals with optimization tasks, discussing optimization by means of ANNs would require a much deeper review of the field, which is out of the scope of this work. It is worth mentioning however, that feed-back ANNs have been claimed to be capable of dealing with very complex combinatorial optimization problems. A very brief discussion on this matter can be found in [64], together with specific applications to typical well known combinatorial optimization problems such as the TSP and “scheduling problems”.

# 3.6 Closure

A brief overview of the field of AI was presented. Its three main paradigms were outlined, not specifically intending to discover their strengths and weaknesses, but to understand their underlying principles, and to recognize the possibilities of dealing with optimization problems by this means.

Thus, the **symbolic paradigm** is argued to consist of a sequence of arithmetic calculations and statements manipulated by crisp logic operators. A tree-like structure guides the process, where each branch corresponds to a crisp logic “if-then” decision. This process saves time to humans by repeatedly applying tedious processes that are programmed based on humans’ knowledge. “Expert systems” are the most successful results of this paradigm.

While the symbolic paradigm relies on mimicking the humans’ behaviour, the **connectionist paradigm** proposes to mimic the massively parallel structure of humans’ brain, which was realized to lead to astonishing results in dealing with the simple tasks that the symbolic paradigm failed in handling. Intelligence is expected to emerge in a bottom-up fashion from the interactions of the basic neuron-like entities. Therefore, a neurophysiologic instead of a behavioural approach is proposed.

The **AL paradigm** goes further and it proposes to create artificial creatures that would display intelligence of their own, without relying on humans’ knowledge. This approach is inspired by the observation that biological organisms learn and evolve so as to deal with extremely complex problems in order to achieve their goals, thus showing different degrees of intelligence. Similarly to ANNs, AL paradigms rely on the overall behaviour that emerges from the interactions of simple entities, instead of programming it in an explicit fashion. The phenomenon of emergence is not fully understood, and cannot be implemented. It is known to be related to the interactions among several entities, but sometimes these interactions cancel each other out. Clearly, like randomness, it is a word to cover up the lack of complete understanding of the real processes. Thus, in the same fashion as biological organisms, the complex behaviour emerges from simple agents that interact by means of simple rules and stochastic operators. Notice that randomness allows making decisions when facing unknown situations, and even when the situation is familiar, it allows discovering new alternatives, whereas learning apprehend the results of those decisions.

AL encompasses many different sub-fields, some of which, in spite of being inspired by processes that biological organisms undergo, they do not aim to create living creatures but to mimic processes that are useful for solving engineering problems (e.g. evolution and social behaviour). This leads to paradigms such as EAs, PSO and ACO, which are well suited to deal with optimization tasks. These paradigms were included here within the AL field from a general point of view, but when dealing specifically with optimization problems, they are viewed as modern-heuristic population-based optimization techniques. Note that ANNs were discussed in some detail here, not as a means of creating AI, but as general function approximators that can be trained by means of an AL paradigm that displays some degree of intelligence.

Focusing specifically on optimization problems, EAs are discussed in detail in **Chapter 4**, and the PSO method from **Chapter 5** forth. It may seem odd not to start discussing about the population-based methods until **Chapter 4**, but dealing with them as isolated methods necessarily leads to a loss of perspective. When first introduced to a population-based method as if it had been designed from scratch, it is inevitable to feel astonished by the fact that someone could have ever possibly come up with such ingenious idea, especially when they so heavily rely on emergent properties that are not fully understood! However, the following of their evolution, and that of some related disciplines, together with the recognition of the links amongst all of them—even when they seem to have been inspired by different natural metaphors—, help to comprehend the lines of thought of the paradigm's creators, as well as to get a sense of the way the algorithms work. This is critical for that understanding their precise behaviour in a perfectly deterministic fashion seems to be rather impossible.

Thus, this general overview allows a wider perspective so as to situate the methods within a bigger frame, as well as to visualize alternatives and improvements that can hardly be thought of from scratch.

# Chapter 4

# EVOLUTIONARY ALGORITHMS

An overview of the evolutionary algorithms is presented, emphasizing their capabilities of handling optimization problems. Some concepts of natural evolution are outlined, and the links between evolution and optimization are discussed. The advantages of the parallelism that population-based methods introduce together with the motives to incorporate stochastic operators into the algorithms are presented. The general procedure of a generic evolutionary algorithm is described, and the main paradigms: "evolution strategies", "evolutionary programming", "genetic algorithms" and "genetic programming" are discussed in some detail. A few possible techniques to handle constraints within population-based methods are considered, while a more detailed discussion about handling constrained optimization problems by means of a particle swarm optimizer is carried out in **Chapter 10**.

## 4.1 Introduction

Some optimal natural structures such as the shape of a shark, which resembles the optimized shape of a submarine, have motivated some researchers in Europe and in the United States, surprisingly independently, to start working some decades ago on the idea of mimicking biological evolution to overcome adaptation and optimization problems.

The EAs are population-based methods developed under the metaphor of natural evolution. Although they are frequently used as models of evolution in AL systems, their origins are not related to that field. In fact, the origins of the different EAs are not even related to each other. Nevertheless, it seems fair to include them within the AL paradigms, since their intelligent behaviour is originated by mimicking natural processes that biological organisms undergo.

There are so many similarities between the EAs and the PSO method that some researchers consider the latter belonging to the former, despite not being inspired by natural evolution. Nonetheless, it evolves a population of individuals taking into account previous experiences, and making use of stochastic operators to introduce new responses, very much like evolution.

Therefore, it seems fair to deal with EAs in some detail, beginning with brief discussion about the biological basis under which the algorithms were originated.

Evolution is a natural process that organisms undergo to adapt to a dynamic environment in order to survive. Since these organisms adapt by seeking the best response to the challenge they are facing, they happen to perform complex optimization processes. This can be viewed as a process of fitness maximization.

Real-world problems do not usually lend themselves to optimization by traditional methods, so that some simplifications to the problems must be made so as to suit the problem to traditional techniques, or more robust optimization methods need to be developed. Bäck et al. [9] suggested that …*even if the general global optimization problem is unsolvable, the identification of an improvement of the actual best solution by optimization is often already a big success for practical problems, and in many cases EAs provide an efficient and effective method to achieve this*.

In nature, organisms have proven successful behaviour in dealing with complex situations at different levels (chromosome's, individual's, species', population of species', etc.), which are characterized by nonlinear interactions, chaos, stochastic processes, nonlinear dynamics, uncertainties, general noise, etc. These are precisely the kinds of problems that traditional methods have failed in dealing with. Thus, scientists' interest in understanding and mimicking the natural mechanisms which nature makes use of to overcome these difficulties is obvious.

## 4.2 Natural evolution background

Since it is not the goal of this thesis to deal with natural evolution, but to briefly understand what the metaphor used by EAs consists of, only a very simplistic discussion on the subject is presented hereafter. Serious efforts were spent in understanding the concepts underlying natural evolution and genetics, and help from people with a more appropriate background was asked when necessary. However, this complex field is way beyond the expertise of the author of this thesis, so that this chapter should be read with care. Mistaken conclusions are possible, for which apologies are presented in advance.

Darwin [19] claims that all organisms descend, with modifications, from a common ancestor, and that the adaptive changes of species occur by means of small apparently random (i.e. not adaptively directed) mutations[1], which if advantageous, are preserved by natural selection[2].

[1] Smaller apparently random mutations are observed in nature much more frequently than larger ones.

The macroscopic process of evolution is driven both by reproduction and by environmental influences. Hence, within a wealthy environment, a population tends to increase exponentially until the resources become insufficient. At this point, the natural selection process is activated and the fittest species[3] are more likely to survive to evolve further.

Notice that the "Darwinian theory of evolution" was developed without any knowledge of genetics. A modern theory, called "neo-Darwinian theory of evolution", incorporated genetics and population biology, recognizing the importance of mutation and variation within a population. Thus, it still recognizes "natural selection" as the mechanism of evolution, but it claims that this mechanism consists of alterations in the frequency of genes[4] in a population. Thus, the microscopic mechanisms of evolution are dealt with too, and living organisms are viewed as a duality of "genetic encoded information" (genotype[5]) and "external observable features" (phenotype[6]). A further step was given recently by replacing the "neo-Darwinian theory of evolution" by the "modern synthesis of genetics and evolution", which offers several other mechanisms in addition to natural selection. Sometimes the "modern synthesis" is referred to as the "neo-Darwinian theory of evolution", what might lead to confusion. All in all, Morán [57] summarizes the three main aspects in what the modern view of evolution differs from Darwinism:

1. *It recognizes several mechanisms of evolution in addition to natural selection. One of these, random genetic drift[7], may be as important as natural selection.*
2. *It recognizes that characteristics are inherited as discrete entities called genes. Variation within a population is due to the presence of multiple alleles[8] of a gene.*
3. *It postulates that speciation[9] is (usually) due to the gradual accumulation of small genetic changes. This is equivalent to saying that macroevolution is simply a lot of microevolution.*

[2] Natural selection is the ability of some individuals to survive and outlast others, to pass their genetic information to the next generation.

[3] The fittest species are the best adapted to the environmental conditions, exploiting the resources in a more efficient manner. From a very simplistic point of view, since they adapt better, they live longer, thus passing their genes to the next generation through a higher number of children.

[4] Genes are the transfer units of heredity.

[5] The genotype is the encoded information inscribed in the DNA, present in every cell of an individual.

[6] The phenotype encompasses all the observable features of the individual (behavior, physiology, morphology, etc.).

[7] Genetic drift: *random fluctuations in the frequency of the appearance of a gene in a small isolated population, presumably owing to chance rather than natural selection* [23].

[8] Alleles are any of the alternative forms of a gene that may occur in a given locus (position).

*In other words, the "Modern Synthesis" is a theory about how evolution works at the level of genes, phenotypes, and populations whereas Darwinism was concerned mainly with organisms, speciation and individuals… The major controversy among evolutionists today concerns the validity of the third point.*

To sum up, and circumscribing to this thesis' specific range of interest[10], the individual is the unit of selection, and its genetic information interacting with the environment gives shape to its phenotype. The unit undergoing evolution is the population of individuals, exposed to the environment, whose whole genetic information is contained in a so-called "genetic pool".

Evolution takes place through generations, during reproduction. The phenotypes can change adapting to the environment within a single individual's life span (for instance, the skin changes its colour when exposed to the sun), but the genetic code remains unchanged. There can only be small changes in the genotype of an individual within a generation by means of infrequent and small mutations, or some kinds of recombination in prokaryotes[11].

The information encoded in the genotype provides the instructions for building different parts of the cell according to the needs, and sets the role each cell is going to play in the large-scale bio-mechanism. All the information encompassing these instructions is inscribed in a macromolecule called "deoxyribonucleic acid" (DNA). For a brief review of genetic concepts that might help to understand the natural evolution processes that inspired the development of the EAs, refer to **Appendix 2**.

*It is a matter of fact that some living beings sometimes happen to imitate other living beings… They do so in order to get some protection from their enemies. But they don't do it consciously on the level of the individuals involved. It is our simplified interpretation on the level of populations or species over several generations of what is really happening as an "evolutionary process". Imitators who resemble the model more closely simply live longer and have more descendants… …Thus, their genetic information spreads within the pool of living genes* [68].

Evolution is a complex natural process that we, humans, artificially divide into small simple usually sequential processes to ease the understanding. Thus, living organisms are viewed as a

[9] Speciation is the evolution of species.

[10] The evolution of the population modifies the environment, so that the population has to adapt again to the new conditions, in a cyclic fashion. This natural process is dynamic, and many different populations interact with the same environment and with other populations, thus altering each other. This complex mechanism is ignored here, since this work is focused on an isolated population of individuals. It is assumed that the dynamic environment, if so, is due to external forces. Feed-back between the population and the environment is out of the scope of this thesis.

[11] Prokaryotes are cells or organisms lacking a membrane-bound nucleus, such as bacteria (they are always haploids).

duality of a genotype and a phenotype. Taking that into account, Lewontin (from [31]) proposed the specification of a genotypic space **G**, a phenotypic space **P**, and four mapping functions ($f_1$, $f_2$, $f_3$, $f_4$), as shown in **Fig. 4. 1**:

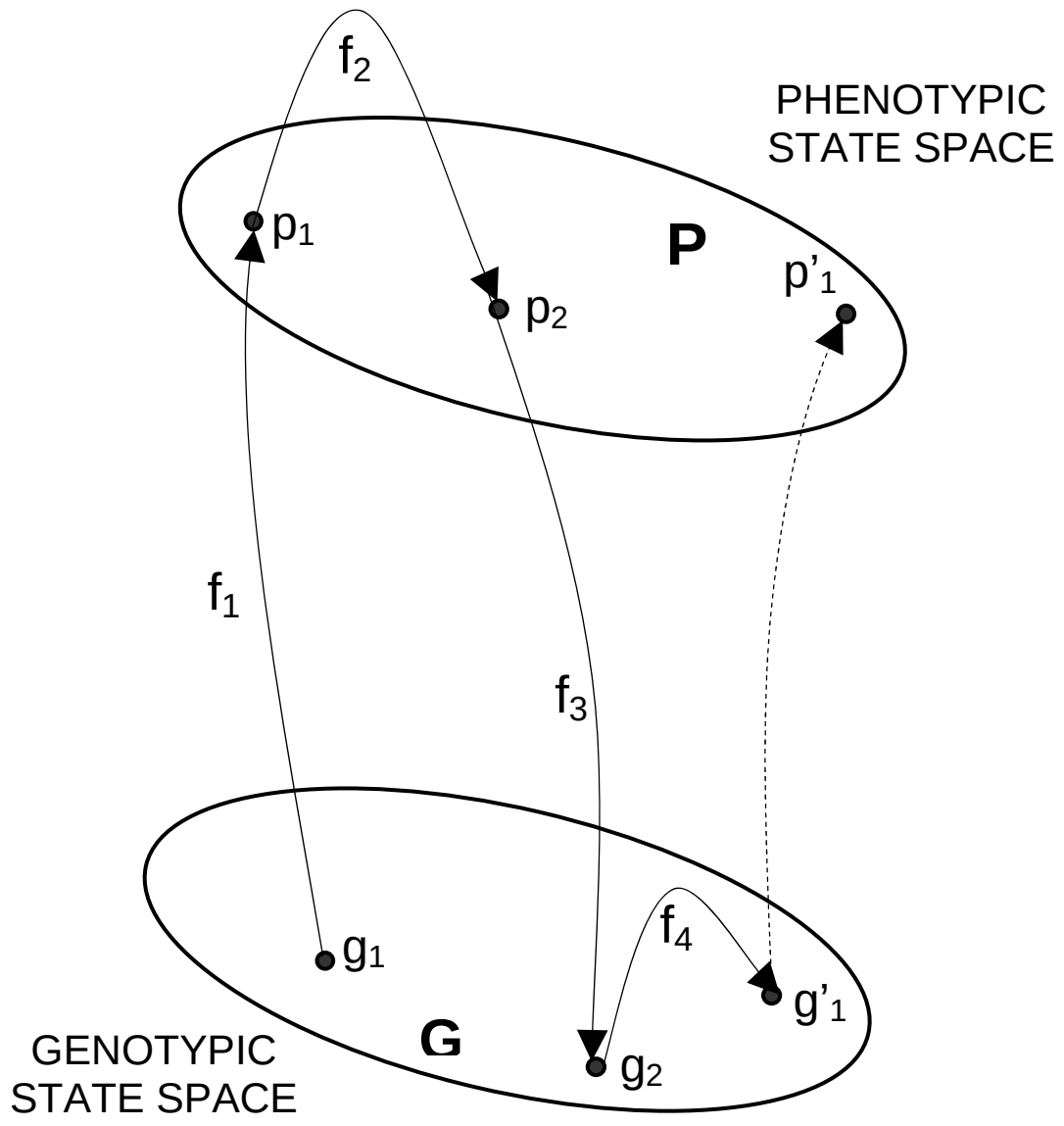

**Fig. 4. 1**: Schematic evolution process of a population within a single generational step. It can be viewed as a succesion of four mapping functions (“epigenesis”, “selection”, “genotypic survival” and “alteration”) relating the genotypic information state space and the phenotypic observable state space (from [30, 31]).

$\mathbf{f_1}$ (epigenesis **G → P**): it maps the population encoded as $g_1$ in the genotypic space **G** into the phenotypic space **P**, as a particular set of traits. Note that the phenotypic space is linked to the environment, so that the observable features of the encoded population $g_1$, represented by the set of phenotypes $p_1$, will depend both on the genetic code and on the environmental features. Notice that there is one subset of phenotypes for each element of the population.

$\mathbf{f_2}$ (selection **P → P**): since natural selection works on phenotypes, the function $f_2$ maps the sets of phenotypes $p_1$ into $p_2$ so that the process of natural selection is performed without any knowledge of the information encoded in $g_1$. The set of phenotypes $p_2$ encompasses the subsets of phenotypes corresponding to the selected individuals.

$\mathbf{f_3}$ (genotypic survival **P → G**): it maps the selected set of phenotypes $p_2$ into the genotypic space **G**. In other words, it encodes back the set of selected subsets of phenotypes, as $g_2$.

$\mathbf{f_4}$ (alteration **G → G**): it maps the genotypes $g_2$ into $g'_1$, comprising all the genetic changes of the process of evolution corresponding to the present iteration. Hence, function $f_4$ contains all the rules handling genetic changes. It is at this point when genetic alterations take place, since previously only a selection process was executed without any genetic information either added (mutation) or further explored (recombination). After the new population $g'_1$ is encoded in **G**, which is related to $p'_1$ in **P** through the mapping function $f'_1$, the single generation is complete. Notice that the evolving unit is the population, which encompasses all the information encoded in the individual in a sort of common genetic pool.

The mapping from the genotypic to the phenotypic state space is "pleiotropic"[12], whereas the mapping following the other direction is "polygenic"[13]. Thus, no useful simplification of the complex mapping between the two state spaces can be made. Notice that selection takes place within the phenotypic state space, whereas genetic alterations do within the genotypic state space. However, computational EAs focus on either one or the other state space, thus developing artificial processes that do not have counterparts in nature.

Although many features of the natural evolution processes remain inscrutable, there are some agreements among the scientific community. Davies [21] claims that evolution takes place on chromosomes: *...here are some general features that are widely accepted:*

- *Evolution is a process that operates on chromosomes rather than on the living beings they encode.*
- *Natural selection is the link between chromosomes and the performance of their decoded structures. Processes of natural selection cause those chromosomes that encode successful structures to reproduce more often than those that do not.*
- *The process of reproduction is the point at which evolution takes place. Mutations may cause the chromosomes of biological children to be different from those of their biological parents, and recombination processes may create quite different chromosomes in the children by combining material from the chromosomes of two parents*[14].
- *Biological evolution has no memory. Whatever it knows about producing individuals that will function well in their environment is contained in the gene pool (the set of chromosomes carried by the current individuals) and in the structure of the chromosome decoders.*

---

[12] More than one phenotypic trait is affected by a genic change.

[13] The modification of a phenotypic trait may be due to simultaneous interactions of several genes' modifications.

[14] This assertion is not accurate. Recombination in eukaryotes occurs during meiosis I, exchanging genetic material between the two sets of chromosomes of a single parent (see **Appendix 2**). When the gametes join to form the offspring, recombination does not occur. Hence, recombination of genetic material when creating offspring, in eukaryotes, takes place between the chromosomes of two grandparents of the child on the one hand, and of the other two grandparents on the other hand. In other words, the two parental sets of chromosomes of a gamete do undergo crossover between corresponding chromosomes.

# 4.3 Evolutionary optimization

Schwefel [68] offers: *Even today, many people think that evolution is a prodigal process and they would model it as a pure random method. They do not realize the fact that life exists on Earth since $10^{17}$ seconds only, and that this would by far not be sufficient to solve the combinatorial task of putting together the DNA of the simplest yeast or bacterium*.

EAs, also known as evolutionary computation (EC), encompass all the optimization methods based upon natural evolution metaphors, where “natural selection” and “survival of the fittest” are the two concepts of utmost importance. There are two different approaches to EAs with regards to the already mentioned view of living adaptive organisms as a duality of their genotype and their phenotype. The “genetic-based” approach focuses on genetic structures, whereas the “phenotypic-based” approach does it on the observable features.

Although all population-based algorithms could be considered to be bottom-up approaches in the sense that the system’s behaviour emerges[15] in a higher level than the individuals’, genotypic-based EAs are typically claimed to be bottom-up approaches in the sense that they emphasize *segregation of individual components and their local interactions, reducing the total system into successively smaller subsystems that are then analyzed in a piecemeal fashion* [31], whereas phenotypic-based EAs are typically claimed to be top-down approaches in the sense that they *emphasize extrinsically imposed forces and their associated physics as they pertain to the modelled system in its entirety* [31][16]. Nevertheless, both approaches are population-based, and their behaviours emerge from the interactions of the members of the population.

In the same fashion as all population-based methods, EAs rely on a population of individuals instead of relying on a single coordinate in the search-space at a time, so that each individual becomes a potential solution to the problem at issue. To represent these solutions in a computer, a data structure must be defined, for which no best choice can be generalized for all problems and for every paradigm. Next, the individuals’ fitness is calculated related to an objective function[17].

---

[15] The key concept of “emergence” has been already discussed in **Chapter 3**, when dealing with the AL paradigm.

[16] Notice the similarity to the “symbolic paradigm”, based on mimicking humans’ behaviours, and the “connectionist paradigm”, based on mimicking humans’ neurophysiology!

[17] As a particular case, the fitness function could be the objective function itself, or at least proportional to it.

For the following iteration, a new population replaces the previous one by means of a manipulation procedure, usually using fitness-based selection operators over the last population and applying probabilistic transformations over the selected individuals, or in the reverse order. This is accomplished by means of two main kinds of operators: "selection" and "alteration". "Selection operators" take the last population and return a new one without modifying the individuals. Each individual's probability of being selected increases with its fitness, and repeated selection of the same individual is allowed. Therefore, new areas of the search-space are not explored. In contrast, "alteration operators" do modify individuals, thus exploring new areas of the search-space.

There are two main types of "alteration operators", although many others could be conceived:

1. Mutation type: It consists of the creation of new individuals by small changes in a single individual. Not only does it allow exploration but it also adds new information to the genetic pool. It mimics nature in the sense that smaller changes occur more often than greater ones[18].

2. Recombination type: They are higher order transformations, which create new individuals by combining parts from two or more individuals. It allows exploration without adding genetic information that was not included in the last population.

As the evolution progresses with the individuals adapting to the environment, the population converges towards a solution. Despite there not being a theoretical basis that guarantees the convergence towards an optimum, the empirical evidence demonstrates that they do so, and that the solutions found are near-optimal (for more details, refer to Bäck [6]). EAs are especially well suited for handling hard and complex real-world problems.

It is worth mentioning that there is still considerable disagreement among scientists with regards to the two widest spread kinds of "alteration operators". The genetic-based algorithms community generally supports the idea that the mutation operator plays a secondary role, while the phenotypic-based algorithms community emphasizes it. Some suggest that the low mutation rate observed in nature is due to the fact that it has reached a dynamic equilibrium close to the optimum, hence diminishing the rate of mutation. Note that bacteria, for instance, show a high mutation rate.

[18] Notice that in bottom-up approaches like genetic-based algorithms, the small changes made by mutation are small in the genotypes, but depending on the mapping to the phenotypes, the steps can take any arbitrary size.

There are numberless types of operators suggested in the literature, but in the same fashion as previously stated for the data structure, no best choice can be generalized for the operators. Once again, they are both problem and algorithm dependent.

Despite there not being complete agreement among scientists with regards to the different mainstreams of EAs, a widely accepted classification divides them into three methodologies:

- **Evolution Strategies (ESs)**
- **Evolutionary Programming (EP)**
- **Genetic Algorithms (GAs)**

Some researchers add a fourth one, whereas some others claim that it is just a branch of GAs:

- **Genetic Programming (GP)**

Furthermore, some people include the PSO method within the EAs[19]. In this work, the PSO paradigm is included within the swarm-intelligence-based methods instead, because it is not based upon the metaphor of natural evolution but upon the metaphor of the intelligence that emerges from a cooperative social behaviour among a group of individuals.

All the EAs' paradigms were initially developed independently, pursuing different purposes. For instance, GAs were aimed at simulating natural evolution[20], EP at creating artificial intelligence (AI), ESs at solving practical optimization problems, and GP at dealing with automatic programming. At present, there are so many variations and problem-specific adaptations, that it is possible to apply any of the paradigms to almost any task, namely biological simulations, AI, machine learning, automatic programming, global optimization problems, etc. The GA paradigm is the most popular of the EAs, and consequently the one with the most work done about. However, since almost every technique can benefit from others, the new tendency is to work on hybrid methods, which, when properly designed, outperform any pure technique. Kennedy et al. [47] *realize that the emphasis on GAs is fading somewhat. In fact, hybrids of the four methodologies are becoming increasingly popular…*

[19] The PSO method is population-based, it makes use of stochastic operators, the search is driven in parallel according to the performance of the individuals, and solutions "evolve" through time. Very much like EAs.

[20] This is rather paradoxical, since canonical GAs are arguably suitable for biological simulations at present. They are bottom-up reductionist models of evolution, whilst no emergent behaviour can be predicted from a bottom-up approach in which too many simplifications are made at the "bottom level". The simplification of the genetic mechanisms would lead to unreliable predictions.

In spite of the different paradigms included into the EAs, the main broad structure of the algorithms remains very much the same, as shown in **Fig. 4. 2**. Typically, the selection stage is claimed to be performed before the "alteration operations" in the genetic-based EAs, and after them in phenotypic-based ones. Thus, to encompass all EAs in a single flow chart, a two-stage selection is proposed.

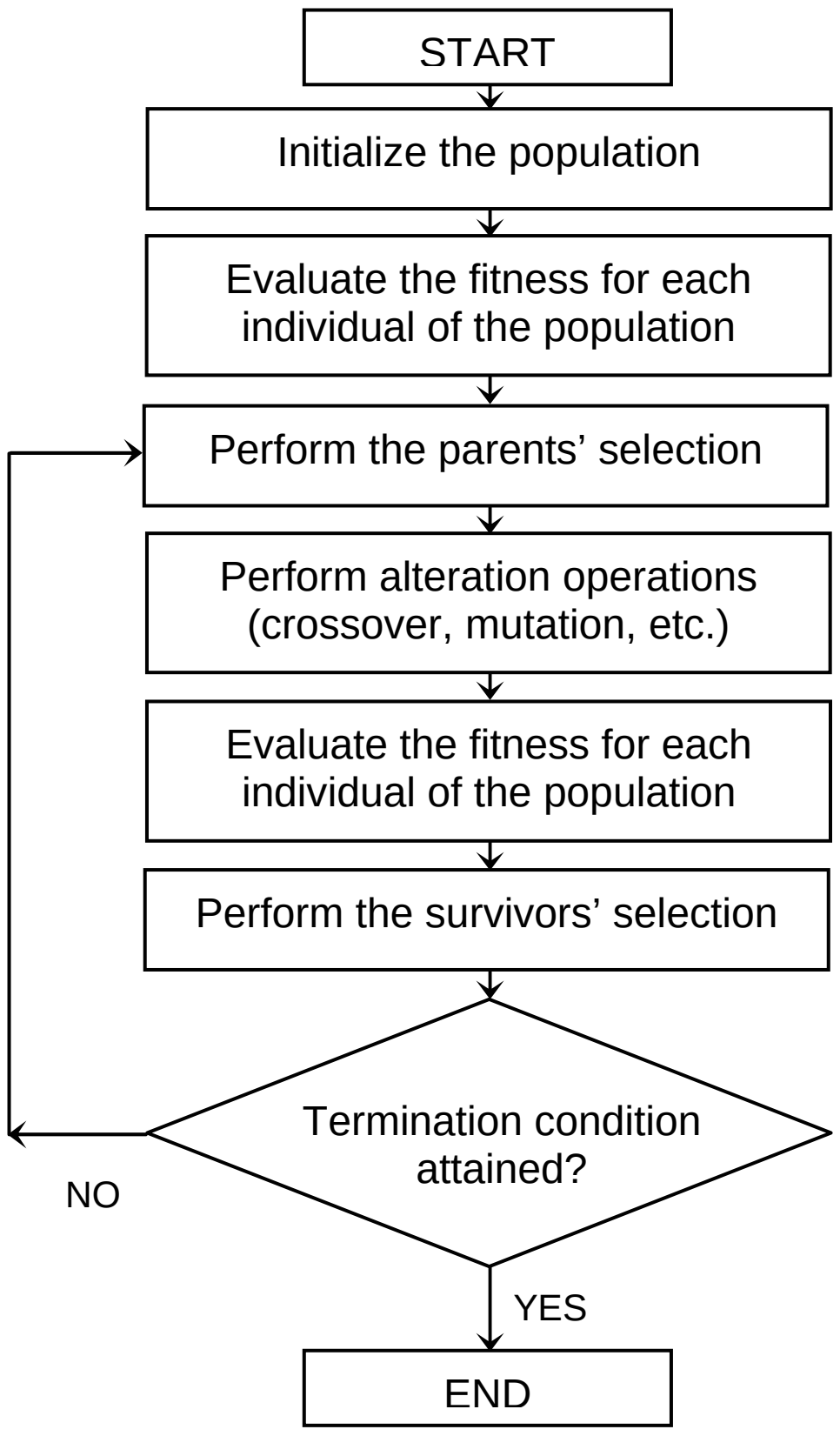

**Fig. 4. 2**: General evolutionary algorithm flow chart.

There are different ways of thinking of and of implementing the selection stages. In this work, the selection of the parents involved in each recombination process is considered to be part of the parents' selection, whereas in the literature, it is typically embedded in the recombination operator. The fundament for doing so is strictly conceptual, and it will be explained in time. Despite sharing the same broad steps, there are important differences between the different mainstreams, and even within the same basic paradigm. Notably, the use of different data structures, of different operators, and even of different attributes within the same operator. Regarding the two selection stages, note that typically only one of them is fitness-based.

## 4.3.1 General Procedure

The general procedure for all EAs is very much the same, as shown in **Fig. 4. 2**. The main differences between them will be explained in the corresponding following chapters.

### 4.3.1.1 Initialization

With regards to the number of individuals involved, Kennedy et al. [47] suggests that …*the total number of individuals chosen to make up the population is both problem and paradigm dependent,* *but it is often in the range of a few dozen to a few hundreds*. The population size is usually kept constant through generations, even though there is no need to make such assumption. It is usually just a question of simplicity.

Regarding the individuals' initial positions, the EAs typically place them stochastically. However, it is sometimes possible and convenient to initialize the population, or a part of it, within promising areas of the search-space. Another interesting alternative consists of initializing the population in such a manner that the individuals achieve a better filling of the whole search-space (assuming it is constrained). In this regard, a well known procedure to do so is the so-called "Latin hypercube sampling" (LHS).

#### 4.3.1.2 Fitness evaluation

The objective function relates the algorithm to the problem being solved. It takes an individual and returns either a scalar or a vector which gives information about the environment's attributes in that particular position of the search-space. The fitness function takes the output from the objective function and returns a scalar that states how good that particular position is. The objective and the fitness functions might be the same, might be proportional, or there might be an arbitrarily complex mapping between them. To sum up, the objective function returns information about the environment, whereas the fitness function establishes how well the individual is performing.

#### 4.3.1.3 Selection operations

**Parents' selection**

It is performed prior to the alteration operations. In GAs, this selection probabilistically picks the fittest individuals within the population, while the same individual is allowed to be picked more than once. Notice that all the selected individuals here are guaranteed to breed. In ESs and EP, the whole population ($\mu$ individuals) is selected at this stage. Notice that all the individuals are then allowed, but not guaranteed, to breed.

In the literature, the parents' selection typically ends up here, while the formation of the "mating couples" is embedded in the recombination operator. In this work, the pairing-up is performed within this stage, since in ESs and EP this is actually the parents' selection.

Therefore, GAs perform a random pairing-up of the individuals already selected.

Differently, ESs successively select $\rho$ parents out of the $\mu$ individuals at random who will breed one descendant per recombination in the next stage. All the individuals typically have the same probability of being selected. This process takes place $\lambda$ times ($\lambda$–fold selection) so that $\lambda$ mating groups composed of $\rho$ parents are created, which will breed one children each per recombination. After the $\lambda$–fold selection procedure, all the picked individuals are indeed guaranteed to breed[21].

**Survivors' selection**

It takes place after the alteration operations. In canonical GAs, all the children produced by the selected parents are selected here, whereas in ESs and EP the children which perform best are selected, in such a manner that the size of the population is typically kept constant.

Notice that the parents' selection is random and the survivors' selection is fitness-based in phenotypic-based EAs, whilst it is the other way round in genotypic-based EAs.

**Selection operations**

One of the most common selection operations is the so-called **roulette wheel selection**, also known as **probabilistic proportional selection**. With this technique, the number of times that an individual is selected is expected to be approximately equal to its present fitness divided by the average fitness of the whole population. Therefore, the probability that each individual has of being selected is calculated by dividing its fitness by the summation of all the individuals' fitness. Then, the individuals are listed, and their cumulative probabilities are calculated so that the first individual's cumulative probability is the probability it has of being picked, whereas the last individual's cumulative probability equals "1". Finally, a random number between "0" and "1" is generated from a uniform distribution, and the first individual whose cumulative probability is higher than the random number generated, is selected. Thus, fitter individuals are more likely to be chosen, although that is not ensured[22]. The name of this procedure is due to the interpretation of the cumulative probabilities as a wheel, where the probability that each individual has of being selected stands for a portion of it, and the random generated number is seen as a spin of the wheel. Each spin selects only one individual.

[21] This is the reason why it has been decided in this work to embed the selection of the $\rho$ individuals involved in each recombination into the parents' selection operator. It is usually claimed in the literature that ESs do not perform selection prior to the alteration operations, or that it selects the whole population, in which case there is an inconsistency with the name "parents' selection", since not all of them are guaranteed to actually become parents. Thus, the parents' selection in this thesis is considered to be a $\lambda$-fold operator that returns the individuals who will certainly breed. Conversely, the selection of the $\rho$ parents is considered in the literature to be an attribute of the recombination operator.

There are some variations of this procedure, one of which guarantees that the fittest individuals are indeed selected. For instance, suppose the same procedure explained before, but the wheel does not have a single pointer but as many pointers as individuals are to be selected, equally separated. Thus, the complete selection is performed in a single spin of the wheel, and the selection of the fittest individuals is ensured. For instance, if an individual has the probability of being selected 1.5 times, at least one time is guaranteed and no more than two are possible, since the distance between pointers equals the hypothetical probability of an individual with an average fitness.

It is very likely in GAs that after a few generations all the population is quite fit, and so the individuals' probabilities are not so different from one another. This makes the roulette wheel work inaccurately. Consider the following example consisting of only three individuals whose fitness values are:

$$f_1 = 38 \quad ; \quad f_1 = 40 \quad ; \quad f_1 = 39 \, .$$

According to the probabilistic proportional selection, the probabilities that the individuals have of being selected by the roulette wheel are:

$$p_1 = \frac{38}{38+40+39} = \frac{38}{117} = 0.325 \quad ; \quad p_2 = \frac{40}{117} = 0.342 \quad ; \quad p_3 = \frac{39}{117} = 0.333$$

As it can be noticed, all of them have been awarded very similar probabilities, making the roulette wheel be almost as likely to select any of them per spin.

There are many different ways of avoiding this problem, such as performing the selection by simply ranking the individuals according to their fitness, providing them with equally spaced probabilities of being selected. This is called the **rank-based roulette wheel selection**, also known as **probabilistic ranking selection**. A simple way of implementing it consists of ranking the individuals according to their fitness, from worst to best (i.e. the worst individual is the one ranked first). Then, the probability that each individual has of being selected is its position in the rank divided by the summation of all the positions. Thus, a probabilistic rank-based selection method is implemented rather than a probabilistic proportional-based one[23].

[22] With this method, the worst individual might survive whilst the best individual might die off, even though the probabilities of any of these happening are negligible.

Making use of the same example as before, the probabilities are now:

$$p_1 = \frac{1}{1+2+3} = 0.167 \quad ; \quad p_2 = \frac{3}{6} = 0.500 \quad ; \quad p_3 = \frac{2}{6} = 0.333$$

Note that the second individual is slightly fitter than the first one, but the probability it has of being selected is noticeably higher.

There are some other techniques to tackle this problem, as well as other problems like the case when one individual is much fitter than the rest, so that it dominates the next progeny, hence losing diversity within the population. Of course, it is not possible to discuss all of them here.

Another common technique for selection is the so-called **tournament selection**. Iteratively, two individuals are randomly selected and their fitness compared, so that the fitter passes to the next generation. This can be generalized to any number of competitors. There are some variations of this method as well, such as considering more than two individuals per tournament, and comparing each one to the rest, counting the number of times each individual's fitness "defeats" the others'. The more successful ones are selected. Clearly, the higher the number of competitors the closer to a deterministic scheme, whereas the lower the more probabilistic the selection becomes. In fact, if the whole population is selected, the method becomes deterministic.

Any selection-scheme may include **elitism**, where only a percentage of the population is replaced, and a small percentage with the fittest individuals is passed to the next generation straightaway, thus guaranteeing the best individuals' survival. Individuals which belong to elitist algorithms do not have their life span limited to a single generation.

#### 4.3.1.4 Alteration operations

In genetic-based EAs, alteration operators are applied to genotypes, attempting to mimic natural genetic transformations as observed in nature. That is, they are applied to the object variables' encoding instead of to the object variables themselves. However, mutation is applied to the genetic code without taking into account the genotype-phenotype mapping, so that the preference of smaller alterations of the traits over larger ones is not guaranteed. Thus, the relation to the natural phenomena they intend to mimic is lost somehow.

[23] Hence, the probability that an individual has of being selected does not depend on its fitness but on its position in the rank. Nevertheless, it is still a fitness-based procedure.

In phenotypic-based EAs, alteration operators are applied to phenotypes in such a way that a desired distribution of new behaviours is obtained. The selection of a particular operator has to be carefully made if the natural analogy is desired to be kept. For instance, although crossover is applied to the behavioural vectors in ESs with proven results, it does not keep the analogy to the biological process, since in nature the collection of phenotypes of an individual is not given by a simple discrete random mixing of their parents' phenotypes, but of their parents' genotypes, followed by the mapping of the recombined genotypes to the phenotypic space. Thus, the link to the natural metaphor is lost here too.

This does not mean that the algorithms should be modified, but that it is not necessary to precisely mimic nature to develop effective and efficient problem-solving techniques. However, it is important to know when natural and computational paths start to bifurcate, since due to the great difficulties of EAs' theories, it is very common to justify approaches, to analyze their accuracy, and even to infer a better performance of one over another, by means of comparisons with nature. Once the algorithm follows its own path, moving away from its original metaphor, these comparisons become less and less applicable.

The traditional assertion that crossover is what makes an algorithm a GA has been widely supported by GAs' practitioners, to the extent that some researchers claim that mutation is not essential, whereas it has been widely criticized by the ESs and EP community. Bäck [5] implemented two GAs, one with and one without crossover, for the optimization of the sphere function (see **Appendix 3**). His experiments showed an almost linear rate of convergence of the latter whereas the former stagnated. Even though it is only a single experiment, which can by no means be generalized, there are many recent experiments arriving to the conclusion that the role of crossover has been traditionally overestimated, whereas the role of the mutations has been underestimated in GAs, through their more than 30 years of history.

For instance, Davis [21] developed a simple metaphor comparing two ideal populations, the "Mutes" and the "Crosses", to illustrate what he thinks might be the reason why crossover is so much more important than mutation. From the point of view of the author of this thesis, the analogy is not appropriate, since Davis awards each population the same mutation rate, with the "Mutes" lacking crossover. In nature, haploids like bacteria do not perform crossover[24], but their mutation rate is much higher than the normal rates in eukaryotes (i.e. in diploids, which reproduce by meiosis thus undergoing crossover). Davis claims that if two different mutations are beneficial for the population, the probabilities of occurrence of one or the other are much higher than the occurrence of both in the same individual, and the crossover would be in charge of eventually putting them together in the same individual. There is no doubt about that, but haploids show a higher rate of mutation in nature, and the probabilities of occurrence of both mutations in the same individual by low-rate mutation plus crossover or just by high-rate mutation are not so clear. The "Mutes" (haploids) should have been awarded a higher rate of mutation than the "Crosses" (diploids) to attempt to mimic natural processes. Furthermore, by applying mutation plus crossover to one population and only mutation at the same rate to the other, only the harmless condition or at most the benefits of crossover can be concluded, but not the importance of one operator over the other. Thus, the comparison seems to be simply not valid.

Furthermore, it seems that the GAs' crossover mechanism does not have a natural counterpart. Haploids reproduce by binary fission, generating clones which might undergo mutation, and can undergo recombination by assimilating genetic material from the environment rather than exchanging it. The way of recombination occurrence most similar to sexual reproduction in prokaryotes is the conjugation, where at least two cells are involved, but there is a cell that gives genetic material and another one that receives it, instead of an exchange. Furthermore, the given genetic material does not belong to the cell's genome. Therefore, the closest analogy to nature might be to think of the chromosomes in GAs as bacteria that reproduce by binary fission and undergo mutation. Crossover is only an artifice to emulate, somehow, recombination. Hence, trying to support the importance of crossover in GAs by making analogies to natural mechanisms might not lead to valid conclusions.

Nevertheless, this does not prove that the claim of the importance of crossover over mutation in GAs is not correct, but that supporting it by natural analogies is not valid. The relative importance of evolutionary mechanisms remains an open question, and only partial conclusions and particular experiments have been developed.

[24] Notice that GAs' chromosomes are haploids. In nature, haploids may undergo different types of recombination, but none of them occurs during sexual reproduction. They reproduce by means of binary fission, thus generating clones. If a clone happens not to be exactly like the original cell is due to mutation only.

Regarding the GAs' low-mutation rate, Schwefel [68] suggests that …*we should not compare EAs for new application tasks with a steady-state near status of organic evolution… …The fact that today mutation rates are very low in most cases observed, may be due to having reached (local) optima or stationary states with respect to current environmental conditions only. In earlier stages it might have been much higher. From theory we learn that optimum mutation rates are inversely proportional to the number of decision variables involved and proportional to the distance from the optimum…*

### 4.3.1.5 Termination condition

The termination condition is problem dependent. Typically, the termination criterion encompasses two sub-criteria, one of which has to be attained:

1. The first one is simply a maximum desired number of trials, generated by multiplying the number of runs of the algorithm by the number of individuals in the population, resulting in the number of individuals generated. Given the desired number of trials, the setting of the population size and the number of runs is always a compromise between these two numbers. A higher number of generations leads to further evolution, whereas a higher number of individuals leads to a wider parallel search, both desirable features.

2. The second sub-criterion consists of terminating the loop if no more improvements are achieved over a number of subsequent iterations. This criterion typically carries the definition of some measures of errors.

Numerous other termination criteria can be thought of according to the problem at hand, although the literature shows lack of interest and developments on this matter.

## 4.3.2 Evolution strategies

ESs are top-down population-based algorithms which belong to the group of phenotypic approaches, so that no attempt is made to model the genetic mechanisms observed in nature.

The individuals are abstracted as vectors of individuals' behavioural traits, whose components are $x_i \in \mathcal{R}$. Following the same broad steps as the rest of the EAs, each individual is evaluated in terms of a predefined fitness function, and the selection is expected to eliminate the individuals with lower performances.

ESs were first introduced by Rechenberg and further developed by Schwefel (from [9]). Although their first applications were on experimental hydrodynamic optimization problems, it was Schwefel himself who first attempted to implement ESs in a computer, though by that time it was not a population-based method but a so-called two-membered evolution strategy.

### 4.3.2.1 The (1+1)-strategy

The method consists of encoding an $n$-dimensional vector, whose components are the object variables to be optimized, usually chosen at random and being mutated to produce a child. The best of the parent and the child is kept for the next generation. This strategy is typically called **(1+1)-strategy** or **(1+1)-ES**, and it can be described by the following 8-tuple[25]:

$$(\mathbf{1}+\mathbf{1})-\mathbf{ES}:\left(\mathbf{P}^{(0)}, m, s, cd, ci, f, g, t\right) \quad \mathbf{(4.\ 1)}$$

Where:

- $\mathbf{P}^{(0)} = \mathbf{a}^{(0)} = \left(\mathbf{x}^{(0)}, \boldsymbol{\sigma}^{(0)}\right) \in I$ initial population consisting of a single individual[26]
- $I = \mathcal{R}^n \text{ x } \mathcal{R}^n = \mathcal{R}^{2n}$ individuals' space (where $\mathcal{R}^n$ is the problem search-space)
- $\mathbf{x}^{(0)} \in \mathcal{R}^n$ initial potential object variables' values
- $\boldsymbol{\sigma}^{(0)} \in \mathcal{R}^n$ initial corresponding standard deviations
- $m : \mathcal{R}^n \to \mathcal{R}^n$ mutation operator (it only affects the object variables)
- $s : I^2 \to I$ survivors' selection operator[27]

[25] It seems fair to remark that despite having been significantly modified and further extended to EP and GAs, this way of representing the evolution strategies by means of $n$-tuples was taken from Bäck et al. [7].

[26] Since this method involves a single individual, the matrix $\mathbf{P}^{(0)}$ is just a (usually row) vector. Commonly, different rows stand for different individuals, whereas different columns do for different coordinates of the individual (object variables). Note that the individual (**a**) is encoded into a vector composed of the object variables vector (**x**) and the standard deviations vector (**σ**) concatenated. Then, it is not the individual (**a**) but a part of it (**x**) which undergoes evolution (i.e. mutation), since the vector of standard deviations (**σ**) is deterministically updated rather than evolved.

[27] Notice that the selection $I^2 \to I$ can be viewed either as taking place from a population of two individuals of the form (**x** , **σ**) $\in I$ to an individual of the same form, or as taking place from an individual (**x** , **σ** , **x'** , **σ'**) $\in I^2$ to an individual (**x** , **σ**) $\in$ $I$. In population-based methods it seems clearer to think of that as different individuals that belong to the space $I$. However, it is fair to clear this up now, for the sake of consistency, since a vector belonging to the space $\mathcal{R}^n$ is usually thought of as an "individual" belonging to an $n$-dimensional space rather than $n$ individuals belonging to the space $\mathcal{R}$ (though this would not be inaccurate).

- $cd, ci \in \mathcal{R}$ step-size control
- $f : \mathcal{R}^n \to \mathcal{R}$ objective function
- $g_j : \mathcal{R}^n \to \mathcal{R}$ $j^{th}$ constraint function
- $t : I^2 \to \{0,1\}$ termination criterion

The initial population, represented by the matrix $\mathbf{P}^{(0)}$ (vector, in this case), is initialized somehow, either by applying some heuristics or randomly. Then, the population's fitness is evaluated and, obviously, the parents' selection does not take place, or it could be trivially viewed as if the only existing individual was selected (see **Fig. 4. 2**).

The only alteration operation in this strategy is the mutation of the object variables $\mathbf{x}^{(t)}$, which is performed by adding a vector of small random numbers, as shown in equation **(4. 2)**. This is in agreement with the fact observed in nature that children are similar to their parents, and that smaller changes occur more frequently than larger ones.

$$\mathbf{x}'^{(t)} = \mathbf{x}^{(t)} + \mathbf{N}_{(0,\boldsymbol{\sigma}'^{(t)})} \tag{4. 2}$$

Where $\mathbf{N}_{(0,\boldsymbol{\sigma}'^{(t)})}$ stands for a vector whose components are independent random numbers generated from Gaussian zero-mean normal distributions with standard deviations equal to the corresponding components of the vector $\boldsymbol{\sigma}'^{(t)}$, which is updated as shown in equation **(4. 6)** before applying the mutation to the object variables.

Thus, an intermediate doubled population is generated by adding together the mutated and the un-mutated individuals:

$$\begin{aligned}
&\mathbf{a}^{(t)} = \mathbf{P}^{(t)} = \left(\mathbf{x}^{(t)}, \boldsymbol{\sigma}^{(t)}\right) \\
&\mathbf{a}'^{(t)} = m\left(\mathbf{P}^{(t)}\right) = \left(m\left(\mathbf{x}^{(t)}\right), \boldsymbol{\sigma}'^{(t)}\right) = \left(\mathbf{x}'^{(t)}, \boldsymbol{\sigma}'^{(t)}\right) \\
&\mathbf{P}'^{(t)} = \begin{pmatrix} \mathbf{a}^{(t)} \\ \mathbf{a}'^{(t)} \end{pmatrix} = \begin{pmatrix} \mathbf{x}^{(t)}, \boldsymbol{\sigma}^{(t)} \\ \mathbf{x}'^{(t)}, \boldsymbol{\sigma}'^{(t)} \end{pmatrix} \in I^2
\end{aligned} \tag{4. 3}$$

Where $\boldsymbol{\sigma}'^{(t)}$ is updated as stated in equation **(4. 6)**.

As shown in equation **(4. 4)**, the survivors' selection operator chooses the fitter individual between the parent and the child, which then survives and becomes parent of the next generation.

$$\mathbf{P}^{(t+1)} = s\left(\mathbf{P'}^{(t)}\right) \Rightarrow \mathbf{a}^{(t+1)} = \begin{cases} \mathbf{a'}^{(t)} \text{ if } f\left(\mathbf{x'}^{(t)}\right) \leq f\left(\mathbf{x}^{(t)}\right) \wedge g_j\left(\mathbf{x'}^{(t)}\right) \leq 0 \quad \forall j \in \{1,\ldots,q\} \\ \mathbf{a}^{(t)} = \mathbf{P}^{(t)} \text{ else} \end{cases} \tag{4. 4}$$

Where $f(\cdot)$ stands for the fitness function and $g_j$ stands for the $j^{\text{th}}$ constraint function.

The iterative process continues until the termination criterion is attained (see **Fig. 4. 2**).

With regards to the "efficacy" of the algorithm, Rechenberg (from [8]) mathematically proved that for regular optimization problems[28] and standard deviations remaining constant over time and the same for every object variable, the global convergence for this strategy can be proven. That is, by having unlimited time, the algorithm is able to find the global optimal solution with probability one (Convergence Theorem), as shown in equation **(4. 5)**.

$$\mathrm{P}\left(\lim_{t\to\infty} \mathbf{x}^{(t)} = \hat{\mathbf{x}}\right) = 1 \quad \text{or} \quad \mathrm{P}\left(\lim_{t\to\infty} f\left(\mathbf{x}^{(t)}\right) = f(\hat{\mathbf{x}})\right) = 1 \quad ; \quad \text{with } \left|f(\hat{\mathbf{x}})\right| < \infty \tag{4. 5}$$

Where $f(\hat{\mathbf{x}})$ is the global optimum and $\hat{\mathbf{x}}$ is the optimum location.

For practical applications, however, it is not the global convergence with probability one what one is most interested in, but rather the capability of the algorithm to find a solution that is better than the best solution known so far.

The "efficiency" is measured by the rate of convergence of the algorithm. Voigt et al. (from [5]) claim that a linear rate of convergence[29] is the best that can be achieved by EAs. Bäck et al. [5] showed that ESs are capable of achieving this rate.

Rechenberg (from [8]) analytically obtained the expressions that rule the "expectations of the convergence rates" and the "probabilities for a successful mutation", for the "sphere" and for the "corridor" model functions, and for $n >> 1$. For the sphere function, refer to **Appendix 3**, whereas a corridor function of width $b$ is given by:

$$f(\mathbf{x}) = f(x_1) = c_0 + c_1 \cdot x_1 \qquad ; \qquad \forall i \in \{2,\ldots,n\} : -\frac{b}{2} \leq x_i \leq \frac{b}{2}$$

[28] For further details on what a regular optimization problem is, refer to [8], page 2.

[29] *The convergence rate is defined by the quotient between the distance covered towards the optimum and the number of trials needed for this distance* [8], page 3.

For these two model functions, he determined the optimum standard deviations in order to maximize the convergence rates[30], realizing that both the maximum convergence rates and the optimum standard deviations were inversely proportional to *n*.

Further, by combining the optimum step sizes (i.e. standard deviations) with the probabilities for a successful mutation for both model functions, and for $n >> 1$, he found out that the optimum probabilities for a successful mutation[31] should be around 1/5. Keep in mind that this value was found for two specific model functions only, for $n >> 1$, and for the **(1+1)-ES**.

Thus, he stated the so-called **1/5-success rule**: *The ratio of successful mutations to all mutations should be 1/5. If it is greater increase the variance and if it is lesser decrease it* (from [8]).

In practice, the standard deviations dynamically adapts as follows:

$$\boldsymbol{\sigma}'^{(t)} = \begin{cases} cd \cdot \boldsymbol{\sigma}^{(t)} & \text{if } p_s^{(t)} < 1/5 \\ ci \cdot \boldsymbol{\sigma}^{(t)} & \text{if } p_s^{(t)} > 1/5 \\ \boldsymbol{\sigma}^{(t)} & \text{if } p_s^{(t)} = 1/5 \end{cases} \qquad \textbf{(4. 6)}$$

Where $p_s^{(t)}$ is the frequency of successful mutations measured over a certain interval of trials, $cd$ is usually less than one, and $ci$ is usually greater than one.

*Schwefel gives reasons to use the factors* $cd = 0.82$ *and* $ci = 1/cd$ *for the adjustment, which should take place every* $n$ *mutations. It should be noted that...the operator* $m$ *consists of a random and a deterministic component, now.* [8]

The equations **(4. 6)** are somehow embedded into the mutation operator *m*, since they update the standard deviations to be used for the object variables' mutation. Notice that the vector of standard deviations of the normal distributions used in equations **(4. 2)** and **(4. 3)** are already updated.

Even though the **1/5-success rule** has been obtained under very particular conditions, it is often used in practice to dynamically adjust the mutation's step sizes for problems that do not match the specifications under which the rule was developed. As it usually happens when seeking the increase of the convergence rate in computer algorithms, the **1/5-success rule** enhances the efficiency in detriment of the robustness[32].

[30] He did so by making the derivative of the "expectations of the convergence rates" with respect to the standard deviations equal to zero.

[31] The optimum probability for a successful mutation is the probability of a successful mutation when the mutation's standard deviation is the optimum (i.e. the one that produces the maximum convergence rate).

### 4.3.2.2 The ($\mu$+1)-strategy

Although the **(1+1)-strategy** is a probabilistic method, it is yet not population-based. Furthermore, it can be thought of as a kind of gradient method because very much like them, it is a point-to-point method where the next point follows the direction stated by the discrete gradient, even though it does not need to use additional information calculated by means of differential calculus. Rechenberg (from [8]) proposed a first population-based algorithm that could therefore mimic sexual reproduction, thus introducing the recombination operator. It works with $\mu$ parents producing only one child per generation, and the best $\mu$ out of the ($\mu$+1) individuals are kept for the next generation. This is called the **($\mu$+1)-strategy,** or **($\mu$+1)-ES**. This strategy can be described by an 11-tuple, as shown in equation **(4. 7)**.

$$(\boldsymbol{\mu}+\mathbf{1})-\mathbf{ES}:\left(\mathbf{P}^{(0)},\mu,s_1,r,m,s_2,cd,ci,f,g,t\right) \quad \textbf{(4. 7)}$$

Where:

- $\mathbf{P}^{(0)} = \begin{pmatrix} \mathbf{a}_1^{(0)} \\ \vdots \\ \mathbf{a}_\mu^{(0)} \end{pmatrix} = \begin{pmatrix} \mathbf{x}_1^{(0)},\boldsymbol{\sigma}_1^{(0)} \\ \vdots \\ \mathbf{x}_\mu^{(0)},\boldsymbol{\sigma}_\mu^{(0)} \end{pmatrix} \in I^\mu$ — initial population consisting of $\mu$ individuals
- $I \;:\; \mathcal{R}^n \text{ x } \mathcal{R}^n = \mathcal{R}^{2n}$ — space of the individuals (where $\mathcal{R}^n$ is the problem search-space)
- $\mu \;>\; 1$ — number of individuals in the population
- $\mathbf{x}_i^{(0)} \in \mathcal{R}^n$ — initial potential object variables’ values for individual $i$
- $\boldsymbol{\sigma}_i^{(0)} \in \mathcal{R}^n$ — initial corresponding standard deviations for individual $i$
- $s_1 \;:\; I^\mu \rightarrow I^2$ — parents’ selection operator
- $r \;:\; I^2 \rightarrow I$ — recombination operator[33]
- $m \;:\; I \rightarrow I$ — mutation operator
- $s_2 \;:\; I^{\mu+1} \rightarrow I^\mu$ — survivors’ selection operator

[32] For further details, refer to Bäck [6] and to Bäck et al. [8].

[33] Note that recombination here is applied between two parents’ phenotypes, whereas in nature recombination occurs between grandparents’ genotypes, during the production of the gametes in meiosis.

- $cd, ci \in \mathcal{R}$ — step-size control
- $f : \mathcal{R}^n \to \mathcal{R}$ — objective function
- $g_j : \mathcal{R}^n \to \mathcal{R}$ — $j^{\text{th}}$ constraint function
- $t : I^\mu \to \{0,1\}$ — termination criterion

The initial population $\mathbf{P}^{(0)}$ is (typically randomly) initialized and its fitness evaluated. Notice that the individuals here are composed by the object variables' vector linked together with the corresponding standard deviations' vector, in the same fashion as in the **(1+1)-strategy**.

During the parents' selection stage (see **Fig. 4. 2**), two out of the $\mu$ individuals are randomly selected, having each member of the population the same probability of being chosen. Thus, a temporary population of two individuals is created $\left(\mathbf{P}'^{(t)}\right)$.

$$\mathbf{P}'^{(t)} = s_1\left(\mathbf{P}^{(t)}\right) \in I^2 \quad ; \quad \text{where } \mathbf{P}^{(t)} \in I^\mu \tag{4. 8}$$

The first alteration operator applied is the recombination between the two selected parents. It generates a child by choosing randomly, with the same probability, each component from either one or the other parent. This kind of recombination is called discrete recombination or uniform crossover, and it is applied both to the object variables and to the standard deviations.

Hence, another temporary intermediate population is created $\left(\mathbf{P}''^{(t)}\right)$.

$$\begin{gathered}
\mathbf{a}''^{(t)}_{\mu+1} = r\left(\mathbf{P}'^{(t)}\right) = r\left(s_1\left(\mathbf{P}^{(t)}\right)\right) = \left(\mathbf{x}''^{(t)}_{\mu+1}, \boldsymbol{\sigma}''^{(t)}_{\mu+1}\right) \in I \\
\mathbf{a}''^{(t)}_{i} = \mathbf{a}^{(t)}_{i} \quad ; \quad i = (1,\ldots,\mu) \\
\mathbf{P}''^{(t)} = \begin{pmatrix} \mathbf{a}''^{(t)}_{1} \\ \vdots \\ \mathbf{a}''^{(t)}_{\mu+1} \end{pmatrix} = \begin{pmatrix} \mathbf{a}^{(t)}_{1} \\ \vdots \\ \mathbf{a}^{(t)}_{\mu} \\ \mathbf{a}''^{(t)}_{\mu+1} \end{pmatrix} \in I^{\mu+1}
\end{gathered} \tag{4. 9}$$

In the same fashion as in the **(1+1)-strategy**, the mutation is applied only to the object variables within the individual ($\mu$+1), whereas the standard deviations update follows the **1/5-success rule**.

It could be arguable whether the individual undergoes evolution (typically, it is said it does not). In any case, the algorithm does not self-adapt.

$$\mathbf{a}'''^{(t)}_{\mu+1} = m\left(\mathbf{a}''^{(t)}_{\mu+1}\right) = m\left(r\left(s_1\left(\mathbf{P}^{(t)}\right)\right)\right) = \left(m\left(\mathbf{x}''^{(t)}_{\mu+1}\right), \boldsymbol{\sigma}'''^{(t)}_{\mu+1}\right) = \left(\mathbf{x}'''^{(t)}_{\mu+1}, \boldsymbol{\sigma}'''^{(t)}_{\mu+1}\right) \in I$$
$$\mathbf{a}'''^{(t)}_{i} = \mathbf{a}''^{(t)}_{i} = \mathbf{a}^{(t)}_{i} \quad ; \quad i = (1,\ldots,\mu)$$
$$\mathbf{P}'''^{(t)} = \begin{pmatrix} \mathbf{a}'''^{(t)}_{1} \\ \vdots \\ \mathbf{a}'''^{(t)}_{\mu+1} \end{pmatrix} = \begin{pmatrix} \mathbf{a}^{(t)}_{1} \\ \vdots \\ \mathbf{a}^{(t)}_{\mu} \\ \mathbf{a}'''^{(t)}_{\mu+1} \end{pmatrix} \in I^{\mu+1} \qquad \textbf{(4. 10)}$$

The survivors' selection operator (see **Fig. 4. 2**) defines the population $\mathbf{P}^{(t+1)}$ ($\mu$ individuals) by eliminating the least fit out of the ($\mu$+1) individuals of the $\mathbf{P}'''^{(t)}$ intermediate population.

$$\mathbf{P}^{(t+1)} = s_2\left(\mathbf{P}'''^{(t)}\right) \in I^{\mu}$$
$$\text{such that } \forall \mathbf{a}'''^{(t)}_{j} = \left(\mathbf{x}'''^{(t)}_{j}, \boldsymbol{\sigma}'''^{(t)}_{j}\right) \ \exists \ \mathbf{a}^{(t+1)}_{i} = \left(\mathbf{x}^{(t+1)}_{i}, \boldsymbol{\sigma}^{(t+1)}_{i}\right) : f\left(\mathbf{x}^{(t+1)}_{i}\right) \le f\left(\mathbf{x}'''^{(t)}_{j}\right) \qquad \textbf{(4. 11)}$$
$$i = 1,\ldots,\mu \quad ; \quad j = 1,\ldots,\mu+1$$

As previously mentioned, the standard deviations are updated before mutation is performed according to the **1/5-success rule**. Thus, mutation is adaptive so that when an optimum is approached, the step size decreases. This is because the successful mutation rate automatically diminishes due to the growing difficulty in achieving further improvements. Of course, the step size does the opposite when the successful rate increases.

Although the standard deviations are altered by recombination, this does not introduce new values but just performs a mixing between the existing ones. The standard deviation vector is dynamically adaptive in these two strategies, although by following a deterministic rule.

### 4.3.2.3 The (*μ*+*λ*)-strategy and the (*μ*, *λ*)-strategy

Later on, Schwefel (from [8]) proposed the **(*μ*, *λ*)-strategy** and the **(*μ*+*λ*)-strategy**, both of which make use of $\mu$ parents to produce $\lambda$ children. From here forth, when referred to ESs, the reference will be to one of these last two strategies.

They both can be described by the same 10-tuple, as shown in equation **(4. 12)**.

$$(\boldsymbol{\mu},\boldsymbol{\lambda})-\mathbf{ES}: \ \left(\mathbf{P}^{(0)}, \mu, \lambda, s_1, r, m, s_2, f, g, t\right)$$
$$(\boldsymbol{\mu}+\boldsymbol{\lambda})-\mathbf{ES}: \ \left(\mathbf{P}^{(0)}, \mu, \lambda, s_1, r, m, s_2, f, g, t\right) \qquad \textbf{(4. 12)}$$

Where:

- $\mathbf{P}^{(0)} = \begin{pmatrix} \mathbf{a}_1^{(0)} \\ \vdots \\ \mathbf{a}_\mu^{(0)} \end{pmatrix} = \begin{pmatrix} \mathbf{x}_1^{(0)}, \boldsymbol{\sigma}_1^{(0)} \\ \vdots \\ \mathbf{x}_\mu^{(0)}, \boldsymbol{\sigma}_\mu^{(0)} \end{pmatrix} \in I^\mu$ initial population consisting of $\mu$ individuals
- $I \; : \; \mathcal{R}^n \text{ x } \mathcal{R}^n = \mathcal{R}^{2n}$ space of the individuals (where $\mathcal{R}^n$ is the problem search-space)
- $\mu \; > \; 1$ number of individuals in the population
- $\lambda \; \geq \; \mu$ number of children for the **($\mu$, $\lambda$)-strategy**
- $\lambda \; \geq \; 1$ number of children for the **($\mu$+$\lambda$)-strategy**
- $\mathbf{x}_i^{(0)} \in \mathcal{R}^n$ initial potential object variables' values for individual *i*
- $\boldsymbol{\sigma}_i^{(0)} \in \mathcal{R}^n$ initial corresponding standard deviations for individual *i*
- $\rho \; > \; 1$ number of individuals involved in the recombination
- $s_1 \; : \; I^\mu \to I^\rho$ parents' selection operator ($\lambda$-fold operator)
- $r \; : \; I^\rho \to I$ recombination operator ($\lambda$-fold operator)
- $m \; : \; I^\lambda \to I^\lambda$ mutation operator
- $s_2 \; : \; I^\lambda \to I^\mu$ survivors' selection operator for the **($\mu$, $\lambda$)-strategy**
- $s_2 \; : \; I^{\mu+\lambda} \to I^\mu$ survivors' selection operator for the **($\mu$+$\lambda$)-strategy**
- $f \; : \; \mathcal{R}^n \to \mathcal{R}$ objective function
- $g_j \; : \; \mathcal{R}^n \to \mathcal{R}$ $j^{th}$ constraint function
- $t \; : \; I^\mu \to \{0,1\}$ termination criterion
- $\Delta\sigma \in \mathcal{R}$ step-size meta control - see equations **(4. 15)** -

Once the elements of the 10-tuple are defined, the algorithm itself is defined, though the tuning of its parameters still remains to be done.

The parameters that are embedded into a certain operator are commonly referred to as attributes of that operator. For instance, $\rho$ and $\Delta\sigma$ are embedded into the parents' selection[34] and into the mutation operators, respectively.

[34] Recall that typically, in the literature, the selection of the $\rho$ individuals involved in each recombination is embedded into the recombination operator, whereas in this work it is embedded into the parents' selection operator, for consistency reasons only (it is the real selection of the individuals which will reproduce!).

#### 4.3.2.3.1 Representation of individuals and fitness evaluation

Following the flow chart in **Fig. 4. 2**, the initial population $\mathbf{P}^{(0)}$ composed of $\mu$ individuals is (typically randomly) initialized, and the individuals' fitness is evaluated.

The main difference with previous ESs is that of evolving the strategy parameters[35]. Previous strategies already included them into the individuals' representation, but only the object variables did actually undergo evolution (i.e. the individuals did not evolve, but a part of them did). The **($\mu$+1)-strategy** made the first step towards the evolution of the strategy parameters, since not only did it include them into the individuals, but it also subjected them to recombination. Recall that it was the first population-based strategy, thus being the first one capable of doing so. However, the step sizes were deterministically updated.

Here the individuals are evolved on their whole, so that the algorithm self-adapts by evolving the parameters that rule the search through the problem space at the same time it evolves the object variables. Thus, both the space of solutions and the space of strategy parameters are searched in parallel, allowing the self-adaptation of the strategy.

#### 4.3.2.3.2 Parents' selection

During the parents' selection stage (see **Fig. 4. 2**), $\rho$ out of the $\mu$ individuals are randomly selected, having each the same probability of being chosen. Thus, a temporary population of $\rho$ individuals is created $\left(\mathbf{P}'^{(t)}\right)$. Typically, $\rho = 2$ or $\rho = \mu$.

$$\mathbf{P}'^{(t)} = s_1\left(\mathbf{P}^{(t)}\right) \in I^{\rho} \tag{4. 13}$$

Where $\mathbf{P}^{(t)} \in I^{\mu}$.

As mentioned before, there are different ways of thinking of and of implementing the parents' selection. Despite the usual claim that parents' selection is not performed in phenotypic-based algorithms, or that the whole population is selected, it is considered here that the selection of the individuals who are to undergo recombination is actually the parents' selection operator.

Notice that this is a $\lambda$–fold operator which selects a subpopulation of $\rho$ individuals per application, among which recombination is to occur.

[35] In fact, only the attributes of the mutation operator are evolved (more precisely, only the standard deviations are).

If only one individual is generated per application of the recombination operator, and $\lambda \geq \mu$, it is very likely that every individual would participate in recombination at least once. For instance, suppose $\mu$ = 20 and $\lambda$ = 30. Then, 60 parents have to be picked among the 20 individuals, having all the same probability of being chosen!

### 4.3.2.3.3 Alteration operations: recombination

Recombination is typically performed $\lambda$ times, generating one child per application. Usually, the recombination operator used for the object variables and the one used for the strategy parameters are different. The child's object variables are generated by choosing randomly, with the same probability, each component from any of the $\rho$ parents (so-called discrete recombination), whereas the child's strategy parameters are commonly generated by the so-called intermediate recombination, consisting of the arithmetic average of the parents' corresponding strategy parameters. The higher the number of parents involved in the recombination, the higher the mixing of the genetic information.

Consider two individuals to be recombined, and the hyper-space[36] spanned between them. By means of discrete recombination, a child can only be placed in one of the vertices of the hyper-polyhedron that contains the mentioned hyper-space. In contrast, intermediate recombination places the child in the middle of the line that joins the parents' positions. Typically, intermediate recombination is capable of overcoming some difficulties of the individuals to reach ridges and ravines (they tend to position themselves either in one or the other side of them), but at the same time it reduces diversity within the population, which is one of the main features EAs rely on. This is the reason why discrete recombination is usually preferred for the object variables and intermediate recombination for the strategy parameters.

Bäck [6] proposes some other alternative recombination techniques to place a child in different areas of the hyper-space spanned between the parents' positions. Notice that a child can by no means be placed outside the mentioned hyper-polyhedron by recombination. For this reason, Bäck [6] claims that recombination causes more or less volume reduction. Conversely, mutation widens the pool.

[36] Hyper-space is the name given to a generic $n$-dimensional space, where $n$ is usually greater than three. However, the terms space and hyper-space are indistinctly used in this thesis. The same is true for similar terms such as plane and hyper-plane, polyhedron and hyper-polyhedron, etc.

Thus, by means of a $\lambda$-fold discrete-intermediate recombination operator, another temporary intermediate population is created $\left(\mathbf{P}''^{(t)}\right)$:

$$\mathbf{P}''^{(t)} = \begin{pmatrix} \mathbf{a}''^{(t)}_1 \\ \vdots \\ \mathbf{a}''^{(t)}_\lambda \end{pmatrix} = r\left(\mathbf{P}'^{(t)}\right)_{\lambda-\text{fold}} = r\left(s_1\left(\mathbf{P}^{(t)}\right)\right)_{\lambda-\text{fold}} = r\begin{pmatrix} \mathbf{a}'^{(t)}_1 \\ \vdots \\ \mathbf{a}'^{(t)}_\rho \end{pmatrix}_{\lambda-\text{fold}} \in I^\lambda$$

$$\mathbf{a}''^{(t)}_i = r\begin{pmatrix} \mathbf{a}'^{(t)}_1 \\ \vdots \\ \mathbf{a}'^{(t)}_\rho \end{pmatrix} \in I \quad ; \quad i = 1,\ldots,\lambda \qquad \textbf{(4. 14)}$$

Note that the parents' selection operator is $\lambda$-fold too. This means that each of the $\lambda$ times the recombination operator is performed, the population $\mathbf{P}'^{(t)}$ (the $\rho$ parents) is resampled anew.

### 4.3.2.3.4 Alteration operations: mutation

Despite the claims in the literature asserting that the strategy parameters self-adapt, only the object variables and the standard deviations of the mutation operator commonly undergo evolution. The evolutions of the recombination operator, the population size, or other attributes of the mutation operator are important issues with very little work done about.

Mutation is typically applied to the object variables and to the standard deviations following different rules, as shown in equations **(4. 15)**.

Thus, each object variable is mutated by simply adding a random number obtained from a Gaussian zero-mean normal distribution, so that smaller perturbations are more frequent than larger ones. Differently, the standard deviations undergo multiplicative logarithmic normal mutations, as proposed by Schwefel (from [68]).

$$\mathbf{P}'''^{(t)} = \begin{pmatrix} \mathbf{a}'''^{(t)}_1 \\ \vdots \\ \mathbf{a}'''^{(t)}_\lambda \end{pmatrix} = \begin{pmatrix} \mathbf{x}'''^{(t)}_1, \boldsymbol{\sigma}'''^{(t)}_1 \\ \vdots \\ \mathbf{x}'''^{(t)}_\lambda, \boldsymbol{\sigma}'''^{(t)}_\lambda \end{pmatrix} = m\left(\mathbf{P}''^{(t)}\right) = m\begin{pmatrix} \mathbf{a}''^{(t)}_1 \\ \vdots \\ \mathbf{a}''^{(t)}_\lambda \end{pmatrix} \in I^\lambda$$

$$\sigma'''^{(t)}_{i,j} = \sigma''^{(t)}_{i,j} \cdot e^{N(0,\Delta\sigma)} \qquad \textbf{(4. 15)}$$

$$x'''^{(t)}_{i,j} = x''^{(t)}_{i,j} + N_{\left(0,\sigma'''^{(t)}_{i,j}\right)} \quad \text{or} \quad \mathbf{x}'''^{(t)}_i = \mathbf{x}''^{(t)}_i + \mathbf{N}_{\left(0,\boldsymbol{\sigma}'''^{(t)}_i\right)}$$

$$i = 1,\ldots,\lambda \quad ; \quad j = 1,\ldots,n$$

Where:

- $N_{(0,\Delta\sigma)}$ : random real number obtained from a Gaussian zero-mean normal distribution with standard deviation $\Delta\sigma$, to be used in the mutation operator, resampled anew for each coordinate, for each individual and for each generation (i.e. resampled anew each time it is referenced)
- $\mathbf{N}_{\left(0,\boldsymbol{\sigma}'''^{(t)}_i\right)}$ : random vector to be used in the mutation operator, resampled anew for each individual and for each generation (i.e. resampled anew each time it is referenced), whose components are random real numbers obtained from Gaussian zero-mean normal distributions with standard deviations equal to the corresponding components of the vector of standard deviations $\boldsymbol{\sigma}'''^{(t)}_i$

Notice that the matrix of standard deviations is mutated before mutating the object variables.

Schwefel claims (from [6, 5, 9]) that the multiplicative logarithmic normal modification of the standard deviations was chosen based on the following heuristic arguments:

- *A multiplicative process preserves positive values.*
- *The median should equal one to guarantee that, on average, a multiplication by a certain value occurs with the same probability as a multiplication by the reciprocal value (i.e., the process would be neutral under absence of selection).*
- *Small modifications should occur more often than large ones.*

### Correlated mutations

By updating the standard deviations independently in every direction of the search-space, the lines of equal probability density to place the offspring would be hyper-spheres if they are the same in every direction, or hyper-ellipsoids if they are not, whose main axes (i.e. preferred directions of search) are aligned with the coordinate axes. In the general case, the best search direction (i.e. the gradient) is not aligned with the coordinate axes, and the trajectory of the population through the search-space zigzags along the gradient. To overcome this effect, which decreases the efficiency of the algorithm, Schwefel (from [8]) introduced correlated standard deviations[37] (see **Fig. 4. 3**). Thus, an individual is now represented as shown in equation **(4. 16)**.

[37] Note that this is exactly the opposite effect looked for in PSO, where the random numbers used to update the velocity of the particles are resampled anew for each component and for each particle, so that the zigzagging diminishes the likelihood of premature convergence into local optima.

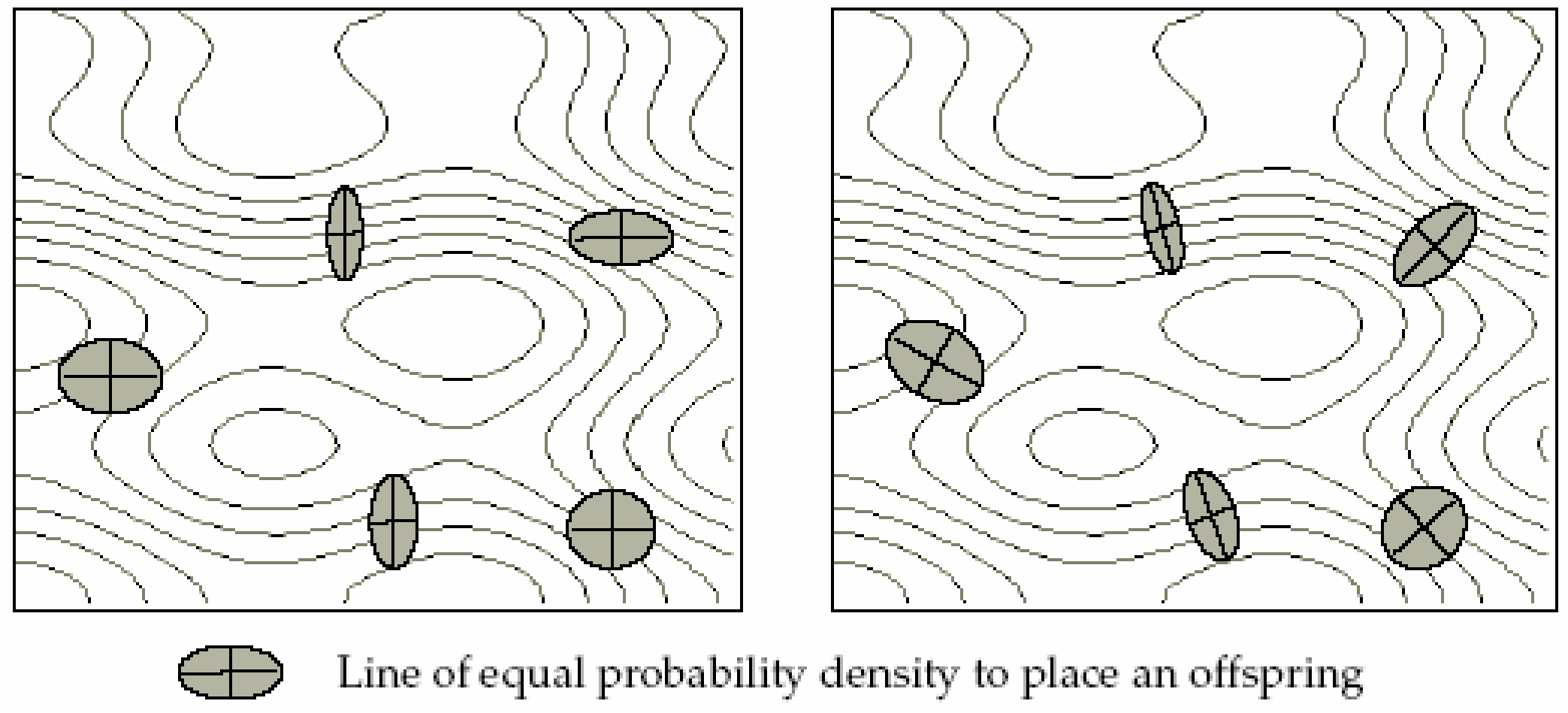

**Fig. 4. 3**: Left: Placing of the offspring by means of independent standard deviations.
Right: Placing of the offspring by means of correlated standard deviations.
The isolines represent equal objective function values, and the grey-shaded ellipses represent individuals' normally distribution of their possible next location, where the centres stand for their present position (from [33]).

$$\mathbf{a}_i^{(t)} = \left(\mathbf{x}_i^{(t)}, \boldsymbol{\sigma}_i^{(t)}, \boldsymbol{\alpha}_i^{(t)}\right) \qquad \textbf{(4. 16)}$$

Where:

- $\mathbf{a}_i^{(t)} \in \mathcal{R}^n \times \mathcal{R}^{n\sigma} \times [-\pi,\pi]^{n\alpha}$ : individual *i* at generation *t*, defined by three float-valued concatenated vectors
- $\mathbf{x}_i^{(t)} \in \mathcal{R}^n$ : point in the problem search-space at generation *t*
- $\boldsymbol{\sigma}_i^{(t)} \in \mathcal{R}^{n\sigma}$ : vector of standard deviations for individual *i* at generation *t*
- $\boldsymbol{\alpha}_i^{(t)} \in [-\pi,\pi]^{n\alpha}$ : vector of rotation angles for individual *i* at generation *t*
- $n\sigma \in \mathcal{N}$ : $1 \le n\sigma \le n$, where *n* is the dimension of the space of solutions[38]

Note that if $n\sigma = 1$, the standard deviations are the same in every direction, whereas if $n\sigma = n$, they are different in each direction of the problem space.

A thorough analysis of the $n\alpha$ parameter is much more complex, and out of the scope of the present work. However, it is worth mentioning that a value of $n\alpha = 0$ eliminates this strategy parameter, so that the vector of standard deviations $\boldsymbol{\sigma}_i^{(t)}$, whose components are uncorrelated, replaces the covariance matrix $\left(\mathbf{C}_i^{(t)}\right)^{-1}$. On the other extreme, a value of $n\alpha = \frac{n\cdot(n-1)}{2}$ means that the vectors $\boldsymbol{\sigma}_i^{(t)}$ and $\boldsymbol{\alpha}_i^{(t)}$ represent the entire covariance matrix of the *n*-dimensional normal distribution, where the vector $\boldsymbol{\alpha}_i^{(t)}$ represents the rotation angles of the hyper-ellipsoids' main axes, necessary to correlate the coordinates of the mutation vector. Note that $\mathbf{C}_i^{(t)} = \mathbf{C}\left(\boldsymbol{\sigma}_i^{(t)}, \boldsymbol{\alpha}_i^{(t)}\right)$.

[38] $\mathcal{N}$ is the set of positive integers.

Recall that the covariance matrix is symmetric and it has $n^2$ components. Thus, apart from the $n$ components of the diagonal (i.e. the squared standard deviations), the number of different components within the covariance matrix is $\frac{n^2-n}{2} = \frac{n\cdot(n-1)}{2}$, which is obviously the same number as the number of rotation angles needed to represent the entire covariance matrix when the standard deviations are different in every direction, and fully correlated.

Mutation is then performed as shown in equations **(4. 17)**.

$$
\begin{aligned}
&\mathbf{P'''}^{(t)} = m\left(\mathbf{P''}^{(t)}\right) = \begin{pmatrix} \mathbf{a'''}_1^{(t)} \\ \vdots \\ \mathbf{a'''}_\lambda^{(t)} \end{pmatrix} = \begin{pmatrix} \mathbf{x'''}_1^{(t)}, \boldsymbol{\sigma'''}_1^{(t)}, \boldsymbol{\alpha'''}_1^{(t)} \\ \vdots \\ \mathbf{x'''}_\lambda^{(t)}, \boldsymbol{\sigma'''}_\lambda^{(t)}, \boldsymbol{\alpha'''}_\lambda^{(t)} \end{pmatrix} \in I^\lambda \quad ; \quad I = \mathcal{R}^n \times \mathcal{R}^{n\sigma} \times [-\pi, \pi]^{n\alpha} \\
&\sigma'''^{(t)}_{i,j} = \sigma''^{(t)}_{i,j} \cdot e^{N_{(0,\Delta\sigma)}} \\
&\alpha'''^{(t)}_{i,k} = \alpha''^{(t)}_{i,k} + N_{(0,\Delta\alpha)} \\
&\mathbf{x'''}_i^{(t)} = \mathbf{x''}_i^{(t)} + \mathbf{N}_{\left(0,\mathbf{C}\left(\boldsymbol{\sigma'''}_i^{(t)}, \boldsymbol{\alpha'''}_i^{(t)}\right)\right)} \\
&i = 1,\ldots,\lambda \quad ; \quad j = 1,\ldots,n\sigma \quad ; \quad k = 1,\ldots,n\alpha
\end{aligned}
\qquad \textbf{(4. 17)}
$$

Where:

- $\mathbf{a}_i^{(t)}$ : individual $i$ at generation $t$, defined by three float-valued concatenated vectors
- $\Delta\alpha$ : step-size meta control for the mutation of the rotation angles; recall that $\alpha_{i,k}^{(t)} \in [-\pi, \pi]$
- $\mathbf{N}_{\left(0,\mathbf{C}\left(\boldsymbol{\sigma}_i^{(t)}, \boldsymbol{\alpha}_i^{(t)}\right)\right)}$ : random vector to be used in the mutation of the object variables, resampled anew for each individual and for each generation, whose components are random real numbers obtained from Gaussian zero-mean normal distributions with correlated standard deviations

It is common to find in the literature the equations **(4. 17)** as shown in equations **(4. 18)**.

$$\mathbf{P}'''^{(t)} = m\left(\mathbf{P}''^{(t)}\right) = \begin{pmatrix} \mathbf{a}'''^{(t)}_1 \\ \vdots \\ \mathbf{a}'''^{(t)}_\lambda \end{pmatrix} = \begin{pmatrix} \mathbf{x}'''^{(t)}_1, \boldsymbol{\sigma}'''^{(t)}_1, \boldsymbol{\alpha}'''^{(t)}_i \\ \vdots \\ \mathbf{x}'''^{(t)}_\lambda, \boldsymbol{\sigma}'''^{(t)}_\lambda, \boldsymbol{\alpha}'''^{(t)}_\lambda \end{pmatrix} \in I^\lambda \quad ; \quad I = \mathcal{R}^n \text{ x } \mathcal{R}^{n\sigma} \text{ x } [-\pi, \pi]^{n\alpha}$$

$$\begin{aligned} &\sigma'''^{(t)}_{i,j} = \sigma''^{(t)}_{i,j} \cdot e^{\tau' \cdot N_{(0,1)} + \tau \cdot \overline{N}_{(0,1)}} \quad ; \quad \text{where } \overline{N}_{(0,1)} = \text{constant} \quad \forall j \\ &\alpha'''^{(t)}_{i,k} = \alpha''^{(t)}_{i,k} + \beta \cdot N_{(0,1)} \\ &\mathbf{x}'''^{(t)}_i = \mathbf{x}''^{(t)}_i + \mathbf{N}_{\left(0, \mathbf{C}\left(\boldsymbol{\sigma}'''^{(t)}_i, \boldsymbol{\alpha}'''^{(t)}_i\right)\right)} \\ &x'''^{(t)}_{i,q} = x''^{(t)}_{i,q} + N_{\left(0, \mathbf{C}\left(\boldsymbol{\sigma}'''^{(t)}_i, \boldsymbol{\alpha}'''^{(t)}_i\right)\right)} \\ &i = 1, \ldots, \lambda \quad ; \quad j = 1, \ldots, n\sigma \quad ; \quad k = 1, \ldots, n\alpha \quad ; \quad q = 1, \ldots, n \end{aligned} \qquad \textbf{(4. 18)}$$

Where:

- $N_{(0,\Delta\alpha)} = \beta \cdot N_{(0,1)}$
- $N_{(0,\Delta\sigma)} = \tau' \cdot N_{(0,1)} + \tau \cdot \overline{N}_{(0,1)}$

Schwefel recommends (from [5]): $\tau \propto \frac{1}{\sqrt{2 \cdot n}}$ ; $\tau' \propto \frac{1}{\sqrt{2 \cdot \sqrt{n}}}$ ; $\beta \cong 0.0873$

Notice that $N_{(0,1)}$ is resampled anew each time it is referenced, while $\overline{N}_{(0,1)}$ is resampled anew for each individual and for each generation, but it is constant within an individual.

The number of strategy parameters included into a single individual can vary from one single parameter for same standard deviation for every problem variable, to $n^2$ parameters for different and correlated standard deviation for each problem variable. It is always a compromise between a great number of parameters and a better adaptation of fewer parameters. Recall that the entire covariance matrix $\left(\mathbf{C}^{(t)}_i\right)^{-1}$ can be represented by the vector of standard deviations $\left(\boldsymbol{\sigma}^{(t)}_i\right)$ plus a vector of rotation angles $\left(\boldsymbol{\alpha}^{(t)}_i\right)$.

Therefore, the maximum number of parameters attached to a single individual (in addition to the object variables) which can undergo mutation, is $n\sigma + n\alpha = n + \frac{n \cdot (n-1)}{2} = \frac{n \cdot (n+1)}{2}$.

Some researchers point out the importance of the strict order in the application of the updates:

The strategy parameters should be mutated first, and then the mutation of the object variables should be performed utilizing the updated strategy parameters. That is, the order of the equations **(4. 15)**, **(4. 17)** and **(4. 18)** should be kept.

#### 4.3.2.3.5 Survivors' selection

In the canonical **($\mu$, $\lambda$)-strategy**, the survivors' selection operator deterministically selects the best $\mu$ out of the $\lambda$ children for the next generation. Note that the best member of a generation might perform worse than the best member of the previous one. This helps sometimes to escape a local optimum by a temporal deterioration. Thus, the life span of an individual is limited to one generation. In the canonical **($\mu$+$\lambda$)-strategy** instead, the survivors' selection operator is deterministic too, but it picks the best $\mu$ out of the ($\mu$+$\lambda$) individuals, thus guaranteeing constant improvement. Hence the life span of an individual is not limited, so that an individual who has reached a local optimum becomes an attractor difficult to escape from.

Alternatively, in non-canonical versions of these two strategies, a tournament selection scheme involving different degrees of determinism can be performed.

$$\begin{aligned}&\text{For } (\boldsymbol{\mu},\boldsymbol{\lambda})-\textbf{strategy}:\\&\mathbf{P}^{(t+1)}=s_2\left(\mathbf{P}'''^{(t)}\right)\in I^{\mu}\quad;\quad s_2:I^{\lambda}\to I^{\mu}\\&\text{such that } \forall \mathbf{a}'''^{(t)}_j=\left(\mathbf{x}'''^{(t)}_j,\boldsymbol{\sigma}'''^{(t)}_j\right)\ \exists\ \mathbf{a}^{(t+1)}_i=\left(\mathbf{x}^{(t+1)}_i,\boldsymbol{\sigma}^{(t+1)}_i\right): f\left(\mathbf{x}^{(t+1)}_i\right)\le f\left(\mathbf{x}'''^{(t)}_j\right)\\&i=1,\ldots,\mu\quad;\quad j=1,\ldots,\lambda\end{aligned} \tag{4. 19}$$

$$\begin{aligned}&\text{For } (\boldsymbol{\mu}+\boldsymbol{\lambda})-\textbf{strategy}:\\&\mathbf{P}^{(t+1)}=s_2\begin{pmatrix}\mathbf{P}^{(t)}\\ \mathbf{P}'''^{(t)}\end{pmatrix}\in I^{\mu}\quad;\quad s_2:I^{\mu+\lambda}\to I^{\mu}\\&\text{such that } \forall \mathbf{a}'''^{(t)}_j=\left(\mathbf{x}'''^{(t)}_j,\boldsymbol{\sigma}'''^{(t)}_j\right)\ \exists\ \mathbf{a}^{(t+1)}_i=\left(\mathbf{x}^{(t+1)}_i,\boldsymbol{\sigma}^{(t+1)}_i\right): f\left(\mathbf{x}^{(t+1)}_i\right)\le f\left(\mathbf{x}'''^{(t)}_j\right)\\&i=1,\ldots,\mu\quad;\quad j=1,\ldots,\mu+\lambda\end{aligned} \tag{4. 20}$$

As opposed to the parents' selection in ESs, the survivors' selection is fitness-based.

The **($\mu$, $\lambda$)-strategy** is usually preferred to the **($\mu$+$\lambda$)-strategy**, although this preference cannot be generalized, since the behaviour of the algorithm is problem dependent.

To summarize, a general form of a modern ES can be described by a 10-tuple, as shown in equations **(4. 21)**. The "tuple" is the same as that one of equation **(4. 12)**, but the definitions of some operators and parameters differ.

$$\begin{aligned}&(\boldsymbol{\mu},\boldsymbol{\lambda})-\textbf{ES}:\ \left(\mathbf{P}^{(0)},\mu,\lambda,s_1,r,m,s_2,f,g,t\right)\\&(\boldsymbol{\mu}+\boldsymbol{\lambda})-\textbf{ES}:\ \left(\mathbf{P}^{(0)},\mu,\lambda,s_1,r,m,s_2,f,g,t\right)\end{aligned} \tag{4. 21}$$

Where:

- $\mathbf{P}^{(0)} = \begin{pmatrix} \mathbf{a}_1^{(0)} \\ \vdots \\ \mathbf{a}_\mu^{(0)} \end{pmatrix} = \begin{pmatrix} \mathbf{x}_1^{(0)}, \boldsymbol{\sigma}_1^{(0)}, \boldsymbol{\alpha}_1^{(0)} \\ \vdots \\ \mathbf{x}_\mu^{(0)}, \boldsymbol{\sigma}_\mu^{(0)}, \boldsymbol{\alpha}_\mu^{(0)} \end{pmatrix} \in I^\mu$ initial population consisting of $\mu$ individuals
- $I$ : $I = \mathcal{R}^n \text{ x } \mathcal{R}^{n\sigma} \text{ x } [-\pi, \pi]^{n\alpha}$ individuals' space (where $\mathcal{R}^n$ is the problem search-space)
- $\mu > 1$ number of individuals in the population
- $\lambda \geq \mu$ number of children for **($\mu$, $\lambda$)-strategy**
- $\lambda \geq 1$ number of children for **($\mu$+$\lambda$)-strategy**
- $\mathbf{x}_i^{(0)} \in \mathcal{R}^n$ initial potential object variables' values for individual *i*
- $\boldsymbol{\sigma}_i^{(0)} \in \mathcal{R}^{n\sigma}$ initial corresponding standard deviations for individual *i*
- $\boldsymbol{\alpha}_i^{(0)} \in [-\pi, \pi]^{n\alpha}$ initial rotation angles for individual *i*
- $\rho > 1$ number of individuals involved in the recombination
- $s_1$ : $I^\mu \to I^\rho$ parents' selection operator ($\lambda$-fold operator)
- $r$ : $I^\rho \to I$ recombination operator ($\lambda$-fold operator)
- $m$ : $I^\lambda \to I^\lambda$ mutation operator
- $s_2$ : $I^\lambda \to I^\mu$ survivors' selection operator for **($\mu$, $\lambda$)-strategy**
- $s_2$ : $I^{\mu+\lambda} \to I^\mu$ survivors' selection operator for **($\mu$+$\lambda$)-strategy**
- $\tau, \tau' \in \mathcal{R}$ parameters that rule the mutation step-size
- $f$ : $\mathcal{R}^n \to \mathcal{R}$ objective / fitness function
- $g_j$ : $\mathcal{R}^n \to \mathcal{R}$ $j^{th}$ constraint function
- $t$ : $I^\mu \to \{0,1\}$ termination criterion

The key development in ESs is the self-adaptation of the strategy and the diversity of the individuals within a population.

*…during self-adaptation of the strategy parameters within evolution strategies, a population can achieve nearly maximal convergence rates as if it knew the optimal parameters a priori. But if one looks, whether at least the best individuals have adapted their internal models consistently to their real world, this is not the case. The near optimal behaviour of the whole stems from the diversity of non-optimal internal models of the individuals … collective intelligence is possible without perfect individual knowledge* [68].

### 4.3.3 Evolutionary programming

EP is, together with ESs, one of the two widest spread population-based algorithms among those which belong to the group of phenotypic approaches. Again, no attempt is made here to model the genetic mechanisms as observed in nature. L. Fogel (from [9]) stated its first roots, whereas D. Fogel (from [9]) reinvented the method quite recently in his PhD. thesis.

EP was originally aimed at AI, but instead of attempting to mimic humans either in their neurophysiologic structure or in their particular behaviours, evolution was thought of as a process of producing increasingly intelligent organisms, under the concept of intelligence as adapting behaviour to achieve goals in a range of environments.

#### 4.3.3.1 Original application: Series prediction

Prediction of the environment together with an appropriate response to that prediction in order to attain a certain goal, were thought to be the keys of the intelligent behaviour. Therefore, a population of finite state machines (FSMs) seemed to be a proper choice to stand for the individuals, whilst the environment was represented by a sequence of symbols taken from a finite alphabet[39]. The process of prediction is performed as follows:

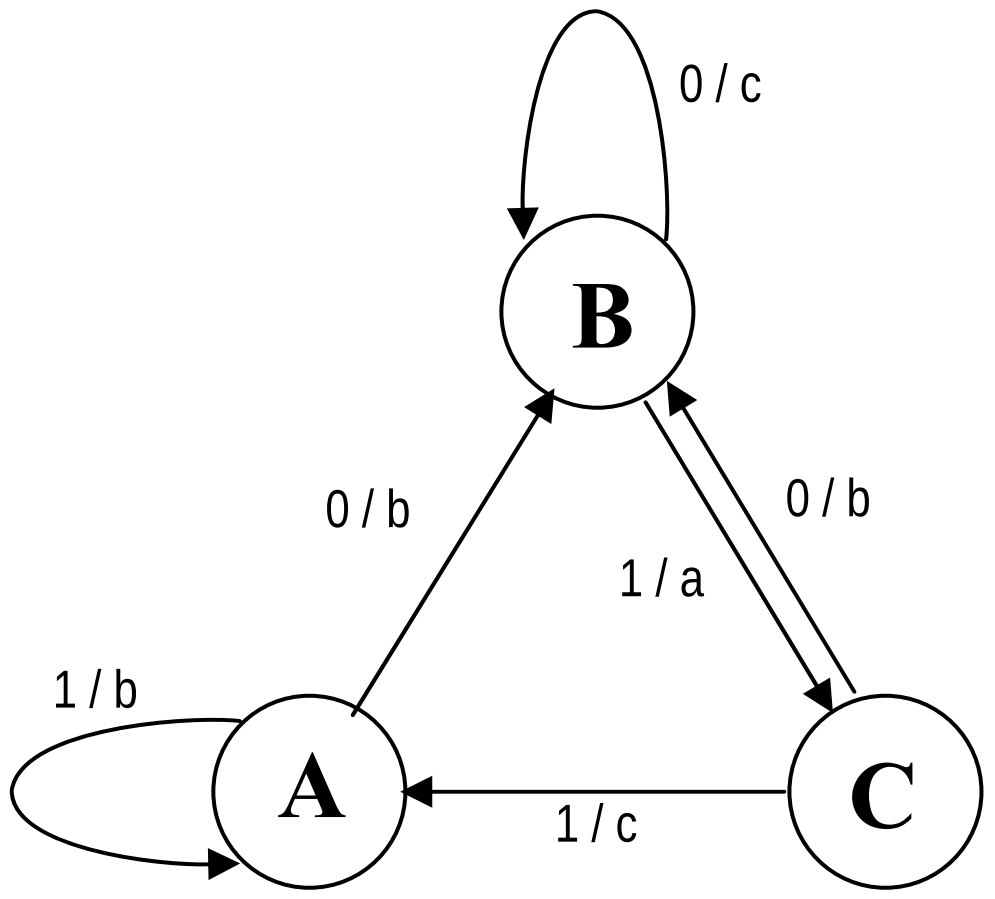

**Fig. 4. 4**: Example of a finite state machine with three states (from [9]).

Only a small part of the environment has already been observed. This experienced environment is represented by a subsequence of symbols. Each FSM's architecture is randomly initialized. An example of a simple FSM (with three possible states: A, B, C; a two-valued input alphabet: 0, 1; and a three-valued output one: a, b, c) is shown in **Fig. 4. 4**. Notice that the output depends both on the input and on the present state of the machine.

Each machine is then exposed to the experienced environment by receiving the inputs which stand for it, as a succession of subsequences, and outputting a symbol for each subsequence that the machine is offered[40]. Each output, which corresponds to each subsequence and to each FSM, is then compared to the next known input symbol. The measure of the prediction's goodness is given by a payoff function for each symbol and for each machine, and the machine's fitness, which stands for the accuracy of the sequence prediction, is calculated after the last prediction is made. Once the fitness of each FSM is calculated, the parents' selection operator selects all the individuals to produce offspring. Since EP does not perform recombination, reproduction can be thought of as a cloning process followed by a mutation process. Another way of thinking of this, supported by some researchers (for instance, Fogel [31]), is to think of the individuals in the algorithm as artificial counterparts of species in nature, since a species does not mate, but it can undergo mutations.

[39] Recall that EP was originally aimed at AI while AI was conducted in the "symbolic paradigm" (see **Chapter 3**), which is based on the concept that all knowledge could be represented in terms of symbols.

Whatever the (rather philosophic) reasons for the algorithm not to perform recombination are, the new population is defined by cloning the last one, and then applying the mutation operator to the cloned population. Both parents' and children's populations consist of $\mu$ individuals. The number of mutations per child might be fixed, or chosen with respect to a probability distribution. There are five possible ways of mutation for FSMs:

- Change an output symbol
- Change a state transition
- Add a state
- Delete a state
- Change the initial state

After the children's population is complete, each FSM's fitness is calculated by exposing it to the environment.

The survivors' selection operator is applied to a population consisting of the adjunction of both the parents' and the children's populations. Thus, the fittest machines among both populations become the next generation. Typically, the fittest half is kept, so that the best $\mu$ individuals are chosen out of the ($\mu$+$\mu$) parents and children. This is called a **($\mu$+$\mu$)-selection** scheme. Note that here the individuals' life span is not limited to one generation.

[40] If $n$ symbols were seen so far, a FSM would predict $n$ symbols as well. The first subsequence is the first symbol known from the experienced environment, and the successive subsequences consist of adding the next symbol so that the last subsequence coincide with the whole sequence known representing the experienced environment.

The process is repeated until a prediction of a new symbol in the environment is required, in which case the best machine predicts it, and the new symbol is added to the experienced environment. From then on, the process is repeated.

The goal here is to predict a series, but before each prediction can be made, some generations are necessary to evolve the machines in the experienced environment. Fogel et al. [34] offer a clear example of the application of this methodology to predict the primeness of the next number of the increasing positive integers' series. Then, the sequence that stands for the environment is a sequence of zeros and ones, where zero means that the number is not a prime number, and one that it is. For instance, suppose that seven symbols are known so far (i.e. the primeness of the first seven increasing integers 1 2 3 4 5 6 7), the subsequence of the experienced environment would be: 0 1 1 0 1 0 1. In his example, Fogel et al. [34] used a population of five FSMs undergoing a single mutation per offspring, and ten generations were performed before predicting a new symbol.

*This general procedure was successfully applied to problems in prediction, identification and automatic control, and was extended to simulate coevolving populations. Additional experiments evolving FSMs (were proposed) for sequence prediction, pattern recognition and gaming* [30].

Whilst the origins of EP were aimed at AI, its present applications are more related to continuous parameter optimization problems, where FSMs are not suitable any longer, and a data structure similar to that of ESs is suggested.

*When EP is applied to problem-solving, it is natural to enquire about its related mathematical properties as an optimization procedure. The most basic result is that the procedure will asymptotically converge to a global optimum of any well-posed problem* (from [30]).

#### 4.3.3.2 Continuous optimization problems

The current existent variants of EP, after Fogel's (from [9]) reinvention of the method, present many similarities to the ESs approach, such as the representation of the individuals, the mutation operator, the selection schemes and the self-adaptation of the strategy parameters.

Analyzing the EP algorithm, it could concluded that it simply does not perform recombination operations, but Fogel [31] claims that this is due to the fact that EP is based upon the behaviours of species, thus making recombination a non-issue.

Thus, the individuals are interpreted as vectors of species' behavioural traits, whose performance is measured by means of a fitness function. Those behaviours with worse performances are more likely to be eliminated by a probabilistic fitness-based selection scheme. New vectors are created by applying mutation operators that mimic the phenotypic traits changes in species.

### 4.3.3.2.1 Representation of individuals and fitness evaluation

A major improvement with respect to the original version of EP is the inclusion of the strategy parameters into the individuals' representation, thus undergoing mutation together with the object variables. In the same fashion as in ESs, it is claimed that the strategy parameters evolve, when in reality only the mutation operator does. Nevertheless, the algorithm self-adapts to the environment.

*The self-adaptation idea was introduced independently from ESs (but 20 years later) into EP* [7].

The representation of the individuals is the same as in ESs, as shown in equation **(4. 22)**.

$$\mathbf{a}_i^{(t)} = \left(\mathbf{x}_i^{(t)}, \boldsymbol{\sigma}_i^{(t)}, \boldsymbol{\alpha}_i^{(t)}\right) \qquad \textbf{(4. 22)}$$

Where:

- $\mathbf{a}_i^{(t)} \in \mathcal{R}^n \text{ x } \mathcal{R}^{n\sigma} \text{ x } [-\pi,\pi]^{n\alpha}$ : individual $i$ at generation $t$, defined by three float-valued concatenated vectors
- $\mathbf{x}_i^{(t)} \in \mathcal{R}^n$ : point in the problem search-space at generation $t$
- $\boldsymbol{\sigma}_i^{(t)} \in \mathcal{R}^{n\sigma}$ : vector of standard deviations for individual $i$ at iteration $t$
- $\boldsymbol{\alpha}_i^{(t)} \in [-\pi,\pi]^{n\alpha}$ : vector of rotation angles for individual $i$ at generation $t$
- $n\sigma \in \mathcal{N}$ : $1 \le n\sigma \le n$, where $n$ is the dimension of the problem space
- $n\alpha \in \mathcal{N}$ : $0 \le n\alpha \le \dfrac{n\cdot(n-1)}{2}$
- $\left(\mathbf{C}_i^{(t)}\right)^{-1}$ : covariance matrix for individual $i$ at generation $t$, whose whole information is contained in the vectors $\boldsymbol{\sigma}_i^{(t)}$ and $\boldsymbol{\alpha}_i^{(t)}$

If $n\sigma = 1$, the standard deviations are the same in every direction, whereas if $n\sigma = n$, they are different in each direction of the problem space. If $n\alpha = 0$, the standard deviations are uncorrelated (and the covariance matrix becomes just a vector of standard deviations), whereas if $n\alpha = \frac{n \cdot (n-1)}{2}$, all the components of the covariance matrix outside the diagonal are independent. The covariance matrix size is $n$ by $n$.

This is exactly the same representation as explained before in ESs, so that all the parameters have the same meaning. Making use of the same nomenclature as in ESs, the EP paradigm could be defined by a 7-tuple, as shown in equation **(4. 23)**.

$$(\boldsymbol{\mu}+\boldsymbol{\mu})-\mathbf{EP}:\left(\mathbf{P}^{(0)},\mu,m,s,f,g,t\right) \tag{4. 23}$$

Where:

- $\mathbf{P}^{(0)} = \begin{pmatrix} \mathbf{a}_1^{(0)} \\ \vdots \\ \mathbf{a}_\mu^{(0)} \end{pmatrix} = \begin{pmatrix} \mathbf{x}_1^{(0)}, \boldsymbol{\sigma}_1^{(0)}, \boldsymbol{\alpha}_1^{(0)} \\ \vdots \\ \mathbf{x}_\mu^{(0)}, \boldsymbol{\sigma}_\mu^{(0)}, \boldsymbol{\alpha}_\mu^{(0)} \end{pmatrix} \in I^\mu$ initial population consisting of $\mu$ individuals

| | |
|---|---|
| - $I \;:\; \mathcal{R}^n \text{ x } \mathcal{R}^{n\sigma} \text{ x } [-\pi,\pi]^{n\alpha}$ | space of the individuals (where $\mathcal{R}^n$ is the problem search-space) |
| - $\mu \;> 1$ | number of individuals in the population |
| - $\mathbf{x}_i^{(0)} \in \mathcal{R}^n$ | initial potential object variables' values for individual $i$ |
| - $\boldsymbol{\sigma}_i^{(0)} \in \mathcal{R}^{n\sigma}$ | initial corresponding standard deviations for individual $i$ |
| - $\boldsymbol{\alpha}_i^{(0)} \in [-\pi,\pi]^{n\alpha}$ | initial rotation angles for individual $i$ |
| - $m \;:\; I^\mu \to I^\mu$ | mutation operator |
| - $s \;:\; I^{\mu+\mu} \to I^\mu$ | survivors' selection operator |
| - $f \;:\; \mathcal{R}^n \to \mathcal{R}$ | objective / fitness function |
| - $g_j \;:\; \mathcal{R}^n \to \mathcal{R}$ | $j^{\text{th}}$ constraint function |
| - $t \;:\; I^\mu \to \{0,1\}$ | termination criterion |

As shown above, there is not much difference between the representation of individuals in ESs and EP for continuous optimization problems. However, numerous other data structures might be used for different kinds of problems.

Following the flow chart shown in **Fig. 4. 2**, a population of $\mu$ individuals is initialized and their fitness values are evaluated. The parents' selection is not performed (or the whole population is selected), so that all the $\mu$ individuals are picked to undergo evolution. More precisely, the $\mu$ individuals are cloned to undergo mutation.

#### 4.3.3.2.2 Alteration operations: mutation

Saravanan and Fogel (from [9, 33]), propose a zero-mean multivariate random variable to mutate the object variables (in the same fashion as in ESs), and an additive equation to self-adapt the algorithm, as shown in equations **(4. 24)**.

$$\mathbf{P'}^{(t)} = m\left(\mathbf{P}^{(t)}\right) = \begin{pmatrix} \mathbf{a'}_1^{(t)} \\ \vdots \\ \mathbf{a'}_\mu^{(t)} \end{pmatrix} = \begin{pmatrix} \mathbf{x'}_1^{(t)}, \boldsymbol{\sigma'}_1^{(t)}, \boldsymbol{\alpha'}_1^{(t)} \\ \vdots \\ \mathbf{x'}_\mu^{(t)}, \boldsymbol{\sigma'}_\mu^{(t)}, \boldsymbol{\alpha'}_\mu^{(t)} \end{pmatrix} \in I^\mu \quad ; \quad I = \mathcal{R}^n \text{ x } \mathcal{R}^{n\sigma} \text{ x } [-\pi, \pi]^{n\alpha}$$

$$\left.\begin{aligned} \boldsymbol{\sigma'}_i^{(t)} &= \boldsymbol{\sigma}_i^{(t)} + k \cdot \boldsymbol{\sigma}_i^{(t)} \cdot N_{(0,1)} \\ \alpha'^{(t)}_{i,j} &= \alpha^{(t)}_{i,j} + \beta \cdot N_{(0,1)} \end{aligned}\right\} \quad ; \quad i = 1,\ldots,\mu \quad ; \quad j = 1,\ldots,n\alpha \qquad \textbf{(4. 24)}$$

$$\mathbf{x'}_i^{(t)} = \mathbf{x}_i^{(t)} + \mathbf{N}_{\left(0,\mathbf{C}\left(\boldsymbol{\sigma'}_i^{(t)}, \boldsymbol{\alpha'}_i^{(t)}\right)\right)}$$

Where:

- $N_{(0,1)}$ : random real number generated from a Gaussian zero-mean normal distribution with standard deviation equal to one, to be used in the mutation operator, resampled anew each time it is referenced.
- $k \in \mathcal{R}$ : scaling parameter ($k \approx 0.2$ according to [7])
- $\mathbf{N}_i^{(t)}{}_{\left(0,\mathbf{C}\left(\boldsymbol{\sigma}_i^{(t)}, \boldsymbol{\alpha}_i^{(t)}\right)\right)}$ : random vector to be used in the mutation of individual *i*, resampled anew for iteration *t*, whose components are random real numbers obtained from Gaussian zero-mean normal distributions with correlated standard deviations

If any $\sigma'^{(t)}_{i,j}$ results in a value less than zero, it is reset to a very small arbitrary positive value.

Notice that the random number $N_{(0,1)}$ is generated anew for each component, for each individual, and for each generation for the update of the rotation angles $\alpha'^{(t)}_{i,j}$, while it is resampled anew only for each individual and for each generation for the update of the standard deviations $\boldsymbol{\sigma'}_i^{(t)}$.

Trials have been performed by exchanging the additive equation to self-adapt the standard deviations for the logarithmic normal distribution previously proposed by Schwefel (from [68]) in ESs. This alternative performs the mutations according to the equations **(4. 25)**.

$$\mathbf{P'}^{(t)} = m\left(\mathbf{P}^{(t)}\right) = \begin{pmatrix} \mathbf{a'}_1^{(t)} \\ \vdots \\ \mathbf{a'}_\mu^{(t)} \end{pmatrix} = \begin{pmatrix} \mathbf{x'}_1^{(t)}, \boldsymbol{\sigma'}_1^{(t)}, \boldsymbol{\alpha'}_1^{(t)} \\ \vdots \\ \mathbf{x'}_\mu^{(t)}, \boldsymbol{\sigma'}_\mu^{(t)}, \boldsymbol{\alpha'}_\mu^{(t)} \end{pmatrix} \in I^\mu \quad ; \quad I = \mathcal{R}^n \text{ x } \mathcal{R}^{n\sigma} \text{ x } [-\pi, \pi]^{n\alpha}$$

$$\sigma'^{(t)}_{i,j} = \sigma^{(t)}_{i,j} \cdot e^{\tau' \cdot N_{(0,1)} + \tau \cdot \overline{N}_{(0,1)}} \quad ; \quad \text{where } \overline{N}_{(0,1)} = \text{constant} \quad \forall j \qquad \textbf{(4. 25)}$$

$$\alpha'^{(t)}_{i,j} = \alpha^{(t)}_{i,j} + \beta \cdot N_{(0,1)}$$

$$\mathbf{x'}^{(t)}_i = \mathbf{x}^{(t)}_i + \mathbf{N}_{\left(0, \mathbf{C}\left(\boldsymbol{\sigma'}^{(t)}_i, \boldsymbol{\alpha'}^{(t)}_i\right)\right)}$$

$$i = 1, \ldots, \mu \quad ; \quad j = 1, \ldots, n\alpha$$

Where $\tau$, $\tau'$ are operator-set parameters that rule the adaptive mutation step-sizes. Schwefel recommends (from [5]): $\tau \propto \frac{1}{\sqrt{2 \cdot n}}$ ; $\tau' \propto \frac{1}{\sqrt{2 \cdot \sqrt{n}}}$ ; $\beta \cong 0.0873$

Notice that $N_{(0,1)}$ is resampled anew each time it is referenced, while $\overline{N}_{(0,1)}$ is resampled anew for each individual and for each generation, but it is constant within an individual.

Both approaches were developed independently. Even though it is widely agreed that Schwefel's (from [68]) approach performs better, some experiments showed that the additive modification of the standard deviations as proposed by Fogel et al. (from [9, 33]) may perform better for noisy objective functions. Furthermore, it has been demonstrated that the addition update derives from the logarithmic update for the same standard deviation in every dimension by a Taylor expansion breaking off after the linear term. Therefore, both approaches should behave very similarly for small settings of the learning rates $\tau$ and $k$. This prediction is confirmed by experiments in [9].

#### 4.3.3.2.3 Survivors' selection

Once the $\mu$ cloned individuals are mutated, an intermediate doubled population consisting of both, the original plus the cloned-mutated populations, is created. For the survivors' selection, $\mu$ out of the ($\mu$+$\mu$) individuals are selected for the next generation. Originally, a probabilistic fitness-based selection scheme was implemented. However, since it is still an open question in EAs whether a deterministic or a probabilistic selection scheme should be preferred, a stochastic competition was introduced, as previously explained in chapter **4.3.1.3**.

Note that there are several parameters embedded in the operators, such as $\tau$, $\tau'$, $k$, $\beta$, $n\alpha$, $n\sigma$, the number of competitors within each tournament, etc. Therefore, once the elements within the 7-tuple in equation **(4. 23)** are completely defined, the optimizer is completely defined itself, but the algorithm's parameters still remain to be tuned. In fact, the main disadvantage of EAs in general is precisely the need of tuning a high number of parameters!

### 4.3.3.3 Other applications

The possible applications for EAs in general are numberless. Two typical problems that can be handled by EP are the TSP and the training of ANNs, as briefly explained hereafter. The representation of individuals depends on the problem at issue, so that a wide variety of representations and operators can be thought of and implemented, provided a strong behavioural link between each parent and its progeny is maintained.

#### 4.3.3.3.1 The travelling salesman problem

The problem consists of finding the permutation of cities that generates the minimum length of a tour that visits all of them only once, and the trip ends in the same city where it started. The problem is NP-hard, and the number of possible solutions increases as $\frac{(n-1)!}{2}$, where $n$ is the number of cities. Clearly, the problem becomes practically unsolvable very soon, and the aim is usually at finding good solutions quickly rather than the global best.

The immediate representation consists of a list of the cities in the order to be visited, and the possible solutions are permutations of this list. The genetic pool does not need to be widen in this case, since all the information is contained within the initialized population (in fact, within each individual!). Hence, the mutation operator is replaced by an inversion operator, which simply swaps two cities in the tour. Thus, a population of random tours is initialized and their fitness is calculated according to the length of the tour, and the inversion operator is applied to produce progeny. A stochastic competition tournament is then implemented to select the tours to be maintained for the next generation. The improvement of the best tour in the population is quite rapid. *Fogel estimated that the number of tours that must be examined in order to achieve solutions which are on average 10% worse than the expected best…increases as the square of the number of cities* (from [34]).

Note that the number of tours to be examined equals the product of the number of individuals by the number of generations that the population is allowed to evolve through. The bigger the population size, the more parallel the search becomes, and the bigger the number of generations allowed, the further the evolution can go. Given a fixed number of tours to examine, it is always a compromise between these two numbers.

#### 4.3.3.3.2 Artificial neural networks

*An artificial neural network is an analysis paradigm that is roughly modelled after the massively parallel structure of the brain. It simulates a highly interconnected, parallel computational structure with many relatively simple individual processing elements* [47] (see section **3.5** for a detailed review).

Basically, they are nonlinear mapping functions that seek to find the relation between some input and output uncorrelated data. It consists of some simple PEs and different ways of interconnecting them. The most common network is the MLP, which is composed of a set of inputs, at least one hidden layer, and an output layer. Each node in one layer is connected to all the nodes belonging to the next layer, but nodes within a layer are not connected to each other. Each node typically performs a weighted summation of its inputs, subtracts off a variable bias, apply a so-called "transfer function", and then passes on the result. It belongs to the family of feed-forward networks because the information is passed in one direction only. There is no feed-back[41].

EP is an alternative to traditional gradient-based training methods. The aim is at recognizing the pattern that relates some known input and output data, so that when the network is offered another set of inputs, supposed to be ruled by the same laws as the training sets are, the ANN is expected to predict the unknown output data. The representation of the individuals here is a real-valued vector, where each component corresponds to a weight or a bias. The mutation is performed by adding a multivariate zero-mean Gaussian random variable to each vector, thus altering all components simultaneously. For each set of weights and bias, the accuracy of the solution is measured for example by calculating the standard mean square error between the known outputs and the ones obtained with these parameters. The rate of change from parents to progeny can be controlled by making the mutations proportional to the mean square error, so that when the solutions are approaching the optimum (i.e. the error is approaching zero) the mutation rate diminishes.

[41] There are many other more complex networks, with less restrictive structures, which are not dealt with in this thesis. For instance, there can be feed-back and connections between non-consecutive layers, as well as connections between nodes within a single layer. The alternatives are unlimited.

EP is a flexible approach that can easily be applied to other problems related to an ANN. For instance, training the latter is a task performed once its architecture, which has to be designed somehow, is already defined. This architecture can also be subject to optimization, thus undergoing evolution itself. Kennedy et al. [47] offer: *Evolutionary computation techniques have most commonly been used to evolve neural network weights, but have sometimes been used to evolve neural network structure or the neural network learning algorithm*.

For further details on the ANNs' paradigm, refer to section **3.5**, and for details on the training of ANNs, refer to **Chapter 12** and to **Appendix 1**. Keep in mind that the ANNs' field has been dealt with for many decades, and the aim of this work is not at doing specific research on the paradigm, but at showing the goodness of the PSO method as an alternative, effective, robust, and easy-to-implement training technique (a major issue in that field).

## 4.3.4 Genetic algorithms

GAs are population-based algorithms that belong to the group of genotypic approaches. They are bottom-up approaches that attempt to mimic genetic processes as observed in nature.

The origins of the GAs are not clear. The first approaches to what could be called the basis of the field go back to the 1950s, when biologists using computers attempted to simulate natural genetic systems, even though they apparently did not foresee the possibilities of applying their methodologies to solve optimization problems. Nowadays, however, these algorithms happen to be mainly problem-solving techniques rather than real life processes simulators.

Mitchell [55] claims that …*genetic algorithms (GAs) were invented by John Holland in the 1960s and were developed by Holland and his students and colleagues at the University of Michigan in the 1960s and the 1970s. In contrast with evolution strategies and evolutionary programming, Holland's original goal was not to design algorithms to solve specific problems, but rather to formally study the phenomenon of adaptation as it occurs in nature and to develop ways in which the mechanisms of natural adaptation might be imported into computer systems. Holland's 1975 book "Adaptation in Natural and Artificial Systems" presented the genetic algorithm as an abstraction of biological evolution and gave a theoretical framework for adaptation under the GA*.

Although the approach is widely used as an optimization tool, it is often claimed to perform adaptation rather than optimization.

Davies [21] suggests that …*with GAs we are not optimizing; we are creating conditions in which optimization occurs, as it may have occurred in the natural world…*

The GAs' are the EAs which have become most popular. Therefore, there are countless variants enhancing different features, which cannot be analyzed in a few pages. For this reason, the field is reduced in this work to its canonical form: a fixed-length-binary representation of individuals, generational replacement (i.e. individuals' life span limited to one generation), a one-point crossover operator, a mutation operator, and a probabilistic proportional selection scheme. A few variants are however mentioned.

Following the same nomenclature as for ESs and EP, what is not frequent in the literature, the canonical GA could be represented by a 9-tuple, as shown in equation **(4. 26)**. Keep in mind that there are numberless variations of the version presented here.

$$(\boldsymbol{\mu},\boldsymbol{\mu})-\mathbf{GA}=\left(\mathbf{P}^{(0)},\mu,s,r,m,h,f,g,t\right) \tag{4. 26}$$

Where:

| | | | |
|---|---|---|---|
| - $\mathbf{P}^{(0)}$ | = | $\begin{pmatrix}\mathbf{a}_1^{(0)}\\ \vdots \\ \mathbf{a}_\mu^{(0)}\end{pmatrix}\in I^\mu$ | initial population consisting of $\mu$ individuals |
| - $I$ | : | $\{0,1\}^m$ | binary space which each individual's genotype belongs to |
| - $\mu$ | $>$ | 1 | number of individuals in the population |
| - $\mathbf{a}_i^{(0)}$ | $\in$ | $I$ | initial genotype of individual $i$ |
| - $s$ | : | $I^\mu \rightarrow I^2$ | parents' selection operator (in this work, but not typically, considered to be a $\mu$-fold operator) |
| - $r$ | : | $I^2 \rightarrow I$ | recombination operator ($\mu$-fold operator) |
| - $m$ | : | $I^\mu \rightarrow I^\mu$ | mutation operator |
| - $\mathbf{x}_i^{(t)}$ | $\in$ | $\mathcal{R}^n$ | object variables (phenotype of individual $i$ at generation $t$) |
| - $h$ | : | $\{0,1\}^m \rightarrow \mathcal{R}^n$ | genotype-to-phenotype mapping operator[42] |

[42] Note that this operator is designed without mimicking the natural links between the genotypic and phenotypic state spaces. This is because GAs work only in the genotypic state space, making use of a convenient genotype-to-phenotype mapping operator (*h*) to evaluate the performance of each individual, but the phenotypes (i.e. the object variables) do not intervene in the genetic processes but only in the survivors' selection stage. In fact, there is no need of a phenotype-to-genotype operator.

- $f$ : $\mathcal{R}^n \rightarrow \mathcal{R}$ objective / fitness function
- $g_j$ : $\mathcal{R}^n \rightarrow \mathcal{R}$ $j^{\text{th}}$ constraint function
- $t$ : $I^\mu \rightarrow \{0,1\}$ termination criterion

### 4.3.4.1 Representation of individuals and fitness evaluation

The GAs' metaphor is the natural evolution through genetic inheritance that occurs at the level of the individual during reproduction. Therefore, and since a GA is a genotypic approach, each individual is represented by its genetic code. Each individual is thought of as a simple organism, whose genetic material is contained within a single chromosome[43]. Each chromosome is encoded in a string[44] representing a set of object variables, which stand for a proposed solution to the problem at issue. Each object variable can be encoded in any number of bits according to the precision required, so that the number of bits composing each chromosome ($m$) would be equal to the product between the number of variables ($n$) and the number of bits desired to represent each variable ($s$).

The total number of bits within each chromosome stands for the dimension of the binary hyper-space to be searched, so that the problem consists of searching an $m$-dimensional hyper-cube, whose vertices represent every possible solution. Many other data structures are also possible. In fact, *...today the possibilities of implementing high-level data-structures as for instance real numbers, have improved greatly, consequently the preferred encoding is no longer binary (at least on real-valued problem-domains)... ...However, binary encoding is highly relevant when faced with a discrete problem. Virtually any discrete problem can be transformed into a bit string representation, and in this light, binary encoding is a general and flexible representation* [78].

For the initial population, a GA creates a set of chromosomes, either randomly or by design. Each chromosome, which stands for the genotype of the individual, is decoded into a set of object variables, which stand for the phenotype of the individual. In turn, the phenotypes are evaluated by means of an objective/fitness function. Notice that a GA works with the representation of the object variables instead of with the variables themselves!

---

43 They are also called "genome", since for organisms composed by a single "chromosome", both terms are equivalent. Notice that due to the peculiarity of GAs' individuals of single-chromosomal beings, they are the artificial counterparts of haploids in nature, such as bacteria (see **Appendix 2**).

44 In a canonical GA, an individual is binary-encoded in a fixed-length-bit-string.

It seems reasonable that mimicking a discrete process by means of discrete entities leads to an algorithm suitable for discrete problems. In fact, applications of GAs to combinatorial problems are straightforward, since almost any discrete problem can be expressed as a pseudoboolean problem[45]. However, surprisingly, many implementations of GAs are not developed to deal with pseudoboolean optimization problems of the form:

$$f : \mathcal{F} \subseteq \{0,1\}^m \rightarrow \mathcal{R} \tag{4. 27}$$

Where $m$ is the dimension of the hyper-cube (i.e. of the bit-string).

Instead, they are frequently aimed at solving continuous optimization problems of the form:

$$f : \mathcal{F} \subseteq \mathcal{R}^n \rightarrow \mathcal{R} \tag{4. 28}$$

Where $n$ is the number of object variables

Of course, some adaptations are required. A typical strategy to deal with continuous search-spaces is shortly described in [21, 5]. The string is subdivided into $n$ segments of equal length, and each binary encoded segment is decoded into a positive integer. For each object variable, the desired range of real values must be set, so that an interval $\lfloor u_j, v_j \rfloor \subset \mathcal{R}$ of real values is defined for each segment of the string. Finally, the integer corresponding to each segment is linearly mapped to that interval.

Let $s$ be the number of bits assigned to each variable, so that $m = n \cdot s$. Then, $2^s$ different integers can be generated in the substring, from 0 to $(2^s - 1)$.

The mapping of the $j^{th}$ object variable for individual $i$ is then performed as follows:

$$x_{ij} = \left( \frac{v_j - u_j}{2^s - 1} \right) \cdot \tilde{x}_{ij} + u_j \tag{4. 29}$$

Where $\tilde{x}_{i,j}$ is the decoded integer, and $x_{i,j}$ is the $j^{th}$ object variable value for individual $i$.

For example, let $u_j = -100$ , $v_j = 100$ and $s = 8$ (each object variable is encoded in 8 bits).

If the decoded integer were zero (minimum possible), then the mapping would be:

$$\left( \frac{200}{255} \right) \cdot 0 - 100 = -100$$

---

[45] In pseudoboolean optimization problems, a hyper-cube (whose vertices represent all possible solutions) is spanned so that the coordinates of the global optimum correspond to one of these vertices.

If the decoded integer were $\left(2^s - 1\right) = 255$ (maximum possible), then the mapping would be:

$$\left(\frac{200}{255}\right) \cdot 255 - 100 = 100$$

However, the discretization of the originally continuous search-space is not avoided, and the distance between consecutive points in the resulting grid depends critically on the number of bits of each segment. In the example above, the number of bits used to represent each variable is clearly insufficient, since only 256 numbers can be generated, to represent an interval containing 201 integers. The resulting grid is extremely coarse. In fact, the distance between possible-to-find consecutive real numbers is $\frac{200}{255} = 0.784313...$

This discretization may lead to a situation where a GA fails to find the global optimum just because its coordinates are not among the grid points. For instance, suppose that the optimum corresponds to a position in the problem space where one coordinate is zero. Then, the best that the algorithm could do is to find the binary string which corresponds to that coordinate decoded into 127 (which, after the mapping shown in equation **(4. 29)**, returns -0.3921…), or the one decoded into 128 (which, after the mapping returns 0.3921…).

Another problem of the binary encoding is that the mutation of a single bit might result in arbitrarily long steps in the phenotypic space, thus failing to mimic nature, where smaller changes occur much more often than larger ones. For the same reason, it makes the search more difficult, since a very small change such us going from one to an adjacent possible number that a coordinate can take, for instance from “011…111” to “100…000”, may require the flip of every bit.

These difficulties are overcome by “Gray codes”, which encode in such a way that it guarantees that the Hamming-distance between adjacent numbers equals one. That is to say, the binary codes of adjacent possible grid numbers differ in a single bit. Notice that a hamming-distance equal to one implies the closest possible distance between two nodes of the hyper-cube (an edge).

*Gray codes play an important role in the application of canonical GAs to parameter optimization problems, because both theoretical and empirical findings indicate that they are preferable if a binary coding of real values is desired* [7].

### 4.3.4.2 Parents' selection

During the parents' selection, two out of the $\mu$ individuals are selected, so that a temporary population of two individuals is created $\left(\mathbf{P'}^{(t)}\right)$.

$$\mathbf{P'}^{(t)} = s\left(\mathbf{P}^{(t)}\right) \in I^2 \tag{4. 30}$$

Where $\mathbf{P}^{(t)} \in I^{\mu}$.

Recall that since the selection is fitness-based, and it always occurs within the phenotypic state space, the genotypic expression of the individuals has to be decoded prior to the fitness evaluation. Although there are many different ways of performing the parents' selection, canonical GAs typically make use of a probabilistic and proportional fitness-based operator, where the probabilities are calculated according to equation **(4. 31)**.

$$p(\mathbf{a}_i) = \frac{f(h(\mathbf{a}_i))}{\sum_{j=1}^{\mu} f(h(\mathbf{a}_j))} = \frac{f(\mathbf{x}_i)}{\sum_{j=1}^{\mu} f(\mathbf{x}_j)} \tag{4. 31}$$

Where:

- $p(\mathbf{a}_i)$ is the probability that the individual $\mathbf{a}_i$ has of being selected
- $f(h(\mathbf{a}_j)) = f(\mathbf{x}_j)$ is the fitness of the individual $\mathbf{a}_j$
- $\mu$ is the number of individuals in the population

The most common method is the roulette wheel selection, as previously explained in the EAs' general procedure. However, several other alternatives to, and variations of the proportional selection are frequently implemented to improve the performance of the algorithm.

The survivors' selection (acording to **Fig. 4. 2**) is not really performed, or it could also be trivially viewed as performed by selecting all the created progeny for the next generation. This procedure is called generational replacement, since the parents' population is completely replaced by the progeny's population.

Variations of the canonical form include the two-stage-selection procedure, or a different size of the children's population. If the number of children ($\lambda$) is higher than the number of parents ($\mu$), the survivors' selection eliminates ($\lambda - \mu$) individuals from the $\lambda$ children, while if ($\lambda - \mu$) is less than zero, the best ($\mu - \lambda$) individuals are passed directly to the next generation without undergoing selection (elitism), so that the individuals' life span is not limited to one generation. In this case, the percentage of the population to be replaced in each generation is called the "generation gap". Elitism guarantees constant improvement in the population since the best individual of the present population returns the best result achieved up to then as well, but it significantly reduces the algorithm's capability of escaping local optima.

### 4.3.4.3 Alteration operations: recombination

Crossover performs a mixing of the existing genetic material, the result of which is an exploration of the search-space within the genetic pool, which is not widened. The canonical version makes use of the so-called one-point crossover operator. Two parents' chromosomes are cut off at a random point, and the parts of the parents are exchanged thus breeding two children, which are allocated in the smallest hyper-cube containing the parents (i.e. in the hyper-cube spanned between the parents' locations). There are some alternative processes such as *n*-point crossover[46], uniform crossover[47], multi-parent crossover, etc. There is no known way to generalize which type of crossover outperforms the rest.

Typically, crossover generates two children out of two parents by exchanging parts of their genetic code, and then chooses one, either randomly or by keeping the fitter. Hence, the operator is $\mu$-fold, defined from $I^2 \rightarrow I$. A quite common alternative version keeps both children generated by crossover, so that the operator becomes $\frac{\mu}{2}$-fold, defined from $I^2 \rightarrow I^2$.

The recombination operator in GAs is typically applied with a high probability of occurrence, usually in the range of 0.6 – 0.95. For each pair of individuals selected by the parents' selection operator, a random number between zero and one is generated from a uniform distribution. If it happens to be below the probability threshold, the crossover is applied by making a random cut off and exchanging the blocks between parents. If the two-point crossover were chosen, the set of genes going from the first to the second point, from left to right, would be exchanged[48].

[46] The two-point crossover is explained in [47], page 151.

[47] Each bit is randomly chosen from the corresponding parental bits.

[48] Notice that in this case the string becomes a ring, so that the last bit is followed by the first one.

Recall that during the parents' selection stage, recurrent selection of the same individual to breed is allowed. Typically, the parents' selection operator is performed from $I^{\mu} \to I^{\mu}$ or from $I^{\mu} \to I^{2 \cdot \mu}$, where fitter individuals appear more than once. Then, individuals who belong to this intermediate population, which has increased its average fitness in relation to the last population in detriment of diversity, need to be randomly paired-up before crossing-over. In contrast, the parents' selection operator is considered in this work to be performed from $I^{\mu} \to I^{2}$, thus being a $\mu$-fold operator, so that the random pairing-up is embedded into the operator.

To conceptually clarify the role that crossover plays, Rennard [66] states:

*The basic idea of GAs is as follows: the genetic pool of a given population contains the solution, or a better solution, to a given adaptive problem. The solution is not "active" because the genetic combination on which it relies is split between several subjects. Only the association of different genomes (i.e. chromosomes) can lead to the solution…*

Clearly, Rennard is not considering mutation in this definition, which is, perhaps wrongly, thought to be a background operator. Nevertheless, it clarifies the concept of the improvement made by the recombination operator. Crossover places the offspring within the hyper-cube spanned between the parents, thus exploring within that volume, whereas mutation introduces new genetic information, thus widening the genetic pool.

#### 4.3.4.4 Alteration operations: mutation

Mutation was introduced by Holland (from [55]) as a background operator of small importance, whose main purpose was to introduce variety when, by accident, all the chromosomes happen to have the same code, in which case no exploration is performed by crossing-over any longer.

Mutation is typically applied after crossover. It stochastically flips some of the genes, moving bit by bit along each string and along the whole population, to create new genetic material[49]. It is applied with a low probability of occurrence, usually in the range of 0.001.

[49] Mutation also takes place spontaneously and randomly in biological organisms. But in nature, mutation might occur at any time (not only during reproduction), both in gametes and in somatic cells. It is the only way of altering an organism's genetic code during its own life time.

*Recent studies have impressively clarified that much larger mutation rates, decreasing over the course of evolution, are often helpful with respect to the convergence reliability and velocity of a GA, and that even so-called self-adaptive mutation rates are effective for pseudoboolean problems* [7].

### 4.3.4.5 The "schema theorem"

In the same fashion as in nature, evolution occurs in GAs by blind manipulations of the information within chromosomes. The algorithm does not know anything about the problem it is solving, but it simply takes each chromosome's performance into account, according to a given fitness function, to select the ones that would become parents. Thus, from simple rules, a complex behaviour capable of solving complex problems, arises.

It is not obvious why GAs actually optimize, successively obtaining populations that are, on average, fitter than the previous ones. *Since all GAs do is work with the strings of chromosomes, there must be something related to the fitness inherent in the strings that are utilized* [47].

Holland (from [47, 21]) described the application of the so-called "schema theorem" in GAs, which attempts to explain why the canonical GA actually indirectly performs optimization. *It should be noted, though, that some researchers have recently found errors in Holland's argument, and the issue is currently controversial* [47]. Nevertheless, the theorem is briefly explained in this work because it has provided a starting point to theoretical work on the field for many years.

Keep in mind that in this context, the word "string" refers to a "schema's string" rather than to an "individual's string", which can either coincide or the former be contained in the latter. In order to define the schemata, the alphabet of the genetic code plus a do-not-care symbol "#" are used to define values at certain locations. If the encoding is binary, then the schemata's alphabet is: {0, 1, #}. The GA procedure increases the probability that the schemata which result in best improvement of the population's fitness will persist to the next generation.

Note that there is a set of possible codes matching each schema. The number of matchings can vary from one to $2^w$, where $w$ is the number of times the symbol # appears in the string. For instance, there are $2^3 = 8$ possible codes that match the three-bits-long schema "# # #", whereas there are only two that match the schema "# 0 1" ("1 0 1" and "0 0 1").

Schemata that belong to an individual fitter than the average are more likely to survive the parents' selection, so that highly fit schemata benefit from the fitness-based selection.

The crossover and mutation operators might disrupt the highly fit schemata that passed the selection, although *Holland argues that crossover among the fittest members of a population will result in the discovery and survival of better schemata* [47].

Therefore, crossover and mutation provide new schemata to guide the search into new unexplored regions. Suppose, for instance, the two following different schemata:

| | | | | | | | | | | | |
|---|---|---|---|---|---|---|---|---|---|---|---|
| # | # | # | 0 | # | # | # | 1 | # | # | # | 0 |
| # | # | # | # | # | # | # | # | # | 0 | 1 | 0 |

The defining length of a schema is the distance between the first and the last specific string positions ("zeros" and "ones"). Thus, the defining length corresponding to the schema to the left is "nine", whereas the one corresponding to the schema to the right is "three".

The order of the schema is the number of fixed positions ("zeros" and "ones") in the schema. Notice that both schemata in the example above have the same order.

The crossover is clearly less likely to disrupt the second schema, which has a smaller defining length (then, it is said to be more compact). Regarding mutation, it occurs at a very low rate, and since it is done bit by bit, it is as likely to disrupt one schema or the other. Hence for different schemata that belong to different individuals equally fit, compact (i.e. with short defining length) and short (i.e. low order) schemata are more likely to persist. Therefore, "compact and short schemata" that are part of highly fit individuals will appear in ever-increasing numbers in future generations. The "schema theorem" predicts the number of times a specific schema will appear in the next generation of a canonical GA.

$$n^{(t+1)}(S) \geq n^{(t)}(S) \cdot \frac{f(S)}{f_{avg}} \cdot \left[ 1 - p_c \cdot \frac{\delta(S)}{L-1} - o(S) \cdot p_m \right] \qquad \textbf{(4. 32)}$$

Where:

- $n^{(t)}(S)$ : total number of individuals at generation $t$ that contain the schema $S$
- $t$ : time step (generation)
- $f(S)$ : average fitness of all the individuals in the population that contain the schema $S$
- $f_{avg}$ : average fitness of the entire population
- $\delta(S)$ : defining length of the schema
- $L$ : total length of every individual's string (i.e. the number of bits it contains)
- $o(S)$ : order of the schema

- $p_c$, $p_m$ : probabilities of occurrence of crossover and mutation, respectively

A thorough analysis of this theorem is out of the scope of this work.

The GA can be thought of as effectively and simultaneously working with a large number of schemata of different lengths. The schema theorem provides a quantitative prediction for all schemata, for a canonical GA only. *Schemata with above-average fitness values will be represented an increasing number of times as generations proceed. Those with below average values will be represented less; they will "die out", just as happens in nature* [47].

*An immediate result of this theorem is that GAs explore the search space by short, low order schemata which, subsequently, are used for information exchange during crossover* [20].

Building Block Hypothesis*: a GA seeks near-optimal performance through the juxtaposition of short, low-order, high-performance schemata, called the building blocks* [20].

*According to the Schema Theorem, such schemata in a population receive an exponentially increasing number of trials in the following generations, such that the promising regions of the search space are sampled with an exponentially increasing number of representatives in the population* [7].

Until recently, the schema theorem was the starting point of every theoretical work on GAs. However, this is no longer the case due to the lack of agreement among the scientific community. Moreover, any change on the canonical version of a GA turns the theorem inapplicable, and the astonishing majority of practical GAs are far from the canonical version.

## 4.3.5 Genetic programming

GP is a population-based algorithm that belongs to the group of genotypic approaches. Some researchers claim that it is merely a branch of GAs, whereas some others state that it is one of the most exciting and promising branches of EAs.

Computer programs are typically written, in a deterministic fashion, by applying human knowledge related to the problem at issue. The GP paradigm was developed in the 90's by Koza [48], who suggested that the desired program should evolve itself, so that instead of evolving a problem-solving algorithm, an algorithm that solves the problem of solving problems is evolved. *Derived from GAs, the GP paradigm characterizes a class of EAs aiming at the automatic generation of computer programs. To achieve this, each individual of a population represents* *a complete computer program in a suitable programming language...* [7]. Koza gives reasons for choosing the LISP (LISt Processing) programming language in [48], although he claims that virtually any other language such as FORTRAN or C can be used. Therefore, *the search space is the hyperspace of LISP "symbolic expressions" (called S-expressions) composed of functions and terminals appropriate to the problem domain* [48].

*Many seemingly different problems in machine learning, artificial intelligence, and symbolic processing can be viewed as requiring the discovery of a computer program that produces some desired output for particular inputs. When viewed in this way, the process of solving these problems becomes equivalent to searching a space of possible computer programs for a highly fit individual computer program. The recently developed genetic programming paradigm ... provides a way to search the space of possible computer programs ... to solve ... a surprising variety of different problems from different fields. In genetic programming, populations of computer programs are genetically bred using the Darwinian principle of survival of the fittest and using a genetic crossover ... operator appropriate for genetically mating computer programs* [49].

Since computer programming is a hard task, automatic programming has been a goal for scientists for a long time. The GP paradigm presents a means of automatically writing computer programs. In practice, a computer program is represented as a tree structure, and GP applies evolutionary search to the hyper-space of possible and valid tree-like structures.

Since GP is a paradigm which the canonical PSO paradigm does not typically compete with[50], only a very brief discussion is presented hereafter for completeness only.

### 4.3.5.1 Representation of individuals and fitness evaluation

Individuals are represented by tree-like structures, each of which stands for a proposed solution of the problem at hand.

The fundamental components of an individual are its genes, of which there can be two kinds:

[50] The PSO paradigm can be applied to the training of an ANN, which in turn would perform the desired input-output mapping, although the final solution is greatly influenced and limited by the pre-defined network structure. In contrast, GP is well capable of solving the whole problem on its own, building up the whole program with virtually no limitations. In fact, Koza [48] states that *...the flexibility we seek includes the ability to create and execute computer programs whose size, shape, and complexity is not specified in advance...* Therefore, the GP algorithm searches the space of possible computer programs, whereas the PSO algorithm searches the space of weights for a pre-defined computer program.

- Terminal genes (or nodes): Keeping the tree-analogy, terminal genes stand for the leaves of the trees, since they are nodes without branches. With regards to the problem, they represent constants and variables.
- Function genes (or nodes): They are intermediate nodes with children. They stand for functions, while their children stand for the arguments of the functions.

The nodes of the tree-structures stand for the genes, and the connections between nodes are given by the data structure representing the individuals.

Prior to the initialization of the population, the set of "terminal" and the set of "function" nodes (i.e. the set of all possible genes) available for the individuals must be specified by the user. This is a very important decision, since it could limit the algorithm to the extent of being not capable of evolving a good solution. It is always preferable to include more rather than fewer genes than necessary, since non-useful genes would simply die off during the evolution.

The maximum and minimum depth[51] of the tree-like individuals must be specified too, according to the kind of initialization process chosen, as explained hereafter.

**Initialization**

Once the set of genes has been specified, Koza (from [48, 49]) proposes three techniques to initialize the population of tree-like individuals.

- Grow: The maximum depth (*md*) must be specified. It creates one individual at a time, beginning from the root of every tree. Every node is first decided to be either a function or a terminal node. If it results being a terminal, a random terminal gene is chosen from the set, whereas if it is a function node, a function gene is randomly selected, which is given as many children as the arity[52] of the function. For each child the initialization process starts again, unless the *md* level has been reached, in which case a terminal gene is randomly selected from the set. This method provides a variety of structures through the population, but it does not guarantee individuals of a certain depth, and individuals of only one level might be created.
- Full: In this case the terminals nodes are guaranteed to be at a specified depth (*d*). Every node with a depth less than *d* is randomly selected from the set of function nodes only. If the depth equals *d*, the node is selected form the set of terminal ones. All the trees have a depth equal to *d*, but for the same reason, diversity is lost in the population.

[51] Following a path in the tree structure, the depth of a node is the minimal number of nodes from the root node, up to the node at issue (e.g. in **Fig. 4. 5**, the parents' depth equals three, whereas the offspring's depth equals four).

[52] The arity of a function is the number of arguments that it requires.

- Ramped-half-and-half: The maximum depth (*md*) must be specified. The population is uniformly divided into (*md*-1) parts. Half of each part is initialized by the grow method and half by the full method. For the first part, the depths *md* and *d* are set to two, and they are increased to three for the second part, and so forth. Thus, diversity in the population is guaranteed, and individuals of only one level cannot be generated.

**Fitness**

The fitness function assigns a value, which stands for the performance of each tree-like individual (i.e. a program). The evaluation is based on a pre-selected set of test cases, whose results are of course well known, and it typically returns the sum of the distances between the correct and the obtained results on all test cases. Obviously, the closer to zero this evaluation results, the higher the performance of the individual. This is a problem-specific issue.

### 4.3.5.2 Parents' selection

The parents' selection procedure is performed in the same manner as in GAs. Any of the selection methods available or newly developed can be used, in principle, here. Originally, as belonging to the same family as GAs, the proportional selection was preferred, but lately the tournament selection has become more popular (for further details, see section **4.3.1.3**).

### 4.3.5.3 Alteration operations: recombination

Despite being more difficult to be implemented than in GAs, the concept of recombination by crossover is very much the same here, provided the programming language used in GP is suitable to the procedure. The crossover operator simply swaps some parents' parts of the tree-like structures, thus mimicking the recombination of genetic material. An example of this is shown in **Fig. 4. 5**, where two structures $e_1$ and $e_2$ represent the expressions $2 \cdot x + 15$ and $x \cdot \cos(\pi)$ respectively. A possible child $e_3$ represents $x \cdot \cos(2 \cdot x)$.

Again, there are many alternatives such as keeping only one or both children generated by crossover, crossing-over two or more parents' genetic code, etc.

Typically, a probability of 0.9 is chosen for the crossover process to occur. The population size is kept constant and there is a generational replacement, so that the individuals have in principle a life-span of one generation. Since GP typically requires huge computational resources, most of which are consumed by the evaluation of the fitness tests, the 10% of individuals which do not undergo crossover saves an important 10% of computer resources.

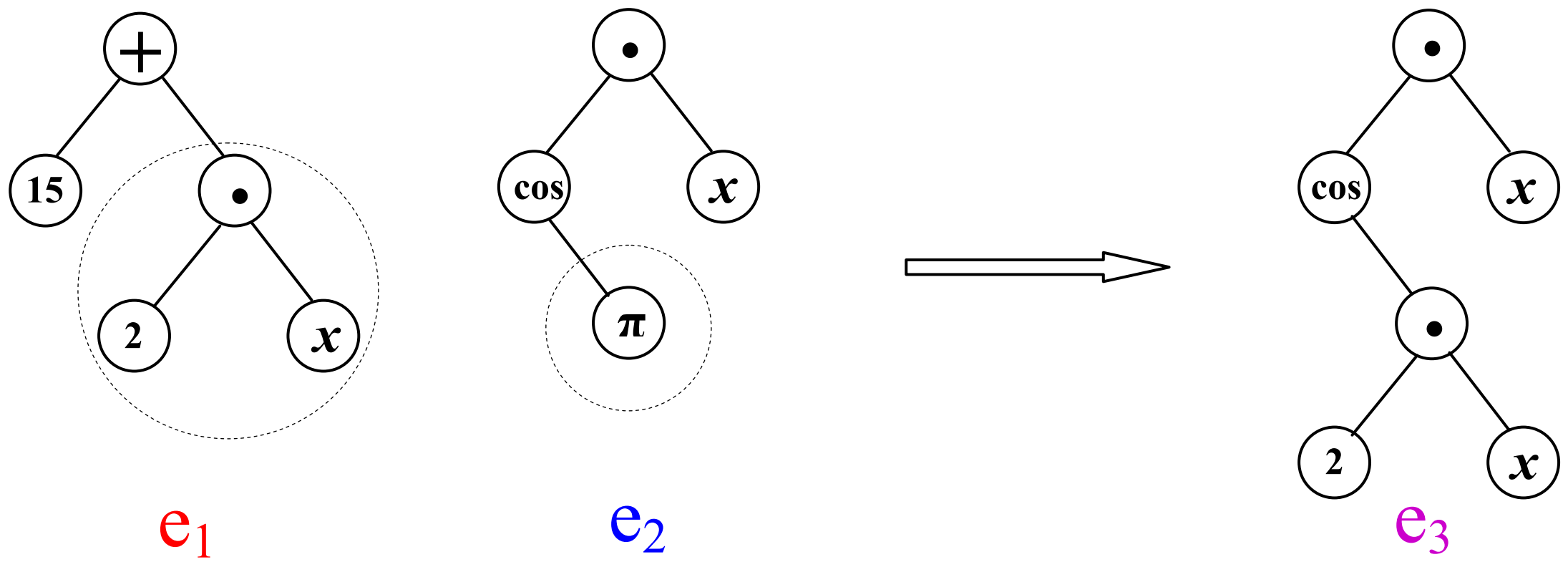

**Fig. 4. 5**: Schematic example of crossover in genetic programming. The dotted lines represent the parts of the parents ($e_1$ and $e_2$) that are exchanged to produce the child $e_3$.

### 4.3.5.4 Alteration operations: mutation

The mutation operator plays here a secondary role, so that it is used with a low probability of occurrence, just to maintain the diversity in the population. However, as opposed to the simplicity of the mutation operator in GAs, several different mutation operators have been proposed in GP. The most common mutation types are:

- **Point mutation**: it consists of the replacement of a single node by another node of the same kind (i.e. “function” or “terminal”) randomly generated.
- **Expansion mutation**: it consists of the replacement of a single “terminal” node by a randomly generated sub-tree.
- **Collapse mutation**: it consists of the replacement of a sub-tree of the individual by a randomly generated “terminal” node.
- **Sub-tree mutation**: it consists of the exchange of a sub-tree of the individual by another randomly generated one.

- **Per-mutation**: it is simply the permutation between nodes of the same kind belonging to the same individual.

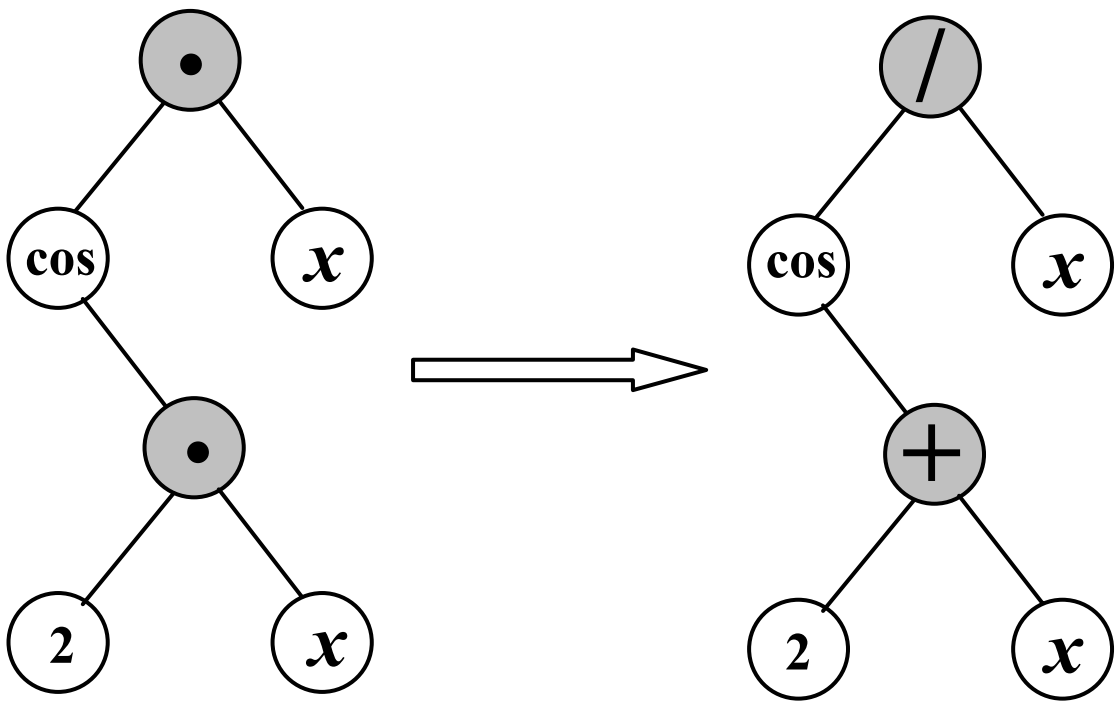

**Fig. 4. 6**: Schematic example of point mutation in two (shaded) nodes of an individual.

After the application of the alteration operators, the survivors' selection should be performed, according to the flow chart in **Fig. 4. 2**. Here, just like in GAs, this selection is not really performed in the canonical version of the algorithm, or it could be trivially viewed as if all the individuals were selected.

The process continues by evaluating the fitness of the progeny, then selecting the individuals to reproduce, and so forth.

Even though this is believed to be one of the most promising paradigms among the EAs, it will not be dealt with in further details here because it is precisely the paradigm least related to the PSO paradigm, which this thesis is focused on.

## 4.3.6 Further comments on evolutionary algorithms

### 4.3.6.1 Miscellaneous remarks

It is quite frequent to find in the literature the statement that a major difference between the three mainstreams is given by the different levels of hierarchy on the metaphor. Thus, GAs are claimed to focus on the evolution of chromosomes, ESs on the evolution of individuals and EP on the evolution of species.

However, since only gametes are generated during meiosis, where crossover might occur, the whole process cannot take place within a single individual[53]. It seems more accurate to consider chromosomes in GAs as analogous to independent single-chromosomal individuals that reproduce by cloning (e.g. mitosis, binary fission), and perform a kind of recombination similar to crossover. In this context, the level of hierarchy of evolution being modelled by

[53] Recall that a gamete only completes the two sets of chromosomes after joining another gamete generated within a different individual.

GAs and ESs is the same, and the differences rely on the fact that the former focus on the genotypes whereas the latter do on the phenotypes. In turn, EP is claimed to mimic the evolution of species instead of the evolution of individuals, hence recombination is not performed because recombination does not occur between species. However, there is not much difference between EP and ESs without recombination, and the statement of the different levels of hierarchy seems to be just conventional. Therefore, the statement of Fogel et al. [34] on the differences of the paradigms based on the level of hierarchy of the natural evolution metaphor is at least arguable.

A major clear division is into phenotypic-based (ESs and EP) and genotypic-based (GAs and GP) approaches. In turn, the main difference between ESs and EP is that the former relies somehow on the recombination operator (even though mutation is the main operator as well)[54]. Regarding the differences between GAs and GP, they do not compete with each other as pure techniques. GAs are classical EAs, which are meant to solve problems, whereas GP is a kind of meta-algorithm aimed at automatic programming by evolving computer programs. GAs can handle automatic programming tasks, but they require adaptations and help from other techniques, such as utilizing a population of neural networks, on the same line as previously explained for the PSO paradigm. GP is believed to be one of the most promising new areas of research on EAs and AI, but they have the inconvenience of still being computationally expensive. *GP uses a phenomenal amount of processing time even for apparently simple problem domains… …The major consumers of processing power are the fitness tests…* [79].

An important aspect in EAs is that of the convergence, since they are expected to have a rate of convergence similar to traditional methods while outperform them with regards to robustness and range of applicability. However, convergence is very difficult to theoretically deal with in EAs. ESs is the approach with most work done about convergence, since this was an issue from the very beginning. Fogel et al. [34] offer: *much of the current effort in evolutionary computation involves gaining a better understanding of the mathematical properties of these techniques*

[54] There is generalized disagreement among the scientific community with regards to the importance of the recombination operator in EAs. *At present, the role of recombination in EAs in not perfectly understood. The advantages or disadvantages of recombination for a particular objective function can hardly be assessed in advance, and certainly no generally useful setting of recombination operators… …exists. Recently, Kursawe has impressively demonstrated that, using an inappropriate setting of the recombination operator, the (15, 100)−ES with $n$ self-adaptable variances might even diverge on a sphere model for $n$ = 100. His work shows that the appropriate choice of the recombination operator not only depends on the objective function topology, but also on the dimension of the objective function and the number of strategy parameters incorporated into the individuals* [7].

*as optimization algorithms. Most of the useful results in convergence have been offered by the German researchers in ESs (Bäck and Schwefel)... ...Similar results with GAs have been difficult to obtain and these had led to many empirical comparisons... ...when the objective criterion is a highly interactive nonlinear function, the results have often been generally favourable to methods of ESs and EP*.

Experimentally comparing the convergence rates between ESs and GAs, Bäck [5] worked on the sphere model obtaining linear convergence for an ES. A canonical GA stagnated after a few generations, whereas a GA without recombination showed a linear convergence (although the rate was much lower than that of the ES). Even though this is only a particular experiment that cannot be generalized, Bäck [5] claims that it may serve as a counterargument against the common claim from many GAs researchers that the recombination operator is of utmost importance while the mutation operator plays a minor role.

Although the different streams of EAs were originated seeking different purposes, all of them are surprisingly based on observations of natural evolution. Some methods are in principle more suitable for some kinds of problems, but they have all reached such a development through decades, that they can easily be adapted to almost any problem. It is not easy to decide which branch of EAs would perform better when handling a certain task. Bäck suggests [5]: *As a rule of thumb, I personally claim that ESs should be applied in cases of continuous problems, whereas GAs serve most useful in case of pseudoboolean problems*[55]*. Obviously, hybridizations of both algorithms are promising for the application to mixed-integer problems involving both discrete and continuous object variables*.

#### 4.3.6.2 Constrained optimization

Nowadays, constrained optimization problems can be dealt with by numerous creative techniques to handle feasible and infeasible solutions within a population. Since none of these techniques has proven to undoubtedly outperform the others in every case, many different techniques are currently in use. A typical one is based on the penalization of infeasible solutions, but even methods within this branch differ in many important aspects. An alternative is to create decoders that avoid creating infeasible individuals, but it is sometimes necessary to evaluate infeasible solutions to guide the search to more promising areas. Two simple techniques are briefly discussed hereafter, and the problem of constrained optimization will be undertaken again when dealing with the PSO algorithm, later in this work.

[55] Notice that virtually any discrete problem can be represented by making use of binary strings, thus becoming a pseudoboolean problem.

#### 4.3.6.2.1 Rejection of infeasible solutions

It is a kind of penalization method sometimes called the "death penalty" due to the elimination of infeasible solutions. The algorithm does not need to evaluate infeasible solutions and compare them to the feasible ones, thus saving computational effort[56]. The method is suitable when the feasible search-space is convex, and when it constitutes an important part of the whole search-space (i.e. feasible plus infeasible). For instance, equality constraints usually cannot be handled by this method. When the feasible search-space is small, the method might be trapped from the very beginning if all the initialized population happens to be infeasible, so that the individuals need to be improved rather than eliminated, or if at any time step all the individuals become infeasible. By elitism, this situation can be avoided, but the improvement might be restricted. Sometimes it is a better strategy to allow the individuals to cross the infeasible region to approach the optimum more easily.

#### 4.3.6.2.2 Penalizing infeasible individuals

In this alternative, the individuals searching the infeasible space are evaluated, but the individual's fitness is increased (for minimization problems) if the solution is infeasible.

$$fp(\mathbf{x}) = f(\mathbf{x}) + Q(\mathbf{x}) \tag{4. 33}$$

Where:
- $fp(\mathbf{x})$: penalized fitness of individual **x**.
- $f(\mathbf{x})$: fitness of individual **x**.
- $Q(\mathbf{x})$: penalty for infeasible individual **x**.

Often penalties are not fixed but linked to the amount of infeasibility of the individual. They might simply be functions of the number of constraints violated, but functions of the distance from feasibility are usually preferred. For instance, for optimization problems of the form:

---

[56] Notice the here "infeasible solutions" refer only to solutions allocated outside the feasible search-space.

$$\text{Minimize } f(\mathbf{x}) \quad \text{with } \mathbf{x} \in \mathcal{R}^n \tag{4.34}$$

Where:

- $g_j(\mathbf{x}) \leq 0 \quad ; \quad j = 1, \ldots, q$
- $g_j(\mathbf{x}) = 0 \quad ; \quad j = q+1, \ldots, m$

The degrees of infeasibility might be taken into account by constraints violation measures:

$$f_j(\mathbf{x}) = \begin{cases} \max\{0, g_j(\mathbf{x})\} & ; \quad 1 \leq j \leq q \\ g_j(\mathbf{x}) & ; \quad q < j \leq m \end{cases} \tag{4.35}$$

Then, the fitness to be considered during the selection process could be calculated as follows:

$$fp(\mathbf{x}) = f(\mathbf{x}) + \sum_{j=1}^{m} \lambda_j^{(t)} \cdot \left(f_j(\mathbf{x})\right)^{\alpha} \tag{4.36}$$

Where $\alpha$ is typically equal to "2", and $\lambda_j^{(t)}$ can be the same for all the constraints, and even through generations. For instance, $\lambda_j^{(t)} = \lambda^{(t)}$ can be updated every generation according to:

$\lambda^{(t+1)} = \dfrac{1}{\beta_1} \cdot \lambda^{(t)}$, if the best individual in the last $k$ generations was always feasible.

$\lambda^{(t+1)} = \beta_2 \cdot \lambda^{(t)}$, if the best individual in the last $k$ generations was never feasible. **(4. 37)**

$\lambda^{(t+1)} = \lambda^{(t)}$, otherwise.

Where $\beta_1, \beta_2 > 1 \wedge \beta_1 \neq \beta_2$.

Notice that if $g_j(\mathbf{x}) \leq 0 \quad \forall j = 1, \ldots, q \quad \wedge \quad h_j(\mathbf{x}) = 0 \quad \forall j = q+1, \ldots, m \quad \Rightarrow \quad f_j(\mathbf{x}) = 0 \ \forall j$
$\Rightarrow \quad fp(\mathbf{x}) = f(\mathbf{x})$.

A high penalization might lead to the situation where the individuals cannot search the infeasible regions, thus converging to a non-optimal but feasible solution. A low penalization might lead to the system evolving individuals that are violating constraints but present themselves as fitter than feasible individuals. The proper definition of the penalty functions is not trivial, and it plays a crucial role in the performance of the algorithm.

# 4.4 Closure

The general procedure of a generic EA was outlined, and the four main paradigms were presented. Their main features were analyzed, and the fundamental underlying principles were discussed. The similarities and differences among them, as well as their strengths, weaknesses, alternatives and applications were also considered. The stress was put on the paradigms closer related to the canonical PSO method, so that not much attention was paid to the GP method. The different techniques developed to handle constrained optimization problems were very briefly mentioned in this chapter, because the subject matter is dealt with in more detail in **Chapter 10**, although focusing on the PSO paradigm. Beware that most of the conclusions within this chapter are also valid for the PSO method.

Evolution is a natural process of fitness maximization that organisms undergo to adapt to a dynamic environment, which has proven to be successful in dealing with complex non-linear and non-stationary problems, characterized by uncertainties and noise. Since these are precisely the kinds of problems that the traditional optimization methods fail in dealing with, EAs were inspired by the processes that biological evolution makes use of to deal with these situations, where it has been realized that **parallelism**, **randomness** and **survival of the fittest** are the key-concepts. Notice that biological organisms do not optimize consciously!

Hence EAs are designed as population-based methods in order to search the problem space in **parallel**, where each individual is a potential solution to the problem. A sort of common dynamic “genetic pool” is created by gathering all the individuals’ genetic information. **Stochastic** genetic-like operators introduce innovation and variation into the system, and they make it less likely to get trapped in periodical phenomena. The “random” numbers are not random but deterministically generated from a certain probability distribution. However, in the same fashion as “randomness” in real life, the laws that rule the generation of such numbers have no connection to the laws that rule the problem at hand, thus showing stochastic behaviour. Finally, a **fitness-based selection** of the individuals ensures that the search does not become random. Therefore, the evolution of the algorithm is not directed by inference but by trial and error, while the trials that led to better responses are learned by storing the corresponding individuals’ genetic structure.

EAs are bottom-up approaches in the sense that the system's intelligent behaviour emerges in a higher level than the individuals'. It is frequently claimed that they do not optimize but adapt to the environment, evolving intelligent solutions without using programmers' expertise on the subject matter. While this makes it difficult to understand the way that optimization is actually performed, the EAs show an astonishing robustness in dealing with many kinds of complex problems that they were not specifically designed for. The individuals really do not know that they are optimizing because they are not programmed to do so!

However, these robust general-purpose optimizers have the disadvantage that their theoretical bases are indeed extremely difficult to understand in a deterministic fashion. Although much theoretical work has been attempted, only problem-specific and partial conclusions have been achieved in such important matters as the tune of the algorithms' parameters, the design of the operators, and the convergence of the paradigms. Thus, different mainstreams coexist, each one supported by a different group of scientists. The truth is that the precise behaviour of each paradigm is not fully understood, what should be of no surprise taking into account that they are not designed in a fully deterministic fashion, so that attempting to understand them in that line of thought appears rather contradictory.

Another big disadvantage of these methods is the difficulties in designing the operators and the great number of parameters that need to be tuned. In contrast, the PSO paradigm does not require the design of operators, and only the tuning of a few parameters is needed. In addition, the implementation of its canonical version is of utmost simplicity.

Thus, **chapter 5** presents the main features of the ACO and PSO paradigms, while only the latter is dealt with from **Chapter 6** forth.

# Chapter 5

# SWARM INTELLIGENCE

Swarm intelligence is the branch of the field of artificial intelligence that involves the study of the collective behaviour that emerges from decentralized and self-organized systems. The basic concepts are discussed, some examples of social behaviour in the animal kingdom are presented, and a few, relevant concepts of social psychology are included, which guide the dissertation towards a thorough understanding of the basic concepts underlying the particle swarm optimization paradigm. Finally, the two most successful swarm-intelligence-based paradigms: "ant colony optimization" and "particle swarm optimization" are outlined.

## 5.1 Introduction

In spite of the fact that human beings have overestimated their role in the universe throughout history, the advances in science have systematically stricken their ego. Therefore, the new tendency is to accept their not so flattering role as mere components of higher level systems.

Nevertheless, humans do not seem to be yet prepared to resign their individuality. From the beliefs of possessing a personal soul to the beliefs of their intelligence being an individual trait, it seems that human beings are not ready to accept a minor role after all. This feeling is so strong that, for decades, almost all the attempts to create AI have consisted of mimicking the individual's behaviours or neurophysiology.

It is extremely difficult, if not impossible, for a member of a natural system to be aware of the system's behaviour, and even more to notice its role within the system. This is so because the system does not have a sense of purpose, and the overall behaviour is an emergent property.

When humans are able to observe a complete system which they do not belong to, it seems easier to accept the relativeness of individuality. This is the case of a colony of social insects, whose individual lives do not appear to be essential for the functioning of the colony. Recall the discussion in **Chapter 3** about the properties that a living organism should display and the different degrees of aliveness: it was pointed out that a single worker bee is not fully alive because it cannot reproduce, whereas the whole colony, viewed as a living superorganism, presents all the required properties to be considered fully alive. In this line of thought, the death of a bee is something like, for instance, the loss of an insignificant organic cell within a dog's body. It is far more difficult to recognize this when the system does not appear self-evident, and the view needs to be widened. Nevertheless, when the solar system is thought of as just one out of many systems within the universe; when the planet Earth is observed just as one of many planets in the solar system; when each human being is seen merely as one out of billions of human beings whose role within the system is sadly not vital (usually not even important); when each cell within a human's body is noticed to be dispensable for the functioning of the system; and so on, the importance of individuality is seriously trivialized.

Now that individuality has been devaluated, it is important to visualize how in the world the interactions of "trivial" individuals can result in complex and adaptive behaviours. Lovelock (quoted in [47]) viewed the planet Earth as a living organism, developing his "Gaia theory". He claimed that *...life, or the biosphere, regulates or maintains the climate and the atmospheric composition at an optimum for itself... Gaia theory holds that the entire planet functions as one single integrated organism, including "inorganic" systems such as the climate and the oceans as well as the biosphere. It is a literally global view that places individual organisms and even species in the roles of cells and subsystems, supporting the larger dynamical system through their breathing, eating, breeding, dying...* [47]. The view of the Earth as a living organism is a good two-fold example: first, it makes clear the concept of the relativeness of the individuality; and second, it gives a first idea of the concept of swarm intelligence (SI), where a system that is composed of different individuals self-organizes by adapting to the environment so as to achieve its goals[1]. Beware that Lovelock's statement must not be interpreted as if the system had a sense of purpose!

Even if individuals do not realize, their "free will" is constrained by the convenience of the system(s) they belong to. For instance, Wynne-Edwards (quoted in [47]) claims that *...many if not all the higher animals can limit their population-densities by intrinsic means*, so that evolution selects against animals that reproduce too much, in order to prevent overpopulation and decimation of species. He suggests that certain behaviours such as noisy group vocalizations might serve the function of informing the members of a group about the size of its population so as to regulate the rate of reproduction. This is another example of swarm-intelligent behaviour, which shows that individual's decisions like breeding might just not be so "individual". In addition, some kinds of behaviour in species simply do not make sense under the typical assumption that individuals seek self-gain, such as the cases of a mother risking her life to protect her offspring from a predator, or animals that give a risky warning call in the presence of a predator. Something similar happens with reproduction itself, where the mother becomes either an easier prey or a less adept predator. The examples could continue ad infinitum. It is believed that there is a tendency in nature for genes to seek their survival even at the cost of individual lives. Therefore, natural beings do not only seek self-gain but more importantly the group's gain (whose members usually share many genes!), and it seems clear that cooperative behaviour brings great advantages to the swarm[2] as a whole.

[1] Recall the definition of intelligence *...as the capability of a system to adapt its behaviour to meet its goals in a range of environments* proposed by Fogel et al. (quoted in [35]).

## 5.2 Fundamental concepts

Some important concepts related to SI are briefly discussed hereafter. Note that the concept of emergence was already defined in section **3.2.3**, and that the concept of self-organization is a whole field in itself, hence the following discussion is necessarily incomplete and simplistic.

### 5.2.1 Emergence

An emergent property is a feature of a swarm of simple lower level entities as a whole, which does not exist at the individual level. The interactions among a number of individual entities might give birth to an emergent property, which is not possible to be inferred by analyzing an isolated individual. Likewise, when designing artificial entities that would display emergent properties, the latter cannot be deterministically implemented. It is extremely difficult even to predict whether a property would emerge from certain kinds of interactions among certain kinds of entities (not to mention which property!), since the interactions, which are executed based on purely local information, must generate a positive feed-back effect. Typically, a lower threshold for the number of entities involved is required for the feed-back to occur. Sometimes the interactions just cancel each other out!

[2] Swarm: - *A large number of insects or other small organisms, especially when in motion.*
- *An aggregation of persons or animals, especially when in turmoil or moving in mass.* [23]

### 5.2.2 Self-organization

*Self-organization is a set of dynamical mechanisms whereby structures appear at the global level of a system from interactions of its lower-level components.*[11]

Despite the obvious fact that the concepts of self-organization (SO) and emergence are closely correlated, emergence is theoretically possible without SO, and vice versa. The precise links between these two not fully understood concepts remain an active research question.

*Self-organization refers to a process in which the internal organization of a system, normally an open system, increases automatically without being guided or managed by an outside source. Self-organizing systems typically (though not always) display emergent properties.* [80]

The SO of a system usually relies on four basic phenomena:

1. **Positive feed-back**: It is the response of a system when an action that affects it induces it to respond in the same direction of change. An example of this is the case of the pheromone-laying and pheromone-following behaviour observed in colonies of ants (see section **5.3.2**). Sometimes a possitive feed-back, if it is not controlled by a negative feed-back, may run out of control resulting in the collapse of the system.

2. **Negative feed-back**: It is the response of the system when the action that acts on it induces it to respond in a reverse direction of change. This is a process that tends to keep things constant, helping to stabilize the emergent collective patterns and to prevent a sysyem from crashing. The exhaustion of food sources, the saturation of available workers, and the pheromone evaporation in ant colonies are good examples of stabilizing negative feed-back.

3. **Fluctuations**: They are (typically stochastic) processes that introduce innovation, thus enabling a system to find new solutions. For example, an insect that gets lost may eventually find new unexploited food sources.

4. **Multiple interactions**: Like emergence, SO can only be generated among a number of individuals. There is usually a lower threshold for this number in order for SO to occur.

*The ancient atomists (among others) believed that a designing intelligence was unnecessary, arguing that given enough time and space and matter, organization was ultimately inevitable, although there would be no preferred tendency for this to happen. What Descartes introduced was the idea that the ordinary laws of nature tend to produce organization.* [80]

## 5.2.3 Division of labour

The division of labour (DoL) is an amazingly decentralized process, which enables a swarm to perform different tasks in parallel. This is a critical issue, since there are certain tasks that cannot be carried out in a sequential fashion. For instance, it would be pretty unsafe for a colony of ants if all its members foraged at the very same time, returning to protect the nest only once the food requirements are fulfilled! Some functions like foraging, breeding, brood-care and protecting the nest, need to be carried out in parallel. In addition, the DoL results in the individuals who repeatedly perform the same kinds of tasks becoming familiar with it, thus "optimizing" their performance.

The DoL is pushed by both genetic pressures and the needs of the swarm. For example, in social insect colonies, the queen breeds while the anatomically different and sterile workers forage and protect the nest. An important aspect of this phenomenon is its plasticity to respond to the needs of the swarm when facing environmental challenges. For instance, some ant colonies have two genetically different kinds of ant workers, one of which is anatomically more suitable for foraging while the other kind is more suitable for protecting the nest. Naturally, they tend to perform the tasks they are better suitable for, but some experiments have shown that if for example the "guardian" kind is removed from the colony, in a few hours some "genetically less suitable" ant foragers start protecting the nest, thus adapting to the needs of the colony with no central control!

## 5.2.4 Stigmergy

Stigmergy is a means of indirect communication among the members of a system, which is performed by individual modifications in their local environment.

*The term was introduced by French biologist Pierre-Paul Grassé in 1959 to refer to termite behavior. He defined it as: "Stimulation of workers by the performance they have achieved"* [80].

The important concepts are that the environment serves as a work-state memory; that the work can be continued by any member of the system; and that the individual work is a behavioural response to the state of the environment, which is in turn modified by such a work. Trail-laying and trail-following in insect colonies is a typical example of stigmergy.

### 5.2.5 Swarm intelligence

Swarm intelligence (SI) is the branch of AI that deals with the collective behaviour that emerges from decentralized and self-organized systems. It is the property of a system whose individual parts interact locally with one another and with their environment, inducing the emergence of coherent functional global patterns.

In agreement with a FAQ document from the Santa Fe Institute (quoted in [47]), the term "swarm" is used …*in a general sense to refer to any such loosely structured collection of interacting agents*[3]*. The classic example of a swarm is a swarm of bees, but the metaphor of a swarm can be extended to other systems with a similar architecture. An ant colony can be thought of as a swarm whose individual agents are ants, a flock of birds is a swarm whose agents are birds, traffic is a swarm of cars, a crowd is a swarm of people, an immune system is a swarm of cells and molecules, and an economy is a swarm of economic agents. Although the notion of a swarm suggests an aspect of collective motion in space, as in the swarm of a flock of birds, we are interested in all types of collective behaviour, not just spatial motion.*

Thus, the term SI is commonly used to refer to …*any attempt to design algorithms and distributed problem-solving devices inspired by the collective behaviour of social insect colonies and other animal societies* [11]. When the swarm is composed of a large number of simple and cheap physical robots, the term is replaced by "swarm robotics". Unlike distributed robotic systems, swarm robotics emphasizes large numbers of dispensable robots that intercommunicate locally. A big advantage of this approach is that each individual is dispensable (provided the size of the population stays above a threshold), so that when performing expensive and risky tasks, the loss of some individuals does not represent important economical losses, and even more important, it does not spoil the whole operation.

The most successful SI-based methods at present are the ACO and the PSO paradigms. The former is a method capable of handling very difficult combinatorial optimization problems, which was developed under the metaphor of the trail-laying and trail-following behaviour observed in some kinds of natural ants. The PSO is a global method capable of dealing with optimization problems whose solutions can be represented as points in an *n*-dimensional space, which was inspired by the behaviour observed in some social animals (e.g. bird flocks, fish schools, and social insects), and in human societies. Thus, the PSO paradigm has strong roots in both, AL and social psychology. In fact, even the behaviour of some social animals are relevant for both AL and social psychology, hence there are some links between these two fields as well. The field of AL was discussed in some detail in **Chapter 3**, and the related field of the **EAs** was discussed in further detail in **Chapter 4**. However, the field of social psychology is considerably far from the expertise of the author of this thesis. Therefore, it is fair to remark that the sections **5.3** (social beings) and **5.4** (social learning and cultural evolution) discussed hereafter are mostly based on the book by James Kennedy and Russell Eberhart [47]: **Swarm Intelligence** (chapters 3, 5 and 6). This is neither due to laziness nor to whim, but to the fact that James Kennedy is both a social psychologist and the co-inventor of the PSO paradigm (Russell Eberhart, the other co-inventor, is an electrical engineer). Hence it is assumed here that the topics of social psychology discussed in [47] are the appropriate ones aiming to guide the work towards the development of the PSO method.

---

[3] Kennedy et al. [47] reject the use of the term "agent" for members of a swarm-intelligent system because such term is typically used in AI for entities that display some degree of autonomy, while the individuals within a swarm-intelligent system are not precisely autonomous. Nevertheless, since this work is not concerned with autonomous entities at all, whenever the term "agent" is used it will mean "individual", unless otherwise specified.

# 5.3 Social beings

Species benefit from sociality in many ways, a few of which were presented in section **5.1**. Pointing the dissertation towards the PSO paradigm, it is convenient to remark especially the advantage that a swarm gains from the sharing of information amongst its members. Perhaps the simplest social behaviour observed in the animal kingdom is that of a group of amoebas that self-organize to "optimize" their individual probability of survival.

## 5.3.1 Amoebas

The amoeba is a single-celled organism that moves by alternating softening and hardening of the protoplasm, feeds on bacteria, and reproduces by cell-division. When food becomes scarce, the amoebas that cannot find nutrients start emitting a chemical substance. In turn, the amoebas that detect the presence of this substance start emitting the substance themselves, while moving towards the areas where the concentration of the substance is higher. When they meet other amoebas, they will merge with them eventually forming an aggregate single organism that can crawl around. This organism produces reproductive spores, which are released when a more favourable location is found. Some spores eventually become amoebas, which are now situated in a more promising environment.

## 5.3.2 Social insects

While the previous example is perhaps the simplest kind of swarm-intelligent behaviour, which “only” shows a kind of SO of individuals that are very much the same, social insects such as termites, ants, bees and wasps also display a DoL that is influenced by both the needs of the swarm and each individual’s genetic structure.

Despite the fact that an ant might have only a few hundreds of brain cells and its individual behaviour is almost random, some of the astonishing achievements of colonies of ants include the building of complex nests and optimized networks of highways connecting food sources to the nest. The amazing thing is that not only are such works performed by simple-minded beings, but also that they are self-organized without a central control. The social insects seem to be equipped with a small set of simple rules, very much like the CA discussed in section **3.2.3.4**, so that the behaviour of the colony is an emergent property with no sense of purpose.

For instance, some termites in the north-east of Argentina are able to build domed structures. The set of rules that these insects follow might be something like:

1. Take some dirt in your mouth and moisten it.
2. Follow the strongest pheromone trail while depositing pheromone as you move.
3. Deposit the moistened dirt where the smell is strongest.

Since termites, like ants, individually display almost random behaviour, the first movements would seem random until a number of pillars are initiated. This process presents a positive feed-back effect, since the pillars are placed where the pheromone concentration is higher, thus becoming more powerful attractors as the pillars grow. Since the termites are attracted by several pillars, they frequently end up in a critical point where the attraction to either one or the other of two pillars is equally strong, making a random selection. Thus, the ants tend to approach both pillars from the sides that face one another, resulting in more dirt deposited on those opposite sides of the pillars. As the pillars ascend, they tend to get closer to one another, eventually meeting and forming an arch. When several pillars meet, the dome is formed!

The pheromone trails evaporates with time, so that a certain minimum number of ants involved in laying a trail is required. This prevents from the formation of a great number of pillars, and from the following of abandoned trails.

There are two main kinds of communication among the individuals in social insect colonies. Direct communication is an individual-to-individual process, such as the dancing of a worker bee when it finds pollen. In contrast, indirect communication consists of modifying the environment in a way that will influence other insects' behaviours (i.e. stigmergy). Some ants display stigmergic behaviour when foraging: the forager ants initially leave the nest in a random quest for food, while a trail-laying and trail-following behaviour allows finding the shortest path from the nest to a food source (see section **5.5**). This is clearly an optimization process performed by decentralized and self-organized simple-minded beings. The ACO paradigm was inspired by this natural behaviour.

In addition to the unaware emergent SO of identical beings, these social insect colonies also display an advantageous DoL: while some ants are in charge of foraging, some others protect the nest, others repair it, and others perform brood-care. As previously discussed, the DoL is self-organized according to the needs of the colony and to some genetic predisposition.

### 5.3.3 Fish schools and bird flocks

As opposed to social insects, some kinds of fishes and birds orderly move about in a rather majestic fashion. For instance, when a predator approaches a fish school, the fishes that first notice the threat change direction, and suddenly, at what seems to be exactly the same instant, all the fishes change direction so as to match their neighbours' new velocities. Some models of this behaviour have been proposed, suggesting that a single fish is attracted to a school, and that the attraction increases, while the rate of increase decreases, with the size of the school.

The behaviour of a bird flock is very similar to that of a fish school. Again, many different models have been proposed. A well-known simulation of bird flocks was developed by Reynolds (quoted in [47]), who proposed three basic rules for each bird to follow.

1. Pull away before crashing into another bird.
2. Try to match your neighbours' velocities.
3. Try to move towards the centre of the flock.

Reynolds named these artificial birds “boids”. The implementation of these rules resulted in realistic flock-like behaviour. Although the rules are entirely artificial, it is self-evident that natural animals try to avoid crashing, and it is believed that a bird within a flock tries to keep the same distance to all its neighbours. Furthermore, it has been noticed that this kind of social behaviour is more frequent in preys than in predators, and since a fish at the edge of the school is more likely to be caught, it is natural to think that they would try to move towards the centre of the group!

Another influential work on the simulation of birds is that of Heppner and Grenander (quoted in [46, 47]), who observed the important issue that natural bird flocks do not have a leader! That is, there is no central control! Heppner and Grenander implemented a simulation similar to that of Reynolds, but now the birds were also attracted to a roost, and an occasional random force was implemented infrequently deflecting the birds’ direction, just like a gust of wind would do. The intensity of the attraction was programmed to increase with the decrease of the distance to the roost. Again, the result was a realistic flock-like choreography.

## 5.3.4 Human beings

Despite possessing a very powerful brain[4], a person that is born and left alone in the world cannot learn much of it in a whole lifetime. It is self-evident that the achievements of the greatest human minds result in benefits for the whole species. Besides, even for simpler activities such as building a house, the DoL in humans is undoubtedly advantageous, mainly due to the resulting specialization, and to the fact that the DoL is performed according to the needs and to the abilities of the individuals involved.

*We humans are the most social of animals: we live together in families, tribes, cities, nations, behaving and thinking according to the rules and norms of our communities, adopting the customs of our fellows, including the facts they believe and the explanations they use to tie those facts together. Even when we are alone, we think about other people, and even when we think about inanimate things, we think using language—the medium of interpersonal communication* [47].

Even human thinking seems to be a social activity. Kennedy et al. [47] claim that thinking and learning take place within a complex cognitive space by means of collaborative processes. For example, people do not deal with huge matrices of words when chatting, yet the interrelations of an immense number of words are understood so that verbal communication is possible. Even not so brilliant-minded people can display these communication skills, and they do not do so by studying the meanings of words from the dictionary, in a top-down definitional symbolic fashion. The meanings of words largely depend on the context, and people learn them from one another in a cooperative fashion.

[4] Even the achievements of the brain rely on the behaviour that emerges from the interactions among its neurons.

### 5.3.5 Cellular robots

In the early days of robotics, the symbolic paradigm dominated the approaches to AI, so that the robot minds were typically implemented in a central executive processor, assigning will to a central role in an artificial mind. Hence the systems required inference engines, so that when a robot's sensors received some inputs, the inference engine would send some precise outputs to the effectors. The problem with the symbolic approach was that, when the environment became more complex, the system had more symbols to retain, and the inference engine more and more complicated chains of logic to analyze. The result is that, even for very specialized robots and for very restricted environments, a robot might need a long time before "deciding" its next step because of the numerous calculations it requires to figure things out.

Brooks (quoted in [47]) proposes a bottom-up approach: simple, independent and specialized robots are developed such that each one performs the tasks it was specifically programmed to do. The differences with previous approaches are that each module can perform its tasks independently from what the other module is doing, and that Brooks' robots respond to the world as it presents to their sensors, thus avoiding maintaining a symbolic representation of it. Thus, the computation time, the length of the encoded programs and memory requirements are remarkably reduced by eliminating the central executive control. The most interesting concept here towards the PSO approach is that the resulting intelligence is decentralized.

However, although Brooks' specialized robots give birth to a decentralized system, each one is an autonomous agent whose behaviour is exactly what it was intentionally programmed to be. The behaviour of a so-called multi-agent system is more or less the sum of each agent's contribution, while the behaviour of a SI-based system emerges from the local interactions of the components, which are all very much alike.

*Fukuda's lab is developing robot swarms. Where Brooks … took the cognitive executive control out of the picture, decentralizing the individual robot's behaviour, Fukuda's robots replace the individual's control with reflex and reactive influence of one robot by another. These decentralized systems have the advantages that the task load can be distributed among a number of workers, the design of each individual can be much simpler than the design of a fully autonomous robot, and the processor required—and the code that runs on it—can be small and inexpensive. Further, individuals are exchangeable* [47]. These kinds of robots are sometimes referred to as "cellular robots".

# 5.4 Social learning and culture evolution

Likewise the ANNs discussed in **Chapter 3** can be viewed either as models of the human brain or as engineering mapping devices, and the EAs discussed in **Chapter 4** can be viewed either as models of biological evolution or as engineering problem-solving techniques, the PSO paradigm can be thought of either as a model of social processes or as a problem-solving technique. Despite the fact that the PSO method is used in practice mainly as an optimization method, some of the principles underlying some simulations of sociocognitive phenomena greatly influenced the development of the PSO paradigm. Therefore, a few related concepts of social psychology and of the simulation of social processes are briefly discussed hereafter, guiding the dissertation towards the basis of the PSO method.

## 5.4.1 Introduction

The concept of "mind" is even more difficult to define than the concepts of "intelligence" and "life" discussed in **Chapter 3**.

*As minds cannot be observed directly, the experience of thinking and feeling can only be described in metaphorical terms…, and throughout history people have used the symbols and technology of their times to describe their experience of thinking* [47].

The systematic demystification of the humans' uniqueness led to see humans as a system that, like any other system, is ruled by the universal laws of physics (in the broadest sense). Therefore, just like the heart is nowadays viewed as an ordinary pump, the brain and the mind are interpreted as the hardware and the software, respectively, of the cognitive system.

In addition, likewise the phenotypes are defined as the observable features that result from the interaction of the genotype and the environment, the behaviour comprises the observable features that result from the interaction of the cognitive system and the environment. Thus, the complexity of the behaviour is believed to be more due to the complexity of the environment than to the complexity of the mind. Simon (quoted in [47]) suggested that the complexity of the irregular path, nearly random but with a general sense of direction, of an ant across a beach is not a characteristic of the ant but of the environment. An ant is a simple-minded entity that is not able to achieve much on its own.

## 5.4.2 Behaviourism

The fact that, as opposed to minds, the behaviour of the organisms can be observed led almost all the research on psychology to be conducted in the "behaviourism" through most of the 20th century. *...Most research was conducted with animals ... though generalizations to humans were offered with great optimism and confidence* [47]. There were two broad mainstreams within the behaviourism. One viewed the behaviour merely as the responses of the organism to stimuli provided by the environment, without even considering the possibility of curiosity as the driving force. The other mainstream *...emphasized "operant conditioning", in which the organism acts on its environment in order to obtain reinforcement* [47]. In either case, behaviourist psychology was characterised by the development of elaborate formulas that linked stimuli to responses, based on strict empirical experimentations and observations.

## 5.4.3 Cognitivism

There is no general consensus with regards to the date of birth of the "cognitive revolution". It is fair to remark that the evolution of the fields of AI and psychology are strongly correlated. Thus, while the connectionist paradigm came back to challenge the symbolic paradigm in AI by mid-1980s (although the first ideas related to it were contemporary to the origins of the symbolic paradigm), some suggest that, if not the cognitive revolution, at least the end of the behaviourism can be claimed to have begun in 1974. By that time, Bandura offered: *It is true that behaviour is regulated by its contingencies, but the contingencies are partly a person's own making. By their actions, people play an active role in producing the reinforcing contingencies that impinge upon* *them* (quoted in [47]). *Bandura argued that cognitive processes must be considered in explaining human behaviour* [47]. Thus, cognitivism inserted internal cerebral computations between the stimuli and the responses, transforming the stimuli into representations that are manipulated to produce the behaviour in a bottom-up fashion. Beware that the ANN paradigm in AI can be viewed as a simplistic model of a biological brain, whose inputs can be viewed as stimuli and its outputs as the behaviour of the organism.

### 5.4.4 Social learning

More or less independently from the dispute between behaviourism and cognitivism, "social psychology" is concerned with the individual in a social context. A few important theories and experiments in social psychology are briefly discussed hereafter, without intending to analyze them in depth, but merely to show the evolution of the ideas in psychology that helped somehow to give birth to some problem-solving methods such as memetic algorithms, cultural algorithms and the PSO paradigm, which were clearly not invented out of nowhere!

An important and influencing theory that was brought into the early social psychology was the "Gestalt theory", which was primarily concerned with the organization of fragmentary perceptions into coherent wholes. Another greatly important theory developed in the early social psychology was the "Lewin's field theory", which portrayed a life-space wherein both the individual and the environment are represented by bounded interconnected regions, while regions that are interconnected are able to influence one another.

In 1936, Sherif (quoted in [47]) reported experiments demonstrating the convergence of individuals' perceptions. He placed subjects in a dark room with a stationary spot of light projected on a wall. When asked in isolation, the individuals tended to report that the spot had been moving, although the range of the movement reported varied from person to person. However, when they were asked to make the report in public, the reports tended to converge!

In 1956, Asch (quoted in [47]) reported that when subjects in an experiment were faced with the dilemma of giving the obvious true answer versus agreeing with the group, about a third of them chose to agree with the group despite knowing the answer was plainly wrong!

In another important contribution, Bandura (quoted in [47]) announced in 1965 the discovery of "no-trial learning", arguing that humans can learn a task without trying it, by observing somebody else doing it with successful results. Kennedy et al. [47] suggest that a person can do something that he or she would not have thought of trying, simply by imitating other people's successful behaviour. This so-called "social learning" *...is a very important form of learning for humans and seems hardly to exist at all among other species.* [47]

Note that the tendency to seek agreement manifested in Sherif's experiment, the conformism observed in Asch's experiment, and Bandura's social learning, all support the belief that when people interact, they become more similar to one another. This is the basic concept underlying the cultural algorithms briefly discussed later in this section.

In another line of thought, it has been asserted that the process of thinking itself is a social process. *For instance, findings that coordinated cognitive activities ... evoke ... shared understanding of the topic, help explain why groups are sometimes better able than individuals to perform certain kinds of tasks—and why they perform worse when they rely too much on shared information* [47]. Levine et al. (quoted in [47]) offer: *Although some might claim that the brain as the physical site of mental processing requires that we treat cognition as a fundamentally individual ... activity, we are prepared to argue that all mental activity—from perceptual recognition to memory to problem solving—involves either representations of other people or the use of artifacts and cultural forms that have a social history.*

Latané (quoted in [47]) suggested in his "social impact theory" that the influence of a group of people over an individual is a function of the strength, immediacy, and the number of people in the group. The strength is just a kind of social persuasiveness, and immediacy is inversely proportional to the distance. While the influence increases, the rate of increase decreases with the number of individuals in the group. Beware that Latané's model strongly resembles some of the models previously discussed for fish schools!

*It may not be the most flattering self-image for us to view ourselves as flocking, schooling conformists. But the literature of social psychology since the 1930s consistently shows us to be ... herding creatures. Whenever people interact they become more similar... Norms and cultures, and, we would say, minds, are the result. It is not the usual case that humans move in synchronously choreographed swoops and dashes across the landscape, as fish and birds do; human physical behaviours are not flocklike..., but the trajectories of human thoughts through high-dimensional cognitive space*[5] *just might be.* [47]

[5] Note that, as opposed to the physical space, collision is not an issue in the cognitive space. In fact, it is something usually desirable because it implies agreement.

Since the rising connectionist paradigm in AI (and in cognitive psychology) appeared to be consistent with traditional social psychology theorizing, the field of social psychology slowly started accepting computer simulations as useful research tools. For instance, *Thagard theorized that people understand events by placing them in the context of a narrative or explanation—a story. ... He supported his theory of explanation by encoding it in a connectionist computer program* [47].

Social psychologists that experimented with Thagard's program arrived to the conclusion that their many theories addressing the tendency or need for individuals to minimize conflicts (or maximize coherence) among their beliefs, attitudes and behaviours can be represented by a properly designed parallel constraint satisfaction network. Thus, the aim is to find the states of the nodes of the network that maximize a coherence function. While the introduction of the connectionism into the social psychology field is a step forward in the simulation of social processes, it still deals with the learning as an individual process.

In 1995, Hutchins (quoted in [47]) represented each of four individuals as a parallel constraint satisfaction network, which has two globally optimal activation patterns that satisfied the constraints equally well. He found that when the individuals were highly interconnected, a kind of unique mega-mind was formed, and the entire population moved towards a solution that could be highly inconsistent. When the individuals were isolated, they tended to converge on one or the other globally optimal activation pattern. However, when the individuals were moderately interconnected, the entire population converged on the same optimal pattern!

*...Hutchins' analysis suggests that moderate ignorance permits not only cognitive consistency, but agreement among members of a group.* [47]

## 5.4.5 Cultural evolution

The same as the "mind" was claimed to be very difficult to define, an accurate definition of "culture" requires a much more profound discussion than it is possible to carry out here. Without aiming to present a polemical definition, the term "culture" in this thesis refers to a supra-individual pattern of beliefs and behaviours that emerge from the individual beliefs and behaviours within a group of people. Thus, the culture can be viewed as a "swarm of minds".

Benedict *...noted that the average person conceptualizes an antagonistic relationship between the individual and society, as if something called "society" was forcing people to obey its rules. Yet, ...the* *individual and the culture are two aspects of a single process: "No individual can arrive even at the threshold of his potentialities without a culture in which he participates. Conversely, no civilization has in it any element which in the last analysis is not the contribution of an individual".* [47]

*With the emergence of complex dynamical systems research through the 1980s, Latané began to expand the predictions of social impact theory to show that the behaviours of individuals could be explained in terms of self-organizing properties of the social system they comprised* [47]. While it had been claimed that any mathematical model of interacting individuals would result in uniformity, Latané's simulations showed that although diversity is reduced as individuals interact, different divergent subpopulations were formed, within each of which individuals' beliefs and behaviours tended to converge. Therefore, a polarization gives birth to different regions, within each of which individuals tend to reach agreement. *Latané's model is at least approximately consistent with findings in the field of social psychology and also in sociology, economics and anthropology. As people interact they persuade one another of things, they show one another how to do things, they impress one another, they copy one another, and the simple, obvious result is they become more similar* [47]. It should be remarked that this model does not consider the interrelations among the different beliefs that a single individual possesses, while they should be logically interrelated for that individuals strive for consistency.

Recall that the adaptation of human beings to the environment is slowly performed through the evolution of their genetic code, and more quickly by learning (see section **3.2.3.5**). While it is self-evident that human beings learn individually from their own experience, it has been discussed in previous sections that they can also learn from the imitation of others' successful behaviours. Besides, experiments have shown that humans' behaviour is directly influenced by the behaviour of their neighbours, and also indirectly influenced by farther people's behaviour by means of social norms and culture. In 1985, Boyd and Richerson (quoted in [47]) developed their dual-inheritance mathematical model of human behaviour, reasoning that an individual's learning from its own experience should be more adaptive for more stable and stationary and homogeneous environments, whereas social learning should be more adaptive for highly heterogeneous and dynamic environments because the individual cannot experience much of it by itself. Since the phenotypes result from the interactions between the genotypes and the environment (refer to section **4.2** and **Appendix 2**), the characteristic human social behaviour must be partly due to the human genetic code. Following a line of thought that seems to be a kind of extension of the Baldwin effect[6] discussed in section **3.2.3.5**, Boyd and Richerson hypothesize that the tendency to learn more individually or socially (cultural transmission) was genetically evolved. *...in humans it appears that social learning was favoured by evolution* [47]. Furthermore, while the predominant type of learning is evolved, it influences the flow of genetic evolution. For instance, if the individual learning prevails, there is more variety in the expressed phenotypes thus reinforcing the importance of genetic evolution, whereas if the cultural transmission prevails, individuals' phenotypes become more similar to one another thus dulling the effect of natural selection[7].

In 1998, Henrich and Boyd (quoted in [47]) conducted computer simulations where the individuals were awarded different degrees of activation of two special genes: the social learning gene (L) and the conformist gene (Δ), which affected the cultural transmission. Thus, L = 0 meant that the individual only displayed individual learning, Δ = 0 meant that the social learning was performed by pure imitation of the observed behaviours, while Δ = 1 meant that the probability of adopting a behaviour depended on the frequency of the behaviour in the population (conformist learning).

The results of the simulations showed that selection favours conformist transmission over a wide range of environments. Kennedy et al. [47] arrived to the conclusion *...that cultures form simply because conformist learning is adaptive*.

## 5.4.5.1 Memetic algorithms

Many researchers have pointed out the profound similarities between biological evolution and human culture. For instance, *Campbell described creative thinking in terms of "blind variation and selective retention" and suggested that mental creativity might be very much analogous to the process of biological evolution* [47]. Similarly, *Fogel theorizes ... that "sociogenetic learning" is an intelligent process in which the basic unit of mutability is the idea, with culture being the reservoir of learned behaviours and beliefs. "Good" adaptive ideas are maintained by the society, much as good genes increase in a population, while poor ideas are forgotten* [47].

---

[6] The Baldwin effect asserts that in an indirect way, *...learning can affect evolution, even if what is learned cannot be transmitted genetically* [56]

[7] Since the latter is the usual case of human beings, it has been polemically suggested through recent history that "artificial selection" should be performed!

Because of these similarities, the same as the biological evolution is thought of in terms of **genetics**, whose unit of transmission is the gene, the cultural evolution is often thought of in terms of **memetics**, whose unit of transmission is the meme. However, the analogy between natural evolution and cultural evolution has been severely criticized. *The lack of a consistent, rigorous and precise definition of a meme remains one of the principal criticisms leveled at memetics... Different definitions of meme generally agree, very roughly, that a meme consists of some sort of a self-propagating unit of cultural evolution having a resemblance to the gene...* [80].

The analogy between genetics and memetics is controversial, there being supporters, detractors, and some who suggest that it is very useful despite the fact that the analogy is not exact and despite several weaknesses in the theory of memes. Although these discussions are way beyond the scope of the present work, it is fair to note that *...cultural change and biological evolution are similar in that they are dynamic, stochastic, adaptive, and occur in populations. But where natural selection is primary in Darwinian evolution, it is a minor aspect in cultural change* [47].

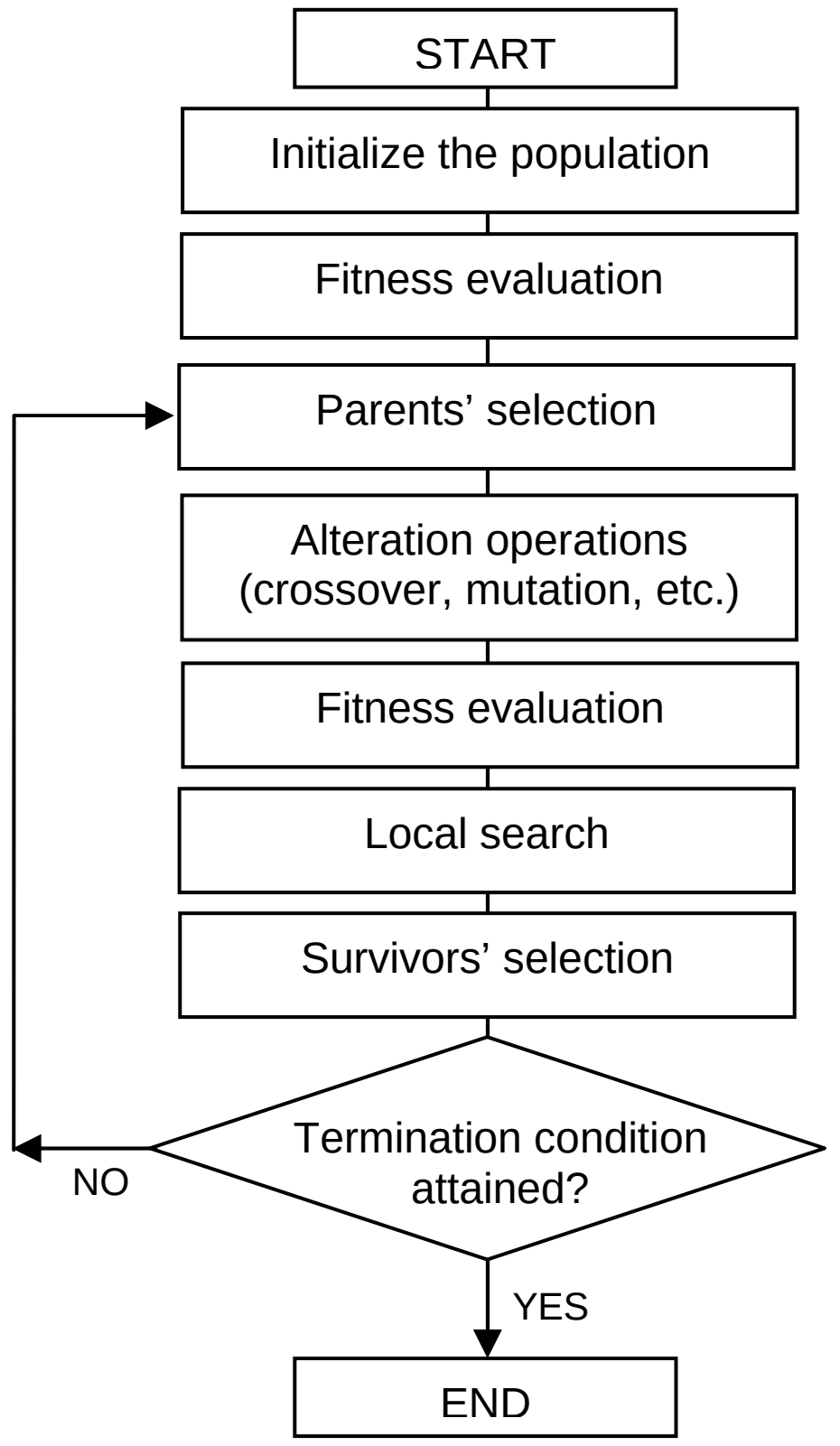

**Fig. 5. 1**: Generic memetic algorithm flow chart.

The so-called **memetic algorithms** (MAs) are basically EAs that are provided with an internal local search strategy. Despite the name, there is little connection to the theory of memes. The fundamental concept underlying MAs is that, while a population evolves through generations, like in any EA, each individual performs its own search during its own lifetime, resembling learning in real life. Hence the MAs resemble the Baldwin effect discussed in section **3.2.3.5**, rather than cultural evolution.

The implementation of a MA is very flexible, so that its structure can be adapted to the problem at hand. The kind of EA to be implemented can be chosen among those discussed in **Chapter 4**, and the local search technique is typically chosen among the hill-climber, SA and TS. A generic MA's flow chart is shown in **Fig. 5. 1**.

### 5.4.5.2 Cultural algorithms

*If culture enables human individuals to adapt to complex environments, then perhaps the processes that drive culture can be used to enable adaptation of artificial individuals to other kinds of fitness landscapes* [47]. Similar to the EAs, Reynolds' (quoted in [47]) **cultural algorithms** are population-based methods, their initial population is typically randomly generated, and they make use of operators that aim to improve the fitness of the population's members. However, as opposed to EAs, cultural algorithms maintain a group memory of the proposed solutions that performed well. In the same fashion as TS, cultural algorithms sometimes also maintain a memory of some poor solutions that should be avoided in the future.

The memetic algorithms perform a slow adaptation by means of evolution, and a more rapid adaptation by means of individual learning (implemented as a local search). Instead, cultural algorithms suggest that the rapid individual-to-individual transmission of traits, beliefs and behaviours is constrained by the culture. The culture, which is formed by the generalization of the traits, beliefs and behaviours of some individuals in the near history, evolves over time in a very slow fashion, thus performing a group-level adaptation.

It is important to remark that although both memetic algorithms and cultural algorithms emphasize a two-level adaptation, the former rely on individual learning during life-time plus biological-like evolution through generations, whereas the cultural algorithms emphasize individual learning by means of transmission of traits, beliefs and behaviours between individuals, constrained by a slowly evolving culture, which is formed and updated based on individual experiences. Although it has been argued that cultural evolution influences the flow of genetic evolution, the latter is not considered in cultural algorithms. The evolution of culture here is due to the influence of individuals who pass an acceptance test.

In summary, cultural algorithms initialize the behaviours of a population of individuals. The performance of each individual is measured by means of a suitable "performance function" (equivalent to the "fitness function" in EAs). An "acceptance function" determines whether the individual at issue should contribute to the culture, which will affect the individuals' behaviours in the next time step.

Lately, it seems *…that every conference has at least a few cultural algorithm papers, and it seems likely that this flexible and powerful approach to adaptive problem-solving has a bright future.* [47]

At first glance, it seems promising to further investigate multi-level adaptive algorithms such as, for instance, population-based methods whose members display individual learning, social learning by mimicking close neighbours (in accordance with Bandura's "no-trial learning"), and being affected by a slowly-evolving culture[8] (thus being influenced by the experiences of distant individuals), while a biological-like evolution is taking place.

### Adaptive culture model

Axelrod's (quoted in [47]) culture model is a computer simulation of the spread of features within a culture. It has been asserted before that by means of interactions, individuals become more similar to one another. However, individuals might choose (to some extent) who to interact with and whose features from the other person to adopt. Axelrod theorized that "similarity" is the driving force for individuals to interact. Therefore, he represented individuals as strings of features, and implemented the probability of interaction between two individuals as a function of their similarity. That is to say that the more similar two persons are, the more likely it is that they will interact thus becoming even more similar. This model is consistent with the observed formations of divergent subpopulations of similar individuals.

Thus, a two-dimensional matrix representing a torus is initialized assigning random features to the individuals. Each individual can interact with its four closest neighbours, while the probability of adopting a feature from a neighbour equals the percentage of matching features between them. If they pass this probability threshold, the individual at issue adopts one feature from its neighbour, randomly selected from the non-matching ones. The algorithm is run iteratively until a stationary state is reached. The result is the formation of homogeneous subpopulations separated by fixed boundaries. That is to say that, as time goes by, the subpopulations tend to become more homogeneous, and more different from one another.

Kennedy et al. [47] assert that *...there is evidence to suggest that the role of similarity is not as important as Axelrod theorizes.* They made some modifications to the original Axelrod's culture model[9] and ran a few interesting experiments:

1. The similarity as a driving force for the interaction was eliminated from the model. The experiment resulted in a uniform population of individuals that displayed identical features.

[8] Beware that while the culture influences individuals' behaviours, some individuals' behaviour influence the culture!

[9] They called "Adaptive culture model" (ACM) the algorithms they derived form "Axelrod's culture model".

2. Kennedy et al. [47] claimed that the similarity as a driving force in Axelrod's model makes it an objective function. Thus, they modified the causes of the interaction: if the summation of the neighbour's features is higher than the summation of the individual's, a non-matching feature is randomly adopted. The result was that the model maximized the summation of the individuals' features (which stands for an objective function!), and again, all the individuals ended up displaying identical features. The conclusion is that the algorithm is able to optimize a simple objective function.

3. While the objective function in the previous experiment has a single global optimum, a slightly harder function with multiple optima was implemented. The strings representing the individuals were split in two parts, and the objective was to find the features that made the summation of the features in a sub-string equal to the summation of the features in the other. Then, the optimization problem consisted of minimizing the differences between the summations of the two sub-strings. The result was that all the individuals in the population found a pattern of features that solved the problem, although a number of solutions corresponding to different local optima were found. Similarly to the original Axelrod's model, different divergent homogeneous subpopulations were formed with fixed boundaries.

*In the ACM paradigm, an individual takes a feature from a randomly selected neighbour if a criterion is met. In Axelrod's writings the criterion is similarity; the …* ACM *… substitutes other criteria and shows that the spread of culture can optimize other functions, resulting, by the way, in similarity among proximal individuals. Similarity, which was a cause in Axelrod's simulations, is now an effect.* [47]

4. In another experiment, Kennedy et al. [47] implemented an eight-city tour TSP. The problem was built up such that the global optimal solution was known, and the cities were defined as two-dimensional Cartesian coordinates. A penalty function was designed for tours going through a city more than once. The ACM was run 20 times, in 11 of which the population found the global optimum. When suboptimal tours dominated the population, each suboptimal solution differed from the global best in that the order of only two neighbour cities were reversed or in that a single city was visited twice. In some other experiments with different TSP with more than one optimum, polarization sometimes occurred.

Kennedy et al. [47] successfully used their ACM model to solve other hard problems such as a "parallel constraint satisfaction problem" and "symbol processing".

To sum up, the ACM is a family of iterative population-based algorithms whose individuals interact until a stable point is reached and change ceases. The different versions of the ACM can find a stable point where the strings end up in unanimity, or where they form regions separated by fixed borders composed of strings that are identical within a region but different between them. In the same fashion as the EAs discussed in **Chapter 4**, these algorithms proved to be able to solve problems they were not specifically designed for. For instance, there is nothing in the algorithm that aims at forming well-defined homogeneous regions!

Thus, the resulting behaviour of a population could be arguably claimed to occur in fourth levels. First, the individuals in quest for solutions learn from their own experiences, they also learn from imitations of others' successful behaviours, although restricted by a slowly evolving culture, which acts as a reservoir of successful behaviours (hence its influence is more profitable for stationary or slowly changing environments). The culture allows individuals to take advantage of the results achieved by individuals they do not have the chance to interact with. Finally, these three levels could be coupled with the biological evolution, which also adapts to the environment seeking to achieve the organism's goals.

# 5.5 Ant colony optimization

## 5.5.1 Introduction

Although the ACO paradigm is clearly a SI-based algorithm, its links to the social psychology field are vague. The very first algorithm was developed by mimicking the behaviours of some colonies of Argentinean ants, which were seen to find the closest path from their nest to a food source without any previous leading clue. *The ant colony optimization algorithm (ACO), introduced by Marco Dorigo in his doctoral thesis in 1992, is a probabilistic technique for solving computational problems which can be reduced to finding good paths through graphs. They are inspired by the behavior of ants in finding paths from the colony to food.* [80]

These ant-based algorithms are especially suitable for solving combinatorial optimization problems. Because the different combinatorial optimization problems greatly differ from one another, there is a family of algorithms resulting from adaptations to the original algorithm so as to handle different combinatorial problems.

The names given to the different algorithms and families of algorithms vary in the literature. For instance, they can be referred to as “ant algorithms”, “ant colony algorithms”, “ACO algorithms”, “ant system”, “ant colony system”, “ACO meta-heuristic”, etc. While some of these names are meant to be synonymous, some refer to different adaptations of the original ACO paradigm. Here, the term “ACO algorithms” encompasses all the different variations.

Experimental research on live Argentinean ants (quoted in [11, 27]) showed that they are capable of finding the shortest path from their nest to a food source without relying on visual cues (see **Fig. 5. 2**).

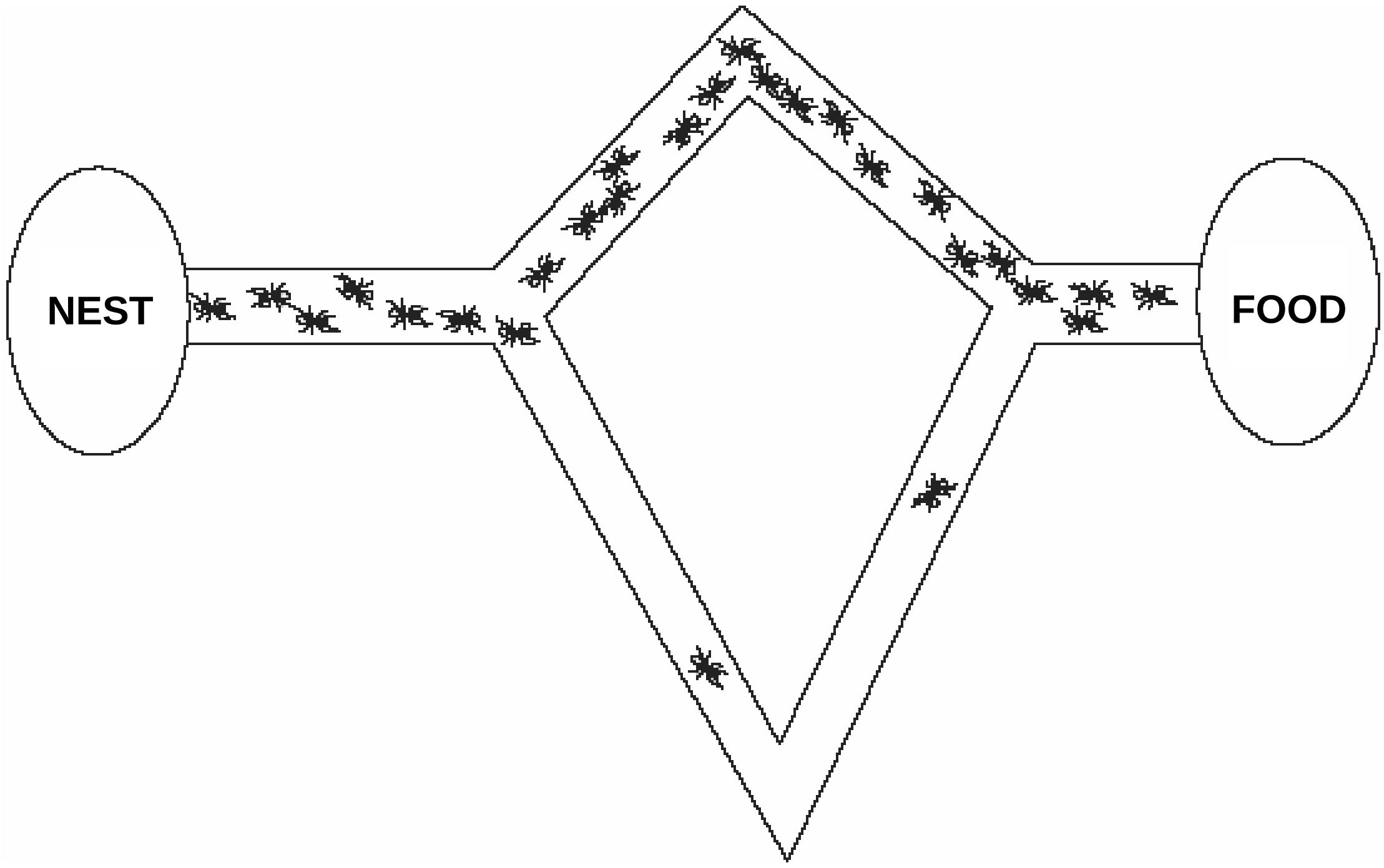

**Fig. 5. 2**: Representation of an experiment showing that a colony of Argentinean ants is able to find the shortest path from the nest to a food source. They collaborate with one another by means of stigmergy (adapted from Goss et al. –quoted in [47]–).

This is accomplished thanks to the ants’ stigmergic pheromone-trail-laying and pheromone-trail-following behaviour, as discussed in section **5.3.2**. Thus, when ants move either from the nest to a food source or from a food source to the nest laying their own pheromone trails, their probability of following a certain trail is a function of the pheromone concentration. Thus, when an ant reaches a decision point (like the ones shown in **Fig. 5. 2**), the initial decision is random. Since the ants move at an approximately constant velocity, the ones that choose the shorter path will get either to the food source or to the nest (and to the other decision point) sooner. In addition, they will also meet the ants coming in the opposite direction sooner. The obvious result is that the concentration of pheromone in the shorter path progressively increases while the concentration in the longer path decreases. Recall that the number of ants involved in the task needs to be greater than a minimum problem-dependent threshold in order for the autocatalytic (i.e. reinforcing) effect to occur. Eventually all the ants ends up following the shortest path, although there is always a small chance that an ant makes a "mistake" and gets lost, enabling it to find a new better food source.

Other experiments showed that once the ants follow the shortest path, if an obstacle is interposed in their way, they also find the shortest path to surround it, as shown in **Fig. 5. 3**.

However, once the ants are following the shortest path they managed to find in the present environmental conditions, if the removal of an obstacle results in a new possible path that is shorter than the one they are following, they are not able to find the new path. Of course, there is a chance that the lost ant mentioned before can find it, but the chance is negligible.

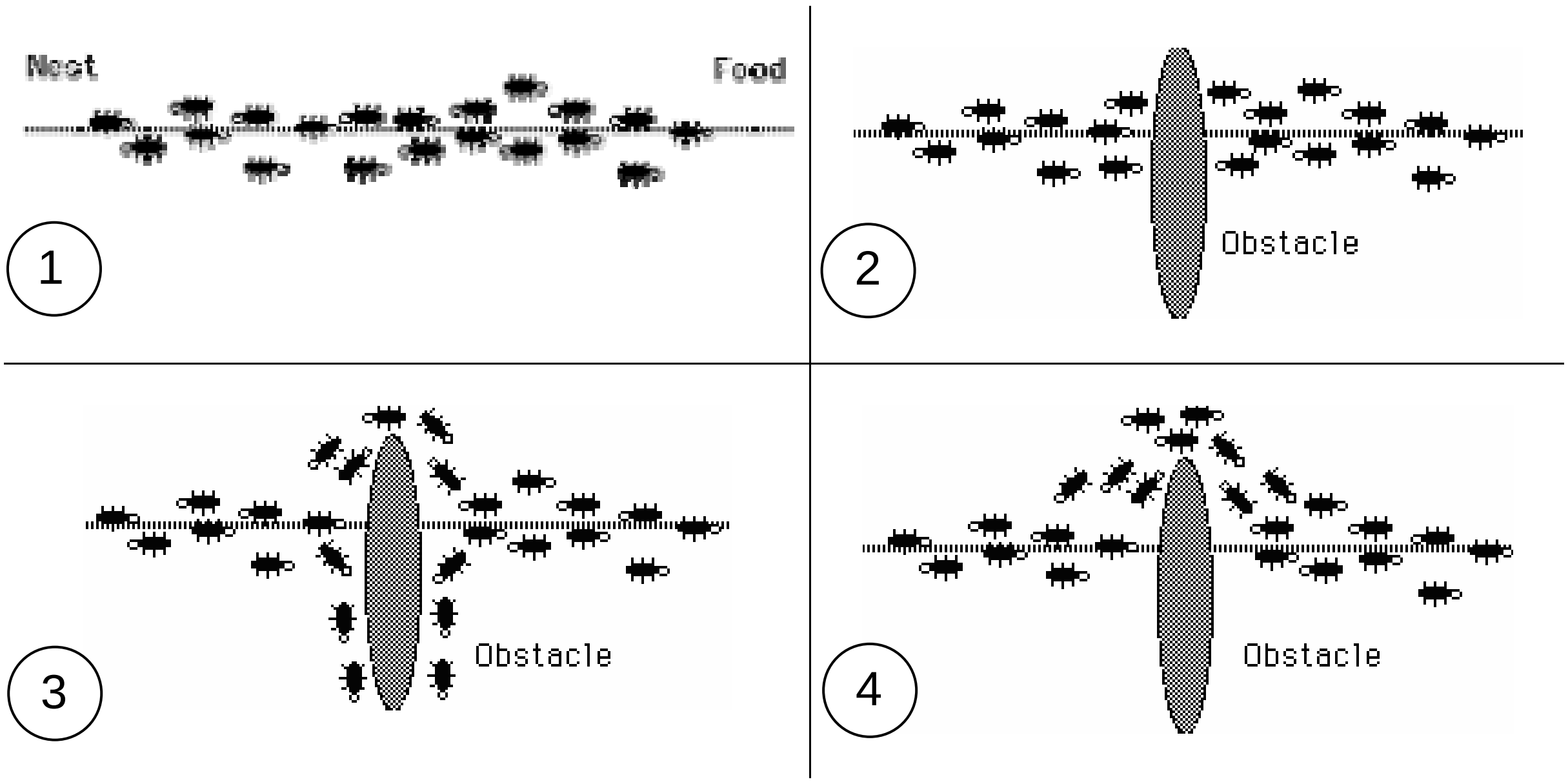

**Fig. 5. 3**: Example of obstacle-avoidance in an ant colony by means of stigmery (adapted from [24]).

If the ants are offered two food sources at the same distance from the nest, they end up exploiting the richer source. However, once the source has been chosen, if the other source is made richer than the one the ants are exploiting, they are not able to switch. Not surprisingly, if both food sources are equally rich, the decision appears to be random.

The idea behind the original ACO algorithm was to mimic the natural ants' behaviour with simulated ants travelling through a graph that represents the problem to be solved. Later developments of the algorithm so as to improve its efficiency led to the use of artificial ants rather than simulated ants, since some of the new behaviours do not have natural counterparts.

*Ant colony optimization algorithms have been used to produce near-optimal solutions to the traveling salesman problem. They have an advantage over simulated annealing and genetic algorithm approaches when the graph may change dynamically; the ant colony algorithm can be run continuously and adapt to changes in real time. This is of interest in network routing.* [80]

The application to the TSP is almost immediate because it is very similar to finding the shortest path from the nest to a food source. Nonetheless, subsequent developments gave birth to other algorithms of the family that are suitable for solving other complex combinatorial problems, which are well beyond the scope of this brief overview.

## 5.5.2 The ant system

The "ant system" (AS) is one of the simplest ACO algorithms. It is fair to remark, however, that an even simpler algorithm can be found in [25]. Due to the similarities to the natural metaphor, the first ACO algorithms were applied to the TSP.

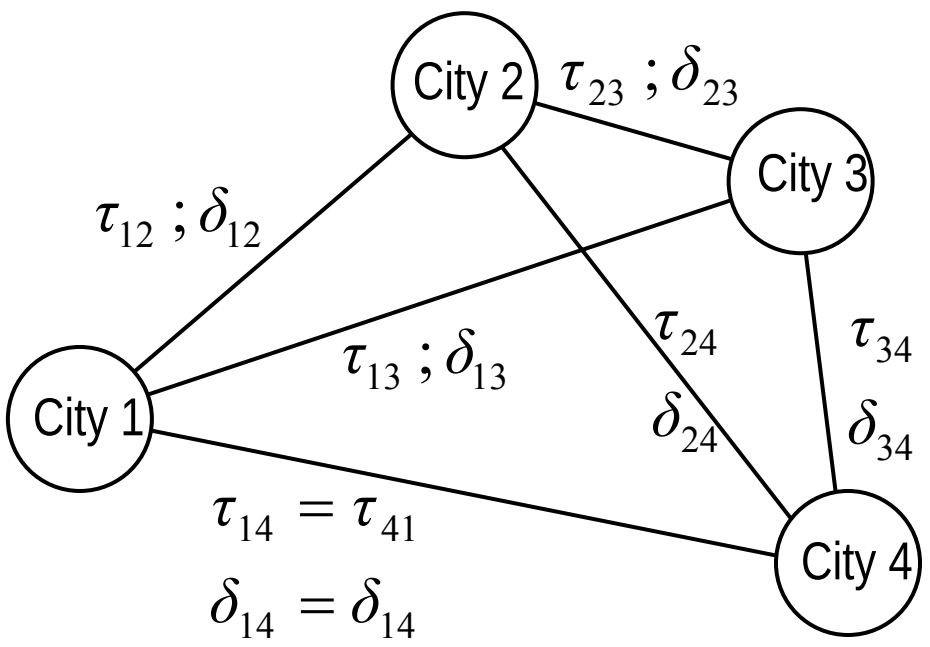

**Fig. 5. 4**: Graph of a four-city symmetric TSP to be solved by an ant algorithm.

An example of a graph of a simple symmetric four-city TSP is shown in **Fig. 5. 4**, where $\tau_{ij} = \tau_{ji}$ is the "desirability" of moving from city *i* to city *j* and vice versa (i.e. the strength of the pheromone trail), and $\delta_{ij} = \delta_{ji}$ is the cost of that movement.

The cities are typically represented by their Cartesian coordinates, and the cost by the Euclidean distance.

The basic algorithm works by randomly initializing a swarm of ants, and each ant is sent to build a complete tour by choosing the cities according to a "state transition rule". Once all the tours are completed, the pheromone trails are updated according to a "global updating rule".

### 5.5.2.1 The state transition rule

An ant in a certain city chooses which city to move to by probabilistically favouring the short edges and the high pheromone concentrations among all the possible valid moves.

Thus, the so-called "random-proportional rule" shown in equation **(5. 1)** is the state transition rule that gives the probability with which the ant $k$ in city $i$ chooses to move to city $j$ at time $t$.

$$P_{kij}^{(t)} = \begin{cases} \dfrac{\tau_{ij}^{(t)} \cdot \left[\eta_{ij}\right]^{\beta}}{\sum\limits_{s \in \mathcal{N}_{ki}} \tau_{is}^{(t)} \cdot \left[\eta_{is}\right]^{\beta}} & \text{if } j \in \mathcal{N}_{ki} \\ 0 & \text{otherwise} \end{cases} \qquad \textbf{(5. 1)}$$

Where:

- $\tau_{ij}^{(t)}$ : pheromone concentration in the edge that links the cities $i$ and $j$ at time step $t$
- $\eta_{ij} = \dfrac{1}{\delta_{ij}}$ : inverse of the Euclidean distance between the cities $i$ and $j$
- $\beta > 0$ : parameter that sets the importance of the pheromone in relation to the distance
- $\mathcal{N}_{ki}$ : set of feasible cities that remain to be visited by ant $k$ located in city $i$

A small amount of pheromone is assigned to all the edges at the beginning. Clearly, equation **(5. 1)** favours the shorter paths, and the ones with higher concentration of pheromone.

### 5.5.2.2 The global updating rule

Once all the ants in the swarm have built up their tours, the concentration of pheromone is updated on all edges according to the following global updating rule:

$$\tau_{ij}^{(t+1)} = (1-\alpha) \cdot \tau_{ij}^{(t)} + \sum_{k=1}^{m} \Delta\tau_{kij}^{(t)} \quad ; \quad \Delta\tau_{kij}^{(t)} = \begin{cases} \dfrac{1}{L_k^{(t)}} & \text{if } l_{ij} \in L_k^{(t)} \\ 0 & \text{otherwise} \end{cases} \qquad \textbf{(5. 2)}$$

Where:

- $0 < \alpha < 1$ : pheromone decay parameter
- $L_k^{(t)}$ : length of the complete tour performed by ant $k$
- $l_{ij}$ : length of the edge that joins the cities $i$ and $j$
- $m$ : number of ants in the swarm

As shown in equation **(5. 2)**, the global updating rule serves both functions, pheromone-laying and pheromone evaporation.

*Although ant system was useful for discovering good or optimal solutions for small TSPs (up to 30 cities), the time required to find such results made it unfeasible for larger problems.* [27]

## 5.5.3 The ant colony system

There are many variations of this simple original algorithm that aim either to improve its efficiency or to adapt it for different combinatorial problems. While a few variations proposed by Dorigo et al. [27] are briefly discussed in this section, more advanced and detailed work on the subject matter can be found in [11, 25, 26].

Acknowledging the limitations of the AS, Dorigo et al. [27] designed the so-called ant colony system (ACS), which differs from the AS in the implementations of the "state transition rule" and of the "global updating rule", and in that it adds a new stage: the "local updating rule".

### 5.5.3.1 The state transition rule

The ACS proposes a state transition rule, known as the "pseudo-random-proportional rule", that provides a balance between exploitation and a biased exploration of new solutions. Thus, an ant that is located in city $i$ decides to move to city $j$ according to the following rule:

$$\text{if } U_{(0,1)} \le q_0 : \quad j = \arg\max_{u \in \mathcal{N}_{ki}} \left\{ \tau_{iu}^{(t)} \cdot \left[\eta_{iu}\right]^{\beta} \right\}$$

$$\text{otherwise}: \begin{cases} P_{kij}^{(t)} = \dfrac{\tau_{ij}^{(t)} \cdot \left[\eta_{ij}\right]^{\beta}}{\sum_{s \in \mathcal{N}_{ki}} \tau_{is}^{(t)} \cdot \left[\eta_{is}\right]^{\beta}} & \text{if } j \in \mathcal{N}_{ki} \\ P_{kij}^{(t)} = 0 & \text{if } j \notin \mathcal{N}_{ki} \end{cases} \qquad \textbf{(5. 3)}$$

Where:

- $U_{(0,1)}$ : random number generated from a uniform distribution in the range [0,1], resampled anew each time it is referenced
- $0 < q_0 < 1$ : parameter of the system that regulates the relative importance of exploitation $(q \le q_0)$ versus a biased exploration $(q > q_0)$
- $\tau_{ij}^{(t)}$ : pheromone concentration in the edge that links the cities $i$ and $j$ at time-step $t$

- $\beta > 0$ : parameter to set the importance of the pheromone in relation to the distance
- $\eta_{ij} = \dfrac{1}{\delta_{ij}}$ : inverse of the Euclidean distance between the cities *i* and *j*
- $\mathcal{N}_{ki}$ : set of feasible cities that remain to be visited by ant *k* located in city *i*

Notice that if $q_0 = 1$, the decision would always consist of choosing the best valid edge (no exploration is performed), while if $q_0 = 0$, the state transition rule becomes almost the one proposed for the AS in equation **(5. 1)**, enabling a biased exploration (the exceptions would be the rare cases when the randomly generated number $q$ equals zero).

Think, for instance, of $q_0 = 0.5$. Then, half the ants, on average, will choose the best edge for their next location, while the other half will perform a probabilistic selection, tending to choose one of the better edges. Thus, this new state transition rule allows a better exploitation of the areas of the graph which, according to previous experiences, seem to be promising. The result is that the performance of the whole process is enhanced in relation to that of the AS (recall that Dorigo et al. [27] reported that the AS becomes too slow for more than 30 cities).

### 5.5.3.2 The global updating rule

As opposed to the AS, here the global updating rule only updates the concentration of pheromone on the edges belonging to the best tour found by any ant up to the current iteration. An alternative is to update the level of pheromone on the edges belonging to the best tour found at the current iteration. Beware that the paths followed by the natural metaphor and by the artificial algorithm are slowly bifurcating!

$$\tau_{ij}^{(t+1)} = (1-\alpha)\cdot\tau_{ij}^{(t)} + \alpha\cdot\Delta\tau_{ij}^{(t)} \quad ; \quad \Delta\tau_{ij}^{(t)} = \begin{cases} \dfrac{1}{L_{gb}} & \text{if } l_{ij} \in L_{gb} \\ 0 & \text{otherwise} \end{cases} \qquad \textbf{(5. 4)}$$

Where:

- $0 < \alpha < 1$ : pheromone-regulation parameter
- $L_{gb}$ : length of the complete globally best tour built from the beginning of the trial
- $l_{ij}$ : length of the edge that joins the cities *i* and *j*

The global updating rule proposed in equation **(5. 4)** together with the state transition rule proposed in equation **(5. 3)** is intended to guide the search towards the neighbourhood of the best tour found up to the current iteration of the algorithm. Notice that the global pheromone updating is performed once all the ants have built their complete tours.

### 5.5.3.3 The local updating rule

The local pheromone updating rule enables ants to modify the concentration of the edges they pass through while building their own tour. Thus, ants influence each other while they construct their solutions. The local updating rule is given by:

$$\tau_{ij}^{(t)} = (1-\rho)\cdot\tau_{ij}^{(t)} + \rho\cdot\Delta\tau \qquad ; \qquad \Delta\tau = \tau_0 \qquad \textbf{(5. 5)}$$

Where:

- $0<\rho<1$ : pheromone-regulation parameter
- $\tau_0$ : initial pheromone level

Dorigo et al. [27] claim that *...the effect of local updating is to make the desirability of the edges change dynamically ... without local-updating all ants would search in a narrow neighbourhood of the best previous tour.*

*Millonas argues that the intelligence of an ant swarm arises during phase transitions—the same transitions that Langton described as defining "the edge of chaos". The movements of ants are essentially random as long as there is no systematic pheromone pattern; activity is a function of two parameters, which are the strength of pheromones and the attractiveness of the pheromones to the ants. If the pheromone distribution is random, or if the attraction of ants to the pheromone is weak, then not pattern will form. On the other hand, if a too-strong pheromone concentration is established, or if the attraction of ants to the pheromone is very intense, then a sub-optimal pattern may emerge, as the ants crowd together in a sort of pointless conformity. At the edge, though, at the very edge of chaos where the parameters are tuned correctly, ... the ants will explore and follow the pheromone signals, and wander from the swarm, and come back to it, and eventually coalesce into a pattern that is, most of the time, the shortest, most efficient path from here to there.* [47]

# 5.6 Particle swarm optimization

## 5.6.1 Introduction

The PSO paradigm was originally designed by social-psychollogist James Kennedy and electrical-engineer Russell Eberhart in 1995 [46]. The method was inspired by previous bird flock simulations, especially that of Heppner and Grenander (see section **5.3.3**). However, it is fair to remark that these studies were framed within the field of social psychology under the sociocognitive view of mind (i.e. thinking and intelligence as social phenomena), so that the PSO paradigm is also closely related to other simulations of social processes (see section **5.4**).

Similar to several paradigms previously discussed in this thesis, the emergent properties of the PSO paradigm result from local interactions amongst the individuals. Kennedy et al. [47] suggest that the behaviour of the individuals can be summarized in terms of three principles:

1. **Evaluate**: The organism evaluates the environment by evaluating the stimuli perceived by its sensors, in order to decide the proper reaction. Suppose, for instance, that each individual's mind is represented by an ANN, and each state of mind is defined by a set of weights. Good examples of this are Kennedy's "EleMentals" [44]. Every individual must be able to receive stimuli from the environment (inputs to the ANN) and make inferences (outputs) at any time, thus evaluating the state of its mind. Note that an ANN can be represented by a particle here.

2. **Compare**: Once the stimuli are evaluated, it is not straightforward to tell good from bad. Experiments and theories in social psychology, a few of which were briefly reviewed in section **5.4**, suggest that humans judge themselves by comparing to others. For instance, the strength in the social impact theory suggests that the persuasiveness of the individuals plays an important role in their influence over other individuals (successful individuals are more persuasive). Another example is that of Bandura's no-trial learning, which suggests that humans also learn socially by imitating the behaviours of other successful individuals.

3. **Imitate**: Humans compare their performances to others' and imitate only those individuals whose performance is superior or somehow desirable. So do the EleMentals [44].

While Kennedy et al. [47] arguably claim that nothing but these three processes occurs within the individual, it is merely noted here that these three processes are implemented in the PSO paradigm with remarkable success: the only sign of individual intelligence shown by the particles is a small memory. However, the PSO paradigm coupled with other paradigms can also display more intelligent beings that can make inferences, **evaluate** the goodness of their own inferences, **compare** them to the goodness of other individuals', and finally **imitate** the most successful ones. Hence the population tends to reach agreement.

Notice that the EAs, which were inspired by processes that organisms undergo in natural evolution, also perform an evaluation of the individual performance, and the "survival of the fittest" requires the comparison between the individuals' performances, while breeding can be viewed as a kind of "imitation", since it produces offspring that resemble their parents.

Kennedy et al. [46] presented the PSO paradigm as a method for optimization of continuous nonlinear functions. It is not clear what they meant by "continuous nonlinear functions", given that the presented paradigm could perfectly handle discontinuous functions provided the object variables are real-valued. Although its development was influenced by some bird flock simulations, bear in mind that the main motive was to model human social behaviour.

Thus, the graceful but unpredictable choreography of a bird flock was modelled in a singular 2-dimensional space, where collision was not an issue. A first simulation was developed so that, at each time step, each artificial bird would adopt the velocity of its nearest neighbour, while a stochastic variable called "craziness" modified some randomly chosen velocities in order to prevent the simulation from settling on a unanimous, unchanging direction. More importantly, the artificial birds in Heppner and Grenander's (quoted in [46, 47]) simulations were attracted to a roost, which led Kennedy and Eberhart [47] to think of optimization:

*Heppner's roost idea provided an inspiration that led us deep into the study of swarm intelligence. In our very first experiments, populations of "birds" flew in orderly flocking patterns. If birds could be programmed to flock toward a roost, then how about having them look for something like birdseed? It seems impossible that birds flying hundreds of feet in the air could see something as tiny as seed on the ground—but they are able to find it. ... The flock does not know where the seed is, but responds to social signals by turning back, flying past the target, circling around, spiralling in cautiously until the birds are sure they have found food in a safe place.* [47]

Encouraged by the belief that there is something about the flock dynamic that enables the birds to capitalize on each other's knowledge and discoveries, successive modifications to the initial bird flock simulation were performed. Kennedy et al. [46] described the evolution of the paradigm, from the initial orderly synchronized flock-like simulation of bird flocks to the swarm-like optimization algorithm. First, Heppner's roost (Kennedy et al. [46] named it the "cornfield vector") was introduced, and the craziness variable was deleted because the cornfield vector made it unnecessary[10]. Further, the "nearest neighbour velocity matching" was removed because Kennedy et al. [46] realized that optimization occurred faster without it, although the choreography became more swarm-like than flock-like. *The flock is now a swarm, but it is well able to find the cornfield* [46]. Thus, the performance of each individual was measured according to its distance to the cornfield, and each bird was programmed to be attracted to both its own best previous experience and the best previous experience of any bird in the flock. After some simulations assigning different relative importance to the individual and to the social experiences, they noticed that the overestimation of the individual experience resulted in excessive wandering of isolated individuals, while the overestimation of the social experience resulted in premature convergence in local optima. Hence the original algorithm was designed so that the equilibrium between individuality and sociality is dynamically and chaotically balanced. Thus, a particle displays sometimes a more individualistic behaviour, sometimes a more social behaviour, and sometimes neither.

The simulations were generalized to $n$-dimensional collision-free search-spaces, where the equations that rule the trajectories of the individuals are as follows:

$$v_{ij}^{(t)} = v_{ij}^{(t-1)} + iw \cdot U_{(0,1)} \cdot \left(pbest_{ij}^{(t-1)} - x_{ij}^{(t-1)}\right) + sw \cdot U_{(0,1)} \cdot \left(gbest_{j}^{(t-1)} - x_{ij}^{(t-1)}\right) \tag{5. 6}$$

$$x_{ij}^{(t)} = x_{ij}^{(t-1)} + v_{ij}^{(t)} \tag{5. 7}$$

Where:

- $x_{ij}^{(t)}$ : coordinate $j$ of the position of particle $i$ at time-step $t$
- $v_{ij}^{(t)}$ : component $j$ of the velocity of particle $i$ at time-step $t$
- $iw = sw = 2$ : individuality and sociality weights, kept constant and equal to 2
- $U_{(0,1)}$ : random number generated from a uniform distribution in the range [0,1], resampled anew each time it is referenced[11]

[10] Recall that the craziness aimed to prevent the simulation from settling on a unanimous, unchanging direction.
[11] Beware that although the stochastic variable craziness was deleted, both the individual and social experiences are affected by stochastic weights in equation **(5. 6)**.

- $pbest_{ij}^{(t-1)}$ : coordinate *j* of the best position found by particle *i* up to time-step (*t*-1)
- $gbest_{j}^{(t-1)}$ : coordinate *j* of the best position found by any particle in the swarm up to time-step (*t*-1)

It is fair to note that the form of the velocity updating rule as shown in some publications (e.g. [46], [47]-page 312, [78]-page 62), sometimes unclear and sometimes mistaken, leads to some misunderstandings with regards to the generation of the random weights in equation **(5. 6)**. Although some expressions seem to suggest that, for each particle and for each time-step, only two random numbers are to be generated: one for the individual acceleration vector and the other for the social acceleration vector, the original algorithm generates the random numbers anew for each term, for each particle, for each time step, and for each component, exactly as shown in equation **(5. 6)**.

It was asserted in section **5.2.2** that although a positive feed-back was required for a system to self-organize, sometimes it could run out of control resulting in the collapse of the system if it is not counterbalanced. The original PSO (O-PSO) tends to expand rather than converge, unless the velocities are damped somehow. The simplest technique to do so consists of limiting the intensity of the velocities' components that result from equation **(5. 6)**:

$$\begin{aligned} &\text{if} \quad v_{ij}^{(t)} > v_{\max} \Rightarrow v_{ij}^{(t)} = v_{\max} \\ &\text{elseif} \quad v_{ij}^{(t)} < -v_{\max} \Rightarrow v_{ij}^{(t)} = -v_{\max} \end{aligned} \qquad \textbf{(5. 8)}$$

With regards to the name given to the paradigm, Kennedy et al. [46] argue that, although each member of the population is mass-less and volume-less, which are typical characteristics of a point, the velocity and its accelerations are more appropriately applied to a **particle**. As to the term **swarm**, it has been already asserted that the emergent behaviour resembles a **swarm** rather than a bird flock. In addition, Kennedy et al. [46] claim that the method adheres to the five principles of **swarm intelligence** (SI) articulated by Millonas (quoted in [46]):

1. *The population should be able to carry out simple space and time computations.*
2. *The population should be able to respond to quality factors in the environment.*
3. *The population should not commit its activities along excessively narrow channels.*
4. *The population should not change its mode of behaviour every time the environment changes.*
5. *The population must be able to change behaviour mode when it's worth the computational price.*

Thus, the paradigm is a SI-based method, which is well able to deal with **optimization** tasks. Hence the name **Particle Swarm Optimization** seems to be a proper denomination.

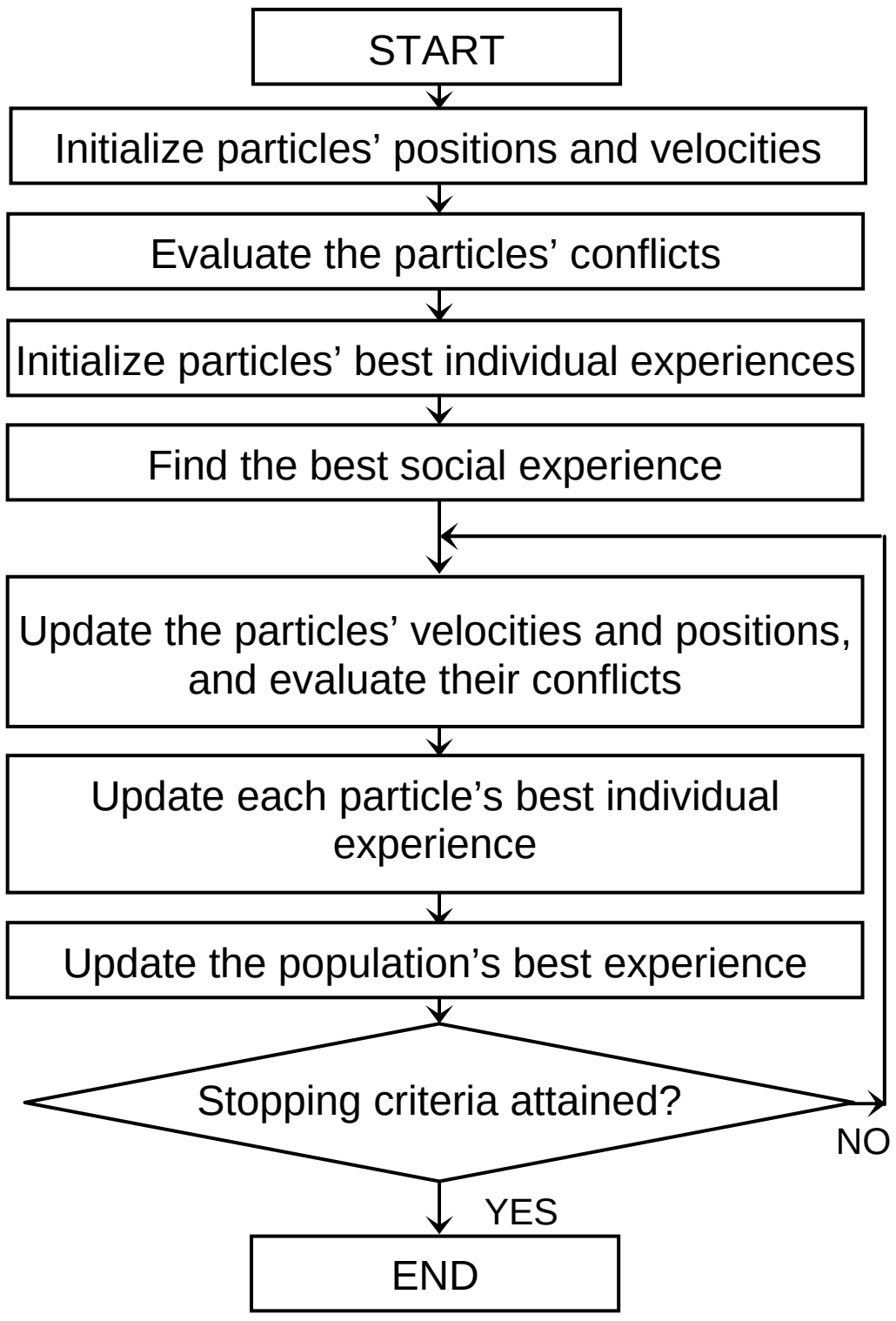

**Fig. 5. 5**: Generic PSO flow chart.

In summary, a number of particles in the form of *n*-dimensional vectors are initialized in the search-space $\mathcal{S} \subseteq \mathcal{R}^n$, typically at random. The same as the fitness function in EAs, a performance function is used here to evaluate each particle's success. Since the PSO method was originally aimed at the simulation of human social behaviour, the name chosen for the performance function in this thesis is **conflict function**, which is to be minimized.

If the search-space is thought of as a space of beliefs, the particles are beings that hold a set of beliefs, while the conflict function measures the inconsistency among them, which the individuals seek to decrease. Therefore the aim is to minimize those conflicts, by doing so the particles tend to reach agreement thus converging in the space of beliefs. The convergence occurs because each individual improves the consistency of its beliefs by imitating the individuals that display lower conflict, while being reluctant to some extent to give up on its present set of beliefs and on its previous best experience. The process is iterated until the maximum number of time steps is reached, until the performance of the particles is good enough, or until no more improvement is observed. The main broad steps are outlined in **Fig. 5. 5**.

## 5.6.2 The basic particle swarm optimizer

Shi et al. [70] theorized that, while the inertia of the particles is given by the first term, and the acceleration of the particles is given by the second and third terms in equation **(5. 6)**, the relative importance of the inertia and acceleration should be adapted to the different problems. Beware that if the inertia term was removed, the search would resemble a big stochastic hill-climber, thus displaying exploitation abilities without being able to escape local optima.

Instead, if the acceleration terms were removed, each particle would keep its initial velocity thus only exploring along a straight line. The problem of deciding the relative importance to confer to individuality and sociality was resolved by adding the stochastic weights, which dynamically modify the tendency throughout the iterations. However, the relative importance between the inertia and the acceleration is kept constant in the O-PSO. Therefore, a new parameter called "inertia weight" ($w$) was introduced into the equation **(5. 6)**, as shown in equation **(5. 9)**. *This $w$ plays the role of balancing the global search and local search. It can be a positive constant or even a positive linear or nonlinear function of time* [70].

$$v_{ij}^{(t)} = w \cdot v_{ij}^{(t-1)} + iw \cdot U_{(0,1)} \cdot \left(pbest_{ij}^{(t-1)} - x_{ij}^{(t-1)}\right) + sw \cdot U_{(0,1)} \cdot \left(gbest_{j}^{(t-1)} - x_{ij}^{(t-1)}\right) \qquad \textbf{(5. 9)}$$

$$x_{ij}^{(t)} = x_{ij}^{(t-1)} + v_{ij}^{(t)} \qquad \textbf{(5. 10)}$$

Where:

- $x_{ij}^{(t)}$ : coordinate $j$ of the position of particle $i$ at time-step $t$
- $v_{ij}^{(t)}$ : component $j$ of the velocity of particle $i$ at time-step $t$
- $U_{(0,1)}$ : random number generated from a uniform distribution in the range [0,1], resampled anew each time it is referenced
- $w$, $iw$, $sw$ : inertia, individuality, and sociality weights
- $pbest_{ij}^{(t-1)}$ : coordinate $j$ of the best position found by particle $i$ up to time-step ($t$-1)
- $gbest_{j}^{(t-1)}$ : coordinate $j$ of the best position found by any particle in the swarm up to time-step ($t$-1)

In addition to $w$, the relative importance between the weights $iw$ and $sw$ also influences the balance between the exploration and exploitation capabilities of the algorithm. The effects of the three weights on the behaviour of the system are discussed in more detail in **Chapter 6**.

There are at present an insurmountable number of versions of the PSO method, each of which shows better performance for specific problems. However, the paradigm defined by equations **(5. 9)** and **(5. 10)** is a very a flexible, general-purpose algorithm, which would perform well in most problems for some standard tuning of its parameters. Besides, the O-PSO can be seen as a particular case of this new version. Hence, from here forth, this will be referred to as the basic PSO (B-PSO) in this thesis.

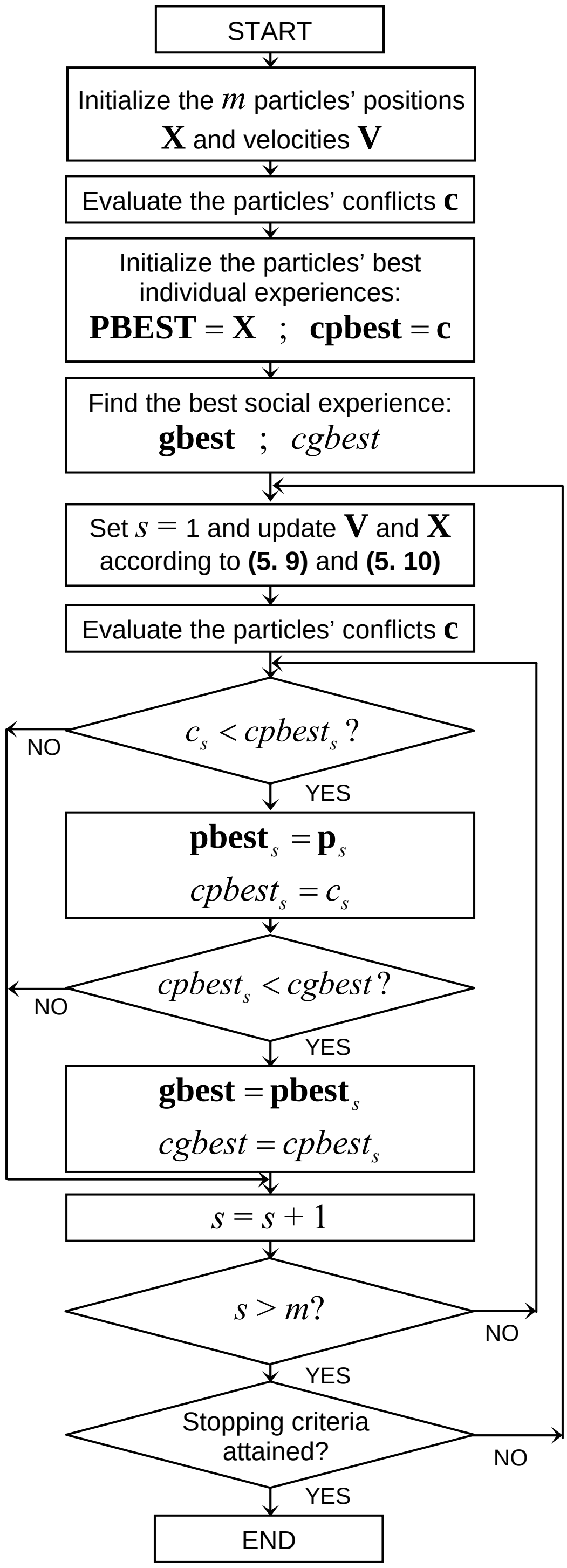

**Fig. 5. 6**: More detailed PSO flow chart.

Shi et al. [70] ran experiments with the B-PSO for the Schaffer f6 test function (see **Appendix 3**), in a 2-dimensional search-space, where each dimension was constrained to the range [-100, 100].

The parameters' setting for the experiments was:

- number of particles: $m = 20$
- $v_{\max} = iw = sw = 2$
- maximum number of time-steps: $t_{\max} = 4000$

Thus, keeping this setting fixed, different inertia weights ($w$) were evaluated. The conclusions were that it is a good idea to choose $w$ from the range [0.9, 1.2], taking into account the failures to find the global optimum and the number of iterations required to find it. It was also reported that a linearly decreasing inertia weight from 1.4 at the beginning to 0 at the 4000$^{\text{th}}$ iteration produced no failure, and the average number of iterations required to find the optimum was less than when using a fixed inertia weight in the range [0.9, 1.2].

Therefore, not only does $w$ allow balancing the relative importance between the particles' inertia and acceleration for different problems, but it also enables the system to perform higher exploration at the beginning, gradually improving its exploitation abilities throughout the iterations.

However, the limitation of the maximum velocity as shown in equation **(5. 8)** acts as a constraint for the exploration abilities of the algorithm: if $v_{\max}$ is set too low, the algorithm behaves as a local search method no matter which $w$ is chosen, while if $v_{\max}$ is set larger, the exploration ability is mainly ruled by the inertia weight.

Since both $v_{\max}$ and $w$ control the exploration abilities of the algorithm (and the explosion), Shi et al. [72] suggest the removal of $v_{\max}$, passing all the control of the global exploration ability to $w$. Since a larger $w$ leads to better exploration and a smaller $w$ leads to better exploitation, a decreasing $w$ seems to be a reasonable choice.

Thus, Shi et al. [72] ran experiments in order to analyze the performance of the B-PSO for different combinations of $w$ and $v_{\max}$. For each setting, the algorithm was run 30 times, optimizing the Schaffer f6 benchmark function. The results showed that for increasing values of $v_{\max}$, decreasing values of $w$ were required to find the global optimum without any failure, and faster. However, the decrease stopped at $w = 0.8$, to the extent that the best $w$ did not change when $v_{\max}$ was virtually removed $(v_{\max} = x_{\max})$). Hence Shi et al. [72] suggest that $w = 1$ is a good choice for a small $v_{\max}$, while $w = 0.8$ is appropriate for a large $v_{\max}$. If a convenient setting for $v_{\max}$ is not evident, they suggest setting $v_{\max} = x_{\max}$ and $w = 0.8$ as a starting point. They also tested a time-decreasing inertia weight, from 1 to 0.4 in the first 1500 time steps, keeping it constant for the remaining time steps. This last setting resulted in the best performance with regards to robustness, convergence rate and variance.

Shi et al. [71] continued to experiment with linearly decreasing inertia weights, and compared their results to those previously obtained by Angline [3], who had compared the O-PSO versus a well-developed EP algorithm. The conclusions were that the results obtained by the B-PSO with a linearly decreasing inertia weight (from 0.9 to 0.4) were noticeably better than those obtained by the O-PSO, and by the EP algorithm in [3]. Surprisingly, the algorithm did not appear to be sensitive to the population size. However, although the algorithm displays a fast convergence, the linearly time-decreasing inertia weight results in the lack of exploration abilities at the end of the run.

Other important issues that need to be defined in order to implement a PSO algorithm are the procedure to follow for the initialization of the particles and the velocities, and the population size. In the simplest algorithm, the particles are initially randomly positioned in the search-space, while their velocities can be initialized randomly within the interval $[-v_{\max}, v_{\max}]$, or they can even be set to zero. As to the population size, Kennedy et al. [47] suggest to choose from 10 to 50 particles, although this setting is obviously problem-dependent.

## 5.6.3 Global and local versions

Kennedy et al. [46] developed the O-PSO by considering that each particle could interact with any other particle in the swarm, forming a fully connected social network. The PSO versions whose social network is fully connected have come to be known as global PSO (PSO-G).

Eberhart et al. [28] proposed the first local version (PSO-L), where each particle could only interact with $k$ other socially connected particles. This so-called "$k$-best topology" takes the form of a ring if $k = 2$, whereas it takes the form of the fully connected social network if $k = m - 1$, where $m$ is the number of particles in the swarm. Note that the neighbourhood is typically defined topologically, so that neighbouring particles are not necessarily near one another in the search-space. Three typical neighbourhood structures are shown in **Fig. 5. 7**.

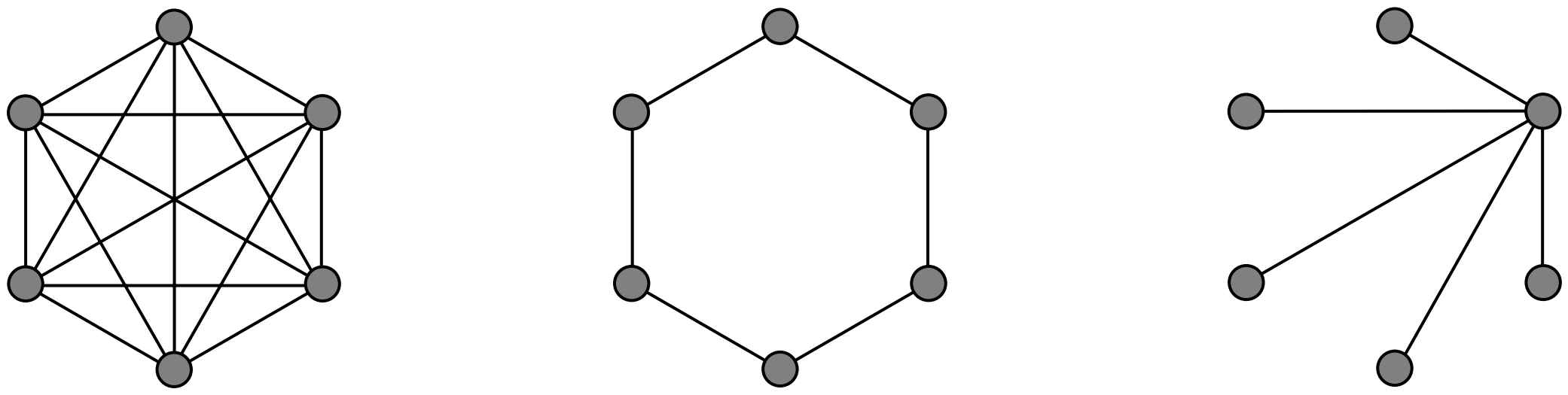

**Fig. 5. 7**: Three typical topological neighbourhoods:

Left: $k$-best topology with $k$ = swarm size - 1 (fully connected topology)
Middle: $k$-best topology with $k$ = 2 (ring topology)
Right: wheel topology

It is self-evident that the PSO-G tends to converge faster, since the PSO-L takes time to transmit the global best positions throughout the overlapping neighbourhoods. However, this allows more exploration because while the "culture transmission" is slowly taking place, each individual searches the areas that it and its neighbours believe to be promising. This enhances the capabilities of the swarm to escape local optima.

With regards to the updating equations for the PSO-L, they are the same as for the PSO-G, except for $gbest_j^{(t-1)}$, which is replaced by $lbest_j^{(t-1)}$.

Ebehart et al. [28] ran experiments with one O-PSO-G and two O-PSO-L ($k$ = 2 and $k$ = 6). They noted that the greater the neighbourhood, the more similar the O-PSO-L became to the O-PSO-G, and suggested *...that the invulnerability of this version to local optima might result from the fact that a number of "groups" of particles spontaneously separate and explore different regions.*

*...though this version rarely if ever becomes entrapped in a local optimum, it clearly requires more iterations on average to find a criterion error level.*

Kennedy et al. [47] also ran experiments to compare the ring and the wheel topologies. The results suggested that the appropriateness of a neighbourhood topology is function-dependent.

## 5.6.4 Other versions of the particle swarm optimizer

While most of the numberless versions of the PSO paradigm are problem-targeted, the aim of this thesis is to develop a robust, general-purpose algorithm. Hence only a few versions are outlined hereafter, especially those which lead to general-purpose optimizers.

### 5.6.4.1 The constricted particle swarm optimizer

Another well known alternative to control the explosion and to ensure the convergence is the addition of the constriction factor $\chi$ to the O-PSO, as proposed by Clerc et al. [16]. They developed a thorough and impressive deterministic analysis of a simplified system composed of a single particle, thus putting aside the inter-particle effects. Furthermore, the particle moved towards two stationary "best" points within a 1-dimensional search-space, removing the $v_{\max}$ constraint, and also the randomness. Then, the particle was moving towards a point that resulted from a weighted average of the two stationary best points:

$$p = \frac{iw \cdot pbest + sw \cdot gbest}{iw + sw} \qquad \textbf{(5. 11)}$$

Clerc et al. [16] showed that for this simplified system, when $iw + sw < 4$, there are some special values of $iw + sw$ for which the behaviour of the particle is cyclic, whereas for other values the behaviour is quasi-cyclic (i.e. the particle cycles unevenly around *p*).

Kennedy et al. [47] performed a similar analysis of the effect of $iw + sw$ in the trajectory of a particle without $v_{\max}$ and without randomness, showing that sine waves crosscut the trajectory of the particle with each time-step. They concluded that, for $0 < iw + sw < 4$, the particle oscillates around the stationary point *p*, with its trajectory varying characteristically with the value of $iw + sw < 4$. Conversely, the particle diverges towards infinity for $iw + sw \geq 4$.

Bear in mind that $U_{(0,1)}$ was simply removed from equation **(5. 6)**!

Clerc et al. [16] developed a generalized model which, by the addition of five coefficients to the O-PSO, allows the manipulation of the particle's trajectory, as well as the development of methods for controlling the explosion that results from randomness in the system. Notice that although the theoretical work was carried out for a single particle and putting randomness aside, once the inter-particle effects and the randomness were re-introduced in the system, the new paradigm was successful in optimizing a set of benchmark functions.

The simplest version of the generalized PSO proposed by Clerc et al. [16] is the so-called Type 1" method, which adds only a single coefficient to the O-PSO:

$$\chi = \begin{cases} \dfrac{2\cdot\kappa}{\left|(iw+sw)-2+\sqrt{(iw+sw)^2-4\cdot(iw+sw)}\right|} & \text{if } (iw+sw)>4 \\ \sqrt{\kappa} & \text{otherwise} \end{cases} \tag{5. 12}$$

$$v_{ij}^{(t)} = \chi\cdot\left(v_{ij}^{(t-1)} + iw\cdot U_{(0,1)}\cdot\left(pbest_{ij}^{(t-1)} - x_{ij}^{(t-1)}\right) + sw\cdot U_{(0,1)}\cdot\left(gbest_{j}^{(t-1)} - x_{ij}^{(t-1)}\right)\right) \tag{5. 13}$$

$$x_{ij}^{(t)} = x_{ij}^{(t-1)} + v_{ij}^{(t)} \tag{5. 14}$$

Where $\chi$ is the so-called constriction factor, and $0<\kappa\leq 1$.

Thus, the behaviour of the system is controlled by the parameter $\kappa$: if $\kappa\to 1$, the system displays more exploration abilities, while if $\kappa\to 0$ the system displays a faster convergence. Kenndey et al. [47] suggest that the setting $\kappa=1$ and $(iw+sw)=4.1$ works fine. This version will be referred to as the constricted PSO (C-PSO) from here forth.

Eberhart et al. [29] ran a series of experiments on test functions, implementing a B-PSO with linearly decreasing $w$ (from 0.9 to 0.4), and a C-PSO with $\kappa=1$ and $(iw+sw)=4.1$. The conclusions were that the C-PSO converges noticeably faster, but with greater variance. In addition, the incorporation of the constraint $v_{\max}=x_{\max}$ makes the C-PSO more robust. Of course, it is easy to set the B-PSO so as to be equivalent to the C-PSO, since the latter can be viewed as a special case of the former. All in all, Eberhart et al. [29] concluded that the best strategy is to use the C-PSO with $v_{\max}=x_{\max}$.

For a complete analysis of a non-random particle's trajectory, and of constriction factors to control the explosion, refer to [16]. Other related analyses can be found in [61, 74].

### 5.6.4.2 The general particle swarm optimizer

The general particle swarm optimizer (G-PSO) is a generalization of the B-PSO, where the inertia, individuality and sociality weights may have different values for different dimensions. Notice that the G-PSO encompasses the O-PSO, the B-PSO and the C-PSO (Type 1").

$$v_{ij}^{(t)} = w_j^{(t)} \cdot v_{ij}^{(t-1)} + iw_j^{(t)} \cdot U_{(0,1)} \cdot \left(pbest_{ij}^{(t-1)} - x_{ij}^{(t-1)}\right) + sw_j^{(t)} \cdot U_{(0,1)} \cdot \left(gbest_j^{(t-1)} - x_{ij}^{(t-1)}\right) \quad \textbf{(5. 15)}$$

$$x_{ij}^{(t)} = a_j^{(t)} \cdot x_{ij}^{(t-1)} + b_j^{(t)} \cdot v_{ij}^{(t)} \quad \textbf{(5. 16)}$$

It can be proven [74] that setting the coefficients $a_j^{(t)} = b_j^{(t)} = 1 \quad \forall j,t$ does not lose generality. Thus, the updating rules of the G-PSO can be expressed as follows:

$$v_{ij}^{(t)} = w_j^{(t)} \cdot v_{ij}^{(t-1)} + iw_j^{(t)} \cdot U_{(0,1)} \cdot \left(pbest_{ij}^{(t-1)} - x_{ij}^{(t-1)}\right) + sw_j^{(t)} \cdot U_{(0,1)} \cdot \left(gbest_j^{(t-1)} - x_{ij}^{(t-1)}\right) \quad \textbf{(5. 17)}$$

$$x_{ij}^{(t)} = x_{ij}^{(t-1)} + v_{ij}^{(t)} \quad \textbf{(5. 18)}$$

### 5.6.4.3 A particle swarm optimizer with selection

Angeline [4] proposes a PSO with the addition of a selection mechanism, which consists of performing the tournament selection typically used in EP (see section **4.3.1.3**) before applying the basic updating rule in the O-PSO (of course, this idea could be adapted to the B-PSO).

Thus, the best half of the individuals is selected and cloned by this procedure, eliminating the other half. However, the individuals of the newly cloned semi-population keep associated the previous best experiences and velocities of the dead particles. Then, the particles' velocities and positions are updated.

It should be noted that the performance of this so-called hybrid PSO[12] was compared to that of the O-PSO [4] by optimizing a set of benchmark functions, without impressive results.

Although the idea seems promising, it is fair to remark that there is not much work reported in the literature about it. Nevertheless, some works oriented to keeping an adaptive population size (e.g. [15]) are being carried out at present, and the incorporation of selection procedures into the PSOs might be a good idea in this respect.

[12] Beware that many completely different algorithms in the literature are referred to as hybrid PSO.

### 5.6.4.4 The binary particle swarm optimizer

The basic concepts of binary optimization were briefly discussed in section **2.2.1.3**, and some further discussions were undertaken in section **4.3.4**, when dealing with the binary GA. In the same fashion as any other binary algorithm, the binary PSO (b-PSO) is especially suitable for handling either binary or combinatorial problems. Although continuous problems could also be dealt with by making a few adaptations, the B-PSO performs better for such problems.

Thus, the search-space is now a binary *n*-dimensional hyper-cube: $\mathcal{S} = \{0,1\}^n$ ; the individuals are represented by binary, fixed-length bit-strings; and the conflict function $c : \{0,1\}^n \to \mathcal{R}$ .

The b-PSO was originally proposed by Kennedy at al. [45] as a variation of the O-PSO. It is fair to note that the metaphor of bird flocks does not help here to understand the optimization process. Instead, the metaphor of cognitive processes still appears applicable.

$$v_{ij}^{(t)} = w \cdot v_{ij}^{(t-1)} + iw \cdot U_{(0,1)} \cdot \left(pbest_{ij}^{(t-1)} - x_{ij}^{(t-1)}\right) + sw \cdot U_{(0,1)} \cdot \left(gbest_j^{(t-1)} - x_{ij}^{(t-1)}\right) \quad \textbf{(5. 19)}$$

$$p_{ij}^{(t)} = \frac{1}{1+e^{-v_{ij}^{(t)}}} \in [0,1] \subset \mathcal{R} \quad \textbf{(5. 20)}$$

$$x_{ij}^{(t)} = \begin{cases} 1 & \text{if} \quad U_{(0,1)} < p_{ij}^{(t)} \\ 0 & \text{otherwise} \end{cases} \quad \textbf{(5. 21)}$$

Where $p_{ij}^{(t)}$ stands for the probability of a bit adopting the state 1.

In other words, $p_{ij}^{(t)}$ is the probability that the individual *i* has of holding the feature[13] *j* at time-step *t*. All the others parameters in equation **(5. 19)** remain the same as in the B-PSO.

Notice that while the particles move in an *n*-dimensional binary hyper-cube, the velocities can still take any real value. It does not make any sense to think of $v_{ij}^{(t)}$ as a velocity any longer, but rather as a measure of the likelihood for individual *i* to hold the belief *j*. Beware that if $v_{ij}^{(t)} < 0 \;\Rightarrow\; p_{ij}^{(t)} < 0.5$, whereas if $v_{ij}^{(t)} > 0 \;\Rightarrow\; p_{ij}^{(t)} > 0.5$. Observe that the function $p_{ij}^{(t)}$ is the same as the transfer function discussed in section **3.5.5.1.3**, and plotted in **Fig. 3.13**.

[13] Feature is a general denomination that can stand for behaviour, belief, attitude, etc., according to the problem. These denominations might be used indistinctly every now and again.

In brief, an individual seeks consistency among its beliefs, which is attained by minimizing the conflicts among them. Thus, the **conflict function** receives the beliefs as inputs and returns a scalar that stands for the level of conflict among those beliefs that are held together.

Although an individual that holds a certain belief does not change its mind immediately when it notices that someone else's beliefs are more consistent, it is influenced by that observation. This is consistent with equation **(5. 19)**, whose first term stands for the tendency to keep the belief the individual has, while the second and third terms tend to move the probability threshold upwards or downwards if the belief is or is not, respectively, held by the best previous individual and social experiences. However, if the present probability is either too high or too low, it might take a long time to change the activation status of the feature.

The tuning of the parameters of the b-PSO is beyond the scope of this dissertation, which is concerned with continuous optimization problems. Kennedy et al. [47] suggest $w = 1$ and $(iw + sw) = 4$. In addition, they suggest the limitation $v_{\max} = \pm 4$, so that there is always at least a chance of $p^{(t)} = 0.018$ that a bit will change state. Notice that $v_{\max}$ serves the function of controlling the positive feed-back in the B-PSO (i.e. it prevents the particles from diverging), whereas in the b-PSO, it prevents the population from a complete loss of diversity in a similar fashion as the mutation does in the binary GA.

Kennedy et al. [47] reported a general better performance of the b-PSO in comparison to a standard GA, a mutation-only GA and a crossover-only GA, when dealing with multimodal problems[14]. Kennedy [44] implemented a b-PSO to optimize a directed S-digraph inference problem, where the aim is to find the optimal state of the binary nodes in a recurrent inference network. The network in [44] was composed of 9 nodes and a complex pattern of connections. The b-PSO was able to optimize such a problem with a population of only 20 particles (Kennedy [44] called them EleMentals), where some member of the population could always find the optimum within the first 20 time steps. Like the ACM discussed in section **5.4.5.2**, the final "states of minds" of the EleMentals showed the formation of cultures.

While a typical application of real-valued PSOs is to optimize the weights of ANNs, a very interesting application of a new paradigm created by coupling the B-PSO with the b-PSO could be to optimize the weights and the structure of ANNs simultaneously.

[14] Multimodal problems are those which present more than one global optimum.

## 5.7 Closure

The fundamental concepts underlying SI were discussed, the relativeness of individuality was pointed out, and the importance of sociality was supported by some natural examples of the impressive achievements of social, simple-minded creatures. The links between SI, AL and social psychology were observed, and the attention was drawn to some work in social psychology that greatly influenced the development of the PSO paradigm. Thus, well-known theories and experiments in social behaviour were briefly presented, guiding the dissertation towards the ACM, which, together with some bird flock simulations, helps understanding the PSO method. In addition, the ACO paradigm was outlined for completeness, despite being more related to AL simulations than to social psychology or the PSO paradigm. It is fair to note that the ACO paradigm is one of the most popular and successful SI-based methods, whose characteristics make it complementary to the PSO method regarding the kinds of problems they can deal with. Finally, the PSO method was presented, and some of its most general variations were discussed.

Both the EAs and the PSO paradigm were discussed in detail in **Chapter 4** and **Chapter 5**, respectively. Although they are both population-based methods that can deal with optimization problems, they were inspired by different forms of intelligence. *Atmar, following Weiner, proposed that there are "three distinct organizational forms of intelligence: ontogenetic, phylogenetic and sociogenetic" …* [57]. Individuals of most species display **ontogenetic** learning (self-arising within the individual), whose minimum unit of mutability is the propensity of a neuron to fire, while the reservoir of learned behaviour is the entire set of engrams[15] that reflects the knowledge of the individual. **Sociogenetic** learning (arising within the group) is the basis for a society to acquire knowledge and communicate, while culture is the reservoir of learned social behaviour. Finally, **phylogenetic** learning (arising from within the lineage) is the most ancient, and the most commonly exhibit, form of intelligence, whose unit of mutability is the nucleotide base pair, while the reservoir of learned behaviour is the genome.

While different problem-solving methods have been developed throughout decades by mimicking ontogenetic (e.g. ANNs) and phylogenetic (e.g. EAs) learning, it took much longer

[15] *An* ***engram*** *is a term for the (hypothesized) means by which memory traces are biologically stored as physical or biochemical change in the brain (and other neural tissue) in response to external stimuli…* [80]

to acknowledge the potentialities of mimicking sociogenetic learning. Much work has been done about this in the last 15 years, resulting, for instance, in the development of the ACO, ACM, and PSO paradigms. Some attempts have also been made to mimic some of these phenomena coupled in the same algorithm. For instance, MAs intend to mimic individual learning coupled with evolution, and the PSO paradigm intends to mimic (very simple) individual learning coupled with social learning. However, in nature, biological evolution occurs simultaneously with individual learning, and with social learning, which includes imitating successful close neighbours, and conforming to close neighbours, while being influenced by farther neighbours by means of culture transmissions. The existing paradigms are far from exploiting the potentialities of all these complementary phenomena.

Regarding the PSO method, **SECTION II** is completely devoted to analyzing the paradigm, while **SECTION III** is devoted to its applications. Within **SECTION II**, **Chapter 6** deals with a preliminary analysis of the parameters of the B-PSO; **Chapter 7** deals with the development of appropriate measures of error as stopping criteria; **Chapter 8** deals with a more advanced analysis of the parameters of the paradigm, once the error measures as stopping criteria are fully implemented; **Chapter 9** deals with further aspects of the paradigm such as the population size and the initialization procedure. Finally, **Chapter 10** is concerned with the adaptations to make to the PSO in order to deal with constrained optimization problems.

It is fair to remark that the aim of this thesis is to develop a robust, general-purpose algorithm that performs well in most problems, displaying a reasonably good convergence rate and an excellent reluctance to getting trapped in poor local optima. Several applications are run in **SECTION III** so as to prove the success of the proposed algorithms. It is evident that problem-specific fine tuning and adaptations of the paradigm would improve its performance when dealing with such specific problems, but they would also decrease its performance when dealing with other problems. Instead, the algorithms proposed in this dissertation perform reasonably well in most problems.

**MASTER EN MÉTODOS NUMÉRICOS**
**PARA CÁLCULO Y DISEÑO EN INGENIERÍA**

Population-Based Methods:

# PARTICLE SWARM OPTIMIZATION

– Development of a General-Purpose Optimizer and Applications –

PART II

Master's Thesis

Submitted by:
*Mauro Sebastián* Innocente

Supervised by:
Dr. *Johann Sienz*
Civil & Computational Engineering Centre
University of Wales Swansea

Barcelona, July, 2006

# SECTION II

# ALGORITHM RESEARCH

# Chapter 6

# PRELIMINARY ANALYSIS OF THE PARAMETERS

A preliminary analysis of the parameters of the basic particle swarm optimizer is undertaken. Some tunings taken from the literature are tried, and some adaptations are proposed. Thus, a few different particle swarm optimizers are implemented, always trying to keep the number of parameters to be tuned to a minimum. The analysis of the performance of each optimizer is more concerned with its behaviour than with the best solution it is able to find. While some of the resulting, robust, most promising optimizers are employed to study the stopping criteria in Chapter 7, some of the others are used to develop general-purpose optimizers in Chapter 8 and Chapter 9.

## 6.1 Introduction

Given that the traditional measures of error used to estimate the goodness of the solution that iterative methods subsequently find are developed for point-to-point search methods, they are not suitable for the population-based ones such as the particle swarm optimizers dealt with here. While the development of some measures of error with the aim to design meaningful stopping criteria is performed in the next chapter, the present one is concerned with the analysis of the influence that the parameters of the particles' velocity updating rule have on the behaviour of the swarm. However, this analysis cannot be complete without a proper definition of the conditions under which the main loop of the algorithm should be terminated. In turn, the development of meaningful measures of error requires a reasonably well tuned, working algorithm. Thus, the problem gets trapped into a vicious circle.

Although a possible approach could be to take some standard tuning of the parameters from the literature, it is useful to comprehend the functioning of the system if meaningful stopping criteria are to be developed. Therefore, a preliminary analysis of the effects that different tunings of the parameters of the algorithm have on the behaviour of the system is carried out within this chapter, where every algorithm is run along a fixed number of time-steps.

Since this chapter is concerned with the B-PSO, its characteristic equations are rewritten hereafter for future reference:

$$v_{ij}^{(t)} = w \cdot v_{ij}^{(t-1)} + iw \cdot U_{(0,1)} \cdot \left(pbest_{ij}^{(t-1)} - x_{ij}^{(t-1)}\right) + sw \cdot U_{(0,1)} \cdot \left(gbest_{j}^{(t-1)} - x_{ij}^{(t-1)}\right) \tag{6. 1}$$

$$x_{ij}^{(t)} = x_{ij}^{(t-1)} + v_{ij}^{(t)} \tag{6. 2}$$

Where:

- $x_{ij}^{(t)}$ : coordinate *j* of the position of particle *i* at time-step *t*
- $v_{ij}^{(t)}$ : component *j* of the velocity of particle *i* at time-step *t*
- $U_{(0,1)}$ : random number generated from a uniform distribution in the range [0,1], resampled anew each time it is referenced
- *w*, *iw*, *sw* : inertia, individuality, and sociality weights
- $pbest_{ij}^{(t-1)}$ : coordinate *j* of the best position found by particle *i* up to time-step (*t*-1)
- $gbest_{j}^{(t-1)}$ : coordinate *j* of the best position found by any particle in the swarm up to time-step (*t*-1)

As it can be observed in equation **(6. 1)**, there are three parameters in the B-PSO that rule the behaviour of the swarm, namely the inertia, the individuality, and the sociality weights[1]. In addition, there is a fourth parameter, the $v_{\max}$ constraint, which also influences the particles' trajectories despite being external to the particles' velocity updating equation. Therefore, the analysis of the influence that the $v_{\max}$ constraint, the inertia weight, and the learning weights have on the behaviour of the swarm is carried out in sections **6.2**, **6.3**, and **6.4**, respectively.

The study of the algorithms in sections **6.2** and **6.3** is carried out using a single benchmark test function, the Schaffer f6 (see **Fig. 6. 1** and **Appendix 3**). Furthermore, the search-spaces are 1 and 2-dimensional, so that a visual analysis of the evolution of the particles' positions can be performed.

The Schaffer f6 function is a very hard function to be optimized, which presents many local optima in the form of ring-like depressions that surround the global optimum. The features of this function are as follows:

[1] Both the individuality and the sociality weights are also referred to as learning weights.

- Function to be minimized: $$f(\mathbf{x}) = \frac{\left[\sin\left(\sqrt{\sum_{i=1}^{n} x_i^{\,2}}\right)\right]^2 - 0.5}{\left(1 + 0.001 \cdot \sum_{i=1}^{n} x_i^{\,2}\right)^2} + 0.5$$
- Region of the search-space: $[-100,100]^n$
- Global optimum: $f(\hat{\mathbf{x}}) = 0$
- Location of the global optimum: $x_i = 0 \quad \forall i$

A surface plot and a colour-map of this function for a 2-dimensional search-space are shown in **Fig. 6. 1**. For further details on this function, refer to **Appendix 3**.

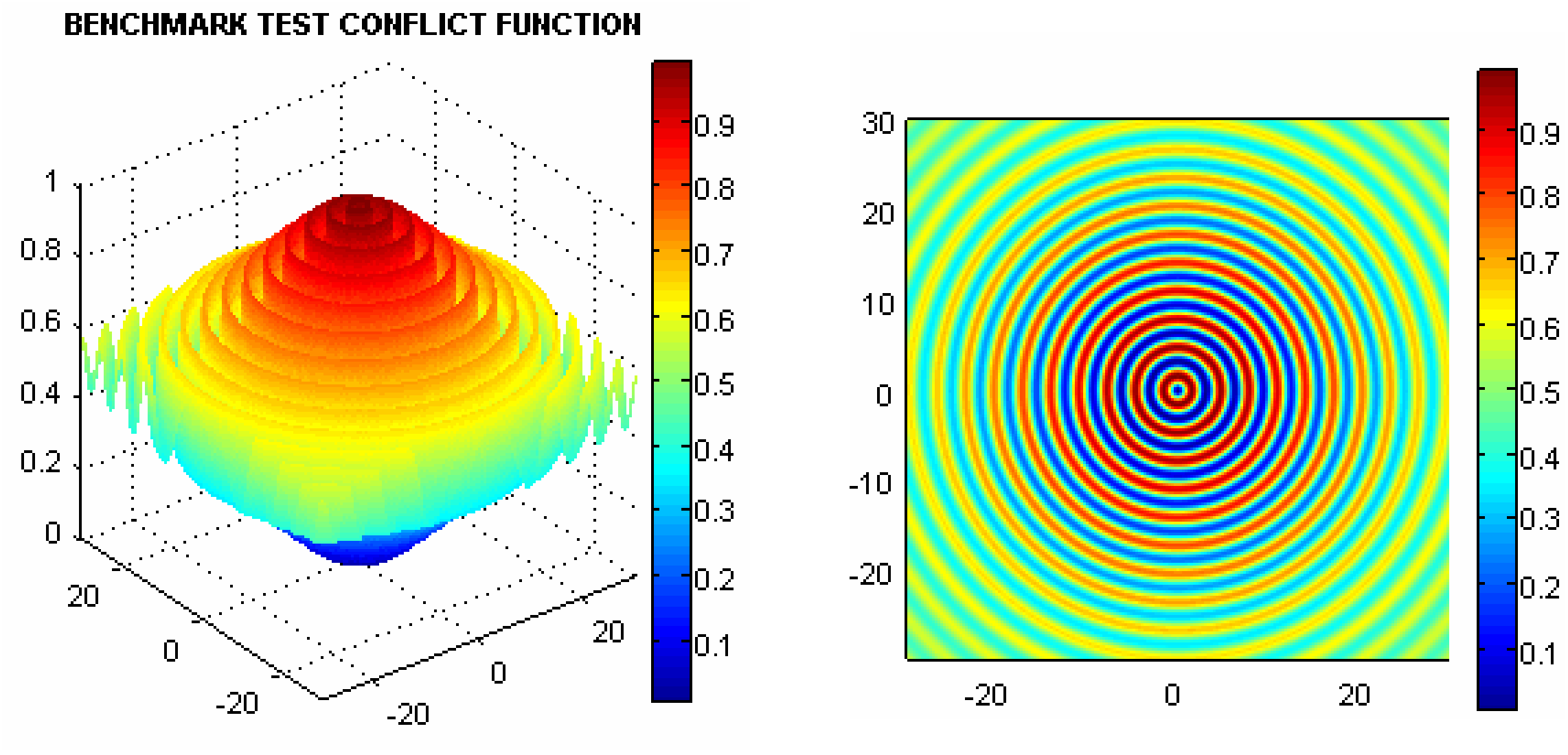

**Fig. 6. 1**: Surface plot and colour-map of the Schaffer f6 benchmark function for 2-dimensional search-spaces in the region $[-30,30]^2$.

The learning weights are initially kept constant and equal to the setting that was first proposed by Kennedy et al. [46] in their O-PSO: $iw = sw = 2$. The $v_{\max}$ constraint is incorporated to the O-PSO in section **6.2**, showing that it is a simple but effective technique to control the explosion. The inertia weight is incorporated to the O-PSO—giving birth to the B-PSO—in section **6.3**, showing that it is an efficient technique both to prevent the particles from massive divergence—or at least to pull them back after the divergence takes place—and to favour the particles’ clustering. Since each section exploits the knowledge gained from the analyses performed in previous sections, the accuracy of the results that are obtained by the algorithms generally goes in crescendo. By the end of section **6.3**, the precision that the optimizers are able to attain is such that a more thorough and quantitative analysis becomes necessary for further improvements. Thus, time-varying learning weights are proposed in section **6.4**, and the algorithms are tested on a full set of benchmark functions and higher-dimensional search-spaces. In addition, given that the probabilistic methods like the PSO obtain different results each time they are run for the same settings[2], the analysis of the optimizers in section **6.4** is performed upon the average behaviour displayed out of 50 runs for each setting.

Although the best solution found by each optimizer is undoubtedly an important piece of information for the analysis of the paradigm, this chapter is more concerned with the analysis of the behaviour of the swarm than with the best solution that it is able to find. Thus, the behaviour of the swarm is analyzed in terms of the evolution of the minimum and of the average conflicts the optimizer is able to find; of its ability to, and speed of, clustering; and of its robustness in the sense of its reluctance to getting trapped in suboptimal solutions. Notice that these last two concepts are two sides of the same coin: optimizers that exhibit pronounced clustering are more likely to get trapped in suboptimal solutions! Nevertheless, bear in mind that they are both considered here to be desirable features.

Other important data that are used in the analyses of the optimizers' performance are the number of failures in attaining an error condition, the number of times the optimizer is able to find the exact global optimum, and the standard deviations of the mean calculations.

# 6.2 Maximum velocity

The direct application of the O-PSO results in the uncontrolled, massive divergence of the particles in the swarm, which is typically referred to as the "explosion" of the system (e.g. [16]). However, if the particles are restrained somehow so as to prevent the explosion, they end up clustering around a potential solution point. The simplest way of preventing the explosion of the system is the incorporation of an upper limit for the particles' velocities:

$$\begin{array}{ll} \text{if} & v_{ij}^{(t)} > v_{\max} \;\Rightarrow\; v_{ij}^{(t)} = v_{\max} \\ \text{elseif} & v_{ij}^{(t)} < -v_{\max} \;\Rightarrow\; v_{ij}^{(t)} = -v_{\max} \end{array} \qquad \textbf{(6. 3)}$$

[2] Of course, if the generator of the random numbers is re-started each time, the results are exactly the same.

In fact, it is not each particle's velocity what is constrained, but its components. Nevertheless, while calculating the magnitude of each particle's velocity at each time-step becomes too expensive for increasing dimensions, the actual value imposed on it is not essential with regards to the prevention of the explosion.

It should be remarked that while a high $v_{\max}$ might delay the clustering and might lead to the recurrent evaluation of the conflict function far from promising regions, a low $v_{\max}$ makes the algorithm more likely to get trapped in local optima. The disappointing fact is that a convenient value for $v_{\max}$ is problem dependent. For instance, imagine a wavy landscape where the wavelengths are much greater than $v_{\max}$: the algorithm will get trapped in a local optimum, unless the space spanned between the initial positions of the particles contains the global optimum, in which case the algorithm might still be able to find it.

The idea of constraining the components of the particles' velocity was originally proposed to control the explosion of the swarm in the O-PSO, which is equivalent to setting $w = 1$ in equation **(6. 1)**. Furthermore, the first proposed algorithms were equipped with $iw = sw = 2$. Thus, the analysis of the effect that the $v_{\max}$ constraint has on the behaviour of the system is performed within this section for the following updating equations:

$$v_{ij}^{(t)} = v_{ij}^{(t-1)} + 2 \cdot U_{(0,1)} \cdot \left(pbest_{ij}^{(t-1)} - x_{ij}^{(t-1)}\right) + 2 \cdot U_{(0,1)} \cdot \left(gbest_{j}^{(t-1)} - x_{ij}^{(t-1)}\right) \qquad \textbf{(6. 4)}$$

$$x_{ij}^{(t)} = x_{ij}^{(t-1)} + v_{ij}^{(t)} \qquad \textbf{(6. 5)}$$

## 6.2.1 Analysis of a single particle

### 6.2.1.1 Explosion

First, consider the case of a single particle flying over a 1-dimensional search-space. The initial position of the particle is set to $x = 100$; no limitation is imposed on the components of its velocities, which are randomly initialized within the interval [–1,1]; the algorithm is run over 100 time-steps; and both best values are set to $pbest = gbest = 0$ for the whole run. Recall that the function to be optimized by the algorithms analyzed within sections **6.2** and **6.3** is the Schaffer f6 benchmark function (see **Fig. 6. 1** and **Appendix 3**).

The evolution of the particle's position through 100 time-steps is shown in **Fig. 6. 2**:

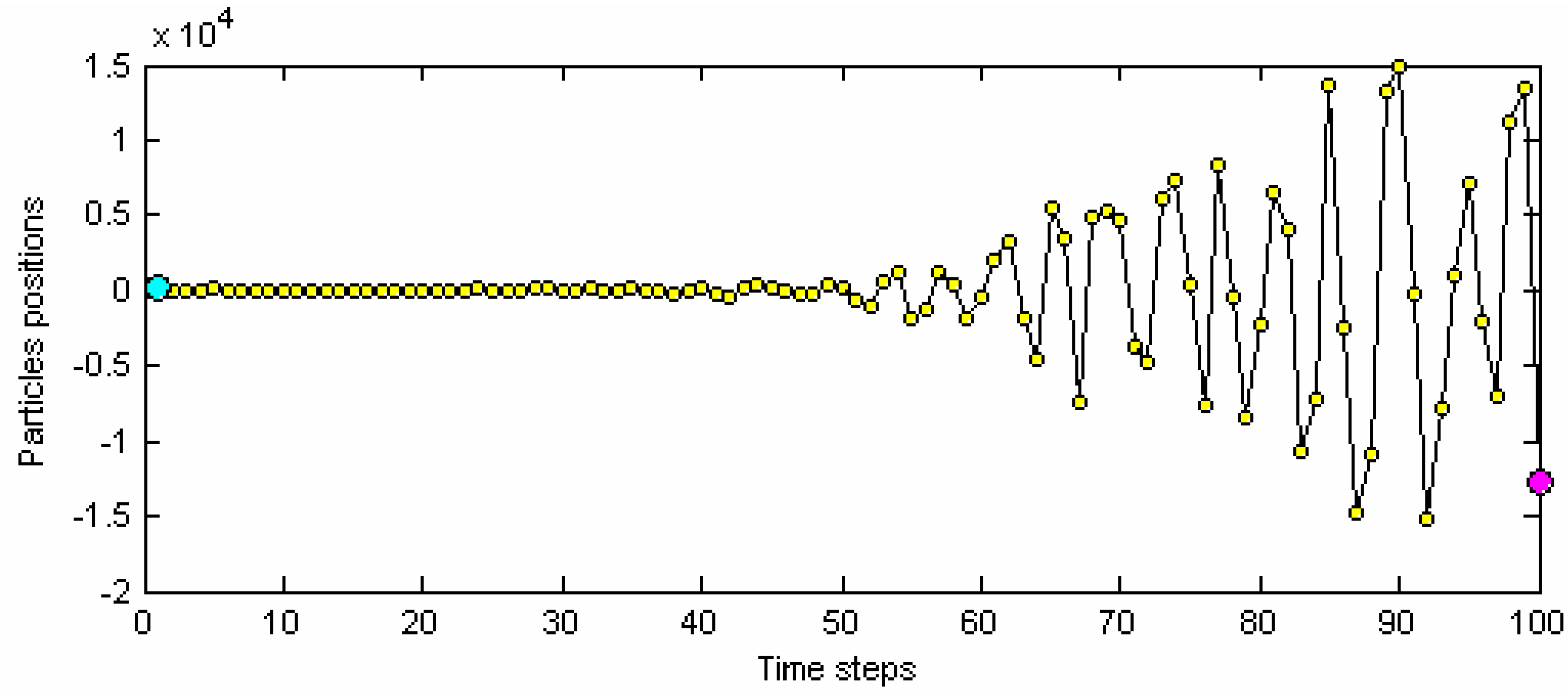

**Fig. 6. 2**: Evolution of a single particle flying over a 1-dimensional search-space, where the two best values are fixed to zero, the particle is initially located at $x = 100$, its velocity is randomly initialized within the interval [–1,1], no $v_{\max}$ is imposed, $iw = sw = 2$, and the function to be optimized is the Schaffer f6. The cyan and magenta dots are the particle's initial and final positions, respectively.

Notice that the amplitudes of the sinus-like wavelengths are time-increasing. Clerc et al. [16] simplified the system in order to study the dynamics of the swarm from the bottom up (i.e. from a particle's point of view). First, starting from the general equations **(6. 6)** and **(6. 7)**,

$$v_{ij}^{(t)} = v_{ij}^{(t-1)} + iw \cdot U_{(0,1)} \cdot \left(pbest_{ij}^{(t-1)} - x_{ij}^{(t-1)}\right) + sw \cdot U_{(0,1)} \cdot \left(gbest_{j}^{(t-1)} - x_{ij}^{(t-1)}\right) \quad \textbf{(6. 6)}$$

$$x_{ij}^{(t)} = x_{ij}^{(t-1)} + v_{ij}^{(t)} \quad \textbf{(6. 7)}$$

they reduced the swarm size to a single particle; considered a 1-dimensional search-space; removed the random weights; and considered that the two best points the particle is attracted to are stationary. Thus, equations **(6. 6)** and **(6. 7)** are turned into equations **(6. 8)** and **(6. 9)**:

$$v^{(t)} = v^{(t-1)} + iw \cdot \left(pbest - x^{(t-1)}\right) + sw \cdot \left(gbest - x^{(t-1)}\right) \quad \textbf{(6. 8)}$$

$$x^{(t)} = x^{(t-1)} + v^{(t)} \quad \textbf{(6. 9)}$$

Since the two points the particle is attracted to are stationary, the particle is in reality attracted to a point that results from the weighted average of *pbest* and *gbest*:

$$iw \cdot pbest + sw \cdot gbest = (iw + sw) \cdot p \quad \Rightarrow \quad p = \frac{iw \cdot pbest + sw \cdot gbest}{iw + sw} \quad \textbf{(6. 10)}$$

Hence the equations that rule the dynamics of this single particle are:

$$v^{(t)} = v^{(t-1)} + (iw + sw)\cdot\left(p - x^{(t-1)}\right) \quad \textbf{(6. 11)}$$

$$x^{(t)} = x^{(t-1)} + v^{(t)} \quad \textbf{(6. 12)}$$

Clerc et al. [16] carried out a thorough analysis of the trajectory of the particle ruled by equations **(6. 11)** and **(6. 12)**, proving that if $iw + sw < 4$, the particle presents a cyclic or quasi-cyclic behaviour. Even further, they found the particular values of $iw + sw$ for which the behaviour is cyclic. Conversely, there is no cyclic behaviour, and the particle diverges from *p*, if $iw + sw \geq 4$. The evolution of the position of a non-random particle ruled by equations **(6. 11)** and **(6. 12)** with $iw + sw = 4$ and $p = 0$ is shown in **Fig. 6. 3**:

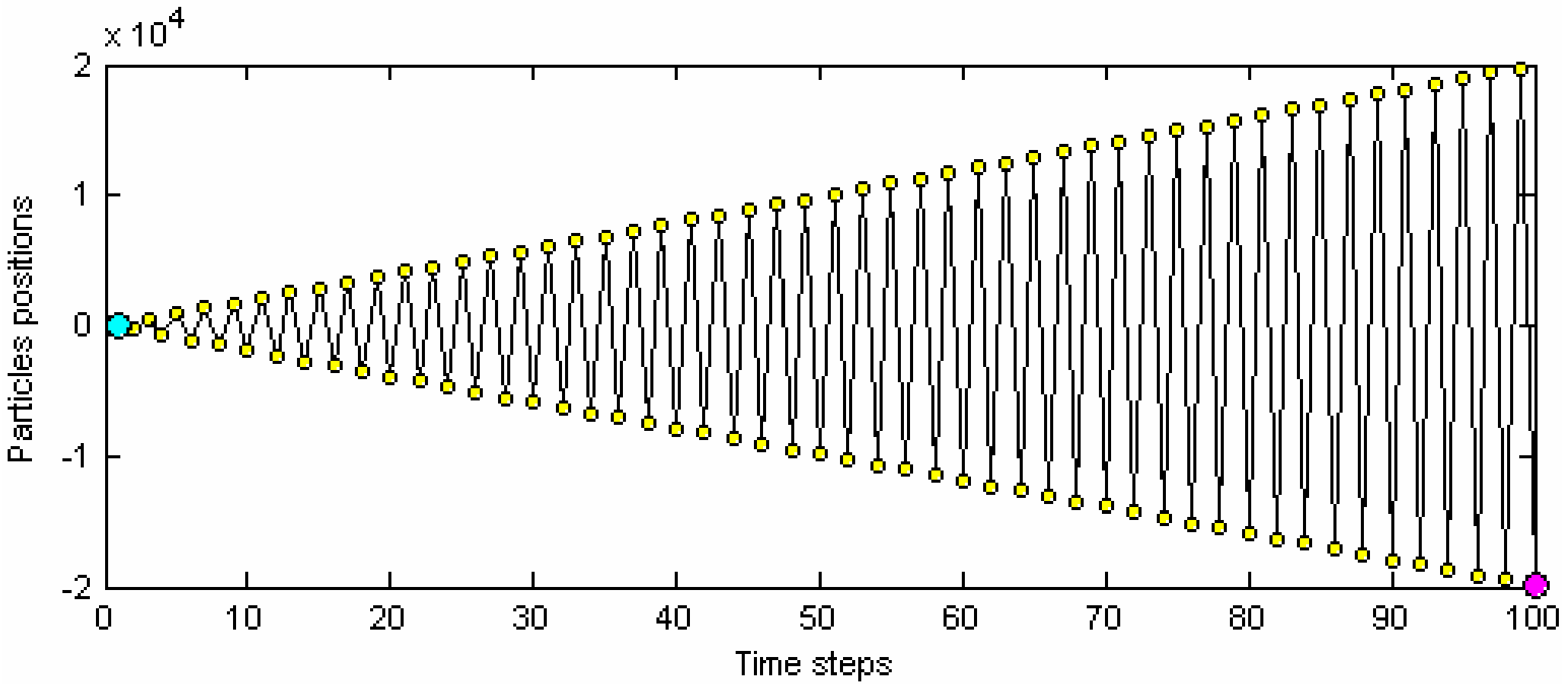

**Fig. 6. 3**: Evolution of a single particle flying over a 1-dimensional search-space, where the two best values are fixed to zero, the particle is initially located at $x = 100$, its velocity is initialized to zero, no $v_{max}$ is imposed, the random weights are removed, $iw = sw = 2$, and the function to be optimized is the Schaffer f6. The cyan and magenta dots are the particle's initial and final positions, respectively.

Clerc et al. [16] analytically developed a constriction factor that ensures the convergence on local optima of the single non-random particle, generalizing the analytic findings to the full multi-particle system with the random weights and with the two non-stationary best values. These generalized algorithms were successfully tested on a set of benchmark functions. Some other researchers have also carried out analyses of the trajectory of a single non-random particle, namely Kennedy et al. [47], Ozcan et al. [61], and Trelea et al. [74].

While there is substantial empirical evidence backing these theoretical developments, bear in mind that, to the knowledge of the author of this thesis, the precise relationship between the multi-particle probabilistic PSO and the single-particle deterministic PSO is still not strictly established.

Notice that although both the explosion observed in **Fig. 6. 2** and the one observed in **Fig. 6. 3** take place for $iw + sw = 4$, the latter is a purely deterministic explosion. While Clerc et al. [16] deal with the mathematical reasons for this deterministic explosion, the dynamics of the explosion once the random weights $0 \le U_{(0,1)} \le 1$ are incorporated is not considered.

If the random weights $U_{(0,1)}$ are replaced by the mean of the uniform distribution used to generate them: $\overline{U}_{(0,1)} = 0.5$, the average behaviour of the O-PSO—which is ruled by equations **(6. 4)** and **(6. 5)**—is cyclic. Imagine that $U_{(0,1)}$ was replaced by $\overline{U}_{(0,1)} = 0.5$ in equation **(6. 4)**, and that the particle's velocity was initialized to 0: the particle in **Fig. 6. 2** would move from its initial position $x = 100$ to $x = -100$ in the second time-step; it would stay in the same position in the third time-step; it would move back to $x = 100$ in the fourth time-step; and so on. This cyclic behaviour is shown in **Fig. 6. 4**:

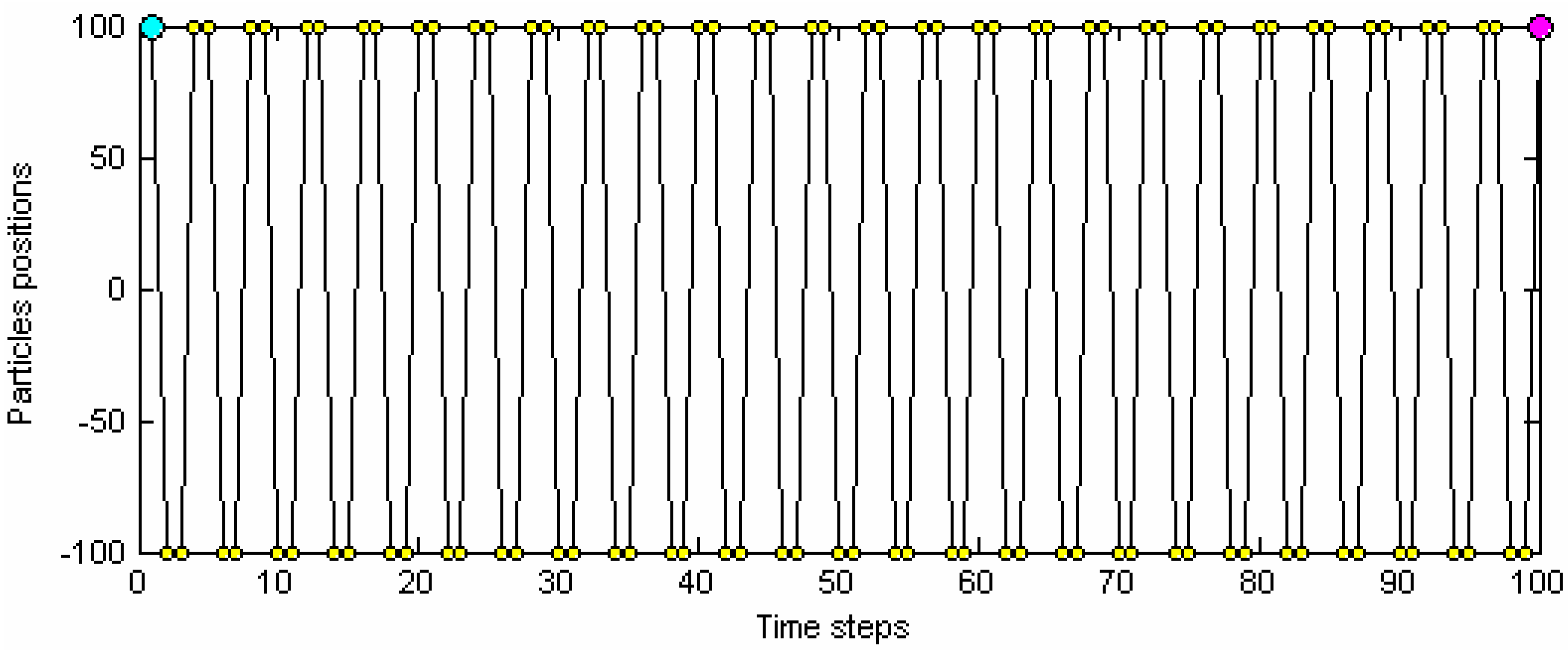

**Fig. 6. 4**: Evolution of a single particle flying over a 1-dimensional search-space, where the two best values are fixed to zero, the particle is initially located at $x = 100$, its velocity is initialized to zero, no $v_{\max}$ is imposed, $U_{(0,1)}$ is replaced by $\overline{U}_{(0,1)} = 0.5$, $iw = sw = 2$, and the function to be optimized is the Schaffer f6. The cyan and magenta dots are the particle's initial and final positions, respectively.

In fact, notice that one of the values of $iw + sw$ that Clerc et al. [16] found to lead to a cyclic behaviour of their non-random particle is $iw + sw = 2$, which is equivalent to $iw + sw = 4$ if the random weights $U_{(0,1)}$ are replaced by $\overline{U}_{(0,1)} = 0.5$ rather than removed (see **Fig. 6. 4**).

Thus, on the one hand, it can be argued that if the random weights are simply incorporated to Clerc et al.'s [16] deterministic O-PSO (see **Fig. 6. 3**), the explosion still occurs (see **Fig. 6. 2**). On the other hand, it can also be argued that if the random weights replace the average of the distribution they are generated from in a deterministic O-PSO with $\overline{U}_{(0,1)} = 0.5$ (see **Fig. 6. 4**), the cyclic behaviour turns into a divergent behaviour (see **Fig. 6. 2**). The reason for this is not self-evident, since each random number generated is as likely to be greater than as it is to be less than 0.5!

Kennedy et al. [47] offer: *It is not immediately obvious why the velocity requires damping. … It turns out that the system explodes because* $\varphi$ (where $\varphi = iw \cdot U_{(0,1)} + sw \cdot U_{(0,1)}$) *is varying due to being weighted with random numbers. This explosion can occur with a random* $\varphi$ *or with* $\varphi$ *fluctuating in almost any way. Some kind of damping is required to control it.*

In fact, if the random weights are incorporated to a deterministic PSO whose learning weights are such that $iw + sw < 4$ (or such that $iw + sw < 8$ if the random weights are replaced by $\overline{U}_{(0,1)} = 0.5$ rather than removed), the explosion might still occur.

The evolution of a non-random particle with $iw + sw = 0.5$ (or with $iw + sw = 1$ if the random weights $U_{(0,1)}$ are replaced by $\overline{U}_{(0,1)} = 0.5$ rather than removed) is shown in **Fig. 6. 5**:

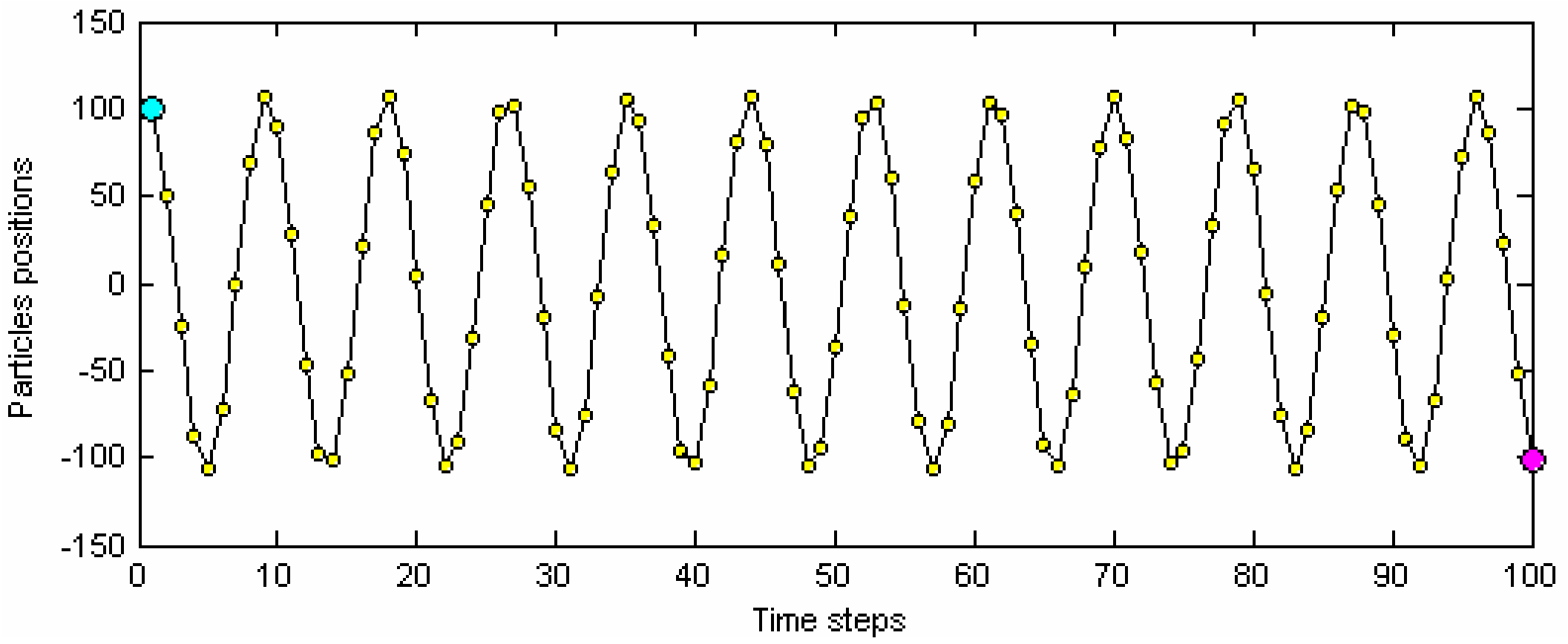

**Fig. 6. 5**: Evolution of a single particle flying over a 1-dimensional search-space, where the two best values are fixed to zero, the particle is initially located at $x = 100$, its velocity is initialized to zero, no $v_{max}$ is imposed, $U_{(0,1)}$ is replaced by $\overline{U}_{(0,1)} = 0.5$, $iw = sw = 0.5$, and the function to be optimized is the Schaffer f6. The cyan and magenta dots are the particle's initial and final positions, respectively.

The evolution of the same particle, but replacing the mean $\overline{U}_{(0,1)} = 0.5$ by the random weights, is shown in **Fig. 6. 6**. Notice that the explosion still occurs in spite of the fact that $iw + sw < 4$ !

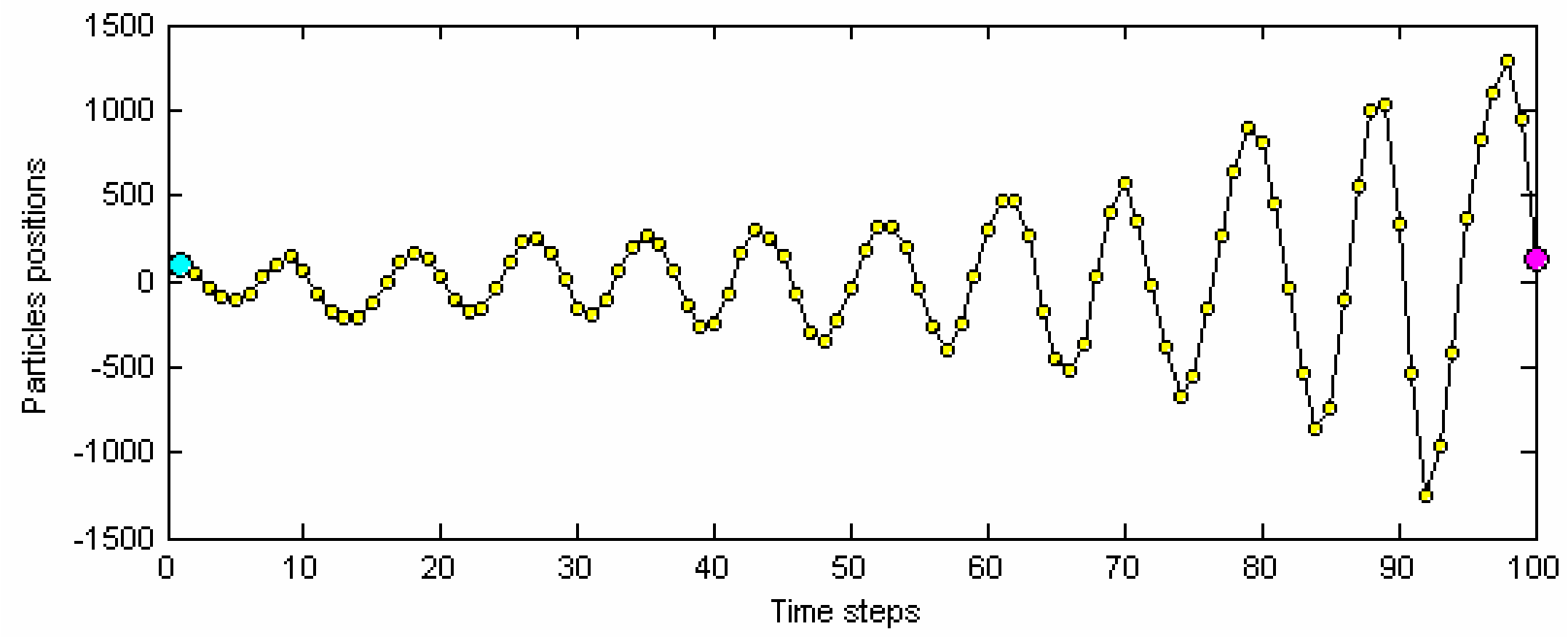

**Fig. 6. 6**: Evolution of a single particle flying over a 1-dimensional search-space, where the two best values are fixed to zero, the particle is initially located at $x = 100$, its velocity is initialized to zero, no $v_{\max}$ is imposed, $iw = sw = 0.5$, and the function to be optimized is the Schaffer f6. The cyan and magenta dots are the particle's initial and final positions, respectively.

**Fig. 6. 5** and **Fig. 6. 6** suggest that replacing a constant coefficient ($\overline{U}_{(0,1)} = 0.5$) by a randomly time-varying one ($U_{(0,1)}$) might result in the explosion of the swarm even if the mean of the time-varying coefficients is equal to the constant coefficient replaced. Thus, it can be argued that it is randomness itself which causes the explosion. Reinforcing this argument, it shows that the explosion might occur even if $iw + sw < 4$ (recall that $iw + sw \geq 4$ is Clerc et al.'s [16] condition for the deterministic explosion, as shown in **Fig. 6. 3**).

Therefore, it seems clear that the random weights are responsible for the explosion. It can be simplistically reasoned that, while the deterministic O-PSO with $\overline{U}_{(0,1)} = 0.5$ is cyclic and the random weights are as likely to be greater than as they are to be less than $\overline{U}_{(0,1)} = 0.5$ in the full probabilistic O-PSO, an explosion is more likely to occur than an implosion just because there is more "area" to explode than to implode to.

#### 6.2.1.2 Controlling the explosion: $v_{\max}$ = 100

As previously mentioned, constraining the components of the particles' velocities to a certain $v_{\max}$ turns out to be a simple, effective way of preventing the particles from diverging. Hence a new experiment is run with the same settings as in the experiment shown in **Fig. 6. 2**, but introducing the constraint $v_{\max} = 100$. The evolution of the particle is shown in **Fig. 6. 7**:

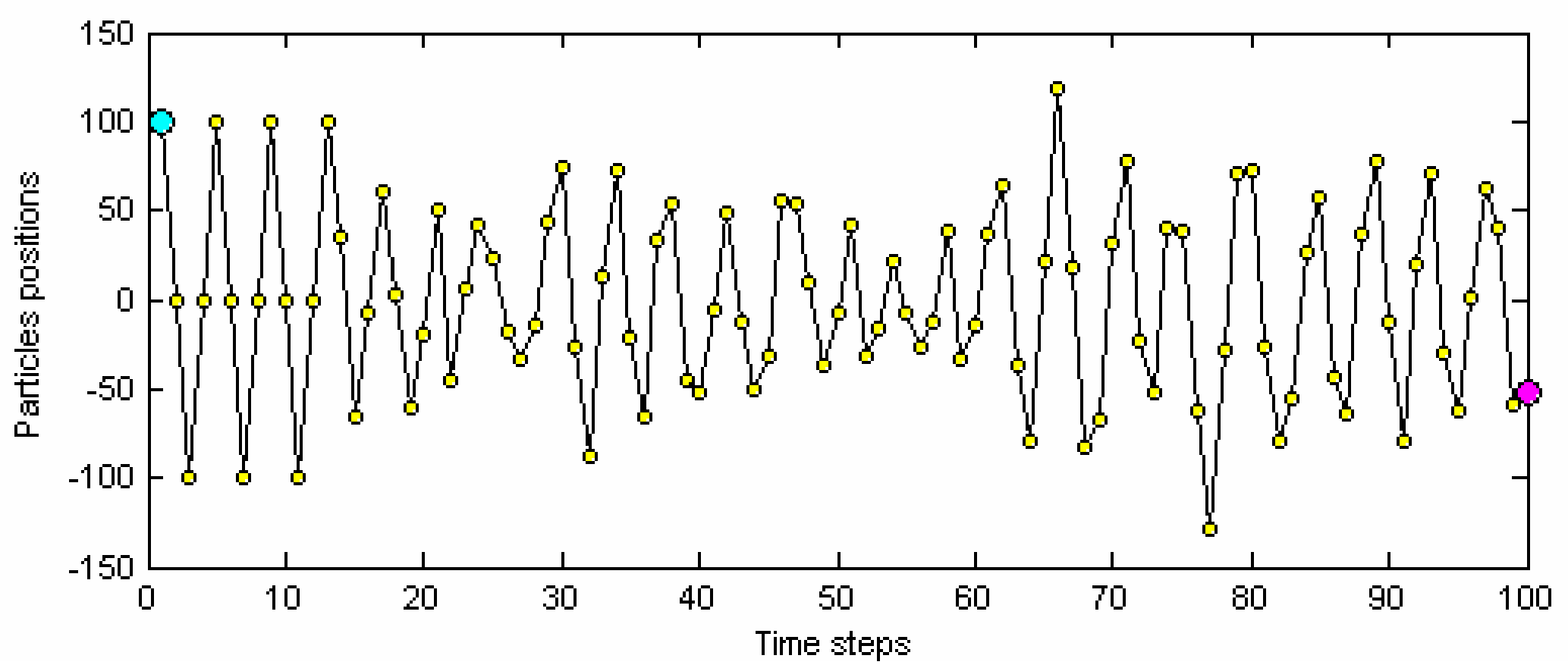

**Fig. 6. 7**: Evolution of a single particle flying over a 1-dimensional search-space, where the two best values are fixed to zero, the particle is initially located at $x = 100$, its velocity is randomly initialized within the interval [–1,1], $v_{\max} = 100$, $iw = sw = 2$, and the function to be optimized is the Schaffer f6. The cyan and magenta dots are the particle's initial and final positions, respectively.

Now, the trajectory of the particle oscillates around the optimum, displaying more controlled wave amplitudes. Notice that the velocities are $v_{\max}$-sized during the first stages of the search, as it can be observed in **Fig. 6. 7**. It appears that the $v_{\max}$ constraint serves the function of effectively controlling the explosion.

## 6.2.2 Analysis of two interacting particles

### 6.2.2.1 Explosion

While the previous experiments are useful to understand the effect that the limitation of the velocities has on the behaviour of the swarm, each particle interacts with other particles in quest for the optimum—which is not introduced by brute force—in the real system. Hence a new experiment is run keeping the same settings as before, but with two interacting particles, and removing the constraint $pbest = gbest = 0$. Therefore, the particles themselves have to find the best positions to be used in the velocity updating equation **(6. 4)**. In the same fashion as in section **6.2.1.1**, no limitation to the particles' velocities is imposed here.

The evolution of the two interacting particles is shown in **Fig. 6. 8**. Displaying a somehow less ordered pattern, the explosion of the system by exhibiting time-increasing amplitudes of the wave lengths is observed again.

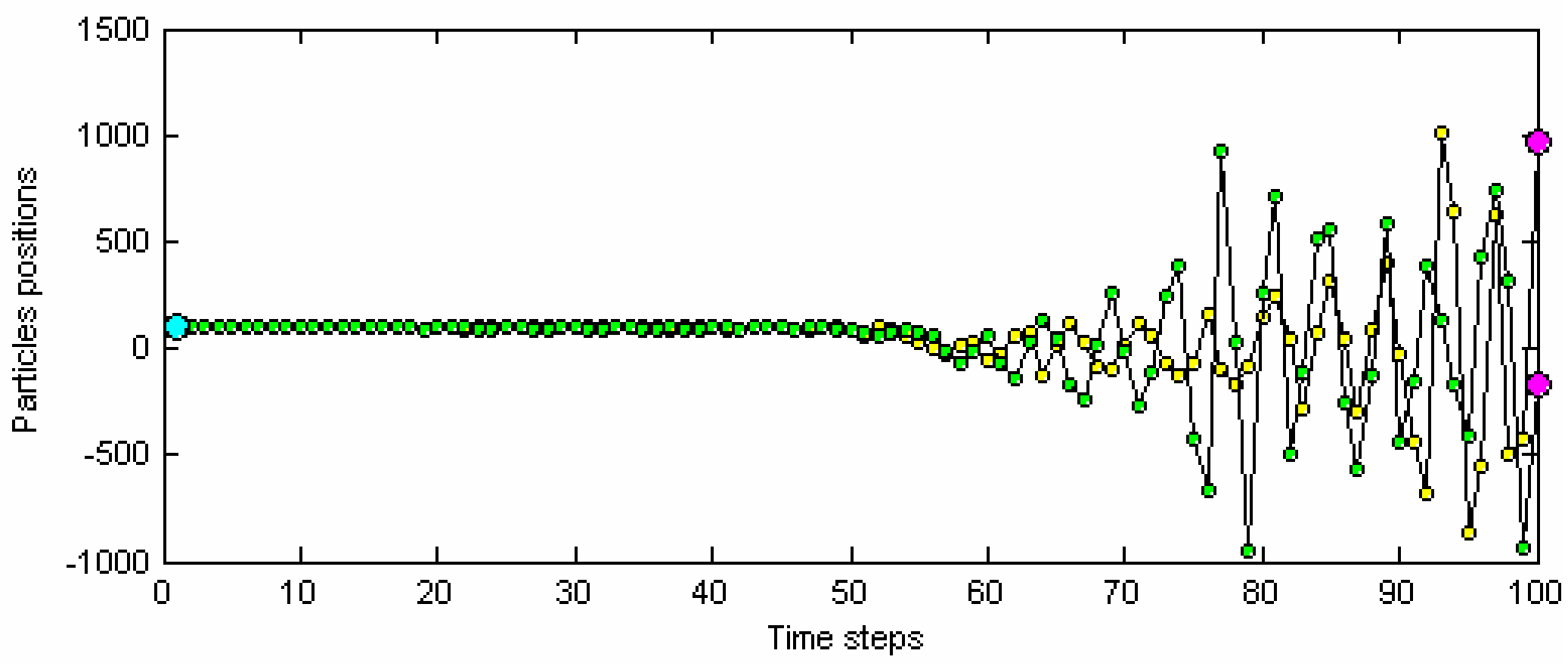

**Fig. 6. 8**: Evolution of two particles flying over a 1-dimensional search-space, where the particles are initially located at $x = 100$, their velocities are randomly initialized within the interval [–1,1], no $v_{max}$ is imposed, and the function to be optimized is the Schaffer f6. The cyan and magenta dots are the particle's initial and final positions, respectively.

### 6.2.2.2 Controlling the explosion: $v_{max}$ = 100

In order to control the explosion, the same experiment is run again but with the incorporation of the constraint $v_{max} = 100$. Beware that, as opposed to the experiment in section **6.2.1.2**, this experiment is run by using a full O-PSO, without forcing the best values to zero.

The evolution of the positions of the two interacting particles is shown in **Fig. 6. 9**. It seems that the limitation of the magnitudes that the velocities of the particles are allowed to take is an effective way of preventing the particles' explosion. However, it also introduces a new parameter to the algorithm, the constraint $v_{max}$, whose appropriate value requires tuning.

### 6.2.2.3 Controlling the explosion: $v_{max}$ = 10

In order to study the influence that the constraint $v_{max}$ has on the behaviour of the system, the previous experiment is run again keeping the same settings as before, but decreasing the value of the upper limit imposed on the particles' velocities to $v_{max} = 10$.

The evolution of the two interacting particles is shown in **Fig. 6. 10**. Notice that, since the particles now take longer to approach the region of interest, the number of time-steps is now increased to 150. It is clear that smaller values of the constraint $v_{max}$ narrow the search, restraining the positions of the particles to regions of more interest. Although this is a desirable feature, the question is now how small the constraint $v_{max}$ should be.

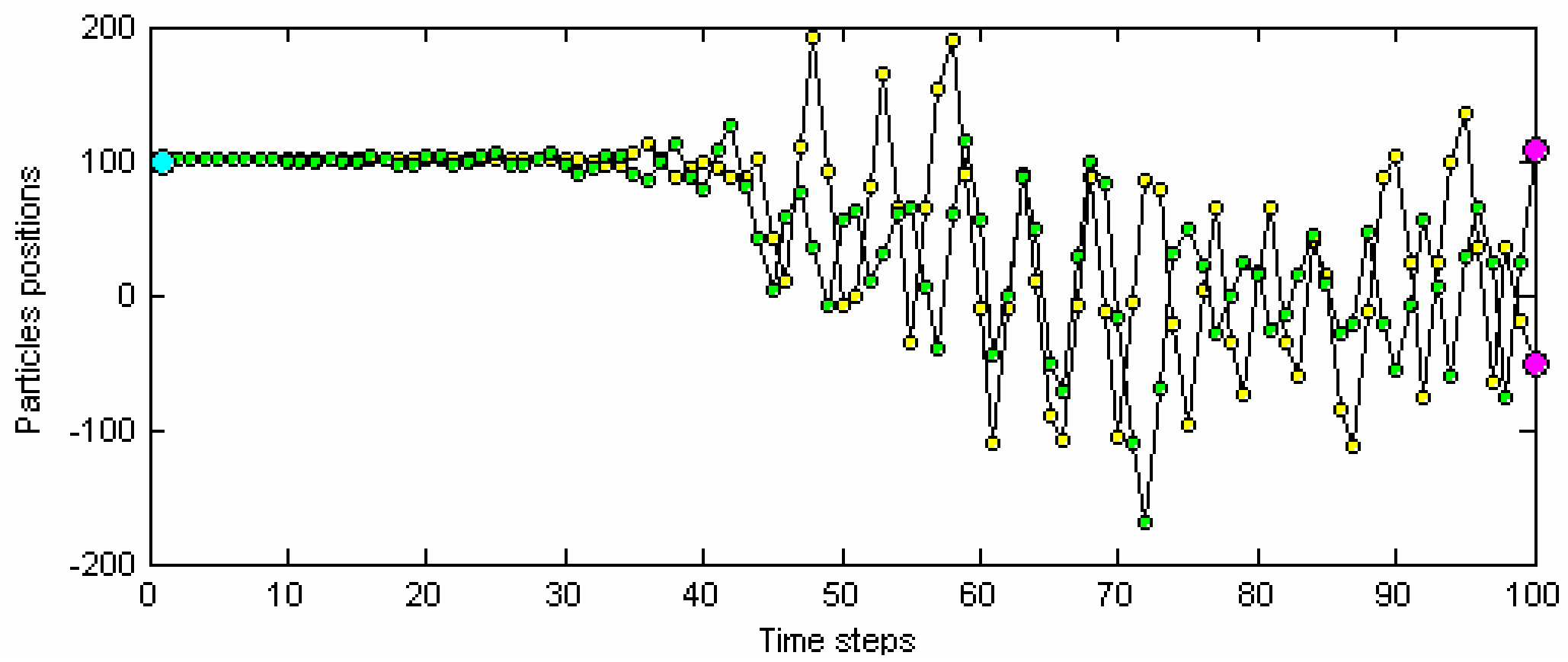

**Fig. 6. 9**: Evolution of two particles flying over a 1-dimensional search-space, where the particles are initially located at $x = 100$, their velocities are randomly initialized within the interval [–1,1] and limited to $v_{max} = 100$, and the function to be optimized is the Schaffer f6. The cyan and magenta dots are the particle's initial and final positions, respectively.

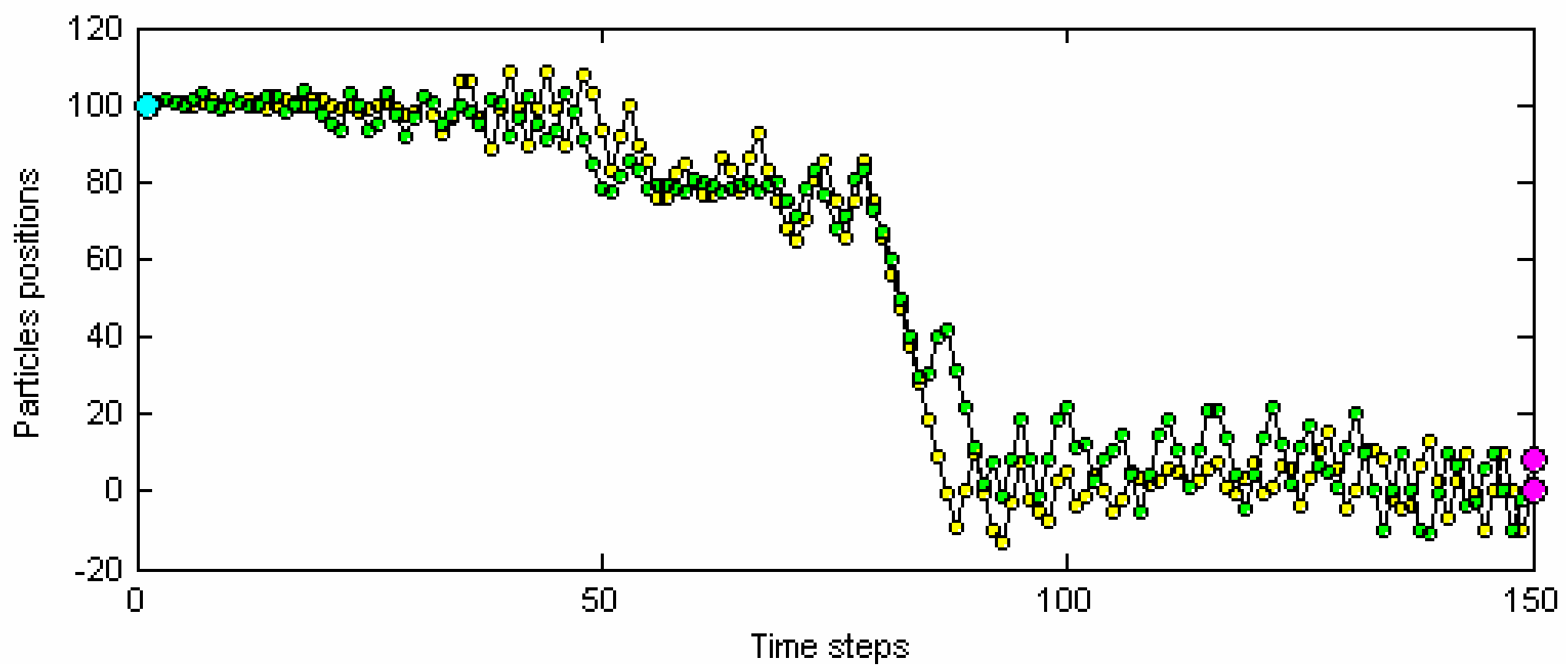

**Fig. 6. 10**: Evolution of two particles flying over a 1-dimensional search-space, where the particles are initially located at $x = 100$, their velocities are randomly initialized within the interval [–1,1] and limited to $v_{max} = 10$, and the function to be optimized is the Schaffer f6. The cyan and magenta dots are the particle's initial and final positions, respectively.

### 6.2.2.4 Controlling the explosion: $v_{max} = 2$

Therefore, the last experiment is run again keeping the same settings as before, but decreasing the value of the upper limit imposed on the particles' velocities to $v_{\max} = 2$. The evolution of the two interacting particles is shown in **Fig. 6. 11**:

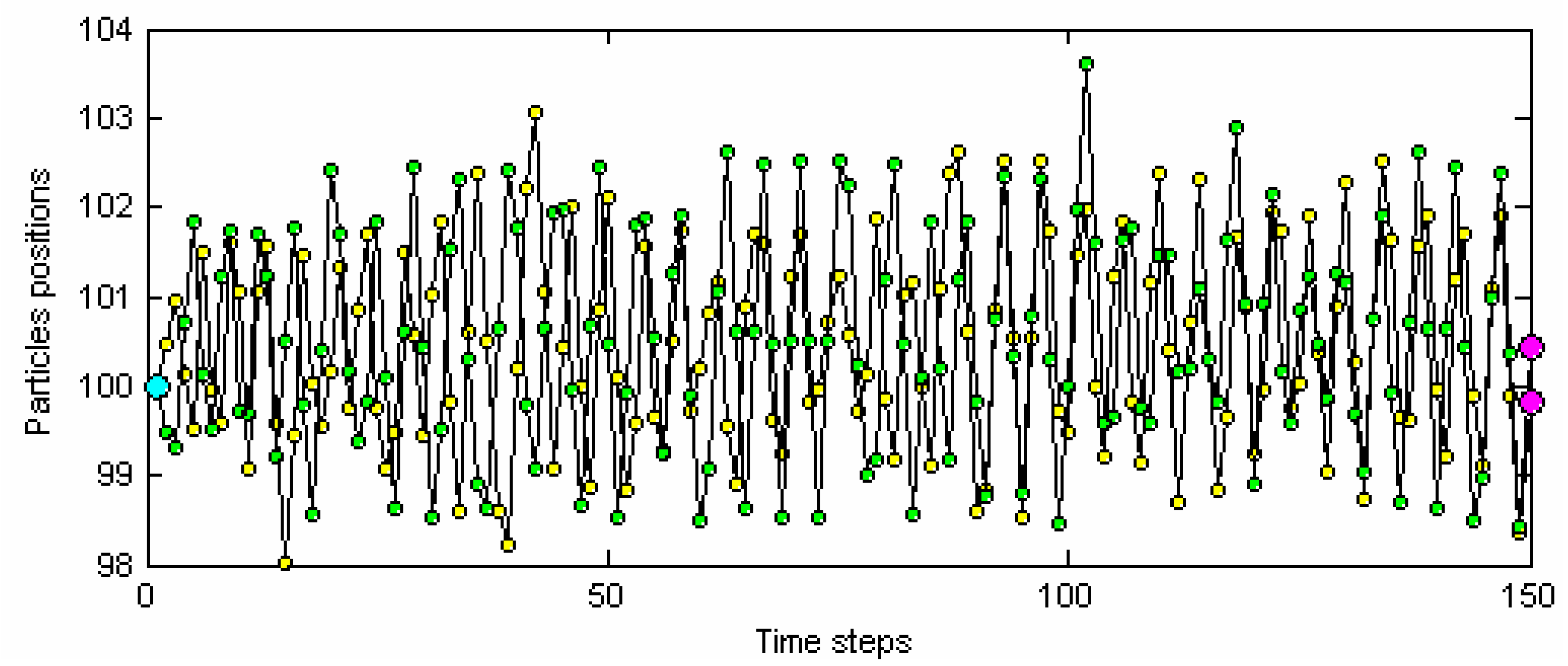

**Fig. 6. 11**: Evolution of two particles flying over a 1-dimensional search-space, where the particles are initially located at $x = 100$, their velocities are randomly initialized within the interval [–1,1] and limited to $v_{\max} = 2$, and the function to be optimized is the Schaffer f6. The cyan and magenta dots are the particle's initial and final positions, respectively.

As it can be observed, the particles oscillate around their initial position, without being able to approach the region of interest. The Schaffer f6 function displays a wavy landscape, and the particles whose velocities are too small may not be able to go through some long waves. Thus, as previously predicted, the use of too small values of $v_{\max}$ might result in the algorithm getting trapped in local optima. Its appropriate setting is undoubtedly function-dependent.

### 6.2.2.5 Controlling the explosion: linearly time-decreasing $v_{max}$

While it has been asserted several times before in this thesis that the aim here is to develop a robust, general-purpose algorithm, a problem-dependent value of $v_{\max}$ is evidently an obstacle in that regard. It is clear from observing the results of the experiments in sections **6.2.2.2** to **6.2.2.4** that a high $v_{\max}$ results in a more robust but less precise algorithm, while a low $v_{\max}$ narrows the search but makes the algorithm more likely to get trapped in local optima. The problem is that there is no general-purpose lower threshold for $v_{\max}$.

Hence, aiming to find a more robust strategy, a linearly time-decreasing $v_{\max}$ constraint is implemented, where $v_{\max} = 100$ at the beginning, and $v_{\max} = 1$ at the end of the run. The evolution of the two interacting particles is shown in **Fig. 6. 12**:

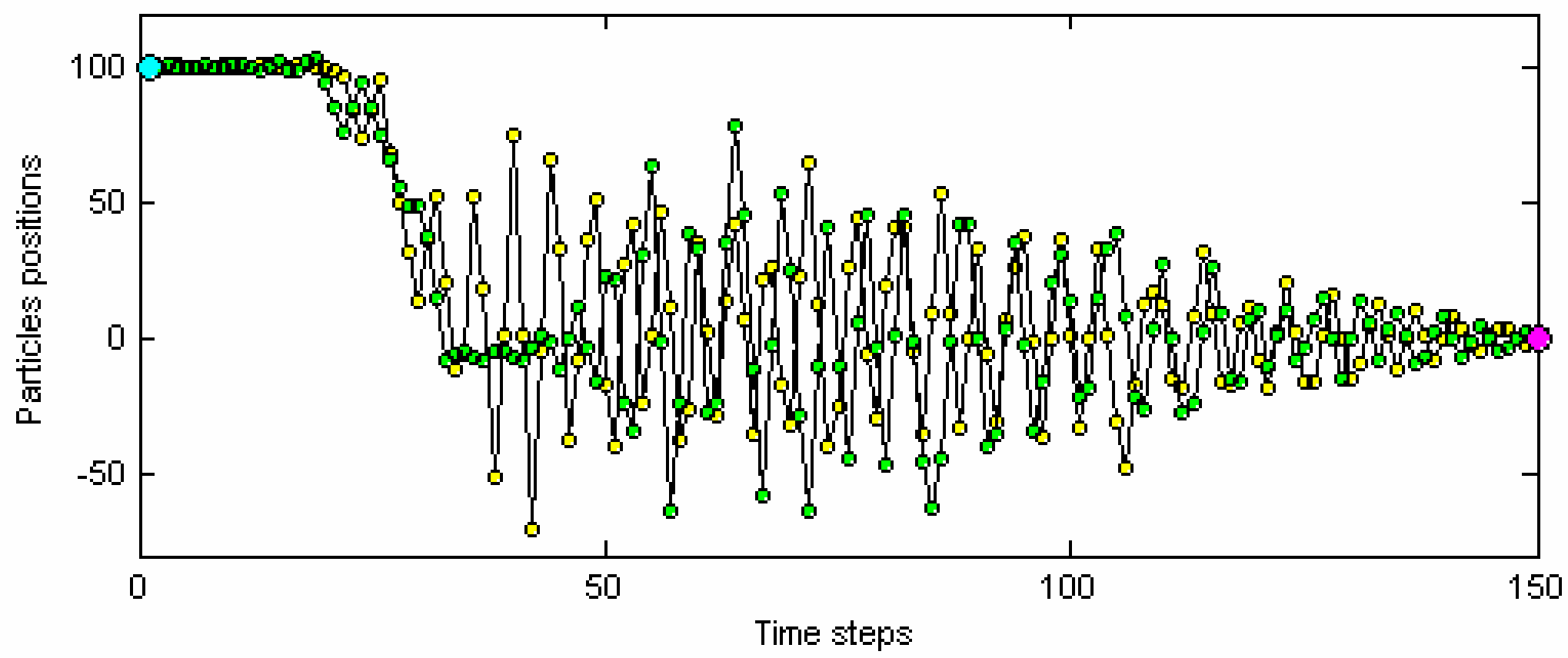

**Fig. 6. 12**: Evolution of two particles flying over a 1-dimensional search-space, where the particles are initially located at $x = 100$, their velocities are randomly initialized within the interval [–1,1] and programmed to linearly time-decrease from 100 to 1, and the function to be optimized is the Schaffer f6. The cyan and magenta dots are the particle's initial and final positions, respectively.

Clearly, a time-decreasing $v_{\max}$ enables the algorithm to perform a more explorative search at the beginning, while the search narrows enhancing exploitation throughout time.

## 6.2.3 Analysis of multiple interacting particles

### 6.2.3.1 Explosion

The trajectories of 10 particles flying over a 2-dimensional search-space along 500 time-steps are shown in **Fig. 6. 13**. The graph on the left corresponds to a deterministic PSO whose velocity updating rule is that of equation **(6. 4)**, but removing the random weights. The graph on the right corresponds to a probabilistic PSO—the O-PSO—whose velocity updating rule is that of equation **(6. 4)**.

It is important to note that the random weights $0 \leq U_{(0,1)} \leq 1$ in equation **(6. 4)**—and more generically in equation **(6. 1)**—make the particles repeatedly over-fly the best coordinates found so far independently from one another. Given that it is very unlikely that both random weights take the same value at a certain time-step like in the deterministic version, and since the global optimum is located at (0,0) for the Schaffer f6 function, the search tends to concentrate along the axes.

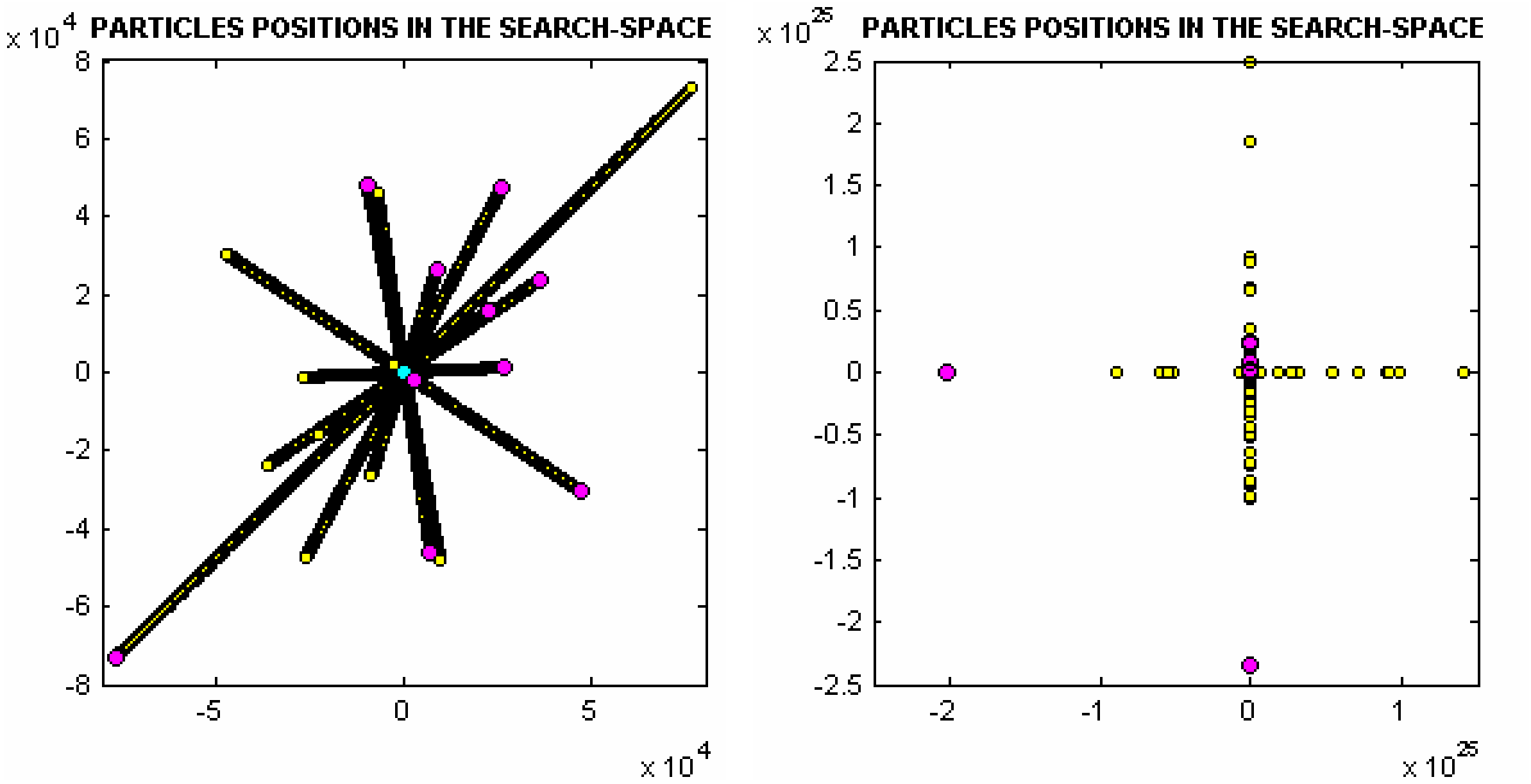

**Fig. 6. 13**: Explosion of a deterministic (left) and a probabilistic (right) PSO composed of 10 particles flying over a 2-dimensional search-space along 500 time-steps, for $w = 1$, $iw = sw = 2$, and no $v_{\max}$, where the particles are randomly initialized within the interval $[-100,100]$ and their velocities within the interval $[-1,1]$, and the function to be optimized is the Schaffer f6. The deterministic PSO differs from the probabilistic one in that the random weights are removed from the velocity updating equation. The cyan and magenta dots are the particle's initial and final positions, respectively.

As it can be seen in **Fig. 6. 13**, the incorporation of the random weights $0 \leq U_{(0,1)} \leq 1$ makes the deterministic explosion even more pronounced, in agreement with the conclusion in section **6.2.1.1** that the randomness favours the explosion.

### 6.2.3.2 General settings for the experiments

Most of the publications about the PSO method are concerned either with particular aspects of the general version, or with problem-specific versions of the algorithm. Similar to the aim of this thesis, Carlisle et al. [13] intend to develop a general-purpose PSO, which they call "an off-the-shelf PSO". While Shi et al. [71] reported that the PSOs are not quite sensitive to the population size, Carlisle et al. [13] suggest that, according to their experiments, a population size of 30 particles is *…small enough to be efficient, yet large enough to produce reliable results*.

Therefore, from here forth, the population size will be of 30 particles unless stated otherwise. Furthermore, taking into account the previous analyses, the next experiments are run only for the cases of constraining the components of the particles' velocities to a fixed 50%; to a fixed 10%; and to a linearly time-decreasing from 50% to 0.5% of the range $[x_{min}, x_{max}]$, where $[x_{min}, x_{max}]^n$ defines the feasible region of the search-space for an *n*-dimensional problem.

The following settings are the same for all the experiments run along the rest of this section:

- Feasible region of the search-space: $[-100,100]^2$
- Function to be optimized: Schaffer f6
- Maximum number of time-steps: $t_{max} = 4000$
- Number of particles: 30

Beware that, since the global optimum for this benchmark function is equal to 0, the conflict and the absolute error values are exactly the same.

### 6.2.3.3 Controlling the explosion: fixed $v_{max}$ = 100

The particles are randomly initialized within the feasible region, setting $v_{max} = 100$ (50% of the range $[-100,100]$), and the updating rule is that of equation **(6. 4)**.

It is fair to note that the Schaffer f6 function displays many local optima that do not differ much from the global optimum. For instance, the best conflict value found after only 3 time-steps is $5389.23\,\text{x}10^{-5}$, which is already close to the global optimum 0. The best solution found after the 4000 time-steps is $975.32\,\text{x}10^{-5}$, corresponding to a point located at the coordinates ($-3216.87\,\text{x}10^{-5}$, $313215.28\,\text{x}10^{-5}$).

The acceptable error for this benchmark function, which is commonly used in the optimizers' test suites, is typically equal to $1\,\text{x}10^{-5}$. Therefore, this implementation is not yet a working optimizer in spite of the fact that the explosion is controlled, and that the solution seems to be quite accurate.

The history of the particles' positions, best, and average conflicts is shown in **Fig. 6. 14**. Notice that although several evaluations of the conflict function are performed in the range $[-5,5]^2$, not a single particle is placed within that region at the final time-step. Furthermore, the final positions are as spread as the initial ones, showing that the particles do not cluster, and the average conflict oscillates around the value of 0.5 without tending to decrease. In conclusion, although the explosion is controlled, there does not seem to be much difference between this optimizer and simply performing $30 \cdot 4000 = 120000$ random evaluations of the conflict function in the range $[-100,100]^2$. The search is not narrowed here at any time.

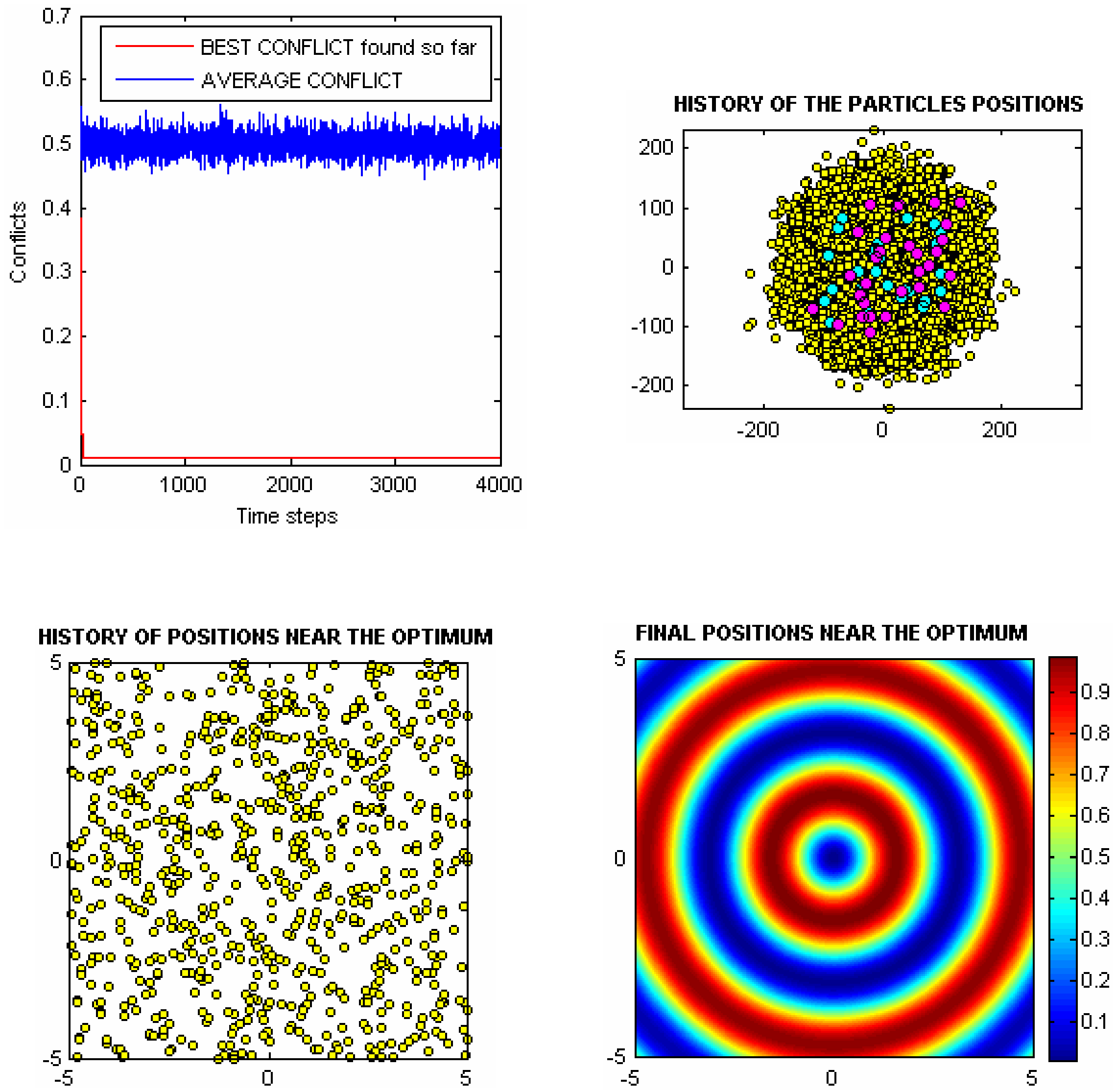

**Fig. 6. 14**: History of the particles' positions, best, and average conflicts for $w=1$, $iw=sw=2$ and $v_{\max}=100$. The cyan and magenta dots are the initial and final particles' positions, respectively.

#### 6.2.3.4 Controlling the explosion: fixed $v_{max}$ = 20

The particles are randomly initialized within the region of interest, setting $v_{max} = 20$ (10% of the range $[-100,100]$), and the updating rule is that of equation **(6. 4)**. The best solution found in this experiment is $338.38\,x10^{-5}$, corresponding to a point located at the coordinates ($737.92\,x10^{-5}$, $-5770.46\,x10^{-5}$). Therefore, the best solution found by this optimizer is still considerably above the acceptable solution $1\,x10^{-5}$.

The history of the particles' positions, best, and average conflicts is shown in **Fig. 6. 15**, where it can be observed that the search is now narrowed to a region around $[-30,30]^2$, and the number of evaluations within the region $[-5,5]^2$ is noticeably higher. In fact, there are now at least 3 particles that are positioned within that region in the final time-step. However, although this experiment already shows some kind of clustering, the results obtained are still not acceptable.

With regards to the evolution of the average conflict values, again there is no improvement throughout the 4000 time-steps despite the fact that the particles now narrow the search. This is because the Schaffer f6 function oscillates around the value of 0.5, so that the average conflict can only decrease when the particles start approaching the valley where the global optimum is located (see **Fig. 6. 1** and **Appendix 3**).

Smaller values of $v_{max}$ would result in better solutions for this particular case. In fact, there are some publications using settings like $v_{max} = 2$, $v_{max} = 3$, $v_{max} = 4$, or so (for instance, refer to [70, 72]). However, small values of $v_{max}$ reduce the algorithm's capability of escaping local optima, especially when the latter is not located within the space spanned by the particles' initial positions. Therefore, fixed values of $v_{max}$ below 10% of $[x_{min}, x_{max}]$ are not considered further within this dissertation.

With the aim to achieve a better exploitation of the region around the global optimum without giving up on the exploration ability, a new experiment is run hereafter using a linearly time-decreasing $v_{max}$ constraint, which exhibited a promising behaviour in the experiment run in section **6.2.2.5** (see **Fig. 6. 12**).

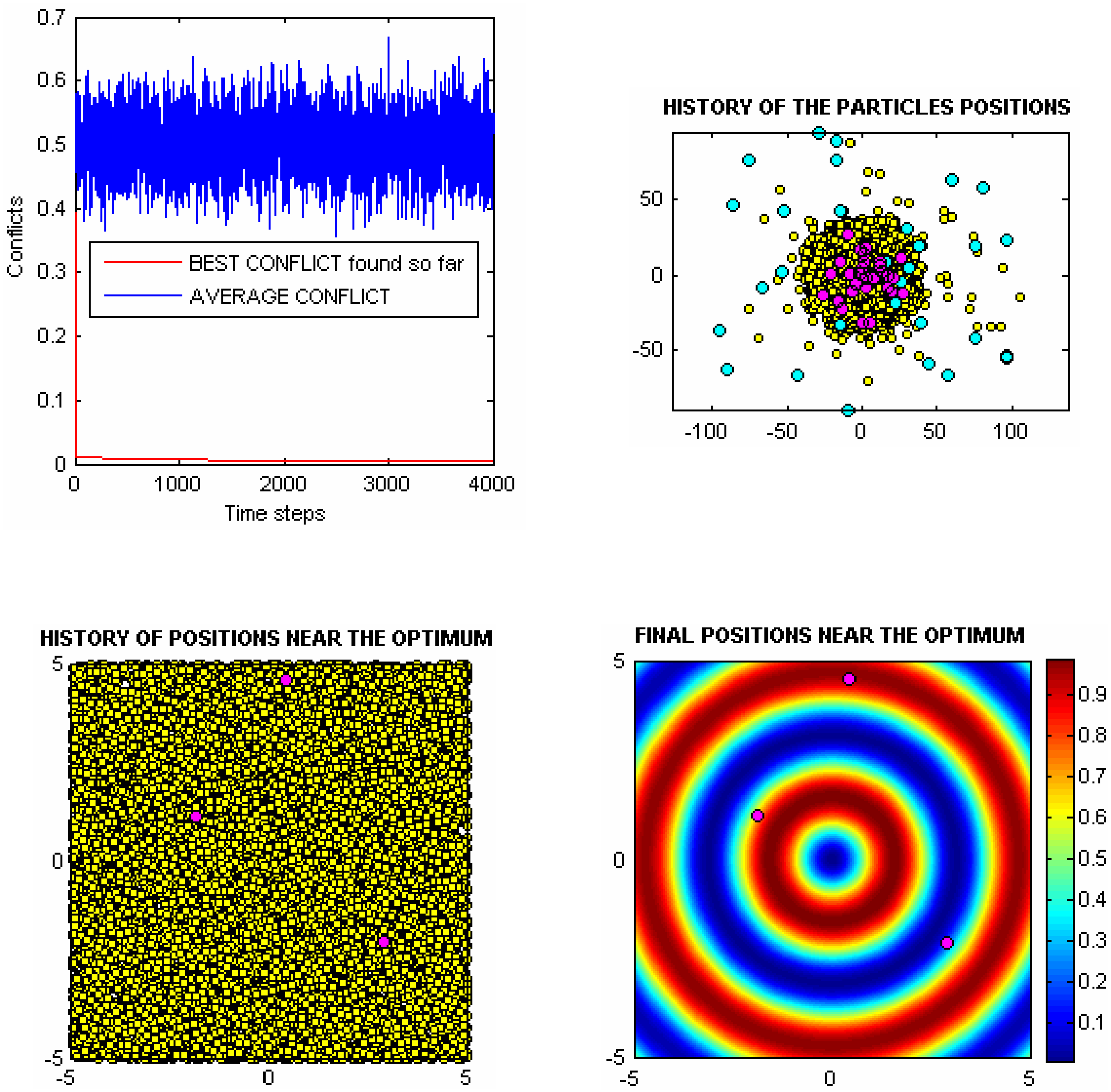

**Fig. 6. 15**: History of the particles' positions, best, and average conflicts for $w = 1$, $iw = sw = 2$ and $v_{max} = 20$. The cyan and magenta dots are the initial and final particles' positions, respectively.

### 6.2.3.5 Controlling the explosion: linearly time-decreasing $v_{max}$

An experiment is run implementing a linearly time-decreasing $v_{max}$, from 100 at the first time-step to 1 at the 4000$^{th}$ time-step (i.e. from 50% to 0.5% of the range $[-100,100]$).

The best solution found in this experiment is $62.56 \text{ x} 10^{-5}$, corresponding to a point located at the coordinates ($1625.73 \text{ x} 10^{-5}$, $-1899.75 \text{ x} 10^{-5}$). Therefore, although the best solution found by this optimizer is noticeably better than those obtained in previous experiments, it is still above the maximum acceptable solution of $1 \text{x} 10^{-5}$.

The history of the particles' positions, best, and average conflicts is shown in **Fig. 6. 16**:

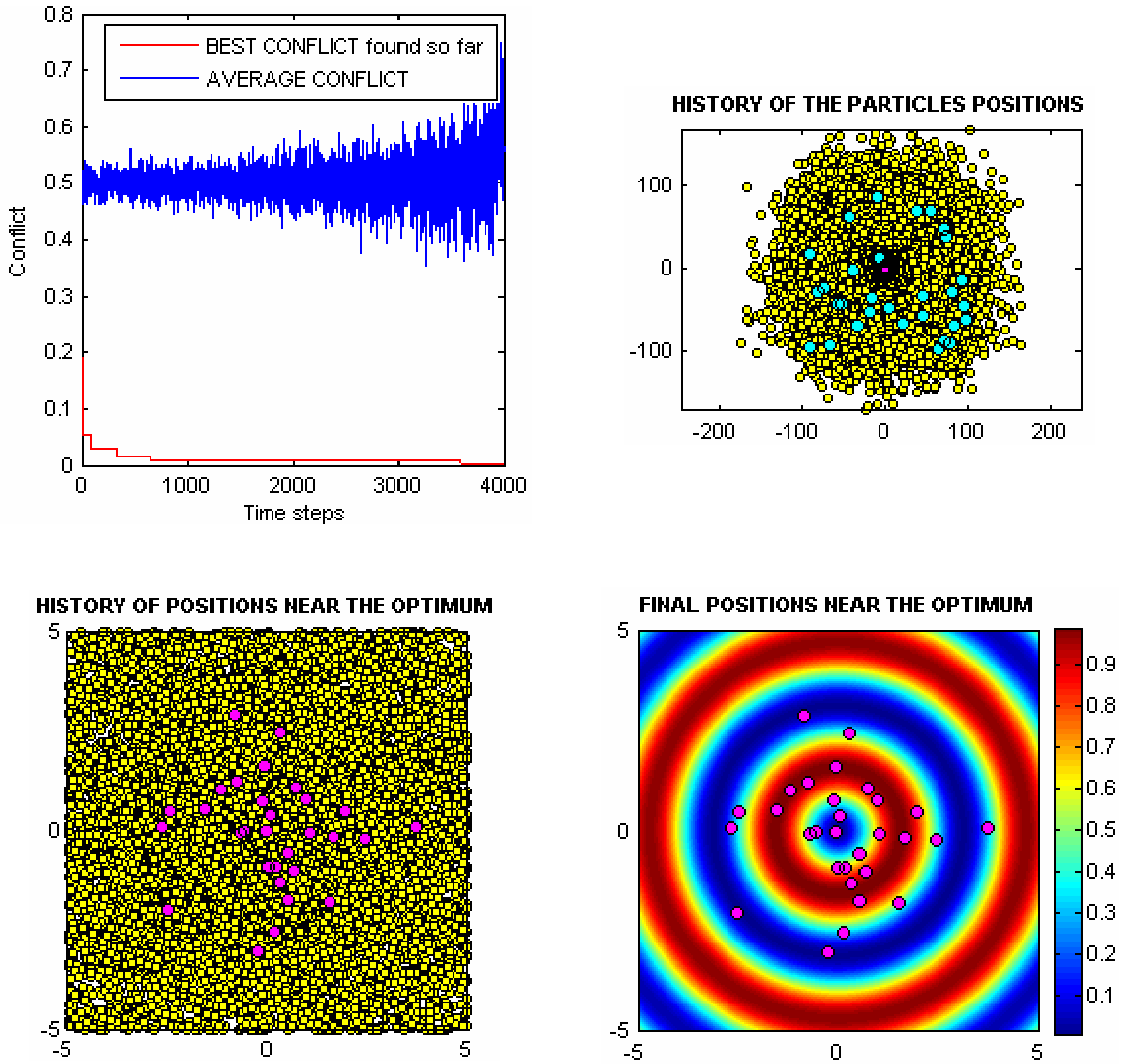

**Fig. 6. 16**: History of the particles' positions, best, and average conflicts for $w = 1$, $iw = sw = 2$ and $v_{\max}$ linearly time-decreasing from 100 at the first time-step to 1 at the 4000$^{th}$ time-step. The cyan and magenta dots are the initial and final particles' positions, respectively.

The resulting behaviour observed here is more promising than those observed in the previous experiments. While a solution better than that of the case with $v_{\max} = 20$ is found here—and all the particles end up within the region $[-5,5]^2$—, the optimizer also performs a much more extensive exploration of the whole region $[-100,100]^2$. The waves of the evolution of the average conflict display increasing amplitudes because the region of the search-space that contains the smallest conflicts also contains the highest ones for this function. Hence, and since all the particles did not yet converge to the deepest valley that contains the global optimum (for visualization, refer to **Appendix 3**), the oscillations of the average conflicts still increase their amplitudes.

In summary, the last is the best algorithm tested so far because it obtains the best solution, while still displays a very extensive exploration of the whole feasible region $[-100,100]^2$.

# 6.3 Inertia weight

It seems evident that the $v_{max}$ constraint is an effective manner of preventing the explosion of the swarm. However, while a relatively small value of $v_{max}$ narrows the search so that a better exploitation of the search-space is carried out, a too small value decreases the particles' reluctance to get trapped in suboptimal areas, especially when the space spanned by the particles' initial positions does not contain the global optimum. Furthermore, the precise quantification of "too small" is disappointingly problem-dependent.

Shi et al. [70] theorized that the relative importance of the particles' inertia and acceleration should be adapted to the different problems. Hence the inertia weight ($w$) was introduced, as shown in equation **(6. 1)**.

Shi et al. [72] carried out experiments in order to analyze the interaction between the inertia weight and the constraint $v_{max}$. They realized that if the value of $v_{max}$ is increased, a smaller value of $w$ leads to better results with regards to both the number of time-steps required to find, and the number of failures in finding an acceptable solution. They also noticed that the decrease of $w$ stagnated at the value of 0.8, concluding that if the convenient setting for $v_{max}$ is not clear, choosing $w$ = 0.8 and $v_{max} = x_{max}$ is a good starting point, which led to no failure in their experiments. Notice, however, that these settings were obtained for the particular case of the Schaffer f6 benchmark function.

## 6.3.1 Fixed inertia weight with no $v_{max}$

In order to investigate the ability of the inertia weight to prevent the explosion of the system, an experiment similar to those of Shi et al. [72] is carried out, but without any limitation to the components of the particles' velocities, and for a 1-dimensional search-space. Although the particles are eventually pulled back to the region of interest, a big explosion is observed, thus repeatedly evaluating positions that are far from the global optimum.

The same experiment is run again, but now setting $w$ = 0.7 so as to pull the particles harder towards the region of interest. Although the explosion occurred once again, the particles are quickly pulled back. The history of the particles' positions, best, and average conflicts is shown in **Fig. 6. 17**:

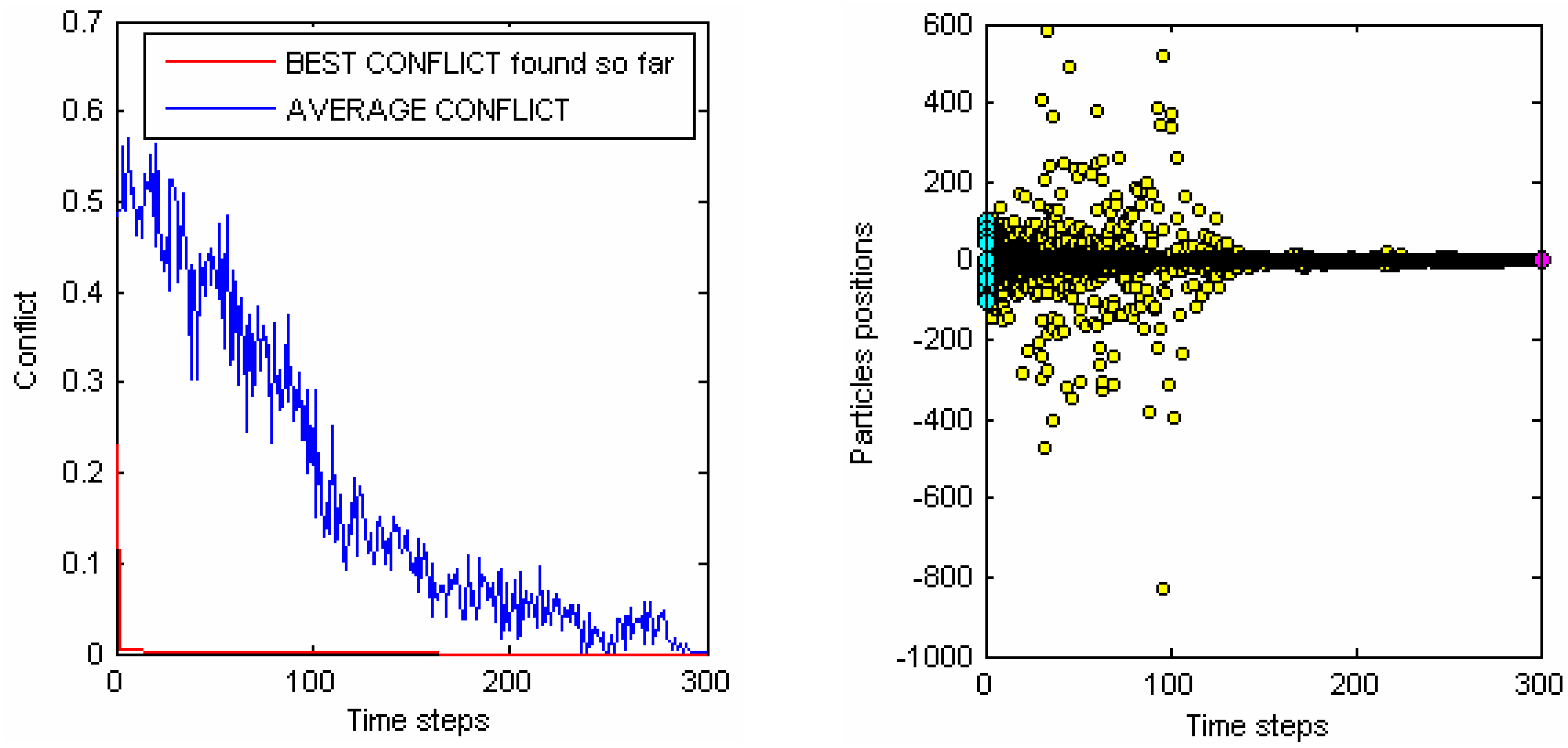

**Fig. 6. 17**: History of the particles' positions, best, and average conflicts for $w = 0.7$, $iw = sw = 2$, no $v_{max}$, and a 1-dimensional search-space. The cyan and magenta dots are the initial and final particles' positions, respectively.

It is evident that the incorporation of the inertia weight noticeably improves the performance of the optimizer. However, although a proper tuning of the inertia weight might make the elimination of the $v_{max}$ constraint possible—thus reducing the number of parameters to be tuned by the user—it is generally agreed among the researchers that it is preferable to keep some upper limit $v_{max}$ for the components of the particles' velocities. Nevertheless, this might be invisible to the user.

Another evident, useful effect of the inertia weight is the astonishing improvement of the precision of the solutions that the optimizer is able to find. In the last experiment, for instance, the error found at the 46th time-step is equal to $0.49\,x10^{-5}$. Even further, the error is null from the 165th time-step on. Notice that the average conflict now tends to decrease through time.

The same experiment is run for a 2-dimensional problem, and the history of the particles' positions, best, and average[3] conflicts is shown in **Fig. 6. 18**. Note that although the explosion is extensive, the convergence of the particles is astonishing in comparison to those of the experiments carried out in section **6.2**. The error at the 280th time-step is of $0.62\,x10^{-5}$, which is already less than the acceptable error of $1\,x10^{-5}$, and it is nullified by the 571st time-step. Beware that this is only a numerical "zero". In fact, the coordinates of the best solution found here are ($0.0004\,x10^{-5}$, $-0.0002\,x10^{-5}$) rather than (0, 0).

The negative feature of this optimizer is that a number of evaluations are performed far from the feasible region $[-100,100]^2$, and that while a thorough exploitation of a region near the global optimum is carried out, the exploration of the whole feasible region is not exactly magnificent. In contrast, the clustering of the particles is noticeably enhanced: all the 30 particles cluster on the region $[-5,5]^2$, and 29 of them do on the region $[-1\,x10^{-6},1\,x10^{-6}]^2$, without any constraint $v_{max}$ imposed. It is important to remark that, although it is not the convergence itself of the particles to a single point what is pursued, the convergence enables the particles to take successively smaller steps, thus fine-tuning the search.

## 6.3.2 Fixed inertia weight with fixed $v_{max}$

For optimization problems with constrained search-spaces where only the global optimum that is located within a hyper-cube-like feasible region is of interest, Shi et al. [72] suggest that setting $v_{max} = x_{max}$ and $w = 0.8$ is a good starting point. However, given that setting $w = 0.7$ led to better performances in previous experiments, this setting is kept here in order to make the comparisons possible.

[3] Bear in mind that, although the word "mean" is reserved for the average among the results obtained from different searches, and the word "average" for the average among the results obtained by all the particles, they might be sometimes used indistinctly.

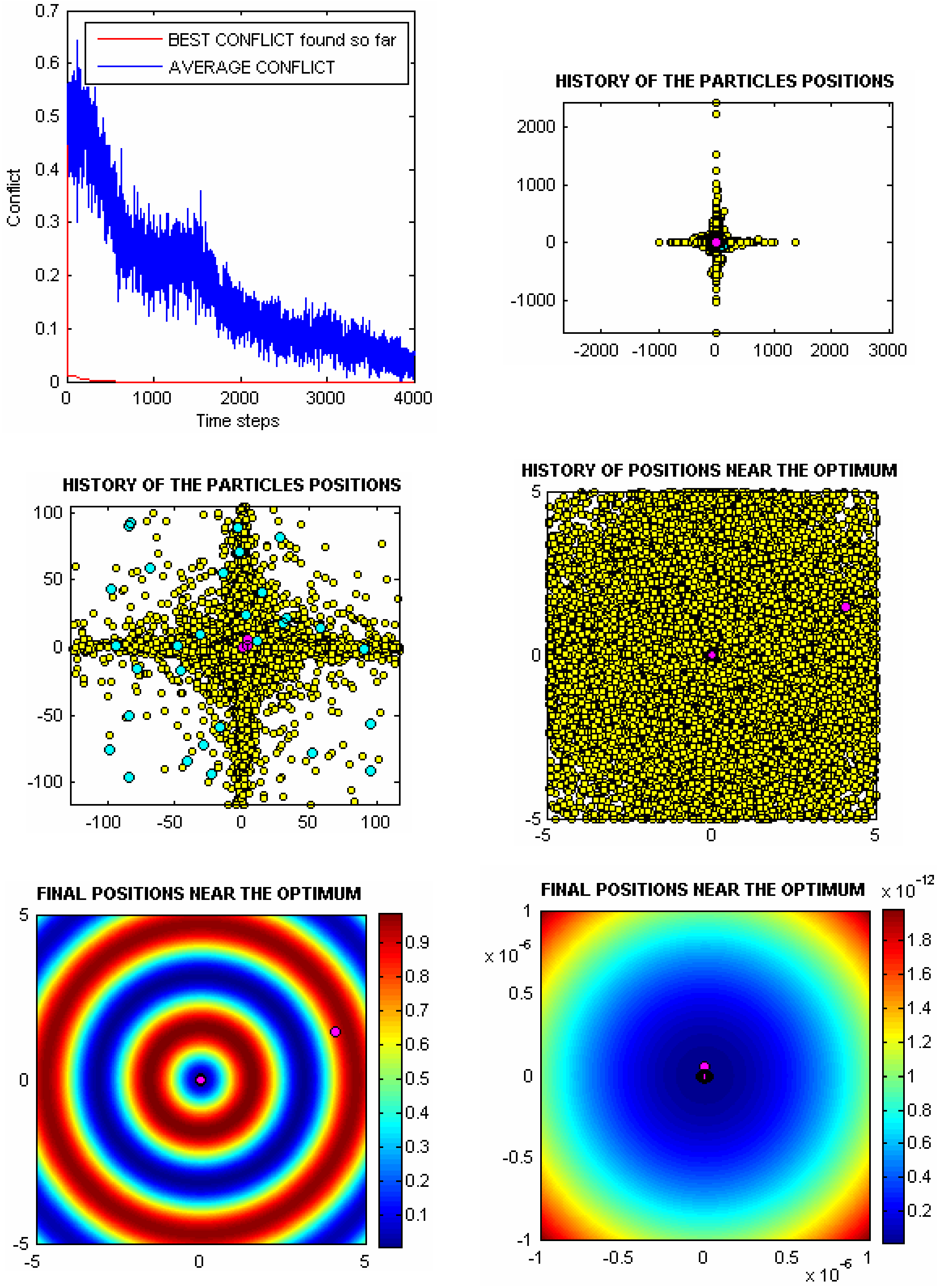

**Fig. 6. 18**: History of the particles' positions, best, and average conflicts for $w = 0.7$, $iw = sw = 2$, and no $v_{\max}$. The cyan and magenta dots are the initial and final particles' positions, respectively.

Therefore, a new experiment is carried out to investigate the effect of the inertia weight incorporated in the last experiment combined with the restriction to the components of the particles' velocities, as suggested by Shi et al. [72]. This limitation is more generically set to $v_{max} = 0.5 \cdot (x_{max} - x_{min})$ for the cases when the initialization of the particles' positions is not symmetric. This setting avoids the danger of setting $v_{max}$ too small, while it also prevents the system from successive evaluations of the conflict function far from the feasible region, as in the previous experiment (see **Fig. 6. 18**). The history of the particles' positions, best, and average conflicts is shown in **Fig. 6. 19**:

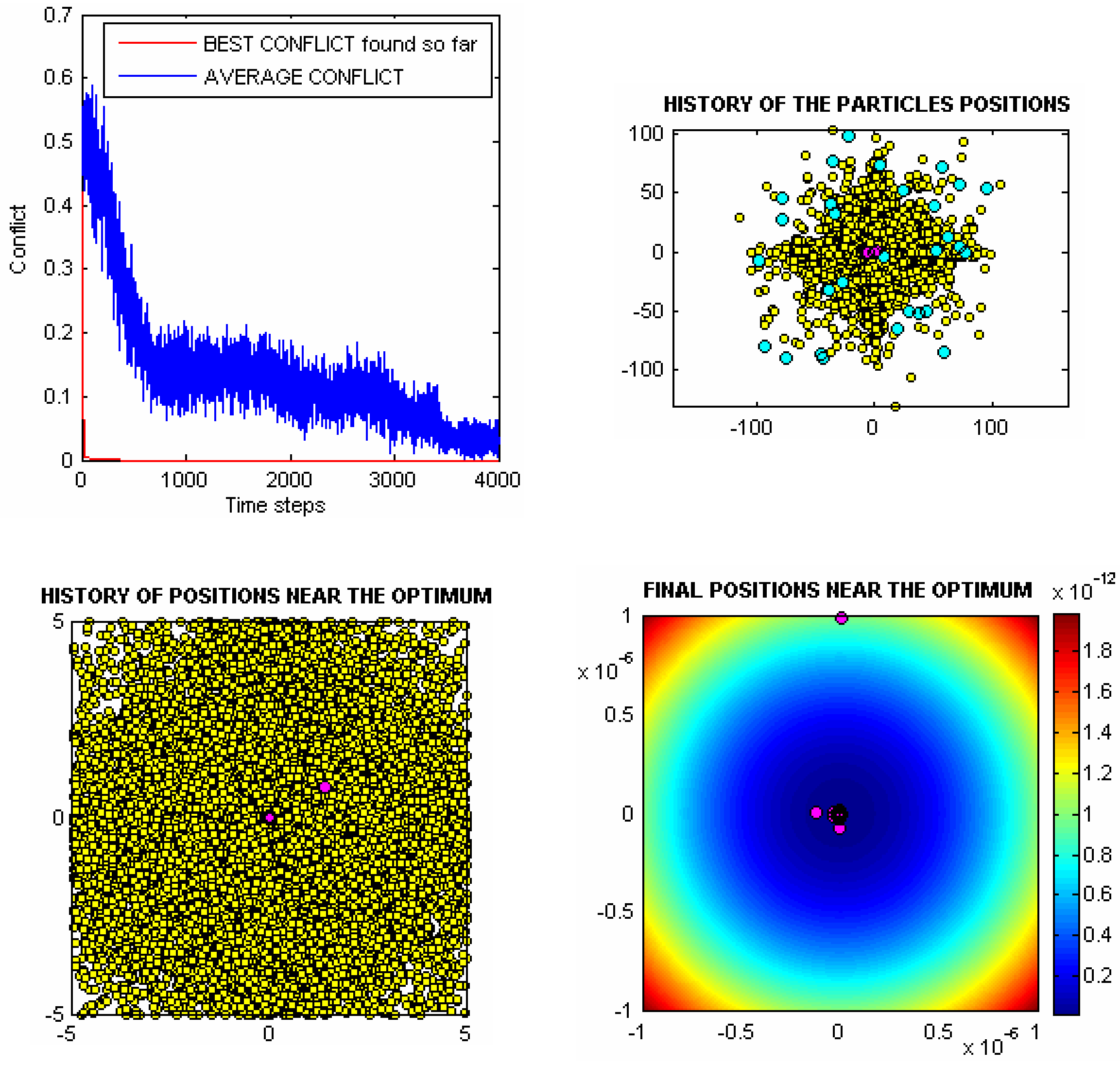

**Fig. 6. 19**: History of the particles' positions, best, and average conflicts for $w = 0.7$, $iw = sw = 2$, and $v_{max} = 100$. The cyan and magenta dots are the initial and final particles' positions, respectively.

The error found at the 134th time-step is of $0.98\,\text{x}10^{-5}$, which is already smaller than the acceptable error of $1\,\text{x}10^{-5}$, and the best solution possible is found at the 389th time-step, corresponding to a point located at the coordinates ($0.0004\,\text{x}10^{-5}$, $0.0002\,\text{x}10^{-5}$).

The differences between the results obtained by the last two experiments are subtle. Both the maximum acceptable error and the error's nullification are found slightly faster, and the clustering of the 30 particles on the region $[-5,5]^2$ is slightly better in this last experiment. In contrast, the clustering of the particles on the region $\left[-1\,\text{x}10^{-6}, 1\,\text{x}10^{-6}\right]^2$ is slightly worse. Perhaps the most important advantage of incorporating $v_{max} = 0.5 \cdot (x_{max} - x_{min})$ is in the better exploration of the whole feasible region $[-100,100]^2$. In conclusion, if the aim is to find the global optimum within a well defined region of the search-space, it seems a better strategy to set $v_{max} = 0.5 \cdot (x_{max} - x_{min})$ than to eliminate the $v_{max}$ constraint.

In order to investigate the combined effect of a more restrictive $v_{max}$ and the inertia weight, the same experiment is run, but now setting $v_{max} = 0.10 \cdot (x_{max} - x_{min}) = 20$. As expected, less exploration of the whole region $[-100,100]^2$ and a better exploitation of a region near the global optimum are exhibited. The history of the particles' positions is shown in **Fig. 6. 20**:

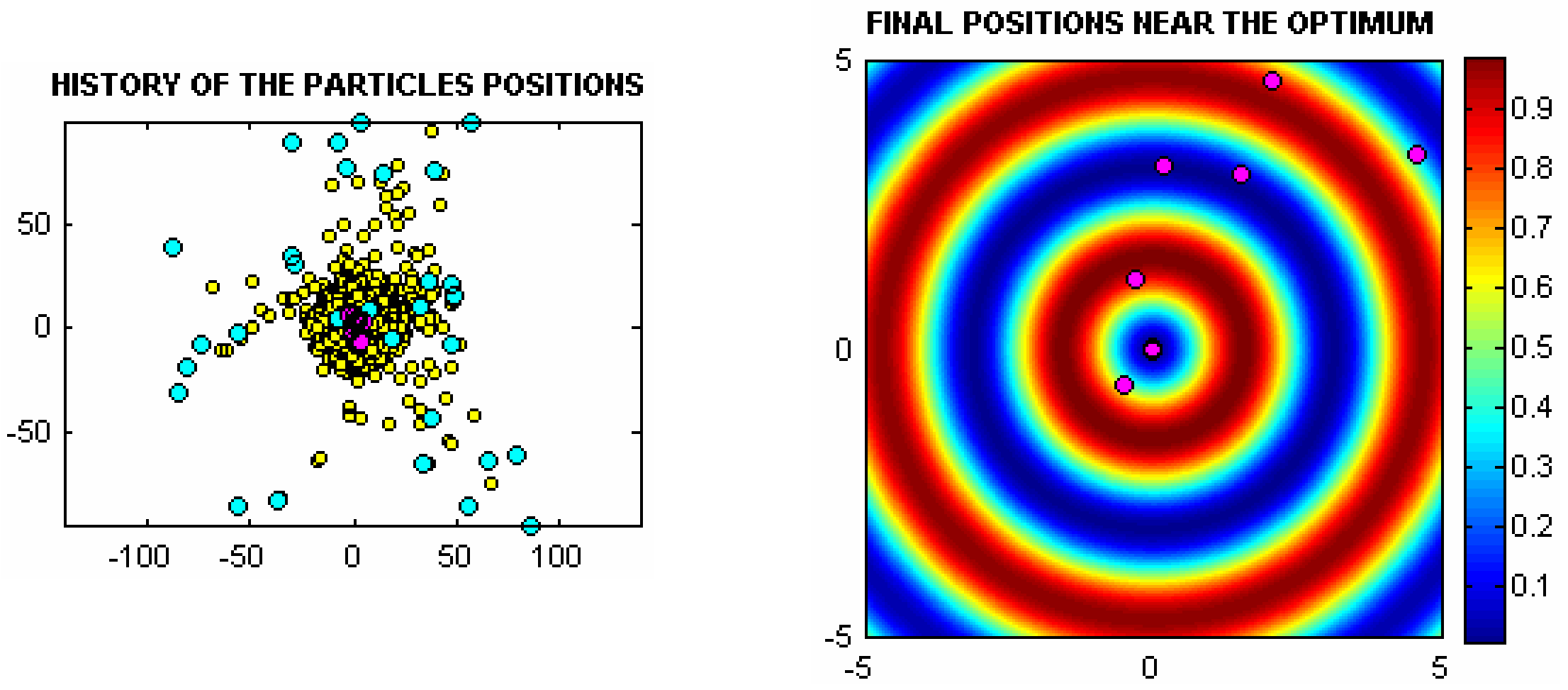

**Fig. 6. 20**: History of the particles' positions for $w = 0.7$, $iw = sw = 2$, and $v_{max} = 20$. The cyan and magenta dots are the initial and final particles' positions, respectively.

The error found at the 1782nd time-step is of $0.29\,\text{x}10^{-5}$, which is already smaller than the acceptable error of $1\,\text{x}10^{-5}$. The best solution possible is found at the 2079th time-step, corresponding to a point located at the coordinates ($0.0005\,\text{x}10^{-5}, -0.0001\,\text{x}10^{-5}$).

This last optimizer takes noticeably longer to attain the acceptable error and the best solution possible. Furthermore, as expected, the exploration is concentrated on a region near the global optimum, thus exhibiting a very poor exploration of the whole feasible region $[-100{,}100]^2$.

### 6.3.3 Fixed inertia weight with linearly time-decreasing $v_{max}$

Nevertheless, given that both settings, independently, led to promising results in previous experiments, a new experiment is run hereafter with an inertia weight fixed to $w = 0.7$ and a linearly time-decreasing $v_{max}$.

The history of the particles' positions, best, and average conflicts is shown in **Fig. 6. 21**. The error condition is attained at the 222nd time-step, when the error is of $0.39\,\text{x}10^{-5}$. The best solution possible is found by the 484th time-step, corresponding to a point located at the coordinates ($0.0002\,\text{x}10^{-5}, -0.0002\,\text{x}10^{-5}$).

The extensive exploration and both the number of time-steps required to attain the error condition and to find the best solution possible are very similar here to the case of $v_{max} = 100$ (see **Fig. 6. 19**). However, this last optimizer exhibits a better clustering of the particles, thus carrying out a better fine-tuning of the search.

Although both this optimizer and the one with $v_{max} = 100$ manage to find the best solution possible for this 2-dimensional problem, the ability of this last optimizer to perform a better exploitation of the search-space near the global optimum without abandoning the explorative behaviour during the early stages of the search might be advantageous for other problems, especially for higher-dimensional ones.

Notice that for higher-dimensional problems where the search-space can not be plotted, the clustering of the particles can be inferred by the merging of the curves of the evolutions of the average and of the best conflicts. Although this inference is not rigorous, it is very unlikely that these curves merge and stagnate without the complete implosion of the particles, which might happen, for instance, if all the particles are stuck in a flat region. Nevertheless, this is not a problem for the benchmark functions, whose "topography" is well known.

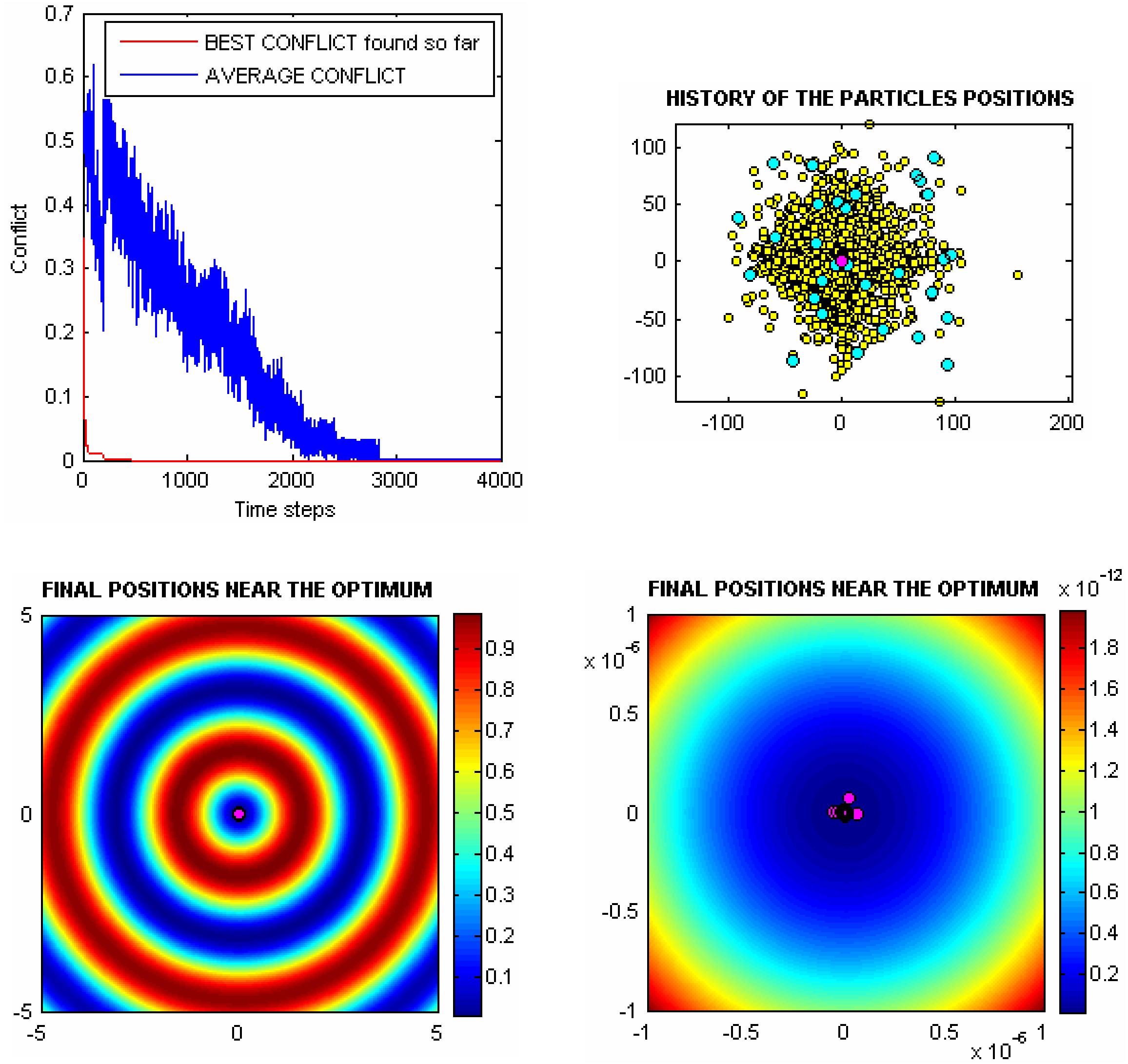

**Fig. 6. 21**: History of the particles' positions, best, and average conflicts for $w = 0.7$, $iw = sw = 2$, and $v_{max}$ linearly time-decreasing from 100 at the first time-step to 0 at the 4000$^{th}$ time-step. The cyan and magenta dots are the initial and final particles' positions, respectively.

## 6.3.4 Linearly time-decreasing inertia weight with no $v_{max}$

Shi et al. [71] ran some experiments with different linearly time-decreasing inertia weights, reporting better performance than for fixed inertia weights. They set $iw = sw = 2$, $v_{max} = x_{max}$, region of interest: $[-100,100]^n$, and a linearly time-decreasing inertia weight from 0.9 to 0.4.

Notice that while the experiments run by Shi et al. [71] optimized the Sphere, Rosenbrock, Rastrigrin, and Griewank functions, only the Schaffer f6 function is tested hereafter. More detailed quantitative analyses will be carried out later, comparing the performances of different optimizers on a “test suite” of benchmark functions, as used by Carlisle et al. [13].

Since the inertia weight sets the importance of the last velocity relative to the acceleration, it is natural to think that a time-decreasing inertia weight would favour exploration at the beginning, conceding increasing importance to the acceleration terms as time goes by. Hence an experiment is run hereafter with the inertia weight linearly time-decreasing from 0.9 at the initial to 0 at the 4000th time-step, and with no constraint imposed to the components of the particles’ velocities.

The history of the particles’ positions, best, and average conflicts is shown in **Fig. 6. 22**. The error condition is attained at the 962nd time-step ($0.88\,x10^{-5} < 1\,x10^{-5}$), and the best solution possible is found at the 1202nd time-step.

In spite of the extensive explosion, the optimizer manages to perform a reasonable exploration of the whole region $[-100{,}100]^2$ and exploitation of the region $[-5{,}5]^2$, finally finding the exact global optimum. It can be noticed by watching an animation of the evolution of the particles’ positions through time that the explosion takes place mainly (but not only) at the beginning of the run. Therefore, if no $v_{max}$ is imposed, it seems convenient to start with a smaller value of the inertia weight. In previous experiments, a constant $w = 0.7$ managed to noticeably reduce the explosion.

Therefore, the previous experiment is run again, but now setting a linearly time-decreasing inertia weight from 0.7 at the initial time-step to 0 at the final time-step. The error condition is attained at the 169th time-step, and the best solution possible is found at the 356th time-step. As expected, the explosion is considerably reduced, and both the error condition and the best solution possible are found much sooner. However, while a thorough exploitation of the region near the global optimum can be observed[4], the exploration of the whole feasible region is slightly worse than in the previous experiment. Some images obtained from this last experiment can be found in **Appendix 4**.

[4] As opposed to the experiment whose results are shown in **Fig. 6. 22**, the graphs of the evolutions of the mean and best conflicts merge here.

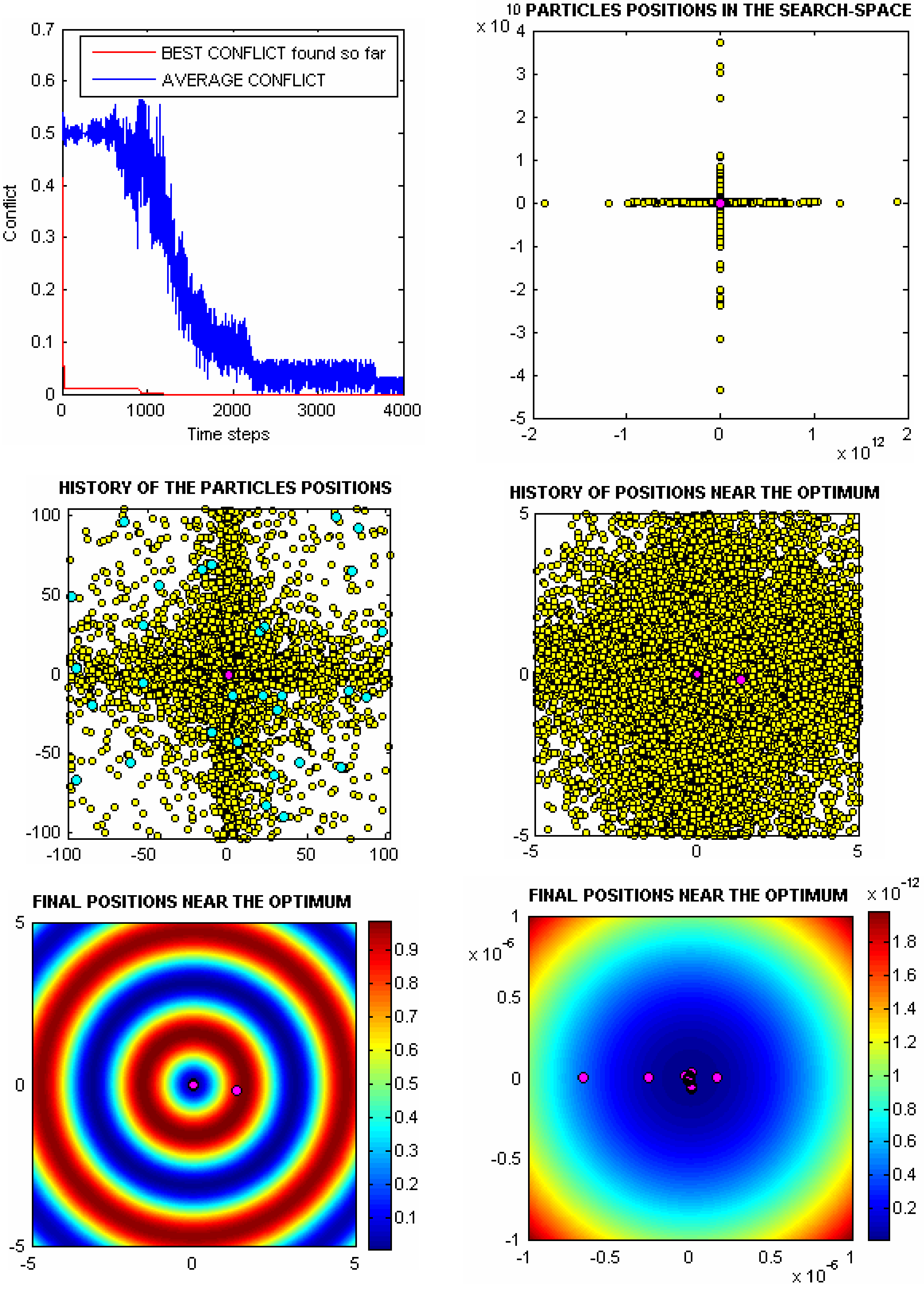

**Fig. 6. 22**: History of the particles' positions, best, and average conflicts for $w$ linearly time-decreasing from 0.9 at the initial time-step to 0 at the 4000[th] time-step, $iw = sw = 2$, and no $v_{\max}$. The cyan and magenta dots are the initial and final particles' positions, respectively.

### 6.3.5 Linearly time-decreasing inertia weight with fixed $v_{max}$

If the optimizer is in quest for a global optimum within a crisply defined finite feasible region, regardless of the existence of a better optimum anywhere else, it is generally a better strategy to constrain the components of the particles' velocities. As it has been argued before, small values of $v_{max}$ are not considered because they reduce the ability of the algorithm to escape local optima.

The $v_{max}$ constraint is external to the equations **(6. 1)** and **(6. 2)**, which rule the trajectories of the particles. Therefore, every time the constraint is violated, the trajectories of the particles are externally forced by equation **(6. 3)**, thus affecting the "natural" behaviour of the swarm. It seems that, in general, it is a better strategy to set a $v_{max}$ constraint small enough to control the explosion, yet large enough not to be frequently violated, thus letting the equations **(6. 1)** and **(6. 2)** control the trajectories of the particles. When only a certain region of the search-space is feasible, a reasonable velocity constraint should be related to the size of that region. A typical setting that led to good results in previous experiments is $v_{max} = 0.5 \cdot (x_{max} - x_{min})$.

Therefore, an experiment is run for a linearly time-decreasing inertia weight from 0.9 to 0, incorporating the constraint $v_{max} = 0.5 \cdot (x_{max} - x_{min})$. The history of the particles' positions, best, and average conflicts is shown in **Fig. 6. 23**. The error condition is attained at the 1040$^{th}$ time-step, and the best solution possible is found at the 1220$^{th}$ time-step. Note that the curves of the evolutions of the average and best conflicts merge.

Since these are probabilistic optimizers, different results are obtained each time the algorithm is run for the same settings. All the graphs presented so far were selected among the graphs corresponding to several runs for each setting, so that they are more or less representative of the average results. Bear in mind that only qualitative analyses are performed for the moment.

It is important to remark that the location of the best solution found by the optimizer does not typically correspond to any of the final particles' positions (magenta dots), and that although the clustering of the particles is not the ultimate objective, it helps to fine-tune the search. It is also important to note that too fast a convergence is not usually desirable for that it implies reducing the exploration of the whole feasible region.

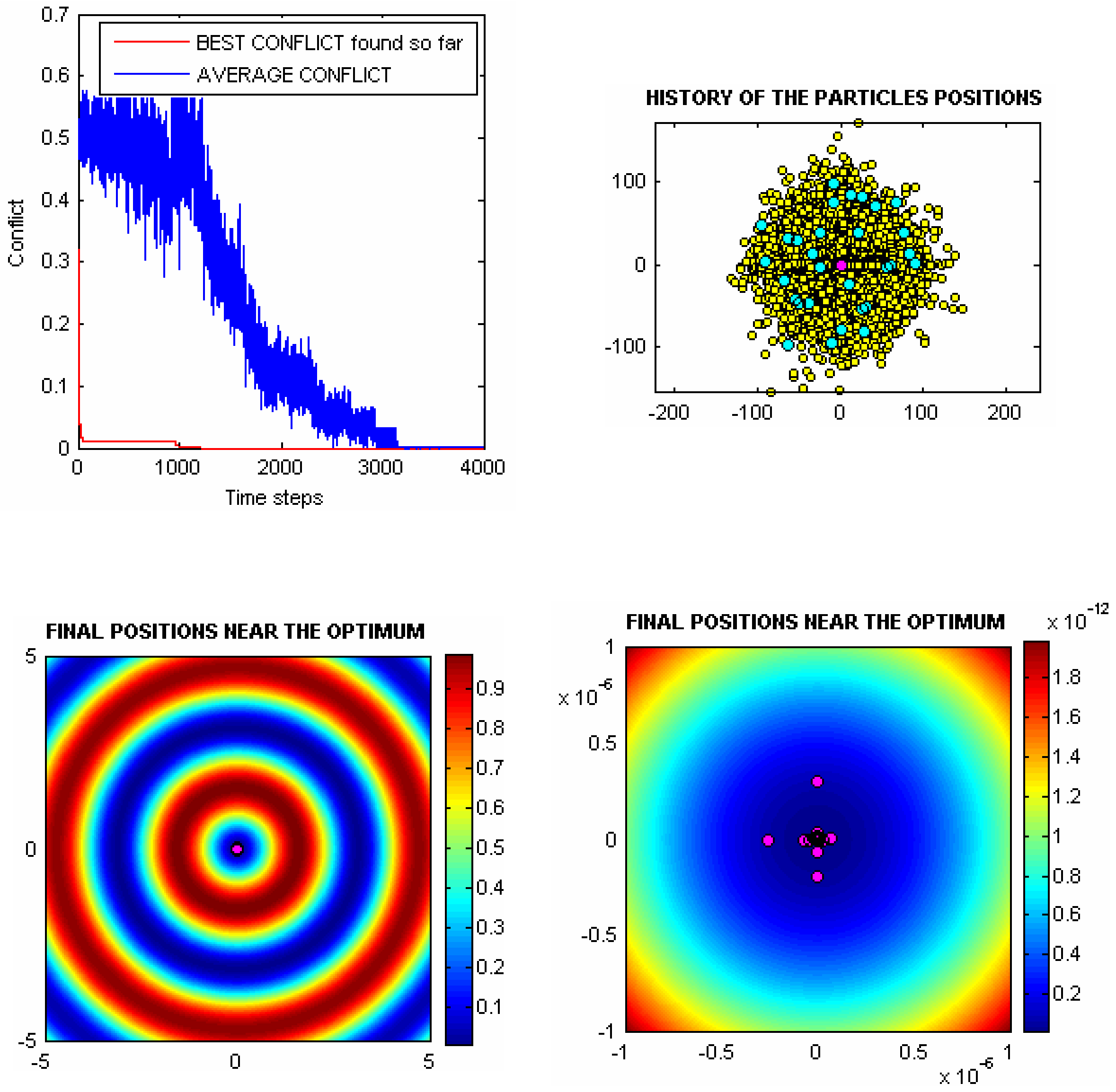

**Fig. 6. 23**: History of the particles' positions, best, and average conflicts for $w$ linearly time-decreasing from 0.9 at the initial time-step to 0 at the 4000th time-step, $iw = sw = 2$, and $v_{\max} = 100$. The cyan and magenta dots are the initial and final particles' positions, respectively.

Most of the optimizers tested so far are able to find the exact global optimum. However, thinking ahead to other benchmark functions and to real-world problems, the qualitative behaviour exhibited by this last optimizer is the best found so far, since it performs a very extensive exploration of the whole feasible region, a very thorough exploitation of a region near the optimum, and an outstanding clustering of the particles.

## 6.3.6 Linearly time-decreasing both inertia weight and $v_{max}$

Although it has been asserted before that it is preferable to leave the algorithm itself to guide the particles' clustering rather than to force it by means of a $v_{\max}$ constraint, previous experiments showed that a linearly time-decreasing $v_{\max}$ might help to fine-tune the search, while still allowing exploration during the early stages of the search. Therefore, a new experiment is run, combining the two promising strategies: a linearly time-decreasing inertia weight from 0.9 to 0, and a linearly time-decreasing $v_{\max}$ from $0.5 \cdot (x_{\max} - x_{\min}) = 100$ to 0.

Although the results are very good indeed, they are of a quality similar to that of the previous experiment. In fact, the history of the particles' positions, best, and average conflicts, as well as the numbers of time-steps required to attain both the error condition and the best solution possible, are, on average, very similar to those of the experiments run with $v_{\max} = 100$. Some images that gather the results obtained in this experiment can be found in **Appendix 4**.

## 6.3.7 Sigmoidly time-decreasing inertia weight

It is self-evident that displaying more explorative behaviour at the beginning of the run, while gradually swapping to more exploitative behaviour as time goes by, is a desirable feature. Thus, some strategies have been tried in the previous experiments aiming to attain this behaviour, namely linearly time-decreasing $v_{\max}$ and linearly time-decreasing inertia weight.

The Schaffer f6 function is indeed a very difficult function to be optimized, which exhibits many local optima, and the only way to get to the global optimum is to pass through them. However, the better a local optimum is, the closer it is located to the global optimum. Therefore, the convergence of a PSO, which is inherently fast, rapidly allocates the particles in the region of the global optimum for the Schaffer f6 function, but the very good local optima near the global optimum are very powerful attractors difficult to escape from. On the contrary, there are real-world optimization problems that possess numerous global or good local optima, which are distant from one another. While it is desirable for these cases that the algorithm displays explorative behaviour for a longer period of time, the linearly time-decreasing inertia weight reinforces the inherent fast convergence of the PSOs.

An interesting alternative could be to maintain the maximum value desired for the inertia weight for some time before starting to linearly decrease it through time. The problem with this approach is that the small inertia weights, which favour exploitation, take place only when the time-steps are close to the maximum allowed. In addition, thinking ahead in the analysis of the paradigm, when a measure of error is implemented as a stopping criterion, it is expected that the algorithm stops before the maximum number of time-steps is reached. In fact, the sooner, the better. Fine-tuning the search for the time-steps close to the maximum allowed does not seem to be the best strategy in that regard!

An alternative could be to force the inertia weight, for instance, to keep its maximum desired value for something like a third of the maximum time-steps allowed (to favour exploration), to keep its minimum value desired constant for the last third of the maximum time-steps allowed (to favour exploitation), and linearly time-decreasing the inertia weight in between. This last approach resembles somehow the sigmoid function used for the transfer function in the nonlinear artificial neuron (see section **3.5.5.1.3**, **Fig. 3. 13**), and for the probability that a bit has of adopting a certain belief in the b-PSO (see section **5.6.4.4**).

Of course, the plain sigmoid function is not suitable for the variation of the inertia weight, but its features of staying close to one for some time and close to zero for some time, while a continuous variation occurs in between, seems very much what it is desired here. Therefore, some variations have been made to that function in order to move the inflection point towards the middle of the time-steps allowed, and to make the variation time-decreasing instead of time-increasing. The constants $w_{\max}$ and $w_{\min}$ are incorporated so that the user can vary the upper and lower limits of the inertia weight, which are approached but never reached. Thus, the function for the sigmoidly time-decreasing inertia weight proposed here is as follows:

$$w(t) = \frac{\left(w_{\max} - w_{\min}\right)}{\left(1 + e^{\left(2 \cdot k \cdot \frac{t}{t_{\max}} - k\right)}\right)} + w_{\min} \qquad \textbf{(6. 13)}$$

The function shown in **(6. 13)** is plotted in **Fig. 6. 24** for $w_{\max} = 0.8$, $w_{\min} = 0$ and $k$ = 10. In order to show how the function scales for different settings of the maximum number of time-steps allowed, the plot is shown for $t_{\max} = 4000$ on the left, and for $t_{\max} = 200$ on the right.

Thus, a new experiment is carried out making use of a sigmoidly time-decreasing inertia weight from $w^{(1)} \to 0.8$ to $w^{(4000)} \to 0$, and $v_{\max} = 0.5 \cdot (x_{\max} - x_{\min}) = 100$. The particles are randomly placed in the region $[-100,100]^2$ for the initial time-step, and their final positions, best, and average conflicts are shown in **Fig. 6. 25**.

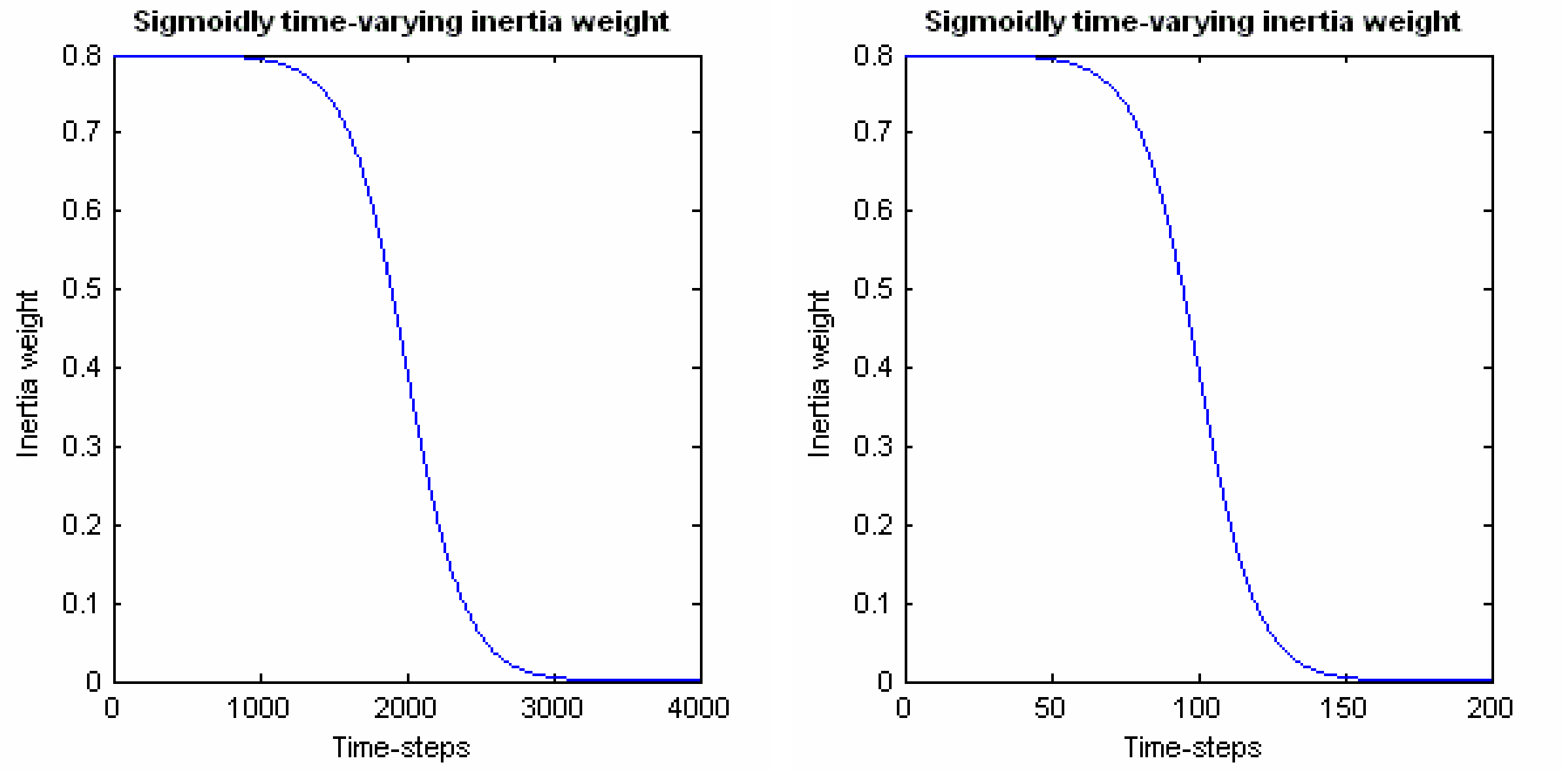

**Fig. 6. 24**: Sigmoidly time-decreasing inertia weight for $w_{\max} = 0.8$, $w_{\min} = 0$, and $k = 10$. The graph on the left shows the variation for a maximum number of time-steps allowed equal to 4000, and the one on the right for a maximum number of time-steps allowed equal to 200.

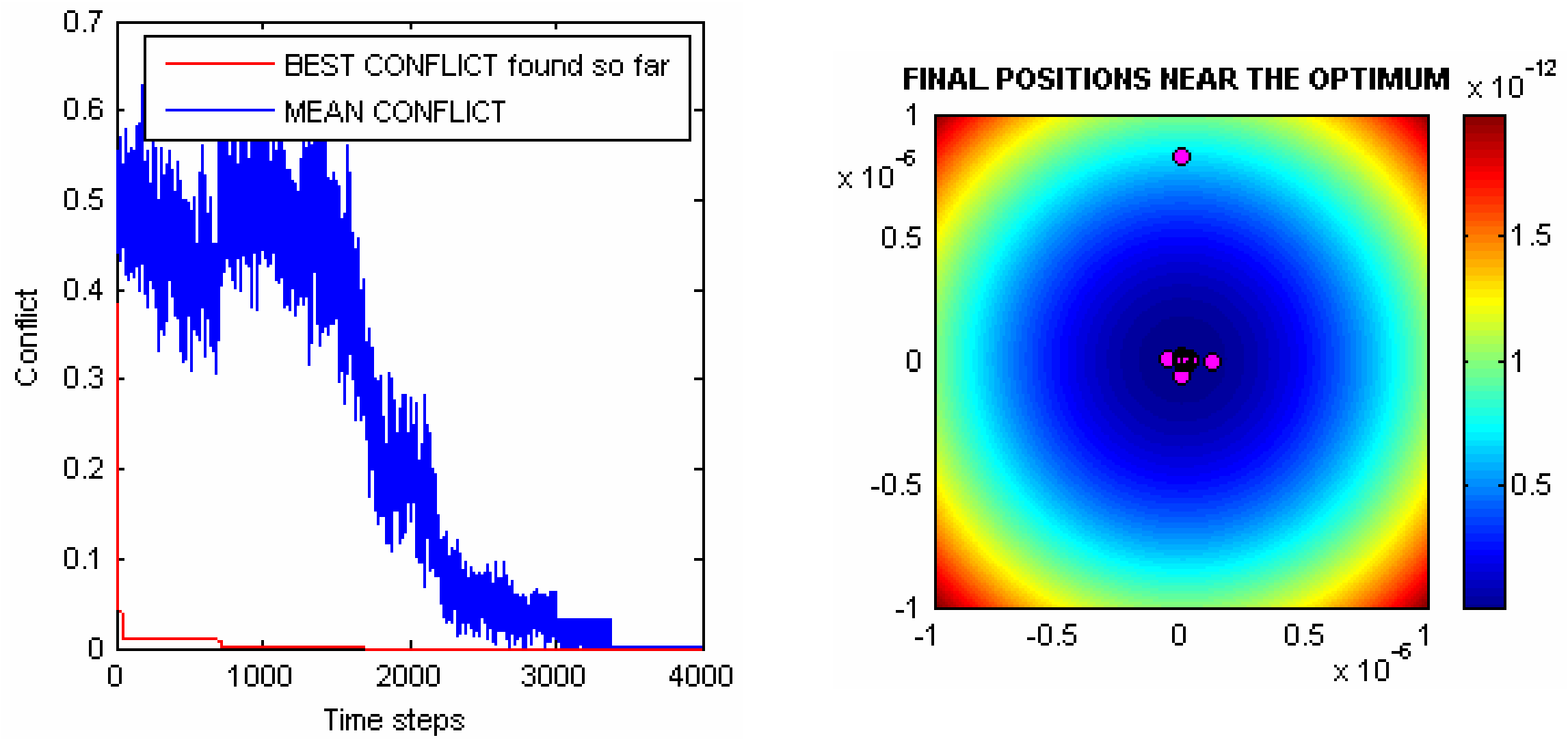

**Fig. 6. 25**: History of the best and average conflicts, and particles' positions at the final time-step, where the inertia weight is sigmoidly time-decreasing from $w^{(1)} \to 0.8$ to $w^{(4000)} \to 0$, $k = 10$, $iw = sw = 2$, and $v_{\max} = 100$.

The error condition is attained by the 1152nd time-step, and the best solution possible is found by the 1715th time-step. While all the particles end up within the region $\left[-1\,x10^{-6}, 1\,x10^{-6}\right]^2$, thus performing a thorough exploitation of a region near the global optimum, an extensive exploration of the region $\left[-100,100\right]^2$ is also carried out. At first glance, the results are similar to those of the experiment shown in **Fig. 6. 23**.

It is evident that at this level of precision, a quantitative, statistical analysis is required for the comparisons between different promising optimizers.

# 6.4 Learning weights

In the experiments with PSOs found in the literature, the individuality and sociality weights are generally set equal to one another, and typically to a value of around 2. It is easy to realize that setting the individuality and sociality weights equal to one another results in the dominance of the individual and of the social behaviour alternating during the search because of the random weights, without any of them taking the lead for too long. Although this feature is highly desirable for a robust, general-purpose optimizer, it is also reasonable to think that setting a high initial individuality weight relative to the sociality weight, while swapping their relative importance through time, would favour exploration in the early stages of the search, while smoothly swapping to more social behaviour as time goes by, so that exploitation of the regions that were found to be promising is enhanced during the late stages of the search.

Some optimizers equipped with different time-varying inertia weights and with learning weights set to $iw = sw = 2$ are implemented in section **6.4.1**. In order to test the influence of swapping the relative importance of the learning weights, the same optimizers are also implemented setting the individuality weight three times greater than the sociality weight at the initial time-step, linearly time-swapping their relative importance so that the sociality weight is three times more important than the individuality weight at the final time-step.

While the optimizers implemented in section **6.4.1** keep the learning weights unrelated to the inertia weight, two different relationships between them—which are expected to favour clustering—are proposed and tested in sections **6.4.2** and **6.4.3**, respectively.

## 6.4.1 Acceleration weight unrelated to the inertia weight

"Acceleration weight" is the name chosen here for the summation of the individuality and the sociality weights: $aw^{(t)} = iw^{(t)} + sw^{(t)}$.

In order to investigate only the effect of swapping the relative importance of the learning weights on the performance of the algorithm, the acceleration weight is kept constant through time, so that $aw^{(t)} = iw^{(t)} + sw^{(t)} = 4 \;\; \forall t$ within this section. Moreover, the $v_{\max}$ constraint is used from here on with the only purpose of helping to prevent the explosion, letting the algorithm itself control the convergence of the particles to fine-tune the search.

Therefore, the general settings for the experiments run within section **6.4.1** are as follows:

- $aw^{(t)} = iw^{(t)} + sw^{(t)} = 4 \;\; \forall t$
- $v_{\max} = 0.5 \cdot (x_{\max} - x_{\min})$
- Number of particles: 30
- $t_{\max} = 10000$

Six different optimizers derived from the B-PSO are proposed and compared. In order to make the comparisons valid, these probabilistic algorithms are run 50 times for each conflict function in the test suite, calculating the mean best solution found ($\overline{cgbest}$); the best solution found in any run (*cgbest*); the mean number of time-steps required to attain a simple error condition—taking advantage of the fact that the global optimum is known for each of the test benchmark functions—; the mean number of time-steps required to find the exact global optimum, if applicable; the number of failures in attaining the error condition; the number of times the global optimum is found; and the corresponding standard deviations.

### 6.4.1.1 First optimizer: Basic, standard PSO

All the optimizers used in section **6.4.1** are based on the B-PSO, whose velocity updating rule is shown in equation **(6. 1)**. Since this first optimizer makes use of a standard setting of the parameters of the algorithm, it is referred to as "basic, standard PSO": **BSt-PSO**.

The settings of the parameters for this optimizer are as follows:

- Inertia weight: $w^{(t)} = 0.7 \quad \forall t$
- Individuality weight: $iw^{(t)} = 2 \quad \forall t$
- Sociality weight: $sw^{(t)} = 2 \quad \forall t$

#### 6.4.1.2 Second optimizer: Basic, swapping PSO

This algorithm differs from the BSt-PSO in that it linearly time-swaps the relative importance of the individuality and sociality weights, while keeping the acceleration weight constant. Hence this optimizer is referred to as “basic, swapping PSO”: **BSw-PSO**.

The settings of the parameters for this optimizer are as follows:

- Inertia weight: $w^{(t)} = 0.7 \quad \forall t$
- Individuality weight: $iw^{(t=1)} = 3$, $iw^{(t=t_{max})} = 1$ (linearly time-decreasing)
- Sociality weight: $sw^{(t=1)} = 1$, $sw^{(t=t_{max})} = 3$ (linearly time-increasing)

#### 6.4.1.3 Third optimizer: Basic, standard, linearly time-decreasing PSO

This optimizer differs from the BSt-PSO in that it linearly time-decrease the inertia weight. Hence it is referred to as “basic, standard, linearly time-decreasing PSO”: **BStLd-PSO**.

Two different optimizers are tested here, which differ in the lower limit of the inertia weight. They are referred to as BStLd-PSO 1 and BStLd-PSO 2.

The settings of the parameters for these optimizers are as follows:

- Inertia weight 1: $w^{(t=1)} = 0.9$, $w^{(t=t_{max})} = 0$ (linearly time-decreasing)
- Inertia weight 2: $w^{(t=1)} = 0.9$, $w^{(t=t_{max})} = 0.4$ (linearly time-decreasing)
- Individuality weight: $iw^{(t)} = 2 \quad \forall t$
- Sociality weight: $sw^{(t)} = 2 \quad \forall t$

#### 6.4.1.4 Fourth optimizer: Basic, standard, sigmoidly time-decreasing PSO

This optimizer differs from the BStLd-PSO in that the inertia weight is sigmoidly instead of linearly time-decreasing (see **Fig. 6. 24**). Hence it is referred to as “basic, standard, sigmoidly time-decreasing PSO”: **BStSd-PSO**.

Two different optimizers are tested here, which differ in the upper and lower limits of the inertia weight. They are referred to as BStSd-PSO 1 and BStSd-PSO 2.

The settings of the parameters for these optimizers are as follows:

- Inertia weight 1: $w^{(t=1)} \rightarrow 0.8$, $w^{(t=t_{\max})} \rightarrow 0$ (sigmoidly time-decreasing)
- Inertia weight 2: $w^{(t=1)} \rightarrow 0.7$, $w^{(t=t_{\max})} \rightarrow 0.4$ (sigmoidly time-decreasing)
- Individuality weight: $iw^{(t)} = 2 \quad \forall t$
- Sociality weight: $sw^{(t)} = 2 \quad \forall t$

#### 6.4.1.5 Fifth optimizer: Basic, swapping, linearly time-decreasing PSO

This optimizer differs from the BStLd-PSO in that it linearly time-swaps the relative importance of the individuality and sociality weights. Therefore, it is referred to as "basic, swapping, linearly time-decreasing PSO": **BSwLd-PSO** (see **Fig. 6. 26**).

Two different optimizers are tested here, which differ in the lower limit of the inertia weight. They are referred to as BSwLd-PSO 1 and BSwLd-PSO 2.

The settings of the parameters for these optimizers are as follows:

- Inertia weight 1: $w^{(t=1)} = 0.9$, $w^{(t=t_{\max})} = 0$ (linearly time-decreasing)
- Inertia weight 2: $w^{(t=1)} = 0.9$, $w^{(t=t_{\max})} = 0.4$ (linearly time-decreasing)
- Individuality weight: $iw^{(t=1)} = 3$, $iw^{(t=t_{\max})} = 1$ (linearly time-decreasing)
- Sociality weight: $sw^{(t=1)} = 1$, $sw^{(t=t_{\max})} = 3$ (linearly time-increasing)

#### 6.4.1.6 Sixth optimizer: Basic, swapping, sigmoidly time-decreasing PSO

This optimizer differs from the BStSd-PSO in that it time-swaps the relative importance of the individuality and sociality weights. Hence it is referred to as "basic, swapping, sigmoidly time-decreasing PSO": **BSwSd-PSO** (see **Fig. 6. 26**).

Two different optimizers are tested here, which differ in the upper and lower limits of the inertia weight. They are referred to as BSwSd-PSO 1 and BSwSd-PSO 2.

The settings of the parameters for these optimizers are as follows:

- Inertia weight 1: $w^{(t=1)} \to 0.8$, $w^{(t=t_{max})} \to 0$ (sigmoidly time-decreasing)
- Inertia weight 2: $w^{(t=1)} \to 0.7$, $w^{(t=t_{max})} \to 0.4$ (sigmoidly time-decreasing)
- Individuality weight: $iw^{(t=1)} = 3$, $iw^{(t=t_{max})} = 1$ (linearly time-decreasing)
- Sociality weight: $sw^{(t=1)} = 1$, $sw^{(t=t_{max})} = 3$ (linearly time-increasing)

#### 6.4.1.7 Experimental results

The benchmark functions in the test suite, the regions of the search-space where the particles are initially spread over, and the acceptable absolute errors are described in **Table 6. 1**. For detailed surface plots and colour-maps of the benchmark test functions in two-dimensional search-spaces, refer to **Appendix 3**.

| | Mathematical expression | Parameters |
|---|---|---|
| Sphere | $f(\mathbf{x}) = \sum_{i=1}^{n} x_i^2$ | - Search-space: $[-100,100]^{30}$<br>- Acceptable error: < 0.01 |
| Rosenbrock | $f(\mathbf{x}) = \sum_{i=1}^{n-1} 100 \cdot \left(x_{i+1} - x_i^{\,2}\right)^2 + \left(x_i - 1\right)^2$ | - Search-space: $[-30,30]^{30}$<br>- Acceptable error: < 100 |
| Rastrigrin | $f(\mathbf{x}) = \sum_{i=1}^{n} \left[x_i^{\,2} - 10 \cdot \cos(2 \cdot \pi \cdot x_i) + 10\right]$ | - Search-space: $[-5.12,5.12]^{30}$<br>- Acceptable error: < 100 |
| Griewank | $f(\mathbf{x}) = \frac{1}{4000} \cdot \sum_{i=1}^{n} x_i^{\,2} - \prod_{i=1}^{n} \cos\left(\frac{x_i}{\sqrt{i}}\right) + 1$ | - Search-space: $[-600,600]^{30}$<br>- Acceptable error: < 0.1 |
| Schaffer f6 2D | $f(\mathbf{x}) = \frac{\left(\sin\sqrt{x_1^{\,2} + x_2^{\,2}}\right)^2 - 0.5}{\left[1 + 0.001 \cdot \left(x_1^{\,2} + x_2^{\,2}\right)\right]^2} + 0.5$ | - Search-space: $[-100,100]^{2}$<br>- Acceptable error: < 0.00001 |
| Schaffer f6 | $f(\mathbf{x}) = \frac{\left[\sin\left(\sqrt{\sum_{i=1}^{n} x_i^{\,2}}\right)\right]^2 - 0.5}{\left(1 + 0.001 \cdot \sum_{i=1}^{n} x_i^{\,2}\right)^2} + 0.5$ | - Search-space: $[-100,100]^{30}$<br>- Acceptable error: < 0.1 |

**Table 6. 1**: Benchmark functions included in the test suite.

It should be remarked that the different functions included in the test suite are meant to test different features of the optimizers. Thus, the Sphere function is included to test the ability of the algorithm to deal with high-dimensional, but simple problems; the Rosenbrock function to test the performance of the algorithm when dealing with problems whose object variables are highly inter-dependent (besides, there is an extensive "flat" region surrounding the global optimum); the Rastrigrin function is used to test the ability of the optimizer to escape local optima; the Griewank function is included to test the ability of the algorithm to optimize simple, but noisy problems; and the 2-dimensional Schaffer f6 function, which is the last function included in Carlisle et al.'s test suite [13], is used here to test the ability of the algorithm to handle very hard, but low-dimensional problems.

Given that all the optimizers tested within this section found no difficulty, not only in attaining the error condition, but also in finding the exact global optimum for the 2-dimensional Schaffer f6 function, its generalized 30-dimensional version is added to Carlisle et al.'s test suite [13]. This last test function also serves the function of further testing the algorithm against problems that exhibit numerous local optima. Although this ability is already tested by the Rastrigrin function, the latter displays a sphere-like trend-hyper-surface, whose gradients point towards the global optimum! In contrast, since the Schaffer f6 function exhibits a topography that oscillates around a value equal to 0.5, all the gradients of its "trend-hyper-surface" are null, thus making the search more difficult. In addition to that, while the local optima displayed by the Rastrigrin function resemble valleys, the ones displayed by the Schaffer f6 function resemble ring-like depressions that surround the single-point global optimum (see **Appendix 3**). Therefore, the 30-dimensional Schaffer f6 benchmark function, which proved itself to be very difficult to be optimized, is added to the test suite. Bear in mind that the acceptable error stated in **Table 6. 1** for this last test function was defined such that the optimizers presented numerous failures.

References:

- BSt-PSO: basic, standard PSO:
  $w^{(t)} = 0.7, \;\; iw^{(t)} = sw^{(t)} = 2 \;\; \forall t$
- BSw-PSO: basic PSO with linearly time-swapping learning weights:
  $w^{(t)} = 0.7 \;\; \forall t$, $iw^{(1)} = sw^{(10000)} = 3$, $iw^{(10000)} = sw^{(1)} = 1$
- BStLd-PSO 1: basic, standard PSO with linearly time-decreasing inertia weight:
  $w^{(1)} = 0.9$, $w^{(10000)} = 0$, $iw^{(t)} = sw^{(t)} = 2 \;\; \forall t$
- BStLd-PSO 2: basic, standard PSO with linearly time-decreasing inertia weight:

$w^{(1)} = 0.9$, $w^{(10000)} = 0.4$, $iw^{(t)} = sw^{(t)} = 2 \quad \forall t$

- BStSd-PSO 1: basic, standard PSO with sigmoidly time-decreasing inertia weight:
  $w^{(1)} \to 0.8$, $w^{(10000)} \to 0$, $iw^{(t)} = sw^{(t)} = 2 \quad \forall t$
- BStSd-PSO 2: basic, standard PSO with sigmoidly time-decreasing inertia weight:
  $w^{(1)} \to 0.7$, $w^{(10000)} \to 0.4$, $iw^{(t)} = sw^{(t)} = 2 \quad \forall t$
- BSwLd-PSO 1: basic PSO with linearly time-swapping learning weights and linearly time-decreasing inertia weight:
  $w^{(1)} = 0.9$, $w^{(10000)} = 0$, $iw^{(1)} = sw^{(10000)} = 3$, $iw^{(10000)} = sw^{(1)} = 1$
- BSwLd-PSO 2: basic PSO with linearly time-swapping learning weights and linearly time-decreasing inertia weight:
  $w^{(1)} = 0.9$, $w^{(10000)} = 0.4$, $iw^{(1)} = sw^{(10000)} = 3$, $iw^{(10000)} = sw^{(1)} = 1$
- BSwSd-PSO 1: basic PSO with linearly time-swapping learning weights and sigmoidly time-decreasing inertia weight:
  $w^{(1)} \to 0.8$, $w^{(10000)} \to 0$, $iw^{(1)} = sw^{(10000)} = 3$, $iw^{(10000)} = sw^{(1)} = 1$
- BSwSd-PSO 2: basic PSO with linearly time-swapping learning weights and sigmoidly time-decreasing inertia weight:
  $w^{(1)} \to 0.7$, $w^{(10000)} \to 0.4$, $iw^{(1)} = sw^{(10000)} = 3$, $iw^{(10000)} = sw^{(1)} = 1$

- $\overline{cgbest}$: mean best solution found along 50 runs of the optimizer
- *cgbest*: best solution found in any of the 50 runs of the optimizer
- $\sigma$: standard deviation
- *nf*: number of failures in attaining the error condition stated in **Table 6. 1** (out of the 50 runs)
- *ngb*: number of times that the global best solution is found along the 50 runs
- $\overline{tsgb}, \overline{tsec}$: mean number of time-steps required to find the global optimum and the error condition, respectively

It must be noticed that the optimizers tested here perform unconstrained optimization—except for the constriction to the components of the particles' velocities—, as opposed to most of the experiments reported in the literature (e.g. [13, 29, 70, 71, 72]), which are typically implemented for search-spaces constricted to the hyper-cubes stated in **Table 6. 1**.

While different techniques to constrain the search-space are discussed in **Chapter 10**, it is convenient to note here that an effective technique consists of allowing the particles to fly over infeasible regions, and to evaluate infeasible positions so as to guide the search, but banning the memory of the infeasible solutions. In other words, the infeasible solutions are not allowed to become the particles' best previous experiences. Therefore, since the best solutions are anyway within the feasible hyper-cube for the benchmark functions included in the test suite, the unconstrained algorithms tested here—which randomly spread the particles over the hyper-cubes stated in **Table 6. 1** for the initial time-step—work just as if this technique to deal with constrained search-spaces was actually implemented.

The evolution of the inertia, individuality, and sociality weights for two of the optimizers that are to be tested, namely the BSwLd-PSO 1 and the BSwSd-PSO 1, is shown in **Fig. 6. 26**:

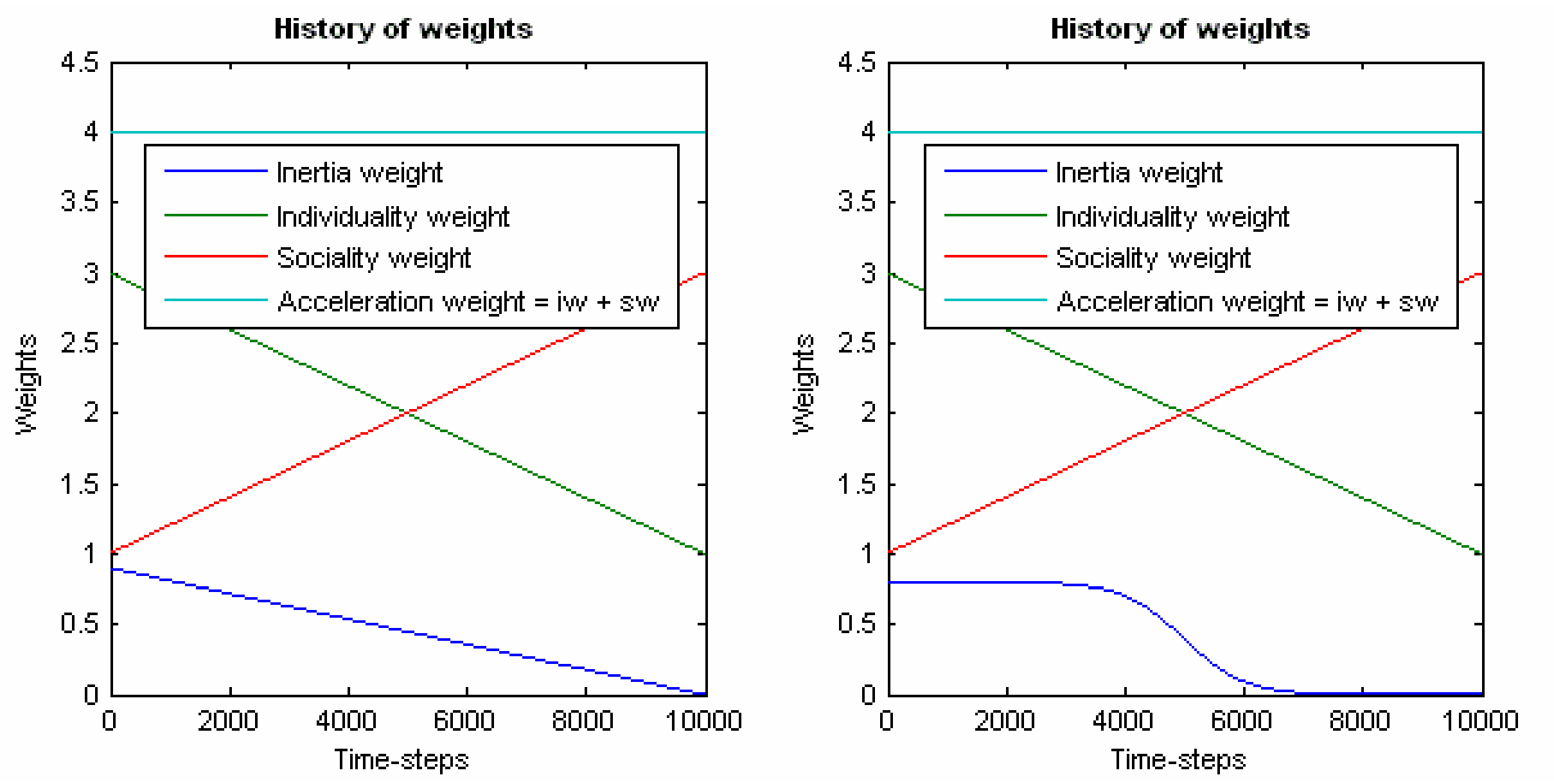

**Fig. 6. 26**: Evolution of the weights for the BSwLd-PSO 1 (left) and the BSwSd-PSO 1 (right).

The performance of each of the 10 optimizers on each of the 6 benchmark test functions is shown in **Table 6. 2** to **Table 6. 7**, and in **Fig. 6. 27** to **Fig. 6. 33**.

### 6.4.1.7.1 SPHERE function

The results of the experiments for the optimization of this function are gathered in **Table 6. 2**, and the evolution of the mean best solution found by each optimizer is plotted in **Fig. 6. 27**.

As expected, none of the optimizers finds much difficulty in attaining the error condition for this function, which is included in the test suite just to make sure that the optimizers are not developed for complex problems only, turning out to be inefficient for the simple ones. However, it is quite intriguing to notice that no optimizer is able to find the exact global optimum for this simple function in any of the 50 runs.

| SPHERE | $\overline{cgbest}\,(\sigma)$ | $cgbest$ | $\overline{tsec}\,(\sigma)$ | $\overline{tsgb}\,(\sigma)$ | $nf$ | $ngb$ |
|---|---|---|---|---|---|---|
| BSt-PSO | **1.24555004E-08** (2.44215030E-08) | **2.34330229E-11** | 4273.50 (492.58) | - - | 0 | 0 |
| BSw-PSO | **2.94750157E-04** (6.10384758E-04) | **7.81983857E-07** | 5737.08 (634.19) | - - | 0 | 0 |
| BStLd-PSO 1 | **1.16023193E-40** (8.20400213E-40) | **1.69280502E-67** | 3345.60 (64.21) | - - | 0 | 0 |
| BStLd-PSO 2 | **2.79569709E-58** (1.85927422E-57) | **2.15561785E-64** | 5386.82 (98.23) | - - | 0 | 0 |
| BStSd-PSO 1 | **1.39512854E-11** (7.19071641E-11) | **3.27163543E-18** | 4753.20 (36.39) | - - | 0 | 0 |
| BStSd-PSO 2 | **2.10086602E-97** (1.28199325E-96) | **1.13692758E-105** | 3791.58 (251.32) | - - | 0 | 0 |
| BSwLd-PSO 1 | **1.73981035E-89** (6.76820882E-89) | **2.07408692E-95** | 3412.18 (60.56) | - - | 0 | 0 |
| BSwLd-PSO 2 | **3.26003024E-39** (9.87041517E-39) | **1.12942586E-43** | 5413.70 (98.11) | - - | 0 | 0 |
| BSwSd-PSO 1 | **1.95282045E-14** (8.04226424E-14) | **3.41158500E-43** | 4743.52 (27.35) | - - | 0 | 0 |
| BSwSd-PSO 2 | **8.98388996E-70** (4.02967016E-69) | **3.73515365E-75** | 4329.60 (111.49) | - - | 0 | 0 |

**Table 6. 2**: Performance of 10 algorithms when optimizing the 30-dimensional Sphere benchmark test function along 10000 time-steps, where the particles are initially randomly spread over the region $[-100,100]^{30}$ , and the statistical data are calculated out of 50 runs.

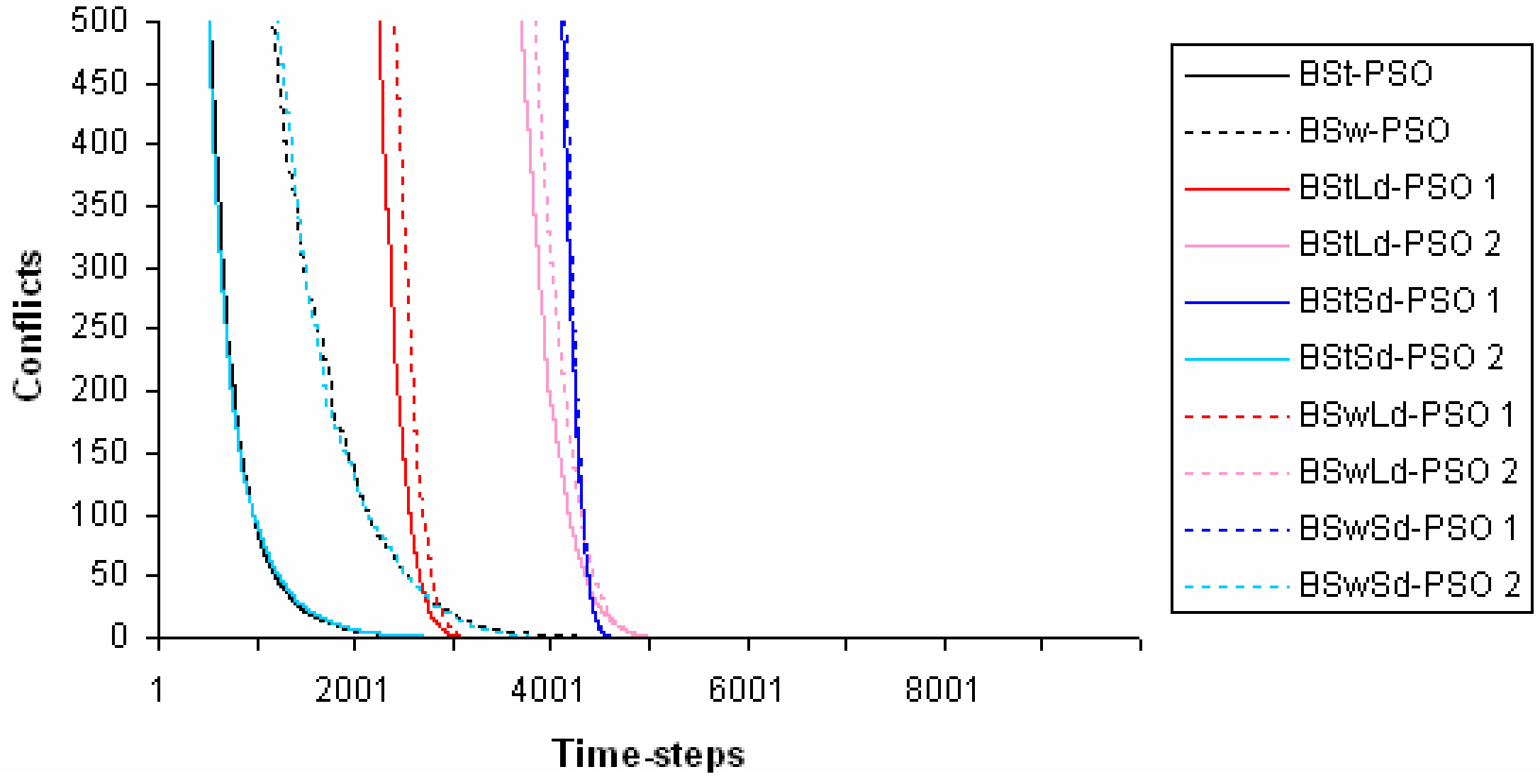

**Fig. 6. 27**: Evolution of the mean best conflicts found by 10 different optimizers for the Sphere benchmark test function out of 50 runs along 10000 time-steps.

Given that the curves of the evolution of the mean average conflicts[5] merge—and flatten—with the corresponding curves of the mean best conflicts for all the optimizers with time-decreasing inertia weights, it can be reasoned that the clustering is so fast that the particles converge to any point, which does not need to be a local optimum. In fact, the Sphere function does not exhibit any local optimum. These curves can be found in digital **Appendix 4**, where the vertical zoom needs to be adjusted to observe the stagnation of the different optimizers.

It can be seen in **Fig. 6. 27** that the optimizers with time-swapping learning weights take longer to approach the optimum than their corresponding standard versions, thus exhibiting explorative behaviour for a longer period of time. However, the benefits of this behaviour cannot be fully appreciated when optimizing these test functions, which do not exhibit numerous global or good local optima that are located far from one another.

The worst mean best conflicts[6] are found by the BSt-PSO, the BSw-PSO, BStSd-PSO 1, and the BSwSd-PSO 1. However, while the first two optimizers—which have constant inertia weights—find it difficult to improve the search because of their lack of clustering ability, the other two optimizers—which have time-decreasing inertia weights—find it difficult to do it because of their strong clustering ability. Therefore, while these four optimizers are the worst with regards to the best solution they are able to find, the first two are the best with regards to their reluctance to cluster around suboptimal solutions, and the other two are the best with regards to their ability to cluster.

Although the BStSd-PSO 2 and the BSwLd-PSO 1 are the optimizers that find the best mean best conflicts ($\overline{cgbest}$), all the other optimizers with time-decreasing inertia weights also find reasonably good mean best conflicts, and also exhibit strong clustering ability. Notice that, while the BStSd-PSO 2 and the BSwSd-PSO 2 display behaviours similar to those of the BSt-PSO and of the BSw-PSO, respectively, during the early stages of the search when the values of their respective inertia weights are similar, their subsequent time-decreasing inertia weights result in noticeable improvement of the clustering ability.

---

[5] The mean average conflicts are the mean among the 50 runs of the average conflicts among the 30 particles.

[6] In order to make the nomenclature clear, it is convenient to remark that the **best conflict** at a given time-step is the best solution found so far by an optimizer along a single search, the **mean best conflict** is the average of the best conflicts found by an optimizer along 50 searches, and the **best** (**worst**) **mean best conflict** refers to the best (worst) of the mean best conflicts found by the 10 optimizers tested. Bear in mind, however, that the **best mean best conflict** might occasionally refer to the mean best conflict found by an optimizer at the final time-step (i.e. once the search is complete), which is undoubtedly the best.

It is important to note that not only do the BStSd-PSO 2 and the BSwLd-PSO 1 find the best mean best conflicts with the smallest corresponding standard deviations, but they also find the best conflicts in any run (*cgbest*). In addition, they are among the three fastest optimizers in attaining the error condition.

### 6.4.1.7.2 ROSENBROCK function

The results of the experiments for the optimization of this function are gathered in **Table 6. 3**:

| ROSENBROCK | $\overline{cgbest}\,(\sigma)$ | $cgbest$ | $\overline{tsec}\,(\sigma)$ | $\overline{tsgb}\,(\sigma)$ | $nf$ | $ngb$ |
|---|---|---|---|---|---|---|
| BSt-PSO | **1.05146453E+02** (1.09351412E+02) | **9.14109398E+00** | 6000.31 (1570.23) | - - | 15 | 0 |
| BSw-PSO | **1.69878269E+02** (4.11626714E+02) | **2.63237875E+01** | 6731.48 (1320.75) | - - | 23 | 0 |
| BStLd-PSO 1 | **3.44681922E+01** (3.04459907E+01) | **4.76697849E-01** | 3697.10 (692.38) | - - | 0 | 0 |
| BStLd-PSO 2 | **3.11244955E+01** (3.17405650E+01) | **8.04585065E-02** | 5921.35 (700.67) | - - | 1 | 0 |
| BStSd-PSO 1 | **5.86494698E+01** (5.29207556E+01) | **2.49195875E+00** | 4976.05 (425.78) | - - | 7 | 0 |
| BStSd-PSO 2 | **3.34077519E+01** (3.34204792E+01) | **2.72631748E-03** | 4546.13 (906.52) | - - | 2 | 0 |
| BSwLd-PSO 1 | **2.97694758E+01** (3.43061953E+01) | **2.69302703E-01** | 4088.35 (1113.73) | - - | 4 | 0 |
| BSwLd-PSO 2 | **3.68520044E+01** (3.80494376E+01) | **1.91562427E-01** | 6125.70 (712.19) | - - | 4 | 0 |
| BSwSd-PSO 1 | **4.31429421E+01** (3.59304777E+01) | **2.17384232E-01** | 5100.74 (904.77) | - - | 3 | 0 |
| BSwSd-PSO 2 | **2.93839226E+01** (3.27118455E+01) | **1.19267816E-02** | 4857.25 (794.88) | - - | 2 | 0 |

**Table 6. 3**: Performance of 10 algorithms when optimizing the 30-dimensional Rosenbrock benchmark test function along 10000 time-steps, where the particles are initially randomly spread over the region $[-30,30]^{30}$, and the statistical data are calculated out of 50 runs.

This function is clearly more difficult to be optimized, and some failures in attaining the error condition now occur. The best mean best conflicts ($\overline{cgbest}$) are found here by the BSwSd-PSO 2 and the BSwLd-PSO 1. However, the best conflict found in any run (*cgbest*) is obtained by the BStSd-PSO 2, while the BStLd-PSO 1 is the fastest—and the only optimizer which exhibits not failure—in attaining the error condition. With regards to the overall performance of all the optimizers, it can be concluded that the best stand-alone optimizers for this function are the BStLd-PSO 1, the BStLd-PSO 2, the BStSd-PSO 2, the BSwLd-PSO 1, and the BSwSd-PSO 2.

As opposed to the optimization of the Sphere function, diversity[7] is maintained here, which can be inferred from the observation of the curves of the evolution of the mean average and of the mean best conflicts in digital **Appendix 4**: when a complete clustering of the particles takes place, the corresponding curves of the evolution of the mean average and of the mean best conflicts merge. This is in agreement with the fact that no stagnation is observed in any of the curves of the evolution of the mean best conflicts found by each optimizer, as shown in **Fig. 6. 28**:

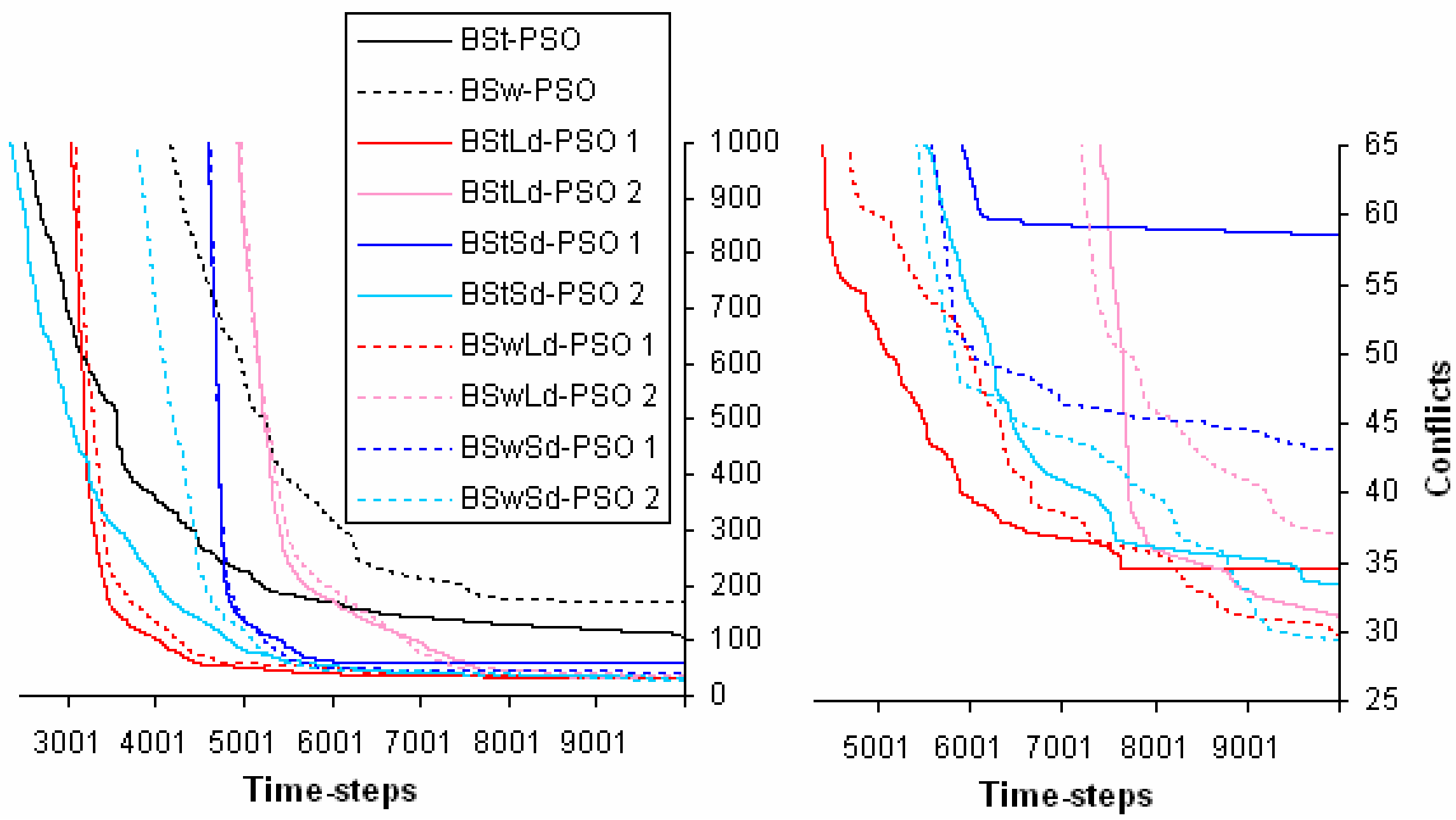

**Fig. 6. 28**: Evolution of the mean best conflicts found by 10 different optimizers for the Rosenbrock benchmark test function out of 50 runs along 10000 time-steps.

It is fair to note that the performances of the optimizers with constant inertia weights are very poor here. With regards to the effect of swapping the relative importance of the learning weights, it appears that it is now beneficial for all the optimizers with time-decreasing inertia weights, except for the BStLd-PSO 2 → BSwLd-PSO 2.

[7] The expression "maintaining diversity" is frequently used in this thesis to mean that the particles do not perform a complete implosion. The expression is imported from the GAs' literature, where it is said that if the chromosomes become exactly equal to one another, diversity of genetic information is lost in the population. When this happens, the genetic pool is reduced to a minimum, and no more improvement is possible by crossover. With regards to the PSO's metaphor, it could be said that when a complete implosion of the particles takes place, all the individuals reach perfect agreement among them, so that diversity of beliefs is lost in the population.

### 6.4.1.7.3 RASTRIGRIN function

The results of the experiments for the optimization of this function are gathered in **Table 6. 4**:

| RASTRIGRIN | $\overline{cgbest}\,(\sigma)$ | $cgbest$ | $\overline{tsec}\,(\sigma)$ | $\overline{tsgb}\,(\sigma)$ | $nf$ | $ngb$ |
|---|---|---|---|---|---|---|
| BSt-PSO | **1.97480492E+01**<br>(6.52306253E+00) | **9.26415334E+00** | 1031.34<br>(435.22) | -<br>- | 0 | 0 |
| BSw-PSO | **2.85174474E+01**<br>(1.08239619E+01) | **7.49858723E+00** | 2829.10<br>(1017.16) | -<br>- | 0 | 0 |
| BStLd-PSO 1 | **2.78190367E+01**<br>(8.81467464E+00) | **9.94959057E+00** | 2587.54<br>(214.15) | -<br>- | 0 | 0 |
| BStLd-PSO 2 | **2.34810282E+01**<br>(6.49798613E+00) | **1.39294218E+01** | 4240.48<br>(270.97) | -<br>- | 0 | 0 |
| BStSd-PSO 1 | **3.39479696E+01**<br>(1.03528581E+01) | **1.69142966E+01** | 4369.60<br>(164.55) | -<br>- | 0 | 0 |
| BStSd-PSO 2 | **2.14712021E+01**<br>(7.83968897E+00) | **1.09445496E+01** | 1147.36<br>(610.84) | -<br>- | 0 | 0 |
| BSwLd-PSO 1 | **2.62943490E+01**<br>(6.66764721E+00) | **1.19395087E+01** | 2834.22<br>(193.31) | -<br>- | 0 | 0 |
| BSwLd-PSO 2 | **2.35208193E+01**<br>(5.80767483E+00) | **1.09445496E+01** | 4341.18<br>(292.84) | -<br>- | 0 | 0 |
| BSwSd-PSO 1 | **3.40673647E+01**<br>(8.51056053E+00) | **1.59193348E+01** | 4349.20<br>(141.59) | -<br>- | 0 | 0 |
| BSwSd-PSO 2 | **2.46550677E+01**<br>(6.56693359E+00) | **1.39294218E+01** | 2802.76<br>(736.30) | -<br>- | 0 | 0 |

**Table 6. 4**: Performance of 10 algorithms when optimizing the 30-dimensional Rastrigrin benchmark test function along 10000 time-steps, where the particles are initially randomly spread over the region $[-5.12, 5.12]^{30}$, and the statistical data are calculated out of 50 runs.

While no optimizer is able to find the exact global optimum, none of them fails in attaining the error condition either. Given that this function tests the ability of the optimizers to escape local optima, and that a time-decreasing inertia weight is advantageous for fine-tuning the search while a constant (not too small) one is advantageous for escaping local optima, it is not surprising that the BSt-PSO finds the best mean best conflict ($\overline{cgbest}$) for this function. It seems that its incapability of fine-clustering makes it more reluctant to get trapped in local optima—and in any other suboptimal solution—than the optimizers which favour fine-clustering. In addition, the BSt-PSO is also the optimizer which requires the smallest mean number of time-steps to attain the error condition stated in **Table 6. 1**. Notice, however, that the best conflict found in any of the 50 runs (*cgbest*) is that obtained by the BSw-PSO, in spite of the fact that it only finds one of the worst mean best conflicts.

The curves of the evolution of the mean best conflict found by each optimizer are shown in **Fig. 6. 29**. While the curves corresponding to all the optimizers with time-decreasing inertia weights almost stagnate, the ones corresponding to the BSt-PSO and the BSw-PSO are still quite steep at the final time-step.

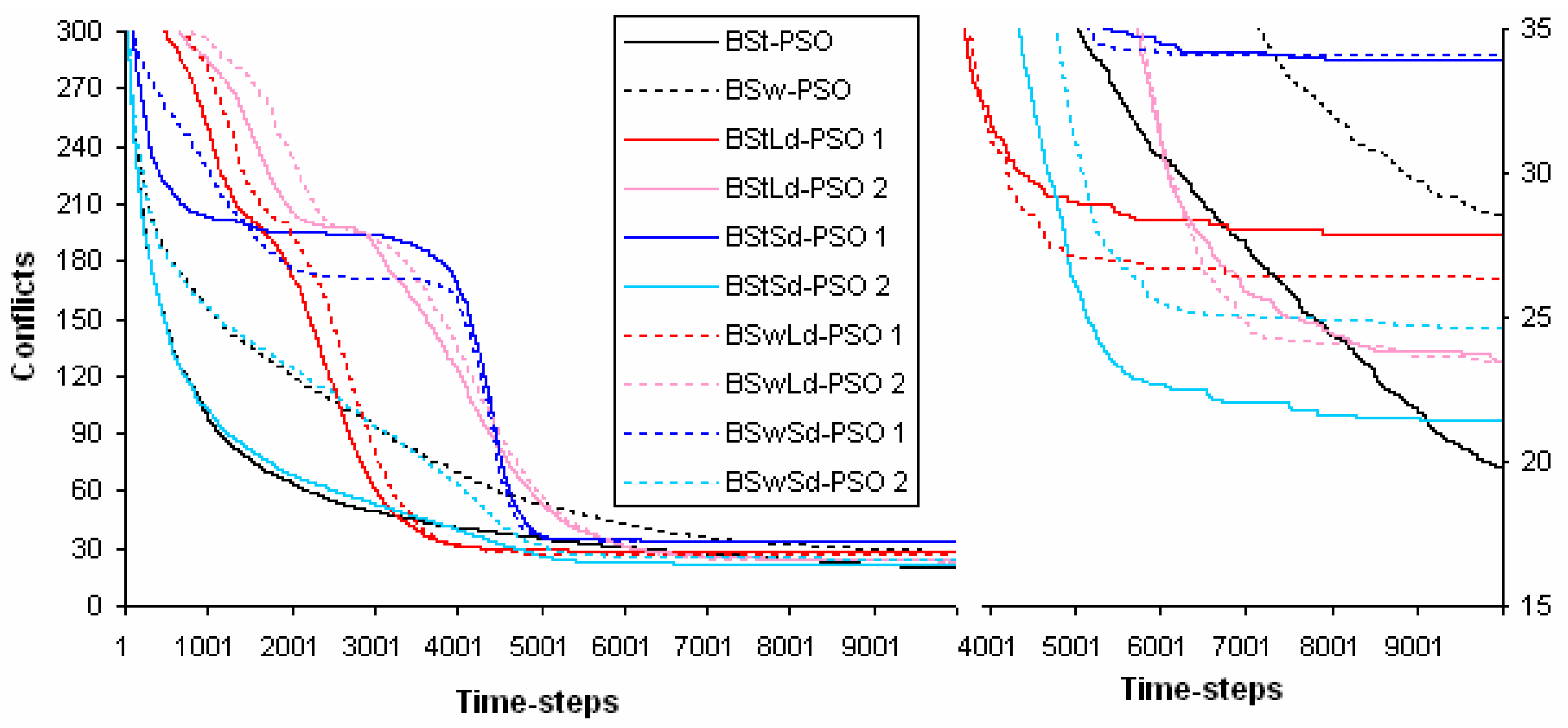

**Fig. 6. 29**: Evolution of the mean best conflicts found by 10 different optimizers for the Rastrigrin benchmark test function out of 50 runs along 10000 time-steps.

It is important to note that the trend-line of the evolution of the mean average conflict of the BStSd-PSO 2—which finds the second best $\overline{cgbest}$—and those of the other optimizers with time-decreasing inertia weights that do not approach zero, do not exhibit stagnation (see **Appendix 4**). This is in accordance with the observation that they find better mean best conflicts than the other four optimizers whose time-decreasing inertia weights do approach zero during the late stages of the search (see **Fig. 6. 29** - right). As expected, it appears that too small inertia weights—which favour exploitation—are not appropriate for stand-alone optimizers which aim to optimize functions with numerous local optima.

It is intriguing to see that no optimizer is able to perform a complete implosion here, so that diversity is not completely lost by any of the optimizers (refer to **Appendix 4**). However, the curves of the evolution of the mean best conflicts of all the optimizers with time-decreasing inertia weights display stagnation.

The trend-lines of the graphs of the evolution of the mean average conflict of the optimizers with constant inertia weights, namely the BSt-PSO and the BSw-PSO, do not approach the global optimum, thus maintaining a very high diversity. This feature seems to be critical when optimizing functions that exhibit numerous local optima.

To sum up, the optimizers with constant inertia weights maintain a very high diversity[8], which seems to result in the non-stagnation of the evolution of their mean best conflicts. While all the other optimizers maintain some (much lower) diversity, it seems to be insufficient to keep on escaping local optima. Thus, the optimizers with time-decreasing inertia weights that do not approach zero seem to be clustering around a local optimum they cannot escape from[9], while the ones whose inertia weights do approach zero seem to be unable to fine-cluster[10].

### 6.4.1.7.4 GRIEWANK function

The results of the experiments for the optimization of this function are gathered in **Table 6. 5**:

| GRIEWANK | $\overline{cgbest}\,(\sigma)$ | $cgbest$ | $\overline{tsec}\,(\sigma)$ | $\overline{tsgb}\,(\sigma)$ | $nf$ | $ngb$ |
|---|---|---|---|---|---|---|
| BSt-PSO | **2.71282371E-02** (2.42179142E-02) | **7.37994110E-11** | 4339.14 (839.07) | - - | 1 | 0 |
| BSw-PSO | **4.33848724E-02** (7.61551617E-02) | **1.05297599E-05** | 5786.73 (834.39) | - - | 5 | 0 |
| BStLd-PSO 1 | **1.42241531E-02** (1.37636316E-02) | **0.00000000E+00** | 3280.00 (81.35) | 4635.93 (69.90) | 0 | 14 |
| BStLd-PSO 2 | **2.32743686E-02** (2.58841965E-02) | **0.00000000E+00** | 5284.29 (124.57) | 7204.18 (88.05) | 2 | 11 |
| BStSd-PSO 1 | **1.57210502E-02** (1.71187144E-02) | **0.00000000E+00** | 4702.06 (43.17) | 6226.82 (1366.43) | 0 | 11 |
| BStSd-PSO 2 | **2.59287671E-02** (2.25989027E-02) | **0.00000000E+00** | 3672.37 (565.83) | 5654.29 (86.76) | 1 | 7 |
| BSwLd-PSO 1 | **1.30433374E-02** (1.42859895E-02) | **0.00000000E+00** | 3359.36 (105.81) | 4643.93 (102.59) | 0 | 14 |
| BSwLd-PSO 2 | **1.70567812E-02** (1.80531614E-02) | **0.00000000E+00** | 5325.28 (118.96) | 7477.38 (269.51) | 0 | 16 |
| BSwSd-PSO 1 | **1.41115095E-02** (1.38002609E-02) | **0.00000000E+00** | 4705.94 (40.54) | 5670.31 (403.51) | 0 | 13 |
| BSwSd-PSO 2 | **2.87758330E-02** (2.61835894E-02) | **0.00000000E+00** | 4345.14 (212.41) | 5896.57 (385.51) | 0 | 7 |

**Table 6. 5**: Performance of 10 algorithms when optimizing the 30-dimensional Griewank benchmark test function along 10000 time-steps, where the particles are initially randomly spread over the region $[-600,600]^{30}$, and the statistical data are calculated out of 50 runs.

8 In fact, the curves of the evolution of their mean average conflicts do not even reach the error condition!

9 The curves of the evolution of their mean best conflicts almost stagnate, while the trend-lines of the graphs of the evolution of their mean average conflicts are still quite steep at the final time-step.

10 The curves of the evolution of their mean best and mean average conflicts almost stagnate without merging.

This function is typically thought of as a noisy sphere-like function. It is surprising to notice that, while this function is supposed to be harder to be optimized than the Sphere function[11], all the optimizers with time-decreasing inertia weights find the exact global optimum at least 7 times, and very few failures in attaining the error condition are observed. The reason for this might be that the noise makes the clustering slightly more difficult, so that higher diversity is maintained for a longer period of time. The evolution of the mean best conflict found by each optimizer is plotted in **Fig. 6. 30**:

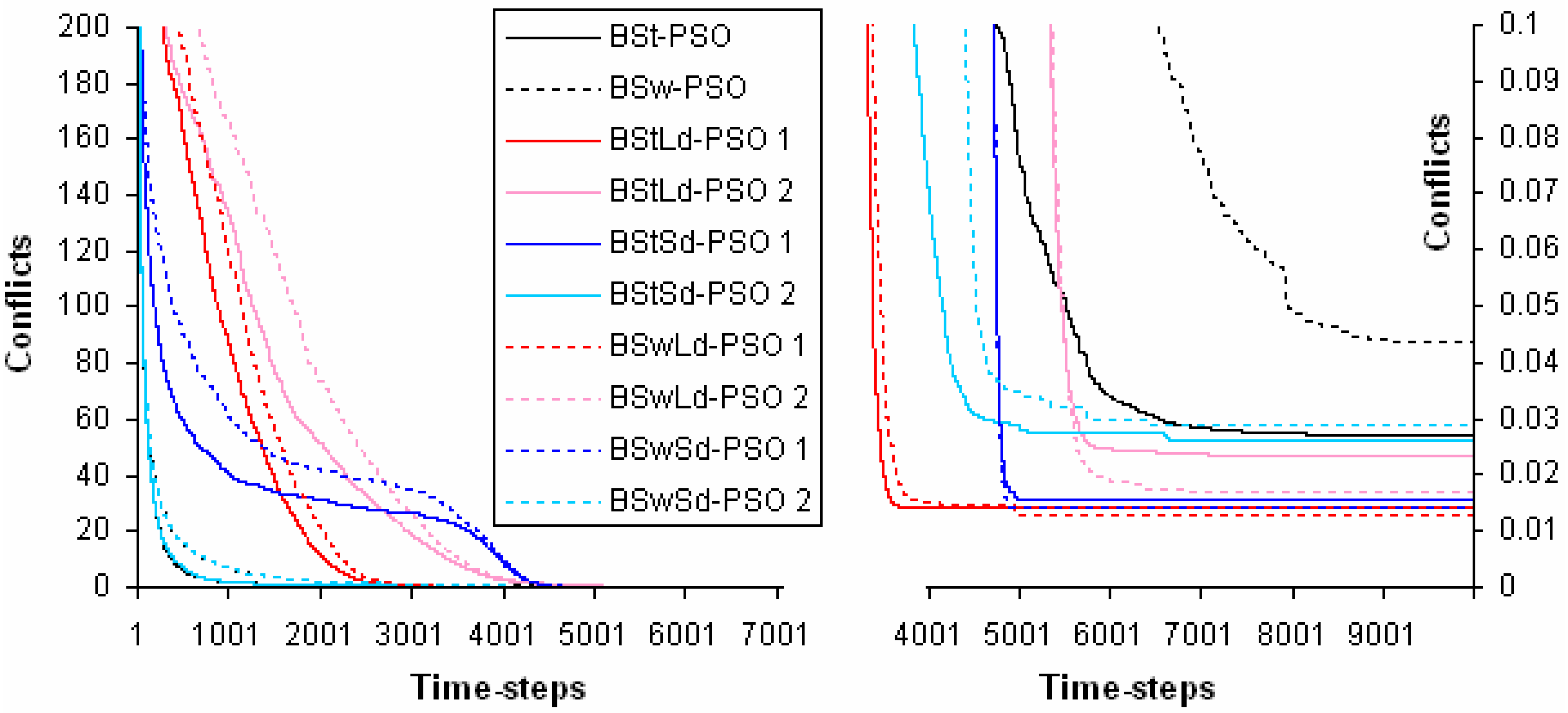

**Fig. 6. 30**: Evolution of the mean best conflicts found by 10 different optimizers for the Griewank benchmark test function out of 50 runs along 10000 time-steps.

Regarding the mean best conflicts found, it is important to note that while inertia weights that approach zero appear to be inconvenient to optimize the Rastrigrin function, they outperform the corresponding inertia weights that do not approach zero to optimize this function. That is to say, the “PSO 1” versions outperform the corresponding “PSO 2” versions here.

The BSwLd-PSO 1 and the BStLd-PSO 1 find two of the three best mean best conflicts. In addition, they are the two fastest optimizers in attaining the error condition, and in finding the exact global optimum. Besides, they are among the three optimizers which manage to find the exact global optimum a higher number of times, and among the optimizers which exhibit no failure in attaining the error condition. Broadly speaking, it can be said that the time-decreasing inertia weights are more convenient for this function than the constant inertia weights, and that the ones which approach zero are more convenient than those which do not.

[11] In fact, the mean best conflicts found by the optimizers here are noticeably worse than those found when optimizing the Sphere function.

Only the BStLd-PSO 1, the BStSd-PSO 1, and the BSwSd-PSO 2 show what seems to be a virtually complete loss of diversity: the corresponding curves of the evolution of their mean average and mean best conflicts merge and stagnate. Although the other optimizers with time-decreasing inertia weights do not exhibit this apparently complete implosion, the curves that lower-bound the graphs of the evolution of their mean average conflicts merge with the corresponding curves of the evolution of their mean best conflicts (refer to **Appendix 4**). Hence it can be inferred that the particles are swarming near the best conflict found so far without being able to improve it either because they are unable to fine-cluster, or because they are unable to carry out further exploration.

The curves of the mean average conflicts found by the BSt-PSO and the BSw-PSO are again very far from the global optimum. However, the trend-line of the graph of the evolution of the mean average conflict found by the former does not stagnate, so that further improvement can be expected for a longer search[12], while the one corresponding to the latter diverges from the global best, so that the search would not be narrowed if the search was time-extended. Bear in mind, however, that while a complete clustering eliminates the possibility of further improvement for a longer search, the incapability of fine-clustering always leaves the door open to find a better solution by chance.

### 6.4.1.7.5 SCHAFFER F6 function (2D)

The results of the experiments for the optimization of this function are gathered in **Table 6. 6**, and the evolution of the mean best conflict found by each optimizer is plotted in **Fig. 6. 31**.

Although all the optimizers find the exact global optimum in every run for this very hard but low-dimensional problem, only the BStLd-PSO 1, the BSwLd-PSO 1, the BSwLd-PSO 2, the BStSd-PSO 2, and the BSwSd-PSO 2, exhibit a complete loss of diversity (see **Appendix 4**): all of their particles implode to the very global optimum. Nonetheless, the curves that lower-bound the graphs of the evolution of the mean average conflicts of the other optimizers with time-decreasing inertia weights reach the exact global optimum by the end of the search.

[12] Note that its weights are not time-dependent.

| SCHAFFER F6 2D | $\overline{cgbest}\,(\sigma)$ | $cgbest$ | $\overline{tsec}\,(\sigma)$ | $\overline{tsgb}\,(\sigma)$ | $nf$ | $ngb$ |
|---|---|---|---|---|---|---|
| BSt-PSO | **0.00000000E+00** (0.00000000E+00) | **0.00000000E+00** | 743.18 (860.92) | 1016.22 (861.99) | 0 | 50 |
| BSw-PSO | **0.00000000E+00** (0.00000000E+00) | **0.00000000E+00** | 659.64 (804.11) | 1071.54 (781.70) | 0 | 50 |
| BStLd-PSO 1 | **0.00000000E+00** (0.00000000E+00) | **0.00000000E+00** | 1383.30 (398.06) | 1897.44 (281.79) | 0 | 50 |
| BStLd-PSO 2 | **0.00000000E+00** (0.00000000E+00) | **0.00000000E+00** | 1815.58 (305.84) | 2686.70 (182.69) | 0 | 50 |
| BStSd-PSO 1 | **0.00000000E+00** (0.00000000E+00) | **0.00000000E+00** | 927.56 (495.26) | 2041.40 (541.02) | 0 | 50 |
| BStSd-PSO 2 | **0.00000000E+00** (0.00000000E+00) | **0.00000000E+00** | 541.00 (571.90) | 804.26 (576.31) | 0 | 50 |
| BSwLd-PSO 1 | **0.00000000E+00** (0.00000000E+00) | **0.00000000E+00** | 1564.14 (263.57) | 2107.22 (175.62) | 0 | 50 |
| BSwLd-PSO 2 | **0.00000000E+00** (0.00000000E+00) | **0.00000000E+00** | 2353.40 (390.50) | 3030.40 (229.16) | 0 | 50 |
| BSwSd-PSO 1 | **0.00000000E+00** (0.00000000E+00) | **0.00000000E+00** | 1411.70 (578.20) | 3123.78 (334.04) | 0 | 50 |
| BSwSd-PSO 2 | **0.00000000E+00** (0.00000000E+00) | **0.00000000E+00** | 489.66 (312.29) | 897.54 (300.95) | 0 | 50 |

**Table 6. 6**: Performance of 10 algorithms when optimizing the 2-dimensional Schaffer f6 benchmark test function along 10000 time-steps, where the particles are initially randomly spread over the region $[-100,100]^2$, and the statistical data are calculated out of 50 runs.

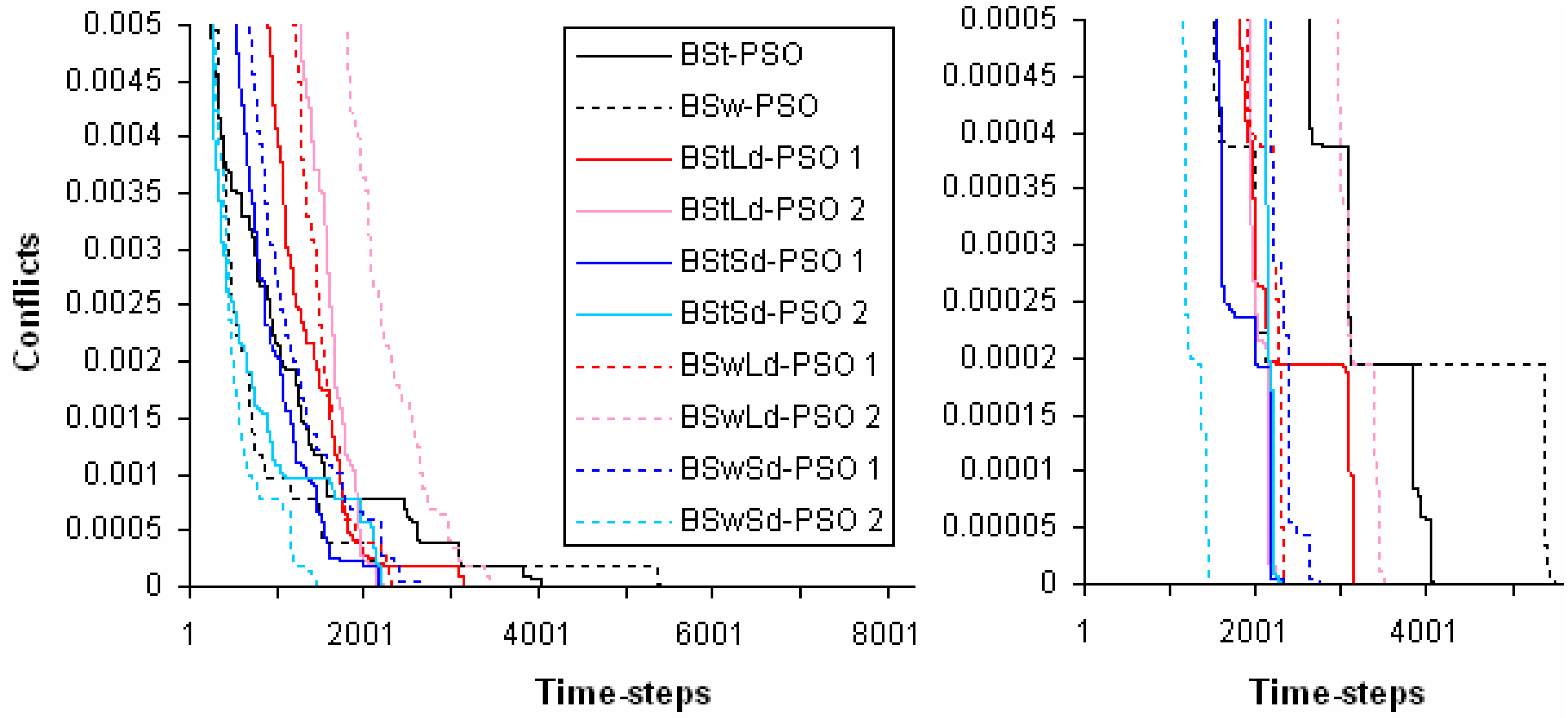

**Fig. 6. 31**: Evolution of the mean best conflicts found by 10 different optimizers for the 2-dimensional Schaffer f6 benchmark test function out of 50 runs along 10000 time-steps.

Note that, despite finding the exact global optimum in each of the 50 runs, the trend-lines of the mean average conflicts of the BSt-PSO and the BSw-PSO are still quite steep and far from the global optimum at the final time-step.

Note that while the BSwSd-PSO 2 and the BStSd-PSO 2 are the fastest optimizers in both attaining the error condition and finding the exact global optimum, the BSt-PSO and the BSw-PSO are also very fast in both, in spite of their incapability of fine-clustering.

### 6.4.1.7.6 SCHAFFER F6 function

The results of the experiments for the optimization of this function are gathered in **Table 6. 7**, and the evolution of the mean best conflict found by each optimizer is plotted in **Fig. 6. 32**.

| SCHAFFER F6 | $\overline{cgbest}\,(\sigma)$ | $cgbest$ | $\overline{tsec}\,(\sigma)$ | $\overline{tsgb}\,(\sigma)$ | $nf$ | $ngb$ |
|---|---|---|---|---|---|---|
| BSt-PSO | **1.52565031E-01**<br>(3.57118391E-02) | **7.81891821E-02** | 8216.50<br>(1133.49) | -<br>- | 48 | 0 |
| BSw-PSO | **1.95993612E-01**<br>(4.74559919E-02) | **7.81891821E-02** | 6757.00<br>- | -<br>- | 49 | 0 |
| BStLd-PSO 1 | **1.05820482E-01**<br>(3.13959893E-02) | **3.72240751E-02** | 5206.29<br>(1549.99) | -<br>- | 26 | 0 |
| BStLd-PSO 2 | **1.02024491E-01**<br>(3.10884136E-02) | **3.72240751E-02** | 6898.04<br>(848.42) | -<br>- | 24 | 0 |
| BStSd-PSO 1 | **1.24613436E-01**<br>(3.39912829E-02) | **7.81891821E-02** | 7147.69<br>(1746.02) | -<br>- | 37 | 0 |
| BStSd-PSO 2 | **1.09567864E-01**<br>(2.95232950E-02) | **7.81891821E-02** | 5820.10<br>(1079.43) | -<br>- | 29 | 0 |
| BSwLd-PSO 1 | **1.12447336E-01**<br>(2.67874782E-02) | **7.81891821E-02** | 5058.71<br>(1188.76) | -<br>- | 33 | 0 |
| BSwLd-PSO 2 | **1.13471971E-01**<br>(2.82926501E-02) | **7.81891821E-02** | 7199.35<br>(948.33) | -<br>- | 33 | 0 |
| BSwSd-PSO 1 | **1.24477706E-01**<br>(3.08892567E-02) | **7.81891821E-02** | 7366.36<br>(1484.79) | -<br>- | 39 | 0 |
| BSwSd-PSO 2 | **1.14653332E-01**<br>(3.08617869E-02) | **3.72240751E-02** | 6623.63<br>(1328.80) | -<br>- | 34 | 0 |

**Table 6. 7**: Performance of 10 algorithms when optimizing the 30-dimensional Schaffer f6 benchmark test function along 10000 time-steps, where the particles are initially randomly spread over the region $[-100,100]^{30}$, and the statistical data are calculated out of 50 runs.

No optimizer is able to find the exact global optimum, and all of them present numerous failures in attaining the demanding error condition set here (refer to **Table 6. 1**).

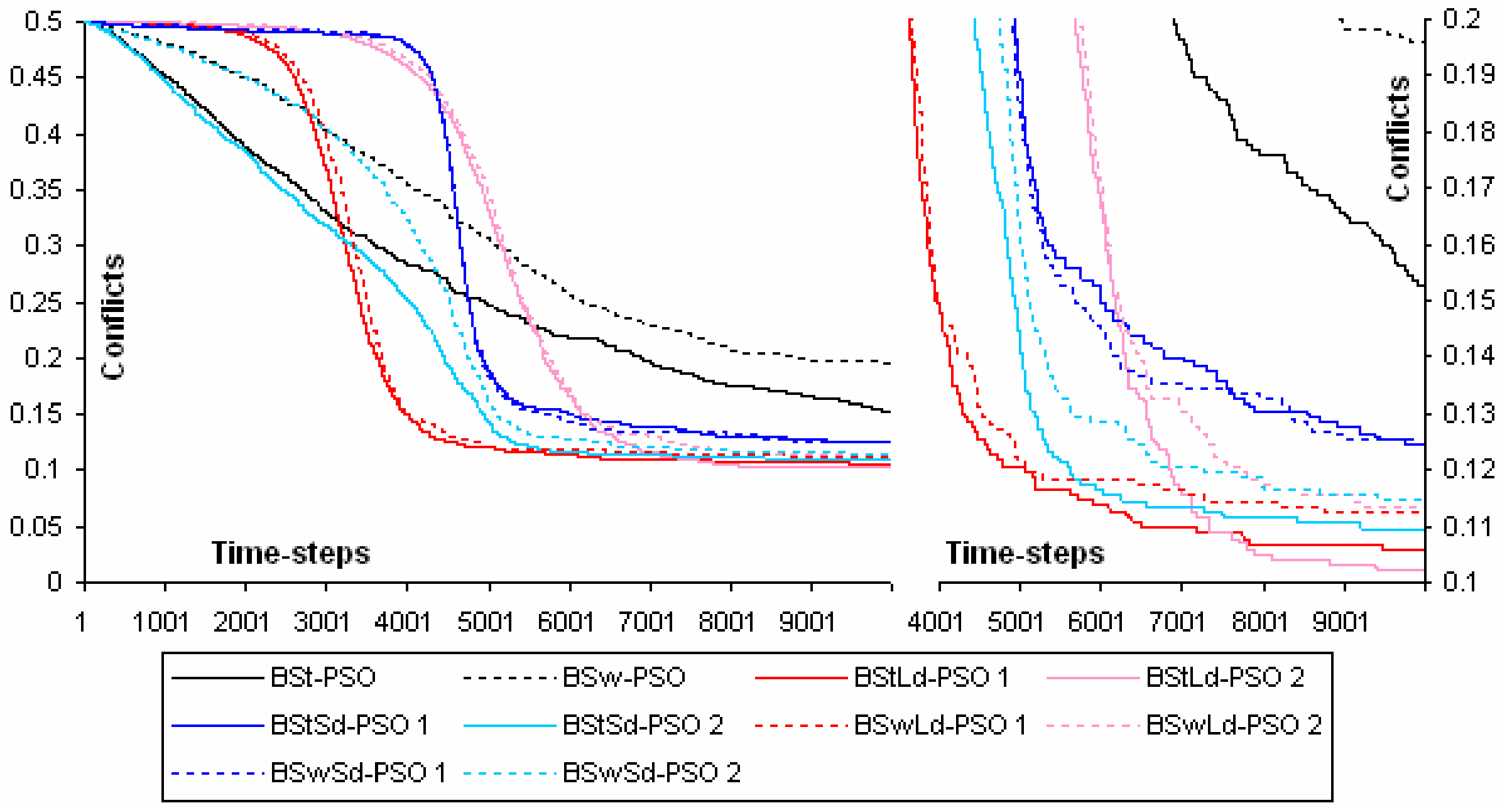

**Fig. 6. 32**: Evolution of the mean best conflicts found by 10 different optimizers for the Schaffer f6 benchmark test function out of 50 runs along 10000 time-steps.

Similar to the case of optimizing the Rastrigrin function—which also exhibits numerous local optima—the curves of the evolution of the mean best conflicts found by the BSt-PSO and the BSw-PSO do not stagnate here, in spite of the fact that they find the worst mean best conflicts after 10000 time-steps. However, further improvement can be expected for a longer search.

All the optimizers with time-decreasing inertia weights are able to find better mean best conflicts than the others, although the rate of improvement dramatically decrease during the late stages of the search. However, no optimizer completely stagnates.

It is fair to note that the swapping strategy appears to be harmful in every case[13], and that the optimizers whose inertia weights do not approach zero are able to escape more local optima than the corresponding optimizers with inertia weights that do approach zero, so that they find better mean best conflicts. In other words, the "St" versions outperform the corresponding "Sw" versions, and the "PSO 2" versions outperform the corresponding "PSO 1" versions when optimizing this function. It can also be noticed that the optimizers with linearly time-decreasing inertia weights outperform those with sigmoidly time-decreasing ones.

[13] Note that the swapping strategy enhances sociality during the late stages of the search thus favouring fine-tuning, but also decreasing the optimizers' capability of escaping local optima. It is reasonable to think that the swapping strategy is convenient to escape wide-spread local optima, but perhaps harmful to escape local optima which are located near one another, like in the case of the Schaffer f6 function.

The graphs of the evolution of the optimizers' mean average conflicts are shown in **Fig. 6. 33**:

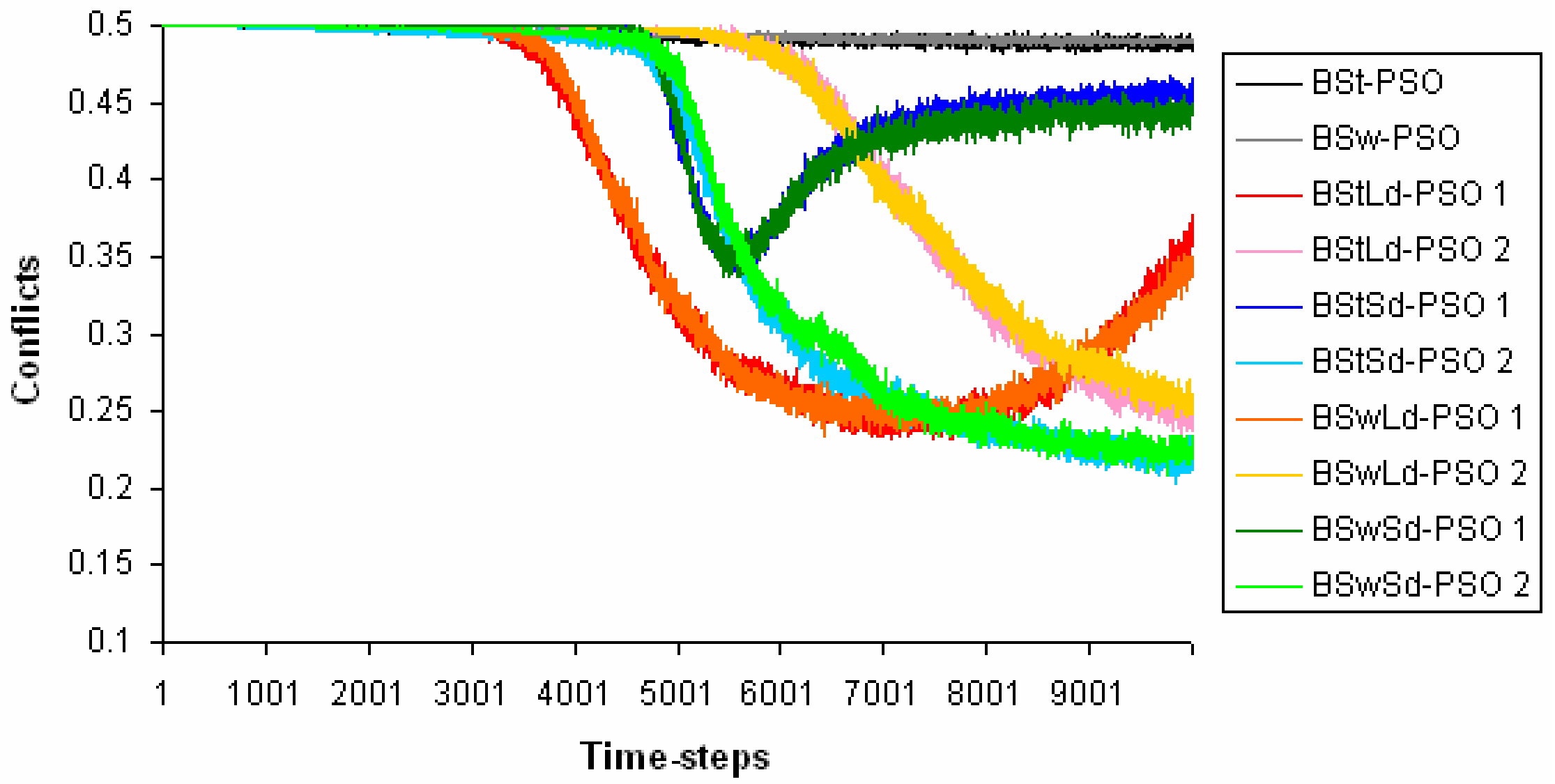

**Fig. 6. 33**: Evolution of the mean average conflicts found by 10 different optimizers for the Schaffer f6 benchmark test function along 10000 time-steps, where the average is calculated out of 30 particles and the mean is calculated out of 50 runs.

Note that the curves corresponding to the BSt-PSO and to the BSw-PSO oscillate near the value of 0.5, so that the trend-lines of their mean average conflicts are nearly horizontal, without approaching the global optimum. It is fair to note that although the BStLd-PSO 1 finds the second best mean best conflict, the trend-line of the evolution of its mean average conflict diverges from the global optimum during the late stages of the search. The same is true for all the other optimizers with inertia weights that approach zero. In contrast, the trend-lines corresponding to the remaining optimizers still do not display stagnation at the final time-step, so that the search can be expected to further narrow.

**Fig. 6. 32** and **Fig. 6. 33** show that the best mean best conflicts found are worth values between 0.1 and 0.2, while the best mean average conflicts found are worth values between 0.2 and 0.5. Hence it can be inferred that no optimizer achieves a complete implosion of its particles.

## 6.4.2 Acceleration weight related to the inertia weight

Not only are the individuality and sociality weights commonly set equal to one another, but they are also commonly set equal to a value of around 2, without further analysis. Originally, Kennedy et al. [46] suggested setting $iw = sw = 2$, so that the multiplication of each of them by the corresponding random weight was, on average, equal to 1. This is reasonable because it makes the part of the trajectory of a particle that is induced by the attractors[14], on average, cyclic. This is sketched in **Fig. 6. 34**, where the random weights are replaced by the average of the uniform distribution used to generate them. That is, the weight $U_{(0,1)}$ in the velocities' updating rule of the O-PSO shown in equation **(6. 4)** is replaced by the average $\overline{U}_{(0,1)} = 0.5$:

$$v_{ij}^{(t)} = v_{ij}^{(t-1)} + \left(pbest_{ij}^{(t-1)} - x_{ij}^{(t-1)}\right) + \left(gbest_{j}^{(t-1)} - x_{ij}^{(t-1)}\right) \quad \textbf{(6. 14)}$$

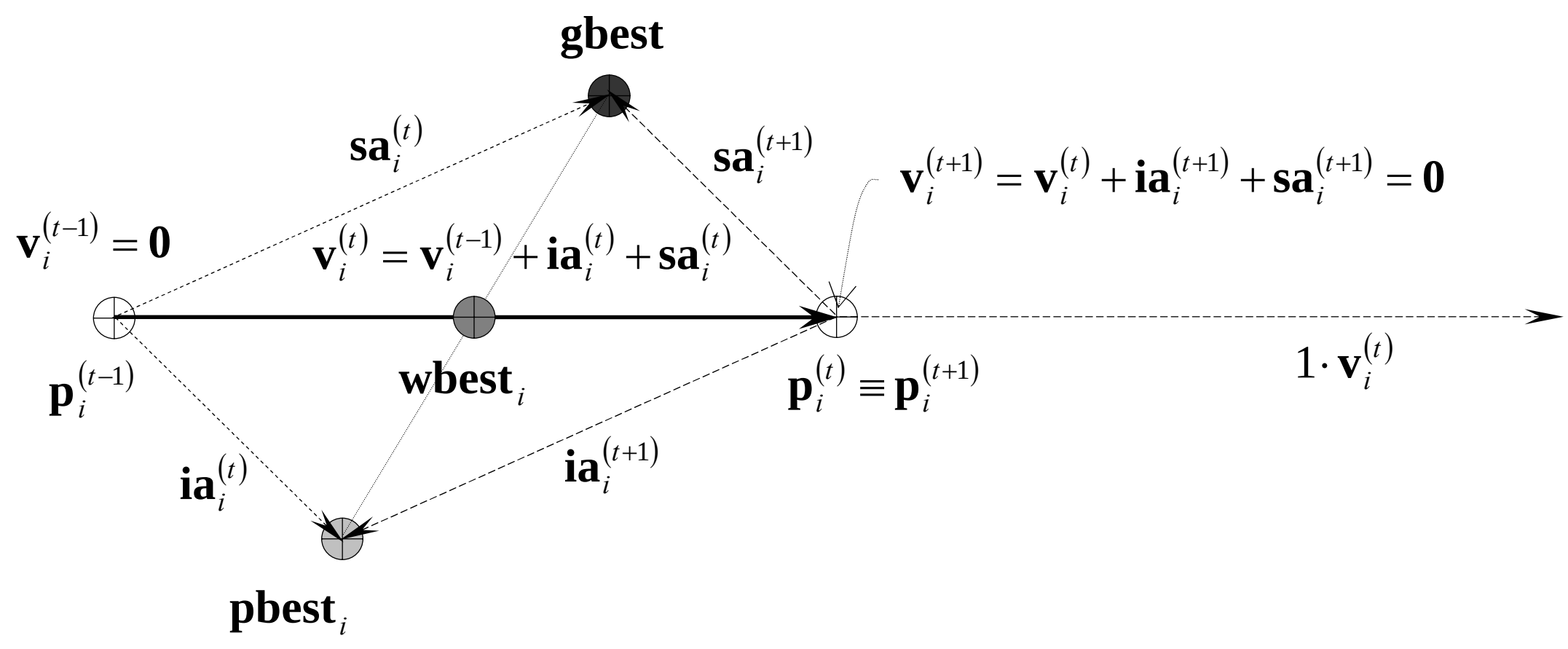

**Fig. 6. 34**: Sketch of the trajectory of a particle $i$, which is attracted towards the points $\mathbf{gbest}$ and $\mathbf{pbest}_i$, corresponding to an optimizer with $w = 1$, $iw = sw = 2$, replacing the random weights $U_{(0,1)}$ in the velocities' updating rule by $\overline{U}_{(0,1)} = 0.5$, and forcing $\mathbf{v}_i^{(t-1)} = \mathbf{0}$. Therefore, this trajectory is in reality the part of the complete trajectory of a generic particle that is induced by the attractors at time-step $(t-1)$. In other words, the particle's trajectory is influenced by the inertia it has at time-step $(t-1)$—omitted in this analysis—which alters this behaviour.

Where:

- $\mathbf{gbest}$ : position of the best solution found so far by any particle in the swarm
- $\mathbf{pbest}_i$ : position of the best solution found so far by particle $i$
- $\mathbf{wbest}_i = \dfrac{2 \cdot 0.5 \cdot \mathbf{pbest}_i + 2 \cdot 0.5 \cdot \mathbf{gbest}}{2 \cdot 0.5 + 2 \cdot 0.5} = \dfrac{\mathbf{pbest}_i + \mathbf{gbest}}{2}$
- $\mathbf{ia}_i^{(t)}$ : individual acceleration of particle $i$ at time-step $t$; here: $\mathbf{ia}_i^{(t)} = \mathbf{pbest}_i - \mathbf{p}_i^{(t-1)}$
- $\mathbf{sa}_i^{(t)}$ : social acceleration of particle $i$ at time-step $t$; here: $\mathbf{sa}_i^{(t)} = \mathbf{gbest} - \mathbf{p}_i^{(t-1)}$

[14] The attractors are the locations of the best solution found by the particle at hand, and of the best solution found by any particle in the swarm, up to the last time-step.

Beware that while **Fig. 6. 34** shows the part of the trajectory that is induced by the attractors only, the existence of a $\mathbf{v}_i^{(t-1)} \neq \mathbf{0}$ must be added for the real trajectory. In addition, the random weights, and the fact that the attractors are non-stationary, alter this simple behaviour. Notice that while this induced cyclic trajectory tends to keep the particle $i$ over-flying the region where the $\mathbf{wbest}_i$ is located, it also makes it difficult to fine-tune the search.

When Shi et al. [70] incorporated the inertia weight to the algorithm, thus giving birth to the B-PSO, the common setting $iw = sw \cong 2$ was traditionally kept, even for time-varying inertia weights. Therefore, when the particles' inertia weights approach zero, as in the BStLd-PSO 1, the BSwLd-PSO 1, the BStSd-PSO 1, and the BSwSd-PSO 1, the effect of the attractors over the particles' trajectories is, on average, similar to the one shown in **Fig. 6. 35**:

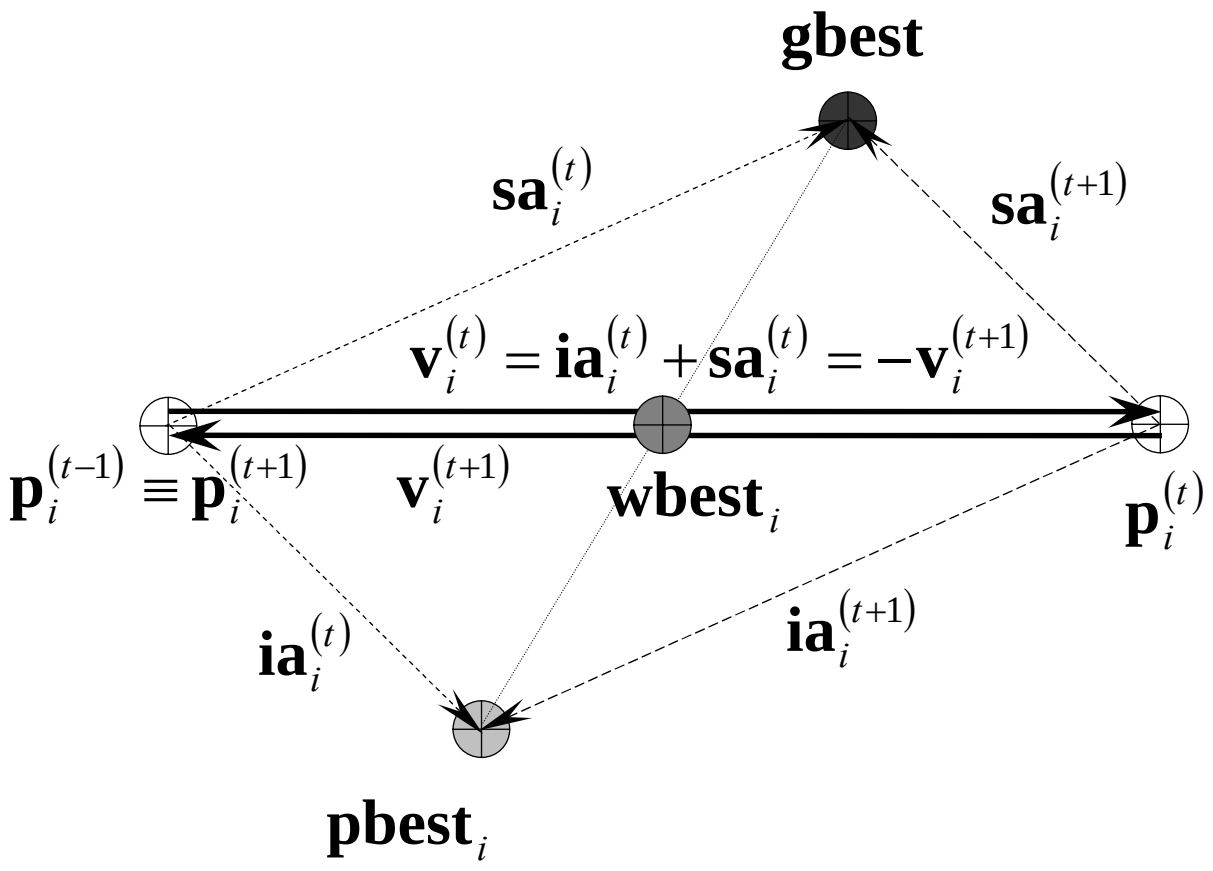

**Fig. 6. 35**: Sketch of the trajectory of a particle $i$, which is attracted towards the points $\mathbf{gbest}$ and $\mathbf{pbest}_i$, corresponding to an optimizer with $w = 0$, $iw = sw = 2$, and replacing the random weights $U_{(0,1)}$ in the velocities' updating rule by the average $\overline{U}_{(0,1)} = 0.5$.

This effect maintains the particles in the neighbourhood of the weighted average of the two attractors (**wbest**), while the random weights enable the particles to vary the step sizes so as to move farther from and closer to **wbest**. Notice that now the part of the trajectory induced by the attractors and the real trajectory are the same, since the inertia was eliminated from the updating rule ($w = 0$).

The experiments run in section **6.4.1** show that the incorporation of a time-decreasing inertia weight $1 \geq w^{(t)} \geq 0$ helps to fine-tune the search. Beware that while both $w$ = 1 and $w$ = 0 lead to average cyclic behaviour, as shown in **Fig. 6. 34** and **Fig. 6. 35**, intermediate settings would help to better exploit the region in between. In fact, if $0 \le w \le 0.5$, $\mathbf{p}_i^{(t+1)}$ would be located between the points $\mathbf{p}_i^{(t-1)}$ and $\mathbf{wbest}_i$, whereas if $0.5 \le w \le 1$, $\mathbf{p}_i^{(t+1)}$ would be located between the points $\mathbf{wbest}_i$ and $\mathbf{p}_i^{(t)}$. The particular case of $w = 0.5$ is sketched in **Fig. 6. 36**:

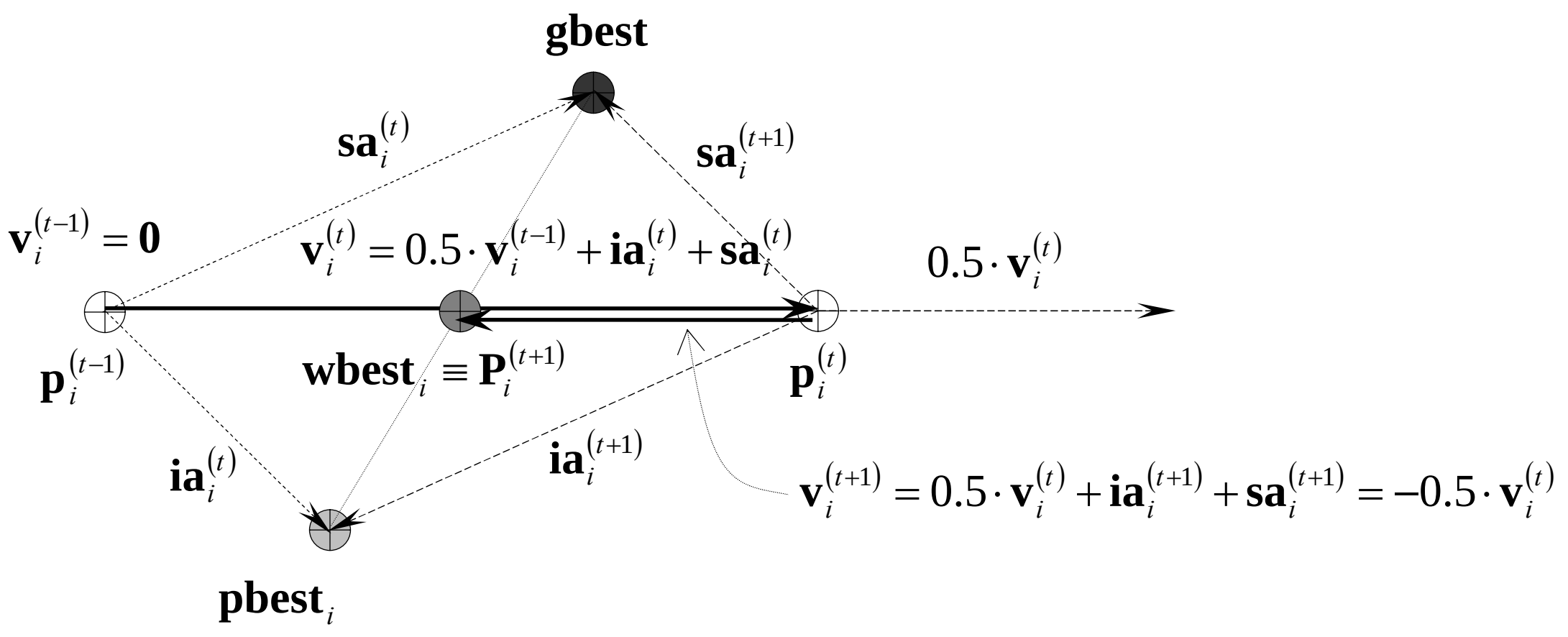

**Fig. 6. 36**: Sketch of the trajectory of a particle $i$, which is attracted towards the points $\mathbf{gbest}$ and $\mathbf{pbest}_i$, corresponding to an optimizer with $w = 0.5$, $iw = sw = 2$, replacing the random weights $U_{(0,1)}$ in the velocities' updating rule by the average $\overline{U}_{(0,1)} = 0.5$, and forcing $\mathbf{v}_i^{(t-1)} = \mathbf{0}$. Therefore, this trajectory is in reality the part of the complete trajectory of a generic particle that is induced by the attractors at time-step $(t-1)$. In other words, the particle's velocity has some inertia at time-step $(t-1)$—omitted in this analysis—which alters this behaviour.

Therefore, this simple analysis of a deterministic PSO suggests that a time-varying inertia weight in the range [0,1] could be a good strategy, which would alter the relative importance between the inertia weight and the acceleration weight during the search.

While higher inertia weights favour wandering around rather than clustering—thus enabling the particles to exhibit more explorative behaviour—, lower inertia weights favour faster changes of direction towards the attractors. It seems reasonable to theorize that increasing the relative importance of the acceleration weight over the inertia weight through time is a good strategy. In fact, this strategy was successfully implemented in previous experiments.

However, the incorporation of the randomness into the deterministic PSO noticeably changes the behaviours sketched in **Fig. 6. 34** to **Fig. 6. 36**. In fact, while the graphical analysis of the average behaviour of the O-PSO shown in **Fig. 6. 34** suggests that the trajectory should be somehow cyclic, the particles diverge, enlarging instead of narrowing the search.

It is reasoned here that the greater the learning weights, the greater the steps that the random weights might generate altering the deterministic trajectories analyzed with $\overline{U}_{(0,1)} = 0.5$. While too small values of the learning weights wipe out the influence of the attractors, too high values—even if the trajectory of the non-random particles is cyclic—lead the particles to continuously and chaotically change directions, exhibiting trajectories that sometimes look very much like a random search. It seems that, as opposed to the deterministic PSO, it is not only the relative importance of the inertia weight and the acceleration weight but also the value itself of the latter which affect the particles' trajectories in the probabilistic PSO.

Previous experiments show that time-decreasing inertia weights are successful in improving exploitation. Since time-decreasing learning weights decrease the influence of the randomness on the particles' trajectories, it might help to further fine-tune the search. Notice that moving from $w = 1$ to $w = 0$ results in moving from cyclic behaviour to cyclic behaviour!

While numerous experiments are run in the literature for time-decreasing inertia weights, the learning weights are usually kept constant. Although some experiments were already run with time-varying learning weights by means of the "swapping strategy" in section **6.4.1**, the acceleration weight was still kept constant. Since the time-decreasing inertia weight appears to be successful, it is wondered here whether there is a relationship between the inertia and the acceleration weights that is convenient to maintain.

### 6.4.2.1 Constant relationship

Clerc et al. [16] carried out a deterministic analysis of the trajectory of a single non-random particle, aiming to solve the problem of the explosion without the need of constraining the components of the particles' velocities. They incorporated a constriction factor ($\chi$) to the O-PSO, thus giving birth to the C-PSO (see section **5.6.4.1**).

The velocities are updated according to equations **(5. 12)** and **(5. 13)**, which are rewritten hereafter for convenience:

$$\chi = \begin{cases} \dfrac{2 \cdot \kappa}{\left|(iw + sw) - 2 + \sqrt{(iw + sw)^2 - 4 \cdot (iw + sw)}\right|} & \text{if } (iw + sw) > 4 \\ \sqrt{\kappa} & \text{otherwise} \end{cases} \qquad \textbf{(6. 15)}$$

$$v_{ij}^{(t)} = \chi \cdot \left(v_{ij}^{(t-1)} + iw \cdot U_{(0,1)} \cdot \left(pbest_{ij}^{(t-1)} - x_{ij}^{(t-1)}\right) + sw \cdot U_{(0,1)} \cdot \left(gbest_{j}^{(t-1)} - x_{ij}^{(t-1)}\right)\right) \quad \textbf{(6. 16)}$$

Where $\chi$ is the constriction factor, and $0 < \kappa \leq 1$.

Kenndey et al. [47] suggest setting $\kappa = 1$ and $aw = 4.1$. Beware that if the C-PSO with these settings is translated to the B-PSO, the settings would be: $w \cong 0.7298$ and $iw = sw \cong 1.49609$.

Carlisle et al. [13] studied different magnitudes of the acceleration weight ($aw$) for the first 5 benchmark functions included in the test suite shown in **Table 6. 1**. They concluded that the setting $aw = 4.1$ is the best choice for a general-purpose C-PSO.

Notice that, while a B-PSO with a time-varying inertia weight alters the relationship between the inertia and the acceleration weights, the C-PSO keeps the same relationship along the whole run (typically, $aw = 4.1 \cdot w$). Although a B-PSO with a constant inertia weight such as the BSt-PSO also maintains the relationship constant, the experiments run along section **6.4.1** showed that the use of time-decreasing inertia weights generally leads to better results[15].

Eberhart et al. [29] compared the performance of the B-PSO to that of the C-PSO, concluding that the best strategy consists of using the C-PSO with $v_{\max} = 0.5 \cdot \left(x_{\max} - x_{\min}\right)$.

It is reasoned here that, while the C-PSO is claimed to ensure convergence towards a local optimum [14, 15, 16], a time-decreasing constriction factor could gather together the advantages of ensuring convergence, and of subsequently reducing both the power of the inertia effect and the sizes of the steps that might take place due to the random weights.

Note that implementing a B-PSO with a time decreasing inertia weight with $w_{\max} = 0.7298$ and keeping the relationship $aw = 4.1 \cdot w$ constant is equivalent to implementing a C-PSO with time-decreasing $\chi$. In other words, with time-decreasing $\kappa$.

Therefore, the same optimizers tested in section **6.4.1** are tested hereafter, but now keeping the learning weights related to the inertia weight like $aw = 4.1 \cdot w$. The BSt-PSO is slightly modified so as to become the C-PSO with $\kappa = 1 \ \Rightarrow \ \chi = 0.7298$, as suggested by Kenndey et al. [47]. Hence the inertia weight for the BSt-PSO here is $w = 0.7298$ instead of $w = 0.7$, as it was in section **6.4.1**.

[15] It is fair to note that only stationary optimization problems are dealt with in this dissertation. At first glance, time-decreasing weights do not seem to be convenient for dynamic problems!

The evolution of the inertia, individuality, and sociality weights for the BSwLd-PSO 1 and the BSwSd-PSO 1—keeping the relationship $aw^{(t)} = 4.1 \cdot w^{(t)} \ \forall t$—is shown in **Fig. 6. 37**:

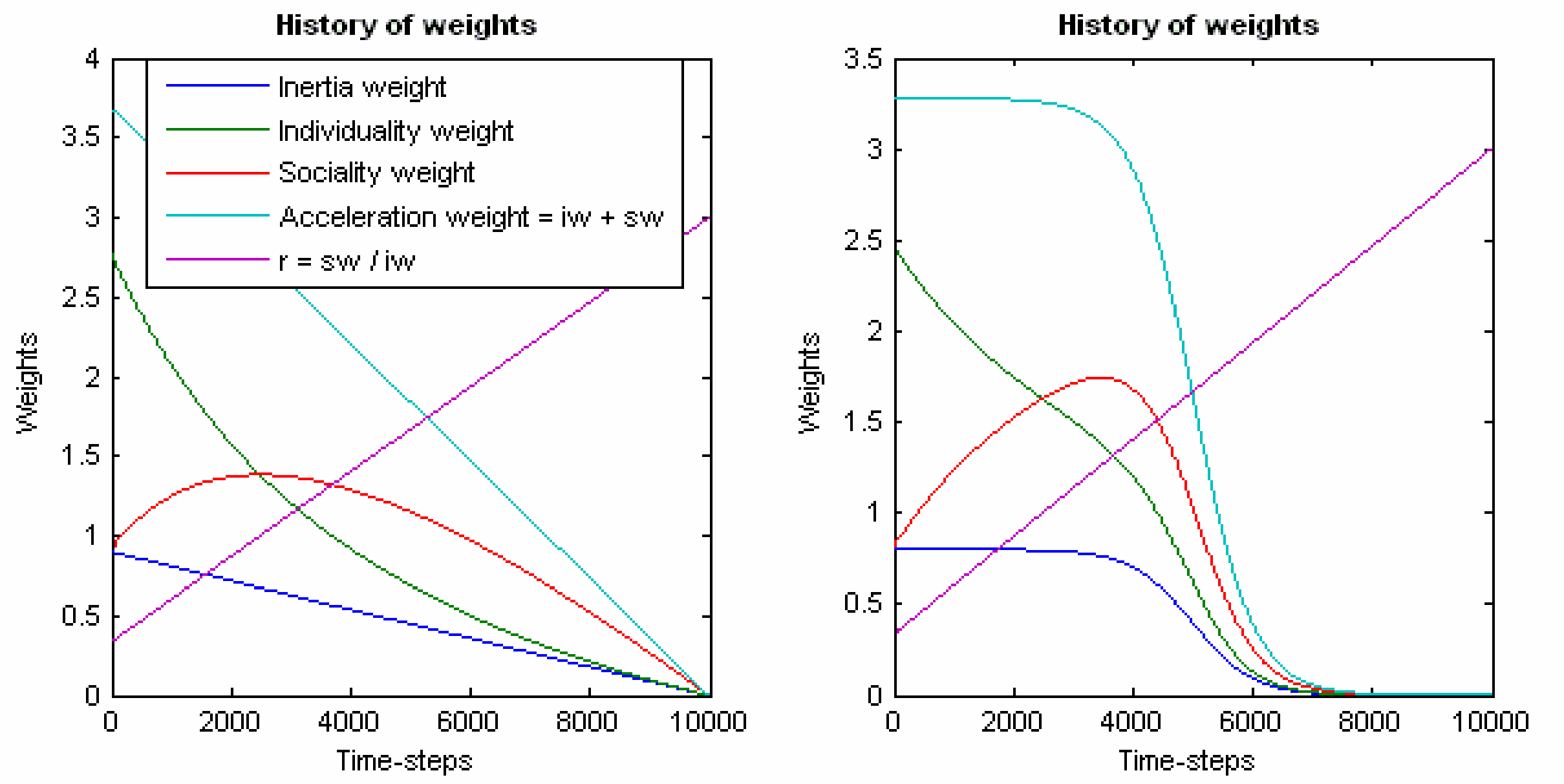

**Fig. 6. 37**: Evolution of the weights for the BSwLd-PSO 1 (left) and the BSwSd-PSO 1 (right), keeping the relationship $aw^{(t)} = 4.1 \cdot w^{(t)} \ \forall t$.

It must be remarked that the stages where the inertia weights of the optimizers tested here are greater than 0.7298 are not equivalent to the C-PSO. During those time-steps, the swarm appears to exhibit less pronounced clustering and more explorative behaviour.

The details of the optimizers tested within this section are as follows:

- **BSt-PSO**: basic, standard PSO:
  $w^{(t)} = 0.7298, \ iw^{(t)} = sw^{(t)} = 2.05 \cdot w^{(t)} \ \forall t$ (it is equivalent to a CSt-PSO)
- **BSw-PSO**: basic PSO with linearly time-swapping learning weights:
  $w^{(t)} = 0.7298 \ \forall t, \ r^{(1)} = \frac{sw^{(1)}}{iw^{(1)}} = \frac{1}{3}, \ r^{(10000)} = \frac{sw^{(10000)}}{iw^{(10000)}} = 3,$
  $aw^{(t)} = iw^{(t)} + sw^{(t)} = 4.1 \cdot w^{(t)} \ \forall t$ (it is equivalent to a CSw-PSO)
- BStLd-PSO 1: basic, standard PSO with linearly time-decreasing inertia weight:
  $w^{(1)} = 0.9, \ w^{(10000)} = 0, \ iw^{(t)} = sw^{(t)} = 2.05 \cdot w^{(t)} \ \forall t$
- BStLd-PSO 2: basic, standard PSO with linearly time-decreasing inertia weight:
  $w^{(1)} = 0.9, \ w^{(10000)} = 0.4, \ iw^{(t)} = sw^{(t)} = 2.05 \cdot w^{(t)} \ \forall t$
- BStSd-PSO 1: basic, standard PSO with sigmoidly time-decreasing inertia weight:
  $w^{(1)} \to 0.8, \ w^{(10000)} \to 0, \ iw^{(t)} = sw^{(t)} = 2.05 \cdot w^{(t)} \ \forall t$

- **BStSd-PSO 2**: basic, standard PSO with sigmoidly time-decreasing inertia weight: $w^{(1)} \to 0.7$, $w^{(10000)} \to 0.4$, $iw^{(t)} = sw^{(t)} = 2.05 \cdot w^{(t)}\ \forall t$ (it is equivalent to a CStSd-PSO 2)
- BSwLd-PSO 1: basic PSO with linearly time-swapping learning weights and linearly time-decreasing inertia weight:
$$w^{(1)} = 0.9,\ w^{(10000)} = 0,\ r^{(1)} = \frac{1}{3},\ r^{(10000)} = 3,\ aw^{(t)} = 4.1 \cdot w^{(t)}\ \forall t$$
- BSwLd-PSO 2: basic PSO with linearly time-swapping learning weights and linearly time-decreasing inertia weight:
$$w^{(1)} = 0.9,\ w^{(10000)} = 0.4,\ r^{(1)} = \frac{1}{3},\ r^{(10000)} = 3,\ aw^{(t)} = 4.1 \cdot w^{(t)}\ \forall t$$
- BSwSd-PSO 1: basic PSO with linearly time-swapping learning weights and sigmoidly time-decreasing inertia weight:
$$w^{(1)} = 0.8,\ w^{(10000)} = 0,\ r^{(1)} = \frac{1}{3},\ r^{(10000)} = 3,\ aw^{(t)} = 4.1 \cdot w^{(t)}\ \forall t$$
- **BSwSd-PSO 2**: basic PSO with linearly time-swapping learning weights and sigmoidly time-decreasing inertia weight: $w^{(1)} = 0.7$, $w^{(10000)} = 0.4$, $r^{(1)} = \frac{1}{3}$, $r^{(10000)} = 3$, $aw^{(t)} = 4.1 \cdot w^{(t)}\ \forall t$ (it is equivalent to a CSwSd-PSO 2)

Note that while the optimizers in bold are pure versions of the C-PSO, the other optimizers start behaving as a C-PSO only once $w^{(t)} \leq 0.7298$.

The performance of each of the 10 optimizers on each of the 6 benchmark test functions is shown in **Table 6. 8** to **Table 6. 13**, and in **Fig. 6. 38** to **Fig. 6. 45**.

### 6.4.2.1.1 SPHERE function

The results of the experiments for the optimization of this function are gathered in **Table 6. 8**, and the evolution of the mean best solution found by each optimizer is plotted in **Fig. 6. 38**.

It is very important to remark that this is a preliminary analysis of the behaviour of the system, which does not intend to develop a “pure strategy” for a general-purpose algorithm. Therefore, no strategy must be discarded just because it is only able to find a very bad mean best conflict ($\overline{cgbest}$), but the reasons that lead to that must be analyzed.

It is interesting to observe that, although the BSt-PSO finds a remarkable $\overline{cgbest}$, once again the exact global optimum could not be found by any optimizer. Besides, while the BSt-PSO and the BSw-PSO have the same acceleration weight, it might be surprising to see that the latter finds a much worse $\overline{cgbest}$ than the former, in spite of the fact that Clerc et al.'s development of the C-PSO [16] relates the acceleration and the inertia weights regardless of the relationship between the individuality and the sociality weights. However, the results are coherent, since the worse $\overline{cgbest}$ found by the BSw-PSO is due to the premature convergence of its particles, as it can be concluded from a visual analysis of **Fig. 6. 38** and **Fig. 6. 39**: the curves of the evolution of the mean best and of the mean average conflicts merge and later stagnate, suggesting that the particles have virtually imploded to a single point. However, it is interesting to note that they still manage to find better solutions for this simple function before stagnating thanks to the inertia weight, but behaving very much like a single particle.

This strategy favours clustering regardless of the topography of the objective function: all the particles fine-cluster around any point—which does not even need to be a local optimum—, noticeably faster than in the experiments run for unrelated inertia and acceleration weights.

| SPHERE | $\overline{cgbest}\ (\sigma)$ | $cgbest$ | $\overline{tsec}\ (\sigma)$ | $\overline{tsgb}\ (\sigma)$ | $nf$ | $ngb$ |
|---|---|---|---|---|---|---|
| **BSt-PSO** | **2.41775603E-157** | **3.14221460E-172** | 363.16 | - | 0 | 0 |
| | (0.00000000E+00) | | (34.38) | - | | |
| **BSw-PSO** | **8.33040094E-02** | **4.42664573E-124** | 1732.48 | - | 19 | 0 |
| | (2.17756204E-01) | | (1693.55) | - | | |
| BStLd-PSO 1 | **5.19778383E-13** | **2.61686502E-20** | 1702.12 | - | 0 | 0 |
| | (2.10471717E-12) | | (37.07) | - | | |
| BStLd-PSO 2 | **1.51633968E-22** | **1.62349216E-37** | 2692.70 | - | 0 | 0 |
| | (1.00788408E-21) | | (48.95) | - | | |
| BStSd-PSO 1 | **2.29450987E-19** | **2.68747235E-25** | 1861.20 | - | 0 | 0 |
| | (1.31088746E-18) | | (149.50) | - | | |
| **BStSd-PSO 2** | **1.52080460E-07** | **7.43537369E-39** | 383.24 | - | 0 | 0 |
| | (9.09583425E-07) | | (104.14) | - | | |
| BSwLd-PSO 1 | **1.08192950E-13** | **8.91176823E-21** | 1672.66 | - | 0 | 0 |
| | (6.68935597E-13) | | (26.47) | - | | |
| BSwLd-PSO 2 | **1.25007990E-27** | **3.40522084E-41** | 2700.14 | - | 0 | 0 |
| | (6.22245811E-27) | | (41.70) | - | | |
| BSwSd-PSO 1 | **1.65996680E-19** | **1.25090335E-25** | 2084.04 | - | 0 | 0 |
| | (5.11411361E-19) | | (175.47) | - | | |
| **BSwSd-PSO 2** | **9.57394824E+01** | **1.84290602E+00** | - | - | 50 | 0 |
| | (9.90527050E+01) | | - | - | | |

**Table 6. 8**: Performance of 10 algorithms when optimizing the 30-dimensional Sphere benchmark test function along 10000 time-steps, where the particles are initially randomly spread over the region $[-100,100]^{30}$, $aw^{(t)} = 4.1 \cdot w^{(t)} \ \forall t$, and the statistical data are calculated out of 50 runs.

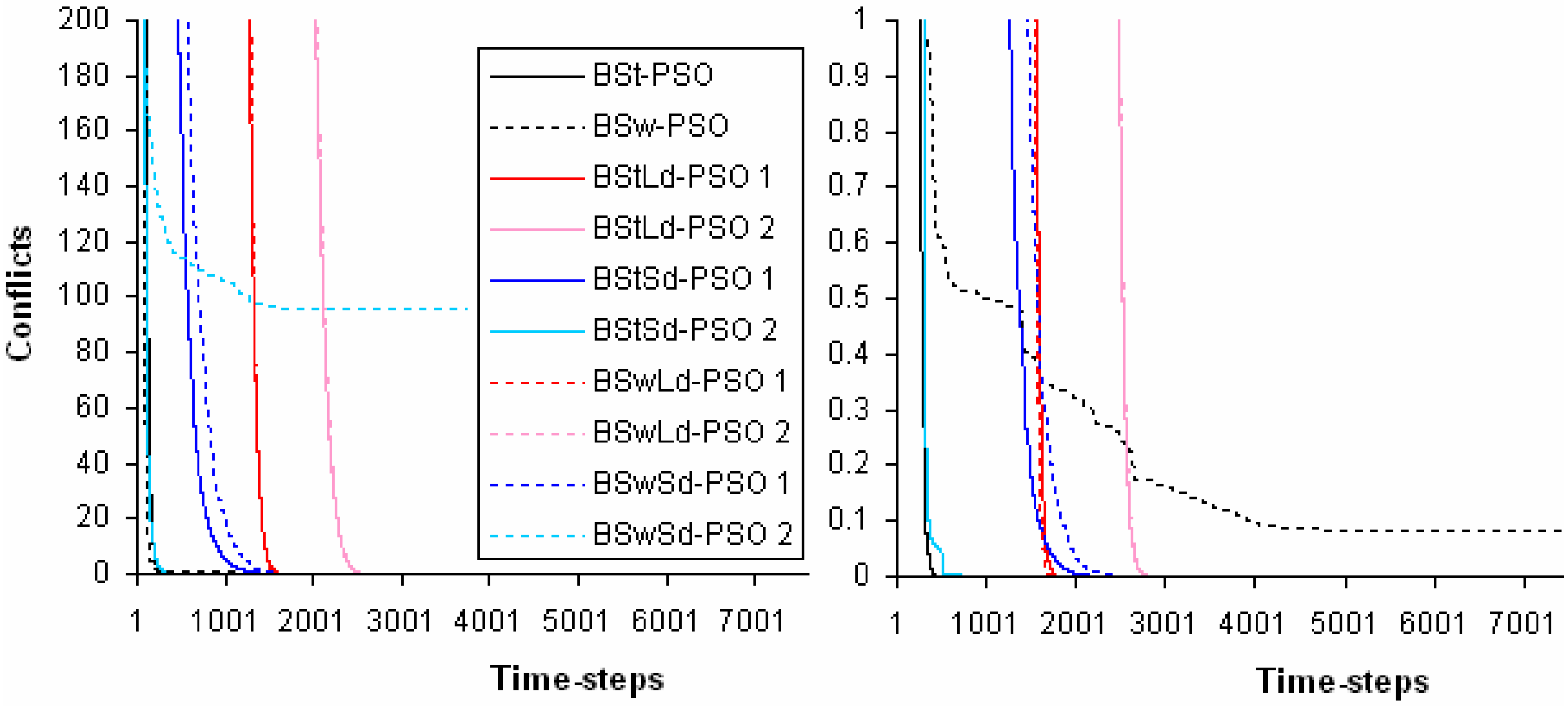

**Fig. 6. 38**: Evolution of the mean best conflicts found by 10 different optimizers for the Sphere benchmark test function out of 50 runs along 10000 time-steps, for $aw^{(t)} = 4.1 \cdot w^{(t)} \quad \forall t$ .

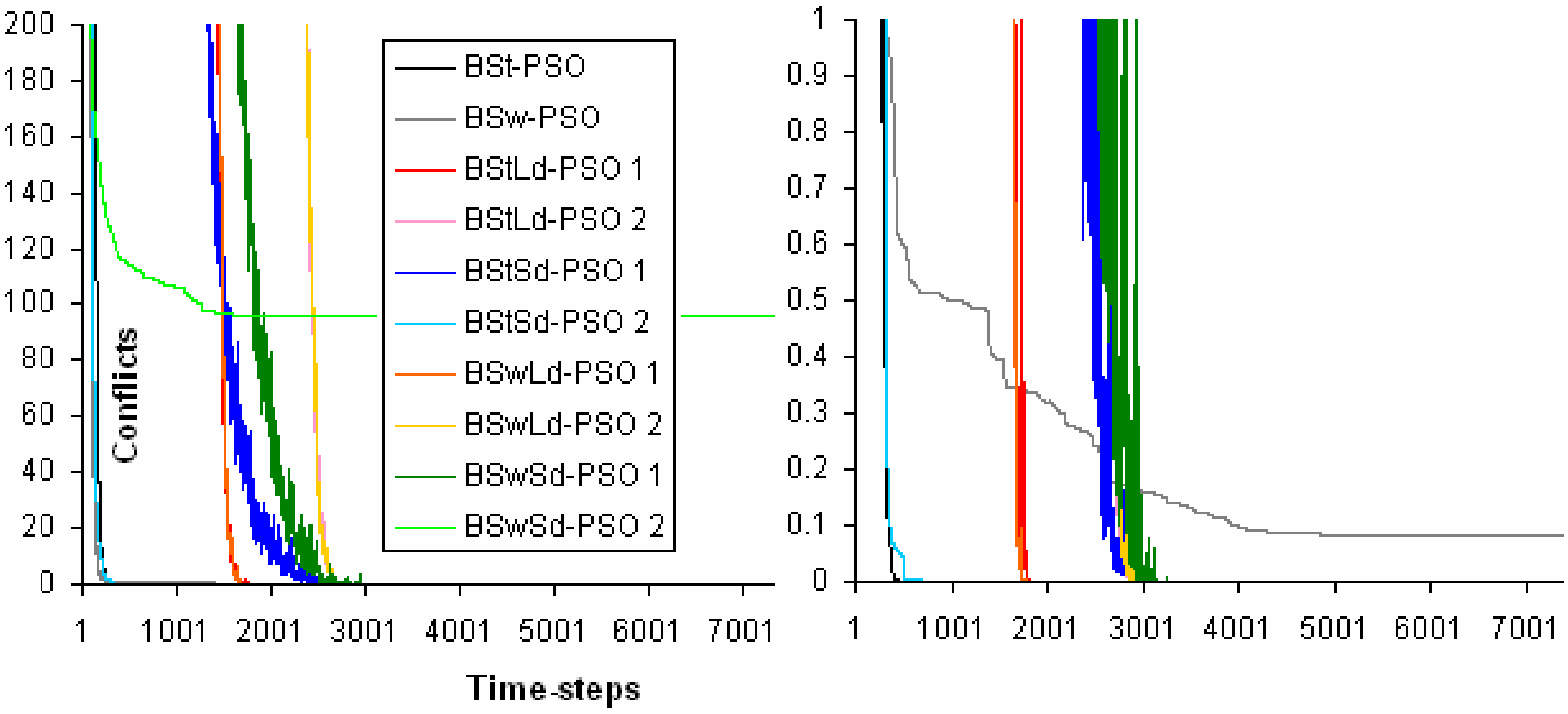

**Fig. 6. 39**: Evolution of the mean average conflicts found by 10 different optimizers for the Sphere benchmark test function along 10000 time-steps, for $aw^{(t)} = 4.1 \cdot w^{(t)} \quad \forall t$ , where the average is calculated out of 30 particles, and the mean is calculated out of 50 runs.

The aim of the swapping strategy is to enhance individuality at the early stages and sociality at the late stages of the search. Hence the clustering is initially delayed in comparison to the corresponding standard versions, but it is enhanced while sociality is. This increase in the pressure to cluster makes the particles more likely to get trapped in suboptimal solutions. Thus, the so bad mean best conflicts found by some optimizers such as the BSw-PSO and the BSwSd-PSO 2 seem to be due to premature clustering. It is reasonable to theorize that premature clustering is also the reason for all the optimizers' failure in finding the exact global optimum. Keeping in mind that the sigmoidly time-decreasing inertia weight also aims to enhance clustering during the late stages of the search, consider only the pure constricted PSOs (in bold in **Table 6. 8**): it can be observed that the swapping versions exhibit more pronounced premature clustering than their corresponding standard versions, and that the optimizers with sigmoidly time-decreasing inertia weights exhibit more pronounced premature clustering than their corresponding ones with constant inertia weights. Thus, the strategies work exactly as intended.

With regards to the other optimizers, which are not pure C-PSOs, their particles' clustering is delayed[16] until their inertia weights start taking values of under 0.7298. Hence they display wider oscillations in the evolution of their mean average conflicts, which can be clearly observed in **Fig. 6. 39**. Nevertheless, they all end up exhibiting outstanding clustering, so that the curves of their mean best conflicts and the corresponding curves of their mean average conflicts end up merging, and finally stagnating. If the graphs of the mean average conflicts in **Appendix 4** are subsequently zoomed in, it can be seen that the particles of every optimizer tested here virtually implode to a single point before the particles of the BSt-PSO do. It can be therefore concluded that the only reason for their poor solutions is their premature clustering, non-desirable for a pure strategy, but perhaps highly desirable for a combined one!

Regarding the best performance as stand-alone optimizers, the BSt-PSO finds the best $\overline{cgbest}$ with the smallest corresponding standard deviation; the best *cgbest*; and it attains the error condition faster than any other and with the second smallest corresponding standard deviation. Conversely, the BSwSd-PSO 2 and the BSw-PSO are the optimizers that find the worst $\overline{cgbest}$ with the largest standard deviations. Furthermore, they are the only optimizers that repeatedly fail in attaining the error condition. In fact, the former fails every time!

In summary, the extremely bad mean best conflicts found by the worst stand-alone optimizers and the incapability of finding the exact global optimum of the best stand-alone optimizers is due to premature clustering. While this effect wipes out the desirable robustness of a PSO— which is its essential characteristic—the effect of the strategy is exactly as expected: the outstanding clustering enables the particles to take extremely small-sized steps, which favour a thorough exploitation of a very small region. It is immediate to think of just a few particles in a swarm displaying this behaviour so as to fine-tune the search, while others display more explorative behaviour. This strategy is proposed in a later chapter.

[16] Notice that, except for the BSwSd-PSO 2—which never attains the error condition due to premature clustering—, all the pure constricted PSOs (in bold in **Table 6. 8**) attain the error condition much faster than the others.

### 6.4.2.1.2 ROSENBROCK function

The results of the experiments for the optimization of this function are gathered in **Table 6. 9**, and the evolution of the mean best conflict found by each optimizer is plotted in **Fig. 6. 40**.

| **ROSENBROCK** | $\overline{cgbest}\,(\sigma)$ | $cgbest$ | $\overline{tsec}\,(\sigma)$ | $\overline{tsgb}\,(\sigma)$ | $nf$ | $ngb$ |
|---|---|---|---|---|---|---|
| **BSt-PSO** | **3.53321340E+00** (3.73341058E+00) | **4.47164501E-06** | 703.04 (911.42) | - - | 0 | 0 |
| **BSw-PSO** | **5.44782882E+01** (4.44031185E+01) | **6.65254089E-03** | 2340.41 (2163.19) | - - | 6 | 0 |
| BStLd-PSO 1 | **5.06228857E+01** (4.21165709E+01) | **1.43570711E+00** | 1868.80 (358.35) | - - | 4 | 0 |
| BStLd-PSO 2 | **2.52875291E+01** (2.96783428E+01) | **1.25648984E-02** | 3010.10 (634.66) | - - | 1 | 0 |
| BStSd-PSO 1 | **3.62988637E+01** (3.60657959E+01) | **1.12640017E-01** | 2423.23 (733.93) | - - | 2 | 0 |
| **BStSd-PSO 2** | **5.43653066E+01** (4.79598944E+01) | **7.15253216E+00** | 475.80 (559.46) | - - | 4 | 0 |
| BSwLd-PSO 1 | **4.31834289E+01** (3.39205587E+01) | **4.09537731E+00** | 1866.30 (336.28) | - - | 4 | 0 |
| BSwLd-PSO 2 | **2.38337650E+01** (2.73146592E+01) | **5.12134471E-08** | 3047.02 (620.11) | - - | 0 | 0 |
| BSwSd-PSO 1 | **3.33803481E+01** (3.37354898E+01) | **2.15298549E-01** | 2632.61 (720.86) | - - | 1 | 0 |
| **BSwSd-PSO 2** | **4.79117089E+03** (6.54773219E+03) | **1.84774866E+02** | - - | - - | 50 | 0 |

**Table 6. 9**: Performance of 10 algorithms when optimizing the 30-dimensional Rosenbrock benchmark test function along 10000 time-steps, where the particles are initially randomly spread over the region $\left[-30,30\right]^{30}$, $aw^{(t)} = 4.1 \cdot w^{(t)}\ \ \forall t$, and the statistical data are calculated out of 50 runs.

First of all, it must be noted that the trend is the same as in the case of optimizing the Sphere function: no optimizer is able to find the exact global optimum; the pure constricted PSOs take much shorter to attain the error condition than the others; the BSt-PSO finds the best, remarkably good $\overline{cgbest}$; the BSwSd-PSO 2 finds the worst, remarkably bad $\overline{cgbest}$, also failing in attaining the error condition every time; the other pure constricted PSOs, namely the BSw-PSO and the BStSd-PSO 2, find the second and third poorest $\overline{cgbest}$ and exhibit the second and third highest numbers of failures in attaining the error condition, respectively.

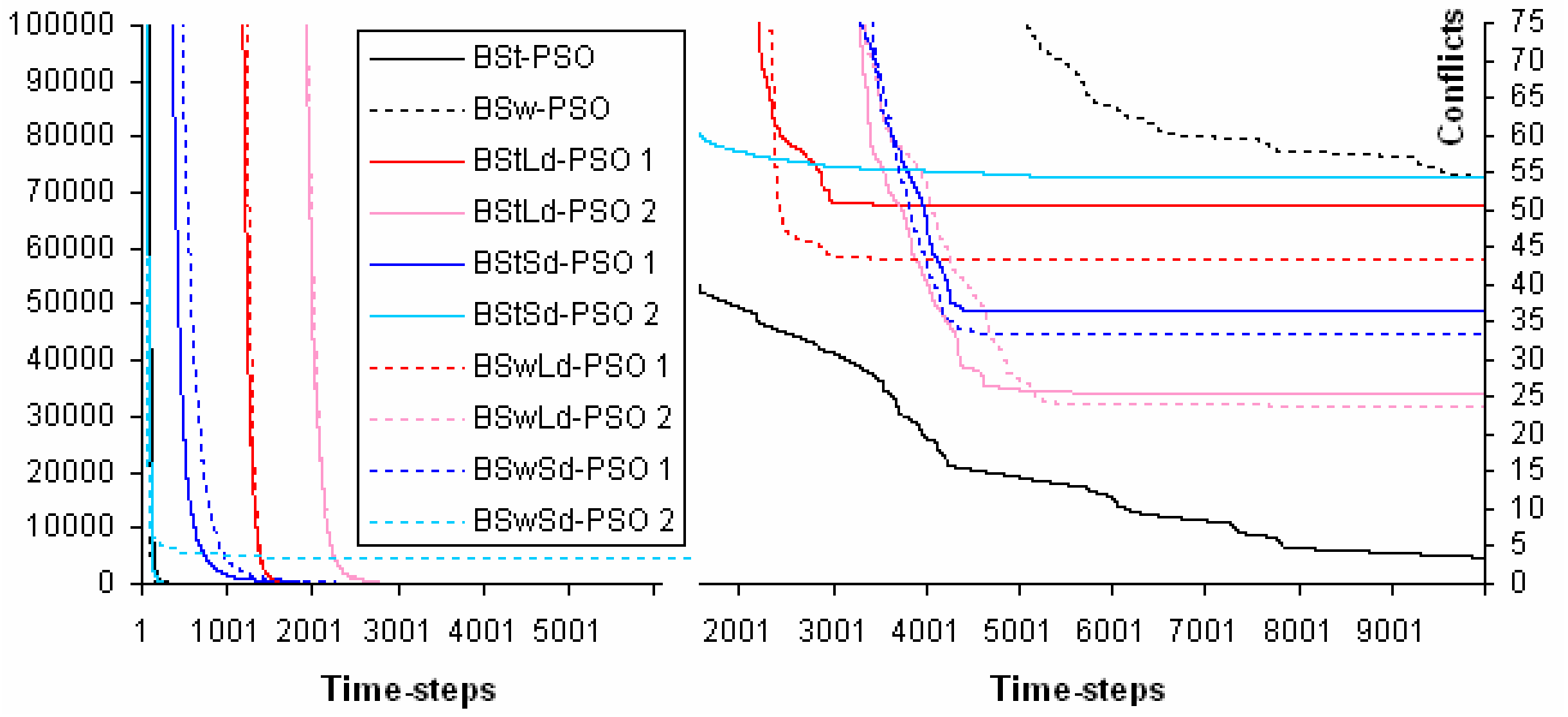

**Fig. 6. 40**: Evolution of the mean best conflicts found by 10 different optimizers for the Rosenbrock benchmark test function out of 50 runs along 10000 time-steps, for $aw^{(t)} = 4.1 \cdot w^{(t)} \;\; \forall t$ .

It can be noticed from observing the curves of the evolution of the mean average conflicts in **Appendix 4** that the BSt-PSO and the BSw-PSO do not exhibit the outstanding clustering they do when optimizing the Sphere function. Although they both already display very small amplitudes in the oscillation of the evolution of their mean average conflicts by around the 500$^{th}$ time-step, the amplitudes start increasing from then on, as if the particles were diverging. Besides, the curves of the evolution of their mean best conflicts do not stagnate (see **Fig. 6. 40**), which is in agreement with the conclusion that diversity is not completely lost. In other words, although the particles exhibit an overall clustering behaviour, they do not fine-cluster as in the case of optimizing the Sphere function. The reason for this is not understood.

Conversely, the clustering of the particles of the BSwSd-PSO 2 is once again striking: all the particles quickly clustered, virtually imploding to a single point—which is worth a very bad conflict approximately equal to 4800—by around the 1500$^{th}$ time-step. The BStLd-PSO 1, the BSwLd-PSO 1, the BStSd-PSO 1, and the BSwSd-PSO 1 also exhibit outstanding clustering behaviour: the curves of the evolution of their mean average conflicts and those of the evolution of the corresponding mean best conflicts virtually merge and stagnate. In other words, a premature implosion of their particles also takes place, although these optimizers take much longer than the BSwSd-PSO 2 to complete it. It is important to remark that no optimizer with unrelated inertia and acceleration weights exhibited the implosion of all their particles (refer to section **6.4.1.7.2** and digital **Appendix 4**).

In spite of the fact that a complete clustering cannot be accomplished, the amplitudes of the oscillations observed in the evolution of the mean average conflicts of the BStSd-PSO 2 are small for the last 5000 time-steps. Besides, the curve that lower-bounds the graph of the evolution of its mean average conflict and the curve of the evolution of its mean best conflict merge and stagnate, showing that the particles are finding it difficult to fine-cluster further.

The BStLd-PSO 2 and the BSwLd-PSO 2 find the second and third best mean best conflicts here, probably thanks to the high diversity they still maintain by the end of the search. In addition, the trend-lines of the evolution of their mean average conflicts are still steeply decreasing, so that further fine-clustering can be expected for a longer search. Since these optimizers take longer to start behaving as C-PSOs, they just seem to need longer to cluster.

In summary, it appears that maintaining some diversity enables the optimizers to find better solutions[17], while time-decreasing inertia weights favour fine-clustering. It can be concluded that, in the same fashion as when optimizing the Sphere function, the BSt-PSO exhibits the best combination of exploration and exploitation abilities, and the BSwSd-PSO 2 exhibits the strongest clustering ability for the optimization of the Rosenbrock function.

### 6.4.2.1.3 RASTRIGRIN function

The results of the experiments for the optimization of this function are gathered in **Table 6. 10**, and the evolution of the mean best conflict found by each optimizer is plotted in **Fig. 6. 41**.

Optimizing this function, which exhibits numerous local optima, is quite a different matter from optimizing the Sphere and the Rosenbrock functions. It can be noticed that the BSt-PSO only finds the second worst mean best conflict ($\overline{cgbest}$) now, while the BSwSd-PSO 1 and the BStSd-PSO 1 find the best and second best ones, respectively. In addition, they also find the best and second best conflicts found in any of the 50 runs (*cgbest*).

[17] The question is how large diversity must be, since clustering and keeping diversity are both desirable features which do not come together: while the particles cluster, diversity is decreased. Note that while the BSt-PSO tested in section **6.4.1** keeps higher diversity than the one tested here, it only finds a very much worse mean best conflict.

| RASTRIGRIN | $\overline{cgbest}\,(\sigma)$ | $cgbest$ | $\overline{tsec}\,(\sigma)$ | $\overline{tsgb}\,(\sigma)$ | $nf$ | $ngb$ |
|---|---|---|---|---|---|---|
| **BSt-PSO** | **6.20057085E+01** (1.58588099E+01) | **2.98487465E+01** | 141.36 (35.69) | - - | 0 | 0 |
| **BSw-PSO** | **4.13504400E+01** (1.15754658E+01) | **2.28840432E+01** | 166.86 (56.10) | - - | 0 | 0 |
| BStLd-PSO 1 | **4.22061090E+01** (1.07462920E+01) | **1.59193428E+01** | 1335.90 (92.09) | - - | 0 | 0 |
| BStLd-PSO 2 | **3.59179800E+01** (1.23327629E+01) | **1.69142939E+01** | 2208.42 (174.67) | - - | 0 | 0 |
| BStSd-PSO 1 | **2.62271085E+01** (6.99439700E+00) | **1.59193399E+01** | 626.86 (206.16) | - - | 0 | 0 |
| **BStSd-PSO 2** | **7.38257760E+01** (1.78865989E+01) | **3.88033528E+01** | 110.06 (36.03) | - - | 3 | 0 |
| BSwLd-PSO 1 | **3.98381129E+01** (1.13281359E+01) | **1.59193449E+01** | 1395.94 (101.30) | - - | 0 | 0 |
| BSwLd-PSO 2 | **3.47837267E+01** (8.16997570E+00) | **2.08941301E+01** | 2176.44 (169.82) | - - | 0 | 0 |
| BSwSd-PSO 1 | **2.41974147E+01** (6.49316142E+00) | **1.09445496E+01** | 996.22 (270.18) | - - | 0 | 0 |
| **BSwSd-PSO 2** | **4.98189553E+01** (1.13272599E+01) | **1.79401387E+01** | 112.28 (29.70) | - - | 0 | 0 |

**Table 6. 10**: Performance of 10 algorithms when optimizing the 30-dimensional Rastrigrin benchmark test function along 10000 time-steps, where the particles are initially randomly spread over the region $[-5.12,5.12]^{30}$, $aw^{(t)} = 4.1 \cdot w^{(t)} \;\; \forall t$, and the statistical data are calculated out of 50 runs.

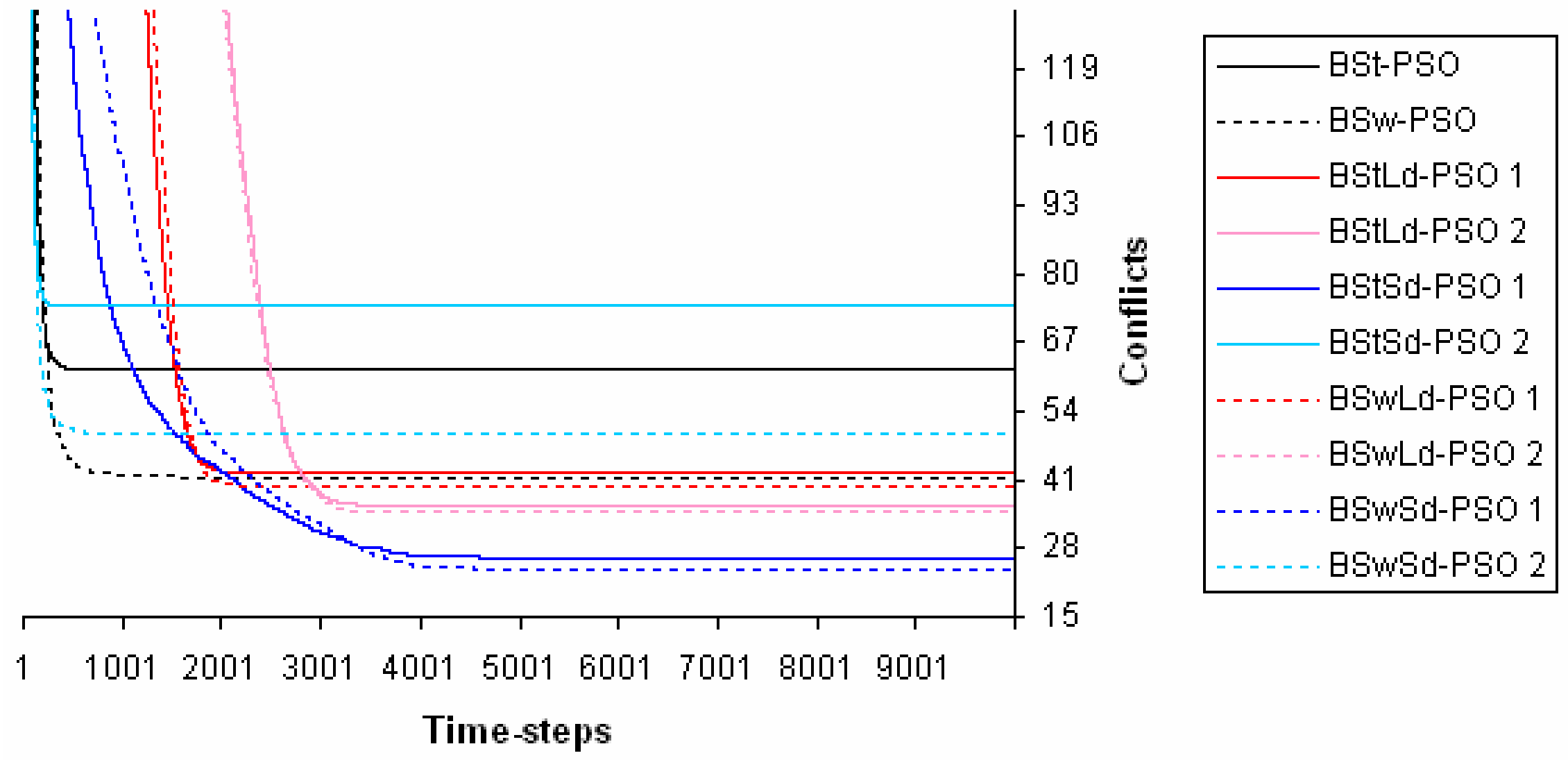

**Fig. 6. 41**: Evolution of the mean best conflicts found by 10 different optimizers for the Rastrigrin benchmark test function out of 50 runs along 10000 time-steps, for $aw^{(t)} = 4.1 \cdot w^{(t)} \;\; \forall t$.

It is self-evident that too strong clustering behaviour makes the algorithms more likely to get trapped in local optima. Therefore, the outstanding clustering of the particles of the BStSd-PSO 2 leads them to a fast premature implosion, thus obtaining the poorest solution and being the only optimizer that presents a few failures in attaining the error condition for this function. Besides, the higher individuality at the beginning of the search seems to enable the particles of the optimizers with time-swapping learning weights to escape more local optima, so that all the swapping versions find better solutions here than their corresponding standard versions.

All the curves of the evolution of the mean best conflicts and the corresponding curves of the evolution of the mean average conflicts, except for those of the BStSd-PSO 1 and of the BSwSd-PSO 1, merge and stagnate (refer to **Appendix 4**). Although the curves of the mean average conflicts of these two optimizers also stagnate and almost eliminate the amplitudes of the oscillations, they do not merge with the corresponding curves of the evolution of the mean best conflicts. Since these optimizers—which find the two best mean best conflicts—are the only ones whose particles do not perform a complete implosion, it can be inferred that maintaining some diversity is critical with regards to the best solution found here.

Regarding the convergence rate, it is clear that the larger the inertia weights the longer the optimizers take to attain the error condition. As it has been claimed before, it seems that despite keeping a certain relationship between the inertia and the acceleration weights, the larger the learning weights the stronger the influence of the random weights, which do not seem to favour clustering. Hence the BStLd-PSO 2 and the BSwLd-PSO 2 take around 2100 time-steps; the BStLd-PSO 1 and the BSwLd-PSO 1 take around 1300 time-steps; the BStSd-PSO 1 and the BSwSd-PSO 1 take between 600 and 1000 time-steps; while the BSt-PSO, the BSw-PSO, the BStSd-PSO 2, and the BSwSd-PSO 2 only take less than 200 time-steps to attain the error condition. Thus, it appears that smaller learning weights favour clustering.

In summary, the BSwSd-PSO 1 and the BStSd-PSO 1 are the two best stand-alone optimizers for the Rastrigrin function, while the BStSd-PSO 2, the BSt-PSO and the BSwSd-PSO 2 are the best optimizers with regards to the ability to cluster.

#### 6.4.2.1.4 GRIEWANK function

The results of the experiments for the optimization of this function are gathered in **Table 6. 11**, and the evolution of the mean best conflict found by each optimizer is plotted in **Fig. 6. 42**:

| GRIEWANK | $\overline{cgbest}(\sigma)$ | $cgbest$ | $\overline{tsec}(\sigma)$ | $\overline{tsgb}(\sigma)$ | $nf$ | $ngb$ |
|---|---|---|---|---|---|---|
| **BSt-PSO** | **2.67443418E-02** (4.85218585E-02) | **0.00000000E+00** | 316.00 (37.77) | 1376.71 (356.18) | 1 | 14 |
| **BSw-PSO** | **9.59908020E-02** (1.59190851E-01) | **0.00000000E+00** | 866.70 (1168.49) | 5627.75 (461.80) | 10 | 4 |
| BStLd-PSO 1 | **1.72351463E-02** (1.99256390E-02) | **2.22044605E-16** | 1669.12 (35.55) | - - | 0 | 0 |
| BStLd-PSO 2 | **1.35179270E-02** (1.51549714E-02) | **0.00000000E+00** | 2637.54 (57.50) | 3757.79 (342.98) | 0 | 14 |
| BStSd-PSO 1 | **1.75757999E-02** (2.35707489E-02) | **0.00000000E+00** | 1775.53 (188.67) | 3917.25 (49.59) | 1 | 8 |
| **BStSd-PSO 2** | **6.27320837E-02** (7.27910307E-02) | **1.11022302E-16** | 309.74 (78.57) | - - | 8 | 0 |
| BSwLd-PSO 1 | **1.30633193E-02** (2.03014230E-02) | **0.00000000E+00** | 1628.90 (30.27) | 2751.00 (410.12) | 1 | 2 |
| BSwLd-PSO 2 | **1.82815263E-02** (2.39583151E-02) | **0.00000000E+00** | 2641.41 (71.31) | 3787.88 (94.22) | 1 | 8 |
| BSwSd-PSO 1 | **1.74658082E-02** (1.49200445E-02) | **0.00000000E+00** | 1978.96 (208.32) | 3961.00 (7.07) | 0 | 2 |
| **BSwSd-PSO 2** | **1.95866213E+00** (1.05752788E+00) | **8.68984756E-01** | - - | - - | 50 | 0 |

**Table 6. 11**: Performance of 10 algorithms when optimizing the 30-dimensional Griewank benchmark test function along 10000 time-steps, where the particles are initially randomly spread over the region $[-600,600]^{30}$, $aw^{(t)} = 4.1 \cdot w^{(t)}\ \forall t$, and the statistical data are calculated out of 50 runs.

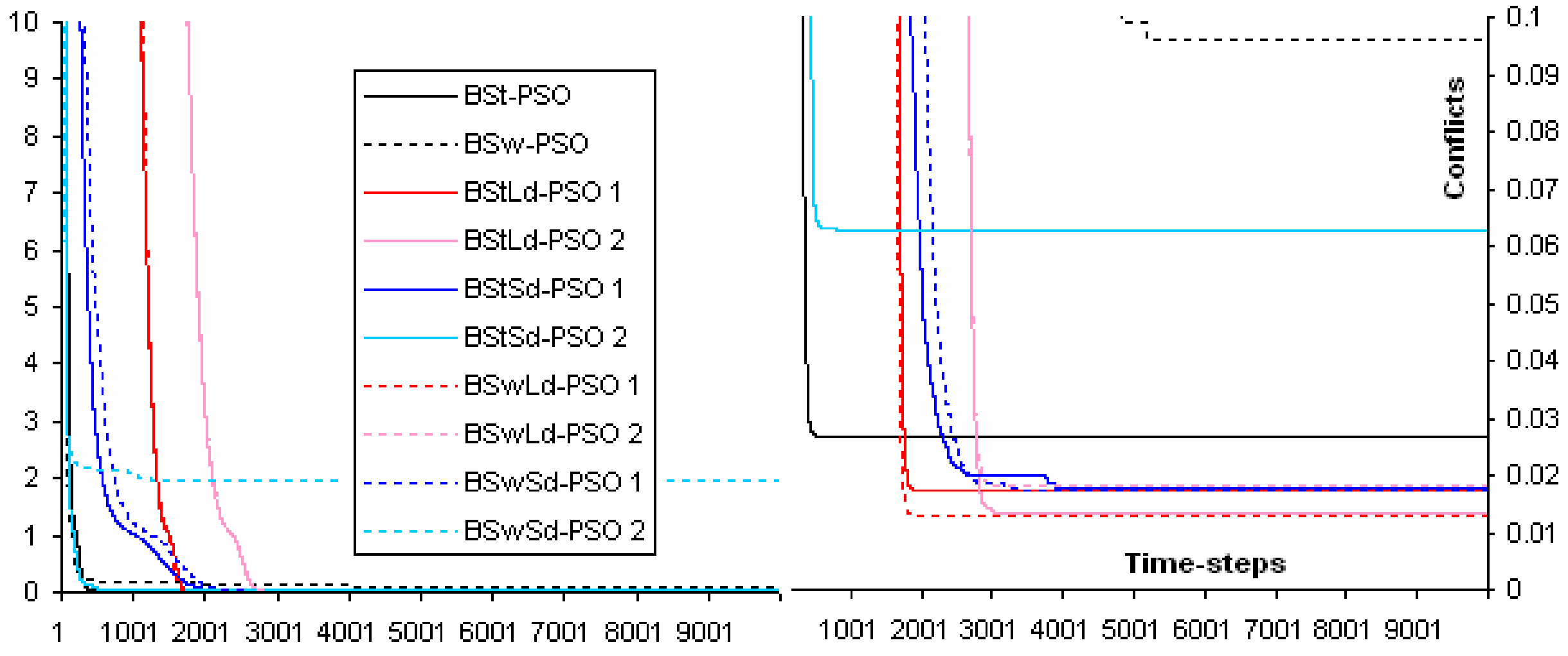

**Fig. 6. 42**: Evolution of the mean best conflicts found by 10 different optimizers for the Griewank benchmark test function out of 50 runs along 10000 time-steps, for $aw^{(t)} = 4.1 \cdot w^{(t)}\ \forall t$.

In the same fashion as when optimizing the Sphere and the Rosenbrock functions, the BSwSd-PSO 2 is the optimizer which exhibits the fastest premature clustering, and the only one which fails in attaining the error condition along every run.

It is interesting to note that the pure constricted PSOs take noticeably shorter to attain the error condition than the other optimizers, except for the BSwSd-PSO 2, which never does! However, they exhibit higher numbers of failures. It seems that the C-PSOs are so fast in clustering that the error condition is quickly attained, but diversity is completely lost shortly after, thus losing all possibility of further improvement of the best conflict found. In fact, the BSwSd-PSO 2 never attains the error condition, and the mean best conflict found by the BSw-PSO is barely better than the error condition itself (refer to **Table 6. 1** and **Table 6. 11**).

The optimizers which are not pure C-PSOs seem to need longer to cluster, thus exhibiting explorative behaviour for a longer period of time. This enables them to escape more local optima, resulting in better solutions found. Nevertheless, the curves of the evolution of their mean best conflicts also end up merging with the corresponding curves of the evolution of their mean average conflicts (see **Appendix 4**).

Therefore, it can be concluded that the stagnation observed in the evolution of the mean best conflicts found by each optimizer is not due to a failure in the fine-tuning but due to the outstanding clustering ability that leads to a premature complete implosion of their particles.

In summary, the best mean best conflict is found by the BSwLd-PSO 1, while the next five best mean best conflicts are found by the other five non-constricted PSOs, whose particles take longer to implode. Conversely, the C-PSOs exhibit striking fast clustering behaviour, where the BSwSd-PSO 2 exhibits the fastest. Notice that although the BSt-PSO does not find an outstanding solution this time, it is the second fastest in attaining the error condition, and the fastest in finding the exact global optimum. In addition, it is one of the two optimizers that find the exact global optimum a higher number of times, while it also exhibits outstanding clustering ability, and it only fails to attain the error condition once out of the 50 runs.

#### 6.4.2.1.5 SCHAFFER F6 function (2D)

The results of the experiments for the optimization of this function are gathered in **Table 6. 12**, and the evolution of the mean best conflict found by each optimizer is plotted in **Fig. 6. 43**:

| SCHAFFER F6 2D | $\overline{cgbest}\,(\sigma)$ | $cgbest$ | $\overline{tsec}\,(\sigma)$ | $\overline{tsgb}\,(\sigma)$ | $nf$ | $ngb$ |
|---|---|---|---|---|---|---|
| **BSt-PSO** | **1.55454558E-03** (3.59807386E-03) | **0.00000000E+00** | 738.14 (1266.39) | 887.74 (1269.33) | 8 | 42 |
| **BSw-PSO** | **5.82954593E-04** (2.33082674E-03) | **0.00000000E+00** | 538.72 (1067.02) | 703.66 (1065.09) | 3 | 47 |
| BStLd-PSO 1 | **3.88636395E-04** (1.92325138E-03) | **0.00000000E+00** | 929.00 (246.48) | 1272.13 (167.99) | 2 | 48 |
| BStLd-PSO 2 | **0.00000000E+00** (0.00000000E+00) | **0.00000000E+00** | 1237.08 (303.66) | 1788.60 (180.44) | 0 | 50 |
| BStSd-PSO 1 | **3.88636395E-04** (1.92325138E-03) | **0.00000000E+00** | 571.96 (573.02) | 872.23 (581.84) | 2 | 48 |
| **BStSd-PSO 2** | **4.46931854E-03** (4.89154705E-03) | **0.00000000E+00** | 159.44 (147.19) | 272.56 (143.62) | 23 | 27 |
| BSwLd-PSO 1 | **1.94318198E-04** (1.37403715E-03) | **0.00000000E+00** | 1036.12 (317.10) | 1384.90 (230.35) | 1 | 49 |
| BSwLd-PSO 2 | **0.00000000E+00** (0.00000000E+00) | **0.00000000E+00** | 1380.14 (369.49) | 1898.10 (261.21) | 0 | 50 |
| BSwSd-PSO 1 | **0.00000000E+00** (0.00000000E+00) | **0.00000000E+00** | 511.90 (603.74) | 892.90 (556.64) | 0 | 50 |
| **BSwSd-PSO 2** | **1.55454558E-03** (3.59807386E-03) | **0.00000000E+00** | 423.93 (629.93) | 566.57 (626.08) | 8 | 42 |

**Table 6. 12**: Performance of 10 algorithms when optimizing the 30-dimensional 2-dimensional Schaffer f6 benchmark test function along 10000 time-steps, where the particles are initially randomly spread over the region $[-100,100]^2$, $aw^{(t)} = 4.1 \cdot w^{(t)}\ \forall t$, and the statistical data are calculated out of 50 runs.

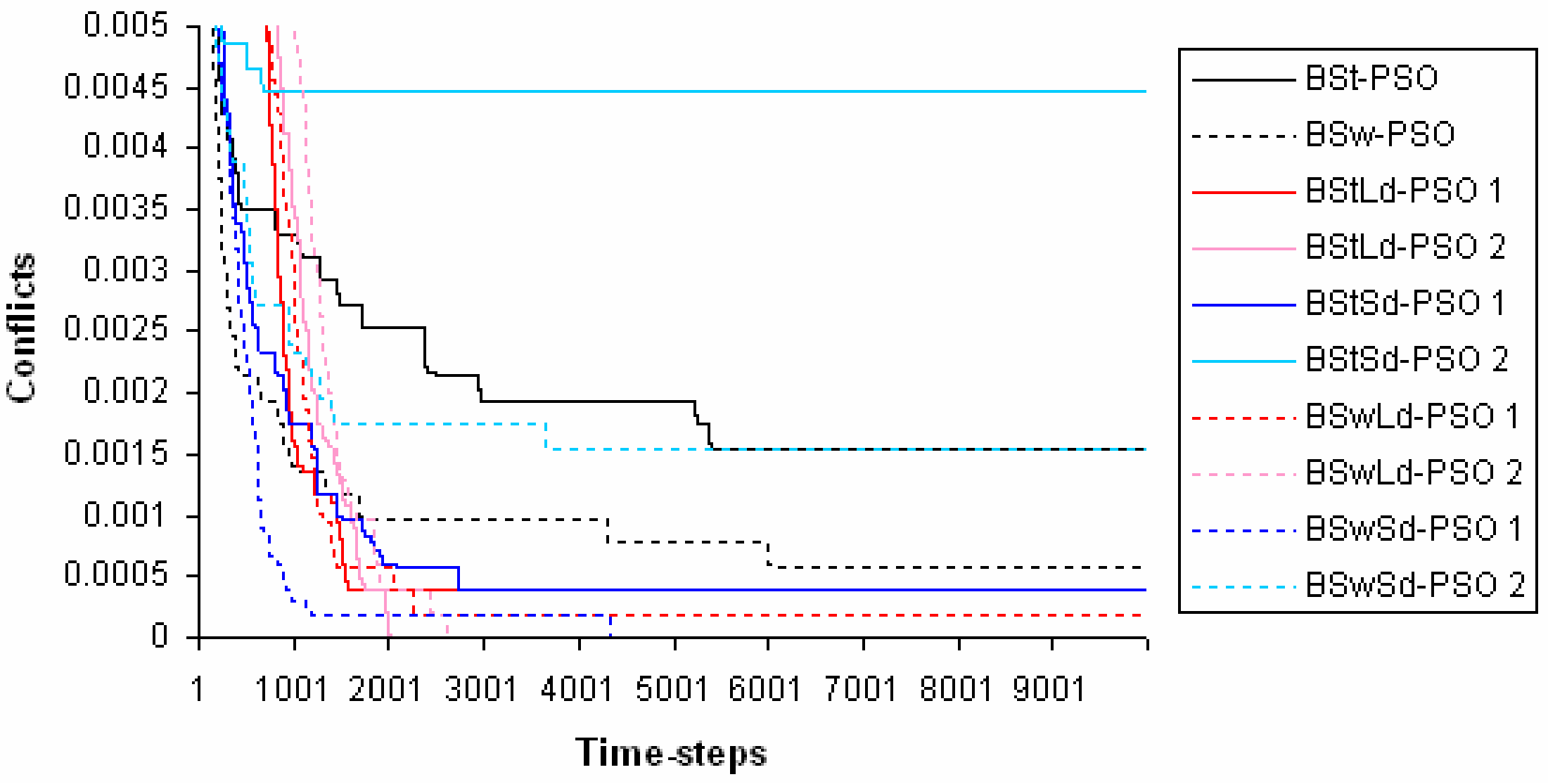

**Fig. 6. 43**: Evolution of the mean best conflicts found by 10 different optimizers for the 2-dimensional Schaffer f6 benchmark test function out of 50 runs along 10000 time-steps, for $aw^{(t)} = 4.1 \cdot w^{(t)}\ \forall t$.

It is surprising to see that, while all the optimizers whose acceleration and inertia weights are kept unrelated find the exact global optimum for this function along each of the 50 runs (refer to **Table 6. 6**), failures not only in finding the exact global optimum but also in attaining the error condition can be observed here. It is even more surprising to see that this is not due to premature clustering, since the only optimizers that perform a complete implosion—namely the BStLd-PSO 2 and the BSwLd-PSO 2—never fail in attaining the error condition; they find the exact global optimum along each of the 50 runs; and their particles completely cluster around the global optimum (see **Appendix 4**). In contrast, although the BSwSd-PSO 1 also finds the global optimum every time, it still maintains some diversity by the end of the search.

Given that all the "Sw" versions find better mean best conflicts than their corresponding "St" versions, the higher initial individuality of the swapping strategy seems to be successful here in helping to escape local optima. In spite of the stagnation of the curves of the evolution of the mean best conflicts, the fact that they do not merge with the corresponding curves of the evolution of the mean average conflicts suggests that the particles of the optimizers that cannot find the exact global optimum do not completely cluster. It seems that they are trapped in local optima. Beware that although the curves of the evolution of the mean average conflicts found by the BStSd-PSO 1 and the BSwSd-PSO 1 display no oscillation by the end of the run, they do not merge with the corresponding curves of the evolution of the mean best conflicts. Hence it can be inferred that they also maintain some diversity (see **Appendix 4**).

Although three of the optimizers tested here—whose inertia and acceleration weights are related like $aw^{(t)} = 4.1 \cdot w^{(t)} \ \forall t$—find the exact global optimum, keeping them unrelated seems to be a better strategy to optimize this function.

### 6.4.2.1.6 SCHAFFER F6 function

The results of the experiments for the optimization of this function are gathered in **Table 6. 13**, while the graphs of the evolution of the mean best and of the mean average conflicts found by each optimizer are plotted in **Fig. 6. 44** and **Fig. 6. 45**, respectively. It is interesting to note that the BStSd-PSO 2 and the BSwSd-PSO 2 are the only optimizers which exhibit a virtually complete implosion of their particles, as it can be concluded from the fact that the curves of the evolution of their mean average conflicts almost merge with the corresponding curves of the evolution of their mean best conflicts by the end of the search.

| SCHAFFER F6 | $\overline{cgbest}\,(\sigma)$ | $cgbest$ | $\overline{tsec}\,(\sigma)$ | $\overline{tsgb}\,(\sigma)$ | $nf$ | $ngb$ |
|---|---|---|---|---|---|---|
| **BSt-PSO** | **1.95553279E-01** (6.73308236E-02) | **3.72240751E-02** | 961.50 (307.59) | - - | 48 | 0 |
| **BSw-PSO** | **1.33615803E-01** (4.24508633E-02) | **7.81891821E-02** | 1913.42 (1698.52) | - - | 38 | 0 |
| BStLd-PSO 1 | **1.36393868E-01** (4.64888288E-02) | **7.81891821E-02** | 2047.27 (285.18) | - - | 39 | 0 |
| BStLd-PSO 2 | **1.32648396E-01** (3.68648462E-02) | **7.81891821E-02** | 3066.45 (277.07) | - - | 39 | 0 |
| BStSd-PSO 1 | **1.20498594E-01** (3.18182344E-02) | **7.81891821E-02** | 3448.57 (280.61) | - - | 36 | 0 |
| **BStSd-PSO 2** | **3.20947828E-01** (9.21032429E-02) | **1.26990519E-01** | - - | - - | 50 | 0 |
| BSwLd-PSO 1 | **1.53665461E-01** (4.85674098E-02) | **7.81891821E-02** | 1949.86 (173.26) | - - | 43 | 0 |
| BSwLd-PSO 2 | **1.36424717E-01** (4.13243260E-02) | **7.81891821E-02** | 3012.38 (79.29) | - - | 42 | 0 |
| BSwSd-PSO 1 | **1.19630683E-01** (3.18403783E-02) | **3.72240751E-02** | 3494.31 (382.40) | - - | 37 | 0 |
| **BSwSd-PSO 2** | **2.54185648E-01** (6.35104887E-02) | **1.78222303E-01** | - - | - - | 50 | 0 |

**Table 6. 13**: Performance of 10 algorithms when optimizing the 30-dimensional Schaffer f6 benchmark test function along 10000 time-steps, where the particles are initially randomly spread over the region $[-100,100]^{30}$, $aw^{(t)} = 4.1 \cdot w^{(t)} \;\; \forall t$, and the statistical data are calculated out of 50 runs.

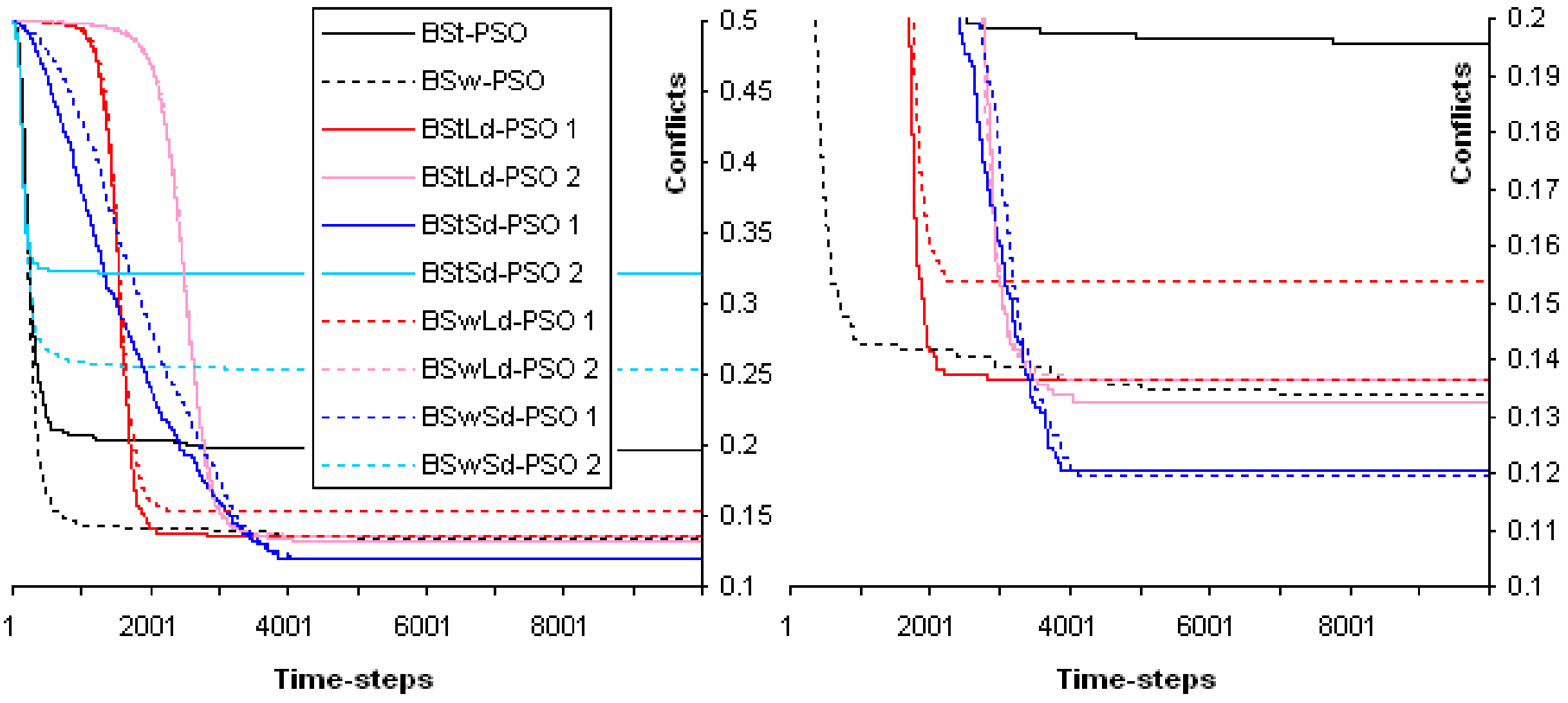

**Fig. 6. 44**: Evolution of the mean best conflicts found by 10 different optimizers for the Schaffer f6 benchmark test function out of 50 runs along 10000 time-steps, for $aw^{(t)} = 4.1 \cdot w^{(t)} \;\; \forall t$.

Beware that although the amplitudes of the oscillations observed in the evolution of the mean average conflicts found by the optimizers whose inertia weights approach zero are reduced to almost zero by the end of the search, they do not merge with the corresponding curves of the evolution of the mean best conflicts. Hence diversity is not lost.

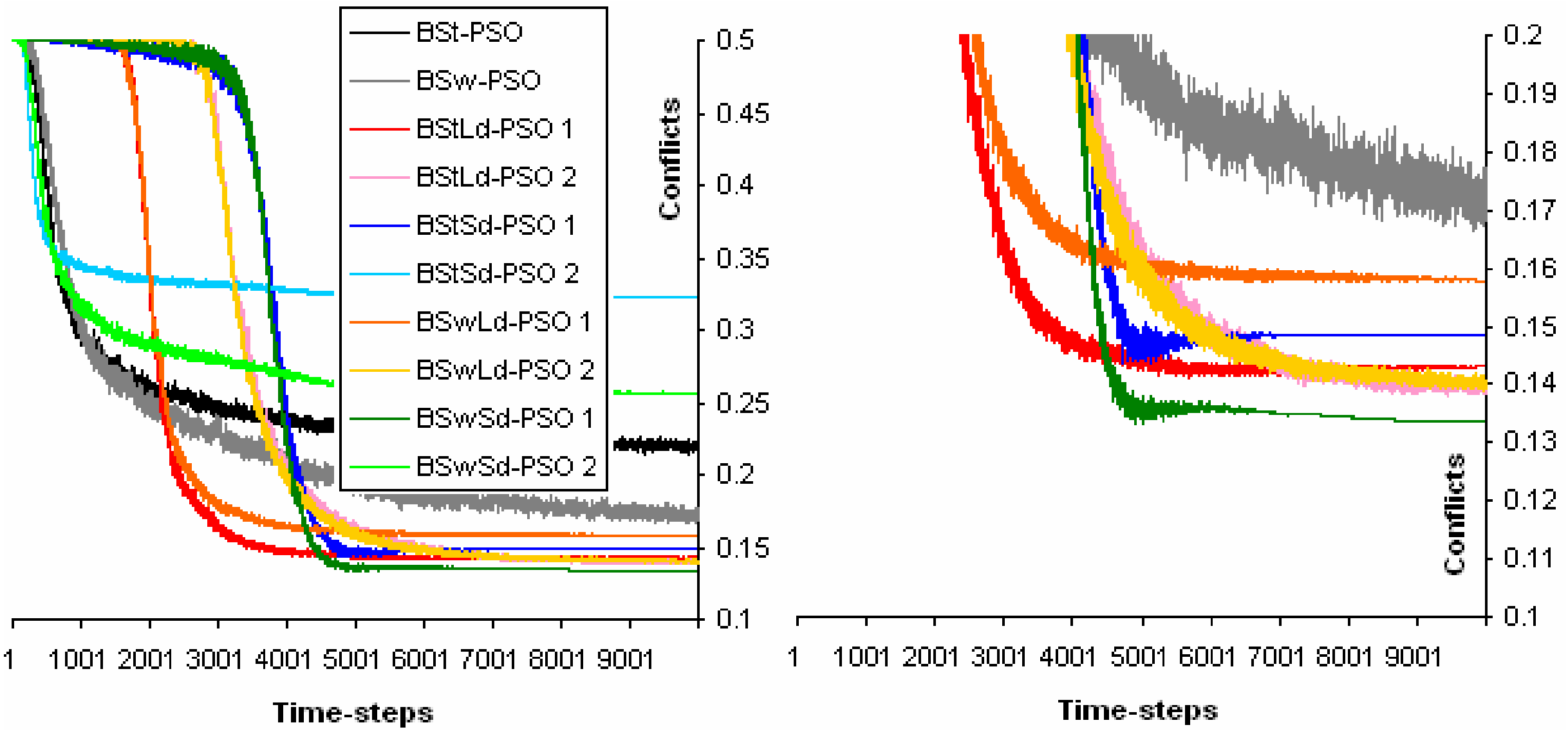

**Fig. 6. 45**: Evolution of the mean average conflicts found by 10 different optimizers for the Schaffer f6 benchmark test function along 10000 time-steps, where the average is calculated out of 30 particles, and the mean is calculated out of 50 runs, keeping the relationship $aw^{(t)} = 4.1 \cdot w^{(t)} \;\; \forall t$ .

The best mean best conflicts are found by the BSwSd-PSO 1 and the BStSd-PSO 1, which maintain inertia weights over the threshold of 0.7298 for a long period of time. Nevertheless, it seems that maintaining the inertia and acceleration weights unrelated is a better strategy if this function is to be optimized by means of stand-alone optimizers: fewer failures in attaining the error condition occur, and no stagnation of the evolution of the mean best conflicts is observed (refer to **Table 6. 7** and **Fig. 6. 32**). Regarding the clustering behaviour, all the pure C-PSOs exhibit the fastest initial clustering of their particles. However, while the BSt-PSO and the BSw-PSO are the fastest in attaining the error condition, their particles are not able to complete the implosion. In contrast, the BStSd-PSO 2 and the BSwSd-PSO 2 fail to attain the error condition every time because of the premature complete implosion of their particles.

### 6.4.2.2 Polynomial relationship

It was shown in **Fig. 6. 36** that setting $w$ = 0.5 for $aw$ = 4 enhances the convergence of a particle towards the weighted average of the two best locations it is attracted to. It is possible to develop a similar analysis for other values of *aw*, so that the behaviour can be maintained for time-varying inertia weights. Therefore, four other analyses similar to the one shown in **Fig. 6. 36** are performed, obtaining the table shown in **Fig. 6. 46** (left). A function that relates *aw* and *w* was obtained by a polynomial interpolation of the values in the table, as shown in **Fig. 6. 46** (right):

| *aw* | *w* |
|---|---|
| 2 | 0 |
| 3 | 1/6 |
| 4 | 1/2 |
| 5 | 9/10 |
| 6 | 4/3 |

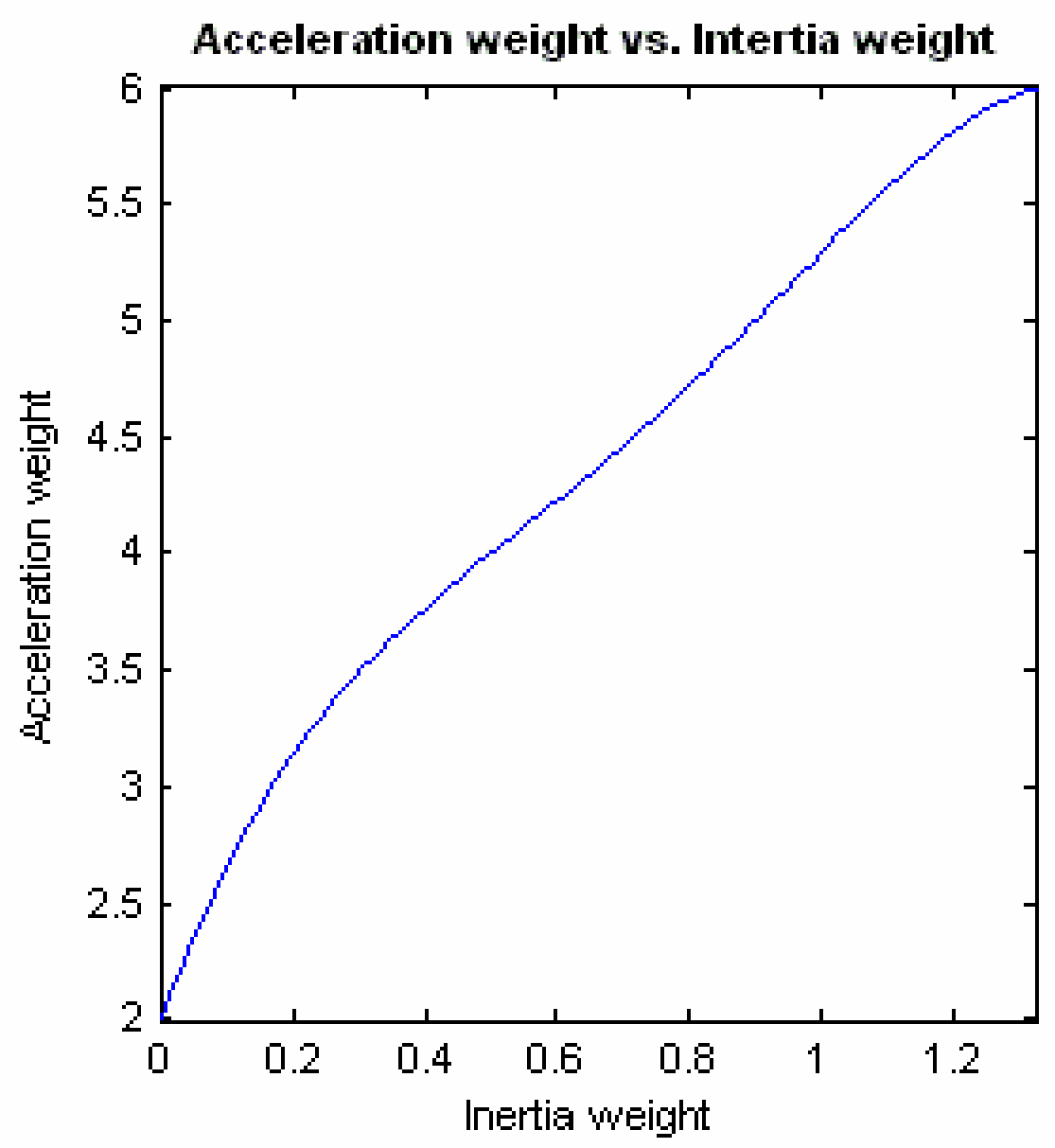

**Fig. 6. 46**: Table (left) and interpolated 4-degree polynomial (right) that relates the acceleration and the inertia weights so as to favour fast clustering.

Bear in mind that this polynomial relationship aims to favour fast clustering, which is not typically a robust strategy for stand-alone optimizers. However, a meta-swarm composed of different specialized swarms such as the ones analyzed within this section can be thought of, which might result in a single optimizer gathering in the beneficial features of different strategies and of different tunings of the parameters. This is proposed in a later chapter.

Thus, the same optimizers tested in section **6.4.1** and in section **6.4.2** are tested hereafter, but now keeping the acceleration and the inertia weights related like the polynomial shown in equation **(6. 17)**:

$$aw^{(t)} = -4.14185814185814 \cdot \left(w^{(t)}\right)^4 + 12.398001998002 \cdot \left(w^{(t)}\right)^3 + \\ -12.76966366966367 \cdot \left(w^{(t)}\right)^2 + 7.80306360306360 \cdot w^{(t)} + 2 = p\left(w^{(t)}\right) \qquad \textbf{(6. 17)}$$

In the same fashion the inertia weight used in the BSt-PSO was slightly modified so as to become the CSt-PSO in section **6.4.2.1**, the one used here for the BSt-PSO is also modified to enhance fast clustering. Thus, the inertia weight is set to a value of 0.5 because a value of 0.7 would imply too high learning weights which, as argued before, generally make the particles' clustering more difficult despite the relationship between the inertia and acceleration weights.

The evolution of the inertia, individuality, and sociality weights for the BSwLd-PSO 1 and the BSwSd-PSO 1—keeping the polynomial relationship between $w$ and $aw$ shown in equation **(6. 17)** —is shown in **Fig. 6. 47**:

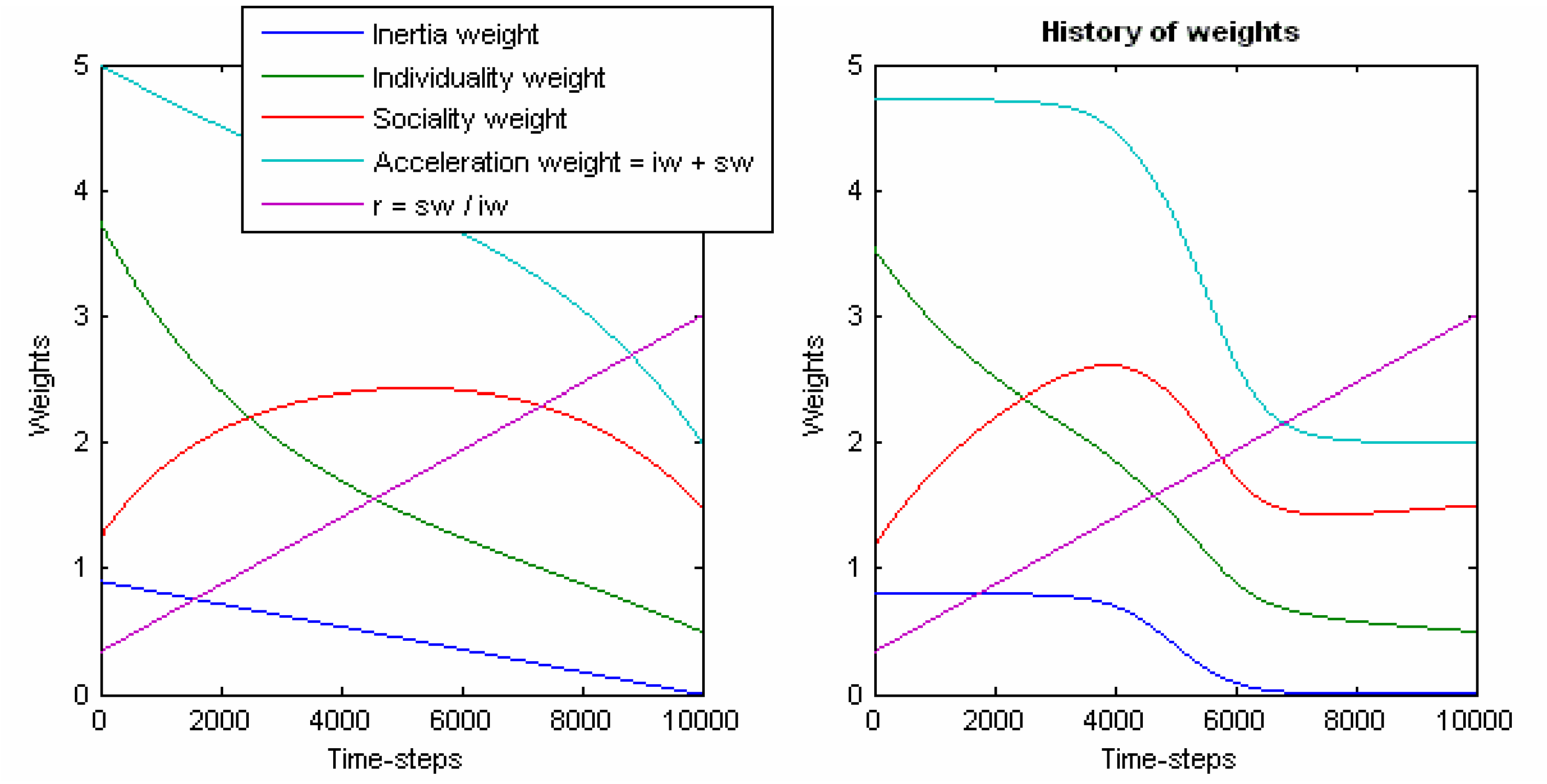

**Fig. 6. 47**: Evolution of the weights for the BSwLd-PSO 1 (left) and the BSwSd-PSO 1 (right), keeping a polynomial relationship between the inertia and acceleration weights that favours clustering.

The details of the optimizers tested within this section are as follows:

- BSt-PSO: basic, standard PSO:
  $w^{(t)} = 0.5,\ iw^{(t)} = sw^{(t)} \ \wedge \ aw^{(t)} = p\left(w^{(t)}\right) \ \forall t$
- BSw-PSO: basic PSO with linearly time-swapping learning weights:
  $w^{(t)} = 0.5 \ \ \forall t\ ,\ r^{(1)} = \frac{sw^{(1)}}{iw^{(1)}} = \frac{1}{3}\ ,\ r^{(10000)} = \frac{sw^{(10000)}}{iw^{(10000)}} = 3\ ,\ aw^{(t)} = p\left(w^{(t)}\right) \ \forall t$
- BStLd-PSO 1: basic, standard PSO with linearly time-decreasing inertia weight:
  $w^{(1)} = 0.9\ ,\ w^{(10000)} = 0\ ,\ iw^{(t)} = sw^{(t)} \ \wedge \ aw^{(t)} = p\left(w^{(t)}\right) \ \forall t$
- BStLd-PSO 2: basic, standard PSO with linearly time-decreasing inertia weight:
  $w^{(1)} = 0.9\ ,\ w^{(10000)} = 0.4\ ,\ iw^{(t)} = sw^{(t)} \ \wedge \ aw^{(t)} = p\left(w^{(t)}\right) \ \forall t$

- BStSd-PSO 1: basic, standard PSO with sigmoidly time-decreasing inertia weight:
  $w^{(1)} \to 0.8,\ w^{(10000)} \to 0,\ iw^{(t)} = sw^{(t)} \ \wedge \ aw^{(t)} = p\left(w^{(t)}\right)\ \forall t$
- BStSd-PSO 2: basic, standard PSO with sigmoidly time-decreasing inertia weight:
  $w^{(1)} \to 0.7,\ w^{(10000)} \to 0.4,\ iw^{(t)} = sw^{(t)} \ \wedge \ aw^{(t)} = p\left(w^{(t)}\right)\ \forall t$
- BSwLd-PSO 1: basic PSO with linearly time-swapping learning weights and linearly time-decreasing inertia weight:
  $w^{(1)} = 0.9,\ w^{(10000)} = 0,\ r^{(1)} = \frac{1}{3},\ r^{(10000)} = 3,\ aw^{(t)} = p\left(w^{(t)}\right)\ \forall t$
- BSwLd-PSO 2: basic PSO with linearly time-swapping learning weights and linearly time-decreasing inertia weight:
  $w^{(1)} = 0.9,\ w^{(10000)} = 0.4,\ r^{(1)} = \frac{1}{3},\ r^{(10000)} = 3,\ aw^{(t)} = p\left(w^{(t)}\right)\ \forall t$
- BSwSd-PSO 1: basic PSO with linearly time-swapping learning weights and sigmoidly time-decreasing inertia weight:
  $w^{(1)} = 0.8,\ w^{(10000)} = 0,\ r^{(1)} = \frac{1}{3},\ r^{(10000)} = 3,\ aw^{(t)} = p\left(w^{(t)}\right)\ \forall t$
- BSwSd-PSO 2: basic PSO with linearly time-swapping learning weights and sigmoidly time-decreasing inertia weight:
  $w^{(1)} = 0.7,\ w^{(10000)} = 0.4,\ r^{(1)} = \frac{1}{3},\ r^{(10000)} = 3,\ aw^{(t)} = p\left(w^{(t)}\right)\ \forall t$

The performance of each of the 10 optimizers on each of the 6 benchmark test functions is shown in **Table 6. 14** to **Table 6. 19**, and in **Fig. 6. 48** to **Fig. 6. 54**.

#### 6.4.2.2.1 SPHERE function

The results of the experiments for the optimization of this function are gathered in **Table 6. 14**, and the evolution of the mean best conflict found by each optimizer is plotted in **Fig. 6. 48**.

In the same fashion as when optimizing this function with optimizers whose $aw^{(t)} = 4.1 \cdot w^{(t)}$ (refer to section **6.4.2.1.1**), the BSt-PSO and the BSw-PSO tested here, whose $aw^{(t)} = p\left(w^{(t)}\right)$, find a remarkably good and a remarkably bad mean best conflicts, respectively[18]. The reason for this is also the same as in the case of the constant relationship: the outstandingly fast clustering of the particles.

[18] Note that, while the BSt-PSO and the BSw-PSO share the same values of the inertia and the acceleration weights, the polynomial relationship is set between them, regardless of the relationship between the individuality and the sociality weights.

| SPHERE | $\overline{cgbest}\,(\sigma)$ | $cgbest$ | $\overline{tsec}\,(\sigma)$ | $\overline{tsgb}\,(\sigma)$ | $nf$ | $ngb$ |
|---|---|---|---|---|---|---|
| BSt-PSO | **2.74916890E-129** (1.94150924E-128) | **2.92190900E-140** | 438.24 (30.50) | - (-) | 0 | 0 |
| BSw-PSO | **1.01053255E+02** (1.97068279E+02) | **4.28972636E-02** | - (-) | - (-) | 50 | 0 |
| BStLd-PSO 1 | **2.40750911E-24** (1.42188383E-23) | **2.66391083E-36** | 4096.14 (53.95) | - (-) | 0 | 0 |
| BStLd-PSO 2 | **1.15249028E-43** (3.73032744E-43) | **2.60632755E-57** | 6872.72 (74.06) | - (-) | 0 | 0 |
| BStSd-PSO 1 | **6.67434714E-05** (4.12569456E-04) | **6.05555318E-09** | 4899.70 (23.31) | - (-) | 0 | 0 |
| BStSd-PSO 2 | **4.90424192E-19** (3.28814094E-18) | **9.77435994E-42** | 5288.52 (34.09) | - (-) | 0 | 0 |
| BSwLd-PSO 1 | **3.75369067E-34** (1.64523788E-33) | **2.06285049E-44** | 4600.96 (59.69) | - (-) | 0 | 0 |
| BSwLd-PSO 2 | **2.76253733E-12** (6.95390117E-12) | **1.86803007E-15** | 8554.66 (113.54) | - (-) | 0 | 0 |
| BSwSd-PSO 1 | **1.32330288E-05** (9.09886479E-05) | **9.92196220E-12** | 5116.62 (23.46) | - (-) | 0 | 0 |
| BSwSd-PSO 2 | **4.04238292E-43** (2.77274573E-42) | **6.69651767E-49** | 5785.58 (52.21) | - (-) | 0 | 0 |

**Table 6. 14**: Performance of 10 algorithms when optimizing the 30-dimensional Sphere benchmark test function along 10000 time-steps, where the particles are initially randomly spread over the region $[-100,100]^{30}$, the inertia and acceleration weights are related like a 4th degree polynomial, and the statistical data are calculated out of 50 runs.

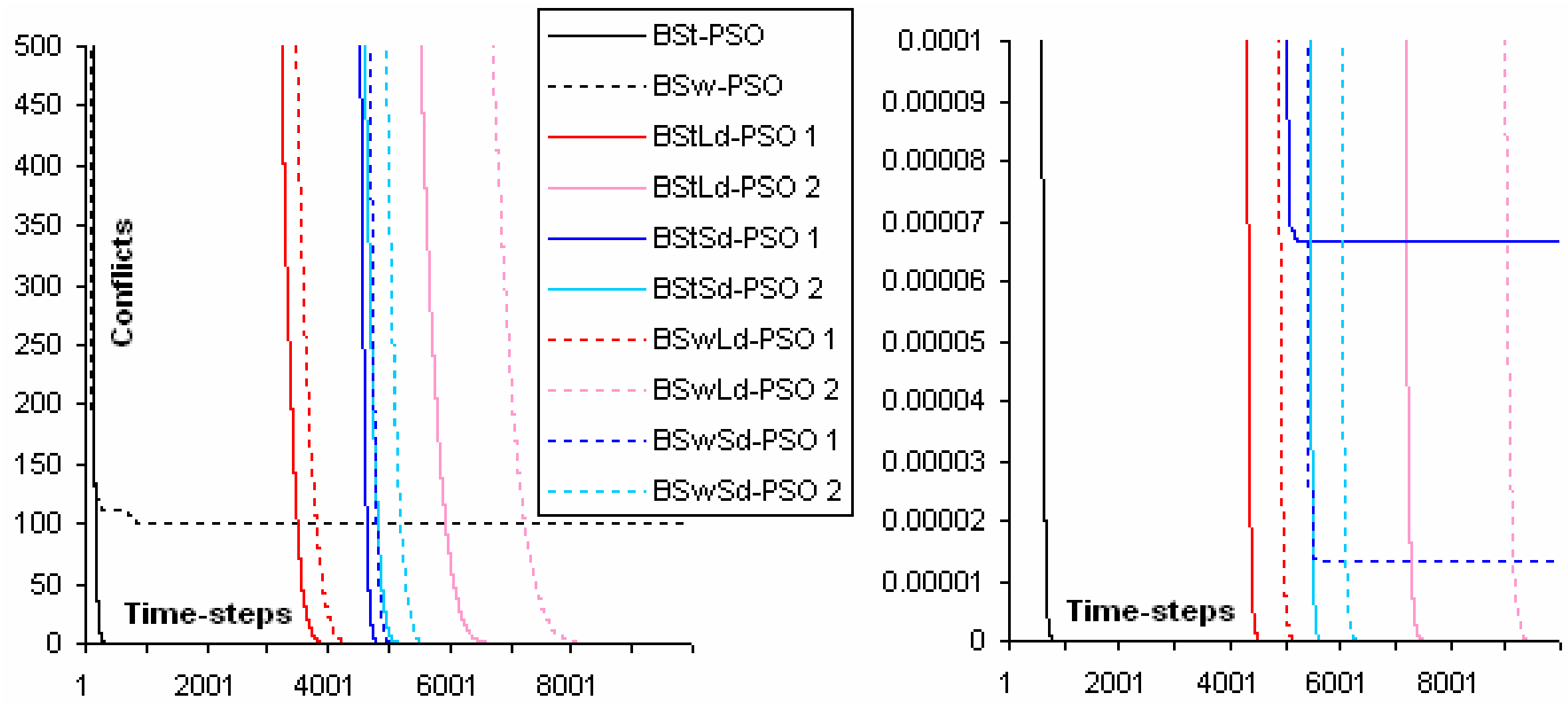

**Fig. 6. 48**: Evolution of the mean best conflicts found by 10 different optimizers for the Sphere benchmark test function out of 50 runs along 10000 time-steps, for a polynomial relationship between the inertia and acceleration weights that favours clustering.

Note that, while the mean best conflict found by the BSw-PSO with a constant relationship is worth a value of around 0.1 (refer to **Table 6. 8**), the one found here is worth a value of around 100, failing to attain the error condition along each of the 50 runs. It is important to note that, in spite of the differences in the mean best conflicts found by the BSt-PSO and the BSw-PSO here, they both present a quick flattening of the curves of the evolution of their mean best conflicts. In contrast, the rest of the optimizers take much longer to reach such stagnation[19]. This is consistent with previous conclusions regarding the influence of the learning weights: the greater they are the less strong the clustering ability that the particles exhibit[20].

It was conjectured before that the reason why the BSt-PSO with $aw^{(t)} = 4.1 \cdot w^{(t)} \;\; \forall t$—which finds the best mean best conflict among all the optimizers tested so far—fails to find the exact global optimum is the premature complete clustering of its particles[21]. The same conjecture stands here, where the BSt-PSO finds the second best one. Given that they both manage to find outstanding mean best conflicts in spite of the quick flattening of the curves of their evolution, it also appears that they present the best compromise between exploration and exploitation abilities for the optimization of this function.

It should be noted that the corresponding curves of the evolution of the mean best and of the mean average conflicts found by each of the optimizers with time-decreasing inertia weights end up merging and stagnating during the late stages of the search. This is consistent with the fact that these optimizers keep a polynomial relationship between the acceleration and the inertia weights, and the learning weights are smaller than 2 during the late stages of the search. Note that both features favour clustering.

#### 6.4.2.2.2 ROSENBROCK function

The results of the experiments for the optimization of this function are gathered in **Table 6. 15**, and the evolution of the mean best conflict found by each optimizer is plotted in **Fig. 6. 49**:

[19] The same is true with regards to the number of time-steps required to attain the error condition, except that the BSw-PSO never attains it due to premature clustering.

[20] Beware that all the optimizers with time-decreasing inertia weights present acceleration weights greater than four during the early stages of the search.

[21] By successively zooming in the graph of the evolution of the mean average conflicts in **Appendix 4**, it can be observed that, although the slopes of the curves of the evolution of its mean best and mean average conflicts are extremely small, they are not strictly zero. It is theorized here that this noticeably small improvement is due to the exploration that the inertia weight still induces, but with the whole swarm behaving very much like a single particle. That is to say that, given that the complete clustering of the particles does not take place at a local optimum, and since the inertia weight is not null, slight improvement is still possible. Nevertheless, the rate is very low.

| ROSENBROCK | $\overline{cgbest}\ (\sigma)$ | $cgbest$ | $\overline{tsec}\ (\sigma)$ | $\overline{tsgb}\ (\sigma)$ | $nf$ | $ngb$ |
|---|---|---|---|---|---|---|
| BSt-PSO | **2.17821562E+01** (2.14489039E+01) | **3.12015304E-01** | 974.52 (1249.21) | - - | 0 | 0 |
| BSw-PSO | **4.14945140E+06** (1.81077834E+06) | **7.61432261E+05** | - - | - - | 50 | 0 |
| BStLd-PSO 1 | **4.27924430E+01** (3.55658978E+01) | **6.55644451E-02** | 4385.91 (476.54) | - - | 7 | 0 |
| BStLd-PSO 2 | **4.81254411E+01** (5.61675510E+01) | **2.87283414E-05** | 7318.93 (542.92) | - - | 7 | 0 |
| BStSd-PSO 1 | **1.29323633E+02** (1.60353178E+02) | **1.58845756E+01** | 4941.94 (88.47) | - - | 17 | 0 |
| BStSd-PSO 2 | **3.96984585E+01** (3.30659519E+01) | **4.03367492E-01** | 5462.23 (350.94) | - - | 2 | 0 |
| BSwLd-PSO 1 | **2.75702177E+01** (3.16012785E+01) | **1.06563250E-01** | 4880.57 (531.14) | - - | 1 | 0 |
| BSwLd-PSO 2 | **9.11102966E+01** (1.54776011E+02) | **8.20915685E-02** | 8719.02 (484.52) | - - | 8 | 0 |
| BSwSd-PSO 1 | **6.44306569E+01** (5.48436569E+01) | **9.14066981E+00** | 5171.93 (153.33) | - - | 5 | 0 |
| BSwSd-PSO 2 | **1.86456170E+01** (1.96767627E+01) | **1.40927428E-02** | 6114.14 (803.88) | - - | 0 | 0 |

**Table 6. 15**: Performance of 10 algorithms when optimizing the 30-dimensional Rosenbrock benchmark test function along 10000 time-steps, where the particles are initially randomly spread over the region $[-30,30]^{30}$, the inertia and the acceleration weights are related like a 4$^{th}$ degree polynomial, and the statistical data are calculated out of 50 runs.

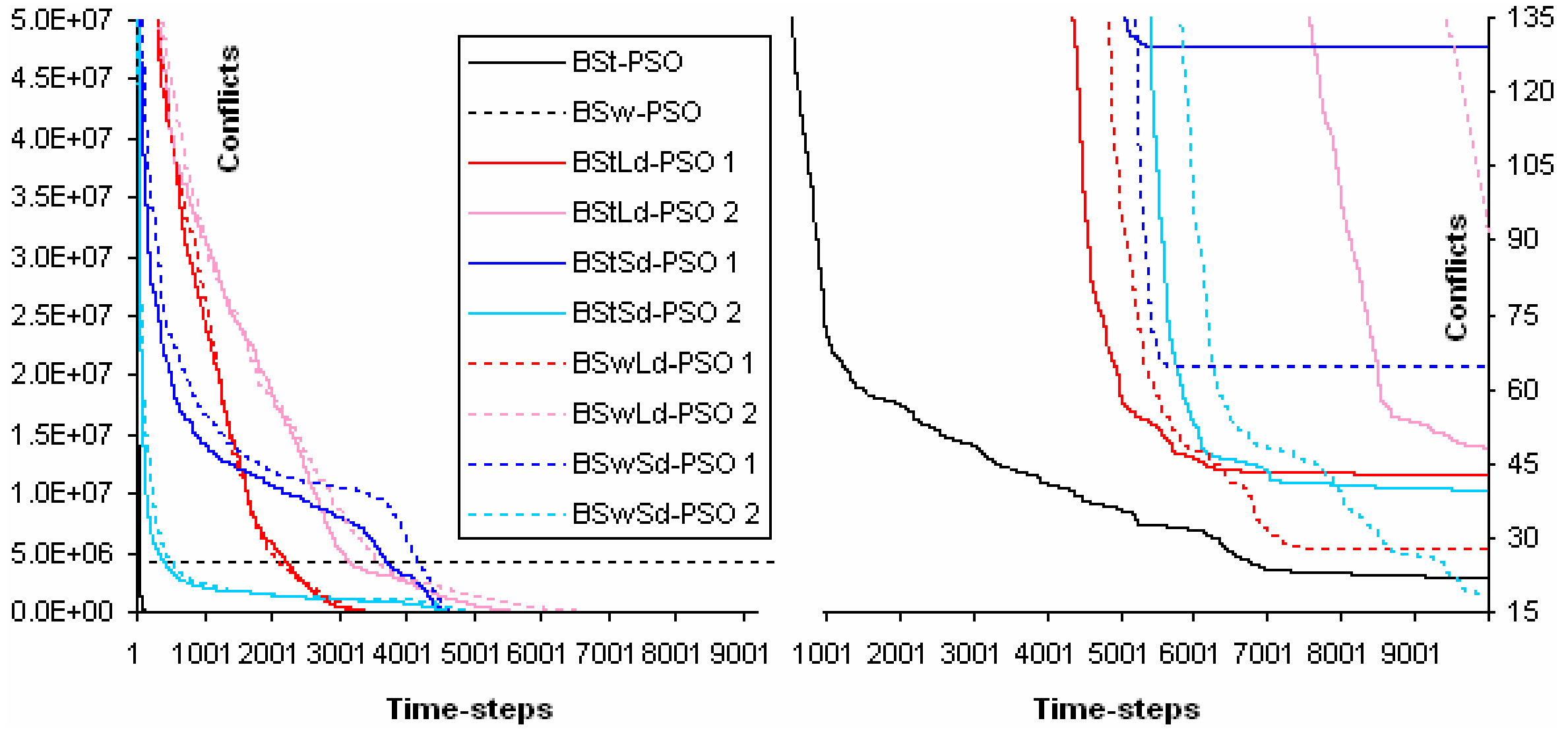

**Fig. 6. 49**: Evolution of the mean best conflicts found by 10 different optimizers for the Rosenbrock benchmark test function out of 50 runs along 10000 time-steps, for a polynomial relationship between the inertia and acceleration weights that favours clustering.

Note that, while the mean best conflict found by the BSt-PSO with polynomial relationship is the second best found by any optimizer tested so far along the whole of **Chapter 6** for the optimization of the Sphere function, the ones found by the BSwSd-PSO 2 and the BSt-PSO are the second and third best, respectively, for the optimization of the Rosenbrock function.

In the same fashion as when optimizing the Sphere function with the polynomial relationship between the inertia and the acceleration weights (see **Table 6. 14** in section **6.4.2.2.1**), and when optimizing the Rosenbrock function keeping them unrelated (see **Table 6. 3** in section **6.4.1.7.2**) and related like $aw^{(t)} = 4.1 \cdot w^{(t)}$ (see **Table 6. 9** in section **6.4.2.1.2**), the BSw-PSO tested here finds the worst mean best conflict, failing to attain the error condition every time. It can be inferred from the observation of the curves of the evolution of the mean best and mean average conflicts that this is also due to premature clustering (refer to **Fig. 6. 49** and **Appendix 4**). Notice that not only is the mean best conflict found by the BSw-PSO the worst one here, but it is also a very bad one, which is way above the error condition!

The behaviour of the BSt-PSO here is very similar to that of the BSt-PSO with the constant relationship: the particles are not capable of fine-clustering; the curve of the evolution of the mean best conflict does not stagnate; and the trend-line of the evolution of the mean average conflict is surprisingly increasing after around 500 time-steps, as if the particles were diverging. The reason for this is not understood. However, it manages to find the second best mean best conflict, and to be the fastest, by far, in attaining the error condition.

As to the other optimizers, the ones with inertia weights that approach zero exhibit the virtually complete implosion of their particles, which can be inferred from the fact that the corresponding curves of the evolution of their mean average and mean best conflicts merge. Although this does not strictly imply that a complete clustering takes place, the odds that the average of the conflicts of all the particles is equal to the best conflict found—and that its evolution stagnates—are indeed negligible. Furthermore, these results are calculated as an average out of 50 runs. In contrast, the optimizers with time-decreasing inertia weights that do not approach zero do not exhibit this outstanding clustering behaviour, which is probably the reason why the curves of the evolution of their mean best conflicts do not completely flatten.

To sum up, the best optimizers with regards to the best solution found are the BSwSd-PSO 2, the BSt-PSO, and the BSwLd-PSO 1, which also outperform all the optimizers with inertia and acceleration weights kept unrelated, and most of the ones with a constant relationship. However, given that this polynomial relationship aims to favour clustering rather than to find the best conflict, it appears that the BSw-PSO and the optimizers with time-decreasing inertia weights that approach zero behave more accordingly with what was aimed for with this strategy. In fact, the BSw-PSO, the BStSd-PSO 1, and the BSwSd-PSO 1 with polynomial relationship—together with the BSwSd-PSO 2 with constant relationship—exhibit the most prominent clustering behaviours among all the optimizers tested within this chapter for the optimization of this function (refer to **Fig. 6. 28**, **Fig. 6. 40**, **Fig. 6. 49**, and **Appendix 4**).

### 6.4.2.2.3 RASTRIGRIN function

The results of the experiments for the optimization of this function are gathered in **Table 6. 16**:

| **RASTRIGRIN** | $\overline{cgbest}\,(\sigma)$ | $cgbest$ | $\overline{tsec}\,(\sigma)$ | $\overline{tsgb}\,(\sigma)$ | $nf$ | $ngb$ |
|---|---|---|---|---|---|---|
| BSt-PSO | **4.77579585E+01** (1.18936954E+01) | **2.28840482E+01** | 152.70 (42.58) | - - | 0 | 0 |
| BSw-PSO | **3.50424182E+01** (7.84020108E+00) | **1.59193399E+01** | 200.10 (59.51) | - - | 0 | 0 |
| BStLd-PSO 1 | **3.48036769E+01** (9.85669855E+00) | **1.39294167E+01** | 3553.20 (207.49) | - - | 0 | 0 |
| BStLd-PSO 2 | **2.63862930E+01** (7.36986720E+00) | **1.39294268E+01** | 6048.64 (308.07) | - - | 0 | 0 |
| BStSd-PSO 1 | **4.43577730E+01** (1.10283124E+01) | **2.28840629E+01** | 4681.12 (71.82) | - - | 0 | 0 |
| BStSd-PSO 2 | **3.36693799E+01** (1.02601834E+01) | **1.59193399E+01** | 4941.64 (165.03) | - - | 0 | 0 |
| BSwLd-PSO 1 | **2.77593316E+01** (6.99003292E+00) | **1.29344677E+01** | 3724.96 (192.04) | - - | 0 | 0 |
| BSwLd-PSO 2 | **1.67153258E+01** (6.17597437E+00) | **4.97540231E+00** | 6885.00 (384.74) | - - | 0 | 0 |
| BSwSd-PSO 1 | **4.75987937E+01** (1.27453934E+01) | **2.78592282E+01** | 4756.66 (73.93) | - - | 0 | 0 |
| BSwSd-PSO 2 | **2.92517650E+01** (8.46399370E+00) | **1.59193449E+01** | 5131.08 (122.51) | - - | 0 | 0 |

**Table 6. 16**: Performance of 10 algorithms when optimizing the 30-dimensional Rastrigrin benchmark test function along 10000 time-steps, where the particles are initially randomly spread over the region $\left[-5.12,5.12\right]^{30}$, the inertia and the acceleration weights are related like a 4$^{th}$ degree polynomial, and the statistical data are calculated out of 50 runs.

Although the best results obtained are not the most important aspect here, it is fair to note that the mean best conflict found by the BSwLd-PSO 2 happen to be the best one found by any of the optimizers tested within this chapter. Given that the Rastrigrin function exhibits numerous local optima spread all over the search-space, the high learning weights that the BStLd-PSO 2 and the BSwLd-PSO 2 have at the beginning of the search ( $w^{(1)} = 0.9 \Rightarrow aw^{(1)} = 5$, as shown in **Fig. 6. 46**)—with a rate of decrease smaller than that of their corresponding "PSO 1" versions—enables them to escape more local optima, while the polynomial relationship favours exploitation, especially when the learning weights take smaller values (i.e. during the late stages of the search). Nevertheless, they are the only optimizers whose particles do not perform a complete implosion, and the ones which take longer to attain the error condition.

The evolution of the mean best conflict found by each optimizer is shown in **Fig. 6. 50**:

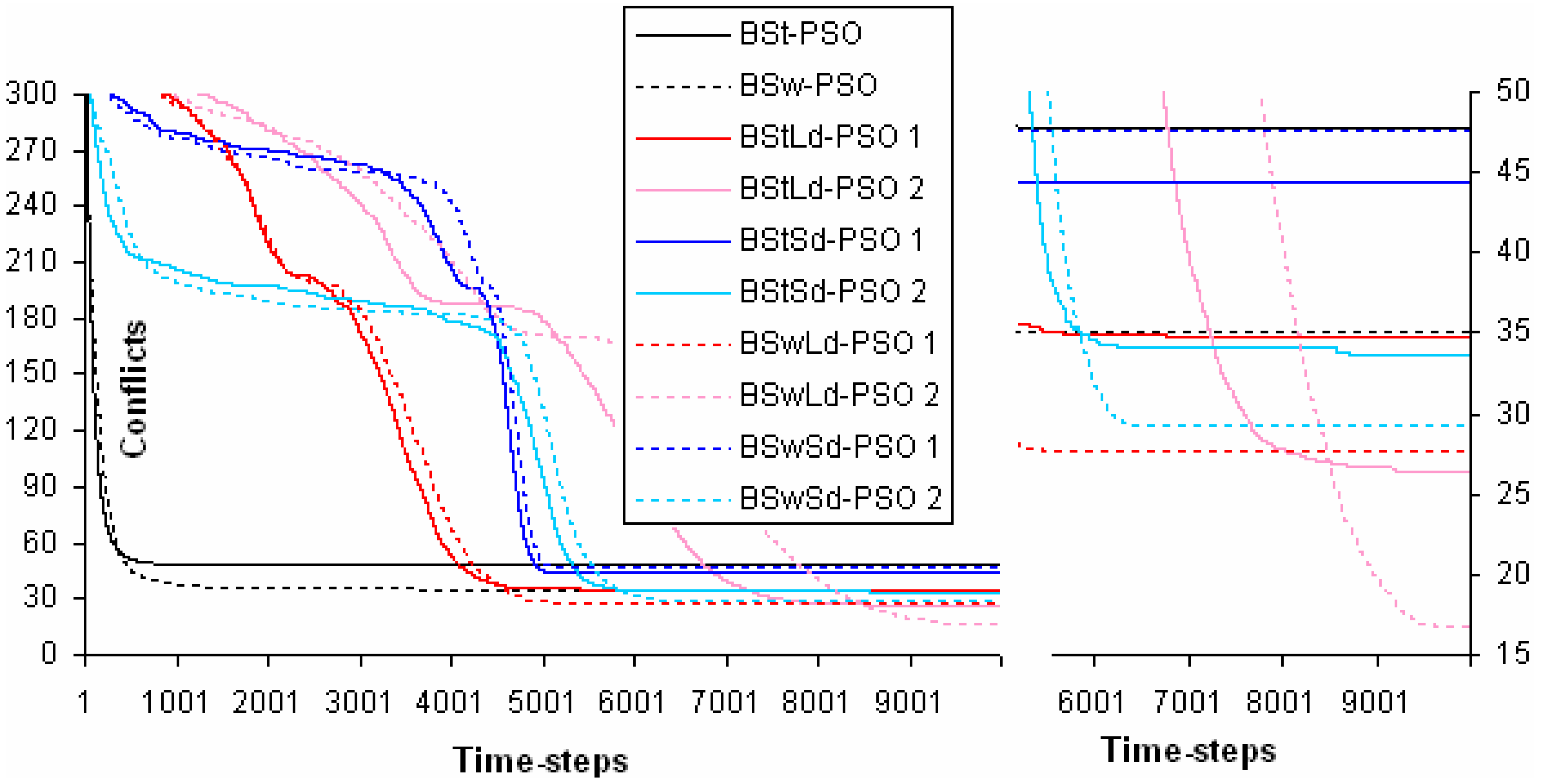

**Fig. 6. 50**: Evolution of the mean best conflicts found by 10 different optimizers for the Rastrigrin benchmark test function out of 50 runs along 10000 time-steps, for a polynomial relationship between the inertia and acceleration weights that favours clustering.

The particles of the BSt-PSO and the BSw-PSO cluster very soon, so that the curves of the evolution of their mean best conflicts flatten very quickly. In fact, they take less than 201 time-steps to attain the error condition, while the BStLd-PSO 2 and the BSwLd-PSO 2 take over 6000 time-steps. Furthermore, the initial individuality of the BSw-PSO higher than that of the BSt-PSO enables the former to escape more local optima before stagnating, so that it is able to find a better best mean best conflict. The particles of the other optimizers also perform a virtually complete implosion, although they take noticeably longer than the particles of the optimizers with constant acceleration weights to do so.

To summarize, the best optimizer with regards to the best mean best conflict found is the BSwLd-PSO 2, while the best ones with regards to the ability to cluster are the BSt-PSO and the BSw-PSO. It can also be concluded that the swapping strategy helps to escape local optima, and that the polynomial relationship—together with acceleration weights smaller than 2—favours clustering. Besides, since the "PSO 1" versions exhibit stronger clustering ability than the corresponding "PSO 2" versions here, it seems that the polynomial relationship together with inertia weights that approach zero towards the end of the search also enhances the swarms' ability to cluster.

### 6.4.2.2.4 GRIEWANK function

The results of the experiments for the optimization of this function are gathered in **Table 6. 17**, and the evolution of the mean best conflict found by each optimizer is plotted in **Fig. 6. 51**:

| **GRIEWANK** | $\overline{cgbest}\,(\sigma)$ | $cgbest$ | $\overline{tsec}\,(\sigma)$ | $\overline{tsgb}\,(\sigma)$ | $nf$ | $ngb$ |
|---|---|---|---|---|---|---|
| BSt-PSO | **1.11115336E-02** (1.50303083E-02) | **0.00000000E+00** | 399.70 (39.91) | 1496.78 (88.00) | 0 | 23 |
| BSw-PSO | **1.27579628E+00** (1.21927528E+00) | **0.00000000E+00** | 487.86 (276.71) | 2407.00 - | 43 | 1 |
| BStLd-PSO 1 | **1.12704067E-02** (1.29144361E-02) | **0.00000000E+00** | 4029.66 (53.12) | 5126.65 (415.30) | 0 | 20 |
| BStLd-PSO 2 | **1.58387559E-02** (1.68919367E-02) | **0.00000000E+00** | 6845.34 (120.68) | 8245.93 (155.00) | 0 | 15 |
| BStSd-PSO 1 | **1.34500267E-02** (1.63074624E-02) | **3.87372027E-08** | 4874.36 (27.39) | - - | 0 | 0 |
| BStSd-PSO 2 | **1.85495441E-02** (1.94292561E-02) | **0.00000000E+00** | 5252.06 (45.50) | 6233.50 (336.33) | 0 | 12 |
| BSwLd-PSO 1 | **1.73827246E-02** (2.65365220E-02) | **0.00000000E+00** | 4511.02 (75.04) | 5858.92 (100.94) | 2 | 12 |
| BSwLd-PSO 2 | **2.66620086E-02** (2.77455413E-02) | **1.38444811E-13** | 8453.23 (104.35) | - - | 2 | 0 |
| BSwSd-PSO 1 | **2.18523417E-02** (4.78408941E-02) | **1.02945419E-09** | 5076.69 (30.66) | - - | 1 | 0 |
| BSwSd-PSO 2 | **1.88589809E-02** (3.11307741E-02) | **0.00000000E+00** | 5711.78 (54.10) | 7309.41 (596.74) | 1 | 17 |

**Table 6. 17**: Performance of 10 algorithms when optimizing the 30-dimensional Griewank benchmark test function along 10000 time-steps, where the particles are initially randomly spread over the region $[-600,600]^{30}$, the inertia and the acceleration weights are related like a 4th degree polynomial, and the statistical data are calculated out of 50 runs.

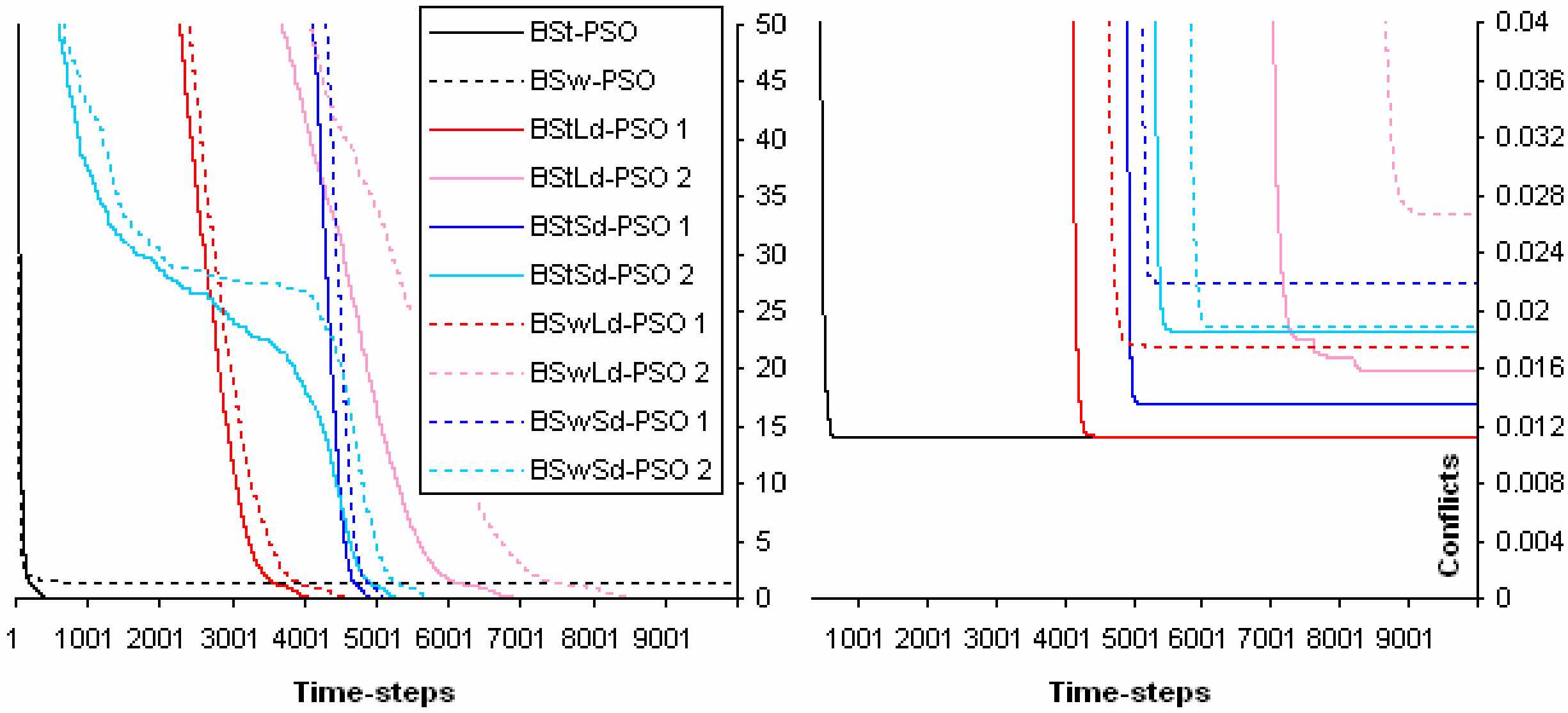

**Fig. 6. 51**: Evolution of the mean best conflicts found by 10 different optimizers for the Griewank benchmark test function out of 50 runs along 10000 time-steps, for a polynomial relationship between the inertia and acceleration weights that favours clustering.

While the aim of the polynomial relationship is to enhance the clustering ability rather than to develop an optimum compromise between the exploration and exploitation abilities, two of the optimizers tested here—namely the BSt-PSO and the BStLd-PSO 1—happen to find the best and second best mean best conflicts, respectively, among all the optimizers tested along **Chapter 6** for the optimization of this function. Besides, they are also the optimizers which find the exact global optimum a higher number of times, while never failing to attain the error condition. In contrast, the BSw-PSO finds the second worst mean best conflict[22] due to premature clustering, as it can be inferred from the observation of the evolution of its mean best and mean average conflicts in **Appendix 4**. Notice that all the optimizers are able to complete the implosion, except for the BStLd-PSO 2, which anyway exhibits very low diversity by the final time-step. However, the ones with time-decreasing inertia weights take noticeably longer both to attain the error condition and to complete the implosion due to the larger initial acceleration weights, which delay the clustering of the particles because of the resulting stronger importance given to the random weights.

In summary, the best optimizers with regards to the best solution they are able to find are the BSt-PSO and the BStLd-PSO 1, while the best ones regarding their ability to cluster are the BSw-PSO and the BSt-PSO, which take less than 500 time-steps to attain the error condition.

[22] The worst one is obtained by the BSwSd-PSO 2 with constant relationship (refer to **Table 6. 11** and **Fig. 6. 42**).

### 6.4.2.2.5 SCHAFFER F6 function (2D)

The results of the experiments for the optimization of this function are gathered in **Table 6. 18**, and the evolution of the mean best conflict found by each optimizer is plotted in **Fig. 6. 52**.

| SCHAFFER F6 2D | $\overline{cgbest}\,(\sigma)$ | $cgbest$ | $\overline{tsec}\,(\sigma)$ | $\overline{tsgb}\,(\sigma)$ | $nf$ | $ngb$ |
|---|---|---|---|---|---|---|
| BSt-PSO | **1.55454558E-03** (3.59807386E-03) | **0.00000000E+00** | 598.26 (1393.12) | 700.12 (1397.38) | 8 | 42 |
| BSw-PSO | **1.94318198E-04** (1.37403715E-03) | **0.00000000E+00** | 740.96 (1395.27) | 862.65 (1397.02) | 1 | 49 |
| BStLd-PSO 1 | **0.00000000E+00** (0.00000000E+00) | **0.00000000E+00** | 2191.32 (247.49) | 2694.86 (146.93) | 0 | 50 |
| BStLd-PSO 2 | **0.00000000E+00** (0.00000000E+00) | **0.00000000E+00** | 3334.12 (387.58) | 4310.42 (198.22) | 0 | 50 |
| BStSd-PSO 1 | **0.00000000E+00** (0.00000000E+00) | **0.00000000E+00** | 3776.72 (346.90) | 4351.66 (92.29) | 0 | 50 |
| BStSd-PSO 2 | **0.00000000E+00** (0.00000000E+00) | **0.00000000E+00** | 715.92 (487.25) | 1394.48 (543.53) | 0 | 50 |
| BSwLd-PSO 1 | **0.00000000E+00** (0.00000000E+00) | **0.00000000E+00** | 2286.14 (315.11) | 2846.60 (183.02) | 0 | 50 |
| BSwLd-PSO 2 | **0.00000000E+00** (0.00000000E+00) | **0.00000000E+00** | 3732.30 (456.97) | 5264.52 (174.61) | 0 | 50 |
| BSwSd-PSO 1 | **0.00000000E+00** (0.00000000E+00) | **0.00000000E+00** | 4090.72 (384.79) | 4589.30 (67.00) | 0 | 50 |
| BSwSd-PSO 2 | **0.00000000E+00** (0.00000000E+00) | **0.00000000E+00** | 741.40 (419.59) | 1464.32 (510.43) | 0 | 50 |

**Table 6. 18**: Performance of 10 algorithms when optimizing the 2-dimensional Schaffer f6 benchmark test function along 10000 time-steps, where the particles are initially randomly spread over the region $[-100,100]^2$, the inertia and the acceleration weights are related like a $4^{th}$ degree polynomial, and the statistical data are calculated out of 50 runs.

It has been observed along this section that this polynomial relationship effectively favours fast clustering, thus fulfilling the purpose it was designed for. Hence the BSt-PSO and the BSw-PSO exhibit a steep initial decrease in the evolution of their mean average conflicts—thus quickly losing diversity—which seems to result in a few failures in finding the exact global optimum. However, given that all the other optimizers present noticeably higher initial acceleration weights, they take much longer to diminish the mean average conflicts of their particles. This results in keeping higher diversity for a longer period of time, which seems to enable them to find the exact global optimum along each of the 50 runs. It is important to note that, although a stronger clustering ability of the optimizers appears to lead to a steeper decrease in the evolution of their mean average conflicts, only those which are capable of finding the exact global optimum appear to be able to perform a complete implosion of their particles. That is to say, they do not completely implode to a local optimum. In fact, the BSwLd-PSO 1 and the BSwSd-PSO 2 with constant relationship are the only 2 out of the 30 optimizers tested within this chapter whose particles are close to implode to a local optimum when optimizing this function (refer to **Table 6. 12**, **Fig. 6. 43**, and **Appendix 4**). It seems that when the particles of an optimizer get trapped in the area of influence of a local optimum, they cannot either escape from or fine-cluster around it. Therefore, the curve of the evolution of the mean best conflict stagnates, but it does not merge with the curve of the evolution of the mean average conflict. It is theorized here that this is due to the fact that the Schaffer f6 function presents a single-point global optimum, while each local optimum presents itself in the form of a ring-like depression (refer to **Appendix 3** for a graphical visualization).

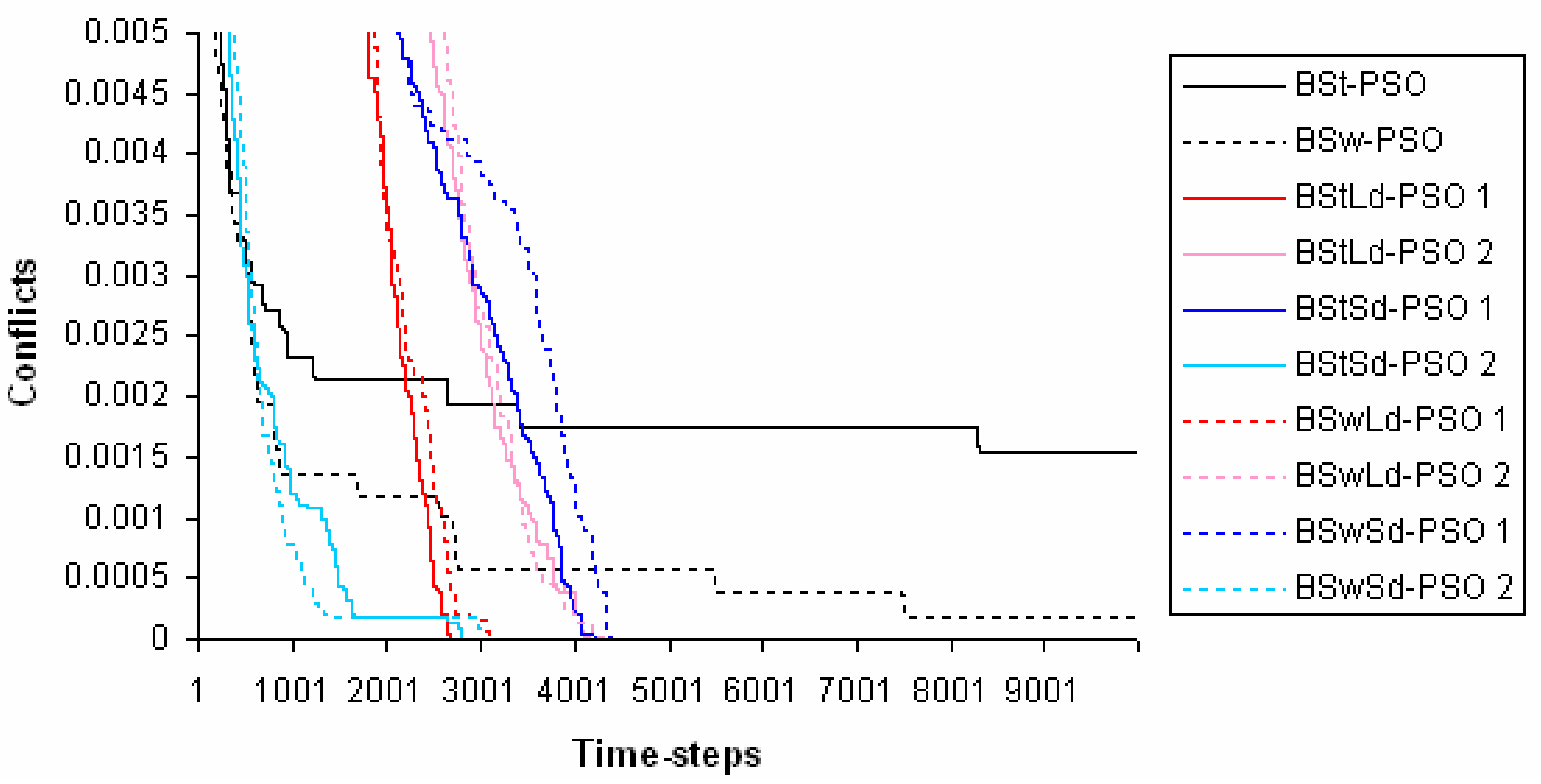

**Fig. 6. 52**: Evolution of the mean best conflicts found by 10 different optimizers for the 2-dimensional Schaffer f6 benchmark test function out of 50 runs along 10000 time-steps, for a polynomial relationship between the inertia and acceleration weights that favours clustering.

Although some of the optimizers with constant and polynomial relationships are able to find the exact global optimum along every run, and some of them are even able to complete the particles' implosion, it seems that, broadly speaking, these relationships lead the particles to lose diversity very quickly, which usually results in them getting trapped in the area of influence of some local optima they are not able to either escape from or fine-cluster around. Therefore, it seems a better strategy to keep the inertia and acceleration weights unrelated if stand-alone optimizers are to be used to deal with this function. Notice that this leads all the optimizers tested within section **6.4.1.7.5** to find the exact global optimum along each of the 50 runs (refer to **Table 6. 6**).

### 6.4.2.2.6 SCHAFFER F6 function

The results of the experiments for the optimization of this function are gathered in **Table 6. 19**, while the graphs of the evolution of the mean best and mean average conflicts found by each optimizer are plotted in **Fig. 6. 53** and **Fig. 6. 54**, respectively.

| SCHAFFER F6 | $\overline{cgbest}\,(\sigma)$ | $cgbest$ | $\overline{tsec}\,(\sigma)$ | $\overline{tsgb}\,(\sigma)$ | $nf$ | $ngb$ |
|---|---|---|---|---|---|---|
| BSt-PSO | **1.19522567E-01** (3.23589283E-02) | **7.81891821E-02** | 2261.53 (2272.33) | - - | 35 | 0 |
| BSw-PSO | **1.10652007E-01** (2.87305408E-02) | **3.72240751E-02** | 2817.17 (2303.88) | - - | 32 | 0 |
| BStLd-PSO 1 | **1.46494137E-01** (3.62091064E-02) | **7.81891821E-02** | 5379.00 (667.51) | - - | 46 | 0 |
| BStLd-PSO 2 | **1.21631346E-01** (3.23223945E-02) | **3.72240751E-02** | 8190.67 (596.45) | - - | 38 | 0 |
| BStSd-PSO 1 | **1.34551840E-01** (3.36744073E-02) | **7.81891821E-02** | 4040.75 (310.36) | - - | 42 | 0 |
| BStSd-PSO 2 | **1.43579125E-01** (3.13834422E-02) | **7.81891821E-02** | 5883.75 (457.66) | - - | 46 | 0 |
| BSwLd-PSO 1 | **1.61753666E-01** (4.46908570E-02) | **7.81891821E-02** | 4688.00 (164.37) | - - | 45 | 0 |
| BSwLd-PSO 2 | **1.46547196E-01** (3.65457969E-02) | **7.81891821E-02** | 8648.75 (197.50) | - - | 46 | 0 |
| BSwSd-PSO 1 | **2.43073706E-01** (6.43577639E-02) | **7.81891821E-02** | 5045.00 - | - - | 49 | 0 |
| BSwSd-PSO 2 | **1.72561040E-01** (4.40425817E-02) | **7.81891821E-02** | 5619.00 (0.00) | - - | 48 | 0 |

**Table 6. 19**: Performance of 10 algorithms when optimizing the 30-dimensional Schaffer f6 benchmark test function along 10000 time-steps, where the particles are initially randomly spread over the region $[-100,100]^{30}$, the inertia and the acceleration weights are related like a $4^{th}$ degree polynomial, and the statistical data are calculated out of 50 runs.

The BSt-PSO and the BSw-PSO find the best—reasonably good—mean best conflicts here, which are noticeably better than those found by the BSt-PSOs and the BSw-PSOs tested in sections **6.4.1.7.6** and **6.4.2.1.6** (see **Table 6. 7** and **Table 6. 13**). In addition, they exhibit fewer failures and reasonably small numbers of time-steps required to attain the error condition[23].

[23] Notice that none of the optimizers tested within **Chapter 6** is able to find the exact global optimum of this function. In fact, all of them present mean best conflicts which are higher than the error condition!

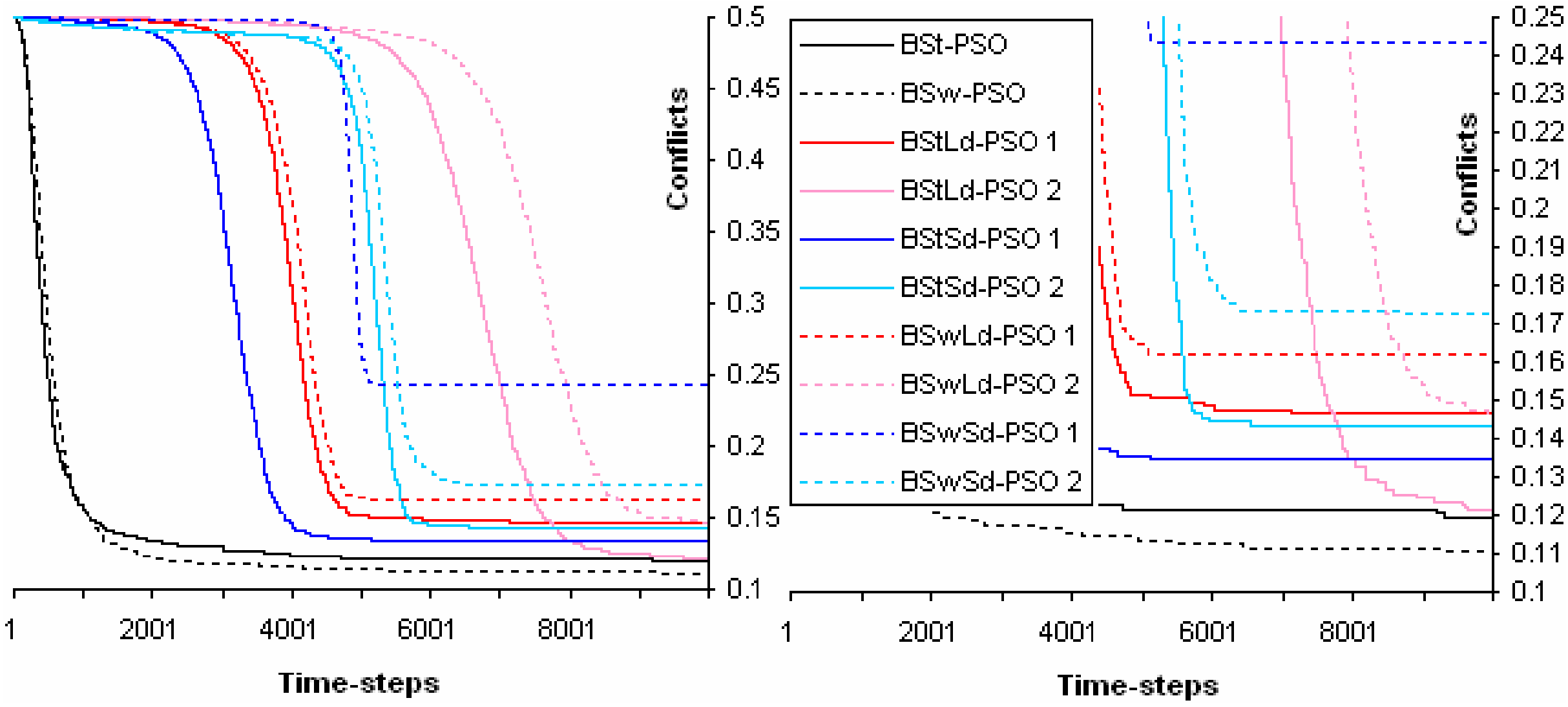

**Fig. 6. 53**: Evolution of the mean best conflicts found by 10 different optimizers for the 30-dimensional Schaffer f6 benchmark test function out of 50 runs along 10000 time-steps, for a polynomial relationship between the inertia and acceleration weights that favours clustering.

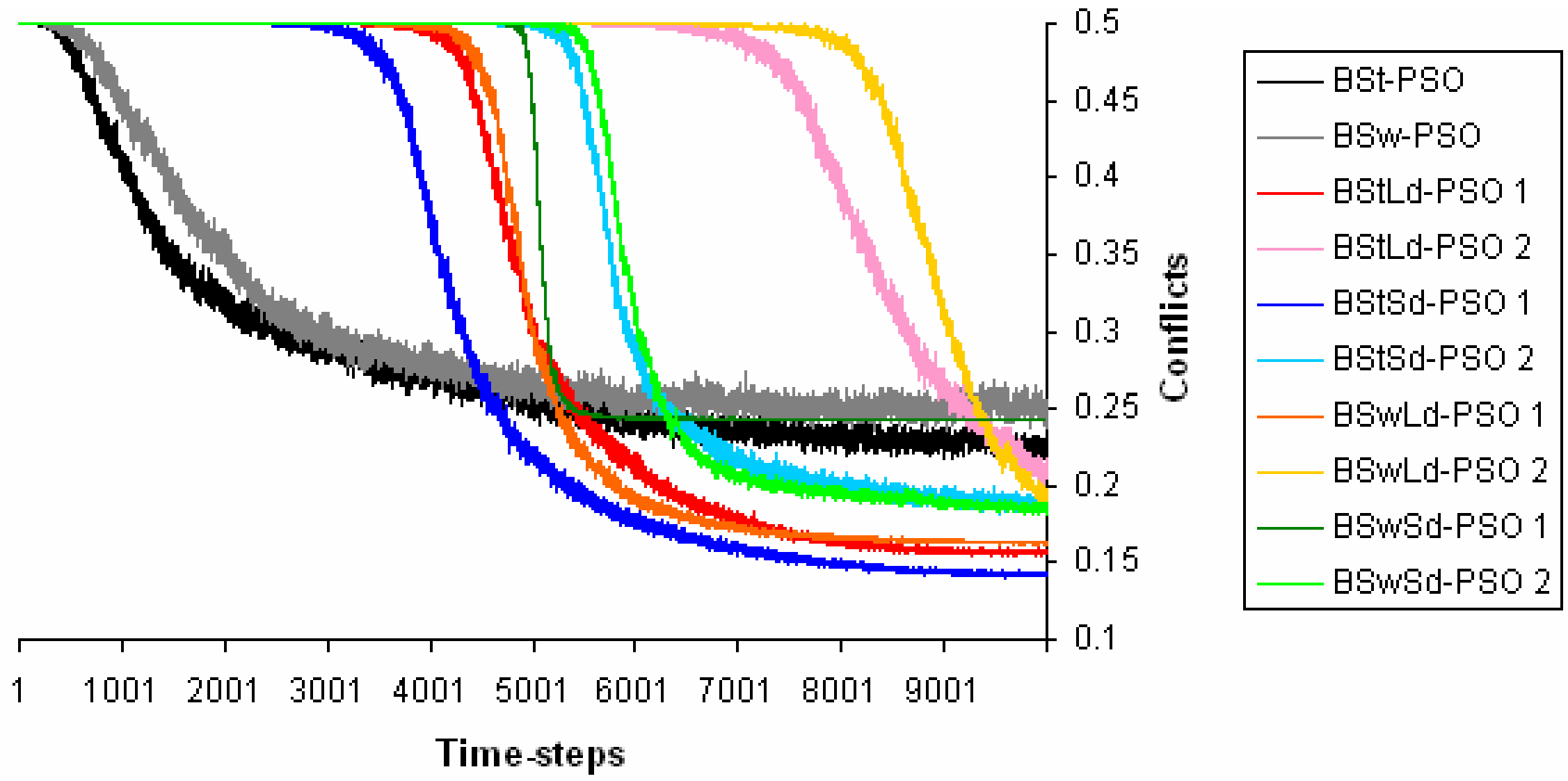

**Fig. 6. 54**: Evolution of the mean average conflicts found by 10 different optimizers for the Schaffer f6 benchmark test function along 10000 time-steps, where the average is calculated out of 30 particles, and the mean is calculated out of 50 runs, keeping a polynomial relationship between the inertia and acceleration weights that favours clustering.

In the same fashion as when optimizing this function in section **6.4.2.1.6** (see **Fig. 6. 44**), and as opposed to when doing so in section **6.4.1.7.6** (see **Fig. 6. 32**), the curves of the evolution of the mean best conflict found by each optimizer with polynomial relationship stagnate during the late stages of the search, except for those obtained by the BStLd-PSO 2 and the BSwLd-PSO 2. Notice that, since the curves of the evolution of their mean best conflict and the trend-lines of the graphs of their mean average conflicts are still quite steep at the final time-step, it is reasonable to expect further improvement and further fine-clustering for a time-extended search. Notice as well that, while the BSt-PSO and the BSw-PSO find the best mean best conflicts here, they are also the ones which keep the higher diversity by the end of the search. It appears that keeping high diversity for a long period of time is critical for the optimization of this function.

With regards to the clustering behaviour, it is fair to observe that the BSwSd-PSO 1 and the BSwLd-PSO 1 are the only optimizers which exhibit a virtually complete implosion of their particles here. Nevertheless, bear in mind the conjecture in the previous section regarding the influence of the local optima that this function presents in the form of ring-like depressions.

In summary, the BSt-PSO and the BSw-PSO are among the best of the 30 optimizers tested within this chapter with regards to the mean best conflicts they are able to find when dealing with the 30-dimensional Schaffer f6 function, while the BSwSd-PSO 1 and the BSwLd-PSO 1 are among the best ones with regards to the clustering behaviour. It can also be concluded that a general clustering behaviour without a complete loss of diversity is the best strategy if stand-alone optimizers are to be used to deal with this function.

### 6.4.3 Summary of the experimental results

The fact that 30 optimizers were tested on 6 different benchmark functions along section **6.4** leads to an evident loss of perspective regarding the overall usefulness or uselessness of the different strategies used to tune the parameters of the B-PSO. Therefore, a brief review of the performances of the most distinctive optimizers when dealing with each function is carried out along sections **6.4.3.1** to **6.4.3.6**. Finally, some of the best optimizers according to three "goodness criteria" are selected in section **6.4.3.7**.

In order to shorten the reference to each optimizer, a superscript is added to the nomenclature previously used for the optimizers so as to identify which of the three strategies to relate the inertia and acceleration weights have been used. Thus, no superscript refers to no relationship between the inertia and acceleration weights, the superscript "$^{(c)}$" refers to the relationship $aw^{(t)} = 4.1 \cdot w^{(t)}$, and "$^{(p)}$" refers to the polynomial relationship shown in equation **(6. 17)**.

### 6.4.3.1 SPHERE function

Surprisingly, none of the 30 optimizers tested along the whole of **Chapter 6** is able to find the global optimum of this very simple function along any of the 50 runs, although some of them find very good solutions. However, only the BSw-PSO$^{(c)}$, the BSwSd-PSO 2$^{(c)}$, and the BSw-PSO$^{(p)}$ exhibit failures in attaining the error condition. These failures are due to the premature clustering of their particles around any suboptimal solution, which is also the reason why no optimizer is able to find the global optimum. By far, the best mean best conflicts are found by the BSt-PSO$^{(c)}$, the BSt-PSO$^{(p)}$, the BStSd-PSO 2, and the BSwLd-PSO 1. It appears that, despite the fact that they also end up stagnating due to the loss of diversity, they present the best compromise between explorative and exploitative behaviour to deal with this function.

The analysis of the clustering ability exhibited by the optimizers is two-fold. On the one hand, it is important to consider the rate of decrease of the mean best conflicts at the early stages of the search: the steeper it is the longer the optimizers can spend in fine-tuning the search. The BSt-PSO$^{(c)}$, the BSw-PSO$^{(c)}$, the BSt-PSO$^{(p)}$, and the BSw-PSO$^{(p)}$ are the best ones in this regard. On the other hand, it is important to consider the fine-clustering ability, which is helpful to fine-tune the search. The BSt-PSO$^{(c)}$, the BSw-PSO$^{(c)}$, BStSd-PSO 2$^{(c)}$, the BSwSd-PSO 2$^{(c)}$, the BSt-PSO$^{(p)}$, and the BSw-PSO$^{(p)}$ appear to be the best ones in this regard.

Another very important aspect to consider when evaluating the goodness of the optimizers' performance is their reluctance to getting trapped in local optima. However, given that the Sphere function does not exhibit local optima, this feature cannot be analyzed here.

To sum up, some of the best optimizers in dealing with this function are as follows:

- Stand-alone optimizers: BSt-PSO$^{(c)}$, BSt-PSO$^{(p)}$, BStSd-PSO 2, and BSwLd-PSO 1.
- Clustering ability: BSt-PSO$^{(c)}$, BSw-PSO$^{(c)}$, BSt-PSO$^{(p)}$, BSw-PSO$^{(p)}$, BStSd-PSO 2$^{(c)}$, and BSwSd-PSO 2$^{(c)}$.

### 6.4.3.2 ROSENBROCK function

Again, no optimizer is able to find the global optimum of this function along any of the 50 runs. However, the curves of the evolution of the mean best conflicts of most optimizers do not stagnate, so that further improvement can be expected for a time-extended search.

The best mean best conflict is found by the BSt-PSO$^{(c)}$ by a large margin, while the second, third, and fourth best ones are found by the BSwSd-PSO 2$^{(p)}$, the BSt-PSO$^{(p)}$, and the BStLd-PSO 2$^{(c)}$, respectively. Notice that these optimizers, together with the BStLd-PSO 1, are the only ones which never fail to attain the error condition.

The reason for the other optimizers' failures can be due to inability to fine-tune the search, or to premature clustering of the particles around a suboptimal solution which happens to be worse than the error condition. For instance, the BSwSd-PSO 2$^{(c)}$ and the BSw-PSO$^{(p)}$ fail every time—and the BStSd-PSO 1$^{(p)}$ fails 17 times—due to the complete premature implosion of their particles, as opposed to the BSt-PSO and the BSw-PSO which fail 15 and 23 times, respectively, due to the their lack of fine-clustering ability.

The optimizers which exhibit the steepest decrease of the mean best conflict at the beginning of the search are the BSt-PSO$^{(p)}$, the BSw-PSO$^{(p)}$, the BSt-PSO$^{(c)}$, the BSw-PSO$^{(c)}$, the BStSd-PSO 2$^{(c)}$, and the BSwSd-PSO 2$^{(c)}$, while the ones which exhibit a complete implosion of their particles are the BSw-PSO$^{(p)}$, the BStSd-PSO 1$^{(p)}$, the BSwSd-PSO 1$^{(p)}$, the BSwSd-PSO 2$^{(c)}$, the BStSd-PSO 1$^{(c)}$, and the BSwSd-PSO 1$^{(c)}$. Some other optimizers whose particles are close to complete the implosion by the final time-step are the BStLd-PSO 1$^{(p)}$, the BSwLd-PSO 1$^{(p)}$, the BStLd-PSO 1$^{(c)}$, and the BSwLd-PSO 1$^{(c)}$. Note that none of the optimizers whose inertia and acceleration weights are unrelated is able to get even close to complete the implosion!

With regards to the reluctance to getting trapped in local optima, this is again not the adequate function to test such ability.

To sum up, some of the best optimizers in dealing with this function are as follows:

- Stand-alone optimizers: BSt-PSO$^{(c)}$, BSwSd-PSO 2$^{(p)}$, BSt-PSO$^{(p)}$, and BStLd-PSO 2$^{(c)}$.
- Clustering ability: BSw-PSO$^{(p)}$, BStSd-PSO 1$^{(p)}$, BSwSd-PSO 1$^{(p)}$, BSwSd-PSO 2$^{(c)}$, BStSd-PSO 1$^{(c)}$, BSwSd-PSO 1$^{(c)}$, and BStSd-PSO 2$^{(c)}$.

#### 6.4.3.3 RASTRIGRIN function

Once again, the global optimum cannot be found by any optimizer along any of the 50 runs.

The best mean best conflicts of this function are found by the BSwLd-PSO 2$^{(p)}$, the BSt-PSO, the BStSd-PSO 2, the BStLd-PSO 2, and the BSwLd-PSO 2. Notice that keeping the inertia and acceleration weights unrelated appears to be a better strategy to deal with this function, which exhibits numerous local optima[24]. Furthermore, while the particles of the BSt-PSO are far from imploding, the evolution of its mean best conflict does not stagnate at all, so that further improvement can be certainly expected for a time-extended search. It seems clear that delaying clustering is critical to find the best mean best conflict possible here.

The BSt-PSO$^{(p)}$ and the BStSd-PSO 2$^{(c)}$ are the only optimizers which present failures in attaining the error condition, which are due to the complete premature clustering of their particles around suboptimal solutions.

As to the clustering ability, all the optimizers with the polynomial and constant relationships except for the BStLd-PSO 2$^{(p)}$, the BSwLd-PSO 2$^{(p)}$, the BStSd-PSO 1$^{(c)}$, and the BSwSd-PSO 1$^{(c)}$, perform the complete implosion of their particles. In contrast, none of the optimizers whose inertia and acceleration weights are kept unrelated is able to complete the implosion. In summary, the optimizer which exhibits the best clustering behaviour is the BSt-PSO$^{(p)}$: the rate of decrease of its mean best conflict is extremely high at the beginning of the search, and its particles virtually complete the implosion in less than 100 time-steps. In fact, it fails to attain the error condition along 43 runs because of premature clustering, while it attains it in only 53 time-steps, on average, along the remaining 7 runs!

With regards to the reluctance to getting trapped in local optima, it seems clear that the best performance is that of the BSt-PSO: the curve of the evolution of its mean best conflict and the trend-line of the evolution of its mean average conflict are monotonically decreasing, while the slope of the former is sill steep and diversity fairly high by the final time-step.

To summarize, some of the best optimizers in dealing with this function are as follows:

- Stand-alone optimizers: BSt-PSO, BSwLd-PSO 2$^{(p)}$, BStSd-PSO 2, and BStLd-PSO 2.
- Clustering ability: BSt-PSO$^{(p)}$, BSw-PSO$^{(p)}$, BSt-PSO$^{(c)}$, BSw-PSO$^{(c)}$, BStSd-PSO 2$^{(c)}$, BSwSd-PSO 2$^{(c)}$.
- Immunity to local optima: BSt-PSO.

[24] Although the best mean best conflict is found by an optimizer whose inertia and acceleration weights are related like the 4$^{th}$ degree polynomial, namely the BSwLd-PSO 2$^{(p)}$, it must be noted that it is precisely one of those which do not take full advantage of the clustering behaviour induced by such a relationship. In fact, it is the optimizer whose inertia weight, acceleration weight, and individuality/sociality ratio are set highest during the early stages of the search, thus counterbalancing the pressure to cluster induced by the polynomial relationship!

### 6.4.3.4 GRIEWANK function

Recall that this is a kind of noisy Sphere function. It is interesting to observe that, while the mean best conflicts found by each optimizer is noticeably higher than the one it finds when dealing with the Sphere function, most of the optimizers are now able to find the global optimum several times. This might be because the noise delays clustering, thus maintaining diversity for a longer period of time.

The best mean best conflicts of this function are found by the BSt-PSO$^{(p)}$, the BStLd-PSO 1$^{(p)}$, the BSwLd-PSO 1, and the BSwLd-PSO 1$^{(c)}$.

Some optimizers exhibit a few failures in attaining the error condition, which are sometimes due to the lack and sometimes to the surplus of clustering ability. However, when numerous failures occur, they are due to the second reason.

Broadly speaking, every optimizer—except for the BSt-PSO and the BSw-PSO—exhibits clustering behaviour. However, the particles of those whose inertia and acceleration weights keep the constant and polynomial relationships—except for the BStLd-PSO 2$^{(p)}$—, the BStLd-PSO 1, the BSwSd-PSO 2, and the BStSd-PSO 1 are the only ones which are able to complete the implosion. Nevertheless, considering both the initial rate of decrease of the mean best conflict and the ability to fine-cluster, the best optimizers are the BSt-PSO$^{(p)}$, the BSw-PSO$^{(p)}$, the BSt-PSO$^{(c)}$, the BSw-PSO$^{(c)}$, the BStSd-PSO 2$^{(c)}$, and the BSwSd-PSO 2$^{(c)}$.

With regards to the reluctance to getting trapped in local optima, it is again the BSt-PSO the one which exhibits the most convenient behaviour: the curve of the evolution of its mean best conflict and the trend-line of the evolution of its mean average conflict are monotonically decreasing, while still maintaining a fairly high diversity by the final time-step.

In summary, some of the best optimizers in dealing with this function are as follows:

- Stand-alone optimizers: BSt-PSO$^{(p)}$, BStLd-PSO 1$^{(p)}$, BSwLd-PSO 1, and BSwLd-PSO 1$^{(c)}$.
- Clustering ability: BSt-PSO$^{(p)}$, BSw-PSO$^{(p)}$, BSt-PSO$^{(c)}$, BSw-PSO$^{(c)}$, BStSd-PSO 2$^{(c)}$, and BSwSd-PSO 2$^{(c)}$.
- Immunity to local optima: BSt-PSO.

### 6.4.3.5 SCHAFFER F6 function (2D)

Although the Schaffer f6 function exhibits numerous local optima in the form of ring-like depressions surrounding the global optimum, this 2-dimensional version does not constitute a serious challenge for the optimizers whose inertia and acceleration weights are not related. In fact, they are able to find the global optimum along each of the 50 runs. In contrast, most of the optimizers with constant relationship, in addition to the BSt-PSO$^{(p)}$ and the BSw-PSO$^{(p)}$, sometimes fail not only in finding the global optimum but also in attaining the error condition.

It is interesting to observe that whenever an optimizer fails to find the global optimum here, it also fails to attain the error condition. The BStSd-PSO 2$^{(c)}$ is the optimizer which exhibits a higher number of failures: 23 out of the 50 runs. In spite of the fact that all the failures occur when the strategies which aim to favour clustering are implemented, none of them is due to the premature complete implosion of the particles. It was theorized before that this is probably because of the local optima having the form of ring-like depressions rather than valleys (refer to **Appendix 3** for graphical visualization).

It seems evident that the constant and polynomial strategies effectively favour clustering, even though no optimizer is able to complete the particles implosion. It appears that although the features of this function make the complete implosion too difficult, each circumference-like local optimum spans an area of influence which traps the swarms whose diversity is not high enough. Therefore, broadly speaking, both the constant and polynomial strategies successfully enhance the ability to cluster, while keeping the inertia and acceleration weights unrelated successfully enhances the reluctance to getting trapped in local optima.

To summarize, some of the best optimizers in dealing with the 2-dimensional Schaffer f6 function are as follows:

- Stand-alone optimizers: All the optimizers whose inertia and acceleration weights are not related, and those which keep the polynomial relationship together with time-decreasing inertia weights.
- Clustering ability: All the optimizers whose $aw^{(t)} = 4.1 \cdot w^{(t)}$, BSt-PSO$^{(p)}$, BSw-PSO$^{(p)}$, BStSd-PSO 2$^{(p)}$, and BSwSd-PSO 2$^{(p)}$.
- Immunity to local optima: The same as the stand-alone optimizers.

#### 6.4.3.6 SCHAFFER F6 function

As previously argued several times within this chapter, keeping high diversity is critical to deal with this function. Hence the optimizers which find six of the seven best mean best conflicts keep their inertia and acceleration weights unrelated. Thus, the best mean best conflicts are found by the BStLd-PSO 2, the BStLd-PSO 1, the BStSd-PSO 2, the BSw-PSO$^{(p)}$, the BSwLd-PSO 1, the BSwLd-PSO 2, and the BSwSd-PSO 2, in that order. Notice that, once again, the global optimum cannot be found by any optimizer.

The demanding error condition set here leads to numerous failures in attaining it. As it can be expected, the optimizers which exhibit the lower number of failures are the same which find the best mean best conflicts. Even though the failures occur because of the lack of clustering ability, there are a few exceptions which are due to premature clustering.

The BStSd-PSO 2$^{(c)}$, the BSwSd-PSO 2$^{(c)}$, and the BSwSd-PSO 1$^{(p)}$ are the only optimizers whose particles virtually implode. Considering the features of this function, these complete implosions to local optima are indeed striking. The optimizers which present the higher initial rate of decrease of the mean best conflict are the BStSd-PSO 2$^{(c)}$, BSwSd-PSO 2$^{(c)}$, the BSt-PSO$^{(c)}$, and the BSw-PSO$^{(c)}$.

Regarding the reluctance to getting trapped in local optima, it can be observed that all the optimizers whose inertia and acceleration weights are unrelated keep high diversity, resulting in the non-stagnation of the evolution of their mean best conflicts. The BSt-PSO$^{(c)}$, the BSw-PSO$^{(c)}$, the BSt-PSO$^{(p)}$, the BSw-PSO$^{(p)}$, the BStLd-PSO 2$^{(p)}$, the BSwLd-PSO 2$^{(p)}$, the BStSd-PSO 2$^{(p)}$, and the BSwSd-PSO 2$^{(p)}$ also maintain a reasonably high diversity.

In summary, some of the best optimizers in dealing with the 30-dimensional Schaffer f6 function are as follows:

- Stand-alone optimizers: BStLd-PSO 2, BStLd-PSO 1, BStSd-PSO 2, and BSw-PSO$^{(p)}$.
- Clustering ability: BStSd-PSO 2$^{(c)}$, BSwSd-PSO 2$^{(c)}$, and BSwSd-PSO 1$^{(p)}$.
- Immunity to local optima: All the optimizers whose inertia and acceleration weights are unrelated, BSt-PSO 1$^{(c)}$, BSw-PSO 1$^{(c)}$, BSt-PSO 1$^{(p)}$, BSw-PSO 1$^{(p)}$, BStLd-PSO 2$^{(p)}$, BSwLd-PSO 2$^{(p)}$, BStSd-PSO 2$^{(p)}$, and BSwSd-PSO 2$^{(p)}$.

### 6.4.3.7 Best overall performances

The general criteria for the analysis of the different algorithms were described by the end of section **6.1**. Thus, the aspects to be considered are divided in three:

1. Best mean best conflict found.
2. Clustering ability.
3. Robustness.

#### 6.4.3.7.1 Best mean best conflict found

The first criterion to be considered for the evaluation of the goodness of an optimizer's performance is the goodness of the mean best conflict it is able to find. The optimizers which obtain the 10 best mean best conflicts of each benchmark function are gathered in **Table 6. 20**. Given that there are numerous optimizers which find the global optimum of the 2-dimensional Schaffer f6 function along each of the 50 runs, the latter is not included in the table.

| Order | SPHERE | ROSENBROCK | RASTRIGRIN | GRIEWANK | SCHAFFER F6 |
|---|---|---|---|---|---|
| **1** | **BSt-PSO$^{(c)}$** | **BSt-PSO$^{(c)}$** | BSwLd-PSO 2$^{(p)}$ | **BSt-PSO$^{(p)}$** | **BStLd-PSO 2** |
| **2** | **BSt-PSO$^{(p)}$** | BSwSd-PSO 2$^{(p)}$ | **BSt-PSO** | BStLd-PSO 1$^{(p)}$ | BStLd-PSO 1 |
| **3** | **BStSd-PSO 2** | **BSt-PSO$^{(p)}$** | **BStSd-PSO 2** | **BSwLd-PSO 1** | **BStSd-PSO 2** |
| **4** | **BSwLd-PSO 1** | BSwLd-PSO 2$^{(c)}$ | **BStLd-PSO 2** | BSwLd-PSO 1$^{(c)}$ | BSw-PSO$^{(p)}$ |
| **5** | **BSwSd-PSO 2** | BStLd-PSO 2$^{(c)}$ | BSwLd-PSO 2 | BStSd-PSO 1$^{(p)}$ | **BSwLd-PSO 1** |
| **6** | **BStLd-PSO 2** | BSwLd-PSO 1$^{(p)}$ | BSwSd-PSO 1$^{(c)}$ | BStLd-PSO 2$^{(c)}$ | BSwLd-PSO 2 |
| **7** | BStLd-PSO 2$^{(p)}$ | **BSwSd-PSO 2** | **BSwSd-PSO 2** | BSwSd-PSO 1 | **BSwSd-PSO 2** |
| **8** | BSwSd-PSO 2$^{(p)}$ | **BSwLd-PSO 1** | BStSd-PSO 1$^{(c)}$ | BStLd-PSO 1 | **BSt-PSO$^{(p)}$** |
| **9** | BStLd-PSO 1 | BStLd-PSO 2 | **BSwLd-PSO 1** | BStSd-PSO 1 | BSwSd-PSO 1$^{(c)}$ |
| **10** | BSwLd-PSO 2 | BSwSd-PSO 1$^{(c)}$ | BStLd-PSO 2$^{(p)}$ | BStLd-PSO 2$^{(p)}$ | BStSd-PSO 1$^{(c)}$ |

**Table 6. 20**: Optimizers which find the 10 best mean best conflicts when optimizing each benchmark function, ordered from best to worse, where some of the optimizers which exhibit some of the best overall performances with regards to the mean best conflicts they are able to find are coloured.

Some of the optimizers which are considered to exhibit the best performances on the whole test suite are coloured so as to facilitate the interpretation of the table. It is fair to remark that, although the **BSt-PSO$^{(c)}$** is among the 10 best optimizers only in dealing with the Sphere and the Rosenbrock functions, it finds the best mean best conflict of both, by a large margin. In turn, the **BSt-PSO$^{(p)}$** finds the best mean best conflict of the Griewank function, while being among the 8 best optimizers in dealing with other 3 functions. Likewise, the **BStLd-PSO 2** finds the best mean best conflict of the Schaffer f6 function, while being among the 6 best optimizers in dealing with other 2 functions.

The optimizers which find the best mean best conflict of each benchmark function except for the Rastrigrin function have been included so far. Although the BSwLd-PSO 2$^{(p)}$ finds the best mean best conflict of the latter, it is decided here to choose the **BSt-PSO** instead. The reason for this is that, while the **BSt-PSO** finds the second best mean best conflict, the curve of its evolution exhibits a noticeably steeper slope. Hence, and given that its weights are independent from the time-steps, further improvement for a longer search is virtually certain. Furthermore, constant weights are probably more convenient to deal with dynamic problems, which are beyond the scope of this dissertation.

Finally, the **BStSd-PSO 2** is chosen because it is among the 3 best optimizers in dealing with 3 functions; the **BSwSd-PSO 2** is selected because it is among the 7 best ones in dealing with 4 functions, and the **BSwLd-PSO 1** is picked out because it is among the 9 best optimizers in dealing with all the 5 benchmark test functions.

Thus, some of the best optimizers with regards to the mean best conflict they are able to find when dealing with the test suite shown in **Table 6. 1** are listed hereafter:

- **BSt-PSO**
- **BSt-PSO$^{(c)}$**
- **BSt-PSO$^{(p)}$**
- **BStLd-PSO 2**
- **BSwLd-PSO 1**
- **BStSd-PSO 2**
- **BSwSd-PSO 2**

Notice that the first 3 optimizers keep their weights constant along the search. This implies that they do not lose their explorative capabilities through time, so that it is reasonable to expect them to be useful for dynamic problems, where the global optimum might not keep its position steady. In addition to that, these optimizers are computationally cheaper. It is also interesting to observe that 5 of these 7 optimizers keep their inertia and acceleration weights unrelated. It is fair to note that these 5 optimizers are also able to find the global optimum of the 2-dimensional Schaffer f6 function, which was put aside within this section.

#### 6.4.3.7.2 Clustering ability

The second criterion to be considered is the clustering ability, which is two-fold: the ability to quickly decrease the mean best conflict found and the ability to fine-cluster. While the higher the initial rate of decrease the longer the particles can spend in fine-clustering, the ability to fine-cluster enables them to take successively smaller step-sizes so as to fine-tune the search.

Considering the analyses carried out along sections **6.4.1** to **6.4.6**, the best optimizers with regards to this criterion are as follows:

- **BSt-PSO$^{(c)}$**
- **BSt-PSO$^{(p)}$**
- **BSw-PSO$^{(c)}$**
- **BSw-PSO$^{(p)}$**
- **BStSd-PSO 2$^{(c)}$**
- **BSwSd-PSO 2$^{(c)}$**
- **BSwSd-PSO 1$^{(p)}$**

#### 6.4.3.7.3 Robustness

The third criterion to be considered for the evaluation of the goodness of an optimizer is its robustness, which refers to its reluctance to getting trapped in suboptimal solutions, while still exhibiting a monotonic decrease of the mean best and mean average conflicts found.

Considering the analyses performed along sections **6.4.1** to **6.4.6**, the optimizer which appears to be the most reluctant to getting trapped in suboptimal solutions are, broadly speaking, those whose inertia and acceleration weights are kept unrelated, being the **BSt-PSO** the one which exhibits the best performance in this regard.

## 6.5 Closure

The parameters of the particles' velocity updating rule of the B-PSO were initially set so as to match the O-PSO, and the concept of the explosion was introduced. It was shown that there are at least two factors which might cause the explosion: a large value of the acceleration weight, and the randomness. Clerc et al. [16] proved that $aw \geq 4$ leads to the uncontrolled divergence of the particles, although the proof was limited to the deterministic, single-particle PSO with fixed attractors. In spite of the second cause not being mathematically proven, there are numerous empirical results which back it up. In addition, a heuristic explanation of the reason why the randomness might lead to the explosion was suggested.

The $v_{max}$ constraint was shown to be successful in controlling the explosion, although this incorporates a new parameter to the algorithm that requires tuning. Thus, small values enhance the ability of the optimizer to fine-cluster but decrease its ability to escape local optima, whilst large values enhance its reluctance to getting trapped in local optima but reduce its ability to fine-cluster. Hence a linearly time-decreasing $v_{max}$ was proposed, which enables the optimizer to exhibit more explorative behaviour during the early stages and more exploitative behaviour during the late stages of the search. However, while this strategy acts as an external restriction to the "natural" behaviour of the swarm, it is usually preferred to control the explosion by means of the particles' velocity updating equation itself. Thus, the inertia weight was incorporated to the velocity updating equation, showing that it is successful in both controlling the explosion—or at least pulling the particles back to the region of interest—and fine-tuning the search. Clerc et al.'s constriction factor [16] is thought of as a particular case. Although the inertia weight makes the restriction to the components of the particles' velocity unnecessary to control the explosion, the latter is still kept in the algorithm so as to avoid subsequent evaluations of the objective function far from the region of interest.

Constant and time-decreasing inertia weights were proposed and tested, showing that the time-decreasing ones typically result in better fine-clustering. It was concluded that while both the restriction to the components of each particle's velocity and an inertia weight smaller than one are able to control the explosion independently from one another, implementing $w < 1$ together with $v_{max} = 0.5 \cdot \left(x_{max} - x_{min}\right)$ comprises a good general-purpose strategy.

The influence of the learning weights on the behaviour of the swarm was discussed, and the swapping strategy was developed. The latter consists of linearly time-swapping the relative importance of the learning weights. It was shown that while this strategy effectively delays the inherent fast clustering of the PSOs and favours clustering during the late stages of the search, this is not always helpful. For instance, small $iw/sw$ ratios are not usually convenient to deal with problems which exhibit numerous local optima located near the global optimum.

Finally, three different strategies were proposed to relate the inertia and acceleration weights:

The first strategy simply makes use of a traditional setting for the acceleration weight, $aw = iw + sw = 4$, keeping no relation to the inertia weight.

The second strategy keeps the inertia and acceleration weights related like $aw = 4.1 \cdot w$. Note that this relationship together with $w^{(t)} \leq 0.7298$ results in Clerc et al.'s C-PSOs [16].

The third strategy keeps the inertia and acceleration weights related like a 4$^{th}$ degree polynomial, which was developed aiming to favour clustering.

For each strategy, 10 different settings of the weights were proposed and tested on a set of benchmark functions. Given the probabilistic nature of the optimizers, each experiment was run 50 times and the analyses of the results were performed on the statistical data.

The criteria used for the evaluation of the goodness of each optimizer's performance involve the best solution it is able to find, its clustering ability, and its reluctance to getting trapped in suboptimal solutions. The results of the experiments showed that none of the 30 optimizers tested exhibits outstanding performance considering the three aspects and every benchmark function. Hence, the idea of a swarm composed of specialized sub-swarms, whose particular abilities are combined to obtain a better overall performance than that of any homogeneous swarm, comes to mind spontaneously. This strategy will be dealt with in a later chapter.

Note that, since it is not possible to cover every possibility, the tunings of the parameters were first developed in a qualitative fashion, and a few quantitative somehow arbitrary settings were tested. Thus, some other settings, for instance of the BSt-PSO's inertia weight, or of the BSw-PSO's initial $iw / sw$ ratio, might eventually lead to better results.

Finally, it is very important to remark that the constant weights should always be preferred—unless a time-decreasing one undoubtedly outperforms it—for at least three reasons:

1. The optimizer does not lose the exploration ability as the search goes by.
2. The optimizer is computationally cheaper.
3. The behaviour does not depend on the maximum number of time-steps allowed.

Thus, several optimizers were developed, and the ones which exhibited the most promising behaviour regarding the three previously mentioned aspects of interest were summarized. The next chapter is entirely devoted to the development of meaningful stopping criteria.

# Chapter 7

# STOPPING CRITERIA

A preliminary but fairly extensive analysis of the parameters of the basic particle swarm optimizer was carried out in the previous chapter. Several optimizers, which only differ in the tuning of their parameters, were developed and tested on a set of benchmark functions. Some of those which performed best with respect to the best solution found, the strongest clustering behaviour, and the ability to escape local optima were selected. However, every algorithm was run along a fixed number of time-steps—regardless of the degree of accuracy achieved—and the goodness of their performance was evaluated thanks to the fact that the global optimum of each benchmark function is already known. Therefore, in order to make the algorithms suitable for the optimization of real-world problems, some measures to somehow evaluate the reliability of the solutions that the optimizers are able to find are developed within this chapter. These measures are thereafter used to develop meaningful sopping criteria.

## 7.1 Introduction

Traditionally, iterative methods are equipped with some stopping criteria which are met when the solution found is good enough, or when further significant improvement is unlikely. This serves the function of both saving computational cost and estimating the reliability of the solution found. Traditional techniques, suitable for traditional methods, involve:

- the difference between the best solution found up to the current time-step and that found up to the preceding one
- the distance between the last two coordinates (typically, the Euclidean norm)
- a permissible maximum number of time-steps

However, these techniques are not suitable for population-based methods for several reasons, some of which are as follows:

- population-based methods present numerous candidate solutions per time-step
- the best solution found up to the current time-step might remain unchanged for quite some time before improving again

- the best solution found up to the current time-step and that found up to the previous one might correspond to different particles, so that the distance between their locations is not an accurate measure of convergence

The influence of the parameters of the basic particle swarm optimizer and their interrelations on both the behaviour and the achievements of the swarm were studied along the previous chapter. Thus, some optimizers were developed and tested on a suite of benchmark functions. However, every algorithm was run along a fixed number of time-steps, regardless of the degree of accuracy achieved. Besides, the measures of error used to test the algorithms were based on the fact that the global optimum of each benchmark function included in the test suite is well known. Hence, if the algorithms are to be used to optimize real-world problems, some measures of the goodness of the solutions found need to be developed. Therefore, some meaningful measures of error—which are expected to serve the function of both developing some stopping criteria and evaluating the reliability of the solutions that the optimizers are able to find—are proposed within this chapter. Their evolution is analyzed and compared to the evolution of the actual absolute error when optimizing a test suite of benchmark functions.

## 7.2 General settings

The experiments run along **Chapter 6** showed that the optimizers which perform best with regards to some criterion are typically not the same as those which do it with regards to some other. Even further, the optimizers which perform best with regards to one single criterion are typically not the same for the different benchmark functions. Hence, taking into account the results obtained from the experiments carried out along the previous chapter, the **BSt-PSO**, the **BSt-PSO$^{(c)}$**, and the **BSt-PSO$^{(p)}$** are selected to undertake the development of some stopping criteria. This is because, when considering each of the six benchmark functions and each of the three criteria assumed to be desirable (refer to section **6.4.3.7**), the performance of at least one of them is among the best ones.

The test suite used within this chapter is that of **Table 6.1**. Each experiment is run 50 times due to the probabilistic nature of the PSO paradigm, and the best solution found, the time-steps at which the error condition and the global optimum—if applicable—are attained, and the values of the different measures of error proposed corresponding to those time-steps are computed for each run for some experiments. In addition to that, the evolution of the mean best conflict, of the mean average conflict, and of the average of the different measures of error proposed are computed and plotted. It is evident that when the evolution of the different measures of error regarding the conflict values exhibit wide, uneven oscillations, their average among the 50 runs smooth such oscillations. Furthermore, the evolution of the mean best conflict may remain unchanged for a longer period of time for a single run than it does for the average of 50 runs. Hence the evolution of the measures of error along a single run is also analyzed.

Because of the huge amount of data that results from the experiments, only some selected results are included in the written part of this thesis. The complete set of results can be found in digital **Appendix 4**. Bear in mind that, whenever the error condition is referred to before the development of the stopping criteria, the reference will be to the error conditions stated in **Table 6.1**. Hence the algorithms are run again along a fixed number of time-steps, regardless of the goodness of the best solution found, except for those experiments run along section **7.6**, where the stopping criteria is already incorporated into the algorithm.

References:

- BSt-PSO : basic, standard PSO: $w^{(t)} = 0.7, \quad iw^{(t)} = sw^{(t)} = 2 \quad \forall t$
- BSt-PSO$^{(c)}$ : basic, standard PSO: $w^{(t)} = 0.7298, \quad iw^{(t)} = sw^{(t)} = 1.49609 \quad \forall t$
- BSt-PSO$^{(p)}$ : basic, standard PSO: $w^{(t)} = 0.5, \quad iw^{(t)} = sw^{(t)} = 2 \quad \forall t$
- $\overline{cgbest}$ : mean best solution found along 50 runs
- $cgbest$ : best solution found in any of the 50 runs
- $\sigma$ : standard deviation
- $\overline{tsgb}$ : mean number of time-steps required to find the global optimum
- $\overline{tsec}$ : mean number of time-steps required to attain the error condition
- length, distance : Euclidean norm

General settings:

- $v_{max} = 0.5 \cdot (x_{max} - x_{min})$
- Number of particles: 30
- $t_{max} = 10000$ for sections **7.4** and **7.5**, while $t_{max} = 30000$ for section **7.6**

Following Gehlhaar and Fogel (quoted in [4]), the experiments are run for two initializations of the particles' positions: one symmetric and one asymmetric about the origin. The areas where the particles are initially spread over are shown in **Table 7. 1**:

| Benchmark function | Initialization |
|---|---|
| Sphere | - Symmetric: $[-100,100]^{30}$<br>- Asymmetric: $[50,100]^{30}$ |
| Rosenbrock | - Symmetric: $[-30,30]^{30}$<br>- Asymmetric: $[15,30]^{30}$ |
| Rastrigrin | - Symmetric: $[-5.12,5.12]^{30}$<br>- Asymmetric: $[2.56,5.12]^{30}$ |
| Griewank | - Symmetric: $[-600,600]^{30}$<br>- Asymmetric: $[300,600]^{30}$ |
| Schaffer f6 2D | - Symmetric: $[-100,100]^{2}$<br>- Asymmetric: $[50,100]^{2}$ |
| Schaffer f6 | - Symmetric: $[-100,100]^{30}$<br>- Asymmetric: $[50,100]^{30}$ |

**Table 7. 1**: Symmetric and asymmetric initializations of the positions of the particles for each benchmark function in the test suite.

Notice that, while the symmetric initialization is the same as the one used for the experiments run along section **6.4** (refer to **Table 6.1**), the asymmetric initialization allows the analysis of the behaviour of the swarm and of the measures of error when the optimum is not contained within the hyper-space spanned by the particles' initial positions. Besides, it also prevents a possible centre-seeking algorithm from accidentally finding the global optimum.

# 7.3 Traditional measures of error

When computing absolute errors in gradient-based methods, it is usually assumed that, if the method converges, the rate of variation of the evaluations of the cost function decreases when approaching the optimum. Since multi-objective optimization problems are beyond the scope of this thesis, the computation of the absolute error with regards to the cost function at the $t^{\text{th}}$ time-step is given by a simple subtraction:

$$e^{(t)} = \text{abs}\left(f\left(\mathbf{x}^{(t)}\right) - f\left(\mathbf{x}^{(t-1)}\right)\right) \tag{7. 1}$$

Likewise, the absolute error regarding the design variables at the $t^{\text{th}}$ time-step can be given by the distance between the location of the last approximate solution and that of the preceding one (i.e. the Euclidean norm):

$$e_c^{(t)} = \sqrt{\sum_{i=1}^{n}\left(x_i^{(t)} - x_i^{(t-1)}\right)^2} \tag{7. 2}$$

Where $n$ is the dimension of the search-space.

Note that thinking of these values as absolute errors is equivalent to assuming that $\mathbf{x}^{(t-1)}$ and $\mathbf{x}^{(t)}$ are the locations of the approximate and exact solutions, respectively.

Because the importance of a given absolute error depends on the value of the true solution itself, relative errors are usually preferred. However, there is the problem of which value the absolute error should be related to. Ideally, it should be related to the exact solution, whose true value is unknown. Such a problem might be dealt with as follows:

1. Relating the absolute error regarding the cost function to the best solution found up to the current time-step:

$$re^{(t)} = \text{abs}\left(\frac{f\left(\mathbf{x}^{(t)}\right) - f\left(\mathbf{x}^{(t-1)}\right)}{f\left(\mathbf{x}^{(t)}\right)}\right) \tag{7. 3}$$

2. Relating the absolute error regarding the design variables to the Euclidean norm of the vector position associated to the best solution found up to the current time-step:

$$re_c^{(t)} = \sqrt{\frac{\sum_{i=1}^{n}\left(x_i^{(t)} - x_i^{(t-1)}\right)^2}{\sum_{i=1}^{n}\left(x_i^{(t)}\right)^2}} \tag{7. 4}$$

However, this strategy results in a dynamic reference point, which is updated every time-step. This makes the accuracy of the relative error computed at each time-step dependent on the accuracy of the best solution found so far. In addition, the relative error becomes huge or infinite when the true solution is near or equal to zero, respectively.

An alternative could be to relate the absolute errors to the values corresponding to the initial rather than to the current time-step, thus keeping the reference point stationary. However, this makes the relative errors dependent on the initialization and on the topography of the cost function. Another strategy could be to compute the relative errors as shown in equations **(7. 3)** and **(7. 4)** while the absolute value of the best solution found so far and the Euclidean norm of its location are greater than one, and to compute them as shown in equations **(7. 1)** and **(7. 2)** otherwise. This allows the definition of meaningful permissible errors in advance, without any knowledge about the function to be optimized.

Sections **7.4** and **7.5** are entirely devoted to the development of measures of error suitable for particle swarm optimizers. In the same fashion as the measures of error previously discussed for traditional iterative search methods, absolute and relative measures of error involving both the conflict values and the particles' positions are proposed, tested, and discussed.

## 7.4 Measures of error regarding the conflict values

As opposed to traditional methods, the particle swarm optimizers present as many candidate solutions per time-step as particles are in the population. Therefore, the absolute error thought of as the simple subtraction shown in equation **(7. 1)** is no longer applicable.

Although the number of solutions per time-step could be reduced to one by considering only the best solution found up to the current time-step, successive best solutions might not belong to the same particle. Therefore, the fact that the best conflict found is not improved from a time-step to the next is not a doubtless indication of convergence. The particles might still be clustering, so that further improvement might still be possible.

Given the population-based nature of the PSOs, several ways of measuring the convergence of the system—and the clustering of the particles—can be conceived, such as the difference between the average conflicts corresponding to consecutive time-steps. Besides, as opposed to traditional methods, some ways of measuring the convergence within a single time-step can be thought of, such as the difference between the current best and worst conflicts. Bear in mind that "convergence" refers to the approximate solution approaching a local optimum, rather than to the clustering of the particles.

## 7.4.1 Absolute errors

An important aspect to be considered for the design of measures of error is that the definition of their permissible values should be straightforward. Besides, given that the aim of this thesis is to develop general-purpose optimizers, problem-independent permissible values need to be set. Therefore, they should depend neither on the features of the conflict function, nor on the number of design variables, nor on the size of the search-space, nor on the size of the swarm.

However, the formulation of absolute errors regarding the conflict values necessarily implies dependence on the features of the conflict function. That is to say that their permissible values cannot be independent from the function. In fact, the permissible value of the exact absolute error for the Rosenbrock function is set 1000 times greater than the one for the Griewank function in **Table 6.1**. Therefore, the absolute errors proposed within this section are merely thought of as a first step towards the design of relative errors. Nonetheless, they should still be independent from the number of design variables, from the size of the search-space, and from the number of particles in the swarm.

### 7.4.1.1 Errors definitions

#### 7.4.1.1.1 Within the current time-step

The population-based nature of the PSOs allows the design of measures that give an idea of the degree of clustering that the particles have achieved up to a given time-step, which in turn gives an idea of the convergence of the method. In spite of the fact that the clustering of the particles does not necessarily imply the convergence of the method, further improvement of the best solution found is unlikely after the complete implosion of the particles due to the loss of diversity. Following the social-psychology-inspiring metaphor of the method, this would indicate that the individuals in the population have reached agreement, which implies that diversity of beliefs is lost. Hence all the individuals exhibit the same set of beliefs, which has associated a given level of conflict among them, resulting in no individual exploring new combinations of beliefs that might result in smaller conflicts. Thus, some measures of error within the current time-step are proposed hereafter. Notice that if all the particles completely clustered around the best solution found, all these measures would become null.

- **abs_c_mse1**: square root of the mean squared error of the particles' current conflicts with respect to the current average conflict.

$$\text{abs_c_mse1}^{(t)} = \sqrt{\frac{\sum_{i=1}^{m}\left[c_i^{(t)} - \bar{c}^{(t)}\right]^2}{m}} \tag{7. 5}$$

where:

- $c_i^{(t)}$ : conflict of particle $i$ at time-step $t$
- $\bar{c}^{(t)}$ : average among the conflicts of all the particles in the swarm at time-step $t$
- $m$ : number of particles in the swarm

- **abs_c_mse2**: square root of the mean squared error of the particles' current conflicts with respect to the best conflict found so far.

$$\text{abs_c_mse2}^{(t)} = \sqrt{\frac{\sum_{i=1}^{m}\left[c_i^{(t)} - cgbest^{(t)}\right]^2}{m}} \tag{7. 6}$$

- **abs_c_maxe**: the greatest of the differences between each particle's current conflict and the best conflict found so far.

$$\text{abs_c_maxe}^{(t)} = \max\left(c_i^{(t)} - cgbest^{(t)}\right) = \max\left(c_i^{(t)}\right) - cgbest^{(t)} \quad , \quad i = 1,\ldots,m \tag{7. 7}$$

- **abs_c_me**: the average of the differences between each particle's current conflict and the best conflict found so far (note that $c_i^{(t)} - cgbest^{(t)} \geq 0 \quad \forall i$ ).

$$\text{abs_c_me}^{(t)} = \frac{\sum_{i=1}^{m}\left(c_i^{(t)} - cgbest^{(t)}\right)}{m} = \bar{c}^{(t)} - cgbest^{(t)} \tag{7. 8}$$

Notice that this is also the difference between the current average conflict and the best conflict found so far, which is always positive for minimization problems.

- **abs_c_b-w**: difference between the current worst conflict and the current best one.

$$\text{abs_c_b-w}^{(t)} = \max\left(c_i^{(t)}\right) - \min\left(c_i^{(t)}\right) \quad , \quad i = 1,\dots,m \tag{7. 9}$$

#### 7.4.1.1.2 Between consecutive time-steps

Even if the implosion of the particles is complete, further improvement might still be possible for simple functions, with the whole swarm behaving similarly to a single particle. Hence the measures of the degree of clustering achieved by the particles should be complemented with measures of the evolution of the conflicts.

- **abs_c_cav**: absolute difference between the current and the preceding average conflicts:

$$\text{abs_c_cav}^{(t)} = \text{abs}\left(\bar{c}^{(t)} - \bar{c}^{(t-1)}\right) \tag{7. 10}$$

- **abs_c_cgbest**: absolute difference between the best conflict found up to the current time-step and that found up to the preceding one:

$$\text{abs_c_cgbest}^{(t)} = \text{abs}\left(cgbest^{(t)} - cgbest^{(t-1)}\right) = cgbest^{(t-1)} - cgbest^{(t)} \tag{7. 11}$$

### 7.4.1.2 Experimental results

#### 7.4.1.2.1 Symmetric initialization

In cases where the optimum within a finite region of the search-space is sought, spreading the particles over the whole feasible region at the initial time-step is the obvious choice. This implies that the solution is contained within the hyper-space[1] spanned by the particles' initial positions. The same symmetric initialization used for the experiments run in section **6.4** (refer to **Table 6.1**) is used within this section. Note that symmetric initialization refers to the fact that the search-space which the particles are initially spread over is symmetric to the origin, and not that the particles are uniformly distributed.

[1] Recall that the words "space" and "hyper-space" are used indistinctly throughout this thesis.

It is important to remark that the global optimum of all the benchmark functions in the test suite is located at the origin, except for that of the Rosenbrock function, which is located at $x_i = 1 \ \ \forall i,$ . Therefore, the symmetric initialization is also symmetric to the global optimum for every function but the Rosenbrock.

As previously mentioned, it is not possible to include all the results from the experiments in the written part of this dissertation, which are nevertheless included in digital **Appendix 4**. Besides, since the absolute errors with regards to the conflict values are developed here as a first step towards the development of relative errors, only qualitative[2] issues of the results obtained from the experiments are discussed hereafter.

#### 7.4.1.2.1.1 Fifty runs

First of all, it can be observed that the evolution of the errors **abs_c_b-w** and **abs_c_maxe** are very much alike, which is due to the fact that there is not much difference in value between $cgbest^{(t)}$ and $\min\left[c_i^{(t)}\right]$, with $i = 1,...,m$ . In other words, all the particles quickly converge to a small area of the search-space, so that the difference between the best solution found along the whole search and the best solution at the current time-step is not significant.

In agreement with the conclusions derived from the experiments run along **Chapter 6**, the evolution of the errors of the **BSt-PSO** exhibit wider oscillations—and for longer periods of time—than those of the **BSt-PSO$^{(c)}$** and of the **BSt-PSO$^{(p)}$**, presumably due to the lower speed of clustering and lack of fine-clustering ability of the particles of the former. This can be seen from the graphs in **Fig. 7. 1**, **Fig. 7. 2**, and **Appendix 4**.

The graphs of the evolution of the proposed measures of error make it possible to infer that not only do the particles not perform a complete implosion but also they end up exhibiting some divergence for the **BSt-PSO$^{(c)}$** and the **BSt-PSO$^{(p)}$** optimizing the Rosenbrock function (refer to **Fig. 7. 3** and to **Appendix 4**). This phenomenon—which is not understood—was also observed in the experiments run along **Chapter 6**. Note that this divergence—which is not clear whether it would continue indefinitely—does not take place for the **BSt-PSO**. However, it is interesting to observe that, while the **BSt-PSO$^{(c)}$** and the **BSt-PSO$^{(p)}$** exhibit this strange, apparently divergent behaviour, they are precisely the optimizers which find the best solutions for this function! In fact, while the measures of error surprisingly tend to increase after an almost complete implosion of the particles (refer to **Fig. 7. 3** and to **Appendix 4**), the best solution found up to the current time-step—which is equal to the actual absolute error for these benchmark functions—is constantly improved (refer to **Fig. 7. 4** and to **Appendix 4**). It is reasonable to infer that this is thanks to the fact that diversity is kept, although neither the reason nor the reach of this apparent divergence are understood.

[2] Recall the previous argument about the permissible quantitative values of the measures of error regarding the conflict values not being independent from the conflict function at issue.

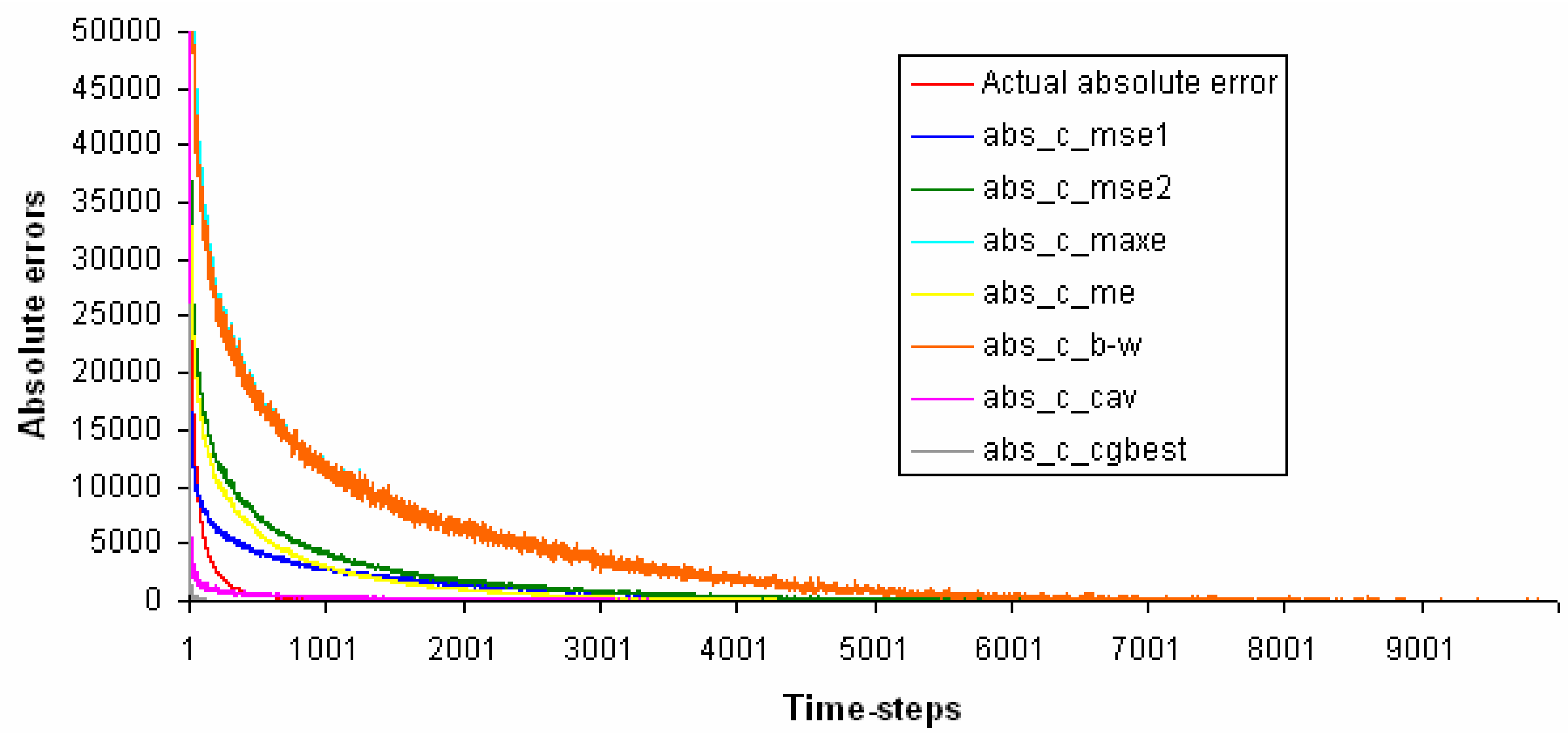

**Fig. 7. 1**: Evolution of the mean absolute errors for the BSt-PSO optimizing the Sphere function.

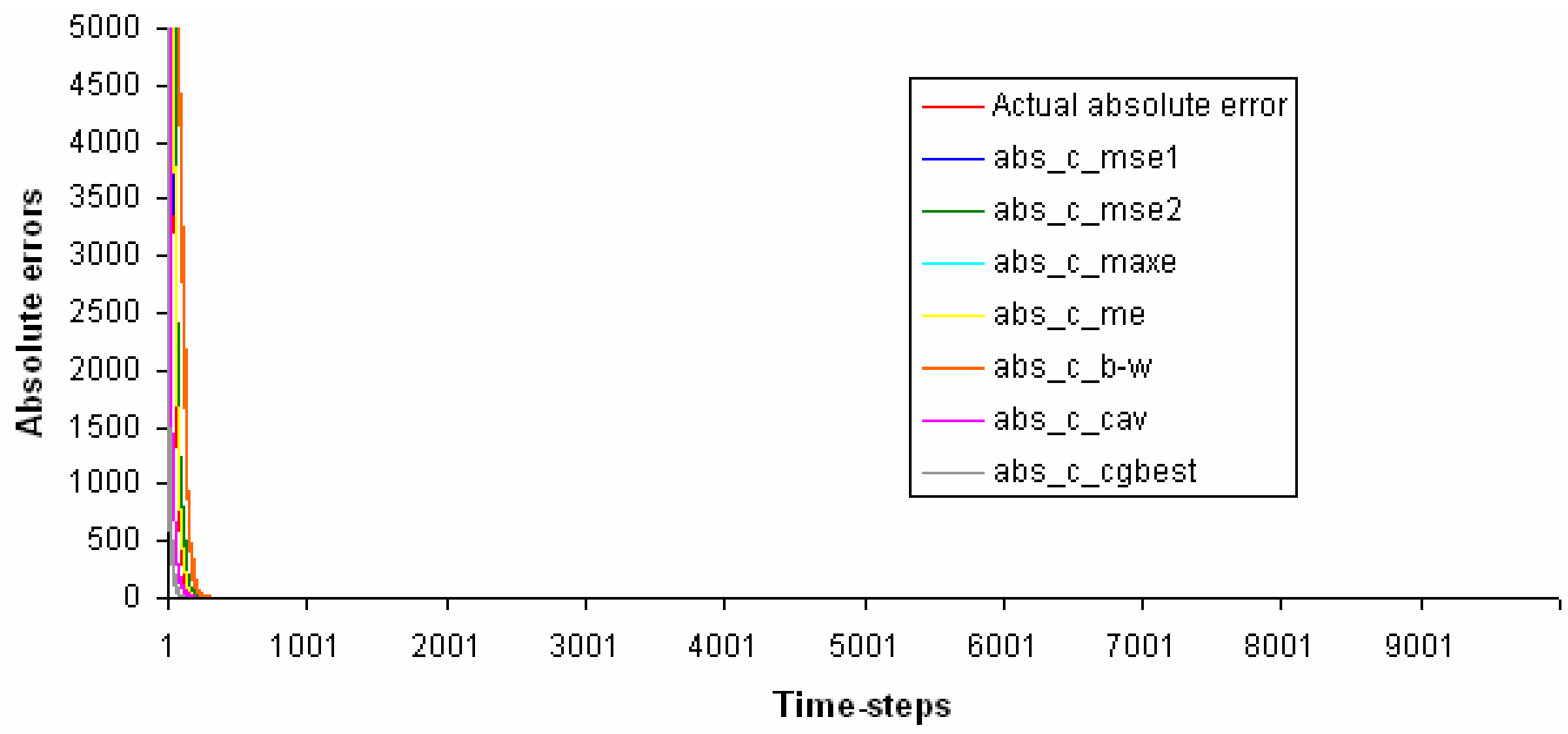

**Fig. 7. 2**: Evolution of the mean absolute errors for the BSt-PSO(c) optimizing the Sphere function.

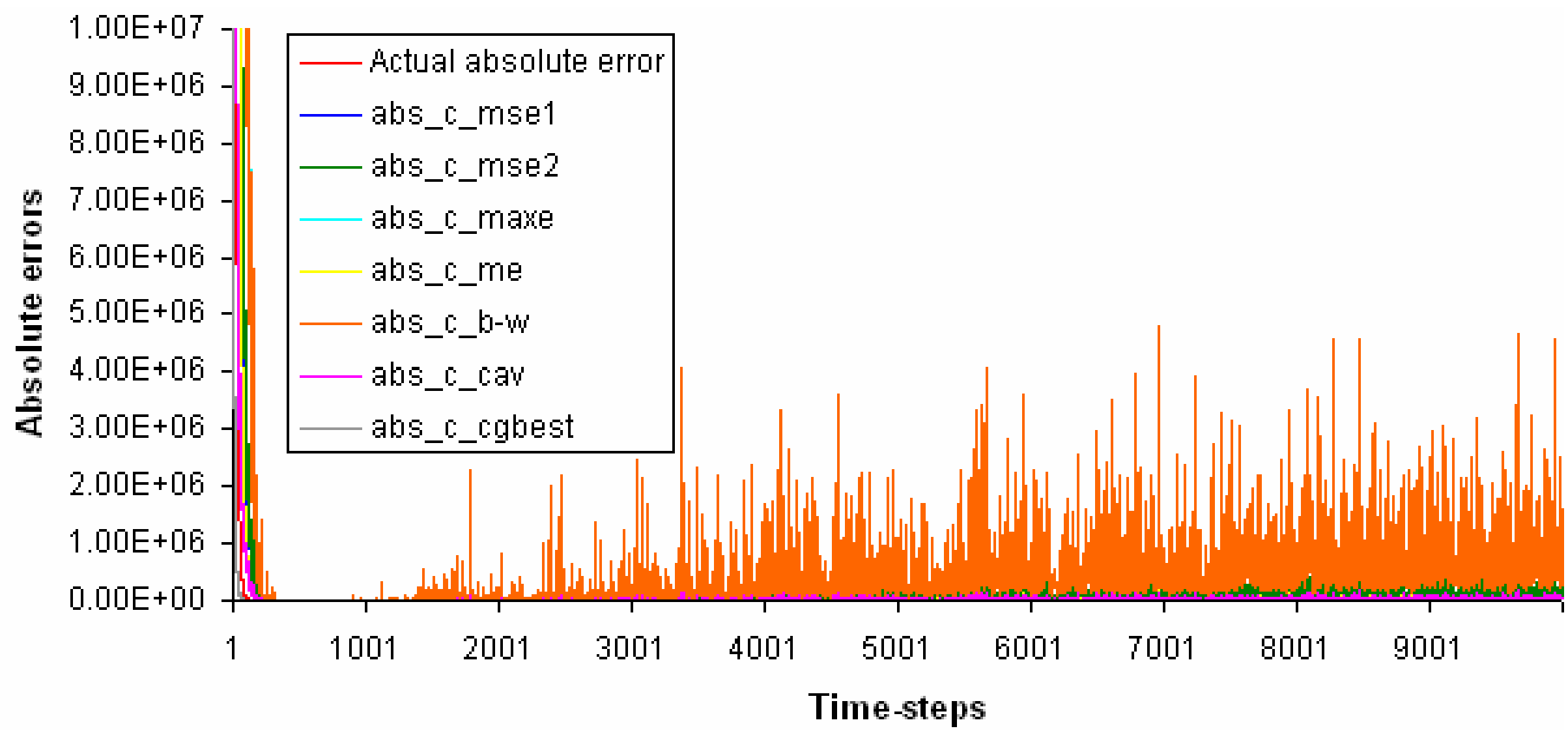

**Fig. 7. 3**: Evolution of the mean absolute errors for the BSt-PSO$^{(c)}$ optimizing the Rosenbrock function.

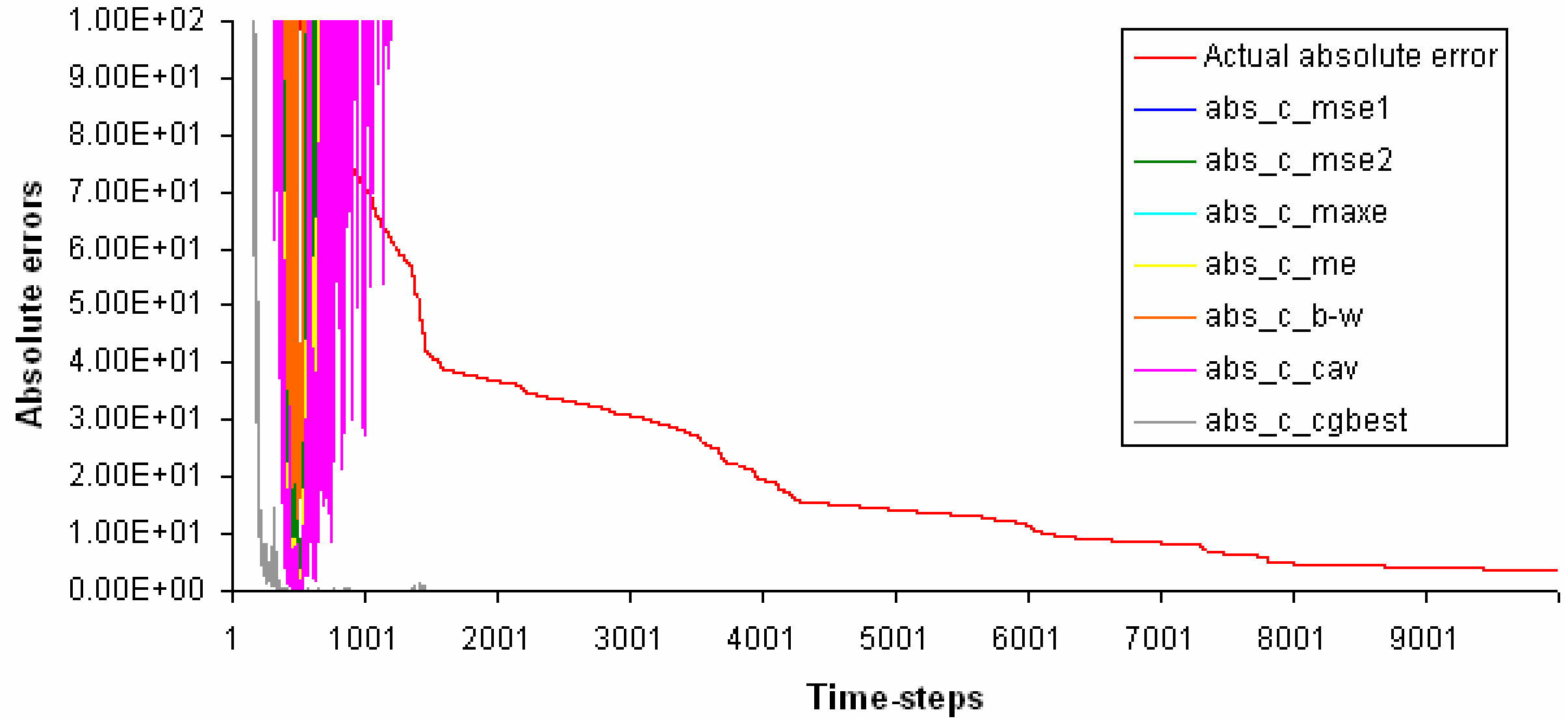

**Fig. 7. 4**: Evolution of the mean absolute errors for the BSt-PSO$^{(c)}$ optimizing the Rosenbrock function.

It is important to note that this strangely divergent behaviour does not take place when dealing with any other of the benchmark functions, nor when making use of a robust optimizer whose particles keep relatively high diversity, such us the **BSt-PSO**.

A similar phenomenon takes place for the **BSt-PSO** optimizing the 30-dimensional Schaffer f6 function, although the reason for this is easily understood. This optimizer does not present outstanding fine-clustering ability, so that the particles end up covering a certain area of the search-space, precisely where the highest and the lowest conflicts are located. Therefore, as the particles approach this area, the measures of error regarding the conflict values tend to increase although the particles are indeed clustering. This can be seen from **Fig. 7. 5**:

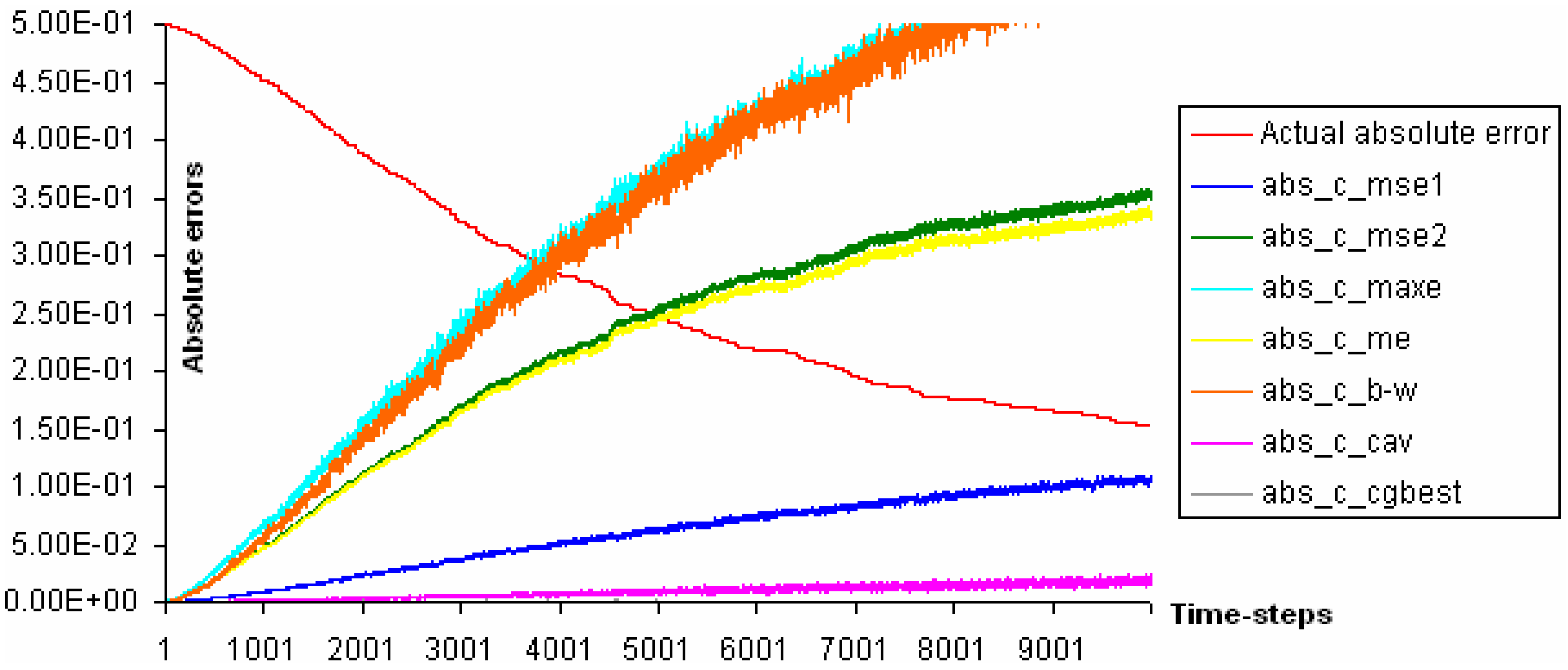

**Fig. 7. 5**: Evolution of the mean absolute errors for the BSt-PSO optimizing the 30-dimensional Schaffer f6 function.

In fact, even the measures of error corresponding to the **BSt-PSO$^{(c)}$** and the **BSt-PSO$^{(p)}$**—which do posses the ability to fine-cluster—increase while their particles approach the area where the best and worst conflicts are located, although they end up decreasing while the particles approach the area where the unique global optimum is placed, as it can be seen from **Fig. 7. 6** (refer to **Appendix 3** for a visualization of the 2-dimensional Schaffer f6 function).

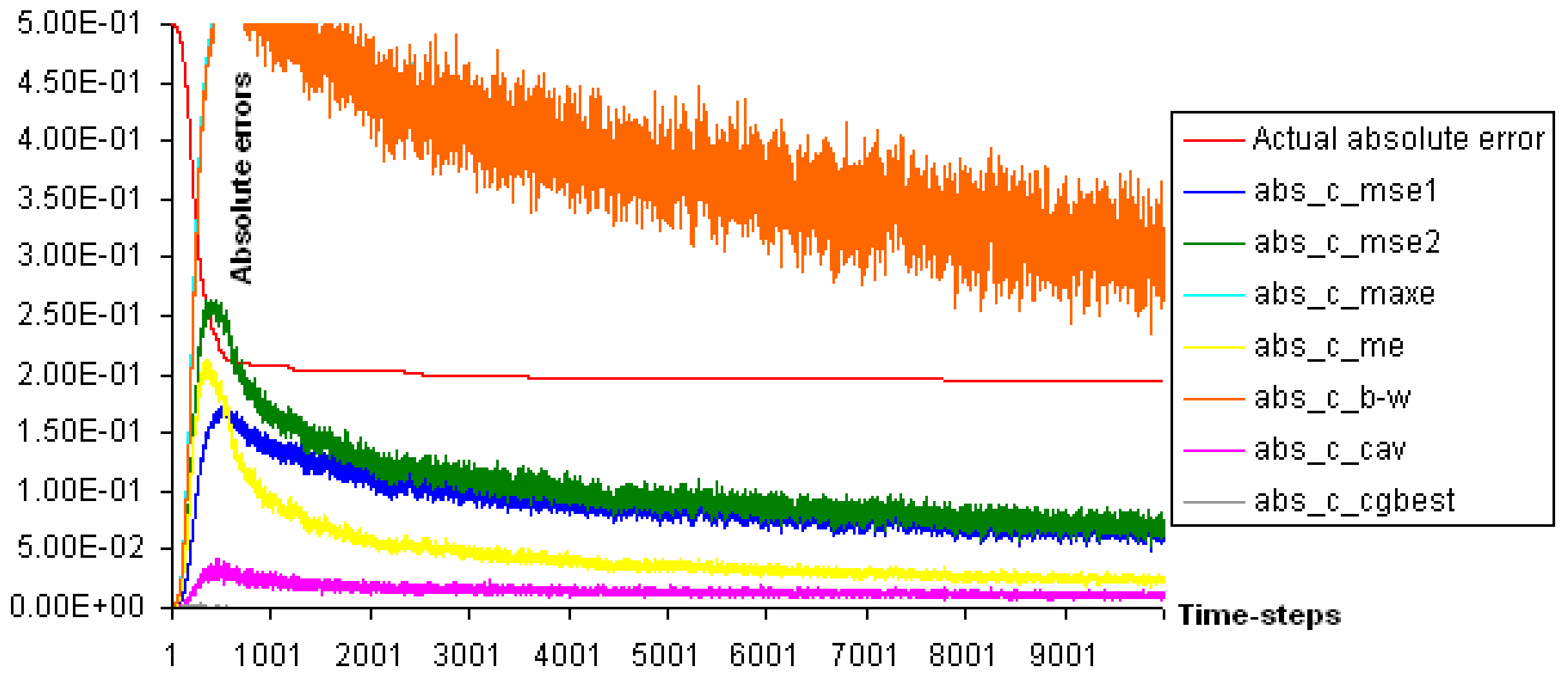

**Fig. 7. 6**: Evolution of the mean absolute errors for the BSt-PSO$^{(c)}$ optimizing the 30-dimensional Schaffer f6 function.

This situation, where the measures of error take very small values even before the search process begins, should be considered when developing stopping criteria for a general-purpose optimizer. For instance, it could be stated that the search cannot be stopped before a given number of time-steps is reached, disregarding the values of the measures of error meanwhile.

The fact that the measures of error regarding the conflict values increase while the particles cluster in **Fig. 7. 5**—and at the early stages of the search in **Fig. 7. 6**—, suggests that the particles of the **BSt-PSO**$^{(c)}$ and of the **BSt-PSO**$^{(p)}$ might not be diverging when optimizing the Rosenbrock function after all. Notice that although the general trend of this function is to present decreasing conflicts as the particles approach the global optimum, there can be some increase during the fine-tuning. A plot of the 2-dimensional Rosenbrock function in the proximity of the global optimum is shown in **Fig. 7. 7** (refer to **Appendix 3** for more detailed graphical visualization). If this was the reason for the increase in the measures of error, the latter would not continue to increase indefinitely.

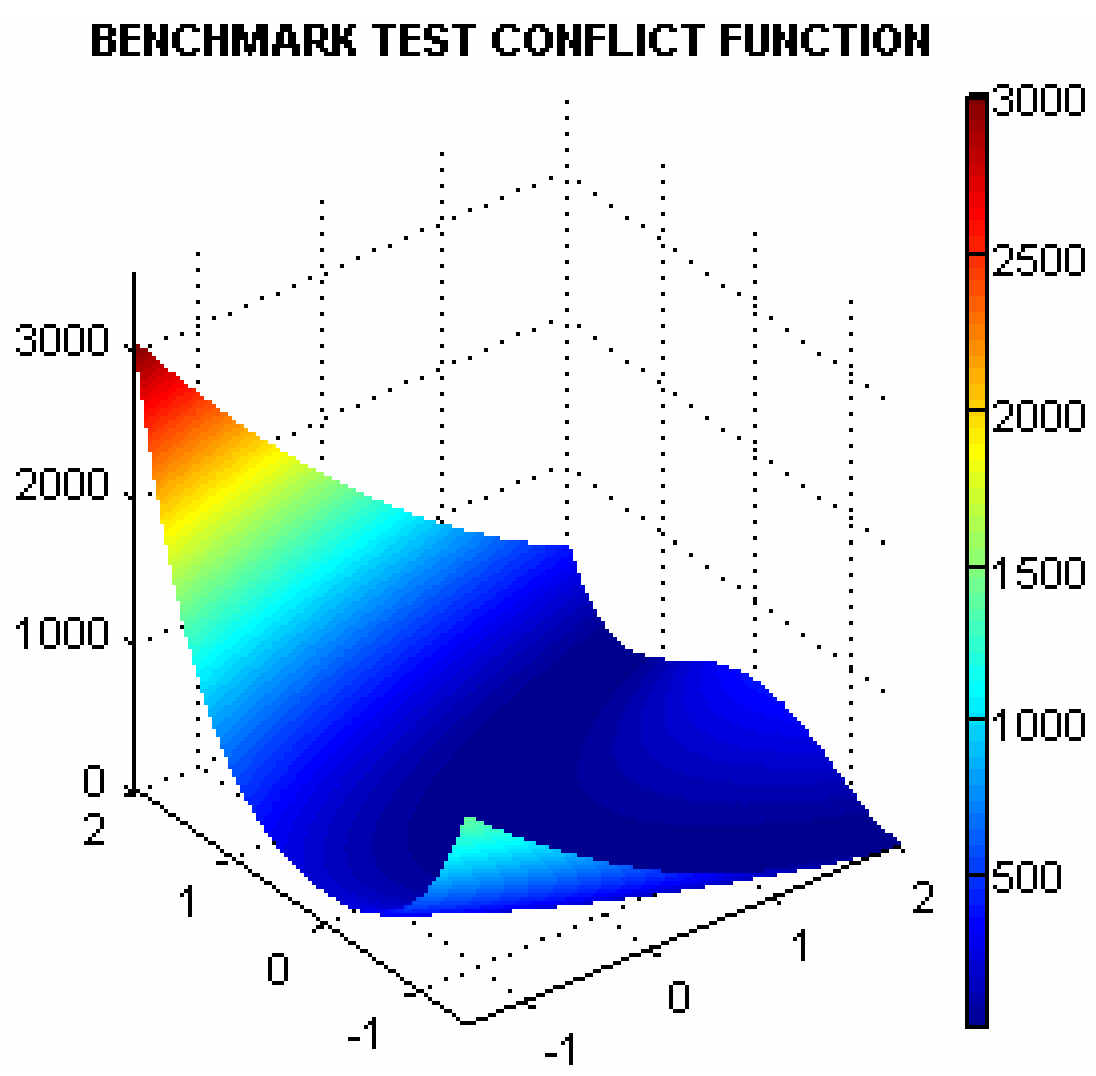

**Fig. 7. 7**: Plot of the 2-dimensional Rosenbrock function in the vicinity of the global optimum, which is located at the coordinates [1,1].

A final conclusion with regards to this increase being due to the divergence of the particles—as it may seem at first glance—or simply due to the shape of the Rosenbrock function in the vicinity of the global optimum will be reached in time, when developing measures of error regarding the particles' positions, later in this chapter.

The analysis of the evolution of the mean best and mean average conflicts in the experiments run along section **6.4** led to inferring that the particles of the **BSt-PSO**$^{(c)}$ and the **BSt-PSO**$^{(p)}$ performed an almost complete implosion when dealing with the Rastrigrin and the Griewank functions, as opposed to the particles of the **BSt-PSO**. This inference is corroborated by the results obtained here, where all the measures of error regarding the conflict values end up virtually nullifying. Thus, the evolution of the mean best conflict found stagnates once the implosion takes place for the **BSt-PSO**$^{(c)}$ and the **BSt-PSO**$^{(p)}$, while improvement never stops for the **BSt-PSO**, which does not lose diversity. As argued several times before, diversity appears to be critical to deal with functions that exhibit numerous local optima. The evolution of the different measures of error proposed and of the actual absolute error for the **BSt-PSO**, the **BSt-PSO**$^{(c)}$, and the **BSt-PSO**$^{(p)}$ optimizing the Rastrigin function is shown in **Fig. 7. 8** to **Fig. 7. 10**. Recall that the mean best conflict equals the actual absolute error.

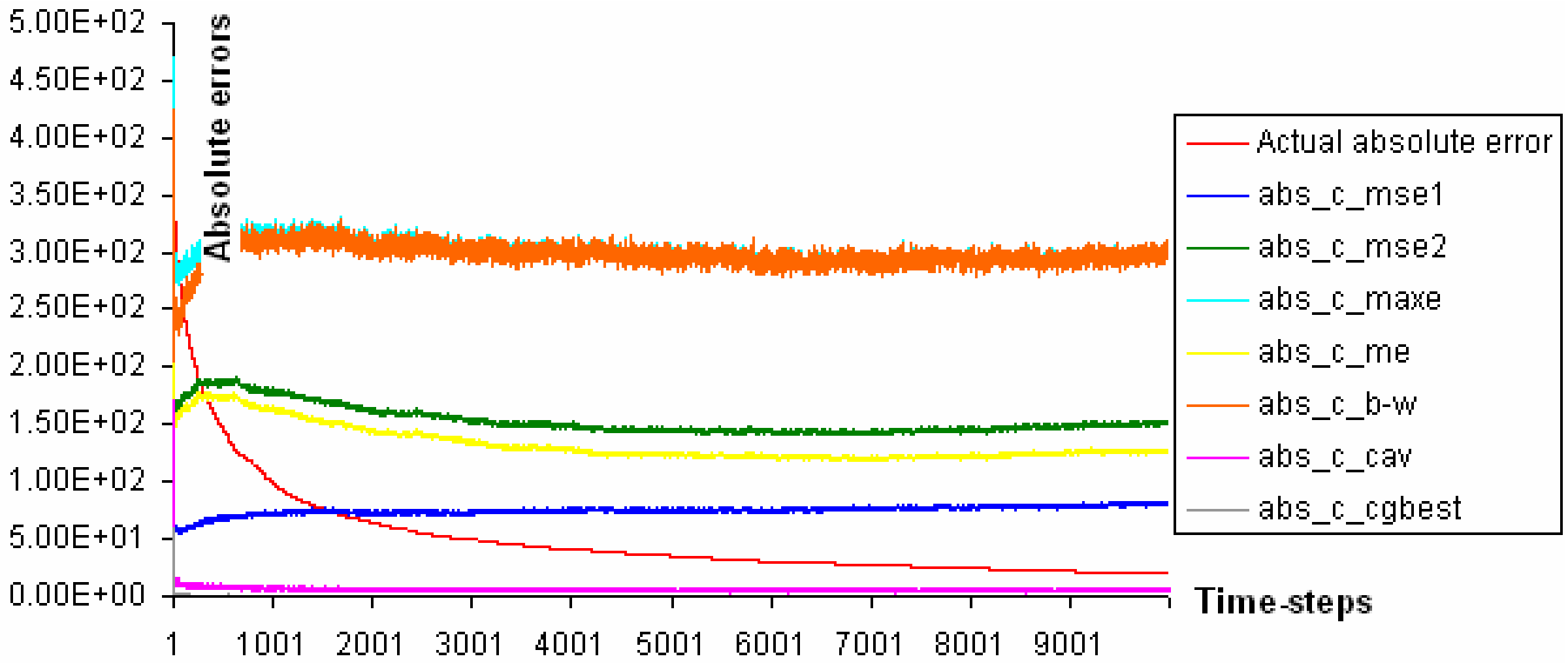

**Fig. 7. 8**: Evolution of the mean absolute errors for the BSt-PSO optimizing the 30-dimensional Rastrigin function.

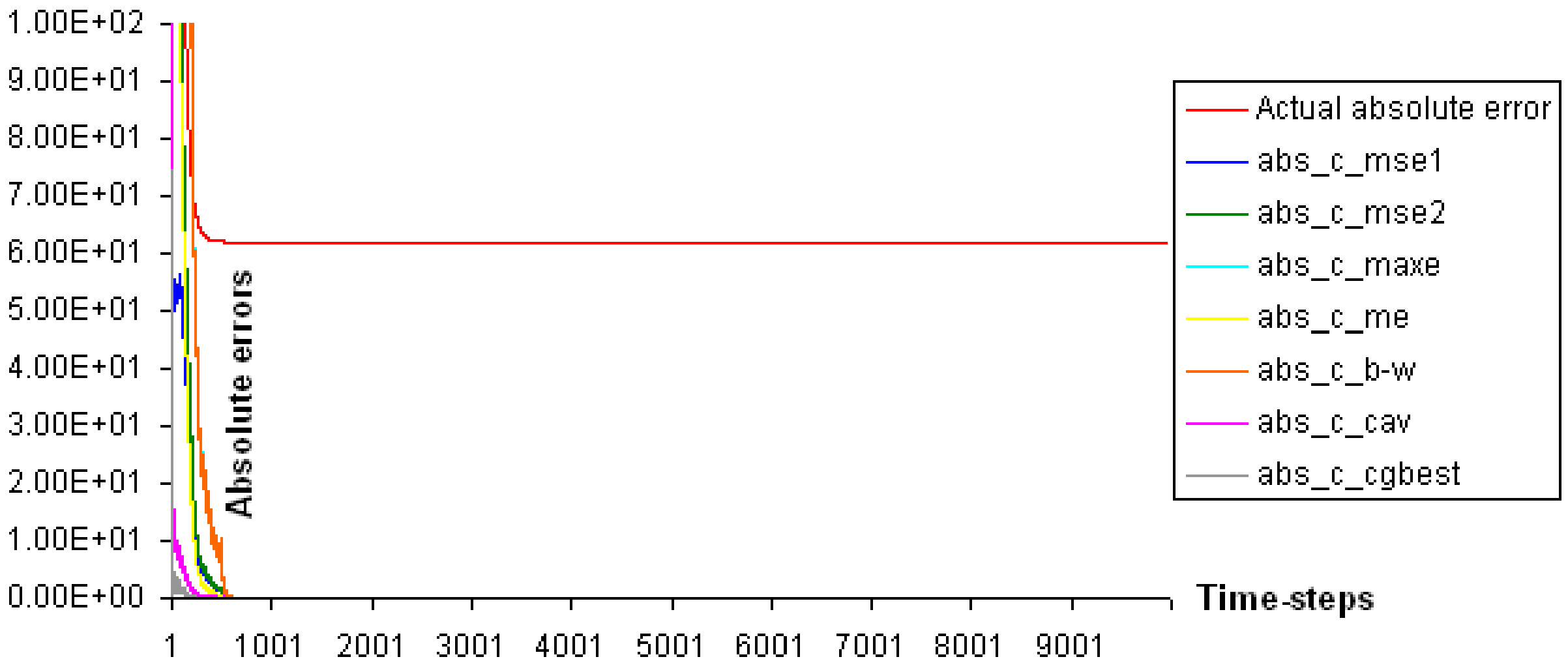

**Fig. 7. 9**: Evolution of the mean absolute errors for the BSt-PSO$^{(c)}$ optimizing the 30-dimensional Rastrigin function.

The evolution of the measures of error when optimizing the Griewank function is qualitatively similar, although the trend-lines of those corresponding to the BSt-PSO continuously decrease rather than stagnate, as oppose to what is observed in **Fig. 7. 8** (refer to digital **Appendix 4**).

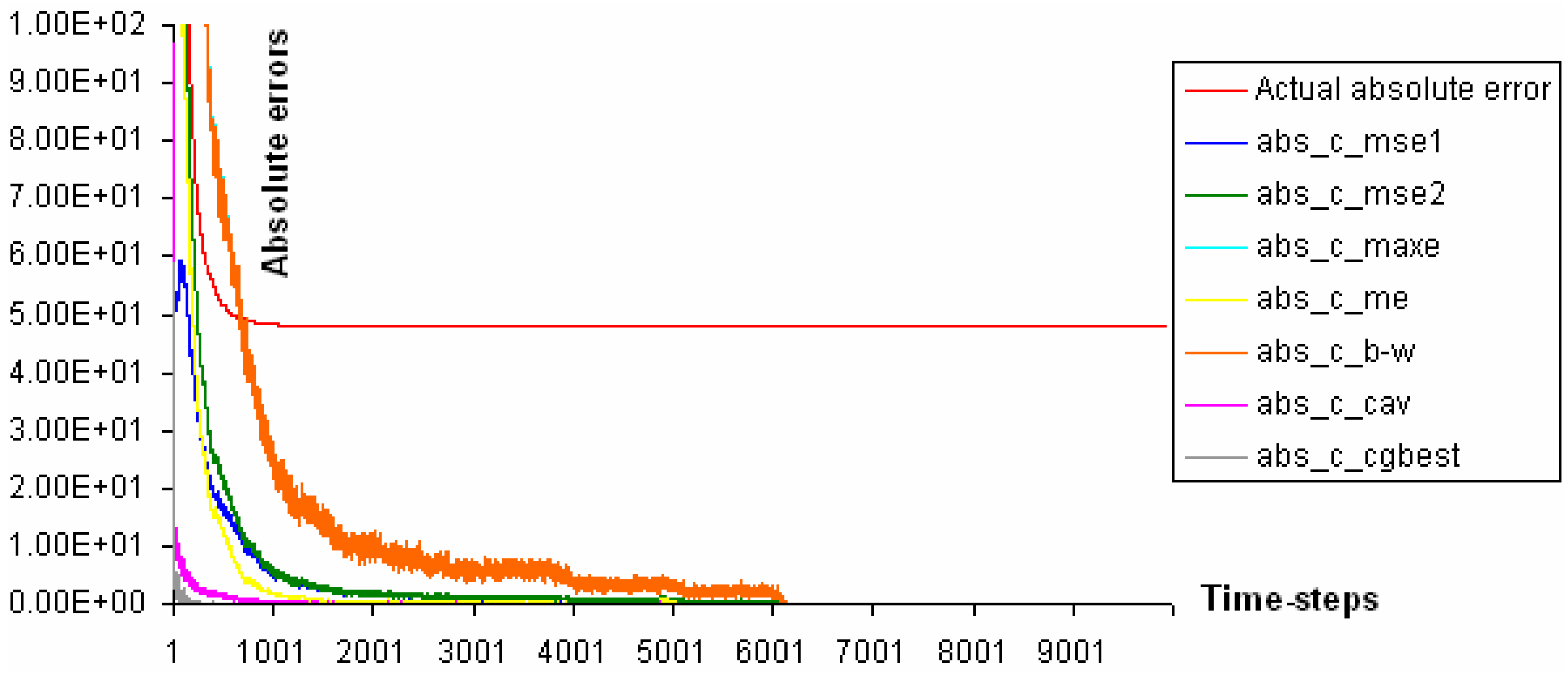

**Fig. 7. 10**: Evolution of the mean absolute errors for the BSt-PSO$^{(p)}$ optimizing the 30-dimensional Rastrigrin function.

The Schaffer f6 function, both in its 2-dimensional and in its 30-dimensional versions, proved once again to present serious difficulties to the fine-clustering of the particles. It has been argued before that this might be due to the local optima in the form of ring-like depressions—rather than in the form of valleys—surrounding the global optimum. Thus, the measures of error proposed do not nullify, although their trend-lines decrease except for the **BSt-PSO** optimizing the 30-dimensional version (see **Fig. 7. 5**).

In summary, very small values of the proposed measures of error imply that the particles have achieved a high degree of clustering, which in turn implies that further improvement of the best solution found is unlikely[3]. Although there are some functions which make the fine-clustering of the particles remarkably difficult, namely the Rosenbrock and the Schaffer f6 functions, they are considered here as extremely hard problems that present serious challenges to the optimizers themselves, rather than to the design of measures of error. However, some robust tunings of the parameters of the basic algorithm result in optimizers that do not posses the ability to fine-cluster (e.g. **BSt-PSO**). Usually, these algorithms exhibit no stagnation of the evolution of the best solution found, although the rate of improvement is typically lower. Furthermore, the values of the measures of error proposed are commonly higher, with wider, uneven oscillations. Therefore, the use of the proposed measures of error within a single time-step seems inconvenient for optimizers which exhibit poor clustering ability.

[3] Note that, in those cases where the measures of error end up almost nullifying, the stagnation of the curve of the evolution of the mean best solution found takes place at an earlier stage.

As previously mentioned, although the probabilistic nature of the PSO method makes it necessary to evaluate the mean evolution of the proposed measures of error, the design of the stopping criteria, which involves the definition of permissible values of the measures of error, makes it necessary to analyze their evolution for a single run. A rougher and more erratic behaviour is to be expected.

#### 7.4.1.2.1.2 Single run

As predicted, the evolution of the proposed measures of error is indeed more erratic, exhibiting much wider, uneven oscillation than in the experiments for the average among 50 runs. It is fair to note that the **abs_c_maxe** and the **abs_c_b-w** are the measures that show the most erratic behaviour. The evolution of the measures of error for the **BSt-PSO$^{(c)}$** optimizing the 30-dimensional Schaffer f6 function is shown in **Fig. 7. 11** (refer to **Fig. 7. 6** to observe the "smoothing effect" of averaging among the 50 runs on the evolution of the absolute errors). The complete set of results, including all the six benchmark functions, can be found in digital **Appendix 4**.

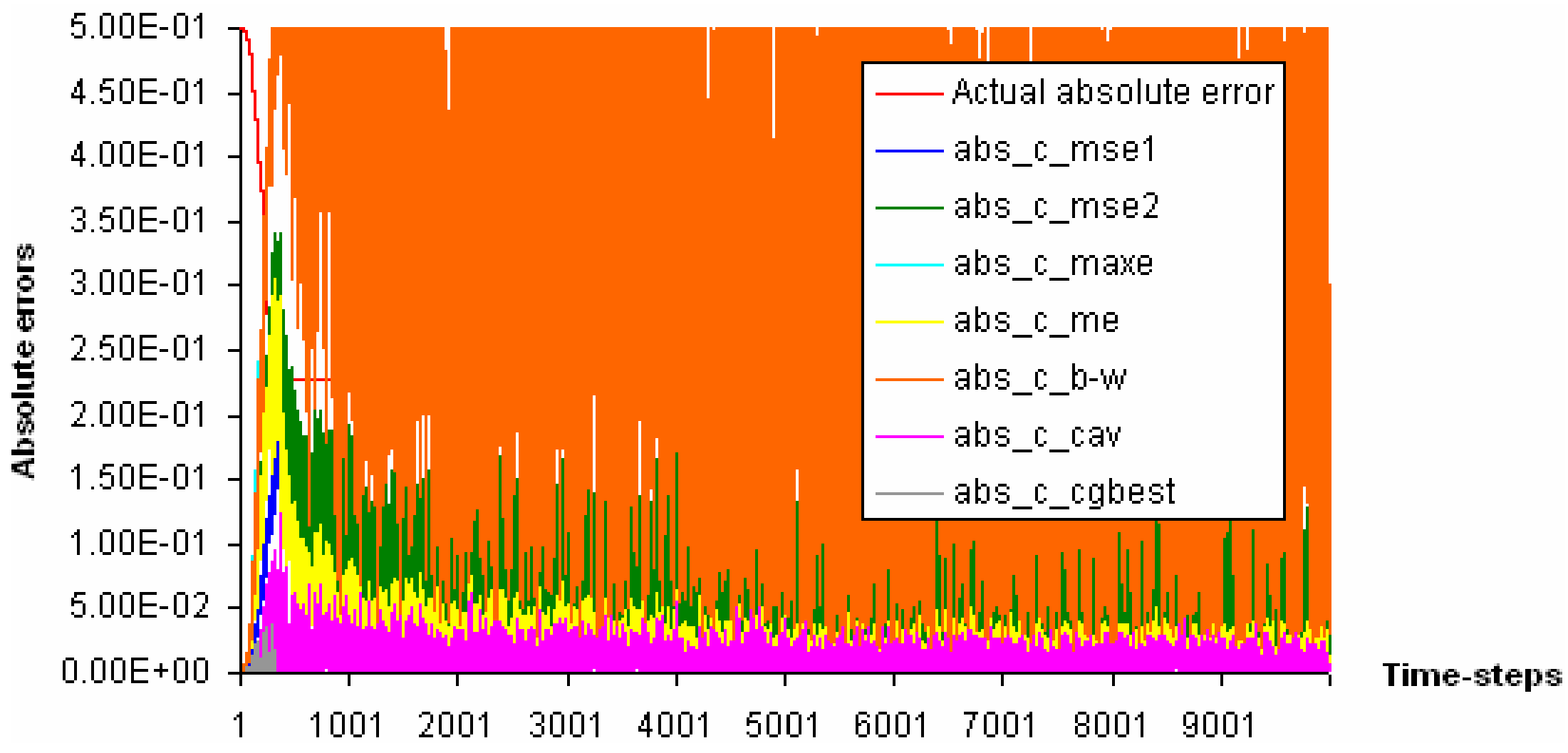

**Fig. 7. 11**: Evolution of the absolute errors for the BSt-PSO$^{(c)}$ optimizing the 30-dimensional Schaffer f6 function.

It is clear that this kind of evolution of the measures of error makes it difficult to design stopping criteria based on their permissible limits. Given that the stopping criteria need to be designed for a single run, it becomes necessary to smooth this erratic behaviour somehow. Some alternatives could be to make use of either the trend-lines or of the lower envelopes of these graphs. However, these alternatives are not easily implemented. It is proposed here to smooth the amplitude of the oscillations by averaging the measures of error corresponding to the last $k$ time-steps. The value of $k$ was arbitrarily taken equal to 50, which is a number big enough to serve its function, while still small in relation to the maximum number of time-steps permitted for the whole search. This strategy is then tried by simply making use of the results obtained from the experiments run for the single and 50 runs. Note that the average of the absolute measures of errors of the last 50 time-steps is defined only from the time-step 50 on. Thus, the average is computed for the first 49 time-steps just as the average of all the measures of error up to the current time-step. The "smoothing effect" of this strategy can be observed by comparing **Fig. 7. 12** to **Fig. 7. 11**.

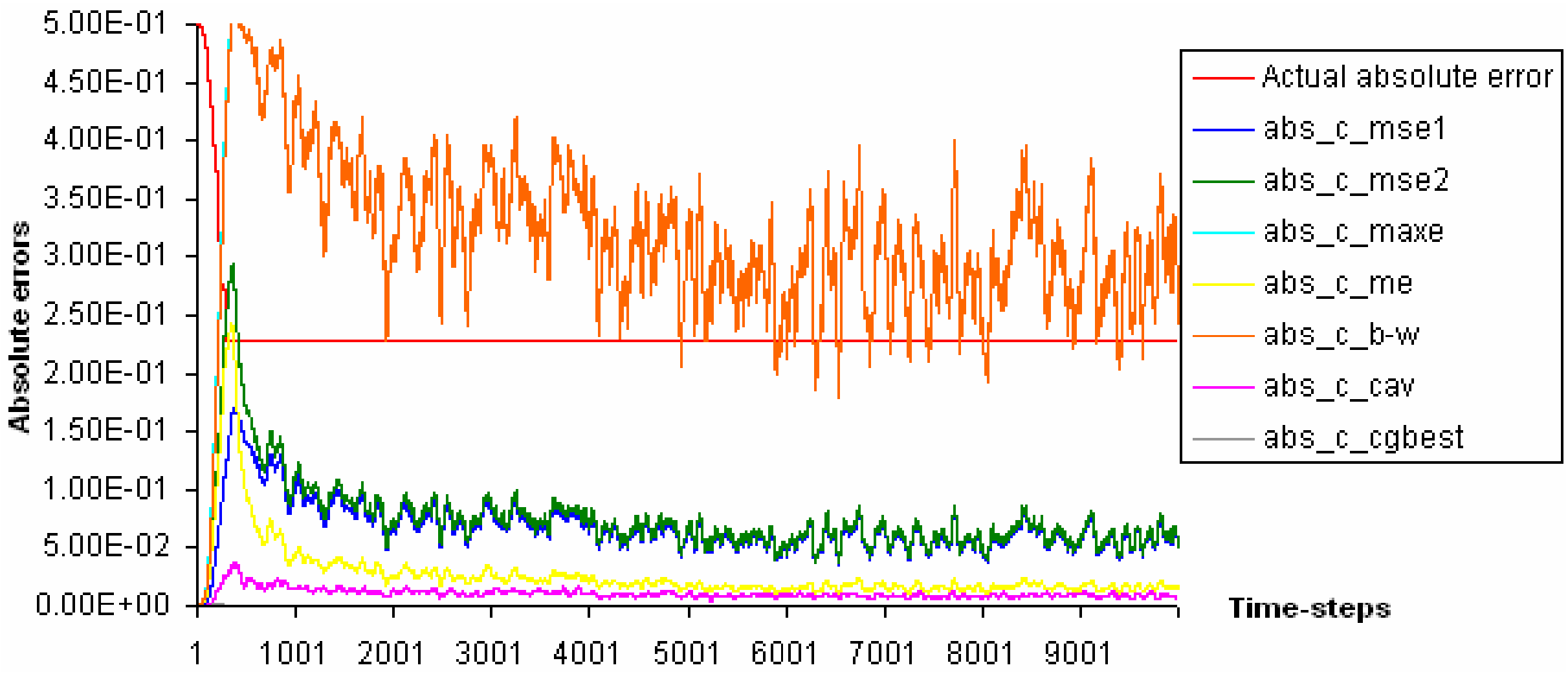

**Fig. 7. 12**: Evolution of the average of the absolute errors of the last 50 time-steps for the BSt-PSO(c) optimizing the 30-dimensional Schaffer f6 function.

### 7.4.1.2.2 Asymmetric initialization

There are some—typically unconstrained problems—where the optimum is not contained within the space spanned by the particles' initial positions. The evolution of the proposed measures of error may differ in these cases with respect to those tested with symmetric initializations that contain the solution. Thus, all the experiments carried out so far within this chapter are repeated for the asymmetric initializations shown in **Table 7. 1**. It is fair to note that the asymmetric initialization is also useful to test the performance of the optimizers without the risk of a centre-seeking algorithm accidentally finding the global optimum.

The evolution of the particles' positions for the **BSt-PSO$^{(c)}$** optimizing the Sphere function and making use of the asymmetric initialization is shown in **Fig. 7. 13**:

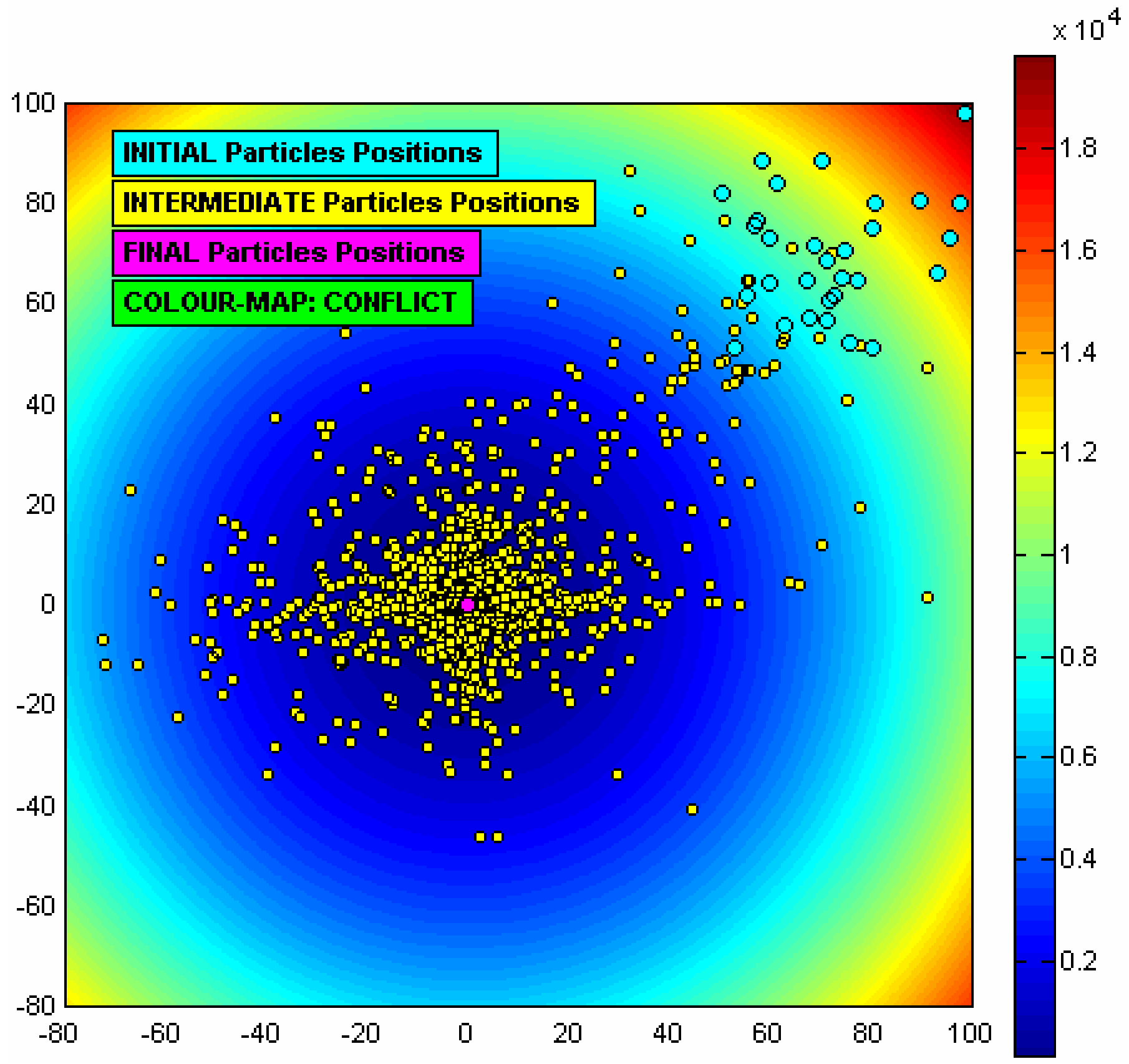

**Fig. 7. 13**: Evolution of the particles' positions for an asymmetric initialization in the range [50,100], where the function to be optimized is the Sphere function, the maximum number of time-steps equals 10000, and the optimizer is the BSt-PSO$^{(c)}$.

As it can be observed from animations of the evolution of the particles' positions, the initial clustering of the particles takes place very quickly, while the fine-tuning of the search takes noticeably longer. This results in the evolution of the proposed measures of error for the asymmetric initialization being qualitatively similar to that of the symmetric one. It is fair to remark, however, that the best solutions the optimizer is able to find may differ substantially, especially for those optimizers which exhibit strong clustering ability dealing with functions that exhibit numerous local optima. As an example, the evolution of the average of the mean absolute errors of the **BSt-PSO$^{(p)}$** optimizing the Rastrigrin function for the symmetric and asymmetric initializations are shown in **Fig. 7. 14** and **Fig. 7. 15**, respectively. Recall that the average is computed among the last 50 time-steps, the mean is computed among 50 runs of the algorithm, and the actual absolute error equals the best solution found for this function.

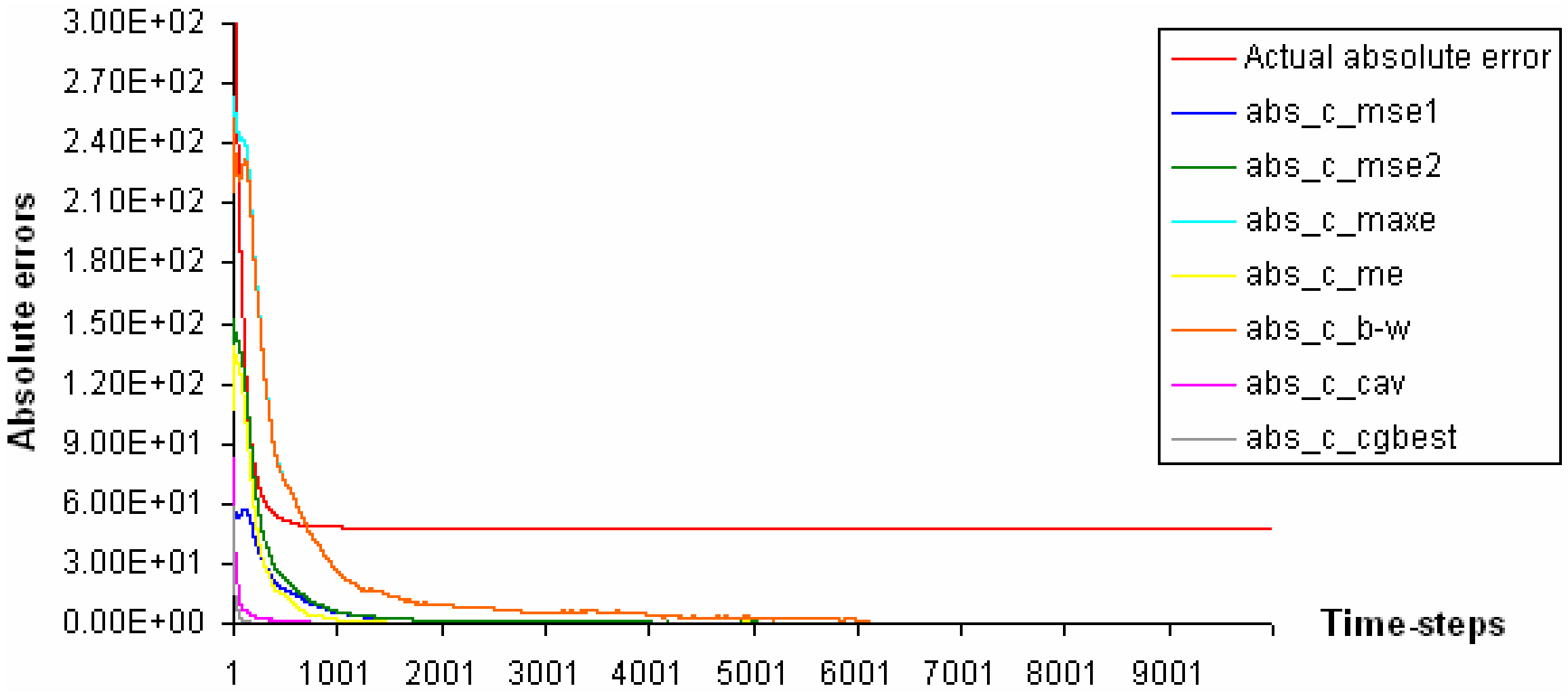

**Fig. 7. 14**: Evolution of the average of the mean absolute errors of the last 50 time-steps for the BSt-PSO(p) optimizing the 30-dimensional Rastrigin function, making use of the symmetric initialization.

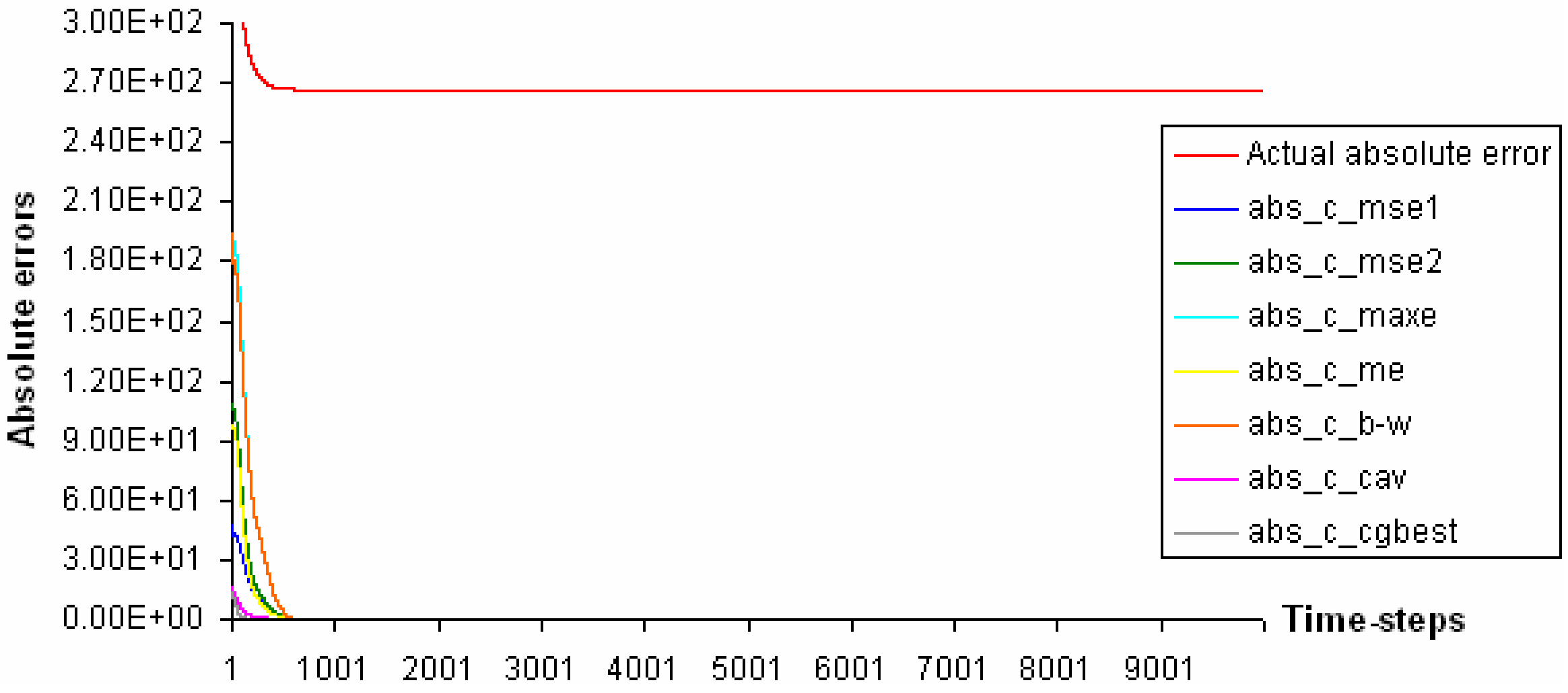

**Fig. 7. 15**: Evolution of the average of the mean absolute errors of the last 50 time-steps for the BSt-PSO(p) optimizing the 30-dimensional Rastrigin function, making use of the asymmetric initialization.

As expected, the particles of the optimizer get trapped in a poorer local optimum for the asymmetric initialization. It is interesting to observe, however, that the measures of the degree of clustering[4] of the particles take longer to approach zero—which would imply a complete implosion—for the symmetric initialization, as opposed to what could be expected. This may be due to the fact that this asymmetric initialization spreads the particles over a smaller region in comparison to the symmetric one, so that the particles' clustering takes place more quickly.

[4] That is to say, the measures of error defined within the current time-step. Note, however, that they are in reality the average of the measures corresponding to the last 50 time-steps.

It is even more surprising to observe that the measures of error defined between consecutive time-steps[5] (e.g. the **abs_c_cav**) exhibit the same behaviour. Clear examples of this are shown in **Fig. 7. 16** and **Fig. 7. 17**:

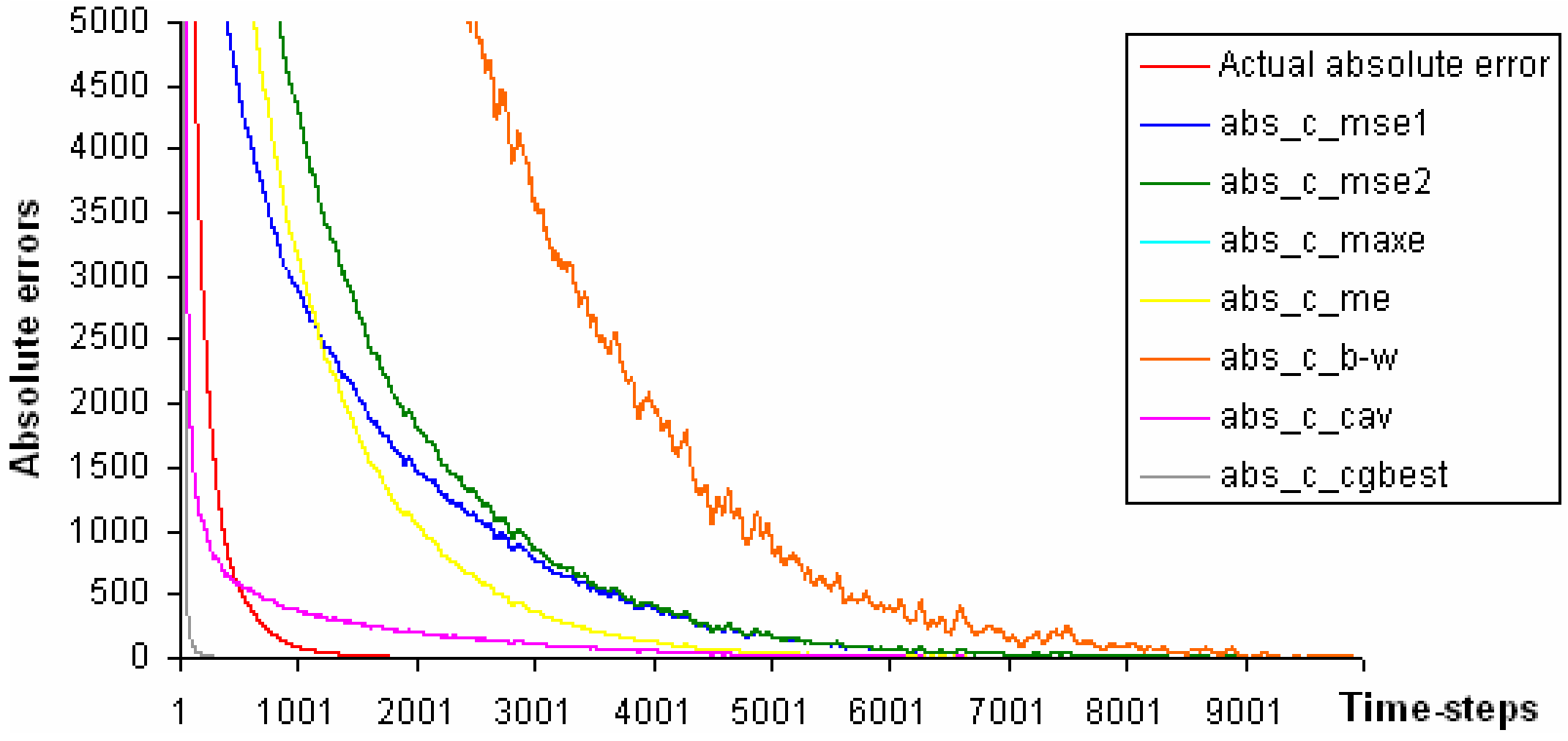

**Fig. 7. 16**: Evolution of the average of the mean absolute errors of the last 50 time-steps for the BSt-PSO$^{(p)}$ optimizing the 30-dimensional Sphere function, making use of the symmetric initialization.

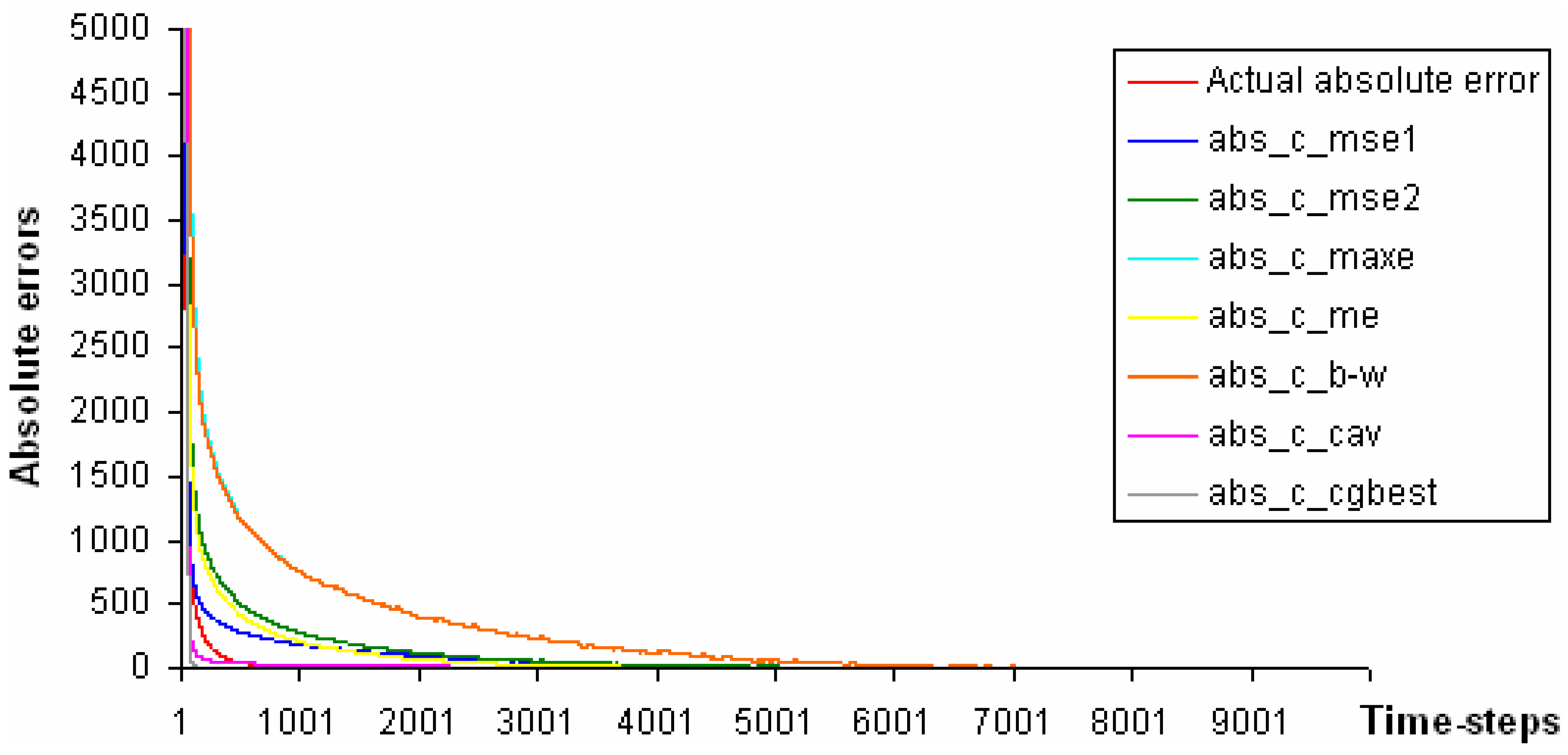

**Fig. 7. 17**: Evolution of the average of the mean absolute errors of the last 50 time-steps for the BSt-PSO$^{(p)}$ optimizing the 30-dimensional Sphere function, making use of the asymmetric initialization.

Nevertheless, as previously stated, the general behaviour of the proposed measures of error is qualitatively similar for both initializations. Regarding the performance of the optimizers, it is clear that the asymmetric initialization makes the problem harder, especially when dealing with functions that exhibit numerous local optima. However, this issue is beyond the scope of this chapter, and general-purpose optimizers whose performances are tested on symmetric and asymmetric initializations will be dealt with along **Chapter 9**.

[5] Again, they are in reality the average of the measures corresponding to the current plus the previous 49 time-steps.

It was argued before that the absolute errors are viewed here as a first step towards the relative errors, so that only the qualitative features of their evolution were discussed within this section. Thus, the experiments run showed that not only are the evolutions of the **abs_c_maxe** and the **abs_c_b-w** very much alike, but they are also the oddest ones among all the proposed measures of error. Therefore, they are removed from the set of proposed absolute errors to be considered for the development of the relative ones hereafter.

## 7.4.2 Relative errors

It is not straightforward to decide on a convenient value to relate the absolute errors to, aiming to make the permissible values of the relative errors independent from the conflict function. The first value that comes to mind is the best solution found so far, so that the permissible relative error can be set as a percentage of the true solution[6]. However, this strategy does not work properly when the optimum is near or equal to zero, as previously discussed in section **7.3**. In addition, it seems reasonable to argue that the error should be limited to a percentage of the maximum error possible rather than to a percentage of the true solution. For instance, consider the Schaffer f6 function (refer to **Appendix 4**) modified by the addition of a very big constant, say 10000. A candidate solution equal to 10000.5 results in a relative error equal to $5 \times 10^{-5}$, which appears to be acceptable at first glance. However, since this function oscillates within the interval $[10000, 10001]$, even a random solution is likely to exhibit such an error. In fact, the maximum relative error possible is less than $1 \times 10^{-4}$. Therefore, it is proposed here to relate the absolute errors to the difference between the best and worst solutions that the algorithm is able to find along the whole search. In order to make this possible, a specialized sub-swarm composed of only five particles is added to the population, which is in quest for the worst rather than for the best conflict. Because the worst conflict is only used for the computation of the relative errors, high precision is not essential. Notice that this strategy is only possible for problems where the search-space is constrained to a finite region. Hence the “preserving feasibility” technique is brought to this section to handle hyper-cube-like constraints to the search-space. This is a robust constraint-handling technique that consists of initializing the particles randomly and repeatedly until the whole population is spread over feasible space, and thereafter simply banning from memory the infeasible solutions. For further details on this constraint-handling technique, refer to [40] and to **Chapter 10**.

[6] This assumes that the best solution found so far equals the exact solution, which is just an approximation.

Considering the results obtained from the experiments run for the analysis of the evolution of the absolute errors, the strategy of involving the current and a number of previous consecutive time-steps in the computation of the errors is adopted hereafter. While the number of time-steps considered for the definition of the absolute errors was set equal to 50, it is decided here to set it equal to 100 so as to smooth the graphs of their evolution even more. Note that this number of time-steps involved in the definition of the relative errors is equal to only 1% of the maximum number of time-steps permitted for the whole search.

The same measures of error defined by equations **(7. 5)** to **(7. 11)** are considered here, except that the **abs_c_maxe** and the **abs_c_b-w** are removed, 100 time-steps are involved in their computation, and the maximum absolute error possible found so far (i.e. $cgworst - cgbest$) is used to normalize the measures of error to a range $[0,1]$.

#### 7.4.2.1 Errors definitions

##### 7.4.2.1.1 Within the current time-step

- **rel_c_mse1**: average of the square root of the mean squared error of the particles’ current conflicts with respect to the current average conflict corresponding to the last 100 time-steps, related to the difference between the worst and the best conflicts found so far:

$$\text{rel_c_mse1}^{(t)} = \frac{\sum_{i=t-99}^{t} \sqrt{\frac{\sum_{j=1}^{m}\left(c_j^{(i)} - \bar{c}^{(i)}\right)^2}{m}}}{100 \cdot \left(cgworst^{(t)} - cgbest^{(t)}\right)} = \frac{\sum_{i=t-99}^{t} \sqrt{\sum_{j=1}^{m}\left(c_j^{(i)} - \bar{c}^{(i)}\right)^2}}{100 \cdot \sqrt{m} \cdot \left(cgworst^{(t)} - cgbest^{(t)}\right)} \quad \textbf{(7. 12)}$$

- **rel_c_mse2**: average of the square root of the mean squared error of the particles' current conflicts with respect to the best conflict found so far corresponding to the last 100 time-steps, related to the difference between the worst and the best conflicts found so far:

$$\mathrm{rel_c_mse2}^{(t)} = \frac{\sum_{i=t-99}^{t} \sqrt{\sum_{j=1}^{m} \left(c_j^{(i)} - cgbest^{(i)}\right)^2}}{100 \cdot \sqrt{m} \cdot \left(cgworst^{(t)} - cgbest^{(t)}\right)} \tag{7. 13}$$

- **rel_c_me**: average of the difference between the current average conflict and the best conflict found so far corresponding to the last 100 time-steps, related to the difference between the worst and the best conflicts found so far:

$$\mathrm{rel_c_me}^{(t)} = \frac{\sum_{i=t-99}^{t} \frac{\sum_{j=1}^{m} \left(c_j^{(i)} - cgbest^{(i)}\right)}{m}}{100 \cdot \left(cgworst^{(t)} - cgbest^{(t)}\right)} = \frac{\sum_{i=t-99}^{t} \left(\bar{c}^{(i)} - cgbest^{(i)}\right)}{100 \cdot \left(cgworst^{(t)} - cgbest^{(t)}\right)} \tag{7. 14}$$

### 7.4.2.1.2 Between consecutive time-steps

- **rel_c_cav**: average of the absolute difference between the current and the preceding average conflicts corresponding to the last 100 time-steps, related to the difference between the worst and the best conflicts found so far:

$$\mathrm{rel_c_cav}^{(t)} = \frac{\sum_{i=t-99}^{t} \mathrm{abs}\left(\bar{c}^{(i)} - \bar{c}^{(i-1)}\right)}{100 \cdot \left(cgworst^{(t)} - cgbest^{(t)}\right)} = \frac{\mathrm{abs}\left(\bar{c}^{(t)} - \bar{c}^{(t-100)}\right)}{100 \cdot \left(cgworst^{(t)} - cgbest^{(t)}\right)} \tag{7. 15}$$

- **rel_c_cgbest**: average of the absolute difference between the best conflicts found up to the current and up to the preceding time-steps corresponding to the last 100 time-steps, related to the difference between the worst and the best conflicts found so far:

$$\mathrm{rel_c_cgbest}^{(t)} = \frac{\mathrm{abs}\left(cgbest^{(t)} - cgbest^{(t-100)}\right)}{100 \cdot \left(cgworst^{(t)} - cgbest^{(t)}\right)} = \frac{cgbest^{(t-100)} - cgbest^{(t)}}{100 \cdot \left(cgworst^{(t)} - cgbest^{(t)}\right)} \tag{7. 16}$$

### 7.4.2.2 Experimental results

It was argued before that, although the asymmetric initialization makes the optimization problem noticeably harder, it does not introduce qualitative differences into the evolution of the proposed measures of error. Therefore, the experiments are carried out hereafter for the symmetric initialization only, although the evolution of the measures of error is analyzed for both the average among fifty runs and a single run. The first case considers the probabilistic nature of the algorithm, while the second one considers the fact that averaging a number of independent runs smoothes the oscillations of the evolution of the proposed measures of error.

#### 7.4.2.2.1 Fifty runs

As opposed to the absolute errors studied along section **7.4.1**, the relative errors proposed in section **7.4.2.1** are constrained to the range $[0,1]$. It is important to observe that the strategy of involving the last 100 time-steps in the computation of the measures of error—as well as the averaging of the fifty runs in this case—effectively serves the function of smoothing the curves of their evolution, which have the inherent tendency to display odd oscillations.

The relatively smooth evolution of these measures of relative errors, in addition to the fact that they are constrained to the range $[0,1]$, makes them well suitable for the development of stopping criteria, which can be done by setting permissible values for them as a percentage of the maximum error possible. It is fair to note that, since the latter—which is estimated as $cgworst - cgbest$—is updated at each time-step, the maximum error assumed to be possible is a dynamic reference point. Nevertheless, its value increases with time, and the real value can never be smaller than the best one found by the algorithm.

The curves of the evolution of the measures of relative errors reveal the **BSt-PSO**’s lack of clustering ability. Thus, these measures—especially those defined within a single time-step—seem to be more useful for their use in the development of stopping criteria for optimizers which exhibit the ability to fine-cluster (e.g. the **BSt-PSO$^{(c)}$** and the **BSt-PSO$^{(p)}$**). Besides, it is clear that even the optimizers which exhibit such ability find it far more difficult to fine-cluster when dealing with the Schaffer f6 function than with the other functions in the test suite (refer to **Fig. 7. 18**, **Fig. 7. 21**, **Fig. 7. 23**, **Fig. 7. 25**, and digital **Appendix 4**). This can be inferred from the fact that all the proposed measures of relative errors display a rather erratic behaviour when optimizing the Schaffer f6 function—both in its 2-dimensional and in its 30-dimensional versions—while they conveniently approach zero when dealing with all the other functions. It is fair to note, however, that the evolution of the measures of relative error corresponding to the Rosenbrock function exhibit a similar behaviour to that of the Sphere function at first glance (refer to **Fig. 7. 18**), but, when zooming the plot in, a small progressive increase in the values of the error can be observed (refer to **Fig. 7. 19**), which is believed to be due to the small increase in the conflict values in the proximity to the global optimum. Since the Rosenbrock function presents a single global optimum, it has to be concluded that the particles also find it more difficult to fine-cluster when dealing with this function than with the Sphere, Rastrigrin, and Griewank functions.

In order to see the **BSt-PSO**'s lack of clustering ability with respect to the **BSt-PSO$^{(c)}$** and the **BSt-PSO$^{(p)}$**, compare **Fig. 7. 20** to **Fig. 7. 21**, **Fig. 7. 22** to **Fig. 7. 23**, and **Fig. 7. 24** to **Fig. 7. 25**. For the complete set of results and plots obtained from the experiments, refer to **Appendix 4**.

The question is now how to define appropriate permissible values for the measures of error so as to design stopping criteria. While this issue is discussed along section **7.6**, quantitative results which will be useful for such purpose are gathered in **Table 7. 2** to **Table 7. 7**. Note that the results obtained for the **BSt-PSO** are not included because of its lack of clustering ability, which results in inappropriate shapes of the curves of the evolution of these measures of error.

While the analysis of the average behaviour among the fifty runs is useful to take into account the probabilistic nature of the particle swarm optimizers—therefore, to set the appropriate permissible values—, it is also necessary to analyze the shape of the curves of the evolution of the proposed relative errors for a single run, given that the stopping criteria must be designed for a single run.

### 7.4.2.2.2 Single run

As expected, the curves of the evolution of the proposed relative errors exhibit rougher shapes for a single run than for the average among 50 runs. Nevertheless, they are still smooth enough to be useful for the design of stopping criteria. It can be observed that the curves of the evolution of the relative errors for the **BSt-PSO$^{(c)}$** and the **BSt-PSO$^{(p)}$** optimizing all the functions in the test suite but the Schaffer f6 function, do not differ greatly whether a single run or the average among 50 runs is carried out (compare **Fig. 7. 19** to **Fig. 7. 26**; **Fig. 7. 21** to **Fig. 7. 27**; **Fig. 7. 23** to **Fig. 7. 28**; and refer to **Appendix 4** for further analyses).

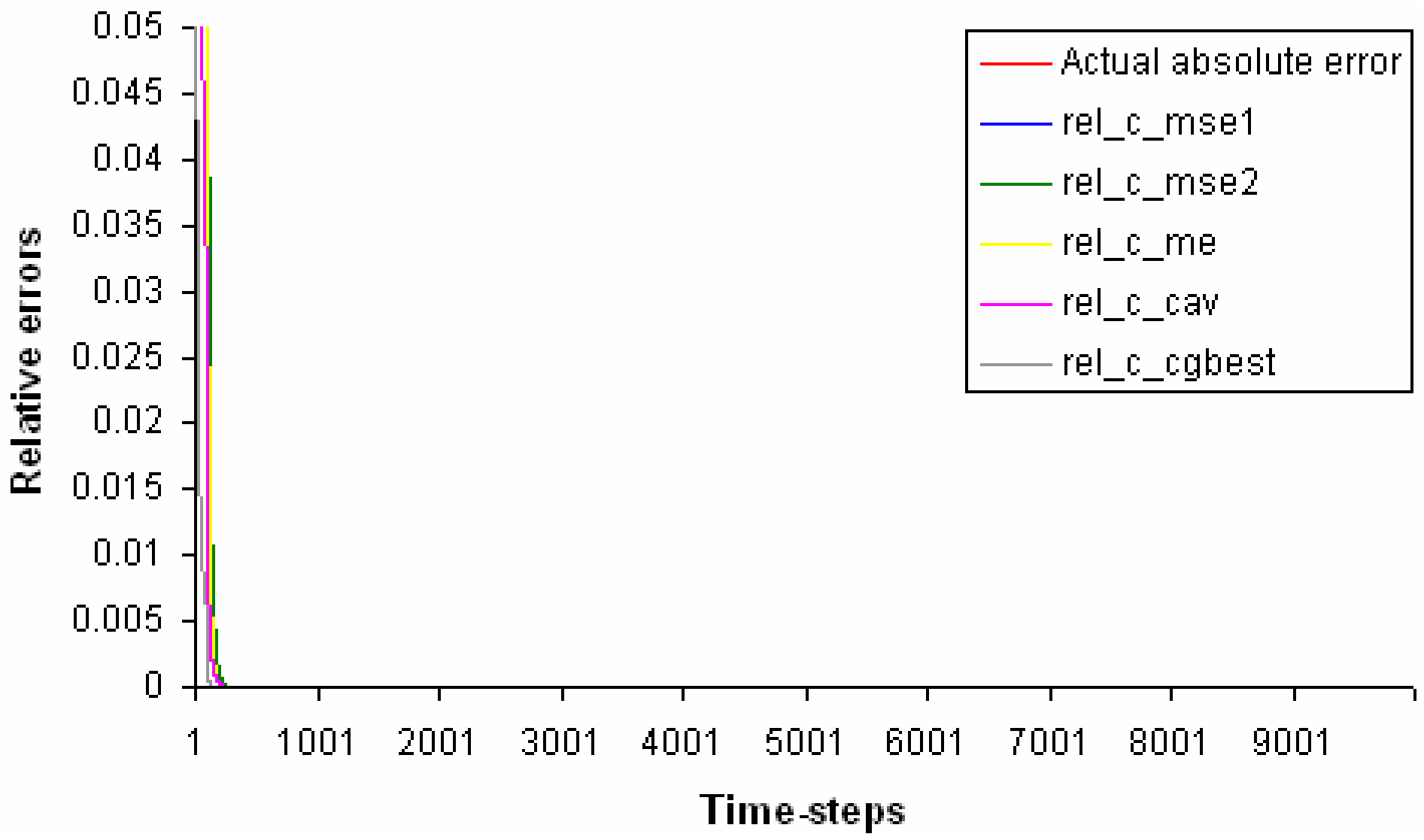

**Fig. 7. 18**: Evolution of the mean relative errors for the BSt-PSO(c) optimizing the 30-dimensional Rosenbrock function.

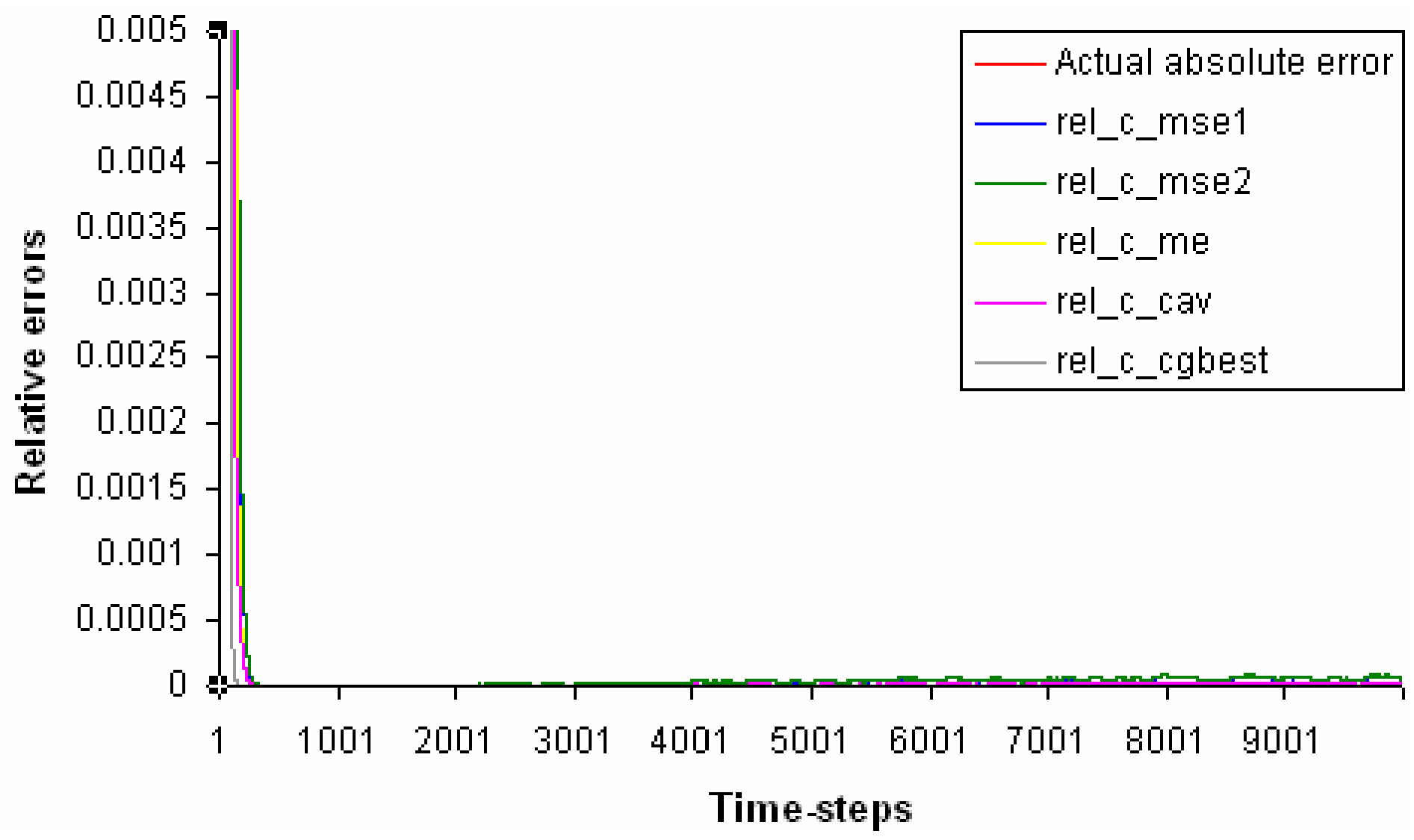

**Fig. 7. 19**: Evolution of the mean relative errors for the BSt-PSO(c) optimizing the 30-dimensional Rosenbrock function.

With regards to the Schaffer f6 function, it was argued before that the curves of the relative errors display rather erratic behaviours due to the fact that the particles of the optimizers find it difficult to fine-cluster when optimizing this function, even for those optimizers with the ability to do so. That is to say, the erratic behaviour is due to the poor performance of the algorithm when dealing with this problem, rather than to the inadequate design of the relative errors. The curves of the evolution of the relative errors for the **BSt-PSO$^{(c)}$** optimizing the 30-dimensional Schaffer f6 function is shown in **Fig. 7. 29** (refer to **Fig. 7. 25** to compare the curve obtained from a single run to that obtained from the average of fifty runs).

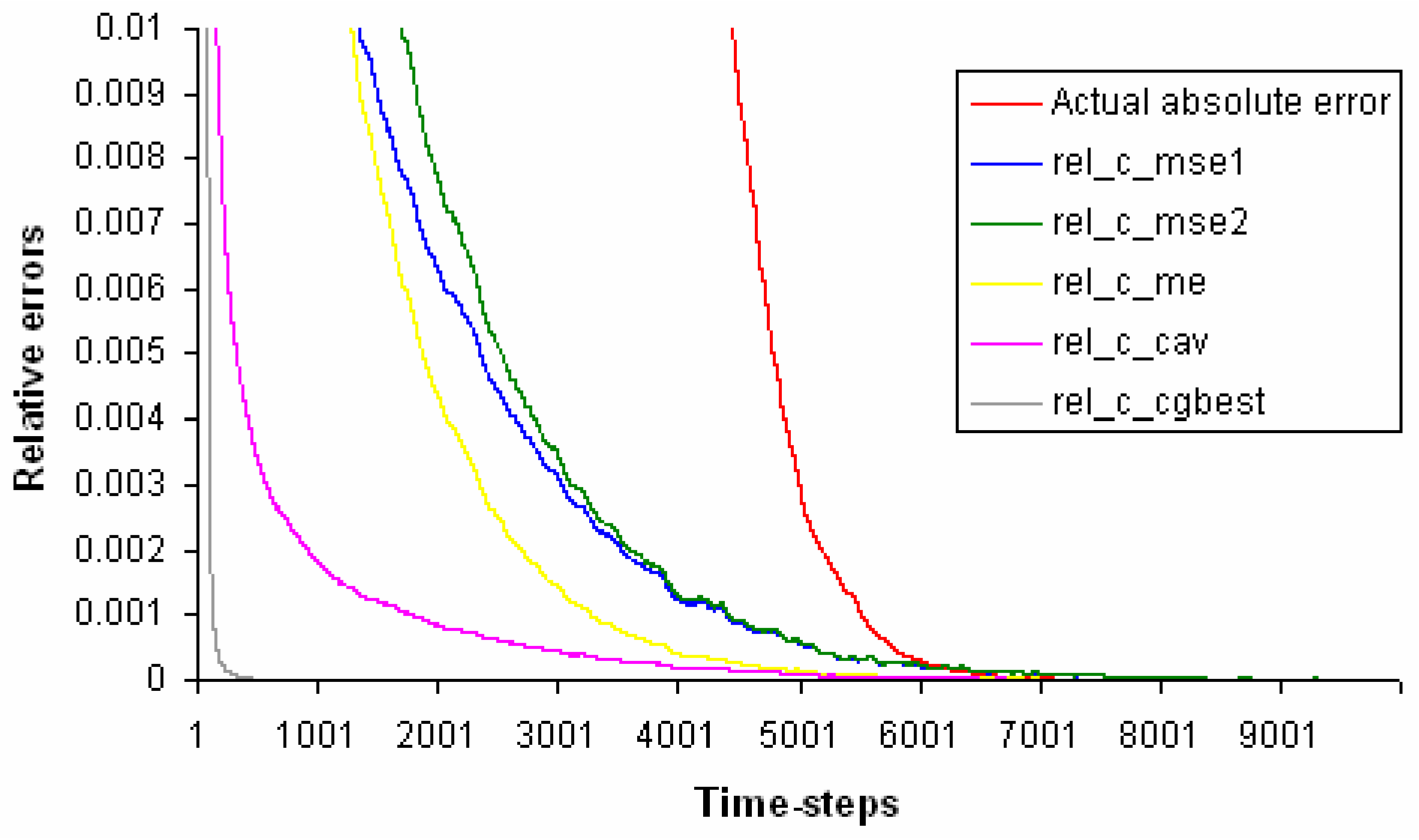

**Fig. 7. 20**: Evolution of the mean relative errors for the BSt-PSO optimizing the 30-dimensional Sphere function.

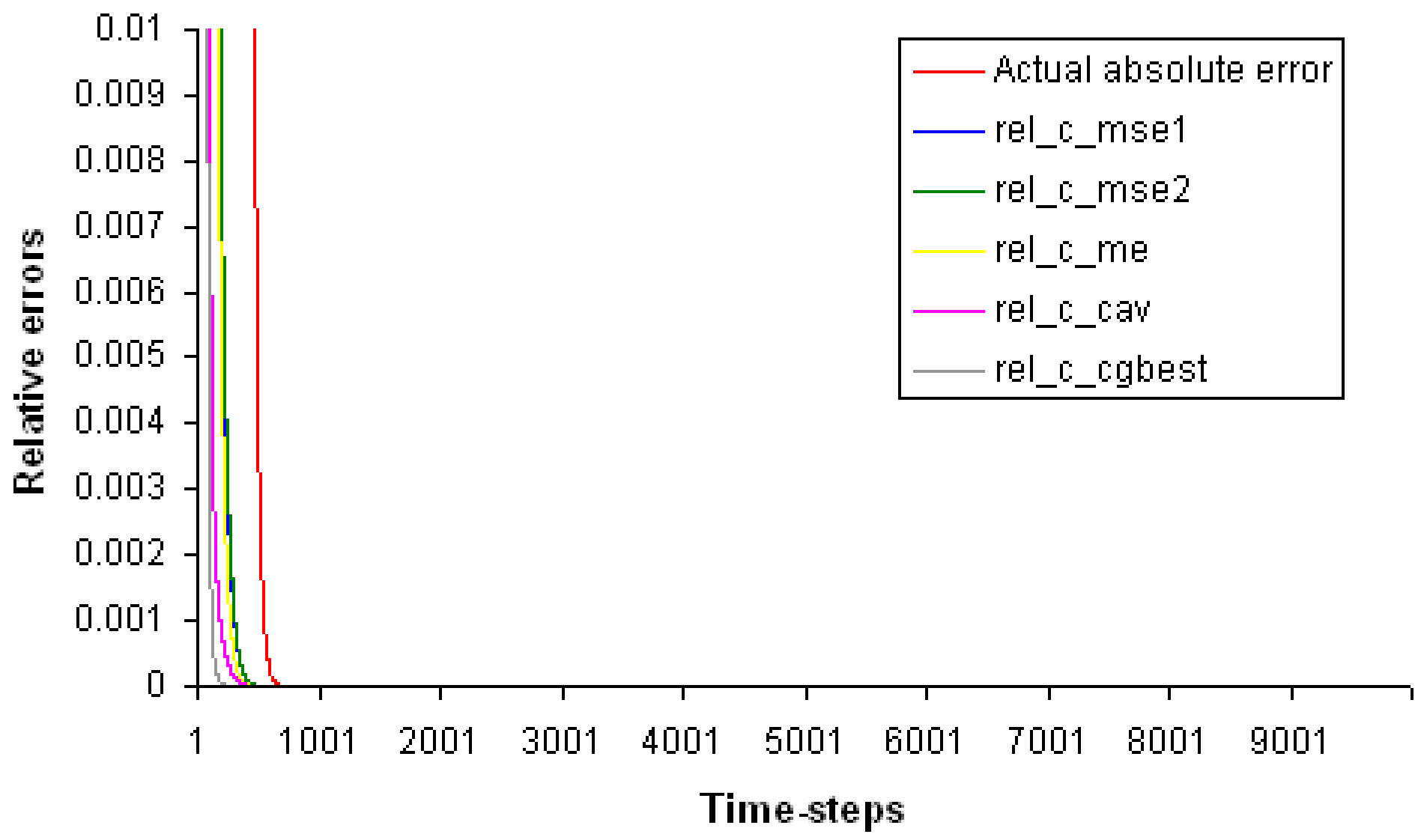

**Fig. 7. 21**: Evolution of the mean relative errors for the BSt-PSO$^{(p)}$ optimizing the 30-dimensional Sphere function.

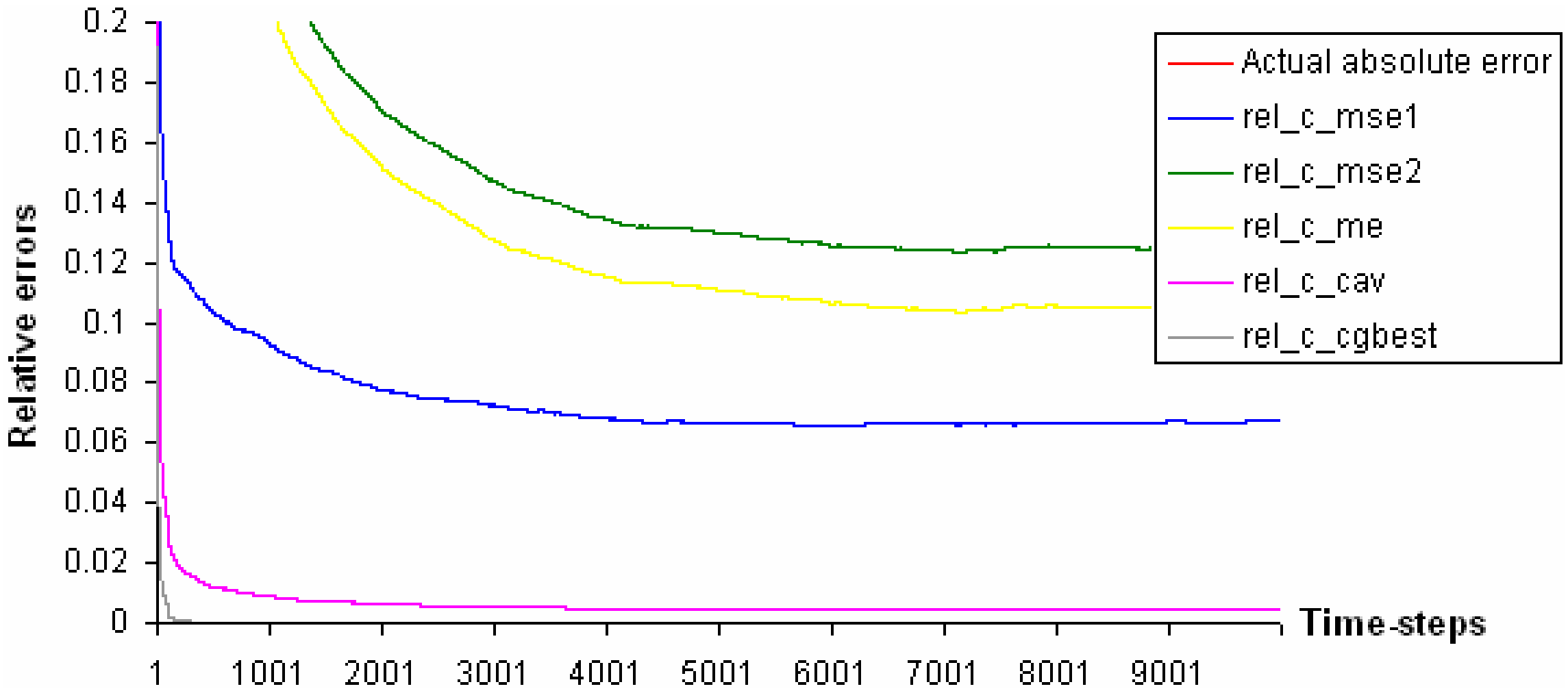

**Fig. 7. 22**: Evolution of the mean relative errors for the BSt-PSO optimizing the 30-dimensional Rastrigrin function.

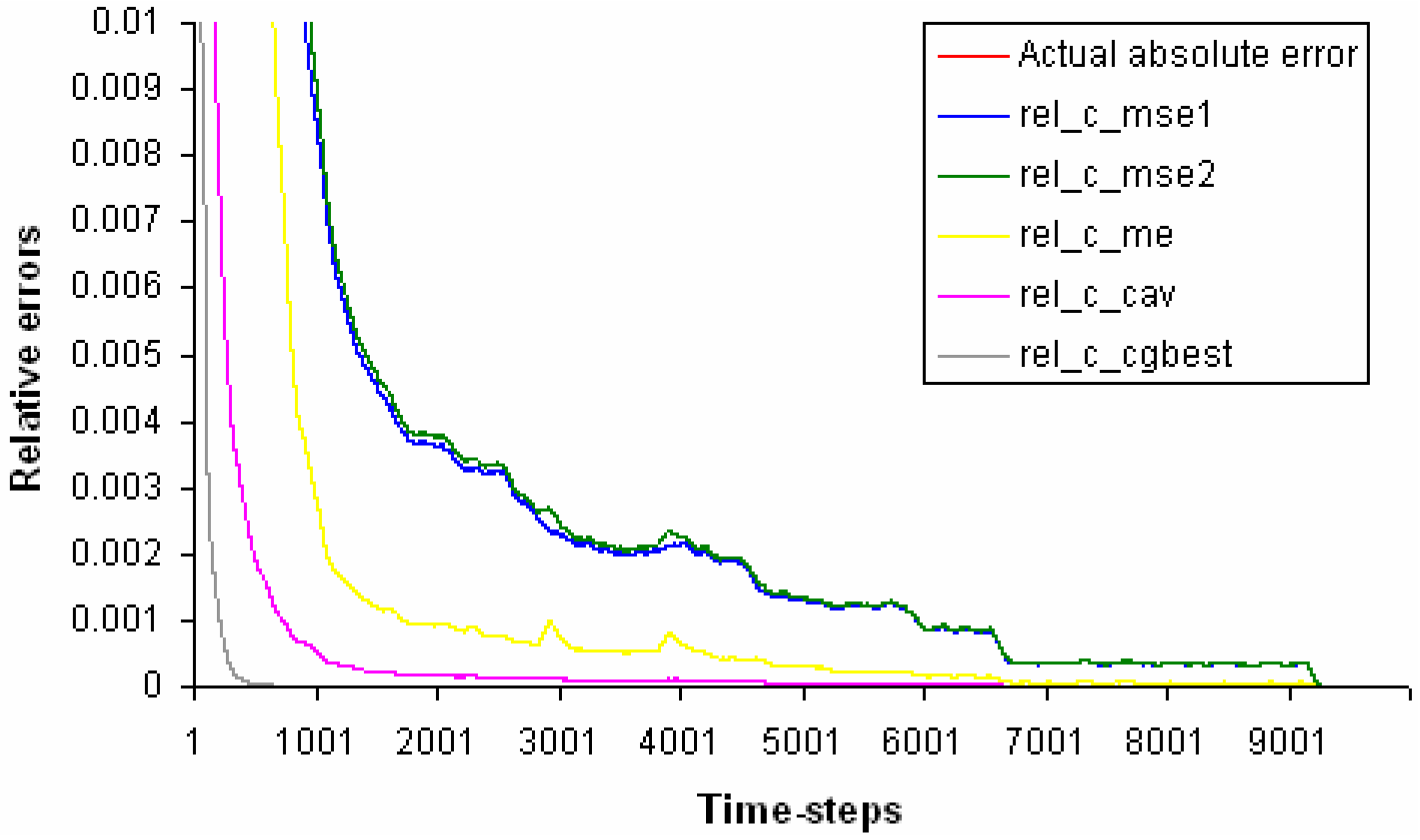

**Fig. 7. 23**: Evolution of the mean relative errors for the BSt-PSO$^{(p)}$ optimizing the 30-dimensional Rastrigrin function.

The study of the quantitative results obtained from a single run of a probabilistic algorithm is not of great practical use. In contrast, the qualitative study is critical for the development of appropriate stopping criteria, given that the latter is to be developed so as to terminate a single run when the solution found so far is good enough, or when further improvement is believed to be unlikely. Following this line of thought, it can be observed that in those cases where the best solution found continues to improve steadily, the **rel_c_cgbest** never equals zero.

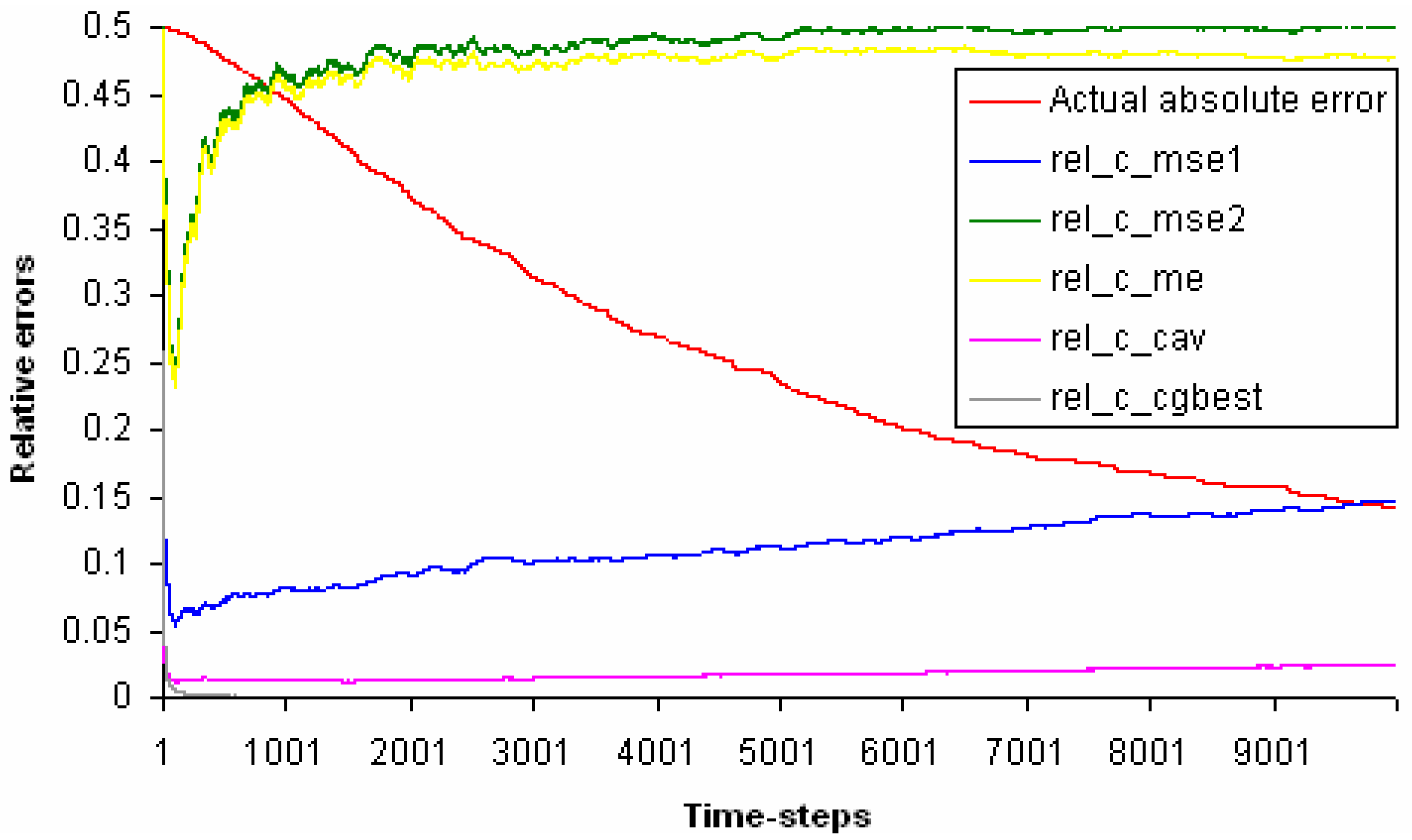

**Fig. 7. 24**: Evolution of the mean relative errors for the BSt-PSO optimizing the 30-dimensional Schaffer f6 function.

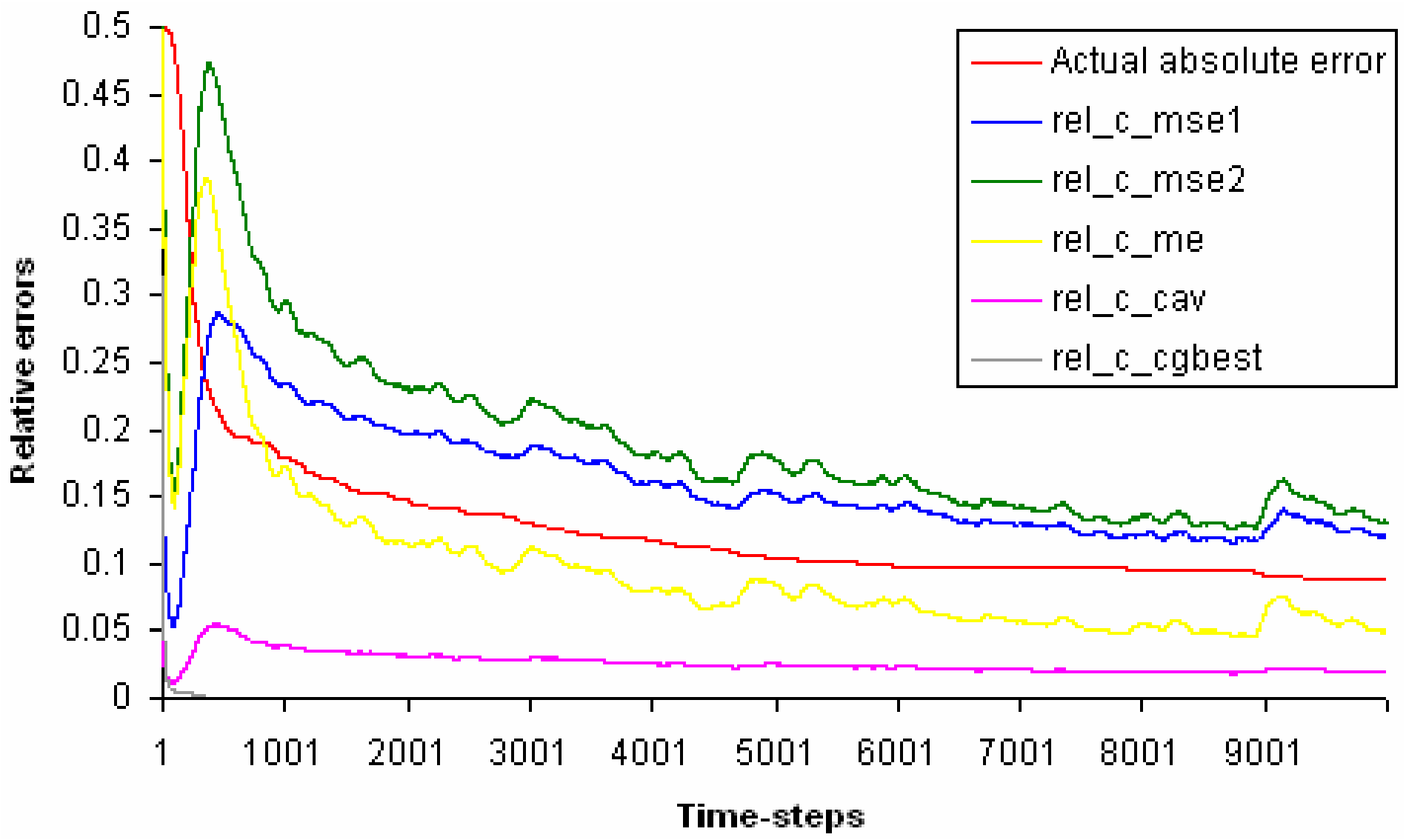

**Fig. 7. 25**: Evolution of the mean relative errors for the BSt-PSO$^{(c)}$ optimizing the 30-dimensional Schaffer f6 function.

Therefore, it seems reasonable to set **rel_c_cgbest** equal to or near zero, so that the iterative process is not terminated while the best solution found is still improving. Assuming that the maximum number of time-steps permitted is set so that it is not unreasonable for the search to take that long, it seems unwise to terminate the search sooner unless further improvement is unlikely or negligible. For instance, being the **BSt-PSO^(c)^** the optimizer which finds the best solution for the Rosenbrock function, it can be observed that, while the best solution found is continuously improved (refer to **Fig. 7. 30**), the **rel_c_cgbest** never equals zero, as can be seen from **Fig. 7. 31**.

| SPHERE | BSt-PSO$^{(c)}$ | | BSt-PSO$^{(p)}$ | |
|---|---|---|---|---|
| | When error condition is attained | When global optimum is found | When error condition is attained | When global optimum is found |
| Actual absolute error (σ) | 1.00E-02 | 0.00E+00 | 1.00E-02 | 0.00E+00 |
| | (0.00E+00) | (0.00E+00) | (0.00E+00) | (0.00E+00) |
| rel_c_mse1 (σ) | 3.63E-06 | - | 2.86E-05 | - |
| | (4.38E-06) | - | (4.63E-05) | - |
| rel_c_mse2 (σ) | 4.08E-06 | - | 2.98E-05 | - |
| | (4.57E-06) | - | (4.73E-05) | - |
| rel_c_me (σ) | 1.67E-06 | - | 7.73E-06 | - |
| | (1.38E-06) | - | (9.30E-06) | - |
| rel_c_cav (σ) | 6.09E-07 | - | 4.60E-06 | - |
| | (7.79E-07) | - | (7.25E-06) | - |
| rel_c_cgbest (σ) | 4.34E-08 | - | 1.51E-08 | - |
| | (3.82E-08) | - | (1.00E-08) | - |

**Table 7. 2**: Values of the mean measures of relative error—and their standard deviations—for two optimizers that exhibit the ability to fine-cluster, corresponding to the time-steps at which the error condition is attained and at which the global optimum is found—if so—, when optimizing the 30-dimensional Sphere function.

| ROSENBROCK | BSt-PSO$^{(c)}$ | | BSt-PSO$^{(p)}$ | |
|---|---|---|---|---|
| | When error condition is attained | When global optimum is found | When error condition is attained | When global optimum is found |
| Actual absolute error (σ) | 1.00E+02 | 0.00E+00 | 1.00E+02 | 0.00E+00 |
| | (0.00E+00) | (0.00E+00) | (0.00E+00) | (0.00E+00) |
| rel_c_mse1 (σ) | 1.28E-05 | - | 2.39E-04 | - |
| | (2.67E-05) | - | (2.28E-04) | - |
| rel_c_mse2 (σ) | 1.32E-05 | - | 2.45E-04 | - |
| | (2.73E-05) | - | (2.34E-04) | - |
| rel_c_me (σ) | 3.17E-06 | - | 5.26E-05 | - |
| | (5.89E-06) | - | (5.22E-05) | - |
| rel_c_cav (σ) | 3.48E-06 | - | 5.79E-05 | - |
| | (7.86E-06) | - | (5.34E-05) | - |
| rel_c_cgbest (σ) | 6.19E-09 | - | 2.14E-09 | - |
| | (1.18E-08) | - | (3.03E-09) | - |

**Table 7. 3**: Values of the mean measures of relative error—and their standard deviations—for two optimizers that exhibit the ability to fine-cluster, corresponding to the time-steps at which the error condition is attained and at which the global optimum is found—if so—, when optimizing the 30-dimensional Rosenbrock function.

In contrast, when the best solution found (*cgbest*) by an optimizer completely stagnates, as in the case of the **BSt-PSO$^{(c)}$** optimizing the Griewank function (see **Fig. 7. 32**) or the Rastrigrin function (refer to **Appendix 4**), the **rel_c_cgbest** equals zero soon after the stagnation (refer to **Fig. 7. 33** and to **Appendix 4**).

| RASTRIGRIN | BSt-PSO$^{(c)}$ | | BSt-PSO$^{(p)}$ | |
|---|---|---|---|---|
| | When error condition is attained | When global optimum is found | When error condition is attained | When global optimum is found |
| Actual absolute error (σ) | 1.00E+02 | 0.00E+00 | 1.00E+02 | 0.00E+00 |
| | (0.00E+00) | (0.00E+00) | (0.00E+00) | (0.00E+00) |
| rel_c_mse1 (σ) | 8.39E-02 | - | 9.53E-02 | - |
| | (1.48E-02) | - | (1.45E-02) | - |
| rel_c_mse2 (σ) | 1.79E-01 | - | 1.97E-01 | - |
| | (4.24E-02) | - | (2.91E-02) | - |
| rel_c_me (σ) | 1.56E-01 | - | 1.71E-01 | - |
| | (4.12E-02) | - | (2.74E-02) | - |
| rel_c_cav (σ) | 1.28E-02 | - | 1.09E-02 | - |
| | (4.94E-03) | - | (2.13E-03) | - |
| rel_c_cgbest (σ) | 3.32E-03 | - | 2.05E-03 | - |
| | (2.22E-03) | - | (8.48E-04) | - |

**Table 7. 4**: Values of the mean measures of relative error—and their standard deviations—for two optimizers that exhibit the ability to fine-cluster, corresponding to the time-steps at which the error condition is attained and at which the global optimum is found—if so—, when optimizing the 30-dimensional Rastrigrin function.

| GREWANK | BSt-PSO$^{(c)}$ | | BSt-PSO$^{(p)}$ | |
|---|---|---|---|---|
| | When error condition is attained | When global optimum is found | When error condition is attained | When global optimum is found |
| Actual absolute error (σ) | 1.00E-01 | 0.00E+00 | 1.00E-01 | 0.00E+00 |
| | (0.00E+00) | (0.00E+00) | (0.00E+00) | (0.00E+00) |
| rel_c_mse1 (σ) | 1.05E-04 | 1.40E-18 | 2.80E-04 | 2.15E-17 |
| | (4.44E-05) | (1.70E-18) | (1.28E-04) | (2.49E-17) |
| rel_c_mse2 (σ) | 1.55E-04 | 1.69E-18 | 4.05E-04 | 2.28E-17 |
| | (5.61E-05) | (2.05E-18) | (1.38E-04) | (2.56E-17) |
| rel_c_me (σ) | 1.08E-04 | 8.97E-19 | 2.72E-04 | 6.93E-18 |
| | (3.33E-05) | (1.11E-18) | (6.11E-05) | (6.04E-18) |
| rel_c_cav (σ) | 1.49E-05 | 2.21E-19 | 3.42E-05 | 3.44E-18 |
| | (6.77E-06) | (2.45E-19) | (1.97E-05) | (3.94E-18) |
| rel_c_cgbest (σ) | 6.21E-06 | 2.91E-20 | 5.06E-06 | 1.94E-20 |
| | (1.05E-06) | (3.57E-20) | (1.14E-06) | (1.55E-20) |

**Table 7. 5**: Values of the mean measures of relative error—and their standard deviations—for two optimizers that exhibit the ability to fine-cluster, corresponding to the time-steps at which the error condition is attained and at which the global optimum is found—if so—, when optimizing the 30-dimensional Griewank function.

| SCHAFFER F6 2D | BSt-PSO^(c) | | BSt-PSO^(p) | |
|---|---|---|---|---|
| | When error condition is attained | When global optimum is found | When error condition is attained | When global optimum is found |
| **Actual absolute error (σ)** | 1.00E-05 | 0.00E+00 | 1.00E-05 | 0.00E+00 |
| | (0.00E+00) | (0.00E+00) | (0.00E+00) | (0.00E+00) |
| **rel_c_mse1 (σ)** | 3.04E-01 | 3.60E-01 | 2.93E-01 | 3.59E-01 |
| | (3.90E-02) | (1.83E-02) | (4.14E-02) | (5.52E-03) |
| **rel_c_mse2 (σ)** | 5.18E-01 | 5.01E-01 | 4.81E-01 | 5.46E-01 |
| | (9.10E-02) | (6.20E-02) | (9.67E-02) | (3.53E-02) |
| **rel_c_me (σ)** | 4.09E-01 | 3.44E-01 | 3.68E-01 | 4.09E-01 |
| | (9.70E-02) | (7.54E-02) | (1.06E-01) | (4.74E-02) |
| **rel_c_cav (σ)** | 6.05E-02 | 5.38E-02 | 5.73E-02 | 6.09E-02 |
| | (9.06E-03) | (7.76E-03) | (9.77E-03) | (6.52E-03) |
| **rel_c_cgbest (σ)** | 5.93E-04 | 3.94E-10 | 1.07E-03 | 5.65E-06 |
| | (1.69E-03) | (1.34E-09) | (2.37E-03) | (1.68E-05) |

**Table 7. 6**: Values of the mean measures of relative error—and their standard deviations—for two optimizers that exhibit the ability to fine-cluster, corresponding to the time-steps at which the error condition is attained and at which the global optimum is found—if so—, when optimizing the 2-dimensional Schaffer f6 function.

| SCHAFFER F6 | BSt-PSO^(c) | | BSt-PSO^(p) | |
|---|---|---|---|---|
| | When error condition is attained | When global optimum is found | When error condition is attained | When global optimum is found |
| **Actual absolute error (σ)** | 1.00E-01 | 0.00E+00 | 1.00E-01 | 0.00E+00 |
| | (0.00E+00) | (0.00E+00) | (0.00E+00) | (0.00E+00) |
| **rel_c_mse1 (σ)** | 1.74E-01 | - | 2.56E-01 | - |
| | (6.51E-02) | - | (2.64E-02) | - |
| **rel_c_mse2 (σ)** | 2.13E-01 | - | 3.71E-01 | - |
| | (1.12E-01) | - | (8.64E-02) | - |
| **rel_c_me (σ)** | 1.14E-01 | - | 2.58E-01 | - |
| | (1.02E-01) | - | (1.09E-01) | - |
| **rel_c_cav (σ)** | 2.95E-02 | - | 4.46E-02 | - |
| | (1.17E-02) | - | (6.80E-03) | - |
| **rel_c_cgbest (σ)** | 5.72E-04 | - | 5.54E-04 | - |
| | (2.57E-04) | - | (2.16E-04) | - |

**Table 7. 7**: Values of the mean measures of relative error—and their standard deviations—for two optimizers that exhibit the ability to fine-cluster, corresponding to the time-steps at which the error condition is attained and at which the global optimum is found—if so—, when optimizing the 30-dimensional Schaffer f6 function.

With regards to the other four proposed relative errors, it is important to remark that the curves of their evolution display very smooth and continuously decreasing shapes—very convenient for their use in the development of stopping criteria—when dealing with very simple functions that do not exhibit numerous local optima, such as the Sphere function.

When dealing with the Rosenbrock, Rastrigrin and Griewank functions, the graphs of their evolution needs to be notably zoomed in to start perceiving their irregularities. Therefore, they are also suitable for the design of stopping criteria.

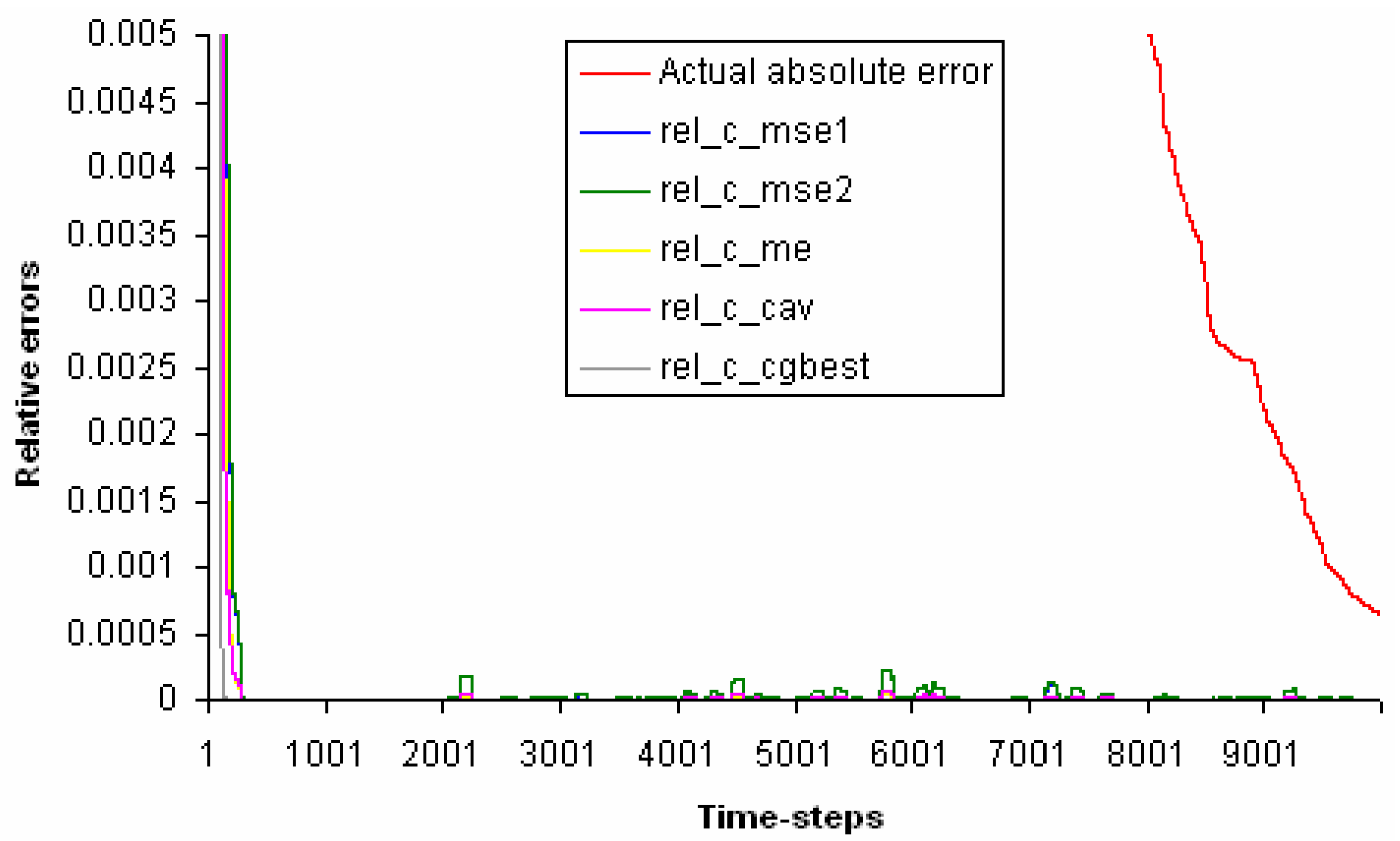

**Fig. 7. 26**: Evolution of the relative errors for the BSt-PSO$^{(c)}$ optimizing the 30-dimensional Rosenbrock function.

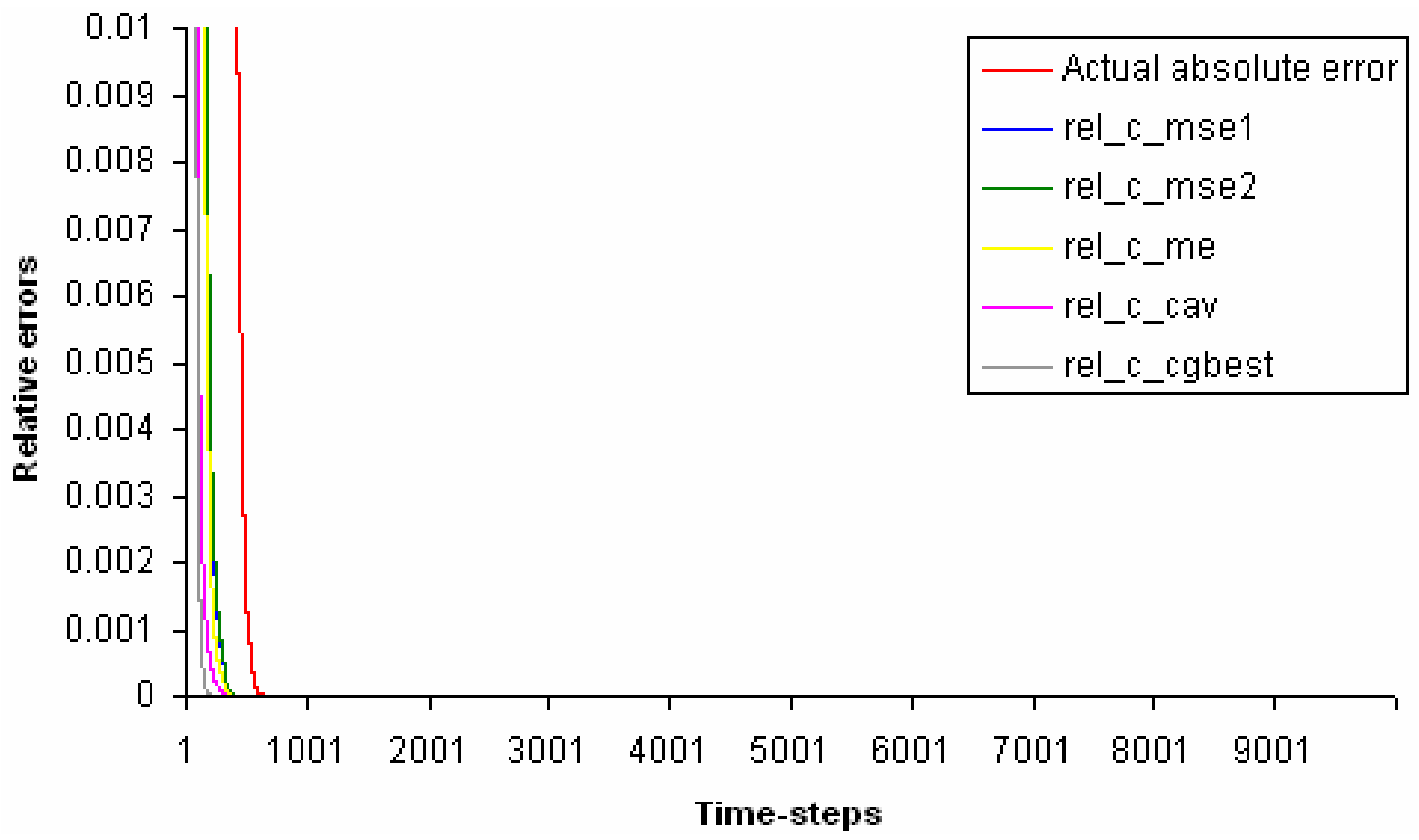

**Fig. 7. 27**: Evolution of the relative errors for the BSt-PSO$^{(p)}$ optimizing the 30-dimensional Sphere function.

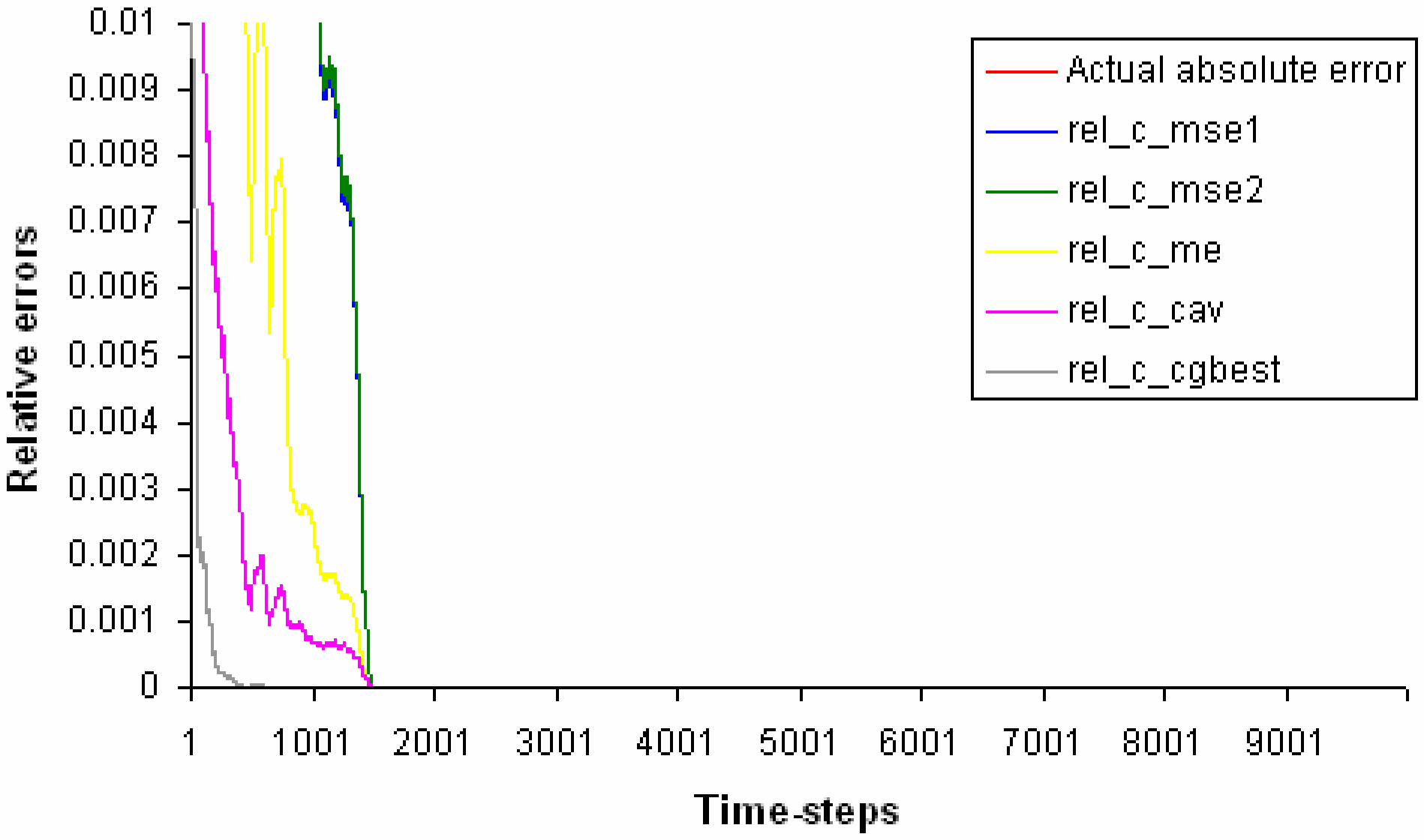

**Fig. 7. 28**: Evolution of the relative errors for the BSt-PSO(p) optimizing the 30-dimensional Rastrigrin function.

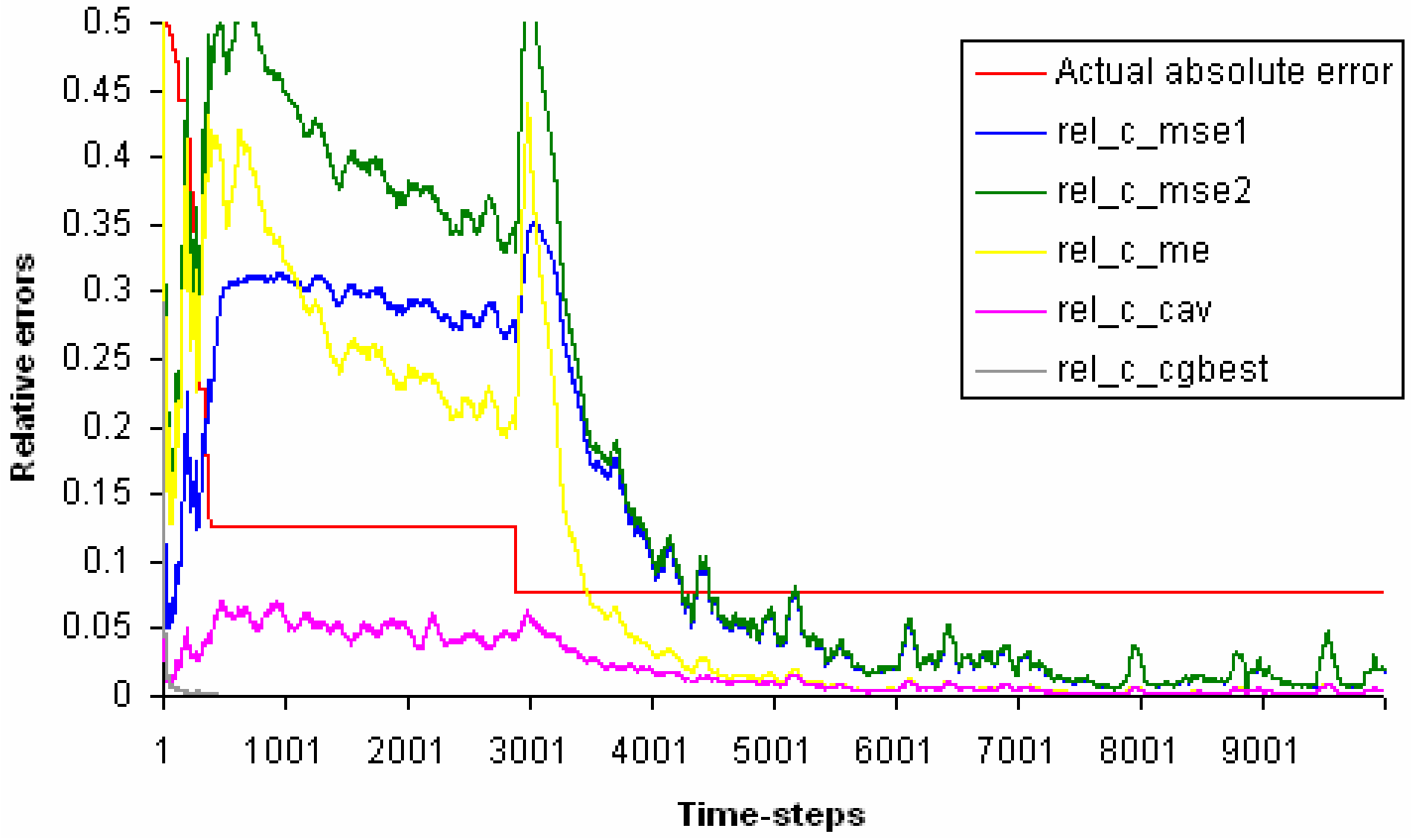

**Fig. 7. 29**: Evolution of the relative errors for the BSt-PSO(c) optimizing the 30-dimensional Schaffer f6 function.

It is fair to remark that the relative errors take very small values very soon for the optimizers with the ability to fine-cluster, when dealing with the Rosenbrock function. However, these errors start increasing and oscillating unevenly again while the particles are fine-clustering. Surprisingly, the best solution found is notably improved during that small explosion. This is an important issue that must be considered when developing the stopping criteria, because, if the degree of clustering required by the termination condition was met, it would be met at the early stages of the search, just before this strange divergence takes place, while it is exactly during this divergence that the best solution found by the optimizer is improved most!

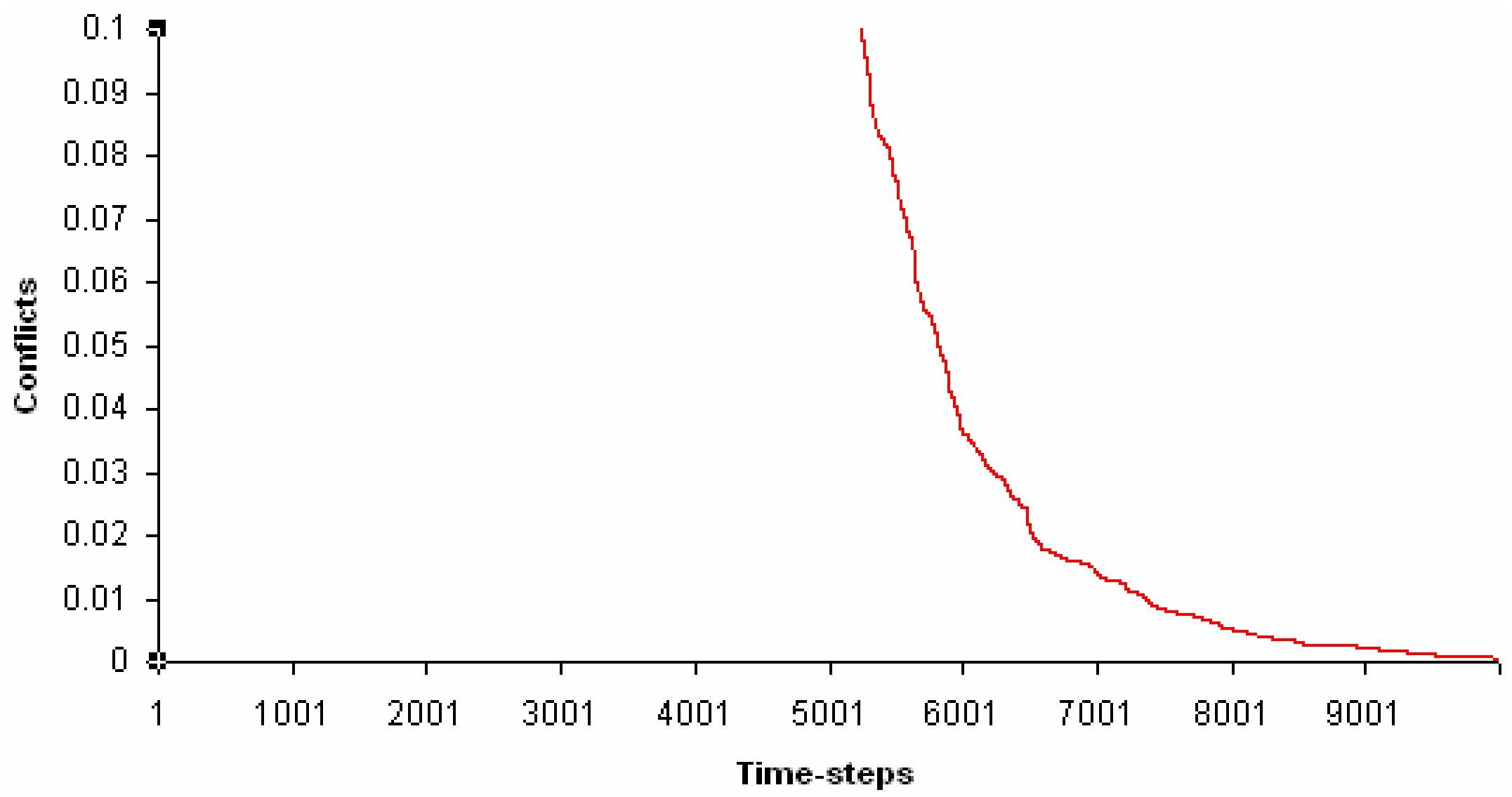

**Fig. 7. 30**: Evolution of the best solution found, “cgbest”, by the BSt-PSO(c) when optimizing the 30-dimensional Rosenbrock function.

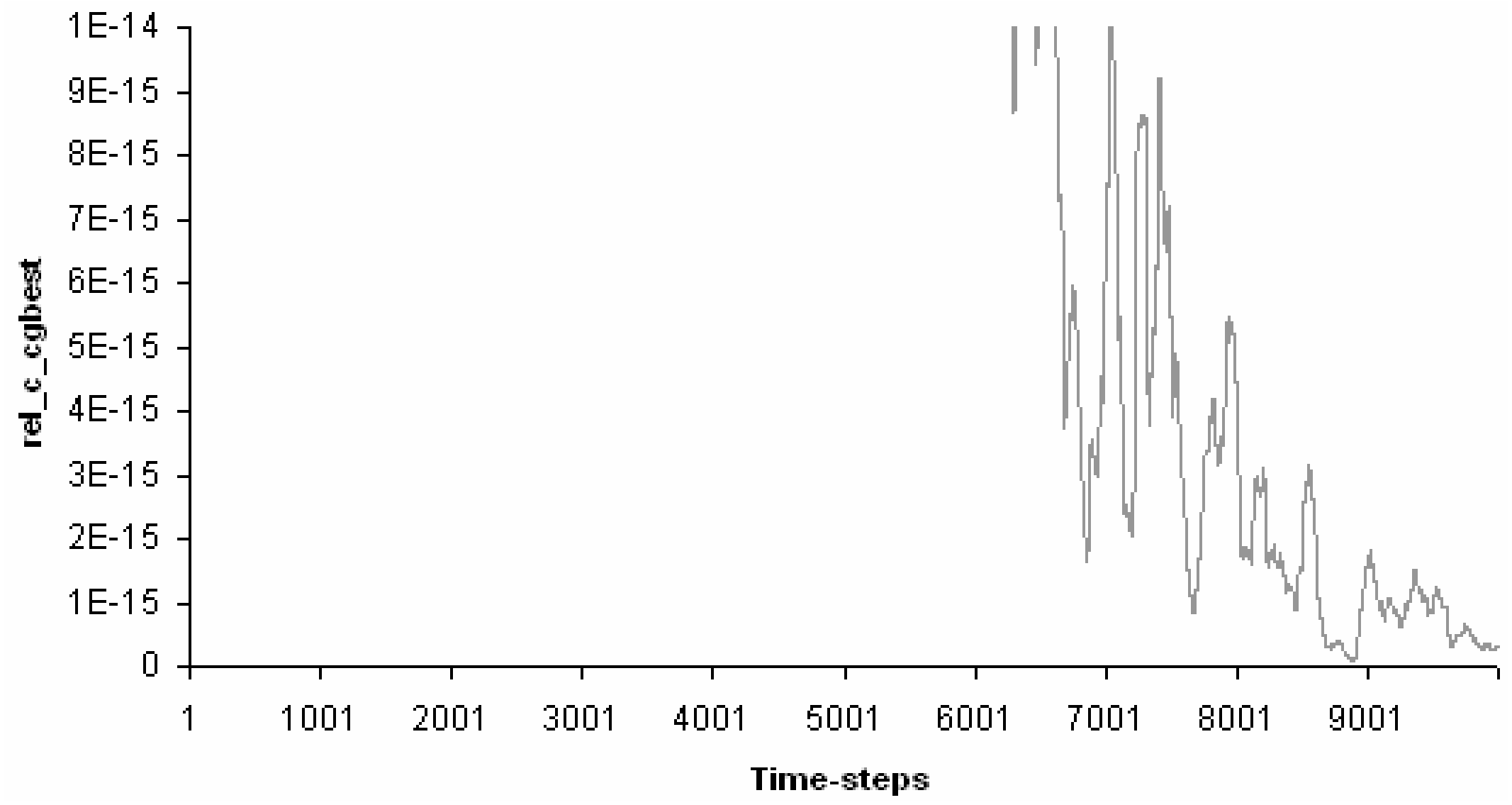

**Fig. 7. 31**: Evolution of the “rel_c_cgbest” for the BSt-PSO(c) optimizing the 30-dimensional Rosenbrock function.

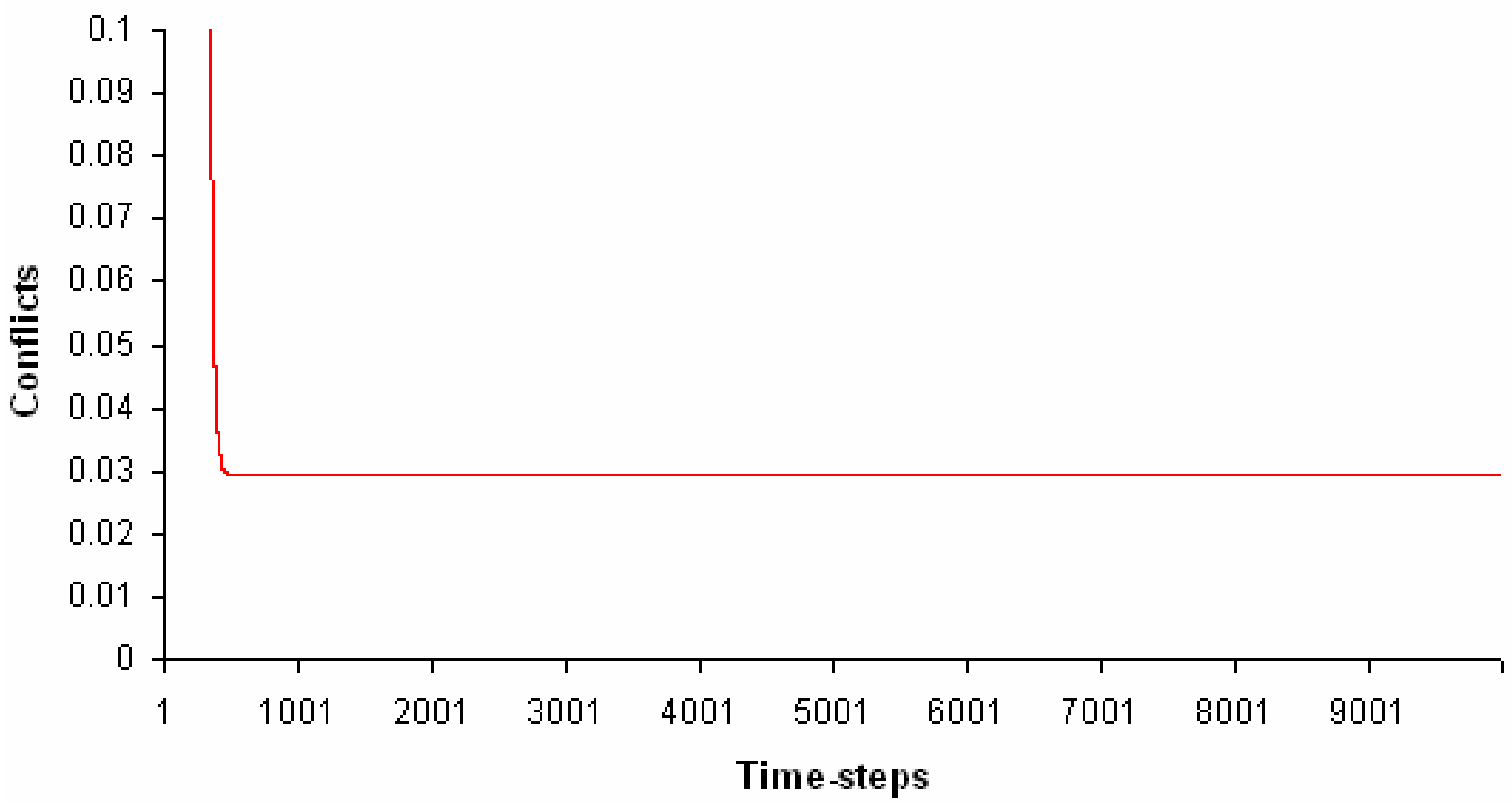

**Fig. 7. 32**: Evolution of the best solution found by the BSt-PSO(c) when optimizing the 30-dimensional Rosenbrock function.

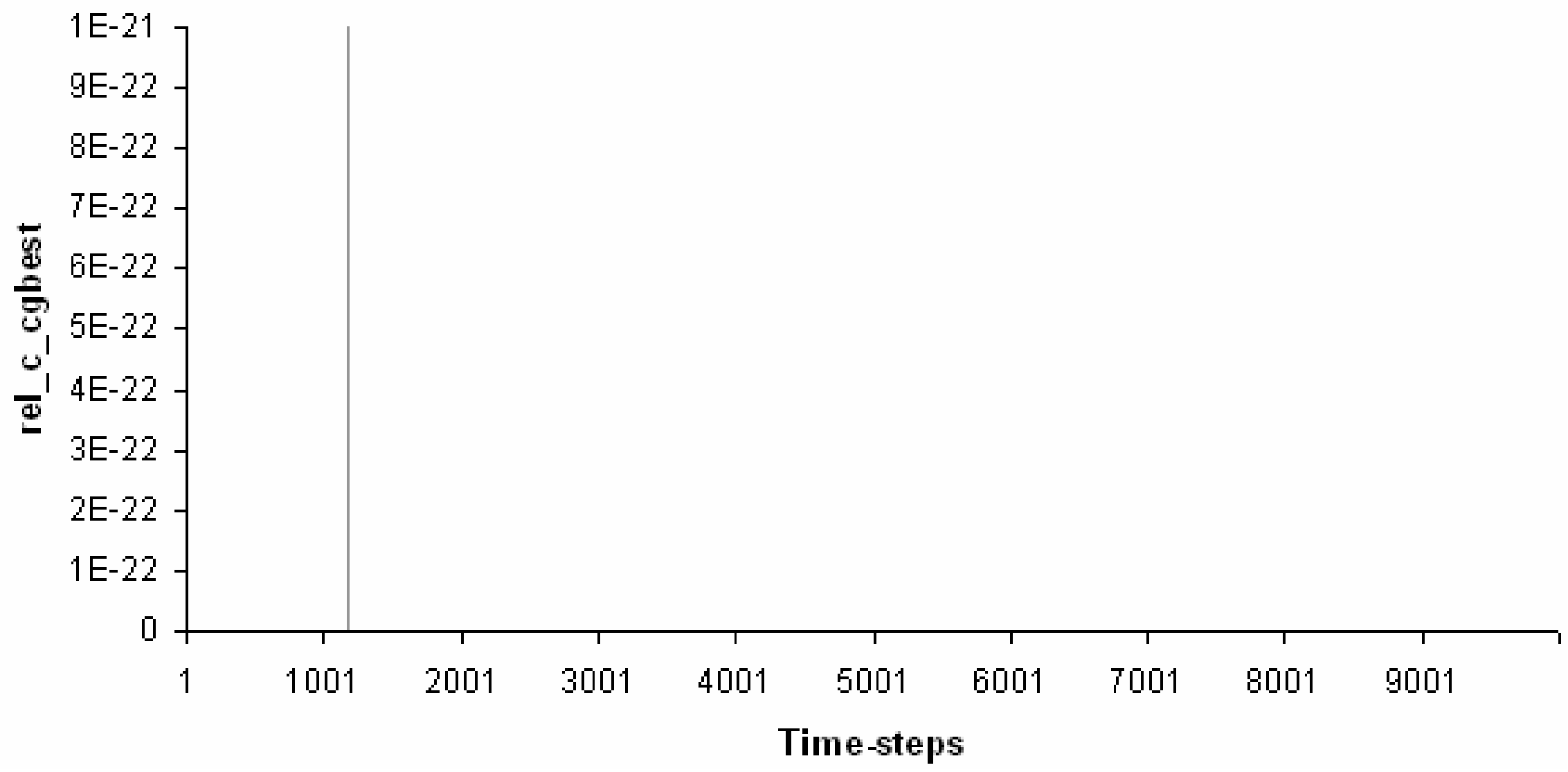

**Fig. 7. 33**: Evolution of the "rel_c_cgbest" for the BSt-PSO(c) optimizing the 30-dimensional Rosenbrock function.

It is also important to observe that, for the optimizers whose particles effectively fine-cluster, the curves of the evolution of the **rel_c_mse1**, the **rel_c_mse2**, the **rel_c_me**, and sometimes even the **rel_c_cav**—which is not defined within a single time-step—look as if each one of them was obtained just by scaling and displacing the very same graph. Examples of this can be seen from **Fig. 7. 29** and **Fig. 7. 35**. Refer to **Appendix 4** for other examples.

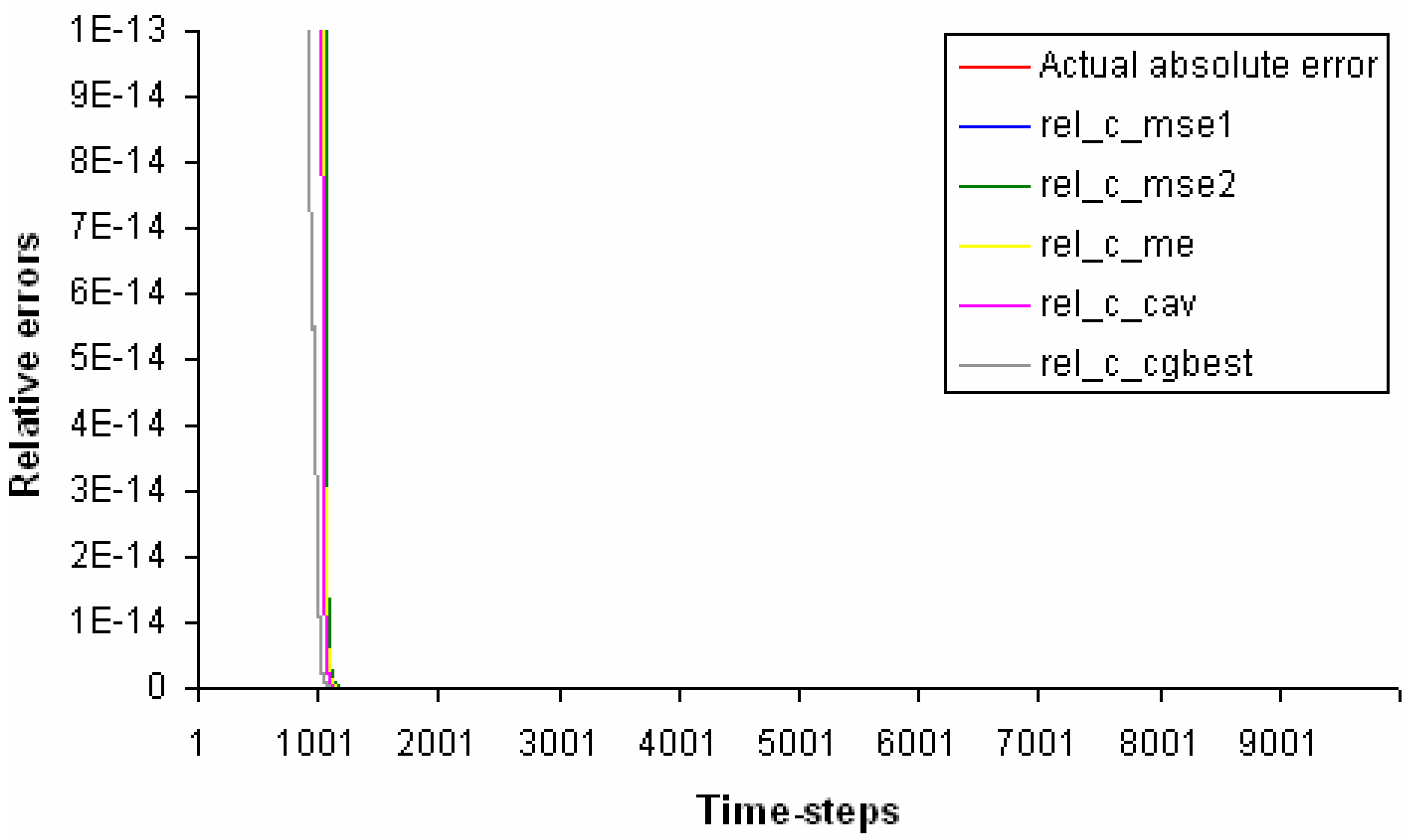

**Fig. 7. 34**: Evolution of the relative errors for the BSt-PSO(c) optimizing the 30-dimensional Rastrigrin function.

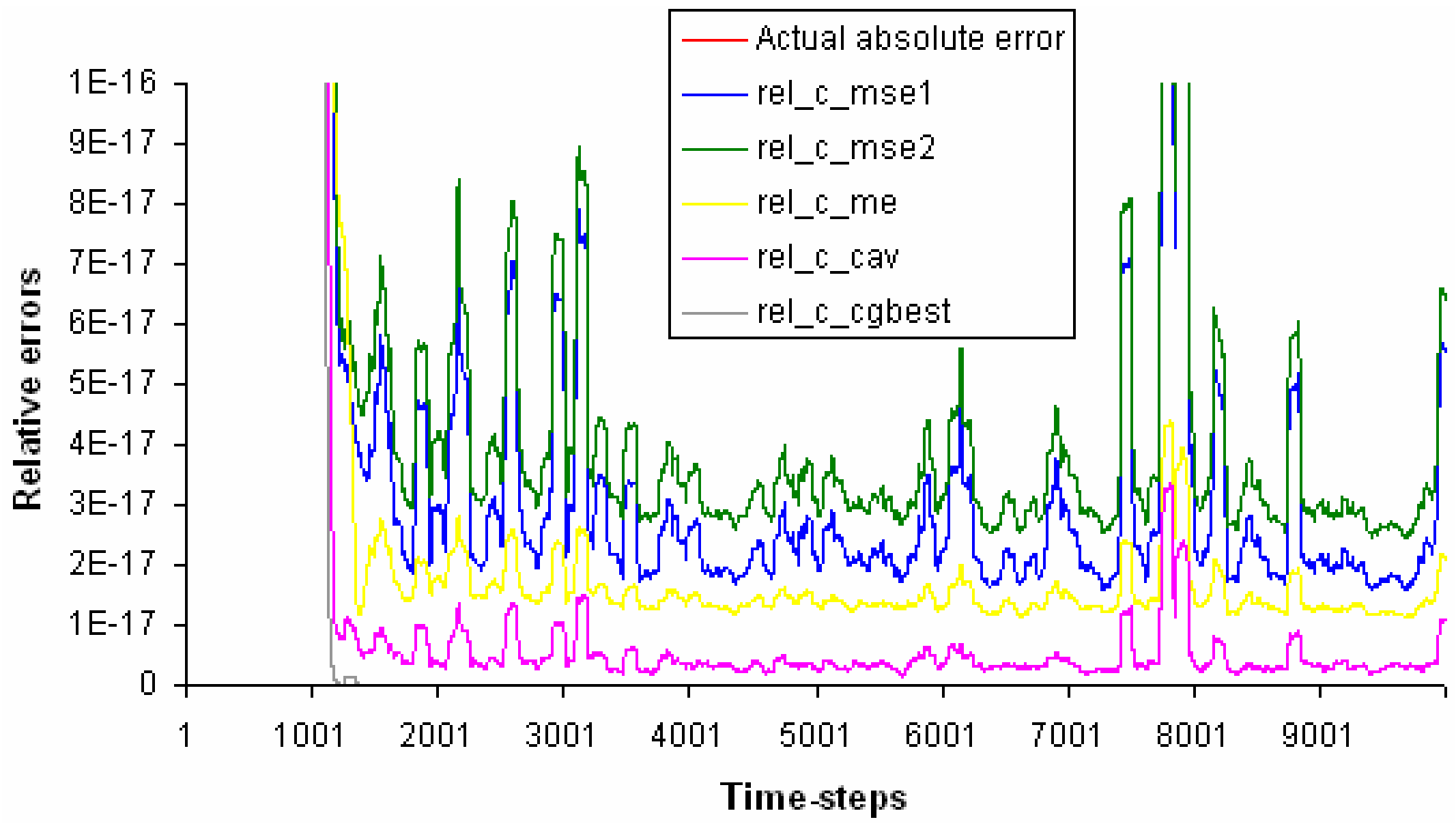

**Fig. 7. 35**: Evolution of the relative errors for the BSt-PSO(c) optimizing the 30-dimensional Rastrigrin function.

In summary, the stopping criteria should be developed for optimizers which exhibit fine-clustering ability, and based on the proposed relative errors rather than on the absolute ones. Furthermore, two of the first three relative errors should be removed because they exhibit similar behaviours. It is reasonable to remove the **rel_c_mse1** and the **rel_c_mse2**, because they are computationally more expensive than the **rel_c_me**. As previously argued, setting **rel_c_cgbest** ≈ 0 might be convenient as a necessary but not sufficient condition for the termination of the iterative process. As to the curves of the evolution of the **rel_c_cav** and the **rel_c_me**, they exhibit similar behaviours, although those corresponding to the former are smoother. Therefore, the measures of error that will be considered for the development of stopping criteria in section **7.6** are the **rel_c_me**, the **rel_c_cav**, and the **rel_c_cgbest**. Finally, it can be observed that there are some functions that exhibit numerous local optima which are located near the global optimum, resulting in the increase of these measures of errors regarding the conflict values as the particles cluster, until they are fine-clustering, provided they are able to do so. An example of this is the Schaffer f6 function, which has, in addition, the attribute that all the conflict values corresponding to coordinates located far from the global optimum are very similar to each other, and near a value of 0.5. This results in small values of the proposed errors regarding the conflict values at the beginning of the search, when the best solution found so far is still far from satisfactory. Therefore, both a minimum and a maximum permissible numbers of time-steps should be set.

The next section is entirely devoted to the design of measures of error regarding the particles' positions. The design of stopping criteria should involve both kinds of measures of error.

## 7.5 Measures of error regarding the particles' positions

Traditional measures of error used to develop stopping criteria typically complement the errors computed with regards to the objective function with those computed with regards to the coordinates' values. The reason for this is that the function to be optimized may exhibit small differences in its evaluation at coordinates which are far from one another, or, in contrast, it may exhibit noticeably great differences in its evaluation at coordinates which are very close to each other. Examples of the first situation are functions that present extensive flat areas, while a clear example of the second situation is the Schaffer f6 function. Thus, it is reasonable to design stopping criteria involving measures of error with regards to the conflict values on the one hand, and measures of error with regards to the particles' positions on the other. The design of the first kind was undertaken along section **7.4**, while the design of the second kind is to be carried out along this section.

### 7.5.1 Errors definitions

In the same fashion as when designing the measures of error regarding the conflict values, the ones regarding the particles' positions can be designed both within the current time-step, and between consecutive ones. Furthermore, because the importance of a given absolute error is problem-dependent, relative errors are usually preferred. In addition to that, since the degree of clustering attained by the particles[7] and the evolution of both the swarm's centre of gravity and the coordinates of the best solution found[8] are independent from the region of the search-space where they take place, relating the errors to the locations where they take place—as traditional techniques do[9]—does not make much sense. Therefore, following the same line of thought as when relating the errors regarding the conflict values to the maximum absolute error found to be possible (i.e. $cgworst - cgbest$), it is proposed here to relate the errors regarding the particles' positions to the maximum range of values that the design variables are permitted to take (i.e. $x_{\max} - x_{\min}$). Thus, the errors are again normalized to the range $[0,1]$.

It was claimed before that problem-independent permissible values of the measures of error need to be set in order to design general-purpose optimizers, which implies that they should depend neither on the features of the conflict function, nor on the number of design variables, nor on the size of the search-space, nor on the number of particles in the swarm. While the number of design variables and the size of the search-space were not an issue when designing errors regarding the conflict values, the features of the conflict function are not an issue now. The problem of making their permissible values independent from the size of the search-space is dealt with by relating the errors to the maximum range of values that the design variables are permitted to take, while the problems of them independent from the number of design variables and from the size of the swarm are dealt with in section **7.5.1.1**.

At first, the measures of error regarding the particles' positions were designed involving either the current time-step or both the current time-step and the preceding one. These measures were implemented and tested on the set of benchmark functions shown in **Table 6.1**.

[7] The degree of clustering attained by the particles is measured by the errors regarding the particles' positions designed within the current time-step.

[8] The evolution of the swarm's centre of gravity and of the location of the best solution found is measured by the errors regarding the particles' positions designed between consecutive time-steps.

[9] Refer to equation **(7. 4)**.

However, their evolutions exhibited wide, uneven oscillations similar to those of the initially proposed measures of error regarding the conflict values, although noticeably smoother (refer to **Fig. 7. 11**). Hence the strategy of involving the last 100 time-steps in the computation of the errors is also adopted hereafter.

In summary, the measures of error regarding the particles' positions are designed in the form of relative errors, where the value of reference is given by $x_{max} - x_{min}$; some measures are designed within the current time-step while some others are designed between consecutive ones, although all of them involve data obtained from the last 100 time-steps[10].

### 7.5.1.1 Relative errors within the current time-step

It has been argued that the measures of error should be independent from the number of dimensions of the search-space. When working with the particles' coordinates rather than with their conflicts, the equivalent to an absolute error is the distance from a given particle to the best solution found so far (assuming the latter is the actual solution). However, the same absolute error of a design variable leads the distance from the corresponding particle to the location of the best solution found to take different scales of values for different number of dimensions of the search-space. Therefore, the distance is normalized as follows:

$$d_i^{(t)} = \sqrt{\sum_{j=1}^{n} \left( x_{ij}^{(t)} - gbest_j^{(t)} \right)^2} = \sqrt{n \cdot \left( d'^{(t)}_i \right)^2} \qquad \textbf{(7. 17)}$$

$$d'^{(t)}_i = \frac{d_i^{(t)}}{\sqrt{n}} = \sqrt{\frac{\sum_{j=1}^{n} \left( x_{ij}^{(t)} - gbest_j^{(t)} \right)^2}{n}} \qquad \textbf{(7. 18)}$$

Where:

- $d_i^{(t)}$ : distance from particle *i* to the location of the best solution found, at time-step *t*
- $gbest_j^{(t)}$ : $j^{th}$ coordinate of the best solution found up to time-step *t*
- $x_{ij}^{(t)}$ : $j^{th}$ coordinate of particle *i* at time-step *t*
- $d'^{(t)}_i$ : normalized distance from particle *i* at time-step *t* to the location of the best solution found up to time-step *t*, defined as a component of a vector with an Euclidean norm equal to $d_i^{(t)}$ and all the components having the same value

[10] Beware that for the errors designed between consecutive time-steps, there are in reality 101 time-steps involved.

- $m$, $n$ : size of the swarm and number of dimensions of the search-space, respectively

Imagine an $m$-dimensional vector, each of whose components is computed as the maximum of the componentwise differences between a particle's coordinates and the coordinates of the location of the best solution found up to the current time-step. If the maximum component of such a vector is smaller than $d'^{(t)}_i$, then all the particles are contained within a hyper-sphere whose radius equals $d_i^{(t)} = \sqrt{n} \cdot d'^{(t)}_i$, and whose centre is $gbest_j^{(t)}$.

- **rel_p_mse**: average of the square root of the mean squared error—where each error is viewed as the normalized distance from each particle's current position to the location of the best solution found so far—corresponding to the last 100 time-steps, related to the maximum range of feasible values for the design variables:

$$\text{rel_p_mse}^{(t)} = \frac{\sum_{i=t-99}^{t} \sqrt{\frac{\sum_{j=1}^{m} \left(d'^{(i)}_j\right)^2}{m}}}{100 \cdot \left(x_{\max} - x_{\min}\right)} = \frac{\sum_{i=t-99}^{t} \sqrt{\frac{\sum_{j=1}^{m} \frac{\sum_{k=1}^{n} \left(x_{jk}^{(i)} - gbest_k^{(i)}\right)^2}{n}}{m}}}{100 \cdot \left(x_{\max} - x_{\min}\right)} \quad \textbf{(7. 19)}$$

$$\text{rel_p_mse}^{(t)} = \frac{\sum_{i=t-99}^{t} \sqrt{\sum_{j=1}^{m} \sum_{k=1}^{n} \left(x_{jk}^{(i)} - gbest_k^{(i)}\right)^2}}{100 \cdot \left(x_{\max} - x_{\min}\right) \cdot \sqrt{m \cdot n}} \quad \textbf{(7. 20)}$$

- **rel_p_mnd**: average of the mean normalized distance from the particles' current positions to the location of the best solution found so far corresponding to the last 100 time-steps, related to the maximum range of feasible values for the design variables:

$$\text{rel_p_mnd}^{(t)} = \frac{\sum_{i=t-99}^{t} \sum_{j=1}^{m} d'^{(i)}_j}{100 \cdot \left(x_{\max} - x_{\min}\right) \cdot m} = \frac{\sum_{i=t-99}^{t} \sum_{j=1}^{m} \sqrt{\frac{\sum_{k=1}^{n} \left(x_{jk}^{(i)} - gbest_k^{(i)}\right)^2}{n}}}{100 \cdot \left(x_{\max} - x_{\min}\right) \cdot m} \quad \textbf{(7. 21)}$$

$$\text{rel_p_mnd}^{(t)} = \frac{\sum_{i=t-99}^{t} \sum_{j=1}^{m} \sqrt{\sum_{k=1}^{n} \left(x_{jk}^{(i)} - gbest_k^{(i)}\right)^2}}{100 \cdot \left(x_{\max} - x_{\min}\right) \cdot m \cdot \sqrt{n}} \quad \textbf{(7. 22)}$$

- **rel_p_maxe**: average of the maximum of the maximum componentwise difference between each particle's current position and the location of the best conflict found up to the current time-step corresponding to the last 100 time-steps, related to the maximum range of feasible values for the design variables:

$$\text{rel_p_maxe}^{(t)} = \frac{\sum_{i=t-99}^{t} \max\left(x_{jk}^{(i)} - gbest_k^{(i)}\right)}{100 \cdot \left(x_{\max} - x_{\min}\right)} \quad , \quad j = 1,\ldots,m \;\wedge\; k = 1,\ldots,n \qquad \textbf{(7. 23)}$$

- **rel_p_me**: average of the mean of the maximum componentwise difference between each particle's current position and the location of the best conflict found up to the current time-step corresponding to the last 100 time-steps, related to the maximum range of feasible values for the design variables:

$$\text{rel_p_me}^{(t)} = \frac{\sum_{i=t-99}^{t} \sum_{j=1}^{m} \max\left(x_{jk}^{(i)} - gbest_k^{(i)}\right)}{100 \cdot \left(x_{\max} - x_{\min}\right) \cdot m} \quad , \quad k = 1,\ldots,n \qquad \textbf{(7. 24)}$$

- **rel_p_cg-gbest**: average of the normalized distance from the centre of gravity (**cg**) of the swarm to the location of the best solution found up to the current time-step (**gbest**) corresponding to the last 100 time-steps, related to the maximum range of feasible values for the design variables:

$$\text{rel_p_cg-gbest}^{(t)} = \frac{\sum_{i=t-99}^{t} \sqrt{\frac{\sum_{j=1}^{n}\left(cg_j^{(i)} - gbest_j^{(i)}\right)^2}{n}}}{100 \cdot \left(x_{\max} - x_{\min}\right)} = \frac{\sum_{i=t-99}^{t} \sqrt{\sum_{j=1}^{n}\left(cg_j^{(i)} - gbest_j^{(i)}\right)^2}}{100 \cdot \left(x_{\max} - x_{\min}\right) \cdot \sqrt{n}} \qquad \textbf{(7. 25)}$$

### 7.5.1.2 Relative errors between consecutive time-steps

- **rel_p_cg**: average of the normalized distance between the current centre of gravity of the swarm and the preceding one corresponding to the last 100 time-steps, related to the maximum range of feasible values for the design variables:

$$\text{rel_p_cg}^{(t)} = \frac{\sum_{i=t-99}^{t} \sqrt{\frac{\sum_{j=1}^{n} \left(cg_j^{(i)} - cg_j^{(i-1)}\right)^2}{n}}}{100 \cdot \left(x_{i\max} - x_{i\min}\right)} = \frac{\sum_{i=t-99}^{t} \sqrt{\sum_{j=1}^{n} \left(cg_j^{(t)} - cg_j^{(t-1)}\right)^2}}{100 \cdot \sqrt{n} \cdot \left(x_{i\max} - x_{i\min}\right)} \qquad \textbf{(7. 26)}$$

- **rel_p_gbest**: average of the normalized distance between the best conflict found up to the current time-step and the preceding one corresponding to the last 100 time-steps, related to the maximum range of feasible values for the design variables:

$$\text{rel_p_gbest}^{(t)} = \frac{\sum_{i=t-99}^{t} \sqrt{\frac{\sum_{j=1}^{n} \left(gbest_j^{(i)} - gbest_j^{(i-1)}\right)^2}{n}}}{100 \cdot \left(x_{i\max} - x_{i\min}\right)} = \frac{\sum_{i=t-99}^{t} \sqrt{\sum_{j=1}^{n} \left(gbest_j^{(t)} - gbest_j^{(t-1)}\right)^2}}{100 \cdot \sqrt{n} \cdot \left(x_{i\max} - x_{i\min}\right)} \qquad \textbf{(7. 27)}$$

## 7.5.2 Experimental results

On the one hand, the proposed relative errors are studied by analyzing the average behaviour among 50 runs in order to take into account the probabilistic nature of the algorithm. On the other hand, their behaviour for a single run is also studied in order to take into account the evolution of these errors without the "smoothing effect" that results from averaging the outcomes of different independent runs. In fact, it can be observed that the curves of the evolution of the different proposed relative errors are noticeably smoother for the fifty runs, although those obtained from the single run are still smooth enough to be considered for the design of stopping criteria.

### 7.5.2.1 Fifty runs

The first important issue that can be noticed is that averaging the results obtained from 50 runs in addition to the strategy of involving 100 time-steps in the computation of the errors results in very smooth curves of the evolution of the proposed measures of error for each of the six benchmark functions.

The curves of the evolution of the mean relative errors regarding the particles' positions designed within the current time-step—and those of the **rel_c_me**—for some optimizers with the ability to fine-cluster dealing with the 30-dimensional Rosenbrock, Sphere, Rastrigrin and Schaffer f6 functions are shown in **Fig. 7. 36** to **Fig. 7. 39**. Refer to **Appendix 4** for the complete set of experimental results.

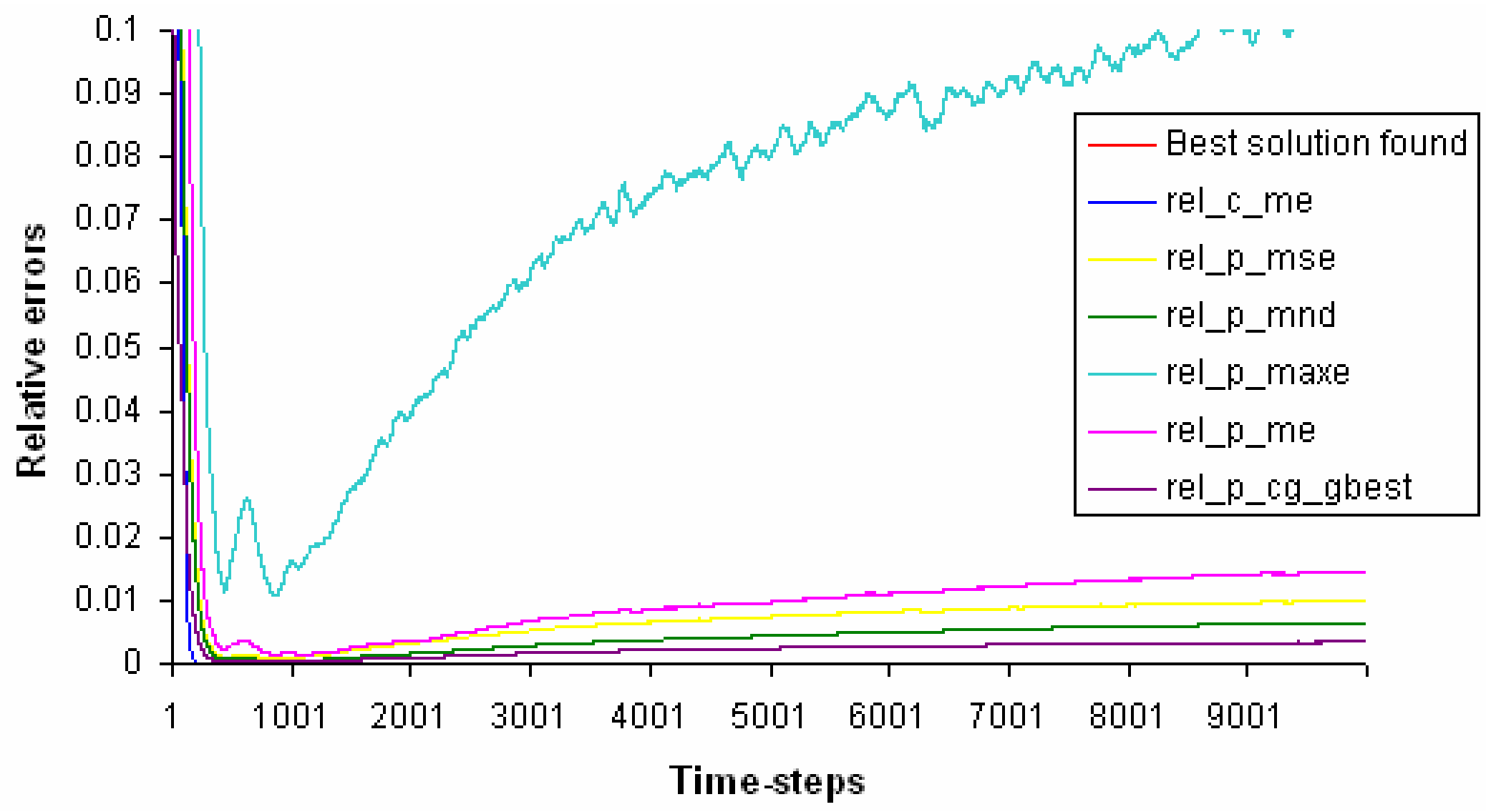

**Fig. 7. 36**: Evolution of the mean relative errors designed within the current time-step for the BSt-PSO$^{(c)}$ optimizing the 30-dimensional Rosenbrock function.

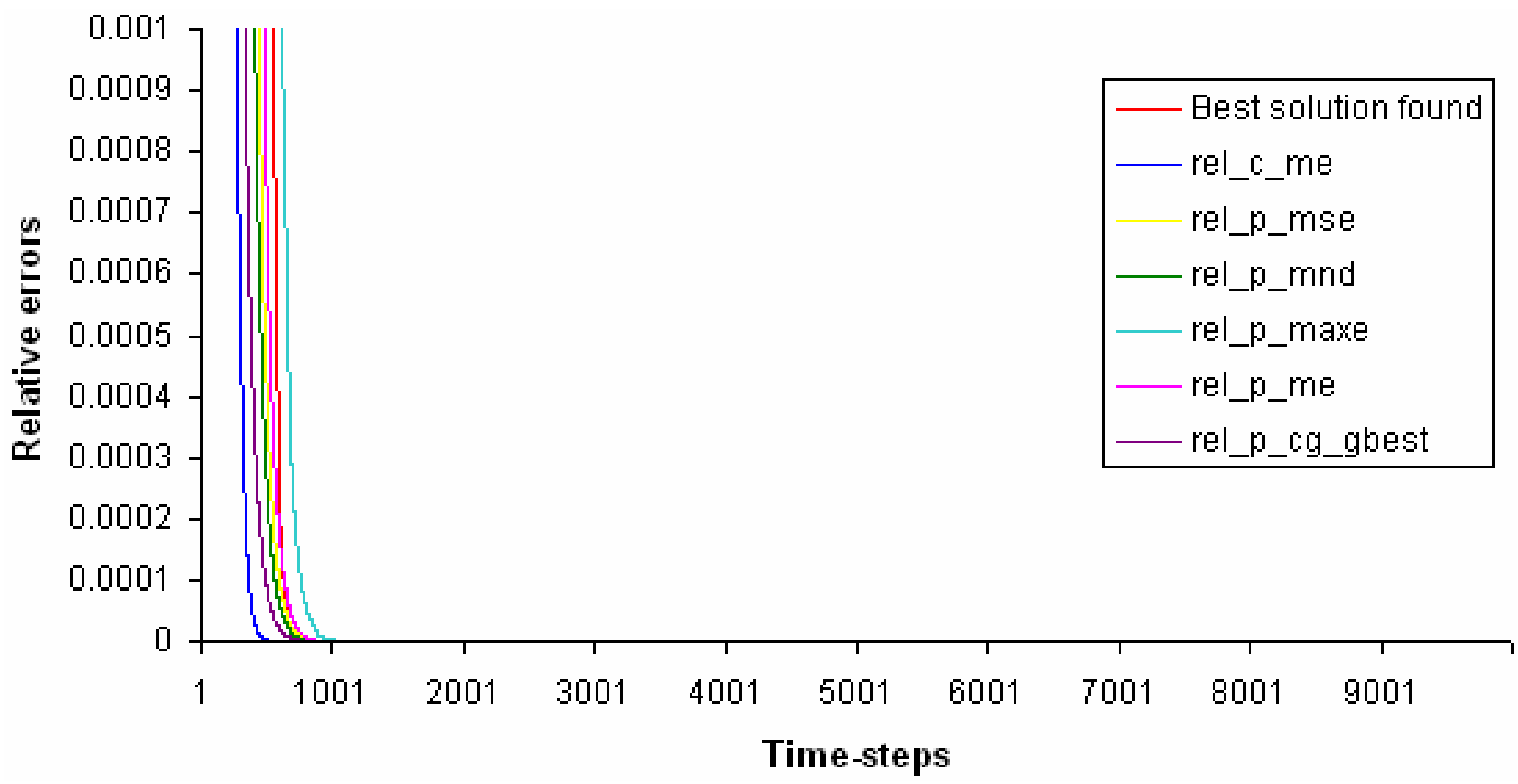

**Fig. 7. 37**: Evolution of the mean relative errors designed within the current time-step for the BSt-PSO$^{(p)}$ optimizing the 30-dimensional Sphere function.

Although the analysis of the behaviour of the optimizer is beyond the scope of this chapter, the curves of the evolution of the errors regarding the particles' positions show that the slight increase in the mean average conflicts (refer to **Appendix 4**) and in the errors regarding the conflict values (refer to section **7.4** and to **Appendix 4**) while the particles of optimizers with the ability to fine-cluster approach the global optimum of the Rosenbrock function is due to a kind of small explosion rather than to the increase in the conflict values in the vicinity of the optimum, as it was conjectured before. The reason for this is not understood.

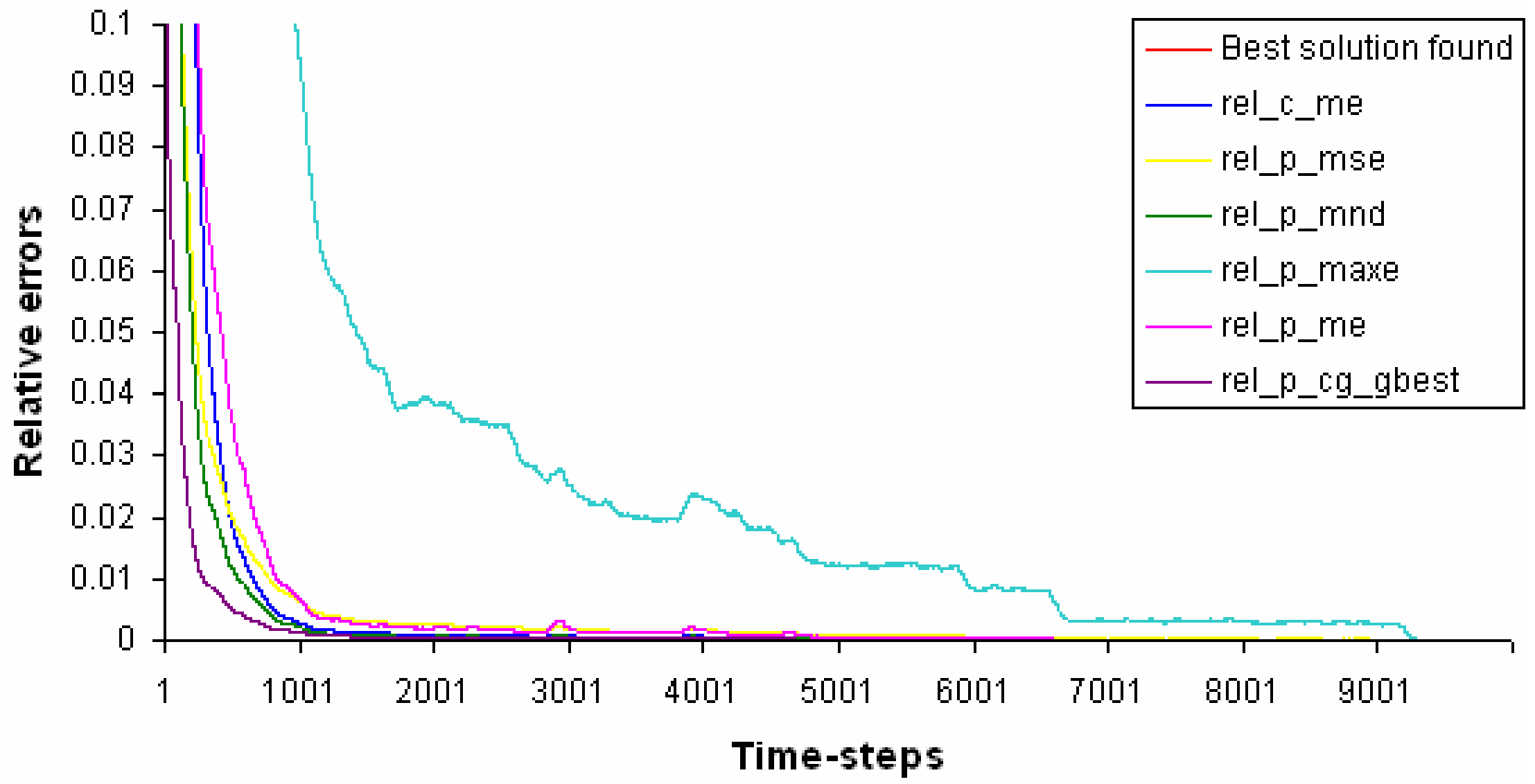

**Fig. 7. 38**: Evolution of the mean relative errors designed within the current time-step for the BSt-PSO$^{(p)}$ optimizing the 30-dimensional Rastrigrin function.

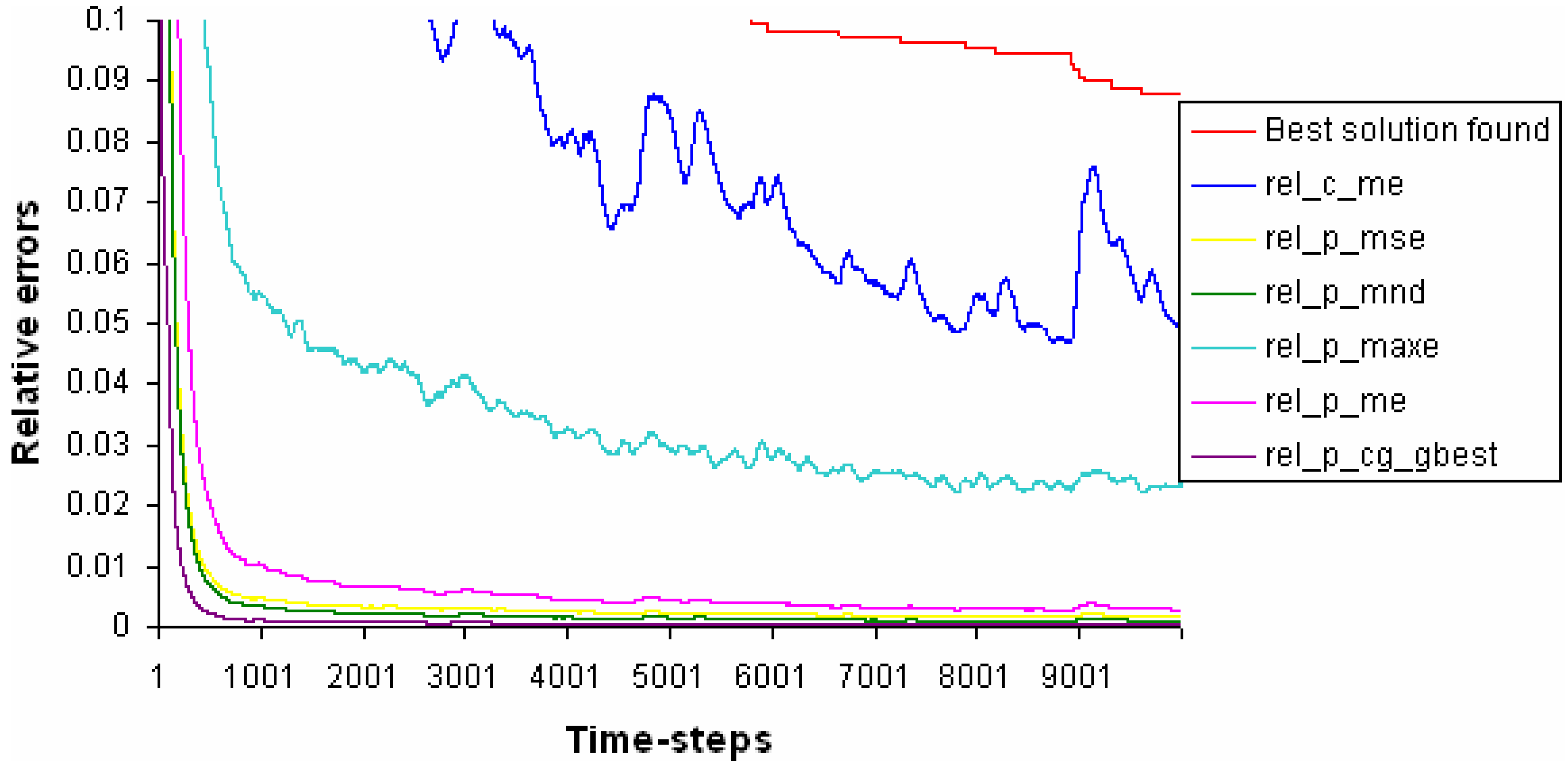

**Fig. 7. 39**: Evolution of the mean relative errors designed within the current time-step for the BSt-PSO$^{(c)}$ optimizing the 30-dimensional Schaffer f6 function.

The curves of the evolution of the mean relative errors regarding the particles' positions designed between consecutive time-steps—in addition to those of the **rel_c_cav** and of the **rel_c_cgbest**—when optimizing the 30-dimensional Rosenbrock, Sphere, Rastrigrin and Schaffer f6 functions are shown in **Fig. 7. 40** to **Fig. 7. 43**. Refer to **Appendix 4** for the complete set of experimental results.

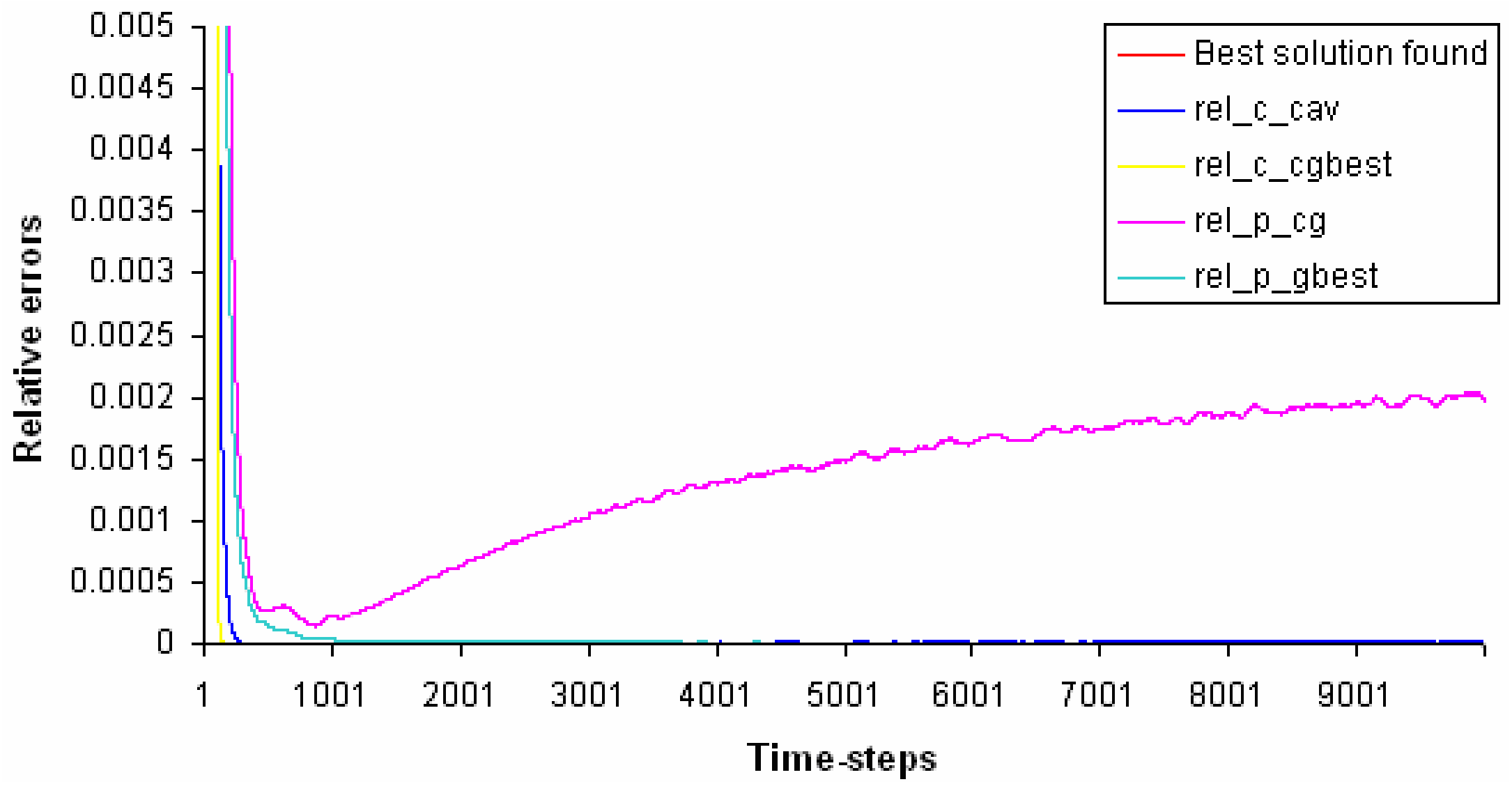

**Fig. 7. 40**: Evolution of the mean relative errors designed between consecutive time-steps for the BSt-PSO$^{(c)}$ optimizing the 30-dimensional Rosenbrock function.

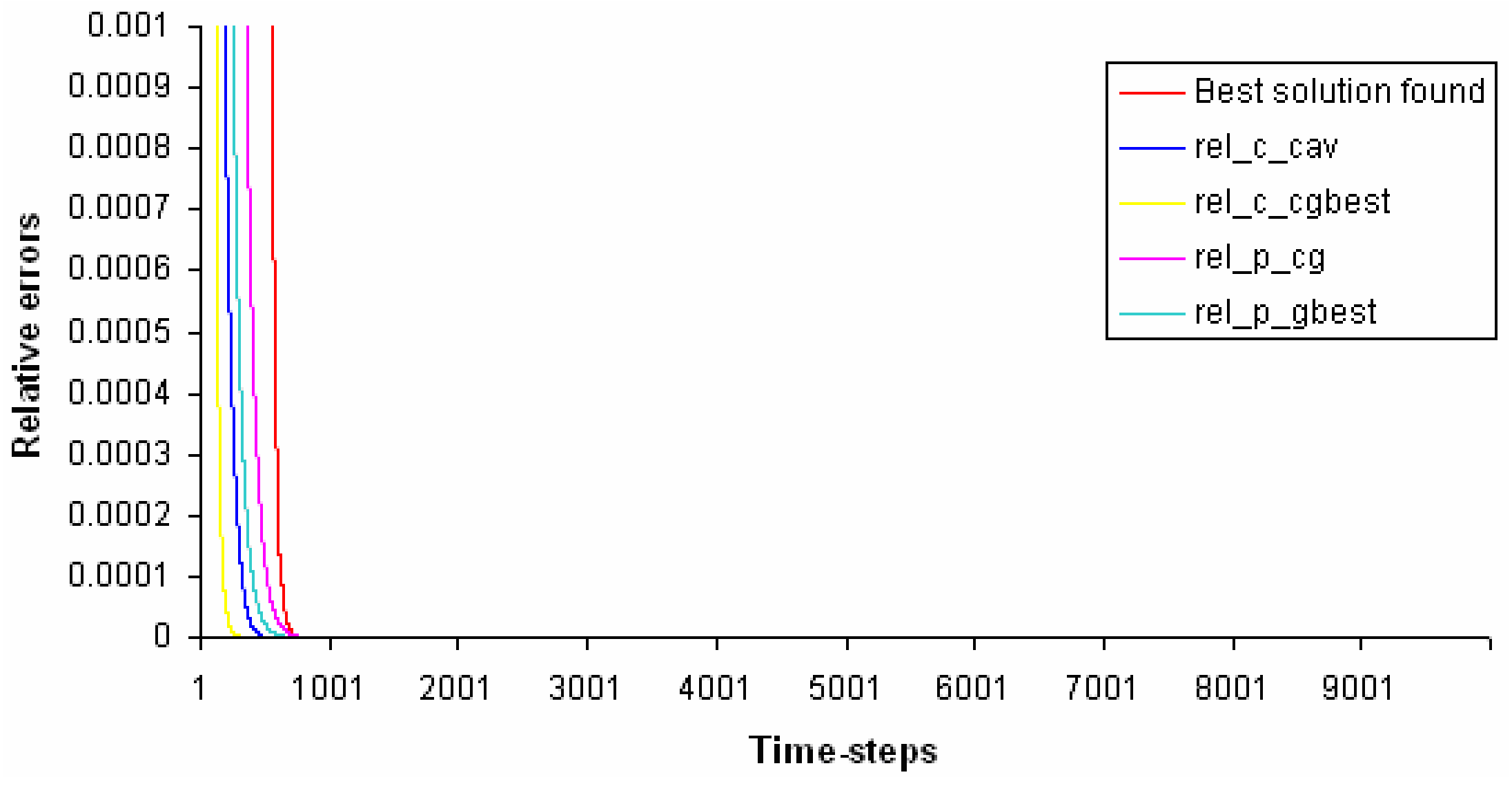

**Fig. 7. 41**: Evolution of the mean relative errors designed between consecutive time-steps for the BSt-PSO$^{(p)}$ optimizing the 30-dimensional Sphere function.

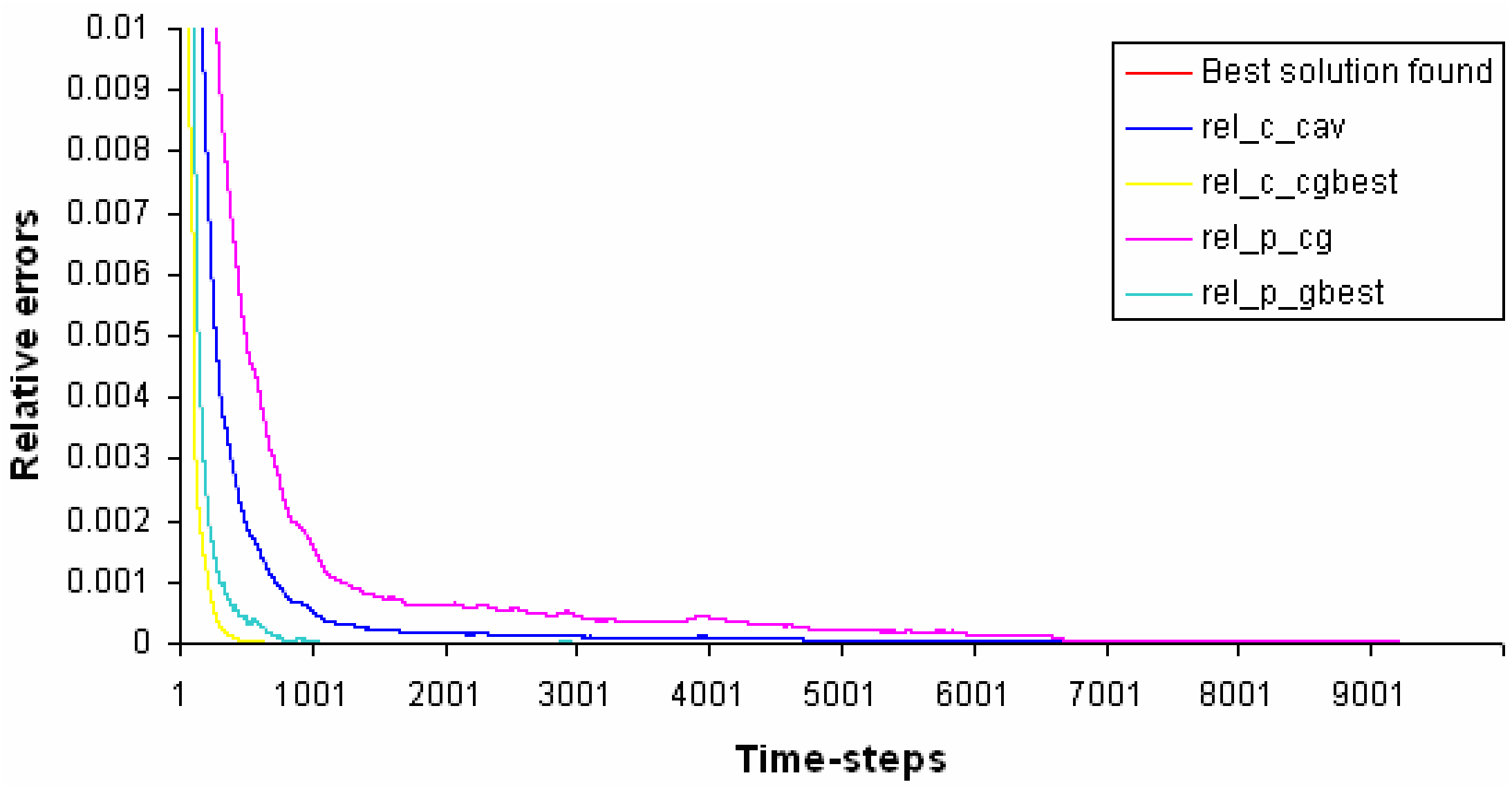

**Fig. 7. 42**: Evolution of the mean relative errors designed between consecutive time-steps for the BSt-PSO$^{(p)}$ optimizing the 30-dimensional Rastrigrin function.

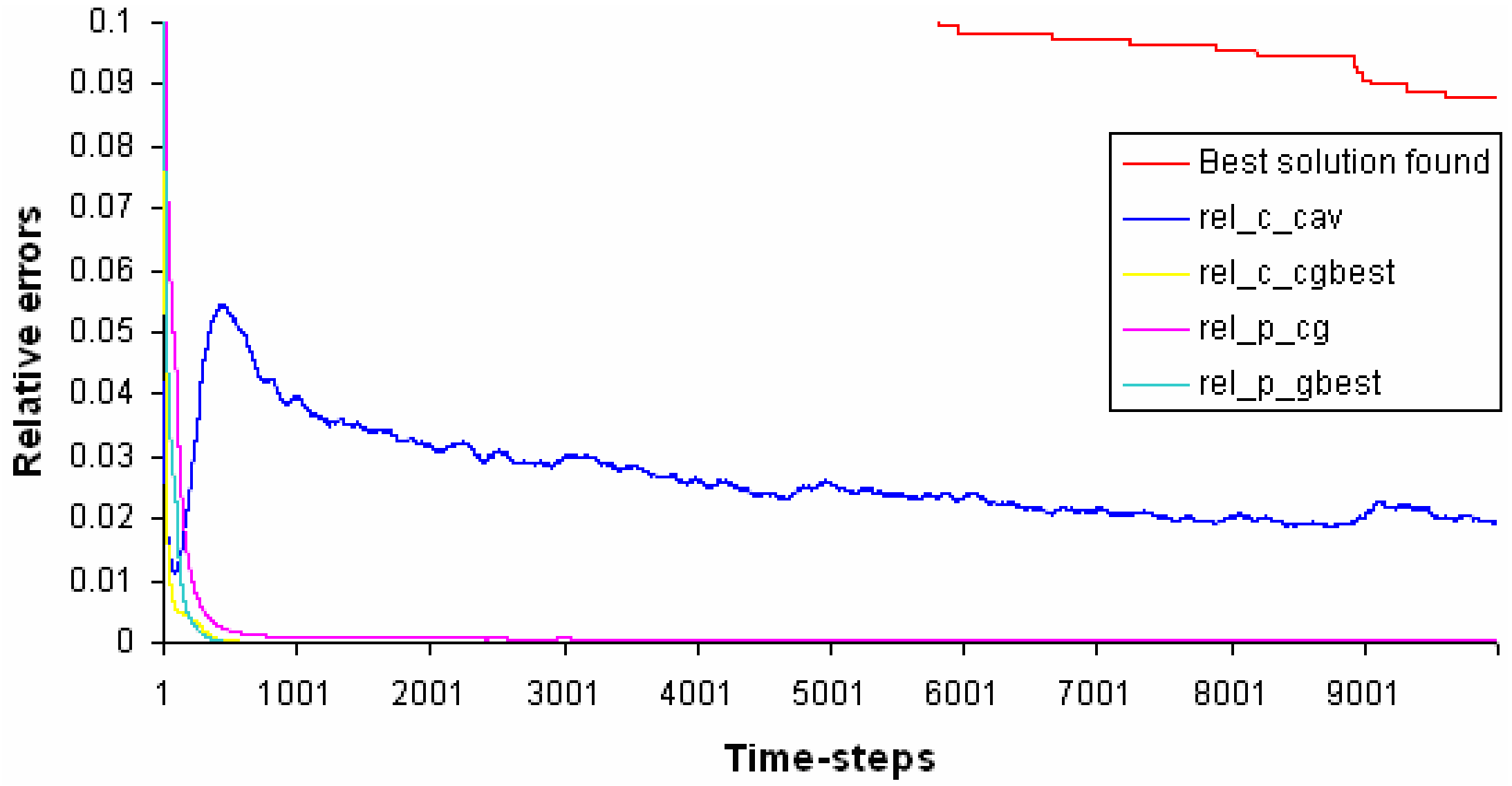

**Fig. 7. 43**: Evolution of the mean relative errors designed between consecutive time-steps for the BSt-PSO$^{(c)}$ optimizing the 30-dimensional Schaffer f6 function.

The graphical visualization of the results shows that the curves of the proposed errors both within the current time-step and between consecutive ones are reasonably smooth, so that it may be possible to set maximum permissible values for some of them so as to terminate the search when such conditions are met. The quantitative analysis of the experimental results is carried out along section **7.6**, while designing the stopping criteria.

### 7.5.2.2 Single run

As previously mentioned, while the analysis of the average behaviour is convenient for the quantitative analysis of the evolution of the errors, the behaviour along a single run is more appropriate for the qualitative analysis, since the stopping criteria is to be implemented for a single run. Of course, rougher curves of the evolution of the errors are to be expected here.

The curves of the evolution of the relative errors regarding the particles' positions designed within the current time-step—in addition to those of the **rel_c_me**—when optimizing the 30-dimensional Rosenbrock, Rastrigrin, and Schaffer f6 functions are shown in **Fig. 7. 44** to **Fig. 7. 46**. The curves of the evolution of the relative errors designed within the current time-step for the BSt-PSO$^{(p)}$ optimizing the 30-dimensional Sphere function are very similar to the corresponding curves of the mean relative errors (see **Fig. 7. 37**), so that they are not included here. The complete set of results obtained from the experiments can be found in **Appendix 4**.

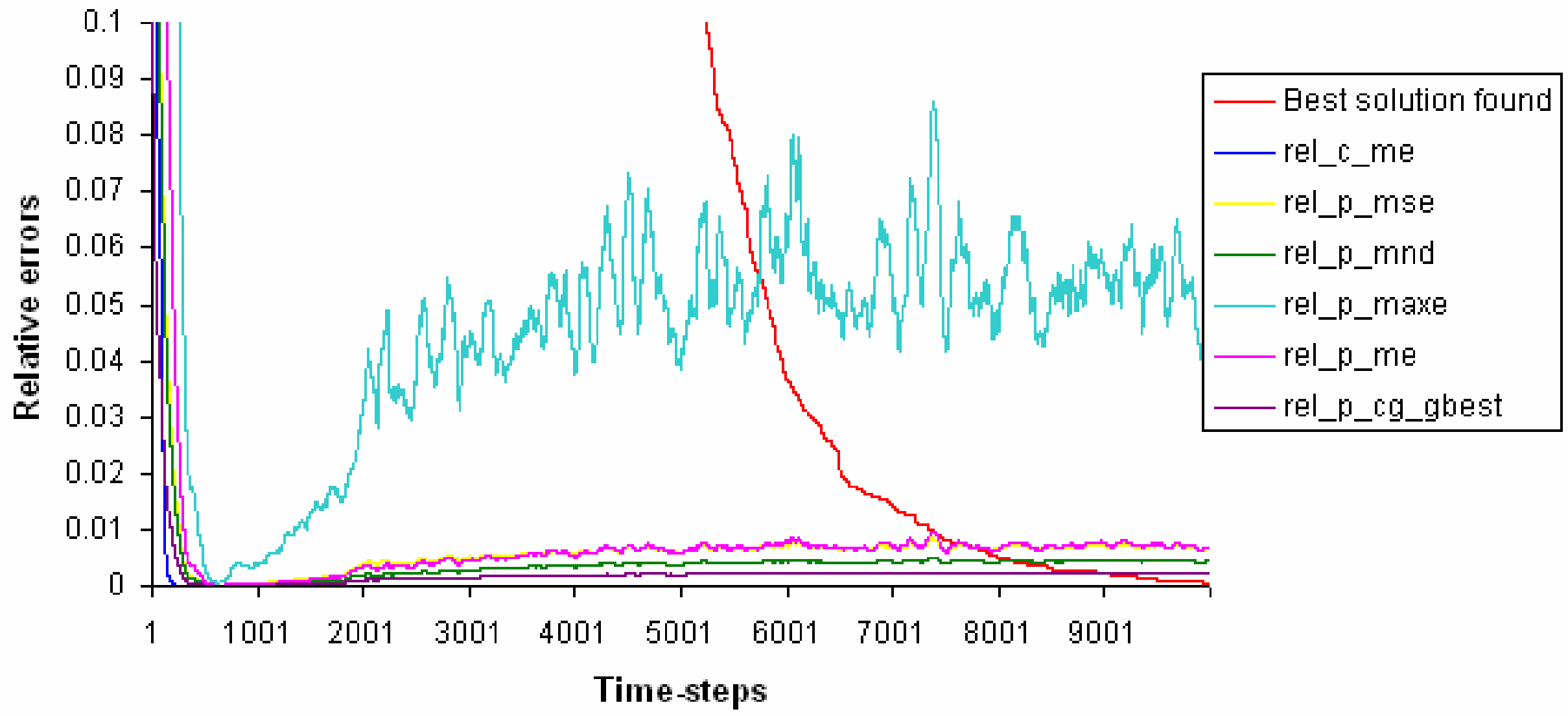

**Fig. 7. 44**: Evolution of the relative errors designed within the current time-step for the BSt-PSO$^{(c)}$ optimizing the 30-dimensional Rosenbrock function.

The small explosion can be clearly observed once more. It is fair to remark that the curves of the evolution of the **rel_p_maxe** display the roughest behaviour, exhibiting uneven oscillation of great amplitudes in comparison to the other proposed errors. The **rel_p_me** exhibits the second worst behaviour with regards to the suitability for their use in the development of stopping criteria.

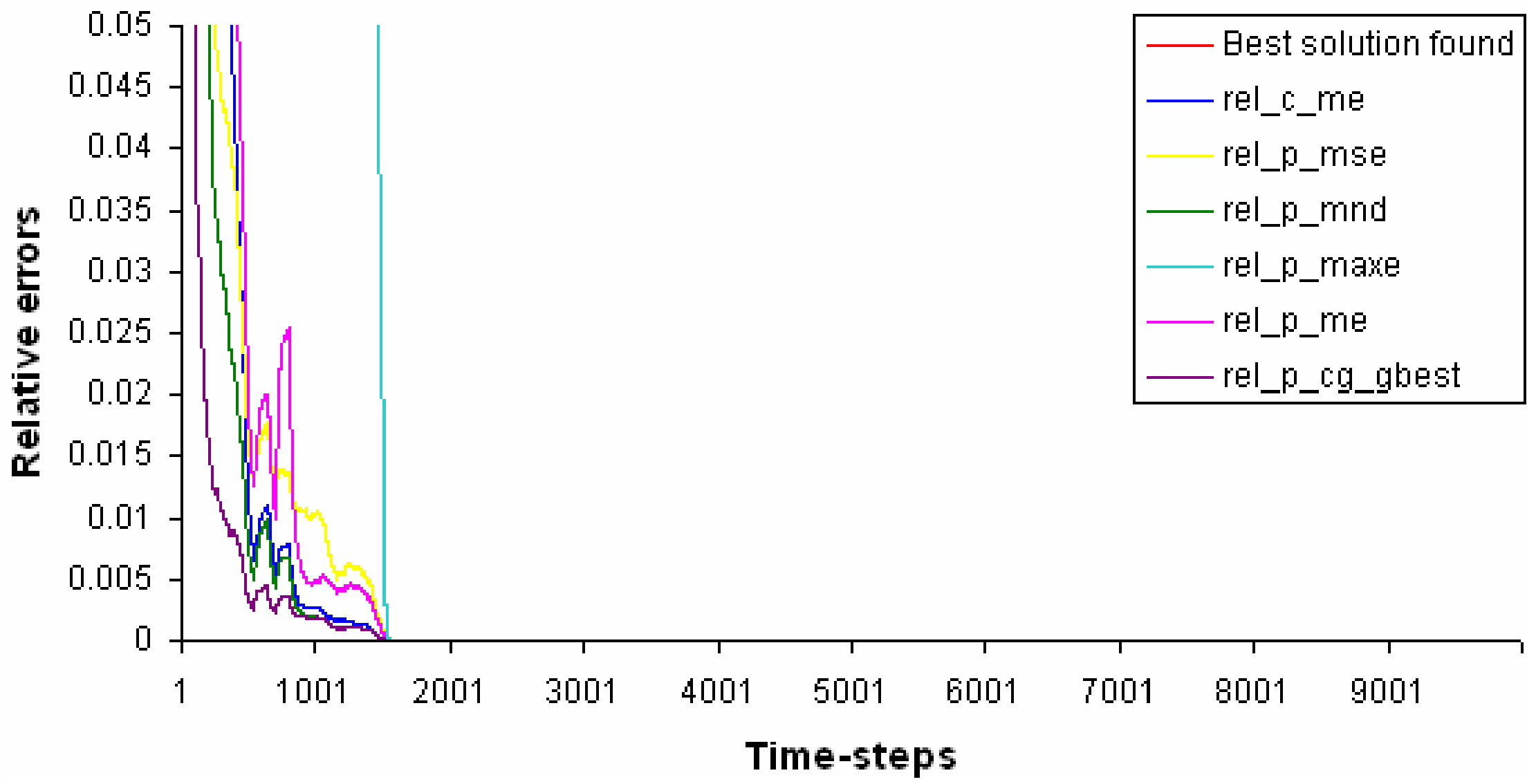

**Fig. 7. 45**: Evolution of the relative errors designed within the current time-step for the BSt-PSO$^{(p)}$ optimizing the 30-dimensional Rastrigrin function.

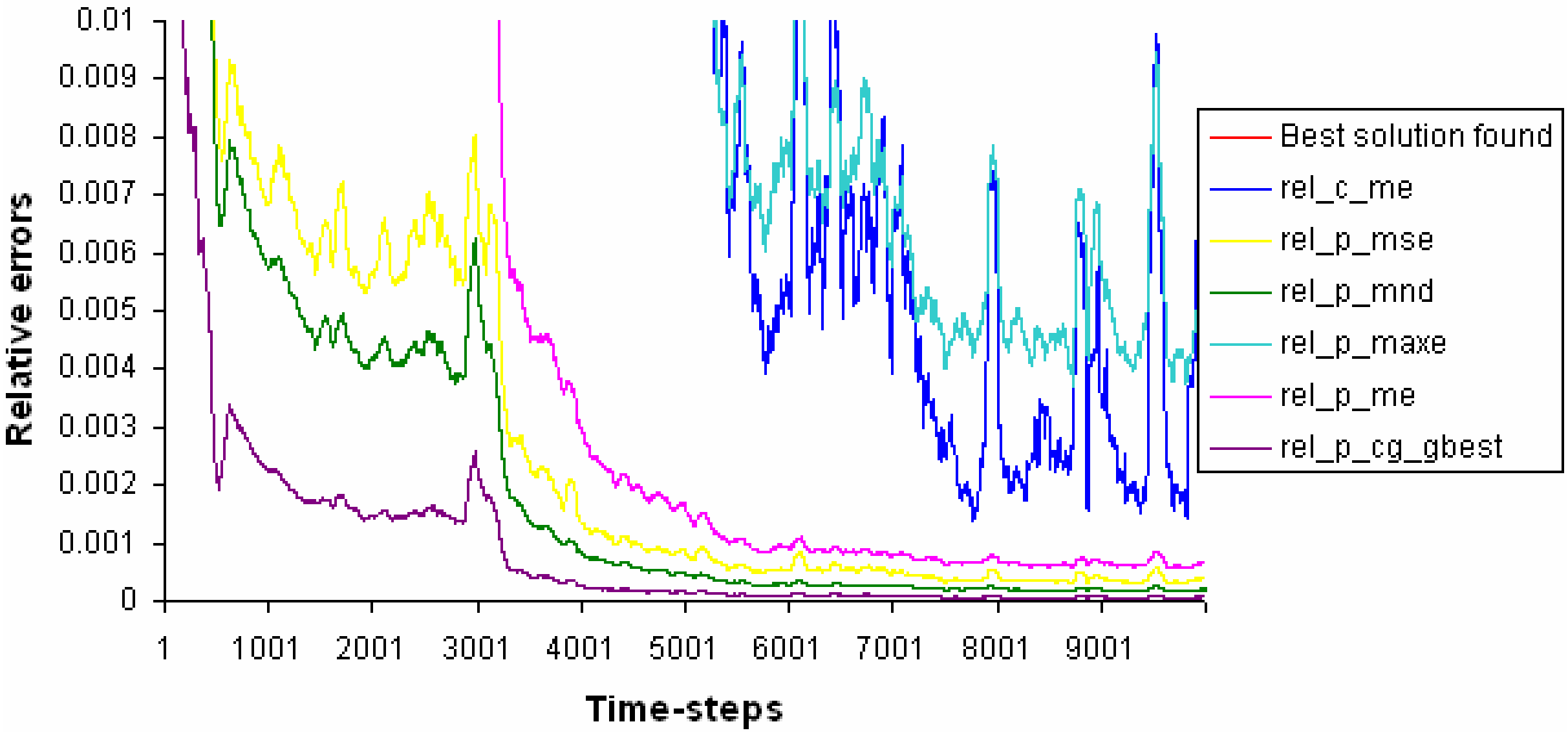

**Fig. 7. 46**: Evolution of the relative errors designed within the current time-step for the BSt-PSO$^{(c)}$ optimizing the 30-dimensional Schaffer f6 function.

The curves of the evolution of the relative errors regarding the particles' positions designed between consecutive time-steps—and those of the **rel_c_cav** and of the **rel_c_cgbest**—when optimizing the 30-dimensional Rosenbrock, Rastrigrin, and Schaffer f6 functions are shown in **Fig. 7. 47** to **Fig. 7. 49**. The curves of the evolution of the relative errors designed between consecutive time-steps for the BSt-PSO$^{(p)}$ optimizing the 30-dimensional Sphere function are very similar to the corresponding curves of the mean relative errors (see **Fig. 7. 41**), so that they are not included here. Refer to **Appendix 4** for the complete set of experimental results.

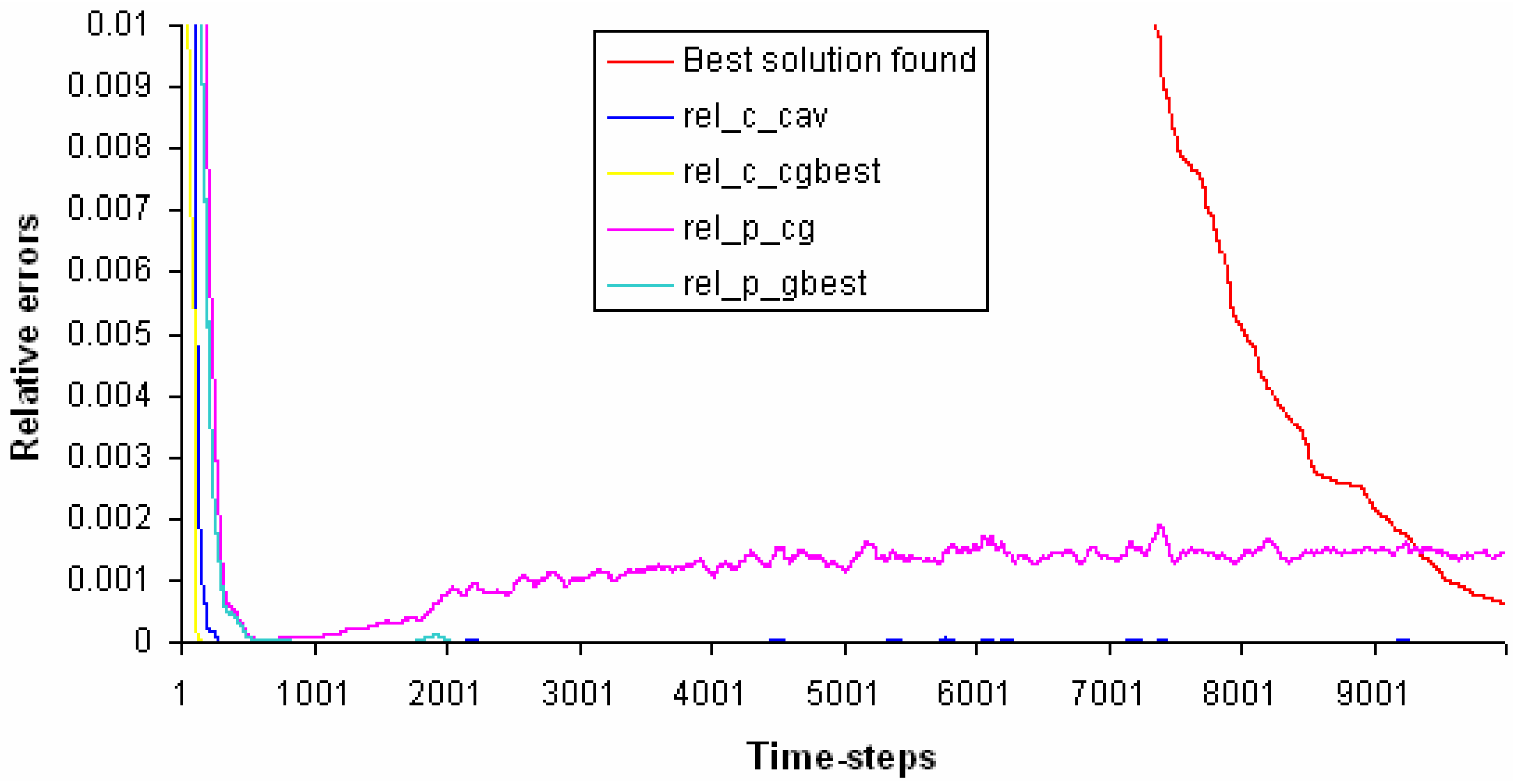

**Fig. 7. 47**: Evolution of the relative errors designed between consecutive time-steps for the BSt-PSO$^{(c)}$ optimizing the 30-dimensional Rosenbrock function.

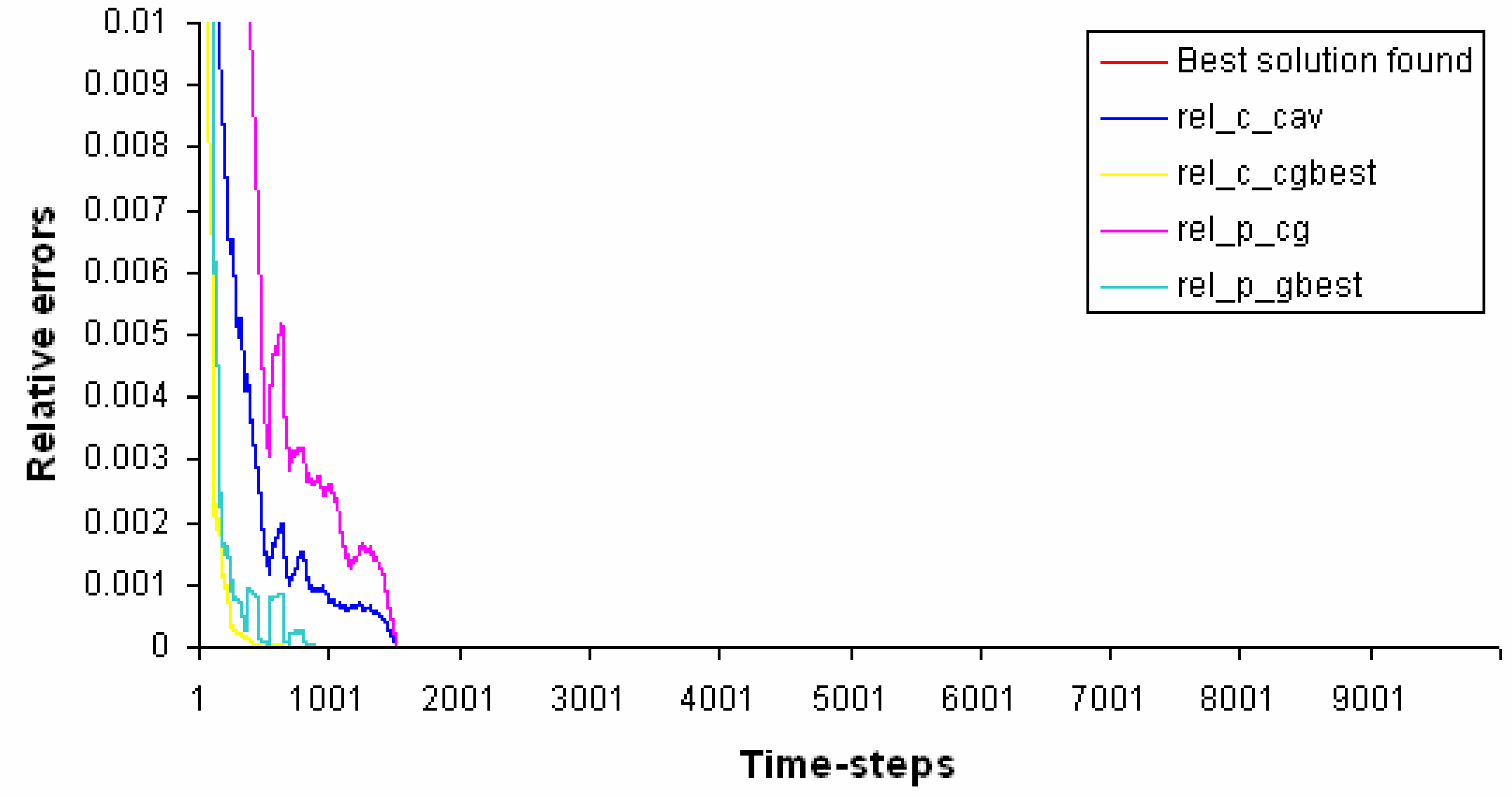

**Fig. 7. 48**: Evolution of the relative errors designed between consecutive time-steps for the BSt-PSO$^{(p)}$ optimizing the 30-dimensional Rastrigrin function.

The figures show that all the curves of the evolution of the proposed measures of error regarding the particles' positions except for those of the **rel_p_maxe** display reasonably smooth shapes, suitable for the development of stopping criteria. Among the other proposed measures, the **rel_p_me** shows the roughest shapes. Therefore, both the **rel_p_maxe** and the **rel_p_me** are removed from the ones to be considered for the design of stopping criteria along the next section.

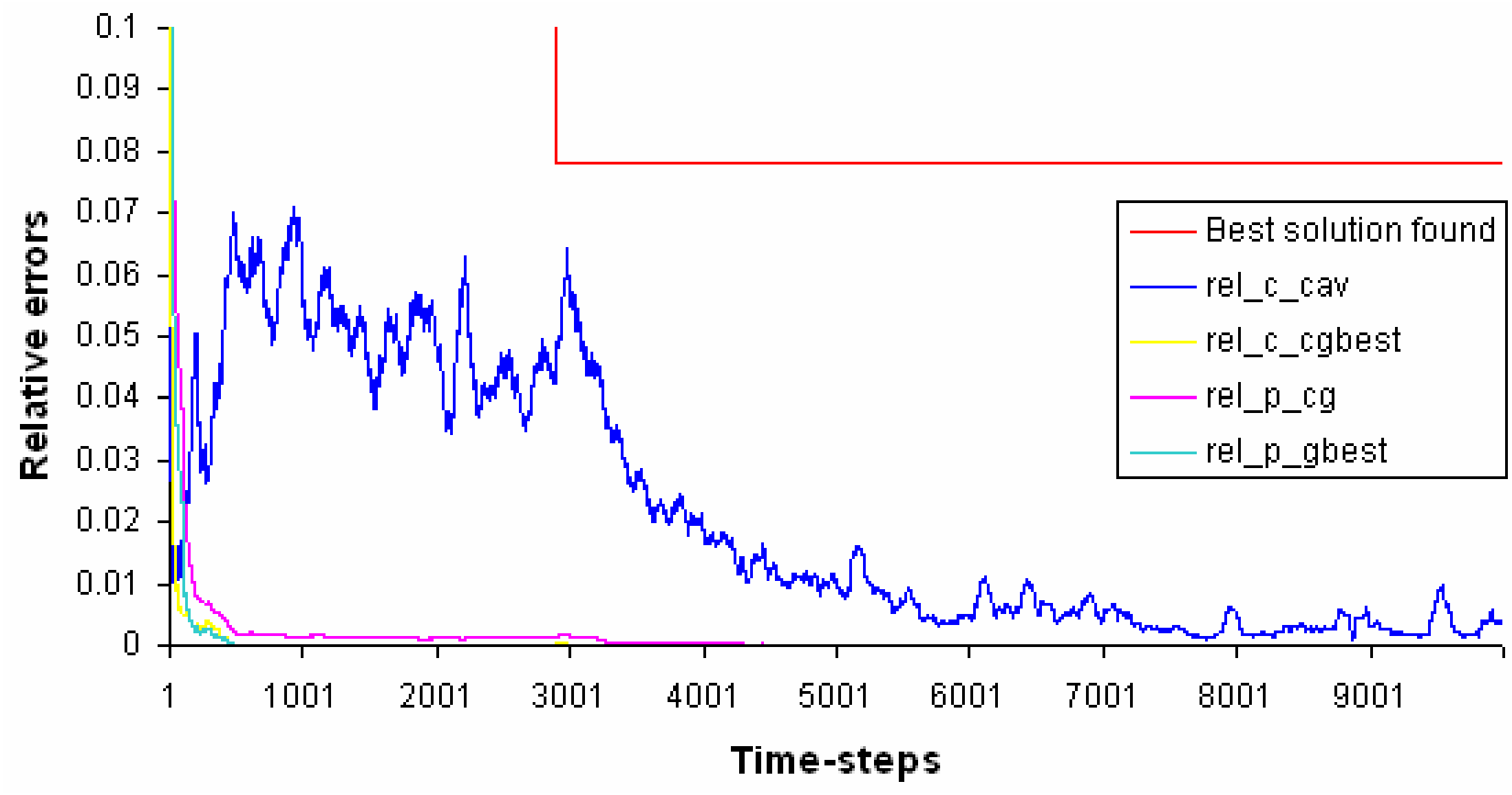

**Fig. 7. 49**: Evolution of the relative errors designed between consecutive time-steps for the BSt-PSO$^{(c)}$ optimizing the 30-dimensional Schaffer f6 function.

In summary, the selected measures of error are designed in the form of relative errors, where the values of reference are $cgworst - cgbest$ for the ones regarding the conflict values, and $x_{max} - x_{min}$ for the ones regarding the particles' positions. Four of the errors selected to be considered for the development of stopping criteria, namely the **rel_c_me**, the **rel_p_mse**, the **rel_p_mnd**, and the **rel_p_cg-gbest**, are designed within the current time-step. The other four, namely the **rel_c_cav**, the **rel_c_cgbest**, the **rel_p_cg**, and the **rel_p_gbest**, are designed between consecutive time-steps. However, the computation of all of them involves data obtained from the last 100 time-steps[11].

## 7.6 Stopping criteria

In spite of the fact of that the clustering of the particles does not necessarily imply the convergence of the algorithm, and that a high degree of clustering and the stagnation of the improvement of the best solution found do not necessarily imply a reliable solution, these are exactly the features considered for the design of the measures of error which are going to be used for the design of the stopping criteria. This is because there is no way to estimate the reliability of the solution found without knowing its real value. Therefore, the strategy adopted within this thesis consists of testing the proposed measures of error on a suite of benchmark functions, and comparing the evolution of their evaluations to the evolution of the reliability of the solution found, profiting from the fact that their global optimum is known.

[11] Recall that for the errors designed between consecutive time-steps, there are in reality 101 time-steps involved.

Note that each benchmark function in the test suite aims to test the algorithm against different challenges. However, there are many other difficulties that may arise in dealing with real-world problems which are not considered by any function in the test suite (e.g. conflict functions which exhibit numerous local optima that are located far from one another).

It is also important to note that the stopping criteria developed hereafter are designed for optimizers which exhibit fine-clustering ability. Thus, on the one hand, the particles may sometimes implode to a poor local optimum, so that the stopping criteria are met despite the algorithm finding a poor suboptimal solution. On the other hand, an optimizer without fine-clustering ability may sometimes find a good solution despite never meeting the stopping criteria. Therefore, the stopping criteria are developed for optimizers with fine-clustering ability, disregarding the problem of premature clustering. That is to say that the latter must be handled by the algorithm itself rather than by the termination condition. Hence, the fulfilment of the stopping criteria gives a certain guarantee with regards to the unlikelihood of finding a better solution, but it gives no direct idea of the reliability of the solution found.

## 7.6.1 Measures of error

As previously mentioned, the measures of error which were found to be promising for the design of the stopping criteria are as follows:

### Within the current time-step

- **rel_c_me**: average of the difference between the current average conflict and the best conflict found so far corresponding to the last 100 time-steps, related to the difference between the worst and the best conflicts found so far:

$$\text{rel_c_me}^{(t)} = \frac{\sum_{i=t-99}^{t}\left(\bar{c}^{(i)} - cgbest^{(i)}\right)}{100 \cdot \left(cgworst^{(t)} - cgbest^{(t)}\right)} \tag{7. 28}$$

- **rel_p_mse**: average of the square root of the mean squared error—where each error is viewed as the normalized distance from each particle's current position to the location of the best solution found so far—corresponding to the last 100 time-steps, related to the maximum range of feasible values for the design variables:

$$\text{rel_p_mse}^{(t)} = \frac{\sum_{i=t-99}^{t}\sqrt{\sum_{j=1}^{m}\sum_{k=1}^{n}\left(x_{jk}^{(i)} - gbest_k^{(i)}\right)^2}}{100 \cdot \left(x_{\max} - x_{\min}\right) \cdot \sqrt{m \cdot n}} \tag{7. 29}$$

- **rel_p_mnd**: average of the mean normalized distance from the particles' current positions to the location of the best solution found so far corresponding to the last 100 time-steps, related to the maximum range of feasible values for the design variables:

$$\text{rel_p_mnd}^{(t)} = \frac{\sum_{i=t-99}^{t}\sum_{j=1}^{m}\sqrt{\sum_{k=1}^{n}\left(x_{jk}^{(i)} - gbest_k^{(i)}\right)^2}}{100 \cdot \left(x_{\max} - x_{\min}\right) \cdot m \cdot \sqrt{n}} \tag{7. 30}$$

- **rel_p_cg-gbest**: average of the normalized distance from the centre of gravity (**cg**) of the swarm to the location of the best solution found so far (**gbest**) corresponding to the last 100 time-steps, related to the maximum range of feasible values for the design variables:

$$\text{rel_p_cg-gbest}^{(t)} = \frac{\sum_{i=t-99}^{t}\sqrt{\sum_{j=1}^{n}\left(cg_j^{(i)} - gbest_j^{(i)}\right)^2}}{100 \cdot \left(x_{\max} - x_{\min}\right) \cdot \sqrt{n}} \tag{7. 31}$$

**Between consecutive time-steps**

- **rel_c_cav**: average of the absolute difference between the current average conflict and the preceding one corresponding to the last 100 time-steps, related to the difference between the worst and the best conflicts found so far:

$$\text{rel_c_cav}^{(t)} = \frac{\sum_{i=t-99}^{t} \text{abs}\left(\bar{c}^{(i)} - \bar{c}^{(i-1)}\right)}{100 \cdot \left(cgworst^{(t)} - cgbest^{(t)}\right)} = \frac{\text{abs}\left(\bar{c}^{(t)} - \bar{c}^{(t-100)}\right)}{100 \cdot \left(cgworst^{(t)} - cgbest^{(t)}\right)} \quad \textbf{(7. 32)}$$

- **rel_c_cgbest**: average of the absolute difference between the best conflict found up to the current time-step and that found up to the preceding one corresponding to the last 100 time-steps, related to the difference between the worst and the best conflicts found so far:

$$\text{rel_c_cgbest}^{(t)} = \frac{\text{abs}\left(cgbest^{(t)} - cgbest^{(t-100)}\right)}{100 \cdot \left(cgworst^{(t)} - cgbest^{(t)}\right)} = \frac{cgbest^{(t-100)} - cgbest^{(t)}}{100 \cdot \left(cgworst^{(t)} - cgbest^{(t)}\right)} \quad \textbf{(7. 33)}$$

- **rel_p_cg**: average of the normalized distance between the current centre of gravity of the swarm and the preceding one corresponding to the last 100 time-steps, related to the maximum range of feasible values for the design variables:

$$\text{rel_p_cg}^{(t)} = \frac{\sum_{i=t-99}^{t} \sqrt{\sum_{j=1}^{n} \left(cg_j^{(t)} - cg_j^{(t-1)}\right)^2}}{100 \cdot \sqrt{n} \cdot \left(x_{i\max} - x_{i\min}\right)} \quad \textbf{(7. 34)}$$

- **rel_p_gbest**: average of the normalized distance between the best conflict found up to the current time-step and that found up to the preceding one corresponding to the last 100 time-steps, related to the maximum range of feasible values for the design variables:

$$\text{rel_p_gbest}^{(t)} = \frac{\sum_{i=t-99}^{t} \sqrt{\sum_{j=1}^{n} \left(gbest_j^{(t)} - gbest_j^{(t-1)}\right)^2}}{100 \cdot \sqrt{n} \cdot \left(x_{i\max} - x_{i\min}\right)} \quad \textbf{(7. 35)}$$

## 7.6.2 Quantitative analysis

The experiments for the analysis of all the proposed measures of error, many of which were already disregarded, were carried out along sections **7.4** and **7.5**. A first, qualitative, analysis was performed along those sections, leading to the selection of the eight relative errors described along section **7.6.1** for the quantitative analysis.

The values of the selected measures of error designed within a single time-step at the 10000th time-step are gathered in **Table 7. 8**, while those of the selected measures of error designed between consecutive time-steps at the 10000th time-step are gathered in **Table 7. 9**:

| RELATIVE ERRORS | | OPTIMIZERS | |
|---|---|---|---|
| | | BSt-PSO (c) | BSt-PSO (p) |
| SPHERE | Mean rel_c_me | 1.05E-161 | 2.02E-136 |
| | Mean rel_p_mse | 3.39E-82 | 2.41E-69 |
| | Mean rel_p_mnd | 2.34E-82 | 1.56E-69 |
| | Mean rel_p_cg-gbest | 9.65E-83 | 5.13E-70 |
| ROSENBROCK | Mean rel_c_me | 1.19E-05 | 4.01E-05 |
| | Mean rel_p_mse | 9.84E-03 | 1.21E-02 |
| | Mean rel_p_mnd | 6.38E-03 | 8.64E-03 |
| | Mean rel_p_cg-gbest | 3.42E-03 | 3.76E-03 |
| RASTRIGRIN | Mean rel_c_me | 2.99E-17 | 6.45E-17 |
| | Mean rel_p_mse | 1.30E-10 | 2.69E-10 |
| | Mean rel_p_mnd | 1.13E-10 | 2.05E-10 |
| | Mean rel_p_cg-gbest | 4.12E-11 | 6.47E-11 |
| GRIEWANK | Mean rel_c_me | 1.29E-19 | 5.56E-19 |
| | Mean rel_p_mse | 1.19E-11 | 2.67E-11 |
| | Mean rel_p_mnd | 1.01E-11 | 1.98E-11 |
| | Mean rel_p_cg-gbest | 3.49E-12 | 6.17E-12 |
| SCHAFFER F6 2D | Mean rel_c_me | 2.30E-03 | 9.84E-17 |
| | Mean rel_p_mse | 1.78E-04 | 2.87E-11 |
| | Mean rel_p_mnd | 4.86E-05 | 1.78E-11 |
| | Mean rel_p_cg-gbest | 3.65E-05 | 7.01E-12 |
| SCHAFFER F6 | Mean rel_c_me | 4.92E-02 | 1.62E-01 |
| | Mean rel_p_mse | 1.77E-03 | 4.52E-03 |
| | Mean rel_p_mnd | 9.79E-04 | 2.84E-03 |
| | Mean rel_p_cg-gbest | 3.58E-04 | 8.89E-04 |

**Table 7. 8**: Mean values of the selected relative errors designed within a single time-step at the 10000th time-step, where the mean is computed out of 50 runs.

| RELATIVE ERRORS | | OPTIMIZERS | |
|---|---|---|---|
| | | BSt-PSO (c) | BSt-PSO (p) |
| SPHERE | Mean rel_c_cav | 5.39E-162 | 1.17E-136 |
| | Mean rel_c_cgbest | 1.59E-163 | 4.27E-139 |
| | Mean rel_p_cg | 1.05E-82 | 7.40E-70 |
| | Mean rel_p_gbest | 5.56E-83 | 1.39E-70 |
| ROSENBROCK | Mean rel_c_cav | 1.31E-05 | 3.98E-05 |
| | Mean rel_c_cgbest | 2.71E-13 | 1.11E-13 |
| | Mean rel_p_cg | 1.98E-03 | 3.18E-03 |
| | Mean rel_p_gbest | 2.51E-06 | 1.56E-06 |
| RASTRIGRIN | Mean rel_c_cav | 5.93E-18 | 2.41E-17 |
| | Mean rel_c_cgbest | 0.00E+00 | 0.00E+00 |
| | Mean rel_p_cg | 2.71E-11 | 7.56E-11 |
| | Mean rel_p_gbest | 1.50E-14 | 0.00E+00 |
| GRIEWANK | Mean rel_c_cav | 2.51E-20 | 2.46E-19 |
| | Mean rel_c_cgbest | 0.00E+00 | 0.00E+00 |
| | Mean rel_p_cg | 2.46E-12 | 7.55E-12 |
| | Mean rel_p_gbest | 0.00E+00 | 0.00E+00 |
| SCHAFFER F6 2D | Mean rel_c_cav | 1.30E-03 | 6.49E-17 |
| | Mean rel_c_cgbest | 0.00E+00 | 0.00E+00 |
| | Mean rel_p_cg | 3.20E-05 | 7.40E-12 |
| | Mean rel_p_gbest | 0.00E+00 | 0.00E+00 |
| SCHAFFER F6 | Mean rel_c_cav | 1.93E-02 | 3.91E-02 |
| | Mean rel_c_cgbest | 0.00E+00 | 0.00E+00 |
| | Mean rel_p_cg | 3.63E-04 | 1.24E-03 |
| | Mean rel_p_gbest | 0.00E+00 | 0.00E+00 |

**Table 7. 9**: Mean values of the selected relative errors designed between consecutive time-steps at the 10000$^{th}$ time-step, where the mean is computed out of 50 runs.

The optimization of the different benchmark functions in the test suite present different degrees of difficulty. Therefore, it cannot be expected that the degree of accuracy of the solution found and the values of the measures of error are similar for all the functions at the 10000th time-step. For instance, the solution never stops being improved in the case of the Sphere function, so that the solution at the 10000th time-step is very accurate. However, the solution was good enough noticeably sooner, and the search could have been stopped so as to save computational cost. Thus, the aim here is to stop the search when the solution is good enough, so that the optimization of the simpler functions simply take shorter. In contrast, the best solution found for the Schaffer f6 function was not satisfactory after 10000 time-steps, so that either the search should be extended, or the optimizers simply failed in dealing with the problem. Hence, the permissible values for the measures of error should be more demanding than those achieved by the particles of the optimizers when dealing with the Schaffer f6 function (refer to **Table 7. 8**, **Table 7. 9**, and **Appendix 4**). The values of the errors achieved when optimizing the Rosenbrock function show that the clustering of the particles is not exactly magnificent, so that the permissible values for the errors should be more demanding than the smallest ones achieved by the particles. Notice that the smallest ones are not the ones corresponding to the last time-step in the case of the Rosenbrock function due to the small explosion whose cause is not understood by the author of this thesis.

Based on the results of the experiments run along sections **7.4** and **7.5**, which are gathered in **Table 7. 8**, **Table 7. 9**, and in more details in **Appendix 4**, the eight selected measures of errors are given the following maximum permissible values:

- $$\text{rel_c_me}^{(t)} = \frac{\sum_{i=t-99}^{t}\left(\bar{c}^{(i)} - cgbest^{(i)}\right)}{100\cdot\left(cgworst^{(t)} - cgbest^{(t)}\right)} \leq 1\text{x}10^{-12}$$

- $$\text{rel_p_mse}^{(t)} = \frac{\sum_{i=t-99}^{t}\sqrt{\sum_{j=1}^{m}\sum_{k=1}^{n}\left(x_{jk}^{(i)} - gbest_k^{(i)}\right)^2}}{100\cdot\left(x_{\max} - x_{\min}\right)\cdot\sqrt{m\cdot n}} \leq 1\text{x}10^{-9}$$

- $$\text{rel_p_mnd}^{(t)} = \frac{\sum_{i=t-99}^{t}\sum_{j=1}^{m}\sqrt{\sum_{k=1}^{n}\left(x_{jk}^{(i)} - gbest_k^{(i)}\right)^2}}{100\cdot\left(x_{\max} - x_{\min}\right)\cdot m\cdot\sqrt{n}} \leq 1\text{x}10^{-9}$$

- $$\text{rel_p_cg-gbest}^{(t)} = \frac{\sum_{i=t-99}^{t}\sqrt{\sum_{j=1}^{n}\left(cg_j^{(i)} - gbest_j^{(i)}\right)^2}}{100\cdot\left(x_{\max} - x_{\min}\right)\cdot\sqrt{n}} \leq 1\text{x}10^{-9}$$

- $\text{rel_c_cav}^{(t)} = \dfrac{\sum_{i=t-99}^{t} \text{abs}\left(\bar{c}^{(i)} - \bar{c}^{(i-1)}\right)}{100 \cdot \left(cgworst^{(t)} - cgbest^{(t)}\right)} = \dfrac{\text{abs}\left(\bar{c}^{(t)} - \bar{c}^{(t-100)}\right)}{100 \cdot \left(cgworst^{(t)} - cgbest^{(t)}\right)} \leq 1\text{x}10^{-12}$

- $\text{rel_c_cgbest}^{(t)} = \dfrac{\text{abs}\left(cgbest^{(t)} - cgbest^{(t-100)}\right)}{100 \cdot \left(cgworst^{(t)} - cgbest^{(t)}\right)} = \dfrac{cgbest^{(t-100)} - cgbest^{(t)}}{100 \cdot \left(cgworst^{(t)} - cgbest^{(t)}\right)} \leq 1\text{x}10^{-15}$

- $\text{rel_p_cg}^{(t)} = \dfrac{\sum_{i=t-99}^{t} \sqrt{\sum_{j=1}^{n} \left(cg_j^{(t)} - cg_j^{(t-1)}\right)^2}}{100 \cdot \sqrt{n} \cdot \left(x_{i\max} - x_{i\min}\right)} \leq 1\text{x}10^{-9}$

- $\text{rel_p_gbest}^{(t)} = \dfrac{\sum_{i=t-99}^{t} \sqrt{\sum_{j=1}^{n} \left(gbest_j^{(t)} - gbest_j^{(t-1)}\right)^2}}{100 \cdot \sqrt{n} \cdot \left(x_{i\max} - x_{i\min}\right)} \leq 1\text{x}10^{-12}$

It should be remarked that some of above permissible values might lead to unacceptable solutions when implemented independently. That is to say, the eight conditions must be met at the same time to allow the termination of the algorithm. This tuning of the errors intends by no means to be definitive, and further work remains to be done to find the appropriate values for a general-purpose optimizer. This is merely the first step in that direction.

### 7.6.3 Termination condition

There are some functions that exhibit numerous local optima located near the global optimum, resulting in the increase of the measures of error regarding the conflict values as the particles cluster. This increase may be reverted while the particles fine-cluster, if they are able to do so. A clear example of this is the Schaffer f6 function, which also has the characteristic that all the conflict values corresponding to coordinates located far from the global optimum are very similar to one another, and near a value of 0.5. This feature results in small values of the errors regarding the conflicts at the beginning of the search, while the best solution found so far is still far from satisfactory. Therefore, both a minimum and a maximum permissible numbers of time-steps should be set for the search.

With regards to the measures of error to be used in the design of stopping criteria, it can be observed that the behaviour of the **rel_p_mse** and of the **rel_p_mnd** are very much alike both qualitatively and quantitatively speaking. Since the latter is computationally more expensive, it is removed from the measures of error to be implemented for the stopping criteria.

In addition, it some experiments showed that even the particles of optimizers with the ability to fine-cluster find it very difficult to do so when optimizing the Rosenbrock function. Thus, it can be observed that the **BSt-PSO$^{(c)}$** is able to find very good solutions, despite the fact that their particles do not perform a complete implosion. This is because the particles do fine-cluster soon, but then they slightly diverge, introducing some diversity in the swarm. Such small divergence is useful to find new better solutions. In fact, the curves of the evolution of the best conflict found by the **BSt-PSO$^{(c)}$** and by the **BSt-PSO$^{(c)}$** do not display stagnation. While this feature is useful for the best solution found by the optimizer, it introduces a problem into the development of the termination condition: if the measures of the clustering[12] were met when the particles achieve the first thorough clustering, the algorithm would be terminated while important further improvement is still possible; if the measures of clustering were more demanding, they would never be met because the particles diverge from there forth. Another clear case of good solutions found despite not exhibiting striking clustering is that of the 2 dimensional Schaffer f6 function, which presents a serious challenge to the clustering of the particles because of its particular shape (refer to **Appendix 3**). Therefore, there are two sets of conditions proposed here, which lead to the termination of the iterative process if any of them are met:

### First set of termination conditions

The search is terminated if the eight following conditions are attained:

1) $t \geq 0.1 \cdot t_{\max}$

2) $\text{rel_c_me}^{(t)} = \dfrac{\sum_{i=t-99}^{t}\left(\bar{c}^{(i)} - cgbest^{(i)}\right)}{100 \cdot \left(cgworst^{(t)} - cgbest^{(t)}\right)} \leq 1\text{x}10^{-12}$

3) $\text{rel_p_mse}^{(t)} = \dfrac{\sum_{i=t-99}^{t}\sqrt{\sum_{j=1}^{m}\sum_{k=1}^{n}\left(x_{jk}^{(i)} - gbest_k^{(i)}\right)^2}}{100 \cdot \left(x_{\max} - x_{\min}\right) \cdot \sqrt{m \cdot n}} \leq 1\text{x}10^{-9}$

[12] That is, the measures of error regarding the particles' positions design within a single time-step.

4) $$\text{rel_p_cg-gbest}^{(t)} = \frac{\sum_{i=t-99}^{t} \sqrt{\sum_{j=1}^{n} \left(cg_j^{(i)} - gbest_j^{(i)}\right)^2}}{100 \cdot (x_{\max} - x_{\min}) \cdot \sqrt{n}} \leq 1\text{x}10^{-9}$$

5) $$\text{rel_c_cav}^{(t)} = \frac{\sum_{i=t-99}^{t} \text{abs}\left(\bar{c}^{(i)} - \bar{c}^{(i-1)}\right)}{100 \cdot \left(cgworst^{(t)} - cgbest^{(t)}\right)} = \frac{\text{abs}\left(\bar{c}^{(t)} - \bar{c}^{(t-100)}\right)}{100 \cdot \left(cgworst^{(t)} - cgbest^{(t)}\right)} \leq 1\text{x}10^{-12}$$

6) $$\text{rel_c_cgbest}^{(t)} = \frac{\text{abs}\left(cgbest^{(t)} - cgbest^{(t-100)}\right)}{100 \cdot \left(cgworst^{(t)} - cgbest^{(t)}\right)} = \frac{cgbest^{(t-100)} - cgbest^{(t)}}{100 \cdot \left(cgworst^{(t)} - cgbest^{(t)}\right)} \leq 1\text{x}10^{-15}$$

7) $$\text{rel_p_cg}^{(t)} = \frac{\sum_{i=t-99}^{t} \sqrt{\sum_{j=1}^{n} \left(cg_j^{(t)} - cg_j^{(t-1)}\right)^2}}{100 \cdot \sqrt{n} \cdot (x_{i\max} - x_{i\min})} \leq 1\text{x}10^{-9}$$

8) $$\text{rel_p_gbest}^{(t)} = \frac{\sum_{i=t-99}^{t} \sqrt{\sum_{j=1}^{n} \left(gbest_j^{(t)} - gbest_j^{(t-1)}\right)^2}}{100 \cdot \sqrt{n} \cdot (x_{i\max} - x_{i\min})} \leq 1\text{x}10^{-12}$$

**Second set of termination conditions**

The search is terminated if the two following conditions are attained:

1) $t > 0.25 \cdot t_{\max}$

2) $cgbest^{(t-0.25 \cdot t_{\max})} - cgbest^{(t)} = 0$

Note that terminating the search due to the attainment of one or the other set of conditions has completely different meanings. Attaining the first set of conditions implies that the particles have achieved a high degree of clustering, and that the rate of improvement of the solution has reached a lower permissible threshold. Fulfilling the second set of conditions implies that, although the particles have not yet achieved the required degree of clustering, further improvement of the best solution found is unlikely. Clearly, both cases justify the termination of the iterative process. Beware that the condition of running the search for at least the 10% of the maximum permissible number of time-steps is guaranteed by both sets of conditions.

It is important to remark that attaining either one or the other set of conditions does not give direct information with regards to the goodness of the solution found. For instance, the second set of conditions might be met by an optimizer which is not able to improve the solutions because its particles do not cluster at all. Again, the latter is a problem of designing the algorithm itself rather than a problem of the design of the measures of error. In any case, the search should be terminated if no further improvement is likely to take place.

A number of experiments were performed aiming to test the above specified termination conditions on the suite of benchmark functions shown in **Table 6.1**. As previously argued, the stopping criteria was in principle developed for optimizers with the ability to fine-cluster[13]. Therefore, the experiments were carried out only for the **BSt-PSO$^{(c)}$** and the **BSt-PSO$^{(p)}$**. The most relevant results obtained from the experiments are gathered in **Table 7. 10**:

| FUNCTIONS | BSt-PSO$^{(c)}$ | | | BSt-PSO$^{(p)}$ | | |
|---|---|---|---|---|---|---|
| | Solution | Time-steps to meet stopping criteria | Set of termination conditions attained | Solution | Time-steps to meet stopping criteria | Set of termination conditions attained |
| Sphere | 1.17E-45 | 3000 | 1 | 1.19E-37 | 3000 | 1 |
| Rosenbrock | 6.48E-10 | - | - | 3.70E+01 | 17829 | 2 |
| Rastrigrin | 5.97E+01 | 3000 | 1 | 3.98E+01 | 3000 | 1 |
| Griewank | 2.95E-02 | 3000 | 1 | 0.00E+00 | 3000 | 1 |
| Schaffer f6 2D | 0.00E+00 | 4802 | 1 | 0.00E+00 | 3222 | 1 |
| Schaffer f6 | 7.82E-02 | 15223 | 2 | 7.82E-02 | 15223 | 2 |

**Table 7. 10**: Results obtained from testing the BSt-PSO$^{(c)}$ and the BSt-PSO$^{(p)}$ with the stopping criteria incorporated on the suite of benchmark functions showed in Table 6.1, where the set of termination conditions attained indicates the reason for the search to be terminated.

It is interesting to observe that the solutions found when the error conditions are attained also satisfy the acceptable exact absolute errors stated in **Table 6.1**, which were taken from Carlisle et al. [13][14].

It was claimed several times before that the Rosenbrock function and the Schaffer f6 function present the highest difficulties to the clustering of the particles, which is corroborated by the fact that the optimizers attained the second set of termination conditions when dealing with those two functions (refer to **Table 7. 10**).

[13] Note, however, that the second set of conditions might well be applicable to optimizers whose particles do not exhibit the ability to fine-cluster.

[14] It should be remarked that the absolute error conditions stated in **Table 6.1**, which can be set because the global optimum is well known for benchmark functions, were not used at all for the development of the proposed stopping criteria.

# 7.7 Closure

A brief review of the simplest traditional measures of error was carried out. The advantages and disadvantages of computing absolute and relative errors were discussed, and some of the reasons why these traditional measures of error are not suitable for population-based methods were outlined.

Several measures of error—suitable for particle swarm optimizers—were proposed, tested, and some of them disregarded. The aim was to develop measures of error to be used in the design of stopping criteria so as to terminate the search either when the best solution found was reliable, or when further improvement was unlikely or negligible. Although the reliability of the solution cannot be directly estimated for real-world problems, it was indirectly related to the degree of clustering achieved by the particles and to the rate of improvement of the best solution found[15]. This was carried out by analyzing the evolution of the measures of error together with the evolution of the best conflict found when optimizing a set of benchmark functions, whose global optima are well known. Thus, appropriate general-purpose settings for the permissible values of the errors were proposed by extrapolating the results obtained from the experiments.

The qualitative analysis of the evolution of the errors was performed on the results obtained from single independent runs of the algorithm. In contrast, the quantitative analysis was performed on the average results obtained from a set of 50 runs for each experiment.

Some of the proposed measures of error were designed within a single time-step, while some others were designed between consecutive ones. The former somehow measure the degree of clustering of the particles, while the latter somehow measure the rate of improvement of the best solution found. Besides, in the same fashion as traditional measures of error, some of the proposed errors were designed involving the conflict values and some others involving the particles' coordinates.

Finally, the stopping criteria were designed by selecting some of the proposed measures of error, and setting some general-purpose permissible values for them. Two sets of termination conditions were proposed, where the first set somehow measures the degree of clustering achieved by the particles and the rate of improvement of the best solution found, while the second set detects the complete stagnation of the improvement of the best solution found throughout a certain number of time-steps. The attainment of any of this sets of termination conditions results in the termination of the iterative search.

[15] It should be remarked that this implies that the termination conditions consider the unlikelihood of further meaningful improvement only, leaving the reliability of the best solution found to the abilities of the optimizers themselves.

The 30 optimizers proposed along section **6.4**—which only differ in the settings of their parameters—but with few slight modifications on some of them and with the incorporation of the termination conditions, are tested again on the same suite of benchmark functions along **Chapter 8**. Since a quantitative analysis is to be carried out, the average of the results obtained from 50 runs will be considered. Thus, the influence of the parameters' setting on the achievements of the particle swarm optimizer is analyzed in terms of the mean best solution found when the stopping criteria are met; the mean number of time-steps required to meet them; which of the two sets of stopping criteria effectively terminates the iterative search; the number of failures in meeting the stopping criteria; and the mean best solution found when the stopping criteria are not attained.

A few adaptations to the algorithm will be carried out along **Chapter 9** by considering the conclusions derived from the experiments performed along **Chapter 8**, and a few constraint-handling techniques are briefly discussed along **Chapter 10**. Applications of the resulting general-purpose optimizers are carried out along **Chapter 11**, **Chapter 12**, and **Chapter 13**, although some of these problems are still rather academic.

# Chapter 8

# FURTHER ANALYSIS OF THE PARAMETERS

A preliminary, fairly extensive analysis of the influence of the parameters' settings of the basic particle swarm optimizer on the behaviour of the swarm was undertaken along **Chapter 6**, although the runs were performed along a fixed number of time-steps. The goodness of the solutions found when testing the algorithms on a suite of benchmark functions was estimated based on permissible absolute errors taken from the literature, but without any consideration of the reasons why such errors are acceptable. Hence **Chapter 7** was entirely devoted to the development of stopping criteria, so that the algorithm can be terminated either when the solution found is good enough or when further improvement is unlikely or negligible, thus saving computational costs. This also allows setting the same termination conditions for all the functions to be optimized, regardless of their features, of the number of dimensions of the search-space, and of the number of particles in the swarm. More importantly, these stopping criteria enable the algorithm to be applied to real-world problems. Therefore, the same 30 optimizers considered along section 6.4 are brought into this chapter with only few slight modifications on some of them and with the stopping criteria incorporated, and they are tested again on the same suite of benchmark functions.

## 8.1 Introduction

The influence that the inertia, the individuality, and the sociality weights—as well as the velocity constraint—have on the behaviour of the swarm was studied along **Chapter 6**. However, every run of every experiment was performed along a fixed number of time-steps due to the lack of termination conditions of the plain B-PSO. Thus, the evolution of the mean[1] best and mean average[2] solutions, and the goodness of the best solution found by the end of the search, were studied and compared for different tunings of the parameters.

However, another very important aspect to consider is the celerity of the optimizers in finding an acceptable solution, which leads to lower computational costs. This is extremely important because the greatest disadvantage of population-based methods with respect to traditional methods is their higher computational costs, which might be critical for problems where the function to be optimized is expensive (e.g. finite element models). Therefore, some stopping criteria were developed along **Chapter 7**, allowing the analysis of the speed of convergence.

[1] Because of the probabilistic nature of the algorithm, the mean among fifty runs was considered in the analyses.
[2] The average conflict is computed among the current conflicts corresponding to all the particles in the population.

Hence the 30 optimizers tested along **Chapter 6** are brought into this chapter; some of them are slightly modified; the stopping criteria are incorporated; and the optimizers are once again tested on the same suite of benchmark functions (refer to **Table 6.1**). Note that the permissible absolute errors stated in **Table 6.1** are no longer applicable.

# 8.2 Experiments

## 8.2.1 Settings

The general settings are kept the same as those of the experiments run along section **6.4**:

- Number of runs per experiment: 50
- $v_{\max} = 0.5 \cdot \left( x_{\max} - x_{\min} \right)$
- Number of particles: 30
- $t_{\max} = 10000$

The settings of the 30 optimizers that are to be tested are now as follows:

**Acceleration weight unrelated to the inertia weight**

- BSt-PSO: basic, standard PSO:
  $w^{(t)} = 0.7, \;\; iw^{(t)} = sw^{(t)} = 2 \;\; \forall t$
- BSw-PSO: basic PSO with linearly time-swapping learning weights:
  $w^{(t)} = 0.7 \;\; \forall t \,,\; iw^{(1)} = 3 \,,\; sw^{(1)} = 1 \,,\; iw^{(t)} = sw^{(t)} = 2 \;\; \forall t \geq 0.1 \cdot t_{\max}$
- BStLd-PSO 1: basic, standard PSO with linearly time-decreasing inertia weight:
  $w^{(1)} = 0.9 \,,\; w^{(t_{\max})} = 0 \,,\; iw^{(t)} = sw^{(t)} = 2 \;\; \forall t$
- BStLd-PSO 2: basic, standard PSO with linearly time-decreasing inertia weight:
  $w^{(1)} = 0.9 \,,\; w^{(t_{\max})} = 0.4 \,,\; iw^{(t)} = sw^{(t)} = 2 \;\; \forall t$
- BStSd-PSO 1: basic, standard PSO with sigmoidly time-decreasing inertia weight:
  $w^{(1)} \to 0.8 \,,\; w^{(t_{\max})} \to 0 \,,\; iw^{(t)} = sw^{(t)} = 2 \;\; \forall t$

- BStSd-PSO 2: basic, standard PSO with sigmoidly time-decreasing inertia weight:
  $w^{(1)} \to 0.7\,,\; w^{(t_{\max})} \to 0.4\,,\; iw^{(t)} = sw^{(t)} = 2 \quad \forall t$
- BSwLd-PSO 1: basic PSO with linearly time-swapping learning weights and linearly time-decreasing inertia weight:
  $w^{(1)} = 0.9\,,\; w^{(t_{\max})} = 0\,,\; iw^{(1)} = 3\,,\; sw^{(1)} = 1\,,\; iw^{(t)} = sw^{(t)} = 2 \quad \forall t \geq 0.1 \cdot t_{\max}$
- BSwLd-PSO 2: basic PSO with linearly time-swapping learning weights and linearly time-decreasing inertia weight:
  $w^{(1)} = 0.9\,,\; w^{(t_{\max})} = 0.4\,,\; iw^{(1)} = 3\,,\; sw^{(1)} = 1\,,\; iw^{(t)} = sw^{(t)} = 2 \quad \forall t \geq 0.1 \cdot t_{\max}$
- BSwSd-PSO 1: basic PSO with linearly time-swapping learning weights and sigmoidly time-decreasing inertia weight:
  $w^{(1)} \to 0.8\,,\; w^{(t_{\max})} \to 0\,,\; iw^{(1)} = 3\,,\; sw^{(1)} = 1\,,\; iw^{(t)} = sw^{(t)} = 2 \quad \forall t \geq 0.1 \cdot t_{\max}$
- BSwSd-PSO 2: basic PSO with linearly time-swapping learning weights and sigmoidly time-decreasing inertia weight:
  $w^{(1)} \to 0.7\,,\; w^{(t_{\max})} \to 0.4\,,\; iw^{(1)} = 3\,,\; sw^{(1)} = 1\,,\; iw^{(t)} = sw^{(t)} = 2 \quad \forall t \geq 0.1 \cdot t_{\max}$

**Acceleration and inertia weights related like a constant**

- BSt-PSO$^{(c)}$: basic, standard PSO with $aw^{(t)} = 4.1 \cdot w^{(t)} \quad \forall t$ :
  $w^{(t)} = 0.7298,\;\; iw^{(t)} = sw^{(t)} = 1.49609 \quad \forall t$
- BSw-PSO$^{(c)}$: basic PSO with $aw^{(t)} = 4.1 \cdot w^{(t)} \quad \forall t$ and linearly time-swapping learning weights:
  $$w^{(t)} = 0.7298 \quad \forall t\,,\; r^{(1)} = \frac{sw^{(1)}}{iw^{(1)}} = \frac{1}{3}\,,\; r^{(t)} = \frac{sw^{(t)}}{iw^{(t)}} = 1 \quad \forall t \geq 0.1 \cdot t_{\max}$$
- BStLd-PSO 1$^{(c)}$: basic, standard PSO with $aw^{(t)} = 4.1 \cdot w^{(t)} \quad \forall t$ and linearly time-decreasing inertia weight:
  $w^{(1)} = 0.7298\,,\; w^{(t_{\max})} = 0\,,\; iw^{(t)} = sw^{(t)} \quad \forall t$
- BStLd-PSO 2$^{(c)}$: basic, standard PSO with $aw^{(t)} = 4.1 \cdot w^{(t)} \quad \forall t$ and linearly time-decreasing inertia weight:
  $w^{(1)} = 0.7298\,,\; w^{(t_{\max})} = 0.4\,,\; iw^{(t)} = sw^{(t)} \quad \forall t$
- BStSd-PSO 1$^{(c)}$: basic, standard PSO with $aw^{(t)} = 4.1 \cdot w^{(t)} \quad \forall t$ and sigmoidly time-decreasing inertia weight:
  $w^{(1)} \to 0.7298\,,\; w^{(t_{\max})} \to 0\,,\; iw^{(t)} = sw^{(t)} \quad \forall t$
- BStSd-PSO 2$^{(c)}$: basic, standard PSO with $aw^{(t)} = 4.1 \cdot w^{(t)} \quad \forall t$ and sigmoidly time-decreasing inertia weight:
  $w^{(1)} \to 0.7298\,,\; w^{(t_{\max})} \to 0.4\,,\; iw^{(t)} = sw^{(t)} \quad \forall t$
- BSwLd-PSO 1$^{(c)}$: basic PSO with $aw^{(t)} = 4.1 \cdot w^{(t)} \quad \forall t$, linearly time-decreasing inertia weight, and linearly time-swapping learning weights:

$$w^{(1)} = 0.7298\,,\ w^{(t_{\max})} = 0\,,\ r^{(1)} = \frac{sw^{(1)}}{iw^{(1)}} = \frac{1}{3}\,,\ r^{(t)} = \frac{sw^{(t)}}{iw^{(t)}} = 1 \quad \forall t \geq 0.1 \cdot t_{\max}$$

- BSwLd-PSO 2$^{(c)}$: basic PSO with $aw^{(t)} = 4.1 \cdot w^{(t)} \quad \forall t$, linearly time-decreasing inertia weight, and linearly time-swapping learning weights:

$$w^{(1)} = 0.7298\,,\ w^{(t_{\max})} = 0.4\,,\ r^{(1)} = \frac{sw^{(1)}}{iw^{(1)}} = \frac{1}{3}\,,\ r^{(t)} = \frac{sw^{(t)}}{iw^{(t)}} = 1 \quad \forall t \geq 0.1 \cdot t_{\max}$$

- BSwSd-PSO 1$^{(c)}$: basic PSO with $aw^{(t)} = 4.1 \cdot w^{(t)} \quad \forall t$, sigmoidly time-decreasing inertia weight, and linearly time-swapping learning weights:

$$w^{(1)} \to 0.7298\,,\ w^{(t_{\max})} \to 0\,,\ r^{(1)} = \frac{sw^{(1)}}{iw^{(1)}} = \frac{1}{3}\,,\ r^{(t)} = \frac{sw^{(t)}}{iw^{(t)}} = 1 \quad \forall t \geq 0.1 \cdot t_{\max}$$

- BSwSd-PSO 2$^{(c)}$: basic PSO with $aw^{(t)} = 4.1 \cdot w^{(t)} \quad \forall t$, sigmoidly time-decreasing inertia weight, and linearly time-swapping learning weights:

$$w^{(1)} \to 0.7298\,,\ w^{(t_{\max})} \to 0.4\,,\ r^{(1)} = \frac{sw^{(1)}}{iw^{(1)}} = \frac{1}{3}\,,$$

$$r^{(t)} = \frac{sw^{(t)}}{iw^{(t)}} = 1 \quad \forall t \geq 0.1 \cdot t_{\max}$$

**Acceleration and inertia weights related like a fourth-degree polynomial**

- BSt-PSO$^{(p)}$: basic, standard PSO with $aw^{(t)} = p\left(w^{(t)}\right) \quad \forall t$:

$$w^{(t)} = 0.5, \quad iw^{(t)} = sw^{(t)} = 2 \quad \forall t$$

- BSw-PSO$^{(p)}$: basic PSO with $aw^{(t)} = p\left(w^{(t)}\right) \quad \forall t$ and linearly time-swapping learning weights:

$$w^{(t)} = 0.5 \quad \forall t\,,\ r^{(1)} = \frac{sw^{(1)}}{iw^{(1)}} = \frac{1}{3}\,,\ r^{(t)} = \frac{sw^{(t)}}{iw^{(t)}} = 1 \quad \forall t \geq 0.1 \cdot t_{\max}$$

- BStLd-PSO 1$^{(p)}$: basic, standard PSO with $aw^{(t)} = p\left(w^{(t)}\right) \quad \forall t$ and linearly time-decreasing inertia weight:

$$w^{(1)} = 0.5\,,\ w^{(t_{\max})} = 0\,,\ iw^{(t)} = sw^{(t)} \quad \forall t$$

- BStLd-PSO 2$^{(p)}$: basic, standard PSO with $aw^{(t)} = p\left(w^{(t)}\right) \quad \forall t$ and linearly time-decreasing inertia weight:

$$w^{(1)} = 0.5\,,\ w^{(t_{\max})} = 0.4\,,\ iw^{(t)} = sw^{(t)} \quad \forall t$$

- BStSd-PSO 1$^{(p)}$: basic, standard PSO with $aw^{(t)} = p\left(w^{(t)}\right) \quad \forall t$ and sigmoidly time-decreasing inertia weight:

$$w^{(1)} \to 0.5\,,\ w^{(t_{\max})} \to 0\,,\ iw^{(t)} = sw^{(t)} \quad \forall t$$

- BStSd-PSO 2$^{(p)}$: basic, standard PSO with $aw^{(t)} = p\left(w^{(t)}\right) \quad \forall t$ and sigmoidly time-decreasing inertia weight:

$$w^{(1)} \to 0.5\,,\ w^{(t_{\max})} \to 0.4\,,\ iw^{(t)} = sw^{(t)} \quad \forall t$$

- BSwLd-PSO 1(p): basic PSO with $aw^{(t)} = p\left(w^{(t)}\right)\ \forall t$, linearly time-decreasing inertia weight, and linearly time-swapping learning weights:

  $$w^{(1)} = 0.5,\ w^{(t_{\max})} = 0,\ r^{(1)} = \frac{sw^{(1)}}{iw^{(1)}} = \frac{1}{3},\ r^{(t)} = \frac{sw^{(t)}}{iw^{(t)}} = 1\quad \forall t \geq 0.1 \cdot t_{\max}$$

- BSwLd-PSO 2(p): basic PSO with $aw^{(t)} = p\left(w^{(t)}\right)\ \forall t$, linearly time-decreasing inertia weight, and linearly time-swapping learning weights:

  $$w^{(1)} = 0.5,\ w^{(t_{\max})} = 0.4,\ r^{(1)} = \frac{sw^{(1)}}{iw^{(1)}} = \frac{1}{3},\ r^{(t)} = \frac{sw^{(t)}}{iw^{(t)}} = 1\quad \forall t \geq 0.1 \cdot t_{\max}$$

- BSwSd-PSO 1(p): basic PSO with $aw^{(t)} = p\left(w^{(t)}\right)\ \forall t$, sigmoidly time-decreasing inertia weight, and linearly time-swapping learning weights:

  $$w^{(1)} \to 0.5,\ w^{(t_{\max})} \to 0,\ r^{(1)} = \frac{sw^{(1)}}{iw^{(1)}} = \frac{1}{3},\ r^{(t)} = \frac{sw^{(t)}}{iw^{(t)}} = 1\quad \forall t \geq 0.1 \cdot t_{\max}$$

- BSwSd-PSO 2(p): basic PSO with $aw^{(t)} = p\left(w^{(t)}\right)\ \forall t$, sigmoidly time-decreasing inertia weight, and linearly time-swapping learning weights:

  $$w^{(1)} \to 0.5,\ w^{(t_{\max})} \to 0.4,\ r^{(1)} = \frac{sw^{(1)}}{iw^{(1)}} = \frac{1}{3},\ r^{(t)} = \frac{sw^{(t)}}{iw^{(t)}} = 1\quad \forall t \geq 0.1 \cdot t_{\max}$$

As previously mentioned, the stopping criteria were developed for optimizers with the ability to fine-cluster[3], although the second set of termination conditions leaves the door open for their application to optimizers without such ability[4]. In order to make the second set of error conditions more demanding, the latter was modified for the experiments run hereafter from $t > 0.25 \cdot t_{\max} \wedge cgbest^{(t-0.25 \cdot t_{\max})} - cgbest^{(t)} = 0$ to $t > 0.35 \cdot t_{\max} \wedge cgbest^{(t-0.35 \cdot t_{\max})} - cgbest^{(t)} = 0$.

The swapping strategy was proposed in **Chapter 6** to favour exploration during the early stages of the search, and exploitation (i.e. fine-clustering) afterwards. However, enhancing the exploitation ability does not appear to be necessary for optimizers which already exhibit the ability to fine-cluster. Therefore, the swapping strategy is modified so that the particles exhibit higher individuality at the beginning, while relatively increasing the sociality until $r = {iw}/{sw} = 1$ from the time-step $t = 0.1 \cdot t_{\max}$ forth, so that the stopping criteria guarantee that the search cannot be terminated while the ratio $r > 1$. Therefore, exploration is expected to be favoured during the first 10% of the maximum number of time-steps permitted for the search to go through, while the learning weights are guaranteed to be equal to one another by the time the search is terminated.

[3] The first 10 optimizers, whose inertia and acceleration weights are kept unrelated, are typically not among them.

[4] The second set of termination conditions takes into account the incapability of the optimizer to achieve further improvement, without considering the degree of clustering achieved by its particles.

The optimizers whose inertia and acceleration weights are unrelated are the same as those tested along section **6.4**, except for the incorporated stopping criteria and the modified swapping strategy. In contrast, the upper limits of the inertia weight of the optimizers whose inertia and acceleration weights are related like $aw^{(t)} = 4.1 \cdot w^{(t)}$ were set to 0.7298 so that they are equivalent to purely constricted PSOs with time-decreasing constriction factors. Likewise, the upper limits of the inertia weight of the optimizers whose inertia and acceleration weights are related like $aw^{(t)} = p\left(w^{(t)}\right)$ were set to 0.5, so that the acceleration weight is never greater than 4 (recall that greater acceleration weights strengthen the influence of the stochastic weights, thus making the clustering of the particles more difficult).

For the sake of visualization of the modified swapping strategy and upper limits of the inertia weights, the curves of the evolution of the parameters of the BSwLd-PSO 1 and of the BSwSd-PSO 1 are shown in **Fig. 8. 1** (compare it to **Fig. 6. 26**); the curves corresponding to the BSwLd-PSO 1$^{(c)}$ and to the BSwSd-PSO 1$^{(c)}$ are shown in **Fig. 8. 2** (compare it to **Fig. 6. 37**), while those corresponding to the BSwLd-PSO 1$^{(p)}$ and to the BSwSd-PSO 1$^{(p)}$ are shown in **Fig. 8. 3** (compare it to **Fig. 6. 47**):

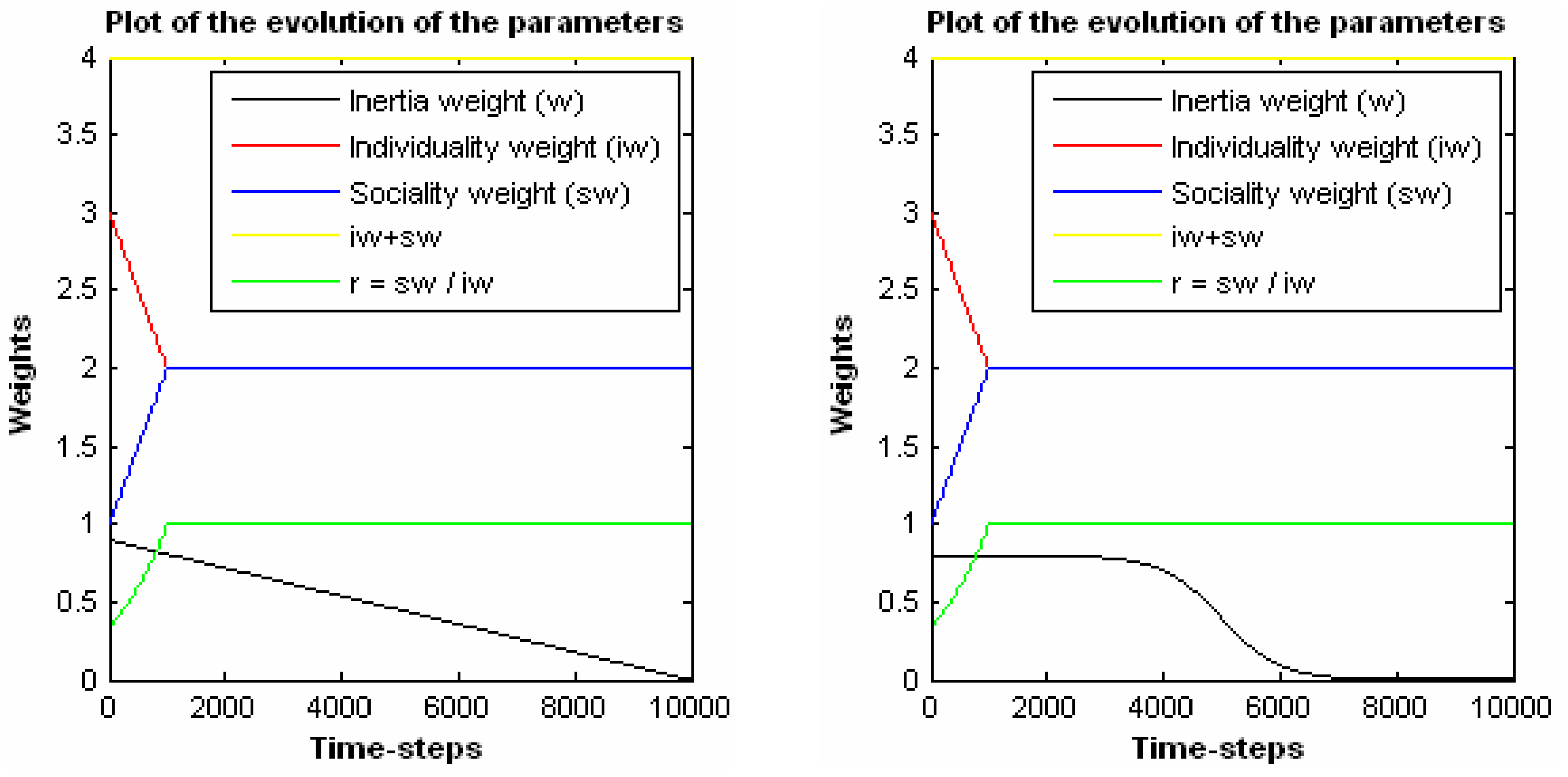

**Fig. 8. 1**: Evolution of the weights for the BSwLd-PSO 1 (left) and the BSwSd-PSO 1 (right).

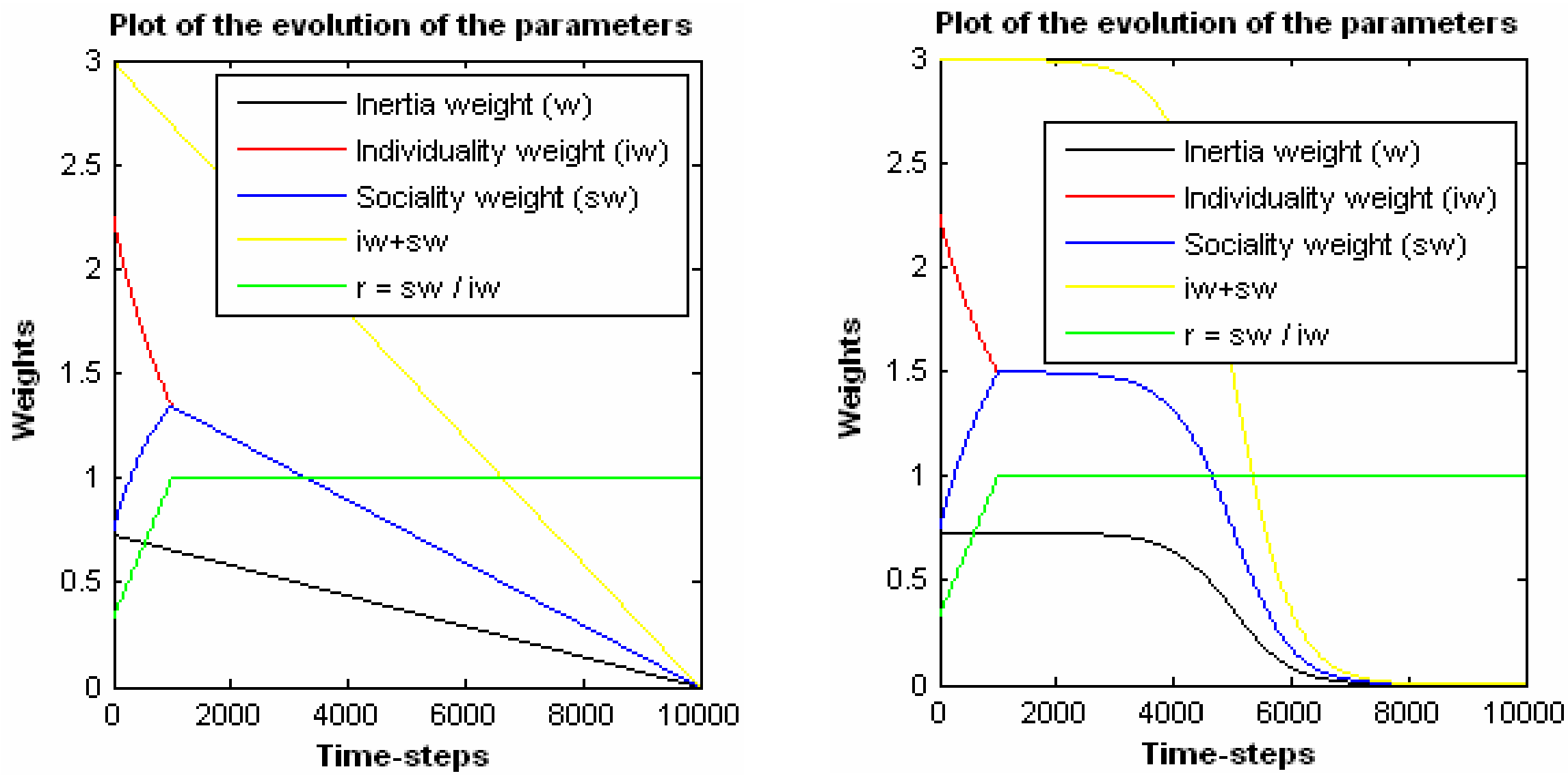

**Fig. 8. 2**: Evolution of the weights for the BSwLd-PSO 1$^{(c)}$ (left) and the BSwSd-PSO 1$^{(c)}$ (right).

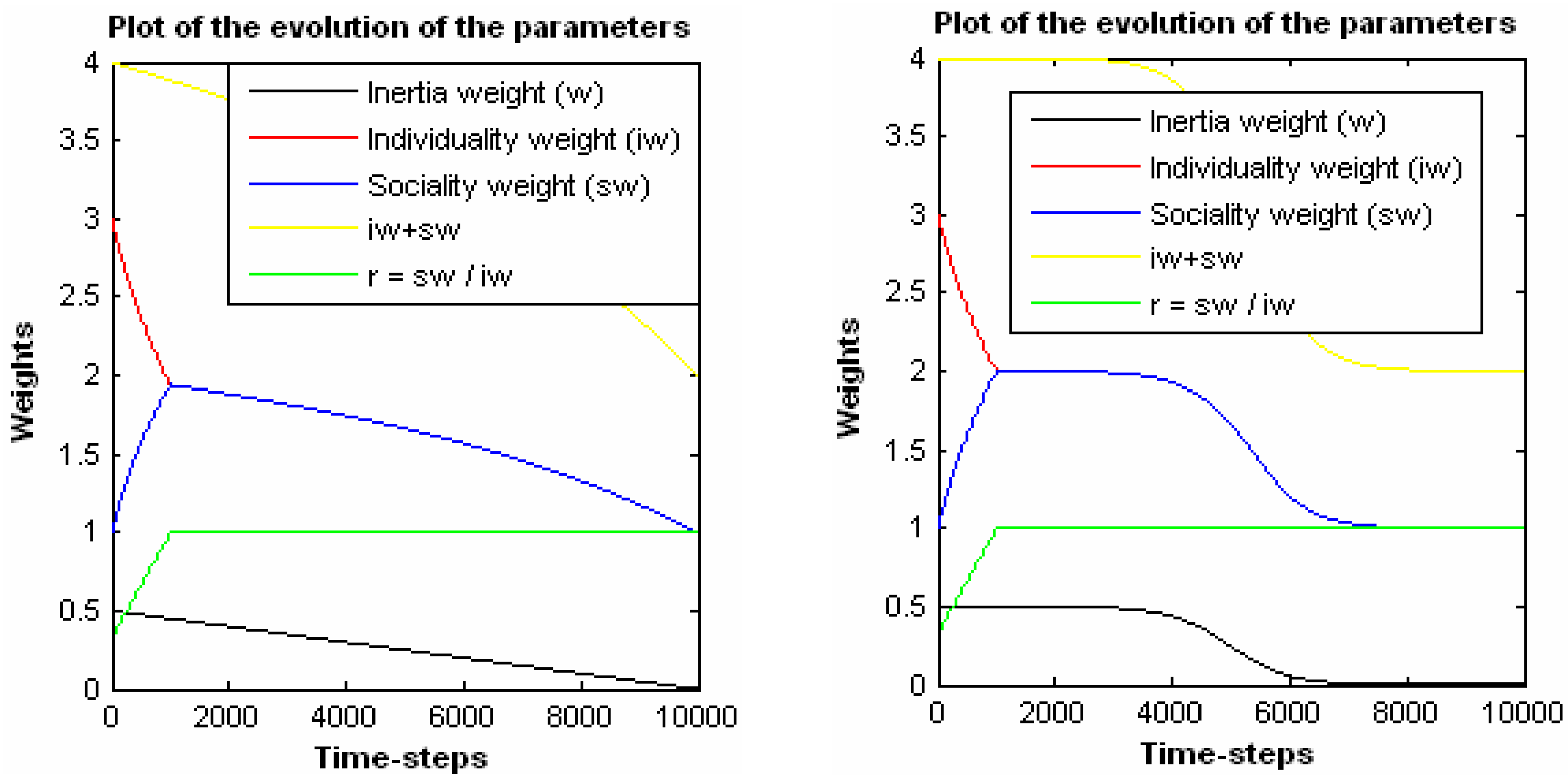

**Fig. 8. 3**: Evolution of the weights for the BSwLd-PSO 1$^{(p)}$ (left) and the BSwSd-PSO 1$^{(p)}$ (right).

## 8.2.2 Experimental results

The most significant experimental results are gathered in **Table 8. 1** to **Table 8. 6**. While only the average results are shown in these tables, the results obtained from each of the 50 runs per experiment, together with the standard deviations, can be found in digital **Appendix 4**.

| SPHERE | | | | | |
|---|---|---|---|---|---|
| OPTIMIZER | Mean best solution found when the error condition is attained | Mean time-steps required to attain the error condition | Mean type of error condition attained | Number of failures in attaining the error condition | Mean best solution found when the error condition is not attained |
| BSt-PSO | - | - | - | 50 | 4.2707E-08 |
| BSw-PSO | - | - | - | 50 | 2.2160E-08 |
| BStLd-PSO 1 | 3.4671E-18 | 4715.10 | 1.00 | 0 | - |
| BStLd-PSO 2 | 1.3092E-17 | 7217.00 | 1.00 | 0 | - |
| BStSd-PSO 1 | 4.6928E-10 | 5503.90 | 1.00 | 0 | - |
| BStSd-PSO 2 | 3.0130E-18 | 5663.24 | 1.00 | 0 | - |
| BSwLd-PSO 1 | 3.7899E-18 | 4721.22 | 1.00 | 0 | - |
| BSwLd-PSO 2 | 1.2046E-17 | 7188.70 | 1.00 | 0 | - |
| BSwSd-PSO 1 | 2.2646E-13 | 5512.72 | 1.00 | 0 | - |
| BSwSd-PSO 2 | 3.5791E-18 | 5648.36 | 1.00 | 0 | - |
| **BSt-PSO(c)** | **9.8522E-18** | **1305.60** | **1.00** | **0** | **-** |
| **BSw-PSO(c)** | **3.8899E-18** | **1455.98** | **1.00** | **0** | **-** |
| BStLd-PSO 1(c) | 1.5259E-05 | 1269.04 | 1.00 | 0 | - |
| BStLd-PSO 2(c) | 2.8142E-11 | 1351.68 | 1.00 | 0 | - |
| **BStSd-PSO 1(c)** | **4.2158E-18** | **1287.22** | **1.00** | **0** | **-** |
| BStSd-PSO 2(c) | 1.8450E-17 | 1302.38 | 1.00 | 0 | - |
| BSwLd-PSO 1(c) | 7.3911E-01 | 1217.82 | 1.00 | 0 | - |
| BSwLd-PSO 2(c) | 7.3857E-02 | 1842.76 | 1.00 | 0 | - |
| **BSwSd-PSO 1(c)** | **3.7159E-18** | **1579.18** | **1.00** | **0** | **-** |
| **BSwSd-PSO 2(c)** | **3.5114E-18** | **1447.14** | **1.00** | **0** | **-** |
| **BSt-PSO(p)** | **5.5345E-18** | **1542.04** | **1.00** | **0** | **-** |
| BSw-PSO(p) | 1.4495E+01 | 1591.54 | 1.00 | 0 | - |
| **BStLd-PSO 1(p)** | **1.5278E-18** | **1243.74** | **1.00** | **0** | **-** |
| **BStLd-PSO 2(p)** | **3.2041E-18** | **1458.84** | **1.00** | **0** | **-** |
| **BStSd-PSO 1(p)** | **5.2442E-18** | **1552.40** | **1.00** | **0** | **-** |
| **BStSd-PSO 2(p)** | **5.0219E-18** | **1548.12** | **1.00** | **0** | **-** |
| BSwLd-PSO 1(p) | 3.3189E+01 | 1014.02 | 1.00 | 0 | - |
| BSwLd-PSO 2(p) | 1.5926E+01 | 1205.68 | 1.00 | 0 | - |
| BSwSd-PSO 1(p) | 1.1183E+01 | 1699.96 | 1.00 | 0 | - |
| BSwSd-PSO 2(p) | 2.5030E+01 | 1626.92 | 1.00 | 0 | - |

**Table 8. 1**: Results obtained from the optimization of the 30-dimensional Sphere function by means of 30 optimizers, where the mean values are calculated out of 50 runs, and the maximum number of time-steps permitted for the search was set to 10000. Given that the first set of termination conditions is referred to as “type 1” and the second one as “type 2”, the “mean type of error condition attained” gives a precise idea of the number of times that the each set of error conditions was attained, which in turn gives an idea of the average degree of clustering achieved. The optimizers in bold are those which perform best with regards to the goodness of—and celerity in finding—the best solutions to the Sphere function. The shaded ones are those performing best when optimizing the complete suite of benchmark functions, or at least showing desirable features for the development of general-purpose optimizers.

| ROSENBROCK | | | | | |
|---|---|---|---|---|---|
| OPTIMIZER | Mean best solution found when the error condition is attained | Mean time-steps required to attain the error condition | Mean type of error condition attained | Number of failures in attaining the error condition | Mean best solution found when the error condition is not attained |
| BSt-PSO | - | - | - | 50 | 8.0383E+01 |
| BSw-PSO | - | - | - | 50 | 7.9745E+01 |
| BStLd-PSO 1 | 5.6191E+01 | 9718.55 | 2.00 | 39 | 3.0771E+01 |
| BStLd-PSO 2 | - | - | - | 50 | 3.5882E+01 |
| BStSd-PSO 1 | 5.2511E+01 | 9457.88 | 1.96 | 26 | 4.9872E+01 |
| BStSd-PSO 2 | - | - | - | 50 | 2.9460E+01 |
| BSwLd-PSO 1 | 4.8183E+01 | 9719.50 | 2.00 | 46 | 5.5180E+01 |
| BSwLd-PSO 2 | - | - | - | 50 | 2.7930E+01 |
| BSwSd-PSO 1 | 4.7402E+01 | 8884.61 | 1.81 | 19 | 5.0465E+01 |
| BSwSd-PSO 2 | - | - | - | 50 | 2.9460E+01 |
| **BSt-PSO(c)** | - | - | - | 50 | 4.1958E+00 |
| **BSw-PSO(c)** | - | - | - | 50 | 2.5005E+00 |
| **BStLd-PSO 1(c)** | 5.9428E+01 | 1847.12 | 1.00 | 0 | - |
| BStLd-PSO 2(c) | 3.8114E+01 | 4659.79 | 1.17 | 8 | 5.0361E+01 |
| **BStSd-PSO 1(c)** | 1.4501E+01 | 8103.16 | 1.92 | 1 | 7.9019E+00 |
| **BStSd-PSO 2(c)** | 1.1348E+01 | 8630.77 | 1.90 | 20 | 1.6803E+01 |
| BSwLd-PSO 1(c) | 2.0543E+02 | 1327.24 | 1.00 | 0 | - |
| BSwLd-PSO 2(c) | 1.3111E+02 | 2239.36 | 1.00 | 0 | - |
| BSwSd-PSO 1(c) | 2.2941E+01 | 7717.70 | 1.78 | 0 | - |
| **BSwSd-PSO 2(c)** | 1.4010E+01 | 8383.31 | 1.79 | 21 | 1.5473E+01 |
| **BSt-PSO(p)** | 5.9050E+01 | 8919.20 | 2.00 | 45 | 1.8097E+01 |
| BSw-PSO(p) | 3.3882E+02 | 4530.44 | 1.11 | 41 | 2.7748E+01 |
| BStLd-PSO 1(p) | 4.0298E+01 | 7542.53 | 1.08 | 12 | 4.3190E+01 |
| BStLd-PSO 2(p) | 3.3340E+01 | 8197.09 | 2.00 | 18 | 2.1920E+01 |
| **BStSd-PSO 1(p)** | 2.5944E+01 | 6989.93 | 1.00 | 35 | 3.8515E+01 |
| BStSd-PSO 2(p) | 1.9604E+01 | 9255.89 | 2.00 | 31 | 2.8983E+01 |
| **BSwLd-PSO 1(p)** | 2.7327E+02 | 2456.11 | 1.00 | 6 | 2.8312E+01 |
| BSwLd-PSO 2(p) | 2.9940E+02 | 4667.40 | 1.40 | 15 | 2.2058E+01 |
| BSwSd-PSO 1(p) | 2.2814E+02 | 6225.42 | 1.13 | 26 | 2.8715E+01 |
| BSwSd-PSO 2(p) | 2.3152E+02 | 6745.27 | 1.55 | 28 | 2.9654E+01 |

**Table 8. 2**: Results obtained from the optimization of the 30-dimensional Rosenbrock function by means of 30 optimizers, where the mean values are calculated out of 50 runs, and the maximum number of time-steps permitted for the search was set to 10000. Given that the first set of termination conditions is referred to as “type 1” and the second one as “type 2”, the “mean type of error condition attained” gives a precise idea of the number of times that the each set of error conditions was attained, which in turn gives an idea of the average degree of clustering achieved. The optimizers in bold are those which perform best with regards to the goodness of—and celerity in finding—the best solutions to the Rosenbrock function. The shaded ones are those performing best when optimizing the complete suite of benchmark functions, or at least showing desirable features for the development of general-purpose optimizers.

| RASTRIGRIN | | | | | |
|---|---|---|---|---|---|
| OPTIMIZER | Mean best solution found when the error condition is attained | Mean time-steps required to attain the error condition | Mean type of error condition attained | Number of failures in attaining the error condition | Mean best solution found when the error condition is not attained |
| BSt-PSO | - | - | - | 50 | 2.1192E+01 |
| BSw-PSO | - | - | - | 50 | 1.9973E+01 |
| BStLd-PSO 1 | 2.7327E+01 | 7104.35 | 1.28 | 7 | 2.4590E+01 |
| BStLd-PSO 2 | 2.3114E+01 | 8673.62 | 1.00 | 37 | 2.6487E+01 |
| BStSd-PSO 1 | 3.9678E+01 | 6498.92 | 1.18 | 11 | 3.1115E+01 |
| BStSd-PSO 2 | 2.4874E+01 | 8556.38 | 1.29 | 29 | 2.4359E+01 |
| BSwLd-PSO 1 | 2.8235E+01 | 7095.71 | 1.36 | 5 | 2.8257E+01 |
| BSwLd-PSO 2 | 2.5007E+01 | 8752.40 | 1.00 | 35 | 2.5985E+01 |
| BSwSd-PSO 1 | 3.3232E+01 | 6550.97 | 1.14 | 15 | 3.5024E+01 |
| BSwSd-PSO 2 | 1.8842E+01 | 8139.94 | 1.13 | 34 | 2.0836E+01 |
| BSt-PSO(c) | 6.2782E+01 | 1289.52 | 1.00 | 0 | - |
| BSw-PSO(c) | 4.3002E+01 | 1427.88 | 1.00 | 0 | - |
| BStLd-PSO 1(c) | 6.1588E+01 | 1177.06 | 1.00 | 0 | - |
| BStLd-PSO 2(c) | 6.4971E+01 | 1300.16 | 1.00 | 0 | - |
| BStSd-PSO 1(c) | 6.2105E+01 | 1320.76 | 1.00 | 0 | - |
| BStSd-PSO 2(c) | 6.0653E+01 | 1318.28 | 1.00 | 0 | - |
| BSwLd-PSO 1(c) | 4.1536E+01 | 1228.12 | 1.00 | 0 | - |
| BSwLd-PSO 2(c) | 4.4356E+01 | 1650.24 | 1.00 | 0 | - |
| BSwSd-PSO 1(c) | 4.5191E+01 | 1456.78 | 1.00 | 0 | - |
| BSwSd-PSO 2(c) | 4.1709E+01 | 1545.08 | 1.00 | 0 | - |
| BSt-PSO(p) | 4.6843E+01 | 1940.26 | 1.00 | 0 | - |
| BSw-PSO(p) | 3.5699E+01 | 2462.84 | 1.08 | 0 | - |
| BStLd-PSO 1(p) | 4.7221E+01 | 1487.74 | 1.02 | 0 | - |
| BStLd-PSO 2(p) | 4.7579E+01 | 1802.88 | 1.00 | 0 | - |
| BStSd-PSO 1(p) | 4.6425E+01 | 2057.74 | 1.00 | 0 | - |
| BStSd-PSO 2(p) | 4.1549E+01 | 2140.80 | 1.02 | 0 | - |
| BSwLd-PSO 1(p) | 3.9500E+01 | 1666.00 | 1.00 | 0 | - |
| BSwLd-PSO 2(p) | 3.5580E+01 | 2315.28 | 1.04 | 0 | - |
| BSwSd-PSO 1(p) | 3.5003E+01 | 2351.40 | 1.00 | 0 | - |
| BSwSd-PSO 2(p) | 3.6455E+01 | 2593.84 | 1.06 | 0 | - |

**Table 8. 3**: Results obtained from the optimization of the 30-dimensional Rastrigrin function by means of 30 optimizers, where the mean values are calculated out of 50 runs, and the maximum number of time-steps permitted for the search was set to 10000. Given that the first set of termination conditions is referred to as “type 1” and the second one as “type 2”, the “mean type of error condition attained” gives a precise idea of the number of times that the each set of error conditions was attained, which in turn gives an idea of the average degree of clustering achieved. The optimizers in bold are those which perform best regarding the goodness of—and celerity in finding—the best solutions to the Rastrigrin function. The shaded ones are those performing best when optimizing the complete suite of benchmark functions, or at least showing desirable features for the development of general-purpose optimizers.

| GRIEWANK | | | | | |
|---|---|---|---|---|---|
| OPTIMIZER | Mean best solution found when the error condition is attained | Mean time-steps required to attain the error condition | Mean type of error condition attained | Number of failures in attaining the error condition | Mean best solution found when the error condition is not attained |
| BSt-PSO | - | - | - | 50 | 1.7019E-02 |
| BSw-PSO | - | - | - | 50 | 2.2362E-02 |
| **BStLd-PSO 1** | **1.4740E-02** | **4790.70** | **1.00** | **0** | **-** |
| BStLd-PSO 2 | 1.9486E-02 | 7323.14 | 1.00 | 0 | - |
| **BStSd-PSO 1** | **1.5545E-02** | **5578.46** | **1.00** | **0** | **-** |
| BStSd-PSO 2 | 2.4758E-02 | 6030.29 | 1.02 | 1 | 9.8573E-03 |
| **BSwLd-PSO 1** | **1.9944E-02** | **4804.00** | **1.00** | **0** | **-** |
| **BSwLd-PSO 2** | **1.6326E-02** | **7249.48** | **1.00** | **0** | **-** |
| **BSwSd-PSO 1** | **1.7535E-02** | **5648.72** | **1.02** | **0** | **-** |
| **BSwSd-PSO 2** | **2.1620E-02** | **5780.22** | **1.00** | **0** | **-** |
| BSt-PSO(c) | 2.6448E-02 | 1236.30 | 1.00 | 0 | - |
| BSw-PSO(c) | 5.0368E-02 | 1200.92 | 1.00 | 0 | - |
| BStLd-PSO 1(c) | 5.1657E-02 | 1216.76 | 1.00 | 0 | - |
| BStLd-PSO 2(c) | 4.9886E-02 | 1275.00 | 1.00 | 0 | - |
| BStSd-PSO 1(c) | 3.7151E-02 | 1280.92 | 1.00 | 0 | - |
| BStSd-PSO 2(c) | 3.4774E-02 | 1255.06 | 1.00 | 0 | - |
| BSwLd-PSO 1(c) | 2.1495E-01 | 1173.88 | 1.00 | 0 | - |
| BSwLd-PSO 2(c) | 1.0093E-01 | 1788.24 | 1.00 | 0 | - |
| BSwSd-PSO 1(c) | 4.6468E-02 | 1423.52 | 1.00 | 0 | - |
| BSwSd-PSO 2(c) | 3.4098E-02 | 1390.24 | 1.00 | 0 | - |
| **BSt-PSO(p)** | **1.4443E-02** | **1576.96** | **1.00** | **0** | **-** |
| BSw-PSO(p) | 5.1224E-01 | 1792.36 | 1.00 | 0 | - |
| **BStLd-PSO 1(p)** | **1.3819E-02** | **1231.46** | **1.00** | **0** | **-** |
| **BStLd-PSO 2(p)** | **1.3465E-02** | **1444.28** | **1.00** | **0** | **-** |
| **BStSd-PSO 1(p)** | **1.2833E-02** | **1582.74** | **1.00** | **0** | **-** |
| **BStSd-PSO 2(p)** | **9.7875E-03** | **1532.66** | **1.00** | **0** | **-** |
| BSwLd-PSO 1(p) | 8.0585E-01 | 1018.14 | 1.00 | 0 | - |
| BSwLd-PSO 2(p) | 5.2271E-01 | 1330.34 | 1.00 | 0 | - |
| BSwSd-PSO 1(p) | 4.9152E-01 | 1572.10 | 1.00 | 0 | - |
| BSwSd-PSO 2(p) | 6.3962E-01 | 1586.42 | 1.00 | 0 | - |

**Table 8. 4**: Results obtained from the optimization of the 30-dimensional Griewank function by means of 30 optimizers, where the mean values are calculated out of 50 runs, and the maximum number of time-steps permitted for the search was set to 10000. Given that the first set of termination conditions is referred to as “type 1” and the second one as “type 2”, the “mean type of error condition attained” gives a precise idea of the number of times that the each set of error conditions was attained, which in turn gives an idea of the average degree of clustering achieved. The optimizers in bold are those which perform best regarding the goodness of—and celerity in finding—the best solutions to the Griewank function. The shaded ones are those performing best when optimizing the complete suite of benchmark functions, or at least showing desirable features for the development of general-purpose optimizers.

| SCHAFFER F6 2D | | | | | |
|---|---|---|---|---|---|
| OPTIMIZER | Mean best solution found when the error condition is attained | Mean time-steps required to attain the error condition | Mean type of error condition attained | Number of failures in attaining the error condition | Mean best solution found when the error condition is not attained |
| BSt-PSO | 1.9432E-04 | 4035.90 | 1.84 | 0 | - |
| **BSw-PSO** | **0.0000E+00** | **4141.94** | **1.82** | **0** | **-** |
| **BStLd-PSO 1** | **0.0000E+00** | **4903.62** | **1.42** | **0** | **-** |
| **BStLd-PSO 2** | **0.0000E+00** | **6065.78** | **1.74** | **0** | **-** |
| BStSd-PSO 1 | 1.9432E-04 | 5233.22 | 1.94 | 0 | - |
| **BStSd-PSO 2** | **0.0000E+00** | **4037.26** | **1.82** | **0** | **-** |
| **BSwLd-PSO 1** | **0.0000E+00** | **5051.44** | **1.52** | **0** | **-** |
| **BSwLd-PSO 2** | **0.0000E+00** | **6113.62** | **1.76** | **0** | **-** |
| **BSwSd-PSO 1** | **0.0000E+00** | **5434.62** | **1.86** | **0** | **-** |
| **BSwSd-PSO 2** | **0.0000E+00** | **4007.34** | **1.76** | **0** | **-** |
| BSt-PSO(c) | 7.7727E-04 | 2566.54 | 1.20 | 0 | - |
| **BSw-PSO(c)** | **0.0000E+00** | **2595.22** | **1.12** | **0** | **-** |
| BStLd-PSO 1(c) | 7.7727E-04 | 2430.52 | 1.20 | 0 | - |
| BStLd-PSO 2(c) | 5.8295E-04 | 2330.64 | 1.06 | 0 | - |
| BStSd-PSO 1(c) | 5.8295E-04 | 2692.94 | 1.16 | 0 | - |
| BStSd-PSO 2(c) | 1.9432E-04 | 2577.86 | 1.14 | 0 | - |
| **BSwLd-PSO 1(c)** | **0.0000E+00** | **2211.70** | **1.06** | **0** | **-** |
| BSwLd-PSO 2(c) | 3.8864E-04 | 2535.60 | 1.06 | 0 | - |
| **BSwSd-PSO 1(c)** | **0.0000E+00** | **2542.00** | **1.02** | **0** | **-** |
| **BSwSd-PSO 2(c)** | **0.0000E+00** | **2398.42** | **1.00** | **0** | **-** |
| **BSt-PSO(p)** | **0.0000E+00** | **3058.46** | **1.34** | **0** | **-** |
| BSw-PSO(p) | 5.8295E-04 | 2815.98 | 1.18 | 0 | - |
| BStLd-PSO 1(p) | 7.7727E-04 | 2687.20 | 1.14 | 0 | - |
| BStLd-PSO 2(p) | 5.8295E-04 | 2952.72 | 1.28 | 0 | - |
| BStSd-PSO 1(p) | 7.7727E-04 | 3005.88 | 1.26 | 0 | - |
| BStSd-PSO 2(p) | 5.8295E-04 | 3262.44 | 1.38 | 0 | - |
| BSwLd-PSO 1(p) | 3.8864E-04 | 2761.50 | 1.10 | 0 | - |
| BSwLd-PSO 2(p) | 1.9432E-04 | 2591.44 | 1.16 | 0 | - |
| BSwSd-PSO 1(p) | 5.8295E-04 | 3026.80 | 1.20 | 0 | - |
| **BSwSd-PSO 2(p)** | **0.0000E+00** | **3111.58** | **1.22** | **0** | **-** |

**Table 8. 5**: Results obtained from the optimization of the 2-dimensional Schaffer f6 function by means of 30 optimizers, where the mean values are calculated out of 50 runs, and the maximum number of time-steps permitted for the search was set to 10000. Given that the first set of termination conditions is referred to as "type 1" and the second one as "type 2", the "mean type of error condition attained" gives a precise idea of the number of times that the each set of error conditions was attained, which in turn gives an idea of the average degree of clustering achieved. The optimizers in bold are those which perform best with regards to the goodness of—and celerity in finding—the best solutions to the 2-dimensional Schaffer f6 function. The shaded ones are those performing best when optimizing the complete suite of benchmark functions, or at least showing desirable features for the development of general-purpose optimizers.

| SCHAFFER F6 | | | | | |
|---|---|---|---|---|---|
| OPTIMIZER | Mean best solution found when the error condition is attained | Mean time-steps required to attain the error condition | Mean type of error condition attained | Number of failures in attaining the error condition | Mean best solution found when the error condition is not attained |
| BSt-PSO | 1.7800E-01 | 8869.63 | 2.00 | 42 | 1.4153E-01 |
| BSw-PSO | 1.9471E-01 | 9160.17 | 2.00 | 44 | 1.5265E-01 |
| BStLd-PSO 1 | 1.0755E-01 | 7990.05 | 2.00 | 13 | 7.9395E-02 |
| BStLd-PSO 2 | 1.1615E-01 | 9576.22 | 2.00 | 32 | 9.5041E-02 |
| BStSd-PSO 1 | 1.2529E-01 | 8689.65 | 2.00 | 27 | 9.5242E-02 |
| BStSd-PSO 2 | 1.0009E-01 | 8780.24 | 2.00 | 16 | 7.9169E-02 |
| BSwLd-PSO 1 | 1.0030E-01 | 8033.19 | 2.00 | 14 | 8.6840E-02 |
| BSwLd-PSO 2 | 1.2731E-01 | 9421.53 | 2.00 | 35 | 8.8397E-02 |
| BSwSd-PSO 1 | 1.1893E-01 | 8960.61 | 2.00 | 27 | 9.4376E-02 |
| BSwSd-PSO 2 | 1.1502E-01 | 8776.31 | 2.00 | 18 | 8.4918E-02 |
| BSt-PSO(c) | 1.1624E-01 | 5949.25 | 2.00 | 2 | 5.7707E-02 |
| **BSw-PSO(c)** | **9.5021E-02** | **5783.17** | **2.00** | **3** | **6.4534E-02** |
| BStLd-PSO 1(c) | 1.3190E-01 | 6801.39 | 2.00 | 4 | 1.0320E-01 |
| BStLd-PSO 2(c) | 1.2413E-01 | 6779.74 | 2.00 | 8 | 9.0390E-02 |
| BStSd-PSO 1(c) | 1.0984E-01 | 6180.68 | 2.00 | 9 | 1.1762E-01 |
| BStSd-PSO 2(c) | 1.0142E-01 | 6431.79 | 2.00 | 3 | 7.8189E-02 |
| BSwLd-PSO 1(c) | 1.4265E-01 | 6557.47 | 2.00 | 7 | 1.0721E-01 |
| BSwLd-PSO 2(c) | 1.2367E-01 | 6699.65 | 2.00 | 2 | 1.0259E-01 |
| BSwSd-PSO 1(c) | 1.0661E-01 | 6691.89 | 2.00 | 5 | 7.2056E-02 |
| BSwSd-PSO 2(c) | 1.0326E-01 | 6358.61 | 2.00 | 4 | 6.7948E-02 |
| BSt-PSO(p) | 1.0218E-01 | 5884.52 | 2.00 | 0 | - |
| **BSw-PSO(p)** | **9.7431E-02** | **5964.13** | **2.00** | **3** | **7.8189E-02** |
| BStLd-PSO 1(p) | 1.0522E-01 | 6745.02 | 2.00 | 8 | 9.0390E-02 |
| **BStLd-PSO 2(p)** | **9.4201E-02** | **6034.89** | **2.00** | **3** | **8.0801E-02** |
| BStSd-PSO 1(p) | 1.0480E-01 | 6073.68 | 2.00 | 9 | 7.5379E-02 |
| **BStSd-PSO 2(p)** | **9.8793E-02** | **6166.83** | **2.00** | **4** | **4.7465E-02** |
| BSwLd-PSO 1(p) | 1.0764E-01 | 6142.06 | 2.00 | 3 | 7.8189E-02 |
| BSwLd-PSO 2(p) | 1.0722E-01 | 5817.61 | 2.00 | 1 | 7.8189E-02 |
| BSwSd-PSO 1(p) | 1.1114E-01 | 5644.19 | 2.00 | 8 | 7.3069E-02 |
| **BSwSd-PSO 2(p)** | **9.7670E-02** | **5905.08** | **2.00** | **2** | **7.8189E-02** |

**Table 8. 6**: Results obtained from the optimization of the 30-dimensional Schaffer f6 function by means of 30 optimizers, where the mean values are calculated out of 50 runs, and the maximum number of time-steps permitted for the search was set to 10000. Given that the first set of termination conditions is referred to as “type 1” and the second one as “type 2”, the “mean type of error condition attained” gives a precise idea of the number of times that the each set of error conditions was attained, which in turn gives an idea of the average degree of clustering achieved. The optimizers in bold are those which perform best with regards to the goodness of—and celerity in finding—the best solutions to the 30-dimensional Schaffer f6 function. The shaded ones are those performing best when optimizing the complete suite of benchmark functions, or at least showing desirable features for the development of general-purpose optimizers.

### 8.2.3 Discussion

As previously observed in **Chapter 6**, no optimizer exhibits both strong fine-clustering ability and strong reluctance to getting trapped in sub-optimal solutions. Therefore, none of these homogeneous[5] optimizers can be claimed to be a "good" general-purpose optimizer, except perhaps for the BSt/SwSd-PSO 2, which manage to find reasonably good solutions for every function in the test suite. However, they always take longer than half the maximum length of time permitted for the search to go through to attain the error condition, when they do so. This is because they behave similarly to the BSt/Sw-PSO ( $w^{(1)} \to 0.7,\ \ aw^{(t)} = 2\ \ \forall t$ ) throughout approximately the first third of the search, while behaving similarly to the BSt-PSO$^{(p)}$ throughout the second half of the search ( $w^{(t_{\max})} \to 0.4$ ). In fact, a possible strategy could be to modify the BSt/Sw-PSO 2 from:

$$w^{(1)} \to 0.7\,,\ w^{(t_{\max})} \to 0.4\,,\ aw^{(t)} = 2\ \ \forall t \qquad \text{to} \qquad w^{(1)} \to 0.7\,,\ w^{(t_{\max})} \to 0.5\,,\ aw^{(t)} = 2\ \ \forall t\,.$$

Thus, the optimizer would resemble a robust BSt/Sw-PSO throughout approximately the first third of the search, while resembling a BSt-PSO$^{(p)}$—which has the ability to fine-cluster—throughout approximately the last third of the search. In between, a smooth but steep decrease of the inertia weight would be performed. Nevertheless, this would still require most of the maximum number of time-steps permitted for the search to go through to find good solutions. This strategy was not implemented within this thesis due to time constraints.

Hence, the following homogeneous optimizers were selected for their use in the development of heterogeneous general-purpose optimizers along the next chapter:

- **BSt-PSO**: Not suitable for a stand-alone general-purpose optimizer. It exhibits a low rate of convergence, and no ability to fine-tune the search. However, its reluctance to getting trapped in sub-optimal solutions might be helpful for keeping diversity when optimizing functions which present numerous local optima (e.g. the Rastrigrin function).

[5] That is to say that all their particles are exactly the same as one another.

- **BSt/SwSd-PSO 2**: As argued before, these are the closest algorithms to a homogeneous general-purpose optimizer. Their weakest point is their low rate of convergence.
- **BSt/Sw-PSO$^{(c)}$**: By far, the best ones in dealing with the Rosenbrock function, but they find quite poor solutions when dealing with the Rastrigrin function. Their performance is just acceptable when optimizing the other functions in the test suite. In summary, they are very good in dealing with functions that do not present numerous local optima.
- **BStLd-PSO 1$^{(c)}$**: Definitely, not suitable for a stand-alone optimizer. It does not find very good solutions to any of the functions in the test suite. However, it is one of the few optimizers which manage to achieve a high degree of clustering when optimizing the Rosenbrock function. In fact, it attains the first set of termination conditions along all the 50 runs. Furthermore, it does so in less than 2000 time-steps[6]. Hence, this optimizer might be useful to enhance the fine-clustering ability.
- **BSt-PSO$^{(p)}$**: It exhibits a behaviour similar to that of the BSt-PSO$^{(c)}$. Since its rate of convergence is usually slightly lower than that of the BSt-PSO$^{(c)}$, it typically finds slightly better solutions to functions with noise and with numerous local optima, and slightly worse solutions to functions which exhibit a single optimum.
- **BStSd-PSO 1$^{(p)}$**: It finds good solutions to the Sphere and to the Griewank functions, and an acceptable solution to the Rosenbrock function. In constrast, the solutions to the Rastrigrin and to the Schaffer f6 functions are quite poor. An important feature is that it is another of the few optimizers which attain the first set of termination conditions (see **Table 8. 2**), and that it takes almost four times longer than the BStLd-PSO 1$^{(c)}$ to do so, thus possibly complementing each other.

[6] Recall that the BSt-PSO$^{(c)}$ and the BSt-PSO$^{(p)}$—among others—displayed a strange, small divergence of their particles after achieving a high degree of clustering, so that they were never able to attain the fist set of termination conditions despite the good solutions they managed to find.

# 8.3 Closure

A preliminary study of the influence of the basic parameters of the plain B-PSO was extensively undertaken along **Chapter 6**. This allowed for the testing of different parameters' tunings, and broadly understanding the influence of each of them on the behaviour and achievements of the swarm. Some stopping criteria were developed along **Chapter 7**, so that the rate of convergence resulting from different parameters' settings could also be incorporated into the analyses carried out along the present chapter. Thus, 30 different settings of the B-PSO were proposed and tested on a suite of benchmark functions along section **8.2.2**, and some of them were selected as being possibly useful for the development of the general-purpose optimizers, which is to be carried out along the next chapter. The desirable features considered for the selection are the best solution they were able to find; their reluctance to getting trapped in sub-optimal solutions; their ability to fine-cluster; the degree of clustering that their particles achieved by the time the search was terminated; and their rate of convergence[7]. The next chapter is entirely devoted to the development and testing of some general-purpose optimizers, profiting from the results obtained up until the present chapter.

With regards to the study of the influence of the parameters of the particles' velocity update equation on the behaviour of the swarm, further research is required. For instance, the explosion caused by the random weights is still not comprehended; further study on the appropriate values for the velocity constraint should be carried out; self-adapting rather than time-varying inertia weights would make the variation of the latter independent from the maximum number of time-steps permitted for the search (e.g. the value of the inertia weight could be linked to the degree of clustering of the particles, which is already computed for the stopping criteria); the addition of a random weight multiplying the inertia weight so as to dynamically and stochastically alter the strength of the particles' reluctance to change direction could also be tried; other distributions for the generation of the random weights such as the Gaussian distribution might be worth trying, which could increase the convergence rate (possibly in detriment of the exploration ability); etc. To summarize, further research on the effect of the parameters' settings, even of the plain B-PSO, is still required.

[7] Note that a high degree of clustering allows fine-tuning the search, while a high rate of convergence allows saving computational cost.

# Chapter 9

# GENERAL-PURPOSE OPTIMIZERS

The influence of the parameters of the particles' velocity update equation on the behaviour and achievements of the swarm was extensively studied along previous chapters. Thus, different parameters' settings were proposed and tested, some of which resulted in outstanding fine-clustering ability while some others resulted in outstanding reluctance to getting trapped in sub-optimal solutions. However, no setting was found to encompass both abilities, together with a high rate of convergence. Therefore, while the design of the optimizers was so far only concerned with the convenient settings for the parameters of the basic PSO, some strategies that alter the canonical version of the algorithm are proposed within this chapter, namely the subdivision of the swarm into sub-swarms with different abilities, the incorporation of a basic stochastic local search embedded into the particles' learning, and the implementation of a local version of the optimizer.

## 9.1 Introduction

A number of different optimizers differing "only" in the settings of their parameters were developed and tested along **Chapter 6** and **Chapter 8**. It was observed that, although the behaviour and achievements of the system greatly depend on the settings of the parameters of the particles' velocity update equation, the settings that result in particles with the ability to fine-cluster typically also result in high rates of convergence and little ability to escape sub-optimal solutions. In contrast, the settings that result in strong reluctance to getting trapped in sub-optimal solutions lead to optimizers without the ability to fine-tune the search and lower rates of convergence. Therefore, swarms composed of sub-swarms whose parameters are tuned so as to exhibit complementary capabilities are proposed within this chapter, profiting from the results obtained from the experiments carried out along previous chapters. The aim is to combine the different abilities they posses in a single optimizer. In addition, the alternative of incorporating a very basic stochastic local search is also proposed and tested, although its implementation is indeed rudimentary. The improvements resulting from embedding the local search should be taken with a "pinch of salt" because they greatly increase the number of function evaluations per time-step. Finally, local versions of two selected optimizers are implemented, which differ from the global version in that the spread of information is carried out through overlapping neighbourhoods composed of three particles each rather than through a single neighbourhood. In other words, every neighbourhood is composed of three particles, and every particle belongs to three neighbourhoods. Hence, a number of candidate general-purpose optimizers are proposed along the next section, and they are tested on the suite of benchmark functions stated in **Table 6.1** along section **9.3**.

## 9.2 General-purpose optimizers

Broadly speaking, the objective pursued with the following designs is to bring together the ability to fine-cluster associated to some parameters' settings and the reluctance to getting trapped in sub-optimal solutions associated to some others. Note that a high convergence rate is usually associated with the ability to fine-cluster.

### 9.2.1 The GP-PSO 1

Aiming to gather together the proven ability to fine-cluster of the BSt-PSO$^{(c)}$ and the proven ability to escape sub-optimal solutions of the BSt-PSO, the first general-purpose optimizer proposed here is simply composed of two sub-swarms of 15 particles each seeking the best conflict, whose parameters' settings are as follows:

- BSt-PSO$^{(c)}$: basic, standard PSO with $aw^{(t)} = 4.1 \cdot w^{(t)} \quad \forall t$ :
  $w^{(t)} = 0.7298, \quad iw^{(t)} = sw^{(t)} = 1.49609 \quad \forall t$

- BSt-PSO: basic, standard PSO:
  $w^{(t)} = 0.7, \quad iw^{(t)} = sw^{(t)} = 2 \quad \forall t$

In addition, there is another small sub-swarm composed of only 5 particles seeking the worst solution, which is used in the computation of the termination conditions:

- BSt-PSO: basic, standard PSO:
  $w^{(t)} = 0.7, \quad iw^{(t)} = sw^{(t)} = 2 \quad \forall t$

The termination conditions are computed considering only the sub-swarm that possesses the ability to fine-cluster. This is because the BSt-PSO typically takes too long in attaining the first set of termination conditions—if it ever does—, so that the search would tend to be long even for simple functions (e.g. see **Fig. 9. 1**).

## 9.2.2 The GP-PSO 2

The design of this optimizer follows the same line of thought as that of the GP-PSO 1: the swarm is composed of two sub-swarms, one with the ability to fine-cluster and the other with the ability to escape sub-optimal solutions. The BSt-PSO$^{(p)}$ is used in replacement of the BSt-PSO$^{(c)}$ because the former also ends up fine-tuning the search, while it typically maintains diversity for a little longer than the BSt-PSO$^{(c)}$. Hence, the second general-purpose optimizer proposed here is made up of two sub-swarms composed of 15 particles each seeking the best conflict, whose parameters' settings are as follows:

- BSt-PSO$^{(p)}$: basic, standard PSO with $aw^{(t)} = p\left(w^{(t)}\right) \ \forall t$:
  $w^{(t)} = 0.5, \ \ iw^{(t)} = sw^{(t)} = 2 \ \ \forall t$

- BSt-PSO: basic, standard PSO:
  $w^{(t)} = 0.7, \ \ iw^{(t)} = sw^{(t)} = 2 \ \ \forall t$

The sub-swarm seeking the worst conflict is the same as before:

- BSt-PSO: basic, standard PSO:
  $w^{(t)} = 0.7, \ \ iw^{(t)} = sw^{(t)} = 2 \ \ \forall t$

The termination conditions are computed considering only the sub-swarm that possesses the ability to fine-cluster (i.e. the BSt-PSO$^{(p)}$).

## 9.2.3 The GP-PSO 3

Although both the BSt-PSO$^{(c)}$ and the BSt-PSO$^{(p)}$ possess the ability to fine-cluster, they outperform each other when optimizing different functions. Therefore, both optimizers in addition to the BSt-PSO are combined aiming to profit from their complementary abilities.

Hence, the third general-purpose optimizer proposed here is composed of three sub-swarms of 10 particles each seeking the best conflict, whose parameters' settings are as follows:

- BSt-PSO$^{(p)}$: basic, standard PSO with $aw^{(t)} = p\left(w^{(t)}\right) \ \forall t$:

$$w^{(t)} = 0.5, \quad iw^{(t)} = sw^{(t)} = 2 \quad \forall t$$

- BSt-PSO[c]: basic, standard PSO with $aw^{(t)} = 4.1 \cdot w^{(t)} \quad \forall t$:
$$w^{(t)} = 0.7298, \quad iw^{(t)} = sw^{(t)} = 1.49609 \quad \forall t$$

- BSt-PSO: basic, standard PSO:
$$w^{(t)} = 0.7, \quad iw^{(t)} = sw^{(t)} = 2 \quad \forall t$$

The sub-swarm seeking the worst conflict is the same as before:

- BSt-PSO: basic, standard PSO:
$$w^{(t)} = 0.7, \quad iw^{(t)} = sw^{(t)} = 2 \quad \forall t$$

The termination conditions are computed considering only the 2 sub-swarms that possess the ability to fine-cluster (20 particles), while only 10 particles are now in charge of keeping diversity and of escaping sub-optimal solutions.

The evolution of the best conflict found so far, of the average conflict among those of the 20 particles with the ability to fine-cluster, and of the average conflict among those of the 30 particles of the minimizer is shown in **Fig. 9. 1**:

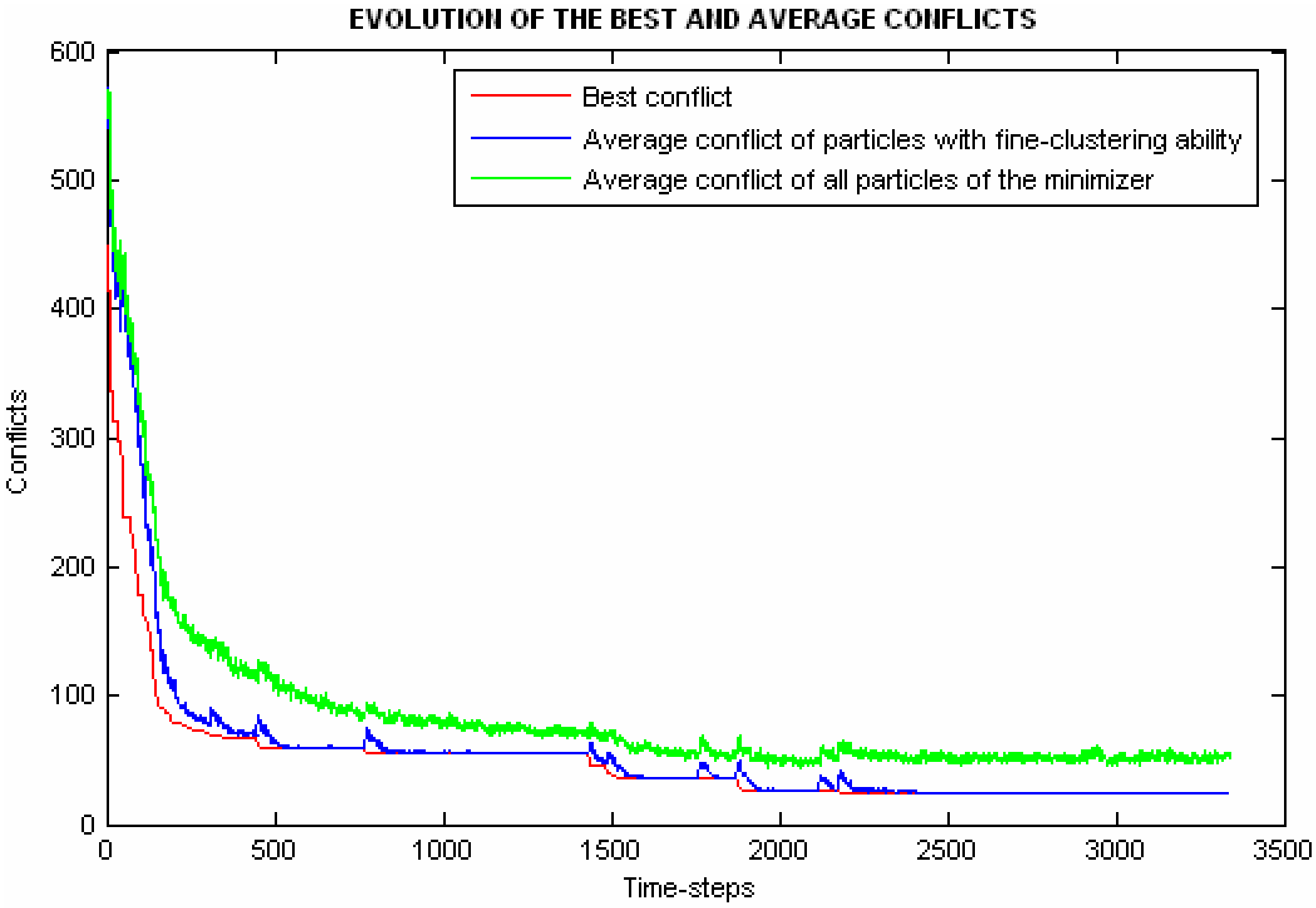

**Fig. 9. 1**: Evolution of the best conflict found so far by the whole swarm, of the average among the current conflicts of all the particles with the ability to fine-cluster (20 particles), and of the average among the current conflicts of all the particles which are in quest for the minimum conflict (30 particles) for the GP-PSO 3 optimizing the 30-dimensional Rastrigrin function.

The difficulty in optimizing the Rastrigrin function lies in the numerous local optima that the function exhibits. The effect of the BSt-PSO helping to escape sub-optimal solutions, and of the BSt-PSO$^{(c)}$ and the BSt-PSO$^{(p)}$ fine-tuning the search, can be clearly observed in **Fig. 9. 1**: a kind of fast local search is carried out until the BSt-PSO finds a better solution, then the local search is performed around the new solution, and so forth. Eventually, the particles get trapped. There is still a need for further investigation of other techniques to keep diversity[1].

## 9.2.4 The GP-PSO 4

The settings of the parameters of the three preceding optimizers are very simple: they are all kept constant throughout the whole search, so that the optimizers' abilities do not vary as the search progresses, which may be especially convenient for dynamic problems. However, both the BSt-PSO$^{(c)}$ and the BSt-PSO$^{(p)}$ proved themselves incapable of performing a complete implosion of their particles when optimizing the Rosenbrock function, although the reason for this was not comprehended. In contrast, the BStLd-PSO 1$^{(c)}$ and the BStSd-PSO 1$^{(p)}$ were two of the few optimizers that were able to perform such implosion, as it can be seen in **Table 8.2** (note that their "mean type of error condition attained" equals 1). Therefore a new optimizer is proposed keeping the BSt-PSO to help escaping sub-optimal solutions, while incorporating the BStLd-PSO 1$^{(c)}$ in replacement of the BSt-PSO$^{(c)}$, and the BStSd-PSO 1$^{(p)}$ in replacement of the BSt-PSO$^{(p)}$. Thus, the GP-PSO 4 is composed of three sub-swarms of 10 particles each seeking the best conflict, whose parameters' settings are as follows (see also **Fig. 9. 2** - left):

- BStLd-PSO 1$^{(c)}$: basic, standard PSO with $aw^{(t)} = 4.1 \cdot w^{(t)} \quad \forall t$ and linearly time-decreasing inertia weight:
  $w^{(1)} = 0.7298,\ w^{(t_{\max})} = 0,\ iw^{(t)} = sw^{(t)} \quad \forall t$

- BStSd-PSO 1$^{(p)}$: basic, standard PSO with $aw^{(t)} = p\left(w^{(t)}\right) \ \forall t$ and sigmoidly time-decreasing inertia weight:
  $w^{(1)} \to 0.5,\ w^{(t_{\max})} \to 0,\ iw^{(t)} = sw^{(t)} \quad \forall t$

- BSt-PSO: basic, standard PSO:
  $w^{(t)} = 0.7,\ \ iw^{(t)} = sw^{(t)} = 2 \quad \forall t$

[1] Note that keeping high diversity may not be helpful in escaping sub-optimal solutions because a better solution cannot be found, and the particles with the ability to fine-cluster attain the stopping criteria. In turn, keeping diversity too low may not be enough to escape the region of attraction of some local optima.

The sub-swarm seeking the worst conflict is the same as before:

- BSt-PSO: basic, standard PSO:
  $w^{(t)} = 0.7, \quad iw^{(t)} = sw^{(t)} = 2 \quad \forall t$

The termination conditions are computed considering only the 2 sub-swarms that possess the ability to fine-cluster (20 particles), while the 10 particles of the third sub-swarm are in charge of maintaining diversity and escaping sub-optimal solutions.

### 9.2.5 The GP-PSO 5

Aiming to delay the attainment of the termination conditions, the incorporation of the BSt-PSO into the computation of the relative errors was considered. However, the BSt-PSO typically exhibits limited or no ability to fine-cluster, resulting in seldom attaining the first set of termination conditions. Therefore, the BSt-PSO in the GP-PSO 3 is replaced by a slightly modified BStSd-PSO 2, whose inertia weight is kept near a value of 0.7 during the early stages of the search, and near a value of 0.5 (rather than 0.4) during the late stages. This results in the GP-PSO 5 behaving similarly to the GP-PSO 3 at the beginning while smoothly changing into a kind of BSt-PSO$^{(p)}$ as the search progresses beyond half of the maximum number of time-steps permitted. Then, all 3 sub-swarms are involved in the computation of the termination conditions, which is expected to delay the attainment of the first set of termination conditions, thus helping to escape sub-optimal solutions.

In summary, the GP-PSO 5 is made up of 3 sub-swarms composed of 10 particles each in quest for the best conflict, whose parameters' settings are as follows (see also **Fig. 9. 2** - right):

- BSt-PSO$^{(p)}$: basic, standard PSO with $aw^{(t)} = p\left(w^{(t)}\right) \ \forall t$:
  $w^{(t)} = 0.5, \quad iw^{(t)} = sw^{(t)} = 2 \quad \forall t$

- BSt-PSO$^{(c)}$: basic, standard PSO with $aw^{(t)} = 4.1 \cdot w^{(t)} \ \forall t$:
  $w^{(t)} = 0.7298, \quad iw^{(t)} = sw^{(t)} = 1.49609 \quad \forall t$

- BStSd-PSO: basic, standard PSO with sigmoidly time-decreasing inertia weight:
  $w^{(1)} \to 0.7, \ w^{(t_{\max})} \to 0.5, \ iw^{(t)} = sw^{(t)} = 2 \quad \forall t$

A BSt-PSO is again in charge of searching for the worst conflict:

- BSt-PSO: basic, standard PSO:

$$w^{(t)} = 0.7, \quad iw^{(t)} = sw^{(t)} = 2 \quad \forall t$$

The evolution of the best and average conflicts corresponding to the GP-PSO 5 optimizing the 30-dimensional Rastrigrin function, where the average is computed considering all the 30 particles of the minimizer, is shown in **Fig. 9. 3** (compare the average conflict to that of the GP-PSO 3, shown in **Fig. 9. 1**).

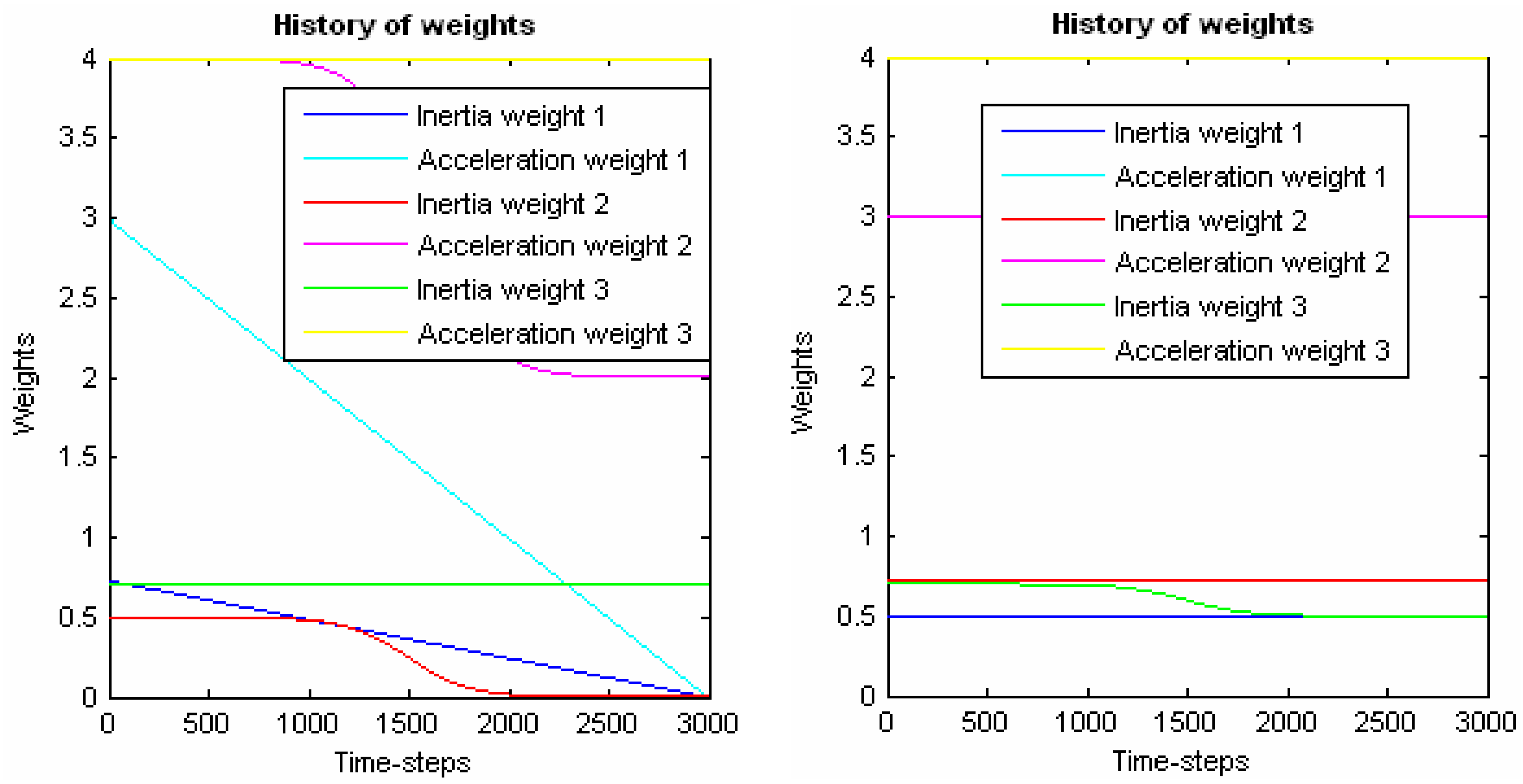

**Fig. 9. 2**: Evolution of the weights corresponding to the 3 sub-swarms composing the GP-PSO 4 (left) and to the 3 sub-swarms composing the GP-PSO 5 (right).

## 9.2.6 The GP-PSO 6

Aiming to delay the attainment of the termination conditions of the GP-PSO 3, a sixth general-purpose optimizer is proposed, which only differs from the GP-PSO 3 in that the BSt-PSO is composed of only 8 particles and considered in the computation of the termination conditions. Thus, the GP-PSO 6 is made up of 3 sub-swarms that are in charge of seeking the best conflict, whose parameters' settings are as follows:

- BSt-PSO$^{(p)}$: basic, standard PSO with $aw^{(t)} = p\left(w^{(t)}\right) \quad \forall t$:
  $w^{(t)} = 0.5, \quad iw^{(t)} = sw^{(t)} = 2 \quad \forall t$ (11 particles)

- BSt-PSO$^{(c)}$: basic, standard PSO with $aw^{(t)} = 4.1 \cdot w^{(t)} \quad \forall t$:
  $w^{(t)} = 0.7298, \quad iw^{(t)} = sw^{(t)} = 1.49609 \quad \forall t$ (11 particles)

- BSt-PSO: basic, standard PSO:
  $w^{(t)} = 0.7, \quad iw^{(t)} = sw^{(t)} = 2 \quad \forall t$ (8 particles)

The sub-swarm in charge of searching for the worst conflict is again the BSt-PSO:

- BSt-PSO: basic, standard PSO:
  $w^{(t)} = 0.7, \quad iw^{(t)} = sw^{(t)} = 2 \quad \forall t$ (5 particles)

All 30 particles of the minimizer are now involved in the computation of the termination conditions.

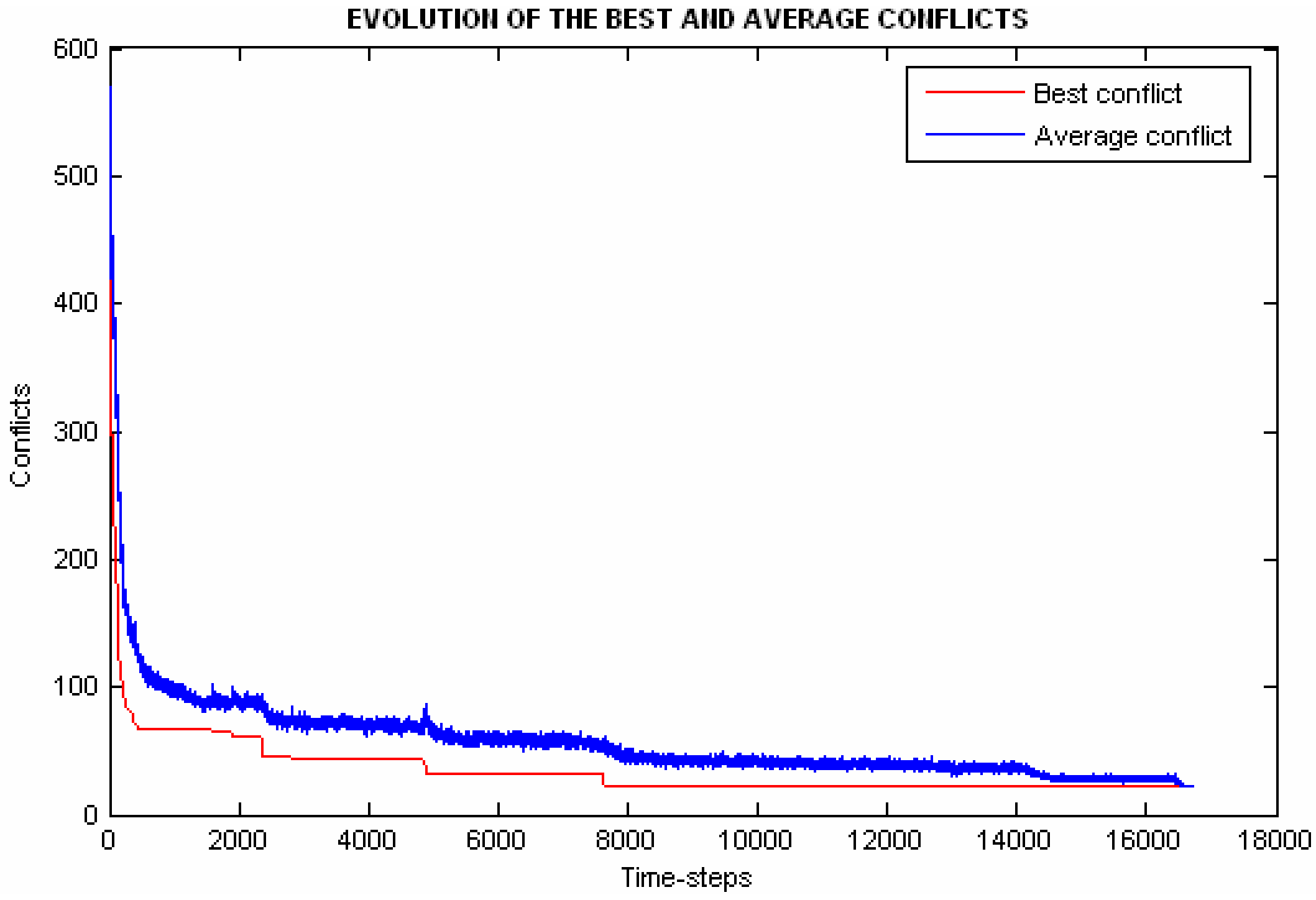

**Fig. 9. 3**: Evolution of the best and average conflicts of the GP-PSO 5 optimizing the 30-dimensional Rastrigrin function, where the average is computed considering all the 30 particles of the minimizer.

### 9.2.7 The GP-PSO (ils)

It is reasonable to expect that the incorporation of a local search into every particle's best previous experience at every time-step would increase the exploration ability of the optimizer while the particles are still widespread over the search-space, and its exploitation ability while the particles are clustering. However, it was observed that this local search was frequently harmful, apparently because it induces too fast a loss of diversity.

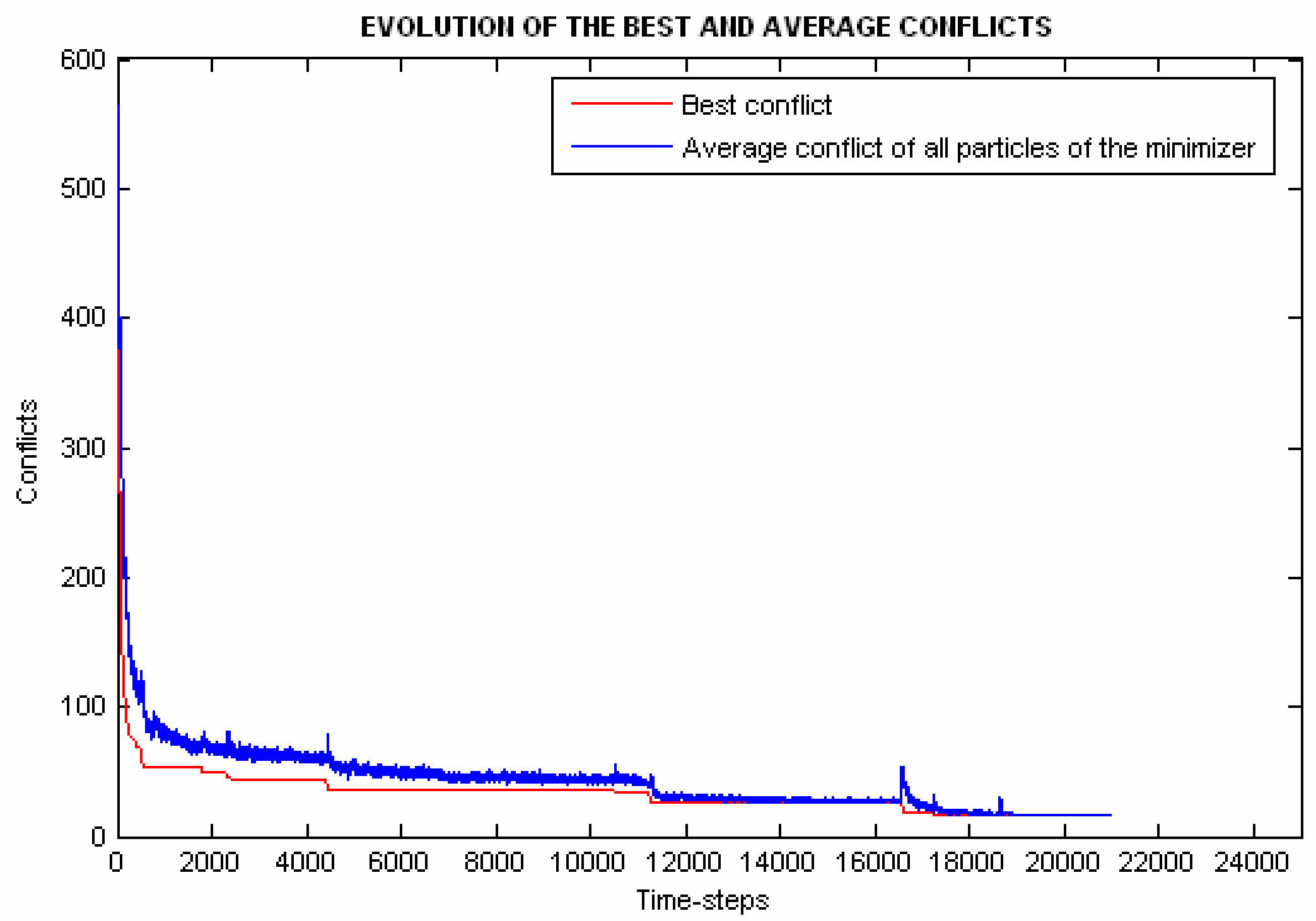

**Fig. 9. 4**: Evolution of the best and average conflicts for the GP-PSO 6 optimizing the 30-dimensional Rastrigrin function, where the average is computed considering all the 30 particles of the minimizer.

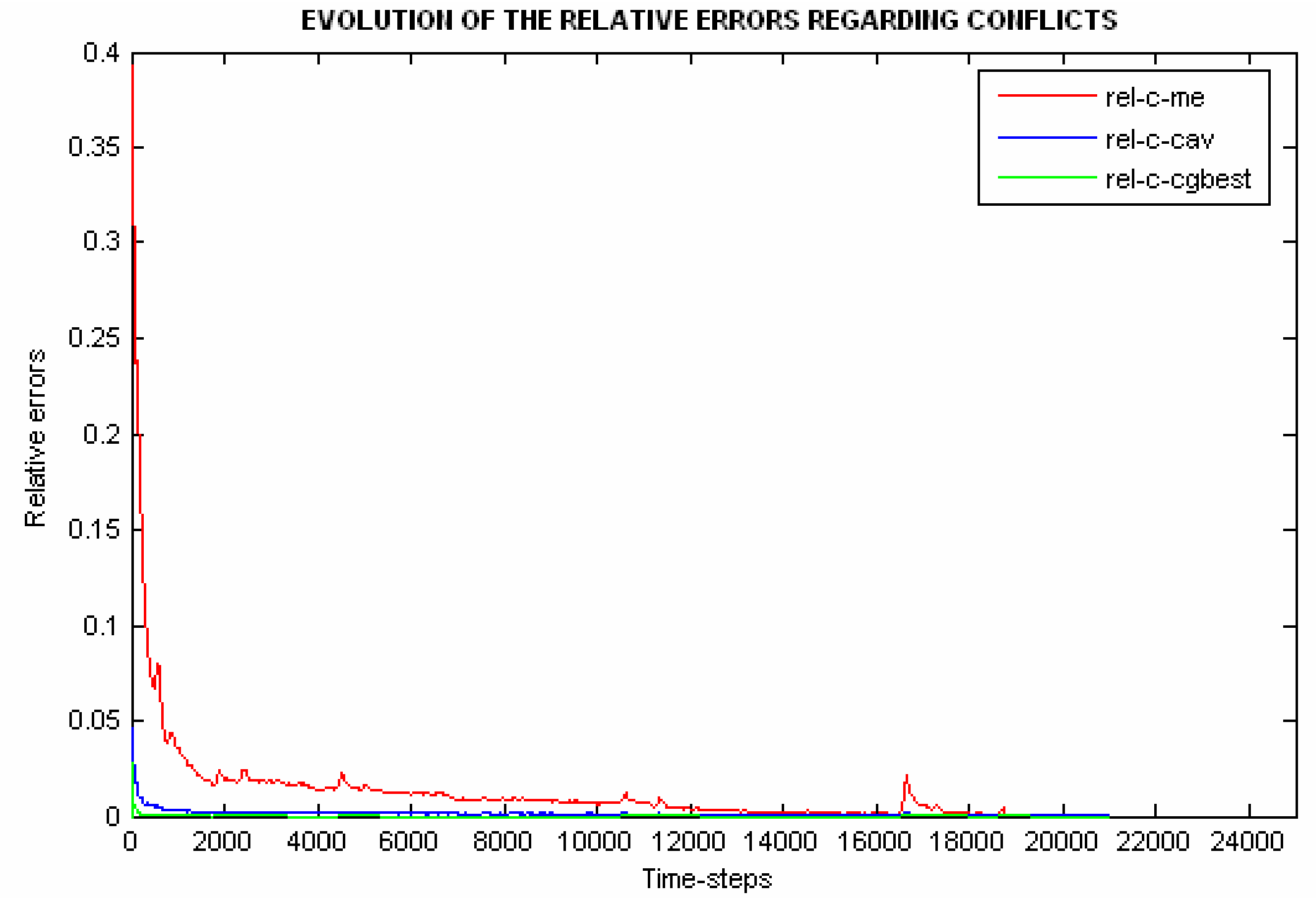

**Fig. 9. 5**: Evolution of the relative errors regarding the conflict values for the GP-PSO 6 optimizing the 30-dimensional Rastrigrin function.

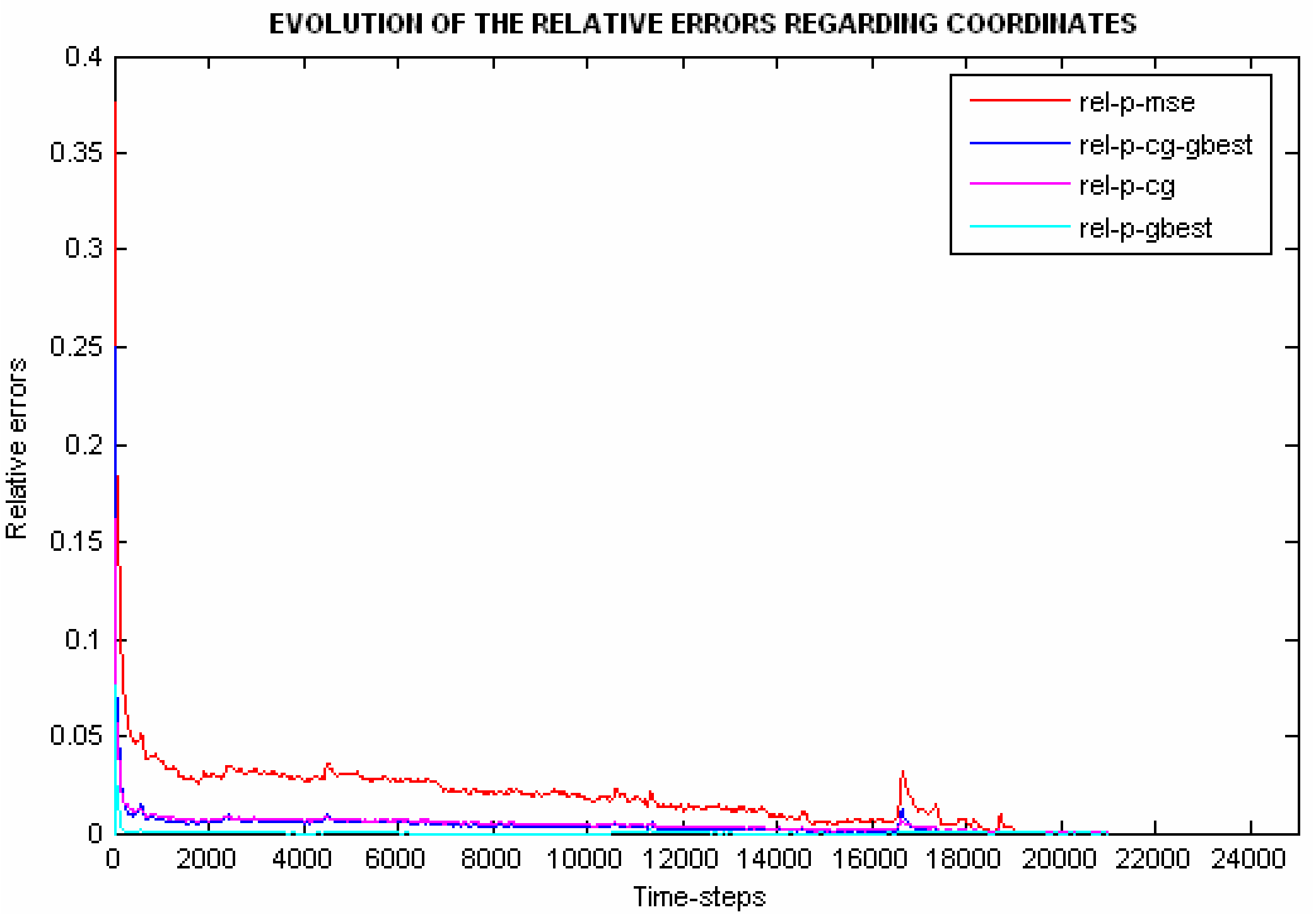

**Fig. 9. 6**: Evolution of the relative errors regarding the particles' positions for the GP-PSO 6 optimizing the 30-dimensional Rastrigrin function.

Given that the PSOs exhibit inherently high convergence rates at the beginning, a local search was implemented so as to be activated only for time-steps $t > 5\% \, t_{max}$, with the purpose to enhance exploitation only. Besides, the local search was only incorporated to 15 out of the 30 particles of the minimizer. Considering the metaphor of individual and social learning that inspired the method, a local search incorporated to the best solution found by a particle would be something like enhancing its individual learning.

A very basic, rudimentary stochastic local search was incorporated into the GP-PSO 5, giving birth to the GP-PSO (ils). The pseudo-code is shown in **Fig. 9. 7**.

## 9.2.8 The GP-PSO (gls)

Two other enhancements of the individual learning in the form of a stochastic local search can be thought of: either a more extended local search can be implemented so as to enhance the swarm's—rather than every particles'—best experience at each time-step, or an even more extended one can be incorporated to the stopping criteria so that, when a set of termination conditions is met, a local search is performed around the best solution found so far, and the search is terminated if and only if this local search is not capable of further improvement. The second option seems to be more convenient with regards to the additional computational costs, and to the previous conjecture that the local search might be sometimes even harmful for the system's exploration ability. However, the rudimentary local search implemented here was seldom able to improve the best solution found by the system by the time the termination conditions were attained. Therefore, only the first alternative was implemented here, and the second is left for future work, involving a more sophisticated local search algorithm.

**if** $t > 5\%\, t_{max}$
  **for** i=1:10
    **for** each of the first 15 particles of the minimizer (particle *j*)
      Generate a new candidate solution:
      $local\ search_{(1,k)} = pbest_{(j,k)} + 0.1\%\left(x_{max} - x_{min}\right)\cdot U_{(-1,1)}$
      **if** *local search* is feasible
        Evaluate the corresponding conflict
        **if** the conflict is smaller than the best conflict found so far by the particle
          Replace the best previous experience of the particle by *local search*
        **end**
      **end**
    **end**
  **end**
**end**

**Fig. 9. 7**: Pseudo-code for the enhancement of the first 15 particles' individual learning, implemented in the form of a very basic stochastic local search with "only" 10 iterations.

Thus, a stochastic local search around the best solution found so far by the swarm at every time-step was incorporated into the GP-PSO 5, giving birth to the GP-PSO (gls). The pseudo-code of this local search is shown in **Fig. 9. 8**.

Notice that the development of a local search to enhance the PSOs can be performed seeking two different purposes: on the one hand, a local search can help to fine-tune the search (e.g. a gradient-based technique, although this would be only suitable for differentiable functions); one the other hand, a local search might help the PSOs to escape sub-optimal solutions when diversity is virtually lost. The second case was the purpose pursued when proposing this rudimentary stochastic local search, although its ability to escape a certain region of attraction is limited by the arbitrary interval considered for the generation of the random numbers.

**if** $t > 5\% \, t_{\max}$
  **for** i=1:150  % the same additional computational cost as the previous local search
    $local\ search_{(1,k)} = gbest_{(1,k)} + 0.1\% \left( x_{\max} - x_{\min} \right) \cdot U_{(-1,1)}$
    **if** *local search* is feasible
      Evaluate the corresponding conflict
      **if** the conflict is smaller than the best conflict found so far
        Replace the best swarm's previous experience by *local search*
      **end**
    **end**
  **end**
**end**

**Fig. 9. 8**: Pseudo-code for the enhancement of the best solution found so far by any particle in the swarm, implemented in the form of a very basic stochastic local search with "only" 150 iterations.

### 9.2.9 The LGP-PSO 1

One of the characteristic features of the PSOs is that they exhibit a high convergence rate at the beginning of the search. Besides, some settings of the parameters of the particles' velocity update equation enable the swarm to fine-tune the search. While another characteristic—and desirable feature of these optimizers—is their ability to escape sub-optimal solutions, their particles eventually get trapped due to the loss of diversity. There is a need of a—typically problem-dependent—appropriate trade-off between the speed of clustering of the particles and their reluctance to getting trapped in sub-optimal solutions.

It is not intended here to assert which such a trade-off should be, but it is evident that there is a need of at least slow down the celerity of clustering. The implementation of a local version of the paradigm seems a reasonable strategy to be considered, since the delay in the flux of information among the population would certainly result in the speed of clustering slowing down. This is expected to lead to a higher number of function evaluations than the global version on the one hand, and to a more pronounced reluctance to getting trapped in sub-optimal solutions on the other.

Four possible neighbourhoods' topologies are depicted in **Fig. 9. 9**. While the above-left (A) topology was the one so far implemented in every code, the below-left topology (C) is now implemented because of its simplicity among the local versions, because the resulting velocity of spread of information is expected to be notably delayed, and because it is one of the most popular topologies among the local versions in the literature.

Thus, a simple local version of the PSO is implemented, where each particle has access to its own memory and to those of its two topologically adjacent neighbours. The topological neighbourhoods are defined at the initial time-step according to the "nearest neighbours" criteria (in the sense of the Euclidean norm), and are kept the same regardless of the particles' relative positions in the search-space at each time-step. The LGP-PSO 1 is implemented as a local version of the GP-PSO 5.

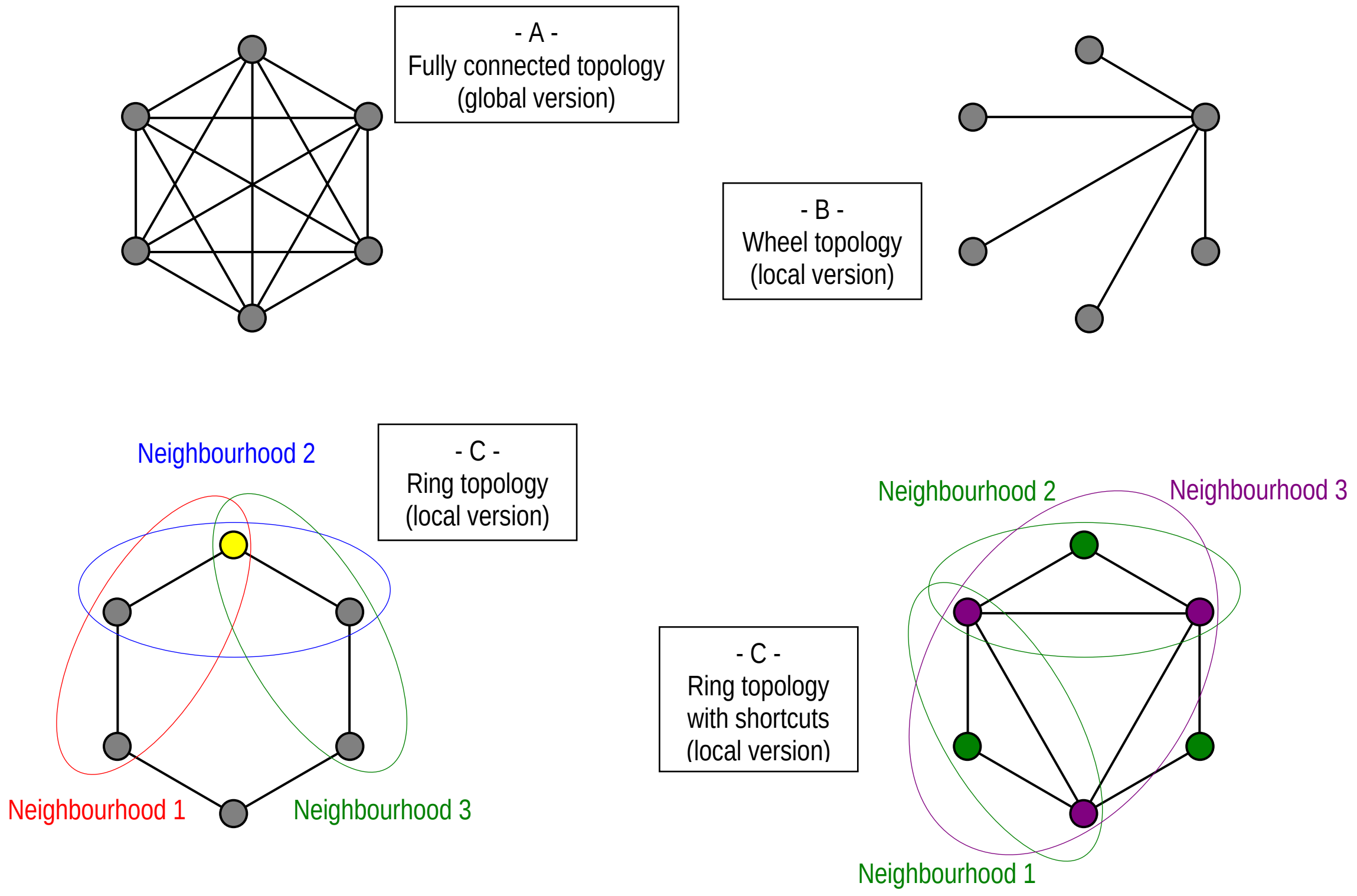

**Fig. 9. 9**: Four possible topological neighbourhoods for the local version of the PSOs:

Above-Left: $k$-best topology with $k$ = swarm size - 1 (fully connected topology)

Above-Right: wheel topology

Below-Left: $k$-best topology with $k$ = 2 (ring topology), which is adopted here

Below-Right: alternative $k$-best topology with $k$ = 2 / 4 (ring topology), where the green partices have access to the 2 topologically nearest neighbours' previous experiences and the purple particles have access to the 4 topologically nearest neighbours' experiences.

### 9.2.10 The LGP-PSO 2

The LGP-PSO 2 is another local PSO implemented using the same criteria as the LGP-PSO 1, but designed as a local version of the GP-PSO 3.

## 9.3 Experimental results

General settings:

- Number of runs per experiment: 50
- $v_{\max} = 0.5 \cdot (x_{\max} - x_{\min})$
- Number of particles of the minimizer: 30
- Number of particles of the maximizer: 5
- $t_{\max} = 30000$

The most significant results are gathered in **Table 9. 1** to **Table 9. 5**:

| SPHERE | | | | | | | |
|---|---|---|---|---|---|---|---|
| OPTIMIZER | Mean best solution found when error condition is attained | Mean time-steps required to attain error condition | Mean type of error condition attained | Number of failures in attaining the error condition | Mean best solution found when error condition not attained | Mean best solution found | Mean time-steps required |
| BSt-PSO | 1.2547E-16 | 1.66E+04 | 1.00 | 0 | - | 1.25E-16 | 1.66E+04 |
| BSt-PSO(c) | 1.0549E-43 | 3.00E+03 | 1.00 | 0 | - | 1.05E-43 | 3.00E+03 |
| BSt-PSO(p) | 2.7360E-37 | 3.00E+03 | 1.00 | 0 | - | 2.74E-37 | 3.00E+03 |
| GP-PSO 1 | 6.8109E-17 | 3.01E+03 | 1.00 | 0 | - | 6.81E-17 | 3.01E+03 |
| GP-PSO 2 | 8.5868E-26 | 3.00E+03 | 1.00 | 0 | - | 8.59E-26 | 3.00E+03 |
| GP-PSO 3 | 1.4219E-33 | 3.00E+03 | 1.00 | 0 | - | 1.42E-33 | 3.00E+03 |
| GP-PSO 4 | 4.3397E-28 | 3.00E+03 | 1.00 | 0 | - | 4.34E-28 | 3.00E+03 |
| GP-PSO 5 | 1.3772E-38 | 4.46E+03 | 1.00 | 0 | - | 1.38E-38 | 4.46E+03 |
| GP-PSO 6 | 3.5879E-42 | 4.47E+03 | 1.00 | 0 | - | 3.59E-42 | 4.47E+03 |
| GP-PSO (ils) | 6.3053E-43 | 4.56E+03 | 1.00 | 0 | - | 6.31E-43 | 4.56E+03 |
| GP-PSO (gls) | 6.3053E-43 | 4.56E+03 | 1.00 | 0 | - | 6.31E-43 | 4.56E+03 |
| LGP-PSO 1 | 5.0158E-55 | 9.30E+03 | 1.00 | 0 | - | 5.02E-55 | 9.30E+03 |
| LGP-PSO 2 | 3.4767E-20 | 3.00E+03 | 1.00 | 0 | - | 3.48E-20 | 3.00E+03 |

**Table 9. 1**: Results obtained from the optimization of the 30-dimensional Sphere function by means of 3 basic and 10 general-purpose PSOs.

| ROSENBROCK | | | | | | | |
|---|---|---|---|---|---|---|---|
| OPTIMIZER | Mean best solution found when error condition is attained | Mean time-steps required to attain error condition | Mean type of error condition attained | Number of failures in attaining the error condition | Mean best solution found when error condition not attained | Mean best solution found | Mean time-steps required |
| BSt-PSO | - | - | - | 50 | 4.2397E+01 | 4.24E+01 | 3.00E+04 |
| BSt-PSO(c) | 4.0059E+00 | 2.92E+04 | 2.00 | 49 | 1.3975E+00 | 1.45E+00 | 3.00E+04 |
| BSt-PSO(p) | 2.3019E+01 | 2.09E+04 | 2.00 | 16 | 4.3836E+00 | 1.71E+01 | 2.38E+04 |
| GP-PSO 1 | 1.8086E+01 | 2.52E+04 | 2.00 | 47 | 1.9029E+00 | 2.87E+00 | 2.97E+04 |
| GP-PSO 2 | 2.0926E+01 | 2.52E+04 | 2.00 | 24 | 1.3060E+01 | 1.72E+01 | 2.75E+04 |
| GP-PSO 3 | 1.7643E+01 | 2.70E+04 | 2.00 | 36 | 9.8587E+00 | 1.20E+01 | 2.92E+04 |
| GP-PSO 4 | 2.1638E+01 | 1.97E+04 | 1.03 | 15 | 4.0785E+01 | 2.74E+01 | 2.28E+04 |
| GP-PSO 5 | 4.3954E+01 | 2.98E+04 | 2.00 | 48 | 6.8154E+00 | 8.30E+00 | 3.00E+04 |
| GP-PSO 6 | 9.4614E+00 | 2.78E+04 | 2.00 | 46 | 8.1717E+00 | 8.27E+00 | 2.98E+04 |
| GP-PSO (ils) | 1.2863E+01 | 2.74E+04 | 2.00 | 48 | 6.7251E+00 | 6.97E+00 | 2.99E+04 |
| GP-PSO (gls) | 1.3909E+01 | 2.61E+04 | 2.00 | 47 | 7.2137E+00 | 7.62E+00 | 2.98E+04 |
| LGP-PSO 1 | - | - | - | 50 | 1.0585E+00 | 1.06E+00 | 3.00E+04 |
| LGP-PSO 2 | - | - | - | 50 | 1.0585E+00 | 1.06E+00 | 3.00E+04 |

**Table 9. 2**: Results obtained from the optimization of the 30-dimensional Rosenbrock function by means of 3 basic and 10 general-purpose PSOs.

| RASTRIGRIN | | | | | | | |
|---|---|---|---|---|---|---|---|
| OPTIMIZER | Mean best solution found when error condition is attained | Mean time-steps required to attain error condition | Mean type of error condition attained | Number of failures in attaining the error condition | Mean best solution found when error condition not attained | Mean best solution found | Mean time-steps required |
| BSt-PSO | - | - | - | 50 | 7.8893E+00 | 7.89E+00 | 3.00E+04 |
| BSt-PSO(c) | 6.5886E+01 | 3.00E+03 | 1.00 | 0 | - | 6.59E+01 | 3.00E+03 |
| BSt-PSO(p) | 5.0683E+01 | 3.20E+03 | 1.00 | 0 | - | 5.07E+01 | 3.20E+03 |
| GP-PSO 1 | 2.8297E+01 | 5.74E+03 | 1.00 | 0 | - | 2.83E+01 | 5.74E+03 |
| GP-PSO 2 | 2.7660E+01 | 5.88E+03 | 1.04 | 0 | - | 2.77E+01 | 5.88E+03 |
| GP-PSO 3 | 3.4744E+01 | 4.93E+03 | 1.04 | 0 | - | 3.47E+01 | 4.93E+03 |
| GP-PSO 4 | 3.2595E+01 | 4.21E+03 | 1.00 | 0 | - | 3.26E+01 | 4.21E+03 |
| GP-PSO 5 | 2.2813E+01 | 2.00E+04 | 1.55 | 8 | 1.3183E+01 | 2.13E+01 | 2.16E+04 |
| GP-PSO 6 | 2.4520E+01 | 2.01E+04 | 1.82 | 5 | 1.3332E+01 | 2.34E+01 | 2.11E+04 |
| GP-PSO (ils) | 2.2933E+01 | 2.13E+04 | 1.71 | 9 | 2.1447E+01 | 2.27E+01 | 2.29E+04 |
| GP-PSO (gls) | 1.5754E+01 | 2.30E+04 | 1.27 | 20 | 1.2421E+01 | 1.44E+01 | 2.58E+04 |
| LGP-PSO 1 | 2.9849E+01 | 2.45E+04 | 2.00 | 42 | 3.1311E+01 | 3.11E+01 | 2.91E+04 |
| LGP-PSO 2 | 4.2228E+01 | 2.34E+04 | 2.00 | 45 | 2.6764E+01 | 2.83E+01 | 2.93E+04 |

**Table 9. 3**: Results obtained from the optimization of the 30-dimensional Rastrigrin function by means of 3 basic and 10 general-purpose PSOs.

| GRIEWANK | | | | | | | |
|---|---|---|---|---|---|---|---|
| OPTIMIZER | Mean best solution found when error condition is attained | Mean time-steps required to attain error condition | Mean type of error condition attained | Number of failures in attaining the error condition | Mean best solution found when error condition not attained | Mean best solution found | Mean time-steps required |
| BSt-PSO | 1.5167E-02 | 2.61E+04 | 1.64 | 28 | 3.6706E-02 | 2.72E-02 | 2.83E+04 |
| BSt-PSO(c) | 3.1750E-02 | 3.00E+03 | 1.00 | 0 | - | 3.17E-02 | 3.00E+03 |
| BSt-PSO(p) | 1.5002E-02 | 3.00E+03 | 1.00 | 0 | - | 1.50E-02 | 3.00E+03 |
| GP-PSO 1 | 7.2862E-02 | 3.05E+03 | 1.00 | 0 | - | 7.29E-02 | 3.05E+03 |
| GP-PSO 2 | 3.4361E-02 | 3.00E+03 | 1.00 | 0 | - | 3.44E-02 | 3.00E+03 |
| GP-PSO 3 | 4.0723E-02 | 3.01E+03 | 1.00 | 0 | - | 4.07E-02 | 3.01E+03 |
| GP-PSO 4 | 2.3699E-02 | 3.00E+03 | 1.00 | 0 | - | 2.37E-02 | 3.00E+03 |
| GP-PSO 5 | 1.9781E-02 | 1.01E+04 | 1.26 | 0 | - | 1.98E-02 | 1.01E+04 |
| GP-PSO 6 | 2.0543E-02 | 1.01E+04 | 1.40 | 0 | - | 2.05E-02 | 1.01E+04 |
| GP-PSO (ils) | 2.1789E-02 | 1.08E+04 | 1.32 | 0 | - | 2.18E-02 | 1.08E+04 |
| GP-PSO (gls) | 2.1789E-02 | 1.08E+04 | 1.32 | 0 | - | 2.18E-02 | 1.08E+04 |
| LGP-PSO 1 | 5.0241E-03 | 1.55E+04 | 1.88 | 1 | 9.8647E-03 | 5.12E-03 | 1.58E+04 |
| LGP-PSO 2 | 4.7721E-03 | 9.09E+03 | 1.18 | 1 | 1.2316E-02 | 4.92E-03 | 9.51E+03 |

**Table 9. 4**: Results obtained from the optimization of the 30-dimensional Griewank function by means of 3 basic and 10 general-purpose PSOs.

| SCHAFFER F6 2D | | | | | | | |
|---|---|---|---|---|---|---|---|
| OPTIMIZER | Mean best solution found when error condition is attained | Mean time-steps required to attain error condition | Mean type of error condition attained | Number of failures in attaining the error condition | Mean best solution found when error condition not attained | Mean best solution found | Mean time-steps required |
| BSt-PSO | 0.0000E+00 | 6.03E+03 | 1.04 | 0 | - | 0.00E+00 | 6.03E+03 |
| BSt-PSO(c) | 1.9432E-04 | 3.54E+03 | 1.02 | 0 | - | 1.94E-04 | 3.54E+03 |
| BSt-PSO(p) | 5.8295E-04 | 4.04E+03 | 1.06 | 0 | - | 5.83E-04 | 4.04E+03 |
| GP-PSO 1 | 0.0000E+00 | 3.30E+03 | 1.00 | 0 | - | 0.00E+00 | 3.30E+03 |
| GP-PSO 2 | 0.0000E+00 | 3.50E+03 | 1.00 | 0 | - | 0.00E+00 | 3.50E+03 |
| GP-PSO 3 | 0.0000E+00 | 3.29E+03 | 1.00 | 0 | - | 0.00E+00 | 3.29E+03 |
| GP-PSO 4 | 0.0000E+00 | 3.42E+03 | 1.00 | 0 | - | 0.00E+00 | 3.42E+03 |
| GP-PSO 5 | 3.8864E-04 | 5.47E+03 | 1.08 | 0 | - | 3.89E-04 | 5.47E+03 |
| GP-PSO 6 | 0.0000E+00 | 4.27E+03 | 1.00 | 0 | - | 0.00E+00 | 4.27E+03 |
| GP-PSO (ils) | 0.0000E+00 | 4.68E+03 | 1.00 | 0 | - | 0.00E+00 | 4.68E+03 |
| GP-PSO (gls) | 1.9432E-04 | 4.85E+03 | 1.04 | 0 | - | 1.94E-04 | 4.85E+03 |
| LGP-PSO 1 | 0.0000E+00 | 9.96E+03 | 1.38 | 0 | - | 0.00E+00 | 9.96E+03 |
| LGP-PSO 2 | 0.0000E+00 | 8.15E+03 | 1.16 | 0 | - | 0.00E+00 | 8.15E+03 |

**Table 9. 5**: Results obtained from the optimization of the 2-dimensional Schaffer f6 function by means of 3 basic and 10 general-purpose PSOs.

| SCHAFFER F6 | | | | | | | |
|---|---|---|---|---|---|---|---|
| OPTIMIZER | Mean best solution found when error condition is attained | Mean time-steps required to attain error condition | Mean type of error condition attained | Number of failures in attaining the error condition | Mean best solution found when error condition not attained | Mean best solution found | Mean time-steps required |
| BSt-PSO | 9.8930E-02 | 2.40E+04 | 2.00 | 10 | 6.5907E-02 | 9.23E-02 | 2.52E+04 |
| BSt-PSO(c) | 8.1977E-02 | 1.67E+04 | 2.00 | 1 | 3.7224E-02 | 8.11E-02 | 1.69E+04 |
| BSt-PSO(p) | 7.0308E-02 | 1.71E+04 | 2.00 | 2 | 3.7224E-02 | 6.90E-02 | 1.76E+04 |
| GP-PSO 1 | 8.3926E-02 | 1.93E+04 | 2.00 | 2 | 3.7224E-02 | 8.21E-02 | 1.97E+04 |
| GP-PSO 2 | 7.1821E-02 | 1.91E+04 | 2.00 | 1 | 3.7224E-02 | 7.11E-02 | 1.94E+04 |
| GP-PSO 3 | 7.9985E-02 | 1.74E+04 | 2.00 | 1 | 3.7224E-02 | 7.91E-02 | 1.77E+04 |
| GP-PSO 4 | 8.2323E-02 | 2.09E+04 | 2.00 | 7 | 4.8928E-02 | 7.76E-02 | 2.22E+04 |
| GP-PSO 5 | 6.7843E-02 | 1.86E+04 | 2.00 | 4 | 3.7224E-02 | 6.54E-02 | 1.95E+04 |
| GP-PSO 6 | 6.9973E-02 | 1.86E+04 | 2.00 | 3 | 3.7224E-02 | 6.80E-02 | 1.93E+04 |
| GP-PSO (ils) | 5.7044E-02 | 1.83E+04 | 2.00 | 0 | - | 5.70E-02 | 1.83E+04 |
| GP-PSO (gls) | 6.2625E-02 | 1.82E+04 | 2.00 | 1 | 3.7224E-02 | 6.21E-02 | 1.85E+04 |
| LGP-PSO 1 | 7.1040E-02 | 2.05E+04 | 2.00 | 10 | 5.6050E-02 | 6.80E-02 | 2.24E+04 |
| LGP-PSO 2 | 7.1443E-02 | 2.06E+04 | 2.00 | 8 | 4.6546E-02 | 6.75E-02 | 2.21E+04 |

**Table 9. 6**: Results obtained from the optimization of the 30-dimensional Schaffer f6 function by means of 3 basic and 10 general-purpose PSOs.

## 9.4 Discussion

All the proposed general-purpose optimizers in addition to three selected B-PSOs were tested on the suite of benchmark functions stated in **Table 6.1**. The B-PSOs are included here to serve the function of points of reference.

It is fair to note that these three B-PSOs are not badly-tuned optimizers specifically selected to show the goodness of the GP-PSOs. On the contrary, they are well-tuned optimizers brought from previous chapters, where the BSt-PSO is very good at escaping sub-optimal solutions, and both the BSt-PSO$^{(c)}$ and the BSt-PSO$^{(p)}$ are very good at fine-clustering, and hence at fine-tuning the search.

The analysis of the performances of the B-PSOs is carried out along section **9.4.1**, and the analysis of the performances of the GP-PSOs is performed along section **9.4.2**. Finally, the conclusions derived from the experiments are drawn along section **9.4.3**.

## 9.4.1 The B-PSOs

### 9.4.1.1 The BSt-PSO

It can be observed that the BSt-PSO performs very well when optimizing the Rastrigrin and the 2-dimensional Schaffer f6 functions. Although the solutions it manages to find for the 30-dimensional Schaffer f6 and the Sphere functions are not noticeably worse than those found by the other optimizers, they are the worst solutions, and the number of time-steps required to find them are the highest (noticeably higher in the case of the Sphere function). The solution it finds for the Griewank function is acceptable, but it again required a number of time-steps considerably greater than those required by the other optimizers (from 2 to 10 times greater). Finally, the solution this optimizer finds for the Rosenbrock function is remarkably bad.

Clearly, the desirable feature of the BSt-PSO is its ability to escape sub-optimal solutions. In fact, the only case where it clearly outperforms any other optimizer is when optimizing the Rastrigrin function.

### 9.4.1.2 The BSt-PSO$^{(c)}$

This optimizer can be considered to be complementary to the BSt-PSO, because it performs very well on those cases where the latter performs very badly, and vice versa. Thus, it finds extremely good solutions for the Sphere function (and 5 times faster than the BSt-PSO) and for the Rosenbrock function. Although the one it finds for the Griewank function is slightly worse than that found by the BSt-PSO, it does so 10 times faster. In contrast, it is among the worst optimizers in dealing with the Rastrigrin function and both the 2-dimensional and the 30-dimensional Schaffer f6 functions. Note that theses are precisely the functions which present numerous local optima.

Clearly, the desirable features of the BSt-PSO are its ability to fine-cluster—which results in the fine-tuning of the search—, and its high convergence rate.

### 9.4.1.3 The BSt-PSO$^{(p)}$

Despite of presenting higher learning weights and lower inertia weight, this optimizer shares the strengths and weaknesses of the BSt-PSO$^{(c)}$, although it typically takes longer to complete the implosion of its particles. Thus, it finds somewhat better solutions for the Rastrigrin, the Griewank, and the 30-dimensional Schaffer f6 functions, but somewhat worse solutions for the other functions. Nonetheless, all the solutions the BSt-PSO$^{(c)}$ and the BSt-PSO$^{(p)}$ find have more or less the same degree of "goodness", except for the case of the Rosenbrock function, where the former clearly outperforms the latter.

## 9.4.2 The GP-PSOs

### 9.4.2.1 The GP-PSO 1

This optimizer is composed of 2 sub-swarms[2], one corresponding to a BSt-PSO and the other to a BSt-PSO$^{(c)}$. It is not so surprising then to observe that the solutions it manages to find are generally better than those found by one and worse than those found by the other. The exception is the Griewank function, for which the GP-PSO 1 finds a solution that is worse than the solutions found by both the BSt-PSO and the BSt-PSO$^{(c)}$. Thus, the results obtained by this optimizer are more general-purpose than those obtained by these B-PSOs, as expected.

With regards to the goodness of the solutions, it could be said that the GP-PSO 1 finds good solutions for the Sphere and the 2-dimensional Schaffer f6 function, and a very good solution for the Rosenbrock function. The solutions it finds for the Griewank and for the 30-dimensional Schaffer f6 functions are acceptable, while the one found for the Rastrigrin function is still not very good, although it is noticeably better than those found by the BSt-PSO$^{(c)}$ and the BSt-PSO$^{(p)}$.

### 9.4.2.2 The GP-PSO 2

This optimizer is composed of 2 sub-swarms, one corresponding to a BSt-PSO and the other to a BSt-PSO$^{(p)}$. Again, the solutions it finds are generally better than those found by one and worse than those found by the other. The exception is also the Griewank function, for which the GP-PSO 2 finds a solution that is worse than the solutions found by both the BSt-PSO and the BSt-PSO$^{(p)}$, although it is still a reasonably good solution. Thus, the results obtained by this optimizer are more general-purpose than those found by the individual optimizers it is composed of, as expected.

[2] In addition to the sub-swarm seeking the worst conflict.

Comparing it to the GP-PSO 1, the GP-PSO 2 finds quite better solutions for the Sphere and Griewank functions, and slightly better solutions for the Rastrigrin and 30-dimensional Schaffer f6 functions. In contrast, the GP-PSO 1 finds a noticeably better solution for the Rosenbrock function in the same fashion as the BSt-PSO$^{(c)}$ does in comparison to the BSt-PSO$^{(p)}$ (even the values of the solutions are very similar).

In summary, the GP-PSO 2 finds good solutions for the Sphere, Griewank and 2-dimensional Schaffer f6 function. The solution it finds for the Rosenbrock function is as good as the one found by the BSt-PSO$^{(p)}$ (i.e. roughly acceptable), while the solution it finds for the Rastrigrin function is not very good, but still noticeably better than those found by the BSt-PSO$^{(c)}$ and the BSt-PSO$^{(p)}$.

### 9.4.2.3 The GP-PSO 3

This optimizer is composed of 3 sub-swarms, one corresponding to a BSt-PSO, another to a BSt-PSO$^{(c)}$, and the other to a BSt-PSO$^{(p)}$. It can be observed that it finds a solution for the Sphere function that is better than those found by both the GP-PSO 1 and the GP-PSO 2, and almost as good as the one found by the BSt-PSO$^{(p)}$. The reasonably good solution it manages to find for the Rosenbrock function is better than the one found by the BSt-PSO$^{(p)}$ and worse than the one found by the BSt-PSO$^{(c)}$. The weak point of this optimizer lies in that it finds the worst solution among those found by the GP-PSOs for the Rastrigrin function. It does not exhibit any problem in finding the exact solution for the 2-dimensional Schaffer f6 function along each of the 50 runs, and the solutions it finds for the Griewank and the 30-dimensional Schaffer f6 functions are slightly better than those found by the GP-PSO 1 but slightly worse than those found by the GP-PSO 2.

In summary, the GP-PSO 3 finds very good solutions for the Sphere and the 2-dimensional Schaffer f6 functions, and roughly acceptable solutions for the Rosenbrock, the Griewank, and the 30-dimensional Schaffer f6 functions. However, the solution it finds for the Rastrigrin function is not very good, although it is again noticeably better than the solutions found by the BSt-PSO$^{(c)}$ and the BSt-PSO$^{(p)}$.

### 9.4.2.4 The GP-PSO 4

This optimizer is also composed of 3 sub-swarms, one corresponding to a BSt-PSO, and the other two corresponding to optimizers with even stronger fine-clustering ability than the BSt-PSO$^{(c)}$ and the BSt-PSO$^{(p)}$ (refer to section **9.2.4**). It can be observed that while it finds very good solutions for the Sphere, the Griewank, and the 2-dimensional Schaffer f6 functions, it only finds a roughly acceptable solution for the 30-dimensional Schaffer f6 function, and very bad solutions for the Rosenbrock and the Rastrigrin functions.

The most interesting feature of this optimizer is the strong and quick fine-clustering of its particles. In fact, it is the optimizer whose "mean type of error condition attained" (refer to **Table 9. 1** to **Table 9. 6**) is closer or equal to one for every function except for the Schaffer f6 function[3]. This feature is most impressive when optimizing the Rosenbrock function: while all the other optimizers are only able to attain the second set of error conditions, the "mean type of error condition attained" by the GP-PSO 4 is very close to 1. In fact, the first set of error conditions is attained 34 times, the second set is attained only once, and the optimizer fails to attain both sets 15 times (refer to digital **Appendix 4**).

In summary, this optimizer finds good solutions for three functions only, while it finds quite bad solutions for the Rosenbrock and the Rastrigrin functions. Hence, despite its outstanding fine-clustering ability, the optimizer is not considered further hereafter. The development of some kind of patch to overcome its weaknesses, namely its quick loss of diversity, might result in a powerful optimizer. However, this is left for future work due to time constraints.

### 9.4.2.5 The GP-PSO 5

This optimizer is composed of 3 sub-swarms, one corresponding to a modified BSt-PSO 2, one corresponding to a BSt-PSO$^{(c)}$, and one corresponding to a BSt-PSO$^{(p)}$ (refer to section **9.2.5**). As opposed to the preceding GP-PSOs, the complete population of the minimizer, including the sub-swarm in charge of maintaining diversity, is considered in the computation of the termination conditions. This results in time-extended searches.

[3] This is because the Schaffer f6 function exhibits local optima in the form of ring-like depressions, which make the clustering of the particles very difficult (refer to **Appendix 4** for graphical visualization).

It can be observed that the GP-PSO 5 finds the best solution for the Sphere function among all those found by the previous GP-PSOs. In fact, it finds a solution that is also better than those of the BSt-PSO and of the BSt-PSO$^{(p)}$, and almost as good as that of the BSt-PSO$^{(c)}$. Although it takes longer than all the previous GP-PSOs, the length of the search is still reasonably short. The GP-PSO 5 also finds very good solutions for the Rosenbrock, the Rastrigrin, the Griewank, and the 30-dimensional Schaffer f6 function. It only presents some problems when optimizing the 2-dimensional Schaffer f6 function. Note, however, that despite finding one of the worst mean best solutions, it only fails to find the exact solution for the 2-dimensional Schaffer f6 function twice out of 50 runs (refer to digital **Appendix 4**).

In summary, the GP-PSO 5 appears to be a very good general-purpose optimizer: it finds either a very good or a quite good solution for every function in the test suite.

### 9.4.2.6 The GP-PSO 6

This optimizer is composed of 3 sub-swarms, one corresponding to a BSt-PSO, one corresponding to a BSt-PSO$^{(c)}$, and one corresponding to a BSt-PSO$^{(p)}$ (refer to section **9.2.6**). This optimizer differs from the GP-PSO 3 only in that the BSt-PSO is now composed of only 8 particles, and it is considered in the computation of the termination conditions. Thus, the other two sub-swarms are composed of 11 particles. In the same fashion as the GP-PSO 5, the searches carried out by this optimizer can be expected to be longer than those carried out by the GP-PSO 1 to the GP-PSO 4.

It can be observed that it is able to find slightly better solutions than the GP-PSO 5 for the Sphere and the Rosenbrock functions, and slightly worse solutions for the Rastrigrin, the Griewank, and the 30-dimensional Schaffer f6 function. Besides, it finds the exact solution for the 2-dimensional Schaffer f6 function along every run, as opposed to the GP-PSO 5.

In summary, the GP-PSO 6 also appears to be a very good general-purpose optimizer: it finds very good solutions for every function in the test suite.

### 9.4.2.7 The GP-PSO (ils)

This optimizer is based on the GP-PSO 5, with the incorporation of a rudimentary, kind of brute-force local search around the best previous experience of each of 15 selected particles (refer to section **9.2.7**, and to **Fig. 9. 7** for the pseudo-code). It should be noted that its computational cost is notably higher than that of the GP-PSO 5.

It can be observed that the solution that the GP-PSO 5 was able to find for the Sphere, the Rosenbrock and both the 2-dimensional and the 30-dimensional Schaffer f6 functions are noticeably improved by this basic local search. In contrast, the solutions that the GP-PSO (ils) finds for the Rastrigin and Griewank functions are slightly worse than those found by the GP-PSO 5.

In summary, the local search is a strategy to be considered to introduce diversity when the search stagnates, which may help to escape some local optima. However, this should be done with great care, since it seems that sometimes it can attempt against exploration.

#### 9.4.2.8 The GP-PSO (gls)

This optimizer is based on the GP-PSO 5, with the incorporation of a rudimentary, kind of brute-force local search around the best previous experience of the whole swarm (refer to section **9.2.8**, and to **Fig. 9. 8** for the pseudo-code). It should be noted that the computational cost is considerably higher than that of the GP-PSO 5, and similar to that of the GP-PSO (ils). Implementing the local search around the best previous experience of the whole swarm rather than around the best previous experience of the individual particles is expected to reduce the problem of attempting against exploration.

It can be observed that the solution this optimizer finds for the Sphere and for the Griewank functions are the same as the ones found by the GP-PSO (ils). Although the solution it finds for the Rosenbrock function is a little worse than the solution found by the GP-PSO (ils), it is still very good, and better than the solution found by the GP-PSO 5 (i.e. the local search is helpful). It is interesting to observe that, while the individual learning implemented in the GP-PSO (ils) was a little harmful, the one implemented here greatly improves the solution that the GP-PSO 5 found for the Rastrigin function. The solutions found for the 2-dimensional and for the 30-dimensional Schaffer f6 functions were also improved with respect to those found by the GP-PSO 5, although the solutions found by the GP-PSO (ils) are a little better.

In summary, this local search was helpful every time, especially for the Rastrigin function, except for the case of optimizing the Griewank function. Nevertheless, the solution it finds for the Griewank function is still very good, and only slightly worse than the one found by the GP-PSO 5. Another (insignificant) drawback is that it fails once out of the 50 runs in finding the exact global optimum of the 2-dimensional Schaffer f6 function.

### 9.4.2.9 The LGP-PSO 1

This optimizer is also based on the GP-PSO 5, with the only difference that the spread of information takes place through overlapping neighbourhoods composed of 3 particles each rather than through a single neighbourhood (refer to section **9.2.9**).

It can be seen that this optimizer finds the best solutions for the Sphere and the Rosenbrock functions, the second best for the Griewank function, the exact solution along every run for the 2-dimensional Schaffer f6 function, and a good solution for the 30-dimensional Schaffer f6 function. However, the mean best solution it finds for the Rastrigrin function is quite poor. It appears that the local version notably delays the particles' clustering: only the BSt-PSO takes longer than this optimizer in attaining the error condition when optimizing the Sphere function; it never attains the error condition when optimizing the Rosenbrock function; it is never able to attain the first set of error conditions when dealing with the Rastrigrin function; and it exhibits the highest "mean type of error condition attained" when optimizing the Griewank and the 2-dimensional Schaffer f6 functions (even higher than those corresponding to the BSt-PSO!).

In summary, this local version is able to find very good solutions for five of the six benchmark functions. However, the solution it finds for the Rastrigrin function is definitely poor. The evolutions of the relative errors and of the best and average conflicts were observed along single runs of the algorithm, showing that the particles' fine-clustering stagnates when optimizing the 30-dimensional Rastrigrin function. Therefore this optimizer is not considered further within this work, and further investigation on the local versions is recommendable.

### 9.4.2.10 The LGP-PSO 2

This optimizer is based on the GP-PSO 3, with the only difference that the spread of information takes place through overlapping neighbourhoods composed of 3 particles each rather than through a single neighbourhood (refer to section **9.2.10**).

Although the solution this optimizer finds for the Sphere function is still acceptable, it is considerably worse than the one found by the GP-PSO 3. In contrast, the solution it finds for the Rosenbrock and the Griewank functions are greatly improved, to the extent that no other optimizer finds better solutions. It finds the exact solution for the 2-dimensional Schaffer f6 function along every run (and a little faster than the LGP-PSO 1), and a good solution for the 30-dimensional Schaffer f6 function (a little better than the one found by the LGP-PSO 1, and notably better than the one found by the GP-PSO 3). Again, its weak point is the not so good solution it is able to find when dealing with the Rastrigrin function, although it improves the solution found by the GP-PSO 3. Nevertheless, it takes much longer for a slight improvement.

In summary, this optimizer is able to find very good solutions for the Rosenbrock, Griewank, and 2-dimensional Schaffer f6 function, and acceptable solutions for the Sphere and the 30-dimensional Schaffer f6 function. In contrast, it finds a poor solution, exhibiting a poor degree of clustering of its particles, for the Rastrigrin function. Same as with the LGP-PSO 1, this optimizer is not considered further within this work and further research is advisable.

## 9.4.3 Conclusion

The two ready-to-use optimizers which exhibit the best overall performances[4]—and therefore the ones to be selected as GP-PSOs—are, undoubtedly, the GP-PSO 5 and the GP-PSO 6. In addition, the GP-PSO 3 and the GP-PSO (gls) should be considered for special cases, since the former finds worse solutions but remarkably faster, and the latter is able to improve the solutions that the GP-PSO 5 finds at the expense of a considerable higher computational cost. It is important to note that the permissible computational cost is problem-dependent, and that the mere extension of the maximum number of time-steps for the search does not guarantee that a better solution is to be found. Therefore, it is an important piece of information to know that the GP-PSO (gls) is able to improve solutions if the required computational cost is available. Nevertheless, a more sophisticated the local search technique should be developed and tested in the future. The one implemented here is indeed rudimentary, and was proposed with the aim to help the optimizers to escape stagnation in sub-optimal solutions[5]. However, the size of the region of attraction that this local search technique is able to escape is deterministically and arbitrarily set (specified as $\pm 0.1\% \left( x_{\max} - x_{\min} \right)$ in **Fig. 9. 7** and **Fig. 9. 8**).

[4] Considering the solutions found for every benchmark function together with the corresponding convergence rate.
[5] A local search algorithm with the aim to fine-tune the search should also be tried.

The local versions clearly and notably delay the clustering of the particles. However, it is sometimes helpful and sometimes harmful, and the length of the search is notably increased without that always resulting in outstanding improvement. It is fair to note that little research has been carried out on the local versions of the algorithm. Namely, the plain version with the neighbourhood composed of three particles (the LGP-PSO 1 and the LGP-PSO 2 tested along the previous section), and a few others such as the local BSt-PSO$^{(p)}$ with the same type of neighbourhoods but incorporating a fourth term to the velocity update equation to consider the best previous experience of the particle, the best experience of the neighbourhood, and the best experience of the population (letting the random weights stochastically alter a more local or more global behaviour), and deterministically time-decreasing the "locality" and time-increasing the "globality" of the versions as the search progressed. However, none of these techniques improved the clustering of the particles, which ended up being very bad for the 30-dimensional Rastrigrin function. Further research on this subject matter is worthwhile.

# 9.5 Closure

A number of optimizers were designed based on the tunings of the parameters of the particles' velocity update equation that were found to be convenient either for the high convergence rate, for the ability to fine-cluster, or for the ability to escape sub-optimal solutions. Thus, six GP-PSOs were proposed and tested on the suite of benchmark functions stated in **Table 6.1**. A basic, rudimentary local search was incorporated thereafter into one of the two best GP-PSOs in two different manners: enhancing each particle's best previous experience, and enhancing the swarm's best previous experience at each time-step. A third alternative can be thought of, consisting of the incorporation of the local search into the stopping criteria, so that when the termination conditions developed along **Chapter 7** are met, the local search is activated around the best solution found, and the search is terminated if and only if the local search cannot improve the solution. This technique is left for future work due to time constraints.

Finally, local versions consisting of overlapping neighbourhoods composed of three particles each were implemented on two different GP-PSOs. The result was the delay of the clustering of the particles thus time-extending the searches. While this led to considerable improvement in some cases, it decreased the quality of the best solution found in some others.

To summarize, the incorporation of the local search and the local versions of the algorithm are strategies that are worth further investigation, and the two algorithms selected as general-purpose optimizers (for boundary-constrained optimization problems) are as follows:

- **GP-PSO$^{(s.d.w)}$**

- BSt-PSO$^{(p)}$: basic, standard PSO with $aw^{(t)} = p\left(w^{(t)}\right) \ \forall t$:
  $w^{(t)} = 0.5, \ \ iw^{(t)} = sw^{(t)} = 2 \ \ \forall t$ (minimizer - 10 particles)

- BSt-PSO$^{(c)}$: basic, standard PSO with $aw^{(t)} = 4.1 \cdot w^{(t)} \ \ \forall t$:
  $w^{(t)} = 0.7298, \ \ iw^{(t)} = sw^{(t)} = 1.49609 \ \ \forall t$ (minimizer - 10 particles)

- BStSd-PSO: basic, standard PSO with sigmoidly time-decreasing inertia weight:
  $w^{(1)} \to 0.7$, $w^{(t_{\max})} \to 0.5$, $iw^{(t)} = sw^{(t)} = 2 \ \ \forall t$ (minimizer - 10 particles)

- BSt-PSO: basic, standard PSO:
  $w^{(t)} = 0.7, \ \ iw^{(t)} = sw^{(t)} = 2 \ \ \forall t$ (maximizer - 5 particles)

- **GP-PSO$^{(c.w)}$**

- BSt-PSO$^{(p)}$: basic, standard PSO with $aw^{(t)} = p\left(w^{(t)}\right) \ \forall t$:
  $w^{(t)} = 0.5, \ \ iw^{(t)} = sw^{(t)} = 2 \ \ \forall t$ (minimizer - 11 particles)

- BSt-PSO$^{(c)}$: basic, standard PSO with $aw^{(t)} = 4.1 \cdot w^{(t)} \ \ \forall t$:
  $w^{(t)} = 0.7298, \ \ iw^{(t)} = sw^{(t)} = 1.49609 \ \ \forall t$ (minimizer - 11 particles)

- BSt-PSO: basic, standard PSO:
  $w^{(t)} = 0.7, \ \ iw^{(t)} = sw^{(t)} = 2 \ \ \forall t$ (minimizer - 8 particles)

- BSt-PSO: basic, standard PSO:
  $w^{(t)} = 0.7, \ \ iw^{(t)} = sw^{(t)} = 2 \ \ \forall t$ (maximizer - 5 particles)

The termination conditions are computed involving all the 30 particles of the minimizer, and the sigmoidly time-decreasing inertia weight is ruled by equation **(6.13)** (refer to **Chapter 6**).

For problems that require a fast solution (of course, in detriment of accuracy), an optimizer able to find solutions that are good in relation to its convergence rate must be selected. The chosen optimizer for such cases is as follows:

- **GP-PSO$^{(fast)}$**

- BSt-PSO$^{(p)}$: basic, standard PSO with $aw^{(t)} = p\left(w^{(t)}\right) \ \forall t$:
  $w^{(t)} = 0.5, \ \ iw^{(t)} = sw^{(t)} = 2 \ \ \forall t$ (minimizer - 10 particles)

- BSt-PSO$^{(c)}$: basic, standard PSO with $aw^{(t)} = 4.1 \cdot w^{(t)} \ \ \forall t$:
  $w^{(t)} = 0.7298, \ \ iw^{(t)} = sw^{(t)} = 1.49609 \ \ \forall t$ (minimizer - 10 particles)

- BSt-PSO: basic, standard PSO:
  $w^{(t)} = 0.7, \ \ iw^{(t)} = sw^{(t)} = 2 \ \ \forall t$ (minimizer - 10 particles)

- BSt-PSO: basic, standard PSO:
  $w^{(t)} = 0.7, \ \ iw^{(t)} = sw^{(t)} = 2 \ \ \forall t$ (maximizer - 5 particles)

The termination conditions are computed considering only the 2 sub-swarms that possess the ability to fine-cluster (20 particles), while the other 10 particles are in charge of keeping diversity and of escaping sub-optimal solutions.

For problems that require a more accurate solution and can afford a considerably higher computational cost, a local search such as the one shown in **Fig. 9. 8** is recommended.

The last issue pending for the development of a general-purpose optimizer is the incorporation of some technique to handle the constraints that real-world problems are typically subject to. Therefore, the next chapter is entirely devoted to the development of constraint-handling techniques that are suitable for particle swarm optimizers.

# Chapter 10

# CONSTRAINED PARTICLE SWARM OPTIMIZATION

The very important issues of the understanding of the behaviour of the system, the development of stopping criteria, and the design of general-purpose unconstrained particle swarm optimizers (except for hypercube-like boundary constraints) were dealt with as far as the previous chapter. Given that real-world problems are typically subject to a number of constraints, the last issue pending to turn the original algorithm into an optimizer suitable for real-world applications is the incorporation of some technique to handle the constraints. Therefore, a few possible techniques are scarcely discussed within this chapter, implemented on a general-purpose unconstrained optimizer, and tested on a suite of three benchmark constrained optimization problems. A more in depth work on constraint-handing techniques is left for future work.

## 10.1 Introduction

Numerous constraint-handling techniques can be found in the literature, most of which are modified versions of preceding ones, such modifications being typically "author-dependent". The present chapter is dedicated to the discussion of the main features of a few existing popular techniques, to the implementation of a few proposed ones, and to their incorporation into a general-purpose unconstrained optimizer brought from the previous chapter (namely, the GP-PSO$^{(s.d.w.)}$). Finally, the resulting general-purpose constrained optimizers are tested on a suite of three benchmarking constrained optimization problems taken from [81].

It is important to remark that this chapter does not intend to be a treatise on the numerous existing methods, nor it intends to re-invent the wheel. The aim is considerably more modest: to develop a general-purpose method that requires a few adaptations or none to deal with most problems, profiting from some already existing methods. Thus, before the implementation of the proposed constraint-handling techniques, it seems fair to briefly discuss the main concepts underlying constrained optimization, and the main features of the most popular methods.

# 10.2 Constrained optimization

Let $\mathcal{S}:\mathcal{F}\cup\mathcal{I}$ be the search-space, where $\mathcal{F}$ and $\mathcal{I}$ are its feasible and infeasible parts, respectively. The feasible part is defined as the aggregation of all the valid candidate solutions to the problem (in other words, all the candidate solutions that do not violate any constraint).

An optimization problem consists of finding the vector $\hat{\mathbf{x}}\in\mathcal{F}$ such that $f(\hat{\mathbf{x}})\leq f(\mathbf{x})\ \ \forall\mathbf{x}\in\mathcal{F}$ (assuming minimization). Therefore, an optimization problem can be formulated as follows:

$$\begin{aligned}&\text{Minimize } f(\mathbf{x})\\&\text{subject to } \mathbf{x}\in\mathcal{F}\end{aligned} \tag{10. 1}$$

More precisely, the problem can be rewritten as shown in equation **(10. 2)**:

$$\begin{aligned}&\text{Minimize } f(\mathbf{x})\\&\text{subject to } \begin{cases} g_j(\mathbf{x})\geq 0 & ; \quad j=1,\ \dots\ ,q\\ g_j(\mathbf{x})=0 & ; \quad j=q+1,\ \dots\ ,m\end{cases}\end{aligned} \tag{10. 2}$$

Where:

- $\mathbf{x}\in\mathcal{S}$ is the vector of object variables
- $f(\cdot):\mathcal{S}\rightarrow\mathcal{E}$ is the function to be optimized
- $g_j(\cdot)$ are the constraint functions
- $\mathcal{S}\subseteq\mathcal{R}^n\ \wedge\ \mathcal{E}\subseteq\mathcal{R}$ (this thesis is only concerned with continuous optimization problems)
- $\mathcal{R}$ is the set of real numbers

For a general constrained optimization problem, $\mathcal{F}$ can be defined as:

$$\mathcal{F}=\{\mathbf{x}\in\mathcal{S}\mid g_j(\mathbf{x})\geq 0\ \ \forall j\in\{1,\dots,q\}\ \wedge\ g_j(\mathbf{x})=0\ \ \forall j\in\{q+1,\dots,m\}\} \tag{10. 3}$$

An optimization problem can be formulated in many ways, being the one stated in equation **(10. 2)** a very general one. In fact, all the constraints can be represented by inequalities only, since two inequalities can represent equality. Besides, since $\min\{f(\mathbf{x})\}=-\max\{-f(\mathbf{x})\}$, the problem can be formulated either as a minimization or a maximization problem. Finally, the constraints of the type "greater than or equal to" can be turned into "less than or equal to" constraints by defining the function $k_j(\mathbf{x})=-g_j(\mathbf{x})$, so that if $g_j(\mathbf{x})\geq 0 \quad\Rightarrow\quad k_j(\mathbf{x})\leq 0$.

### 10.2.1 Types of constraints

The most appropriate techniques to handle a constraint frequently depend on the type of constraint at issue. An important aspect in this regard is to define which of the following groups the constraint belongs to:

- **Inequality constraint**

This constraint consists of a function of the object variables that must be "greater than or equal to" a constant, as shown in equations **(10. 2)** and **(10. 3)**.

- **Equality constraint**

This constraint consists of a function of the object variables that must be equal to a constant, as shown in equations **(10. 2)** and **(10. 3)**.

- **Boundary constraint**

This is typically a particular case of the inequality constraints, which consists of interval limits that define the boundaries of the feasible search-space. The most frequent boundary constraints consist of setting a feasible hyper-cube $x_{\min} \leq x_i \leq x_{\max}$ (or $\mathbf{x} \in [x_{\min}, x_{\max}]^n$, with $n$ being the dimension of the search-space), where the interval confinement is the same for every variable. If the cases of $x_{\min} = -\infty$ and $x_{\max} = \infty$ are considered, every optimization problem can be viewed as having boundary constraints. It is fair to note, however, that the boundaries can also be defined by a hyper-rectangle, or by any other set of arbitrary functions.

While a constrained optimization problem has been defined as the problem of finding the combination of variables that minimize the objective function and satisfy all the constraints, real-world problems sometimes do not lend themselves to such strict conditions. Frequently, all the constraints cannot be strictly satisfied at the same time, so that the problem turns into finding the trade-off between minimizing the conflict function and minimizing the constraint-violations. Thus, depending on the problem at hand, a constraint can be also classified as:

- **Hard constraint**

This is a kind of constraint that does not admit any degree of violation. That is to say that, if there is no solution to the problem that can comply with all the constraints, so that a trade-off between the minimizations of the conflict function and of the constraint-violations needs to be searched for, a violation to this type of constraint is still unacceptable.

- **Soft constraint**

This is a type of constraint that should be complied with as much as possible. However, it accepts some degree of violation when the problem cannot be solved otherwise. Although the problems with soft constraints can be viewed and handled as multi-objective optimization problems—where one objective is to minimize the conflict function and another to minimize the violation of the constraints[1]—, multi-objective optimization is beyond the scope of this thesis. The appropriate technique to handle a set of constraints, and sometimes even the settings of the parameters of a given technique, are problem-dependent. Some of the most popular techniques are outlined hereafter, although each of them presents numerous versions.

## 10.2.2 Constraint-handling techniques

Only some methods which are suitable for particle swarm optimizers are discussed hereafter, some of which are extensively used in evolutionary algorithms. Notice that, although some methods are regarded as belonging to different groups, the boundaries between them are frequently blurry. In fact, similar methods are often called differently and different methods are called similarly by different authors in the literature.

- **Rejection of infeasible solutions**

The infeasible solutions are simply eliminated from the population. It can be seen as a kind of "penalty method", where the penalization is not a "fine" but the "death penalty". There are different possible implementations such as removing the infeasible solution and randomly generating a replacement within a feasible region. This method presents some problems such as the fact that the particles need to be initialized within feasible space, and cannot pass through infeasible regions to enter another feasible disjointed region of the search-space. In addition to that, equality constraints cannot be handled by this method.

[1] Note that the concept of minimizing the violation of constraints is not an easy task. For instance, is it a more serious violation to slightly violate a high number of constraints or to strongly violate a few ones?

- **Preserving feasibility**

This technique allows the particles to fly over infeasible regions of the search-space, although their ability to explore it is limited because they are not allowed to remember experiences over infeasible space as best experiences. Therefore, the particles are quickly pulled back to the feasible region. The method could be easily seen as belonging to the group of methods that "reject infeasible solutions", but it is called a "preserving feasibility" technique here in agreement with Hu et al. [40].

The drawbacks are that the particles need to be initialized within feasible space (although it is not strictly necessary that all of them are), that it is typically inefficient in dealing with small and disjointed feasible spaces, and that it cannot handle equality constraints. The advantage of the method is in that it only requires a slight adaptation from the unconstrained algorithm, and there is no parameter to be tuned differently for different problems.

- **Cut off at the boundary**

This technique is straightforward for hyper-cube-like boundary constraints, and it resembles the velocity constraint to the components of the particles' velocity.

When a particle moves from a feasible location to an infeasible one, the vector of displacement is cut off. Two alternatives can be thought of in this regard: either to locate the particle in the feasible position that is nearest to the infeasible position the particle intended to move to in the first place (refer to **Fig. 10. 1** - left), or to keep the direction of the vector of displacement unmodified so as to interfere in the dynamics of the swarm the least possible (refer to **Fig. 10. 1** - right). As previously mentioned, this technique resembles the velocity constraint of the canonical PSO, which considered the first alternative: the components of the particles' velocity were limited so that the direction of the vector of displacement was almost always altered every time the velocity constraint was activated. Likewise, the first alternative is usually preferred here, whose algorithm is given by equation **(10. 4)**:

$$\begin{aligned} &\textbf{if} \quad x_{ij}^{(t)} > x_{\max} \;\Rightarrow\; x_{ij}^{(t)} = x_{\max} \\ &\textbf{elseif} \quad x_{ij}^{(t)} < x_{\min} \;\Rightarrow\; x_{ij}^{(t)} = x_{\min} \end{aligned} \qquad \textbf{(10. 4)}$$

Notice that, while the application of this technique is indeed straightforward for hyper-cube-like boundary constraints, it needs to be specifically adapted for other types of constraints.

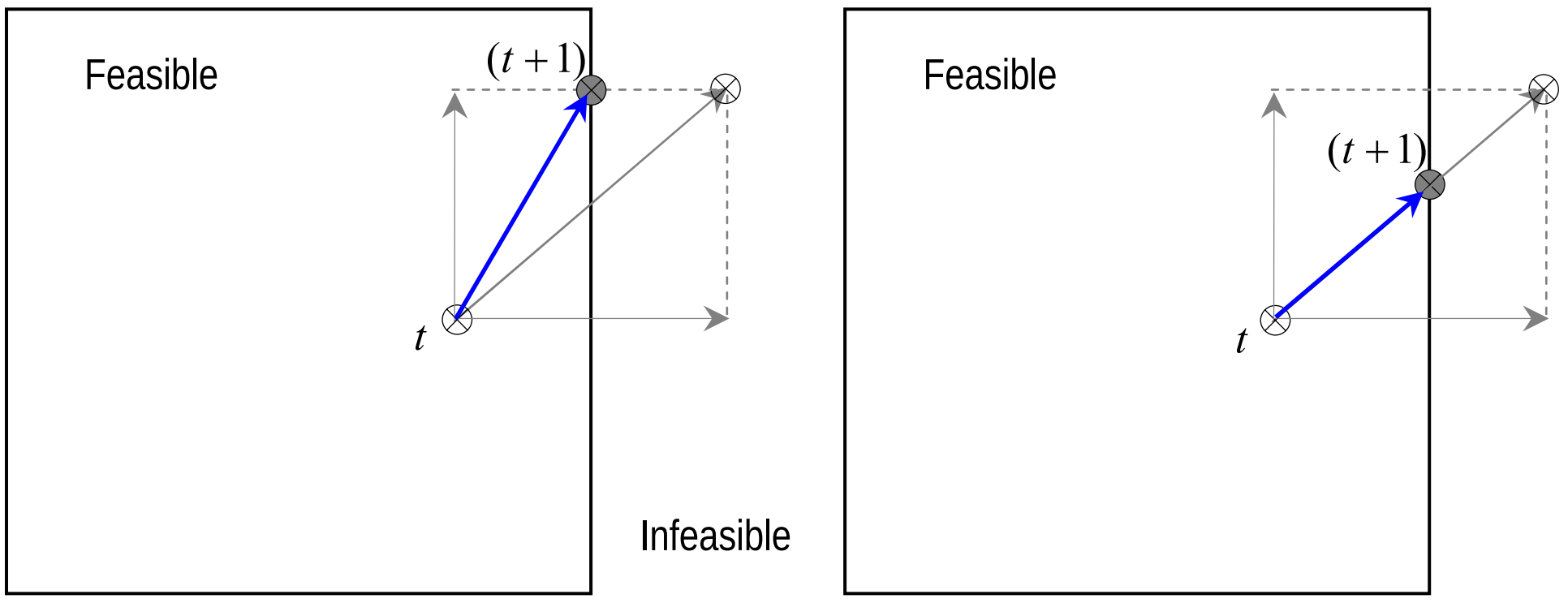

**Fig. 10. 1**: Two alternative "cut off at the boundary" techniques. On the left, the particle is relocated in the feasible point that is nearest to the infeasible solution the particle originally intended to move to. On the right, the vector of displacement is cut off in such a way that its direction remains unchanged. Both particles end up located on the boundary.

- **Bisection method**

The "cut off at the boundary" technique works very well when the boundaries are constrained by a hyper-cube, and when the solution is located on the boundary. However, the particles usually get trapped in the boundaries when the solution is located near them. Foryś et al. [36] proposed a "reflection technique" rather than cutting off the trajectory. Thus, a number of control points are placed between the particle's last feasible position and the intersection of the displacement vector with the boundary, and the objective function is evaluated on every point. The one with the best conflict value is selected as the next location.

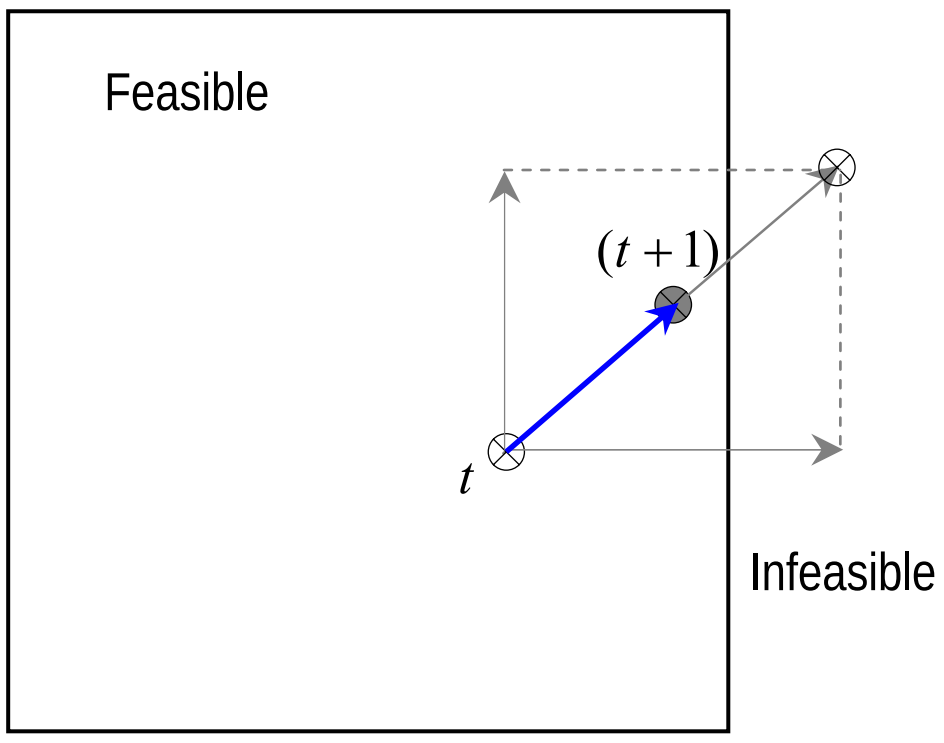

**Fig. 10. 2**: Sketch of a possible constraint-violation handled by the bisection method.

A simpler strategy is proposed here, which resembles both Foryś et al.'s control points [36] and the classical bisection method used for root-finding: if the new position of the particle is located over infeasible space, the displacement vector is split in halves, and the constraints are verified in the middle. If the new position is still infeasible, the vector is split again, and so forth. Eventually, a feasible position will be found, unless the particle is already on the boundary. Hence an upper limit of iterations is stated, and the particle keeps its last feasible position if no new feasible location can be found.

The drawbacks of this method are that it cannot pass through infeasible space to search for new, disjointed feasible space; that the particles need to be initialized within feasible space[2]; and that the method is not applicable for equality constraints straightaway, although some adaptation can be thought of. These drawbacks are shared with the "cut off at the boundary" technique and to some extent with the "preserving feasibility" method[3].

The advantage of the "bisection" method—also shared with the "preserving feasibility" method—is that it does not require adaptations to handle different inequality constraints. In other words, it is a general-purpose technique that is in principle applicable to most problems, although its capabilities can be expected to be limited. Another characteristic feature of the method is that it results in a notably fast clustering of the particles.

With regards to the classification of the different techniques, it is fair to note that, according to the definitions given by Engelbrecht [81], the "preserving feasibility" methods are those which ensure that adjustments to the particles do not violate any constraint. Hence the "cut off at the boundary" technique and the "bisection" method proposed here could be considered as "preserving feasibility" methods, while the "preserving feasibility" method discussed above could be considered either as a method that "rejects infeasible solutions" or as a "repair method", which Engelbrecht [81] defines as methods that …*allow particles to move into infeasible space, but special operators (methods) are then applied to either change the particle into a feasible one or to direct the particle to move towards feasible space. These methods usually start with initial feasible particles*. As mentioned before, the nomenclature is not uniform.

- **Penalization methods**

It is sometimes necessary to evaluate infeasible solutions in order to guide the search towards more promising areas. In this alternative, the particles searching the infeasible space are evaluated, but the particle's conflict is increased if the solution is infeasible. The algorithm works just as if the problem was unconstrained. Besides, assigning different penalizations to the different constraint violations allows handling soft and hard constraints, so that the optimizers can deal with real-world problems where all the theoretical constraints cannot be complied with simultaneously. The drawback is that it requires a problem-dependent tuning of its parameters. Thus, the (penalized) function to be minimized is given by:

$$fp(\mathbf{x}) = f(\mathbf{x}) + Q(\mathbf{x}) \qquad \textbf{(10. 5)}$$

[2] This is a serious drawback for problems subject to numerous and restrictive constraints. It might be practically impossible to randomly initialize a swarm of feasible solutions (a deterministic initialization might be considered).

[3] The "preserving feasibility" method could be able to pass through small areas of infeasible space thanks to the inertia of the particles, but without using the information gained from flying over the infeasible space.

Where:
- $fp(\mathbf{x})$: penalized conflict of particle **x**.
- $f(\mathbf{x})$: conflict of particle **x**.
- $Q(\mathbf{x})$: penalty for infeasible particle **x**.

Often penalties are not fixed but linked to the amount of infeasibility of the particle. They might simply be functions of the number of constraints violated, but functions of the distance from feasibility are usually preferred. For instance, for optimization problems of the form:

$$\begin{aligned}&\text{Minimize } f(\mathbf{x})\\&\text{with } \mathbf{x} \in \mathcal{R}^n\end{aligned} \qquad \textbf{(10. 6)}$$

Where:
- $g_j(\mathbf{x}) \le 0 \quad ; \quad j = 1, \dots, q$
- $g_j(\mathbf{x}) = 0 \quad ; \quad j = q+1, \dots, m$

The degrees of infeasibility might be taken into account by constraint-violation measures:

$$f_j(\mathbf{x}) = \begin{cases} \max\{0, g_j(\mathbf{x})\} & ; \quad 1 \le j \le q \\ g_j(\mathbf{x}) & ; \quad q < j \le m \end{cases} \qquad \textbf{(10. 7)}$$

Then, the conflict function to be optimized might be computed, for instance, as follows:

$$fp(\mathbf{x}) = f(\mathbf{x}) + \sum_{j=1}^{m} \lambda_j^{(t)} \cdot \left(f_j(\mathbf{x})\right)^{\alpha} \qquad \textbf{(10. 8)}$$

Where $\lambda_j^{(t)}$ and $\alpha$ are the coefficients of penalization. .

Typically, $\lambda_j^{(t)}$ is set to a high value and $\alpha$ to a small one. The coefficient $\lambda_j^{(t)}$ also serves the function of setting the relative importance desired for the different constraints, for instance, to deal with soft and hard constraints.

Note that if $g_j(\mathbf{x}) \leq 0 \quad \forall j = 1, \ldots, q \quad \wedge \quad h_j(\mathbf{x}) = 0 \quad \forall j = q+1, \ldots, m \quad \Rightarrow \quad f_j(\mathbf{x}) = 0 \;\; \forall j$,

$$\Rightarrow \quad fp(\mathbf{x}) = f(\mathbf{x}).$$

A high penalization might lead to converging towards a sub-optimal but feasible solution, while a low penalization might lead to the particles violating constraints but exhibiting lower conflicts than those of feasible particles. The proper definition of the penalty functions is not trivial, and it plays a critical role in the performance of the algorithm.

Notice that the computation of the penalized function as shown in **(10. 8)** is only one of many penalization techniques. For instance, $\lambda$ might be the same for every constraint; it might be kept the same throughout the whole search; some deterministic update can be thought of; or some self-adapting (ideally, parameter-free) penalization coefficients can be developed.

# 10.3 Benchmark constrained optimization problems

A few constraint-handling techniques are incorporated into the GP-PSO$^{(s.d.w)}$ (for the details of this optimizer, refer to **Chapter 9**), and the resulting optimizers are tested on the following suite of three benchmark problems (taken from [81]):

## 10.3.1 First benchmark constrained optimization problem

Minimize $f(\mathbf{x}) = 100 \cdot \left(x_2 - x_1^{\,2}\right)^2 + \left(1 - x_1\right)^2$ **(10. 9)**

$$\text{Subject to} \begin{cases} -x_1 - x_2^{\,2} \leq 0 \\ -x_1^{\,2} - x_2 \leq 0 \\ -x_1 - 0.5 \leq 0 \\ x_1 - 0.5 \leq 0 \\ x_2 - 1 \leq 0 \end{cases} \qquad \textbf{(10. 10)}$$

The following constraints are added here: $x_i \in \left[x_{\min}, x_{\max}\right]^n = \left[-0.5, 1.5\right]^2$ **(10. 11)**

The solution is given by: $\hat{\mathbf{x}} = (0.5, 0.25) \quad \wedge \quad f(\hat{\mathbf{x}}) = 0.25$ **(10. 12)**

## 10.3.2 Second benchmark constrained optimization problem

Minimize $f(\mathbf{x}) = (x_1 - 2)^2 - (x_2 - 1)^2$ **(10. 13)**

Subject to $\begin{cases} x_1{}^2 - x_2 \le 0 \\ x_1 + x_2 - 2 \le 0 \end{cases}$ **(10. 14)**

The following constraints are added here: $x_i \in [-2,2]^2$ **(10. 15)**

The solution is given by: $\hat{\mathbf{x}} = (1,1) \quad \wedge \quad f(\hat{\mathbf{x}}) = 1$ **(10. 16)**

## 10.3.3 Third benchmark constrained optimization problem

Minimize $f(\mathbf{x}) = 5 \cdot \left( x_1 + x_2 + x_3 + x_4 - \sum_{i=1}^{4} x_i{}^2 \right) - \sum_{i=5}^{13} x_i$ **(10. 17)**

Subject to
$$\begin{cases} 2 \cdot x_1 + 2 \cdot x_2 + x_{10} + x_{11} - 10 \le 0 \\ 2 \cdot x_2 + 2 \cdot x_3 + x_{11} + x_{12} - 10 \le 0 \\ 2 \cdot x_1 + 2 \cdot x_3 + x_{10} + x_{12} - 10 \le 0 \\ -8 \cdot x_2 + x_{11} \le 0 \\ -8 \cdot x_1 + x_{10} \le 0 \\ -8 \cdot x_3 + x_{12} \le 0 \\ -2 \cdot x_4 - x_5 + x_{10} \le 0 \\ -2 \cdot x_8 - x_9 + x_{12} \le 0 \\ -2 \cdot x_6 - x_7 + x_{11} \le 0 \\ -x_j \le 0 \quad \text{for} \quad j = 1,\ldots,13 \\ x_j - 1 \le 0 \quad \text{for} \quad j = 1,\ldots,9 \\ x_j - 100 \le 0 \quad \text{for} \quad j = 10, 11, 12 \\ x_{13} - 1 \le 0 \end{cases}$$ **(10. 18)**

Whose solution is given by:

$\hat{\mathbf{x}} = (1,1,1,1,1,1,1,1,1,3,3,3,1) \quad \wedge \quad f(\hat{\mathbf{x}}) = -15$ **(10. 19)**

# 10.4 Proposed constraint-handling techniques

Following the main concepts discussed along section **10.2.2**, seven constraint-handling techniques are implemented and tested on the three benchmark constrained optimization problems stated along section **10.3**.

## 10.4.1 Preserving feasibility method

This is the original "preserving feasibility" technique, as proposed by Hu et al. [40]: all the particles are iteratively, randomly initialized within feasible space, search the whole search-space, but only keep track of feasible solutions.

The history of the particles' positions for the GP-PSO$^{(s.d.w.)}$ with the incorporation of the "preserving feasibility" technique optimizing the first benchmark constrained optimization problem—together with a colour-map of the conflict function—can be seen in **Fig. 10. 3**. Recall that the solution found by the minimizer is the solution to the problem, while the solution found by the maximizer is used for the computation of the stopping criteria.

It can be observed that the particles are initialized within feasible search-space (blue and cyan dots), that the swarms of the maximizer and of the minimizer split, and that they both manage to converge. Notice that the particles extensively fly over infeasible space, and that they successively fly past the solution from different angles. The region around the solution is conveniently explored, which could be very beneficial if the solution was near but not on the boundary.

## 10.4.2 Preserving feasibility + resetting velocity method

This is a slight modification to the original "preserving feasibility" technique: the particles are iteratively, randomly initialized within feasible space, fly over the whole search-space only keeping track of feasible solutions, and their velocities are re-set to zero when they are flying over infeasible space. Thus, they are pulled back into feasible space faster, although they lose the already limited capability of exploring infeasible space, and so the corresponding chances of finding new (perhaps disjointed) feasible regions.

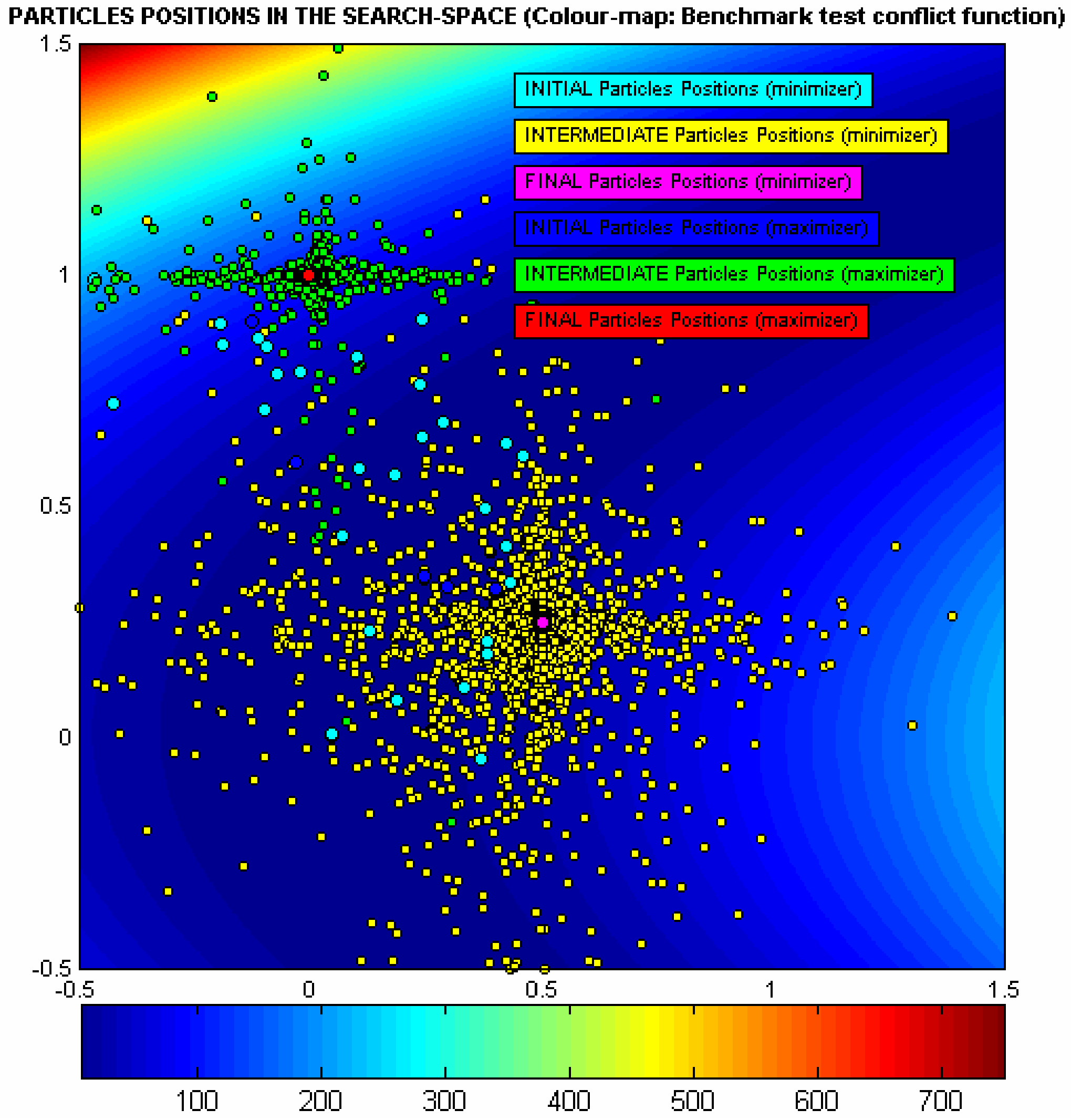

**Fig. 10. 3**: The GP-PSO$^{(s.d.w.)}$ optimizing the first benchmark constrained optimization problem, where the constraint-handling method is the pure "preserving feasibility" technique.

The history of the particles' positions for the GP-PSO$^{(s.d.w.)}$ with the incorporation of the "preserving feasibility + re-setting velocity" technique optimizing the first benchmark constrained optimization problem—together with a colour-map of the conflict function—can be seen in **Fig. 10. 4**.

It can be observed that the particles are initialized within feasible search-space (blue and cyan dots), that the swarms of the maximizer and of the minimizer split, and that they both converge. The behaviour of the swarm is very similar to that of the pure "preserving feasibility" technique, although the particles perform a less extensive exploration of infeasible space, and concentrate more around the solution.

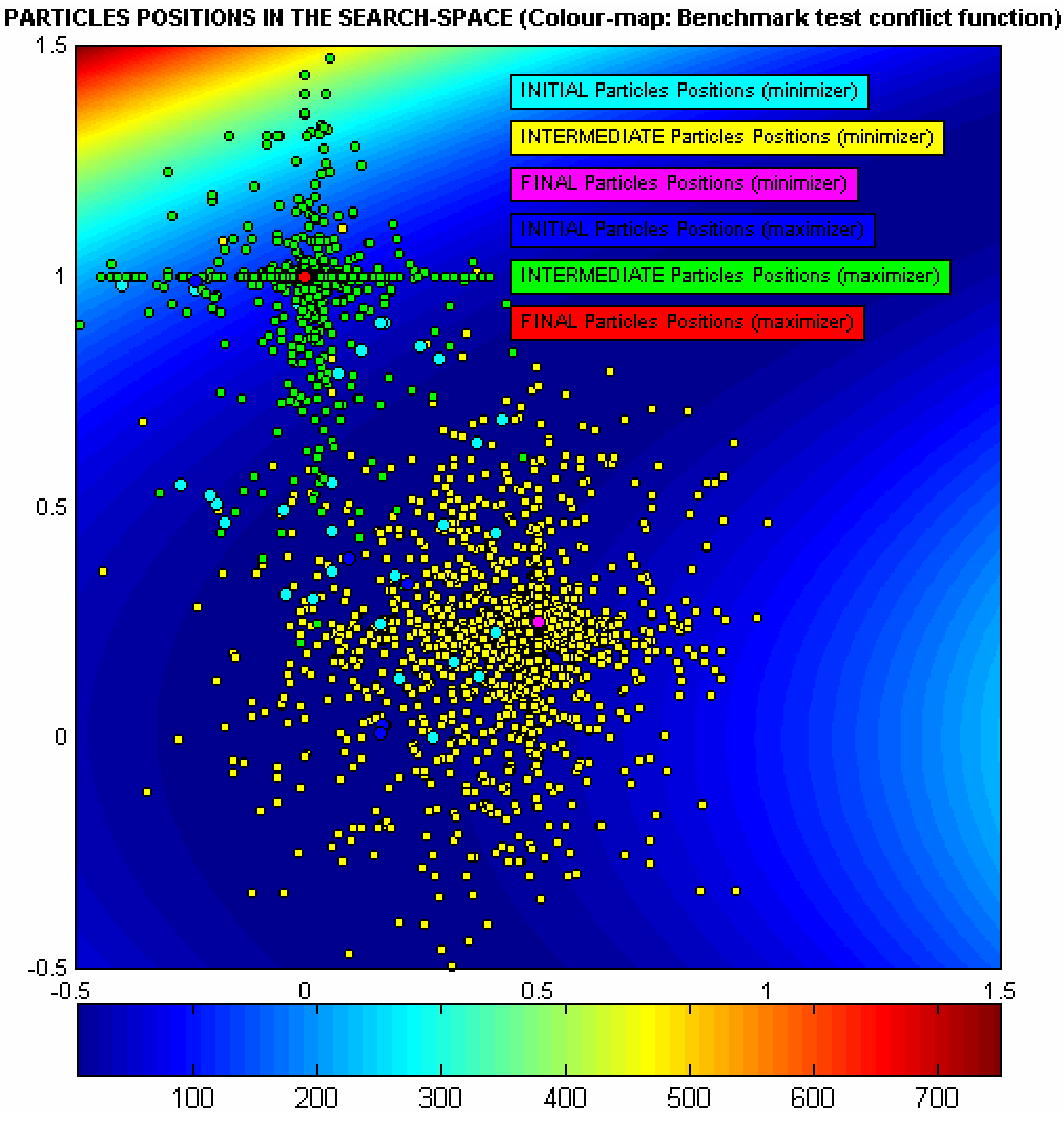

**Fig. 10. 4**: The GP-PSO$^{(s.d.w.)}$ optimizing the first benchmark constrained optimization problem, where the constraint-handling method is the "preserving feasibility + re-setting velocity" technique.

## 10.4.3 Preserving feasibility + cut off + resetting velocity method

This technique brings together the flexibility of the "preserving feasibility" technique to handle general inequality constraints with the "cut off at the boundary" technique to handle hyper-cube-like boundary constraints. The particles need to be initialized iteratively until they are all placed within feasible space. However, if the initialization places a particle out of the hyper-cube-like boundaries, the "cut off at the boundary" technique relocates the particle on a boundary. While this makes the random initialization a little easier, it might happen that most particles are located on the boundaries at the initial time-step, which would result in the lack of ability to explore the search-space. Nevertheless, the interval constraints to the object variables are usually clearly set for real-world problems, making the initialization of the particles within those intervals quite straightforward. The initialization still needs to be successively carried out until all the other constraints are complied with, although it is also possible to specify only a subset of the swarm that must be initialized within feasible space (and the other particles would eventually follow) so as to ease the initialization process.

The history of the particles' positions for the GP-PSO$^{(s.d.w.)}$ with the incorporation of the "preserving feasibility + cut off + re-setting velocity" technique optimizing the first benchmark constrained optimization problem—together with a colour-map of the conflict function—can be seen in **Fig. 10. 5**:

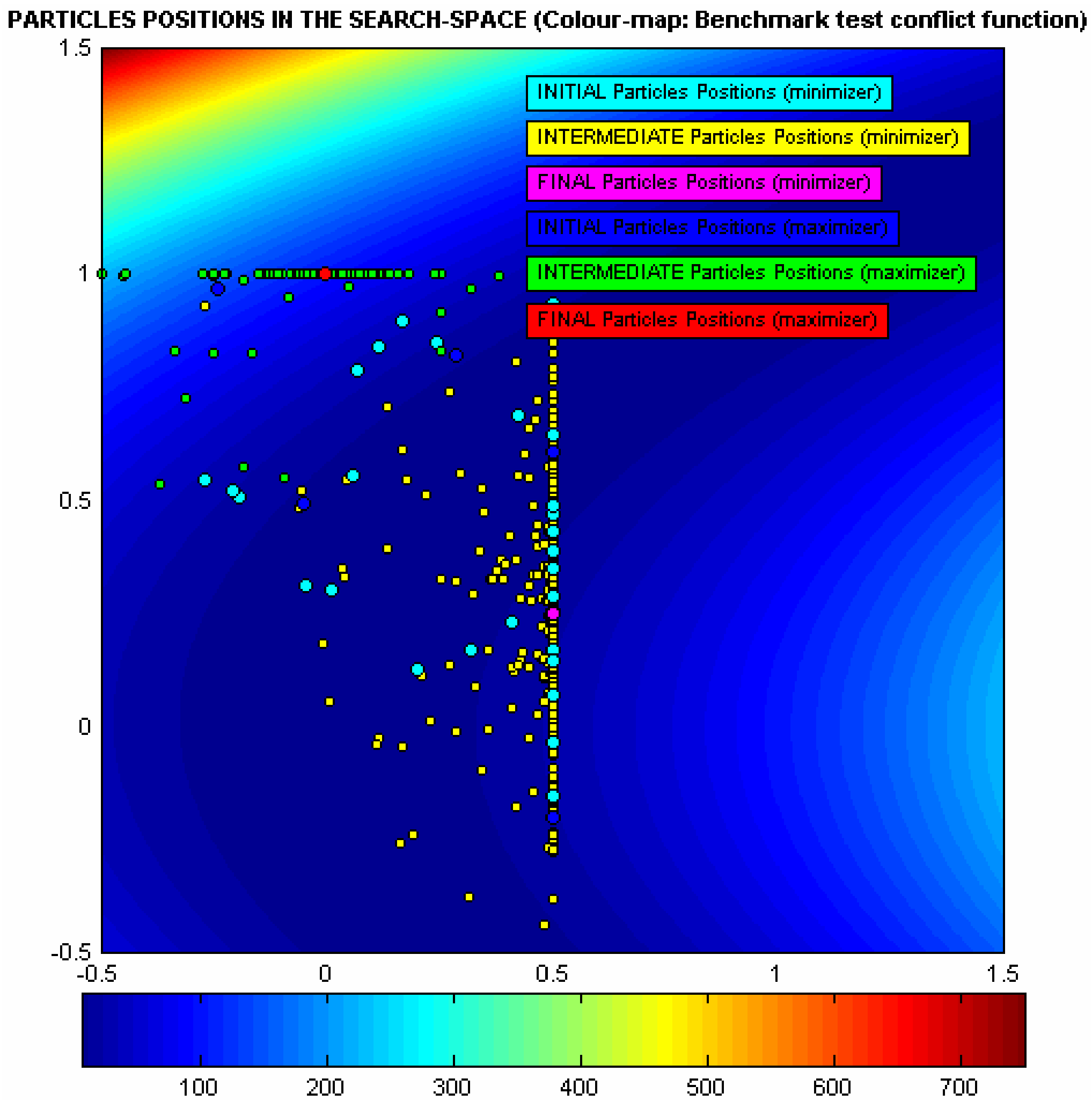

**Fig. 10. 5**: The GP-PSO$^{(s.d.w.)}$ optimizing the first benchmark constrained optimization problem, where the constraint-handling method is the "preserving feasibility + cut off + re-setting velocity" technique.

Note that the particles are not allowed to fly out of the boundaries. Since the solution is on the boundary, the technique works very well. However, when the solution is located near the boundaries, the particles might find it impossible to escape them, thus moving only along the boundaries.

### 10.4.4 Bisection method

This technique is similar to the "preserving feasibility" method in that it does not require adaptations to handle different inequality constraints, and it is similar to the "cut of at the boundary" technique in that it does not allow the particles to fly over infeasible space. The particles are iteratively, randomly initialized within feasible space. Then, whenever a particle attempts to move to an infeasible position, the displacement vector is successively cut in halves until a feasible solution is found.

The history of the particles' positions for the GP-PSO$^{(s.d.w.)}$ with the incorporation of the "bisection" method optimizing the first benchmark constrained optimization problem—together with a colour-map of the conflict function—can be seen in **Fig. 10. 6**.

### 10.4.5 Penalization method

In the same fashion as the "preserving feasibility" method does, the "penalization" method allows the particles to fly over infeasible space. However, in contrast to the former, the latter explores the infeasible search-space by profiting from the information extracted from it.

Since the preset work does not intend to develop a sophisticated penalization method but to show its applicability to the particle swarm optimizers, a very simple version is implemented hereafter. Namely, it is a static penalization method, whose coefficients are the same for every object variable. Hence, the computation of the penalization function is as follows:

$$f_j(\mathbf{x}) = \max\{0, g_j(\mathbf{x})\} \quad \text{(only inequality constraints are considered here)} \qquad \textbf{(10. 20)}$$

$$fp(\mathbf{x}) = f(\mathbf{x}) + 10^8 \cdot \sum_{j=1}^{m} \left(f_j(\mathbf{x})\right)^2 \qquad \textbf{(10. 21)}$$

This very simple, static penalization method was previously used by Venter et al. [77]. More sophisticated methods would consider low penalizations at the early stages of the search to favour exploration of infeasible space, and higher penalizations as the search progresses to favour exploration of the feasible regions. Even more sophisticated methods would consider self-adapting coefficients.

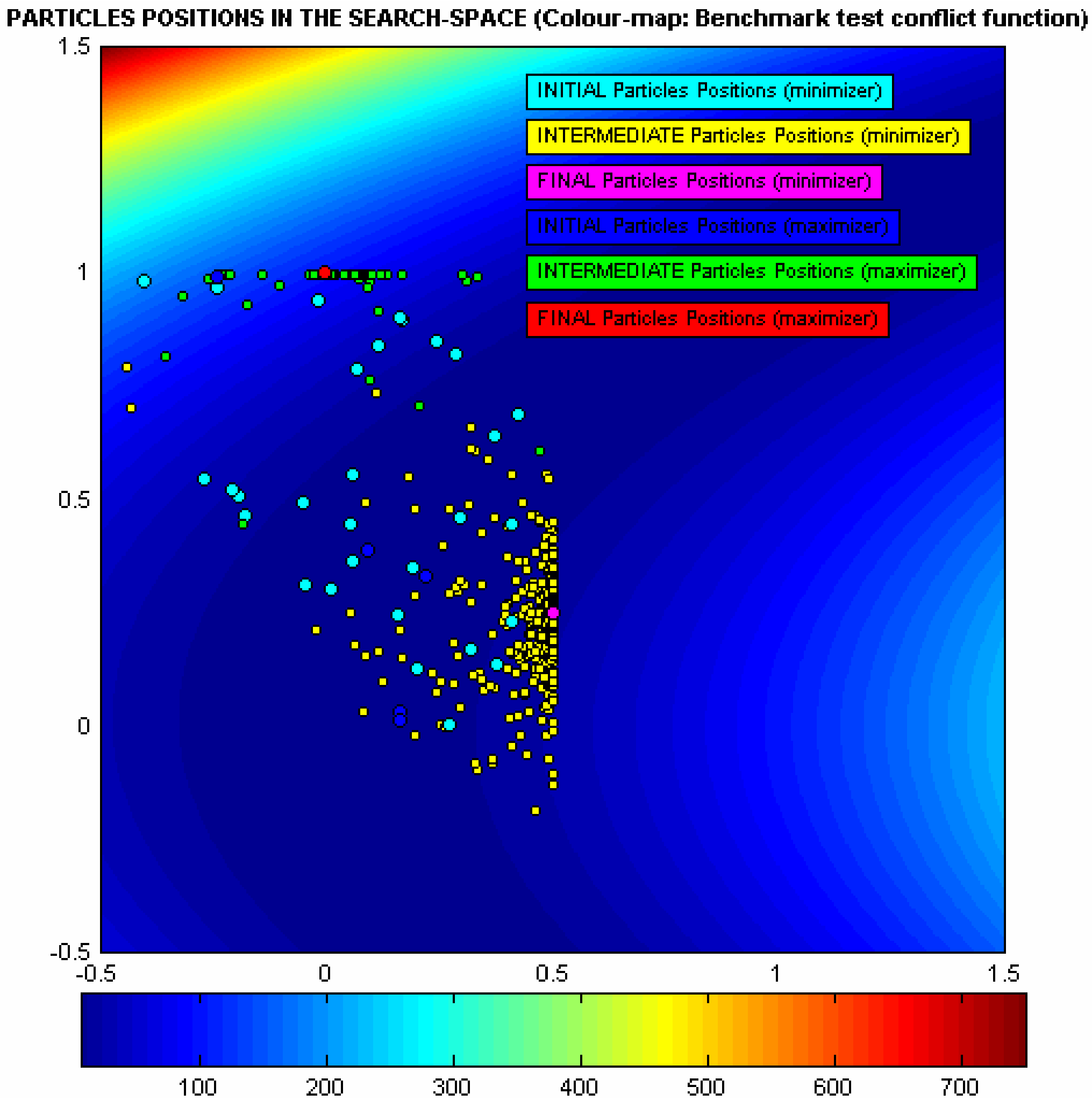

**Fig. 10. 6**: The GP-PSO$^{(s.d.w.)}$ optimizing the first benchmark constrained optimization problem, where the constraint-handling method is the "bisection" technique.

It is important to note that the penalization method makes some modifications to the optimizer necessary. Namely, the sub-swarm that seeks the best and the one that seeks the worst conflicts were sub-swarms that had access to each other's memories. Instead, the sub-swarms need to be turned now into swarms completely independent from one another to incorporate the penalization method. This is because an infeasible particle of the minimizer is penalized by adding, while an infeasible particle of the maximizer is penalized by subtracting a high penalty. Thus, a penalized particle of the maximizer might present itself as having a better conflict than all particles of the minimizer, which would then follow an infeasible particle. Likewise, the particles of the maximizer might follow an infeasible particle of the minimizer.

The history of the particles' positions for the GP-PSO$^{(s.d.w.)}$ with the incorporation of this "penalization" method optimizing the first benchmark constrained optimization problem—together with a colour-map of the conflict function—can be seen in **Fig. 10. 7**:

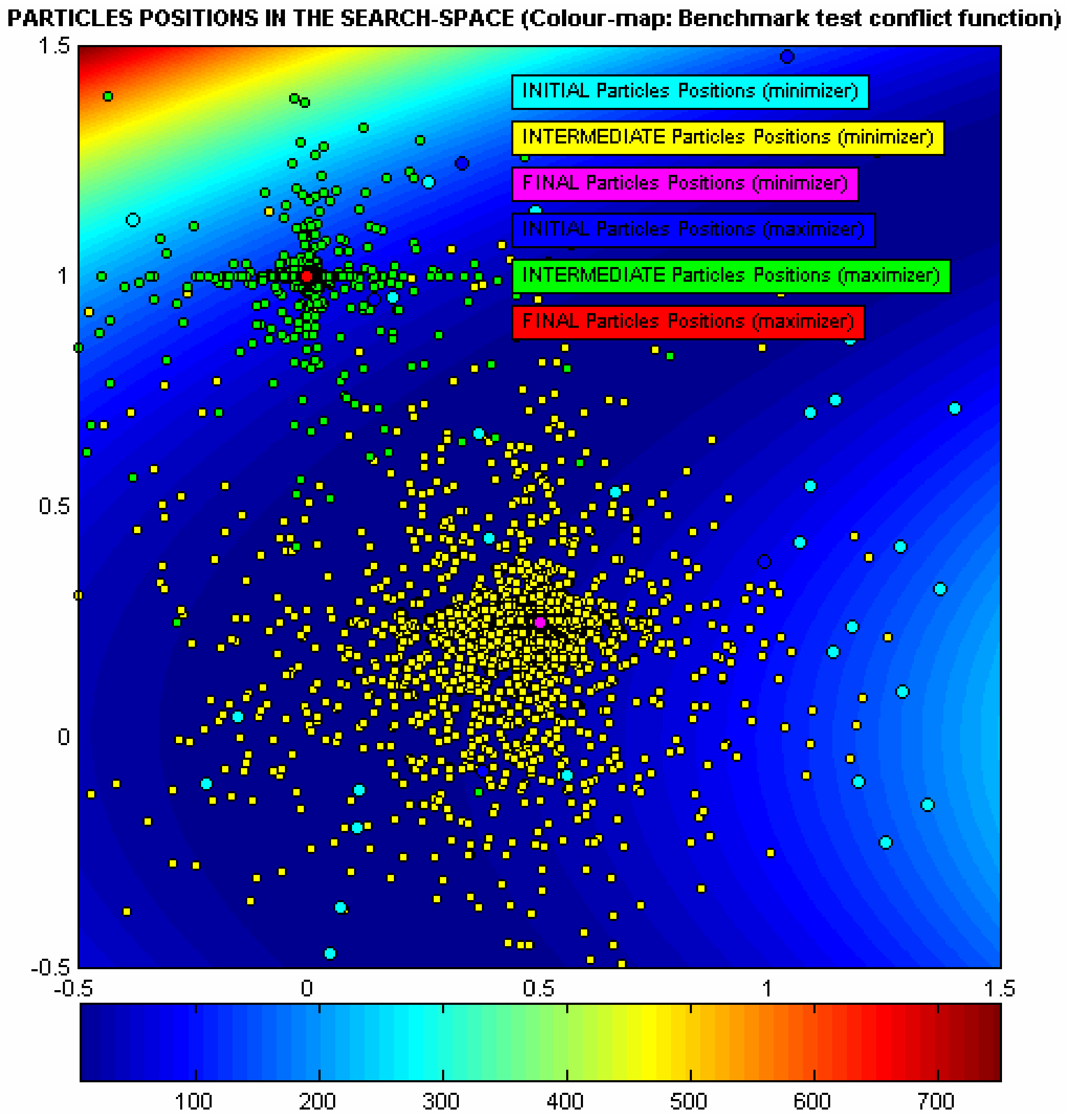

**Fig. 10. 7**: The GP-PSO$^{(s.d.w.)}$ optimizing the first benchmark constrained optimization problem, where the constraint-handling method is the "penalization" method.

It can be observed that an extensive exploration of the infeasible space is performed, and that the particles are initialized both within feasible and infeasible space (blue and cyan dots).

### 10.4.6 Penalization + cut off + resetting velocity method

This technique consists of the penalization method for every constraint except for the hyper-cube-like boundary constraints, which are handled by the "cut off at the boundary" technique. Thus, when a particle attempts to leave the hyper-cube, it is relocated on the boundary, and its velocity is re-set to zero. Note that the particles need to be initialized within the hyper-cube.

The history of the particles' positions for the GP-PSO$^{(s.d.w.)}$ with the incorporation of the "penalization + cut off + re-setting velocity" technique optimizing the first benchmark constrained optimization problem—together with a colour-map of the conflict function—can be seen in **Fig. 10. 8**:

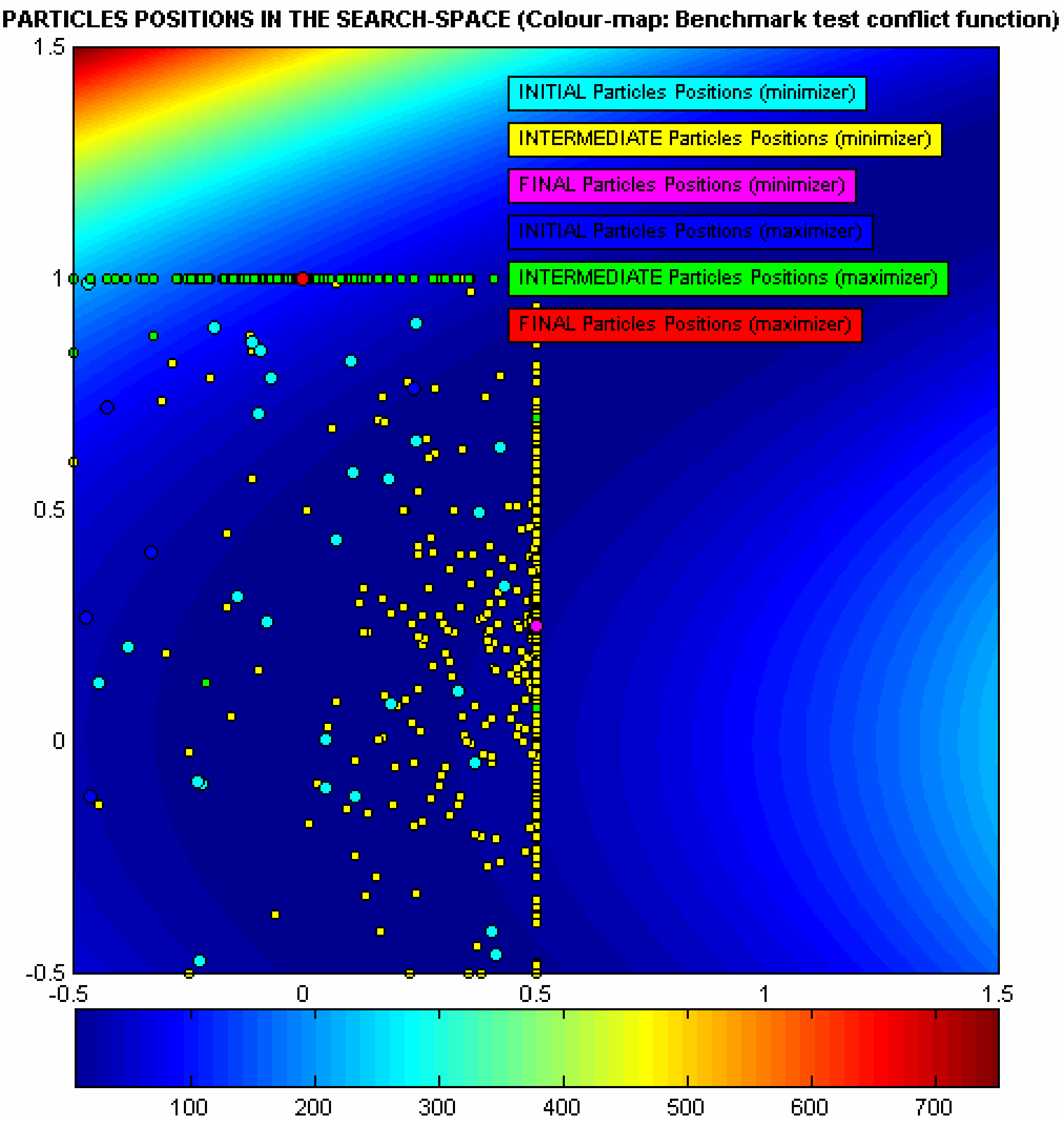

**Fig. 10. 8**: The GP-PSO$^{(s.d.w.)}$ optimizing the first benchmark constrained optimization problem, where the constraint-handling method is the "penalization + cut off + re-setting velocity" method.

### 10.4.7 Penalization + cut off + resetting velocity method (bis)

This technique is a variation of the previous one. The only difference is in that the particles are not iteratively initialized but forced by the "cut off at the boundary" method. That is to say, if a particle is initialized out of the hyper-cube-like boundary, the method automatically relocates it on the boundary. The benefits and drawbacks of incorporating this "cut off at the boundary" technique were already discussed in section **10.4.3**. The history of the particles' positions for the GP-PSO$^{(s.d.w.)}$ with the incorporation of the "penalization + cut off + re-setting velocity (bis)" technique optimizing the first benchmark constrained optimization problem—together with a colour-map of the conflict function—can be seen in **Fig. 10. 9**:

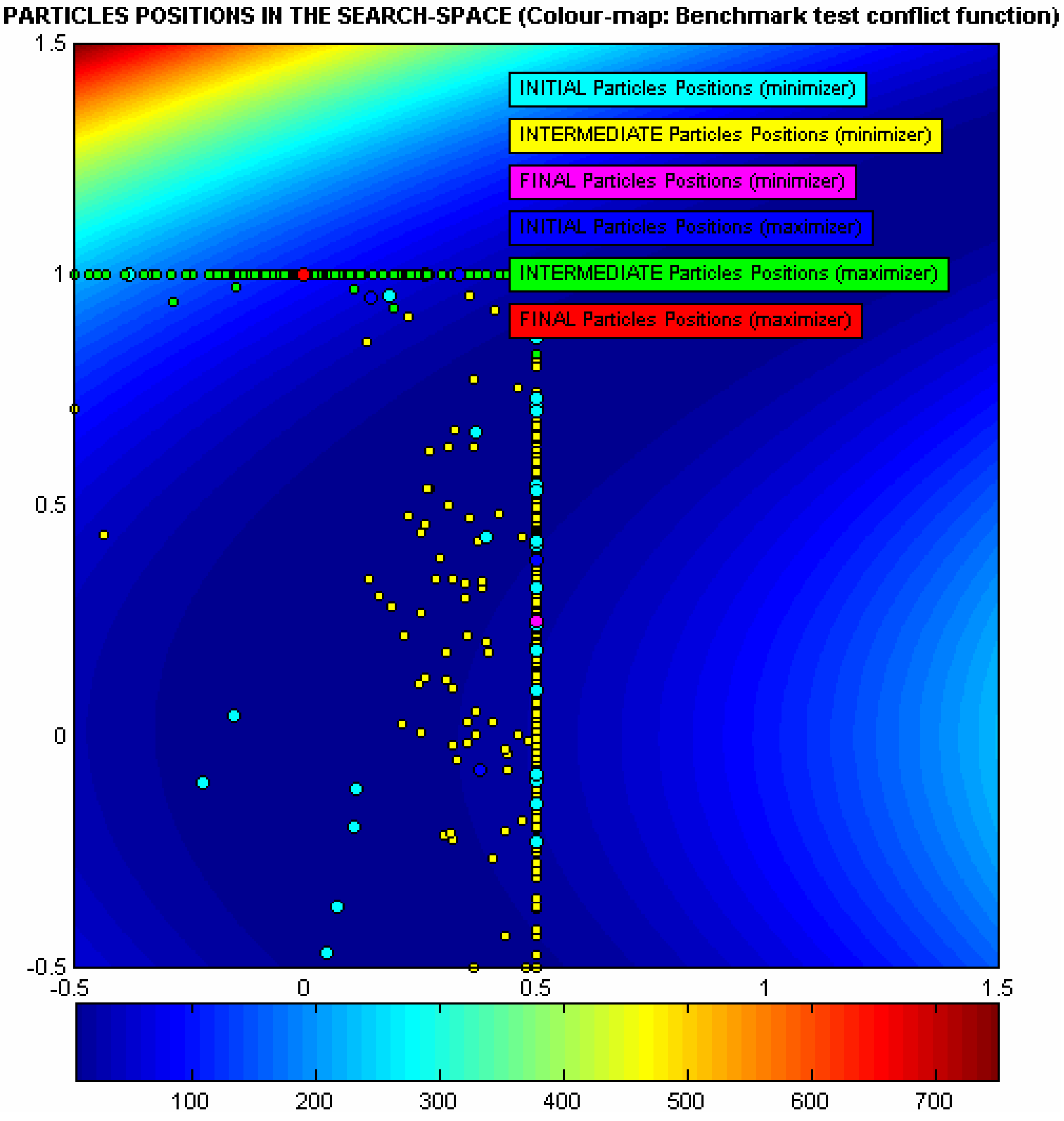

**Fig. 10. 9**: The GP-PSO$^{(s.d.w.)}$ optimizing the first benchmark constrained optimization problem, where the constraint-handling method is the "penalization + cut off + re-setting velocity" method (bis).

Animations of the evolution of the particles' positions corresponding to the experiments shown from **Fig. 10. 3** to **Fig. 10. 9** can be found in digital **Appendix 4**.

# 10.5 Experimental results

The GP-PSO$^{(s.d.w.)}$ is implemented with the incorporation of each of the seven constraint-handling techniques proposed in section **10.4**, and tested on the suite of three benchmark constrained optimization problems stated in section **10.3**.

General settings:
- Number of runs per experiment: 50
- $v_{max} = 0.5 \cdot (x_{max} - x_{min})$
- Number of particles of the minimizer: 30
- Number of particles of the maximizer: 5
- Optimizer: GP-PSO$^{(s.d.w.)}$
- $t_{max} = 10000$

It is important to note that the optimizer developed here requires that the feasible search-space is finite. Otherwise, the maximum (*cgworst*) and/or the minimum (*cgbest*) solution found might depend on the length of the search, and the values of $x_{max} - x_{min}$ might not be defined for every variable, making some of the relative errors meaningless. In addition to that, the value of $x_{max} - x_{min}$ is kept the same for every dimension within this work.

## 10.5.1 First benchmark constrained optimization problem

The additional constraint $x_i \in [x_{min}, x_{max}]^n = [-0.5, 1.5]^2$ does not introduce any alteration to the feasible space defined by the original problem. The objective is to define a square-like boundary constraint that contains the feasible space and can be used by the "cut off at the boundary" technique to reduce the number of evaluations far from the region of interest. For instance, refer to **Fig. 10. 5** and **Fig. 10. 6**. In the first case, the particles are allowed to fly over the space limited by the rectangle-like constraints that result from the intersection of the feasible intervals of the original problem ($-0.5 \le x_1 \le 0.5$, $-\infty \le x_2 \le 1$) and the ones added here ($-0.5 \le x_1 \le 1.5$, $-0.5 \le x_2 \le 1.5$). Then the particles are allowed to fly over the space defined as ($-0.5 \le x_1 \le 0.5$, $-0.5 \le x_2 \le 1$). In the second case, the particles are allowed to fly only over feasible space (i.e. over space that satisfy all constraints), which is clearly contained within the square-like boundary added here. In fact the feasible space happens to be contained within a rectangle defined as ($-0.5 \le x_1 \le 0.5$, $-0.25 \le x_2 \le 1$). Therefore, the additional constraints do not affect the original feasible space.

Another purpose pursued with the additional constraints is to define the value of $x_{\max} - x_{\min}$ which the relative errors regarding the particles' positions will be related to, as well as the maximum size of the search-space where the best and worst conflicts can be sought in cases where the feasible space is not finite. In addition to that, the particles are randomly initialized within this additional square-like boundary.

The results obtained from the optimization of the first benchmark constrained optimization problem by means of the GP-PSO$^{(s.d.w.)}$ with the incorporation of seven different constraint-handling techniques are gathered in **Table 10. 1**:

| FIRST BENCHMARK PROBLEM | | | |
|---|---|---|---|
| OPTIMIZER | Best solution<br>Worst solution | Mean best solution<br>(Standard deviation) | Mean time-steps<br>(Standard deviation) |
| PRESERVING FEASIBILITY METHOD | 0.25<br>0.25 | 0.25<br>(0.00) | 1010.48<br>(24.20) |
| PRESERVING FEASIBILITY + RESETTING VELOCITY | 0.25<br>0.25 | 0.25<br>(0.00) | 1003.54<br>(15.74) |
| PRESERVING FEASIBILITY + CUT OFF + RESETTING VELOCITY | 0.25<br>0.25 | 0.25<br>(0.00) | 1003.52<br>(18.38) |
| BISECTION METHOD | 0.25<br>0.25 | 0.25<br>(0.00) | 1007.54<br>(25.37) |
| PENALIZATION METHOD | 0.25<br>0.25 | 0.25<br>(0.00) | 1009.84<br>(33.95) |
| PENALIZATION + CUT OFF + RESETTING VELOCITY | 0.25<br>0.25 | 0.25<br>(0.00) | 1013.80<br>(43.84) |
| PENALIZATION + CUT OFF + RESETTING VELOCITY (BIS) | 0.25<br>0.25 | 0.25<br>(0.00) | 1009.44<br>(29.51) |

**Table 10. 1**: Results obtained from the optimization of the first benchmark constrained optimization problem by means of the GP-PSO$^{(s.d.w.)}$ with the incorporation of seven different constraint-handling techniques.

Clearly, none of the techniques exhibit any trouble in finding the exact solution to this simple 2-dimensional problem along any of the 50 runs.

## 10.5.2 Second benchmark constrained optimization problem

The additional constraint $x_i \in [-2,2]^2$ does not introduce any alteration to the solution to the original problem. However, the feasible search-space is cut off, as it can be seen in **Fig. 10. 10**. That is, the worst conflict of the original problem is not located within the region $[-2,2]^2$, so that the computation of the relative errors is affected. However, this was not corrected because it is not easy to determine whether the constraints effectively define a finite region for more complex problems, and whether a certain hyper-cube that contains the complete feasible space can be defined. Notice that this second benchmark constrained optimization problem does not specifically define intervals permitted for each variable, and the feasible space is defined by two equations. It is not straightforward to tell whether the feasible space is finite, and which the maximum and minimum values permitted for each variable are. Therefore, the additional constraint is kept as the region defined as $x_i \in [-2,2]^2$, and the particles are initialized within this same region.

The results obtained from the optimization of the second benchmark constrained optimization problem by means of the GP-PSO$^{(s.d.w.)}$ with the incorporation of seven different constraint-handling techniques are gathered in **Table 10. 2**:

| SECOND BENCHMARK PROBLEM | | | |
|---|---|---|---|
| OPTIMIZER | Best solution | Mean best solution | Mean time-steps |
| | Worst solution | (Standard deviation) | (Standard deviation) |
| PRESERVING FEASIBILITY METHOD | 1.00 | 1.00 | 1151.96 |
| | 1.00 | (0.00) | (144.00) |
| PRESERVING FEASIBILITY + RESETTING VELOCITY | 1.00 | 1.00 | 1011.36 |
| | 1.00 | (0.00) | (27.57) |
| PRESERVING FEASIBILITY + CUT OFF + RESETTING VELOCITY | 1.00 | 1.00 | 1012.04 |
| | 1.00 | (0.00) | (33.80) |
| BISECTION METHOD | 1.00 | 1.00 | 1000.00 |
| | 1.00 | (0.00) | (0.00) |
| PENALIZATION METHOD | 1.00 | 1.00 | 1040.66 |
| | 1.00 | (0.00) | (78.20) |
| PENALIZATION + CUT OFF + RESETTING VELOCITY | 1.00 | 1.00 | 1036.14 |
| | 1.00 | (0.00) | (61.11) |
| PENALIZATION + CUT OFF + RESETTING VELOCITY (BIS) | 1.00 | 1.00 | 1049.78 |
| | 1.00 | (0.00) | (70.92) |

**Table 10. 2**: Results obtained from the optimization of the second benchmark constrained optimization problem by means of the GP-PSO$^{(s.d.w.)}$ with the incorporation of seven different constraint-handling techniques.

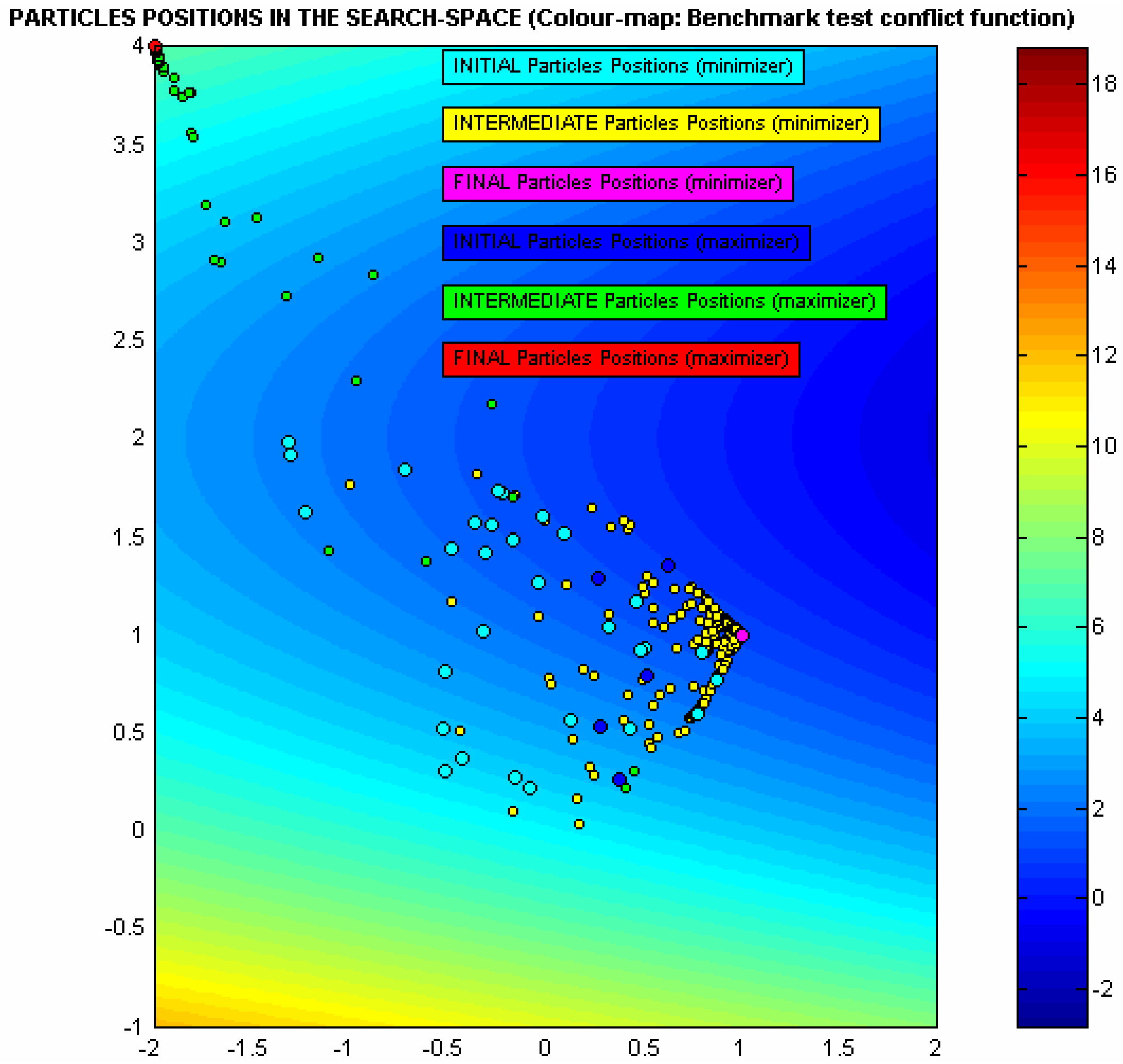

**Fig. 10. 10**: The GP-PSO$^{(s.d.w.)}$ optimizing the second benchmark constrained optimization problem, where the constraint-handling method is the "bisection" method, and the additional square-like constraint is set as $[-4,4]^2$, while the particles are still initialized within the region $[-2,2]^2$.

## 10.5.3 Third benchmark constrained optimization problem

There is no additional constraint incorporated into this problem because the values permitted for every variable are clearly limited by interval constraints. Hence the interval defined as $(x_{\max} - x_{\min}) = 100 - 0 = 100$, which is the greatest interval, is used for the computation of the relative errors. However, the particles are initialized within a region defined as $x_i \in [0,2]^{13}$ because the methods which have to find an initial feasible population were not able to randomly generate one within a reasonable amount of time when the initialization region was set as $x_i \in [0,100]^{13}$ so as to contain all the permitted values for all the variables. This already reveals a serious weakness of such methods. Note, however, that this weakness can be either eliminated or at least notably improved if the particles were randomly initialized within the interval specifically permitted for every variable. For instance, the permitted values for 10 variables are limited by an interval of $[0,1]$, while the permitted values for only 3 variables are limited by the interval $[0,100]$.

The results obtained from the optimization of the second benchmark constrained optimization problem by means of the GP-PSO$^{(s.d.w.)}$ with the incorporation of seven different constraint-handling techniques are gathered in **Table 10. 3**:

| THIRD BENCHMARK PROBLEM | | | |
|---|---|---|---|
| OPTIMIZER | Best solution<br>Worst solution | Mean best solution<br>(Standard deviation) | Mean time-steps<br>(Standard deviation) |
| PRESERVING FEASIBILITY METHOD | -15.00<br>-12.38 | -14.53<br>(0.90) | 9026.74<br>(1098.93) |
| PRESERVING FEASIBILITY + RESETTING VELOCITY | -15.00<br>-12.07 | -14.53<br>(0.97) | 7140.42<br>(1053.15) |
| PRESERVING FEASIBILITY + CUT OFF + RESETTING VELOCITY | -15.00<br>-13.00 | -14.96<br>(0.28) | 1091.36<br>(88.30) |
| BISECTION METHOD | -15.00<br>-8.45 | -12.82<br>(1.61) | 1039.74<br>(72.35) |
| PENALIZATION METHOD | -15.00<br>-12.45 | -14.55<br>(0.86) | 7796.48<br>(1470.75) |
| PENALIZATION + CUT OFF + RESETTING VELOCITY | -15.00<br>-12.00 | -14.50<br>(0.97) | 1173.48<br>(134.23) |
| PENALIZATION + CUT OFF + RESETTING VELOCITY (BIS) | -15.00<br>-13.00 | -14.96<br>(0.28) | 1223.12<br>(174.33) |

**Table 10. 3**: Results obtained from the optimization of the third benchmark constrained optimization problem by means of the GP-PSO$^{(s.d.w.)}$ with the incorporation of seven different constraint-handling techniques.

## 10.5.4 Discussion

First of all, it should be remarked that the conclusions derived from the quantitative results obtained from the experiments carried out optimizing only three benchmark problems, and from the visualization of the evolution of the particles' positions on 2-dimensional problems only, can by no means be considered final. As it was previously asserted, the work carried out here with regards to constrained particle swarm optimizers has a very limited scope. More extensive and in depth work is required in the future.

Three main methods were proposed here. Some additional, complementary techniques were incorporated into two of them, giving birth to four other methods. Hence, the “cut of at the boundary” technique is thought of as a possible enhancement to be incorporated into these three methods, rather than as a constraint-handling technique on its own. Another possible enhancement consists of resetting the velocities to zero when the inertia effect is believed to require damping (e.g. when a particle is flying over infeasible space). Therefore, the three constraint-handling methods considered here are the “preserving feasibility”, the “bisection”, and the “penalization” methods.

Considering only these three pure techniques, it is clear that the “bisection” method is notably faster than the others. While this is convenient in terms of function evaluations, the solutions found by the “preserving feasibility” and by the “penalization” methods are more accurate, especially for the 13-dimensional problem. Although the “bisection” method manages to find a very accurate solution to this problem along some runs, it exhibits a remarkably poorer mean solution, and a notably higher corresponding standard deviation (refer to **Table 10. 3**).

Two other important aspects must be considered here. On the one hand, the “bisection” and the “preserving feasibility” methods are parameter-free methods which require no adaptation to deal with different problems. However, they do require an initial feasible population, and cannot extensively explore the infeasible space. On the other hand, the “penalization” method does not require that the particles are initialized within feasible space, and can extensively explore the infeasible space in quest for some other (possibly disjointed) unexplored feasible regions. However, it does require a problem-dependent tuning of its parameters.

With regards to the additional techniques, it appears that resetting the velocity and cutting off a particle’s displacement vector when it intends to fly out of the hyper-rectangle-like boundary result in remarkably faster convergence, especially for the 13-dimensional problem, without any decrease in the accuracy of the solution found.

Therefore, the “bisection”, the “preserving feasibility + cut off + v = 0”, and the “penalization + cut off + v = 0” methods are the selected techniques for the general-purpose optimizers. It is not possible to assert that any of them would normally outperform the others. In contrast, the “bisection” method would typically exhibit faster convergence, would guarantee that the solution found is feasible, would require no adaptation for different problems, but would lack the ability to explore the infeasible space, and would require an initial swarm of feasible particles. The “preserving feasibility + cut off + v = 0” method would also guarantee that the solution found is feasible, and would require no adaptation for different problems, but would exhibit limited ability to explore the infeasible space (without gaining any information from doing so), and would also require an initial swarm of feasible particles. The “cut off + v = 0” technique enhances its convergence rate. Finally, the “penalization + cut off + v = 0” would exhibit the advantage that an initial swarm of feasible particles is not required, that it can extensively explore infeasible space, but it cannot guarantee that the solution found is 100% feasible, and it requires problem-dependent tuning of its parameters.

Some other alternatives such as incorporating the “bisection” method into the “preserving feasibility” or into the “penalty” methods in order to handle the boundary constraints were not even tried. Another option could be to combine two sub-swarms, one handling the constraints by means of the “penalization” method, and the other by means of the “preserving feasibility” method. If the system is implemented so that the solution to the problem is not the best experience of the whole swarm—which is used in the particles’ velocity update equation—but the best among the experiences corresponding to all the particles equipped with the “preserving feasibility” method, the solution can be guaranteed to be feasible. The infeasible space is explored, nevertheless, and even the sub-swarm equipped with the “preserving feasibility” technique could be able to pass through infeasible space thanks to the sub-swarm equipped with the “penalization” technique. This strategy is left for future work. For the time being, the “bisection”, the “preserving feasibility + cut off + v = 0”, and the “penalization + cut off + v = 0” methods are selected, independently, giving birth to three different, complementary general-purpose optimizers. Ideally, every problem would be solved by the three of them independently, and the best solution would be picked.

# 10.6 Closure

The basic formulation of a constrained optimization problem was briefly introduced within this chapter. The different types of constraints that might limit the portion of the search-space that is feasible were discussed, and a few popular constraint-handling methods were outlined. Profiting from these pre-existing techniques, seven methods to handle inequality constraints were proposed, implemented, and tested on a small suite of three benchmark functions taken from [81][4]. Considering both the quantitative and the qualitative results obtained from the experiments, the "preserving feasibility + cut off + v = 0", the "bisection", and the "penalization + cut off + v = 0" methods were selected to be incorporated into the GP-PSO$^{(s.d.w.)}$. Hence, three general-purpose optimizers which are expected to complement each other were developed. Ideally, the different, complementary, desirable features of these three optimizers should be brought together into a single optimizer. However, this ambitious objective is left for future work due to time constraints.

In summary, three general-purpose optimizers emerge from this thesis, all of them being based on the GP-PSO$^{(s.d.w.)}$, with the incorporation of different constraint-handling techniques. The final codes are modified with respect to the ones tested within this chapter, so that hyper-rectangle-like rather than hyper-cube-like boundary constraints can be set, thus eliminating the problem of deciding the region for the initialization of the particles: they are now initialized by being randomly placed within the hyper-rectangle-like boundary. This leads to another modification: the velocity constraint is not the same for every dimension but computed according to the interval constraints associated to the variable at issue. Hence, the "bisection" and the "preserving feasibility + cut off + v = 0" methods "only" have to successively, randomly initialize its particles until they all comply with the remaining constraints. These final codes can be found in digital **Appendix 4**. The path is as follows:

X:\ Appendix 4\Matlab PSO codes\FINAL GENERAL-PURPOSE OPTIMIZERS

The next three chapters are devoted to the illustration of the capabilities of the three general-purpose constrained particle swarm optimizers by means of their application to a number of different problems. Thus, **Chapter 11** is concerned with the optimization of a number of unconstrained and constrained benchmark functions; **Chapter 12** with the training of a very simple, rather academic, artificial neural network; and **Chapter 13** with the optimization of a few simple benchmark engineering problems.

[4] Technically, the "penalization" method can handle equality constraints straightaway, while the "bisection" and the "preserving feasibility" methods would require a number of modifications. Nevertheless, no problem with equality constraints was tested within this work.

# SECTION III

# A P P L I C A T I O N S

# Chapter 11

# FUNCTION OPTIMIZATION

This chapter is dedicated to illustrating the ability of some of the proposed general-purpose algorithms to optimize a number of challenging benchmark functions. While the main purpose is to show the efficacy of the optimizers in dealing with complex problems, the experiments run hereafter also serve the function of further testing the algorithms, thus inducing further conclusions. The optimizers are first applied to the suite of benchmark functions shown in Table 6.1 for a symmetric initialization all over the feasible search-space. Then, they are applied to the same suite of functions but for the asymmetric initialization stated in Table 7.1. Finally, the optimizers are applied to the three benchmark functions used along Chapter 10, in addition to a set of other three benchmark functions taken from the literature.

## 11.1 Introduction

The efficacy of the GP-PSO$^{(s.d.w.)}$ equipped with the pure "preserving feasibility" technique in optimizing the six benchmark functions stated in **Table 6.1**, for the symmetric initialization and hyper-cube-like boundary constraints, was already tested along **Chapter 9**. In turn, the efficacy of the same optimizer equipped with seven different constraint-handling techniques in optimizing three benchmark constrained optimization functions was already tested along **Chapter 10**. Three versions of the GP-PSO$^{(s.d.w.)}$—one equipped with the "bisection" method, another with the "penalization + cut off + v = 0" method, and the third with the "preserving feasibility + cut off + v = 0" method—were selected by the end of **Chapter 10** as the proposed final general-purpose codes. Since these codes are expected to be able to deal with most problems efficiently, it seems reasonable to subject them to a final test on a number of benchmark functions that present different challenges in order to prove their effectiveness.

Thus, these optimizers are first tested on the six benchmark functions stated in **Table 6.1**, where the particles are symmetrically[1] initialized within the whole feasible search-space.

[1] Recall that "symmetric initialization" refers to the the particles being randomly initialized within a region that is symmetric to the origin. This does not imply that they are uniformly distributed, nor that they are symmetrically located with respect to the global optimum.

Then, the optimizers are tested on the same suite of functions, but with their particles being asymmetrically initialized according to **Table 7.1**. Finally, they are tested on the set of three benchmark functions used along **Chapter 10**, in addition to another function taken from [81] and to other two functions taken from [82].

The general settings for all the experiments are as follows:

- Number of runs per experiment: 50
- $v_{\max} = 0.5 \cdot \left( x_{\max} - x_{\min} \right)$
- Number of particles of the minimizer: 30
- Number of particles of the maximizer: 5
- Optimizer: GP-PSO$^{(s.d.w.)}$
- $t_{\max} = 30000$

## 11.2 First test suite of benchmark functions

The efficacy of the optimizers is first tested on the suite of benchmark functions stated in **Table 6.1**, which only present hyper-cube-like boundary constraints.

### 11.2.1 Symmetric initialization

The GP-PSO$^{(s.d.w.)}$ equipped with the three constraint-handling techniques selected by the end of **Chapter 10**—namely the “preserving feasibility + cut off + v = 0” method, the “bisection” method, and the “penalization + cut off + v = 0” method—is now tested on this suite of benchmark functions, whose challenge lies in the objective function itself rather than on the constraints. Surprisingly, all of them present a few severe problems when optimizing the Rosenbrock function. This is surprising, since the only difference in relation to the optimizer equipped with the pure “preserving feasibility” technique tested along **Chapter 9** is the constraint-handling method, while the only constraints are hyper-cube-like boundaries located far from the global optimum! Therefore, the GP-PSO$^{(s.d.w.)}$ equipped with the pure “preserving feasibility” and with the pure “penalization” methods is also tested along this section.

The main results obtained from the experiments are gathered from **Table 11. 1** to **Table 11. 6**:

| SPHERE | | | |
|---|---|---|---|
| OPTIMIZER | Best solution / Worst solution | Mean best solution (Standard deviation) | Mean time-steps (Standard deviation) |
| PRESERVING FEASIBILITY + CUT OFF + V = 0 | 2.43E-84 | 1.31E-41 | 4.58E+03 |
| | 5.26E-40 | (7.52E-41) | (7.20E+02) |
| BISECTION METHOD | 2.59E-81 | 3.29E-44 | 4.62E+03 |
| | 5.13E-43 | (1.17E-43) | (6.18E+02) |
| PENALIZATION + CUT OFF + V = 0 | 7.49E-78 | 1.97E-38 | 4.44E+03 |
| | 9.86E-37 | (1.39E-37) | (6.84E+02) |
| PRESERVING FEASIBILITY | 3.94E-79 | 1.38E-38 | 4.46E+03 |
| | 6.73E-37 | (9.52E-38) | (6.95E+02) |
| PENALIZATION | 6.33E-81 | 4.57E-47 | 4.58E+03 |
| | 2.05E-45 | (2.89E-46) | (6.01E+02) |

**Table 11. 1**: Summary of the most significant results obtained from the optimization of the 30-dimensional Sphere function by means of five selected general-purpose optimizers.

| ROSENBROCK | | | |
|---|---|---|---|
| OPTIMIZER | Best solution / Worst solution | Mean best solution (Standard deviation) | Mean time-steps (Standard deviation) |
| PRESERVING FEASIBILITY + CUT OFF + V = 0 | 6.12E-08 | 1.81E+03 | 2.98E+04 |
| | 9.00E+04 | (1.27E+04) | (7.73E+02) |
| BISECTION METHOD | 1.33E-11 | 5.67E+03 | 2.85E+04 |
| | 9.00E+04 | (2.15E+04) | (3.98E+03) |
| PENALIZATION + CUT OFF + V = 0 | 1.41E-09 | 1.87E+03 | 2.97E+04 |
| | 9.00E+04 | (1.27E+04) | (1.24E+03) |
| PRESERVING FEASIBILITY | 1.33E-10 | 8.30E+00 | 3.00E+04 |
| | 7.30E+01 | (1.65E+01) | (4.80E+01) |
| PENALIZATION | 4.08E-10 | 7.77E+00 | 2.97E+04 |
| | 7.29E+01 | (1.43E+01) | (1.14E+03) |

**Table 11. 2**: Summary of the most significant results obtained from the optimization of the 30-dimensional Rosenbrock function by means of five selected general-purpose optimizers.

The incorporation of the “cut off + v = 0” technique into the pure “preserving feasibility” and “penalization” methods appears to be sometimes convenient while sometimes harmful. Nonetheless, all the results are reasonably good except for those obtained by the optimizers with the “cut off + v = 0” techniques optimizing the Rosenbrock function.

This is worth an individual analysis. The optimizer with the “preserving feasibility + cut off + v = 0” technique finds very good solutions along 49 out of the 50 runs. However, the solution found for the remaining run is over 90000! In turn, the solutions found by the optimizer with the "penalization + cut off + v = 0" technique are very good for 48 out of the 50 runs, but the remaining solutions are one over 90000, and the other over 3000! Finally, the solutions found by the optimizer with the "bisection" technique fails to find good solutions along 9 out of the 50 runs: three times the solution is over 90000, four times over 3000, and twice over 500. If all these runs are removed, the resulting mean conflicts found by these three optimizers outperform those found by the pure "preserving feasibility" and "penalization" methods.

| **RASTRIGRIN** | | | |
|---|---|---|---|
| OPTIMIZER | Best solution | Mean best solution | Mean time-steps |
| | Worst solution | (Standard deviation) | (Standard deviation) |
| PRESERVING FEASIBILITY + CUT OFF + V = 0 | 5.97E+00 | 2.45E+01 | 2.06E+04 |
| | 6.37E+01 | (1.40E+01) | (5.59E+03) |
| BISECTION METHOD | 9.95E+00 | 3.53E+01 | 1.98E+04 |
| | 8.36E+01 | (2.06E+01) | (6.76E+03) |
| PENALIZATION + CUT OFF + V = 0 | 5.97E+00 | 3.08E+01 | 1.96E+04 |
| | 9.17E+01 | (1.79E+01) | (4.53E+03) |
| PRESERVING FEASIBILITY | 7.96E+00 | 2.13E+01 | 2.16E+04 |
| | 4.48E+01 | (8.36E+00) | (5.85E+03) |
| PENALIZATION | 9.95E+00 | 2.32E+01 | 2.15E+04 |
| | 3.98E+01 | (7.15E+00) | (5.55E+03) |

**Table 11. 3**: Summary of the most significant results obtained from the optimization of the 30-dimensional Rastrigrin function by means of five selected general-purpose optimizers.

| **GRIEWANK** | | | |
|---|---|---|---|
| OPTIMIZER | Best solution | Mean best solution | Mean time-steps |
| | Worst solution | (Standard deviation) | (Standard deviation) |
| PRESERVING FEASIBILITY + CUT OFF + V = 0 | 0.00E+00 | 2.87E-02 | 1.04E+04 |
| | 1.28E-01 | (2.93E-02) | (2.71E+03) |
| BISECTION METHOD | 0.00E+00 | 2.39E-02 | 1.00E+04 |
| | 1.13E-01 | (2.95E-02) | (2.75E+03) |
| PENALIZATION + CUT OFF + V = 0 | 0.00E+00 | 3.17E-02 | 1.08E+04 |
| | 1.03E-01 | (2.90E-02) | (2.39E+03) |
| PRESERVING FEASIBILITY | 0.00E+00 | 1.98E-02 | 1.01E+04 |
| | 8.29E-02 | (2.17E-02) | (2.55E+03) |
| PENALIZATION | 0.00E+00 | 2.21E-02 | 1.02E+04 |
| | 9.32E-02 | (2.42E-02) | (2.60E+03) |

**Table 11. 4**: Summary of the most significant results obtained from the optimization of the 30-dimensional Griewank function by means of five selected general-purpose optimizers.

| SCHAFFER F6 2D | | | |
|---|---|---|---|
| OPTIMIZER | Best solution<br>Worst solution | Mean best solution<br>(Standard deviation) | Mean time-steps<br>(Standard deviation) |
| PRESERVING FEASIBILITY + CUT OFF + V = 0 | 0.00E+00<br>0.00E+00 | 0.00E+00<br>(0.00E+00) | 4.14E+03<br>(1.69E+03) |
| BISECTION METHOD | 0.00E+00<br>9.72E-03 | 1.94E-04<br>(1.37E-03) | 4.73E+03<br>(2.15E+03) |
| PENALIZATION + CUT OFF + V = 0 | 0.00E+00<br>9.72E-03 | 1.94E-04<br>(1.37E-03) | 5.65E+03<br>(2.65E+03) |
| PRESERVING FEASIBILITY | 0.00E+00<br>2.50E-01 | 3.89E-04<br>(1.92E-03) | 5.47E+03<br>(2.59E+03) |
| PENALIZATION | 0.00E+00<br>9.72E-03 | 3.89E-04<br>(1.92E-03) | 5.23E+03<br>(2.77E+03) |

**Table 11. 5**: Summary of the most significant results obtained from the optimization of the 2-dimensional Schaffer f6 function by means of five selected general-purpose optimizers.

| SCHAFFER F6 | | | |
|---|---|---|---|
| OPTIMIZER | Best solution<br>Worst solution | Mean best solution<br>(Standard deviation) | Mean time-steps<br>(Standard deviation) |
| PRESERVING FEASIBILITY + CUT OFF + V = 0 | 3.72E-02<br>1.27E-01 | 7.14E-02<br>(2.44E-02) | 1.96E+04<br>(4.48E+03) |
| BISECTION METHOD | 3.72E-02<br>1.27E-01 | 7.27E-02<br>(2.87E-02) | 1.87E+04<br>(4.79E+03) |
| PENALIZATION + CUT OFF + V = 0 | 3.72E-02<br>1.78E-01 | 1.10E-01<br>(2.90E-02) | 1.69E+04<br>(4.93E+03) |
| PRESERVING FEASIBILITY | 3.72E-02<br>1.27E-01 | 6.54E-02<br>(2.34E-02) | 1.95E+04<br>(4.99E+03) |
| PENALIZATION | 3.72E-02<br>1.27E-01 | 1.01E-01<br>(2.61E-02) | 1.77E+04<br>(5.87E+03) |

**Table 11. 6**: Summary of the most significant results obtained from the optimization of the 30-dimensional Schaffer f6 function by means of five selected general-purpose optimizers.

Because of these unexpected results, the GP-PSO$^{(s.d.w.)}$ equipped with the pure "preserving feasibility" method and with the pure "penalization" method are also tested, obtaining more consistent results. Considering that the selection of the optimizers equipped with the "cut off + v = 0" technique was based on the faster convergence without the loss of accuracy rather than on higher accuracy, these optimizers are removed from the selected ones. Hence, only two algorithms are chosen as the proposed general-purpose optimizers: the GP-PSO$^{(s.d.w.)}$ equipped with the pure "preserving feasibility" method, and the GP-PSO$^{(s.d.w.)}$ equipped with the pure "penalization" method. Therefore, the experiments on the asymmetric initialization are carried out for these two optimizers only.

### 11.2.2 Asymmetric initialization

The asymmetric initialization is considered here to show that the region of the search-space where the particles are initially spread over does not need to contain the solution for the optimizers to be able to find it, as well as to show the convenient effect of keeping diversity for a longer period of time in order to delay premature convergence. Recall that premature convergence on a poor sub-optimal solution occurred for the BSt-PSO$^{(c)}$ and the BSt-PSO$^{(p)}$ optimizing the 30-dimensional Rastrigrin function, when using the asymmetric initialization. For instance, compare **Fig. 6.50** to **Fig. 7.15**, keeping in mind that the "actual absolute error" in **Fig. 7.15** is equivalent to the "mean best conflict". Refer to **Appendix 4** for the visualization of the curves corresponding to the BSt-PSO$^{(c)}$.

The most significant results obtained from these experiments are gathered in **Table 11. 7**.

### 11.2.3 Discussion

The incorporation of the "cut off + v = 0" technique into the "preserving feasibility" and into the "penalization" methods leads to notably higher standard deviations. Notice that the best solutions found by the optimizers with and without the "cut off + v = 0" technique do not differ much. In contrast, the difference between their mean solutions is sometimes shocking.

Given that the aim is to develop a general-purpose optimizer, more stable techniques are preferred. Therefore, the pure "preserving feasibility" and the pure "penalization" methods are selected as the constraint-handling techniques for the proposed general-purpose optimizer.

As it can be observed, the GP-PSO$^{(s.d.w.)}$ equipped with the pure "preserving feasibility" and with the pure "penalization" methods find very good solutions for all the six benchmark functions included in this test suite, and for both the symmetric and the asymmetric initializations. This is exactly what is sought for a general-purpose optimizer. Both techniques are kept because, while both find very good solutions, the "preserving feasibility" method guarantees a feasible solution, does not require parameters' tuning, but does require a feasible initial population. In contrast, the "penalization method" does not guarantee a feasible solution, does require the tuning of its parameters, but does not need an initial feasible population. Clearly, both techniques are complementary.

| OPTIMIZER | Best solution / Worst solution | Mean best solution (Standard deviation) | Mean time-steps (Standard deviation) |
|---|---|---|---|
| **SPHERE** | | | |
| PRESERVING FEASIBILITY | 2.02E-93 | 3.46E-41 | 4.59E+03 |
| | 1.48E-39 | (2.11E-40) | (7.73E+02) |
| PENALIZATION | 5.41E-71 | 4.68E-43 | 4.52E+03 |
| | 1.23E-41 | (2.12E-42) | (5.68E+02) |
| **ROSENBROCK** | | | |
| PRESERVING FEASIBILITY | 7.08E-12 | 5.40E+00 | 2.97E+04 |
| | 7.63E+01 | (1.15E+01) | (1.38E+03) |
| PENALIZATION | 2.17E-08 | 3.66E+00 | 2.99E+04 |
| | 1.56E+01 | (4.07E+00) | (8.67E+02) |
| **RASTRIGRIN** | | | |
| PRESERVING FEASIBILITY | 3.98E+00 | 1.97E+01 | 2.02E+04 |
| | 4.48E+01 | (9.86E+00) | (5.14E+03) |
| PENALIZATION | 3.98E+00 | 2.04E+01 | 2.02E+04 |
| | 4.18E+01 | (8.33E+00) | (6.00E+03) |
| **GRIEWANK** | | | |
| PRESERVING FEASIBILITY | 0.00E+00 | 2.27E-02 | 9.86E+03 |
| | 1.00E-01 | (2.60E-02) | (2.79E+03) |
| PENALIZATION | 0.00E+00 | 3.26E-02 | 1.07E+04 |
| | 1.10E-01 | (3.21E-02) | (3.01E+03) |
| **SCHAFFER F6 2D** | | | |
| PRESERVING FEASIBILITY | 0.00E+00 | 0.00E+00 | 4.29E+03 |
| | 0.00E+00 | (0.00E+00) | (1.58E+03) |
| PENALIZATION | 0.00E+00 | 3.89E-04 | 6.38E+03 |
| | 9.72E-03 | (1.92E-03) | (3.25E+03) |
| **SCHAFFER F6** | | | |
| PRESERVING FEASIBILITY | 3.72E-02 | 7.05E-02 | 2.07E+04 |
| | 1.27E-01 | (2.30E-02) | (4.95E+03) |
| PENALIZATION | 3.72E-02 | 1.06E-01 | 1.67E+04 |
| | 1.78E-01 | (3.26E-02) | (4.92E+03) |

**Table 11. 7**: Summary of the most significant results obtained from the optimization of the first suite of benchmark functions by means of two selected general-purpose optimizers, making use of the asymmetric initialization.

It seems that both the “cut off + v = 0” technique and the “bisection” method decrease the mobility of the particles, which is critical to escape sub-optimal solutions. The positive effect is that this frequently also results in faster convergence. Hence, they might still be convenient for problems where the function evaluations are too expensive or too time-consuming, so that less accurate but faster solutions are sought. Notice that the GP-PSO$^{(fast)}$ should also be considered for such cases (refer to **Chapter 9**).

# 11.3 Second test suite of benchmark functions

A second suite of benchmark functions is presented here so as to test the algorithm against objective functions that are typically less challenging than those of the first suite, but whose valid solutions are limited by a number of constraints in addition to the boundary ones. In addition, a benchmark function which is completely unconstrained is incorporated to the suite.

Except for the unconstrained problem, all the other functions have hyper-rectangle-like boundary constraints, which are used for the initialization of the particles. That is to say, the particles are initialized by being randomly spread all over the feasible space. Then, although the first three functions of this suite are the same as the functions used to test the algorithms along **Chapter 10**, they are re-written hereafter for convenience with the addition of the smallest hyper-rectangle-like boundary constraints that contains the feasible space, if the original problem does not explicitly state them.

## 11.3.1 First benchmark function

Minimize $f(\mathbf{x}) = 100 \cdot \left(x_2 - x_1^{\,2}\right)^2 + \left(1 - x_1\right)^2$ **(11. 1)**

Subject to
$$\begin{cases} -x_1 - x_2^{\,2} \le 0 \\ -x_1^{\,2} - x_2 \le 0 \\ -x_1 - 0.5 \le 0 \\ x_1 - 0.5 \le 0 \\ x_2 - 1 \le 0 \\ -x_2 - 0.25 \le 0 \end{cases}$$ **(11. 2)**

This benchmark problem is taken from [81], and its solution is given by:

$$\hat{\mathbf{x}} = (0.5, 0.25) \quad \wedge \quad f(\hat{\mathbf{x}}) = 0.25 \tag{11. 3}$$

## 11.3.2 Second benchmark function

$$\text{Minimize } f(\mathbf{x}) = (x_1 - 2)^2 - (x_2 - 1)^2 \tag{11. 4}$$

$$\text{Subject to } \begin{cases} x_1^{\ 2} - x_2 \leq 0 \\ x_1 + x_2 - 2 \leq 0 \\ x_1 - 1 \leq 0 \\ -x_1 - 2 \leq 0 \\ x_2 - 4 \leq 0 \\ -x_2 \leq 0 \end{cases} \tag{11. 5}$$

This benchmark problem is taken from [81], and its solution is given by:

$$\hat{\mathbf{x}} = (1,1) \quad \wedge \quad f(\hat{\mathbf{x}}) = 1 \tag{11. 6}$$

## 11.3.3 Third benchmark function

$$\text{Minimize } f(\mathbf{x}) = 5 \cdot \left( x_1 + x_2 + x_3 + x_4 - \sum_{i=1}^{4} x_i^{\ 2} \right) - \sum_{i=5}^{13} x_i \tag{11. 7}$$

$$\text{Subject to } \begin{cases} 2 \cdot x_1 + 2 \cdot x_2 + x_{10} + x_{11} - 10 \leq 0 \\ 2 \cdot x_2 + 2 \cdot x_3 + x_{11} + x_{12} - 10 \leq 0 \\ 2 \cdot x_1 + 2 \cdot x_3 + x_{10} + x_{12} - 10 \leq 0 \\ -8 \cdot x_2 + x_{11} \leq 0 \\ -8 \cdot x_1 + x_{10} \leq 0 \\ -8 \cdot x_3 + x_{12} \leq 0 \\ -2 \cdot x_4 - x_5 + x_{10} \leq 0 \\ -2 \cdot x_8 - x_9 + x_{12} \leq 0 \\ -2 \cdot x_6 - x_7 + x_{11} \leq 0 \end{cases} \tag{11. 8}$$

And also subject to $$\begin{cases} -x_j \le 0 & \text{for} \quad j = 1,...,13 \\ x_j - 1 \le 0 & \text{for} \quad j = 1,...,9 \\ x_j - 100 \le 0 & \text{for} \quad j = 10, 11, 12 \\ x_{13} - 1 \le 0 \end{cases} \tag{11. 9}$$

This benchmark problem is taken from [81], and its solution is given by:

$$\hat{\mathbf{x}} = (1,1,1,1,1,1,1,1,1,3,3,3,1) \quad \wedge \quad f(\hat{\mathbf{x}}) = -15 \tag{11. 10}$$

## 11.3.4 Fourth benchmark function

Minimize $$f(\mathbf{x}) = -10.5 \cdot x_1 - 7.5 \cdot x_2 - 3.5 \cdot x_3 - 2.5 \cdot x_4 - 1.5 \cdot x_5 - 10 \cdot x_6 - 0.5 \cdot \sum_{i=1}^{5} x_i^2 \tag{11. 11}$$

Subject to $$\begin{cases} 6 \cdot x_1 + 3 \cdot x_2 + 3 \cdot x_3 + 2 \cdot x_4 + x_5 - 6.5 \le 0 \\ 10 \cdot x_1 + 10 \cdot x_3 + x_6 - 20 \le 0 \\ x_i - 1 \le 0 \quad \forall i = 1,...,5 \\ -x_i \le 0 \quad \forall i = 1,...,5 \\ x_6 - 20 \le 0 \\ -x_6 \le 0 \end{cases} \tag{11. 12}$$

This benchmark problem is taken from [81], and the best known solution is:

$$f(\hat{\mathbf{x}}) = -213 \tag{11. 13}$$

## 11.3.5 Fifth benchmark function

This benchmark problem is taken from [82]. It …*features a linear objective function subject to two nonlinear inequality polynomial constraints. The bounds on the two variables introduce four additional inequality constraints. The feasible region is almost disconnected.* [82]

Minimize $$f(\mathbf{x}) = -x_1 - x_2 \tag{11. 14}$$

Subject to
$$\begin{cases} x_2 - 2\cdot x_1^{\;4} + 8\cdot x_1^{\;3} - 8\cdot x_1^{\;2} - 2 \le 0 \\ x_2 - 4\cdot x_1^{\;4} + 32\cdot x_1^{\;3} - 88\cdot x_1^{\;2} + 96\cdot x_1 - 36 \le 0 \\ x_1 - 3 \le 0 \\ -x_1 \le 0 \\ x_2 - 4 \le 0 \\ -x_2 \le 0 \end{cases} \tag{11. 15}$$

The best known solution for this problem is given by:

$$\hat{\mathbf{x}} = (2.3295, 3.1783) \quad \wedge \quad f(\hat{\mathbf{x}}) = -5.5079 \tag{11. 16}$$

## 11.3.6 Sixth benchmark function

This benchmark problem is taken from [82]. *It features the unconstrained minimization of a non-convex function in two variables.* [82]

Minimize
$$\begin{aligned} f(\mathbf{x}) = &\left(1+(x_1+x_2+1)^2\cdot\left(19-14\cdot x_1+3\cdot x_1^{\;2}-14\cdot x_2+6\cdot x_1\cdot x_2+3\cdot x_2^{\;2}\right)\right)\cdot \\ &\cdot\left(30+(2\cdot x_1-3\cdot x_2)^2\cdot\left(18-32\cdot x_1+12\cdot x_1^{\;2}+48\cdot x_2-36\cdot x_1\cdot x_2+27\cdot x_2^{\;2}\right)\right) \end{aligned} \tag{11. 17}$$

The solution is given by:

$$\hat{\mathbf{x}} = (0, -1) \quad \wedge \quad f(\hat{\mathbf{x}}) = 3 \tag{11. 18}$$

Note that it is sufficient to set the penalization equal to zero, or the feasibility equal to one (i.e. always feasible) to deal with an unconstrained optimization problem.

Thus, the two general-purpose optimizers are tested on these six benchmark functions. The most relevant results obtained from the experiments are gathered in **Table 11. 8**. As it can be observed, both optimizers perform very well on the whole test suite. It is important to note that, since $t_{max} = 30000$, the search is not allowed to be terminated earlier than the 3000th time-step. Another important aspect to be considered that cannot be observed in the table is that the preserving feasibility method finds it very difficult to generate the initial feasible population for the third benchmark function in this test suite.

| FIRST BENCHMARK FUNCTION | | | |
|---|---|---|---|
| OPTIMIZER | Best solution<br>Worst solution | Mean best solution<br>(Standard deviation) | Mean time-steps<br>(Standard deviation) |
| PRESERVING FEASIBILITY | 2.500E-01<br>2.500E-01 | 2.500E-01<br>0.000E+00 | 3.023E+03<br>5.729E+01 |
| PENALIZATION | 2.500E-01<br>2.500E-01 | 2.500E-01<br>9.520E-09 | 3.011E+03<br>3.180E+01 |
| **SECOND BENCHMARK FUNCTION** | | | |
| OPTIMIZER | Best solution<br>Worst solution | Mean best solution<br>(Standard deviation) | Mean time-steps<br>(Standard deviation) |
| PRESERVING FEASIBILITY | 1.000E+00<br>1.000E+00 | 1.000E+00<br>0.000E+00 | 3.000E+03<br>0.000E+00 |
| PENALIZATION | 1.000E+00<br>1.000E+00 | 1.000E+00<br>0.000E+00 | 3.000E+03<br>0.000E+00 |
| **THIRD BENCHMARK FUNCTION** | | | |
| OPTIMIZER | Best solution<br>Worst solution | Mean best solution<br>(Standard deviation) | Mean time-steps<br>(Standard deviation) |
| PRESERVING FEASIBILITY | -1.500E+01<br>-1.183E+01 | -1.468E+01<br>7.847E-01 | 2.467E+04<br>5.469E+03 |
| PENALIZATION | -1.500E+01<br>-1.245E+01 | -1.421E+01<br>1.019E+00 | 1.728E+04<br>3.034E+03 |
| **FOURTH BENCHMARK FUNCTION** | | | |
| OPTIMIZER | Best solution<br>Worst solution | Mean best solution<br>(Standard deviation) | Mean time-steps<br>(Standard deviation) |
| PRESERVING FEASIBILITY | -2.130E+02<br>-2.130E+02 | -2.130E+02<br>0.000E+00 | 1.063E+04<br>9.906E+02 |
| PENALIZATION | -2.130E+02<br>-2.130E+02 | -2.130E+02<br>0.000E+00 | 2.536E+04<br>4.983E+03 |
| **FIFTH BENCHMARK FUNCTION** | | | |
| OPTIMIZER | Best solution<br>Worst solution | Mean best solution<br>(Standard deviation) | Mean time-steps<br>(Standard deviation) |
| PRESERVING FEASIBILITY | -5.508E+00<br>-5.508E+00 | -5.508E+00<br>2.725E-07 | 3.173E+03<br>1.118E+03 |
| PENALIZATION | -5.508E+00<br>-5.508E+00 | -5.508E+00<br>0.000E+00 | 3.051E+03<br>3.609E+02 |
| **SIXTH BENCHMARK FUNCTION** | | | |
| OPTIMIZER | Best solution<br>Worst solution | Mean best solution<br>(Standard deviation) | Mean time-steps<br>(Standard deviation) |
| PRESERVING FEASIBILITY | 3.000E+00<br>3.000E+00 | 3.000E+00<br>1.770E-07 | 3.008E+03<br>2.882E+01 |
| PENALIZATION | 3.000E+00<br>3.000E+00 | 3.000E+00<br>1.770E-07 | 3.003E+03<br>1.252E+01 |

**Table 11. 8**: Summary of the most significant results obtained from the optimization of the second suite of benchmark functions by means of two selected general-purpose optimizers.

### 11.3.7 Discussion

This second suite of benchmark functions tests other capabilities of the optimizers than the first suite does. The functions are noticeably less challenging than those of the first test suite, but there are a number of constraints in addition to the boundary constraints which the optimizers have to deal with. It shows that these optimizers have no trouble in solving the problems in this second test suite, except perhaps for the third benchmark problem, for which the solutions found are least accurate. Nevertheless, they are still good enough. Perhaps the greatest drawback is the difficulty encountered by the "preserving feasibility" technique in generating the initial population, which happened to be too time consuming.

## 11.4 Closure

The algorithms selected as the proposed general-purpose optimizers by the end of **Chapter 10** were tested on a first suite of challenging benchmark functions with only boundary constraints with doubtful success. Hence, they were replaced by the GP-PSO$^{(s.w.d.)}$ equipped with the "preserving feasibility" method on the one hand, and with the "penalization" method on the other. These two optimizers were tested on two sets of benchmark functions composed of six functions each, including very complex and high-dimensional functions with hyper-cube-like boundary constraints; one 2-dimensional, unconstrained, non-convex function; and five low-dimensional functions with numerous constraints. The selected optimizers performed very well on all tests. Therefore, they are proposed here as the outcome of this thesis.

It must be noted, however, that the "penalization" method used here is a very basic one, where the penalty coefficients are arbitrarily set, the same for every variable, and constant along the whole search. Typically, the coefficients should be lower during the early stages of the search so as to favour exploration of the infeasible search-space, and be progressively increased as the search progresses so as to favour exploration of the feasible search-space during the late stages of the search. A too high constant penalization might result in premature convergence towards sub-optimal solutions, while a too low constant penalization might result in infeasible solutions presenting themselves with lower conflicts than the feasible ones. A more in-depth analysis of the constraint-handling techniques is left for future work.

# Chapter 12

# TRAINING OF A SIMPLE ARTIFICIAL NEURAL NETWORK

A fairly extensive introduction to the artificial neural networks was carried out along Chapter 3. It was claimed that they can be applied to different kinds of problems such as the simulation of the human brain, function approximations, and classification tasks. Perhaps the simplest and most graphical example that reveals the limited capabilities of a single artificial neuron is that of the logical "xor" problem, whose two classes are not linearly separable. Hence at least one hidden layer is required. However, the training of multilayer networks is a hard task, which in fact delayed the popularization of the paradigm. Traditional techniques are gradient-based, with the corresponding risk of getting trapped in poor local optima. Besides, these techniques need to be adapted for different kinds of networks, and even for different transfer functions. Non-traditional search techniques such as evolutionary algorithms and particle swarm optimization are especially suitable for this task. They are able to escape local optima, and require no adaptation to deal with different kinds of networks. Therefore, the simple logical "xor" problem is selected here to illustrate the suitability of the particle swarm optimizers for the training of artificial neural networks. The training of more complex networks is left for future work.

## 12.1 Introduction

An overview of the field of artificial intelligence (AI) was carried out along **Chapter 3**, with the stress put on the artificial neural networks (ANNs). The reason for this was two-fold: first, the connectionist paradigm comprises one of the main non-traditional paradigms of AI, which heavily relies on the concept of emergence, very much like the PSO paradigm; second, the particle swarm optimizers are especially suitable for dealing with the very hard task of training these networks.

The use of the PSO paradigm for the optimization of the architecture of the networks is still not as widespread as its use for the training of their weights. Nevertheless, numerous attempts have been already performed. For instance, Kennedy et al. [47] indirectly optimize the structure of the network by adding the slope of the sigmoid transfer function as yet another variable. Thus, if the slope is sufficiently small, the output of a neuron is rather constant regardless of the aggregated signal. Hence the neuron can be removed, since its effect can be replaced by conveniently modifying the values of the bias of the other neurons in the same layer. In turn, if the slope is sufficiently high, the sigmoid resembles a step activation function. Binary or discrete versions of the algorithm can also be used to train the network's architecture. The issue of optimizing the structure of ANNs is beyond the scope of this thesis.

One of the most interesting features of the multi-layer perceptron (refer to **Chapter 3**) is its ability to approximate any function with finitely many discontinuities to any degree of precision, provided the network is made up of at least one hidden layer and a sufficient number of non-linear neurons within it (e.g. with a sigmoid transfer function). One problem is that if the number of neurons is too big, the training of the network might result in over-fitting, since the network approximates so well that it even considers the noise that is so common in the experimental data typically used for the training data set. In contrast, if the network is not complex enough, it cannot approximate many functions. The issue of the proper complexity of the feed-forward neural networks is also beyond the scope of this work.

The logical "xor" problem is perhaps the simplest and most straightforward example to illustrate the need of at least one hidden layer. Therefore, it is frequently addressed in the literature as a first benchmark problem to test training algorithms. Since the aim here is not to deal with neural networks in details but to illustrate the suitability of the PSO paradigm for training their weights, the interesting problem of optimizing networks designed for function approximation is left for future work. The scope of the present chapter is limited to training a simple two-layer network composed of three artificial neurons to learn the logical "xor" problem. Further work is required to deal with more complex real-world problems, and to compare this algorithm to practical, existing training methods.

## 12.2 The logical "xor" problem

As discussed along **Chapter 3**, the use of linear artificial neurons for classification tasks is limited to patterns that are linearly separable. Since some problems such as the logical "xor" problem are not linearly separable, different linear artificial neurons can successively subdivide the space so that even classes which comprise disjointed regions of the search-space can be recognized and delimitated. Thus, two straight lines are sufficient to separate the two classes corresponding to the “xor” problem, as it can be observed in **Fig. 12. 1** – left: one class belongs within the two lines, while the other outside that region. The network sketched in **Fig. 12. 1** - middle can represent these two regions, where each neuron in the hidden layer performs a linear separation, and the neuron in the output layer makes the final decision.

There are infinite solutions to this problem, corresponding to different pairs of straight lines that can separate these two classes. Hence there are infinite combinations of weights which can solve the problem by means of this ANN. A possible solution obtained by common sense was shown in **Fig. 3.14**.

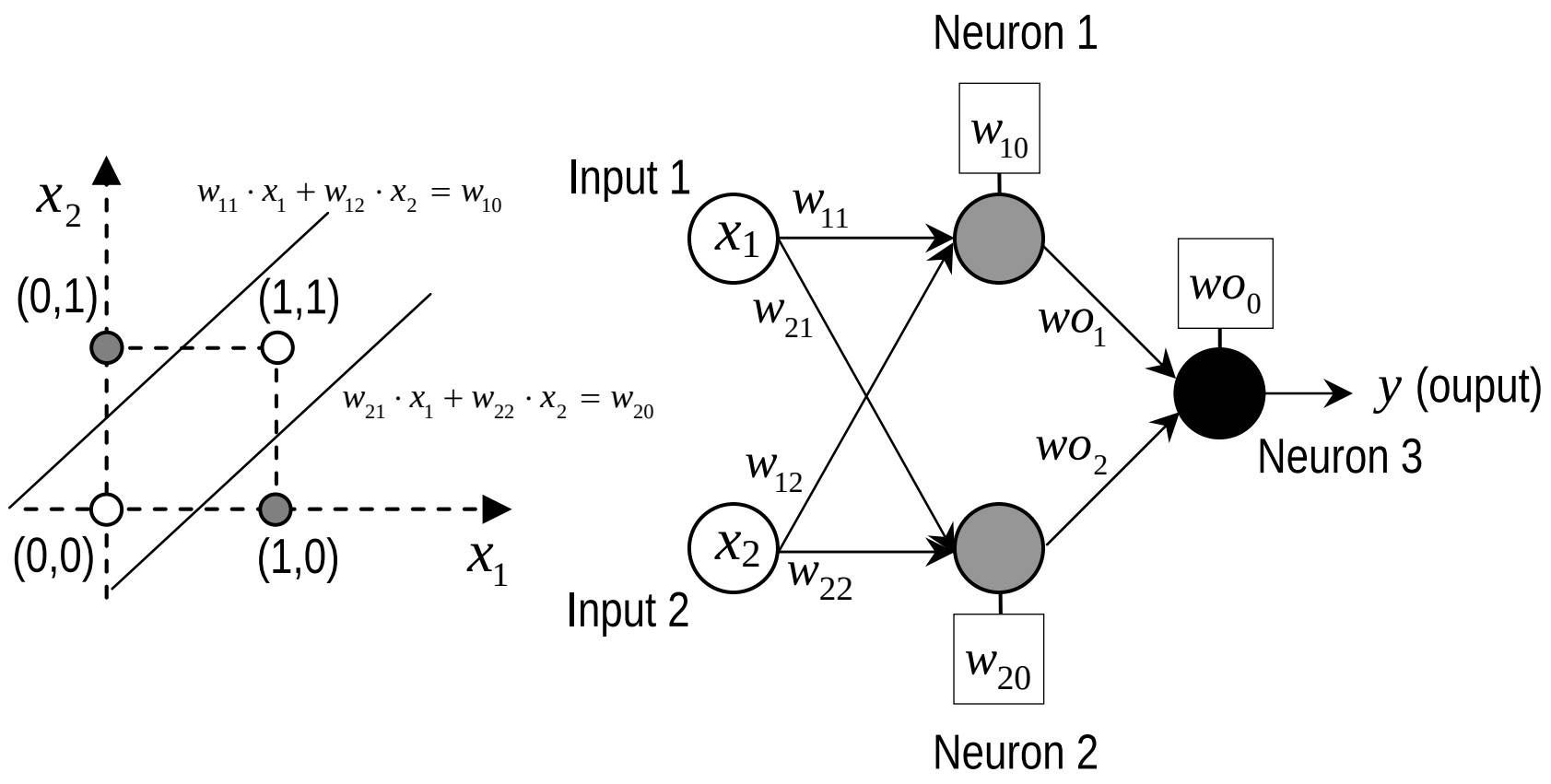

| $x_1$ | $x_2$ | *t* (target) |
|---|---|---|
| 1 | 0 | 1 |
| 0 | 1 | 1 |
| 1 | 1 | 0 |
| 0 | 0 | 0 |

**Fig. 12. 1**: Left: The logical operation “xor”. The two classes cannot be separated by a single perceptron. A proposed solution is given by the two straight lines: one class is between the lines, while the other is outside.

Middle: A proposed ANN that solves the problem. The upper artificial neuron, whose decision boundary is the upper straight line on the left, makes a decision. The lower artificial neuron, whose decision boundary is the lower straight line, makes a decision as well. A third artificial neuron receives both decisions and makes its own (final) decision.

Right: Truth table and training (also validation) data set for the logical operation “xor”.

To solve the problem by means of a search algorithm such as a particle swarm optimizer, the space of the nine weights of the network needs to be explored. Thus, the position of every particle in the search-space stands for a potential solution to the problem. The conflict function is given by an error function, which is typically computed as the squared difference between the output that results from a given set of weights (i.e. from a position in the search-space) when a set of inputs is presented to the network, and the correct output corresponding to those inputs. Therefore, the network requires a training data set of inputs and their correct corresponding outputs to learn its weights.

The error function, which stands for the conflict function to be minimized, is usually computed as the average of the squared errors corresponding to all the patterns presented to the network. Typically, numerous patterns are presented—especially for noisy data—, and another group of patterns are kept aside during the training in order to be used for the validation of the trained network. This is not possible for the logical “xor” problem because there are a small, finite number of patterns (see **Fig. 12. 1** - right). Nevertheless, these patterns do not contain noise. Hence the training and the validation data sets are the same in this case.

The problem was solved by two different networks, both with the architecture and with the training data set shown in **Fig. 12. 1**. The only difference between the two networks is the type of transfer function. The first network simply uses three perceptrons (i.e. artificial neurons with a step activation function), while the second uses three sigmoid transfer functions. The training was successful in both cases. Since the training data set does not contain noise, the optimum conflict value should equal zero.

The weights were randomly initialized within the interval $[-20,20]$ for both cases, because the sigmoid function saturates for aggregated inputs outside this interval.

## 12.2.1 Three perceptrons

For the case of the network composed of three perceptrons, the search was terminated at the 1067$^{th}$ time-step because the second set of termination conditions was attained. That is to say, the particles did not achieve a high degree of clustering. This can also be inferred from the evolution of the best and average conflicts shown in **Fig. 12. 2**, and from the evolution of the relative errors regarding the conflict values and regarding the particles’ positions shown in **Fig. 12. 3** and **Fig. 12. 4**, respectively.

The nine weights obtained for this network, which resulted from a single run of the optimizer, are as follows:

- $w_{10} = 15.8042549716352$
- $w_{11} = 7.51536132271753$
- $w_{12} = 13.1234408094416$
- $w_{20} = 7.6427733795084$

- $w_{21} = 17.5852335288703$
- $w_{22} = 14.6609333927224$
- $wo_0 = 4.38071818907651$
- $wo_1 = -13.5666224959979$
- $wo_2 = 15.129119311375$

Since the artificial neurons consist of perceptrons, the best conflict found equals zero, and the outputs that result from entering the inputs from the training data set into the trained network to validate the results are the exact outputs from training the data set. That is to say:

- 0 xor 0 $\Rightarrow$ 0
- 1 xor 1 $\Rightarrow$ 0
- 0 xor 1 $\Rightarrow$ 1
- 1 xor 0 $\Rightarrow$ 1

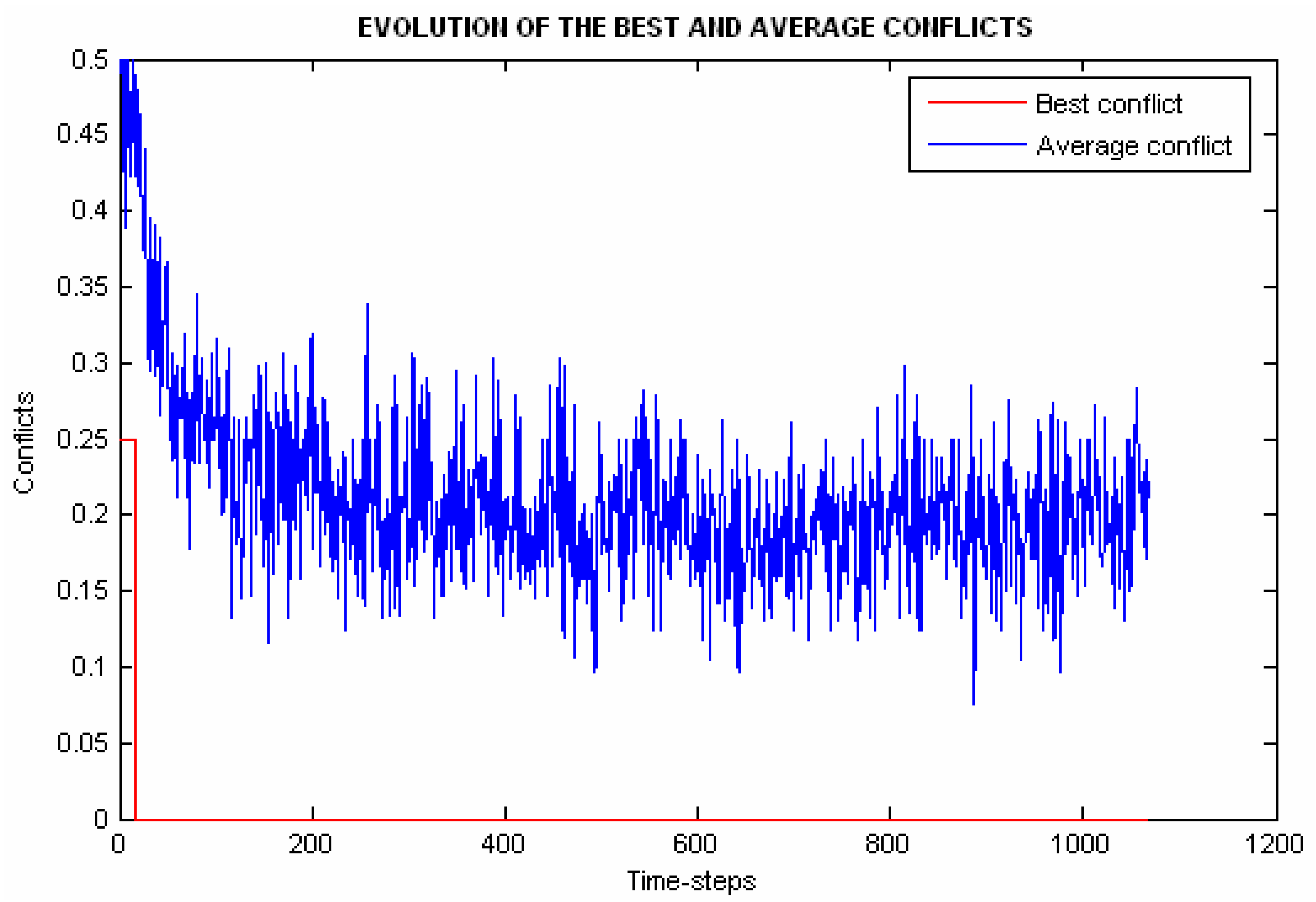

**Fig. 12. 2**: Evolution of the best and average conflicts corresponding to a PSO training an ANN composed of two perceptrons in the hidden layer and one perceptron in the output layer for the logical "xor" problem.

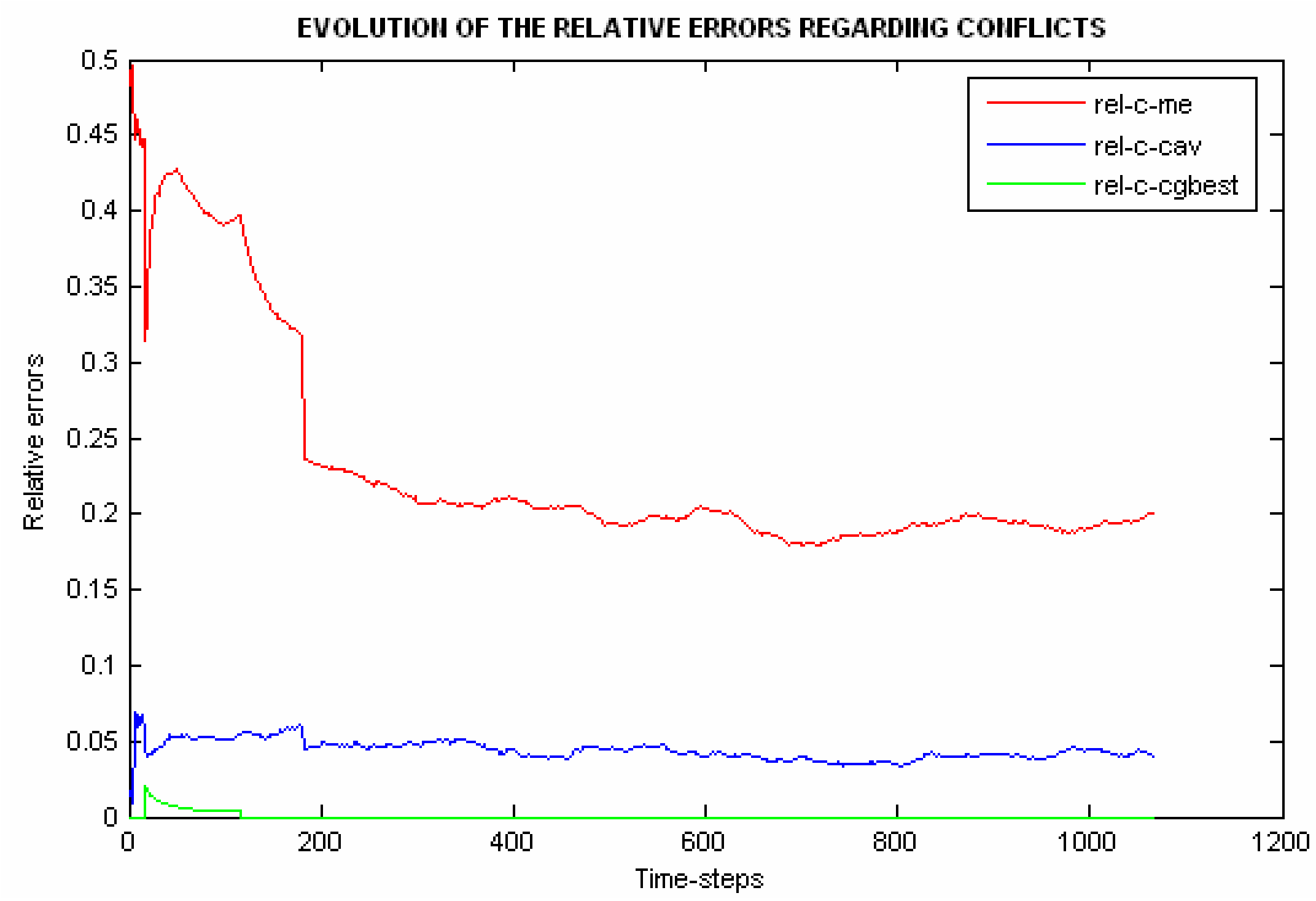

**Fig. 12. 3**: Evolution of the relative errors regarding the conflict values corresponding to a PSO training an ANN composed of two perceptrons in the hidden layer and one perceptron in the output layer for the logical “xor” problem.

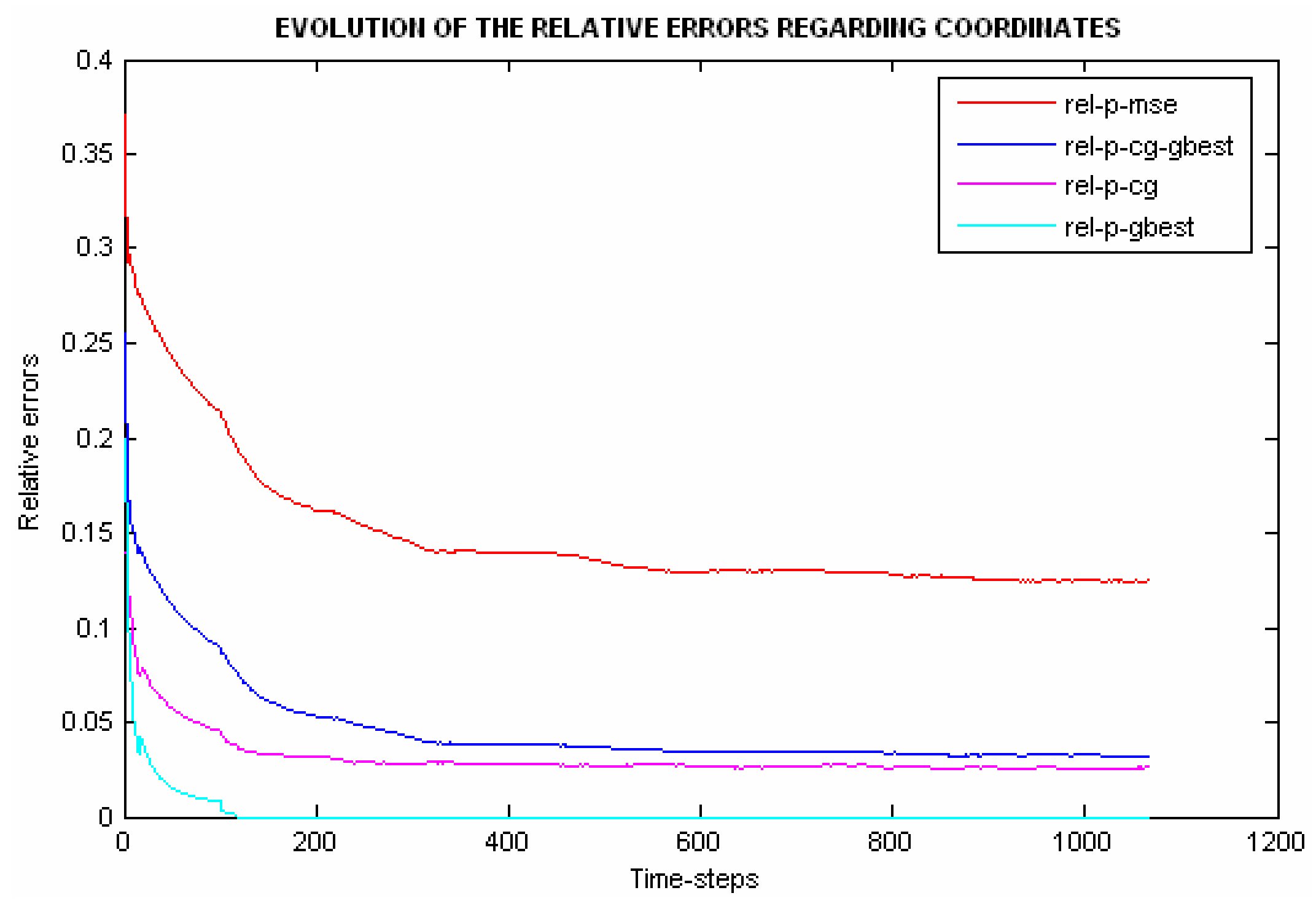

**Fig. 12. 4**: Evolution of the relative errors regarding the particles’ positions corresponding to a PSO training an ANN composed of two perceptrons in the hidden layer and one perceptron in the output layer for the logical “xor” problem.

### 12.2.2 Three sigmoid transfer functions

For the case of the network composed of three nonlinear artificial neurons equipped with sigmoid transfer functions, the search was terminated after the 3000 time-steps permitted without attaining the error condition. Nevertheless, the particles achieved a high degree of clustering, as it can be observed in **Fig. 12. 5** to **Fig. 12. 7**. The best conflict found is equal to 2.14412353741792E-09, and the nine weights resulting from a single run are as follows:

- $w_{10} = 9.6413882483196$
- $w_{11} = 19.9999999994233$
- $w_{12} = 19.9999999997995$
- $w_{20} = 19.9999999999967$
- $w_{21} = 13.3411633764489$
- $w_{22} = 13.3411694711353$
- $wo_0 = 9.9935242984494$
- $wo_1 = 20$
- $wo_2 = -19.9999999999999$

The validation of the trained network is performed by offering the inputs of the truth table to the network, and then checking the correctness of the outputs. Since the transfer functions consist of sigmoid functions, the output can now take continuous values within the open interval $]0,1[$. The outputs corresponding to the inputs of the training data set are as follows:

- 0 xor 0 $\Rightarrow$ 4.575221e-005
- 1 xor 1 $\Rightarrow$ 4.685066e-005
- 0 xor 1 $\Rightarrow$ 9.999537e-001
- 1 xor 0 $\Rightarrow$ 9.999537e-001

Extending the search does not result in improvement, unless the permitted interval for the values of the variables, $[-20,20]$, is enlarged. This is because of the features of the sigmoid function (refer to **Fig. 3.13**). The evolution of the best and average conflicts, of the relative errors regarding the conflict values, and of the relative errors regarding the particles' positions is shown in **Fig. 12. 5**, in **Fig. 12. 6**, and in **Fig. 12. 7**, respectively:

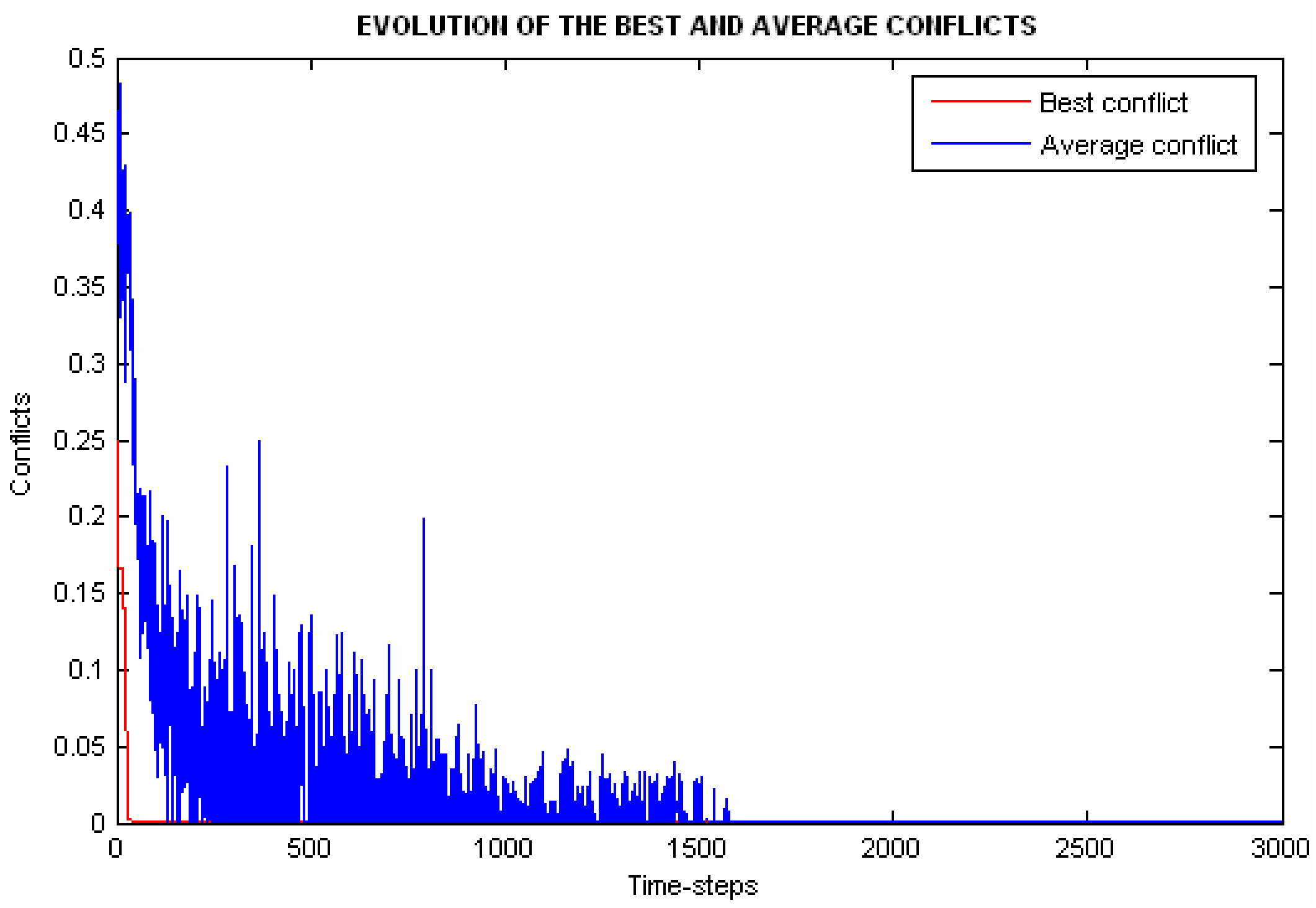

**Fig. 12. 5**: Evolution of the best and average conflicts corresponding to a PSO training an ANN for the logical "xor" problem, where the network is composed of two artificial neurons in the hidden layer and one in the output layer, all of them equipped with a sigmoid transfer function.

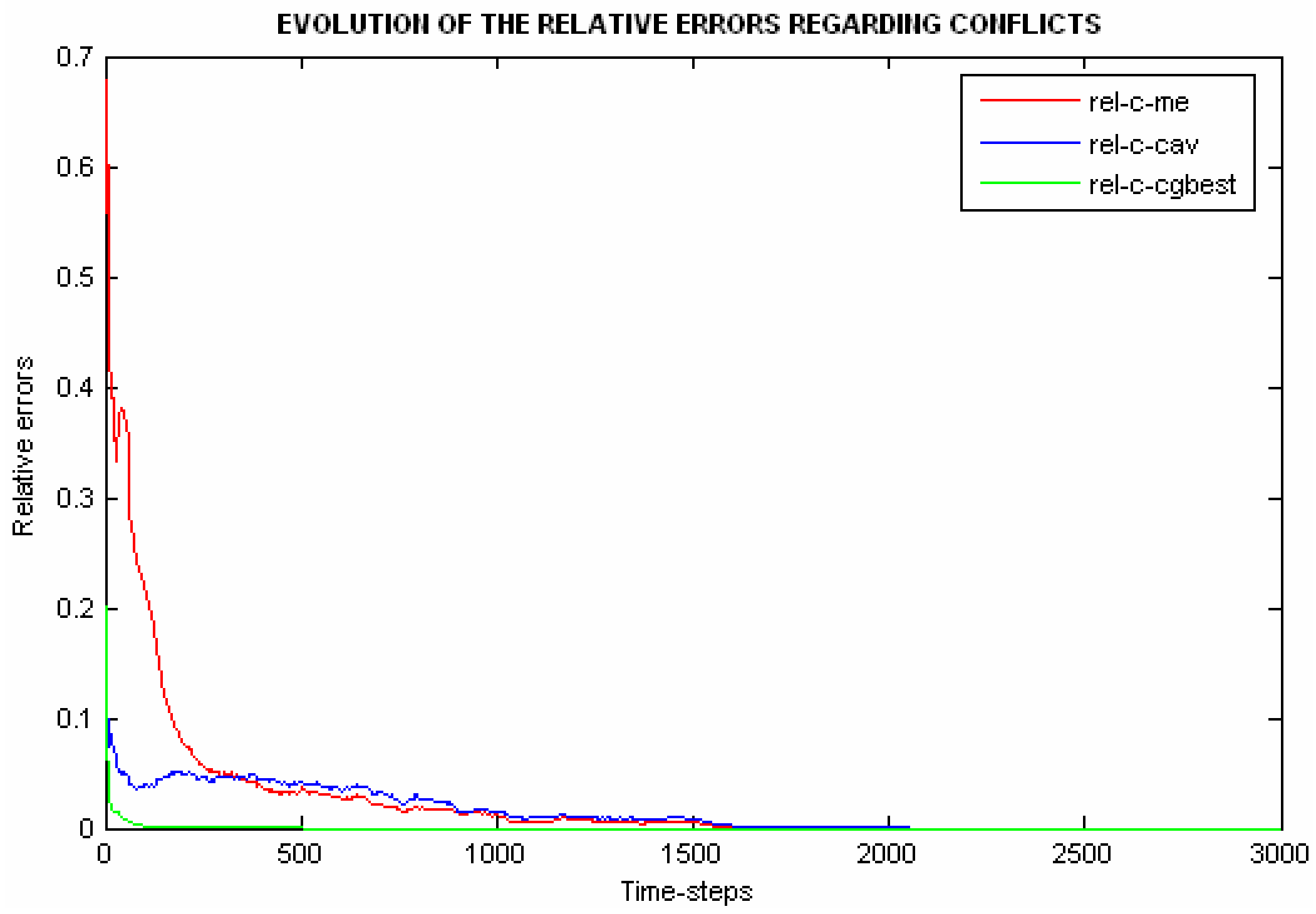

**Fig. 12. 6**: Evolution of the relative errors regarding the conflict values corresponding to a PSO training an ANN for the logical "xor" problem, where the network is composed of two artificial neurons in the hidden layer and one in the output layer, all of them equipped with a sigmoid transfer function.

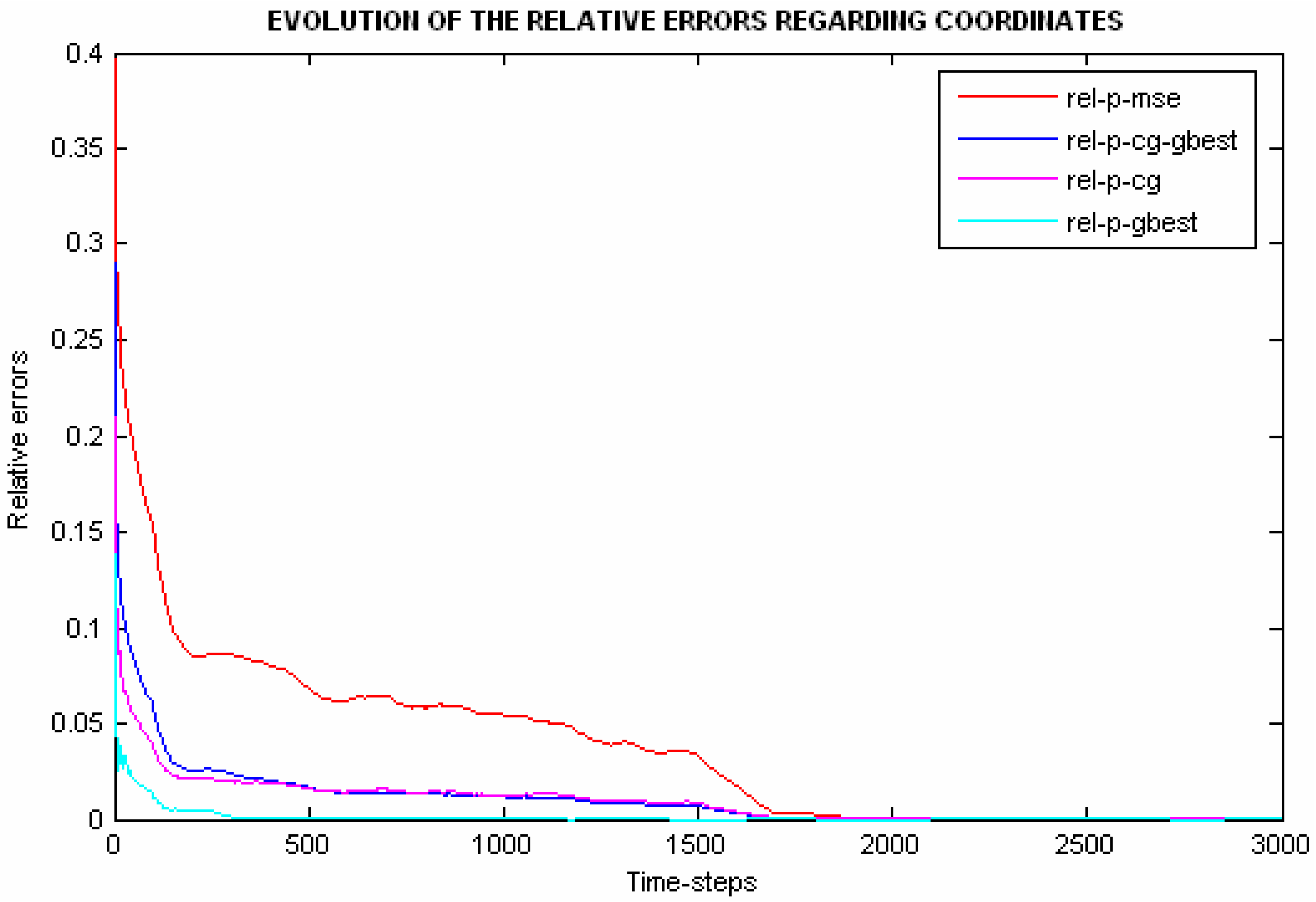

**Fig. 12. 7**: Evolution of the relative errors regarding the particles' positions corresponding to a particle swarm optimizer training an artificial neural network for the logical "xor" problem, where the network is composed of two artificial neurons in the hidden layer and one in the output layer, all of them equipped with a sigmoid transfer function.

# 12.3 Closure

The suitability of the particle swarm optimizers for the training of artificial neural networks was illustrated by successfully training two simple networks, composed of three artificial neurons each, for the logical "xor" problem. This problem is indeed too simple to illustrate the real power of the paradigm. In fact, one of the most interesting features is that the same paradigm can very well train a feed-forward network with continuous transfer functions, a feed-forward network with discontinuous activation functions, and recurrent networks. It can also optimize linear and non-linear, and convex and non-convex optimization problems. It clearly comprises a very robust, general-purpose problem-solving technique.

The study of the training of more complex neural networks as well as a thorough comparative study of the performance of this paradigm in relation to traditional training techniques and to non-traditional methods such as evolutionary algorithms is undoubtedly worthwhile. This is left for future work due to time constraints.

# Chapter 13

# ENGINEERING PROBLEMS

The proposed general-purpose optimizers equipped with the “preserving feasibility” and with the “penalization” methods are finally applied to a set of three engineering problems. The first one is rather academic, while the other two are typical benchmark optimization problems. The effectiveness of the proposed optimizers is illustrated by comparing the results obtained here for the last two problems to some results reported in the literature.

## 13.1 Introduction

The study of the particle swarm optimization paradigm and the development of a general-purpose optimizer were carried out from **Chapter 6** to **Chapter 10**. Some applications to function optimization were carried out along **Chapter 11**, and the application to the training of a simple two-layer and three-neuron artificial neural network was presented in **Chapter 12**. This thesis is concluded by showing the effectiveness of the two proposed general-purpose optimizers—the GP-PSO$^{(s.d.w.)}$ with the “preserving feasibility” technique on the one hand, and with the “penalization” method on the other—in solving three engineering problems.

Although the problems are rather academic in the sense that they are not complex enough to show the real capabilities of the algorithm, they serve the function of illustrating its suitability for dealing with real-world engineering problems. As it can be seen from the next sections, the optimizers are well able to deal with the problems presented here.

It should be noted that this chapter is, after all, just an extension to the function optimization applications carried out along **Chapter 11**. The only difference is that the physical problem represented by the formulation of the optimization problem is also presented.

It is also important to note that, since the aim here is to illustrate the applications rather than to analyze the quantitative results, only a single run is carried out for each problem, unless stated otherwise. The maximum number of time-steps allowed for the search to go through is set to 3000 for the first two problems, and to 10000 for the last one.

## 13.2 The scaffolding system

This problem is taken from [63]. Consider the scaffolding system of **Fig. 13. 1**, where loads $x_1$ and $x_2$ are applied at certain points of beams 2 and 3, respectively. Ropes A and B can bear a maximum weight of 300 kg each, C and D of 200 Kg, and E and F, a maximum of 100 Kg each. Find the maximum load $x_1 + x_2$ the system can bear without failure in equilibrium of forces and moments, the optimal loads $x_1$ and $x_2$, and the optimal points $x_3$ and $x_4$ where the loads must be applied, assuming that the weight of ropes and beams is negligible.

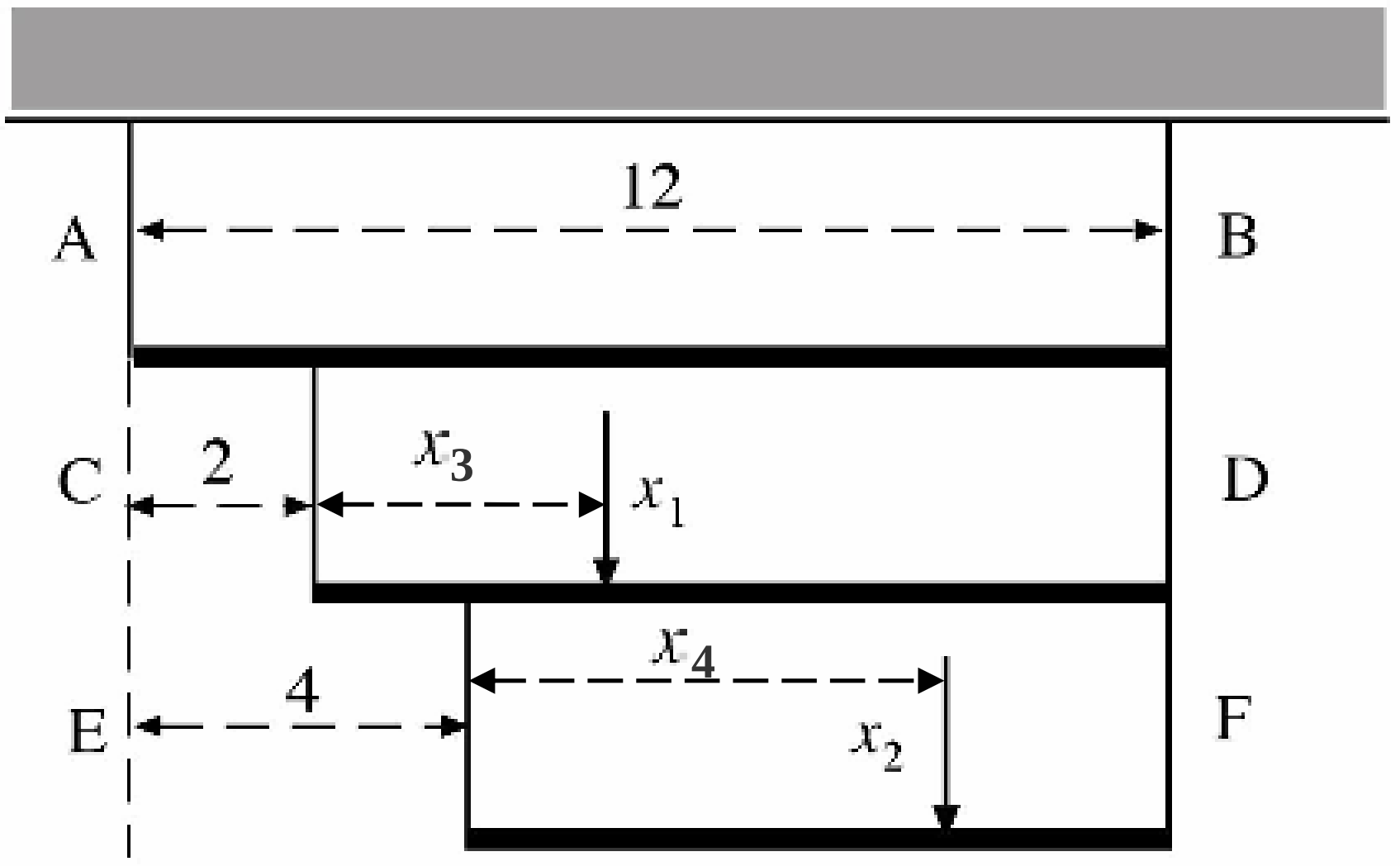

**Fig. 13. 1**: Sketch of the "scaffolding system problem", taken from [63] and slightly modified.

The problem can be formulated as follows:

Maximize $f(\mathbf{x}) = x_1 + x_2$ **(13. 1)**

$$\text{Subject to} \begin{cases} -x_1 \le 0 \\ x_1 - 400 \le 0 \\ -x_2 \le 0 \\ x_2 - 200 \le 0 \\ -x_3 \le 0 \\ x_3 - 10 \le 0 \\ -x_4 \le 0 \\ x_4 - 8 \le 0 \end{cases} \quad \textbf{(13. 2)}$$

And also subject to
$$\begin{cases} x_2 \cdot x_4 - 800 \le 0 \\ 8 \cdot x_2 - x_2 \cdot x_4 - 800 \le 0 \\ 2 \cdot x_2 + x_1 \cdot x_3 + x_2 \cdot x_4 - 2000 \le 0 \\ 10 \cdot x_1 + 8 \cdot x_2 - x_1 \cdot x_3 - x_2 \cdot x_4 - 2000 \le 0 \\ 2 \cdot x_1 + 4 \cdot x_2 + x_1 \cdot x_3 + x_2 \cdot x_4 - 3600 \le 0 \\ 10 \cdot x_1 + 8 \cdot x_2 - x_1 \cdot x_3 - x_2 \cdot x_4 - 3600 \le 0 \end{cases} \quad \textbf{(13. 3)}$$

The upper limit for $x_1$ and $x_2$ are constraints added here to the problem formulated by Pedregal [63] in order to define a hyper-rectangle that contains the feasible search-space. Note that these limits are consistent with the maximum loads that the corresponding ropes can bear.

Since the implemented optimizers are minimizers, the problem was turned into minimizing $f'(\mathbf{x}) = -x_1 - x_2$. Hence, the maximum load is given by $-cgbest$.

The problem was solved by the GP-PSO$^{(s.d.w.)}$ equipped with the "preserving feasibility" technique on the one hand, and with the "penalization" technique on the other.

## 13.2.1 Preserving feasibility

The results obtained by this optimizer are as follows:

- $x_1 = 257.752960223542$
- $x_2 = 142.247039776458$
- $x_3 = 4.50800708531484$
- $x_4 = 3.89149574127093$
- $x_1 + x_2 = 400$

The evolution of the best and average conflicts is shown in **Fig. 13. 2**.

## 13.2.2 Penalization

The results obtained by this optimizer are as follows:

- $x_1 = 278.191757619473$

- $x_2 = 121.808242380625$
- $x_3 = 5.1191220086385$
- $x_4 = 2.72794319738439$
- $x_1 + x_2 = 400$

As it can be observed, the maximum load equals 400 in both cases, which is equal to the sum of the loads that the ropes C and D can bear. However, there is more than one position of the loads that satisfy all the constraints. In fact, different values of the object variables are obtained for different runs of the same algorithm.

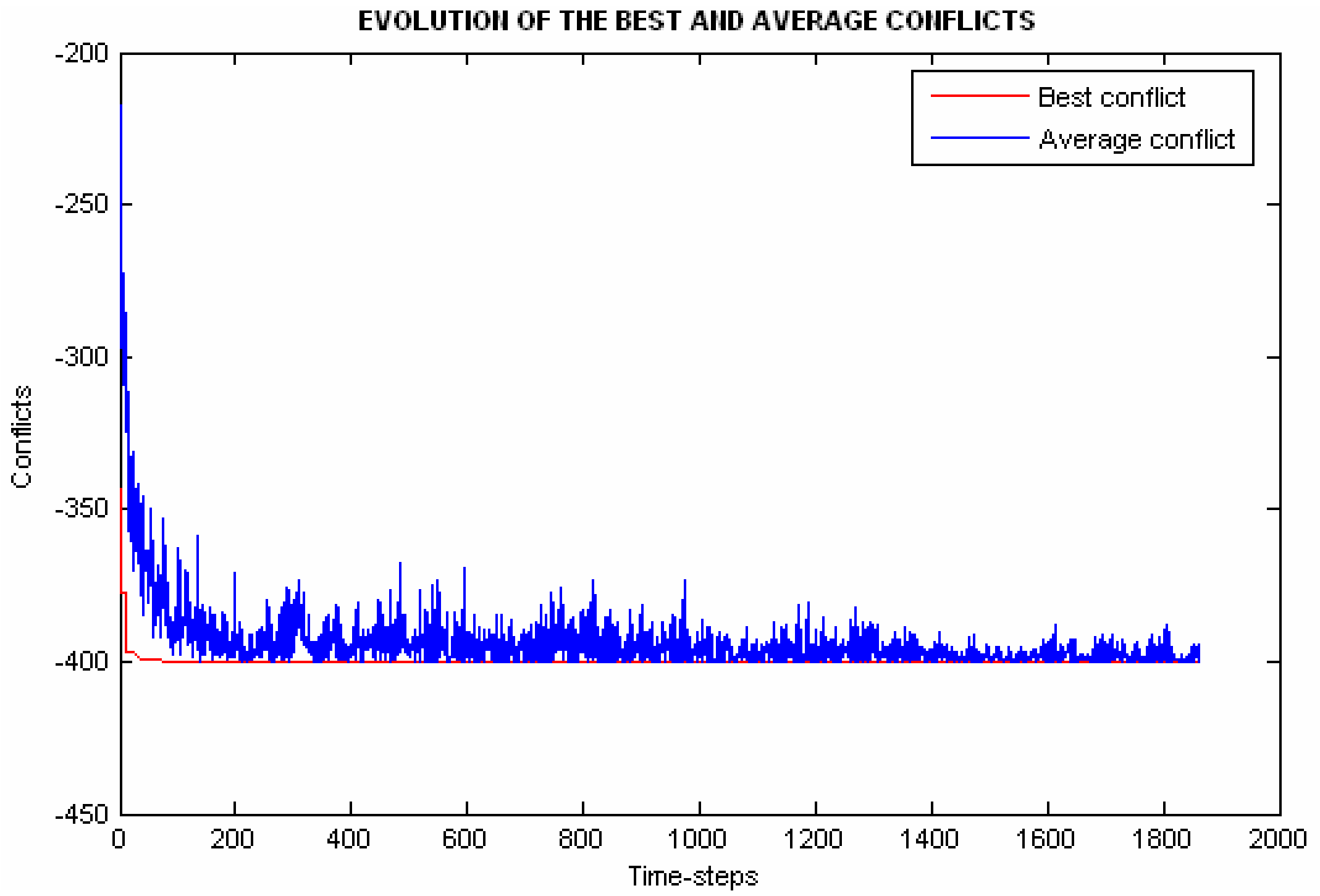

**Fig. 13. 2**: Evolution of the best and average conflicts for the GP-PSO$^{(s.d.w.)}$ equipped with the "preserving feasibility" technique optimizing the scaffolding system.

The evolution of the best and average conflicts for the GP-PSO$^{(s.d.w.)}$ equipped with the "penalization" method is shown in **Fig. 13. 3**. Since the penalization of the infeasible solutions leads to huge conflict values, a zoom of **Fig. 13. 3** is shown in **Fig. 13. 4**.

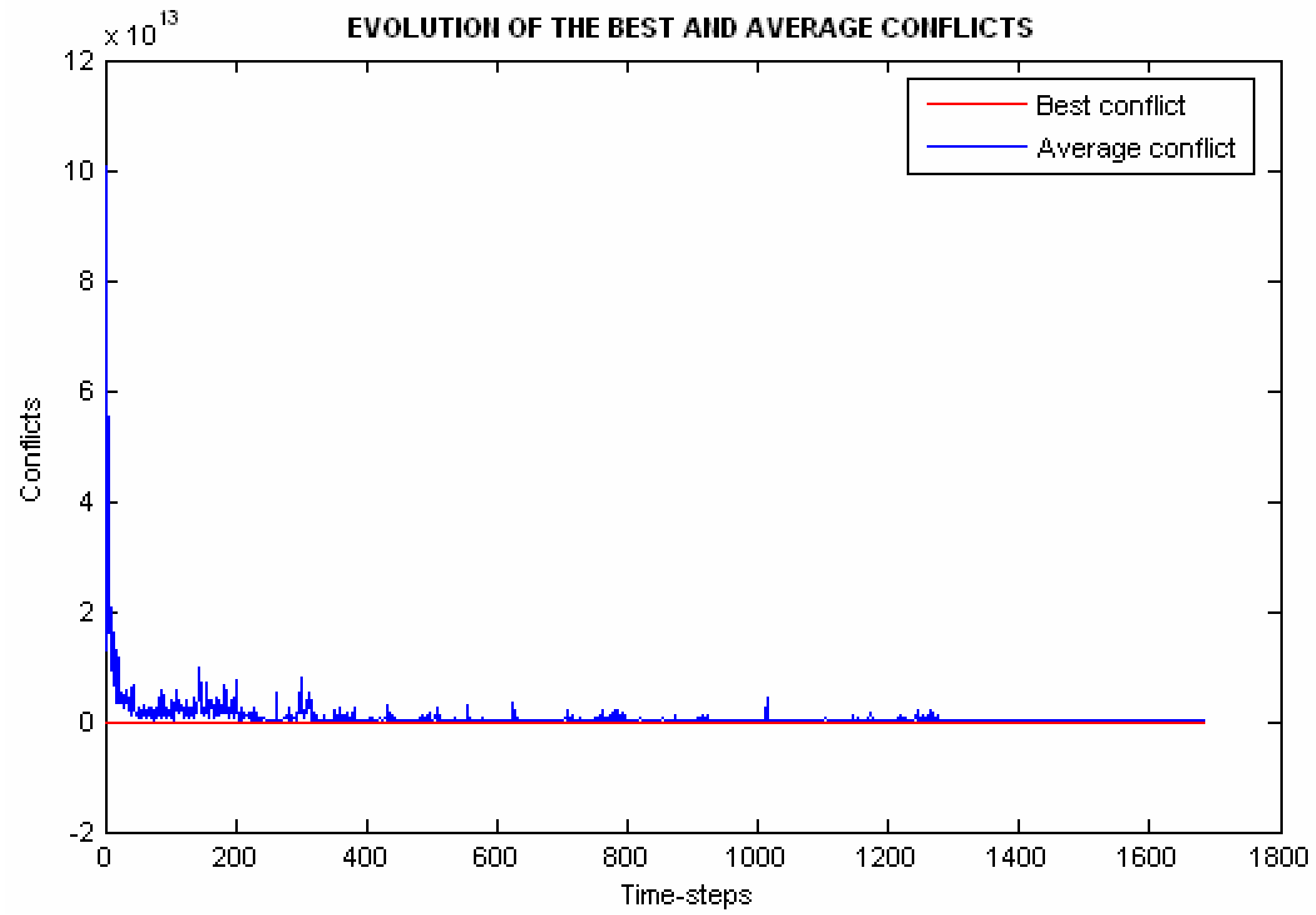

**Fig. 13. 3**: Evolution of the best and average conflicts for the GP-PSO$^{(s.d.w.)}$ equipped with the "penalization" technique optimizing the scaffolding system.

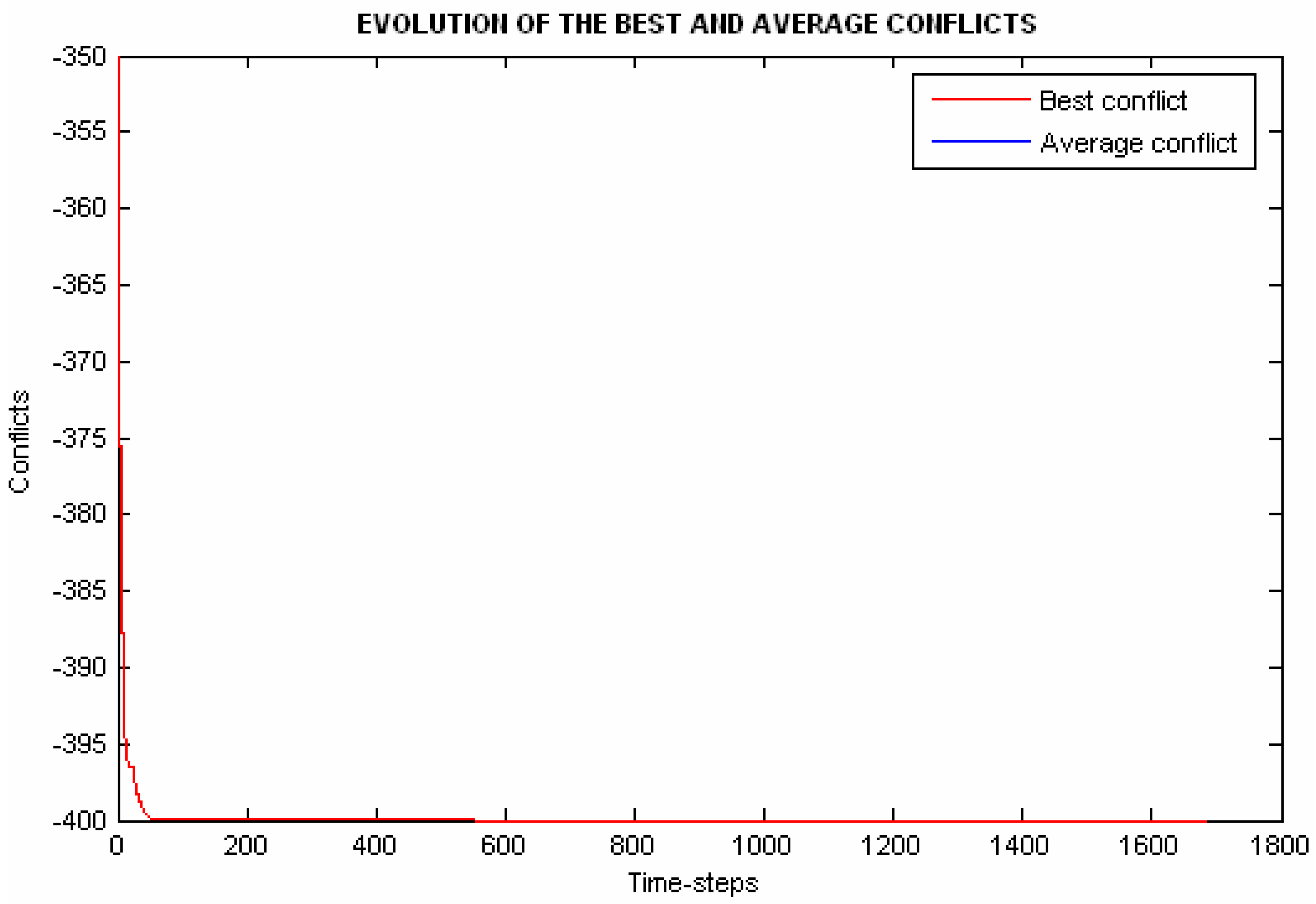

**Fig. 13. 4**: Evolution of the best and average conflicts for the GP-PSO$^{(s.d.w.)}$ equipped with the "penalization" technique optimizing the scaffolding system (zoomed).

# 13.3 The two-bar truss

This problem is taken from Foryś [36]. *The design of a two-bar truss was one of the examples presented in the original paper by Svanberg (1987). The characteristic feature of the problem, underlined by the author, is that neither CONLIN (Fleury and Braibant, 1986) nor "traditional" Sequential Linear Programming can converge to the optimal solution.* [36]

A sketch of the problem can be observed in **Fig. 13. 5**. The cross-sectional area of the bars and half the distance between supports that minimize the weight of the truss are sought for.

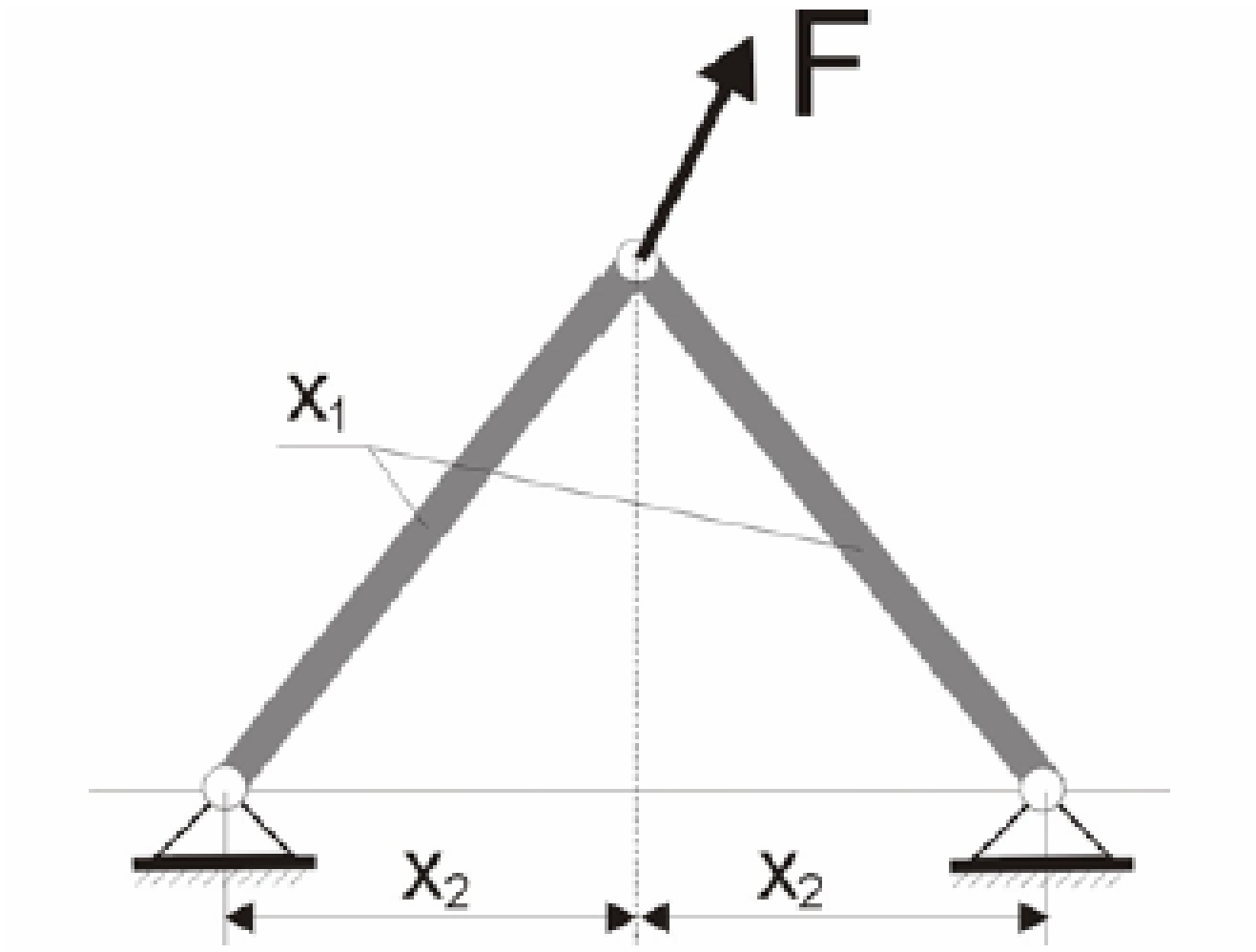

**Fig. 13. 5**: Sketch of the "two-bar truss problem", taken from [36].

The problem can be formulated as follows:

Minimize $f(\mathbf{x}) = x_1 \cdot \sqrt{1 + x_2^{\ 2}}$ **(13. 4)**

$$\text{Subject to} \begin{cases} -x_1 + 0.2 \le 0 \\ x_1 - 4 \le 0 \\ -x_2 + 0.1 \le 0 \\ x_2 - 1.6 \le 0 \\ 0.124 \cdot \sqrt{1 + x_2^{\ 2}} \cdot \left( \dfrac{8}{x_1} + \dfrac{1}{x_1 \cdot x_2} \right) - 1 \le 0 \\ 0.124 \cdot \sqrt{1 + x_2^{\ 2}} \cdot \left( \dfrac{8}{x_1} - \dfrac{1}{x_1 \cdot x_2} \right) - 1 \le 0 \end{cases} \qquad \textbf{(13. 5)}$$

## 13.3.1 Preserving feasibility

The results obtained by this optimizer are as follows:

- $x_1 = 1.41162976487056$
- $x_2 = 0.377075377432421$
- $cgbest = 1.50865241753044$

The history of the particles' positions is shown in **Fig. 13. 6**:

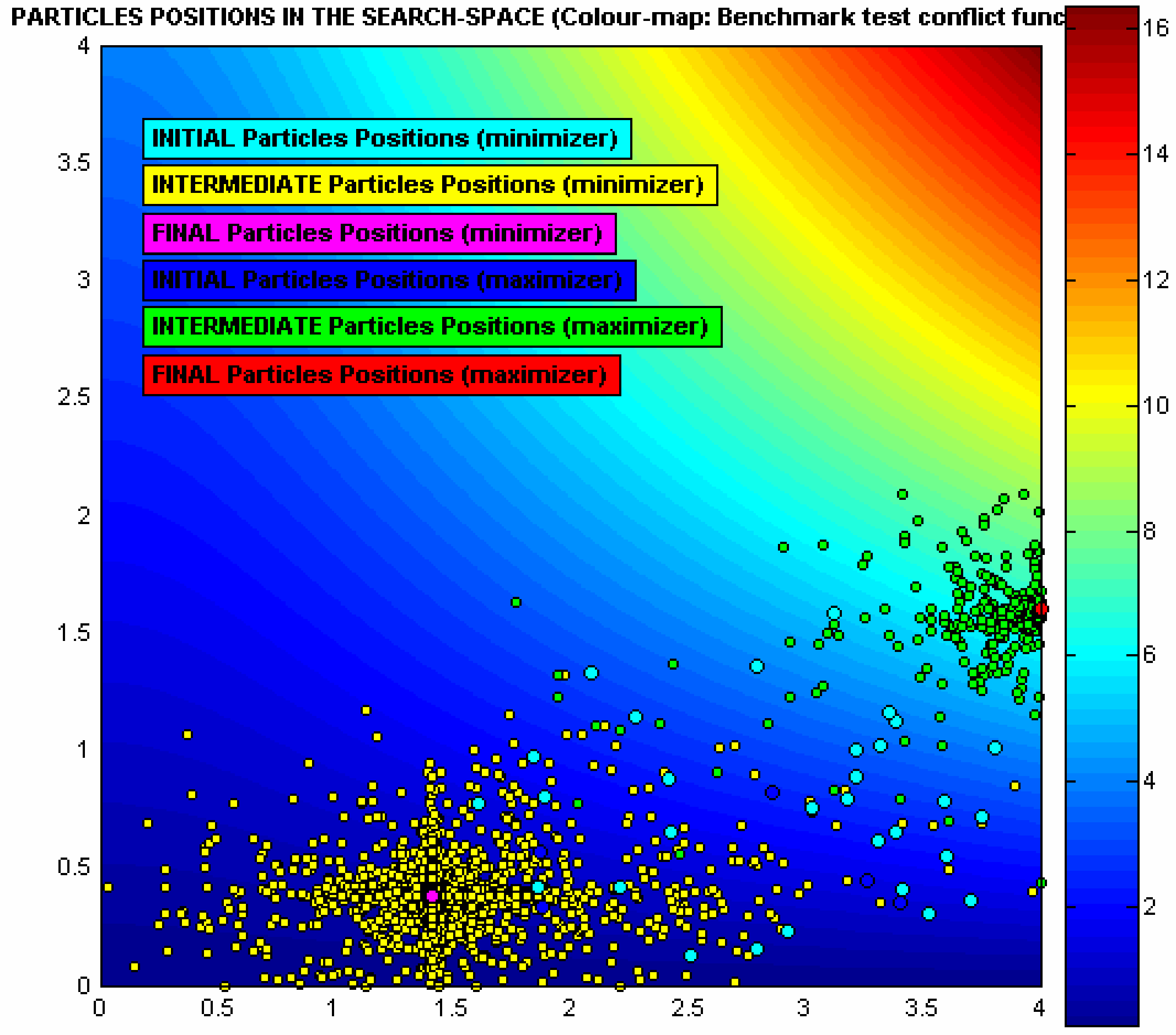

**Fig. 13. 6**: History of the particles' postions for the GP-PSO$^{(s.d.w.)}$ equipped with the "preserving feasibility" method optimizing the two-bar truss problem.

The evolution of the best and average conflicts is shown in **Fig. 13. 7**, the evolution of the relative errors regarding the conflict values is shown in **Fig. 13. 8**, and the evolution of the relative errors regarding the particles' positions is shown in **Fig. 13. 9**:

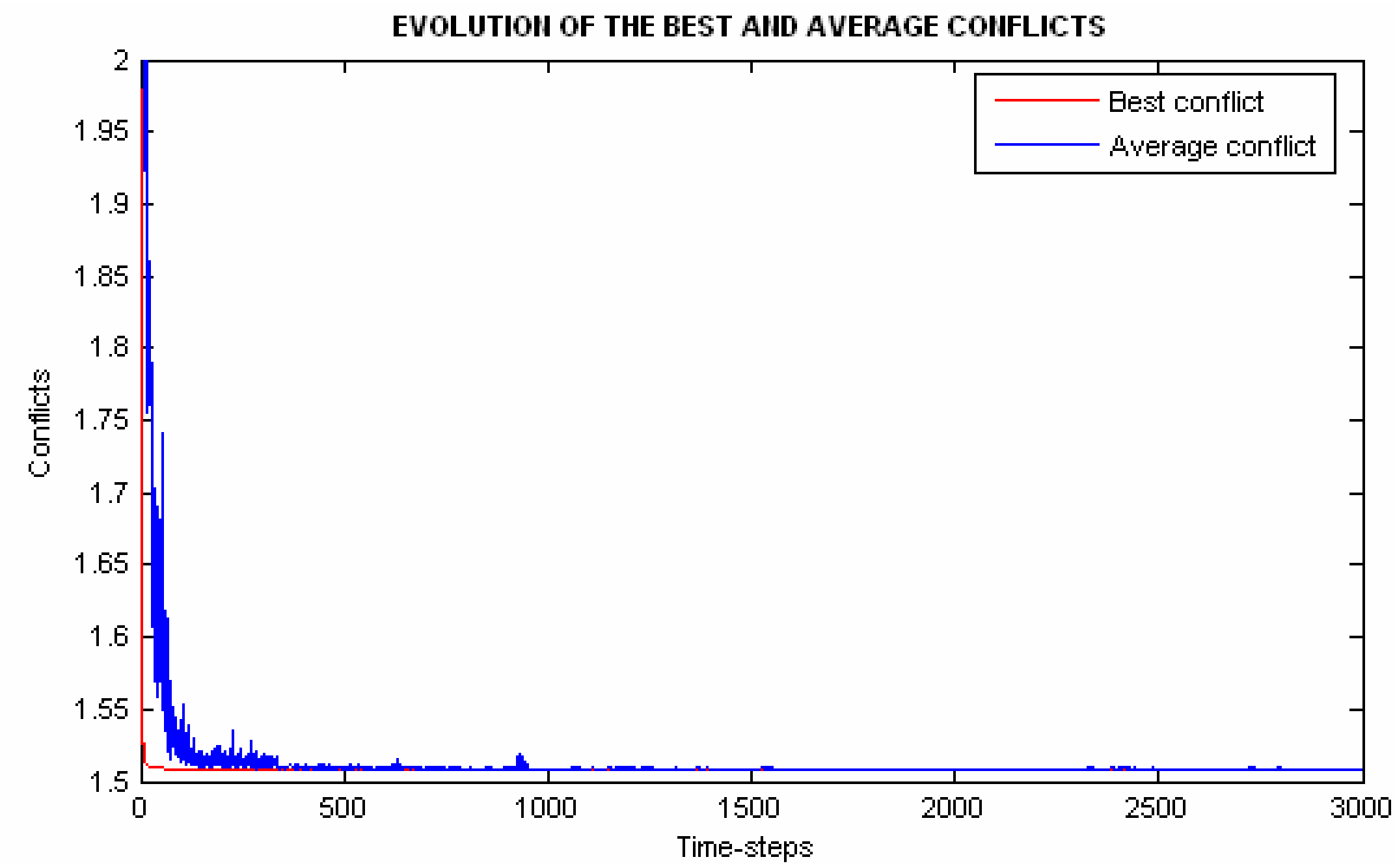

**Fig. 13. 7**: Evolution of the best and average conflicts for the GP-PSO$^{(s.d.w.)}$ equipped with the "preserving feasibility" method optimizing the two-bar truss problem.

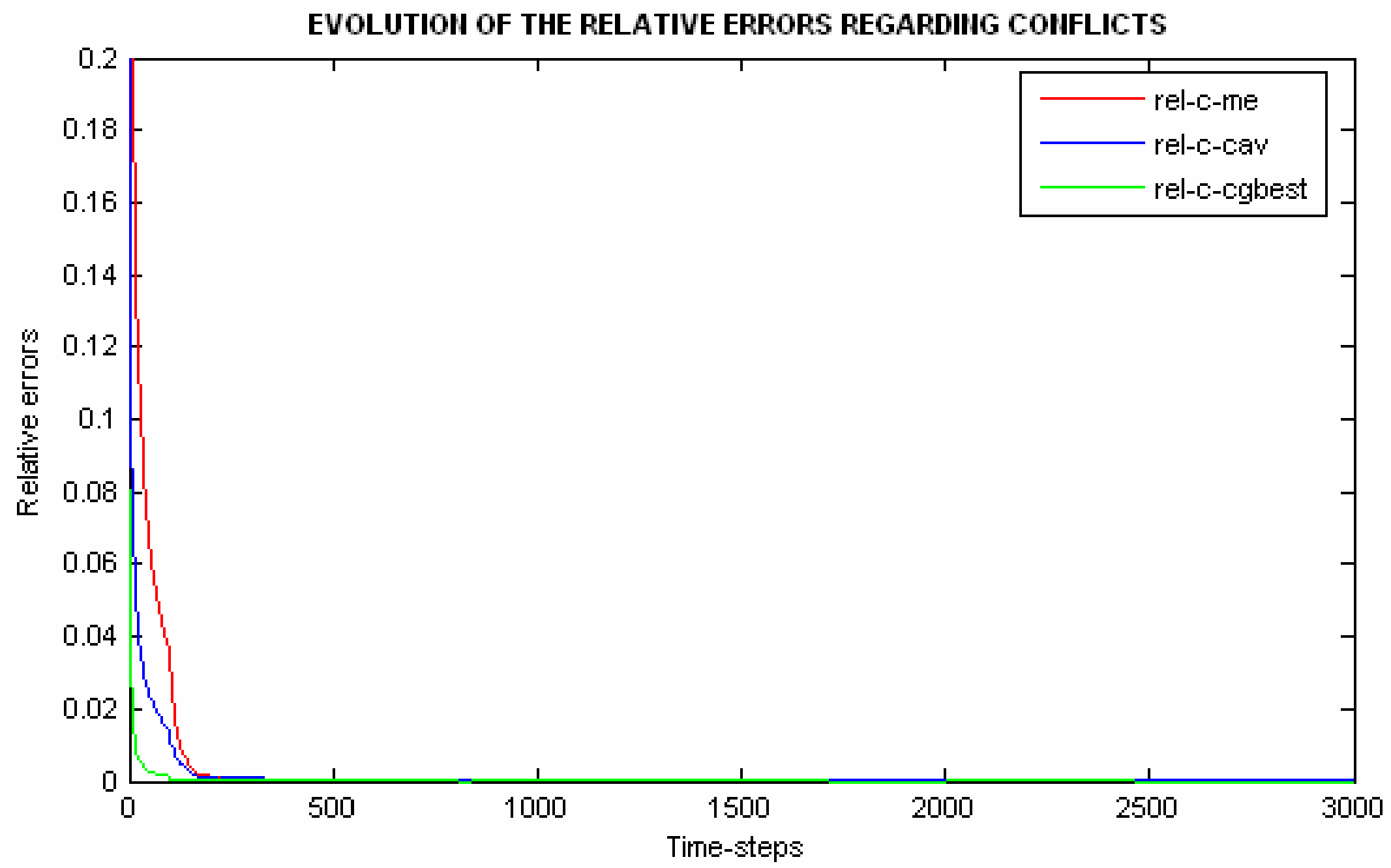

**Fig. 13. 8**: Evolution of the relative errors regarding the conflict values for the GP-PSO$^{(s.d.w.)}$ equipped with the "preserving feasibility" method optimizing the two-bar truss problem.

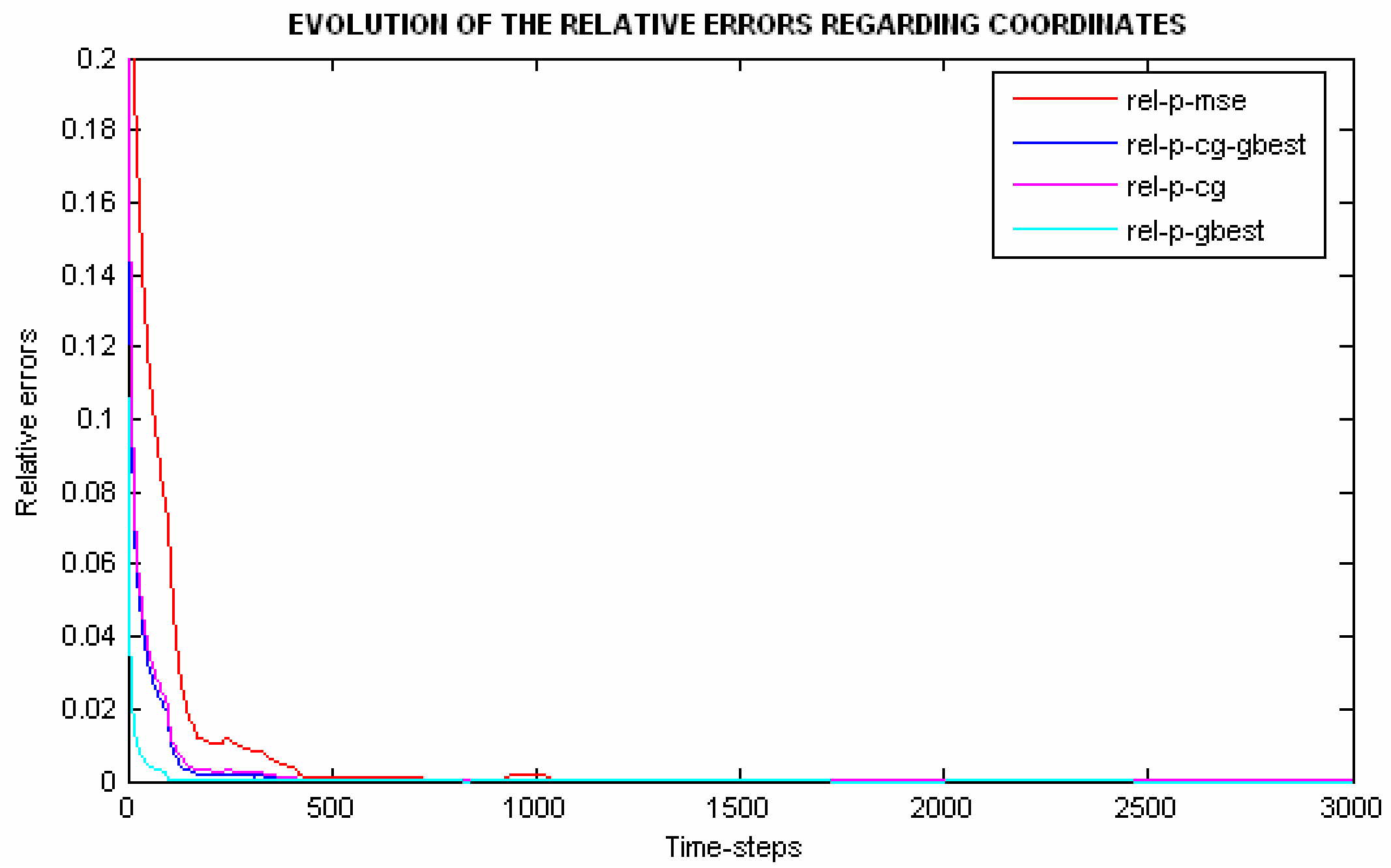

**Fig. 13. 9**: Evolution of the relative errors regarding the particles' positions for the GP-PSO$^{(s.d.w.)}$ equipped with the "preserving feasibility" method optimizing the two-bar truss problem.

## 13.3.2 Penalization

The results obtained by this optimizer are as follows:

- $x_1 = 1.41161348525941$
- $x_2 = 0.377110295696288$
- $cgbest = 1.50865241654898$

The history of the particles' positions is shown in **Fig. 13. 10**, the evolution of the best and average conflicts is shown in **Fig. 13. 11**, the evolution of the relative errors regarding the conflict values is shown in **Fig. 13. 12**, and the evolution of the relative errors regarding the particles' positions is shown in **Fig. 13. 13**.

The results obtained here for the two-bar problem by both optimizers are in agreement with the results reported by Foryś [36].

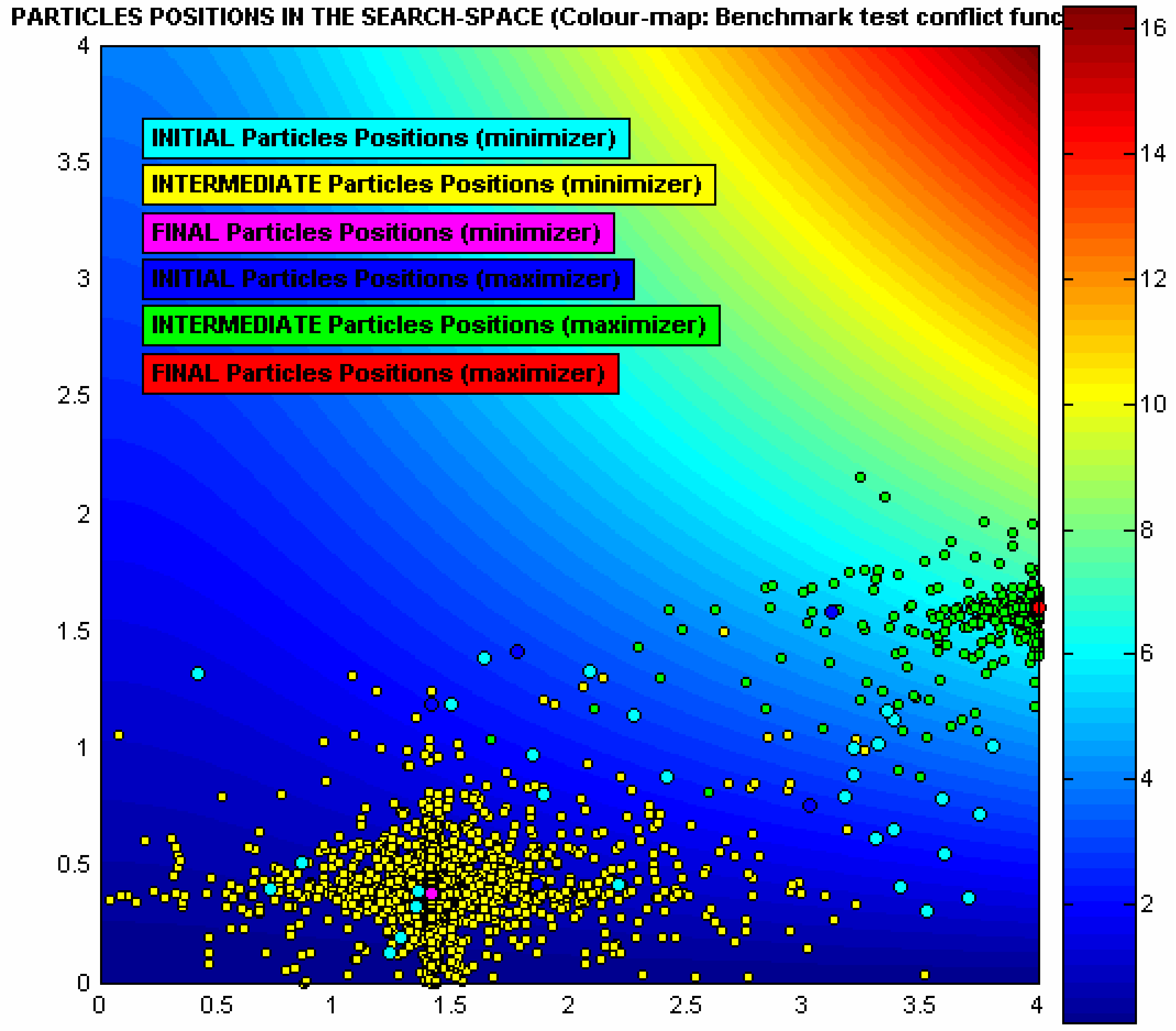

**Fig. 13. 10**: History of the particles' postions for the GP-PSO$^{(s.d.w.)}$ equipped with the "penalization" method optimizing the two-bar truss problem.

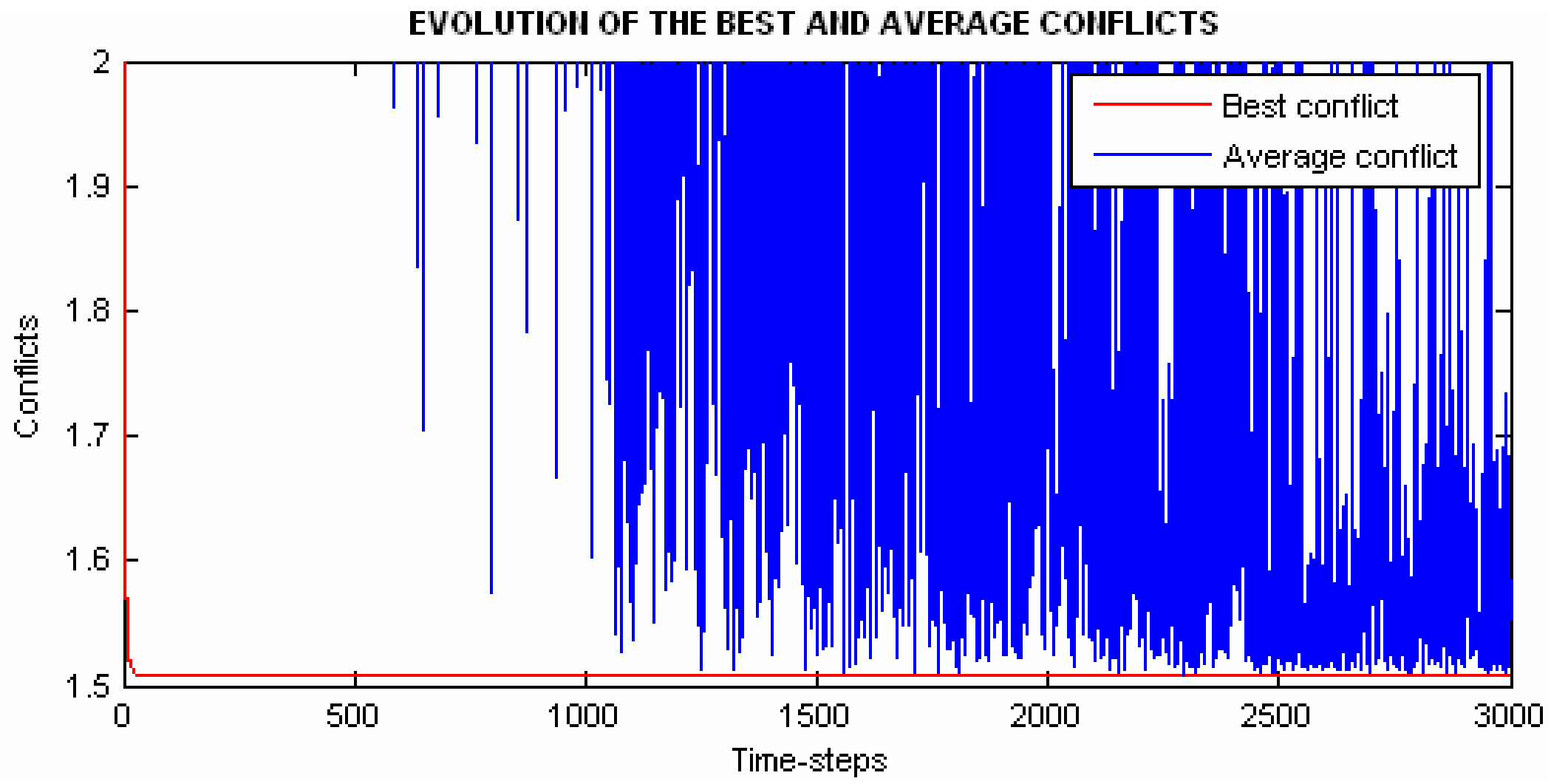

**Fig. 13. 11**: Evolution of the best and average conflicts for the GP-PSO$^{(s.d.w.)}$ equipped with the "penalization" method optimizing the two-bar truss problem.

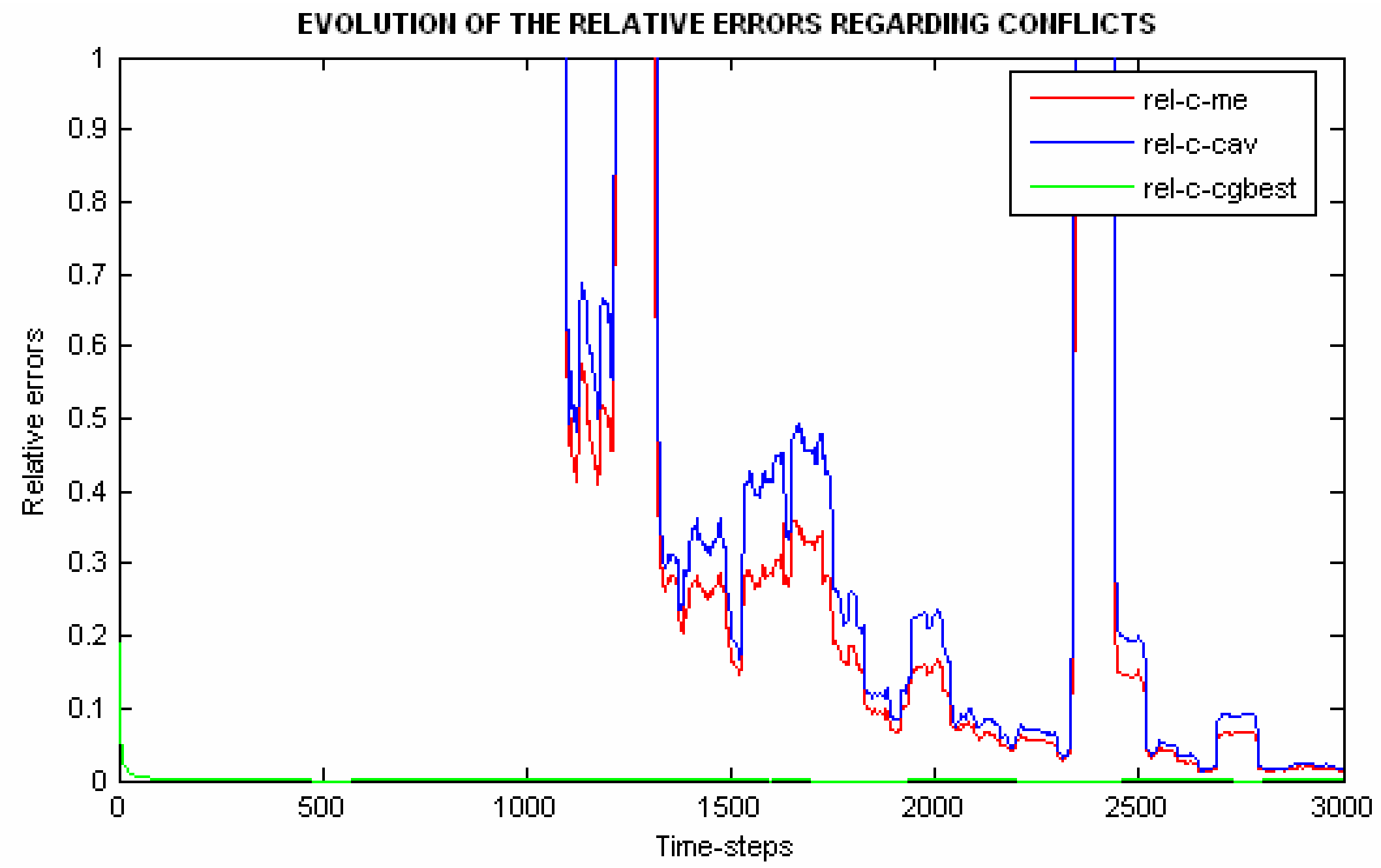

**Fig. 13. 12**: Evolution of the relative errors regarding the conflict values for the GP-PSO$^{(s.d.w.)}$ equipped with the "penalization" method optimizing the two-bar truss problem.

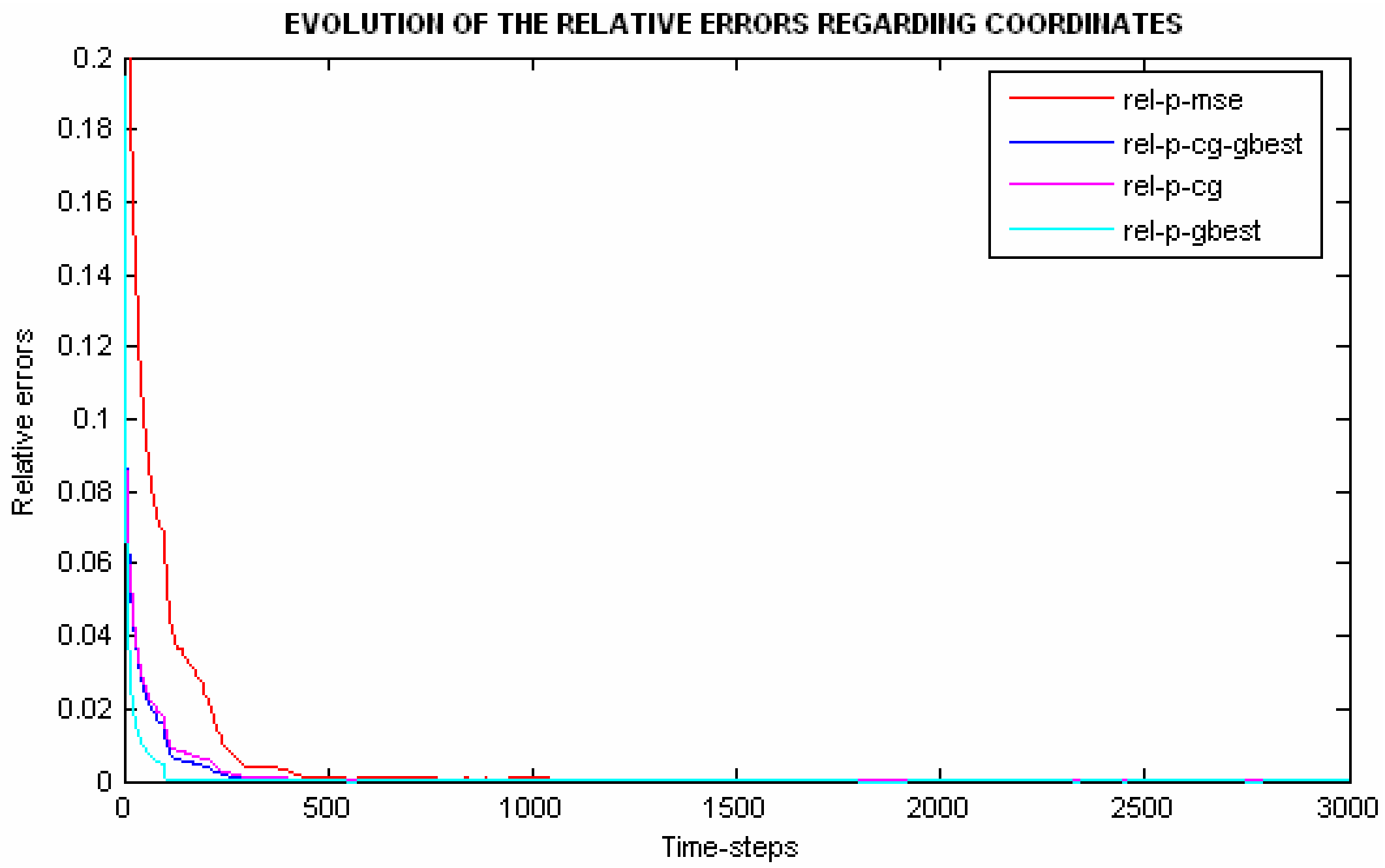

**Fig. 13. 13**: Evolution of the relative errors regarding the particles' positions for the GP-PSO$^{(s.d.w.)}$ equipped with the "penalization" method optimizing the two-bar truss problem.

## 13.4 Design of a pressure vessel

This problem is taken from Hu et al. [41] and from Foryś [36]. *The objective of the problem is to minimize the total cost of a material, forming and welding of a cylindrical vessel. There are four design variables: thickness of a shell ($x_1$), thickness of the head ($x_2$), inner radius ($x_3$) and length of the cylindrical section of the vessel ($x_4$).* [36]

The problem can be formulated as follows:

Minimize

$$f(\mathbf{x}) = 0.6224 \cdot x_1 \cdot x_3 \cdot x_4 + 1.7781 \cdot x_2 \cdot x_3^{\,2} + 3.1661 \cdot x_1^{\,2} \cdot x_4 + 19.84 \cdot x_1^{\,2} \cdot x_3 \tag{13. 6}$$

$$\text{Subject to } \begin{cases} -x_1 \le 0 \\ x_1 - 99 \le 0 \\ -x_2 \le 0 \\ x_2 - 99 \le 0 \\ -x_3 + 10 \le 0 \\ x_3 - 200 \le 0 \\ -x_4 + 10 \le 0 \\ x_4 - 200 \le 0 \\ -x_1 + 0.0193 \cdot x_3 \le 0 \\ -x_2 + 0.00954 \cdot x_3 \le 0 \\ -\pi \cdot x_3^{\,2} \cdot x_4 - \dfrac{4}{3} \cdot \pi \cdot x_3^{\,3} + 1296000 \le 0 \end{cases} \tag{13. 7}$$

### 13.4.1 Preserving feasibility

As previously mentioned, it is not the aim here to perform a quantitative analysis of the results. Therefore, only single runs of the algorithms are being performed for every problem. However, in order to compare the results obtained here to those obtained by Hu et al. [41], 11 runs for a maximum length of 10000 time-steps were carried out. The best result obtained by Hu et al. [41] among 11 runs were reported to equal 6059.131296, which were claimed to be better than previous results reported in the literature (refer to [41]). For the same maximum number of time-steps and the same number of runs, the best result obtained here equals 5988.02058052663. It is fair to note, however, that the optimizer implemented by Hu et al. [41] was composed of 20 particles whereas the one here is composed of 30.

The values of the object variables and of the best conflict are as follows:

- $cgbest = 5988.02058052663$
- $x_1 = 0.834123619260632$
- $x_2 = 0.412307737189654$
- $x_3 = 43.2188403761969$
- $x_4 = 163.230730336041$

Foryś [36] claims that the best result obtained by his optimizer equals 5885.49, but he does not specify the details of the experiments. For instance, a single run of the optimizer proposed here along 30000 time-steps returned the following solution:

- $cgbest = 5912.21829914392$
- $x_1 = 0.79358972000913$
- $x_2 = 0.392271809786896$
- $x_3 = 41.1186383424419$
- $x_4 = 189.168324788703$

## 13.4.2 Penalization

The best result obtained among 11 runs carried out for a maximum length of 10000 time-steps and using the penalization method are as follows:

- $cgbest = 5938.96603298293$
- $x_1 = 0.808402688051258$
- $x_2 = 0.399595408275591$
- $x_3 = 41.8879462780464$
- $x_4 = 179.262594459951$

In turn, the results obtained by using the penalization method to handle the constraints for a single run along 30000 time-steps are as follows:

- $cgbest = 5911.83156099849$
- $x_1 = 0.793419823369527$
- $x_2 = 0.392191380484682$
- $x_3 = 41.1116811058979$
- $x_4 = 189.260189344975$

Further improvement can be expected for time-extended searches.

# 13.5 Closure

The two proposed general-purpose optimizers, which are the outcome of this thesis, were successfully applied to three engineering problems. The first one happened to be multimodal, and the best solution could be found for different combinations of the object variables. The other two problems consist of typical benchmark problems. Hence the results obtained here could be compared to those reported in the literature. The solutions found by these optimizers were very good, outperforming most of the solutions previously reported by other authors. It must be noted, however, that the optimizers were not extensively tested along numerous runs per experiment as in previous chapters. The aim was not to test them but to illustrate possible applications.

Finally, it is fair to remark that, while the optimizer equipped with the “preserving feasibility” technique guarantees that the solution found is feasible, the optimizers equipped with the “penalization” technique usually exhibit small constraint violations when the location of the solution implies the activation of some constraints.

# SECTION IV

# FINAL CONCLUSIONS

# Chapter 14

# FINAL CONCLUSIONS

The achievements attained within this thesis are summarized, and some concluding remarks indicating the limitations of the work carried out and a number of possible directions for future research are pointed out.

## 14.1 Achievements

The main objective pursued in this thesis was the development of an optimizer with the ability to perform well on most problems. Thus, the main achievement attained here is given by two general-purpose optimizers, whose codes can be found in **Appendix 4**. The path is as follows:

X:\ Appendix 4\Matlab PSO codes\TWO SELECTED GENERAL-PURPOSE OPTIMIZERS

The capabilities of these optimizers were illustrated by optimizing a number of challenging functions along **Chapter 11**, by training a small artificial neural network for the logical "xor" problem in **Chapter 12**, and by optimizing three engineering problems along **Chapter 13**.

However, the achievements are not limited to these two black-box optimizers, which were built up by incorporating two different constraint-handling techniques into the **GP-PSO$^{(s.d.w.)}$** developed in **Chapter 9**. The **GP-PSO$^{(c.w.)}$** was also found to perform well on a first suite of six benchmark functions with hyper-cube-like boundary constraints only, whose difficulty is on the objective functions themselves rather than on the constraints. In addition, it was shown that the **GP-PSO$^{(fast)}$** also performed well, finding good—although less accurate—solutions, but taking a notably smaller number of time-steps in doing so. Hence, this optimizer could be more appropriate for cases where the evaluation of the objective function is too expensive. In addition, it was also shown that the incorporation of a local search could be convenient for problems where the degree of accuracy is of utmost importance, while the computational cost derived from the function evaluations is not an issue. It appears that incorporating the local search around the best solution found so far leads to better results than incorporating it into each particle's individual learning. However, this last assertion is not yet conclusive.

Less evident achievements are summarized along the following conclusions:

- Although the use of the inertia weight or of the constriction factor can guarantee the clustering of the particles, it is convenient to keep the velocity constraint to avoid successive function evaluations far from the regions of interest.

- The $v_{max}$ constraint is an easy, effective way of controlling the explosion. Time-varying values of this constraint might also help to fine-tune the search. However, controlling the swarm behaviour by the parameters within the velocity update equation rather than by an external constraint seems to lead to better results. This also appears to be the case for the use of the "cut off at the boundary" technique to handle the boundary constraints: the "preserving feasibility" and "penalization" techniques—which do not introduce external constraints to the "natural" dynamics of the swarm—seem to lead to more stable results, as discovered during the application of the paradigm to function optimization along **Chapter 11**.

- The relationship between the inertia weight and the learning weights greatly influences the capabilities of the optimizer. Some settings that reinforce the ability to escape sub-optimal solutions and some others that favour fine-clustering were found along **Chapter 6**. However, no setting was found to encompass both abilities. This problem can be tackled by splitting the main swarm into sub-swarms whose parameters' settings favour complementary capabilities, as discussed in **Chapter 9**.

- The acceleration weight should never be greater than four, regardless of the value of the inertia weight. Otherwise, the search becomes rather random.

- It seems that keeping the individuality and the sociality weights equal to one another is the best choice. A higher individuality sometimes delays the convergence thus keeping the ability to escape sub-optimal solutions for a longer period of time. However, it seems to be harmful for some functions, apparently due to the fact that a higher individuality may lead to less swarming-like—and hence less explorative—behaviour.

- The incorporation of a rudimentary stochastic local search around the best solution found at every time-step usually leads to better results, although the degree of improvement is not always worth the additional computational cost.

- The decision upon the best constraint-handling method is problem-dependent, being the "preserving feasibility" and the "penalization" methods probably the most suitable for a general-purpose optimizer. Some of the features of a few constraint-handling techniques are summarized hereafter:

  - Preserving feasibility method: It is very good for general-purpose optimizers in the sense that it requires no adaptation to deal with different inequality constraints. Besides, it guarantees that the solution found is feasible. However, it requires an initial feasible swarm, which is difficult to be obtained at random for problems where the size of the feasible space is small in relation to the infeasible one. Furthermore, it does not explore the infeasible space searching for new—perhaps disjointed—regions of the feasible one. The particles are allowed to fly over infeasible space, but they do not gain knowledge from doing so.

  - Penalization method: The main advantages of this method are that it does not require a feasible initial swarm, and that it explores the infeasible space searching for new, promising feasible regions. However, it does require the problem-dependent tuning of its parameters, although numerous attempts have been performed to develop general-purpose "penalization" methods with either time-varying or self-adapting parameters. The other drawback is that the solution is not guaranteed to be feasible. In fact, it frequently presents some degree of constraint-violations when the location of the solution carries the activation of one or more constraints. This last drawback turns into an advantage for problems where all the constraints cannot be complied with simultaneously.

  - Bisection method: This method presents a high convergence rate. A feasible solution is guaranteed, but a feasible initial swarm is required. The particles never fly over infeasible space. This method appears to be convenient for the cases where the computational cost of the function evaluations is high, the size of the feasible space is not too small in relation to the size of the infeasible one, and the feasible space is not composed of disjointed sub-spaces.

  - Cut-off technique: This technique was only considered to handle hyper-rectangle-like boundary constraints, and appears to be convenient for problems where the solution is located on the boundary. In addition, it accelerates the convergence. However, it seems that the alterations to the "natural" dynamics of the swarm that result from this constraint, which is external to the velocity update equation, sometimes lead to less stable behaviour: it usually finds good solutions, but a few extremely bad solutions might be found every now and again, which seriously damage the mean solutions. This can be clearly observed in **Appendix 4** for the case of the Rosenbrock function subject to hyper-cube-like boundary constraints only, optimized by the GP-PSO$^{(s.d.w.)}$ equipped with the "preserving feasibility", the "cut off at the boundary", and the "resetting of the velocity" techniques. The path is as follows:

X:\Appendix 4\Experiments carried out along chapter 11\First test suite of benchmark functions\Rosenbrock - symmetric.xls

Another drawback is that the adaptation of this technique to constraints other than hyper-rectangle-like boundaries is not straightforward.

- Resetting velocity technique: This is not a constraint-handling method but a possible technique to be incorporated into other methods, and it consists of resetting the velocity of an infeasible particle to zero. It saves computational cost because the particles are pulled back to the feasible space faster, thus speeding up the convergence.

In addition, some measures of the degree of clustering of the particles and some measures of the convergence of the algorithm were incorporated into the canonical optimizer along **Chapter 6** with the aim to estimate the reliability of the solution found, and to terminate the search when the solution is good enough, when further significant improvement is unlikely, or when a maximum number of time-steps permitted is reached.

## 14.2 Concluding remarks and future work

The work carried out along this thesis was limited to continuous, static, and single-objective optimization problems. Therefore, the corresponding adaptations of the optimizers proposed here to deal with discrete, mixed-discrete, combinatorial, dynamic, and/or multi-objective problems comprise worthwhile directions for future research. For instance, the discrete problems could be handled by the binary PSO introduced in **Chapter 5**, or simply by rounding off the trajectories of the particles; the combinatorial problems could be handled by defining a discrete space, where locations associated to repeated values of the variables are penalized; the dynamic problems could be handled by forcing the particles to eventually forget some previous experiences, which might be out of date; multi-objective problems might be handled by simply optimizing a weighted summation of the measures of attainment of the different objectives. However, more sophisticated methods should be investigated, especially for problems where no degree of constraint violations is permitted.

Following Carlisle et al.'s findings [13], the swarm size was kept equal to 30 along most of the experiments run within this thesis. However, the influence of the swarm size should be investigated, since the number of function evaluations directly depends on it. An equation that relates the number of particles in the swarm to the number of dimensions of the problem might be developed. Besides, for the same number of function evaluations, the best trade-off between the swarm size and the maximum number of time-steps permitted for the search should be investigated: a larger swarm would lead to a more parallel search, while a longer search would give each particle more time to find better solutions.

The incorporation of the $v_{max}$ constraint into the O-PSO was shown to control the explosion. Time-varying values were shown to be able to improve the fine-clustering, although it was conjectured that enhancing the fine-clustering ability by means of appropriate settings of the parameters of the velocity update equation was a better strategy. Once the explosion was controlled and the fine-clustering ability enhanced by convenient settings of these parameters, the $v_{max}$ constraint was simply set equal to $0.5 \cdot (x_{max} - x_{min})$[1] with the aim to avoid numerous function evaluations far from the region of interest, without excessively altering the "natural" dynamics of the swarm. Further research on the influence of different settings of the $v_{max}$ constraint over the behaviour of the system when the inertia weight or the constriction factor is incorporated should be carried out.

Only random initializations of the particles were considered within this work. Although some experiments (not included in this thesis) were performed for uniform distributions and for Latin hyper-cube samplings without noticeable improvement in the optimizer's achievements, the experiments were only carried out for 2-dimensional problems and for single runs of the algorithm[2]. Thus, further research on the influence of the particles' initialization procedure should be considered for future research. Some deterministic initialization might be important for improving the efficiency of the algorithm, and also for problems where the constraint-handling technique requires an initial feasible swarm.

[1] When the interval constraints to the variables' values are different for the different dimensions, the constraints to the components of the particles' velocities are also different. Hence, the constraint is given by a vector rather than by a scalar.

[2] That is to say, the probabilistic nature of the paradigm was not regarded.

The global version of the algorithm was extensively tested, and a local version consisting of a ring topology with three-particle neighbourhoods was implemented and tested in **Chapter 9**. It was observed that the convergence was delayed, which sometimes resulted in improvement and sometimes in the lack of fine-clustering ability. Different topologies and neighbourhood sizes should be investigated further. For instance, greater neighbourhood sizes would lead to versions that are less global than the global version, yet less local than the local version tested here. In turn, some shortcuts in the implemented ring topology would speed up the transfer of information. Implementing shortcuts between the first and last particles of each sub-swarm should be investigated, since the different sub-swarms composing the optimizers tested here exhibit different kinds of behaviour.

Incorporating a third level of learning could also act as a trade-off between the local and the global versions. Then, each particle would learn from its own experiences, from the experiences of the particles comprising its neighbourhood (resembling learning from the observation of other individuals' experiences), and from the whole swarm's experiences (resembling learning from culture). Setting the local social learning weight (third term) higher would result in a more local version, while setting the global social learning weight (fourth term) higher would result in a more global version. The acceleration weight would then be given by the sum of the individual, the local social, and the global social learning weights.

A simple, stochastic local search was incorporated into the optimizer aiming to improve the individual learning. Two alternatives were implemented and tested: one consisted of a local search around each particle's best previous experience at each time-step, while the other consisted of a local search around the best solution found so far by any particle at each time-step. Generally speaking, the second alternative led to better results. A third alternative was mentioned: the incorporation of a local search into the termination conditions, which would be performed only when all the other termination conditions were attained, so that the algorithm is terminated if and only if this local search is not able to improve the best solution found so far. This last strategy was implemented without success due to the fact that the local search was too rudimentary. Hence an improved local search should be incorporated into the algorithm, and tested for the three proposed strategies: performing the local search around each particle's best experience at each time-step, around the swarm's best experience at each time-step, and around the best solution found when the termination conditions are attained. Besides, the implemented local search consisted of successively, randomly altering the position around which the local search is being performed, evaluating the new position, and keeping the best between the new and the previous positions. However, the alterations were generated from a rectangular distribution, where the limits were arbitrarily set. If a better solution is located outside these limits, this rudimentary local search has no chance to find it. A Gaussian distribution was unsuccessfully tried with the aim to overcome this issue.

In addition to the stochastic local search, which was intended to help the optimizer to escape sub-optimal solutions, a gradient-based local search could be implemented with the aim to improve the fine-tuning of the search. Note, however, that this would turn the algorithm suitable only for those kinds of functions that the gradient-based local search is suitable for.

Another strategy which should be considered for future work consists of incorporating this social-learning-based algorithm into an evolutionary algorithm, so that the individual learning, the social learning, and the biological-like evolution can be considered together.

The best individual and the best social experiences were updated in parallel along this thesis. That is to say, they were updated only after the positions of all the particles in the swarm were updated and their new conflicts evaluated. Alternatively, the updates of the best experiences could be performed in a sequential fashion. That is to say, they can be updated as soon as a particle's position is updated and its new conflict evaluated, so that the next particle can profit from this knowledge straightaway. In the parallel alternative, the particle has to wait for the next time-step to have access to this information.

The parameters of the particles' velocity update equation were either constant or time-varying according to some deterministic rule which consisted of a function of the current time-step. The self-adaptation of the parameters, like in evolution strategies, would be an important step forward for the development of general-purpose optimizers. Alternatively, a deterministic adaptation could be implemented, which could profit from the computation of the relative errors proposed in **Chapter 7**, so that additional function evaluations would not be required. For instance, the decrease of the inertia weight could be computed as a function of the degree of clustering of the particles.

The particle swarm optimizers have the characteristic of exhibiting a high initial convergence rate. Keeping higher diversity for a longer period of time frequently leads to better results because it helps to avoid premature convergence. The approach proposed here consisted of implementing sub-swarms within the main swarm, where some were in charge of maintaining higher diversity. However, they eventually end up clustering. Some other alternatives to keep higher diversity should be investigated. For instance, some particles might be also equipped with a gravity-like repulsive (rather than attractive) acceleration, so that the repulsive force (or acceleration) becomes greater as the particles cluster. Thus, these particles would be incapable of achieving a high degree of clustering, and diversity would be kept. A similar approach, but with attractive rather than repulsive forces, was proposed by Hendtlass et al. [38]. The attractive forces result in keeping diversity rather than in speeding up the clustering due to the discrete nature of time in computer implementations.

The constraint-handling methods were briefly studied within this thesis. For instance, the "cut off" technique was not considered for constraints other than hyper-rectangle-like boundaries, the "bisection" method was not considered to handle only the boundary constraints while leaving the other constraints to other methods, and the "penalization" method implemented was equipped with constant and arbitrarily set penalization coefficients. Furthermore, only inequality constraints were considered, although the "penalization" method could, in theory, handle equality constraints. The constraint-handling techniques should clearly be investigated further in future works.

Finally, a thorough comparative study of the computational cost of the different alternative algorithms proposed, as well as of other optimization methods, should be carried out in order to complete the evaluation of the algorithm's performance.

It is fair to note that the work undertaken here was mostly based on common sense, heuristics, geometrical analyses, and empirical findings. A more in-depth deterministic analysis of the dynamics of the swarm would be required for further understanding and improvement of the paradigm. Some works in this direction can be found in [14], [15], [16], [61], [74], [76], and [81], to name a few.

# REFERENCES

[1]. ***ACCESS EXCELLENCE @ the national health museum***, (http://www.accessexcellence.org/).

[2]. ***An Introduction to Artificial Neural Networks*** – Building an artificial brain, (http://cerium.raunvis.hi.is/~tpr/courseware/rl/slides/total.pdf), 2002.

[3]. Angeline, P., ***Evolutionary Optimization Versus Particle Swarm Optimization: Philosophy and Performance Differences***, Annual Conference on Evolutionary Programming, San Diego, 1998.

[4]. Angeline, P., ***Using Selection to Improve Particle Swarm Optimization***, 0-7803-4869-9/98, IEEE, 1998.

[5]. Bäck, T., ***Evolution Strategies: An Alternative Evolutionary Algorithm***, in Alliot, J.-M. and Lutton, E. and Ronald, E. and Schoenauer, M. and Snyers, D., *Artificial Evolution (Lecture Notes in Computer Science)*, Springer, 1996.

[6]. Bäck, T., ***Evolutionary Algorithms in Theory and Practice***, Oxford University Press, 1996.

[7]. Bäck, T., ***Evolutionary Computation***, Leiden Institute of Advanced Computer Science, Leiden University, NL and divis digital solutions GmbH, Dortmund, D., (http://www2.cs.uh.edu/~ceick/ai/EC_Intro.pdf).

[8]. Bäck, T. and Hoffmeister, F. and Schwefel, H.-P., ***A Survey of Evolution Strategies***, Proceedings of the Fourth International Conference on Genetic Algorithms, San Mateo, CA: Morgan Kaufmann, 2-9.

[9]. Bäck, T. and Schwefel, H.-P., ***Evolutionary Computation: An Overview***, IEEE, 1996.

[10]. Batista, B. M., ***Metaheuristics***, Universidad de La Laguna (Spain), Slides created by the Intelligence Computation Group, 2005.

[11]. Bonabeau, E. and Dorigo, M. and Theraulaz, G., ***Swarm Intelligence: From Natural to Artificial Systems***, Oxford University Press, 1999.

[12]. Brule', J., ***Fuzzy Systems: A Tutorial***, (http://www.austinlinks.com/Fuzzy/tutorial.html).

[13]. Carlisle, A. and Dozier, G., ***An Off-The-Shelf PSO***, Proceedings of the workshop on particle swarm optimization, Purdue school of engineering and technology, Indianapolis, 2001.

[14]. Clerc, M., ***The Swarm and the Queen: Towards a Deterministic and Adaptive Particle Swarm Optimization***, 0-7803-5536-9/99, IEEE, 1999.

[15]. Clerc, M., ***Think locally, act locally: The Way of Life of a Cheap-PSO, an Adaptive Particle Swarm Optimizer***, (http://clerc.maurice.free.fr/pso/).

[16]. Clerc, M. and Kennedy, J., ***The Particle Swarm—Explosion, Stability, and Convergence in a Multidimensional Complex Space***, IEEE Transactions on Evolutionary Computation, Vol.6, No.1, 58-73, February 2002.

[17]. Corne, D. and Dorigo, M. and Glover, F. (Eds.), ***New Ideas in Optimization***, McGraw-Hill International (UK), 1999.

[18]. Cox, E., ***Fuzzy fundamentals***, Advanced Technology/Circuits, 0018-9235, IEEE Spectrum. October, 1992.

[19]. Darwin, Ch., ***On the Origin of Species by Means of Natural Selection, or the Preservation of Favoured Races in the Struggle for Life***, 1859. This reference was not read by the author of this thesis but just taken as a necessary reference.

[20]. Dasgupta, D. and Michalewicz, Z., ***Evolutionary Algorithms – An Overview***, in Dasgupta, D. and Michalewicz, Z. (Eds.), *Evolutionary Algorithms in Engineering Applications*, Springer-Verlag Berlin Heidelberg, 1997.

[21]. Davies, L. (Ed.), ***Handbook of Genetic Algorithms***, New York: Van Nostrand Reinhold, 1991 - International Thomsom Computer Press, 1996.

[22]. Davies, L. and Steenstrup, M., ***Genetic Algorithms and Simulated Annealing: An Overview***, in Davies, L. (Ed.), *Genetic Algorithms and Simulated Annealing (Research notes in Artificial Intelligence)*, Pitman Publishing, London, 1987.

[23]. ***Dictionary.com***, (http://dictionary.reference.com/).

[24]. Dorigo, M., ***An introduction to Ant Colony Algorithms***, Iridia, Universite' Libre de Bruxelles, Belgium, (http://www.aco-metaheuristic.org/RealAnts.html).

[25]. Dorigo, M. and Di Caro, G., ***Ant Algorithms for Discrete Optimization***, Massachusetts Institute of Technology, Artificial Life 5: 137-172 (1999).

[26]. Dorigo, M. and Di Caro, G., ***The Ant Colony Optimization Meta-Heuristic***, In Corne, D. and Dorigo, M. and Glover, F. (Eds.), *New Ideas in Optimization*, McGraw-Hill International (UK), 1999.

[27]. Dorigo, M. and Gambardella, L., ***Ant Colony System: A Cooperative Learning Approach to the Traveling Salesman Problem***, IEEE Transactions on Evolutionary Computation, Vol.1, No.1, 1997. In Press.

[28]. Eberhart, R. and Kennedy, J., ***A New Optimizer Using Particle Swarm Theory***, Sixth International Symposium on Micro Machine and Human Science, 0-7803-2676-8/95, IEEE, 1995.

[29]. Eberhart, R. and Shi, Y., ***Comparing Inertia Weights and Constriction Factors in Particle Swarm Optimization***, 0-7803-6375-2/00, IEEE, 2000.

[30]. Fogel, D., ***An overview of Evolutionary Programming***, Evolutionary Algorithms, The IMA volumes in mathematics and its applications (volume111), Springer-Verlag New York Berlin Heidelberg, 1999.

[31]. Fogel, D., ***Phenotypes, Genotypes, and Operators in Evolutionary Computation***, IEEE, 1995.

[32]. Fogel, D., ***What is evolutionary computation?***, IEEE Spectrum, 2000.

[33]. Fogel, L. and Angeline, P. and Fogel, D., ***An Evolutionary Programming Approach To Self-Adaptation on Finite State Machines***, In McDonnel, J. and Reynolds, R. and

Fogel, D. (Eds.), *Proceeding of the 4th Annual Conference on Evolutionary Programming*, MIT Press, 1995.

[34]. Fogel, D. and Fogel, L., ***An introduction to Evolutionary Programming***, in Alliot, J.-M. and Lutton, E. and Ronald, E. and Schoenauer, M. and Snyers, D., *Artificial Evolution (Lecture Notes in Computer Science)*, Springer, 1996.

[35]. Fogel, D. and Fogel, L., ***Evolution and Computational Intelligence***, IEEE Spectrum, 1995.

[36]. Foryś, P. and Bochenek, B., ***A New Particle Swarm Optimizer and its Application to Structural Optimization***, In Proceedings of 5th ASMO UK / ISSMO Conference on Engineering Design Optimization, Stratford upon Avon 2004 (CD-ROM).

[37]. Gallagher, M., ***Explicit Modelling in Metaheuristic Optimization***, University of Queensland, Slides created by the School of Information Technology and Electrical Engineering.

[38]. Hendtlass, T. and Rogers, T., ***Discrete Evaluation and the Particle Swarm Algorithm***, In Proceedings of 7th Asia-Pacific Conference on Complex systems, Cairns – Australia, 2004.

[39]. Hoffmann, A., ***Paradigms of Artificial Intelligence***, Springer-Verlag Singapore, 1998.

[40]. Hu, X. and Eberhart, R., ***Solving Constrained Nonlinear Optimization Problems with Particle Swarm Optimization***, In Proceedings of the 6th World Multi-conference on Systemics, Cybernetics and Informatics, 2002. (http://www.swarmintelligence.org/papers/SCI2002Constrained.pdf).

[41]. Hu, X. and Eberhart, R. and Shi, Y., ***Engineering Optimization with Particle Swarm***, in Swarm Intelligence Symposium, 2003. SIS '03. Proceedings of the 2003 IEEE.

[42]. Jain, A. K. and Mao, J. and Mohiuddin, K. M., ***Artificial Neural Networks: A Tutorial***, Computer archive, Volume 29, Issue 3, Special issue: neural computing: companion issue to Spring 1996 IEEE Computational Science & Engineering – Page 31-44, 1996.

[43]. Jantzen, J., ***Tutorial on Fuzzy Logic***, (http://www.iau.dtu.dk/~jj/pubs/logic.pdf).

[44]. Kennedy, J., ***Methods of Agreement: Inference among EleMentals***, Proceedings of the 1998 IEEE ISIC/CIRA/ISAS Joint Conference, Gaithersburg, MD, 0-7803-4423-5/98.

[45]. Kennedy, J. and Eberhart, R., ***A Discrete Binary Version of the Particle Swarm Algorithm***, 0-7803-4053-1/97 – IEEE, 1997.

[46]. Kennedy, J. and Eberhart, R., ***Particle Swarm optimization***, In Proceedings of IEEE International Conference on Neural Networks, Piscataway – New Jersey, 1995.

[47]. Kennedy, J. and Eberhart, R., ***Swarm Intelligence***, Morgan Kaufmann Publishers, 2001.

[48]. Koza, J. R., ***Genetic Programming: A Paradigm for Genetically Breeding Populations of Computer Programs to Solve Problems***, Stanford University Computer Science Department technical report STAN-CS-90-1314. June 1990.

[49]. Koza, J. R., ***Genetic programming as a means for programming computers by natural Selection***, Statistics and Computing, 4(2): 87 - 112, June 1994.

[50]. Langton, Ch. (Ed.), ***Artificial Life. An Overview***, MIT Press, 2000. This reference was not read by the author of this thesis but just taken as a necessary reference.

[51]. ***Lecture 6: Artificial Neural Networks***, (http://ocw.mit.edu/NR/rdonlyres/Sloan-School-of-Management/15-062Data-MiningSpring2003/650A194A-828C-4990-98CE-7EB966628437/0/NeuralNet2002.pdf), 2002).

[52]. Levy, S., ***Artificial Life: The Quest for a New Creation***, Pantheon, 1992. This work was not read by the author of this thesis but just taken as a necessary reference.

[53]. ***Merriam-Webster OnLine***, (http://www.m-w.com/dictionary.htm).

[54]. Michalewicz, Z. and Fogel, D., ***How to Solve It: Modern Heuristics***, Springer-Verlag Berlin Heidelberg, 2000.

[55]. Mitchell, M., ***An Introduction to Genetic algorithms***, The MIT Press, 1999.

[56]. Mitchell, M. and Forrest, S., ***Genetic Algorithms and Artificial Life***, Santa Fe Institute Working Paper 93-11-072, Journal: *Artificial Life – Vol. 1*, 1994.

[57]. Morán, L., ***The Modern Synthesis of Genetics and Evolution***, (http://www.talkorigins.org/).

[58]. Muñoz Zavala, A. and Hernández Aguirre, A. and Villa Diharce, E., ***Constrained Optimization via Particle Evolutionary Swarm Optimization Algorithm (PESO)***, GECCO ’05, June 25-29, 2005, Washington, DC, USA. Copyright 2005 ACM 1-59593-010-8/05/0006.

[59]. Neural networks tutorial, ***Neural Networks***, (http://www.iiit.net/~vikram/nn_intro.html).

[60]. Ofria, Ch., ***Artificial Life used for Artificial Intelligence***, (http://www.krl.caltech.edu/~charles/stories/alife.html).

[61]. Ozcan, E. and Mohan, C., ***Particle Swarm Optimization: Surfing the Waves***, Proceedings of IEEE Congress on Evolutionary Computation, Piscataway – New Jersey, 1999.

[62]. Parsopoulos, K. and Vrahatis, M., ***Particle Swarm Optimization Method for Constrained Optimization Problems***, In Proceeding of the Euro-International Symposium on Computational Intelligence, 2002.

[63]. Pedregal, P., ***Introduction to Optimization – Texts in Applied Mathematics –***, Springer-Verlag New York Inc., 2004.

[64]. Peterson, C. and Söderberg, B., ***Artificial Neural Networks***, In Reeves, C. (Ed.), *Modern Heuristic Techniques for Combinatorial Problems*, McGraw-Hill International (UK) Limited, 1995.

[65]. Reeves, C. and Beasley, J., ***Introduction***, In Reeves, C. (Ed.), *Modern Heuristic Techniques for Combinatorial Problems*, McGraw-Hill International (UK) Limited, 1995.

[66]. Rennard, J. P., ***Genetic Algorithm Viewer: Demonstration of a Genetic Algorithm***, (http://www.rennard.org/alife/english/gavintrgb.html), 2000.

[67]. Schutte, J. and Groenwold, A., ***The Optimal Sizing Design of Truss Structures using the Particle Swarm Optimization Algrithm***, In Proceedings of 9th AIAA/ISSMO Symposium on Multidisciplinary Analysis and Optimization, Atlanta – Georgia, 2002.

[68]. Schwefel, H.-P., ***On the Evolution of Evolutionary Computation***, In Zurada, J. M. et al (Eds.), *Computational Intelligence: Imitating Life*. Piscataway, Nj: IEEE Press, 1994.

[69]. Settles, M. and Rylander, B., ***Neural Network Learning using Particle Swarm Optimizers***, (http://upibm9.egr.up.edu/contrib/rylander/studpubs/matt.pdf).

[70]. Shi, Y. and Eberhart, R., ***A Modified Particle Swarm Optimizer***, IEEE International Conference on Evolutionary Computation, Anchorage, Alaska, 0-7803-4869-9/98.

[71]. Shi, Y. and Eberhart, R., ***Empirical Study of Particle Swarm Optimization***, 0-7803-5536-9/99, IEEE, 1999.

[72]. Shi, Y. and Eberhart, R., ***Parameter Selection in Particle Swarm Optimization***, Evolutionary Programming VII, Springer, Lecture Notes in Computer Science 1447, 591-600, 1998.

[73]. Taylor, Ch., ***Fleshing Out Artificial Life***, in Langton, Ch. and Taylor, Ch. and Farmer, J. (Eds.), ***Artificial Life II***, Addison Wesley Longman. This reference was not read by the author of this thesis but just taken as a necessary reference.

[74]. Trelea, I., ***The particle swarm optimization algorithm: convergence analysis and parameter selection***, ELSEVIER Science, 2002 – Information Processing Letters 85, 317-325, 2003.

[75]. Tutorial from the University of Mining and Metallurgy in Cracow – Poland (from The Board of European Students of Technology), ***An Introduction to Artificial Neural Networks***, (http://www.best.agh.edu.pl/summer/summer2002/m/introduction_to_nets.pdf).

[76]. van den Bergh, F., ***An Analysis of Particle Swarm Optimizers***, Ph.D. thesis – University of Pretoria, 2001.

[77]. Venter, G. and Sobieszczanski-Sobieski, J., ***Particle Swarm Optimization***, Copyright © 2002 by Gerhard Venter – Published by the American Institute of Aeronautics and Astronautics, Inc. with permission.

[78]. Vesterstrøm, J. and Riget, J., ***Particle Swarms***, Master's Thesis, (http://www.evalife.dk/publications/JSV_JR_thesis_2002.pdf).

[79]. Walker, Matthew, ***Introduction to Genetic Programming***, (http://www.massey.ac.nz/~mgwalker/gp/intro/introduction_to_gp.pdf), 2001.

[80]. ***Wikipedia – The Free Encyclopedia*** (http://en.wikipedia.org/wiki/Main_Page).

[81]. Engelbrecht, A. P., ***Fundamentals of Computational Swarm Intelligence***, John Wiley & Sons Ltd, England, 2005.

[82]. Floudas, C. A. and Pardalos, P. M., ***A Collection of Test Problems for Constrained Global Optimization Algorithms***, Springer – U.S.A., 1990.

# BIBLIOGRAPHY

1. Carbón Pose, E., ***La Teoría del Caos. ¿Caprichosas leyes del azar?***, Longseller – Buenos Aires, Argentina, 1995.
2. Demuth, H. and Beale, M., ***Neural Network Toolbox For Use with Matlab***, User's Guide – Version 4 – The MathWorks, Inc., 2000.
3. Díaz Hernandez, A. and Novo Sanjurjo, V. and Perán Mazón, J., ***Optimización. Casos Prácticos***, UNED, Madrid, 2000.
4. Eberhart, R. C., ***Overview of Computational Intelligence***, IEEE, 1998.
5. Fogel, D., ***An Introduction to Simulated Evolutionary Optimization***, IEEE Transactions on Neural Networks, Vol. 5, No. 1, January 1994.
6. Fogel, D., ***Evolutionary Optimization***, Proc. of the 26th Asilomar Conf. on Signals, Systems and Computers, R. R. Chen (ed.), Pacific Grove, CA, pp. 409-414., IEEE, 1992.
7. Fogel, D., ***Simulated Evolution: A 30-Year Perspective***, MAPLE PRESS, 1990.
8. ***Genetic Science Learning Centre, University of Utah***, (http://gslc.genetics.utah.edu/).
9. Guttmacher, A. E. and Collins, F. S., ***Genomic Medicine – A Primer***, The New England Journal of Medicine, Vol 347, No. 19, 2002.
10. Haupt, R. and Haupt, S., ***Practical Genetic Algorithms***, John Wiley & Sons, New Jersey, 2004.
11. Hopfield, J., ***Artificial Neural Networks***, 87755-3996/88/0900-0003$01.00 © 1988 IEEE.
12. Koza, J. R., ***Genetic Programming***, In Kent, A. and Williams (Eds.), G., Encyclopedia of Computer Science and Technology, 1997.
13. Luke, S. and Hamahashi, Sh. And Hiroaki, K., ***Genetic Programming***, Proceedings of the Genetic and Evolutionary Computation Conference – Vol.2, Morgan Kauffmann, San Francisco, USA, 1999.
14. Novo Sanjurjo, V., ***Teoría de la Optimización***, UNED, Madrid, 1999.
15. ***Particle Swarm Optimization***, (http://www.swarmintelligence.org/).
16. Pomeroy, P., ***An Introduction to Particle Swarm Optimization***, (http://www.adaptiveview.com/articles/ipsoprnt.html), 2003.
17. Ramos Méndez, E., ***Programación Lineal y Métodos de Optimización***, UNED, Madrid, 1993.
18. Spears, W. and De Jong, K. and Bäck, T. and Fogel, D. and de Garis, H., ***An Overview of Evolutionary Computation***, Proceedings of the 1993 European Conference on Machine Learning.
19. Steels, L. and Brooks, R. (Eds.), ***The Artificial Life Route to Artificial Intelligence***, Lawrence Erlbaum Associates, Publishers – New Jersey, USA, 1995.

20. Streichert, F., ***Introduction to Evolutionary Algorithms***, (http://www-ra.informatik.uni-tuebingen.de/mitarb/streiche/publications/Introduction_to_Evolutionary_Algorithms.pdf).

21. Vergara, H., ***Monografías.com – cromosomas***, (http://www.monografias.com).

# SECTION V

# A P P E N D I C E S

# Appendix 1

# TRADITIONAL TRAINING OF ARTIFICIAL NEURONS

## A1.1 Introduction

This appendix is not meant to be independent or self-contained, but a complement of section **3.5.5.1**. A few illustrative examples are discussed in order to enhance the understanding of traditional training methods for artificial neurons. Keep in mind that the goal of this thesis is not to deal with traditional training methods but to show the goodness of the PSO paradigm as an alternative training technique.

## A1.2 Learning of a perceptron

The ***δ*-rule** proposes to update each weight $\left(\Delta w_i^{(k)}\right)$ in proportion to the error $\left(\delta^{(k)}\right)$, and to the value of the corresponding input $\left(x_i^{(k)}\right)$.

$$\left.\begin{array}{l} \Delta w_i^{(k)} = \eta\cdot\delta^{(k)}\cdot x_i^{(k)} = \eta\cdot\left(t^{(k)}-y^{(k)}\right)\cdot x_i^{(k)} \\ w_i^{(k+1)} = w_i^{(k)} + \Delta w_i^{(k)} \quad\Rightarrow\quad \mathbf{w}^{(k+1)} = \mathbf{w}^{(k)} + \mathbf{\Delta w}^{(k)} \end{array}\right\} \text{ where } \quad i = 0, 1, 2, \ldots, n \qquad \textbf{(A1. 1)}$$

The response of the perceptron when a set of inputs is offered is stated in equation **(A1. 2)**.

$$y^{(k)} = 1 \text{ if } lc^{(k)} = \sum_{i=0}^{n} w_i^{(k)}\cdot x_i^{(k)} \geq 0 \quad ; \quad y^{(k)} = 0 \text{ if } lc^{(k)} = \sum_{i=0}^{n} w_i^{(k)}\cdot x_i^{(k)} < 0 \qquad \textbf{(A1. 2)}$$

A simple example of a perceptron's training has been given in section **3.5.5.1.1**, where the input values could only be binary, so that $\delta^{(k)} \in \{-1,0,1\}$ and $x_i \in \{1,0\}$. Hence the influence of the input values over the correction rule could not be appreciated.

Thus, in order to illustrate the ***δ*-rule** for perceptrons with real-valued inputs, a simple example with only 10 patterns is presented hereafter, for $\eta = 1$.

| Pattern | $x_0$ | $x_1$ | $x_2$ | $w_0$ | $w_1$ | $w_2$ | $t$ | $lc$ | $y$ | $\delta$ | $\Delta w_0$ | $\Delta w_1$ | $\Delta w_2$ |
|---|---|---|---|---|---|---|---|---|---|---|---|---|---|
| 1 | -1 | 0.80 | 0.40 | 0.00 | 0.00 | 0.00 | 1.00 | 0.00 | 1.00 | 0.00 | 0.00 | 0.00 | 0.00 |
| 2 | -1 | 1.00 | 1.00 | 0.00 | 0.00 | 0.00 | 1.00 | 0.00 | 1.00 | 0.00 | 0.00 | 0.00 | 0.00 |
| 3 | -1 | 1.40 | 0.70 | 0.00 | 0.00 | 0.00 | 1.00 | 0.00 | 1.00 | 0.00 | 0.00 | 0.00 | 0.00 |
| 4 | -1 | 2.00 | 1.50 | 0.00 | 0.00 | 0.00 | 1.00 | 0.00 | 1.00 | 0.00 | 0.00 | 0.00 | 0.00 |
| 5 | -1 | 0.90 | 0.10 | 0.00 | 0.00 | 0.00 | 0.00 | 0.00 | 1.00 | -1.00 | 0.10 | -0.09 | -0.01 |
| 6 | -1 | 2.50 | 1.00 | 0.10 | -0.09 | -0.01 | 0.00 | -0.34 | 0.00 | 0.00 | 0.00 | 0.00 | 0.00 |
| 7 | -1 | 1.60 | 0.50 | 0.10 | -0.09 | -0.01 | 0.00 | -0.25 | 0.00 | 0.00 | 0.00 | 0.00 | 0.00 |
| 8 | -1 | 3.00 | 3.00 | 0.10 | -0.09 | -0.01 | 1.00 | -0.40 | 0.00 | 1.00 | -0.10 | 0.30 | 0.30 |
| 9 | -1 | -1.00 | -3.00 | 0.00 | 0.21 | 0.29 | 0.00 | -1.08 | 0.00 | 0.00 | 0.00 | 0.00 | 0.00 |
| 10 | -1 | -1.00 | -0.20 | 0.00 | 0.21 | 0.29 | 1.00 | -0.27 | 0.00 | 1.00 | -0.10 | -0.10 | -0.02 |
| 1 | -1 | 0.80 | 0.40 | -0.10 | 0.11 | 0.27 | 1.00 | 0.30 | 1.00 | 0.00 | 0.00 | 0.00 | 0.00 |
| 2 | -1 | 1.00 | 1.00 | -0.10 | 0.11 | 0.27 | 1.00 | 0.48 | 1.00 | 0.00 | 0.00 | 0.00 | 0.00 |
| 3 | -1 | 1.40 | 0.70 | -0.10 | 0.11 | 0.27 | 1.00 | 0.44 | 1.00 | 0.00 | 0.00 | 0.00 | 0.00 |
| 4 | -1 | 2.00 | 1.50 | -0.10 | 0.11 | 0.27 | 1.00 | 0.73 | 1.00 | 0.00 | 0.00 | 0.00 | 0.00 |
| 5 | -1 | 0.90 | 0.10 | -0.10 | 0.11 | 0.27 | 0.00 | 0.23 | 1.00 | -1.00 | 0.10 | -0.09 | -0.01 |
| 6 | -1 | 2.50 | 1.00 | 0.00 | 0.02 | 0.26 | 0.00 | 0.31 | 1.00 | -1.00 | 0.10 | -0.25 | -0.10 |
| 7 | -1 | 1.60 | 0.50 | 0.10 | -0.23 | 0.16 | 0.00 | -0.39 | 0.00 | 0.00 | 0.00 | 0.00 | 0.00 |
| 8 | -1 | 3.00 | 3.00 | 0.10 | -0.23 | 0.16 | 1.00 | -0.31 | 0.00 | 1.00 | -0.10 | 0.30 | 0.30 |
| 9 | -1 | -1.00 | -3.00 | 0.00 | 0.07 | 0.46 | 0.00 | -1.45 | 0.00 | 0.00 | 0.00 | 0.00 | 0.00 |
| 10 | -1 | -1.00 | -0.20 | 0.00 | 0.07 | 0.46 | 1.00 | -0.16 | 0.00 | 1.00 | -0.10 | -0.10 | -0.02 |
| 1 | -1 | 0.80 | 0.40 | -0.10 | -0.03 | 0.44 | 1.00 | 0.25 | 1.00 | 0.00 | 0.00 | 0.00 | 0.00 |
| 2 | -1 | 1.00 | 1.00 | -0.10 | -0.03 | 0.44 | 1.00 | 0.51 | 1.00 | 0.00 | 0.00 | 0.00 | 0.00 |
| 3 | -1 | 1.40 | 0.70 | -0.10 | -0.03 | 0.44 | 1.00 | 0.37 | 1.00 | 0.00 | 0.00 | 0.00 | 0.00 |
| 4 | -1 | 2.00 | 1.50 | -0.10 | -0.03 | 0.44 | 1.00 | 0.70 | 1.00 | 0.00 | 0.00 | 0.00 | 0.00 |
| 5 | -1 | 0.90 | 0.10 | -0.10 | -0.03 | 0.44 | 0.00 | 0.12 | 1.00 | -1.00 | 0.10 | -0.09 | -0.01 |
| 6 | -1 | 2.50 | 1.00 | 0.00 | -0.12 | 0.43 | 0.00 | 0.13 | 1.00 | -1.00 | 0.10 | -0.25 | -0.10 |
| 7 | -1 | 1.60 | 0.50 | 0.10 | -0.37 | 0.33 | 0.00 | -0.53 | 0.00 | 0.00 | 0.00 | 0.00 | 0.00 |
| 8 | -1 | 3.00 | 3.00 | 0.10 | -0.37 | 0.33 | 1.00 | -0.22 | 0.00 | 1.00 | -0.10 | 0.30 | 0.30 |
| 9 | -1 | -1.00 | -3.00 | 0.00 | -0.07 | 0.63 | 0.00 | -1.82 | 0.00 | 0.00 | 0.00 | 0.00 | 0.00 |
| 10 | -1 | -1.00 | -0.20 | 0.00 | -0.07 | 0.63 | 1.00 | -0.06 | 0.00 | 1.00 | -0.10 | -0.10 | -0.02 |
| 1 | -1 | 0.80 | 0.40 | -0.10 | -0.17 | 0.61 | 1.00 | 0.21 | 1.00 | 0.00 | 0.00 | 0.00 | 0.00 |
| 2 | -1 | 1.00 | 1.00 | -0.10 | -0.17 | 0.61 | 1.00 | 0.54 | 1.00 | 0.00 | 0.00 | 0.00 | 0.00 |
| 3 | -1 | 1.40 | 0.70 | -0.10 | -0.17 | 0.61 | 1.00 | 0.29 | 1.00 | 0.00 | 0.00 | 0.00 | 0.00 |
| 4 | -1 | 2.00 | 1.50 | -0.10 | -0.17 | 0.61 | 1.00 | 0.68 | 1.00 | 0.00 | 0.00 | 0.00 | 0.00 |
| 5 | -1 | 0.90 | 0.10 | -0.10 | -0.17 | 0.61 | 0.00 | 0.01 | 1.00 | -1.00 | 0.10 | -0.09 | -0.01 |
| 6 | -1 | 2.50 | 1.00 | 0.00 | -0.26 | 0.60 | 0.00 | -0.05 | 0.00 | 0.00 | 0.00 | 0.00 | 0.00 |
| 7 | -1 | 1.60 | 0.50 | 0.00 | -0.26 | 0.60 | 0.00 | -0.12 | 0.00 | 0.00 | 0.00 | 0.00 | 0.00 |
| 8 | -1 | 3.00 | 3.00 | 0.00 | -0.26 | 0.60 | 1.00 | 1.02 | 1.00 | 0.00 | 0.00 | 0.00 | 0.00 |
| 9 | -1 | -1.00 | -3.00 | 0.00 | -0.26 | 0.60 | 0.00 | -1.54 | 0.00 | 0.00 | 0.00 | 0.00 | 0.00 |
| 10 | -1 | -1.00 | -0.20 | 0.00 | -0.26 | 0.60 | 1.00 | 0.14 | 1.00 | 0.00 | 0.00 | 0.00 | 0.00 |
| 1 | -1 | 0.80 | 0.40 | 0.00 | -0.26 | 0.60 | 1.00 | 0.03 | 1.00 | 0.00 | 0.00 | 0.00 | 0.00 |
| 2 | -1 | 1.00 | 1.00 | 0.00 | -0.26 | 0.60 | 1.00 | 0.34 | 1.00 | 0.00 | 0.00 | 0.00 | 0.00 |
| 3 | -1 | 1.40 | 0.70 | 0.00 | -0.26 | 0.60 | 1.00 | 0.06 | 1.00 | 0.00 | 0.00 | 0.00 | 0.00 |
| 4 | -1 | 2.00 | 1.50 | 0.00 | -0.26 | 0.60 | 1.00 | 0.38 | 1.00 | 0.00 | 0.00 | 0.00 | 0.00 |
| 5 | -1 | 0.90 | 0.10 | 0.00 | -0.26 | 0.60 | 0.00 | -0.17 | 0.00 | 0.00 | 0.00 | 0.00 | 0.00 |
| 6 | -1 | 2.50 | 1.00 | 0.00 | -0.26 | 0.60 | 0.00 | -0.05 | 0.00 | 0.00 | 0.00 | 0.00 | 0.00 |
| 7 | -1 | 1.60 | 0.50 | 0.00 | -0.26 | 0.60 | 0.00 | -0.12 | 0.00 | 0.00 | 0.00 | 0.00 | 0.00 |
| 8 | -1 | 3.00 | 3.00 | 0.00 | -0.26 | 0.60 | 1.00 | 1.02 | 1.00 | 0.00 | 0.00 | 0.00 | 0.00 |
| 9 | -1 | -1.00 | -3.00 | 0.00 | -0.26 | 0.60 | 0.00 | -1.54 | 0.00 | 0.00 | 0.00 | 0.00 | 0.00 |
| 10 | -1 | -1.00 | -0.20 | 0.00 | -0.26 | 0.60 | 1.00 | 0.14 | 1.00 | 0.00 | 0.00 | 0.00 | 0.00 |

**Table A1. 1**: Process of $\delta$-rule correction for a perceptron with real-valued inputs and $\eta$ = 1. The training data set consists of ten patterns only. This simple case required only 5 updates for each pattern to eliminate all the errors.

Notice that after going through the 10 patterns 5 times, all the errors $\delta^{(k)}$ are equal to zero. That is to say that the perceptron is already trained for classifying the patterns offered. Typically, there must be another set of patterns, whose classifications are known in advance but were not used for training, to verify the results.

# A1.3 Learning of a linear artificial neuron

The perceptron's output is binary, so that the error $\delta^{(k)} \in \{-1,0,1\}$. In contrast, the linear artificial neuron receives real-valued inputs and returns a real-valued output.

$$y^{(k)} = f\left(lc^{(k)}\right) = f\left(\sum_{i=0}^{n} w_i^{(k)} \cdot x_i^{(k)}\right) = \sum_{i=0}^{n} w_i^{(k)} \cdot x_i^{(k)} \qquad \textbf{(A1. 3)}$$

Therefore, in order to follow the evolution of the weights all along the correction rule, an example of the training of a linear artificial neuron is presented below. The transfer function is $y^{(k)} = lc^{(k)} = \sum_{i=0}^{n} w_i^{(k)} \cdot x_i^{(k)}$, so that the approximation function is a hyper-plane. The training patterns were provided so that each one belongs to the following plane: $y = -0.5 \cdot x_1 + x_2 - 1$.

The results are gathered in **Table A1. 2**.

| Pattern | $x_0$ | $x_1$ | $x_2$ | $w_0$ | $w_1$ | $w_2$ | $t$ | $y$ | $\delta$ | $\Delta w_0$ | $\Delta w_1$ | $\Delta w_2$ |
|---|---|---|---|---|---|---|---|---|---|---|---|---|
| 1 | -1 | 0.80 | 0.40 | 0.00 | 0.00 | 0.00 | -1.00 | 0.00 | -1.00 | 0.10 | -0.08 | -0.04 |
| 2 | -1 | 1.00 | 1.00 | 0.10 | -0.08 | -0.04 | -0.50 | -0.22 | -0.28 | 0.03 | -0.03 | -0.03 |
| 3 | -1 | 1.40 | 0.70 | 0.13 | -0.11 | -0.07 | -1.00 | -0.33 | -0.67 | 0.07 | -0.09 | -0.05 |
| 4 | -1 | 2.00 | 1.50 | 0.20 | -0.20 | -0.12 | -0.50 | -0.77 | 0.27 | -0.03 | 0.05 | 0.04 |
| 5 | -1 | 0.90 | 0.10 | 0.17 | -0.15 | -0.07 | -1.35 | -0.31 | -1.04 | 0.10 | -0.09 | -0.01 |
| 6 | -1 | 2.50 | 1.00 | 0.27 | -0.24 | -0.08 | -1.25 | -0.96 | -0.29 | 0.03 | -0.07 | -0.03 |
| 7 | -1 | 1.60 | 0.50 | 0.30 | -0.31 | -0.11 | -1.30 | -0.86 | -0.44 | 0.04 | -0.07 | -0.02 |
| 8 | -1 | 3.00 | 3.00 | 0.35 | -0.38 | -0.14 | 0.50 | -1.90 | 2.40 | -0.24 | 0.72 | 0.72 |
| 9 | -1 | -1.00 | -3.00 | 0.10 | 0.34 | 0.59 | -3.50 | -2.20 | -1.30 | 0.13 | 0.13 | 0.39 |
| 10 | -1 | -1.00 | -0.20 | 0.23 | 0.47 | 0.98 | -0.70 | -0.90 | 0.20 | -0.02 | -0.02 | 0.00 |
| 1 | -1 | 0.80 | 0.40 | 0.22 | 0.45 | 0.97 | -1.00 | 0.53 | -1.53 | 0.15 | -0.12 | -0.06 |
| 2 | -1 | 1.00 | 1.00 | 0.37 | 0.32 | 0.91 | -0.50 | 0.87 | -1.37 | 0.14 | -0.14 | -0.14 |
| 3 | -1 | 1.40 | 0.70 | 0.50 | 0.19 | 0.77 | -1.00 | 0.30 | -1.30 | 0.13 | -0.18 | -0.09 |
| 4 | -1 | 2.00 | 1.50 | 0.64 | 0.01 | 0.68 | -0.50 | 0.40 | -0.90 | 0.09 | -0.18 | -0.14 |
| 5 | -1 | 0.90 | 0.10 | 0.73 | -0.17 | 0.55 | -1.35 | -0.83 | -0.52 | 0.05 | -0.05 | -0.01 |
| 6 | -1 | 2.50 | 1.00 | 0.78 | -0.22 | 0.54 | -1.25 | -0.79 | -0.46 | 0.05 | -0.12 | -0.05 |
| 7 | -1 | 1.60 | 0.50 | 0.82 | -0.34 | 0.50 | -1.30 | -1.11 | -0.19 | 0.02 | -0.03 | -0.01 |
| 8 | -1 | 3.00 | 3.00 | 0.84 | -0.37 | 0.49 | 0.50 | -0.48 | 0.98 | -0.10 | 0.29 | 0.29 |
| 9 | -1 | -1.00 | -3.00 | 0.74 | -0.07 | 0.78 | -3.50 | -3.02 | -0.48 | 0.05 | 0.05 | 0.15 |
| 10 | -1 | -1.00 | -0.20 | 0.79 | -0.02 | 0.93 | -0.70 | -0.95 | 0.25 | -0.03 | -0.03 | -0.01 |
| 1 | -1 | 0.80 | 0.40 | 0.77 | -0.05 | 0.92 | -1.00 | -0.44 | -0.56 | 0.06 | -0.04 | -0.02 |
| 2 | -1 | 1.00 | 1.00 | 0.82 | -0.09 | 0.90 | -0.50 | -0.02 | -0.48 | 0.05 | -0.05 | -0.05 |
| 3 | -1 | 1.40 | 0.70 | 0.87 | -0.14 | 0.85 | -1.00 | -0.48 | -0.52 | 0.05 | -0.07 | -0.04 |
| 4 | -1 | 2.00 | 1.50 | 0.92 | -0.22 | 0.81 | -0.50 | -0.13 | -0.37 | 0.04 | -0.07 | -0.05 |
| 5 | -1 | 0.90 | 0.10 | 0.96 | -0.29 | 0.76 | -1.35 | -1.14 | -0.21 | 0.02 | -0.02 | 0.00 |
| 6 | -1 | 2.50 | 1.00 | 0.98 | -0.31 | 0.76 | -1.25 | -0.99 | -0.26 | 0.03 | -0.06 | -0.03 |
| 7 | -1 | 1.60 | 0.50 | 1.01 | -0.37 | 0.73 | -1.30 | -1.24 | -0.06 | 0.01 | -0.01 | 0.00 |
| 8 | -1 | 3.00 | 3.00 | 1.01 | -0.38 | 0.73 | 0.50 | 0.03 | 0.47 | -0.05 | 0.14 | 0.14 |
| 9 | -1 | -1.00 | -3.00 | 0.97 | -0.24 | 0.87 | -3.50 | -3.34 | -0.16 | 0.02 | 0.02 | 0.05 |
| 10 | -1 | -1.00 | -0.20 | 0.98 | -0.22 | 0.92 | -0.70 | -0.94 | 0.24 | -0.02 | -0.02 | 0.00 |
| 1 | -1 | 0.80 | 0.40 | 0.96 | -0.25 | 0.91 | -1.00 | -0.79 | -0.21 | 0.02 | -0.02 | -0.01 |
| 2 | -1 | 1.00 | 1.00 | 0.98 | -0.26 | 0.91 | -0.50 | -0.34 | -0.16 | 0.02 | -0.02 | -0.02 |
| 3 | -1 | 1.40 | 0.70 | 1.00 | -0.28 | 0.89 | -1.00 | -0.77 | -0.23 | 0.02 | -0.03 | -0.02 |
| 4 | -1 | 2.00 | 1.50 | 1.02 | -0.31 | 0.87 | -0.50 | -0.34 | -0.16 | 0.02 | -0.03 | -0.02 |
| 5 | -1 | 0.90 | 0.10 | 1.03 | -0.35 | 0.85 | -1.35 | -1.26 | -0.09 | 0.01 | -0.01 | 0.00 |
| 6 | -1 | 2.50 | 1.00 | 1.04 | -0.35 | 0.85 | -1.25 | -1.08 | -0.17 | 0.02 | -0.04 | -0.02 |
| 7 | -1 | 1.60 | 0.50 | 1.06 | -0.40 | 0.83 | -1.30 | -1.28 | -0.02 | 0.00 | 0.00 | 0.00 |
| 8 | -1 | 3.00 | 3.00 | 1.06 | -0.40 | 0.83 | 0.50 | 0.23 | 0.27 | -0.03 | 0.08 | 0.08 |
| 9 | -1 | -1.00 | -3.00 | 1.04 | -0.32 | 0.91 | -3.50 | -3.45 | -0.05 | 0.00 | 0.00 | 0.01 |
| 10 | -1 | -1.00 | -0.20 | 1.04 | -0.31 | 0.93 | -0.70 | -0.91 | 0.21 | -0.02 | -0.02 | 0.00 |
| 1 | -1 | 0.80 | 0.40 | 1.02 | -0.33 | 0.92 | -1.00 | -0.92 | -0.08 | 0.01 | -0.01 | 0.00 |
| 2 | -1 | 1.00 | 1.00 | 1.03 | -0.34 | 0.92 | -0.50 | -0.45 | -0.05 | 0.01 | -0.01 | -0.01 |
| 3 | -1 | 1.40 | 0.70 | 1.03 | -0.35 | 0.91 | -1.00 | -0.88 | -0.12 | 0.01 | -0.02 | -0.01 |
| 4 | -1 | 2.00 | 1.50 | 1.04 | -0.36 | 0.90 | -0.50 | -0.41 | -0.09 | 0.01 | -0.02 | -0.01 |
| 5 | -1 | 0.90 | 0.10 | 1.05 | -0.38 | 0.89 | -1.35 | -1.31 | -0.04 | 0.00 | 0.00 | 0.00 |
| 6 | -1 | 2.50 | 1.00 | 1.06 | -0.38 | 0.89 | -1.25 | -1.13 | -0.12 | 0.01 | -0.03 | -0.01 |
| 7 | -1 | 1.60 | 0.50 | 1.07 | -0.42 | 0.88 | -1.30 | -1.29 | -0.01 | 0.00 | 0.00 | 0.00 |
| 8 | -1 | 3.00 | 3.00 | 1.07 | -0.42 | 0.88 | 0.50 | 0.32 | 0.18 | -0.02 | 0.05 | 0.05 |
| 9 | -1 | -1.00 | -3.00 | 1.05 | -0.36 | 0.93 | -3.50 | -3.49 | -0.01 | 0.00 | 0.00 | 0.00 |
| 10 | -1 | -1.00 | -0.20 | 1.05 | -0.36 | 0.94 | -0.70 | -0.88 | 0.18 | -0.02 | -0.02 | 0.00 |
| 1 | -1 | 0.80 | 0.40 | 1.04 | -0.38 | 0.93 | -1.00 | -0.96 | -0.04 | 0.00 | 0.00 | 0.00 |
| 2 | -1 | 1.00 | 1.00 | 1.04 | -0.38 | 0.93 | -0.50 | -0.49 | -0.01 | 0.00 | 0.00 | 0.00 |
| 3 | -1 | 1.40 | 0.70 | 1.04 | -0.38 | 0.93 | -1.00 | -0.92 | -0.08 | 0.01 | -0.01 | -0.01 |
| 4 | -1 | 2.00 | 1.50 | 1.05 | -0.39 | 0.92 | -0.50 | -0.45 | -0.05 | 0.01 | -0.01 | -0.01 |
| 5 | -1 | 0.90 | 0.10 | 1.05 | -0.40 | 0.92 | -1.35 | -1.32 | -0.03 | 0.00 | 0.00 | 0.00 |
| 6 | -1 | 2.50 | 1.00 | 1.06 | -0.41 | 0.92 | -1.25 | -1.15 | -0.10 | 0.01 | -0.02 | -0.01 |
| 7 | -1 | 1.60 | 0.50 | 1.06 | -0.43 | 0.91 | -1.30 | -1.30 | 0.00 | 0.00 | 0.00 | 0.00 |
| 8 | -1 | 3.00 | 3.00 | 1.06 | -0.43 | 0.91 | 0.50 | 0.37 | 0.13 | -0.01 | 0.04 | 0.04 |
| 9 | -1 | -1.00 | -3.00 | 1.05 | -0.39 | 0.95 | -3.50 | -3.50 | 0.00 | 0.00 | 0.00 | 0.00 |
| 10 | -1 | -1.00 | -0.20 | 1.05 | -0.39 | 0.95 | -0.70 | -0.85 | 0.15 | -0.02 | -0.02 | 0.00 |
| 1 | -1 | 0.80 | 0.40 | 1.04 | -0.41 | 0.94 | -1.00 | -0.98 | -0.02 | 0.00 | 0.00 | 0.00 |
| 2 | -1 | 1.00 | 1.00 | 1.04 | -0.41 | 0.94 | -0.50 | -0.50 | 0.00 | 0.00 | 0.00 | 0.00 |
| 3 | -1 | 1.40 | 0.70 | 1.04 | -0.41 | 0.94 | -1.00 | -0.95 | -0.05 | 0.01 | -0.01 | 0.00 |
| 4 | -1 | 2.00 | 1.50 | 1.04 | -0.41 | 0.94 | -0.50 | -0.46 | -0.04 | 0.00 | -0.01 | -0.01 |
| 5 | -1 | 0.90 | 0.10 | 1.05 | -0.42 | 0.93 | -1.35 | -1.33 | -0.02 | 0.00 | 0.00 | 0.00 |
| 6 | -1 | 2.50 | 1.00 | 1.05 | -0.42 | 0.93 | -1.25 | -1.17 | -0.08 | 0.01 | -0.02 | -0.01 |
| 7 | -1 | 1.60 | 0.50 | 1.06 | -0.44 | 0.93 | -1.30 | -1.30 | 0.00 | 0.00 | 0.00 | 0.00 |
| 8 | -1 | 3.00 | 3.00 | 1.06 | -0.44 | 0.93 | 0.50 | 0.39 | 0.11 | -0.01 | 0.03 | 0.03 |
| 9 | -1 | -1.00 | -3.00 | 1.05 | -0.41 | 0.96 | -3.50 | -3.51 | 0.01 | 0.00 | 0.00 | 0.00 |
| 10 | -1 | -1.00 | -0.20 | 1.04 | -0.41 | 0.96 | -0.70 | -0.82 | 0.12 | -0.01 | -0.01 | 0.00 |
| 1 | -1 | 0.80 | 0.40 | 1.03 | -0.42 | 0.95 | -1.00 | -0.99 | -0.01 | 0.00 | 0.00 | 0.00 |
| 2 | -1 | 1.00 | 1.00 | 1.03 | -0.42 | 0.95 | -0.50 | -0.51 | 0.01 | 0.00 | 0.00 | 0.00 |
| 3 | -1 | 1.40 | 0.70 | 1.03 | -0.42 | 0.95 | -1.00 | -0.96 | -0.04 | 0.00 | -0.01 | 0.00 |
| 4 | -1 | 2.00 | 1.50 | 1.04 | -0.43 | 0.95 | -0.50 | -0.47 | -0.03 | 0.00 | -0.01 | 0.00 |
| 5 | -1 | 0.90 | 0.10 | 1.04 | -0.44 | 0.95 | -1.35 | -1.34 | -0.01 | 0.00 | 0.00 | 0.00 |
| 6 | -1 | 2.50 | 1.00 | 1.04 | -0.44 | 0.95 | -1.25 | -1.19 | -0.06 | 0.01 | -0.02 | -0.01 |
| 7 | -1 | 1.60 | 0.50 | 1.05 | -0.45 | 0.94 | -1.30 | -1.30 | 0.00 | 0.00 | 0.00 | 0.00 |
| 8 | -1 | 3.00 | 3.00 | 1.05 | -0.45 | 0.94 | 0.50 | 0.41 | 0.09 | -0.01 | 0.03 | 0.03 |
| 9 | -1 | -1.00 | -3.00 | 1.04 | -0.43 | 0.96 | -3.50 | -3.51 | 0.01 | 0.00 | 0.00 | 0.00 |
| 10 | -1 | -1.00 | -0.20 | 1.04 | -0.43 | 0.96 | -0.70 | -0.80 | 0.10 | -0.01 | -0.01 | 0.00 |
| 1 | -1 | 0.80 | 0.40 | 1.03 | -0.44 | 0.96 | -1.00 | -0.99 | -0.01 | 0.00 | 0.00 | 0.00 |
| 2 | -1 | 1.00 | 1.00 | 1.03 | -0.44 | 0.96 | -0.50 | -0.51 | 0.01 | 0.00 | 0.00 | 0.00 |
| 3 | -1 | 1.40 | 0.70 | 1.03 | -0.44 | 0.96 | -1.00 | -0.97 | -0.03 | 0.00 | 0.00 | 0.00 |
| 4 | -1 | 2.00 | 1.50 | 1.03 | -0.44 | 0.96 | -0.50 | -0.48 | -0.02 | 0.00 | 0.00 | 0.00 |
| 5 | -1 | 0.90 | 0.10 | 1.03 | -0.45 | 0.96 | -1.35 | -1.34 | -0.01 | 0.00 | 0.00 | 0.00 |
| 6 | -1 | 2.50 | 1.00 | 1.03 | -0.45 | 0.96 | -1.25 | -1.20 | -0.05 | 0.01 | -0.01 | -0.01 |
| 7 | -1 | 1.60 | 0.50 | 1.04 | -0.46 | 0.95 | -1.30 | -1.30 | 0.00 | 0.00 | 0.00 | 0.00 |
| 8 | -1 | 3.00 | 3.00 | 1.04 | -0.46 | 0.95 | 0.50 | 0.43 | 0.07 | -0.01 | 0.02 | 0.02 |
| 9 | -1 | -1.00 | -3.00 | 1.03 | -0.44 | 0.97 | -3.50 | -3.51 | 0.01 | 0.00 | 0.00 | 0.00 |
| 10 | -1 | -1.00 | -0.20 | 1.03 | -0.44 | 0.97 | -0.70 | -0.79 | 0.09 | -0.01 | -0.01 | 0.00 |
| 1 | -1 | 0.80 | 0.40 | 1.02 | -0.45 | 0.97 | -1.00 | -0.99 | -0.01 | 0.00 | 0.00 | 0.00 |
| 2 | -1 | 1.00 | 1.00 | 1.02 | -0.45 | 0.97 | -0.50 | -0.51 | 0.01 | 0.00 | 0.00 | 0.00 |
| 3 | -1 | 1.40 | 0.70 | 1.02 | -0.45 | 0.97 | -1.00 | -0.97 | -0.03 | 0.00 | 0.00 | 0.00 |
| 4 | -1 | 2.00 | 1.50 | 1.03 | -0.45 | 0.97 | -0.50 | -0.48 | -0.02 | 0.00 | 0.00 | 0.00 |
| 5 | -1 | 0.90 | 0.10 | 1.03 | -0.46 | 0.96 | -1.35 | -1.34 | -0.01 | 0.00 | 0.00 | 0.00 |
| 6 | -1 | 2.50 | 1.00 | 1.03 | -0.46 | 0.96 | -1.25 | -1.21 | -0.04 | 0.00 | -0.01 | 0.00 |
| 7 | -1 | 1.60 | 0.50 | 1.03 | -0.47 | 0.96 | -1.30 | -1.30 | 0.00 | 0.00 | 0.00 | 0.00 |
| 8 | -1 | 3.00 | 3.00 | 1.03 | -0.47 | 0.96 | 0.50 | 0.44 | 0.06 | -0.01 | 0.02 | 0.02 |
| 9 | -1 | -1.00 | -3.00 | 1.03 | -0.45 | 0.98 | -3.50 | -3.51 | 0.01 | 0.00 | 0.00 | 0.00 |
| 10 | -1 | -1.00 | -0.20 | 1.03 | -0.45 | 0.97 | -0.70 | -0.77 | 0.07 | -0.01 | -0.01 | 0.00 |
| 1 | -1 | 0.80 | 0.40 | 1.02 | -0.46 | 0.97 | -1.00 | -1.00 | 0.00 | 0.00 | 0.00 | 0.00 |
| 2 | -1 | 1.00 | 1.00 | 1.02 | -0.46 | 0.97 | -0.50 | -0.50 | 0.00 | 0.00 | 0.00 | 0.00 |
| 3 | -1 | 1.40 | 0.70 | 1.02 | -0.46 | 0.97 | -1.00 | -0.98 | -0.02 | 0.00 | 0.00 | 0.00 |
| 4 | -1 | 2.00 | 1.50 | 1.02 | -0.46 | 0.97 | -0.50 | -0.48 | -0.02 | 0.00 | 0.00 | 0.00 |
| 5 | -1 | 0.90 | 0.10 | 1.02 | -0.46 | 0.97 | -1.35 | -1.34 | -0.01 | 0.00 | 0.00 | 0.00 |
| 6 | -1 | 2.50 | 1.00 | 1.02 | -0.46 | 0.97 | -1.25 | -1.21 | -0.04 | 0.00 | -0.01 | 0.00 |
| 7 | -1 | 1.60 | 0.50 | 1.03 | -0.47 | 0.97 | -1.30 | -1.30 | 0.00 | 0.00 | 0.00 | 0.00 |
| 8 | -1 | 3.00 | 3.00 | 1.03 | -0.47 | 0.97 | 0.50 | 0.45 | 0.05 | 0.00 | 0.01 | 0.01 |
| 9 | -1 | -1.00 | -3.00 | 1.02 | -0.46 | 0.98 | -3.50 | -3.50 | 0.00 | 0.00 | 0.00 | 0.00 |
| 10 | -1 | -1.00 | -0.20 | 1.02 | -0.46 | 0.98 | -0.70 | -0.76 | 0.06 | -0.01 | -0.01 | 0.00 |
| 1 | -1 | 0.80 | 0.40 | 1.02 | -0.46 | 0.98 | -1.00 | -1.00 | 0.00 | 0.00 | 0.00 | 0.00 |
| 2 | -1 | 1.00 | 1.00 | 1.02 | -0.47 | 0.98 | -0.50 | -0.50 | 0.00 | 0.00 | 0.00 | 0.00 |
| 3 | -1 | 1.40 | 0.70 | 1.02 | -0.46 | 0.98 | -1.00 | -0.98 | -0.02 | 0.00 | 0.00 | 0.00 |
| 4 | -1 | 2.00 | 1.50 | 1.02 | -0.47 | 0.98 | -0.50 | -0.49 | -0.01 | 0.00 | 0.00 | 0.00 |
| 5 | -1 | 0.90 | 0.10 | 1.02 | -0.47 | 0.97 | -1.35 | -1.34 | -0.01 | 0.00 | 0.00 | 0.00 |
| 6 | -1 | 2.50 | 1.00 | 1.02 | -0.47 | 0.97 | -1.25 | -1.22 | -0.03 | 0.00 | -0.01 | 0.00 |
| 7 | -1 | 1.60 | 0.50 | 1.02 | -0.48 | 0.97 | -1.30 | -1.30 | 0.00 | 0.00 | 0.00 | 0.00 |
| 8 | -1 | 3.00 | 3.00 | 1.02 | -0.48 | 0.97 | 0.50 | 0.46 | 0.04 | 0.00 | 0.01 | 0.01 |
| 9 | -1 | -1.00 | -3.00 | 1.02 | -0.47 | 0.98 | -3.50 | -3.50 | 0.00 | 0.00 | 0.00 | 0.00 |
| 10 | -1 | -1.00 | -0.20 | 1.02 | -0.47 | 0.98 | -0.70 | -0.75 | 0.05 | 0.00 | 0.00 | 0.00 |
| 1 | -1 | 0.80 | 0.40 | 1.01 | -0.47 | 0.98 | -1.00 | -1.00 | 0.00 | 0.00 | 0.00 | 0.00 |
| 2 | -1 | 1.00 | 1.00 | 1.01 | -0.47 | 0.98 | -0.50 | -0.50 | 0.00 | 0.00 | 0.00 | 0.00 |
| 3 | -1 | 1.40 | 0.70 | 1.01 | -0.47 | 0.98 | -1.00 | -0.99 | -0.01 | 0.00 | 0.00 | 0.00 |
| 4 | -1 | 2.00 | 1.50 | 1.01 | -0.47 | 0.98 | -0.50 | -0.49 | -0.01 | 0.00 | 0.00 | 0.00 |
| 5 | -1 | 0.90 | 0.10 | 1.02 | -0.47 | 0.98 | -1.35 | -1.35 | 0.00 | 0.00 | 0.00 | 0.00 |
| 6 | -1 | 2.50 | 1.00 | 1.02 | -0.48 | 0.98 | -1.25 | -1.23 | -0.02 | 0.00 | -0.01 | 0.00 |
| 7 | -1 | 1.60 | 0.50 | 1.02 | -0.48 | 0.98 | -1.30 | -1.30 | 0.00 | 0.00 | 0.00 | 0.00 |
| 8 | -1 | 3.00 | 3.00 | 1.02 | -0.48 | 0.98 | 0.50 | 0.47 | 0.03 | 0.00 | 0.01 | 0.01 |
| 9 | -1 | -1.00 | -3.00 | 1.02 | -0.47 | 0.99 | -3.50 | -3.50 | 0.00 | 0.00 | 0.00 | 0.00 |
| 10 | -1 | -1.00 | -0.20 | 1.02 | -0.47 | 0.99 | -0.70 | -0.74 | 0.04 | 0.00 | 0.00 | 0.00 |
| 1 | -1 | 0.80 | 0.40 | 1.01 | -0.48 | 0.98 | -1.00 | -1.00 | 0.00 | 0.00 | 0.00 | 0.00 |
| 2 | -1 | 1.00 | 1.00 | 1.01 | -0.48 | 0.98 | -0.50 | -0.50 | 0.00 | 0.00 | 0.00 | 0.00 |
| 3 | -1 | 1.40 | 0.70 | 1.01 | -0.48 | 0.98 | -1.00 | -0.99 | -0.01 | 0.00 | 0.00 | 0.00 |
| 4 | -1 | 2.00 | 1.50 | 1.01 | -0.48 | 0.98 | -0.50 | -0.49 | -0.01 | 0.00 | 0.00 | 0.00 |
| 5 | -1 | 0.90 | 0.10 | 1.01 | -0.48 | 0.98 | -1.35 | -1.35 | 0.00 | 0.00 | 0.00 | 0.00 |
| 6 | -1 | 2.50 | 1.00 | 1.01 | -0.48 | 0.98 | -1.25 | -1.23 | -0.02 | 0.00 | -0.01 | 0.00 |
| 7 | -1 | 1.60 | 0.50 | 1.02 | -0.48 | 0.98 | -1.30 | -1.30 | 0.00 | 0.00 | 0.00 | 0.00 |
| 8 | -1 | 3.00 | 3.00 | 1.02 | -0.48 | 0.98 | 0.50 | 0.47 | 0.03 | 0.00 | 0.01 | 0.01 |
| 9 | -1 | -1.00 | -3.00 | 1.01 | -0.48 | 0.99 | -3.50 | -3.50 | 0.00 | 0.00 | 0.00 | 0.00 |
| 10 | -1 | -1.00 | -0.20 | 1.01 | -0.48 | 0.99 | -0.70 | -0.73 | 0.03 | 0.00 | 0.00 | 0.00 |
| 1 | -1 | 0.80 | 0.40 | 1.01 | -0.48 | 0.99 | -1.00 | -1.00 | 0.00 | 0.00 | 0.00 | 0.00 |
| 2 | -1 | 1.00 | 1.00 | 1.01 | -0.48 | 0.99 | -0.50 | -0.50 | 0.00 | 0.00 | 0.00 | 0.00 |
| 3 | -1 | 1.40 | 0.70 | 1.01 | -0.48 | 0.99 | -1.00 | -0.99 | -0.01 | 0.00 | 0.00 | 0.00 |
| 4 | -1 | 2.00 | 1.50 | 1.01 | -0.48 | 0.99 | -0.50 | -0.49 | -0.01 | 0.00 | 0.00 | 0.00 |
| 5 | -1 | 0.90 | 0.10 | 1.01 | -0.48 | 0.99 | -1.35 | -1.35 | 0.00 | 0.00 | 0.00 | 0.00 |
| 6 | -1 | 2.50 | 1.00 | 1.01 | -0.48 | 0.99 | -1.25 | -1.23 | -0.02 | 0.00 | 0.00 | 0.00 |
| 7 | -1 | 1.60 | 0.50 | 1.01 | -0.49 | 0.98 | -1.30 | -1.30 | 0.00 | 0.00 | 0.00 | 0.00 |
| 8 | -1 | 3.00 | 3.00 | 1.01 | -0.49 | 0.98 | 0.50 | 0.48 | 0.02 | 0.00 | 0.01 | 0.01 |
| 9 | -1 | -1.00 | -3.00 | 1.01 | -0.48 | 0.99 | -3.50 | -3.50 | 0.00 | 0.00 | 0.00 | 0.00 |
| 10 | -1 | -1.00 | -0.20 | 1.01 | -0.48 | 0.99 | -0.70 | -0.73 | 0.03 | 0.00 | 0.00 | 0.00 |
| 1 | -1 | 0.80 | 0.40 | 1.01 | -0.48 | 0.99 | -1.00 | -1.00 | 0.00 | 0.00 | 0.00 | 0.00 |
| 2 | -1 | 1.00 | 1.00 | 1.01 | -0.48 | 0.99 | -0.50 | -0.50 | 0.00 | 0.00 | 0.00 | 0.00 |
| 3 | -1 | 1.40 | 0.70 | 1.01 | -0.48 | 0.99 | -1.00 | -0.99 | -0.01 | 0.00 | 0.00 | 0.00 |
| 4 | -1 | 2.00 | 1.50 | 1.01 | -0.48 | 0.99 | -0.50 | -0.49 | -0.01 | 0.00 | 0.00 | 0.00 |
| 5 | -1 | 0.90 | 0.10 | 1.01 | -0.49 | 0.99 | -1.35 | -1.35 | 0.00 | 0.00 | 0.00 | 0.00 |
| 6 | -1 | 2.50 | 1.00 | 1.01 | -0.49 | 0.99 | -1.25 | -1.24 | -0.01 | 0.00 | 0.00 | 0.00 |
| 7 | -1 | 1.60 | 0.50 | 1.01 | -0.49 | 0.99 | -1.30 | -1.30 | 0.00 | 0.00 | 0.00 | 0.00 |
| 8 | -1 | 3.00 | 3.00 | 1.01 | -0.49 | 0.99 | 0.50 | 0.48 | 0.02 | 0.00 | 0.01 | 0.01 |
| 9 | -1 | -1.00 | -3.00 | 1.01 | -0.48 | 0.99 | -3.50 | -3.50 | 0.00 | 0.00 | 0.00 | 0.00 |
| 10 | -1 | -1.00 | -0.20 | 1.01 | -0.48 | 0.99 | -0.70 | -0.72 | 0.02 | 0.00 | 0.00 | 0.00 |
| 1 | -1 | 0.80 | 0.40 | 1.01 | -0.49 | 0.99 | -1.00 | -1.00 | 0.00 | 0.00 | 0.00 | 0.00 |
| 2 | -1 | 1.00 | 1.00 | 1.01 | -0.49 | 0.99 | -0.50 | -0.50 | 0.00 | 0.00 | 0.00 | 0.00 |
| 3 | -1 | 1.40 | 0.70 | 1.01 | -0.49 | 0.99 | -1.00 | -0.99 | -0.01 | 0.00 | 0.00 | 0.00 |
| 4 | -1 | 2.00 | 1.50 | 1.01 | -0.49 | 0.99 | -0.50 | -0.50 | 0.00 | 0.00 | 0.00 | 0.00 |
| 5 | -1 | 0.90 | 0.10 | 1.01 | -0.49 | 0.99 | -1.35 | -1.35 | 0.00 | 0.00 | 0.00 | 0.00 |
| 6 | -1 | 2.50 | 1.00 | 1.01 | -0.49 | 0.99 | -1.25 | -1.24 | -0.01 | 0.00 | 0.00 | 0.00 |
| 7 | -1 | 1.60 | 0.50 | 1.01 | -0.49 | 0.99 | -1.30 | -1.30 | 0.00 | 0.00 | 0.00 | 0.00 |
| 8 | -1 | 3.00 | 3.00 | 1.01 | -0.49 | 0.99 | 0.50 | 0.48 | 0.02 | 0.00 | 0.00 | 0.00 |
| 9 | -1 | -1.00 | -3.00 | 1.01 | -0.49 | 0.99 | -3.50 | -3.50 | 0.00 | 0.00 | 0.00 | 0.00 |
| 10 | -1 | -1.00 | -0.20 | 1.01 | -0.49 | 0.99 | -0.70 | -0.72 | 0.02 | 0.00 | 0.00 | 0.00 |
| 1 | -1 | 0.80 | 0.40 | 1.01 | -0.49 | 0.99 | -1.00 | -1.00 | 0.00 | 0.00 | 0.00 | 0.00 |
| 2 | -1 | 1.00 | 1.00 | 1.01 | -0.49 | 0.99 | -0.50 | -0.50 | 0.00 | 0.00 | 0.00 | 0.00 |
| 3 | -1 | 1.40 | 0.70 | 1.01 | -0.49 | 0.99 | -1.00 | -0.99 | -0.01 | 0.00 | 0.00 | 0.00 |
| 4 | -1 | 2.00 | 1.50 | 1.01 | -0.49 | 0.99 | -0.50 | -0.50 | 0.00 | 0.00 | 0.00 | 0.00 |
| 5 | -1 | 0.90 | 0.10 | 1.01 | -0.49 | 0.99 | -1.35 | -1.35 | 0.00 | 0.00 | 0.00 | 0.00 |
| 6 | -1 | 2.50 | 1.00 | 1.01 | -0.49 | 0.99 | -1.25 | -1.24 | -0.01 | 0.00 | 0.00 | 0.00 |
| 7 | -1 | 1.60 | 0.50 | 1.01 | -0.49 | 0.99 | -1.30 | -1.30 | 0.00 | 0.00 | 0.00 | 0.00 |
| 8 | -1 | 3.00 | 3.00 | 1.01 | -0.49 | 0.99 | 0.50 | 0.49 | 0.01 | 0.00 | 0.00 | 0.00 |
| 9 | -1 | -1.00 | -3.00 | 1.01 | -0.49 | 0.99 | -3.50 | -3.50 | 0.00 | 0.00 | 0.00 | 0.00 |
| 10 | -1 | -1.00 | -0.20 | 1.01 | -0.49 | 0.99 | -0.70 | -0.72 | 0.02 | 0.00 | 0.00 | 0.00 |
| 1 | -1 | 0.80 | 0.40 | 1.00 | -0.49 | 0.99 | -1.00 | -1.00 | 0.00 | 0.00 | 0.00 | 0.00 |
| 2 | -1 | 1.00 | 1.00 | 1.00 | -0.49 | 0.99 | -0.50 | -0.50 | 0.00 | 0.00 | 0.00 | 0.00 |
| 3 | -1 | 1.40 | 0.70 | 1.00 | -0.49 | 0.99 | -1.00 | -1.00 | 0.00 | 0.00 | 0.00 | 0.00 |
| 4 | -1 | 2.00 | 1.50 | 1.00 | -0.49 | 0.99 | -0.50 | -0.50 | 0.00 | 0.00 | 0.00 | 0.00 |
| 5 | -1 | 0.90 | 0.10 | 1.01 | -0.49 | 0.99 | -1.35 | -1.35 | 0.00 | 0.00 | 0.00 | 0.00 |
| 6 | -1 | 2.50 | 1.00 | 1.01 | -0.49 | 0.99 | -1.25 | -1.24 | -0.01 | 0.00 | 0.00 | 0.00 |
| 7 | -1 | 1.60 | 0.50 | 1.01 | -0.49 | 0.99 | -1.30 | -1.30 | 0.00 | 0.00 | 0.00 | 0.00 |
| 8 | -1 | 3.00 | 3.00 | 1.01 | -0.49 | 0.99 | 0.50 | 0.49 | 0.01 | 0.00 | 0.00 | 0.00 |
| 9 | -1 | -1.00 | -3.00 | 1.00 | -0.49 | 1.00 | -3.50 | -3.50 | 0.00 | 0.00 | 0.00 | 0.00 |
| 10 | -1 | -1.00 | -0.20 | 1.00 | -0.49 | 1.00 | -0.70 | -0.71 | 0.01 | 0.00 | 0.00 | 0.00 |
| 1 | -1 | 0.80 | 0.40 | 1.00 | -0.49 | 1.00 | -1.00 | -1.00 | 0.00 | 0.00 | 0.00 | 0.00 |
| 2 | -1 | 1.00 | 1.00 | 1.00 | -0.49 | 1.00 | -0.50 | -0.50 | 0.00 | 0.00 | 0.00 | 0.00 |
| 3 | -1 | 1.40 | 0.70 | 1.00 | -0.49 | 1.00 | -1.00 | -1.00 | 0.00 | 0.00 | 0.00 | 0.00 |
| 4 | -1 | 2.00 | 1.50 | 1.00 | -0.49 | 0.99 | -0.50 | -0.50 | 0.00 | 0.00 | 0.00 | 0.00 |
| 5 | -1 | 0.90 | 0.10 | 1.00 | -0.49 | 0.99 | -1.35 | -1.35 | 0.00 | 0.00 | 0.00 | 0.00 |
| 6 | -1 | 2.50 | 1.00 | 1.00 | -0.49 | 0.99 | -1.25 | -1.24 | -0.01 | 0.00 | 0.00 | 0.00 |
| 7 | -1 | 1.60 | 0.50 | 1.00 | -0.50 | 0.99 | -1.30 | -1.30 | 0.00 | 0.00 | 0.00 | 0.00 |
| 8 | -1 | 3.00 | 3.00 | 1.00 | -0.50 | 0.99 | 0.50 | 0.49 | 0.01 | 0.00 | 0.00 | 0.00 |
| 9 | -1 | -1.00 | -3.00 | 1.00 | -0.49 | 1.00 | -3.50 | -3.50 | 0.00 | 0.00 | 0.00 | 0.00 |
| 10 | -1 | -1.00 | -0.20 | 1.00 | -0.49 | 1.00 | -0.70 | -0.71 | 0.01 | 0.00 | 0.00 | 0.00 |
| 1 | -1 | 0.80 | 0.40 | 1.00 | -0.49 | 1.00 | -1.00 | -1.00 | 0.00 | 0.00 | 0.00 | 0.00 |
| 2 | -1 | 1.00 | 1.00 | 1.00 | -0.49 | 1.00 | -0.50 | -0.50 | 0.00 | 0.00 | 0.00 | 0.00 |
| 3 | -1 | 1.40 | 0.70 | 1.00 | -0.49 | 1.00 | -1.00 | -1.00 | 0.00 | 0.00 | 0.00 | 0.00 |
| 4 | -1 | 2.00 | 1.50 | 1.00 | -0.49 | 1.00 | -0.50 | -0.50 | 0.00 | 0.00 | 0.00 | 0.00 |
| 5 | -1 | 0.90 | 0.10 | 1.00 | -0.49 | 1.00 | -1.35 | -1.35 | 0.00 | 0.00 | 0.00 | 0.00 |
| 6 | -1 | 2.50 | 1.00 | 1.00 | -0.49 | 1.00 | -1.25 | -1.24 | -0.01 | 0.00 | 0.00 | 0.00 |
| 7 | -1 | 1.60 | 0.50 | 1.00 | -0.50 | 0.99 | -1.30 | -1.30 | 0.00 | 0.00 | 0.00 | 0.00 |
| 8 | -1 | 3.00 | 3.00 | 1.00 | -0.50 | 0.99 | 0.50 | 0.49 | 0.01 | 0.00 | 0.00 | 0.00 |
| 9 | -1 | -1.00 | -3.00 | 1.00 | -0.49 | 1.00 | -3.50 | -3.50 | 0.00 | 0.00 | 0.00 | 0.00 |
| 10 | -1 | -1.00 | -0.20 | 1.00 | -0.49 | 1.00 | -0.70 | -0.71 | 0.01 | 0.00 | 0.00 | 0.00 |
| 1 | -1 | 0.80 | 0.40 | 1.00 | -0.49 | 1.00 | -1.00 | -1.00 | 0.00 | 0.00 | 0.00 | 0.00 |
| 2 | -1 | 1.00 | 1.00 | 1.00 | -0.49 | 1.00 | -0.50 | -0.50 | 0.00 | 0.00 | 0.00 | 0.00 |
| 3 | -1 | 1.40 | 0.70 | 1.00 | -0.49 | 1.00 | -1.00 | -1.00 | 0.00 | 0.00 | 0.00 | 0.00 |
| 4 | -1 | 2.00 | 1.50 | 1.00 | -0.50 | 1.00 | -0.50 | -0.50 | 0.00 | 0.00 | 0.00 | 0.00 |
| 5 | -1 | 0.90 | 0.10 | 1.00 | -0.50 | 1.00 | -1.35 | -1.35 | 0.00 | 0.00 | 0.00 | 0.00 |
| 6 | -1 | 2.50 | 1.00 | 1.00 | -0.50 | 1.00 | -1.25 | -1.25 | 0.00 | 0.00 | 0.00 | 0.00 |
| 7 | -1 | 1.60 | 0.50 | 1.00 | -0.50 | 1.00 | -1.30 | -1.30 | 0.00 | 0.00 | 0.00 | 0.00 |
| 8 | -1 | 3.00 | 3.00 | 1.00 | -0.50 | 1.00 | 0.50 | 0.49 | 0.01 | 0.00 | 0.00 | 0.00 |
| 9 | -1 | -1.00 | -3.00 | 1.00 | -0.49 | 1.00 | -3.50 | -3.50 | 0.00 | 0.00 | 0.00 | 0.00 |
| 10 | -1 | -1.00 | -0.20 | 1.00 | -0.49 | 1.00 | -0.70 | -0.71 | 0.01 | 0.00 | 0.00 | 0.00 |
| 1 | -1 | 0.80 | 0.40 | 1.00 | -0.50 | 1.00 | -1.00 | -1.00 | 0.00 | 0.00 | 0.00 | 0.00 |
| 2 | -1 | 1.00 | 1.00 | 1.00 | -0.50 | 1.00 | -0.50 | -0.50 | 0.00 | 0.00 | 0.00 | 0.00 |
| 3 | -1 | 1.40 | 0.70 | 1.00 | -0.50 | 1.00 | -1.00 | -1.00 | 0.00 | 0.00 | 0.00 | 0.00 |
| 4 | -1 | 2.00 | 1.50 | 1.00 | -0.50 | 1.00 | -0.50 | -0.50 | 0.00 | 0.00 | 0.00 | 0.00 |
| 5 | -1 | 0.90 | 0.10 | 1.00 | -0.50 | 1.00 | -1.35 | -1.35 | 0.00 | 0.00 | 0.00 | 0.00 |
| 6 | -1 | 2.50 | 1.00 | 1.00 | -0.50 | 1.00 | -1.25 | -1.25 | 0.00 | 0.00 | 0.00 | 0.00 |
| 7 | -1 | 1.60 | 0.50 | 1.00 | -0.50 | 1.00 | -1.30 | -1.30 | 0.00 | 0.00 | 0.00 | 0.00 |
| 8 | -1 | 3.00 | 3.00 | 1.00 | -0.50 | 1.00 | 0.50 | 0.50 | 0.00 | 0.00 | 0.00 | 0.00 |
| 9 | -1 | -1.00 | -3.00 | 1.00 | -0.50 | 1.00 | -3.50 | -3.50 | 0.00 | 0.00 | 0.00 | 0.00 |
| 10 | -1 | -1.00 | -0.20 | 1.00 | -0.50 | 1.00 | -0.70 | -0.71 | 0.01 | 0.00 | 0.00 | 0.00 |
| 1 | -1 | 0.80 | 0.40 | 1.00 | -0.50 | 1.00 | -1.00 | -1.00 | 0.00 | 0.00 | 0.00 | 0.00 |
| 2 | -1 | 1.00 | 1.00 | 1.00 | -0.50 | 1.00 | -0.50 | -0.50 | 0.00 | 0.00 | 0.00 | 0.00 |
| 3 | -1 | 1.40 | 0.70 | 1.00 | -0.50 | 1.00 | -1.00 | -1.00 | 0.00 | 0.00 | 0.00 | 0.00 |
| 4 | -1 | 2.00 | 1.50 | 1.00 | -0.50 | 1.00 | -0.50 | -0.50 | 0.00 | 0.00 | 0.00 | 0.00 |
| 5 | -1 | 0.90 | 0.10 | 1.00 | -0.50 | 1.00 | -1.35 | -1.35 | 0.00 | 0.00 | 0.00 | 0.00 |
| 6 | -1 | 2.50 | 1.00 | 1.00 | -0.50 | 1.00 | -1.25 | -1.25 | 0.00 | 0.00 | 0.00 | 0.00 |
| 7 | -1 | 1.60 | 0.50 | 1.00 | -0.50 | 1.00 | -1.30 | -1.30 | 0.00 | 0.00 | 0.00 | 0.00 |
| 8 | -1 | 3.00 | 3.00 | 1.00 | -0.50 | 1.00 | 0.50 | 0.50 | 0.00 | 0.00 | 0.00 | 0.00 |
| 9 | -1 | -1.00 | -3.00 | 1.00 | -0.50 | 1.00 | -3.50 | -3.50 | 0.00 | 0.00 | 0.00 | 0.00 |
| 10 | -1 | -1.00 | -0.20 | 1.00 | -0.50 | 1.00 | -0.70 | -0.71 | 0.01 | 0.00 | 0.00 | 0.00 |
| 1 | -1 | 0.80 | 0.40 | 1.00 | -0.50 | 1.00 | -1.00 | -1.00 | 0.00 | 0.00 | 0.00 | 0.00 |
| 2 | -1 | 1.00 | 1.00 | 1.00 | -0.50 | 1.00 | -0.50 | -0.50 | 0.00 | 0.00 | 0.00 | 0.00 |
| 3 | -1 | 1.40 | 0.70 | 1.00 | -0.50 | 1.00 | -1.00 | -1.00 | 0.00 | 0.00 | 0.00 | 0.00 |
| 4 | -1 | 2.00 | 1.50 | 1.00 | -0.50 | 1.00 | -0.50 | -0.50 | 0.00 | 0.00 | 0.00 | 0.00 |
| 5 | -1 | 0.90 | 0.10 | 1.00 | -0.50 | 1.00 | -1.35 | -1.35 | 0.00 | 0.00 | 0.00 | 0.00 |
| 6 | -1 | 2.50 | 1.00 | 1.00 | -0.50 | 1.00 | -1.25 | -1.25 | 0.00 | 0.00 | 0.00 | 0.00 |
| 7 | -1 | 1.60 | 0.50 | 1.00 | -0.50 | 1.00 | -1.30 | -1.30 | 0.00 | 0.00 | 0.00 | 0.00 |
| 8 | -1 | 3.00 | 3.00 | 1.00 | -0.50 | 1.00 | 0.50 | 0.50 | 0.00 | 0.00 | 0.00 | 0.00 |
| 9 | -1 | -1.00 | -3.00 | 1.00 | -0.50 | 1.00 | -3.50 | -3.50 | 0.00 | 0.00 | 0.00 | 0.00 |
| 10 | -1 | -1.00 | -0.20 | 1.00 | -0.50 | 1.00 | -0.70 | -0.70 | 0.00 | 0.00 | 0.00 | 0.00 |

**Table A1. 2**: Process of $\delta$-rule correction for a linear artificial neuron, for $\eta$ = 0.1. The training data set consists of ten patterns. In this case, 25 updates for each pattern were necessary before the exact set of weights was learned.

This last example is in fact an interpolation problem, since the training data set does not contain noise, so that all the target values are exact (hence there is only one solution). The same neuron with the same training patterns was trained again, but this time artificial noise was introduced into the target values. After the same 25 updates as in the last example, the error could not be eliminated, but it was minimized. Further updates following the same rule does not lead to any improvement. The **Table A1. 3** shows only the 25th update.

| Pattern | $x_0$ | $x_1$ | $x_2$ | $w_0$ | $w_1$ | $w_2$ | $t$ | $y$ | $\delta$ | $\Delta w_0$ | $\Delta w_1$ | $\Delta w_2$ |
|---|---|---|---|---|---|---|---|---|---|---|---|---|
| 1 | -1 | 0.80 | 0.40 | 1.01 | -0.45 | 0.97 | -1.02 | -0.98 | -0.04 | 0.00 | 0.00 | 0.00 |
| 2 | -1 | 1.00 | 1.00 | 1.01 | -0.46 | 0.96 | -0.55 | -0.50 | -0.05 | 0.00 | 0.00 | 0.00 |
| 3 | -1 | 1.40 | 0.70 | 1.02 | -0.46 | 0.96 | -0.99 | -0.99 | 0.00 | 0.00 | 0.00 | 0.00 |
| 4 | -1 | 2.00 | 1.50 | 1.02 | -0.46 | 0.96 | -0.53 | -0.50 | -0.03 | 0.00 | -0.01 | -0.01 |
| 5 | -1 | 0.90 | 0.10 | 1.02 | -0.47 | 0.96 | -1.30 | -1.34 | 0.04 | 0.00 | 0.00 | 0.00 |
| 6 | -1 | 2.50 | 1.00 | 1.02 | -0.46 | 0.96 | -1.15 | -1.22 | 0.07 | -0.01 | 0.02 | 0.01 |
| 7 | -1 | 1.60 | 0.50 | 1.01 | -0.45 | 0.96 | -1.31 | -1.24 | -0.07 | 0.01 | -0.01 | 0.00 |
| 8 | -1 | 3.00 | 3.00 | 1.02 | -0.46 | 0.96 | 0.52 | 0.49 | 0.03 | 0.00 | 0.01 | 0.01 |
| 9 | -1 | -1.00 | -3.00 | 1.01 | -0.45 | 0.97 | -3.47 | -3.47 | 0.00 | 0.00 | 0.00 | 0.00 |
| 10 | -1 | -1.00 | -0.20 | 1.01 | -0.45 | 0.97 | -0.71 | -0.76 | 0.05 | 0.00 | 0.00 | 0.00 |

**Table A1. 3**: Process of $\delta$-rule correction for a linear artificial neuron with noisy inputs, for $\eta$ = 0.1. The training data set consists of ten patterns with artificial noise added. In this case, 25 updates for each pattern were necessary to learn the optimum weights, although the errors cannot be eliminated.

# Appendix 2

# BRIEF REVIEW OF GENETICS

## A2.1 Deoxyribonucleic acid

The deoxyribonucleic acid (DNA) is organized as a double-stranded macromolecule of helical structure, sometimes known as "the molecule of heredity", resembling a twisted ladder.

The DNA is encoded with four "building blocks" called "bases" (A: Adenine, T: Thymine, C: Cytosine, G: Guanine) which constitute the base-four alphabet of the genetic information. The strands are connected to each other by the rungs of the "twisted ladder", which are built with the four bases plus hydrogen bonds. Each rung can only pair up an Adenine with a Thymine, or a Cytosine with a Guanine, by means of two or three hydrogen bonds, respectively (see **Fig. A2. 1**). Note that the order does matter. That is to say that T + A ≠ A + T, so that four different combinations are possible).

However, since each base can only combine with only another specific one, the sequence can be described by only specifying one base on one conventionally chosen strand. The order of the bases in the DNA encompasses the genetic instructions to build an organism.

Each strand of DNA consists of a chain of nucleotides, each of which is composed by sugar (deoxyribose), phosphate, and one of the four bases. As shown in **Fig. A2. 1**, the pairing up of the bases forces them to face the axis of the helix, letting the sugar and phosphate run along the outside forming what is known as the "backbones" of the helix. Its function consists of linking one nucleotide base to the next in the same DNA strand.

Replication is performed by disassociating the double strand and building the other half of each new single strand by exposing the latter to an environment consisting of the four bases. Since each base can only combine with another specific one, the base on the old strand directs the building of the new one, which ideally would result in an exact replication of the original. However, errors (mutations) might occur during this process.

Thus, it can be simplistically said that the DNA consists of a sequence of letters (representing the nucleotide bases) such as: ATGCTCAACTGTTCAGGCACGTAT. This sequence of letters comprises the information for the building plan.

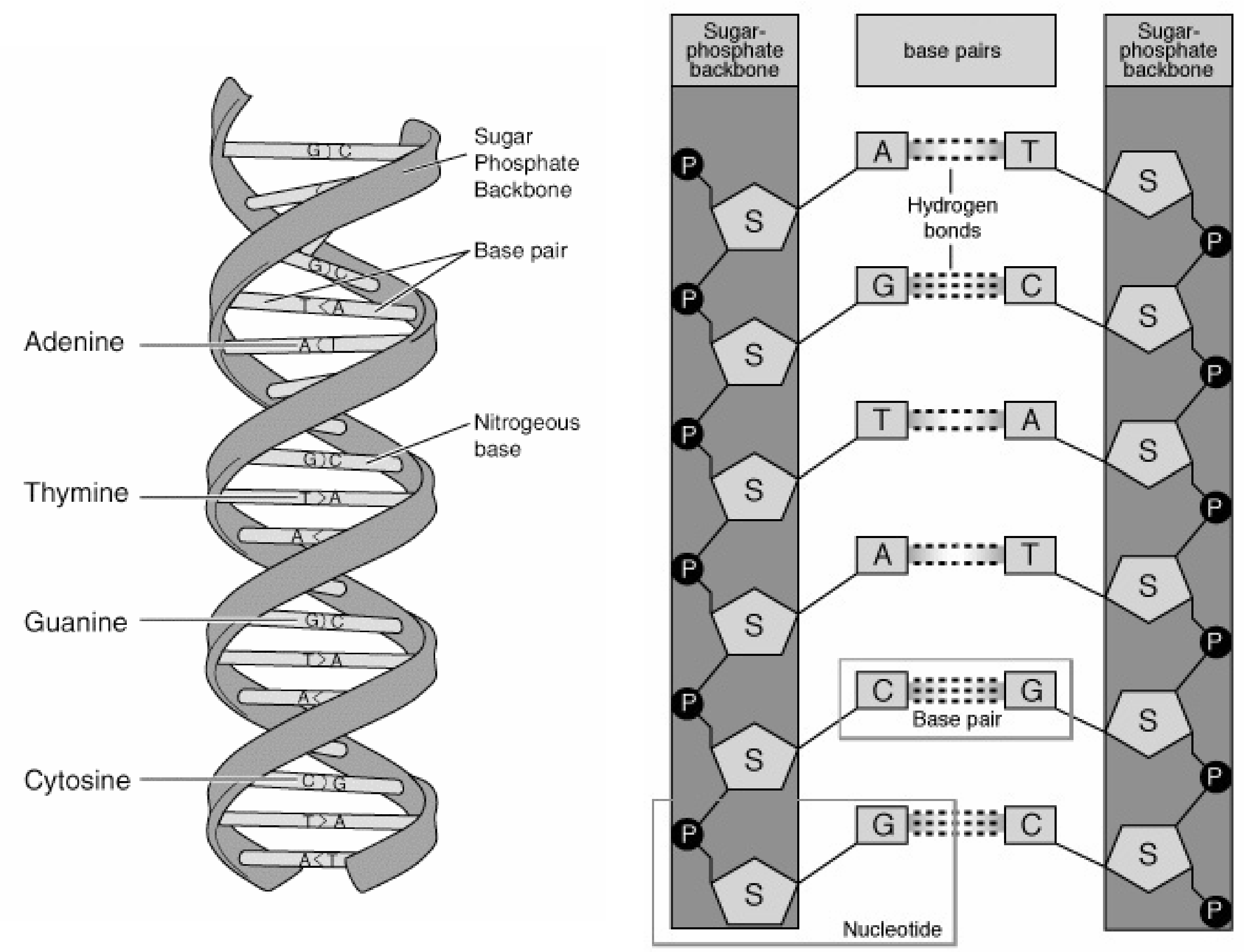

**Fig. A2. 1**: Schematic representation of the DNA, which illustrates its double helix structure and the base pairs by means of hydrogen bonds (from [80, 1]).

Once agreed that the DNA contains the necessary information to create life, it is immediate to question about the mechanisms to create organisms from the genotypes encoded in the DNA. In other words, the mapping functions relating the building plan to the observable features of the organisms. This is a major subject which will not be dealt with in this work. It is only mentioned that the mapping consists of a nonlinear function, where any phenotype's modification may be due to several genes' change, and a single gene's modification can lead to many observable features being modified. This fact leads to the EAs not intending to mimic such a complex nonlinear mapping, but implementing their own with no biological counterpart.

# A2.2 Genes

Genes are the units of heredity that parents pass to their offspring during reproduction. They rule the organisms that contain them, so that the genes that exist today are those which have been successfully passed through generations from parents to progeny. Often, many individual organisms contain the same allele of a particular gene, making it possible for an allele to survive despite the death of an individual.

As mentioned before, the DNA could be interpreted as a sequence of "letters", such us ATGCTCAACTGTTCAGGCACGTAT, which contains the essential information to build a biological organism. A sequence of three consecutive nucleotides called "codon" is the protein-coding vocabulary. Keeping the sequence of "letters" in the DNA, the "letters" make "words" by grouping them in threes (i.e. forming codons): ATG CTC AAC TGT TCA GGC ACG TAT. The "words", in turn, make "sentences": [ATG CTC AAC TGT] [TCA GGC ACG TAT]. These sentences are called "genes".

The sequence of codons in a gene specifies the amino-acid sequence of the proteins it encodes. Therefore, genes are the actual instruction manual for building the necessary proteins for the growth and regulation of the biological organism. Proteins enable a cell to perform special functions.

The analogy of DNA as "letters", codons as "words" and genes as "sentences" is quite illustrative: Genes are made of DNA. However, each "piece" of DNA contains not only genes but also areas that regulate them and areas whose function in not known yet. In most eukaryotic species[1], only a small part of the DNA in the genome encodes proteins (in fact, less than two percent of the human genome does it). Hence, genes may be separated by vast sequences of so-called "junk-DNA". Furthermore, the genes are often fragmented internally by non-coding sequences called "introns", while only "exons" actually do encode proteins.

Proteins are the devices in charge of making all biological organisms function, so that they can be seen as different parts of an engine. An organism is made up of specialized cells, each of which contains thousands of different proteins working together performing different tasks needed for the cell's job. Cells use the directives contained in their genes to build proteins.

[1] Eukaryotes are organisms having cells with membrane-bound nucleus.

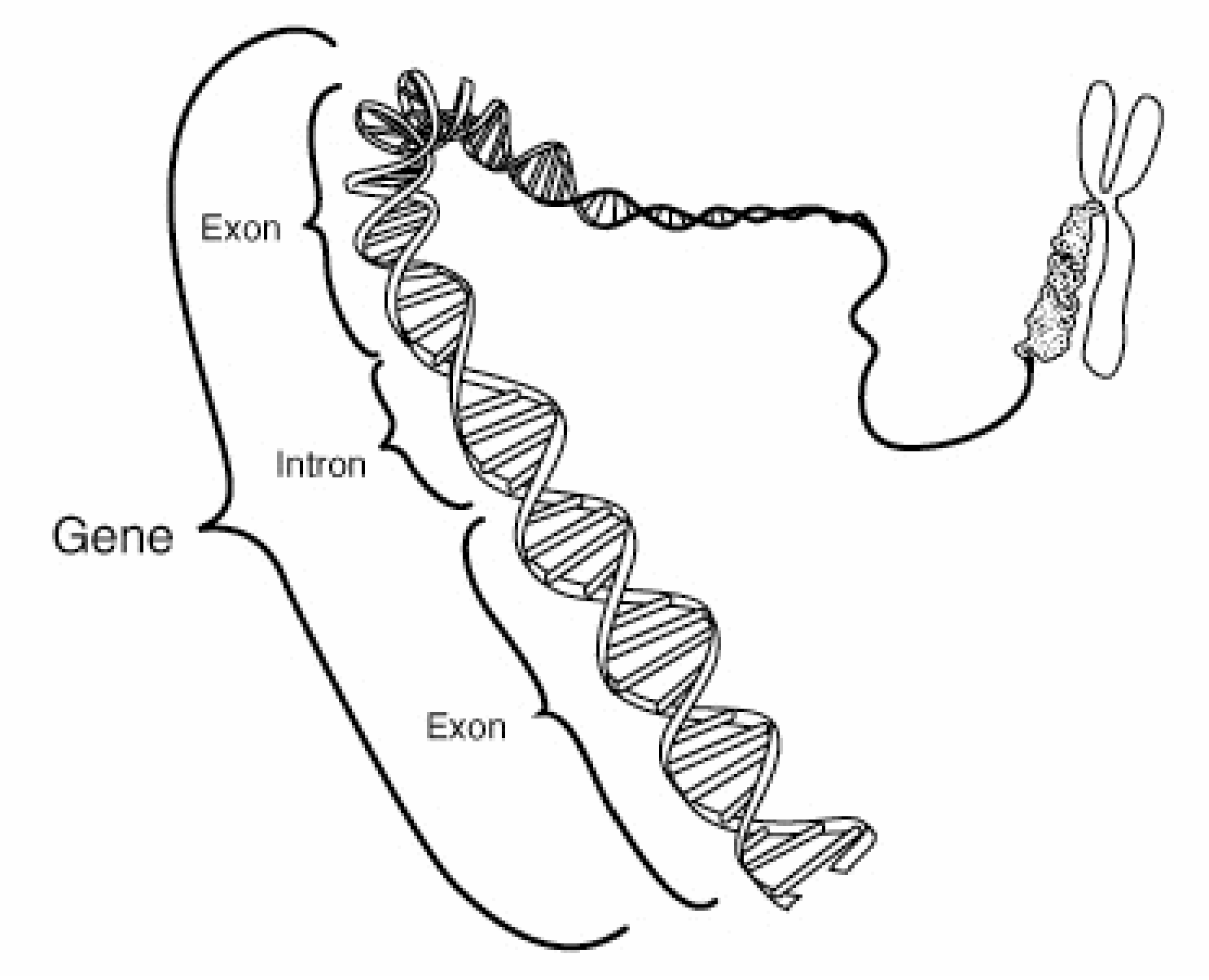

**Fig. A2. 2**: Schematic diagram showing a gene in relation to the double helix structure of DNA and to a chromosome (right). Introns are regions often found in eukaryotes' genes which are removed in the splicing process: only the exons encode proteins. This diagram labels a region of only forty or so bases as a gene. In reality many genes are much larger (from [80]).

When a certain protein is needed, the information encoded in the gene is read and a molecular message is produced in the form of "ribonucleic acid" (RNA), which is a molecule similar to the DNA[2]. The RNA moves from the nucleus to the cytoplasm of the cell and the ribosome (cell's protein-making machinery) reads the message and produces the protein matching the specifications. Then the protein travels to wherever it is needed.

# A2.3 Chromosomes

A chromosome is a very long continuous piece of DNA that contains many genes, regulatory elements and other intervening nucleotide sequences (see **Fig. A2. 2** and **Fig. A2. 3**).

The "genome" of an organism encompasses all the genetic information, which in many eukaryotic species is divided among several chromosomes. Genes are made of DNA, and reside in chromosomes. Organisms that only present a single set of chromosomes, are called "haploids". Species that carry more than one copy of their genome within each somatic (i.e. non-reproductive) cells are called "diploids" if they have two copies, or "polyploids" if they have more. In such organisms, the copies are practically never identical.

[2] The nucleotide "Uracil" in RNA replaces the nucleotide "Thymine" in DNA. Moreover, RNA is single stranded, as opposed to DNA.

Each human cell's DNA is organized into two complete sets of 23 chromosomes: 22 pairs plus the sexual pair (XX: female, XY: male). A mosquito, instead, has three pairs of chromosomes, and bacteria have one chromosome only, to name some examples.

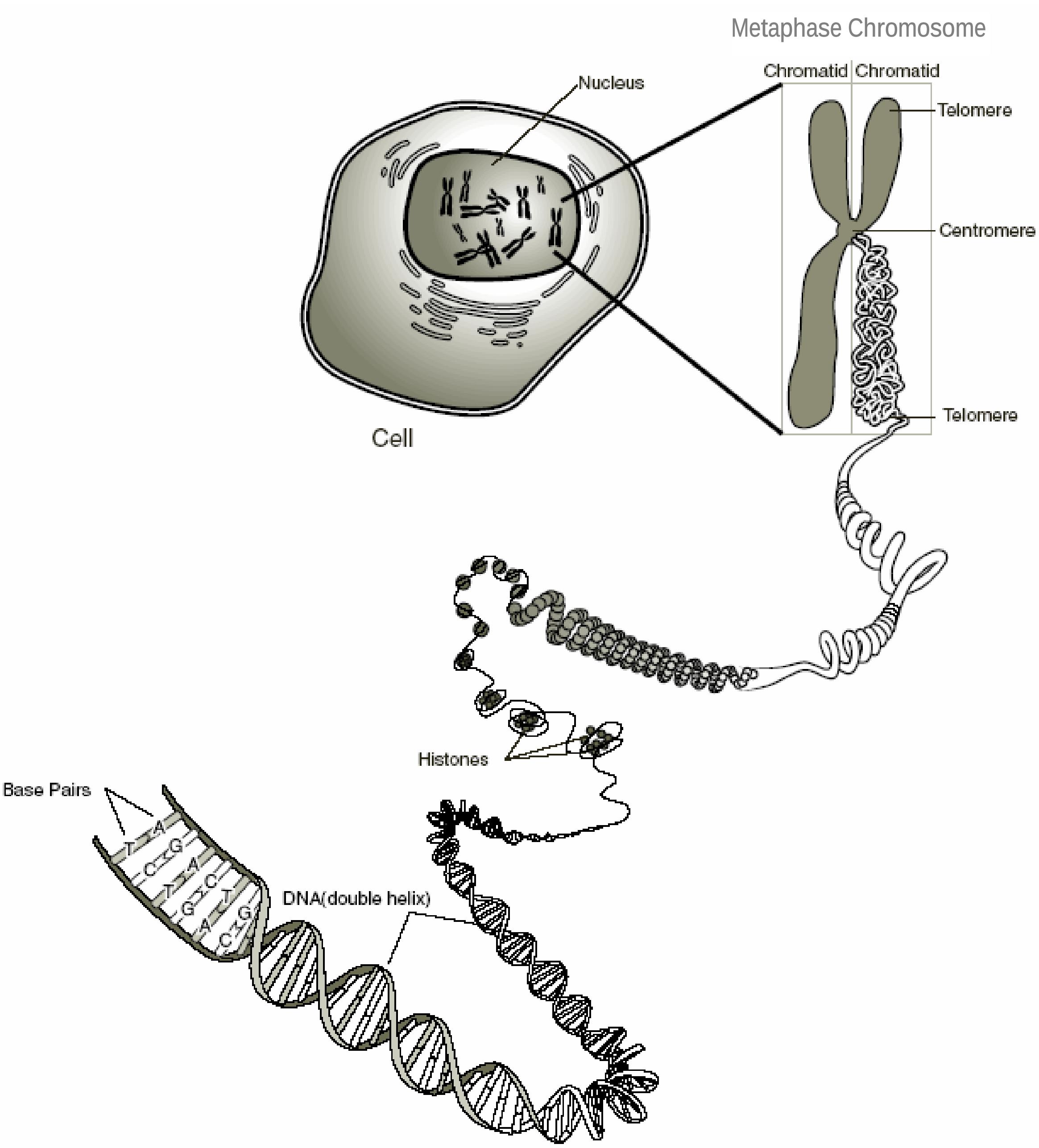

**Fig. A2. 3**: Different scales of the genetic information encoding and storage. Genes are made of DNA, and they reside in chromosomes. Chromosomes reside inside the nucleus of every cell (from [1]).

# A2.4 Cellular reproduction

### Asexual reproduction

It is a form of duplication by means of replication and cell division. The usual methods for asexual reproduction are "mitosis" in eukaryotes, and "binary fission" in prokaryotes, both of which are processes of cell division that produce two cells from a single parent. The offspring individuals are called "clones" because each one is an exact copy of the original cell, except for rare spontaneous mutations (and recombination in eukaryotes).

In eukaryotes, after replication, the cell division takes place so that each new cell receives one maternal and one paternal analogous chromosome[3]. Some typical examples are a new plant growing out of the root, and the repair and growth of different tissues.

In prokaryotes, once the only chromosome is replicated, each copy is attached to a different part of the cell's membrane, so that the two "sister" chromatids are separated during the cell division. A typical example is the reproduction of haploids organisms, such as bacteria.

### Sexual reproduction

It consists of the formation of a new individual by a combination of two haploid cells called "gametes", which come from separate parents. They are produced by "meiosis", which is the type of cell division by which four reproductive cells are produced from a somatic parent. The process involves a reduction in the amount of genetic material, since the children are haploids. It takes place in two steps: in "meiosis I", a DNA replication occurs, so that each duplicated chromosome is composed by two "sister" chromatids (see **Fig. A2. 3**). Then, a first division produces two new cells, each of which gets either the maternal or the paternal duplicated chromosome. During this division, when the homologous chromosomes are pressed together and eventually break, the genetic recombination occurs by exchanging portions of genetic material (so-called crossover). The process of "meiosis II" leads to the gametes formation by the division of the duplicated chromosomes. Hence, one parent diploid cell produces four daughter haploid cells which have half of the number of chromosomes found in the original parent cell. Furthermore, due to recombination, they are genetically different. Meiosis converts a diploid cell into four haploid gametes, and causes a change in the genetic information to increase diversity in the progeny. It differs from mitosis in that there are two cell divisions and recombination of genetic material, resulting in haploid and unequal cells.

[3] Notice that maternal and paternal chromosomes here refer to the chromosomes of the parents of the cell being cloned. That is to say that the cell which a clone is created from is not called a parent!

When diploid organisms conceive a child, each parent contributes one complete set to it during conception, provided by the sperm and egg cells. When these two reproductive cells join, a single cell is created (zygote), which has now two sets of chromosomes. One of the two new sets can contain chromosomes from only one parent (provided by the sperm cell), and the other set from the other parent (provided by the egg cell). However, due to the recombination that occurred during meiosis I, the set of chromosomes contributed by each reproductive cell is composed by some chromosomes from one, some from the other and some from a recombination of both sets of chromosomes belonging to that parent. Thus, every child inherits a unique genome composed by two sets of chromosomes, resulting in a unique set of traits. Some traits will resemble one parent, some the other, and some will be unique.

Mutations might occur either in the gametes or in somatic cells, but only the ones occurring in gametes are taken into account in (some) EAs.

## A2.5 Traits

Traits are the observable features of an organism, shaped by the genetic code and by the environment. Diploid organisms have paired homologous chromosomes in their somatic cells, thus containing two copies of each gene. Scientists define an allele as any of the alternative forms of a gene that may occur at a given locus in a chromosome. If the two alleles in the same position are identical, the organism is said to be "homozygus" for that gene, otherwise, it is called "heterozygous". They are generally different, and the traits are expressed according to the "dominant", which masks the "recessive" allele. However, the recessive allele can be passed to the next generation and become active, so that the trait (not the gene) skips a generation. The parents also have two alleles in each locus, passing one each to the offspring, randomly chosen. Alleles can also interact to produce "incomplete dominance", therefore the trait would be something in between (a red and a white parents can produce pink progeny). Another possibility is "co-dominance", where both alleles are active at the same time, such the case of red and white flowers in the same plant, for instance.

Traits influenced by only one gene are very rare. Usually the modification of a single trait is due to simultaneous interactions of several genes' modifications (polygenic behaviour).

The non-genetic influence, i.e. the environment, is just as important as the genotypes in defining and shaping traits. For instance, the hair colour of a person depends on its genetic material as well as on the atmospheric conditions. The same is true for the skin colour.

# A2.6 Prokaryotes

Prokaryotes are organisms whose cells lack a membrane-bound nucelus. Prime examples are bacteria and blue-green algae. They reproduce by mitosis, so that the children are exact copies (i.e. clones) of the original cell. Since they are always haploids, any change or mutation is expressed straightaway, and their rate of mutation is noticeably higher than that of eukaryotes.

They possess a genetic system much simpler than eukaryotes'. Commonly, their chromosome is circular, and during reproduction its shape resembles the greek letter "θ" (theta).

In spite of the fact that they are haploids[4] and that their reproduction is asexual (binary fission[5]) resulting in clones colonies, bacteria can recombine their genes in different ways. In eukaryotes, the recombination occurs between chromosomes coming from different parents, whereas in prokaryotes it occurs between a cell's genome and a small piece of genetic material coming from another cell.

### Recombination in prokaryotes

- Conjugation: it is a kind of mating process involving two bacteria. The transference of genetic material takes place from one bacterial cell (F+) to the other (F-), requiring physical contact via a protein tube called "F pilus". The difference between an F+ and and F- bacteria is the presence of the fertility factor "F" in the former, which is a DNA molecule independent from the chromosomal DNA. Thus, the "F+" cell initiates the process by extending an "F pilus" tube towards the host cell ("F-"). The genetic material transferred is the "F factor" itself, which contains among its genes the ones encoding the

[4] Recombination by crossover requires homologous chromosomes to occur.

[5] Binary fission: *method of asexual reproduction which involves the splitting of a parent cell into two approximately equal parts* [23].

"F pilus" building. In the same fashion as recombination occurs during meiosis in eukaryotes by exchanging genes, once the "F factor" is inside the receiver, their DNA exchange genes, so that the genes received from the "F factor" become part of the cell's genome. Notice that while the transfer is being carried out, the "F factor" replicates itself, so that in the end both cells are "F+".

- Transformation: it is a process by which a bacterium can assimilate (i.e. incorporate to its genome) genetic material from the environment, mainly coming from dead bacteria.
- Transduction: it involves the exchange of DNA between bacteria using bacterial viruses as an intermediate. This process involves bacterial infection by viruses together with the reproduction of the infected cell, the spread of the infection and sometimes the recombination between the added virus' DNA and the cell's DNA. This is way beyond the scope of this work, but it is mentioned here for completeness.

# Appendix 3

# OPTIMIZERS' BENCHMARK TEST FUNCTIONS

## A3.1 Introduction

There are many different benchmark functions used by different authors to test the capabilities of their optimization algorithms. The "test suit" used to test the capabilities of the different particle swarm optimizers developed throughout this thesis is composed of five benchmark functions, which are described hereafter. All the functions map from $\mathcal{R}^n \to \mathcal{R}$.

## A3.2 Sphere

The optimization algorithms do not only have to perform well when optimizing extremely complex functions, but also when the function is simple. Hence the simple sphere function is often found in the test suits. The features of this function are as follows:

- Function to be minimized: $f(\mathbf{x}) = \sum_{i=1}^{n} x_i^2$
- Region of the search-space: $[-100,100]^n$
- Dimensions: $n = 30$
- Global optimum: $f(\hat{\mathbf{x}}) = 0$
- Location of the global optimum: $x_i = 0 \quad \forall i$

This is a very simple function that displays a single global optimum (uni-modal) and no local optimum. The function is ideal for gradient-descent techniques because all gradients point towards the global optimum.

The Sphere benchmark function is plotted for a two-dimensional search-space within the region $[-100,100]^2$ in **Fig. A3. 1**:

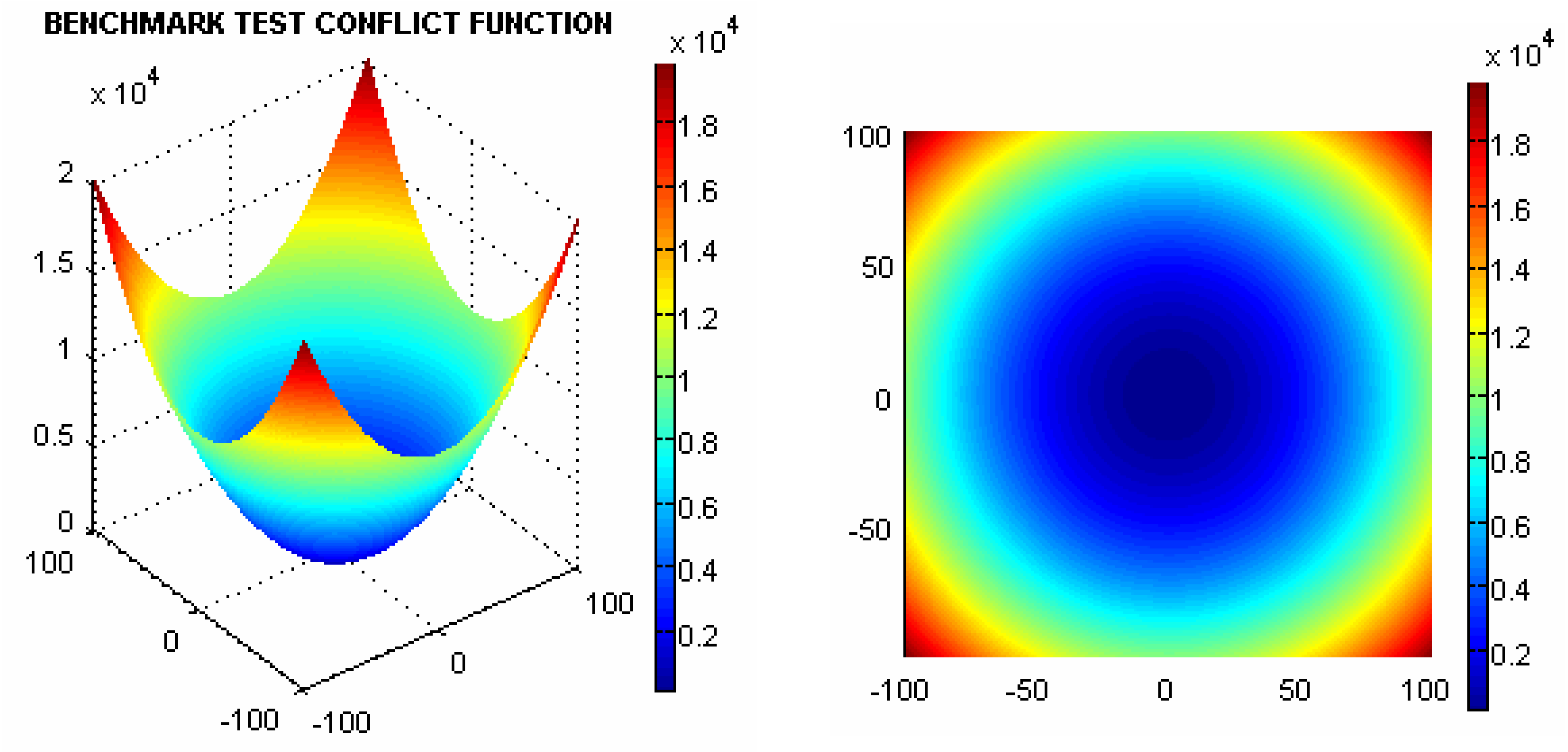

**Fig. A3. 1**: Surface plot and colour-map of the **Sphere benchmark function** for 2-dimensional search-spaces in the region $[-100,100]^2$.

## A3.3 Rosenbrock

Since the Rosenbrock function is very difficult to be optimized by population-based methods, it is frequently found in the test suits used to evaluate these algorithms. It is a uni-modal function, suitable to test optimizers in search-spaces of more than one dimension, which displays an extensive flat surface around the optimum. As opposed to the Sphere function, the variables here are strongly dependent. The features of this function are as follows:

- Function to be minimized: $f(\mathbf{x}) = \sum_{i=1}^{n-1} 100 \cdot \left(x_{i+1} - x_i^{\,2}\right)^2 + (x_i - 1)^2$
- Region of the search-space: $[-30,30]^n$
- Dimensions: $n = 30$
- Global optimum: $f(\hat{\mathbf{x}}) = 0$
- Location of the global optimum: $x_i = 1 \quad \forall i$

The function is plotted for a two-dimensional search-space within the region $[-30,30]^2$ in **Fig. A3. 2**, within the region $[-1.5,2]^2$ in **Fig. A3. 3**, and within the region $[0.8,1.2]^2$ in **Fig. A3. 4**:

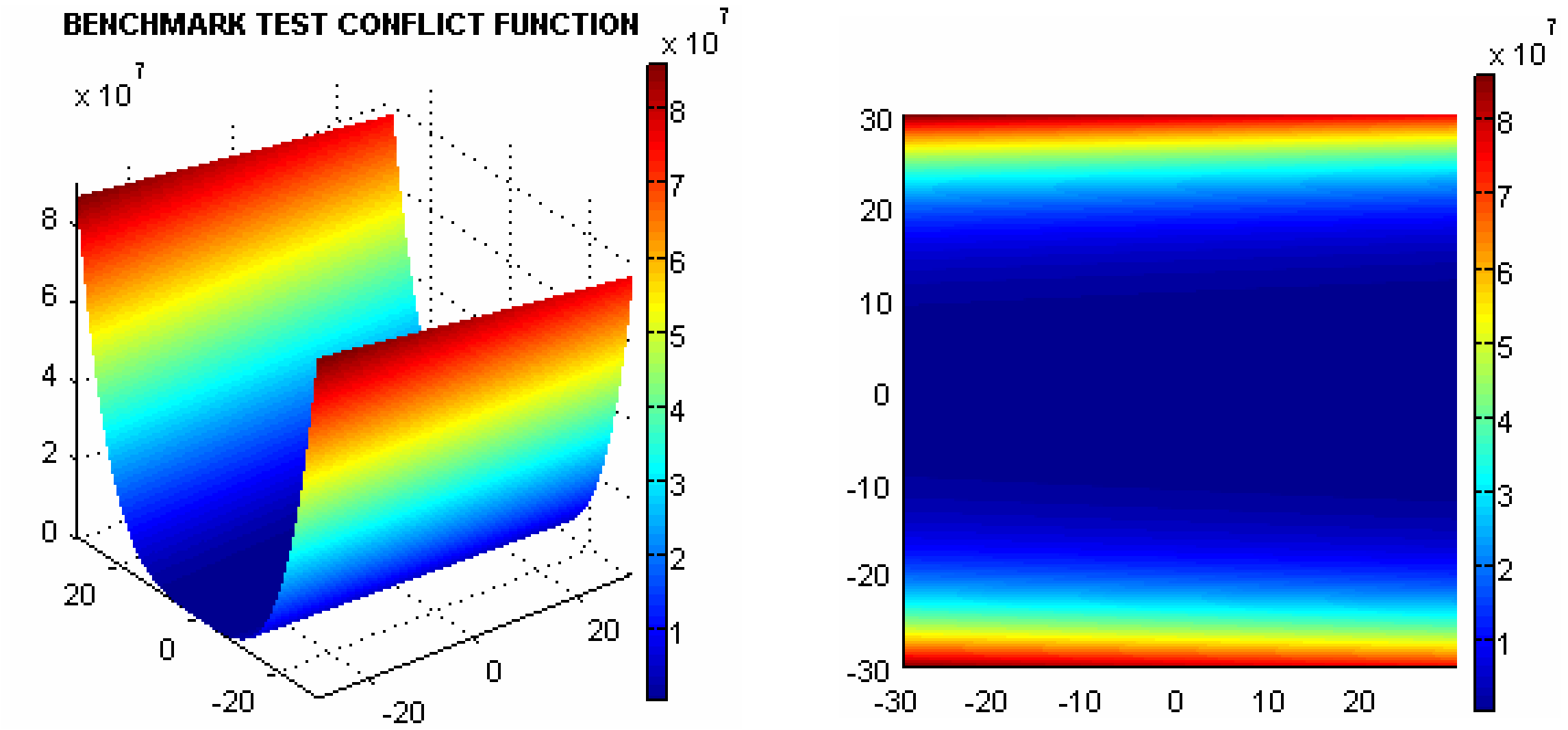

**Fig. A3. 2**: Surface plot and colour-map of the **Rosenbrock benchmark function** for 2-dimensional search-spaces in the region $[-30,30]^2$.

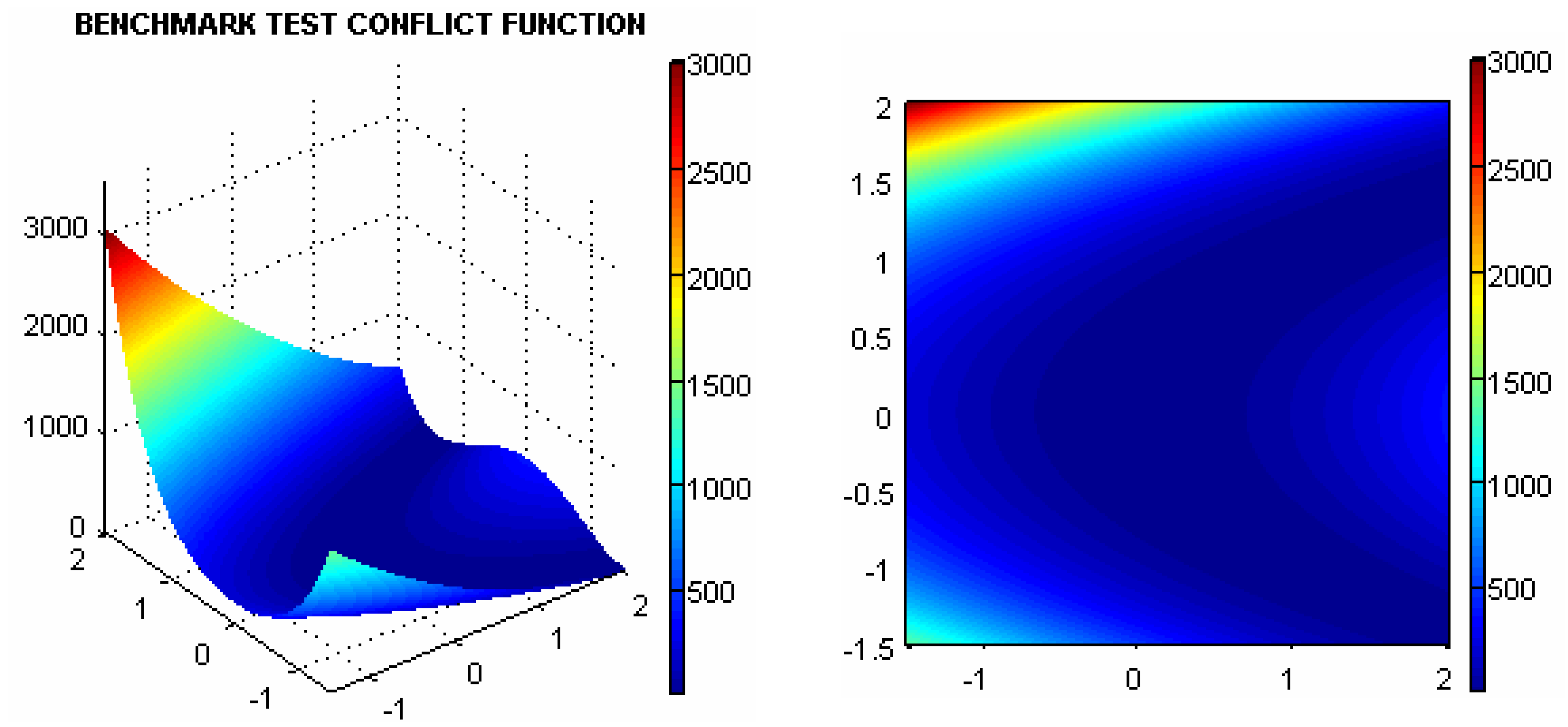

**Fig. A3. 3**: Surface plot and colour-map of the **Rosenbrock benchmark function** for 2-dimensional search-spaces in the region $[-1.5,2]^2$.

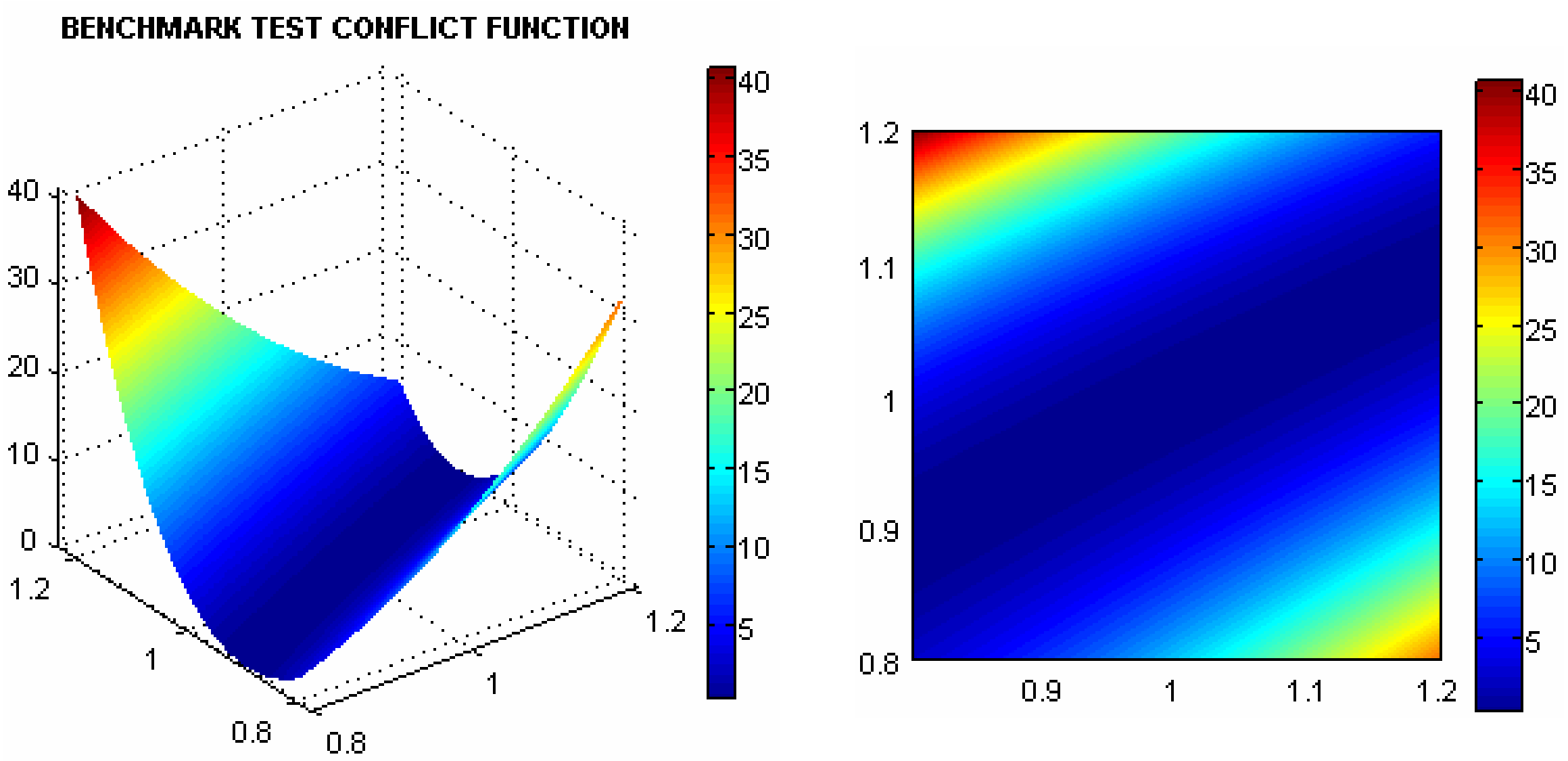

**Fig. A3. 4**: Surface plot and colour-map of the **Rosenbrock benchmark function** for 2-dimensional search-spaces in the region $[0.8,1.2]^2$.

## A3.4 Rastrigrin

The Rastrigrin function displays a single global optimum (uni-modal) and many local optima in the form of valleys. The difficulty in finding the global optimum of this function is in that the very many local optima are worth conflict values that are very similar to the global optimum. Therefore, the particles find it difficult to discover which of the many valleys the one which contains the global optimum is. The features of this function are as follows:

- Function to be minimized: $f(\mathbf{x}) = \sum_{i=1}^{n} \left[ x_i^2 - 10 \cdot \cos(2 \cdot \pi \cdot x_i) + 10 \right]$
- Region of the search-space: $[-5.12, 5.12]^n$
- Dimensions: $n = 30$
- Global optimum: $f(\hat{\mathbf{x}}) = 0$
- Location of the global optimum: $x_i = 0 \quad \forall i$

The function is plotted for a two-dimensional search-space within the regions $[-5.12,5.12]^2$, $[-1,1]^2$, and $[-0.5,0.5]^2$ in **Fig. A3. 5**, **Fig. A3. 6**, and **Fig. A3. 7**, respectively:

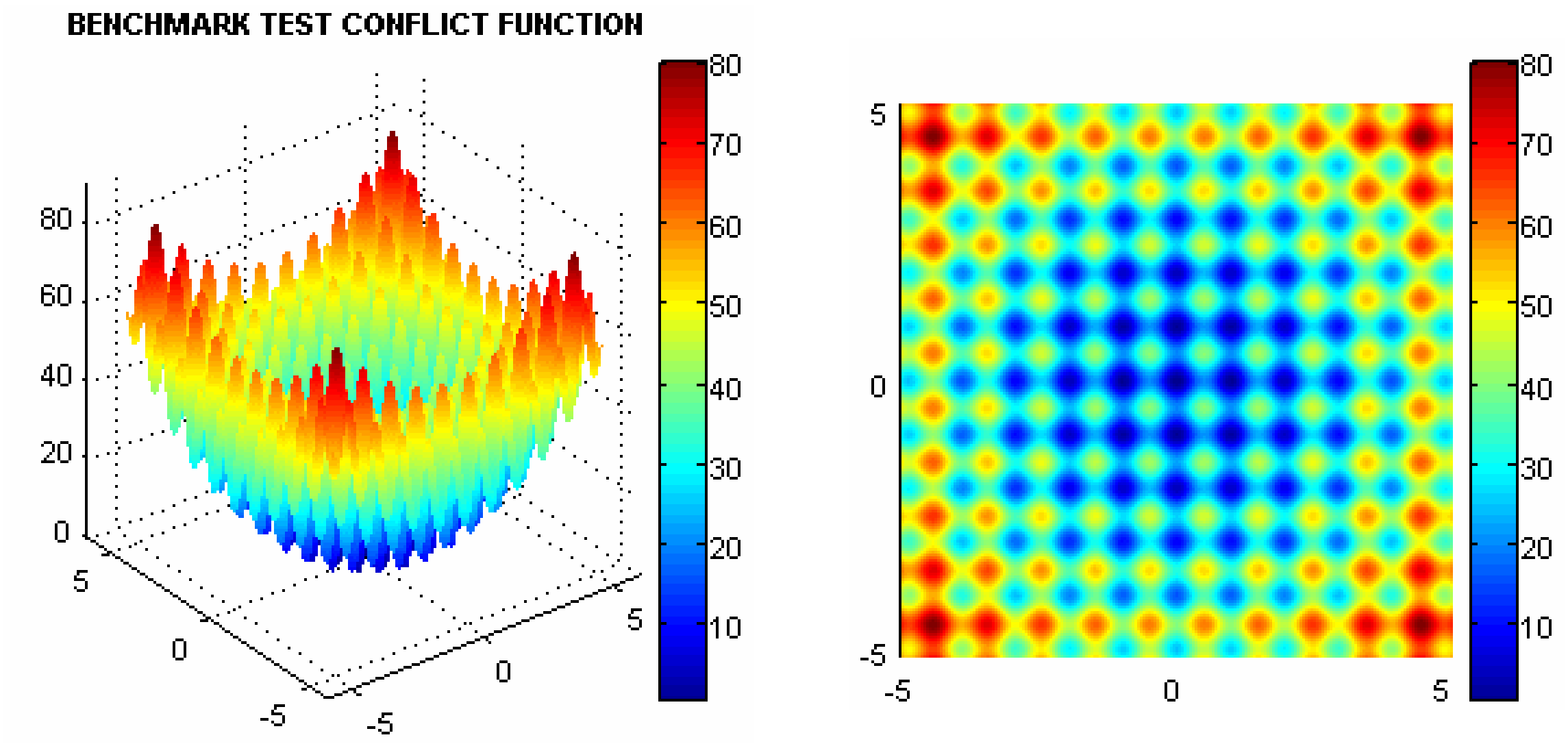

**Fig. A3. 5**: Surface plot and colour-map of the **Rastrigrin benchmark function** for 2-dimensional search-spaces in the region $[-5.12,5.12]^2$.

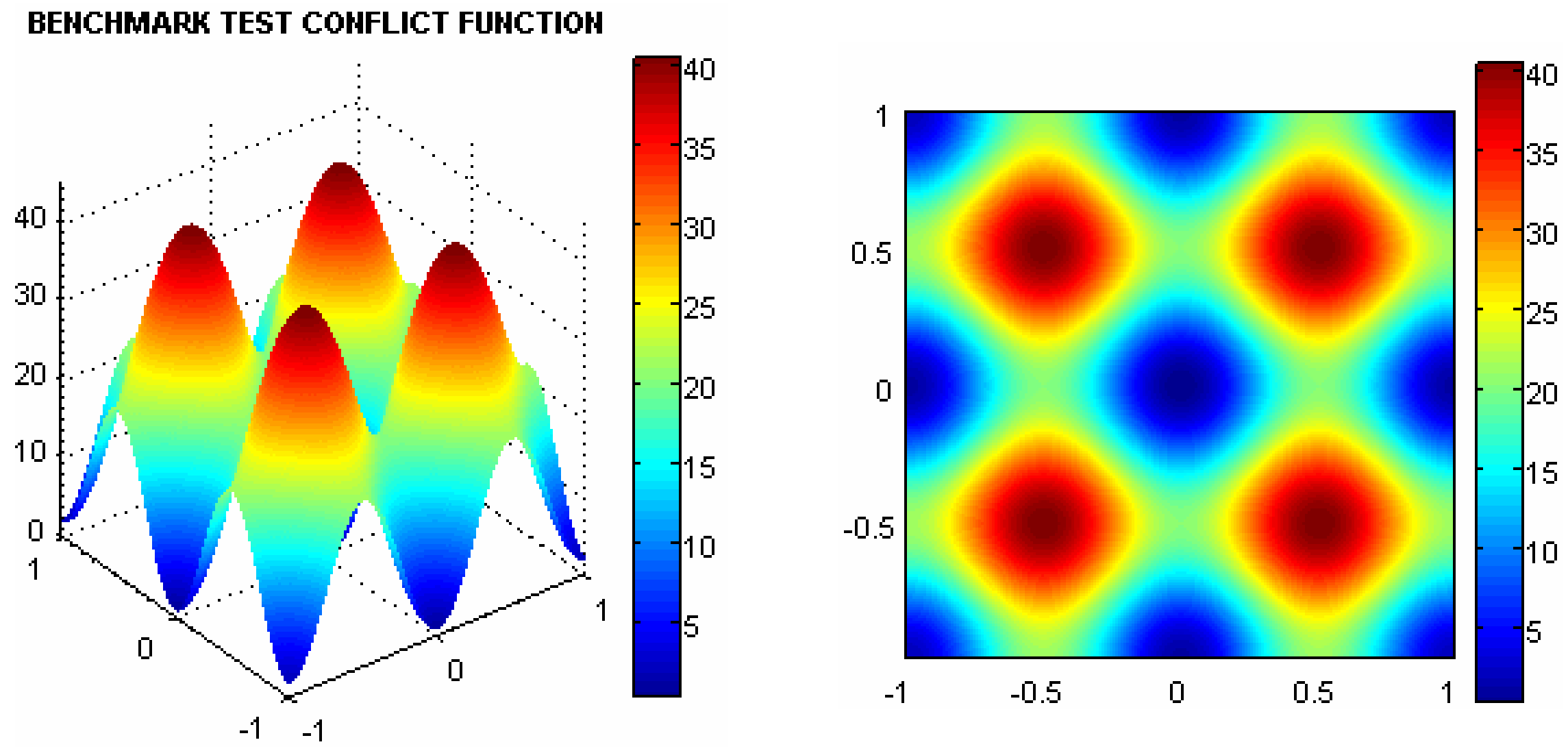

**Fig. A3. 6**: Surface plot and colour-map of the **Rastrigrin benchmark function** for 2-dimensional search-spaces in the region $[-1,1]^2$.

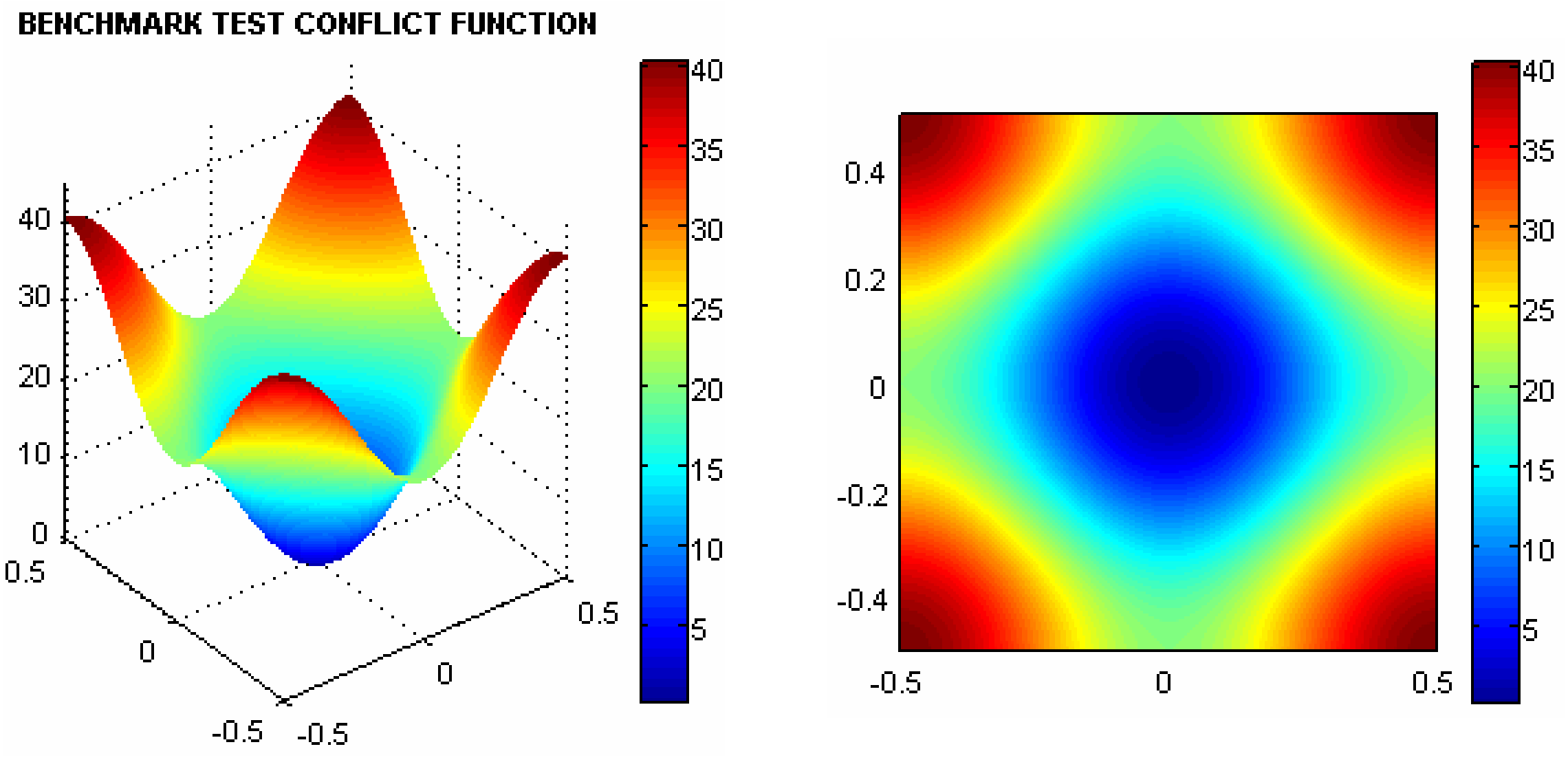

**Fig. A3. 7**: Surface plot and colour-map of the **Rastrigrin benchmark function** for 2-dimensional search-spaces in the region $[-0.5,0.5]^2$.

# A3.5 Griewank

The Griewank function also displays a single global optimum (uni-modal) and many local optima in the form of valleys. The features of this function are as follows:

- Function to be minimized: $f(\mathbf{x}) = \frac{1}{4000} \cdot \sum_{i=1}^{n} x_i^2 - \prod_{i=1}^{n} \cos\left(\frac{x_i}{\sqrt{i}}\right) + 1$
- Region of the search-space: $[-600,600]^n$
- Dimensions: $n = 30$
- Global optimum: $f(\hat{\mathbf{x}}) = 0$
- Location of the global optimum: $x_i = 0 \quad \forall i$

As opposed to the Rastrigrin function, here the influence of the term $\prod_{i=1}^{n} \cos\left(\frac{x_i}{\sqrt{i}}\right)$ decreases with the increase of the dimensionality of the problem, thus turning the function as easy to be optimized as the Sphere function. It is sometimes argued that finding the global optimum of the Griewank function resembles solving a simple optimization problem with a lot of noise.

The function is plotted for a two-dimensional search-space within the region $[-600,600]^2$ in **Fig. A3. 8**, within the region $[-60,60]^2$ in **Fig. A3. 9**, within the region $[-10,10]^2$ in **Fig. A3. 10**, and within the region $[-2,2]^2$ in **Fig. A3. 11**:

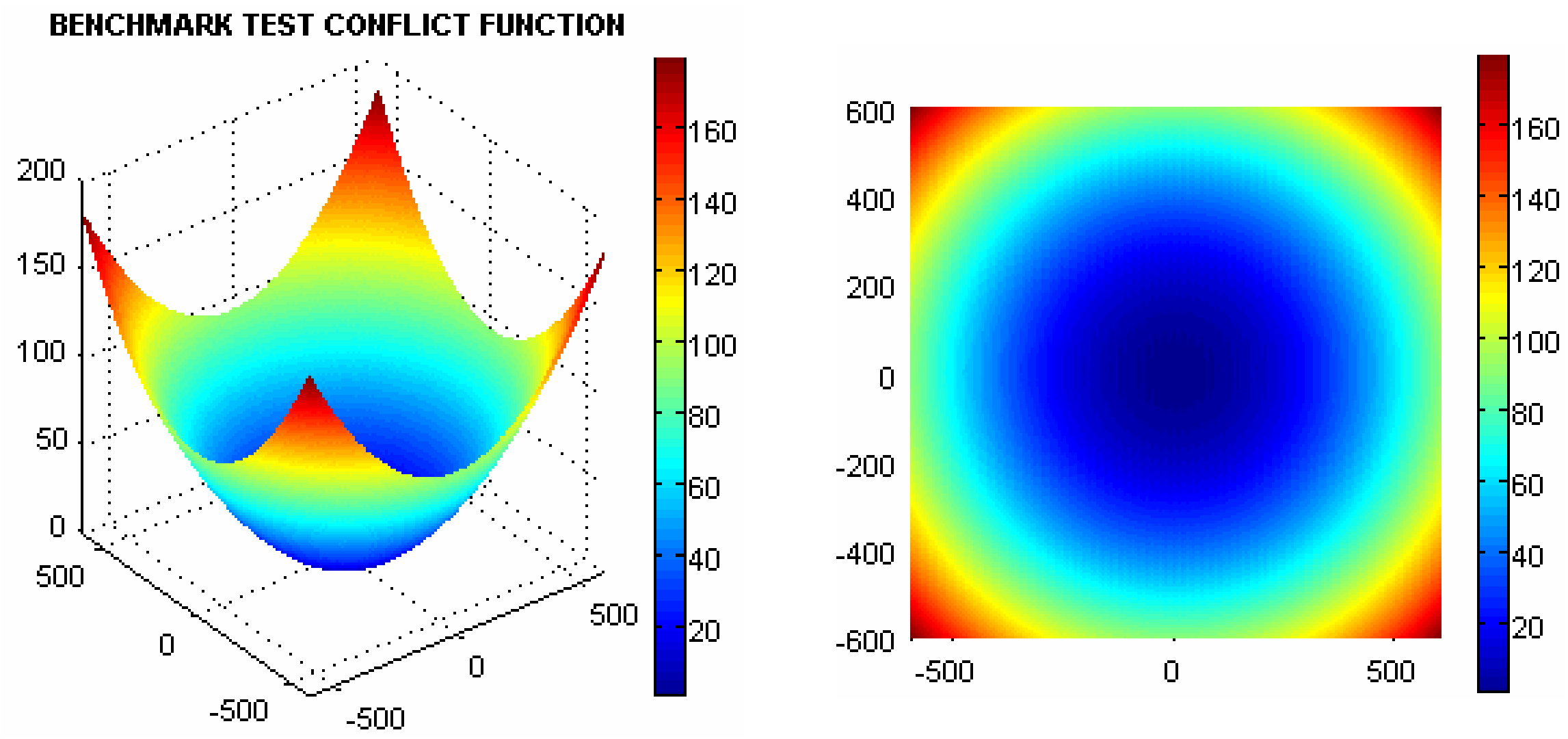

**Fig. A3. 8**: Surface plot and colour-map of the **Griewank benchmark function** for 2-dimensional search-spaces in the region $[-600,600]^2$.

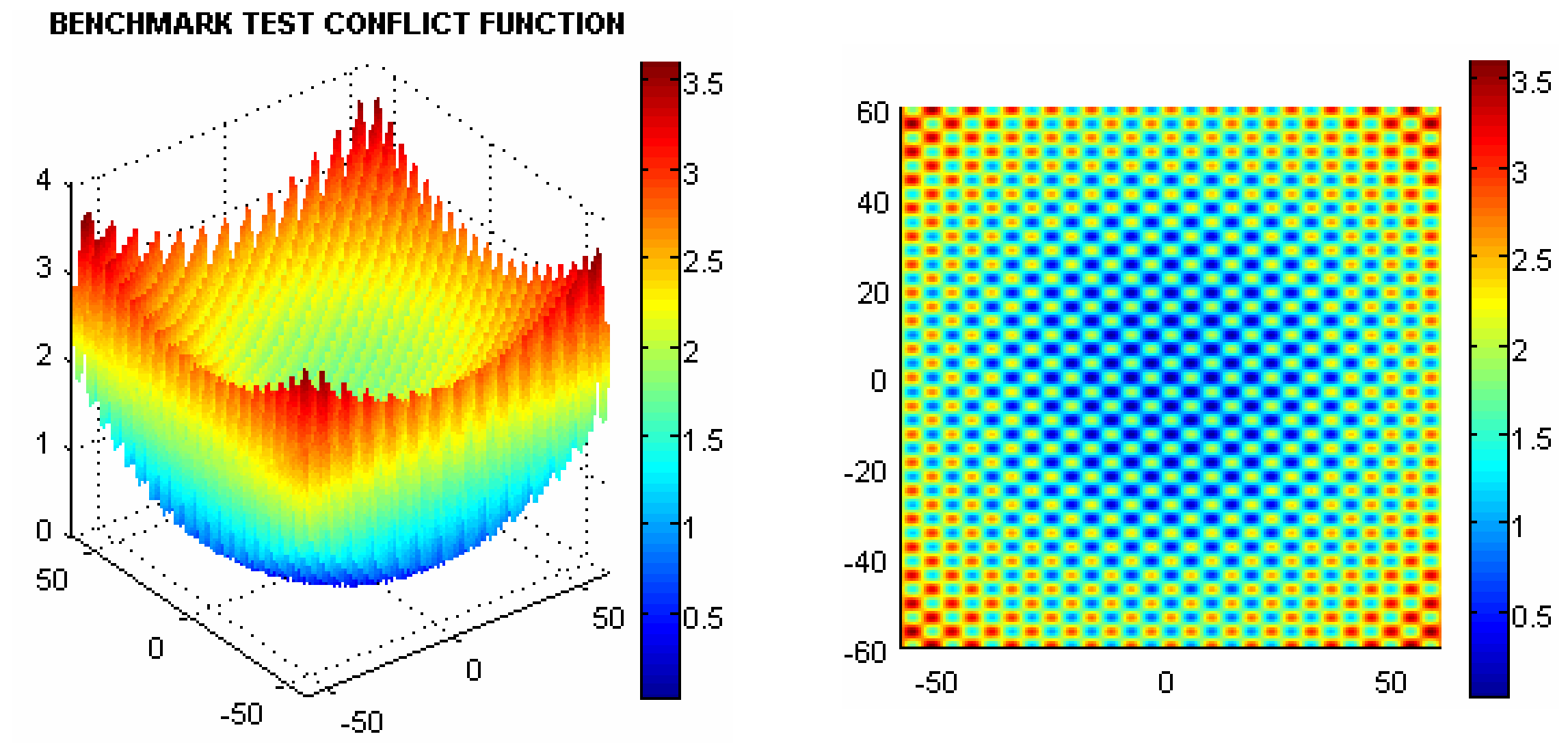

**Fig. A3. 9**: Surface plot and colour-map of the **Griewank benchmark function** for 2-dimensional search-spaces in the region $[-60,60]^2$.

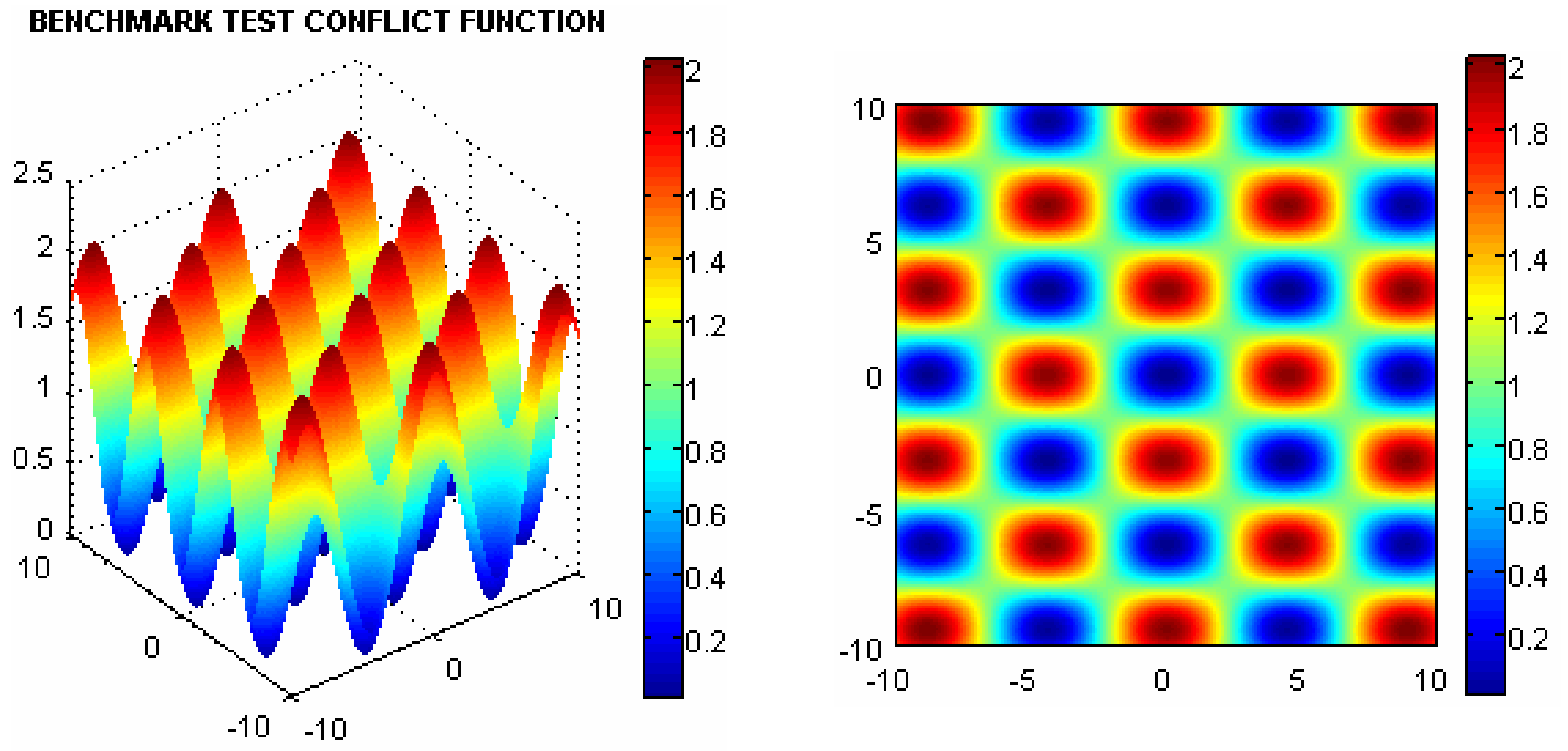

**Fig. A3. 10**: Surface plot and colour-map of the **Griewank benchmark function** for 2-dimensional search-spaces in the region $[-10,10]^2$.

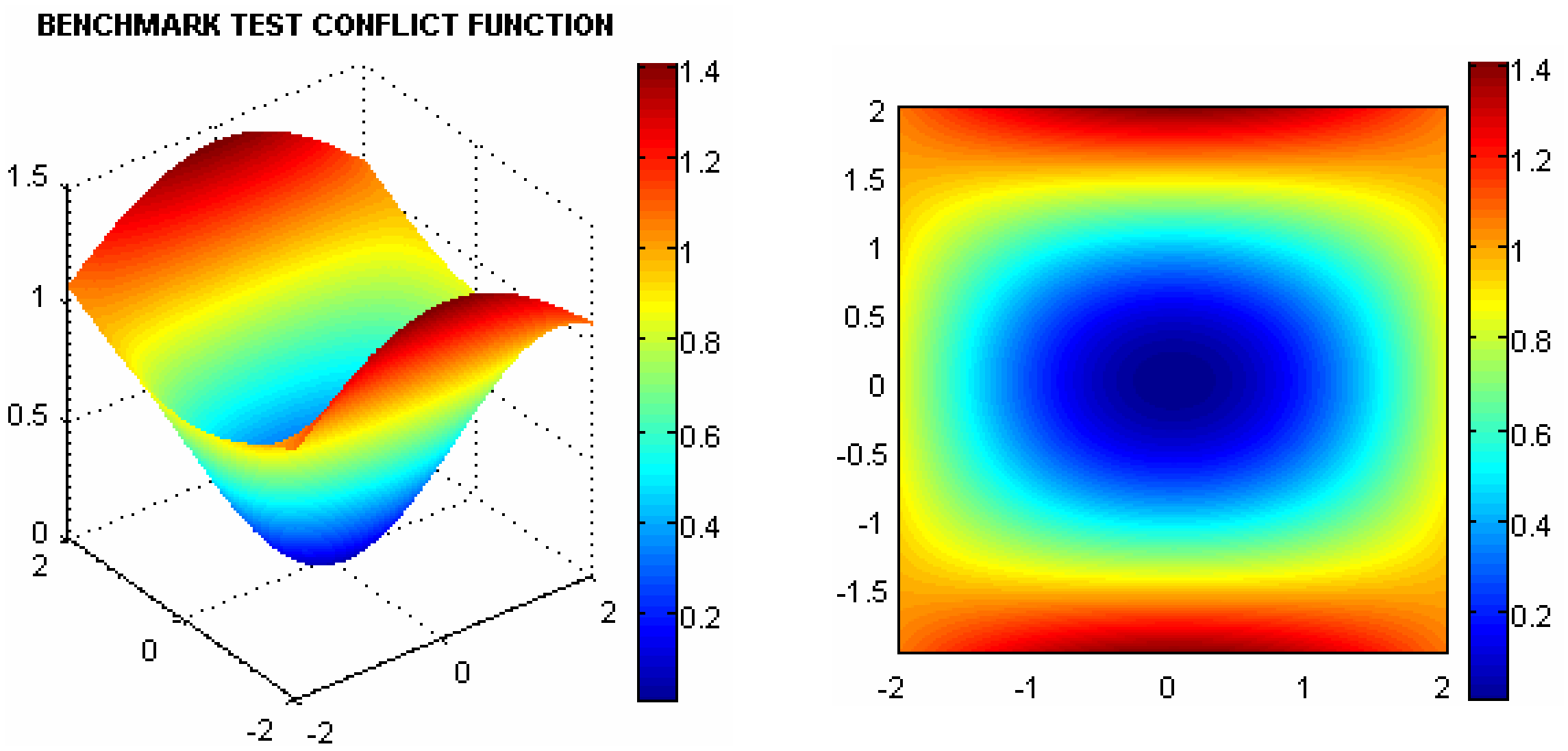

**Fig. A3. 11**: Surface plot and colour-map of the **Griewank benchmark function** for 2-dimensional search-spaces in the region $[-2,2]^2$.

# A3.6 Schaffer f6

The Schaffer f6 function is a very hard function to be optimized, which presents many local optima in the form of ring-like depressions. The features of this function are as follows:

- Function to be minimized: $$f(\mathbf{x}) = \frac{\left[\sin\left(\sqrt{\sum_{i=1}^{n} x_i^2}\right)\right]^2 - 0.5}{\left(1 + 0.001 \cdot \sum_{i=1}^{n} x_i^2\right)^2} + 0.5$$
- Region of the search-space: $[-100,100]^n$
- Dimensions: $n = 2$ and $n = 30$
- Global optimum: $f(\hat{\mathbf{x}}) = 0$
- Location of the global optimum: $x_i = 0 \quad \forall i$

The function is plotted for a two-dimensional search-space within the region $[-100,100]^2$ in **Fig. A3. 12**, within the region $[-30,30]^2$ in **Fig. A3. 13**, within the region $[-5,5]^2$ in **Fig. A3. 14**, and within the region $[-1,1]^2$ in **Fig. A3. 15**:

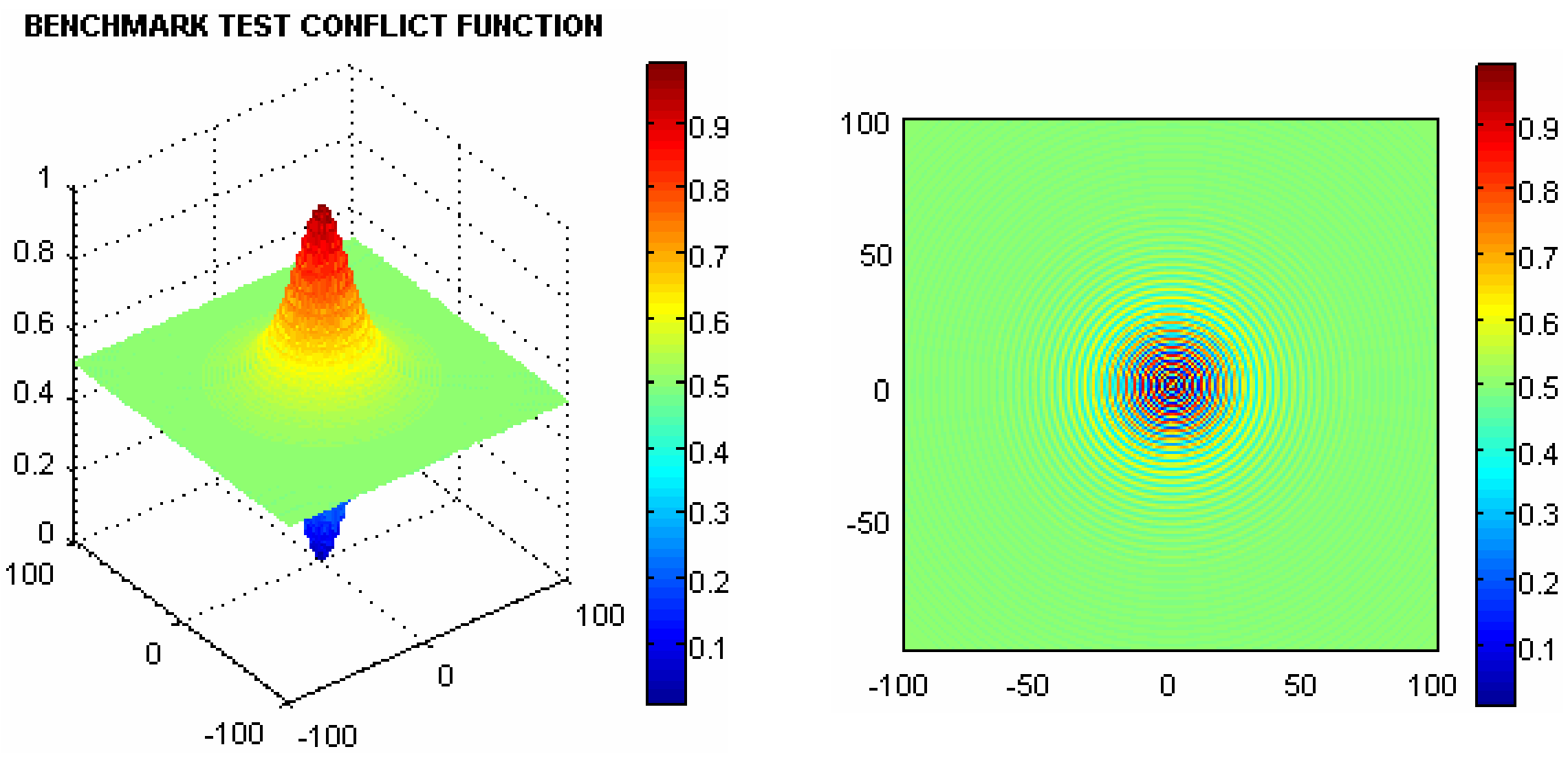

**Fig. A3. 12**: Surface plot and colour-map of the **Schaffer f6 benchmark function** for 2-dimensional search-spaces in the region $[-100,100]^2$.

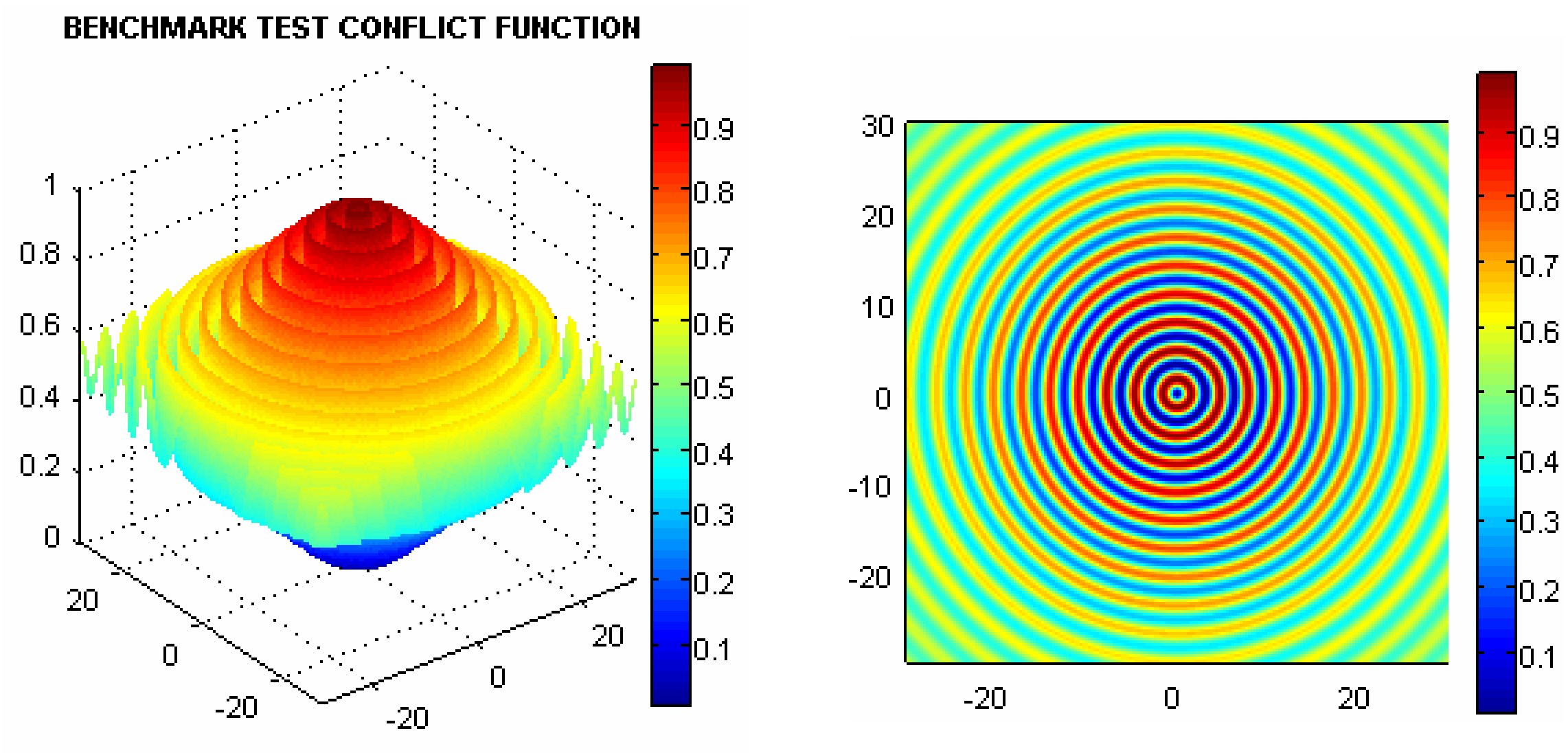

**Fig. A3. 13**: Surface plot and colour-map of the **Schaffer f6 benchmark function** for 2-dimensional search-spaces in the region $[-30,30]^2$.

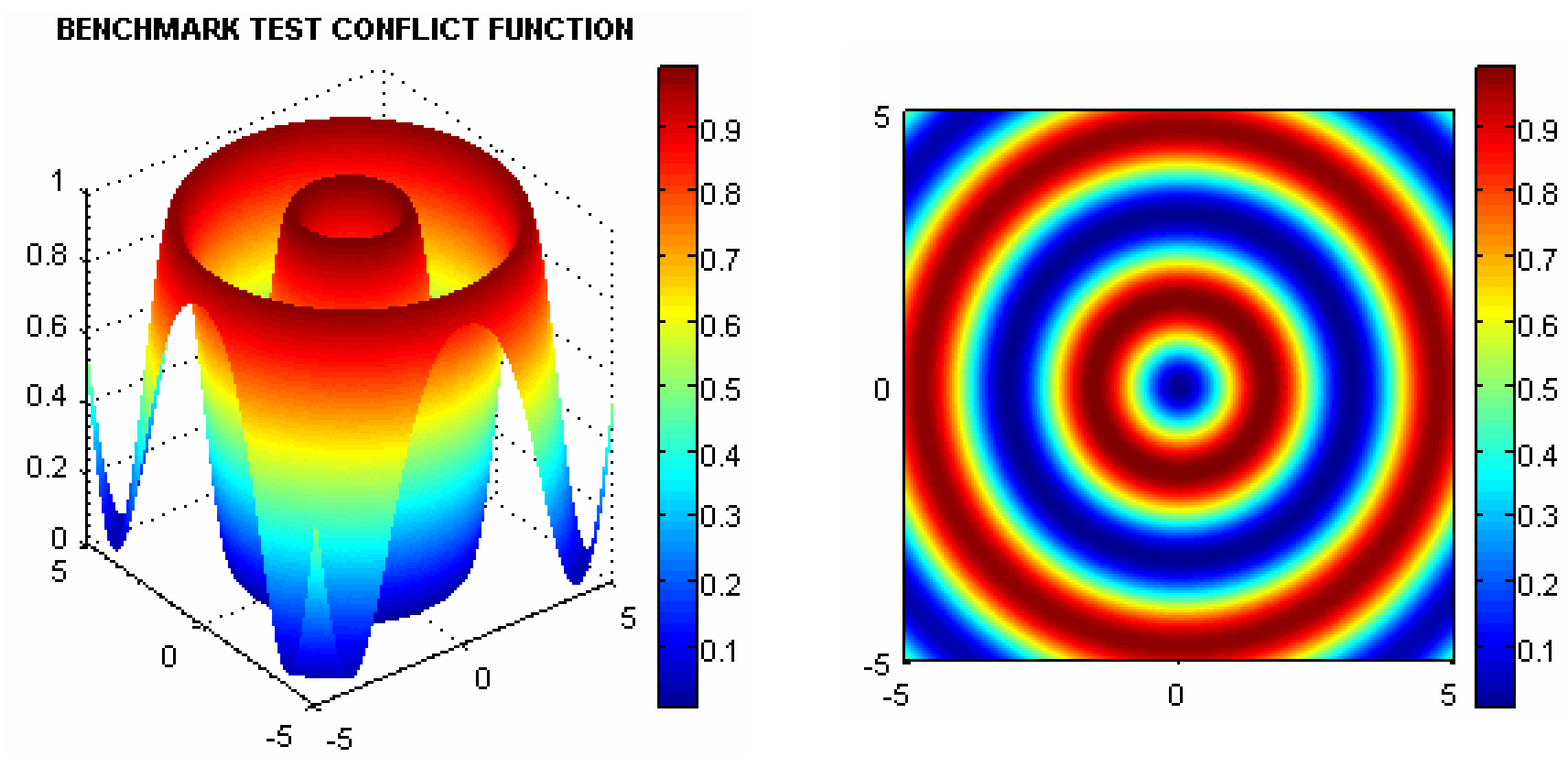

**Fig. A3. 14**: Surface plot and colour-map of the **Schaffer f6 benchmark function** for 2-dimensional search-spaces in the region $[-5,5]^2$.

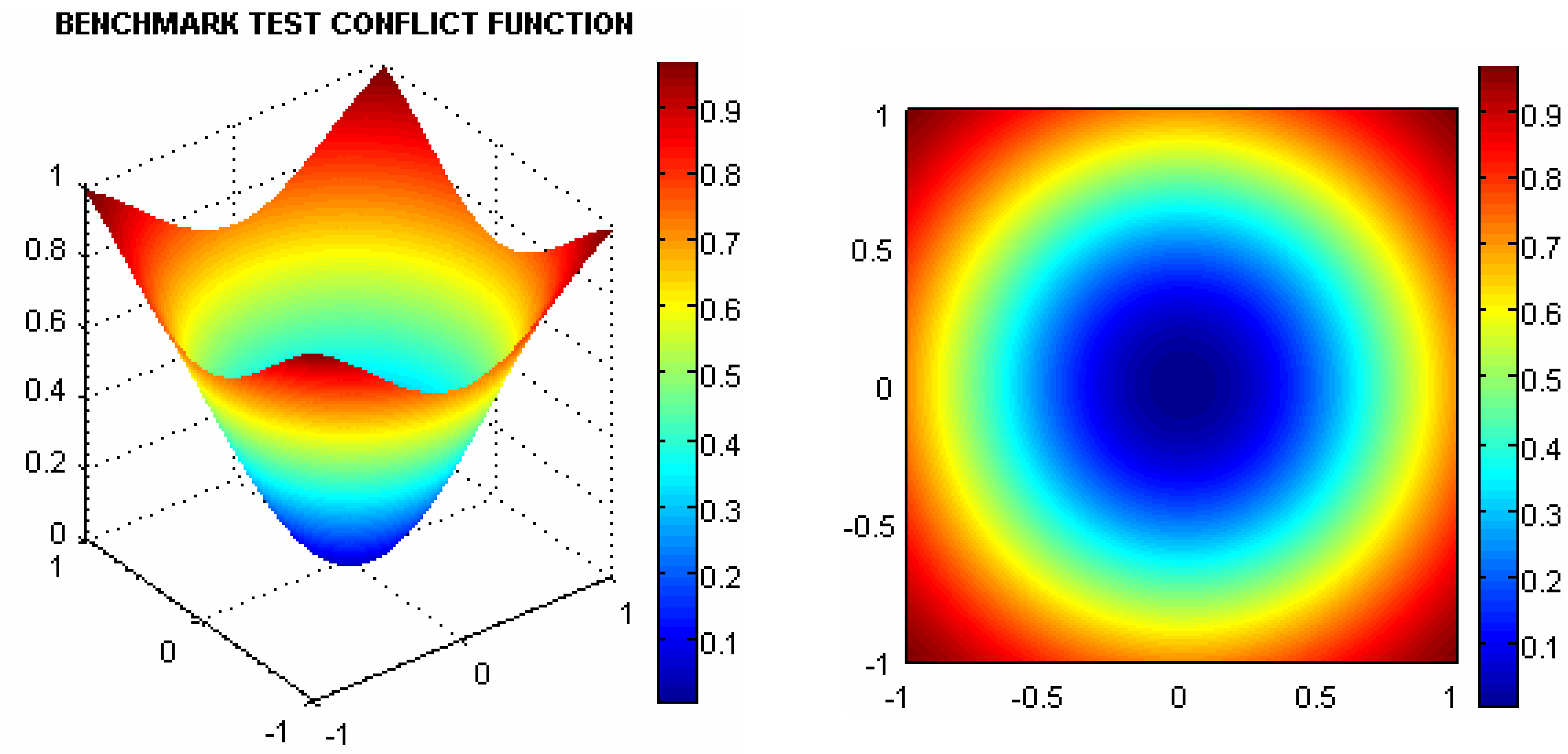

**Fig. A3. 15**: Surface plot and colour-map of the **Schaffer f6 benchmark function** for 2-dimensional search-spaces in the region $[-1,1]^2$ .

It is not very difficult for a PSO to find the region that contains the global optimum when optimizing this function, because the closer to the global optimum the local optima are located, the better the local optima are. However, the existence of very good local optima (i.e. similar in value to the global optimum) near the global optimum makes the fine-tuning of the search extremely difficult. Besides, the only way to arrive to the global optimum is to pass through the local optima.